\documentclass{article}%
\usepackage[round,semicolon]{natbib}
\usepackage{amsmath}
\usepackage{amsfonts}
\usepackage{amssymb}
\usepackage{graphicx}%
\providecommand{\U}[1]{\protect\rule{.1in}{.1in}}

\begin{document}

\title{ Design of Data-Driven Mathematical Laws for Optimal Statistical
Classification Systems}
\author{Denise M. Reeves}
\date{}
\maketitle

\begin{abstract}
This article will devise data-driven, mathematical laws that generate optimal,
statistical classification systems which achieve minimum error rates for data
distributions with unchanging statistics. Thereby, I will design learning
machines that minimize the expected risk or probability of misclassification.
I\ will devise a system of fundamental equations of binary classification for
a classification system in statistical equilibrium. I will use this system of
equations to formulate the problem of learning unknown, linear and quadratic
discriminant functions from data as a locus problem, thereby formulating
geometric locus methods within a statistical framework. Solving locus problems
involves finding equations of curves or surfaces defined by given properties
and finding graphs or loci of given equations. I will devise three systems of
data-driven, locus equations that generate optimal, statistical classification
systems. Each class of learning machines satisfies fundamental statistical
laws for a classification system in statistical equilibrium. Thereby, I\ will
formulate three classes of learning machines that are scalable modules for
optimal, statistical pattern recognition systems, all of which are capable of
performing a wide variety of statistical pattern recognition tasks, where any
given $M$-class statistical pattern recognition system exhibits optimal
generalization performance for an $M$-class feature space.

\end{abstract}

\textbf{Keywords:} geometric locus, geometric locus dilemma, bias/variance
dilemma, optimal decision function, optimal decision boundary, risk, counter
risk, binary classification theorem, discrete linear classification theorem,
discrete quadratic classification theorem, linear eigenlocus, linear
eigenlocus transform, quadratic eigenlocus, quadratic eigenlocus transform,
expected risk, total allowed eigenenergy, conservation of eigenenergy,
conservation of risk, critical minimum eigenenergy, symmetrical balance,
statistical equilibrium, optimal learning machine, optimal ensemble system of
learning machines, optimal statistical pattern recognition system, linear
kernel support vector machine, polynomial kernel support vector machine,
Gaussian kernel support vector machine.

\section{Introduction}

The discoveries presented in this paper are motivated by deep-seated and
long-standing problems in machine learning and statistics. In this article,
I\ will formulate the problem of learning unknown, linear and quadratic
discriminant functions from data as a locus problem, thereby formulating
geometric locus methods within a statistical framework. I will devise
fundamental, data-driven, locus equations of binary classification for linear
and quadratic classification systems in statistical equilibrium, where the
opposing forces and influences of a system are balanced with each other, and
the eigenenergy and the corresponding expected risk of a classification system
are minimized. Geometric locus problems involve equations of curves or
surfaces, where the coordinates of any given point on a curve or surface
satisfy an equation, and all of the points on any given curve or surface
possess a uniform geometric property. For example, a circle is a locus of
points, all of which are at the same distance (the radius) from a fixed point
(the center). Classic geometric locus problems involve algebraic equations and
Cartesian coordinate systems
\citep{Nichols1893,Tanner1898,Whitehead1911,Eisenhart1939}%
.

The primary purpose of the article is to devise data-driven, mathematical laws
that generate optimal, statistical classification systems which achieve
minimum error rates for statistical distributions with unchanging statistics.
Such optimal decision rules divide two-class feature spaces into decision
regions that have minimal conditional probabilities of classification error.
The data-driven, mathematical laws involve finding a solution of locus
equations, subject to fundamental statistical laws for a classification system
in statistical equilibrium. The data-driven, mathematical laws are based on
unexpected relations between likelihood ratio tests, geometric locus methods,
statistical methods, Hilbert space methods, and reproducing kernel Hilbert
space methods. Moreover, the data-driven, mathematical laws govern learning
machine architectures and generalization performance.

The other purpose of the article is to introduce new ways of thinking about
learning machines: in terms of fundamental \emph{statistical laws} that
\emph{unknown discriminant functions} of \emph{data} are \emph{subject to}.
Mathematical models of physical systems are often derived from physical laws
that physical systems in equilibrium are subject to. For example, Kirchhoff's
circuit laws and Newton's laws of motion are both based on the law of
conservation of energy: In a closed system, i.e., a system that is isolated
from its surroundings, the total energy of the system is conserved. Forms of
energy include: $\left(  1\right)  $ kinetic energy or energy associated with
motion, $\left(  2\right)  $ potential energy or energy of location with
respect to some reference point, $\left(  3\right)  $ chemical energy or
energy stored in chemical bonds, which can be released in chemical reactions,
$\left(  4\right)  $ electrical energy or energy created by separating
charges, e.g., energy stored in a battery, and $\left(  5\right)  $ thermal
energy or energy given off as heat, such as friction
\citep[see][]{Strang1986}%
.

In this paper, I will introduce a statistical form of energy that is
conserved: statistical eigenenergy or \emph{eigenenergy of location with
respect to the primary reference point for a geometric locus}, where a primary
reference point is the locus of the principal eigenaxis of a conic curve or a
quadratic surface.

Thinking about learning machines in terms of statistical laws will lead to the
discovery of equations of statistical equilibrium along with equations of
minimization of eigenenergy and expected risk that \emph{model-based
architectures} of learning machines \emph{are subject to}. I\ will derive
three model-based architectures from fundamental statistical laws that
classification systems in statistical equilibrium are subject to.

My discoveries are summarized below.

I\ will devise a system of fundamental equations of binary classification for
a classification system in statistical equilibrium that must be satisfied by
likelihood ratios and decision boundaries that achieve minimum error rates. I
will demonstrate that classification systems seek a point of statistical
equilibrium where the opposing forces and influences of a classification
system are balanced with each other, and the eigenenergy and the corresponding
expected risk of a classification system are minimized.

I\ will use these results to rigorously define three classes of learning
machines that are scalable modules for optimal, statistical classification or
pattern recognition systems, where each class of learning machines exhibits
optimal generalization performance for a category of statistical
distributions. One class of learning machines achieves minimum error rates for
data sets drawn from statistical distributions that have unchanging statistics
and similar covariance matrices. The other two classes of learning machines
achieve minimum error rates for any given data sets drawn from statistical
distributions that have either similar or dissimilar covariance matrices and
unchanging statistics.

All three classes of learning machines are solutions to fundamental integral
equations of likelihood ratios and corresponding decision boundaries, so that
each class of learning machines finds a point of statistical equilibrium where
the opposing forces and influences of a statistical classification system are
balanced with each other, and the eigenenergy and the corresponding expected
risk of the learning machine are minimized. Thereby, for each class of
learning machines, the generalization error of any given learning machine is a
function of the amount of overlap between data distributions. Thus, the
generalization error of each class of learning machines is determined by
\emph{the minimum probability of classification error}, which is the
\emph{lowest error rate} that can be achieved by a discriminant function and
the \emph{best generalization error} that can be achieved by a learning
machine. I\ will also define optimal ensemble systems for each class of
learning machines so that any given ensemble system exhibits optimal
generalization performance.

I will devise three systems of data-driven, vector-based locus equations that
generate optimal discriminant functions and decision boundaries. The three
systems of locus equations involve solving variants of the inequality
constrained optimization problem for linear, polynomial, and Gaussian kernel
support vector machines (SVMs). All three classes of learning machines are
capable of performing a wide variety of statistical pattern recognition tasks,
where any given learning machine exhibits optimal generalization performance
for a two-class feature space. For each class of learning machines, I will
demonstrate that any given learning machine is a scalable, individual
component of an optimal ensemble system, where any given ensemble system of
learning machines exhibits optimal generalization performance for an $M$-class
feature space.

By way of motivation, I will begin with an overview of long-standing problems
and unresolved questions in machine learning.

\section{Long-standing Problems in Machine Learning}

Machine learning is concerned with the design and development of computer
programs or algorithms that enable computers to identify and discover patterns
or trends contained within collections of digitized signals, images,
documents, or networks. For example, deep learning algorithms enable computers
to recognize objects contained within digitized videos.

Machine learning algorithms are said to enable computers to "learn from data."
Supervised machine learning algorithms, called "learning with a teacher,"
involve estimating unknown, input-output mappings or functions from training
examples, where each example consists of a unique input signal and a desired
(target) response or output. Accordingly, computers "learn" from training
examples by constructing input-output mappings for unknown functions of data.
Unsupervised machine learning algorithms involve estimating correlations or
connections between training examples, e.g., clusters or groupings of training
data
\citep{Geman1992,Hastie2001,Haykin2009}%
.

Supervised machine learning problems are generally considered extrapolation
problems for unknown functions, e.g., nontrivial, black box estimates. Black
boxes are defined in terms of inputs, subsequent outputs, and the mathematical
functions that relate them. Because training points will never cover a space
of possible inputs, practical learning machines must extrapolate in manners
that provide effective generalization performance. The generalization
performance of any given learning machine depends on the quality and quantity
of the training data, the complexity of the underlying problem, the learning
machine architecture, and the learning algorithm used to train the network.
\citep{Geman1992,Gershenfeld1999,Haykin2009}%
.

Fitting learning machine architectures to unknown functions of data involves
multiple and interrelated difficulties. Learning machine architectures are
sensitive to algebraic and topological structures that include functionals,
reproducing kernels, kernel parameters, and constraint sets
\citep
[see, e.g.,][]{Geman1992,Burges1998,Gershenfeld1999,Byun2002,Haykin2009,Reeves2015resolving}
as well as regularization parameters that determine eigenspectra of data
matrices
\citep[see, e.g.,][]{Haykin2009,Reeves2009,Reeves2011,Reeves2015resolving}%
. Identifying the correct form of an equation for a statistical model is also
a large concern
\citep{Daniel1979,Breiman1991,Geman1992,Gershenfeld1999,Duda2001}%
.

Fitting all of the training data is generally considered bad statistical
practice. Learning machine architectures that correctly interpolate
collections of noisy training points tend to fit the idiosyncrasies of the
noise and are not expected to exhibit good generalization performance. Highly
flexible architectures with indefinite parameter sets are said to overfit the
training data
\citep
{Wahba1987,Breiman1991,Geman1992,Barron1998,Boser1992,Gershenfeld1999,Duda2001,Hastie2001,Haykin2009}%
. Then again, learning machine architectures that interpolate insufficient
numbers of data points exhibit underfitting difficulties
\citep{Guyon1992,Ivanciuc2007applications}%
. Architectures with too few parameters ignore both the noise and the
meaningful behavior of the data
\citep{Gershenfeld1999}%
.

All of the above difficulties indicate that learning unknown functions from
training data involves trade-offs between underfittings and overfittings of
data points. The bias/variance dilemma describes statistical facets of these
trade-offs
\citep{Geman1992,Gershenfeld1999,Duda2001,Scholkopf2002,Haykin2009}%
.

\subsection{The Bias/Variance Dilemma}

All learning machine architectures are composed of training data. Moreover,
the estimation error between a learning machine and its target function
depends on the training data in a twofold manner.
\citet*{Geman1992}
examined these dual sources of estimation error in their article
titled\textit{\ Neural Networks and the Bias/Variance Dilemma}. The crux of
the dilemma is that estimation error is composed of two distinct components
termed a bias and a variance. Large numbers of parameter estimates raise the
variance, whereas incorrect statistical models increase the bias
\citep{Geman1992,Gershenfeld1999,Duda2001,Hastie2001,Haykin2009}%
.

The bias/variance dilemma can be summarized as follows. Model-free
architectures based on insufficient data samples are unreliable and have slow
convergence speeds. However, model-based architectures based on incorrect
statistical models are also unreliable. Except, model-based architectures
based on accurate statistical models are reliable and have reasonable
convergence speeds. Even so, proper statistical models for model-based
architectures are difficult to identify.

\subsubsection{Prewiring of Important Generalizations}

I\ have considered the arguments regarding the bias/variance dilemma presented
by
\citet{Geman1992}%
, and I have come to the overall conclusion that learning an unknown function
from data requires prewiring important generalizations into a learning
machine's architecture, where generalizations involve suitable representations
of statistical decision systems. For statistical classification systems,
I\ will show that generalizations involve \emph{joint representations} of
discriminant functions \emph{and} decision boundaries.

So how do we identify the important generalizations for a given problem? How
should these generalizations be prewired? How do important generalizations
represent key aspects of statistical decision systems? What does it really
mean to introduce a carefully designed bias into a learning machine's
architecture? How does the introduction of a proper bias involve a model-based
architecture? How do we discover model-based architectures? What is a
model-based architecture? I will consider all of these problems in terms of
the fundamental modeling question posed next. The general problem is outlined
in
\citet{Naylor1971}%
.

\subsection{The System Representation Problem}

I propose that effective designs of learning machine architectures involve an
underlying system modeling problem. In particular, effective designs of
model-based, learning machine architectures involve the formulation of a
mathematical system which simulates essential stochastic behavior and models
key aspects of a real statistical system. Accordingly, statistical model
formulation for learning machine architectures involves the development of a
mathematically tractable, statistical model that provides a useful
representation of a statistical decision system.

In general terms, a system is an interconnected set of elements which are
coherently organized in a manner that achieves a useful function or purpose
\citep{Meadows2008}%
. This indicates that prewiring relevant aspects of statistical decision
systems within learning machine architectures involves effective
interconnections between suitable sets of coherently organized data points. I
will devise three classes of learning machine architectures that provide
substantial examples of effective interconnections between suitable sets of
coherently organized data points.

\subsection{Suitable Representations for Learning Machines}

The matter of identifying suitable representations for learning machine
architectures is important. Indeed, for many scientific and engineering
problems, there is a natural and elegant way to represent the solution. For
example, each of the well-known special functions, e.g., Legendre polynomials,
Bessel functions, Fourier series, Fourier integrals, etc., have the common
motivation of being most appropriate for certain problems and quite unsuitable
for others, where each special function represents the relevant aspects of a
physical system
\citep{Keener2000}%
. Moreover, most mathematical models of physical systems are based on a
fundamental principle that nature acts to minimize energy. Accordingly,
physical systems seek a point of equilibrium where the opposing forces and
influences of the system are balanced with each other, and the energy of the
system is minimized
\citep[see][]{Strang1986}%
.

Yet, most machine learning methods attempt to approximate unknown functions
with methods that assume no sort of representation, e.g., nonparametric
inference methods
\citep{Geman1992,Cherkassky1998,Duda2001,Hastie2001,Haykin2009}
or assume representations that are tentative and ill-defined, e.g., indefinite
interpolations of SVM margin hyperplanes in unknown, high-dimensional spaces
\citep{Boser1992,Cortes1995}%
. Likewise, consider the notion of the asymptotic convergence of a learning
machine architecture to some unknown function. Can we picture what this
actually means?

So how do we devise mathematically tractable, statistical models for learning
machine architectures?

\subsection{Learning Statistical Laws from Training Data}

Tangible representations provide objects and forms which can be seen and
imagined, along with a perspective for seeing and imagining them
\citep{Hillman2012}%
. I will devise effective hyperparameters and substantial geometric
architectures for linear, polynomial, and Gaussian kernel SVMs, each of which
is based on a mathematically tractable, statistical model that can be depicted
and understood in two and three-dimensional Euclidean spaces and fully
comprehended in higher dimensions. Each statistical model is derived from
fundamental laws in mathematics and statistics, whereby computers "learn"
optimal decision rules from data by finding a solution of locus equations,
subject to fundamental statistical laws that are satisfied by classification
systems in statistical equilibrium.

In this paper, I\ will devise model-based architectures for learning machines
that determine optimal decision functions and boundaries for training data
drawn from any two statistical distributions that have unchanging statistics
and $\left(  1\right)  $ similar covariance matrices, $\left(  2\right)  $
dissimilar covariance matrices, and $\left(  3\right)  $ homogeneous
distributions which are completely overlapping with each other. Any given
discriminant function and decision boundary satisfies fundamental statistical
laws for a binary classification system in statistical equilibrium. Thereby,
each class of learning machines achieves the lowest possible error rate, i.e.,
the learning machine minimizes the average risk $\mathfrak{R}_{\mathfrak{\min
}}$: which is the lowest error rate that can be achieved by any linear or
quadratic discriminant function
\citep{VanTrees1968,Fukunaga1990,Duda2001}%
.

All of the problems outlined above indicate that learning unknown functions
from data involves indeterminate problems which remain largely unidentified. I
have identified a problem which I\ have named the geometric locus dilemma
\citep{,Reeves2015resolving}%
. The geometric locus dilemma is summarized below.

\subsection{The Geometric Locus Dilemma}

Any given conic section or quadratic surface is a \emph{predetermined}
geometric \emph{configuration} of points, i.e., endpoints of directed line
segments called vectors, whose Cartesian coordinate locations satisfy, i.e.,
are \emph{determined by}, an algebraic \emph{equation}. Moreover, curves or
surfaces of standard locus equations are determined by properties of geometric
loci with respect to coordinate axes of \emph{arbitrary} Cartesian coordinate
systems. Thereby, an algebraic equation of a classical locus of points
generates an \emph{explicit} curve or surface in an \emph{arbitrarily
specified} Cartesian space. It follows that any point on a classical geometric
locus \emph{naturally} exhibits the uniform property of the locus.

So, consider fitting a collection of training data to standard locus
equations. Given the correlated, algebraic and geometric constraints on a
traditional locus of points, it follows that any attempt to fit\ an
$N$-dimensional set of $d$-dimensional, random data points to the standard
equation(s) of a geometric locus, involves the unsolvable problem of
determining an effective constellation of an $\left(  N-M\right)  \times d$
subset of $N\times d$ random vector coordinates that $(1)$ inherently satisfy
preset, and thus fixed, length constraints on each of the respective $d$
Cartesian coordinate axes and thereby $(2)$ generate predetermined curves or
surfaces in Cartesian space $%
\mathbb{R}
^{d}$. Such estimation tasks are not possible.

It follows that fitting collections of random data points to standard locus
equations is \emph{an impossible estimation task}.

Given the limitations imposed by the geometric locus dilemma, the design and
development of learning machine architectures has primarily been based on
curve and surface fitting methods of interpolation or regression, alongside
statistical methods of reducing data to minimum numbers of relevant
parameters. For example, multilayer artificial neural networks (ANNs) estimate
nonlinear regressions with optimally pruned architectures. Good generalization
performance for ANNs is considered an effect of a good nonlinear interpolation
of the training data
\citep{Geman1992,Haykin2009}%
. Alternatively, support vector machines (SVMs) fit linear curves or surfaces
to minimum numbers of training points. Good generalization performance for
SVMs is largely attributed to maximally separated, linear decision borders
\citep{Boser1992,Cortes1995,Bennett2000,Cristianini2000,Scholkopf2002}%
.

\subsection{The Geometric Locus Dilemma for SVMs}

SVMs estimate linear and nonlinear decision boundaries by solving a quadratic
programming problem. SVM methods specify a pair of linear borders, termed
margin hyperplanes, that pass through data points called support vectors.

The capacity or complexity of SVM decision boundaries is regulated by means of
a geometric margin of separation between a pair of margin hyperplanes. The SVM
method minimizes the capacity of a separating hyperplane by maximizing the
distance between a pair of margin hyperplanes or linear borders. Large
distances between margin hyperplanes $(1)$ allow for considerably fewer
hyperplane orientations and $(2)$ enforce a limited capacity to separate
training data. Thus, maximizing the distance between margin hyperplanes
regulates the complexity of separating hyperplane estimates
\citep
{Boser1992,Cortes1995,Burges1998,Bennett2000,Cristianini2000,Scholkopf2002}%
.

\subsubsection{Separation of Overlapping Data}

Identifying interpolation methods that provide effective fits of separating
lines, planes, or hyperplanes involves the long-standing problem of fitting
linear decision boundaries to overlapping sets of data points
\citep[see, e.g.,][]{Cover1965,Cortes1995}%
. Soft margin, linear kernel SVMs\ are said to resolve this problem by means
of non-negative, random slack variables $\xi_{i}\geq0$, each of which allows a
correlated data point $\mathbf{x}_{i}$ that lies between a pair of linear
decision borders to satisfy a linear border. Nonlinear kernel SVMs also employ
non-negative, random slack variables, each of which allows a
transformed,\ correlated data point to satisfy a hyperplane\textit{\ }decision
border in some higher dimensional feature space
\citep{Cortes1995,Bennett2000,Cristianini2000,Hastie2001}%
.

\subsubsection{An Impossible Estimation Task}

SVM applications of slack variables imply that non-negative, random slack
variables \emph{specify distances} of data points \emph{from unknown} linear
curves or surfaces.

Clearly, this is an \emph{impossible estimation task}. Therefore, given $l$
overlapping data points $\left\{  \mathbf{x}_{i}\right\}  _{i=1}^{l}$ and $l$
non-negative, random slack variables $\left\{  \xi_{i}|\xi_{i}\geq0\right\}
_{i=1}^{l}$, it is concluded that computing effective values for $l$
non-negative, random slack variables $\left\{  \xi_{i}|\xi_{i}\geq0\right\}
_{i=1}^{l}$ is an impossible estimation task.

\subsubsection{Hyperparameter Tuning}

SVMs are widely reported to perform well on statistical classification tasks.
However, SVMs require \emph{computationally expensive} hyperparameter
\emph{tuning}
\citep{Byun2002,Liang2011}%
. For nonlinear kernel SVMs, the user must select the polynomial degree or the
kernel width. In addition, the user must select regularization parameters for
linear and nonlinear kernel SVMs. Generally speaking, the performance of SVMs
depends on regularization parameters, the type of kernel, and the kernel
parameter for nonlinear kernels. The choice of a nonlinear kernel and its
parameter for a given problem is a research issue
\citep{Burges1998}%
.

In this paper, I\ will introduce new ways of thinking about linear,
polynomial, and Gaussian kernel SVMs. In the process, I will resolve the
geometric locus dilemma for all three classes of SVMs. I\ will define
effective hyperparameters for polynomial and Gaussian kernel SVMs. I\ will
also define regularization methods for linear, polynomial, and Gaussian kernel
SVMs. Thereby, I will devise three classes of high-performance learning
machines that are scalable modules for statistical pattern recognition systems.

\subsection{Statistical Pattern Recognition Systems}

Statistical pattern recognition systems provide an automated means to identify
or discover information and objects contained within large collections of
digitized $(1)$ images or videos, $(2)$ waveforms, signals, or sequences and
$(3)$ documents. Statistical pattern recognition systems provide automated
processes such as optical character recognition, geometric object recognition,
speech recognition, spoken language identification, handwriting recognition,
waveform recognition, face recognition, system identification, spectrum
identification, fingerprint identification, and DNA sequencing
\citep{Srinath1996,Jain2000statistical,Duda2001}%
.

\subsubsection{Design of Statistical Pattern Recognition Systems}

Statistical pattern recognition systems divide pattern spaces into decision
regions that are separated by decision boundaries; pattern spaces are commonly
known as feature spaces. The design of statistical pattern recognition systems
involves two fundamental problems. The first problem concerns identifying
measurements or numerical features of the objects being classified and using
these measurements to form pattern or feature vectors for each pattern class.
For $M$ classes of patterns, a pattern space is composed of $M$ regions, where
each region contains the pattern vectors of a class. The second problem
involves generating decision boundaries that divide a pattern space into $M$
regions. Functions that determine decision boundaries are called discriminant
functions
\citep{VanTrees1968,Srinath1996,Duda2001}%
.

\section{Theory of Binary Classification}

The problem of generating decision boundaries involves specifying discriminant
functions. The fundamental problem of explicitly defining discriminant
functions is termed the binary classification problem. I\ will present an
overview of existing criterion for the binary classification problem, and
I\ will use these results to develop new criteria and a theorem for binary classification.

\subsubsection{The Binary Classification Problem}

The basic classification problem of deciding between two pattern or feature
vectors is essentially one of partitioning a feature space $Z$ into two
suitable regions $Z_{1}$ and $Z_{2}$, such that whenever a pattern vector
$\mathbf{x}$ lies in region $Z_{1}$, the classifier decides that $\mathbf{x}$
belongs to class $\omega_{1}$ and whenever $\mathbf{x}$ lies in region $Z_{2}
$, the classifier decides that $\mathbf{x}$ belongs to class $\omega_{2}$.
When the classifier makes a wrong decision, the classifier is said to make an
error. A suitable criterion is necessary to determine the best possible
partitioning for a given feature space $Z$
\citep{VanTrees1968,Srinath1996,Duda2001}%
.

\subsection{Bayes' Criterion}

Bayes' criterion divides a feature space $Z$ in a manner that minimizes the
probability of classification error $\mathcal{P}_{\min_{e}}\left(  Z\right)
$. Bayes' decision rule is determined by partitioning a feature space $Z$ into
two regions $Z_{1}$ and $Z_{2}$, where the average risk $\mathfrak{R}%
_{\mathfrak{B}}\left(  Z\right)  $ of the total probability of making a
decision error $\mathcal{P}_{\min_{e}}\left(  Z\right)  $ is minimized. The
overall cost $C$ that determines the Bayes' risk $\mathfrak{R}_{\mathfrak{B}%
}\left(  Z\right)  $ is controlled by assigning points to two suitable regions
$Z_{1}$ and $Z_{2}$, where each region $Z_{1}$ or $Z_{2}$ has a Bayes' risk
$\mathfrak{R}_{\mathfrak{B}}\left(  Z_{1}\right)  $ or $\mathfrak{R}%
_{\mathfrak{B}}\left(  Z_{2}\right)  $ that determines the minimum probability
of error $\mathcal{P}_{\min_{e}}\left(  Z_{1}\right)  $ or $\mathcal{P}%
_{\min_{e}}\left(  Z_{2}\right)  $ for the region.

A Bayes' test is based on two assumptions. The first is that prior
probabilities $P\left(  \omega_{1}\right)  $ and $P\left(  \omega_{2}\right)
$ of the pattern classes $\omega_{1}$ and $\omega_{2}$ represent information
about the pattern vector sources. The second assumption is that a cost is
assigned to each of the four possible outcomes associated with a decision.
Denote the costs for the four possible outcomes by $C_{11}$, $C_{21}$,
$C_{22}$, and $C_{12}$, where the first subscript indicates the chosen class
and the second subscript indicates the true class. Accordingly, each time that
the classifier makes a decision, a certain cost will be incurred
\citep{VanTrees1968,Srinath1996,Duda2001}%
.

\subsection{Bayes' Likelihood Ratio Test}

Minimization of the Bayes' risk $\mathfrak{R}_{\mathfrak{B}}\left(  Z\right)
$ produces decision regions defined by the following expression:

If%
\[
P\left(  \omega_{1}\right)  \left(  C_{21}-C_{11}\right)  p\left(
\mathbf{x}|\omega_{1}\right)  \geq P\left(  \omega_{2}\right)  \left(
C_{12}-C_{22}\right)  p\left(  \mathbf{x}|\omega_{2}\right)  \text{,}%
\]
then assign the pattern vector $\mathbf{x}$ to region $Z_{1}$ and say that
class $\omega_{1}$ is true. Otherwise, assign the pattern vector $\mathbf{x} $
to region $Z_{2}$ and say that class $\omega_{2}$ is true.

\begin{figure}[ptb]%
\centering
\fbox{\includegraphics[
height=2.5875in,
width=3.4411in
]%
{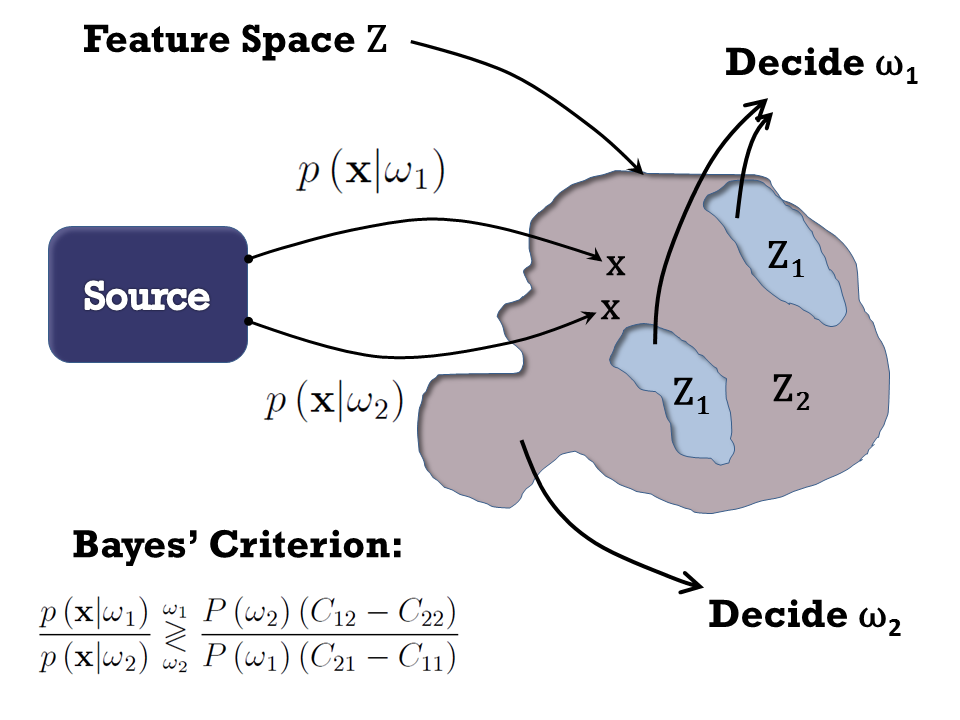}%
}\caption{Bayes' likelihood ratio test divides a feature space $Z$ into
decision regions $Z_{1}$ and $Z_{2}$ which minimize the Bayes' risk
$\mathfrak{R}_{\mathfrak{B}}\left(  Z|\Lambda\right)  $.}%
\label{Bayes' Decision Rule}%
\end{figure}

The resulting discriminant function is Bayes' likelihood ratio test:%

\begin{equation}
\Lambda\left(  \mathbf{x}\right)  \triangleq\frac{p\left(  \mathbf{x}%
|\omega_{1}\right)  }{p\left(  \mathbf{x}|\omega_{2}\right)  }\overset{\omega
_{1}}{\underset{\omega_{2}}{\gtrless}}\frac{P\left(  \omega_{2}\right)
\left(  C_{12}-C_{22}\right)  }{P\left(  \omega_{1}\right)  \left(
C_{21}-C_{11}\right)  }\text{,} \label{Bayes' LRT}%
\end{equation}
where $p\left(  \mathbf{x}|\omega_{1}\right)  $ and $p\left(  \mathbf{x}%
|\omega_{2}\right)  $ are class-conditional density functions. Given Bayes'
likelihood ratio test, a\ decision boundary divides a feature space $Z$ into
two decision regions $Z_{1}$ and $Z_{2}$, where a decision rule $\Lambda
\left(  \mathbf{x}\right)  \overset{\omega_{1}}{\underset{\omega_{2}%
}{\gtrless}}\eta$ is based on a threshold $\eta$. Figure
$\ref{Bayes' Decision Rule}$ illustrates how Bayes' decision rule divides a
feature space\ $Z$ into decision regions
\citep{VanTrees1968,Srinath1996}%
.%

Bayes' decision rule in Eq. (\ref{Bayes' LRT}) computes the likelihood ratio
for a feature vector $\mathbf{x}$%
\[
\Lambda\left(  \mathbf{x}\right)  =\frac{p\left(  \mathbf{x}|\omega
_{1}\right)  }{p\left(  \mathbf{x}|\omega_{2}\right)  }%
\]
and makes a decision by comparing the ratio $\Lambda\left(  \mathbf{x}\right)
$ to the threshold $\eta$%
\[
\eta=\frac{P\left(  \omega_{2}\right)  \left(  C_{12}-C_{22}\right)
}{P\left(  \omega_{1}\right)  \left(  C_{21}-C_{11}\right)  }\text{.}%
\]

Costs and prior probabilities are usually based on educated guesses.
Therefore, it is common practice to determine a likelihood ratio
$\Lambda\left(  \mathbf{x}\right)  $ that is independent of costs and prior
probabilities and let $\eta$ be a variable threshold that accommodates changes
in estimates of cost assignments and prior probabilities
\citep{,VanTrees1968,Srinath1996}%
.

Bayes' classifiers are difficult to design because the class-conditional
density functions and the decision threshold are usually not known. Instead, a
collection of "training data" is used to estimate either decision boundaries
or class-conditional density functions
\citep{Fukunaga1990,Duda2001,Hastie2001,Haykin2009}%
.

\subsubsection{Minimal Probability of Error Criterion}

If $C_{11}=C_{22}=0$ and $C_{21}=C_{12}=1$, then the average risk
$\mathfrak{R}_{\mathfrak{B}}\left(  Z\right)  $ is given by the expression%
\begin{equation}
\mathfrak{R}_{\mathfrak{B}}\left(  Z\right)  =P\left(  \omega_{2}\right)
\int_{Z_{1}}p\left(  \mathbf{x}|\omega_{2}\right)  d\mathbf{x}+P\left(
\omega_{1}\right)  \int_{Z_{2}}p\left(  \mathbf{x}|\omega_{1}\right)
d\mathbf{x} \label{Bayes' Risk}%
\end{equation}
which is the total probability of making an error. Given this cost assignment,
the Bayes' test $\ln\Lambda\left(  \mathbf{x}\right)  \overset{\omega
_{1}}{\underset{\omega_{2}}{\gtrless}}\ln\eta$%
\[
\ln\frac{p\left(  \mathbf{x}|\omega_{1}\right)  }{p\left(  \mathbf{x}%
|\omega_{2}\right)  }\overset{\omega_{1}}{\underset{\omega_{2}}{\gtrless}}%
\ln\frac{P\left(  \omega_{2}\right)  }{P\left(  \omega_{1}\right)  }\text{,}%
\]
which can be written as%
\[
\ln p\left(  \mathbf{x}|\omega_{1}\right)  -\ln p\left(  \mathbf{x}|\omega
_{2}\right)  \overset{\omega_{1}}{\underset{\omega_{2}}{\gtrless}}\ln P\left(
\omega_{2}\right)  -\ln P\left(  \omega_{2}\right)  \text{,}%
\]
is minimizing the total probability of error. When two given pattern classes
are equally likely, the decision threshold is zero: $\eta=0$. The probability
of misclassification is called the Bayes' error
\citep{VanTrees1968,Fukunaga1990,Srinath1996,Duda2001}%
.

\subsection{Bayes' Error}

The performance of a discriminant function can be determined by evaluating
errors associated with making decisions. For a binary classification problem,
there are two types of decision errors. A classifier may decide class
$\omega_{2}$ when the true class is $\omega_{1}$ (denoted by $D_{2}|\omega
_{1}$) or a classifier may decide class $\omega_{1}$ when the true class is
$\omega_{2}$ (denoted by $D_{1}|\omega_{2}$). Each type of decision error has
a probability associated with it which depends on the class-conditional
densities and the decision rule.

Let $P_{1}$ denote the probability of error $P\left(  D_{2}|\omega_{1}\right)
$ corresponding to deciding class $\omega_{2}$ when the true class is
$\omega_{1}$. The $P_{1}$ Bayes' error is given by the integral%
\begin{align}
P_{1}  &  =P\left(  D_{2}|\omega_{1}\right)  =\int_{-\infty}^{\eta}p\left(
\mathbf{x}|\omega_{1}\right)  d\mathbf{x}\label{Type One Error}\\
&  =\int_{Z_{2}}p\left(  \mathbf{x}|\omega_{1}\right)  d\mathbf{x}\nonumber
\end{align}
which is a conditional probability given the class-conditional density
$p\left(  \mathbf{x}|\omega_{1}\right)  $ and the decision region $Z_{2}$.

Let $P_{2}$ denote the probability of error $P\left(  D_{1}|\omega_{2}\right)
$ corresponding to deciding class $\omega_{1}$ when the true class is
$\omega_{2}$. The $P_{2}$ Bayes' error is given by the integral%
\begin{align}
P_{2}  &  =P\left(  D_{1}|\omega_{2}\right)  =\int_{\eta}^{\infty}p\left(
\mathbf{x}|\omega_{2}\right)  d\mathbf{x}\label{Type Two Error}\\
&  =\int_{Z_{1}}p\left(  \mathbf{x}|\omega_{2}\right)  d\mathbf{x}\nonumber
\end{align}
which is a conditional probability given the class-conditional density
$p\left(  \mathbf{x}|\omega_{2}\right)  $ and the decision region $Z_{1}$.

Once the $Z_{1}$ and $Z_{2}$ decision regions are chosen, the values of the
error integrals in Eqs (\ref{Type One Error}) and (\ref{Type Two Error}) are
determined
\citep{VanTrees1968,Srinath1996}%
.

I\ will redefine decision regions in the next section.

\subsection{Practical Decision Regions}

Bayes' decision rule defines the $Z_{1}$ and $Z_{2}$ decision regions to
consist of values of $\mathbf{x}$ for which the likelihood ratio
$\Lambda\left(  \mathbf{x}\right)  $ is, respectively, greater than or less
than a threshold $\eta$, where any given set of $Z_{1}$ and $Z_{2}$ decision
regions spans an entire feature space over the interval of $\left(
-\infty,\infty\right)  $. Accordingly, the lower limit of the integral for the
$P_{1}$ Bayes' error in Eq. (\ref{Type One Error}) is $-\infty$, and the upper
limit of the integral for the $P_{2}$ Bayes' error in Eq.
(\ref{Type Two Error}) is $\infty$. Figure $\ref{Bayes' Decision Rule}$
depicts an example for which the $Z_{1}$ and $Z_{2}$ decision regions span an
entire feature space $Z$, where%
\[
Z=Z_{1}+Z_{2}=Z_{1}\cup Z_{2}\text{.}%
\]

I\ will now redefine decision regions based solely on regions that are
associated with decision errors or lack thereof. Accordingly, regions
associated with decision errors involve regions associated with overlapping
data distributions and regions associated with no decision errors involve
regions associated with non-overlapping data distributions.

\begin{figure}[ptb]%
\centering
\fbox{\includegraphics[
height=2.5875in,
width=3.4411in
]%
{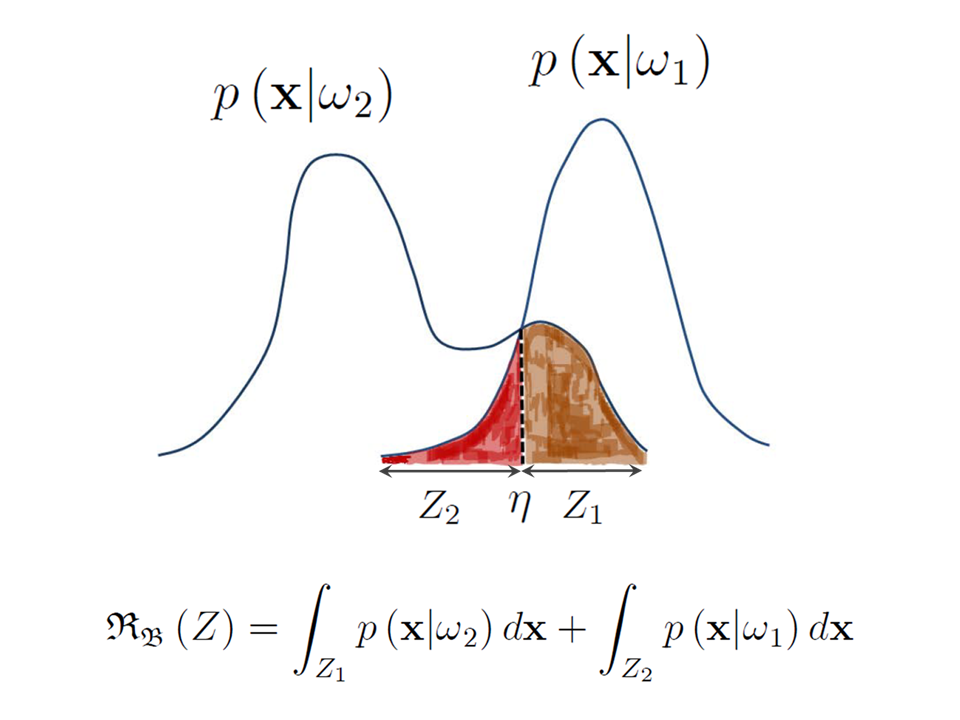}%
}\caption{The average risk $\mathfrak{R}_{\mathfrak{B}}\left(  Z\right)  $,
i.e., the decision error $\mathcal{P}_{\min_{e}}\left(  Z\right)  $, is
determined by class-conditional densities $p\left(  \mathbf{x}|\omega
_{2}\right)  $ and $p\left(  \mathbf{x}|\omega_{1}\right)  $ and decision
regions $Z_{1}$ and $Z_{2}$ that may or may not be contiguous, where region
$Z_{1}$ or $Z_{2}$ has a conditional risk $\mathfrak{R}\left(  Z_{1}\right)  $
or $\mathfrak{R}\left(  Z_{2}\right)  $ that determines the conditional
probability of error $\mathcal{P}_{\min_{e}}\left(  Z_{1}\right)  $ or
$\mathcal{P}_{\min_{e}}\left(  Z_{2}\right)  $ for the region.}%
\label{Bayes' Error and Decision Regions Overlapping Data}%
\end{figure}

For overlapping data distributions, Fig.
$\ref{Bayes' Error and Decision Regions Overlapping Data}$ illustrates how
values of the $P_{1}$ and $P_{2}$ error integrals in Eqs (\ref{Type One Error}%
) and (\ref{Type Two Error}) are determined by a decision threshold $\eta$ and
the $Z_{1}$ and $Z_{2}$ decision regions. Accordingly, the risk $\mathfrak{R}%
_{\mathfrak{B}}\left(  Z\right)  $ is determined by the class-conditional
densities $p\left(  \mathbf{x}|\omega_{2}\right)  $ and $p\left(
\mathbf{x}|\omega_{1}\right)  $ and the corresponding $Z_{1}$ and $Z_{2}$
decision regions.%

\subsubsection{Decision Regions for Overlapping Data Distributions}

For overlapping data distributions, decision regions are defined to be those
regions that span regions of \emph{data distribution overlap}. Accordingly,
the $Z_{1}$ decision region, which is associated with class $\omega_{1}$,
spans a finite region between the decision threshold $\eta$ and the region of
distribution overlap between $p\left(  \mathbf{x}|\omega_{1}\right)  $ and
$p\left(  \mathbf{x}|\omega_{2}\right)  $, whereas the $Z_{2}$ decision
region, which is associated with class $\omega_{2}$, spans a finite region
between the region of distribution overlap between $p\left(  \mathbf{x}%
|\omega_{2}\right)  $ and $p\left(  \mathbf{x}|\omega_{1}\right)  $ and the
decision threshold $\eta$ (see, e.g., Fig.
$\ref{Bayes' Error and Decision Regions Overlapping Data}$). It follows that
the risk $\mathfrak{R}\left(  Z_{1}\right)  $ in the $Z_{1}$ decision region
involves pattern vectors $\mathbf{x}$ that are generated according to
$p\left(  \mathbf{x}|\omega_{1}\right)  $ and $p\left(  \mathbf{x}|\omega
_{2}\right)  $, where pattern vectors $\mathbf{x}$ that are generated
according to $p\left(  \mathbf{x}|\omega_{2}\right)  $ contribute to the
$P_{2}$ decision error in Eq. (\ref{Type Two Error}) and the total risk
$\mathfrak{R}_{\mathfrak{B}}\left(  Z\right)  $ in Eq. (\ref{Bayes' Risk}). It
also follows that the risk $\mathfrak{R}\left(  Z_{2}\right)  $ in the $Z_{2}$
decision region involves pattern vectors $\mathbf{x}$ that are generated
according to $p\left(  \mathbf{x}|\omega_{2}\right)  $ and $p\left(
\mathbf{x}|\omega_{1}\right)  $, where pattern vectors $\mathbf{x}$ that are
generated according to $p\left(  \mathbf{x}|\omega_{1}\right)  $ contribute to
the $P_{1}$ decision error in Eq. (\ref{Type One Error}) and the total risk
$\mathfrak{R}_{\mathfrak{B}}\left(  Z\right)  $ in Eq. (\ref{Bayes' Risk}).
Accordingly, the risk $\mathfrak{R}_{\mathfrak{B}}\left(  Z\right)  $ in Eq.
(\ref{Bayes' Risk}) is determined by the risks $\mathfrak{R}_{\mathfrak{B}%
}\left(  Z_{1}\right)  $ and $\mathfrak{R}_{\mathfrak{B}}\left(  Z_{2}\right)
$ in the corresponding $Z_{1}$ and $Z_{2}$ decision regions, which involve
functionals of pattern vectors $\mathbf{x}$ that lie in overlapping regions of
data distributions.

\subsubsection{Decision Regions for Non-overlapping Data Distributions}

For non-overlapping data distributions, the $Z_{1}$ decision region, which is
associated with class $\omega_{1}$, spans a finite region between the decision
threshold $\eta$ and the tail region of $p\left(  \mathbf{x}|\omega
_{1}\right)  $, whereas the $Z_{2}$ decision region, which is associated with
class $\omega_{2}$, spans a finite region between the tail region of $p\left(
\mathbf{x}|\omega_{2}\right)  $ and the decision threshold $\eta$. Because the
risk $\mathfrak{R}_{\mathfrak{B}}\left(  Z_{1}\right)  $ in the $Z_{1}$
decision region only involves functionals of pattern vectors $\mathbf{x}$ that
are generated according to $p\left(  \mathbf{x}|\omega_{1}\right)  $, the risk
$\mathfrak{R}_{\mathfrak{B}}\left(  Z_{1}\right)  $ in region $Z_{1}$ is zero:
$\mathfrak{R}_{\mathfrak{B}}\left(  Z_{1}\right)  =0$. Likewise, because the
risk $\mathfrak{R}_{\mathfrak{B}}\left(  Z_{2}\right)  $ in the $Z_{2}$
decision region only involves functionals of pattern vectors $\mathbf{x}$ that
are generated according to $p\left(  \mathbf{x}|\omega_{2}\right)  $, the risk
$\mathfrak{R}_{\mathfrak{B}}\left(  Z_{2}\right)  $ in region $Z_{2}$ is zero:
$\mathfrak{R}_{\mathfrak{B}}\left(  Z_{2}\right)  =0$. Accordingly, the risk
$\mathfrak{R}_{\mathfrak{B}}\left(  Z\right)  $ in the decision space in $Z$
is zero$\ $because $\mathfrak{R}_{\mathfrak{B}}\left(  Z_{1}\right)
+\mathfrak{R}_{\mathfrak{B}}\left(  Z_{2}\right)  =0$, where the risks
$\mathfrak{R}_{\mathfrak{B}}\left(  Z_{1}\right)  $ and $\mathfrak{R}%
_{\mathfrak{B}}\left(  Z_{2}\right)  $ in the $Z_{1}$ and $Z_{2}$ decision
regions involve functionals of pattern vectors $\mathbf{x}$ that lie in tail
regions of data distributions.%
\begin{figure}[ptb]%
\centering
\fbox{\includegraphics[
height=2.5875in,
width=3.4411in
]%
{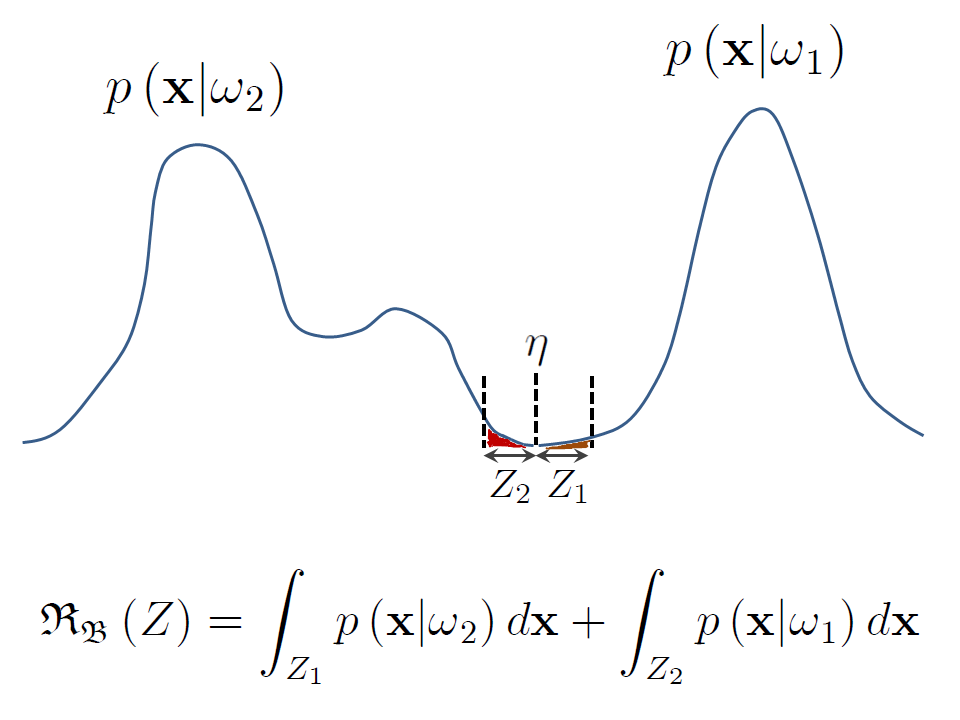}%
}\caption{For non-overlapping data distributions, the average risk
$\mathfrak{R}_{\mathfrak{B}}\left(  Z\right)  $ in the decision space $Z$ is
zero because the conditional risks $\mathfrak{R}\left(  Z_{1}\right)  $ and
$\mathfrak{R}\left(  Z_{2}\right)  $ in the $Z_{1}$ and $Z_{2}$ decision
regions are zero: $\mathfrak{R}\left(  Z_{1}\right)  =\mathfrak{R}\left(
Z_{2}\right)  =0$.}%
\label{Bayes' Error and Decision Regions Non-overlapping Data}%
\end{figure}

For non-overlapping data distributions, Fig.
$\ref{Bayes' Error and Decision Regions Non-overlapping Data}$ illustrates
that values of the $P_{1}$ and $P_{2}$ error integrals in Eqs
(\ref{Type One Error}) and (\ref{Type Two Error}), which are determined by the
risks $\mathfrak{R}_{\mathfrak{B}}\left(  Z_{2}\right)  $ and $\mathfrak{R}%
_{\mathfrak{B}}\left(  Z_{1}\right)  $ in the corresponding $Z_{2}$ and
$Z_{1}$ decision regions, are effectively zero, so that the risk
$\mathfrak{R}_{\mathfrak{B}}\left(  Z\right)  $ in the decision space $Z$ is
zero: $\mathfrak{R}_{\mathfrak{B}}\left(  Z\right)  =0$.

\subsection{Decision Error in Terms of the Likelihood Ratio}

The $Z_{1}$ and $Z_{2}$ decision regions consist of values of $\mathbf{x}$ for
which the likelihood ratio $\Lambda\left(  \mathbf{x}\right)  $ is,
respectively, less than or greater than the decision threshold $\eta$.
Accordingly, the error $P\left(  D_{2}|\omega_{1}\right)  $ in Eq.
(\ref{Type One Error}) can be written in terms of the likelihood ratio
$\Lambda\left(  \mathbf{x}\right)  $ as%
\begin{equation}
P_{1}=\int_{Z_{2}}p\left(  \Lambda\left(  \mathbf{x}\right)  |\omega
_{1}\right)  d\Lambda\text{,} \label{Type One Error Likelihood Ratio}%
\end{equation}
and the error $P\left(  D_{1}|\omega_{2}\right)  $ in Eq.
(\ref{Type Two Error}) can be written in terms of the likelihood ratio
$\Lambda\left(  \mathbf{x}\right)  $ as%
\begin{equation}
P_{2}=\int_{Z_{1}}p\left(  \Lambda\left(  \mathbf{x}\right)  |\omega
_{2}\right)  d\Lambda\text{,} \label{Type Two Error Likelihood Ratio}%
\end{equation}
where $P_{1}$ and $P_{2}$ are conditional probabilities given the decision
regions $Z_{1}$ and $Z_{2}$ and the likelihood ratio $\Lambda\left(
\mathbf{x}\right)  $
\citep{VanTrees1968,Srinath1996}%
.

\begin{figure}[ptb]%
\centering
\fbox{\includegraphics[
height=2.5875in,
width=3.4411in
]%
{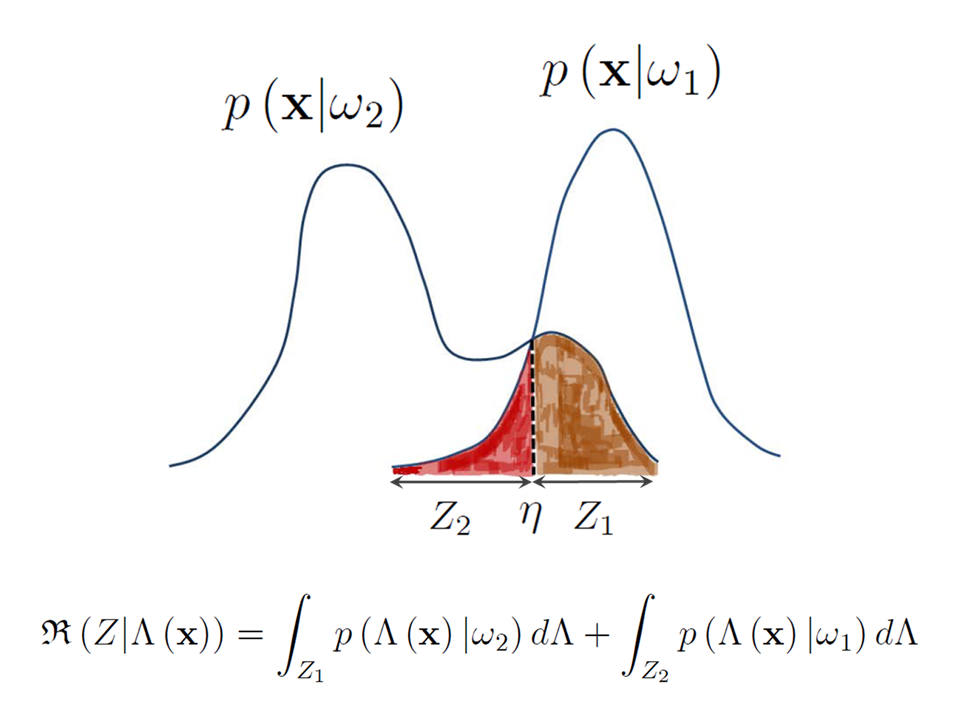}%
}\caption{Illustration of a decision rule that partitions a feature space $Z$
in a manner that minimizes the conditional probabilities of decision error
$P_{1}$ and $P_{2}$ given the likelihood ratio $\Lambda\left(  \mathbf{x}%
\right)  $ and the decision regions $Z_{1}$ and $Z_{2}$.}%
\label{Bayes' Decision Rule and Risk}%
\end{figure}

Figure $\ref{Bayes' Decision Rule and Risk}$ illustrates how the decision
error is a function of the likelihood ratio $\Lambda\left(  \mathbf{x}\right)
$ and the $Z_{1}$ and $Z_{2}$ decision regions.%

\subsection{Bayes' Minimax Criterion}

The Bayes' criterion requires that costs be assigned to various decisions and
that values be assigned to prior probabilities for pattern classes. If not
enough is known about the sources or mechanisms generating a set of pattern or
feature vectors, then prior probabilities cannot be determined, and the Bayes'
criterion cannot be applied. However, a decision rule can be obtained by using
a criterion known as the minimax risk, which involves minimization of the
maximum risk. A decision rule that satisfies Bayes' minimax risk is called a
minimax test. In some cases, the threshold for the minimax test is identical
to the threshold for the minimum probability of error test
\citep{VanTrees1968,Poor1996,Srinath1996}%
.

\subsubsection{Equalizer Rules}

A minimax test is also called an \emph{equalizer rule}, where any given
equalizer rule performs well over a range of prior probabilities
\citep{Poor1996}%
. Bayes' minimax risk $\mathfrak{R}_{mm}\left(  Z\right)  $, which is given by
the integral equation%
\begin{align}
\mathfrak{R}_{mm}\left(  Z\right)   &  =C_{22}+\left(  C_{12}-C_{22}\right)
\int\nolimits_{Z_{1}}p\left(  \mathbf{x}|\omega_{2}\right)  d\mathbf{x}%
\label{Minimax Risk}\\
&  =C_{11}+\left(  C_{21}-C_{11}\right)  \int\nolimits_{Z_{2}}p\left(
\mathbf{x}|\omega_{1}\right)  d\mathbf{x}\text{,}\nonumber
\end{align}
determines a decision rule for which the $Z_{1}$ and $Z_{2}$ decision regions
have equal Bayes' risks: $\mathfrak{R}_{\mathfrak{B}}\left(  Z_{1}\right)
=\mathfrak{R}_{\mathfrak{B}}\left(  Z_{2}\right)  $
\citep{VanTrees1968,Srinath1996}%
.

\subsection{Bayes' Minimax Theorem}

For the case when $C_{11}=C_{22}=0$ and $C_{21}=C_{12}=1$, the minimax
integral equation reduces to $P_{2}=P_{1}$, where the minimax cost is the
average probability of error. It follows that the integral equation for Bayes'
minimax risk $\mathfrak{R}_{mm}\left(  Z\right)  $ is given by%
\begin{align}
\mathfrak{R}_{mm}\left(  Z\right)   &  =\int\nolimits_{Z_{1}}p\left(
\mathbf{x}|\omega_{2}\right)  d\mathbf{x}\label{Minimax Criterion}\\
&  =\int\nolimits_{Z_{2}}p\left(  \mathbf{x}|\omega_{1}\right)  d\mathbf{x}%
\text{,}\nonumber
\end{align}
where the conditional probability of making an error $P_{2}$ for class
$\omega_{2}$ is equal to the conditional probability of making an error
$P_{1}$ for class $\omega_{1}$:%
\[
\int\nolimits_{Z_{1}}p\left(  \mathbf{x}|\omega_{2}\right)  d\mathbf{x}%
=\int\nolimits_{Z_{2}}p\left(  \mathbf{x}|\omega_{1}\right)  d\mathbf{x}%
\text{.}%
\]

Accordingly, Bayes' minimax risk $\mathfrak{R}_{mm}\left(  Z\right)  $
involves solving the integral equation in Eq. (\ref{Minimax Criterion}) where
the conditional probabilities of decision error $P\left(  D_{1}|\omega
_{2}\right)  $ and $P\left(  D_{2}|\omega_{1}\right)  $ for class $\omega_{2}$
and class $\omega_{1}$ are symmetrically balanced with each other%
\[
P\left(  D_{1}|\omega_{2}\right)  \equiv P\left(  D_{2}|\omega_{1}\right)
\]
about a decision threshold $\eta$.

\subsection{Bayes' Decision Rules for Gaussian Data}

For Gaussian data, Bayes' decision rule and boundary are defined by the
likelihood ratio test:%
\begin{align}
\Lambda\left(  \mathbf{x}\right)   &  =\frac{\left\vert \mathbf{\Sigma}%
_{2}\right\vert ^{1/2}\exp\left\{  -\frac{1}{2}\left(  \mathbf{x}%
-\boldsymbol{\mu}_{1}\right)  ^{T}\mathbf{\Sigma}_{1}^{-1}\left(
\mathbf{x}-\boldsymbol{\mu}_{1}\right)  \right\}  }{\left\vert \mathbf{\Sigma
}_{1}\right\vert ^{1/2}\exp\left\{  -\frac{1}{2}\left(  \mathbf{x}%
-\boldsymbol{\mu}_{2}\right)  ^{T}\mathbf{\Sigma}_{2}^{-1}\left(
\mathbf{x}-\boldsymbol{\mu}_{2}\right)  \right\}  }%
\label{Likelihood Ratio Gaussian Data}\\
&  \overset{\omega_{1}}{\underset{\omega_{2}}{\gtrless}}\frac{P\left(
\omega_{2}\right)  \left(  C_{12}-C_{22}\right)  }{P\left(  \omega_{1}\right)
\left(  C_{21}-C_{11}\right)  }\text{,}\nonumber
\end{align}
where $\boldsymbol{\mu}_{1}$ and $\boldsymbol{\mu}_{2}$ are $d$-component mean
vectors, $\mathbf{\Sigma}_{1}$ and $\mathbf{\Sigma}_{2}$ are $d$-by-$d$
covariance matrices, $\mathbf{\Sigma}^{-1}$ and $\left\vert \mathbf{\Sigma
}\right\vert $ denote the inverse and determinant of a covariance matrix, and
$\omega_{1}$ or $\omega_{2}$ is the true data category.

The transform $\ln\left(  \Lambda\left(  \mathbf{x}\right)  \right)
\overset{\omega_{1}}{\underset{\omega_{2}}{\gtrless}}$ $\ln\left(
\eta\right)  $ of the likelihood ratio test in Eq.
(\ref{Likelihood Ratio Gaussian Data}):%
\begin{align}
\ln\left(  \Lambda\left(  \mathbf{x}\right)  \right)   &  =\mathbf{x}%
^{T}\left(  \mathbf{\Sigma}_{1}^{-1}\boldsymbol{\mu}_{1}-\mathbf{\Sigma}%
_{2}^{-1}\boldsymbol{\mu}_{2}\right)  +\frac{1}{2}\mathbf{x}^{T}\left(
\mathbf{\Sigma}_{2}^{-1}\mathbf{x}-\mathbf{\Sigma}_{1}^{-1}\mathbf{x}\right)
\label{Likelihood Ratio General Gaussian}\\
&  +\frac{1}{2}\boldsymbol{\mu}_{2}^{T}\mathbf{\Sigma}_{2}^{-1}\boldsymbol{\mu
}_{2}-\frac{1}{2}\boldsymbol{\mu}_{1}^{T}\mathbf{\Sigma}_{1}^{-1}%
\boldsymbol{\mu}_{1}\nonumber\\
&  +\frac{1}{2}\ln\left(  \left\vert \mathbf{\Sigma}_{2}\right\vert
^{1/2}\right)  -\frac{1}{2}\ln\left(  \left\vert \mathbf{\Sigma}%
_{1}\right\vert ^{1/2}\right)  \overset{\omega_{1}}{\underset{\omega
_{2}}{\gtrless}}\ln\left(  \eta\right)  \text{,}\nonumber
\end{align}
where the decision threshold $\ln\left(  \eta\right)  $ is defined by the
algebraic expression%
\[
\ln\left(  \eta\right)  \triangleq\ln\left(  P\left(  \omega_{2}\right)
\right)  -\ln\left(  P\left(  \omega_{1}\right)  \right)  \text{,}%
\]
defines the general form of the discriminant function for the general
Gaussian, binary classification problem, where no costs ($C_{11}=C_{22}=0$)
are associated with correct decisions, and $C_{12}=C_{21}=1$ are costs
associated with incorrect decisions
\citep{VanTrees1968,Duda2001}%
.

The likelihood ratio test in Eq. (\ref{Likelihood Ratio General Gaussian})
generates decision boundaries $D\left(  \mathbf{x}\right)  $ that are
determined by the vector equation:%
\begin{align}
D\left(  \mathbf{x}\right)   &  :\mathbf{x}^{T}\mathbf{\Sigma}_{1}%
^{-1}\boldsymbol{\mu}_{1}-\frac{1}{2}\mathbf{x}^{T}\mathbf{\Sigma}_{1}%
^{-1}\mathbf{x}-\frac{1}{2}\boldsymbol{\mu}_{1}^{T}\mathbf{\Sigma}_{1}%
^{-1}\boldsymbol{\mu}_{1}-\frac{1}{2}\ln\left(  \left\vert \mathbf{\Sigma}%
_{1}\right\vert ^{1/2}\right) \label{General Gaussian Decision Boundary}\\
&  -\mathbf{x}^{T}\mathbf{\Sigma}_{2}^{-1}\boldsymbol{\mu}_{2}+\frac{1}%
{2}\mathbf{x}^{T}\mathbf{\Sigma}_{2}^{-1}\mathbf{x+}\frac{1}{2}\boldsymbol{\mu
}_{2}^{T}\mathbf{\Sigma}_{2}^{-1}\boldsymbol{\mu}_{2}+\frac{1}{2}\ln\left(
\left\vert \mathbf{\Sigma}_{2}\right\vert ^{1/2}\right) \nonumber\\
&  =\ln\left(  P\left(  \omega_{2}\right)  \right)  -\ln\left(  P\left(
\omega_{1}\right)  \right) \nonumber
\end{align}
and are characterized by the class of hyperquadric decision surfaces which
include hyperplanes, pairs of hyperplanes, hyperspheres, hyperellipsoids,
hyperparaboloids, and hyperhyperboloids
\citep{VanTrees1968,Duda2001}%
.

Let $\widehat{\Lambda}\left(  \mathbf{x}\right)  $ denote the transform
$\ln\left(  \Lambda\left(  \mathbf{x}\right)  \right)  $ of the likelihood
ratio $\Lambda\left(  \mathbf{x}\right)  $ in Eq.
(\ref{Likelihood Ratio Gaussian Data}). I will now use Eqs
(\ref{Likelihood Ratio General Gaussian}) and
(\ref{General Gaussian Decision Boundary}) to derive statistical equilibrium
laws that are satisfied by classification systems.

\subsection{Equilibrium Laws for Classification Systems}

It can be argued that the decision threshold $\eta$ of a likelihood test
$\Lambda\left(  \mathbf{x}\right)  \overset{\omega_{1}}{\underset{\omega
_{2}}{\gtrless}}\eta$ for a binary classification system does not depend on
prior probabilities $P\left(  \omega_{1}\right)  $ and $P\left(  \omega
_{2}\right)  $ and costs $C_{11}$, $C_{21}$, $C_{22}$, and $C_{12}$. It can
also be argued that the expected risk $\mathfrak{R}_{\mathfrak{\min}}$ for a
binary classification system is minimized by letting $\eta=1$.

Therefore, let $\eta=0$ in Eq. (\ref{Likelihood Ratio General Gaussian}). It
follows that the likelihood ratio test $\widehat{\Lambda}\left(
\mathbf{x}\right)  \overset{\omega_{1}}{\underset{\omega_{2}}{\gtrless}}0$
generates decision boundaries $D\left(  \mathbf{x}\right)  $ that satisfy the
vector equation:%
\begin{align}
D\left(  \mathbf{x}\right)   &  :\mathbf{x}^{T}\mathbf{\Sigma}_{1}%
^{-1}\boldsymbol{\mu}_{1}-\frac{1}{2}\mathbf{x}^{T}\mathbf{\Sigma}_{1}%
^{-1}\mathbf{x}-\frac{1}{2}\boldsymbol{\mu}_{1}^{T}\mathbf{\Sigma}_{1}%
^{-1}\boldsymbol{\mu}_{1}-\frac{1}{2}\ln\left(  \left\vert \mathbf{\Sigma}%
_{1}\right\vert ^{1/2}\right)
\label{Vector Equation Binary Decision Boundary}\\
&  -\mathbf{x}^{T}\mathbf{\Sigma}_{2}^{-1}\boldsymbol{\mu}_{2}+\frac{1}%
{2}\mathbf{x}^{T}\mathbf{\Sigma}_{2}^{-1}\mathbf{x+}\frac{1}{2}\boldsymbol{\mu
}_{2}^{T}\mathbf{\Sigma}_{2}^{-1}\boldsymbol{\mu}_{2}+\frac{1}{2}\ln\left(
\left\vert \mathbf{\Sigma}_{2}\right\vert ^{1/2}\right) \nonumber\\
&  =0\text{,}\nonumber
\end{align}
where the likelihood ratio $p\left(  \widehat{\Lambda}\left(  \mathbf{x}%
\right)  |\omega_{1}\right)  $ given class $\omega_{1}$ satisfies the vector
expression:%
\begin{align*}
p\left(  \widehat{\Lambda}\left(  \mathbf{x}\right)  |\omega_{1}\right)   &
=\mathbf{x}^{T}\mathbf{\Sigma}_{1}^{-1}\boldsymbol{\mu}_{1}-\frac{1}%
{2}\mathbf{x}^{T}\mathbf{\Sigma}_{1}^{-1}\mathbf{x}-\frac{1}{2}\boldsymbol{\mu
}_{1}^{T}\mathbf{\Sigma}_{1}^{-1}\boldsymbol{\mu}_{1}-\frac{1}{2}\ln\left(
\left\vert \mathbf{\Sigma}_{1}\right\vert ^{1/2}\right) \\
&  =p\left(  \mathbf{x}|\omega_{1}\right)  \text{,}%
\end{align*}
and the likelihood ratio $p\left(  \widehat{\Lambda}\left(  \mathbf{x}\right)
|\omega_{2}\right)  $ given class $\omega_{2}$ satisfies the vector
expression:%
\begin{align*}
p\left(  \widehat{\Lambda}\left(  \mathbf{x}\right)  |\omega_{2}\right)   &
=\mathbf{x}^{T}\mathbf{\Sigma}_{2}^{-1}\boldsymbol{\mu}_{2}-\frac{1}%
{2}\mathbf{x}^{T}\mathbf{\Sigma}_{2}^{-1}\mathbf{x-}\frac{1}{2}\boldsymbol{\mu
}_{2}^{T}\mathbf{\Sigma}_{2}^{-1}\boldsymbol{\mu}_{2}-\frac{1}{2}\ln\left(
\left\vert \mathbf{\Sigma}_{2}\right\vert ^{1/2}\right) \\
&  =p\left(  \mathbf{x}|\omega_{2}\right)  \text{.}%
\end{align*}

Therefore, the class-conditional likelihood ratios $p\left(  \widehat{\Lambda
}\left(  \mathbf{x}\right)  |\omega_{1}\right)  $ and $p\left(
\widehat{\Lambda}\left(  \mathbf{x}\right)  |\omega_{2}\right)  $ satisfy the
vector equation:%
\begin{equation}
p\left(  \widehat{\Lambda}\left(  \mathbf{x}\right)  |\omega_{1}\right)
-p\left(  \widehat{\Lambda}\left(  \mathbf{x}\right)  |\omega_{2}\right)  =0
\label{Equilibrium Equation for Class Conditional Densities}%
\end{equation}
such that the probability density functions $p\left(  \mathbf{x}|\omega
_{1}\right)  $ and $p\left(  \mathbf{x}|\omega_{2}\right)  $ for class
$\omega_{1}$ and class $\omega_{2}$ are balanced with each other:%
\[
p\left(  \mathbf{x}|\omega_{1}\right)  =p\left(  \mathbf{x}|\omega_{2}\right)
\text{,}%
\]
and the class-conditional likelihood ratios $p\left(  \widehat{\Lambda}\left(
\mathbf{x}\right)  |\omega_{1}\right)  $ and $p\left(  \widehat{\Lambda
}\left(  \mathbf{x}\right)  |\omega_{2}\right)  $ satisfy the statistical
equilibrium equation:%
\[
p\left(  \widehat{\Lambda}\left(  \mathbf{x}\right)  |\omega_{1}\right)
=p\left(  \widehat{\Lambda}\left(  \mathbf{x}\right)  |\omega_{2}\right)
\text{.}%
\]

Accordingly, the likelihood ratio test $\widehat{\Lambda}\left(
\mathbf{x}\right)  \overset{\omega_{1}}{\underset{\omega_{2}}{\gtrless}}0$%
\begin{align}
\widehat{\Lambda}\left(  \mathbf{x}\right)   &  =\mathbf{x}^{T}\left(
\mathbf{\Sigma}_{1}^{-1}\boldsymbol{\mu}_{1}-\mathbf{\Sigma}_{2}%
^{-1}\boldsymbol{\mu}_{2}\right) \label{General Gaussian Equalizer Rule}\\
&  +\frac{1}{2}\mathbf{x}^{T}\left(  \mathbf{\Sigma}_{2}^{-1}\mathbf{x}%
-\mathbf{\Sigma}_{1}^{-1}\mathbf{x}\right) \nonumber\\
&  +\frac{1}{2}\boldsymbol{\mu}_{2}^{T}\mathbf{\Sigma}_{2}^{-1}\boldsymbol{\mu
}_{2}-\frac{1}{2}\boldsymbol{\mu}_{1}^{T}\mathbf{\Sigma}_{1}^{-1}%
\boldsymbol{\mu}_{1}\nonumber\\
&  +\frac{1}{2}\ln\left(  \left\vert \mathbf{\Sigma}_{2}\right\vert
^{1/2}\right)  -\frac{1}{2}\ln\left(  \left\vert \mathbf{\Sigma}%
_{1}\right\vert ^{1/2}\right)  \overset{\omega_{1}}{\underset{\omega
_{2}}{\gtrless}}0\nonumber
\end{align}
generates decision boundaries $D\left(  \mathbf{x}\right)  $%
\begin{align*}
\widehat{\Lambda}\left(  \mathbf{x}\right)   &  :\mathbf{x}^{T}\mathbf{\Sigma
}_{1}^{-1}\boldsymbol{\mu}_{1}-\frac{1}{2}\mathbf{x}^{T}\mathbf{\Sigma}%
_{1}^{-1}\mathbf{x}-\frac{1}{2}\boldsymbol{\mu}_{1}^{T}\mathbf{\Sigma}%
_{1}^{-1}\boldsymbol{\mu}_{1}-\frac{1}{2}\ln\left(  \left\vert \mathbf{\Sigma
}_{1}\right\vert ^{1/2}\right) \\
&  -\mathbf{x}^{T}\mathbf{\Sigma}_{2}^{-1}\boldsymbol{\mu}_{2}+\frac{1}%
{2}\mathbf{x}^{T}\mathbf{\Sigma}_{2}^{-1}\mathbf{x+}\frac{1}{2}\boldsymbol{\mu
}_{2}^{T}\mathbf{\Sigma}_{2}^{-1}\boldsymbol{\mu}_{2}+\frac{1}{2}\ln\left(
\left\vert \mathbf{\Sigma}_{2}\right\vert ^{1/2}\right) \\
&  =0
\end{align*}
that determine decision regions $Z_{1}$ and $Z_{2}$ for which \emph{the
likelihood ratio} $\widehat{\Lambda}\left(  \mathbf{x}\right)  $ \emph{is in
statistical equilibrium}:%
\[
p\left(  \widehat{\Lambda}\left(  \mathbf{x}\right)  |\omega_{1}\right)
=p\left(  \widehat{\Lambda}\left(  \mathbf{x}\right)  |\omega_{2}\right)
\text{,}%
\]
and the expected risk $\mathfrak{R}_{\mathfrak{\min}}\left(
Z|\widehat{\Lambda}\left(  \mathbf{x}\right)  \right)  $ over the decision
space $Z$%
\[
\mathfrak{R}_{\mathfrak{\min}}\left(  Z|\widehat{\Lambda}\left(
\mathbf{x}\right)  \right)  =\int_{Z_{2}}p\left(  \widehat{\Lambda}\left(
\mathbf{x}\right)  |\omega_{1}\right)  d\widehat{\Lambda}+\int_{Z_{1}}p\left(
\widehat{\Lambda}\left(  \mathbf{x}\right)  |\omega_{2}\right)
d\widehat{\Lambda}%
\]
is determined by the \emph{total allowed probability of error} that is given
by the $P_{2}$ integral for the decision error $P\left(  D_{1}|\omega
_{2}\right)  $:
\[
P_{2}=\int_{Z_{1}}p\left(  \widehat{\Lambda}\left(  \mathbf{x}\right)
|\omega_{2}\right)  d\widehat{\Lambda}=P\left(  D_{1}|\omega_{2}\right)
\text{,}%
\]
over the $Z_{1}$ decision region, and the \emph{total allowed probability of
error} that is given by the $P_{1}$ integral for the decision error $P\left(
D_{2}|\omega_{1}\right)  $:%
\[
P_{1}=\int_{Z_{2}}p\left(  \widehat{\Lambda}\left(  \mathbf{x}\right)
|\omega_{1}\right)  d\widehat{\Lambda}=P\left(  D_{2}|\omega_{1}\right)
\text{,}%
\]
over the $Z_{2}$ decision region, where the $Z_{1}$ and $Z_{2}$ decision
regions involve either regions of data distribution overlap or tail regions of
data distributions.

Returning to Eq. (\ref{Equilibrium Equation for Class Conditional Densities}),
the vector equation%
\[
p\left(  \widehat{\Lambda}\left(  \mathbf{x}\right)  |\omega_{1}\right)
-p\left(  \widehat{\Lambda}\left(  \mathbf{x}\right)  |\omega_{2}\right)  =0
\]
indicates that the total allowed probabilities of decision errors $P\left(
D_{1}|\omega_{2}\right)  $ and $P\left(  D_{2}|\omega_{1}\right)  $ for a
classification system involve \emph{opposing forces} that depend on the
likelihood ratio test $\widehat{\Lambda}\left(  \mathbf{x}\right)  =p\left(
\widehat{\Lambda}\left(  \mathbf{x}\right)  |\omega_{1}\right)  -p\left(
\widehat{\Lambda}\left(  \mathbf{x}\right)  |\omega_{2}\right)
\overset{\omega_{1}}{\underset{\omega_{2}}{\gtrless}}0$ and the corresponding
decision boundary $p\left(  \widehat{\Lambda}\left(  \mathbf{x}\right)
|\omega_{1}\right)  -p\left(  \widehat{\Lambda}\left(  \mathbf{x}\right)
|\omega_{2}\right)  =0$. This indicates that classification systems $p\left(
\widehat{\Lambda}\left(  \mathbf{x}\right)  |\omega_{1}\right)  -p\left(
\widehat{\Lambda}\left(  \mathbf{x}\right)  |\omega_{2}\right)
\overset{\omega_{1}}{\underset{\omega_{2}}{\gtrless}}0$ that are determined by
the likelihood ratio test in Eq. (\ref{General Gaussian Equalizer Rule}) are
in \emph{statistical equilibrium}.

I will now motivate the concept of forces associated with binary
classification systems in terms of forces that are associated with positions
and potential locations of feature vectors. By way of motivation, I will
define feature vectors in terms of samples of electromagnetic forces.

\subsection{Sampled Sources of Electromagnetic Forces}

Pattern vectors or feature vectors are generated by a wide variety of physical
sources or mechanisms. Generally speaking, pattern vectors are generated by
\emph{sampled sources} of electromagnetic radiation which include radio waves,
infrared radiation, visible light, ultraviolet radiation, X-rays, and gamma rays.

For example, hydrophones and microphones are acoustic mechanisms that convert
sound waves into electrical signals, whereas seismometers measure motion of
the ground. Radar systems use radio waves to detect objects such as aircraft,
ships, guided missiles, weather formations, and terrains. Lidar systems use
light waves in the form of pulsed laser to make high resolution maps of the
Earth. Multispectral or hyperspectral sensors use infrared and ultraviolet
radiation to collect and process information to find objects, identify
materials, and detect processes. Accordingly, feature vectors can be defined
as samples of electromagnetic forces, where any given sample of an
electromagnetic force is a pattern vector that has a magnitude and a direction.

Pattern vectors may also be generated by responses to electromagnetic
radiation. For example, electrophoresis is a method that sorts proteins
according to their response to an electric field: electrophoresis may be used
to determine DNA sequence genotypes, or genotypes that are based on the length
of specific DNA fragments.

Electromagnetic radiation sources are characterized in terms of energy,
wavelength, or frequency, where any given source of electromagnetic radiation
is determined by electromagnetic forces. Generally speaking, a force is
anything that can change an object's speed or direction of motion. Therefore,
let a feature vector be the locus of a directed, straight line segment that
starts at the origin and terminates at the endpoint of the feature vector.
Accordingly, the endpoint of a feature vector is due to an electromagnetic
force that changed an object's speed and direction of motion, where the
magnitude of the feature vector represents the distance an object traveled
from the origin.

So, we can think about forces associated with positions and potential
locations of endpoints of feature vectors, where the forces involve sampled
sources of electromagnetic radiation: feature vectors that have magnitudes and directions.

I will now derive a system of fundamental equations of binary classification
for a classification system in statistical equilibrium.

\subsection{Fundamental Equations of Binary Classification}

Take any given collection of data that is drawn from class-conditional
probability density functions $p\left(  \mathbf{x}|\omega_{1}\right)  $and
$p\left(  \mathbf{x}|\omega_{2}\right)  $ that have unknown statistical
distributions. Let the decision space $Z$ and the corresponding $Z_{1}$ and
$Z_{2}$ decision regions be determined by either overlapping or
non-overlapping data distributions, as depicted in Figs
$\ref{Bayes' Error and Decision Regions Overlapping Data}$ or
$\ref{Bayes' Error and Decision Regions Non-overlapping Data}$.

For overlapping data distributions, the $Z_{1}$ and $Z_{2}$ decision regions
involve pattern vectors $\mathbf{x}$ that are generated according to $p\left(
\mathbf{x}|\omega_{1}\right)  $ \emph{and} $p\left(  \mathbf{x}|\omega
_{2}\right)  $, where pattern vectors $\mathbf{x}$ that are generated
according to $p\left(  \mathbf{x}|\omega_{2}\right)  $ and are located in
region $Z_{1}$ contribute to the $P_{2}$ decision error and the average risk
$\mathfrak{R}_{\mathfrak{\min}}\left(  Z\right)  $ in the decision space $Z$,
and pattern vectors $\mathbf{x}$ that are generated according to $p\left(
\mathbf{x}|\omega_{1}\right)  $ and are located in region $Z_{2}$ contribute
to the $P_{1}$ decision error and the average risk $\mathfrak{R}%
_{\mathfrak{\min}}\left(  Z\right)  $ in the decision space $Z$. For
non-overlapping data distributions, the $Z_{1}$ and $Z_{2}$ decision regions
involve pattern vectors $\mathbf{x}$ that are generated according to either
$p\left(  \mathbf{x}|\omega_{1}\right)  $ \emph{or} $p\left(  \mathbf{x}%
|\omega_{2}\right)  $ so that the average risk $\mathfrak{R}_{\mathfrak{\min}%
}\left(  Z\right)  $ in the decision space $Z$ is zero.

Therefore, let the $P_{2}$ integral for the probability of decision error
$P\left(  D_{1}|\omega_{2}\right)  $ be given by the integral%
\begin{align}
\mathfrak{R}_{\mathfrak{\min}}\left(  Z|p\left(  \widehat{\Lambda}\left(
\mathbf{x}\right)  |\omega_{2}\right)  \right)   &  =\int_{Z_{1}}p\left(
\widehat{\Lambda}\left(  \mathbf{x}\right)  |\omega_{2}\right)
d\widehat{\Lambda}+\int_{Z_{2}}p\left(  \widehat{\Lambda}\left(
\mathbf{x}\right)  |\omega_{2}\right)  d\widehat{\Lambda}%
\label{Total Risk of Class Two}\\
&  =\mathfrak{R}_{\mathfrak{\min}}\left(  Z_{1}|p\left(  \widehat{\Lambda
}\left(  \mathbf{x}\right)  |\omega_{2}\right)  \right)  +\overline
{\mathfrak{R}}_{\mathfrak{\min}}\left(  Z_{2}|p\left(  \widehat{\Lambda
}\left(  \mathbf{x}\right)  |\omega_{2}\right)  \right) \nonumber\\
&  =\int_{Z}p\left(  \widehat{\Lambda}\left(  \mathbf{x}\right)  |\omega
_{2}\right)  d\widehat{\Lambda}\text{,}\nonumber
\end{align}
over the $Z_{1}$ and $Z_{2}$ decision regions, where the integral $\int%
_{Z_{2}}p\left(  \widehat{\Lambda}\left(  \mathbf{x}\right)  |\omega
_{2}\right)  d\widehat{\Lambda}$ over the $Z_{2}$ decision region, which
involves pattern vectors $\mathbf{x}$ that are generated according to
$p\left(  \mathbf{x}|\omega_{2}\right)  $ and are located in region $Z_{2}$,
and is denoted by $\overline{\mathfrak{R}}_{\mathfrak{\min}}\left(
Z_{2}|p\left(  \widehat{\Lambda}\left(  \mathbf{x}\right)  |\omega_{2}\right)
\right)  $, involves forces that \emph{oppose} forces associated with the
average risk $\mathfrak{R}_{\mathfrak{\min}}\left(  Z\right)  $ in the
decision space $Z$: because the expected risk $\mathfrak{R}_{\mathfrak{\min}%
}\left(  Z_{1}|p\left(  \widehat{\Lambda}\left(  \mathbf{x}\right)
|\omega_{2}\right)  \right)  $ for class $\omega_{2}$ involves pattern vectors
$\mathbf{x}$ that are generated according to $p\left(  \mathbf{x}|\omega
_{2}\right)  $ and are located in region $Z_{1}$.

Thus, Eq. (\ref{Total Risk of Class Two}) involves opposing forces for pattern
vectors $\mathbf{x}$ that are generated according to $p\left(  \mathbf{x}%
|\omega_{2}\right)  $, where opposing forces are associated with risks and
counter risks that depend on positions and potential locations of pattern
vectors $\mathbf{x}$ in the $Z_{1}$ and $Z_{2}$ decision regions. Therefore,
Eq. (\ref{Total Risk of Class Two}) also represents the \emph{total allowed
eigenenergy} of a classification system for class $\omega_{2}$, where the
total allowed eigenenergy is the eigenenergy associated with the position or
location of the likelihood ratio $p\left(  \widehat{\Lambda}\left(
\mathbf{x}\right)  |\omega_{2}\right)  $ given class $\omega_{2}$.

Accordingly, let $E_{2}\left(  Z|p\left(  \widehat{\Lambda}\left(
\mathbf{x}\right)  |\omega_{2}\right)  \right)  $ denote the total allowed
eigenenergy of a classification system for class $\omega_{2}$:%
\[
E_{2}\left(  Z|p\left(  \widehat{\Lambda}\left(  \mathbf{x}\right)
|\omega_{2}\right)  \right)  =E_{2}\left(  Z_{1}|p\left(  \widehat{\Lambda
}\left(  \mathbf{x}\right)  |\omega_{2}\right)  \right)  +E_{2}\left(
Z_{2}|p\left(  \widehat{\Lambda}\left(  \mathbf{x}\right)  |\omega_{2}\right)
\right)  \text{,}%
\]
where $E_{2}\left(  Z_{1}|p\left(  \widehat{\Lambda}\left(  \mathbf{x}\right)
|\omega_{2}\right)  \right)  $ and $E_{2}\left(  Z_{2}|p\left(
\widehat{\Lambda}\left(  \mathbf{x}\right)  |\omega_{2}\right)  \right)  $ are
eigenenergies associated with the risk $\mathfrak{R}_{\mathfrak{\min}}\left(
Z_{1}|p\left(  \widehat{\Lambda}\left(  \mathbf{x}\right)  |\omega_{2}\right)
\right)  $ and the counter risk $\overline{\mathfrak{R}}_{\mathfrak{\min}%
}\left(  Z_{2}|p\left(  \widehat{\Lambda}\left(  \mathbf{x}\right)
|\omega_{2}\right)  \right)  $ in the $Z_{1}$ and $Z_{2}$ decision regions for
class $\omega_{2}$ that depend on positions and potential locations of pattern
vectors $\mathbf{x}$ that are generated according to $p\left(  \mathbf{x}%
|\omega_{2}\right)  $. It follows that the integral in Eq.
(\ref{Total Risk of Class Two}) also represents the total allowed eigenenergy
$E_{2}\left(  Z|p\left(  \widehat{\Lambda}\left(  \mathbf{x}\right)
|\omega_{2}\right)  \right)  $ of a classification system for class
$\omega_{2}$:%
\begin{align}
E_{2}\left(  Z|p\left(  \widehat{\Lambda}\left(  \mathbf{x}\right)
|\omega_{2}\right)  \right)   &  =\int_{Z_{1}}p\left(  \widehat{\Lambda
}\left(  \mathbf{x}\right)  |\omega_{2}\right)  d\widehat{\Lambda}+\int%
_{Z_{2}}p\left(  \widehat{\Lambda}\left(  \mathbf{x}\right)  |\omega
_{2}\right)  d\widehat{\Lambda}\label{Energy of Class Two}\\
&  =E_{2}\left(  Z_{1}|p\left(  \widehat{\Lambda}\left(  \mathbf{x}\right)
|\omega_{2}\right)  \right)  +E_{2}\left(  Z_{2}|p\left(  \widehat{\Lambda
}\left(  \mathbf{x}\right)  |\omega_{2}\right)  \right) \nonumber\\
&  =\int_{Z}p\left(  \widehat{\Lambda}\left(  \mathbf{x}\right)  |\omega
_{2}\right)  d\widehat{\Lambda}\text{,}\nonumber
\end{align}
which is given by the area under the class-conditional likelihood ratio
$p\left(  \widehat{\Lambda}\left(  \mathbf{x}\right)  |\omega_{2}\right)  $
over the decision space $Z$, where the total allowed eigenenergy is the
eigenenergy associated with the position or location of the likelihood ratio
$p\left(  \widehat{\Lambda}\left(  \mathbf{x}\right)  |\omega_{2}\right)  $
given class $\omega_{2}$.

Likewise, let the $P_{1}$ integral for the probability of decision error
$P\left(  D_{2}|\omega_{1}\right)  $ be given by the integral%
\begin{align}
\mathfrak{R}_{\mathfrak{\min}}\left(  Z|p\left(  \widehat{\Lambda}\left(
\mathbf{x}\right)  |\omega_{1}\right)  \right)   &  =\int_{Z_{2}}p\left(
\widehat{\Lambda}\left(  \mathbf{x}\right)  |\omega_{1}\right)
d\widehat{\Lambda}+\int_{Z_{1}}p\left(  \widehat{\Lambda}\left(
\mathbf{x}\right)  |\omega_{1}\right)  d\widehat{\Lambda}%
\label{Total Risk of Class One}\\
&  =\mathfrak{R}_{\mathfrak{\min}}\left(  Z_{2}|p\left(  \widehat{\Lambda
}\left(  \mathbf{x}\right)  |\omega_{1}\right)  \right)  +\overline
{\mathfrak{R}}_{\mathfrak{\min}}\left(  Z_{1}|p\left(  \widehat{\Lambda
}\left(  \mathbf{x}\right)  |\omega_{1}\right)  \right) \nonumber\\
&  =\int_{Z}p\left(  \widehat{\Lambda}\left(  \mathbf{x}\right)  |\omega
_{1}\right)  d\widehat{\Lambda}\text{,}\nonumber
\end{align}
over the $Z_{1}$ and $Z_{2}$ decision regions, where the integral $\int%
_{Z_{1}}p\left(  \widehat{\Lambda}\left(  \mathbf{x}\right)  |\omega
_{1}\right)  d\widehat{\Lambda}$ over the $Z_{1}$ decision region, which
involves pattern vectors $\mathbf{x}$ that are generated according to
$p\left(  \mathbf{x}|\omega_{1}\right)  $ and are located in region $Z_{1}$,
and is denoted by $\overline{\mathfrak{R}}_{\mathfrak{\min}}\left(
Z_{1}|p\left(  \widehat{\Lambda}\left(  \mathbf{x}\right)  |\omega_{1}\right)
\right)  $, involves forces that \emph{oppose} forces associated with the
average risk $\mathfrak{R}_{\mathfrak{\min}}\left(  Z\right)  $ in the
decision space $Z$: because the expected risk $\mathfrak{R}_{\mathfrak{\min}%
}\left(  Z_{2}|p\left(  \widehat{\Lambda}\left(  \mathbf{x}\right)
|\omega_{1}\right)  \right)  $ for class $\omega_{1}$ involves pattern vectors
$\mathbf{x}$ that are generated according to $p\left(  \mathbf{x}|\omega
_{1}\right)  $ and are located in region $Z_{2}$.

Thus, Eq. (\ref{Total Risk of Class One}) involves opposing forces for pattern
vectors $\mathbf{x}$ that are generated according to $p\left(  \mathbf{x}%
|\omega_{1}\right)  $, where opposing forces are associated with the risk
$\mathfrak{R}_{\mathfrak{\min}}\left(  Z_{2}|p\left(  \widehat{\Lambda}\left(
\mathbf{x}\right)  |\omega_{1}\right)  \right)  $ and the counter risk
$\overline{\mathfrak{R}}_{\mathfrak{\min}}\left(  Z_{1}|p\left(
\widehat{\Lambda}\left(  \mathbf{x}\right)  |\omega_{1}\right)  \right)  $ for
class $\omega_{1}$: which depend on positions and potential locations of
pattern vectors $\mathbf{x}$ in the $Z_{2}$ and $Z_{1}$ decision regions.
Therefore, Eq. (\ref{Total Risk of Class One}) also represents the \emph{total
allowed eigenenergy} of a classification system for class $\omega_{1}$, where
the total allowed eigenenergy is the eigenenergy associated with the position
or location of the likelihood ratio $p\left(  \widehat{\Lambda}\left(
\mathbf{x}\right)  |\omega_{1}\right)  $ given class $\omega_{1}$.

Accordingly, let $E_{1}\left(  Z|p\left(  \widehat{\Lambda}\left(
\mathbf{x}\right)  |\omega_{1}\right)  \right)  $ denote the total allowed
eigenenergy of a classification system for class $\omega_{1}$:%
\[
E_{1}\left(  Z|p\left(  \widehat{\Lambda}\left(  \mathbf{x}\right)
|\omega_{1}\right)  \right)  =E_{1}\left(  Z_{1}|p\left(  \widehat{\Lambda
}\left(  \mathbf{x}\right)  |\omega_{1}\right)  \right)  +E_{1}\left(
Z_{2}|p\left(  \widehat{\Lambda}\left(  \mathbf{x}\right)  |\omega_{1}\right)
\right)  \text{,}%
\]
where $E_{1}\left(  Z_{1}|p\left(  \widehat{\Lambda}\left(  \mathbf{x}\right)
|\omega_{1}\right)  \right)  $ and $E_{1}\left(  Z_{2}|p\left(
\widehat{\Lambda}\left(  \mathbf{x}\right)  |\omega_{1}\right)  \right)  $ are
eigenenergies associated with the counter risk $\overline{\mathfrak{R}%
}_{\mathfrak{\min}}\left(  Z_{1}|p\left(  \widehat{\Lambda}\left(
\mathbf{x}\right)  |\omega_{1}\right)  \right)  $ and the risk $\mathfrak{R}%
_{\mathfrak{\min}}\left(  Z_{2}|p\left(  \widehat{\Lambda}\left(
\mathbf{x}\right)  |\omega_{1}\right)  \right)  $ in the $Z_{1}$ and $Z_{2}$
decision regions for class $\omega_{1}$ that depend on positions and potential
locations of pattern vectors $\mathbf{x}$ that are generated according to
$p\left(  \mathbf{x}|\omega_{1}\right)  $. It follows that the integral in Eq.
(\ref{Total Risk of Class One}) also represents the total allowed eigenenergy
$E_{1}\left(  Z|p\left(  \widehat{\Lambda}\left(  \mathbf{x}\right)
|\omega_{1}\right)  \right)  $ of a classification system for class
$\omega_{1}$:%
\begin{align}
E_{1}\left(  Z|\omega_{1}\right)   &  =\int_{Z_{2}}p\left(  \widehat{\Lambda
}\left(  \mathbf{x}\right)  |\omega_{1}\right)  d\widehat{\Lambda}+\int%
_{Z_{1}}p\left(  \widehat{\Lambda}\left(  \mathbf{x}\right)  |\omega
_{1}\right)  d\widehat{\Lambda}\label{Energy of Class One}\\
&  =E_{1}\left(  Z_{2}|p\left(  \widehat{\Lambda}\left(  \mathbf{x}\right)
|\omega_{1}\right)  \right)  +E_{1}\left(  Z_{1}|p\left(  \widehat{\Lambda
}\left(  \mathbf{x}\right)  |\omega_{1}\right)  \right) \nonumber\\
&  =\int_{Z}p\left(  \widehat{\Lambda}\left(  \mathbf{x}\right)  |\omega
_{1}\right)  d\widehat{\Lambda}\text{,}\nonumber
\end{align}
which is given by the area under the class-conditional likelihood ratio
$p\left(  \widehat{\Lambda}\left(  \mathbf{x}\right)  |\omega_{1}\right)  $
over the decision space $Z$, where the total allowed eigenenergy is the
eigenenergy associated with the position or location of the likelihood ratio
$p\left(  \widehat{\Lambda}\left(  \mathbf{x}\right)  |\omega_{1}\right)  $
given class $\omega_{1}$.

Using Eqs (\ref{Total Risk of Class Two}) and (\ref{Total Risk of Class One}),
it follows that the expected risk $\mathfrak{R}_{\mathfrak{\min}}$ of a binary
classification system is given by the integral:%
\begin{align}
\mathfrak{R}_{\mathfrak{\min}}\left(  Z|\widehat{\Lambda}\left(
\mathbf{x}\right)  \right)   &  =\int_{Z}p\left(  \widehat{\Lambda}\left(
\mathbf{x}\right)  |\omega_{1}\right)  d\widehat{\Lambda}+\int_{Z}p\left(
\widehat{\Lambda}\left(  \mathbf{x}\right)  |\omega_{2}\right)
d\widehat{\Lambda}\label{Total Allowed Risk of Classification System}\\
&  =\mathfrak{R}_{\mathfrak{\min}}\left(  Z|p\left(  \widehat{\Lambda}\left(
\mathbf{x}\right)  |\omega_{1}\right)  \right)  +\mathfrak{R}_{\mathfrak{\min
}}\left(  Z|p\left(  \widehat{\Lambda}\left(  \mathbf{x}\right)  |\omega
_{2}\right)  \right)  \text{,}\nonumber
\end{align}
over a finite decision space $Z$, where the expected risk $\mathfrak{R}%
_{\mathfrak{\min}}$ of a classification system is the probability of
misclassification or decision error.

Using Eqs (\ref{Energy of Class Two}) and (\ref{Energy of Class One}), it
follows that the total allowed eigenenergy of a classification system is given
by the corresponding integral:%
\begin{align}
E\left(  Z|\widehat{\Lambda}\left(  \mathbf{x}\right)  \right)   &  =\int%
_{Z}p\left(  \widehat{\Lambda}\left(  \mathbf{x}\right)  |\omega_{1}\right)
d\widehat{\Lambda}+\int_{Z}p\left(  \widehat{\Lambda}\left(  \mathbf{x}%
\right)  |\omega_{2}\right)  d\widehat{\Lambda}%
\label{Total Allowed Energy of Classification System}\\
&  =E_{1}\left(  Z|p\left(  \widehat{\Lambda}\left(  \mathbf{x}\right)
|\omega_{1}\right)  \right)  +E_{2}\left(  Z|p\left(  \widehat{\Lambda}\left(
\mathbf{x}\right)  |\omega_{2}\right)  \right)  \text{,}\nonumber
\end{align}
over a finite decision space $Z$, where the total allowed eigenenergy of a
classification system is the eigenenergy associated with the position or
location of the likelihood ratio $p\left(  \widehat{\Lambda}\left(
\mathbf{x}\right)  |\omega_{1}\right)  -p\left(  \widehat{\Lambda}\left(
\mathbf{x}\right)  |\omega_{2}\right)  $ and the corresponding decision
boundary $D\left(  \mathbf{x}\right)  $: $p\left(  \widehat{\Lambda}\left(
\mathbf{x}\right)  |\omega_{1}\right)  -p\left(  \widehat{\Lambda}\left(
\mathbf{x}\right)  |\omega_{2}\right)  =0$.

\begin{figure}[ptb]%
\centering
\fbox{\includegraphics[
height=2.5875in,
width=3.4411in
]%
{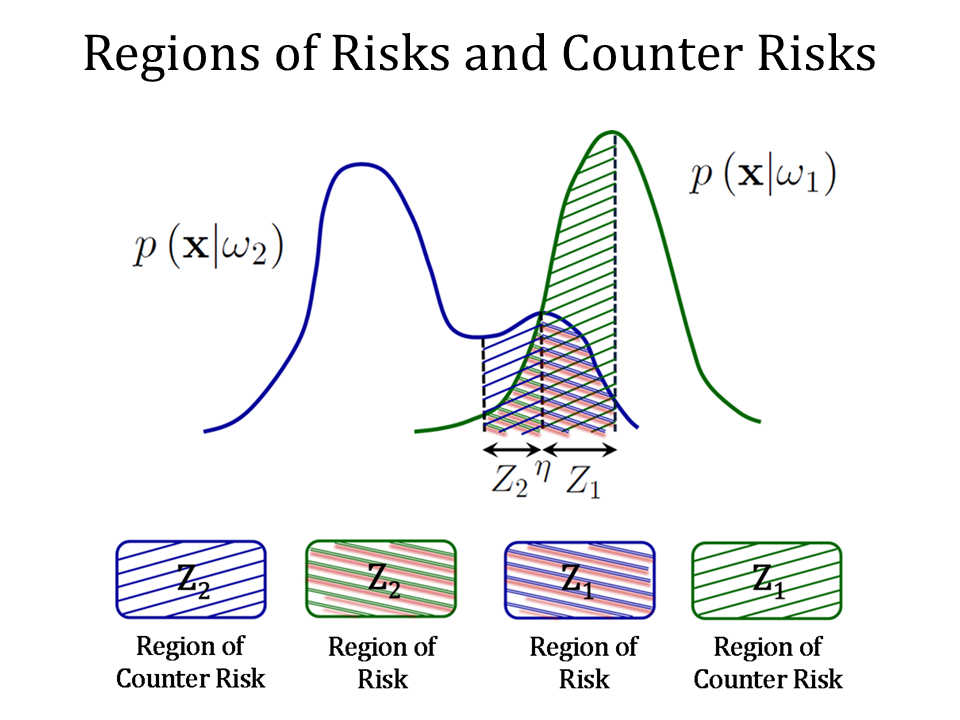}%
}\caption{Counter risks $\overline{\mathfrak{R}}_{\mathfrak{\min}}\left(
Z_{1}|p\left(  \protect\widehat{\Lambda}\left(  \mathbf{x}\right)  |\omega
_{1}\right)  \right)  $ and $\overline{\mathfrak{R}}_{\mathfrak{\min}}\left(
Z_{2}|p\left(  \protect\widehat{\Lambda}\left(  \mathbf{x}\right)  |\omega
_{2}\right)  \right)  $ in the $Z_{1}$ and $Z_{2}$ decisions regions are
\emph{opposing forces} for the risks or decision errors\emph{ }$\mathfrak{R}%
_{\mathfrak{\min}}\left(  Z_{1}|p\left(  \protect\widehat{\Lambda}\left(
\mathbf{x}\right)  |\omega_{2}\right)  \right)  $ and $\mathfrak{R}%
_{\mathfrak{\min}}\left(  Z_{2}|p\left(  \protect\widehat{\Lambda}\left(
\mathbf{x}\right)  |\omega_{1}\right)  \right)  $ in the $Z_{1}$ and $Z_{2}$
decisions regions.}%
\label{Regions of Risks and Counter Risks}%
\end{figure}

Figure $\ref{Regions of Risks and Counter Risks}$ provides an overview of
regions of risks and counter risks in the $Z_{1}$ and $Z_{2}$ decisions
regions, where the counter risks for class $\omega_{1}$ and class $\omega_{2}%
$:%
\[
\overline{\mathfrak{R}}_{\mathfrak{\min}}\left(  Z_{1}|p\left(
\widehat{\Lambda}\left(  \mathbf{x}\right)  |\omega_{1}\right)  \right)
\text{ \ and \ }\overline{\mathfrak{R}}_{\mathfrak{\min}}\left(
Z_{2}|p\left(  \widehat{\Lambda}\left(  \mathbf{x}\right)  |\omega_{2}\right)
\right)
\]
are \emph{opposing forces} for the risks or decision errors for class
$\omega_{1}$ and class $\omega_{2}$:%
\[
\mathfrak{R}_{\mathfrak{\min}}\left(  Z_{1}|p\left(  \widehat{\Lambda}\left(
\mathbf{x}\right)  |\omega_{2}\right)  \right)  \text{ \ and \ }%
\mathfrak{R}_{\mathfrak{\min}}\left(  Z_{2}|p\left(  \widehat{\Lambda}\left(
\mathbf{x}\right)  |\omega_{1}\right)  \right)  \text{.}%
\]%

I\ will now show that classification systems seek a point of statistical
equilibrium where the opposing forces and influences of a system are balanced
with each other, and the eigenenergy and the corresponding expected risk of a
classification system are minimized. As a result, I\ will devise a system of
fundamental equations of binary classification that must be satisfied by
classification systems in statistical equilibrium.

\subsection{Classification Systems in Statistical Equilibrium}

Take any given decision boundary $D\left(  \mathbf{x}\right)  $%
\begin{align*}
D\left(  \mathbf{x}\right)   &  :\mathbf{x}^{T}\mathbf{\Sigma}_{1}%
^{-1}\boldsymbol{\mu}_{1}-\frac{1}{2}\mathbf{x}^{T}\mathbf{\Sigma}_{1}%
^{-1}\mathbf{x}-\frac{1}{2}\boldsymbol{\mu}_{1}^{T}\mathbf{\Sigma}_{1}%
^{-1}\boldsymbol{\mu}_{1}-\frac{1}{2}\ln\left(  \left\vert \mathbf{\Sigma}%
_{1}\right\vert ^{1/2}\right) \\
&  -\mathbf{x}^{T}\mathbf{\Sigma}_{2}^{-1}\boldsymbol{\mu}_{2}+\frac{1}%
{2}\mathbf{x}^{T}\mathbf{\Sigma}_{2}^{-1}\mathbf{x+}\frac{1}{2}\boldsymbol{\mu
}_{2}^{T}\mathbf{\Sigma}_{2}^{-1}\boldsymbol{\mu}_{2}+\frac{1}{2}\ln\left(
\left\vert \mathbf{\Sigma}_{2}\right\vert ^{1/2}\right) \\
&  =0\text{,}%
\end{align*}
that is generated according to the likelihood ratio test $\widehat{\Lambda
}\left(  \mathbf{x}\right)  \overset{\omega_{1}}{\underset{\omega
_{2}}{\gtrless}}0$%
\begin{align*}
\widehat{\Lambda}\left(  \mathbf{x}\right)   &  =\mathbf{x}^{T}\left(
\mathbf{\Sigma}_{1}^{-1}\boldsymbol{\mu}_{1}-\mathbf{\Sigma}_{2}%
^{-1}\boldsymbol{\mu}_{2}\right) \\
&  +\frac{1}{2}\mathbf{x}^{T}\left(  \mathbf{\Sigma}_{2}^{-1}\mathbf{x}%
-\mathbf{\Sigma}_{1}^{-1}\mathbf{x}\right) \\
&  +\frac{1}{2}\boldsymbol{\mu}_{2}^{T}\mathbf{\Sigma}_{2}^{-1}\boldsymbol{\mu
}_{2}-\frac{1}{2}\boldsymbol{\mu}_{1}^{T}\mathbf{\Sigma}_{1}^{-1}%
\boldsymbol{\mu}_{1}\\
&  +\frac{1}{2}\ln\left(  \left\vert \mathbf{\Sigma}_{2}\right\vert
^{1/2}\right)  -\frac{1}{2}\ln\left(  \left\vert \mathbf{\Sigma}%
_{1}\right\vert ^{1/2}\right)  \overset{\omega_{1}}{\underset{\omega
_{2}}{\gtrless}}0\text{,}%
\end{align*}
where $\boldsymbol{\mu}_{1}$ and $\boldsymbol{\mu}_{2}$ are $d$-component mean
vectors, $\mathbf{\Sigma}_{1}$ and $\mathbf{\Sigma}_{2}$ are $d$-by-$d$
covariance matrices, $\mathbf{\Sigma}^{-1}$ and $\left\vert \mathbf{\Sigma
}\right\vert $ denote the inverse and determinant of a covariance matrix, and
$\omega_{1}$ or $\omega_{2}$ is the true data category.

Given that the decision boundary $D\left(  \mathbf{x}\right)  $ and the
likelihood ratio $\widehat{\Lambda}\left(  \mathbf{x}\right)  $ satisfy the
vector equation%
\begin{equation}
p\left(  \widehat{\Lambda}\left(  \mathbf{x}\right)  |\omega_{1}\right)
-p\left(  \widehat{\Lambda}\left(  \mathbf{x}\right)  |\omega_{2}\right)
=0\text{,} \label{Vector Equation of Likelihood Ratio and Decision Boundary}%
\end{equation}
where the decision boundary $D\left(  \mathbf{x}\right)  $ and the likelihood
ratio $\widehat{\Lambda}\left(  \mathbf{x}\right)  $ are in statistical
equilibrium%
\begin{equation}
p\left(  \widehat{\Lambda}\left(  \mathbf{x}\right)  |\omega_{1}\right)
=p\left(  \widehat{\Lambda}\left(  \mathbf{x}\right)  |\omega_{2}\right)
\text{,}
\label{Equilibrium Equation of Likelihood Ratio and Decision Boundary}%
\end{equation}
it follows that the decision boundary $D\left(  \mathbf{x}\right)  $ and the
likelihood ratio $\widehat{\Lambda}\left(  \mathbf{x}\right)  $ satisfy the
integral equation%
\begin{equation}
\int_{Z}p\left(  \widehat{\Lambda}\left(  \mathbf{x}\right)  |\omega
_{1}\right)  d\widehat{\Lambda}=\int_{Z}p\left(  \widehat{\Lambda}\left(
\mathbf{x}\right)  |\omega_{2}\right)  d\widehat{\Lambda}\text{,}
\label{Integral Equation of Likelihood Ratio and Decision Boundary}%
\end{equation}
over the decision space $Z$.

Therefore, the decision boundary $D\left(  \mathbf{x}\right)  $ and the
likelihood ratio $\widehat{\Lambda}\left(  \mathbf{x}\right)  $ satisfy the
fundamental integral equation of binary classification%
\begin{align}
f\left(  \widehat{\Lambda}\left(  \mathbf{x}\right)  \right)   &  =\int%
_{Z_{1}}p\left(  \widehat{\Lambda}\left(  \mathbf{x}\right)  |\omega
_{1}\right)  d\widehat{\Lambda}+\int_{Z_{2}}p\left(  \widehat{\Lambda}\left(
\mathbf{x}\right)  |\omega_{1}\right)  d\widehat{\Lambda}%
\label{Equalizer Rule}\\
&  =\int_{Z_{1}}p\left(  \widehat{\Lambda}\left(  \mathbf{x}\right)
|\omega_{2}\right)  d\widehat{\Lambda}+\int_{Z_{2}}p\left(  \widehat{\Lambda
}\left(  \mathbf{x}\right)  |\omega_{2}\right)  d\widehat{\Lambda}%
\text{,}\nonumber
\end{align}
over the $Z_{1}$ and $Z_{2}$ decision regions, such that the expected risk
$\mathfrak{R}_{\mathfrak{\min}}\left(  Z|\widehat{\Lambda}\left(
\mathbf{x}\right)  \right)  $ and the corresponding eigenenergy $E_{\min
}\left(  Z|\widehat{\Lambda}\left(  \mathbf{x}\right)  \right)  $ of the
classification system are minimized, and the classification system is in
statistical equilibrium.

Given Eqs (\ref{Total Allowed Risk of Classification System}) and
(\ref{Equalizer Rule}), it follows that the forces associated with the counter
risk $\overline{\mathfrak{R}}_{\mathfrak{\min}}\left(  Z_{1}|p\left(
\widehat{\Lambda}\left(  \mathbf{x}\right)  |\omega_{1}\right)  \right)  $ and
the risk $\mathfrak{R}_{\mathfrak{\min}}\left(  Z_{2}|p\left(
\widehat{\Lambda}\left(  \mathbf{x}\right)  |\omega_{1}\right)  \right)  $ in
the $Z_{1}$ and $Z_{2}$ decision regions: which are related to positions and
potential locations of pattern vectors $\mathbf{x}$ that are generated
according to $p\left(  \mathbf{x}|\omega_{1}\right)  $, must be \emph{equal}
to the forces associated with the risk $\mathfrak{R}_{\mathfrak{\min}}\left(
Z_{1}|p\left(  \widehat{\Lambda}\left(  \mathbf{x}\right)  |\omega_{2}\right)
\right)  $ and the counter risk $\overline{\mathfrak{R}}_{\mathfrak{\min}%
}\left(  Z_{2}|p\left(  \widehat{\Lambda}\left(  \mathbf{x}\right)
|\omega_{2}\right)  \right)  $ in the $Z_{1}$ and $Z_{2}$ decision regions:
which are related to positions and potential locations of pattern vectors
$\mathbf{x}$ that are generated according to $p\left(  \mathbf{x}|\omega
_{2}\right)  $.

Given Eqs (\ref{Total Allowed Energy of Classification System}) and
(\ref{Equalizer Rule}), it follows that the eigenenergy associated with the
position or location of the likelihood ratio $p\left(  \widehat{\Lambda
}\left(  \mathbf{x}\right)  |\omega_{2}\right)  $ given class $\omega_{2}$
must be \emph{equal} to the eigenenergy associated with the position or
location of the likelihood ratio $p\left(  \widehat{\Lambda}\left(
\mathbf{x}\right)  |\omega_{1}\right)  $ given class $\omega_{1}$.

I will now use Eq. (\ref{Equalizer Rule}) to develop an integral equation of
binary classification, where the forces associated with the counter risk
$\overline{\mathfrak{R}}_{\mathfrak{\min}}\left(  Z_{1}|p\left(
\widehat{\Lambda}\left(  \mathbf{x}\right)  |\omega_{1}\right)  \right)  $ for
class $\omega_{1}$ and the risk $\mathfrak{R}_{\mathfrak{\min}}\left(
Z_{1}|p\left(  \widehat{\Lambda}\left(  \mathbf{x}\right)  |\omega_{2}\right)
\right)  $ for class $\omega_{2}$ in the $Z_{1}$ decision region are
\emph{balanced} with the forces associated with the counter risk
$\overline{\mathfrak{R}}_{\mathfrak{\min}}\left(  Z_{2}|p\left(
\widehat{\Lambda}\left(  \mathbf{x}\right)  |\omega_{2}\right)  \right)  $ for
class $\omega_{2}$ and the risk $\mathfrak{R}_{\mathfrak{\min}}\left(
Z_{2}|p\left(  \widehat{\Lambda}\left(  \mathbf{x}\right)  |\omega_{1}\right)
\right)  $ for class $\omega_{1}$ in the $Z_{2}$ decision region.

Given that the counter risks $\overline{\mathfrak{R}}_{\mathfrak{\min}}\left(
Z_{1}|p\left(  \widehat{\Lambda}\left(  \mathbf{x}\right)  |\omega_{1}\right)
\right)  $ and $\overline{\mathfrak{R}}_{\mathfrak{\min}}\left(
Z_{2}|p\left(  \widehat{\Lambda}\left(  \mathbf{x}\right)  |\omega_{2}\right)
\right)  $ are opposing forces for the risks $\mathfrak{R}_{\mathfrak{\min}%
}\left(  Z_{1}|p\left(  \widehat{\Lambda}\left(  \mathbf{x}\right)
|\omega_{2}\right)  \right)  $ and $\mathfrak{R}_{\mathfrak{\min}}\left(
Z_{2}|p\left(  \widehat{\Lambda}\left(  \mathbf{x}\right)  |\omega_{1}\right)
\right)  $, let the counter risks $\overline{\mathfrak{R}}_{\mathfrak{\min}%
}\left(  Z_{1}|p\left(  \widehat{\Lambda}\left(  \mathbf{x}\right)
|\omega_{1}\right)  \right)  $ and $\overline{\mathfrak{R}}_{\mathfrak{\min}%
}\left(  Z_{2}|p\left(  \widehat{\Lambda}\left(  \mathbf{x}\right)
|\omega_{2}\right)  \right)  $ for class $\omega_{1}$ and class $\omega_{2}$
in the $Z_{1}$ and $Z_{2}$ decision regions be positive forces:%
\[
\overline{\mathfrak{R}}_{\mathfrak{\min}}\left(  Z_{1}|p\left(
\widehat{\Lambda}\left(  \mathbf{x}\right)  |\omega_{1}\right)  \right)
>0\text{ \ and \ }\overline{\mathfrak{R}}_{\mathfrak{\min}}\left(
Z_{2}|p\left(  \widehat{\Lambda}\left(  \mathbf{x}\right)  |\omega_{2}\right)
\right)  >0\text{,}%
\]
and let the risks $\mathfrak{R}_{\mathfrak{\min}}\left(  Z_{1}|p\left(
\widehat{\Lambda}\left(  \mathbf{x}\right)  |\omega_{2}\right)  \right)  $ and
$\mathfrak{R}_{\mathfrak{\min}}\left(  Z_{2}|p\left(  \widehat{\Lambda}\left(
\mathbf{x}\right)  |\omega_{1}\right)  \right)  $ for class $\omega_{2}$ and
class $\omega_{1}$ in the $Z_{1}$ and $Z_{2}$ decision regions be negative
forces:%
\[
\mathfrak{R}_{\mathfrak{\min}}\left(  Z_{1}|p\left(  \widehat{\Lambda}\left(
\mathbf{x}\right)  |\omega_{2}\right)  \right)  <0\text{ \ and \ }%
\mathfrak{R}_{\mathfrak{\min}}\left(  Z_{2}|p\left(  \widehat{\Lambda}\left(
\mathbf{x}\right)  |\omega_{1}\right)  \right)  <0\text{.}%
\]

Given these assumptions and using Eqs (\ref{Total Risk of Class Two}) -
(\ref{Total Allowed Energy of Classification System}) and Eq.
(\ref{Equalizer Rule}), it follows that the \emph{expected risk}
$\mathfrak{R}_{\mathfrak{\min}}\left(  Z|\widehat{\Lambda}\left(
\mathbf{x}\right)  \right)  $%
\begin{align*}
\mathfrak{R}_{\mathfrak{\min}}\left(  Z|\widehat{\Lambda}\left(
\mathbf{x}\right)  \right)   &  =\overline{\mathfrak{R}}_{\mathfrak{\min}%
}\left(  Z_{1}|p\left(  \widehat{\Lambda}\left(  \mathbf{x}\right)
|\omega_{1}\right)  \right)  +\mathfrak{R}_{\mathfrak{\min}}\left(
Z_{2}|p\left(  \widehat{\Lambda}\left(  \mathbf{x}\right)  |\omega_{1}\right)
\right) \\
&  +\mathfrak{R}_{\mathfrak{\min}}\left(  Z_{1}|p\left(  \widehat{\Lambda
}\left(  \mathbf{x}\right)  |\omega_{2}\right)  \right)  +\overline
{\mathfrak{R}}_{\mathfrak{\min}}\left(  Z_{2}|p\left(  \widehat{\Lambda
}\left(  \mathbf{x}\right)  |\omega_{2}\right)  \right)
\end{align*}
and the corresponding \emph{eigenenergy} $E_{\min}\left(  Z|\widehat{\Lambda
}\left(  \mathbf{x}\right)  \right)  $%
\begin{align*}
E_{\min}\left(  Z|\widehat{\Lambda}\left(  \mathbf{x}\right)  \right)   &
=E_{\min}\left(  Z_{1}|p\left(  \widehat{\Lambda}\left(  \mathbf{x}\right)
|\omega_{1}\right)  \right)  +E_{\min}\left(  Z_{2}|p\left(  \widehat{\Lambda
}\left(  \mathbf{x}\right)  |\omega_{1}\right)  \right) \\
&  +E_{\min}\left(  Z_{1}|p\left(  \widehat{\Lambda}\left(  \mathbf{x}\right)
|\omega_{2}\right)  \right)  +E_{\min}\left(  Z_{2}|p\left(  \widehat{\Lambda
}\left(  \mathbf{x}\right)  |\omega_{2}\right)  \right)
\end{align*}
of classification systems are minimized in the following manner:%
\begin{align}
\mathfrak{R}_{\mathfrak{\min}}\left(  Z|\widehat{\Lambda}\left(
\mathbf{x}\right)  \right)   &  :\;\int_{Z_{1}}p\left(  \widehat{\Lambda
}\left(  \mathbf{x}\right)  |\omega_{1}\right)  d\widehat{\Lambda}-\int%
_{Z_{1}}p\left(  \widehat{\Lambda}\left(  \mathbf{x}\right)  |\omega
_{2}\right)  d\widehat{\Lambda}%
\label{Balancing of Bayes' Risks and Counteracting Risks}\\
&  =\int_{Z_{2}}p\left(  \widehat{\Lambda}\left(  \mathbf{x}\right)
|\omega_{2}\right)  d\widehat{\Lambda}-\int_{Z_{2}}p\left(  \widehat{\Lambda
}\left(  \mathbf{x}\right)  |\omega_{1}\right)  d\widehat{\Lambda}%
\text{,}\nonumber
\end{align}
where positive forces associated with the counter risk $\overline
{\mathfrak{R}}_{\mathfrak{\min}}\left(  Z_{1}|p\left(  \widehat{\Lambda
}\left(  \mathbf{x}\right)  |\omega_{1}\right)  \right)  $ for class
$\omega_{1}$ and negative forces associated with the risk $\mathfrak{R}%
_{\mathfrak{\min}}\left(  Z_{1}|p\left(  \widehat{\Lambda}\left(
\mathbf{x}\right)  |\omega_{2}\right)  \right)  $ for class $\omega_{2}$ in
the $Z_{1}$ decision region are \emph{balanced} with positive forces
associated with the counter risk $\overline{\mathfrak{R}}_{\mathfrak{\min}%
}\left(  Z_{2}|p\left(  \widehat{\Lambda}\left(  \mathbf{x}\right)
|\omega_{2}\right)  \right)  $ for class $\omega_{2}$ and negative forces
associated with the risk $\mathfrak{R}_{\mathfrak{\min}}\left(  Z_{2}|p\left(
\widehat{\Lambda}\left(  \mathbf{x}\right)  |\omega_{1}\right)  \right)  $ for
class $\omega_{1}$ in the $Z_{2}$ decision region:%
\begin{align*}
\mathfrak{R}_{\mathfrak{\min}}\left(  Z|\widehat{\Lambda}\left(
\mathbf{x}\right)  \right)  :  &  \overline{\mathfrak{R}}_{\mathfrak{\min}%
}\left(  Z_{1}|p\left(  \widehat{\Lambda}\left(  \mathbf{x}\right)
|\omega_{1}\right)  \right)  -\mathfrak{R}_{\mathfrak{\min}}\left(
Z_{1}|p\left(  \widehat{\Lambda}\left(  \mathbf{x}\right)  |\omega_{2}\right)
\right) \\
&  =\overline{\mathfrak{R}}_{\mathfrak{\min}}\left(  Z_{2}|p\left(
\widehat{\Lambda}\left(  \mathbf{x}\right)  |\omega_{2}\right)  \right)
-\mathfrak{R}_{\mathfrak{\min}}\left(  Z_{2}|p\left(  \widehat{\Lambda}\left(
\mathbf{x}\right)  |\omega_{1}\right)  \right)  \text{,}%
\end{align*}
and the \emph{eigenenergies} associated with the counter risk $\overline
{\mathfrak{R}}_{\mathfrak{\min}}\left(  Z_{1}|p\left(  \widehat{\Lambda
}\left(  \mathbf{x}\right)  |\omega_{1}\right)  \right)  $ for class
$\omega_{1}$ and the risk $\mathfrak{R}_{\mathfrak{\min}}\left(
Z_{1}|p\left(  \widehat{\Lambda}\left(  \mathbf{x}\right)  |\omega_{2}\right)
\right)  $ for class $\omega_{2}$ in the $Z_{1}$ decision region are
\emph{balanced} with the \emph{eigenenergies} associated with the counter risk
$\overline{\mathfrak{R}}_{\mathfrak{\min}}\left(  Z_{2}|p\left(
\widehat{\Lambda}\left(  \mathbf{x}\right)  |\omega_{2}\right)  \right)  $ for
class $\omega_{2}$ and the risk $\mathfrak{R}_{\mathfrak{\min}}\left(
Z_{2}|p\left(  \widehat{\Lambda}\left(  \mathbf{x}\right)  |\omega_{1}\right)
\right)  $ for class $\omega_{1}$ in the $Z_{2}$ decision region:%
\begin{align*}
E_{\min}\left(  Z|\widehat{\Lambda}\left(  \mathbf{x}\right)  \right)  :  &
E_{\min}\left(  Z_{1}|p\left(  \widehat{\Lambda}\left(  \mathbf{x}\right)
|\omega_{1}\right)  \right)  -E_{\min}\left(  Z_{1}|p\left(  \widehat{\Lambda
}\left(  \mathbf{x}\right)  |\omega_{2}\right)  \right) \\
&  =E_{\min}\left(  Z_{2}|p\left(  \widehat{\Lambda}\left(  \mathbf{x}\right)
|\omega_{2}\right)  \right)  -E_{\min}\left(  Z_{2}|p\left(  \widehat{\Lambda
}\left(  \mathbf{x}\right)  |\omega_{1}\right)  \right)  \text{.}%
\end{align*}

\begin{figure}[ptb]%
\centering
\fbox{\includegraphics[
height=2.5875in,
width=3.4411in
]%
{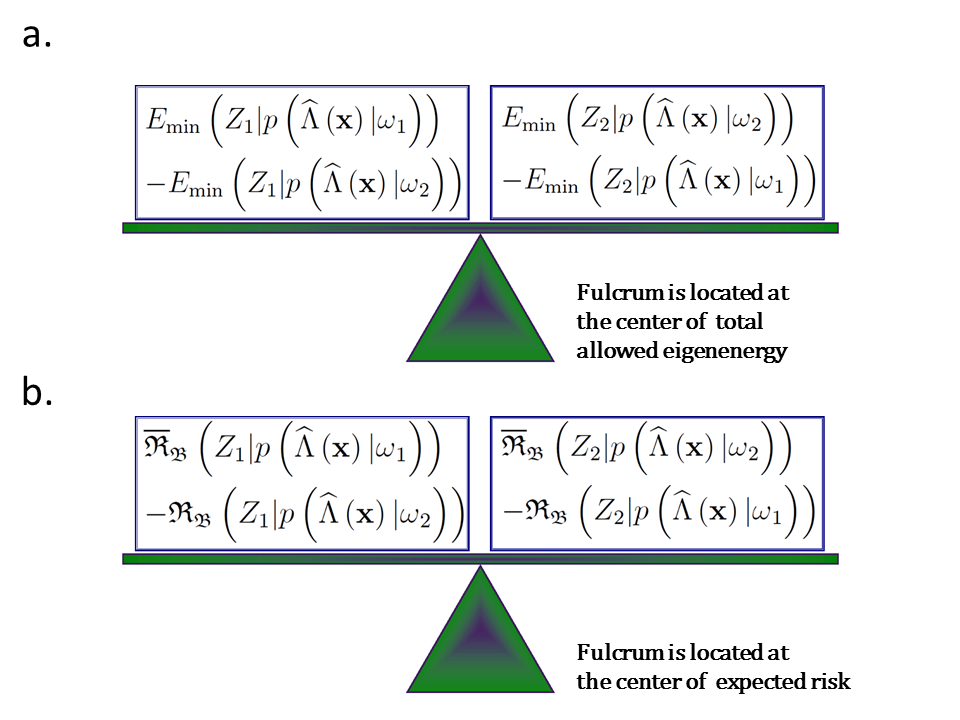}%
}\caption{The equilibrium point of a binary classification system determines a
statistical fulcrum which is located at $\left(  a\right)  $ a center of total
allowed eigenenergy $E_{\min}\left(  Z|\protect\widehat{\Lambda}\left(
\mathbf{x}\right)  \right)  $ and $\left(  b\right)  $ a corresponding center
of expected risk $\mathfrak{R}_{\mathfrak{\min}}\left(
Z|\protect\widehat{\Lambda}\left(  \mathbf{x}\right)  \right)  $.}%
\label{Balancing Eiigenenergies Risks and Counter Risks}%
\end{figure}

Therefore, it is concluded that the expected risk $\mathfrak{R}%
_{\mathfrak{\min}}\left(  Z|\widehat{\Lambda}\left(  \mathbf{x}\right)
\right)  $ and the corresponding eigenenergy $E_{\min}\left(
Z|\widehat{\Lambda}\left(  \mathbf{x}\right)  \right)  $ of a classification
system $p\left(  \widehat{\Lambda}\left(  \mathbf{x}\right)  |\omega
_{1}\right)  -p\left(  \widehat{\Lambda}\left(  \mathbf{x}\right)  |\omega
_{2}\right)  \overset{\omega_{1}}{\underset{\omega_{2}}{\gtrless}}0$ are
governed by the equilibrium point%
\[
p\left(  \widehat{\Lambda}\left(  \mathbf{x}\right)  |\omega_{1}\right)
-p\left(  \widehat{\Lambda}\left(  \mathbf{x}\right)  |\omega_{2}\right)  =0
\]
of the integral equation%
\begin{align*}
f\left(  \widehat{\Lambda}\left(  \mathbf{x}\right)  \right)   &  =\int%
_{Z_{1}}p\left(  \widehat{\Lambda}\left(  \mathbf{x}\right)  |\omega
_{1}\right)  d\widehat{\Lambda}+\int_{Z_{2}}p\left(  \widehat{\Lambda}\left(
\mathbf{x}\right)  |\omega_{1}\right)  d\widehat{\Lambda}\\
&  =\int_{Z_{1}}p\left(  \widehat{\Lambda}\left(  \mathbf{x}\right)
|\omega_{2}\right)  d\widehat{\Lambda}+\int_{Z_{2}}p\left(  \widehat{\Lambda
}\left(  \mathbf{x}\right)  |\omega_{2}\right)  d\widehat{\Lambda}\text{,}%
\end{align*}
over the $Z_{1}$ and $Z_{2}$ decision regions, where the opposing forces and
influences of the classification system are balanced with each other, such
that the eigenenergy and the expected risk of the classification system are
minimized, and the classification system is in statistical equilibrium. Figure
$\ref{Balancing Eiigenenergies Risks and Counter Risks}$ illustrates that the
equilibrium point of the integral equation $f\left(  \widehat{\Lambda}\left(
\mathbf{x}\right)  \right)  $ in Eq. (\ref{Equalizer Rule}) determines a
statistical fulcrum which is located at a center of total allowed eigenenergy
and a corresponding center of risk.%

Equations (\ref{Vector Equation of Likelihood Ratio and Decision Boundary}) -
(\ref{Balancing of Bayes' Risks and Counteracting Risks}) are the fundamental
equations of binary classification for a classification system in statistical
equilibrium. Because Eq.
(\ref{Balancing of Bayes' Risks and Counteracting Risks}) is derived from Eq.
(\ref{Equalizer Rule}), I\ will refer to Eq. (\ref{Equalizer Rule}) as the
fundamental integral equation of binary classification for a classification
system in statistical equilibrium.

Thus, it is concluded that classification systems seek a point of statistical
equilibrium where the opposing forces and influences of any given system are
balanced with each other, and the eigenenergy and the expected risk of the
classification system are minimized.

I will now consider likelihood ratio tests and decision boundaries within the
mathematical framework of a \emph{geometric locus}. I will show that the
\emph{point of statistical equilibrium} sought by classification systems is
\emph{a locus of points} that jointly satisfies Eqs
(\ref{Vector Equation of Likelihood Ratio and Decision Boundary}) -
(\ref{Balancing of Bayes' Risks and Counteracting Risks}).

\subsection{Geometric Locus}

The general idea of a curve or surface which at any point of it exhibits some
uniform property is expressed in geometry by the term \emph{locus}
\citep{Whitehead1911}%
. Generally speaking, a geometric locus is a curve or surface formed by
points, \emph{all} of which \emph{possess} some \emph{uniform property}. Any
given point on a geometric locus possesses a property which is common to all
points on the locus and \emph{no other points}. Any given geometric locus is
determined by either an algebraic or a vector equation, where the locus of an
algebraic or a vector equation is the location of all those points whose
coordinates are solutions of the equation. Standard locus methods involve
\emph{algebraic} equations of elliptic, parabolic, hyperbolic, circular, and
linear curves
\citep{Nichols1893,Tanner1898,Whitehead1911,Eisenhart1939}%
.

I will now use the notion of a geometric locus to show that the
\emph{equilibrium point} of a classification system involves a \emph{locus} of
\emph{points} $\mathbf{x}$ that \emph{jointly} satisfy the likelihood ratio
test in Eq. (\ref{General Gaussian Equalizer Rule}), the decision boundary in
Eq. (\ref{Vector Equation Binary Decision Boundary}), and the fundamental
equations of binary classification for a classification system in statistical
equilibrium in Eqs
(\ref{Vector Equation of Likelihood Ratio and Decision Boundary}) -
(\ref{Balancing of Bayes' Risks and Counteracting Risks}).

\subsection{Loci of Likelihood Ratios and Decision Boundaries}

Any given decision boundary $D\left(  \mathbf{x}\right)  $ in Eq.
(\ref{Vector Equation Binary Decision Boundary}) that is determined by the
likelihood ratio test $\widehat{\Lambda}\left(  \mathbf{x}\right)
\overset{\omega_{1}}{\underset{\omega_{2}}{\gtrless}}0$ in Eq.
(\ref{General Gaussian Equalizer Rule}), where the likelihood ratio
$\widehat{\Lambda}\left(  \mathbf{x}\right)  $%
\begin{align*}
\widehat{\Lambda}\left(  \mathbf{x}\right)   &  =p\left(  \widehat{\Lambda
}\left(  \mathbf{x}\right)  |\omega_{1}\right)  -p\left(  \widehat{\Lambda
}\left(  \mathbf{x}\right)  |\omega_{2}\right) \\
&  =\left(  \mathbf{x}^{T}\mathbf{\Sigma}_{1}^{-1}\boldsymbol{\mu}_{1}%
-\frac{1}{2}\mathbf{x}^{T}\mathbf{\Sigma}_{1}^{-1}\mathbf{x}-\frac{1}%
{2}\boldsymbol{\mu}_{1}^{T}\mathbf{\Sigma}_{1}^{-1}\boldsymbol{\mu}_{1}%
-\frac{1}{2}\ln\left(  \left\vert \mathbf{\Sigma}_{1}\right\vert
^{1/2}\right)  \right) \\
&  -\left(  \mathbf{x}^{T}\mathbf{\Sigma}_{2}^{-1}\boldsymbol{\mu}_{2}%
-\frac{1}{2}\mathbf{x}^{T}\mathbf{\Sigma}_{2}^{-1}\mathbf{x-}\frac{1}%
{2}\boldsymbol{\mu}_{2}^{T}\mathbf{\Sigma}_{2}^{-1}\boldsymbol{\mu}_{2}%
-\frac{1}{2}\ln\left(  \left\vert \mathbf{\Sigma}_{2}\right\vert
^{1/2}\right)  \right)
\end{align*}
and the decision boundary $D\left(  \mathbf{x}\right)  $ satisfy the vector
equation:%
\[
p\left(  \widehat{\Lambda}\left(  \mathbf{x}\right)  |\omega_{1}\right)
-p\left(  \widehat{\Lambda}\left(  \mathbf{x}\right)  |\omega_{2}\right)
=0\text{,}%
\]
the statistical equilibrium equation:%
\[
p\left(  \widehat{\Lambda}\left(  \mathbf{x}\right)  |\omega_{1}\right)
=p\left(  \widehat{\Lambda}\left(  \mathbf{x}\right)  |\omega_{2}\right)
\text{,}%
\]
the corresponding integral equation:%
\[
\int_{Z}p\left(  \widehat{\Lambda}\left(  \mathbf{x}\right)  |\omega
_{1}\right)  d\widehat{\Lambda}=\int_{Z}p\left(  \widehat{\Lambda}\left(
\mathbf{x}\right)  |\omega_{2}\right)  d\widehat{\Lambda}\text{,}%
\]
and the fundamental integral equation of binary classification:%
\begin{align*}
f\left(  \widehat{\Lambda}\left(  \mathbf{x}\right)  \right)   &  =\int%
_{Z_{1}}p\left(  \widehat{\Lambda}\left(  \mathbf{x}\right)  |\omega
_{1}\right)  d\widehat{\Lambda}+\int_{Z_{2}}p\left(  \widehat{\Lambda}\left(
\mathbf{x}\right)  |\omega_{1}\right)  d\widehat{\Lambda}\\
&  =\int_{Z_{1}}p\left(  \widehat{\Lambda}\left(  \mathbf{x}\right)
|\omega_{2}\right)  d\widehat{\Lambda}+\int_{Z_{2}}p\left(  \widehat{\Lambda
}\left(  \mathbf{x}\right)  |\omega_{2}\right)  d\widehat{\Lambda}\text{,}%
\end{align*}
for a classification system in statistical equilibrium:%
\begin{align*}
f\left(  \widehat{\Lambda}\left(  \mathbf{x}\right)  \right)   &
:\;\int_{Z_{1}}p\left(  \widehat{\Lambda}\left(  \mathbf{x}\right)
|\omega_{1}\right)  d\widehat{\Lambda}-\int_{Z_{1}}p\left(  \widehat{\Lambda
}\left(  \mathbf{x}\right)  |\omega_{2}\right)  d\widehat{\Lambda}\\
&  =\int_{Z_{2}}p\left(  \widehat{\Lambda}\left(  \mathbf{x}\right)
|\omega_{2}\right)  d\widehat{\Lambda}-\int_{Z_{2}}p\left(  \widehat{\Lambda
}\left(  \mathbf{x}\right)  |\omega_{1}\right)  d\widehat{\Lambda}\text{,}%
\end{align*}
is the \emph{locus}, i.e., the position or location, of all of the
\emph{endpoints} of the vectors $\mathbf{x}$ whose \emph{coordinates} are
\emph{solutions} of the vector \emph{equation}:%
\begin{align*}
D\left(  \mathbf{x}\right)   &  :\mathbf{x}^{T}\mathbf{\Sigma}_{1}%
^{-1}\boldsymbol{\mu}_{1}-\frac{1}{2}\mathbf{x}^{T}\mathbf{\Sigma}_{1}%
^{-1}\mathbf{x}-\frac{1}{2}\boldsymbol{\mu}_{1}^{T}\mathbf{\Sigma}_{1}%
^{-1}\boldsymbol{\mu}_{1}-\frac{1}{2}\ln\left(  \left\vert \mathbf{\Sigma}%
_{1}\right\vert ^{1/2}\right) \\
&  -\mathbf{x}^{T}\mathbf{\Sigma}_{2}^{-1}\boldsymbol{\mu}_{2}+\frac{1}%
{2}\mathbf{x}^{T}\mathbf{\Sigma}_{2}^{-1}\mathbf{x+}\frac{1}{2}\boldsymbol{\mu
}_{2}^{T}\mathbf{\Sigma}_{2}^{-1}\boldsymbol{\mu}_{2}+\frac{1}{2}\ln\left(
\left\vert \mathbf{\Sigma}_{2}\right\vert ^{1/2}\right) \\
&  =0\text{.}%
\end{align*}

It follows that any given point $\mathbf{x}$ that satisfies the above vector
equation must also satisfy the vector equation in Eq.
(\ref{Vector Equation of Likelihood Ratio and Decision Boundary}), the
statistical equilibrium and corresponding integral equations in Eqs
(\ref{Equilibrium Equation of Likelihood Ratio and Decision Boundary}) and
(\ref{Integral Equation of Likelihood Ratio and Decision Boundary}), and the
fundamental integral and corresponding integral equation of binary
classification in Eqs (\ref{Equalizer Rule}) and
(\ref{Balancing of Bayes' Risks and Counteracting Risks}), where endpoints of
vectors $\mathbf{x}$ whose coordinates are solutions of Eqs
(\ref{Vector Equation of Likelihood Ratio and Decision Boundary}) -
(\ref{Balancing of Bayes' Risks and Counteracting Risks}) are located in
regions that are either $\left(  1\right)  $ associated with decision errors
due to overlapping distributions or $\left(  2\right)  $ associated with no
decision errors due to non-overlapping distributions.

This indicates that the \emph{equilibrium point }of a classification system
involves \emph{a locus of points} that are determined by a nonlinear system of
locus equations. Given this assumption, I\ will now define a dual locus of
binary classifiers.

\subsection{Dual Locus of Binary Classifiers}

Let $\widehat{\Lambda}\left(  \mathbf{x}\right)  $ denote the likelihood ratio
$p\left(  \widehat{\Lambda}\left(  \mathbf{x}\right)  |\omega_{1}\right)
-p\left(  \widehat{\Lambda}\left(  \mathbf{x}\right)  |\omega_{2}\right)  $
for the binary classification system in Eq.
(\ref{General Gaussian Equalizer Rule}). Using the general idea of a geometric
locus, it follows that the likelihood ratio test $\widehat{\Lambda}\left(
\mathbf{x}\right)  \overset{\omega_{1}}{\underset{\omega_{2}}{\gtrless}}0$ in
Eq. (\ref{General Gaussian Equalizer Rule}) determines the \emph{dual locus}
of a decision boundary $D\left(  \mathbf{x}\right)  $ and a likelihood ratio
$\widehat{\Lambda}\left(  \mathbf{x}\right)  $, where any given
point\ $\mathbf{x}$ that satisfies the likelihood ratio $\widehat{\Lambda
}\left(  \mathbf{x}\right)  $ \emph{and} the decision boundary $D\left(
\mathbf{x}\right)  $ \emph{also} satisfies the fundamental integral equation
of binary classification for a classification system in statistical
equilibrium:%
\begin{align*}
f\left(  \widehat{\Lambda}\left(  \mathbf{x}\right)  \right)   &  =\int%
_{Z_{1}}p\left(  \widehat{\Lambda}\left(  \mathbf{x}\right)  |\omega
_{1}\right)  d\widehat{\Lambda}+\int_{Z_{2}}p\left(  \widehat{\Lambda}\left(
\mathbf{x}\right)  |\omega_{1}\right)  d\widehat{\Lambda}\\
&  =\int_{Z_{1}}p\left(  \widehat{\Lambda}\left(  \mathbf{x}\right)
|\omega_{2}\right)  d\widehat{\Lambda}+\int_{Z_{2}}p\left(  \widehat{\Lambda
}\left(  \mathbf{x}\right)  |\omega_{2}\right)  d\widehat{\Lambda}\text{,}%
\end{align*}
where%
\[
p\left(  \widehat{\Lambda}\left(  \mathbf{x}\right)  |\omega_{1}\right)
=p\left(  \widehat{\Lambda}\left(  \mathbf{x}\right)  |\omega_{2}\right)
\]
and%
\begin{align*}
f\left(  \widehat{\Lambda}\left(  \mathbf{x}\right)  \right)   &  :\int%
_{Z_{1}}p\left(  \widehat{\Lambda}\left(  \mathbf{x}\right)  |\omega
_{1}\right)  d\widehat{\Lambda}-\int_{Z_{1}}p\left(  \widehat{\Lambda}\left(
\mathbf{x}\right)  |\omega_{2}\right)  d\widehat{\Lambda}\\
&  =\int_{Z_{2}}p\left(  \widehat{\Lambda}\left(  \mathbf{x}\right)
|\omega_{2}\right)  d\widehat{\Lambda}-\int_{Z_{2}}p\left(  \widehat{\Lambda
}\left(  \mathbf{x}\right)  |\omega_{1}\right)  d\widehat{\Lambda}\text{,}%
\end{align*}

I\ will show that Eq. (\ref{General Gaussian Equalizer Rule}) is the basis of
a data-driven \emph{dual locus of binary classifiers} for Gaussian data:%
\[
\widehat{\Lambda}\left(  \mathbf{x}\right)  =\sum\nolimits_{i=1}^{l_{1}}%
\psi_{1_{i}}k_{\mathbf{x}_{1_{i}}}-\sum\nolimits_{i=1}^{l_{2}}\psi_{2_{i}%
}k_{\mathbf{x}_{2_{i}}}%
\]
where $\mathbf{x}_{1i}\sim p\left(  \mathbf{x}|\omega_{1}\right)  $,
$\mathbf{x}_{2i}\sim p\left(  \mathbf{x}|\omega_{2}\right)  $, $k_{\mathbf{x}%
_{1_{i}}}$ and $k_{\mathbf{x}_{2_{i}}}$ are reproducing kernels for respective
data points $\mathbf{x}_{1_{i}}$ and $\mathbf{x}_{2_{i}}$, and $\psi_{1i}$ and
$\psi_{2i}$ are scale factors that provide measures of likelihood for
respective data points $\mathbf{x}_{1i}$ and $\mathbf{x}_{2i}$ which lie in
either overlapping regions or tails regions of data distributions. The
coefficients $\left\{  \psi_{1i}\right\}  _{i=1}^{l_{1}}$ and $\left\{
\psi_{2i}\right\}  _{i=1}^{l_{2}}$ and the locus of data points $%
{\textstyle\sum\nolimits_{i=1}^{l_{1}}}
\psi_{1i}\mathbf{x}_{1i}-%
{\textstyle\sum\nolimits_{i=1}^{l_{2}}}
\psi_{2i}\mathbf{x}_{2i}$ are determined by finding the equilibrium point of
Eq. (\ref{Equalizer Rule}).

Therefore, take any given likelihood ratio $\widehat{\Lambda}\left(
\mathbf{x}\right)  $ and decision boundary $D\left(  \mathbf{x}\right)  $ that
are determined by the likelihood ratio test in Eq.
(\ref{General Gaussian Equalizer Rule}). It follows that all of the points
$\mathbf{x}$ that satisfy the decision boundary $D\left(  \mathbf{x}\right)  $
and the likelihood ratio $\widehat{\Lambda}\left(  \mathbf{x}\right)  $
\emph{must} possess a geometric and statistical \emph{property} which is
\emph{common} to all points $\mathbf{x}$ that jointly satisfy the fundamental
equations of binary classification in Eqs
(\ref{Vector Equation of Likelihood Ratio and Decision Boundary}) -
(\ref{Balancing of Bayes' Risks and Counteracting Risks}).

I\ will identify this essential property later on.

\subsection{Learning Dual Loci of Binary Classifiers}

I\ have conducted simulation studies which demonstrate that properly
regularized, linear kernel SVMs learn optimal linear decision boundaries for
\emph{any} two classes of Gaussian data, where $\mathbf{\Sigma}_{1}=$
$\mathbf{\Sigma}_{2}=\mathbf{\Sigma}$
\citep[see ][]{Reeves2009,Reeves2011}%
, including \emph{completely overlapping} data distributions
\citep[see ][]{Reeves2015resolving}%
. I\ have also conducted simulation studies which demonstrate that properly
regularized, second-order, polynomial kernel SVMs learn optimal decision
boundaries for data drawn from \emph{any} two Gaussian distributions,
including completely overlapping data distributions
\citep{,Reeves2015resolving}%
. In addition, given an effective value for the hyperparameter $\gamma$ of a
Gaussian reproducing kernel $\exp\left(  -\gamma\left\Vert \left(
\mathbf{\cdot}\right)  -\mathbf{s}\right\Vert ^{2}\right)  $, I\ have
conducted simulation studies which demonstrate that properly regularized,
Gaussian kernel SVMs learn optimal decision boundaries for data drawn from
\emph{any} two Gaussian distributions, including completely overlapping data distributions.

Suppose that the expression for the likelihood ration test in Eq.
(\ref{General Gaussian Equalizer Rule}) is not considered within the context
of a binary classification system in statistical equilibrium. Then my findings
are both unexpected and surprising. Using Eq.
(\ref{General Gaussian Equalizer Rule}), for any given pair of homogeneous
distributions, where $\mathbf{\Sigma}_{1}=\mathbf{\Sigma}_{2}=\mathbf{\Sigma}$
and $\boldsymbol{\mu}_{1}=\boldsymbol{\mu}_{2}=\boldsymbol{\mu}$, it follows
that the likelihood ratio test%
\begin{align*}
\widehat{\Lambda}\left(  \mathbf{x}\right)   &  =\mathbf{x}^{T}\mathbf{\Sigma
}^{-1}\left(  \boldsymbol{\mu}-\boldsymbol{\mu}\right)  +\frac{1}{2}%
\mathbf{x}^{T}\left(  \mathbf{\Sigma}^{-1}\mathbf{x}-\mathbf{\Sigma}%
^{-1}\mathbf{x}\right) \\
&  +\frac{1}{2}\boldsymbol{\mu}^{T}\mathbf{\Sigma}^{-1}\boldsymbol{\mu}%
-\frac{1}{2}\boldsymbol{\mu}^{T}\mathbf{\Sigma}^{-1}\boldsymbol{\mu}\\
&  +\frac{1}{2}\ln\left(  \left\vert \mathbf{\Sigma}\right\vert ^{1/2}\right)
-\frac{1}{2}\ln\left(  \left\vert \mathbf{\Sigma}\right\vert ^{1/2}\right)
\overset{\omega_{1}}{\underset{\omega_{2}}{\gtrless}}0\\
&  =0\overset{\omega_{1}}{\underset{\omega_{2}}{\gtrless}}0
\end{align*}
reduces to the constant $0$ and \emph{is undefined}. Therefore, the decision
rule \emph{and} the decision boundary \emph{are also undefined}. However, the
decision rule and the decision boundary \emph{are both defined} in terms of an
equilibrium point of a binary classification system for which \emph{a locus of
points}, which is determined by a nonlinear system of locus equations,
satisfies a statistical equilibrium equation.

I\ have conducted simulation studies which show that an optimal decision
function and decision boundary \emph{are both determined} by a system of
data-driven, locus equations, where \emph{data points} satisfy a
\emph{data-driven,} \emph{dual locus }of a likelihood ratio \emph{and} a
decision boundary. The system of data-driven, locus equations is based on the
support-vector network algorithm developed by
\citet*{Boser1992}
and
\citet*{Cortes1995}%
.

\subsubsection{Data-Driven Dual Locus of Optimal Linear Classifiers}

In this paper, I\ will devise data-driven, locus equations of likelihood ratio
tests and linear decision boundaries that satisfy data-driven versions of the
fundamental equations of binary classification in Eqs
(\ref{Vector Equation of Likelihood Ratio and Decision Boundary}) -
(\ref{Balancing of Bayes' Risks and Counteracting Risks}). The data-driven,
locus equations are based on variants of the inequality constrained
optimization problem for linear, polynomial, and Gaussian kernel SVMs.

I will formulate a system of data-driven, locus equations that determines an
optimal likelihood ratio test $\widehat{\Lambda}\left(  \mathbf{x}\right)
\overset{\omega_{1}}{\underset{\omega_{2}}{\gtrless}}0$ and a linear decision
boundary $D\left(  \mathbf{x}\right)  $, for any two classes of data: where
$\mathbf{\Sigma}_{1}=$ $\mathbf{\Sigma}_{2}=\mathbf{\Sigma}$, where a
\emph{data-driven, dual locus of points} determines the \emph{locus} of the
decision boundary in Eq. (\ref{Vector Equation Binary Decision Boundary})
\emph{and} the likelihood ratio in Eq. (\ref{General Gaussian Equalizer Rule}%
). The data-driven, dual locus of points satisfies a data-driven version of
the fundamental integral equation of binary classification in Eq.
(\ref{Equalizer Rule}). Thereby, the decision rule determines the locus of a
linear decision boundary for \emph{any} two classes of pattern vectors.

\subsubsection{Data-Driven Dual Locus of Optimal Quadratic Classifiers}

In this paper, I\ will also devise data-driven, locus equations of optimal
likelihood ratio tests and quadratic decision boundaries that satisfy
data-driven versions of the fundamental equations of binary classification in
Eqs (\ref{Vector Equation of Likelihood Ratio and Decision Boundary}) -
(\ref{Balancing of Bayes' Risks and Counteracting Risks}). The data-driven,
locus equations are based on variants of the inequality constrained
optimization problem for polynomial and Gaussian kernel SVMs.

I will formulate a system of data-driven, locus equations that determines an
optimal likelihood ratio test $\widehat{\Lambda}\left(  \mathbf{x}\right)
\overset{\omega_{1}}{\underset{\omega_{2}}{\gtrless}}0$ and a quadratic
decision boundary, for any two classes of data: where $\mathbf{\Sigma}_{1}%
\neq$ $\mathbf{\Sigma}_{2}$, where a \emph{data-driven, dual locus of points}
determines the locus of the decision boundary in\ Eq.
(\ref{Vector Equation Binary Decision Boundary}) and the likelihood ratio in
Eq. (\ref{General Gaussian Equalizer Rule}). The data-driven, dual locus of
points satisfies a data-driven version of the fundamental integral equation of
binary classification in Eq. (\ref{Equalizer Rule}). Thereby, the decision
rule determines the locus of a quadratic decision boundary for any two classes
of pattern vectors that have dissimilar covariance matrices. Furthermore, the
decision rule provides an estimate of the locus of a linear decision boundary
for any two classes of pattern vectors that have similar covariance matrices
and different mean vectors.

By way of motivation, graphical illustrations of decision regions in optimal
decision spaces are presented next.

\subsection{Decision Regions in Optimal Decision Spaces}

Take any two sets of data that have been drawn from any two Gaussian
distributions, where both data sets are completely characterized by the mean
vectors and the covariance matrices of the Gaussian distributions. I\ will
demonstrate that decision regions in optimal decision spaces satisfy
symmetrical border constraints in relation to circular, parabolic, elliptical,
hyperbolic, or linear decision boundaries. Thus, each decision region is
determined by a \emph{decision border} in relation to a \emph{decision
boundary}. Given that decision boundaries and borders are both determined by a
data-driven version of the fundamental integral equation of binary
classification in Eq. (\ref{Equalizer Rule}), it follows that the decision
boundaries and decision borders for a given classification system involve
similar types of curves or surfaces.

\subsubsection{Decision Regions for Dissimilar Covariance Matrices}

For data distributions that have dissimilar covariance matrices, I\ will
demonstrate that decision regions satisfy border constraints in relation to
circular, parabolic, elliptical, or hyperbolic decision boundaries. Thus, each
decision region is determined by a nonlinear decision border \emph{in relation
to} a nonlinear decision boundary.%
\begin{figure}[ptb]%
\centering
\fbox{\includegraphics[
height=2.5875in,
width=3.2206in
]%
{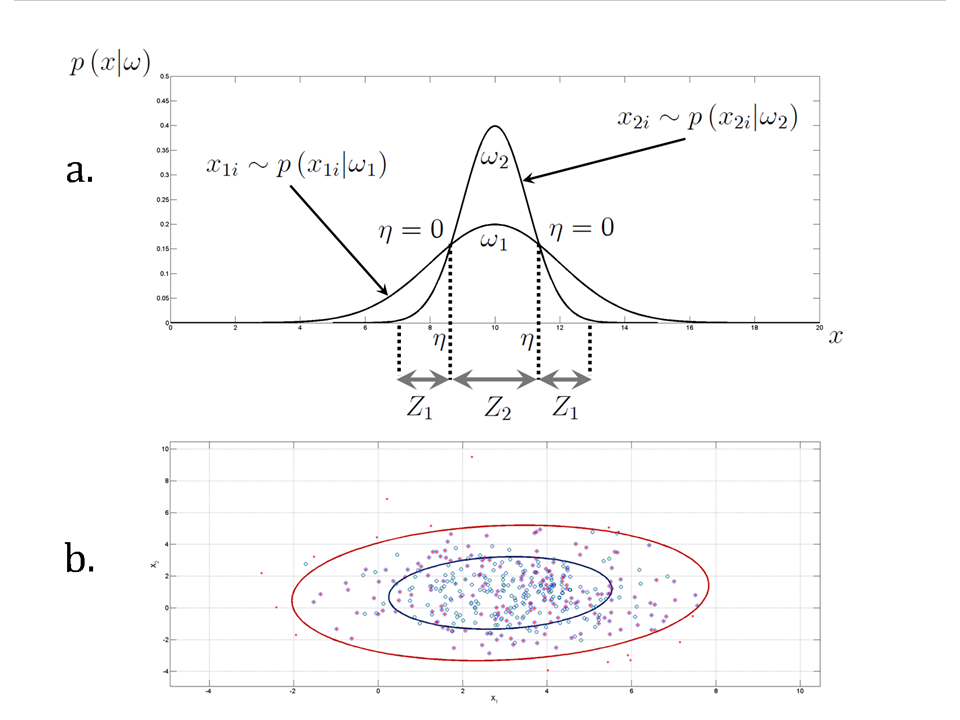}%
}\caption{For overlapping data distributions that have dissimilar covariance
matrices and common means, optimal decision functions divide feature spaces
$Z$ into decision regions $Z_{1}$ and $Z_{2}$ that have minimum expected risks
$\mathfrak{R}_{\mathfrak{\min}}\left(  Z_{1}\right)  $ and $\mathfrak{R}%
_{\mathfrak{\min}}\left(  Z_{2}\right)  $.}%
\label{Overlapping Data Same Mean One}%
\end{figure}

\paragraph{Example One}

Figure $\ref{Overlapping Data Same Mean One}$a illustrates how the likelihood
ratio test $\widehat{\Lambda}\left(  \mathbf{x}\right)  \overset{\omega
_{1}}{\underset{\omega_{2}}{\gtrless}}0$ in Eq.
(\ref{General Gaussian Equalizer Rule}) determines decision regions $Z_{1}$
and $Z_{2}$ for overlapping data distributions that have dissimilar covariance
matrices and common means, where the risk $\mathfrak{R}_{\mathfrak{\min}%
}\left(  Z_{1}|\omega_{2}\right)  $ for class $\omega_{2}$ and the risk
$\mathfrak{R}_{\mathfrak{\min}}\left(  Z_{2}|\omega_{1}\right)  $ for class
$\omega_{1}$ are determined by the statistical balancing feat:%
\[
\mathfrak{R}_{\mathfrak{\min}}\left(  Z|\widehat{\Lambda}\left(
\mathbf{x}\right)  \right)  :\overline{\mathfrak{R}}_{\mathfrak{\min}}\left(
Z_{1}|\omega_{1}\right)  -\mathfrak{R}_{\mathfrak{\min}}\left(  Z_{1}%
|\omega_{2}\right)  =\overline{\mathfrak{R}}_{\mathfrak{\min}}\left(
Z_{2}|\omega_{2}\right)  -\mathfrak{R}_{\mathfrak{\min}}\left(  Z_{2}%
|\omega_{1}\right)  \text{,}%
\]
over the $Z_{1}$ and $Z_{2}$ decision regions. Because the means are equal:
$\boldsymbol{\mu}_{1}=\boldsymbol{\mu}_{2}$, the decision border of region
$Z_{2}$ \emph{satisfies} the decision boundary, so that the decision border of
region $Z_{2}$ \emph{contains} region $Z_{2}$. So, if the decision boundary is
an elliptical curve or surface, then the decision border of region $Z_{1}$ is
also an elliptical curve or surface, so that region $Z_{1}$ is constrained by
the elliptical decision border of region $Z_{1}$ and the elliptical decision
boundary. For example, Fig. $\ref{Overlapping Data Same Mean One}$b depicts an
overview of a decision space that has been generated in MATLAB, where the
elliptical decision border of region $Z_{1}$ is red, and the elliptical
decision border of region $Z_{2}$ and the corresponding elliptical decision
boundary is dark blue.%
\begin{figure}[ptb]%
\centering
\fbox{\includegraphics[
height=2.5875in,
width=3.2015in
]%
{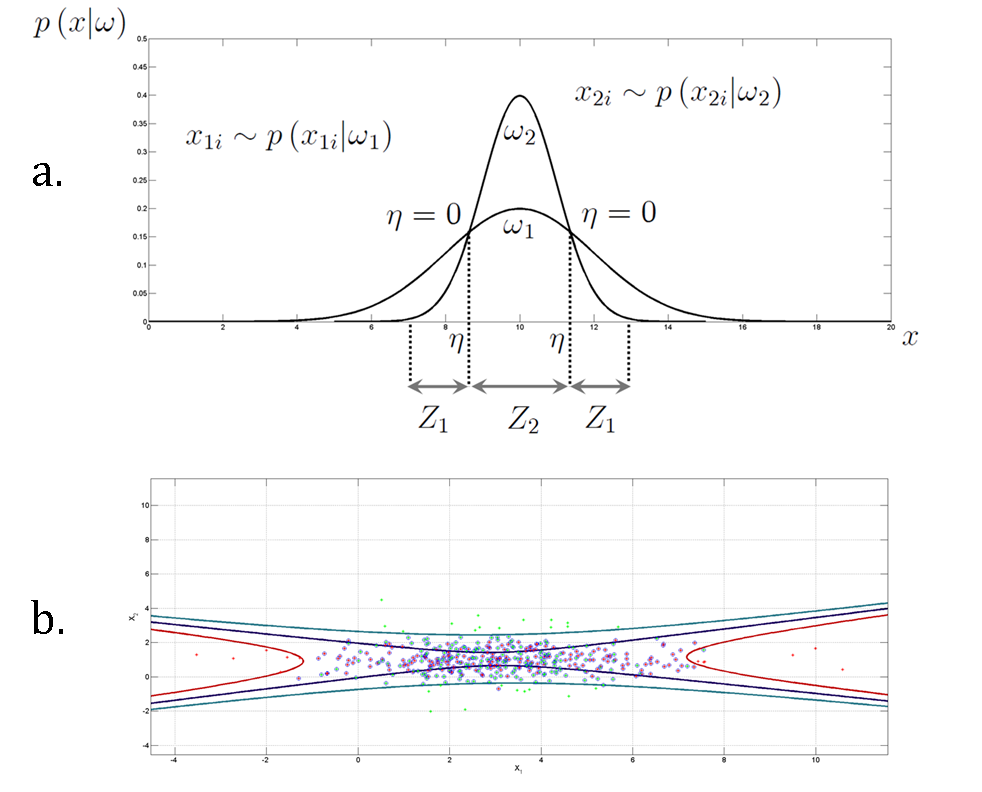}%
}\caption{Illustration of a binary classification system of symmetrically
balanced hyperbolic curves that divides a feature space $Z$ into decision
regions $Z_{1}$ and $Z_{2}$ that have minimum expected risks $\mathfrak{R}%
_{\mathfrak{\min}}\left(  Z_{1}\right)  $ and $\mathfrak{R}_{\mathfrak{\min}%
}\left(  Z_{2}\right)  $.}%
\label{Overlapping Data Same Mean Two}%
\end{figure}

If the decision boundary is a hyperbolic curve or surface, it follows that the
$Z_{1}$ and $Z_{2}$ decision regions are each constrained by two hyperbolic
curves or surfaces that satisfy symmetrical boundary conditions in relation to
the decision boundary. For example, Fig. $\ref{Overlapping Data Same Mean Two}%
$b depicts an overview of a decision space that has been generated in MATLAB,
where the hyperbolic decision borders of region $Z_{1}$ are red, the
hyperbolic decision borders of region $Z_{2}$ are aqua blue, and the
hyperbolic curves of the decision boundary are dark blue.

\paragraph{Example Two}

\begin{figure}[ptb]%
\centering
\fbox{\includegraphics[
height=2.5875in,
width=3.0995in
]%
{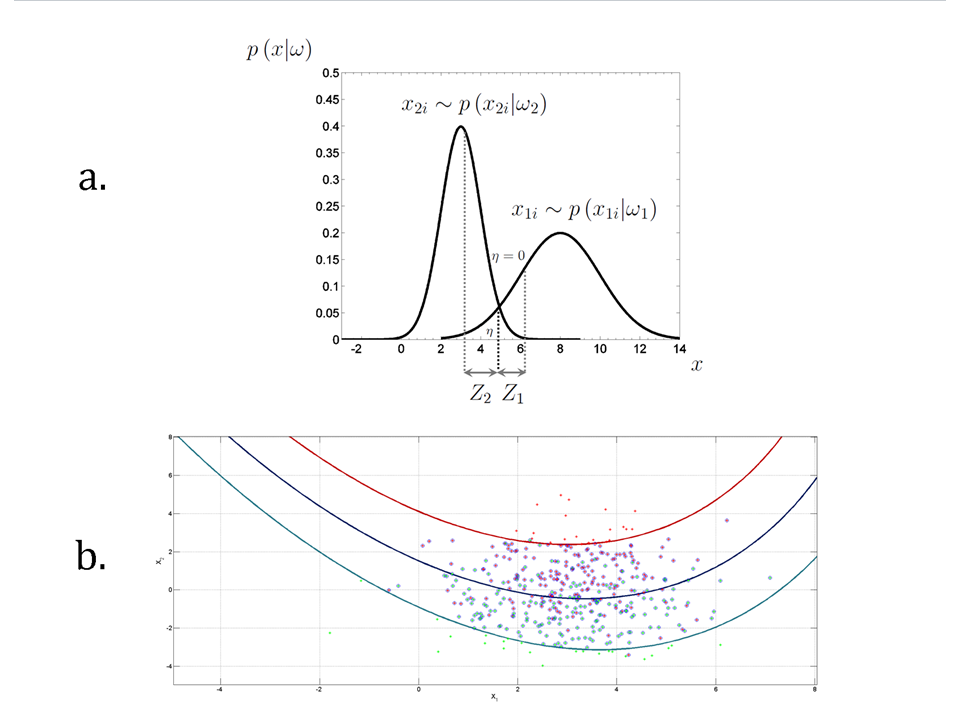}%
}\caption{Optimal likelihood ratio tests divide feature spaces $Z$ into
decision regions $Z_{1}$ and $Z_{2}$ that have minimum expected risks
$\mathfrak{R}_{\mathfrak{\min}}\left(  Z_{1}\right)  $ and $\mathfrak{R}%
_{\mathfrak{\min}}\left(  Z_{2}\right)  $.}%
\label{Equalizer Rule for Overlapping Data Two}%
\end{figure}

Figure $\ref{Equalizer Rule for Overlapping Data Two}$a illustrates how the
likelihood ratio test $\widehat{\Lambda}\left(  \mathbf{x}\right)
\overset{\omega_{1}}{\underset{\omega_{2}}{\gtrless}}0$ in Eq.
(\ref{General Gaussian Equalizer Rule}) determines decision regions $Z_{1}$
and $Z_{2}$ for overlapping data distributions that have dissimilar covariance
matrices and different means, where the risk $\mathfrak{R}_{\mathfrak{\min}%
}\left(  Z_{1}|\omega_{2}\right)  $ for class $\omega_{2}$ and the risk
$\mathfrak{R}_{\mathfrak{\min}}\left(  Z_{2}|\omega_{1}\right)  $ for class
$\omega_{1}$ are determined by a statistical balancing feat:%
\[
\mathfrak{R}_{\mathfrak{\min}}\left(  Z|\widehat{\Lambda}\left(
\mathbf{x}\right)  \right)  :\overline{\mathfrak{R}}_{\mathfrak{\min}}\left(
Z_{1}|\omega_{1}\right)  -\mathfrak{R}_{\mathfrak{\min}}\left(  Z_{1}%
|\omega_{2}\right)  =\overline{\mathfrak{R}}_{\mathfrak{\min}}\left(
Z_{2}|\omega_{2}\right)  -\mathfrak{R}_{\mathfrak{\min}}\left(  Z_{2}%
|\omega_{1}\right)  \text{,}%
\]
for which the risks in the decision regions are similar:%
\[
\mathfrak{R}_{\mathfrak{\min}}\left(  Z_{1}|\omega_{2}\right)  \sim
\mathfrak{R}_{\mathfrak{\min}}\left(  Z_{2}|\omega_{1}\right)  \text{.}%
\]

Accordingly, the expected risk $\mathfrak{R}_{\mathfrak{\min}}\left(
Z|\widehat{\Lambda}\left(  \mathbf{x}\right)  \right)  $ of the classification
system is minimized when the counter risks are similar in the $Z_{1}$ and
$Z_{2}$ decision regions:%
\[
\overline{\mathfrak{R}}_{\mathfrak{\min}}\left(  Z_{1}|\omega_{1}\right)
\sim\overline{\mathfrak{R}}_{\mathfrak{\min}}\left(  Z_{2}|\omega_{2}\right)
\text{.}%
\]%

For example, Fig. $\ref{Equalizer Rule for Overlapping Data Two}$b depicts an
overview of a decision space that has been generated in MATLAB, where the
parabolic decision border of region $Z_{1}$ is red, the parabolic decision
border of region $Z_{2}$ is aqua blue, and the parabolic curve of the decision
boundary is dark blue.

If the counter risks are different in the decision regions are different, then
the likelihood ratio test $\widehat{\Lambda}\left(  \mathbf{x}\right)
\overset{\omega_{1}}{\underset{\omega_{2}}{\gtrless}}0$ in Eq.
(\ref{General Gaussian Equalizer Rule}) determines decision regions $Z_{1}$
and $Z_{2}$ that satisfy border constraints in relation to parabolic,
elliptical, or hyperbolic decision boundaries such that the risks
$\mathfrak{R}_{\mathfrak{\min}}\left(  Z_{1}|\omega_{2}\right)  $ and
$\mathfrak{R}_{\mathfrak{\min}}\left(  Z_{2}|\omega_{1}\right)  $ and the
counter risks $\overline{\mathfrak{R}}_{\mathfrak{\min}}\left(  Z_{1}%
|\omega_{1}\right)  $ and $\overline{\mathfrak{R}}_{\mathfrak{\min}}\left(
Z_{2}|\omega_{2}\right)  $ are effectively balanced with each other.

\subsubsection{Decision Regions for Similar Covariance Matrices}

\begin{figure}[ptb]%
\centering
\fbox{\includegraphics[
height=2.5875in,
width=3.1955in
]%
{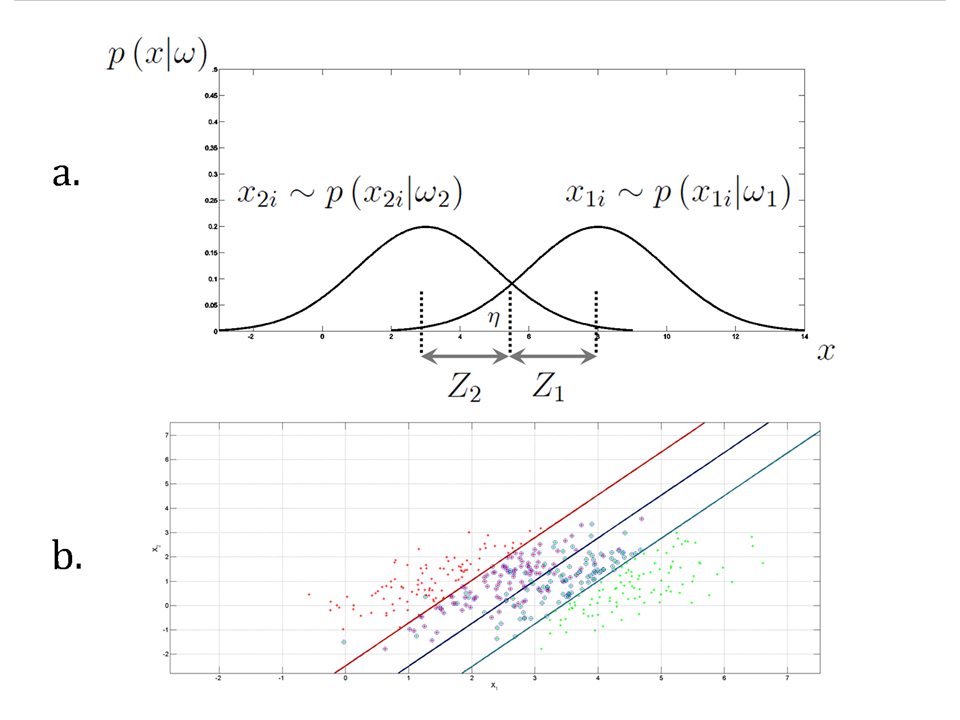}%
}\caption{For overlapping distributions that have similar covariance matrices,
optimal likelihood ratio tests $\protect\widehat{\Lambda}\left(
\mathbf{x}\right)  \protect\overset{\omega_{1}}{\protect\underset{\omega
_{2}}{\gtrless}}$ $0$ divide feature spaces $Z$ into decision regions $Z_{1}$
and $Z_{2}$ that have equal and minimum expected risks: $\mathfrak{R}%
_{\mathfrak{\min}}\left(  Z_{1}\right)  =\mathfrak{R}_{\mathfrak{\min}}\left(
Z_{2}\right)  $.}%
\label{Equalizer Rule for Overlapping Data Similar Covariance}%
\end{figure}

For data distributions that have similar covariance matrices, the risks in the
decision regions are equal. Figure
$\ref{Equalizer Rule for Overlapping Data Similar Covariance}$a illustrates
how the decision rule $\widehat{\Lambda}\left(  \mathbf{x}\right)
\overset{\omega_{1}}{\underset{\omega_{2}}{\gtrless}}$ $0$ in in Eq.
(\ref{General Gaussian Equalizer Rule}) determines decision regions for
overlapping data drawn from Gaussian distributions that have similar
covariance matrices, where the risks $\mathfrak{R}_{\mathfrak{\min}}\left(
Z_{1}|\omega_{2}\right)  $ and $\mathfrak{R}_{\mathfrak{\min}}\left(
Z_{2}|\omega_{1}\right)  $ in the $Z_{1}$ and $Z_{2}$ decision regions are
equal for similar covariance matrices. Because the risks in the decision
regions are equal:%
\[
\mathfrak{R}_{\mathfrak{\min}}\left(  Z_{1}|\omega_{2}\right)  =\mathfrak{R}%
_{\mathfrak{\min}}\left(  Z_{2}|\omega_{1}\right)  \text{,}%
\]
it follows that the counter risks in the $Z_{1}$ and $Z_{2}$ decision regions
are also equal:%
\[
\overline{\mathfrak{R}}_{\mathfrak{\min}}\left(  Z_{1}|\omega_{1}\right)
=\overline{\mathfrak{R}}_{\mathfrak{\min}}\left(  Z_{2}|\omega_{2}\right)
\text{.}%
\]%

Figure $\ref{Equalizer Rule for Overlapping Data Similar Covariance}$b depicts
an overview of a decision space that has been generated in MATLAB, where the
linear decision border of region $Z_{1}$ is red, the linear decision border of
region $Z_{2}$ is aqua blue, and the linear curve of the decision boundary is
dark blue. The linear decision borders exhibit bilateral symmetry with respect
to the linear decision boundary.

In order to develop data-driven locus equations of optimal, statistical
classification systems, I\ need to devise fundamental locus equations for
linear and quadratic loci. By way of motivation, an overview of locus methods
will be followed by a summary of geometric methods in Hilbert spaces.

\section{Locus Methods}

The graph of an equation is the locus (the \emph{place}) of all points whose
coordinates are solutions of the equation. Any given point on a locus
possesses a geometric property which is common to all points of the locus and
no other points. For example, a\ circle is a locus of points $\left(
x,y\right)  $, all of which are at the same distance, the radius $r$, from a
fixed point $\left(  x_{0},y_{0}\right)  $, the center. The algebraic equation
of the locus of a circle in Cartesian coordinates is%
\begin{equation}
\left(  x-x_{0}\right)  ^{2}+\left(  y-y_{0}\right)  ^{2}=r^{2}\text{.}
\label{Coordinate Equation of Circle}%
\end{equation}
Only those coordinates $\left(  x,y\right)  $ that satisfy Eq.
(\ref{Coordinate Equation of Circle}) contribute to the geometric locus of a
specified circle
\citep{Eisenhart1939}%
. Classic examples of geometric loci include circles, ellipses, hyperbolas,
parabolas, lines, and points. Locus problems for all of the second-order
curves have been widely studied in analytic (coordinate) geometry
\citep{Nichols1893,Tanner1898,Eisenhart1939}%
.

Solving a locus problem requires finding the equation of a curve defined by a
given property and drawing the graph or locus of a given equation. Methods for
solving locus problems are based on two fundamental problems in analytic
geometry, both of which involve the graph or locus of an equation.

The identification of the geometric property of a locus of points is a central
problem in coordinate geometry. The inverse problem finds the algebraic form
of an equation whose solutions give the coordinates of all of the points on a
locus which has been defined geometrically. A geometric figure is any set of
points which exhibit a uniform property. Accordingly, any point, line, line
segment, angle, polygon, curve, region, plane, surface, solid, etc. is a
geometric figure. Geometric figures are defined in two ways: $(1)$ as a figure
with certain known properties and $(2)$ as the path of a point which moves
under known conditions
\citep{Nichols1893,Tanner1898,Eisenhart1939}%
.

Finding the algebraic form of an equation for a given geometric figure or
locus is usually a difficult problem. Solving locus problems involves
identifying algebraic and geometric constraints for a given locus of points.
The algebraic form of an equation of a locus \emph{hinges on} both the
\emph{geometric property} and the frame of reference (the \emph{coordinate
system}) of the locus. Moreover, changing the positions of the coordinate axes
of any given locus changes the algebraic form of the locus that references the
axes and the coordinates of any point on the locus. The equation of a locus
and the identification of the geometric property of the locus can be greatly
simplified by changing the positions of the axes to which the locus of points
is referenced
\citep{Nichols1893,Tanner1898,Eisenhart1939}%
.

\subsection{Locus of a Straight Line}

The equations of a straight line in the coordinate plane have been widely
studied in analytical and coordinate geometry. Standard forms of the equation
of a linear locus are outlined below.

\subsubsection{Standard Equations of the First Degree}

The geometric locus of every equation of the first degree is a straight line.
The general equation of the first degree in two coordinate variables $x $ and
$y$ has the form%
\[
Ax+By+C=0\text{,}%
\]
where $A$, $B$, $C$ are constants which may have any real values, subject to
the restriction that $A$ and $B$ cannot both be zero. Only two geometric
conditions are deemed necessary to determine the equation of a particular
line. Either a line should pass through two given points, or should pass
through a given point and have a given slope. Standard equations of a straight
line include the point-slope, slope-intercept, two-point, intercept, and
normal forms
\citep{Nichols1893,Tanner1898,Eisenhart1939}%
.

Excluding the point, a straight line appears to be the simplest type of
geometric locus. Yet, the locus of \emph{a point} is ill-defined because a
locus has \emph{more} than \emph{one point}. Moreover, the uniform geometric
property of a straight line remains unidentified.

I will identify several, correlated, uniform properties exhibited by all of
the points on a linear locus shortly. I will now define the locus of a point
in terms of the locus of a position vector.

\subsection{Locus of a Position Vector}

The locus of a position vector will play a significant role in analyses that
follow. A position vector $\mathbf{x}=%
\begin{pmatrix}
x_{1}, & x_{2}, & \cdots, & x_{d}%
\end{pmatrix}
^{T}$ is defined to be the locus of a directed, straight line segment formed
by two points $P_{\mathbf{0}}%
\begin{pmatrix}
0, & 0, & \cdots, & 0
\end{pmatrix}
$ and $P_{\mathbf{x}}%
\begin{pmatrix}
x_{1}, & x_{2}, & \cdots, & x_{d}%
\end{pmatrix}
$ which are at a distance of%
\[
\left\Vert \mathbf{x}\right\Vert =\left(  x_{1}^{2}+x_{2}^{2}+\cdots+x_{d}%
^{2}\right)  ^{1/2}%
\]
from each other, where $\left\Vert \mathbf{x}\right\Vert $ denotes the length
of a position vector $\mathbf{x}$, such that each point coordinate $x_{i}$ or
vector component $x_{i}$ is at a signed distance of $\left\Vert \mathbf{x}%
\right\Vert \cos\mathbb{\alpha}_{ij}$ from the origin $P_{\mathbf{0}}$, along
the direction of an orthonormal coordinate axis $\mathbf{e}_{j}$, where
$\cos\mathbb{\alpha}_{ij}$ is the direction cosine between the vector
component $x_{i}$ and the orthonormal coordinate axis $\mathbf{e}_{j}$.

It follows that the locus of a position vector $\mathbf{x}$ is specified by an
ordered set of signed magnitudes%
\begin{equation}
\mathbf{x}\triangleq%
\begin{pmatrix}
\left\Vert \mathbf{x}\right\Vert \cos\mathbb{\alpha}_{x_{1}1}, & \left\Vert
\mathbf{x}\right\Vert \cos\mathbb{\alpha}_{x_{2}2}, & \cdots, & \left\Vert
\mathbf{x}\right\Vert \cos\mathbb{\alpha}_{x_{d}d}%
\end{pmatrix}
^{T} \label{Geometric Locus of Vector}%
\end{equation}
along the axes of the standard set of basis vectors%
\[
\left\{  \mathbf{e}_{1}=\left(  1,0,\ldots,0\right)  ,\ldots,\mathbf{e}%
_{d}=\left(  0,0,\ldots,1\right)  \right\}  \text{,}%
\]
all of which describe a unique, ordered $d$-tuple of geometric locations on
$d$ axes $\mathbf{e}_{j}$, where $\left\Vert \mathbf{x}\right\Vert $ is the
length of the vector $\mathbf{x}$, $\left(  \cos\alpha_{x_{1}1},\cdots
,\cos\alpha_{x_{d}d}\right)  $ are the direction cosines of the components $%
\begin{pmatrix}
x_{1}, & \cdots, & x_{d}%
\end{pmatrix}
$ of the vector $\mathbf{x}$ relative to the standard set of orthonormal
coordinate axes $\left\{  \mathbf{e}_{j}\right\}  _{j=1}^{d}$, and each vector
component $x_{i}$ specifies a point coordinate $x_{i}$ of the endpoint
$P_{\mathbf{x}}$ of the vector $\mathbf{x}$.

Using Eq. (\ref{Geometric Locus of Vector}), a point is the endpoint on the
locus of a position vector, such that a correlated point $P_{\mathbf{x}}$ and
position vector $\mathbf{x}$ both describe an ordered pair of real numbers in
the real Euclidean plane or an ordered $d$-tuple of real numbers in real
Euclidean space, all of which jointly determine a geometric location in $%
\mathbb{R}
^{2}$ or $%
\mathbb{R}
^{d}$. In the analyses that follow, the term vector refers to a position vector.

The definition of the locus of a vector is based on inner product statistics
of vectors in a Hilbert space $\mathfrak{H}$. I will now demonstrate that the
multiplication of any two vectors in Hilbert space determines a rich system of
geometric and topological relationships between the loci of the two vectors.
The inner product statistics defined next determine a Hilbert space
$\mathfrak{H}$.

\subsection{Inner Product Statistics}

The inner product expression $\mathbf{x}^{T}\mathbf{x}$ defined by%
\[
\mathbf{x}^{T}\mathbf{x}=x_{1}x_{1}+x_{2}x_{2}+\cdots+x_{d}x_{d}%
\]
generates the norm $\left\Vert \mathbf{x}\right\Vert $ of the vector
$\mathbf{x}$%
\[
\left\Vert \mathbf{x}\right\Vert =\left(  x_{1}^{2}+x_{2}^{2}+\cdots+x_{d}%
^{2}\right)  ^{1/2}%
\]
which determines the Euclidean distance between the endpoint $P_{\mathbf{x}}$
of $\mathbf{x}$ and the origin $P_{\mathbf{o}}$, where the norm $\left\Vert
\mathbf{x}\right\Vert $ measures the length of the vector $\mathbf{x}$, which
is also the magnitude of $\mathbf{x}$. For any given scalar $\zeta\in%
\mathbb{R}
^{1}$, $\left\Vert \zeta\mathbf{x}\right\Vert =|\zeta|\left\Vert
\mathbf{x}\right\Vert $
\citep{Naylor1971}%
.

The inner product function $\mathbf{x}^{T}\mathbf{y}$ also determines the
angle between two vectors $\mathbf{x}$ and $\mathbf{y}$ in $%
\mathbb{R}
^{d}$. Given any two vectors $\mathbf{x}$ and $\mathbf{y}$, the inner product
expression%
\begin{equation}
\mathbf{x}^{T}\mathbf{y}=\ x_{1}y_{1}+x_{2}y_{2}+\cdots+x_{d}y_{d}
\label{Inner Product Expression1}%
\end{equation}
is equivalent to the vector relationship%
\begin{equation}
\mathbf{x}^{T}\mathbf{y}=\left\Vert \mathbf{x}\right\Vert \left\Vert
\mathbf{y}\right\Vert \cos\theta\text{,} \label{Inner Product Expression2}%
\end{equation}
where $\theta$ is the angle between the vectors $\mathbf{x}$ and $\mathbf{y}%
$\textbf{.}

If $\theta=90^{\circ}$, then $\mathbf{x}^{T}\mathbf{y}=0$, and the vectors
$\mathbf{x}$ and $\mathbf{y}$ are said to be orthogonal to each other.
Accordingly, the inner product function $\mathbf{x}^{T}\mathbf{y}$ allows us
to determine vectors which are orthogonal or perpendicular to each other;
orthogonal vectors are denoted by $\mathbf{x}\perp\mathbf{y}$
\citep{Naylor1971}%
.

\subsubsection{Energy of a Vector}

The functional $\mathbf{x}^{T}\mathbf{x}=\left\Vert \mathbf{x}\right\Vert
^{2}$ determines the energy of the vector $\mathbf{x}$. Using Eq.
(\ref{Geometric Locus of Vector}), a vector $\mathbf{x}$ exhibits an energy
$\left\Vert \mathbf{x}\right\Vert ^{2}$ according to its locus, so that the
scaled vector $\zeta\mathbf{x}$ exhibits the scaled energy $\zeta
^{2}\left\Vert \mathbf{x}\right\Vert ^{2}$ of its scaled locus. Principal
eigenaxes of quadratic curves and surfaces exhibit an eigenenergy according to
the locus of a major axis.

The relationships in Eqs (\ref{Inner Product Expression1}) and
(\ref{Inner Product Expression2}) are derived from second-order distance
statistics. I will now demonstrate that second-order distance statistics
determine a rich system of geometric and topological relationships between the
loci of two vectors.

\subsection{Second-order Distance Statistics}

The relationship $\boldsymbol{\upsilon}^{T}\boldsymbol{\nu}=\left\Vert
\boldsymbol{\upsilon}\right\Vert \left\Vert \boldsymbol{\nu}\right\Vert
\cos\varphi$ between two vectors $\boldsymbol{\upsilon}$ and $\boldsymbol{\nu
}$ can be derived by using the law of cosines
\citep{Lay2006}%
:%
\begin{equation}
\left\Vert \boldsymbol{\upsilon}-\boldsymbol{\nu}\right\Vert ^{2}=\left\Vert
\boldsymbol{\upsilon}\right\Vert ^{2}+\left\Vert \boldsymbol{\nu}\right\Vert
^{2}-2\left\Vert \boldsymbol{\upsilon}\right\Vert \left\Vert \boldsymbol{\nu
}\right\Vert \cos\varphi\label{Inner Product Statistic}%
\end{equation}
which reduces to%
\begin{align*}
\left\Vert \boldsymbol{\upsilon}\right\Vert \left\Vert \boldsymbol{\nu
}\right\Vert \cos\varphi &  =\upsilon_{1}\nu_{1}+\upsilon_{2}\nu_{2}%
+\cdots+\upsilon_{d}\nu_{d}\\
&  =\boldsymbol{\upsilon}^{T}\boldsymbol{\nu}=\boldsymbol{\nu}^{T}%
\boldsymbol{\upsilon}\text{.}%
\end{align*}

The vector relationships in Eq. (\ref{Inner Product Statistic}) indicate that
the inner product statistic $\boldsymbol{\upsilon}^{T}\boldsymbol{\nu}$
determines the length $\left\Vert \boldsymbol{\upsilon}-\boldsymbol{\nu
}\right\Vert $ of the vector from $\boldsymbol{\nu}$ to $\boldsymbol{\upsilon
}$, i.e., the vector $\boldsymbol{\upsilon}-\boldsymbol{\nu}$, which is the
distance between the endpoints of $\boldsymbol{\upsilon}$ and $\boldsymbol{\nu
}$, so that%
\[
\boldsymbol{\upsilon}^{T}\boldsymbol{\nu}=\left\Vert \boldsymbol{\upsilon
}\right\Vert \left\Vert \boldsymbol{\nu}\right\Vert \cos\varphi=\left\Vert
\boldsymbol{\upsilon}-\boldsymbol{\nu}\right\Vert \text{.}%
\]

Because second-order distance statistics are symmetric, the law of cosines%
\[
\left\Vert \boldsymbol{\nu}-\boldsymbol{\upsilon}\right\Vert ^{2}=\left\Vert
\boldsymbol{\nu}\right\Vert ^{2}+\left\Vert \boldsymbol{\upsilon}\right\Vert
^{2}-2\left\Vert \boldsymbol{\nu}\right\Vert \left\Vert \boldsymbol{\upsilon
}\right\Vert \cos\varphi
\]
also determines the length $\left\Vert \boldsymbol{\nu}-\boldsymbol{\upsilon
}\right\Vert $ of the vector from $\boldsymbol{\upsilon}$ to $\boldsymbol{\nu
}$ (the vector $\boldsymbol{\nu}-\boldsymbol{\upsilon}$), which is also the
distance between the endpoints of $\boldsymbol{\upsilon}$ and $\boldsymbol{\nu
}$.

Therefore, the inner product statistic $\boldsymbol{\upsilon}^{T}%
\boldsymbol{\nu}$ between two vectors $\boldsymbol{\upsilon}$ and
$\boldsymbol{\nu}$ in Hilbert space $\mathfrak{H}$%
\begin{align}
\boldsymbol{\upsilon}^{T}\boldsymbol{\nu}  &  =\upsilon_{1}\nu_{1}%
+\upsilon_{2}\nu_{2}+\cdots+\upsilon_{d}\nu_{d}%
\label{Locus Statistics in Hilbert Space}\\
&  =\left\Vert \boldsymbol{\upsilon}\right\Vert \left\Vert \boldsymbol{\nu
}\right\Vert \cos\varphi\nonumber\\
&  =\left\Vert \boldsymbol{\upsilon}-\boldsymbol{\nu}\right\Vert \nonumber
\end{align}
determines the distance between the geometric loci%
\[%
\begin{pmatrix}
\left\Vert \boldsymbol{\upsilon}\right\Vert \cos\mathbb{\alpha}_{\upsilon
_{1}1}, & \left\Vert \boldsymbol{\upsilon}\right\Vert \cos\mathbb{\alpha
}_{\upsilon_{2}2}, & \cdots, & \left\Vert \boldsymbol{\upsilon}\right\Vert
\cos\mathbb{\alpha}_{\upsilon_{d}d}%
\end{pmatrix}
\]
and%
\[%
\begin{pmatrix}
\left\Vert \boldsymbol{\nu}\right\Vert \cos\mathbb{\alpha}_{\nu_{1}1}, &
\left\Vert \boldsymbol{\nu}\right\Vert \cos\mathbb{\alpha}_{\nu_{2}2}, &
\cdots, & \left\Vert \boldsymbol{\nu}\right\Vert \cos\mathbb{\alpha}_{\nu
_{d}d}%
\end{pmatrix}
\]
of the given vectors.

\begin{figure}[ptb]%
\centering
\fbox{\includegraphics[
height=2.5875in,
width=3.4368in
]%
{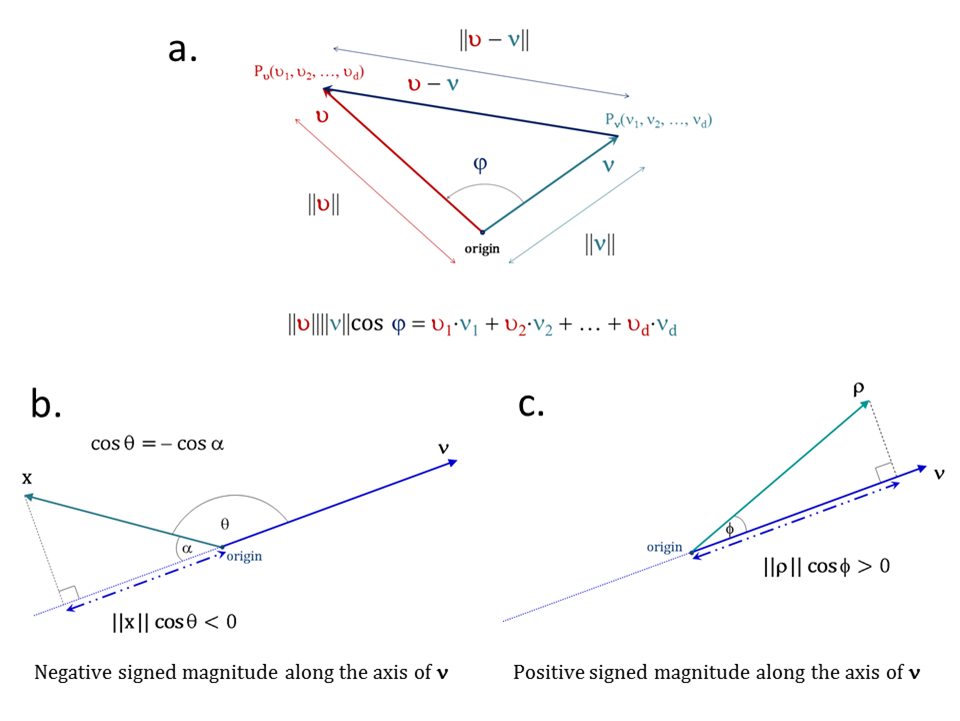}%
}\caption{$\left(  a\right)  $ Inner product statistics specify angles and
corresponding distances between the geometric loci of vectors. Scalar
projection statistics specify $\left(  b\right)  $ negative signed magnitudes
or $\left(  c\right)  $ positive signed magnitudes along the axes of given
vectors.}%
\label{Second-order Distance Statisitcs}%
\end{figure}

Thus, it is concluded that the vector relationships contained within Eq.
(\ref{Inner Product Statistic}) determine a rich system of topological
relationships between the loci of two vectors. Figure
$\ref{Second-order Distance Statisitcs}$ depicts correlated algebraic,
geometric, and topological structures determined by an inner product statistic
(see Fig. $\ref{Second-order Distance Statisitcs}$a).

Inner product statistics include the component of a vector along another
vector, which is also known as a scalar projection. Figure
$\ref{Second-order Distance Statisitcs}$ illustrates the geometric nature of
scalar projections for obtuse angles (see Fig.
$\ref{Second-order Distance Statisitcs}$b) and acute angles (see Fig.
$\ref{Second-order Distance Statisitcs}$c) between vectors.%

\subsubsection{Scalar Projection Statistics}

Scalar projection statistics specify signed magnitudes along the axes of given
vectors. The inner product statistic $\mathbf{x}^{T}\mathbf{y}=\left\Vert
\mathbf{x}\right\Vert \left\Vert \mathbf{y}\right\Vert \cos\theta$ can be
interpreted as the length $\left\Vert \mathbf{x}\right\Vert $ of $\mathbf{x}$
times the scalar projection of $\mathbf{y}$ onto $\mathbf{x}$:%
\begin{equation}
\mathbf{x}^{T}\mathbf{y}=\left\Vert \mathbf{x}\right\Vert \times\left[
\left\Vert \mathbf{y}\right\Vert \cos\theta\right]  \text{,}
\label{Scalar Projection}%
\end{equation}
where the scalar projection of $\mathbf{y}$ onto $\mathbf{x}$, also known as
the component of $\mathbf{y}$ along $\mathbf{x}$, is defined to be the signed
magnitude $\left\Vert \mathbf{y}\right\Vert \cos\theta$ of the vector
projection, where $\theta$ is the angle between $\mathbf{x}$ and $\mathbf{y}$
\citep{Stewart2009}%
. Scalar projections are denoted by $\operatorname{comp}%
_{\overrightarrow{\mathbf{x}}}\left(  \overrightarrow{\mathbf{y}}\right)  $,
where $\operatorname{comp}_{\overrightarrow{\mathbf{x}}}\left(
\overrightarrow{\mathbf{y}}\right)  <0$ if $\pi/2<\theta\leq\pi$. The scalar
projection statistic $\left\Vert \mathbf{y}\right\Vert \cos\theta$ satisfies
the inner product relationship $\left\Vert \mathbf{y}\right\Vert \cos
\theta=\frac{\mathbf{x}^{T}\mathbf{y}}{\left\Vert \mathbf{x}\right\Vert }$
between the unit vector $\frac{\mathbf{x}}{\left\Vert \mathbf{x}\right\Vert }$
and the vector $\mathbf{y}$.

\subsubsection{Locus Methods in Dual Hilbert Spaces}%

\citet{Naylor1971}
note that a truly amazing number of problems in engineering and science can be
fruitfully treated with geometric methods in Hilbert space. In this paper, I
will devise three systems of data-driven, locus equations, subject to
statistical laws that satisfy a binary classification theorem, that involve
geometric and statistical methods in dual Hilbert spaces. In the process, I
will devise data-driven, mathematical laws that generate optimal statistical
classification systems for digital data.

In the next section, I\ will formulate equations and identify geometric
properties of linear loci.

\section{Loci of Lines, Planes, and Hyperplanes}

I will now devise the fundamental locus equation that determines loci of
lines, planes, and hyperplanes. I will use this equation to identify
correlated, uniform properties exhibited by all of the points on a linear
locus. I will also devise a general eigen-coordinate system that determines
all forms of linear loci.

The analysis that follows will denote both points and vectors by $\mathbf{x}$.

\subsection{Vector Equation of a Linear Locus}

Let $\boldsymbol{\nu}\triangleq%
\begin{pmatrix}
\nu_{1}, & \nu_{2}%
\end{pmatrix}
^{T}$ be a \emph{fixed vector} in the real Euclidean plane and consider the
line $l$ at the endpoint of $\boldsymbol{\nu}$ that is perpendicular to
$\boldsymbol{\nu}$. It follows that the endpoint of the vector
$\boldsymbol{\nu}$ is a point on the line $l$. Thereby, the coordinates
$\left(  \nu_{1},\nu_{2}\right)  $ of $\boldsymbol{\nu}$ delineate and satisfy
$l$. In addition, consider an arbitrary vector $\mathbf{x}\triangleq%
\begin{pmatrix}
x_{1}, & x_{2}%
\end{pmatrix}
^{T}$ whose endpoint is also on the line $l$. Thereby, the coordinates
$\left(  x_{1},x_{2}\right)  $ of the point $\mathbf{x}$ also delineate and
satisfy $l$. Finally, let $\phi$ be the acute angle between the vectors
$\boldsymbol{\nu}$ and $\mathbf{x}$, satisfying the constraints $0\leq\phi
\leq\pi/2$ and the relationship $\cos\phi=\frac{\left\Vert \boldsymbol{\nu
}\right\Vert }{\left\Vert \mathbf{x}\right\Vert }$.

\begin{figure}[ptb]%
\centering
\fbox{\includegraphics[
height=2.5875in,
width=3.4411in
]%
{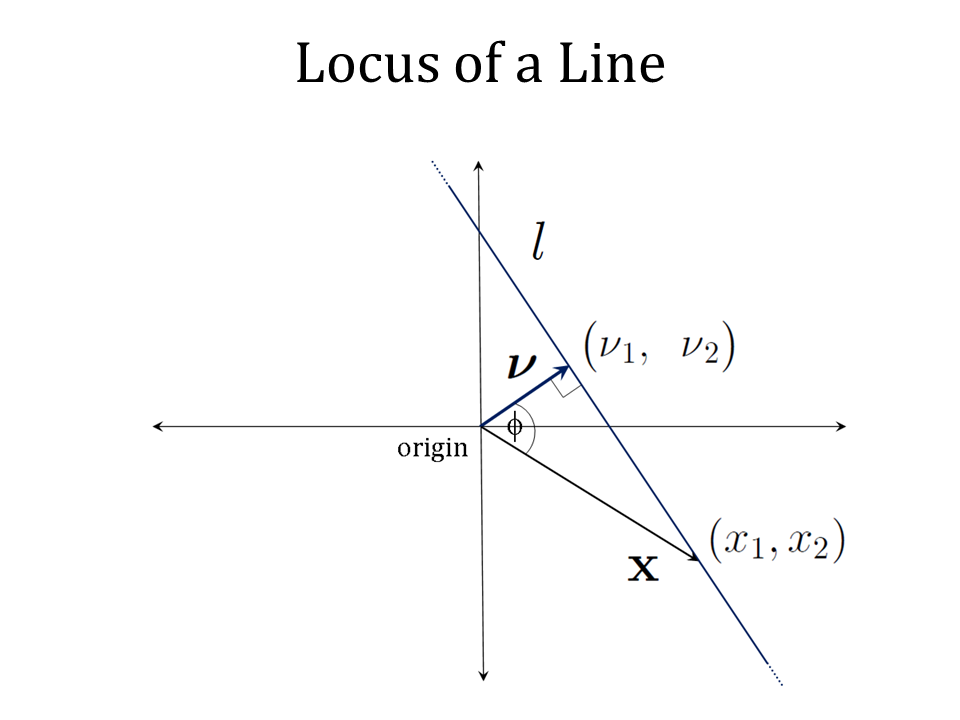}%
}\caption{An elegant, general eigen-coordinate system for lines that is
readily generalized to planes and hyperplanes. Any vector $\mathbf{x}=%
\protect\begin{pmatrix}
x_{1}, & x_{2}%
\protect\end{pmatrix}
^{T}$ whose endpoint $%
\protect\begin{pmatrix}
x_{1}, & x_{2}%
\protect\end{pmatrix}
$ is on the line $l$ explicitly and exclusively references the fixed vector
$\boldsymbol{\nu}=%
\protect\begin{pmatrix}
\nu_{1}, & \nu_{2}%
\protect\end{pmatrix}
^{T}$ which is shown to be the principal eigenaxis of the line $l$.}%
\label{Geometric Locus of Line}%
\end{figure}

Using all of the above assumptions, it follows that the locus of points
$\left(  x_{1},x_{2}\right)  $ on a line $l$ is determined by the equation:%
\begin{equation}
\mathbf{x}^{T}\boldsymbol{\nu}=\left\Vert \mathbf{x}\right\Vert \left\Vert
\boldsymbol{\nu}\right\Vert \cos\phi\label{Linear Locus Functional}%
\end{equation}
which is the vector equation of a line
\citep{Davis1973}%
. By way of illustration, Fig. $\ref{Geometric Locus of Line}$ depicts the
geometric locus of a line in the real Euclidean plane $%
\mathbb{R}
^{2}$.%

\subsection{Fundamental Equation of a Linear Locus}

Take any fixed vector $\boldsymbol{\nu}$, and consider the line $l$ that is
determined by Eq. (\ref{Linear Locus Functional}), where the axis of
$\boldsymbol{\nu}$ is perpendicular to the specified line $l$, and the
endpoint of $\boldsymbol{\nu}$ is on $l$. Given that any vector $\mathbf{x}$
with its endpoint on the given line $l$ satisfies the identity $\left\Vert
\mathbf{x}\right\Vert \cos\phi=\left\Vert \boldsymbol{\nu}\right\Vert $ with
the fixed vector $\boldsymbol{\nu}$, it follows that the locus of a line $l$
is also determined by the vector equation:%
\begin{equation}
\mathbf{x}^{T}\boldsymbol{\nu}=\left\Vert \boldsymbol{\nu}\right\Vert
^{2}\text{.} \label{Normal Eigenaxis Functional}%
\end{equation}
Equations (\ref{Linear Locus Functional}) and
(\ref{Normal Eigenaxis Functional}) are readily generalized to planes $p$ and
hyperplanes $h$ in $%
\mathbb{R}
^{d}$ by letting $\boldsymbol{\nu}\triangleq%
\begin{pmatrix}
\nu_{1}, & \nu_{2}, & \cdots, & \nu_{d}%
\end{pmatrix}
^{T}$ and $\mathbf{x}\triangleq%
\begin{pmatrix}
x_{1}, & x_{2}, & \cdots, & x_{d}%
\end{pmatrix}
^{T}$.

Because Eq. (\ref{Normal Eigenaxis Functional}) contains no constants or
parameters, Eq. (\ref{Normal Eigenaxis Functional}) is the fundamental
equation of a linear locus. So, take any line, plane, or hyperplane in $%
\mathbb{R}
^{d}$. It follows that the linear locus is delineated by a fixed vector
$\boldsymbol{\nu}$, where the axis of $\boldsymbol{\nu}$ is perpendicular to
the linear locus and the endpoint of $\boldsymbol{\nu}$ is on the linear locus.

Assuming that $\left\Vert \boldsymbol{\nu}\right\Vert \neq0$, Eq.
(\ref{Normal Eigenaxis Functional}) can also be written as:%
\begin{equation}
\frac{\mathbf{x}^{T}\boldsymbol{\nu}}{\left\Vert \boldsymbol{\nu}\right\Vert
}=\left\Vert \boldsymbol{\nu}\right\Vert \text{.}
\label{Normal Form Normal Eigenaxis}%
\end{equation}
The axis $\boldsymbol{\nu}/\left\Vert \boldsymbol{\nu}\right\Vert $ has length
$1$ and points in the direction of the vector $\boldsymbol{\nu}$, such that
$\left\Vert \boldsymbol{\nu}\right\Vert $ is the distance of a specified line
$l$, plane $p$, or hyperplane $h$ to the origin. Using Eq.
(\ref{Normal Form Normal Eigenaxis}), it follows that the distance
$\mathbf{\Delta}$ of a line, plane, or hyperplane from the origin is specified
by the magnitude $\left\Vert \boldsymbol{\nu}\right\Vert $ of the axis
$\boldsymbol{\nu}$.

I will now argue that the fixed vector $\boldsymbol{\nu}$ is the
\emph{principal eigenaxis} of linear loci.

\subsection{Principal Eigenaxes of Linear Loci}

Figure $\ref{Geometric Locus of Line}$ shows how the locus of a fixed vector
$\boldsymbol{\nu}$ specifies the locus of a line $l$. I will show that such
fixed vectors $\boldsymbol{\nu}$ provide exclusive, intrinsic reference axes
for loci of lines, planes, and hyperplanes. Intrinsic reference axes, i.e.,
coordinate axes, of geometric loci are inherently specified by locus
equations. Therefore, an intrinsic reference axis is an integral component of
a locus of points. Furthermore, intrinsic reference axes are principal
eigenaxes of conic sections and quadratic surfaces
\citep{Hewson2009}%
.

I will now demonstrate that the axis denoted by $\boldsymbol{\nu}$ in Eqs
(\ref{Linear Locus Functional}), (\ref{Normal Eigenaxis Functional}), and
(\ref{Normal Form Normal Eigenaxis}) is the principal eigenaxis of linear loci.

All of the major axes of conic sections and quadratic surfaces are major
intrinsic axes which may also coincide as exclusive, fixed reference axes.
Major axes have been referred to as "the axis of the curve"
\citep{Nichols1893}%
, "the principal axis," and "the principal axis of the curve"
\citep{Tanner1898}%
. In order to demonstrate that the vector $\boldsymbol{\nu}$ is the principal
eigenaxis of linear loci, it must be shown that $\boldsymbol{\nu}$ is a major
intrinsic axis which is also an exclusive, fixed reference axis. It will first
be argued that $\boldsymbol{\nu}$ is a major intrinsic axis for a linear locus
of points.

Using the definitions of Eqs (\ref{Linear Locus Functional}),
(\ref{Normal Eigenaxis Functional}), or (\ref{Normal Form Normal Eigenaxis}),
it follows that the axis $\boldsymbol{\nu}$ is a major intrinsic axis because
all of the points $\mathbf{x}$ on a linear locus satisfy similar algebraic and
geometric\ constraints related to the locus of $\boldsymbol{\nu}$ that are
inherently specified by Eqs (\ref{Linear Locus Functional}),
(\ref{Normal Eigenaxis Functional}), and (\ref{Normal Form Normal Eigenaxis}).
Therefore, $\boldsymbol{\nu}$ is a major intrinsic axis of a linear locus. The
uniform algebraic and geometric constraints satisfied by all of the points on
a linear locus specify the properties exhibited by each point on the linear locus.

Again, using the definitions of Eqs (\ref{Linear Locus Functional}),
(\ref{Normal Eigenaxis Functional}), or (\ref{Normal Form Normal Eigenaxis}),
it follows that the axis $\boldsymbol{\nu}$ is an exclusive, fixed reference
axis because the uniform properties possessed by all of the points
$\mathbf{x}$ on a linear locus are defined solely by their relation to the
axis of $\boldsymbol{\nu}$. Therefore, all of the points $\mathbf{x}$ on any
given line, plane, or hyperplane explicitly and exclusively reference the
major intrinsic axis $\boldsymbol{\nu}$ of the linear locus. It follows that
the vector $\boldsymbol{\nu}$ provides an exclusive, fixed reference axis for
a linear locus.

Thus, it is concluded that the vector $\boldsymbol{\nu}$ is a major intrinsic
axis that coincides as an exclusive, fixed reference axis for a linear locus.
It follows that the vector $\boldsymbol{\nu}$ is the major axis of linear
loci. Therefore, the vector denoted by $\boldsymbol{\nu}$ in Eqs
(\ref{Linear Locus Functional}), (\ref{Normal Eigenaxis Functional}), and
(\ref{Normal Form Normal Eigenaxis}) is the principal eigenaxis of linear
curves and plane or hyperplane surfaces in $%
\mathbb{R}
^{d}$.

I will now use Eq. (\ref{Normal Form Normal Eigenaxis}) to devise a coordinate
form locus equation which I will use to identify a uniform property exhibited
by any point on a linear curve or surface.

\subsubsection{Coordinate Form Equation of a Linear Locus}

Using Eq. (\ref{Normal Form Normal Eigenaxis}), it follows that any line $l$
in the Euclidean plane $%
\mathbb{R}
^{2}$ and any plane $p$ or hyperplane $h$ in Euclidean space $%
\mathbb{R}
^{d}$ is determined by the vector equation%
\begin{equation}
\mathbf{x}^{T}\mathbf{u}_{P_{e}}\mathbf{=\Delta}\text{\textbf{,}}
\label{normal form linear functional}%
\end{equation}
where $\mathbf{u}_{P_{e}}$ is a unit length principal eigenaxis that is
perpendicular to $l$, $p$, or $h$ and $\mathbf{\Delta}$ denotes the distance
of $l$, $p$, or $h$ to the origin. The unit eigenvector $\mathbf{u}_{P_{e}}$
specifies the direction of the principal eigenaxis $\boldsymbol{\nu}$ of a
linear curve or surface, while the distance $\mathbf{\Delta}$ of a line,
plane, or hyperplane from the origin is specified by the magnitude $\left\Vert
\boldsymbol{\nu}\right\Vert $ of its principal eigenaxis $\boldsymbol{\nu}$.

Express $\mathbf{u}_{P_{e}}$ in terms of standard orthonormal basis vectors%
\[
\left\{  \mathbf{e}_{1}=\left(  1,0,\ldots,0\right)  ,\ldots,\mathbf{e}%
_{d}=\left(  0,0,\ldots,1\right)  \right\}
\]
so that%
\[
\mathbf{u}_{P_{e}}=\cos\alpha_{1}\mathbf{e}_{1}+\cos\alpha_{2}\mathbf{e}%
_{2}+\cdots+\cos\alpha_{d}\mathbf{e}_{d}\text{,}%
\]
where $\cos\alpha_{i}$ are the direction cosines between $\mathbf{u}_{P_{e}}$
and $\mathbf{e}_{i}$. The term $\cos\alpha_{i}$ is the $i^{\text{th}}$
component of the unit principal eigenaxis $\mathbf{u}_{P_{e}}$ along the
coordinate axis $\mathbf{e}_{i}$, where each scale factor $\cos\alpha_{i}$ is
said to be normalized.

Substitution of the expression for $\mathbf{u}_{P_{e}}$ into Eq.
(\ref{normal form linear functional}) produces a coordinate form locus
equation%
\begin{equation}
x_{1}\cos\alpha_{1}+x_{2}\cos\alpha_{2}+\cdots+x_{d}\cos\alpha_{d}%
=\mathbf{\Delta} \label{Unit Normal Coordinate Form Equation}%
\end{equation}
which is satisfied by the transformed coordinates $\cos\alpha_{i}x_{i}$ of all
of the points $\mathbf{x}$ on the locus of a line, plane, or hyperplane.

Equation (\ref{Unit Normal Coordinate Form Equation}) is similar to the
well-known coordinate equation version of a linear locus%
\[
\alpha_{1}x_{1}+\alpha_{2}x_{2}+\ldots+\alpha_{n}x_{n}=p
\]
which is a unique equation for a given linear locus if and only if it contains
the components $\cos\alpha_{i}$ of the unit principal eigenaxis $\mathbf{u}%
_{P_{e}}$ of the linear locus.

I will now use Eq. (\ref{Unit Normal Coordinate Form Equation}) to define a
uniform property which is exhibited by any point on a linear locus.

\subsubsection{Uniform Property of a Linear Locus}

Using Eq. (\ref{Unit Normal Coordinate Form Equation}), it follows that
a\ line, plane, or hyperplane is a locus of points $\mathbf{x}$, all of which
possess a set of transformed coordinates:%
\[
\mathbf{u}_{P_{e}}^{T}\mathbf{x}=\left(  \cos\alpha_{1}x_{1},\cos\alpha
_{2}x_{2},\ldots,\cos\alpha_{d}x_{d}\right)  ^{T}\text{,}%
\]
such that the sum of those transformed coordinates equals the distance
$\mathbf{\Delta}$ that the line, plane, or hyperplane is from the origin
$\left(  0,0,\ldots,0\right)  $:%
\begin{equation}
\sum\nolimits_{i=1}^{d}\cos\alpha_{i}x_{i}=\mathbf{\Delta}\text{,}
\label{Geometric Property of Linear Loci}%
\end{equation}
where $x_{i}$ are point coordinates or vector components, and $\cos\alpha_{i}$
are the direction cosines between a unit principal eigenaxis $\mathbf{u}%
_{P_{e}}$ and the coordinate axes $\mathbf{e}_{i}:\left\{  \mathbf{e}%
_{1}=\left(  1,0,\ldots,0\right)  ,\ldots,\mathbf{e}_{d}=\left(
0,0,\ldots,1\right)  \right\}  $.

Therefore, a point $\mathbf{x}$ is on the locus of a line $l$, plane $p$, or
hyperplane $h$ if and only if the transformed coordinates of $\mathbf{x}$
satisfy Eq. (\ref{Geometric Property of Linear Loci}); otherwise, the point
$\mathbf{x}$ is not on the locus of points determined by Eqs
(\ref{Linear Locus Functional}), (\ref{Normal Eigenaxis Functional}), and
(\ref{normal form linear functional}).

Thereby, it is concluded that all of the points $\mathbf{x}$ on a linear locus
possess a characteristic set of transformed coordinates, such that the inner
product of each vector $\mathbf{x}$ with the unit principal eigenaxis
$\mathbf{u}_{P_{e}}$ satisfies the distance $\mathbf{\Delta}$ of the linear
locus from the origin. Likewise, it is concluded that the sum of transformed
coordinates of any point on a linear locus satisfies the magnitude of the
principal eigenaxis of the linear locus.

Properties of principal eigenaxes are examined next.

\subsection{Properties of Principal Eigenaxes}

Take any line, plane, or hyperplane in $%
\mathbb{R}
^{d}$. Given the line, plane, or hyperplane and Eqs
(\ref{Linear Locus Functional}) or (\ref{Normal Eigenaxis Functional}), it
follows that a principal eigenaxis $\boldsymbol{\nu}$ exists, such that the
endpoint of $\boldsymbol{\nu}$ is on the line, plane, or hyperplane and the
axis of $\boldsymbol{\nu}$ is perpendicular to the line, plane, or hyperplane.
Using Eq. (\ref{Normal Form Normal Eigenaxis}), it follows that the length
$\left\Vert \boldsymbol{\nu}\right\Vert $ of $\boldsymbol{\nu}$\ is specified
by the line, plane, or hyperplane. Using Eq.
(\ref{Unit Normal Coordinate Form Equation}), it follows that the unit
principal eigenaxis $\mathbf{u}_{P_{e}}$ of the linear curve or surface is
characterized by a unique set of direction cosines $\left\{  \cos\alpha
_{i}\right\}  _{i=1}^{d}$ between $\mathbf{u}_{P_{e}}$ and the standard set of
basis vectors $\left\{  \mathbf{e}_{i}\right\}  _{i=1}^{d}$.

Next, take any principal eigenaxis $\boldsymbol{\nu}$ in $%
\mathbb{R}
^{d}$. Given the principal eigenaxis $\boldsymbol{\nu}$ and Eqs
(\ref{Linear Locus Functional}) or (\ref{Normal Eigenaxis Functional}), it
follows that a line, plane, or hyperplane exists that is perpendicular to
$\boldsymbol{\nu}$, such that the endpoint of the principal eigenaxis
$\boldsymbol{\nu}$ is on the line, plane, or hyperplane. Using Eq.
(\ref{Normal Form Normal Eigenaxis}), it follows that the distance of the
line, plane, or hyperplane from the origin is specified by the magnitude
$\left\Vert \boldsymbol{\nu}\right\Vert $ of the principal eigenaxis
$\boldsymbol{\nu}$.

Thus, it is concluded that the principal eigenaxis of a linear locus is
determined by a unique set of direction cosines, and that the principal
eigenaxis of a linear locus determines a unique, linear locus.

I will now show that the principal eigenaxis of any linear locus satisfies the
linear locus in terms of its eigenenergy.

\subsubsection{Characteristic Eigenenergy}

Take the principal eigenaxis $\boldsymbol{\nu}$ of any line, plane, or
hyperplane in $%
\mathbb{R}
^{d}$. Therefore, the principal eigenaxis $\boldsymbol{\nu}$ satisfies Eqs
(\ref{Linear Locus Functional}), (\ref{Normal Eigenaxis Functional}), and
(\ref{Normal Form Normal Eigenaxis}). Using Eqs (\ref{Linear Locus Functional}%
) or (\ref{Normal Eigenaxis Functional}), it follows that the principal
eigenaxis $\boldsymbol{\nu}$ satisfies a linear locus in terms of its squared
length $\left\Vert \boldsymbol{\nu}\right\Vert ^{2}$:%
\begin{equation}
\boldsymbol{\nu}^{T}\boldsymbol{\nu}=\left\Vert \boldsymbol{\nu}\right\Vert
^{2}=\sum\nolimits_{i=1}^{d}\nu_{i\ast}^{2}\text{,}
\label{Characteristic Eigenenergy}%
\end{equation}
where $\nu_{i\ast}$ are the eigen-coordinates of $\boldsymbol{\nu}$ and
$\left\Vert \boldsymbol{\nu}\right\Vert ^{2}$ is the eigenenergy exhibited by
the locus of $\boldsymbol{\nu}$. Thus, the principal eigenaxis
$\boldsymbol{\nu}$ of any given line, plane, or hyperplane exhibits a
characteristic eigenenergy $\left\Vert \boldsymbol{\nu}\right\Vert ^{2}$ that
is unique for the linear locus. It follows that the locus of any given line,
plane, or hyperplane is determined by the eigenenergy $\left\Vert
\boldsymbol{\nu}\right\Vert ^{2}$ exhibited by the locus of its principal
eigenaxis $\boldsymbol{\nu}$.

\subsection{Properties Possessed by Points on Linear Loci}

Take any point $\mathbf{x}$ on any linear locus. Given Eq.
(\ref{Linear Locus Functional}) and the point $\mathbf{x}$ on the linear
locus, it follows that the length of the component $\left\Vert \mathbf{x}%
\right\Vert \cos\phi$ of the vector $\mathbf{x}$ along the principal eigenaxis
$\boldsymbol{\nu}$ of the linear locus satisfies the length $\left\Vert
\boldsymbol{\nu}\right\Vert $ of $\boldsymbol{\nu}$, i.e., $\left\Vert
\mathbf{x}\right\Vert \cos\phi=\left\Vert \boldsymbol{\nu}\right\Vert $, where
the length $\left\Vert \boldsymbol{\nu}\right\Vert $ of $\boldsymbol{\nu}$
specifies the distance $\mathbf{\Delta}$ of the linear locus from the origin.
Accordingly, the signed magnitude of any given vector $\mathbf{x}$ along the
principal eigenaxis $\boldsymbol{\nu}$ of the linear locus satisfies the
length $\left\Vert \boldsymbol{\nu}\right\Vert $ of $\boldsymbol{\nu}$.

Given Eq. (\ref{normal form linear functional}) and the point $\mathbf{x}$ on
the linear locus, it follows that the inner product $\mathbf{x}^{T}%
\mathbf{u}_{P_{e}}$ of the vector $\mathbf{x}$ with the unit principal
eigenaxis $\mathbf{u}_{P_{e}}$ of the linear locus satisfies the distance
$\mathbf{\Delta}$ of the linear locus from the origin, i.e., $\mathbf{x}%
^{T}\mathbf{u}_{P_{e}}\mathbf{=\Delta}$. Likewise, using Eq.
(\ref{Geometric Property of Linear Loci}), it follows that the sum of the
normalized, transformed coordinates of the point $\mathbf{x}$ also satisfies
the distance $\mathbf{\Delta}$ of the linear locus from the origin, i.e.,
$\sum\nolimits_{i=1}^{d}\cos\alpha_{i}x_{i}=\mathbf{\Delta}$.

Finally, using Eq. (\ref{Normal Eigenaxis Functional}), it follows that the
inner product $\mathbf{x}^{T}\boldsymbol{\nu}$ of the vector $\mathbf{x}$ with
the principal eigenaxis $\boldsymbol{\nu}$ of the linear locus%
\[
\mathbf{x}^{T}\boldsymbol{\nu}=\left\Vert \boldsymbol{\nu}\right\Vert ^{2}%
\]
satisfies the eigenenergy $\left\Vert \boldsymbol{\nu}\right\Vert ^{2}$ of the
principal eigenaxis $\boldsymbol{\nu}$ of the linear locus. Therefore, any
given point $\mathbf{x}$ on any given linear locus satisfies the eigenenergy
$\left\Vert \boldsymbol{\nu}\right\Vert ^{2}$ exhibited by the principal
eigenaxis $\boldsymbol{\nu}$ of the linear locus.

\subsection{Inherent Property of a Linear Locus}

It has been shown that the principal eigenaxis of any given linear locus
satisfies the linear locus in terms of its eigenenergy. It has also been shown
that any given point on any given linear locus satisfies the eigenenergy
exhibited by the principal eigenaxis of the linear locus. Thereby, it has been
demonstrated that the locus of any given line, plane, or hyperplane is
determined by the eigenenergy of its principal eigenaxis.

Therefore, the inherent property of a linear locus and its principal eigenaxis
$\boldsymbol{\nu}$ is the eigenenergy $\left\Vert \boldsymbol{\nu}\right\Vert
^{2}$ exhibited by the locus of the principal eigenaxis $\boldsymbol{\nu}$. In
the next section, I will show that a principal eigenaxis $\boldsymbol{\nu}$ of
a linear locus is the \emph{focus} of the linear locus.

\subsection{Overall Conclusions for Linear Loci}

It has been shown that the uniform properties exhibited by all of the points
$\mathbf{x}$ on any given linear locus are specified by the locus of its
principal eigenaxis $\boldsymbol{\nu}$, where each point $\mathbf{x}$ on the
linear locus and the principal eigenaxis $\boldsymbol{\nu}$ of the linear
locus satisfy the linear locus in terms of the eigenenergy $\left\Vert
\boldsymbol{\nu}\right\Vert ^{2}$ exhibited by its principal eigenaxis
$\boldsymbol{\nu}$. Thereby, it is concluded that the vector components of a
principal eigenaxis specify all forms of linear curves and surfaces, and that
all of the points $\mathbf{x}$ on any given line, plane, or hyperplane
explicitly and exclusively reference the principal eigenaxis $\boldsymbol{\nu
}$ in Eqs (\ref{Linear Locus Functional}), (\ref{Normal Eigenaxis Functional}%
), and (\ref{Normal Form Normal Eigenaxis}).

Therefore, the\ important generalizations for a linear locus are specified by
the eigenenergy exhibited by the locus of its principal eigenaxis. Moreover, a
principal eigenaxis is an exclusive and distinctive coordinate axis that
specifies all of the points on a linear locus. Thereby, it is concluded that
the principal eigenaxis of a linear locus provides an elegant, general
eigen-coordinate system for a linear locus of points.

In the next section, I\ will formulate fundamental locus equations and
identify geometric properties of quadratic loci.

\section{Loci of Quadratic Curves and Surfaces}

I will now devise fundamental locus equations that determine loci of conic
sections and quadratic surfaces. I will use these equation to identify uniform
properties exhibited by all of the points on a quadratic locus. I will also
devise general eigen-coordinate systems that determine all forms of quadratic loci.

\subsection{Solving Quadratic Locus Problems}

Methods for solving geometric locus problems hinge on the identification of
algebraic and geometric correlations for a given locus of points. Geometric
correlations between a set of points which lie on a definite curve or surface
correspond to geometric and algebraic constraints that are satisfied by the
coordinates of any point on a given locus or geometric figure.

Geometric figures are defined in two ways: $(1)$ as a figure with certain
known properties and $(2)$ as the path of a point which moves under known
conditions
\citep{Nichols1893,Tanner1898}%
. The path of a point which moves under known conditions is called a
\emph{generatrix}. Common generatrices involve curves or surfaces called
quadratics which include conic sections and quadratic surfaces. Quadratic
surfaces are also called quadrics
\citep{Hilbert1952}%
.

I\ will now use the definition of a generatrix of a quadratic to devise
fundamental locus equations for conic sections and quadratic surfaces.

\subsection{Generatrices of Quadratic Curves and Surfaces}

A generatrix is a point $P_{\mathbf{x}}$ which moves along a given path such
that the path generates a curve or surface. Three of the quadratics are traced
by a point $P_{\mathbf{x}}$ which moves so that its distance from a fixed
point $P_{\mathbf{f}}$ always bears a constant ratio to its distance from a
fixed line, plane, or hyperplane $D$. Quadratics which are generated in this
manner include $d$-dimensional parabolas, hyperbolas, and ellipses. The
geometric nature of the generatrix of these quadratics is considered next.

\begin{figure}[ptb]%
\centering
\fbox{\includegraphics[
height=2.5875in,
width=3.4411in
]%
{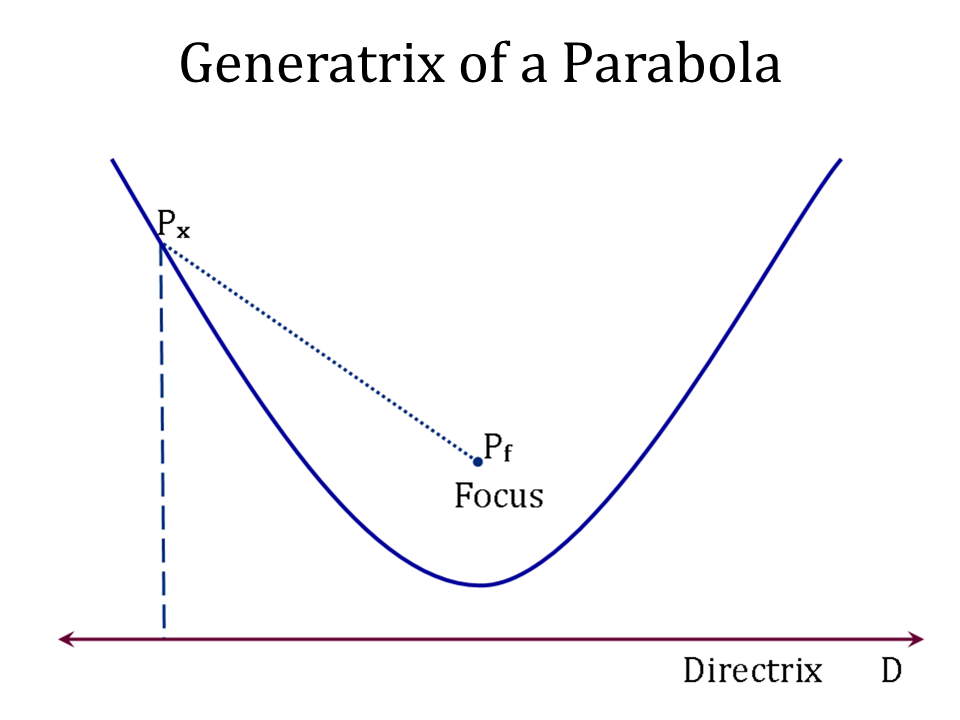}%
}\caption{A parabolic conic section is the set of points $P_{\mathbf{x}}$ such
that the distance from a point $P_{\mathbf{x}}$ to the focus $P_{\mathbf{f}}$
is $1$ times the shortest distance from a point $P_{\mathbf{x}}$ to the
directrix $D$, where distance is measured along a perpendicular.}%
\label{Parabolic Conic Section}%
\end{figure}

Take a fixed point $P_{\mathbf{f}}$ in the Euclidean plane, a line $D$ not
going through $P_{\mathbf{f}}$, and a positive real number $e$. The set of
points $P_{\mathbf{x}}$ such that the distance from $P_{\mathbf{x}}$ to
$P_{\mathbf{f}}$ is $e$ times the shortest distance from $P_{\mathbf{x}}$ to
$D$, where distance is measured along a perpendicular, is a locus of points
termed a conic section. For any given conic section, the point $P_{\mathbf{f}%
}$ is called the focus, the line $D$ is called the directrix, and the term $e$
is called the eccentricity. If $e<1$, the conic is an ellipse; if $e=1$, the
conic is a parabola; if $e>1$, the conic is an hyperbola. The quantities which
determine the size and shape of a conic are its eccentricity $e$ and the
distance of the focus $P_{\mathbf{f}}$ from the directrix $D$
\citep{Zwillinger1996}%
. Figure $\ref{Parabolic Conic Section}$ depicts a parabolic conic section in
the Euclidean plane.%

The definition of a conic section is readily generalized to quadratic surfaces
by taking a fixed point $P_{\mathbf{f}}$ in $%
\mathbb{R}
^{d}$, a $\left(  d-1\right)  $-dimensional hyperplane not going through
$P_{\mathbf{f}}$, and a positive real number $e$.

I will now devise a vector equation of conic generatrices.

\subsection{Vector Equation of Conic Generatrices}

Consider the generatrix of a conic section $c$, where a point $P_{\mathbf{x}}$
moves so that its distance from a fixed point $P_{\mathbf{f}}$ bears a
constant ratio to its distance from a fixed line $D$. Accordingly, take any
given conic $c$.

\subsubsection{Assumptions}

Denote the major axis of the conic $c$ by $\boldsymbol{\nu}$. Let the focus
$P_{\mathbf{f}}$ of the conic $c$ be the endpoint of the major axis
$\boldsymbol{\nu}$ of the conic $c$. Accordingly, the focus $P_{\mathbf{f}}$
and the major axis $\boldsymbol{\nu}$ describe the same geometric location,
where the distance between the focus $P_{\mathbf{f}}$ and the directrix $D$ is
the length $\left\Vert \boldsymbol{\nu}\right\Vert $ of the major axis
$\boldsymbol{\nu}$, i.e., $\boldsymbol{\nu}\perp D$. Any given point
$P_{\mathbf{x}}$ on the conic $c$ is the endpoint of a vector $\mathbf{x}$.
Therefore, any given point $P_{\mathbf{x}}$ and correlated vector $\mathbf{x}
$ describe the same geometric location.

The analysis that follows will denote both points and vectors by $\mathbf{x}$.

Let $\boldsymbol{\nu}\triangleq%
\begin{pmatrix}
\nu_{1}, & \nu_{2}%
\end{pmatrix}
^{T}$ be the major axis of a conic $c$ in the real Euclidean plane. Consider
an arbitrary vector $\mathbf{x}\triangleq%
\begin{pmatrix}
x_{1}, & x_{2}%
\end{pmatrix}
^{T}$ whose endpoint is on the conic $c$ and let $\theta$ be the angle between
$\boldsymbol{\nu}$ and $\mathbf{x}$. Given the above assumptions, it follows
that a conic $c$ is a set of points $\mathbf{x}$ for which the distance
$\left\Vert \mathbf{x-}\boldsymbol{\nu}\right\Vert $ between any given point
$\mathbf{x}$ and the focus $\boldsymbol{\nu}$ is equal to $e$ times the
shortest distance between the given point $\mathbf{x}$ and the directrix $D$.

Now take any given point $\mathbf{x}$ on any given conic $c$. Using Eq.
(\ref{Scalar Projection}), the inner product statistic $\mathbf{x}%
^{T}\boldsymbol{\nu}=\left\Vert \boldsymbol{\nu}\right\Vert \left\Vert
\mathbf{x}\right\Vert \cos\theta$ can be interpreted as the length $\left\Vert
\boldsymbol{\nu}\right\Vert $ of $\boldsymbol{\nu}$ times the scalar
projection of $\mathbf{x}$ onto $\boldsymbol{\nu}$%
\[
\mathbf{x}^{T}\boldsymbol{\nu}=\left\Vert \boldsymbol{\nu}\right\Vert
\times\left[  \left\Vert \mathbf{x}\right\Vert \cos\theta\right]  \text{,}%
\]
where the scalar projection of $\mathbf{x}$ onto $\boldsymbol{\nu}$, also
known as the component of $\mathbf{x}$ along $\boldsymbol{\nu}$, is defined to
be the signed magnitude $\left\Vert \mathbf{x}\right\Vert \cos\theta$ of the
vector projection, where $\theta$ is the angle between $\boldsymbol{\nu}$ and
$\mathbf{x}$. It follows that the shortest distance between the point
$\mathbf{x}$ and the directrix $D$ is specified by the signed magnitude
$\left\Vert \mathbf{x}\right\Vert \cos\theta$ of the component of the vector
$\mathbf{x}$ along the major axis $\boldsymbol{\nu}$.

\begin{figure}[ptb]%
\centering
\fbox{\includegraphics[
height=2.5875in,
width=3.4411in
]%
{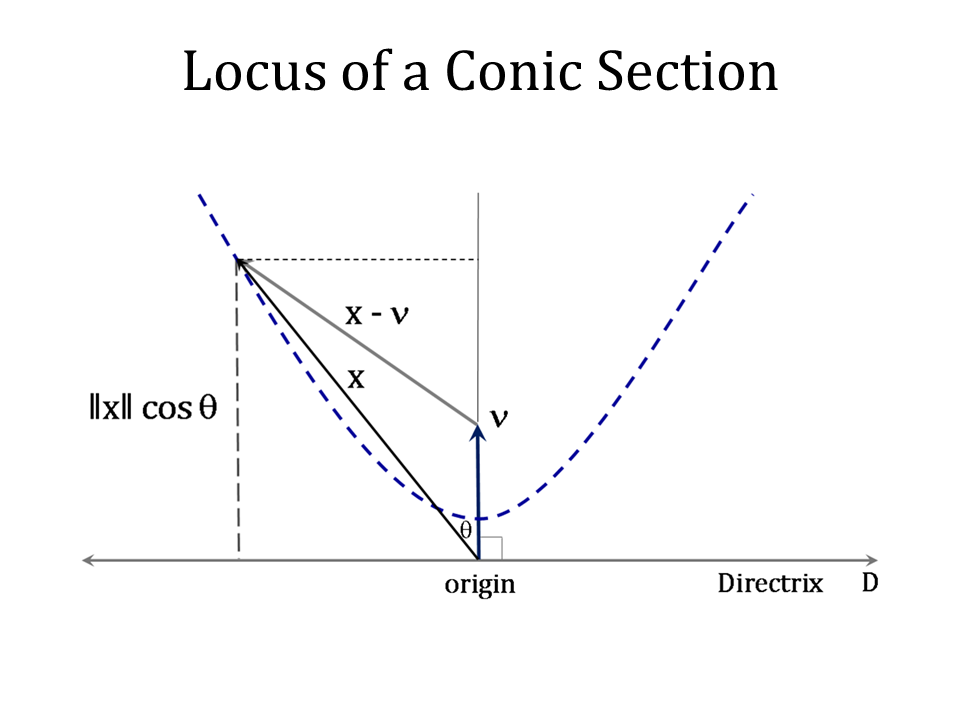}%
}\caption{The geometric locus of a parabola involves a rich system of
geometric and topological relationships between the locus of a major axis
$\boldsymbol{\nu}$ \ and the loci of points $\mathbf{x}$ on a parabola.}%
\label{Geometric Locus of Conic Section}%
\end{figure}

Accordingly, the geometric locus of a conic section $c$ involves a rich system
of geometric and topological relationships between the locus of its major axis
$\boldsymbol{\nu}$ and the loci of points $\mathbf{x}$ on the conic section
$c$. By way of illustration, Fig. $\ref{Geometric Locus of Conic Section}$
depicts the system of geometric and topological relationships between the
locus of a major axis $\boldsymbol{\nu}$ and the loci of points $\mathbf{x}$
on a parabola.%

\subsection{Locus Equation of Conic Generatrices}

Using the law of cosines, it follows that a locus of points $\mathbf{x}$ on an
ellipse, hyperbola, or parabola is determined by the vector equation:%
\begin{equation}
\left\Vert \mathbf{x-}\boldsymbol{\nu}\right\Vert ^{2}=\left\Vert
\mathbf{x}\right\Vert ^{2}+\left\Vert \boldsymbol{\nu}\right\Vert
^{2}-2\left\Vert \mathbf{x}\right\Vert \left\Vert \boldsymbol{\nu}\right\Vert
\cos\theta\text{,} \label{Equation of Generatrices}%
\end{equation}
where $\boldsymbol{\nu}$ is the major eigenaxis of a conic section $c$,
$\mathbf{x}$ is an arbitrary point on $c$, and $\theta$ is the angle between
$\boldsymbol{\nu}$ and $\mathbf{x}$.

Given the geometric and topological relationships depicted in Fig.
$\ref{Geometric Locus of Conic Section}$, it follows that the uniform
geometric property exhibited by a locus of points on a conic section $c$ is
determined by the vector equation%
\begin{equation}
\left\Vert \mathbf{x-}\boldsymbol{\nu}\right\Vert =e\times\left\Vert
\mathbf{x}\right\Vert \cos\theta\text{,} \label{Geometric Property of Conics}%
\end{equation}
where the scaled $e$ signed magnitude $\left\Vert \mathbf{x}\right\Vert
\cos\theta$ of the vector projection of $\mathbf{x}$ onto the major axis
$\boldsymbol{\nu}$ specifies the distance between a point $\mathbf{x}$ on a
given conic $c$ and its directrix $D$.

\subsection{Fundamental Equation of Conic Generatrices}

Substitution of Eq. (\ref{Geometric Property of Conics}) into Eq.
(\ref{Equation of Generatrices}) produces the vector equation of a conic
section $c$%
\[
e^{2}\left\Vert \mathbf{x}\right\Vert ^{2}\cos^{2}\theta=\left\Vert
\mathbf{x}\right\Vert ^{2}+\left\Vert \boldsymbol{\nu}\right\Vert
^{2}-2\left\Vert \mathbf{x}\right\Vert \left\Vert \boldsymbol{\nu}\right\Vert
\cos\theta
\]
which reduces to%
\begin{equation}
2\mathbf{x}^{T}\boldsymbol{\nu}+\left(  e^{2}\cos^{2}\theta-1\right)
\left\Vert \mathbf{x}\right\Vert ^{2}=\left\Vert \boldsymbol{\nu}\right\Vert
^{2}\text{,} \label{Vector Equation of a Conic}%
\end{equation}
where $\boldsymbol{\nu}$ is the major axis of a conic section $c$, $\left\Vert
\boldsymbol{\nu}\right\Vert ^{2}$ is the eigenenergy exhibited by the major
axis $\boldsymbol{\nu}$, $\mathbf{x}$ is a point on a conic $c$, $\left\Vert
\mathbf{x}\right\Vert ^{2}$ is the energy exhibited by the vector $\mathbf{x}%
$, $e$ is the eccentricity of the conic $c$, and $\theta$ is the angle between
the vector $\mathbf{x}$ and the major axis $\boldsymbol{\nu}$ of the conic
$c$. It follows that any point $\mathbf{x}$ on a conic section $c$ is the
endpoint of the locus of a vector $\mathbf{x}$, such that the energy
$\left\Vert \mathbf{x}\right\Vert ^{2}$ exhibited by $\mathbf{x}$ is scaled by
$\left(  e^{2}\cos^{2}\theta-1\right)  $.

Equations (\ref{Equation of Generatrices}),
(\ref{Geometric Property of Conics}), and (\ref{Vector Equation of a Conic})
are readily generalized to $d$-dimensional ellipses, hyperbolas, and parabolas
by letting%
\[
\boldsymbol{\nu}\triangleq%
\begin{pmatrix}
\nu_{1}, & \nu_{2}, & \cdots, & \nu_{d}%
\end{pmatrix}
^{T}%
\]
and%
\[
\mathbf{x}\triangleq%
\begin{pmatrix}
x_{1}, & x_{2}, & \cdots, & x_{d}%
\end{pmatrix}
^{T}\text{.}%
\]

Equation (\ref{Vector Equation of a Conic}) and the system of vector and
topological relationships depicted in Fig.
$\ref{Geometric Locus of Conic Section}$ jointly indicate that all of the
points on a $d$-dimensional ellipse, hyperbola, or parabola exclusively
reference the major axis $\boldsymbol{\nu}$ of the second degree locus.

I will now demonstrate that all of the points $\mathbf{x}$ on a $d$%
-dimensional ellipse, hyperbola, or parabola are essentially characterized by
the geometric locus of its major axis $\boldsymbol{\nu}$.

\subsubsection{Eigen-transformed Coordinates}

Given Eq. (\ref{Vector Equation of a Conic}) and assuming that $\left\Vert
\boldsymbol{\nu}\right\Vert \neq0$, it follows that a locus of points on a $d
$-dimensional ellipse, hyperbola, or parabola is determined by the vector
equation:%
\begin{equation}
\frac{2\mathbf{x}^{T}\boldsymbol{\nu}}{\left\Vert \boldsymbol{\nu}\right\Vert
}+\frac{\left(  e^{2}\cos^{2}\theta-1\right)  \left\Vert \mathbf{x}\right\Vert
^{2}}{\left\Vert \boldsymbol{\nu}\right\Vert }=\left\Vert \boldsymbol{\nu
}\right\Vert \text{,} \label{Normal Form Second-order Locus}%
\end{equation}
where the major axis $\boldsymbol{\nu}/\left\Vert \boldsymbol{\nu}\right\Vert
$ has length $1$ and points in the direction of the principal eigenaxis
$\boldsymbol{\nu}$. Given Eq. (\ref{Normal Form Second-order Locus}), it
follows that all of the points on a $d$-dimensional ellipse, hyperbola, or
parabola explicitly and exclusively reference the major axis $\boldsymbol{\nu
}$ of the second degree locus.

I\ will now use Eq. (\ref{Normal Form Second-order Locus}) to devise a
coordinate form locus equation which I will use to identify a uniform property
exhibited by any point on a $d$-dimensional ellipse, hyperbola, or parabola.

Let $\mathbf{u}_{\boldsymbol{\nu}}$ denote the unit major eigenaxis
$\frac{\boldsymbol{\nu}}{\left\Vert \boldsymbol{\nu}\right\Vert }$ in Eq.
(\ref{Normal Form Second-order Locus}) and express $\mathbf{u}%
_{\boldsymbol{\nu}}$ in terms of orthonormal basis vectors%
\[
\left\{  \mathbf{e}_{1}=\left(  1,0,\ldots,0\right)  ,\ldots,\mathbf{e}%
_{d}=\left(  0,0,\ldots,1\right)  \right\}
\]
so that%
\[
\mathbf{u}_{\boldsymbol{\nu}}=\cos\alpha_{1}\mathbf{e}_{1}+\cos\alpha
_{2}\mathbf{e}_{2}+\cdots+\cos\alpha_{d}\mathbf{e}_{d}\text{,}%
\]
where $\cos\alpha_{i}$ are the direction cosines between $\mathbf{u}%
_{\boldsymbol{\nu}}$ and $\mathbf{e}_{i}$. Each $\cos\alpha_{i}$ is the
$i^{\text{th}}$ component of the unit major eigenaxis $\mathbf{u}%
_{\boldsymbol{\nu}}$ along the coordinate axis $\mathbf{e}_{i}$. Projection of
a vector $\mathbf{x}$ onto $\mathbf{u}_{\boldsymbol{\nu}}$ transforms the
coordinates $\left(  x_{1},\ldots,x_{d}\right)  $ of $\mathbf{x}$ by $\left(
\cos\alpha_{1}x_{1},\ldots,\cos\alpha_{d}x_{d}\right)  $.

Substitution of the expression for $\mathbf{u}_{\boldsymbol{\nu}}$ into Eq.
(\ref{Normal Form Second-order Locus}) produces a coordinate form locus
equation%
\begin{equation}
2%
\begin{pmatrix}
x_{1}, & \cdots, & x_{d}%
\end{pmatrix}
^{T}%
\begin{pmatrix}
\cos\alpha_{1}, & \cdots, & \cos\alpha_{d}%
\end{pmatrix}
+\frac{\left(  e^{2}\cos^{2}\theta-1\right)  }{\sum\nolimits_{i=1}^{d}\nu_{i}%
}\sum\nolimits_{i=1}^{d}x_{i}^{2}=\sum\nolimits_{i=1}^{d}\nu_{i}
\label{Eigen-Coordinate Equation Conic Section}%
\end{equation}
which is satisfied by the transformed coordinates $\left(  \cos\alpha_{1}%
x_{1},\ldots,\cos\alpha_{d}x_{d}\right)  $ of all of the points $\mathbf{x}$
on the geometric locus of a $d$-dimensional ellipse, hyperbola, or parabola.

It follows that all of the points $\mathbf{x}$ on a $d$-dimensional ellipse,
hyperbola, or parabola possess a characteristic set of coordinates, such that
the inner product of each vector $\mathbf{x}$ with the unit major eigenaxis
$\mathbf{u}_{\boldsymbol{\nu}}$ of a given locus satisfies the vector equation%
\[
\sum\nolimits_{i=1}^{d}\cos\alpha_{i}x_{i}=\frac{1}{2}\left\Vert
\boldsymbol{\nu}\right\Vert -\frac{\left(  e^{2}\cos^{2}\theta-1\right)
}{2\left\Vert \boldsymbol{\nu}\right\Vert }\left\Vert \mathbf{x}\right\Vert
^{2}\text{,}%
\]
where the length $\left\Vert \boldsymbol{\nu}\right\Vert $ of $\boldsymbol{\nu
}$ is scaled by $\frac{1}{2}$ and the energy $\left\Vert \mathbf{x}\right\Vert
^{2}$ of $\mathbf{x}$ is scaled by $\frac{\left(  e^{2}\cos^{2}\theta
-1\right)  }{2\left\Vert \boldsymbol{\nu}\right\Vert }$. Therefore, all of the
points $\mathbf{x}$ on any given $d$-dimensional ellipse, hyperbola, or
parabola possess a characteristic set of transformed coordinates, such that
the sum of those coordinates satisfies half the length $\left\Vert
\boldsymbol{\nu}\right\Vert $ of the major eigenaxis $\boldsymbol{\nu}$ minus
the scaled $\frac{\left(  e^{2}\cos^{2}\theta-1\right)  }{2\left\Vert
\boldsymbol{\nu}\right\Vert }$ energy $\left\Vert \mathbf{x}\right\Vert ^{2}$
of $\mathbf{x}$.

Thus, the uniform properties exhibited by all of the points $\mathbf{x}$ on a
$d$-dimensional ellipse, hyperbola, or parabola are specified by the locus of
its principal eigenaxis $\boldsymbol{\nu}$, where the statistic $\sum
\nolimits_{i=1}^{d}\cos\alpha_{i}x_{i}$ varies with the eccentricity $e$ of
the second degree locus.

\subsection{Properties of Major Axes}

All of the major axes of conic sections and quadratic surfaces are major
intrinsic axes which may also coincide as exclusive, fixed reference axes.
Moreover, major axes are principal eigenaxes of quadratic forms
\citep{Hewson2009}%
. Recall that major axes have been referred to as "the axis of the curve"
\citep{Nichols1893}%
, "the principal axis," and "the principal axis of the curve"
\citep{Tanner1898}%
.

I will now show that the principal eigenaxis of any $d$-dimensional ellipse,
hyperbola, or parabola satisfies the quadratic locus in terms of its
eigenenergy. I will also demonstrate that any given point on a quadratic locus
satisfies the eigenenergy $\left\Vert \boldsymbol{\nu}\right\Vert ^{2}$
exhibited by the principal eigenaxis $\boldsymbol{\nu}$ of the quadratic locus.

\subsection{Characteristic Eigenenergy}

Take the principal eigenaxis $\boldsymbol{\nu}$ of any given $d$-dimensional
ellipse, hyperbola, or parabola. Accordingly, the principal eigenaxis
$\boldsymbol{\nu}$ satisfies Eq. (\ref{Vector Equation of a Conic}). Using Eq.
(\ref{Vector Equation of a Conic}), it follows that the principal eigenaxis
$\boldsymbol{\nu}$ satisfies the quadratic locus in terms of its squared
length $\left\Vert \boldsymbol{\nu}\right\Vert ^{2}$:%
\begin{equation}
2\mathbf{x}^{T}\boldsymbol{\nu}+\left(  e^{2}\cos^{2}\theta-1\right)
\left\Vert \mathbf{x}\right\Vert ^{2}=\left\Vert \boldsymbol{\nu}\right\Vert
^{2} \label{Characteristic Eigenenergy of Quadratic}%
\end{equation}
so that the principal eigenaxis $\boldsymbol{\nu}$ and any given vector
$\mathbf{x}$ which satisfy the vector expression%
\[
2\mathbf{x}^{T}\boldsymbol{\nu}+\left(  e^{2}\cos^{2}\theta-1\right)
\left\Vert \mathbf{x}\right\Vert ^{2}%
\]
equally satisfy the vector expression%
\[
\left\Vert \boldsymbol{\nu}\right\Vert ^{2}=\sum\nolimits_{i=1}^{d}\nu_{i\ast
}^{2}\text{,}%
\]
where $\nu_{i\ast}$ are the eigen-coordinates of $\boldsymbol{\nu}$ and
$\left\Vert \boldsymbol{\nu}\right\Vert ^{2}$ is the eigenenergy exhibited by
$\boldsymbol{\nu}$. It follows that the principal eigenaxis $\boldsymbol{\nu}$
of any given $d$-dimensional ellipse, hyperbola, or parabola and any given
point $\mathbf{x}$ on the quadratic locus satisfy the eigenenergy $\left\Vert
\boldsymbol{\nu}\right\Vert ^{2}$ exhibited by the principal eigenaxis
$\boldsymbol{\nu}$ of the quadratic locus.

Therefore, the principal eigenaxis $\boldsymbol{\nu}$ of any given
$d$-dimensional ellipse, hyperbola, or parabola exhibits a characteristic
eigenenergy $\left\Vert \boldsymbol{\nu}\right\Vert ^{2}$ that is unique for
the quadratic locus. It follows that the locus of any given $d$-dimensional
ellipse, hyperbola, or parabola is determined by the eigenenergy $\left\Vert
\boldsymbol{\nu}\right\Vert ^{2}$ of its principal eigenaxis $\boldsymbol{\nu
}$.

\subsection{Inherent Property of $d$-Dimensional Conics}

Let a quadratic locus be a $d$-dimensional ellipse, hyperbola, or parabola. It
has been shown that the principal eigenaxis $\boldsymbol{\nu}$ of any given
quadratic locus satisfies the quadratic locus in terms of its eigenenergy. It
has also been shown that any given point on any given quadratic locus
satisfies the eigenenergy exhibited by the principal eigenaxis of the
quadratic locus. Thereby, it has been demonstrated that the locus of any given
$d$-dimensional ellipse, hyperbola, or parabola is determined by the
eigenenergy of its principal eigenaxis.

Therefore, the inherent property of a quadratic locus and its principal
eigenaxis $\boldsymbol{\nu}$ is the eigenenergy $\left\Vert \boldsymbol{\nu
}\right\Vert ^{2}$ exhibited by the locus of the principal eigenaxis
$\boldsymbol{\nu}$.

\subsection{Overall Conclusions for $d$-Dimensional Conics}

Let a quadratic locus be a $d$-dimensional ellipse, hyperbola, or parabola. It
has been shown that the uniform properties exhibited by all of the points
$\mathbf{x}$ on any given quadratic locus are specified by the locus of its
principal eigenaxis $\boldsymbol{\nu}$, where each point $\mathbf{x}$ on the
quadratic locus and the principal eigenaxis $\boldsymbol{\nu}$ of the
quadratic locus satisfy the quadratic locus in terms of the eigenenergy
$\left\Vert \boldsymbol{\nu}\right\Vert ^{2}$ exhibited by its principal
eigenaxis $\boldsymbol{\nu}$.

Thereby, it is concluded that the vector components of a principal eigenaxis
specify all forms of conic curves and quadratic surfaces, and that all of the
points $\mathbf{x}$ on any given quadratic locus explicitly and exclusively
reference the principal eigenaxis $\boldsymbol{\nu}$ in Eqs
(\ref{Vector Equation of a Conic}) and (\ref{Normal Form Second-order Locus}).

Therefore, the\ important generalizations for a quadratic locus are specified
by the eigenenergy exhibited by the locus of its principal eigenaxis.
Moreover, a principal eigenaxis is an exclusive and distinctive coordinate
axis that specifies all of the points on a quadratic locus. Accordingly, a
principal eigenaxis is the focus of a quadratic locus. Thereby, it is
concluded that the principal eigenaxis of a quadratic locus provides an
elegant, general eigen-coordinate system for a quadratic locus of points.

Circles and lines are considered special cases of conic sections
\citep{Zwillinger1996}%
. In the next section, I will devise fundamental locus equations that
determine loci of circles and $d$-dimensional spheres. I\ will use these
equations to identify uniform geometric properties exhibited by all of the
points on a spherically symmetric, quadratic locus. I will also devise a
general eigen-coordinate system that determines all forms of spherically
symmetric, quadratic loci.

By way of motivation, I will first define the eccentricity of lines, planes,
and hyperplanes.

\subsection{Eccentricity of Lines, Planes, and Hyperplanes}

Let $e$ denote the eccentricity of any given conic section $c$. Returning to
Eq. (\ref{Geometric Property of Conics}), it has been shown that the uniform
geometric property exhibited by a locus of points on a conic section $c$ is
determined by the vector equation%
\[
\left\Vert \mathbf{x-}\boldsymbol{\nu}\right\Vert =e\times\left\Vert
\mathbf{x}\right\Vert \cos\theta\text{,}%
\]
where $e$ is the eccentricity of a conic section $c$, and the signed magnitude
$\left\Vert \mathbf{x}\right\Vert \cos\theta$ of the vector projection of
$\mathbf{x}$ onto the major axis $\boldsymbol{\nu}$ specifies the distance
between a point $\mathbf{x}$ on a given conic $c$ and its directrix $D$.

Given the vector equation of a line $l$ in Eq. (\ref{Linear Locus Functional})%
\[
\mathbf{x}^{T}\boldsymbol{\nu}=\left\Vert \mathbf{x}\right\Vert \left\Vert
\boldsymbol{\nu}\right\Vert \cos\phi\text{,}%
\]
where $\phi$ is the acute angle between the vectors $\boldsymbol{\nu}$ and
$\mathbf{x}$, and using the inner product relationship in Eq.
(\ref{Locus Statistics in Hilbert Space})%
\[
\mathbf{x}^{T}\boldsymbol{\nu}=\left\Vert \mathbf{x-}\boldsymbol{\nu
}\right\Vert \text{,}%
\]
it follows that the eccentricity $e$ of a line $l$ is determined by the vector
equation%
\begin{align*}
\left\Vert \mathbf{x-}\boldsymbol{\nu}\right\Vert  &  =e\times\left\Vert
\mathbf{x}\right\Vert \cos\phi\\
&  =\left\Vert \boldsymbol{\nu}\right\Vert \left\Vert \mathbf{x}\right\Vert
\cos\phi
\end{align*}
so that the eccentricity $e$ of a line $l$ is the length $\left\Vert
\boldsymbol{\nu}\right\Vert $ of the principal eigenaxis $\boldsymbol{\nu}$ of
the line $l$.

Therefore, the eccentricity $e$ of a line, plane, or hyperplane is the length
$\left\Vert \boldsymbol{\nu}\right\Vert $ of the principal eigenaxis
$\boldsymbol{\nu}$ of the linear locus%
\[
e=\left\Vert \boldsymbol{\nu}\right\Vert \text{.}%
\]
It follows that the principal eigenaxis $\boldsymbol{\nu}$ of a linear locus
is the \emph{focus} of the linear locus.

I will now devise fundamental locus equations that determine loci of circles
and $d$-dimensional spheres.

\subsection{Eigen-Centric Equations of Circles and Spheres}

Circles and lines are special cases of parabolas, hyperbolas, and ellipses.
A\ circle is a locus of points $\left(  x,y\right)  $, all of which are at the
same distance, the radius $r$, from a fixed point $\left(  x_{0},y_{0}\right)
$, the center. Using Eq. (\ref{Coordinate Equation of Circle}), the algebraic
equation for the geometric locus of a circle in Cartesian coordinates is:%
\[
\left(  x-x_{0}\right)  ^{2}+\left(  y-y_{0}\right)  ^{2}=r^{2}\text{.}%
\]

I\ will now define the eccentricity of circles and spheres.

\subsection{Eccentricity of Circles and Spheres}

A circle or $d$-dimensional sphere is considered a special case of a
$d$-dimensional ellipse, where the eccentricity $e\approx0$ in the limit
$e\rightarrow0$
\citep{Zwillinger1996}%
.

However, $e$ \emph{cannot be zero}. Indeed, if $e\approx0$, then%
\begin{align*}
\left\Vert \mathbf{x-}\boldsymbol{\nu}\right\Vert  &  =e\times\left\Vert
\mathbf{x}\right\Vert \cos\theta\\
&  \approx0
\end{align*}
which indicates that the radius $r\approx0$ because $\left\Vert \mathbf{x-}%
\boldsymbol{\nu}\right\Vert \approx0$ specifies that $\left\Vert
\mathbf{r}\right\Vert \approx0$.

Instead, the eccentricity $e$ for a circle or a sphere \emph{varies} with
$\left\Vert \mathbf{x}\right\Vert $ \emph{and} $\arccos\theta$:%
\[
e=\frac{\left\Vert \mathbf{r}\right\Vert }{\left\Vert \mathbf{x}\right\Vert
}\arccos\theta\text{,}%
\]
where the length $\left\Vert \mathbf{r}\right\Vert $ of the radius
$\mathbf{r}$ is fixed.

Therefore, consider again the generatrix of a conic section $c$, where a point
$P_{\mathbf{x}}$ moves so that its distance from a fixed point $P_{\mathbf{f}%
}$ bears a constant ratio to its distance from a fixed line $D$. The
generatrix of a circle differs in the following manner.

The generatrix of a circle involves a point $P_{\mathbf{x}}$ which moves so
that its distance from a fixed point $P_{\mathbf{f}}$ is constant and its
distance from a fixed line $D$ varies with the locus of the point
$P_{\mathbf{x}}$.

So, take a fixed point $P_{\mathbf{f}}$ in the Euclidean plane and a line $D$
not going through $P_{\mathbf{f}}$. The set of points $P_{\mathbf{x}}$ such
that the distance from $P_{\mathbf{x}}$ to $P_{\mathbf{f}}$ is \emph{constant}
and the distance from $P_{\mathbf{x}}$ to the fixed line $D$ \emph{varies}
with the \emph{locus} of $P_{\mathbf{x}}$ is a circle. For any given circle,
the point $P_{\mathbf{f}}$ is called the focus, the line $D$ is called the
directrix, and the fixed distance between the focus and a point on the circle
is called the radius.

The definition of a circle is readily generalized to $d$-dimensional spheres
by taking a fixed point $P_{\mathbf{f}}$ in $%
\mathbb{R}
^{d}$ and a $\left(  d-1\right)  $-dimensional hyperplane not going through
$P_{\mathbf{f}}$.

I will now devise the vector equation of circle generatrices.

\subsection{Vector Equation of Circle Generatrices}

Consider the generatrix of a circle, where a point $P_{\mathbf{x}}$ moves so
that its distance from a fixed point $P_{\mathbf{f}}$ is constant and its
distance from a fixed line $D$ varies with the locus of $P_{\mathbf{x}}$.
Accordingly, take any given circle.

\subsubsection{Assumptions}

Denote the major axis of the circle by $\boldsymbol{\nu}$. Let the center
$P_{\mathbf{f}}$ of the circle be the endpoint of the major axis
$\boldsymbol{\nu}$ of the circle. It follows that the center $P_{\mathbf{f}}$
and the major axis $\boldsymbol{\nu}$ of the circle describe the same
geometric location, where the distance between the center $P_{\mathbf{f}}$ and
the directrix $D$ is the length $\left\Vert \boldsymbol{\nu}\right\Vert $ of
the major axis $\boldsymbol{\nu}$, i.e., $\boldsymbol{\nu}\perp D$. Let any
given point $P_{\mathbf{x}}$ on the circle be the endpoint of a vector
$\mathbf{x}$. It follows that any given point $P_{\mathbf{x}}$ and correlated
vector $\mathbf{x}$ describe the same geometric location.

The analysis that follows will denote both points and vectors by $\mathbf{x}$.

Let $\boldsymbol{\nu}\triangleq%
\begin{pmatrix}
\nu_{1}, & \nu_{2}%
\end{pmatrix}
^{T}$ be the major axis of a circle in the real Euclidean plane. Let the
vector $\mathbf{r}\triangleq%
\begin{pmatrix}
r_{1}, & r_{2}%
\end{pmatrix}
^{T}$ be the radius of the circle. In addition, consider an arbitrary vector
$\mathbf{x}\triangleq%
\begin{pmatrix}
x_{1}, & x_{2}%
\end{pmatrix}
^{T}$ whose endpoint is on the circle. Let $\theta$ be the angle between
$\boldsymbol{\nu}$ and $\mathbf{x}$; let $\phi$ be the angle between
$\boldsymbol{\nu}$ and $\mathbf{r}$.

Using all of the above assumptions, it follows that a circle is a set of
points $\mathbf{x}$ for which the distance $\left\Vert \mathbf{x-}%
\boldsymbol{\nu}\right\Vert $ between any given point $\mathbf{x}$ and the
center $\boldsymbol{\nu}$ is equal to the length $\left\Vert \mathbf{r}%
\right\Vert $ of the radius $\mathbf{r}$ of the circle, where the radius
$\mathbf{r}$ of the circle satisfies the expression $\mathbf{r=x-}%
\boldsymbol{\nu}$. Thereby, for any given circle, the focus (center) of the
circle is determined by the locus of its major axis $\boldsymbol{\nu}$, and
the diameter of the circle is determined by its radius $\mathbf{r}$.

\begin{figure}[ptb]%
\centering
\fbox{\includegraphics[
height=2.5875in,
width=3.4411in
]%
{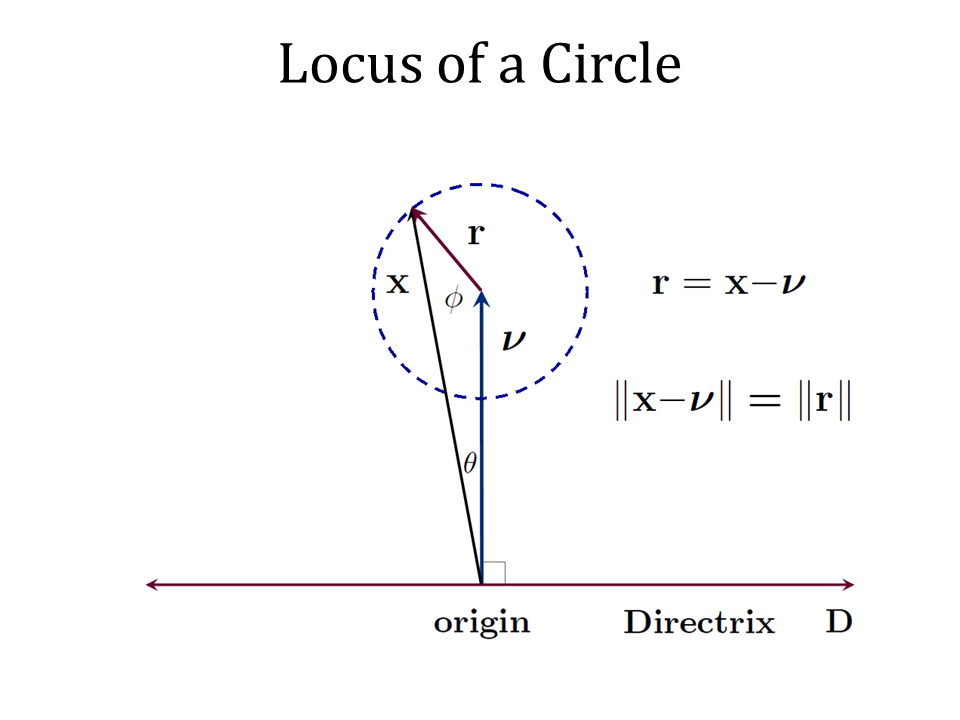}%
}\caption{The geometric locus of a circle involves a rich system of vector and
topological relationships between the locus of a major axis $\boldsymbol{\nu}%
$, the loci of a radius $\mathbf{r}$, and the loci of points $\mathbf{x}$ on a
circle. The distance between the focus $\boldsymbol{\nu}$ \ of a circle and
any point $\mathbf{x}$ on the circle is determined by the length $\left\Vert
\mathbf{r}\right\Vert $ of the radius $\mathbf{r}$.}%
\label{Geometric Locus of Circle}%
\end{figure}

Accordingly, the geometric locus of any given circle involves a rich system of
vector and topological relationships between the locus of its major axis
$\boldsymbol{\nu}$, the loci of its radius $\mathbf{r}$, and the loci of
points $\mathbf{x}$ on the circle. Figure $\ref{Geometric Locus of Circle}$
depicts the system of vector and topological relationships between the locus
of a major axis $\boldsymbol{\nu}$, the loci of a radius $\mathbf{r}$, and the
loci of points $\mathbf{x}$ on a circle.%

\subsection{Vector Equation of Circle Generatrices}

Using the law of cosines, it follows that the geometric locus of points
$\mathbf{x}$ on a circle is determined by the vector equation:%
\[
\left\Vert \mathbf{r}\right\Vert ^{2}=\left\Vert \mathbf{x}\right\Vert
^{2}+\left\Vert \boldsymbol{\nu}\right\Vert ^{2}-2\left\Vert \mathbf{x}%
\right\Vert \left\Vert \boldsymbol{\nu}\right\Vert \cos\theta\text{,}%
\]
where $\theta$ is the angle between $\mathbf{x}$ and $\boldsymbol{\nu}$, the
radius $\mathbf{r}$ of the circle has a fixed length $\left\Vert
\mathbf{r}\right\Vert $ and a constant energy $\left\Vert \mathbf{r}%
\right\Vert ^{2}$, and the major axis $\boldsymbol{\nu}$ of the circle has a
fixed length $\left\Vert \boldsymbol{\nu}\right\Vert $ and a constant energy
$\left\Vert \boldsymbol{\nu}\right\Vert ^{2}$. The above equation is readily
generalized to $d$-dimensional spheres by letting $\boldsymbol{\nu}\triangleq%
\begin{pmatrix}
\nu_{1}, & \nu_{2}, & \cdots, & \nu_{d}%
\end{pmatrix}
^{T}$, $\mathbf{r}\triangleq%
\begin{pmatrix}
r_{1}, & r_{2}, & \cdots, & r_{d}%
\end{pmatrix}
^{T}$, and $\mathbf{x}\triangleq%
\begin{pmatrix}
x_{1}, & x_{2}, & \cdots, & x_{d}%
\end{pmatrix}
^{T}$.

\subsection{Fundamental Equation of Circle Generatrices}

Using the geometric properties of the circle depicted in Fig.
$\ref{Geometric Locus of Circle}$, it follows that any vector $\mathbf{x}$
whose endpoint is on a circle or a $d$-dimensional sphere satisfies the
equation $\mathbf{x}=\boldsymbol{\nu}+\mathbf{r}$. Substituting this equation
for $\mathbf{x}$ into the above equation produces a vector equation that
determines the geometric loci of circles or $d$-dimensional spheres%
\begin{align*}
\left\Vert \mathbf{r}\right\Vert ^{2}  &  =\left\Vert \boldsymbol{\nu
}+\mathbf{r}\right\Vert ^{2}+\left\Vert \boldsymbol{\nu}\right\Vert
^{2}-2\left\Vert \mathbf{x}\right\Vert \left\Vert \boldsymbol{\nu}\right\Vert
\cos\theta\text{,}\\
&  =\left\Vert \boldsymbol{\nu}\right\Vert ^{2}+2\left\Vert \mathbf{r}%
\right\Vert \left\Vert \boldsymbol{\nu}\right\Vert \cos\phi+\left\Vert
\mathbf{r}\right\Vert ^{2}-2\left\Vert \mathbf{x}\right\Vert \left\Vert
\boldsymbol{\nu}\right\Vert \cos\theta
\end{align*}
which reduces to:%
\[
2\left(  \left\Vert \mathbf{x}\right\Vert \left\Vert \boldsymbol{\nu
}\right\Vert \cos\theta-\left\Vert \mathbf{r}\right\Vert \left\Vert
\boldsymbol{\nu}\right\Vert \cos\phi\right)  =\left\Vert \boldsymbol{\nu
}\right\Vert ^{2}\text{.}%
\]

Thus, it is concluded that the geometric locus of a circle denoted by $c$ or a
$d$-dimensional sphere denoted by $S$ is determined by the vector equation%
\begin{equation}
2\left(  \left\Vert \mathbf{x}\right\Vert \left\Vert \boldsymbol{\nu
}\right\Vert \cos\theta-\left\Vert \mathbf{r}\right\Vert \left\Vert
\boldsymbol{\nu}\right\Vert \cos\phi\right)  =\left\Vert \boldsymbol{\nu
}\right\Vert ^{2}\text{,} \label{Vector Equation of Circles and Spheres}%
\end{equation}
where $\boldsymbol{\nu}$ is the major axis and $\mathbf{r}$ is the radius of
$c$ or $S$, $\mathbf{x}$ is a point on $c$ or $S$, $\theta$ is the angle
between $\mathbf{x}$ and $\boldsymbol{\nu}$, $\phi$ is the angle between
$\mathbf{r}$ and $\boldsymbol{\nu}$, and $\left\Vert \boldsymbol{\nu
}\right\Vert ^{2}$ is the eigenenergy exhibited by the locus of the major axis
$\boldsymbol{\nu}$.

Figure $\ref{Geometric Locus of Circle}$ and Eq.
(\ref{Vector Equation of Circles and Spheres}) jointly indicate that all of
the points on a circle or a $d$-dimensional sphere explicitly reference the
major axis $\boldsymbol{\nu}$.

I will now demonstrate that all of the points $\mathbf{x}$ on a circle or a
$d$-dimensional sphere are essentially characterized by the geometric locus of
its major axis $\boldsymbol{\nu}$.

\subsubsection{Eigen-transformed Coordinates}

Given Eq. (\ref{Vector Equation of Circles and Spheres}) and assuming that
$\left\Vert \boldsymbol{\nu}\right\Vert \neq0$, a locus of points on a circle
or a $d$-dimensional sphere is determined by the vector equation:%
\begin{equation}
\frac{2\left(  \mathbf{x-r}\right)  ^{T}\boldsymbol{\nu}}{\left\Vert
\boldsymbol{\nu}\right\Vert }=\left\Vert \boldsymbol{\nu}\right\Vert \text{,}
\label{Normal Form Circle and Sphere}%
\end{equation}
where the major eigenaxis $\boldsymbol{\nu}/\left\Vert \boldsymbol{\nu
}\right\Vert $ has length $1$ and points in the direction of the principal
eigenvector $\boldsymbol{\nu}$.

I\ will now use Eq. (\ref{Normal Form Circle and Sphere}) to devise a
coordinate form locus equation which I will use to identify a uniform property
exhibited by any point on a circle or a $d$-dimensional sphere.

Let $\mathbf{u}_{\boldsymbol{\nu}}$ denote the unit major eigenaxis
$\frac{\boldsymbol{\nu}}{\left\Vert \boldsymbol{\nu}\right\Vert }$ in Eq.
(\ref{Normal Form Circle and Sphere}) and express $\mathbf{u}_{\boldsymbol{\nu
}}$ in terms of orthonormal basis vectors%
\[
\left\{  \mathbf{e}_{1}=\left(  1,0,\ldots,0\right)  ,\ldots,\mathbf{e}%
_{d}=\left(  0,0,\ldots,1\right)  \right\}
\]
so that%
\[
\mathbf{u}_{\boldsymbol{\nu}}=\cos\alpha_{1}\mathbf{e}_{1}+\cos\alpha
_{2}\mathbf{e}_{2}+\cdots+\cos\alpha_{d}\mathbf{e}_{d}\text{,}%
\]
where $\cos\alpha_{i}$ are the direction cosines between $\mathbf{u}%
_{\boldsymbol{\nu}}$ and $\mathbf{e}_{i}$. Each $\cos\alpha_{i}$ is the
$i^{\text{th}}$ component of the unit major eigenaxis $\mathbf{u}%
_{\boldsymbol{\nu}}$ along the coordinate axis $\mathbf{e}_{i}$. Projection of
a vector $\mathbf{x}$ onto $\mathbf{u}_{\boldsymbol{\nu}}$ transforms the
coordinates $\left(  x_{1},\ldots,x_{d}\right)  $ of $\mathbf{x}$ by $\left(
\cos\alpha_{1}x_{1},\ldots,\cos\alpha_{d}x_{d}\right)  $.

Substitution of the expression for $\mathbf{u}_{\boldsymbol{\nu}}$ into Eq.
(\ref{Normal Form Circle and Sphere}) produces a coordinate form locus
equation%
\begin{equation}%
\begin{pmatrix}
x_{1}-r_{1}, & \cdots, & x_{d}-r_{d}%
\end{pmatrix}
^{T}%
\begin{pmatrix}
\cos\alpha_{1}, & \cdots, & \cos\alpha_{d}%
\end{pmatrix}
=\frac{1}{2}\sum\nolimits_{i=1}^{d}\nu_{i}\text{,}
\label{Eigen-Coordinate Equation Circle}%
\end{equation}
where $\frac{1}{2}\sum\nolimits_{i=1}^{d}\nu_{i}=\frac{1}{2}\left\Vert
\boldsymbol{\nu}\right\Vert $, which is satisfied by the transformed
coordinates $\left(  \cos\alpha_{1}\left(  x_{1}-r_{1}\right)  ,\ldots
,\cos\alpha_{d}\left(  x_{d}-r_{d}\right)  \right)  $ of all of the points
$\mathbf{x}$ on the geometric locus of a circle or a $d$-dimensional sphere.

It follows that all of the points $\mathbf{x}$ on any given circle or
$d$-dimensional sphere possess a characteristic set of transformed
coordinates, such that the sum of those coordinates satisfies half the length
$\left\Vert \boldsymbol{\nu}\right\Vert $ of the principal eigenaxis
$\boldsymbol{\nu}$ of a spherically symmetric, quadratic locus%
\begin{equation}
\sum\nolimits_{i=1}^{d}\cos\alpha_{i}\left(  x_{i}-r_{i}\right)  =\frac{1}%
{2}\left\Vert \boldsymbol{\nu}\right\Vert \text{.}
\label{Geometric Property Exhibited by Points on Circles}%
\end{equation}

Therefore, it is concluded that the uniform properties exhibited by the
transformed points $\mathbf{x-r}$ on any given circle or $d$-dimensional
sphere are specified by the locus of its principal eigenaxis $\boldsymbol{\nu
}$.

Thus, all of the points on any circle or $d$-dimensional sphere explicitly
reference the major axis $\boldsymbol{\nu}$ of the spherically symmetric,
quadratic locus. Thereby, the principal eigenaxis denoted by $\boldsymbol{\nu
}$ in Eqs (\ref{Vector Equation of Circles and Spheres}) and
(\ref{Normal Form Circle and Sphere}) is an exclusive and distinctive
coordinate axis that inherently characterizes all of the points on a circle or
a $d$-dimensional sphere.

I will now show that the principal eigenaxis of any given circle or
$d$-dimensional sphere satisfies the locus in terms of its eigenenergy. I will
also demonstrate that any given point on a spherically symmetric, quadratic
locus satisfies the eigenenergy $\left\Vert \boldsymbol{\nu}\right\Vert ^{2}$
exhibited by the principal eigenaxis $\boldsymbol{\nu}$ of the quadratic locus.

\subsection{Characteristic Eigenenergy}

Take the principal eigenaxis $\boldsymbol{\nu}$ of any given circle or
$d$-dimensional sphere. Accordingly, the principal eigenaxis $\boldsymbol{\nu
}$ satisfies Eq. (\ref{Vector Equation of Circles and Spheres}). Using Eq.
(\ref{Vector Equation of Circles and Spheres})%
\[
2\left(  \left\Vert \mathbf{x}\right\Vert \left\Vert \boldsymbol{\nu
}\right\Vert \cos\theta-\left\Vert \mathbf{r}\right\Vert \left\Vert
\boldsymbol{\nu}\right\Vert \cos\phi\right)  =\left\Vert \boldsymbol{\nu
}\right\Vert ^{2}%
\]
and the vector relationship $\mathbf{r}=\mathbf{x-}\boldsymbol{\nu}$, it
follows that the principal eigenaxis $\boldsymbol{\nu}$ satisfies the circle
or $d$-dimensional sphere in terms of its squared length $\left\Vert
\boldsymbol{\nu}\right\Vert ^{2}$:%
\begin{equation}
2\left(  \left\Vert \mathbf{x}\right\Vert \left\Vert \boldsymbol{\nu
}\right\Vert \cos\theta-\left\Vert \mathbf{x-}\boldsymbol{\nu}\right\Vert
\left\Vert \boldsymbol{\nu}\right\Vert \cos\phi\right)  =\left\Vert
\boldsymbol{\nu}\right\Vert ^{2}
\label{Characteristic Eigenenergy of Quadratic 2}%
\end{equation}
so that the principal eigenaxis $\boldsymbol{\nu}$ and any given vector
$\mathbf{x}$ which satisfy the vector expression%
\[
2\left(  \left\Vert \mathbf{x}\right\Vert \left\Vert \boldsymbol{\nu
}\right\Vert \cos\theta-\left\Vert \mathbf{x-}\boldsymbol{\nu}\right\Vert
\left\Vert \boldsymbol{\nu}\right\Vert \cos\phi\right)
\]
equally satisfy the vector expression%
\[
\left\Vert \boldsymbol{\nu}\right\Vert ^{2}=\sum\nolimits_{i=1}^{d}\nu_{i\ast
}^{2}\text{,}%
\]
where $\nu_{i\ast}$ are the eigen-coordinates of $\boldsymbol{\nu}$ and
$\left\Vert \boldsymbol{\nu}\right\Vert ^{2}$ is the eigenenergy exhibited by
$\boldsymbol{\nu}$: It follows that the principal eigenaxis $\boldsymbol{\nu}
$ of any given circle or $d$-dimensional sphere and any given point
$\mathbf{x}$ on the spherically symmetric, quadratic locus satisfy the
eigenenergy $\left\Vert \boldsymbol{\nu}\right\Vert ^{2}$ exhibited by the
principal eigenaxis $\boldsymbol{\nu}$ of the quadratic locus.

Therefore, the principal eigenaxis $\boldsymbol{\nu}$ of any given circle or
$d$-dimensional sphere exhibits a characteristic eigenenergy $\left\Vert
\boldsymbol{\nu}\right\Vert ^{2}$ that is unique for the spherically
symmetric, quadratic locus. It follows that the locus of any given circle or
$d$-dimensional sphere is determined by the eigenenergy $\left\Vert
\boldsymbol{\nu}\right\Vert ^{2}$ exhibited by the locus of its principal
eigenaxis $\boldsymbol{\nu}$.

\subsection{Inherent Property of Circles and Spheres}

Let a quadratic locus be a circle or a $d$-dimensional sphere. It has been
shown that the principal eigenaxis $\boldsymbol{\nu}$ of any given spherically
symmetric, quadratic locus satisfies the quadratic locus in terms of its
eigenenergy. It has also been shown that any given point on any given
spherically symmetric, quadratic locus satisfies the eigenenergy exhibited by
the principal eigenaxis of the quadratic locus. Thereby, it has been
demonstrated that the locus of any given circle or $d$-dimensional sphere is
determined by the eigenenergy of its principal eigenaxis.

Therefore, the inherent property of a spherically symmetric, quadratic locus
and its principal eigenaxis $\boldsymbol{\nu}$ is the eigenenergy $\left\Vert
\boldsymbol{\nu}\right\Vert ^{2}$ exhibited by the locus of the principal
eigenaxis $\boldsymbol{\nu}$.

\subsection{Overall Conclusions for Circles and Spheres}

Let a quadratic locus be a circle or a $d$-dimensional sphere. It has been
shown that the uniform properties exhibited by all of the points $\mathbf{x}$
on any given spherically symmetric, quadratic locus are specified by the locus
of its principal eigenaxis $\boldsymbol{\nu}$, where each point $\mathbf{x}$
on the quadratic locus and the principal eigenaxis $\boldsymbol{\nu}$ of the
quadratic locus satisfy the quadratic locus in terms of the eigenenergy
$\left\Vert \boldsymbol{\nu}\right\Vert ^{2}$ exhibited by its principal
eigenaxis $\boldsymbol{\nu}$.

Thereby, it is concluded that the vector components of a principal eigenaxis
specify all forms of spherically symmetric, conic curves and quadratic
surfaces, and that all of the points $\mathbf{x}$ on any given quadratic locus
explicitly and exclusively reference the principal eigenaxis $\boldsymbol{\nu
}$ in Eqs (\ref{Vector Equation of Circles and Spheres}) and
(\ref{Normal Form Circle and Sphere}).

Therefore, the\ important generalizations for a spherically symmetric,
quadratic locus are specified by the eigenenergy exhibited by the locus of its
principal eigenaxis. Moreover, a principal eigenaxis is an exclusive and
distinctive coordinate axis that specifies all of the points on a spherically
symmetric, quadratic locus. Accordingly, a principal eigenaxis is the focus of
a spherically symmetric, quadratic locus. Thereby, it is concluded that the
principal eigenaxis of a spherically symmetric, quadratic locus provides an
elegant, general eigen-coordinate system for a spherically symmetric,
quadratic locus of points.

I\ will now summarize my primary findings for quadratic loci.

\subsection{Inherent Property of Quadratic Loci}

It has been shown that the principal eigenaxis of any given quadratic locus
satisfies the quadratic curve or surface in terms of its eigenenergy. It has
also been shown that any given point on any given quadratic curve or surface
satisfies the eigenenergy exhibited by the principal eigenaxis of the
quadratic locus. Thereby, it has been demonstrated that the locus of any given
quadratic curve or surface is determined by the eigenenergy of its principal eigenaxis.

Therefore, the inherent property of a quadratic locus and its principal
eigenaxis $\boldsymbol{\nu}$ is the eigenenergy $\left\Vert \boldsymbol{\nu
}\right\Vert ^{2}$ exhibited by the locus of the principal eigenaxis
$\boldsymbol{\nu}$, where $\boldsymbol{\nu}$ is the focus of the quadratic locus.

\subsection{Overall Conclusions for Quadratic Loci}

It has been shown that the uniform properties exhibited by all of the points
$\mathbf{x}$ on any given quadratic locus are specified by the locus of its
principal eigenaxis $\boldsymbol{\nu}$, where each point $\mathbf{x}$ on the
quadratic locus and the principal eigenaxis $\boldsymbol{\nu}$ of the
quadratic locus satisfy the quadratic locus in terms of the eigenenergy
$\left\Vert \boldsymbol{\nu}\right\Vert ^{2}$ exhibited by its principal
eigenaxis $\boldsymbol{\nu}$. Thereby, it is concluded that the vector
components of a principal eigenaxis specify all forms of quadratic curves and
surfaces, and that all of the points $\mathbf{x}$ on any given quadratic curve
or surface explicitly and exclusively reference the principal eigenaxis
$\boldsymbol{\nu}$ in Eqs (\ref{Vector Equation of a Conic}),
(\ref{Normal Form Second-order Locus}),
(\ref{Vector Equation of Circles and Spheres}), and
(\ref{Normal Form Circle and Sphere}).

Therefore, the\ important generalizations for a quadratic locus are specified
by the eigenenergy exhibited by the locus of its principal eigenaxis.
Moreover, a principal eigenaxis is an exclusive and distinctive coordinate
axis that specifies all of the points on a quadratic locus, where the
principal eigenaxis is the focus of the quadratic locus. Thereby, it is
concluded that the principal eigenaxis of a quadratic locus provides an
elegant, general eigen-coordinate system for a quadratic locus of points.

Conic sections and quadratic surfaces involve \emph{first and second degree}
point coordinates or vector components. A reproducing kernel Hilbert space is
a Hilbert space that specifies reproducing kernels for points or vectors,
where reproducing kernels determine first and second degree coordinates or
components for points or vectors. In the next section, I\ will define how a
reproducing kernel Hilbert space extends algebraic and topological properties
for vectors and second-order distance statistics between the loci of two vectors.

\section{Reproducing Kernel Hilbert Spaces}

A Hilbert space $\mathfrak{H}$ is a reproducing kernel Hilbert space
\emph{(}$\mathfrak{rkH}$\emph{)} if the members of $\mathfrak{H}$ are
functions $f$ on some set $T$, and if there is a function called a reproducing
kernel $K(\mathbf{s,t})$ defined on $T\times T$ such that%
\[
K(\cdot\mathbf{,t})\in\mathcal{H}\text{ for all }\mathbf{t}\in T
\]
and%
\[
\left\langle f,K(\cdot\mathbf{,t})\right\rangle =f\left(  \mathbf{t}\right)
\text{ for all }\mathbf{t}\in T\text{ and }f\in\mathcal{H}\text{.}%
\]
A $\mathfrak{rkH}$ space defined on vectors $\mathbf{q}\in%
\mathbb{R}
^{d}$ has a reproducing kernel $K(\mathbf{s,q})$ such that%
\[
K\left(  \cdot,\mathbf{q}\right)  \in\mathcal{H}\text{ for all }\mathbf{q}\in%
\mathbb{R}
^{d}%
\]
and%
\[
\left\langle f\left(  \cdot\right)  ,k(\cdot\mathbf{,q})\right\rangle
=f\left(  \mathbf{q}\right)  \text{ for any }\mathbf{q}\in%
\mathbb{R}
^{d}\text{ and any }f\in\mathcal{H}\left(  k\right)  \text{.}%
\]

So, given a $\mathfrak{rkH}$ space, take any vector $\mathbf{y}\in%
\mathbb{R}
^{d}$. Then there exists a unique vector $k_{\mathbf{y}}\in\mathcal{H}=%
\mathbb{R}
^{d}$ such that for every $f\in\mathcal{%
\mathbb{R}
}^{d}$, $f(\mathbf{y})=\left\langle f,k_{\mathbf{y}}\right\rangle $. The
function $k_{\mathbf{y}}$ is called the reproducing kernel for the point
$\mathbf{y}$. The $2$-variable function defined by $K(\mathbf{x,y}%
)=k_{\mathbf{y}}(\mathbf{x})$ is called the reproducing kernel for
$\mathfrak{H}$
\citep{Aronszajn1950,Small1994}%
.

It follows that%
\begin{align*}
K(\mathbf{x},\mathbf{y})  &  =k_{\mathbf{y}}(\mathbf{x})\\
&  =\left\langle k_{\mathbf{y}}(\mathbf{x}),k_{\mathbf{x}}(\mathbf{y}%
)\right\rangle \\
&  =\left\langle K\left(  .,\mathbf{y}\right)  ,K\left(  .,\mathbf{x}\right)
\right\rangle \text{.}%
\end{align*}
Accordingly, a reproducing kernel $K\left(  \mathbf{t},\mathbf{s}\right)  $
implements the inner product%
\[
\left\langle K\left(  .,\mathbf{s}\right)  ,K\left(  .,\mathbf{t}\right)
\right\rangle =K\left(  \mathbf{t},\mathbf{s}\right)  =K\left(  \mathbf{s}%
,\mathbf{t}\right)
\]
of the vectors $\mathbf{s}$ and $\mathbf{t}$
\citep{Aronszajn1950,Small1994}%
.

\paragraph{Enriched Coordinate Systems}

Given Eq. (\ref{Geometric Locus of Vector}), it follows that the geometric
locus of any given vector $\mathbf{y}$ contains first degree point
coordinates$\ y_{i}$ or vector components $y_{i}$. A\text{ reproducing kernel
}$K\left(  \mathbf{\cdot},\mathbf{y}\right)  $ for a point $\mathbf{y}$
defines an enriched point $P_{k_{\mathbf{y}}}$ in terms of a vector
$k_{\mathbf{y}}$ that contains\ first $y_{i}$\emph{, }second $y_{i}^{2}$,
third $y_{i}^{3}$, and up to $d$\ degree point coordinates or vector
components. Reproducing kernels extend algebraic and topological properties of
the geometric loci of vectors in the manner outlined next.

\subsection{Sinuous Approximations of Vectors}

Topology is the study of geometric properties and spatial relations which are
unaffected by the continuous change of shape or size of a figure. Thus,
topology deals with geometric properties of figures which are unchanged by
continuous transformations. Henri Poincare noted that geometric properties of
figures would remain true even if figures were copied by a draftsman who
grossly changed proportions of figures and \emph{replaced} all \emph{straight
lines} by lines more or less \emph{sinuous}
\citep{Rapport1963}%
. It follows that geometric properties of vectors in Hilbert spaces
$\mathfrak{H}$ are unchanged by continuous transformations. Therefore,
topological properties exhibited by geometric loci of vectors are unchanged if
directed line segments of vectors are replaced by sinuous curves. Given this
assumption, it follows that directed, straight line segments of vectors are
best approximated by second-order curves.

\subsection{Inner Product Statistics for Reproducing Kernels}

Reproducing kernels approximate vectors, which are directed line segments,
with continuous curves. Therefore, reproducing kernels satisfy algebraic and
topological relationships which are similar to those satisfied by vectors.
Using the inner product expression $\mathbf{x}^{T}\mathbf{y}=\left\Vert
\mathbf{x}\right\Vert \left\Vert \mathbf{y}\right\Vert \cos\theta$ in Eq.
(\ref{Inner Product Expression2}) satisfied by any two vectors $\mathbf{x}$
and $\mathbf{y}$ in Hilbert space, it follows that any two reproducing kernels
$k_{\mathbf{s}}(\mathbf{x})$ and $k_{\mathbf{x}}(\mathbf{s})$ for any two
points $\mathbf{s}$ and $\mathbf{x}$ in a reproducing kernel Hilbert space
satisfy the following relationship%
\begin{equation}
K(\mathbf{x},\mathbf{s})=\left\Vert k_{\mathbf{s}}(\mathbf{x})\right\Vert
\left\Vert k_{\mathbf{x}}(\mathbf{s})\right\Vert \cos\varphi\text{,}
\label{Inner Product Relationship for Reproducing Kernels}%
\end{equation}
where $K(\mathbf{x},\mathbf{s})=$ $k_{\mathbf{s}}(\mathbf{x})$ is the
reproducing kernel for $\mathcal{H}$, $k_{\mathbf{s}}(\mathbf{x})$ is the
reproducing kernel for the point $\mathbf{s}$, $k_{\mathbf{x}}(\mathbf{s})$ is
the reproducing kernel for the point $\mathbf{x}$, and $\varphi$ is the angle
between the reproducing kernels $k_{\mathbf{s}}(\mathbf{x})$ and
$k_{\mathbf{x}}(\mathbf{s})$.

\subsection{Why Reproducing Kernels Matter}

Practically speaking, reproducing kernels $K(\mathbf{x},\mathbf{s})$ replace
directed, straight line segments of vectors $\mathbf{s}$ with curves, such
that vectors and correlated points contain first degree vectors components and
point coordinates, \emph{as well as} \emph{second }$x_{i}^{2}$\emph{, third
}$x_{i}^{3}$\emph{, and up to }$\emph{d}$\emph{\ degree} vector components and
point coordinates, where the highest degree $d$ depends on the reproducing
kernel $K(\mathbf{x},\mathbf{s})$.

\subsubsection{Replacements of Directed Straight Line Segments}

Polynomial reproducing kernels $k_{\mathbf{q}\left(  d\right)  }=\left(
\mathbf{x}^{T}\mathbf{q}+1\right)  ^{d}$ replace directed, straight line
segments of vectors with $d$-order polynomial curves. Given that geometric and
topological properties of Hilbert spaces remain true when straight\emph{\ }%
lines are replaced by lines which are more or less sinuous, it follows that
the directed, straight line segment of any given vector $\mathbf{q}$ is best
approximated by a second-order, polynomial curve.

Therefore, second-order, polynomial reproducing kernels%
\[
k_{\mathbf{q}\left(  2\right)  }=\left(  \mathbf{x}^{T}\mathbf{q}+1\right)
^{2}%
\]
are vectors such that the relationships contained within Eq.
(\ref{Inner Product Statistic Reproducing Kernels}) determine a rich system of
topological relationships between the geometric loci of reproducing kernels,
where topological and geometric relationships between vectors $\left(
\mathbf{x}^{T}\mathbf{q}+1\right)  ^{2}$ and $\left(  \mathbf{x}^{T}%
\mathbf{s}+1\right)  ^{2}$ and corresponding points include topological and
geometric relationships between first and second degree point coordinates or
vector components (see Fig. $\ref{Second-order Distance Statisitcs RKHS}$).

Second-order, polynomial reproducing kernels $k_{\mathbf{q}\left(  2\right)
}=\left(  \mathbf{x}^{T}\mathbf{q}+1\right)  ^{2}$ replace straight lines of
vectors $\mathbf{q}$ with second-order, polynomial curves. It will now be
demonstrated that second-order, polynomial reproducing kernels $k_{\mathbf{q}%
\left(  2\right)  }=\left(  \mathbf{x}^{T}\mathbf{q}+1\right)  ^{2}$ contain
first $q_{i}$ and second $q_{i}^{2}$ degree point coordinates or vector
components of vectors $\mathbf{q}$.

\subsection{Second-order Polynomial Reproducing Kernels}

A second-order, polynomial reproducing kernel $\left(  \mathbf{x}%
^{T}\mathbf{q}+1\right)  ^{2}$ for a point or vector $\mathbf{q}\in%
\mathbb{R}
^{d}$ specifies a transformed point $P_{k_{\mathbf{q}}}$ $\in%
\mathbb{R}
^{d}$ or vector $k_{\mathbf{q}}\in%
\mathbb{R}
^{d}$ that contains first and second degree point coordinates or vector
components. The\ reproducing kernel $\left(  \mathbf{x}^{T}\mathbf{q}%
+1\right)  ^{2}$ replaces the straight line segment of the vector $\mathbf{q}$
with a second-order, polynomial curve%
\begin{align*}
\left(  \mathbf{x}^{T}\mathbf{q}+1\right)  ^{2}  &  =\left(  \mathbf{x}%
^{T}\mathbf{q}+1\right)  \left(  \mathbf{x}^{T}\mathbf{q}+1\right) \\
&  =\left(  \left(  \mathbf{\cdot}\right)  \mathbf{^{T}}\left(  \mathbf{\cdot
}\right)  \right)  \left(  \mathbf{q^{T}q}\right)  +2\left(  \mathbf{\cdot
}\right)  ^{T}\mathbf{q}+1\text{,}%
\end{align*}
where $\left(  \mathbf{\cdot}\right)  $ denotes the argument of the
reproducing kernel $\left(  \left(  \mathbf{\cdot}\right)  ^{T}\mathbf{q}%
+1\right)  ^{2}$ of the point $\mathbf{q}$. Because the space $V$ of all
vectors is closed under addition and scaling, such that for any $\mathbf{u,v}%
\in V$ and number $\lambda$, $\mathbf{u+v}\in V$ and $\lambda\mathbf{v}\in V$,
it follows that the reproducing kernel $\left(  \left(  \mathbf{\cdot}\right)
^{T}\mathbf{q}+1\right)  ^{2}$ specifies first and second degree vector
components and point coordinates of $\mathbf{q}$ in the following manner:%
\begin{align*}
\left(  \mathbf{q}+1\right)  ^{2}  &  =\left(  \mathbf{q}+1\right)  \left(
\mathbf{q}+1\right)  =\mathbf{q}^{2}+2\mathbf{q}+\mathbf{1}\\
&  =%
\begin{pmatrix}
q_{1}^{2}+2q_{1}+1, & q_{2}^{2}+2q_{2}+1, & \cdots, & q_{d}^{2}+2q_{d}+1
\end{pmatrix}
^{T}\text{.}%
\end{align*}

Accordingly, each point coordinate or vector component of the reproducing
kernel $\left(  \left(  \mathbf{\cdot}\right)  ^{T}\mathbf{q}+1\right)  ^{2}$
contains first $q_{i}$ and second degree $q_{i}^{2}$ components.

Let $P_{k_{\mathbf{q}}}$ denote the transformed $\left(  \mathbf{q}+1\right)
^{2}$ point $\mathbf{q}$. Because the square root of the inner product
$\sqrt{\left(  \mathbf{q}^{T}\mathbf{q}+1\right)  ^{2}}$ describes the
distance between the endpoint $P_{k_{\mathbf{q}}}$ of the reproducing kernel
$\left(  \left(  \mathbf{\cdot}\right)  ^{T}\mathbf{q}+1\right)  ^{2}$ and the
origin $P_{\mathbf{o}}$, it follows that the distance between the point
$P_{k_{\mathbf{q}}}$ and the origin $P_{\mathbf{o}}$, which is the length
$\left\Vert k_{\mathbf{q}}\right\Vert $ of the vector $k_{\mathbf{q}}$, is
$\mathbf{q}^{T}\mathbf{q}+1=\left\Vert \mathbf{q}\right\Vert ^{2}+1$.

\subsection{Gaussian Reproducing Kernels}

Gaussian reproducing kernels $k_{\mathbf{s}}=\exp\left(  -\gamma\left\Vert
\left(  \mathbf{\cdot}\right)  -\mathbf{s}\right\Vert ^{2}\right)  $ replace
directed, straight line segments of vectors $\mathbf{s}$ with second-order
curves $k_{\mathbf{s}}(\mathbf{x})$, where the hyperparameter $\gamma$ is a
scale factor for the second-order, distance statistic $\left\Vert \left(
\mathbf{\cdot}\right)  -\mathbf{s}\right\Vert ^{2}$. Given that geometric and
topological properties of Hilbert spaces remain true when straight lines are
replaced by lines which are more or less sinuous, it follows that the
directed, straight line segment of any given vector $\mathbf{s}$ is naturally
approximated by a second-order curve. Therefore, given an effective
hyperparameter $\gamma$, it follows that a Gaussian reproducing kernel
$k_{\mathbf{s}}=\exp\left(  -\gamma\left\Vert \left(  \mathbf{\cdot}\right)
-\mathbf{s}\right\Vert ^{2}\right)  $ naturally approximates any given vector
$\mathbf{s}$.

Gaussian reproducing kernels $k_{\mathbf{s}}=\exp\left(  -\gamma\left\Vert
\mathbf{x}-\mathbf{s}\right\Vert ^{2}\right)  $ of points $\mathbf{s}$ are
also vectors $k_{\mathbf{s}}(\mathbf{x})$, where topological and geometric
relationships between vectors $\exp\left(  -\gamma\left\Vert \left(
\mathbf{\cdot}\right)  -\mathbf{x}\right\Vert ^{2}\right)  $ and $\exp\left(
-\gamma\left\Vert \left(  \mathbf{\cdot}\right)  -\mathbf{s}\right\Vert
^{2}\right)  $ and corresponding reproducing kernels of points $\mathbf{x}$
and $\mathbf{s}$ include topological and geometric relationships between first
and second degree point coordinates or vector components (see Fig.
$\ref{Second-order Distance Statisitcs RKHS}$).

I\ will now develop an effective value for the hyperparameter $\gamma$ of a
Gaussian reproducing kernel $k_{\mathbf{s}}=\exp\left(  -\gamma\left\Vert
\left(  \mathbf{\cdot}\right)  -\mathbf{s}\right\Vert ^{2}\right)  $.

\subsubsection{An Effective Value for the Gaussian Hyperparameter}

Later on, I\ will examine an elegant, statistical balancing feat which
involves inner product statistics of feature vectors. Therefore, Gaussian
reproducing kernels $k_{\mathbf{s}}=\exp\left(  -\gamma\left\Vert \left(
\mathbf{\cdot}\right)  -\mathbf{s}\right\Vert ^{2}\right)  $ need to specify
effective inner product statistics of vectors.

By way of motivation, consider the expression for the inner product statistic
$K\left(  \mathbf{x},\mathbf{s}\right)  $%
\begin{align}
K\left(  \mathbf{x},\mathbf{s}\right)   &  =\exp\left(  -\gamma\left\Vert
\mathbf{x}-\mathbf{s}\right\Vert ^{2}\right)
\label{Inner Product Gaussian Reproducing Kernel}\\
&  =\exp\left(  -\gamma\left\{  \left\Vert \mathbf{x}\right\Vert
^{2}+\left\Vert \mathbf{s}\right\Vert ^{2}-2\mathbf{x}^{T}\mathbf{s}\right\}
\right) \nonumber
\end{align}
of a Gaussian reproducing kernel. I propose that an effective value for the
hyperparameter $\gamma$ of a Gaussian reproducing kernel $k_{\mathbf{s}}%
=\exp\left(  -\gamma\left\Vert \left(  \mathbf{\cdot}\right)  -\mathbf{s}%
\right\Vert ^{2}\right)  $ determines an effective scaling factor $\gamma$ for
the inner product statistics determined by $\exp\left(  -\gamma\left\Vert
\mathbf{x}-\mathbf{s}\right\Vert ^{2}\right)  $.

Equation (\ref{Inner Product Gaussian Reproducing Kernel}) indicates that
inner product statistics determined by Gaussian reproducing kernels%
\begin{align*}
K\left(  \mathbf{x},\mathbf{s}\right)   &  =\exp\left(  -\gamma\left\Vert
\mathbf{x}\right\Vert ^{2}\right)  \times\exp\left(  -\gamma\left\Vert
\mathbf{s}\right\Vert ^{2}\right)  \times\exp\left(  \gamma2\mathbf{x}%
^{T}\mathbf{s}\right) \\
&  =k_{\mathbf{s}}(\mathbf{x})\times k_{\mathbf{x}}(\mathbf{s})
\end{align*}
involve the multiplication of three, exponential transforms:%
\[
\exp\left(  -\gamma\left\Vert \mathbf{x}\right\Vert ^{2}\right)  \text{, }%
\exp\left(  -\gamma\left\Vert \mathbf{s}\right\Vert ^{2}\right)  \text{, and
}\exp\left(  \gamma2\mathbf{x}^{T}\mathbf{s}\right)
\]
of three, corresponding, \emph{scaled}, inner product statistics:%
\[
-\gamma\left\Vert \mathbf{x}\right\Vert ^{2}\text{, }-\gamma\left\Vert
\mathbf{s}\right\Vert ^{2}\text{, and }\gamma2\mathbf{x}^{T}\mathbf{s}\text{.}%
\]

Therefore, we need to choose a value for $\gamma$ which specifies
\emph{effective proportions} for the inner product statistics determined by
inner products of Gaussian reproducing kernels $\exp\left(  -\gamma\left\Vert
\mathbf{x}-\mathbf{s}\right\Vert ^{2}\right)  $.

I will now define a value for $\gamma$ which specifies effective proportions
for the inner product statistics determined by the Gaussian reproducing kernel
$\exp\left(  -\gamma\left\Vert \mathbf{x}-\mathbf{s}\right\Vert ^{2}\right)  $.

First, consider the statistics $\exp\left(  -\gamma\left\Vert \mathbf{x}%
\right\Vert ^{2}\right)  $ and $\exp\left(  -\gamma\left\Vert \mathbf{s}%
\right\Vert ^{2}\right)  $. Take the statistic $\exp\left(  -\gamma\left\Vert
\mathbf{z}\right\Vert ^{2}\right)  $ and let $\gamma$ be greater than or equal
to $0.1$. If $\gamma\geq0.1$, then $\exp\left(  -\gamma\left\Vert
\mathbf{z}\right\Vert ^{2}\right)  \ll1$, which indicates that the inner
product statistics $\left\Vert \mathbf{x}\right\Vert ^{2}$ and $\left\Vert
\mathbf{s}\right\Vert ^{2}$ are substantially diminished for any value of
$\gamma\geq0.1$.

Next, consider the statistic $\exp\left(  \gamma2\mathbf{x}^{T}\mathbf{s}%
\right)  $. If $\gamma\geq0.1$, then $\exp\left(  \gamma2\mathbf{x}%
^{T}\mathbf{s}\right)  \gg1$, which indicates that the inner product statistic
$2\mathbf{x}^{T}\mathbf{s}$ is substantially magnified for any value of
$\gamma\geq0.1$.

Therefore, it is concluded that values of $\gamma\geq0.1$ do not specify
effective proportions for inner product statistics determined by the Gaussian
reproducing kernel $\exp\left(  -\gamma\left\Vert \mathbf{x}-\mathbf{s}%
\right\Vert ^{2}\right)  $.

Now, suppose that we let $\gamma=1/100$. Because%
\[
\exp\left(  -0.01\times\left(  0.001\right)  \right)  =1\text{,}%
\]%
\[
\exp\left(  -0.01\times\left(  1\right)  \right)  =0.99\text{,}%
\]%
\[
\exp\left(  -0.01\times\left(  100\right)  \right)  =0.369\text{,}%
\]
and%
\[
\exp\left(  -0.01\times\left(  100000\right)  \right)  =0\text{,}%
\]
it follows that%
\[
0\leq\exp\left(  -0.01\left\Vert \mathbf{x}\right\Vert ^{2}\right)
\leq1\text{.}%
\]

So, if $\gamma=1/100$, the inner product statistics $\exp\left(
-0.01\left\Vert \mathbf{x}\right\Vert ^{2}\right)  $ and $\exp\left(
-0.01\left\Vert \mathbf{s}\right\Vert ^{2}\right)  $ have reasonable
proportions. Therefore, let $\gamma=1/100$ and consider the statistic
$\exp\left\{  \gamma2\mathbf{x}^{T}\mathbf{s}\right\}  $.

If $\gamma=1/100$, the inner product statistic $\exp\left(  0.02\times
\mathbf{x}^{T}\mathbf{s}\right)  $ is not substantially magnified. For
example:%
\[
\exp\left(  0.02\times\left(  100\right)  \right)  =7.39\text{.}%
\]

Moreover, the magnification of the inner product statistic $\exp\left(
0.02\times\left(  \mathbf{x}^{T}\mathbf{s}\right)  \right)  $ is balanced with
diminished proportions of the inner product statistics $\exp\left(
-0.01\left\Vert \mathbf{x}\right\Vert ^{2}\right)  $ and $\exp\left(
-0.01\left\Vert \mathbf{s}\right\Vert ^{2}\right)  $. Therefore, the value of
$\gamma=1/100$ specifies effective proportions for inner product statistics
determined by the Gaussian reproducing kernel $\exp\left(  -\gamma\left\Vert
\mathbf{x}-\mathbf{s}\right\Vert ^{2}\right)  $.

Thus, it is concluded that an effective choice for the hyperparameter $\gamma$
of a Gaussian reproducing kernel $k_{\mathbf{s}}=\exp\left(  -\gamma\left\Vert
\left(  \mathbf{\cdot}\right)  -\mathbf{s}\right\Vert ^{2}\right)  $ is
$\gamma=1/100$, where inner products between Gaussian reproducing kernels
$\exp\left(  -0.01\left\Vert \left(  \mathbf{\cdot}\right)  -\mathbf{x}%
\right\Vert ^{2}\right)  $ and $\exp\left(  -0.01\left\Vert \left(
\mathbf{\cdot}\right)  -\mathbf{s}\right\Vert ^{2}\right)  $ satisfy the
vector expression:%
\begin{align*}
\left\langle k_{\mathbf{x}},k_{\mathbf{s}}\right\rangle  &  =\exp\left(
-0.01\left\Vert \mathbf{x}-\mathbf{s}\right\Vert ^{2}\right) \\
&  =\exp\left(  -0.01\times\left\{  \left\Vert \mathbf{x}\right\Vert
^{2}+\left\Vert \mathbf{s}\right\Vert ^{2}-2\mathbf{x}^{T}\mathbf{s}\right\}
\right) \\
&  =\exp\left(  -0.01\times\left\{  \left\Vert \mathbf{x}\right\Vert
^{2}+\left\Vert \mathbf{s}\right\Vert ^{2}-2\left\Vert \mathbf{x}\right\Vert
\left\Vert \mathbf{s}\right\Vert \cos\varphi\right\}  \right) \\
&  =\exp\left(  -0.01\times\left\Vert \mathbf{x}\right\Vert ^{2}\right)
\times\exp\left(  -0.01\times\left\Vert \mathbf{s}\right\Vert ^{2}\right)
\times\exp\left(  0.02\times\mathbf{x}^{T}\mathbf{s}\right)  \text{.}%
\end{align*}

I will now define the locus of a reproducing kernel. The notion of the locus
of a reproducing kernel will play a significant role in analyses that follow.

\subsection{Locus of a Reproducing Kernel}

A reproducing kernel $K(\mathbf{x},\mathbf{s})=k_{\mathbf{s}}(\mathbf{x})\in%
\mathbb{R}
^{d}$ for a point $\mathbf{s}$ is defined to be the geometric locus of a curve
formed by two points $P_{\mathbf{0}}$ and $P_{k_{\mathbf{s}}}$ which are at a
distance of $\left\Vert k_{\mathbf{s}}\right\Vert =\left(  K(\mathbf{s}%
,\mathbf{s})\right)  ^{1/2}$ from each other, such that each point coordinate
or vector component $k\left(  s_{i}\right)  $ is at a signed distance of
$\left\Vert k_{\mathbf{s}}\right\Vert \cos\mathbb{\alpha}_{ij}$ from the
origin $P_{\mathbf{0}}$, along the direction of an orthonormal coordinate axis
$\mathbf{e}_{j}$, where $\cos\mathbb{\alpha}_{k\left(  s_{i}\right)  j}$ is
the direction cosine between a vector component $k\left(  s_{i}\right)  $ and
an orthonormal coordinate axis $\mathbf{e}_{j}$. It follows that the geometric
locus of a reproducing kernel $k_{\mathbf{s}}$ is specified by an ordered set
of signed magnitudes%
\[
k_{\mathbf{s}}\triangleq%
\begin{pmatrix}
\left\Vert k_{\mathbf{s}}\right\Vert \cos\mathbb{\alpha}_{k\left(
s_{1}\right)  1}, & \left\Vert k_{\mathbf{s}}\right\Vert \cos\mathbb{\alpha
}_{k\left(  s_{2}\right)  2}, & \cdots, & \left\Vert k_{\mathbf{s}}\right\Vert
\cos\mathbb{\alpha}_{k\left(  s_{d}\right)  d}%
\end{pmatrix}
^{T}%
\]
along the axes of the standard set of basis vectors%
\[
\left\{  \mathbf{e}_{1}=\left(  1,0,\ldots,0\right)  ,\ldots,\mathbf{e}%
_{d}=\left(  0,0,\ldots,1\right)  \right\}  \text{,}%
\]
all of which describe a unique, ordered $d$-tuple of geometric locations on
$d$ axes $\mathbf{e}_{j}$, where $\left\Vert k_{\mathbf{s}}\right\Vert $ is
the length of the vector $k_{\mathbf{s}}$, $\left(  \cos\mathbb{\alpha
}_{k\left(  s_{1}\right)  1}\cdots,\cos\mathbb{\alpha}_{k\left(  s_{d}\right)
d}\right)  $ are the direction cosines of the components $k\left(
s_{i}\right)  $ of the vector $k_{\mathbf{s}}$ relative to the standard set of
orthonormal coordinate axes $\left\{  \mathbf{e}_{j}\right\}  _{j=1}^{d}$, and
each vector component $k\left(  s_{i}\right)  $ specifies a point coordinate
$k\left(  s_{i}\right)  $ of the endpoint $P_{k_{\mathbf{s}}}$ of the vector
$k_{\mathbf{s}}$.

Given the above assumptions and notation, the reproducing kernel $\left(
\mathbf{x}^{T}\mathbf{q}+1\right)  ^{2}$%
\[
\left(  \mathbf{x}^{T}\mathbf{q}+1\right)  ^{2}=%
\begin{pmatrix}
q_{1}^{2}+2q_{1}+1, & q_{2}^{2}+2q_{2}+1, & \cdots, & q_{d}^{2}+2q_{d}+1
\end{pmatrix}
^{T}%
\]
is defined to be the geometric locus of a second-order, polynomial curve
formed by two points%
\[
P_{\mathbf{0}}%
\begin{pmatrix}
0, & 0, & \cdots, & 0
\end{pmatrix}
\]
and%
\[
P_{k_{\mathbf{q}}}%
\begin{pmatrix}
q_{1}^{2}+2q_{1}+1, & q_{2}^{2}+2q_{2}+1, & \cdots, & q_{d}^{2}+2q_{d}+1
\end{pmatrix}
\]
which are at a distance of $\left\Vert \mathbf{q}\right\Vert ^{2}+1$ from each
other, such that each point coordinate $x_{i}^{2}+2x_{i}+1$ or vector
component $x_{i}^{2}+2x_{i}+1$ is at a signed distance of $\left(  \left\Vert
\mathbf{x}\right\Vert ^{2}+1\right)  \cos\mathbb{\alpha}_{k\left(
q_{i}\right)  j}$ from the origin $P_{\mathbf{0}}$, along the direction of an
orthonormal coordinate axis $\mathbf{e}_{j}$, where $\cos\mathbb{\alpha
}_{k\left(  q_{i}\right)  j}$ is the direction cosine between the vector
component $x_{i}^{2}+2x_{i}+1$ and the orthonormal coordinate axis
$\mathbf{e}_{j}$.

It follows that the geometric locus of the reproducing kernel $\left(
\mathbf{x}^{T}\mathbf{q}+1\right)  ^{2}$ is specified by an ordered set of
signed magnitudes%
\[
\left(  \mathbf{x}^{T}\mathbf{q}+1\right)  ^{2}\triangleq%
\begin{pmatrix}
\left(  \left\Vert \mathbf{q}\right\Vert ^{2}+1\right)  \cos\mathbb{\alpha
}_{k\left(  q_{1}\right)  1}, & \cdots, & \left(  \left\Vert \mathbf{q}%
\right\Vert ^{2}+1\right)  \cos\mathbb{\alpha}_{k\left(  q_{d}\right)  d}%
\end{pmatrix}
^{T}%
\]
along the axes of the standard set of basis vectors%
\[
\left\{  \mathbf{e}_{1}=\left(  1,0,\ldots,0\right)  ,\ldots,\mathbf{e}%
_{d}=\left(  0,0,\ldots,1\right)  \right\}  \text{,}%
\]
all of which describe a unique, ordered $d$-tuple of geometric locations on
$d$ axes $\mathbf{e}_{j}$, where $\left\Vert \mathbf{q}\right\Vert ^{2}+1$ is
the length of the reproducing kernel $\left(  \mathbf{x}^{T}\mathbf{q}%
+1\right)  ^{2}$, $\left(  \cos\mathbb{\alpha}_{k\left(  q_{1}\right)
1},\cdots,\cos\mathbb{\alpha}_{k\left(  q_{d}\right)  d}\right)  $ are the
direction cosines of the components $x_{i}^{2}+2x_{i}+1$ of the reproducing
kernel $\left(  \mathbf{x}^{T}\mathbf{q}+1\right)  ^{2}$ relative to the
standard set of orthonormal coordinate axes $\left\{  \mathbf{e}_{j}\right\}
_{j=1}^{d}$, and each vector component $x_{i}^{2}+2x_{i}+1$ specifies a point
coordinate $x_{i}^{2}+2x_{i}+1$ of the endpoint $P_{k_{\mathbf{q}}}$ of
$\left(  \mathbf{x}^{T}\mathbf{q}+1\right)  ^{2}$.

It follows that the endpoint of the geometric locus of a reproducing kernel
and the vector determined by the reproducing kernel both describe an ordered
pair of real numbers in the real Euclidean plane or an ordered $d$-tuple of
real numbers in real Euclidean space, all of which jointly determine a
geometric location in $%
\mathbb{R}
^{2}$ or $%
\mathbb{R}
^{d}$.

Therefore, it is concluded that the reproducing kernel $\left(  \mathbf{x}%
^{T}\mathbf{s}+1\right)  ^{2}$ is a vector $k_{\mathbf{s}}\in%
\mathbb{R}
^{d}$ that has a magnitude and a direction such that the endpoint of the
reproducing kernel $\left(  \mathbf{x}^{T}\mathbf{s}+1\right)  ^{2}$%
\[
P_{k_{\mathbf{s}}}%
\begin{pmatrix}
s_{1}^{2}+2s_{1}+1, & s_{2}^{2}+2s_{2}+1, & \cdots, & s_{d}^{2}+2s_{d}+1
\end{pmatrix}
\]
is specified by the ordered set of signed magnitudes%
\begin{equation}
\left(  \mathbf{x}^{T}\mathbf{q}+1\right)  ^{2}\triangleq%
\begin{pmatrix}
\left(  \left\Vert \mathbf{q}\right\Vert ^{2}+1\right)  \cos\mathbb{\alpha
}_{k\left(  q_{1}\right)  1}, & \cdots, & \left(  \left\Vert \mathbf{q}%
\right\Vert ^{2}+1\right)  \cos\mathbb{\alpha}_{k\left(  q_{d}\right)  d}%
\end{pmatrix}
^{T}\text{.} \label{Geometric Locus of Second-order Reproducing Kernel}%
\end{equation}
Depending on the geometric context, reproducing kernels will be referred to as
points or vectors. Vectors $\mathbf{s}$ which are specified by reproducing
kernels $K(\mathbf{x},\mathbf{s})$ will be denoted by $k_{\mathbf{s}}$.

It will now be demonstrated that second-order distance statistics determine a
rich system of algebraic and topological relationships between the geometric
loci of two reproducing kernels.

\subsection{Second-order Distance Statistics for Kernels}

Returning to Eq. (\ref{Inner Product Relationship for Reproducing Kernels}),
the vector relationship%
\[
K(\mathbf{x},\mathbf{s})=\left\Vert k_{\mathbf{x}}(\mathbf{s})\right\Vert
\left\Vert k_{\mathbf{s}}(\mathbf{x})\right\Vert \cos\varphi
\]
between two reproducing kernels $k_{\mathbf{x}}(\mathbf{s})$ and
$k_{\mathbf{s}}(\mathbf{x})$ can be derived by using the law of cosines
\citep[see, e.g.,][]{Lay2006}%
:%
\begin{equation}
\left\Vert k_{\mathbf{x}}-k_{\mathbf{s}}\right\Vert ^{2}=\left\Vert
k_{\mathbf{x}}\right\Vert ^{2}+\left\Vert k_{\mathbf{s}}\right\Vert
^{2}-2\left\Vert k_{\mathbf{x}}\right\Vert \left\Vert k_{\mathbf{s}%
}\right\Vert \cos\varphi\label{Inner Product Statistic Reproducing Kernels}%
\end{equation}
which reduces to%
\begin{align*}
\left\Vert k_{\mathbf{x}}\right\Vert \left\Vert k_{\mathbf{s}}\right\Vert
\cos\varphi &  =k\left(  x_{1}\right)  k\left(  s_{1}\right)  +k\left(
x_{2}\right)  k\left(  s_{2}\right)  +\cdots+k\left(  x_{d}\right)  k\left(
s_{d}\right) \\
&  =K(\mathbf{x},\mathbf{s})=K(\mathbf{s},\mathbf{x})\text{.}%
\end{align*}

The vector relationships in Eq.
(\ref{Inner Product Statistic Reproducing Kernels}) indicate that the inner
product statistic $K(\mathbf{x},\mathbf{s}) $ determines the length
$\left\Vert k_{\mathbf{x}}-k_{\mathbf{s}}\right\Vert $ of the vector from
$k_{\mathbf{s}}$ to $k_{\mathbf{x}}$, i.e., the vector $k_{\mathbf{x}%
}-k_{\mathbf{s}}$, which is the distance between the endpoints of
$k_{\mathbf{x}}$ and $k_{\mathbf{s}}$, so that%
\[
K(\mathbf{x},\mathbf{s})=\left\Vert k_{\mathbf{x}}(\mathbf{s})\right\Vert
\left\Vert k_{\mathbf{s}}(\mathbf{x})\right\Vert \cos\varphi=\left\Vert
k_{\mathbf{x}}-k_{\mathbf{s}}\right\Vert \text{.}%
\]

Because second-order distance statistics are symmetric, the law of cosines%
\[
\left\Vert k_{\mathbf{s}}-k_{\mathbf{x}}\right\Vert ^{2}=\left\Vert
k_{\mathbf{s}}\right\Vert ^{2}+\left\Vert k_{\mathbf{x}}\right\Vert
^{2}-2\left\Vert k_{\mathbf{s}}\right\Vert \left\Vert k_{\mathbf{x}%
}\right\Vert \cos\varphi
\]
also determines the length $\left\Vert k_{\mathbf{s}}-k_{\mathbf{k}%
}\right\Vert $ of the vector from $k_{\mathbf{x}}$ to $k_{\mathbf{s}}$, i.e.,
the vector $k_{\mathbf{s}}-k_{\mathbf{k}}$, which is the distance between the
endpoints of $k_{\mathbf{s}}$ and $k_{\mathbf{x}}$.

Therefore, the inner product statistic $K(\mathbf{x},\mathbf{s})$ between two
reproducing kernels $k_{\mathbf{x}}(\mathbf{s})$ and $k_{\mathbf{s}%
}(\mathbf{x})$ in a $\mathfrak{rkH}$ space%
\begin{align}
K(\mathbf{x},\mathbf{s})  &  =k\left(  x_{1}\right)  k\left(  s_{1}\right)
+k\left(  x_{2}\right)  k\left(  s_{2}\right)  +\cdots+k\left(  x_{d}\right)
k\left(  s_{d}\right) \label{Locus Statistics in RKHS}\\
&  =\left\Vert k_{\mathbf{x}}\right\Vert \left\Vert k_{\mathbf{s}}\right\Vert
\cos\varphi\nonumber\\
&  =\left\Vert k_{\mathbf{x}}-k_{\mathbf{s}}\right\Vert \nonumber
\end{align}
determines the distance between the geometric loci%
\[%
\begin{pmatrix}
\left\Vert k_{\mathbf{x}}\right\Vert \cos\mathbb{\alpha}_{k\left(
\mathbf{x}_{1}\right)  1}, & \left\Vert k_{\mathbf{x}}\right\Vert
\cos\mathbb{\alpha}_{k\left(  \mathbf{x}_{2}\right)  2}, & \cdots, &
\left\Vert k_{\mathbf{x}}\right\Vert \cos\mathbb{\alpha}_{k\left(
\mathbf{x}_{d}\right)  d}%
\end{pmatrix}
\]
and%
\[%
\begin{pmatrix}
\left\Vert k_{\mathbf{s}}\right\Vert \cos\mathbb{\alpha}_{k\left(
\mathbf{s}_{1}\right)  1}, & \left\Vert k_{\mathbf{s}}\right\Vert
\cos\mathbb{\alpha}_{k\left(  \mathbf{s}_{2}\right)  2}, & \cdots, &
\left\Vert k_{\mathbf{s}}\right\Vert \cos\mathbb{\alpha}_{k\left(
\mathbf{s}_{d}\right)  d}%
\end{pmatrix}
\]
of the given reproducing kernels.

\begin{figure}[ptb]%
\centering
\fbox{\includegraphics[
height=2.5875in,
width=3.4411in
]%
{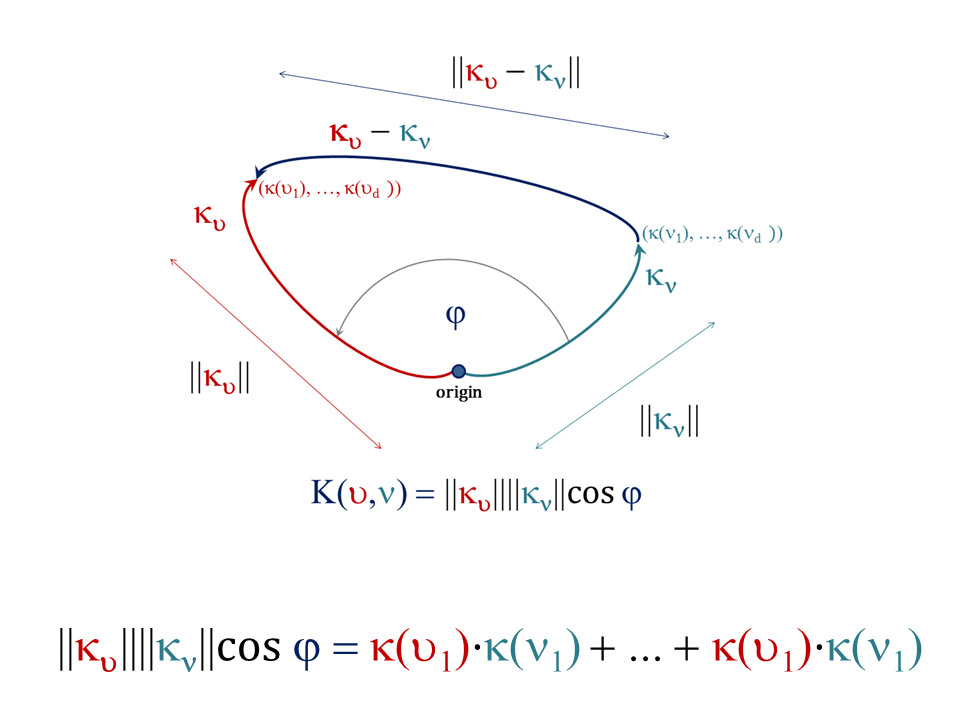}%
}\caption{Inner product statistics $K(\boldsymbol{\upsilon},\boldsymbol{\nu})$
in reproducing kernel Hilbert spaces determine angles and corresponding
distances between the geometric loci of reproducing kernels
$k_{\boldsymbol{\upsilon}}$ and $k_{\boldsymbol{\nu}}$.}%
\label{Second-order Distance Statisitcs RKHS}%
\end{figure}

Thus, it is concluded that the vector relationships contained within Eq.
(\ref{Inner Product Statistic Reproducing Kernels}) determine a rich system of
topological relationships between the loci of two reproducing kernels. Figure
$\ref{Second-order Distance Statisitcs RKHS}$ depicts correlated algebraic,
geometric, and topological structures determined by an inner product statistic
$K(\boldsymbol{\upsilon},\boldsymbol{\nu})$ in a $\mathfrak{rkH}$ space.%
\textbf{\ }

\subsubsection{Scalar Projection Statistics for Kernels}

Scalar projection statistics specify signed magnitudes along the axes of given
reproducing kernels. The inner product statistic $\left\Vert k_{\mathbf{x}%
}(\mathbf{s})\right\Vert \left\Vert k_{\mathbf{s}}(\mathbf{x})\right\Vert
\cos\theta$ can be interpreted as the length $\left\Vert k_{\mathbf{x}%
}(\mathbf{s})\right\Vert $ of $k_{\mathbf{x}}$ times the scalar projection of
$k_{\mathbf{s}}$ onto $k_{\mathbf{x}}$%
\begin{equation}
K(\mathbf{x},\mathbf{s})=\left\Vert k_{\mathbf{x}}(\mathbf{s})\right\Vert
\times\left[  \left\Vert k_{\mathbf{s}}(\mathbf{x})\right\Vert \cos
\theta\right]  \text{,} \label{Scalar Projection Reproducing Kernels}%
\end{equation}
where the scalar projection of $k_{\mathbf{s}}$ onto $k_{\mathbf{x}}$, also
known as the component of $k_{\mathbf{s}}$ along $k_{\mathbf{x}}$, is defined
to be the signed magnitude $\left\Vert k_{\mathbf{s}}(\mathbf{x})\right\Vert
\cos\theta$ of the vector projection, where $\theta$ is the angle between
$k_{\mathbf{x}}$ and $k_{\mathbf{s}}$
\citep{Stewart2009}%
. Scalar projections are denoted by $\operatorname{comp}%
_{\overrightarrow{k_{\mathbf{x}}}}\left(  \overrightarrow{k_{\mathbf{s}}%
}\right)  $, where $\operatorname{comp}_{\overrightarrow{k_{\mathbf{x}}}%
}\left(  \overrightarrow{k_{\mathbf{s}}}\right)  <0$ if $\pi/2<\theta\leq\pi$.
The scalar projection statistic $\left\Vert k_{\mathbf{s}}(\mathbf{x}%
)\right\Vert \cos\theta$ satisfies the inner product relationship $\left\Vert
k_{\mathbf{s}}(\mathbf{x})\right\Vert \cos\theta=\frac{K(\mathbf{x}%
,\mathbf{s})}{\left\Vert k_{\mathbf{x}}\right\Vert }$ between the unit vector
$\frac{k_{\mathbf{x}}}{\left\Vert k_{\mathbf{x}}\right\Vert }$ and the vector
$k_{\mathbf{s}}$.

In the next part of the paper, I\ will devise three systems of data-driven,
locus equations which generate optimal likelihood ratio tests, i.e., optimal
binary classification systems, for digital data. The data-driven, mathematical
laws provide a constructive solution to the fundamental integral equation of
binary classification for a classification system in statistical equilibrium
in Eq. (\ref{Equalizer Rule}). I\ have devised a theorem that states the
essential aspects of the binary classification problem. The binary
classification theorem is stated below.

\section*{Binary Classification Theorem}

Let $\widehat{\Lambda}\left(  \mathbf{x}\right)  =p\left(  \widehat{\Lambda
}\left(  \mathbf{x}\right)  |\omega_{1}\right)  -p\left(  \widehat{\Lambda
}\left(  \mathbf{x}\right)  |\omega_{2}\right)  \overset{\omega_{1}%
}{\underset{\omega_{2}}{\gtrless}}0$ denote the likelihood ratio test for a
binary classification system, where $\omega_{1}$ or $\omega_{2}$ is the true
data category, and $d$-component random vectors $\mathbf{x}$ from class
$\omega_{1}$ and class $\omega_{2}$ are generated according to probability
density functions $p\left(  \mathbf{x}|\omega_{1}\right)  $ and $p\left(
\mathbf{x}|\omega_{2}\right)  $ related to statistical distributions of random
vectors $\mathbf{x}$ that have constant or unchanging statistics.

The discriminant function%
\[
\widehat{\Lambda}\left(  \mathbf{x}\right)  =p\left(  \widehat{\Lambda}\left(
\mathbf{x}\right)  |\omega_{1}\right)  -p\left(  \widehat{\Lambda}\left(
\mathbf{x}\right)  |\omega_{2}\right)
\]
is the solution to the integral equation%
\begin{align*}
f\left(  \widehat{\Lambda}\left(  \mathbf{x}\right)  \right)   &  =\int%
_{Z_{1}}p\left(  \widehat{\Lambda}\left(  \mathbf{x}\right)  |\omega
_{1}\right)  d\widehat{\Lambda}+\int_{Z_{2}}p\left(  \widehat{\Lambda}\left(
\mathbf{x}\right)  |\omega_{1}\right)  d\widehat{\Lambda}\\
&  =\int_{Z_{1}}p\left(  \widehat{\Lambda}\left(  \mathbf{x}\right)
|\omega_{2}\right)  d\widehat{\Lambda}+\int_{Z_{2}}p\left(  \widehat{\Lambda
}\left(  \mathbf{x}\right)  |\omega_{2}\right)  d\widehat{\Lambda}\text{,}%
\end{align*}
over the decision space $Z=Z_{1}+Z_{2}$, such that the expected risk
$\mathfrak{R}_{\mathfrak{\min}}\left(  Z|\widehat{\Lambda}\left(
\mathbf{x}\right)  \right)  $ and the corresponding eigenenergy $E_{\min
}\left(  Z|\widehat{\Lambda}\left(  \mathbf{x}\right)  \right)  $ of the
classification system $p\left(  \widehat{\Lambda}\left(  \mathbf{x}\right)
|\omega_{1}\right)  -p\left(  \widehat{\Lambda}\left(  \mathbf{x}\right)
|\omega_{2}\right)  \overset{\omega_{1}}{\underset{\omega_{2}}{\gtrless}}0$
are governed by the equilibrium point%
\[
p\left(  \widehat{\Lambda}\left(  \mathbf{x}\right)  |\omega_{1}\right)
-p\left(  \widehat{\Lambda}\left(  \mathbf{x}\right)  |\omega_{2}\right)  =0
\]
of the integral equation $f\left(  \widehat{\Lambda}\left(  \mathbf{x}\right)
\right)  $.

Therefore, the forces associated with the counter risk $\overline
{\mathfrak{R}}_{\mathfrak{\min}}\left(  Z_{1}|p\left(  \widehat{\Lambda
}\left(  \mathbf{x}\right)  |\omega_{1}\right)  \right)  $ and the risk
$\mathfrak{R}_{\mathfrak{\min}}\left(  Z_{2}|p\left(  \widehat{\Lambda}\left(
\mathbf{x}\right)  |\omega_{1}\right)  \right)  $ in the $Z_{1}$ and $Z_{2}$
decision regions, which are related to positions and potential locations of
random vectors $\mathbf{x}$ that are generated according to $p\left(
\mathbf{x}|\omega_{1}\right)  $, are equal to the forces associated with the
risk $\mathfrak{R}_{\mathfrak{\min}}\left(  Z_{1}|p\left(  \widehat{\Lambda
}\left(  \mathbf{x}\right)  |\omega_{2}\right)  \right)  $ and the counter
risk $\overline{\mathfrak{R}}_{\mathfrak{\min}}\left(  Z_{2}|p\left(
\widehat{\Lambda}\left(  \mathbf{x}\right)  |\omega_{2}\right)  \right)  $ in
the $Z_{1}$ and $Z_{2}$ decision regions, which are related to positions and
potential locations of random vectors $\mathbf{x}$ that are generated
according to $p\left(  \mathbf{x}|\omega_{2}\right)  $.

Furthermore, the eigenenergy $E_{\min}\left(  Z|p\left(  \widehat{\Lambda
}\left(  \mathbf{x}\right)  |\omega_{1}\right)  \right)  $ associated with the
position or location of the likelihood ratio $p\left(  \widehat{\Lambda
}\left(  \mathbf{x}\right)  |\omega_{1}\right)  $ given class $\omega_{1}$ is
equal to the eigenenergy $E_{\min}\left(  Z|p\left(  \widehat{\Lambda}\left(
\mathbf{x}\right)  |\omega_{2}\right)  \right)  $ associated with the position
or location of the likelihood ratio $p\left(  \widehat{\Lambda}\left(
\mathbf{x}\right)  |\omega_{2}\right)  $ given class $\omega_{2}$:%
\[
E_{\min}\left(  Z|p\left(  \widehat{\Lambda}\left(  \mathbf{x}\right)
|\omega_{1}\right)  \right)  =E_{\min}\left(  Z|p\left(  \widehat{\Lambda
}\left(  \mathbf{x}\right)  |\omega_{2}\right)  \right)  \text{.}%
\]

Thus, the total eigenenergy $E_{\min}\left(  Z|\widehat{\Lambda}\left(
\mathbf{x}\right)  \right)  $ of the binary classification system $p\left(
\widehat{\Lambda}\left(  \mathbf{x}\right)  |\omega_{1}\right)  -p\left(
\widehat{\Lambda}\left(  \mathbf{x}\right)  |\omega_{2}\right)
\overset{\omega_{1}}{\underset{\omega_{2}}{\gtrless}}0$ is equal to the
eigenenergies associated with the position or location of the likelihood ratio
$p\left(  \widehat{\Lambda}\left(  \mathbf{x}\right)  |\omega_{1}\right)
-p\left(  \widehat{\Lambda}\left(  \mathbf{x}\right)  |\omega_{2}\right)  $
and the locus of a corresponding decision boundary $p\left(  \widehat{\Lambda
}\left(  \mathbf{x}\right)  |\omega_{1}\right)  -p\left(  \widehat{\Lambda
}\left(  \mathbf{x}\right)  |\omega_{2}\right)  =0$:%
\[
E_{\min}\left(  Z|\widehat{\Lambda}\left(  \mathbf{x}\right)  \right)
=E_{\min}\left(  Z|p\left(  \widehat{\Lambda}\left(  \mathbf{x}\right)
|\omega_{1}\right)  \right)  +E_{\min}\left(  Z|p\left(  \widehat{\Lambda
}\left(  \mathbf{x}\right)  |\omega_{2}\right)  \right)  \text{.}%
\]

It follows that the binary classification system $p\left(  \widehat{\Lambda
}\left(  \mathbf{x}\right)  |\omega_{1}\right)  -p\left(  \widehat{\Lambda
}\left(  \mathbf{x}\right)  |\omega_{2}\right)  \overset{\omega_{1}%
}{\underset{\omega_{2}}{\gtrless}}0$ is in statistical equilibrium:%
\begin{align*}
f\left(  \widehat{\Lambda}\left(  \mathbf{x}\right)  \right)   &  :\int%
_{Z_{1}}p\left(  \widehat{\Lambda}\left(  \mathbf{x}\right)  |\omega
_{1}\right)  d\widehat{\Lambda}-\int_{Z_{1}}p\left(  \widehat{\Lambda}\left(
\mathbf{x}\right)  |\omega_{2}\right)  d\widehat{\Lambda}\\
&  =\int_{Z_{2}}p\left(  \widehat{\Lambda}\left(  \mathbf{x}\right)
|\omega_{2}\right)  d\widehat{\Lambda}-\int_{Z_{2}}p\left(  \widehat{\Lambda
}\left(  \mathbf{x}\right)  |\omega_{1}\right)  d\widehat{\Lambda}\text{,}%
\end{align*}
where the forces associated with the counter risk $\overline{\mathfrak{R}%
}_{\mathfrak{\min}}\left(  Z_{1}|p\left(  \widehat{\Lambda}\left(
\mathbf{x}\right)  |\omega_{1}\right)  \right)  $ for class $\omega_{1}$ and
the risk $\mathfrak{R}_{\mathfrak{\min}}\left(  Z_{1}|p\left(
\widehat{\Lambda}\left(  \mathbf{x}\right)  |\omega_{2}\right)  \right)  $ for
class $\omega_{2}$ in the $Z_{1}$ decision region are balanced with the forces
associated with the counter risk $\overline{\mathfrak{R}}_{\mathfrak{\min}%
}\left(  Z_{2}|p\left(  \widehat{\Lambda}\left(  \mathbf{x}\right)
|\omega_{2}\right)  \right)  $ for class $\omega_{2}$ and the risk
$\mathfrak{R}_{\mathfrak{\min}}\left(  Z_{2}|p\left(  \widehat{\Lambda}\left(
\mathbf{x}\right)  |\omega_{1}\right)  \right)  $ for class $\omega_{1}$ in
the $Z_{2}$ decision region such that the expected risk $\mathfrak{R}%
_{\mathfrak{\min}}\left(  Z|\widehat{\Lambda}\left(  \mathbf{x}\right)
\right)  $ of the classification system is minimized, and the eigenenergies
associated with the counter risk $\overline{\mathfrak{R}}_{\mathfrak{\min}%
}\left(  Z_{1}|p\left(  \widehat{\Lambda}\left(  \mathbf{x}\right)
|\omega_{1}\right)  \right)  $ for class $\omega_{1}$ and the risk
$\mathfrak{R}_{\mathfrak{\min}}\left(  Z_{1}|p\left(  \widehat{\Lambda}\left(
\mathbf{x}\right)  |\omega_{2}\right)  \right)  $ for class $\omega_{2}$ in
the $Z_{1}$ decision region are balanced with the eigenenergies associated
with the counter risk $\overline{\mathfrak{R}}_{\mathfrak{\min}}\left(
Z_{2}|p\left(  \widehat{\Lambda}\left(  \mathbf{x}\right)  |\omega_{2}\right)
\right)  $ for class $\omega_{2}$ and the risk $\mathfrak{R}_{\mathfrak{\min}%
}\left(  Z_{2}|p\left(  \widehat{\Lambda}\left(  \mathbf{x}\right)
|\omega_{1}\right)  \right)  $ for class $\omega_{1}$ in the $Z_{2}$ decision
region such that the eigenenergy $E_{\min}\left(  Z|\widehat{\Lambda}\left(
\mathbf{x}\right)  \right)  $ of the classification system is minimized.

Thus, any given binary classification system $p\left(  \widehat{\Lambda
}\left(  \mathbf{x}\right)  |\omega_{1}\right)  -p\left(  \widehat{\Lambda
}\left(  \mathbf{x}\right)  |\omega_{2}\right)  \overset{\omega_{1}%
}{\underset{\omega_{2}}{\gtrless}}0$ exhibits an error rate that is consistent
with the expected risk $\mathfrak{R}_{\mathfrak{\min}}\left(
Z|\widehat{\Lambda}\left(  \mathbf{x}\right)  \right)  $ and the corresponding
eigenenergy $E_{\min}\left(  Z|\widehat{\Lambda}\left(  \mathbf{x}\right)
\right)  $ of the classification system: for all random vectors $\mathbf{x}$
that are generated according to $p\left(  \mathbf{x}|\omega_{1}\right)  $ and
$p\left(  \mathbf{x}|\omega_{2}\right)  $, where $p\left(  \mathbf{x}%
|\omega_{1}\right)  $ and $p\left(  \mathbf{x}|\omega_{2}\right)  $ are
related to statistical distributions of random vectors $\mathbf{x}$ that have
constant or unchanging statistics.

Therefore, a binary classification system $p\left(  \widehat{\Lambda}\left(
\mathbf{x}\right)  |\omega_{1}\right)  -p\left(  \widehat{\Lambda}\left(
\mathbf{x}\right)  |\omega_{2}\right)  \overset{\omega_{1}}{\underset{\omega
_{2}}{\gtrless}}0$ seeks a point of statistical equilibrium where the opposing
forces and influences of the classification system are balanced with each
other, such that the eigenenergy and the expected risk of the classification
system are minimized, and the classification system is in statistical equilibrium.

\subsection*{Proof}

It can be shown that the general form of the decision function for a binary
classification system is given by the likelihood ratio test:%
\begin{equation}
\Lambda\left(  \mathbf{x}\right)  \triangleq\frac{p\left(  \mathbf{x}%
|\omega_{1}\right)  }{p\left(  \mathbf{x}|\omega_{2}\right)  }\overset{\omega
_{1}}{\underset{\omega_{2}}{\gtrless}}1\text{,}
\label{General Form of Decision Function}%
\end{equation}
where $\omega_{1}$ or $\omega_{2}$ is the true data category and $p\left(
\mathbf{x}|\omega_{1}\right)  $ and $p\left(  \mathbf{x}|\omega_{2}\right)  $
are class-conditional probability density functions.

Now, take the transform $\ln\left(  \Lambda\left(  \mathbf{x}\right)  \right)
\overset{\omega_{1}}{\underset{\omega_{2}}{\gtrless}}$ $\ln\left(  1\right)  $
of the likelihood ratio test for a binary classification system%
\[
\ln\Lambda\left(  \mathbf{x}\right)  \triangleq\ln p\left(  \mathbf{x}%
|\omega_{1}\right)  -\ln p\left(  \mathbf{x}|\omega_{2}\right)
\overset{\omega_{1}}{\underset{\omega_{2}}{\gtrless}}\ln1\text{,}%
\]
where $\ln\exp\left(  x\right)  =x$, and let $\widehat{\Lambda}\left(
\mathbf{x}\right)  $ denote the transform $\ln\left(  \Lambda\left(
\mathbf{x}\right)  \right)  $ of the likelihood ratio $\Lambda\left(
\mathbf{x}\right)  =\frac{p\left(  \mathbf{x}|\omega_{1}\right)  }{p\left(
\mathbf{x}|\omega_{2}\right)  }$.

It follows that the general form of the decision function for a binary
classification system can be written as:%
\begin{align}
\widehat{\Lambda}\left(  \mathbf{x}\right)   &  \triangleq\ln p\left(
\mathbf{x}|\omega_{1}\right)  -\ln p\left(  \mathbf{x}|\omega_{2}\right)
\overset{\omega_{1}}{\underset{\omega_{2}}{\gtrless}}%
0\label{General Form of Decision Function II}\\
&  =p\left(  \widehat{\Lambda}\left(  \mathbf{x}\right)  |\omega_{1}\right)
-p\left(  \widehat{\Lambda}\left(  \mathbf{x}\right)  |\omega_{2}\right)
\overset{\omega_{1}}{\underset{\omega_{2}}{\gtrless}}0\text{.}\nonumber
\end{align}

Given Eq. (\ref{General Form of Decision Function}), the general form of the
decision function and the decision boundary for Gaussian data are completely
defined by the likelihood ratio test:%
\[
\Lambda\left(  \mathbf{x}\right)  =\frac{\left\vert \mathbf{\Sigma}%
_{2}\right\vert ^{1/2}\exp\left\{  -\frac{1}{2}\left(  \mathbf{x}%
-\boldsymbol{\mu}_{1}\right)  ^{T}\mathbf{\Sigma}_{1}^{-1}\left(
\mathbf{x}-\boldsymbol{\mu}_{1}\right)  \right\}  }{\left\vert \mathbf{\Sigma
}_{1}\right\vert ^{1/2}\exp\left\{  -\frac{1}{2}\left(  \mathbf{x}%
-\boldsymbol{\mu}_{2}\right)  ^{T}\mathbf{\Sigma}_{2}^{-1}\left(
\mathbf{x}-\boldsymbol{\mu}_{2}\right)  \right\}  }\overset{\omega
_{1}}{\underset{\omega_{2}}{\gtrless}}1\text{,}%
\]
where $\boldsymbol{\mu}_{1}$ and $\boldsymbol{\mu}_{2}$ are $d$-component mean
vectors, $\mathbf{\Sigma}_{1}$ and $\mathbf{\Sigma}_{2}$ are $d$-by-$d$
covariance matrices, $\mathbf{\Sigma}^{-1}$ and $\left\vert \mathbf{\Sigma
}\right\vert $ denote the inverse and determinant of a covariance matrix, and
$\omega_{1}$ or $\omega_{2}$ is the true data category.

So, take the transform $\ln\left(  \Lambda\left(  \mathbf{x}\right)  \right)
\overset{\omega_{1}}{\underset{\omega_{2}}{\gtrless}}$ $\ln\left(  1\right)  $
of the likelihood ratio test for Gaussian data. Accordingly, take any given
decision boundary $D\left(  \mathbf{x}\right)  $:%
\begin{align*}
D\left(  \mathbf{x}\right)   &  :\mathbf{x}^{T}\mathbf{\Sigma}_{1}%
^{-1}\boldsymbol{\mu}_{1}-\frac{1}{2}\mathbf{x}^{T}\mathbf{\Sigma}_{1}%
^{-1}\mathbf{x}-\frac{1}{2}\boldsymbol{\mu}_{1}^{T}\mathbf{\Sigma}_{1}%
^{-1}\boldsymbol{\mu}_{1}-\frac{1}{2}\ln\left(  \left\vert \mathbf{\Sigma}%
_{1}\right\vert ^{1/2}\right) \\
&  -\mathbf{x}^{T}\mathbf{\Sigma}_{2}^{-1}\boldsymbol{\mu}_{2}+\frac{1}%
{2}\mathbf{x}^{T}\mathbf{\Sigma}_{2}^{-1}\mathbf{x+}\frac{1}{2}\boldsymbol{\mu
}_{2}^{T}\mathbf{\Sigma}_{2}^{-1}\boldsymbol{\mu}_{2}+\frac{1}{2}\ln\left(
\left\vert \mathbf{\Sigma}_{2}\right\vert ^{1/2}\right) \\
&  =0
\end{align*}
that is generated according to the likelihood ratio test:%
\begin{align*}
\widehat{\Lambda}\left(  \mathbf{x}\right)   &  =\mathbf{x}^{T}\mathbf{\Sigma
}_{1}^{-1}\boldsymbol{\mu}_{1}-\frac{1}{2}\mathbf{x}^{T}\mathbf{\Sigma}%
_{1}^{-1}\mathbf{x}-\frac{1}{2}\boldsymbol{\mu}_{1}^{T}\mathbf{\Sigma}%
_{1}^{-1}\boldsymbol{\mu}_{1}-\frac{1}{2}\ln\left(  \left\vert \mathbf{\Sigma
}_{1}\right\vert ^{1/2}\right) \\
&  -\mathbf{x}^{T}\mathbf{\Sigma}_{2}^{-1}\boldsymbol{\mu}_{2}-\frac{1}%
{2}\mathbf{x}^{T}\mathbf{\Sigma}_{2}^{-1}\mathbf{x-}\frac{1}{2}\boldsymbol{\mu
}_{2}^{T}\mathbf{\Sigma}_{2}^{-1}\boldsymbol{\mu}_{2}-\frac{1}{2}\ln\left(
\left\vert \mathbf{\Sigma}_{2}\right\vert ^{1/2}\right)  \overset{\omega
_{1}}{\underset{\omega_{2}}{\gtrless}}0\text{.}%
\end{align*}

It follows that the decision space $Z$ and the corresponding decision regions
$Z_{1}$ and $Z_{2}$ of the classification system%
\[
p\left(  \widehat{\Lambda}\left(  \mathbf{x}\right)  |\omega_{1}\right)
-p\left(  \widehat{\Lambda}\left(  \mathbf{x}\right)  |\omega_{2}\right)
\overset{\omega_{1}}{\underset{\omega_{2}}{\gtrless}}0
\]
are determined by either overlapping or non-overlapping data distributions.

It also follows that the decision boundary $D\left(  \mathbf{x}\right)  $ and
the likelihood ratio $\widehat{\Lambda}\left(  \mathbf{x}\right)  =p\left(
\widehat{\Lambda}\left(  \mathbf{x}\right)  |\omega_{1}\right)  -p\left(
\widehat{\Lambda}\left(  \mathbf{x}\right)  |\omega_{2}\right)  $ satisfy the
vector equation:%
\[
p\left(  \widehat{\Lambda}\left(  \mathbf{x}\right)  |\omega_{1}\right)
-p\left(  \widehat{\Lambda}\left(  \mathbf{x}\right)  |\omega_{2}\right)
=0\text{,}%
\]
where the likelihood ratio $p\left(  \widehat{\Lambda}\left(  \mathbf{x}%
\right)  |\omega_{1}\right)  $ given class $\omega_{1}$ is given by the vector
expression:%
\[
p\left(  \widehat{\Lambda}\left(  \mathbf{x}\right)  |\omega_{1}\right)
=\mathbf{x}^{T}\mathbf{\Sigma}_{1}^{-1}\boldsymbol{\mu}_{1}-\frac{1}%
{2}\mathbf{x}^{T}\mathbf{\Sigma}_{1}^{-1}\mathbf{x}-\frac{1}{2}\boldsymbol{\mu
}_{1}^{T}\mathbf{\Sigma}_{1}^{-1}\boldsymbol{\mu}_{1}-\frac{1}{2}\ln\left(
\left\vert \mathbf{\Sigma}_{1}\right\vert ^{1/2}\right)
\]
of a class-conditional probability density function $p\left(  \mathbf{x}%
|\omega_{1}\right)  $, and the likelihood ratio $p\left(  \widehat{\Lambda
}\left(  \mathbf{x}\right)  |\omega_{2}\right)  $ given class $\omega_{2}$ is
given by the vector expression:%
\[
p\left(  \widehat{\Lambda}\left(  \mathbf{x}\right)  |\omega_{2}\right)
=\mathbf{x}^{T}\mathbf{\Sigma}_{2}^{-1}\boldsymbol{\mu}_{2}-\frac{1}%
{2}\mathbf{x}^{T}\mathbf{\Sigma}_{2}^{-1}\mathbf{x-}\frac{1}{2}\boldsymbol{\mu
}_{2}^{T}\mathbf{\Sigma}_{2}^{-1}\boldsymbol{\mu}_{2}-\frac{1}{2}\ln\left(
\left\vert \mathbf{\Sigma}_{2}\right\vert ^{1/2}\right)
\]
of a class-conditional probability density function $p\left(  \mathbf{x}%
|\omega_{2}\right)  $.

Therefore, the loci of the decision boundary $D\left(  \mathbf{x}\right)  $
and the likelihood ratio $\widehat{\Lambda}\left(  \mathbf{x}\right)  $ are
determined by a locus of principal eigenaxis components and likelihoods:%
\begin{align*}
\widehat{\Lambda}\left(  \mathbf{x}\right)   &  =p\left(  \widehat{\Lambda
}\left(  \mathbf{x}\right)  |\omega_{1}\right)  -p\left(  \widehat{\Lambda
}\left(  \mathbf{x}\right)  |\omega_{2}\right) \\
&  \triangleq\sum\nolimits_{i=1}^{l_{1}}\psi_{1_{i}}k_{\mathbf{x}_{1_{i}}%
}-\sum\nolimits_{i=1}^{l_{2}}\psi_{2_{i}}k_{\mathbf{x}_{2_{i}}}\text{,}%
\end{align*}
where $k_{\mathbf{x}_{1_{i}}}$ and $k_{\mathbf{x}_{2_{i}}}$ are reproducing
kernels for respective data points $\mathbf{x}_{1_{i}}$ and $\mathbf{x}%
_{2_{i}}$: the reproducing kernel $K(\mathbf{x,s})=k_{\mathbf{s}}(\mathbf{x})$
is either $k_{\mathbf{s}}(\mathbf{x})\triangleq\left(  \mathbf{x}%
^{T}\mathbf{s}+1\right)  ^{2}$ or $k_{\mathbf{s}}(\mathbf{x})\triangleq
\exp\left(  -0.01\left\Vert \mathbf{x}-\mathbf{s}\right\Vert ^{2}\right)  $,
$\mathbf{x}_{1i}\sim p\left(  \mathbf{x}|\omega_{1}\right)  $, $\mathbf{x}%
_{2i}\sim p\left(  \mathbf{x}|\omega_{2}\right)  $, and $\psi_{1i}$ and
$\psi_{2i}$ are scale factors that provide measures of likelihood for
respective data points $\mathbf{x}_{1i}$ and $\mathbf{x}_{2i}$ which lie in
either overlapping regions or tails regions of data distributions related to
$p\left(  \mathbf{x}|\omega_{1}\right)  $ and $p\left(  \mathbf{x}|\omega
_{2}\right)  $, that is in statistical equilibrium:%
\[
p\left(  \widehat{\Lambda}\left(  \mathbf{x}\right)  |\omega_{1}\right)
\rightleftharpoons p\left(  \widehat{\Lambda}\left(  \mathbf{x}\right)
|\omega_{2}\right)
\]
such that the locus of principal eigenaxis components and likelihoods
$\widehat{\Lambda}\left(  \mathbf{x}\right)  =\sum\nolimits_{i=1}^{l_{1}}%
\psi_{1_{i}}k_{\mathbf{x}_{1_{i}}}-\sum\nolimits_{i=1}^{l_{2}}\psi_{2_{i}%
}k_{\mathbf{x}_{2_{i}}}$ satisfies an equilibrium equation:%
\[
\sum\nolimits_{i=1}^{l_{1}}\psi_{1_{i}}k_{\mathbf{x}_{1_{i}}}%
\rightleftharpoons\sum\nolimits_{i=1}^{l_{2}}\psi_{2_{i}}k_{\mathbf{x}_{2_{i}%
}}%
\]
in an optimal manner.

Thus, the discriminant function%
\begin{align*}
\widehat{\Lambda}\left(  \mathbf{x}\right)   &  =p\left(  \widehat{\Lambda
}\left(  \mathbf{x}\right)  |\omega_{1}\right)  -p\left(  \widehat{\Lambda
}\left(  \mathbf{x}\right)  |\omega_{2}\right) \\
&  \triangleq\sum\nolimits_{i=1}^{l_{1}}\psi_{1_{i}}k_{\mathbf{x}_{1_{i}}%
}-\sum\nolimits_{i=1}^{l_{2}}\psi_{2_{i}}k_{\mathbf{x}_{2_{i}}}\text{,}%
\end{align*}
is the solution to the integral equation%
\begin{align*}
f\left(  \widehat{\Lambda}\left(  \mathbf{x}\right)  \right)   &  =\int%
_{Z_{1}}p\left(  \widehat{\Lambda}\left(  \mathbf{x}\right)  |\omega
_{1}\right)  d\widehat{\Lambda}+\int_{Z_{2}}p\left(  \widehat{\Lambda}\left(
\mathbf{x}\right)  |\omega_{1}\right)  d\widehat{\Lambda}\\
&  =\int_{Z_{1}}p\left(  \widehat{\Lambda}\left(  \mathbf{x}\right)
|\omega_{2}\right)  d\widehat{\Lambda}+\int_{Z_{2}}p\left(  \widehat{\Lambda
}\left(  \mathbf{x}\right)  |\omega_{2}\right)  d\widehat{\Lambda}\text{,}%
\end{align*}
over the decision space $Z=Z_{1}+Z_{2}$, where the equilibrium point $p\left(
\widehat{\Lambda}\left(  \mathbf{x}\right)  |\omega_{1}\right)  -p\left(
\widehat{\Lambda}\left(  \mathbf{x}\right)  |\omega_{2}\right)  =0$:%
\[
\sum\nolimits_{i=1}^{l_{1}}\psi_{1_{i}}k_{\mathbf{x}_{1_{i}}}-\sum
\nolimits_{i=1}^{l_{2}}\psi_{2_{i}}k_{\mathbf{x}_{2_{i}}}=0
\]
is the focus of a decision boundary $D\left(  \mathbf{x}\right)  $, and
$Z_{1}$ and $Z_{2}$ are decision regions that have respective risks for class
$\omega_{2}$ and class $\omega_{1}$:%
\[
\mathfrak{R}_{\mathfrak{\min}}\left(  Z_{1}|p\left(  \widehat{\Lambda}\left(
\mathbf{x}\right)  |\omega_{2}\right)  \right)  \text{ \ and \ }%
\mathfrak{R}_{\mathfrak{\min}}\left(  Z_{2}|p\left(  \widehat{\Lambda}\left(
\mathbf{x}\right)  |\omega_{1}\right)  \right)
\]
and respective counter risks for class $\omega_{1}$ and class $\omega_{2}$:%
\[
\overline{\mathfrak{R}}_{\mathfrak{\min}}\left(  Z_{1}|p\left(
\widehat{\Lambda}\left(  \mathbf{x}\right)  |\omega_{1}\right)  \right)
\text{ \ and \ }\overline{\mathfrak{R}}_{\mathfrak{\min}}\left(
Z_{2}|p\left(  \widehat{\Lambda}\left(  \mathbf{x}\right)  |\omega_{2}\right)
\right)  \text{,}%
\]
where the forces associated with the counter risk $\overline{\mathfrak{R}%
}_{\mathfrak{\min}}\left(  Z_{1}|p\left(  \widehat{\Lambda}\left(
\mathbf{x}\right)  |\omega_{1}\right)  \right)  $ for class $\omega_{1}$ in
the $Z_{1}$ decision region and the risk $\mathfrak{R}_{\mathfrak{\min}%
}\left(  Z_{2}|p\left(  \widehat{\Lambda}\left(  \mathbf{x}\right)
|\omega_{1}\right)  \right)  $ for class $\omega_{1}$ in the $Z_{2}$ decision
region are balanced with the forces associated with the risk $\mathfrak{R}%
_{\mathfrak{\min}}\left(  Z_{1}|p\left(  \widehat{\Lambda}\left(
\mathbf{x}\right)  |\omega_{2}\right)  \right)  $ for class $\omega_{2}$ in
the $Z_{1}$ decision region and the counter risk $\overline{\mathfrak{R}%
}_{\mathfrak{\min}}\left(  Z_{2}|p\left(  \widehat{\Lambda}\left(
\mathbf{x}\right)  |\omega_{2}\right)  \right)  $ for class $\omega_{2}$ in
the $Z_{2}$ decision region:%
\begin{align*}
f\left(  \widehat{\Lambda}\left(  \mathbf{x}\right)  \right)   &
:\overline{\mathfrak{R}}_{\mathfrak{\min}}\left(  Z_{1}|p\left(
\widehat{\Lambda}\left(  \mathbf{x}\right)  |\omega_{1}\right)  \right)
+\mathfrak{R}_{\mathfrak{\min}}\left(  Z_{2}|p\left(  \widehat{\Lambda}\left(
\mathbf{x}\right)  |\omega_{1}\right)  \right) \\
&  =\mathfrak{R}_{\mathfrak{\min}}\left(  Z_{1}|p\left(  \widehat{\Lambda
}\left(  \mathbf{x}\right)  |\omega_{2}\right)  \right)  +\overline
{\mathfrak{R}}_{\mathfrak{\min}}\left(  Z_{2}|p\left(  \widehat{\Lambda
}\left(  \mathbf{x}\right)  |\omega_{2}\right)  \right)  \text{,}%
\end{align*}
and the eigenenergy $E_{\min}\left(  Z|p\left(  \widehat{\Lambda}\left(
\mathbf{x}\right)  |\omega_{1}\right)  \right)  $ associated with the position
or location of the likelihood ratio $p\left(  \widehat{\Lambda}\left(
\mathbf{x}\right)  |\omega_{1}\right)  $ given class $\omega_{1}$ is balanced
with the eigenenergy $E_{\min}\left(  Z|p\left(  \widehat{\Lambda}\left(
\mathbf{x}\right)  |\omega_{2}\right)  \right)  $ associated with the position
or location of the likelihood ratio $p\left(  \widehat{\Lambda}\left(
\mathbf{x}\right)  |\omega_{2}\right)  $ given class $\omega_{2}$:%
\[
E_{\min}\left(  Z|p\left(  \widehat{\Lambda}\left(  \mathbf{x}\right)
|\omega_{1}\right)  \right)  \rightleftharpoons E_{\min}\left(  Z|p\left(
\widehat{\Lambda}\left(  \mathbf{x}\right)  |\omega_{2}\right)  \right)
\text{,}%
\]
where $E_{\min}\left(  Z|p\left(  \widehat{\Lambda}\left(  \mathbf{x}\right)
|\omega_{1}\right)  \right)  \triangleq\left\Vert \sum\nolimits_{i=1}^{l_{1}%
}\psi_{1_{i}}k_{\mathbf{x}_{1_{i}}}\right\Vert ^{2}$ and $E_{\min}\left(
Z|p\left(  \widehat{\Lambda}\left(  \mathbf{x}\right)  |\omega_{2}\right)
\right)  \triangleq\left\Vert \sum\nolimits_{i=1}^{l_{2}}\psi_{2_{i}%
}k_{\mathbf{x}_{2_{i}}}\right\Vert ^{2}$, in such a manner that the expected
risk $\mathfrak{R}_{\mathfrak{\min}}\left(  Z|\widehat{\Lambda}\left(
\mathbf{x}\right)  \right)  $ and the corresponding eigenenergy $E_{\min
}\left(  Z|\widehat{\Lambda}\left(  \mathbf{x}\right)  \right)  $ of the
classification system $p\left(  \widehat{\Lambda}\left(  \mathbf{x}\right)
|\omega_{1}\right)  -p\left(  \widehat{\Lambda}\left(  \mathbf{x}\right)
|\omega_{2}\right)  \overset{\omega_{1}}{\underset{\omega_{2}}{\gtrless}}0$
are governed by the equilibrium point%
\[
p\left(  \widehat{\Lambda}\left(  \mathbf{x}\right)  |\omega_{1}\right)
-p\left(  \widehat{\Lambda}\left(  \mathbf{x}\right)  |\omega_{2}\right)  =0
\]
of the integral equation $f\left(  \widehat{\Lambda}\left(  \mathbf{x}\right)
\right)  $.

It follows that the classification system $p\left(  \widehat{\Lambda}\left(
\mathbf{x}\right)  |\omega_{1}\right)  -p\left(  \widehat{\Lambda}\left(
\mathbf{x}\right)  |\omega_{2}\right)  \overset{\omega_{1}}{\underset{\omega
_{2}}{\gtrless}}0$ is in statistical equilibrium:%
\begin{align*}
f\left(  \widehat{\Lambda}\left(  \mathbf{x}\right)  \right)   &  :\int%
_{Z_{1}}p\left(  \widehat{\Lambda}\left(  \mathbf{x}\right)  |\omega
_{1}\right)  d\widehat{\Lambda}-\int_{Z_{1}}p\left(  \widehat{\Lambda}\left(
\mathbf{x}\right)  |\omega_{2}\right)  d\widehat{\Lambda}\\
&  =\int_{Z_{2}}p\left(  \widehat{\Lambda}\left(  \mathbf{x}\right)
|\omega_{2}\right)  d\widehat{\Lambda}-\int_{Z_{2}}p\left(  \widehat{\Lambda
}\left(  \mathbf{x}\right)  |\omega_{1}\right)  d\widehat{\Lambda}\text{,}%
\end{align*}
where the forces associated with the counter risk $\overline{\mathfrak{R}%
}_{\mathfrak{\min}}\left(  Z_{1}|p\left(  \widehat{\Lambda}\left(
\mathbf{x}\right)  |\omega_{1}\right)  \right)  $ for class $\omega_{1}$ and
the risk $\mathfrak{R}_{\mathfrak{\min}}\left(  Z_{1}|p\left(
\widehat{\Lambda}\left(  \mathbf{x}\right)  |\omega_{2}\right)  \right)  $ for
class $\omega_{2}$ in the $Z_{1}$ decision region are balanced with the forces
associated with the counter risk $\overline{\mathfrak{R}}_{\mathfrak{\min}%
}\left(  Z_{2}|p\left(  \widehat{\Lambda}\left(  \mathbf{x}\right)
|\omega_{2}\right)  \right)  $ for class $\omega_{2}$ and the risk
$\mathfrak{R}_{\mathfrak{\min}}\left(  Z_{2}|p\left(  \widehat{\Lambda}\left(
\mathbf{x}\right)  |\omega_{1}\right)  \right)  $ for class $\omega_{1}$ in
the $Z_{2}$ decision region:%
\begin{align*}
f\left(  \widehat{\Lambda}\left(  \mathbf{x}\right)  \right)   &
:\overline{\mathfrak{R}}_{\mathfrak{\min}}\left(  Z_{1}|p\left(
\widehat{\Lambda}\left(  \mathbf{x}\right)  |\omega_{1}\right)  \right)
-\mathfrak{R}_{\mathfrak{\min}}\left(  Z_{1}|p\left(  \widehat{\Lambda}\left(
\mathbf{x}\right)  |\omega_{2}\right)  \right) \\
&  =\overline{\mathfrak{R}}_{\mathfrak{\min}}\left(  Z_{2}|p\left(
\widehat{\Lambda}\left(  \mathbf{x}\right)  |\omega_{2}\right)  \right)
-\mathfrak{R}_{\mathfrak{\min}}\left(  Z_{2}|p\left(  \widehat{\Lambda}\left(
\mathbf{x}\right)  |\omega_{1}\right)  \right)
\end{align*}
such that the expected risk $\mathfrak{R}_{\mathfrak{\min}}\left(
Z|\widehat{\Lambda}\left(  \mathbf{x}\right)  \right)  $ of the classification
system is minimized, and the eigenenergies associated with the counter risk
$\overline{\mathfrak{R}}_{\mathfrak{\min}}\left(  Z_{1}|p\left(
\widehat{\Lambda}\left(  \mathbf{x}\right)  |\omega_{1}\right)  \right)  $ for
class $\omega_{1}$ and the risk $\mathfrak{R}_{\mathfrak{\min}}\left(
Z_{1}|p\left(  \widehat{\Lambda}\left(  \mathbf{x}\right)  |\omega_{2}\right)
\right)  $ for class $\omega_{2}$ in the $Z_{1}$ decision region are balanced
with the eigenenergies associated with the counter risk $\overline
{\mathfrak{R}}_{\mathfrak{\min}}\left(  Z_{2}|p\left(  \widehat{\Lambda
}\left(  \mathbf{x}\right)  |\omega_{2}\right)  \right)  $ for class
$\omega_{2}$ and the risk $\mathfrak{R}_{\mathfrak{\min}}\left(
Z_{2}|p\left(  \widehat{\Lambda}\left(  \mathbf{x}\right)  |\omega_{1}\right)
\right)  $ for class $\omega_{1}$ in the $Z_{2}$ decision region:%
\begin{align*}
f\left(  \widehat{\Lambda}\left(  \mathbf{x}\right)  \right)   &  :E_{\min
}\left(  Z_{1}|p\left(  \widehat{\Lambda}\left(  \mathbf{x}\right)
|\omega_{1}\right)  \right)  -E_{\min}\left(  Z_{1}|p\left(  \widehat{\Lambda
}\left(  \mathbf{x}\right)  |\omega_{2}\right)  \right) \\
&  =E_{\min}\left(  Z_{2}|p\left(  \widehat{\Lambda}\left(  \mathbf{x}\right)
|\omega_{2}\right)  \right)  -E_{\min}\left(  Z_{2}|p\left(  \widehat{\Lambda
}\left(  \mathbf{x}\right)  |\omega_{1}\right)  \right)
\end{align*}
such that the eigenenergy $E_{\min}\left(  Z|\widehat{\Lambda}\left(
\mathbf{x}\right)  \right)  $ of the classification system is minimized.

Therefore, it is concluded that the expected risk $\mathfrak{R}%
_{\mathfrak{\min}}\left(  Z|\widehat{\Lambda}\left(  \mathbf{x}\right)
\right)  $ and the corresponding eigenenergy $E_{\min}\left(
Z|\widehat{\Lambda}\left(  \mathbf{x}\right)  \right)  $ of the classification
system $p\left(  \widehat{\Lambda}\left(  \mathbf{x}\right)  |\omega
_{1}\right)  -p\left(  \widehat{\Lambda}\left(  \mathbf{x}\right)  |\omega
_{2}\right)  \overset{\omega_{1}}{\underset{\omega_{2}}{\gtrless}}0$ are
governed by the equilibrium point%
\[
p\left(  \widehat{\Lambda}\left(  \mathbf{x}\right)  |\omega_{1}\right)
-p\left(  \widehat{\Lambda}\left(  \mathbf{x}\right)  |\omega_{2}\right)  =0
\]
of the integral equation%
\begin{align*}
f\left(  \widehat{\Lambda}\left(  \mathbf{x}\right)  \right)   &  =\int%
_{Z_{1}}p\left(  \widehat{\Lambda}\left(  \mathbf{x}\right)  |\omega
_{1}\right)  d\widehat{\Lambda}+\int_{Z_{2}}p\left(  \widehat{\Lambda}\left(
\mathbf{x}\right)  |\omega_{1}\right)  d\widehat{\Lambda}\\
&  =\int_{Z_{1}}p\left(  \widehat{\Lambda}\left(  \mathbf{x}\right)
|\omega_{2}\right)  d\widehat{\Lambda}+\int_{Z_{2}}p\left(  \widehat{\Lambda
}\left(  \mathbf{x}\right)  |\omega_{2}\right)  d\widehat{\Lambda}\text{,}%
\end{align*}
over the decision space $Z=Z_{1}+Z_{2}$, where the opposing forces and
influences of the classification system are balanced with each other, such
that the eigenenergy and the expected risk of the classification system are
minimized, and the classification system is in statistical equilibrium.

These results are readily generalized for any given class-conditional
probability density functions $p\left(  \mathbf{x}|\omega_{1}\right)  $ and
$p\left(  \mathbf{x}|\omega_{2}\right)  $ related to statistical distributions
of random vectors $\mathbf{x}$ that have constant or unchanging statistics.

\subsection*{Generalization of Proof}

Returning to Eq. (\ref{General Form of Decision Function II}), take the
general form of the decision function for a binary classification system%
\begin{align*}
\widehat{\Lambda}\left(  \mathbf{x}\right)   &  \triangleq\ln p\left(
\mathbf{x}|\omega_{1}\right)  -\ln p\left(  \mathbf{x}|\omega_{2}\right)
\overset{\omega_{1}}{\underset{\omega_{2}}{\gtrless}}0\\
&  =p\left(  \widehat{\Lambda}\left(  \mathbf{x}\right)  |\omega_{1}\right)
-p\left(  \widehat{\Lambda}\left(  \mathbf{x}\right)  |\omega_{2}\right)
\overset{\omega_{1}}{\underset{\omega_{2}}{\gtrless}}0\text{,}%
\end{align*}
where $\omega_{1}$ or $\omega_{2}$ is the true data category and $d$-component
random vectors $\mathbf{x}$ from class $\omega_{1}$ and class $\omega_{2}$ are
generated according to probability density functions $p\left(  \mathbf{x}%
|\omega_{1}\right)  $ and $p\left(  \mathbf{x}|\omega_{2}\right)  $ related to
statistical distributions of random vectors $\mathbf{x}$ that have constant or
unchanging statistics.

Now take any given decision boundary $D\left(  \mathbf{x}\right)  $:%
\[
D\left(  \mathbf{x}\right)  :p\left(  \widehat{\Lambda}\left(  \mathbf{x}%
\right)  |\omega_{1}\right)  -p\left(  \widehat{\Lambda}\left(  \mathbf{x}%
\right)  |\omega_{2}\right)  =0
\]
that is generated according to the likelihood ratio test:%
\[
\widehat{\Lambda}\left(  \mathbf{x}\right)  =p\left(  \widehat{\Lambda}\left(
\mathbf{x}\right)  |\omega_{1}\right)  -p\left(  \widehat{\Lambda}\left(
\mathbf{x}\right)  |\omega_{2}\right)  \overset{\omega_{1}}{\underset{\omega
_{2}}{\gtrless}}0\text{,}%
\]
where $d$-component random vectors $\mathbf{x}$ from class $\omega_{1}$ and
class $\omega_{2}$ are generated according to the probability density
functions $p\left(  \mathbf{x}|\omega_{1}\right)  $ and $p\left(
\mathbf{x}|\omega_{2}\right)  $ related to statistical distributions of random
vectors $\mathbf{x}$ that have constant or unchanging statistics.

It follows that the decision space $Z$ and the corresponding decision regions
$Z_{1}$ and $Z_{2}$ of the classification system $p\left(  \widehat{\Lambda
}\left(  \mathbf{x}\right)  |\omega_{1}\right)  -p\left(  \widehat{\Lambda
}\left(  \mathbf{x}\right)  |\omega_{2}\right)  \overset{\omega_{1}%
}{\underset{\omega_{2}}{\gtrless}}0$ are determined by either overlapping or
non-overlapping data distributions.

It also follows that the decision boundary $D\left(  \mathbf{x}\right)  $ and
the likelihood ratio $\widehat{\Lambda}\left(  \mathbf{x}\right)  =p\left(
\widehat{\Lambda}\left(  \mathbf{x}\right)  |\omega_{1}\right)  -p\left(
\widehat{\Lambda}\left(  \mathbf{x}\right)  |\omega_{2}\right)  $ satisfy the
vector equation:%
\[
p\left(  \widehat{\Lambda}\left(  \mathbf{x}\right)  |\omega_{1}\right)
-p\left(  \widehat{\Lambda}\left(  \mathbf{x}\right)  |\omega_{2}\right)
=0\text{,}%
\]
where the likelihood ratio $p\left(  \widehat{\Lambda}\left(  \mathbf{x}%
\right)  |\omega_{1}\right)  $ given class $\omega_{1}$ is given by a vector
expression $p\left(  \widehat{\Lambda}\left(  \mathbf{x}\right)  |\omega
_{1}\right)  $ of the class-conditional probability density function $p\left(
\mathbf{x}|\omega_{1}\right)  $, and the likelihood ratio $p\left(
\widehat{\Lambda}\left(  \mathbf{x}\right)  |\omega_{2}\right)  $ given class
$\omega_{2}$ is given by a vector expression $p\left(  \widehat{\Lambda
}\left(  \mathbf{x}\right)  |\omega_{2}\right)  $ of the class-conditional
probability density function $p\left(  \mathbf{x}|\omega_{2}\right)  $.

Therefore, the loci of the decision boundary $D\left(  \mathbf{x}\right)  $
and the likelihood ratio $\widehat{\Lambda}\left(  \mathbf{x}\right)  $ are
determined by a locus of principal eigenaxis components and likelihoods:%
\begin{align*}
\widehat{\Lambda}\left(  \mathbf{x}\right)   &  =p\left(  \widehat{\Lambda
}\left(  \mathbf{x}\right)  |\omega_{1}\right)  -p\left(  \widehat{\Lambda
}\left(  \mathbf{x}\right)  |\omega_{2}\right) \\
&  \triangleq\sum\nolimits_{i=1}^{l_{1}}\psi_{1_{i}}k_{\mathbf{x}_{1_{i}}%
}-\sum\nolimits_{i=1}^{l_{2}}\psi_{2_{i}}k_{\mathbf{x}_{2_{i}}}\text{,}%
\end{align*}
where $k_{\mathbf{x}_{1_{i}}}$ and $k_{\mathbf{x}_{2_{i}}}$ are reproducing
kernels for respective data points $\mathbf{x}_{1_{i}}$ and $\mathbf{x}%
_{2_{i}}$: the reproducing kernel $K(\mathbf{x,s})=k_{\mathbf{s}}(\mathbf{x})$
is either $k_{\mathbf{s}}(\mathbf{x})\triangleq\left(  \mathbf{x}%
^{T}\mathbf{s}+1\right)  ^{2}$ or $k_{\mathbf{s}}(\mathbf{x})\triangleq
\exp\left(  -0.01\left\Vert \mathbf{x}-\mathbf{s}\right\Vert ^{2}\right)  $,
$\mathbf{x}_{1i}\sim p\left(  \mathbf{x}|\omega_{1}\right)  $, $\mathbf{x}%
_{2i}\sim p\left(  \mathbf{x}|\omega_{2}\right)  $, and $\psi_{1i}$ and
$\psi_{2i}$ are scale factors that provide measures of likelihood for
respective data points $\mathbf{x}_{1i}$ and $\mathbf{x}_{2i}$ which lie in
either overlapping regions or tails regions of data distributions related to
$p\left(  \mathbf{x}|\omega_{1}\right)  $ and $p\left(  \mathbf{x}|\omega
_{2}\right)  $, that is in statistical equilibrium:%
\[
p\left(  \widehat{\Lambda}\left(  \mathbf{x}\right)  |\omega_{1}\right)
\rightleftharpoons p\left(  \widehat{\Lambda}\left(  \mathbf{x}\right)
|\omega_{2}\right)
\]
such that the locus of principal eigenaxis components and likelihoods
$\widehat{\Lambda}\left(  \mathbf{x}\right)  =\sum\nolimits_{i=1}^{l_{1}}%
\psi_{1_{i}}k_{\mathbf{x}_{1_{i}}}-\sum\nolimits_{i=1}^{l_{2}}\psi_{2_{i}%
}k_{\mathbf{x}_{2_{i}}}$ satisfies the equilibrium equation:%
\[
\sum\nolimits_{i=1}^{l_{1}}\psi_{1_{i}}k_{\mathbf{x}_{1_{i}}}%
\rightleftharpoons\sum\nolimits_{i=1}^{l_{2}}\psi_{2_{i}}k_{\mathbf{x}_{2_{i}%
}}%
\]
in an optimal manner.

Thus, the discriminant function $\widehat{\Lambda}\left(  \mathbf{x}\right)
=\sum\nolimits_{i=1}^{l_{1}}\psi_{1_{i}}k_{\mathbf{x}_{1_{i}}}-\sum
\nolimits_{i=1}^{l_{2}}\psi_{2_{i}}k_{\mathbf{x}_{2_{i}}}$ is the solution to
the integral equation%
\begin{align*}
f\left(  \widehat{\Lambda}\left(  \mathbf{x}\right)  \right)   &  =\int%
_{Z_{1}}p\left(  \widehat{\Lambda}\left(  \mathbf{x}\right)  |\omega
_{1}\right)  d\widehat{\Lambda}+\int_{Z_{2}}p\left(  \widehat{\Lambda}\left(
\mathbf{x}\right)  |\omega_{1}\right)  d\widehat{\Lambda}\\
&  =\int_{Z_{1}}p\left(  \widehat{\Lambda}\left(  \mathbf{x}\right)
|\omega_{2}\right)  d\widehat{\Lambda}+\int_{Z_{2}}p\left(  \widehat{\Lambda
}\left(  \mathbf{x}\right)  |\omega_{2}\right)  d\widehat{\Lambda}\text{,}%
\end{align*}
over the decision space $Z=Z_{1}+Z_{2}$, where the equilibrium point $p\left(
\widehat{\Lambda}\left(  \mathbf{x}\right)  |\omega_{1}\right)  -p\left(
\widehat{\Lambda}\left(  \mathbf{x}\right)  |\omega_{2}\right)  =0$:%
\[
\sum\nolimits_{i=1}^{l_{1}}\psi_{1_{i}}k_{\mathbf{x}_{1_{i}}}-\sum
\nolimits_{i=1}^{l_{2}}\psi_{2_{i}}k_{\mathbf{x}_{2_{i}}}=0
\]
is the focus of a decision boundary $D\left(  \mathbf{x}\right)  $, and the
forces associated with the counter risk $\overline{\mathfrak{R}}%
_{\mathfrak{\min}}\left(  Z_{1}|p\left(  \widehat{\Lambda}\left(
\mathbf{x}\right)  |\omega_{1}\right)  \right)  $ for class $\omega_{1}$ in
the $Z_{1}$ decision region and the risk $\mathfrak{R}_{\mathfrak{\min}%
}\left(  Z_{2}|p\left(  \widehat{\Lambda}\left(  \mathbf{x}\right)
|\omega_{1}\right)  \right)  $ for class $\omega_{1}$ in the $Z_{2}$ decision
region are balanced with the forces associated with the risk $\mathfrak{R}%
_{\mathfrak{\min}}\left(  Z_{1}|p\left(  \widehat{\Lambda}\left(
\mathbf{x}\right)  |\omega_{2}\right)  \right)  $ for class $\omega_{2}$ in
the $Z_{1}$ decision region and the counter risk $\overline{\mathfrak{R}%
}_{\mathfrak{\min}}\left(  Z_{2}|p\left(  \widehat{\Lambda}\left(
\mathbf{x}\right)  |\omega_{2}\right)  \right)  $ for class $\omega_{2}$ in
the $Z_{2}$ decision region:%
\begin{align*}
f\left(  \widehat{\Lambda}\left(  \mathbf{x}\right)  \right)   &
:\overline{\mathfrak{R}}_{\mathfrak{\min}}\left(  Z_{1}|p\left(
\widehat{\Lambda}\left(  \mathbf{x}\right)  |\omega_{1}\right)  \right)
+\mathfrak{R}_{\mathfrak{\min}}\left(  Z_{2}|p\left(  \widehat{\Lambda}\left(
\mathbf{x}\right)  |\omega_{1}\right)  \right) \\
&  =\mathfrak{R}_{\mathfrak{\min}}\left(  Z_{1}|p\left(  \widehat{\Lambda
}\left(  \mathbf{x}\right)  |\omega_{2}\right)  \right)  +\overline
{\mathfrak{R}}_{\mathfrak{\min}}\left(  Z_{2}|p\left(  \widehat{\Lambda
}\left(  \mathbf{x}\right)  |\omega_{2}\right)  \right)  \text{,}%
\end{align*}
and the eigenenergy $E_{\min}\left(  Z|p\left(  \widehat{\Lambda}\left(
\mathbf{x}\right)  |\omega_{1}\right)  \right)  $ associated with the position
or location of the likelihood ratio $p\left(  \widehat{\Lambda}\left(
\mathbf{x}\right)  |\omega_{1}\right)  $ given class $\omega_{1}$ is balanced
with the eigenenergy $E_{\min}\left(  Z|p\left(  \widehat{\Lambda}\left(
\mathbf{x}\right)  |\omega_{2}\right)  \right)  $ associated with the position
or location of the likelihood ratio $p\left(  \widehat{\Lambda}\left(
\mathbf{x}\right)  |\omega_{2}\right)  $ given class $\omega_{2}$:
\[
E_{\min}\left(  Z|p\left(  \widehat{\Lambda}\left(  \mathbf{x}\right)
|\omega_{1}\right)  \right)  \rightleftharpoons E_{\min}\left(  Z|p\left(
\widehat{\Lambda}\left(  \mathbf{x}\right)  |\omega_{2}\right)  \right)
\text{,}%
\]
where $E_{\min}\left(  Z|p\left(  \widehat{\Lambda}\left(  \mathbf{x}\right)
|\omega_{1}\right)  \right)  \triangleq\left\Vert \sum\nolimits_{i=1}^{l_{1}%
}\psi_{1_{i}}k_{\mathbf{x}_{1_{i}}}\right\Vert ^{2}$ and $E_{\min}\left(
Z|p\left(  \widehat{\Lambda}\left(  \mathbf{x}\right)  |\omega_{2}\right)
\right)  \triangleq\left\Vert \sum\nolimits_{i=1}^{l_{2}}\psi_{2_{i}%
}k_{\mathbf{x}_{2_{i}}}\right\Vert ^{2}$, in such a manner that the risk
$\mathfrak{R}_{\mathfrak{\min}}\left(  Z|\widehat{\Lambda}\left(
\mathbf{x}\right)  \right)  $ and the corresponding eigenenergy $E_{\min
}\left(  Z|\widehat{\Lambda}\left(  \mathbf{x}\right)  \right)  $ of the
classification system $p\left(  \widehat{\Lambda}\left(  \mathbf{x}\right)
|\omega_{1}\right)  -p\left(  \widehat{\Lambda}\left(  \mathbf{x}\right)
|\omega_{2}\right)  \overset{\omega_{1}}{\underset{\omega_{2}}{\gtrless}}0$
are governed by the equilibrium point%
\[
p\left(  \widehat{\Lambda}\left(  \mathbf{x}\right)  |\omega_{1}\right)
-p\left(  \widehat{\Lambda}\left(  \mathbf{x}\right)  |\omega_{2}\right)  =0
\]
of the integral equation $f\left(  \widehat{\Lambda}\left(  \mathbf{x}\right)
\right)  $.

It follows that the classification system $p\left(  \widehat{\Lambda}\left(
\mathbf{x}\right)  |\omega_{1}\right)  -p\left(  \widehat{\Lambda}\left(
\mathbf{x}\right)  |\omega_{2}\right)  \overset{\omega_{1}}{\underset{\omega
_{2}}{\gtrless}}0$ is in statistical equilibrium:%
\begin{align*}
f\left(  \widehat{\Lambda}\left(  \mathbf{x}\right)  \right)   &  :\int%
_{Z_{1}}p\left(  \widehat{\Lambda}\left(  \mathbf{x}\right)  |\omega
_{1}\right)  d\widehat{\Lambda}-\int_{Z_{1}}p\left(  \widehat{\Lambda}\left(
\mathbf{x}\right)  |\omega_{2}\right)  d\widehat{\Lambda}\\
&  =\int_{Z_{2}}p\left(  \widehat{\Lambda}\left(  \mathbf{x}\right)
|\omega_{2}\right)  d\widehat{\Lambda}-\int_{Z_{2}}p\left(  \widehat{\Lambda
}\left(  \mathbf{x}\right)  |\omega_{1}\right)  d\widehat{\Lambda}\text{,}%
\end{align*}
where the forces associated with the counter risk $\overline{\mathfrak{R}%
}_{\mathfrak{\min}}\left(  Z_{1}|p\left(  \widehat{\Lambda}\left(
\mathbf{x}\right)  |\omega_{1}\right)  \right)  $ for class $\omega_{1}$ and
the risk $\mathfrak{R}_{\mathfrak{\min}}\left(  Z_{1}|p\left(
\widehat{\Lambda}\left(  \mathbf{x}\right)  |\omega_{2}\right)  \right)  $ for
class $\omega_{2}$ in the $Z_{1}$ decision region are balanced with the forces
associated with the counter risk $\overline{\mathfrak{R}}_{\mathfrak{\min}%
}\left(  Z_{2}|p\left(  \widehat{\Lambda}\left(  \mathbf{x}\right)
|\omega_{2}\right)  \right)  $ for class $\omega_{2}$ and the risk
$\mathfrak{R}_{\mathfrak{\min}}\left(  Z_{2}|p\left(  \widehat{\Lambda}\left(
\mathbf{x}\right)  |\omega_{1}\right)  \right)  $ for class $\omega_{1}$ in
the $Z_{2}$ decision region:%
\begin{align*}
f\left(  \widehat{\Lambda}\left(  \mathbf{x}\right)  \right)   &
:\overline{\mathfrak{R}}_{\mathfrak{\min}}\left(  Z_{1}|p\left(
\widehat{\Lambda}\left(  \mathbf{x}\right)  |\omega_{1}\right)  \right)
-\mathfrak{R}_{\mathfrak{\min}}\left(  Z_{1}|p\left(  \widehat{\Lambda}\left(
\mathbf{x}\right)  |\omega_{2}\right)  \right) \\
&  =\overline{\mathfrak{R}}_{\mathfrak{\min}}\left(  Z_{2}|p\left(
\widehat{\Lambda}\left(  \mathbf{x}\right)  |\omega_{2}\right)  \right)
-\mathfrak{R}_{\mathfrak{\min}}\left(  Z_{2}|p\left(  \widehat{\Lambda}\left(
\mathbf{x}\right)  |\omega_{1}\right)  \right)
\end{align*}
such that the expected risk $\mathfrak{R}_{\mathfrak{\min}}\left(
Z|\widehat{\Lambda}\left(  \mathbf{x}\right)  \right)  $ of the classification
system is minimized, and the eigenenergies associated with the counter risk
$\overline{\mathfrak{R}}_{\mathfrak{\min}}\left(  Z_{1}|p\left(
\widehat{\Lambda}\left(  \mathbf{x}\right)  |\omega_{1}\right)  \right)  $ for
class $\omega_{1}$ and the risk $\mathfrak{R}_{\mathfrak{\min}}\left(
Z_{1}|p\left(  \widehat{\Lambda}\left(  \mathbf{x}\right)  |\omega_{2}\right)
\right)  $ for class $\omega_{2}$ in the $Z_{1}$ decision region are balanced
with the eigenenergies associated with the counter risk $\overline
{\mathfrak{R}}_{\mathfrak{\min}}\left(  Z_{2}|p\left(  \widehat{\Lambda
}\left(  \mathbf{x}\right)  |\omega_{2}\right)  \right)  $ for class
$\omega_{2}$ and the risk $\mathfrak{R}_{\mathfrak{\min}}\left(
Z_{2}|p\left(  \widehat{\Lambda}\left(  \mathbf{x}\right)  |\omega_{1}\right)
\right)  $ for class $\omega_{1}$ in the $Z_{2}$ decision region:%
\begin{align*}
f\left(  \widehat{\Lambda}\left(  \mathbf{x}\right)  \right)   &  :E_{\min
}\left(  Z_{1}|p\left(  \widehat{\Lambda}\left(  \mathbf{x}\right)
|\omega_{1}\right)  \right)  -E_{\min}\left(  Z_{1}|p\left(  \widehat{\Lambda
}\left(  \mathbf{x}\right)  |\omega_{2}\right)  \right) \\
&  =E_{\min}\left(  Z_{2}|p\left(  \widehat{\Lambda}\left(  \mathbf{x}\right)
|\omega_{2}\right)  \right)  -E_{\min}\left(  Z_{2}|p\left(  \widehat{\Lambda
}\left(  \mathbf{x}\right)  |\omega_{1}\right)  \right)
\end{align*}
such that the eigenenergy $E_{\min}\left(  Z|\widehat{\Lambda}\left(
\mathbf{x}\right)  \right)  $ of the classification system is minimized.

Therefore, it is concluded that the risk $\mathfrak{R}_{\mathfrak{\min}%
}\left(  Z|\widehat{\Lambda}\left(  \mathbf{x}\right)  \right)  $ and the
corresponding eigenenergy $E_{\min}\left(  Z|\widehat{\Lambda}\left(
\mathbf{x}\right)  \right)  $ of the classification system $p\left(
\widehat{\Lambda}\left(  \mathbf{x}\right)  |\omega_{1}\right)  -p\left(
\widehat{\Lambda}\left(  \mathbf{x}\right)  |\omega_{2}\right)
\overset{\omega_{1}}{\underset{\omega_{2}}{\gtrless}}0$ are governed by the
equilibrium point%
\[
p\left(  \widehat{\Lambda}\left(  \mathbf{x}\right)  |\omega_{1}\right)
-p\left(  \widehat{\Lambda}\left(  \mathbf{x}\right)  |\omega_{2}\right)  =0
\]
of the integral equation%
\begin{align*}
f\left(  \widehat{\Lambda}\left(  \mathbf{x}\right)  \right)   &  =\int%
_{Z_{1}}p\left(  \widehat{\Lambda}\left(  \mathbf{x}\right)  |\omega
_{1}\right)  d\widehat{\Lambda}+\int_{Z_{2}}p\left(  \widehat{\Lambda}\left(
\mathbf{x}\right)  |\omega_{1}\right)  d\widehat{\Lambda}\\
&  =\int_{Z_{1}}p\left(  \widehat{\Lambda}\left(  \mathbf{x}\right)
|\omega_{2}\right)  d\widehat{\Lambda}+\int_{Z_{2}}p\left(  \widehat{\Lambda
}\left(  \mathbf{x}\right)  |\omega_{2}\right)  d\widehat{\Lambda}\text{,}%
\end{align*}
over the decision space $Z=Z_{1}+Z_{2}$, where the opposing forces and
influences of the classification system are balanced with each other, such
that the eigenenergy and the expected risk of the classification system are
minimized, and the classification system is in statistical equilibrium.

I\ will now show that the eigenenergy of classification systems is conserved
and remains relatively constant, so that the eigenenergy and the corresponding
expected risk of a binary classification system cannot be created or
destroyed, but only transferred from one classification system to another.

\section*{Law of Conservation of Eigenenergy:}

\subsection*{For Binary Classification Systems}

Let $\widehat{\Lambda}\left(  \mathbf{x}\right)  =p\left(  \widehat{\Lambda
}\left(  \mathbf{x}\right)  |\omega_{1}\right)  -p\left(  \widehat{\Lambda
}\left(  \mathbf{x}\right)  |\omega_{2}\right)  \overset{\omega_{1}%
}{\underset{\omega_{2}}{\gtrless}}0$ denote the likelihood ratio test for a
binary classification system, where $\omega_{1}$ or $\omega_{2}$ is the true
data category and $d$-component random vectors $\mathbf{x}$ from class
$\omega_{1}$ and class $\omega_{2}$ are generated according to probability
density functions $p\left(  \mathbf{x}|\omega_{1}\right)  $ and $p\left(
\mathbf{x}|\omega_{2}\right)  $ related to statistical distributions of random
vectors $\mathbf{x}$ that have constant or unchanging statistics.

The expected risk $\mathfrak{R}_{\mathfrak{\min}}\left(  Z|\widehat{\Lambda
}\left(  \mathbf{x}\right)  \right)  $ and the corresponding eigenenergy
$E_{\min}\left(  Z|\widehat{\Lambda}\left(  \mathbf{x}\right)  \right)  $ of a
binary classification system $p\left(  \widehat{\Lambda}\left(  \mathbf{x}%
\right)  |\omega_{1}\right)  -p\left(  \widehat{\Lambda}\left(  \mathbf{x}%
\right)  |\omega_{2}\right)  \overset{\omega_{1}}{\underset{\omega
_{2}}{\gtrless}}0$ are governed by the equilibrium point%
\[
p\left(  \widehat{\Lambda}\left(  \mathbf{x}\right)  |\omega_{1}\right)
-p\left(  \widehat{\Lambda}\left(  \mathbf{x}\right)  |\omega_{2}\right)  =0
\]
of the integral equation%
\begin{align*}
f\left(  \widehat{\Lambda}\left(  \mathbf{x}\right)  \right)   &  =\int%
_{Z_{1}}p\left(  \widehat{\Lambda}\left(  \mathbf{x}\right)  |\omega
_{1}\right)  d\widehat{\Lambda}+\int_{Z_{2}}p\left(  \widehat{\Lambda}\left(
\mathbf{x}\right)  |\omega_{1}\right)  d\widehat{\Lambda}\\
&  =\int_{Z_{1}}p\left(  \widehat{\Lambda}\left(  \mathbf{x}\right)
|\omega_{2}\right)  d\widehat{\Lambda}+\int_{Z_{2}}p\left(  \widehat{\Lambda
}\left(  \mathbf{x}\right)  |\omega_{2}\right)  d\widehat{\Lambda}\text{,}%
\end{align*}
over the decision space $Z=Z_{1}+Z_{2}$, where the forces associated with the
counter risk $\overline{\mathfrak{R}}_{\mathfrak{\min}}\left(  Z_{1}|p\left(
\widehat{\Lambda}\left(  \mathbf{x}\right)  |\omega_{1}\right)  \right)  $ and
the risk $\mathfrak{R}_{\mathfrak{\min}}\left(  Z_{2}|p\left(
\widehat{\Lambda}\left(  \mathbf{x}\right)  |\omega_{1}\right)  \right)  $ for
class $\omega_{1}$ in the $Z_{1}$ and $Z_{2}$ decision regions are equal to
the forces associated with the risk $\mathfrak{R}_{\mathfrak{\min}}\left(
Z_{1}|p\left(  \widehat{\Lambda}\left(  \mathbf{x}\right)  |\omega_{2}\right)
\right)  $ and the counter risk $\overline{\mathfrak{R}}_{\mathfrak{\min}%
}\left(  Z_{2}|p\left(  \widehat{\Lambda}\left(  \mathbf{x}\right)
|\omega_{2}\right)  \right)  $ for class $\omega_{2}$ in the $Z_{1}$ and
$Z_{2}$ decision regions, and the eigenenergy $E_{\min}\left(  Z|p\left(
\widehat{\Lambda}\left(  \mathbf{x}\right)  |\omega_{1}\right)  \right)  $
associated with the position or location of the likelihood ratio $p\left(
\widehat{\Lambda}\left(  \mathbf{x}\right)  |\omega_{1}\right)  $ given class
$\omega_{1}$ is equal to the eigenenergy $E_{\min}\left(  Z|p\left(
\widehat{\Lambda}\left(  \mathbf{x}\right)  |\omega_{2}\right)  \right)  $
associated with the position or location of the likelihood ratio $p\left(
\widehat{\Lambda}\left(  \mathbf{x}\right)  |\omega_{2}\right)  $ given class
$\omega_{2}$.

The eigenenergy $E_{\min}\left(  Z|\widehat{\Lambda}\left(  \mathbf{x}\right)
\right)  $ is the state of a binary classification system $p\left(
\widehat{\Lambda}\left(  \mathbf{x}\right)  |\omega_{1}\right)  -p\left(
\widehat{\Lambda}\left(  \mathbf{x}\right)  |\omega_{2}\right)
\overset{\omega_{1}}{\underset{\omega_{2}}{\gtrless}}0$ that is associated
with the position or location of a likelihood ratio in statistical
equilibrium: $p\left(  \widehat{\Lambda}\left(  \mathbf{x}\right)  |\omega
_{1}\right)  -p\left(  \widehat{\Lambda}\left(  \mathbf{x}\right)  |\omega
_{2}\right)  =0$ and the locus of a corresponding decision boundary: $D\left(
\mathbf{x}\right)  :$ $p\left(  \widehat{\Lambda}\left(  \mathbf{x}\right)
|\omega_{1}\right)  -p\left(  \widehat{\Lambda}\left(  \mathbf{x}\right)
|\omega_{2}\right)  =0$.

Thus, any given binary classification system $p\left(  \widehat{\Lambda
}\left(  \mathbf{x}\right)  |\omega_{1}\right)  -p\left(  \widehat{\Lambda
}\left(  \mathbf{x}\right)  |\omega_{2}\right)  \overset{\omega_{1}%
}{\underset{\omega_{2}}{\gtrless}}0$ exhibits an error rate that is consistent
with the expected risk $\mathfrak{R}_{\mathfrak{\min}}\left(
Z|\widehat{\Lambda}\left(  \mathbf{x}\right)  \right)  $ and the corresponding
eigenenergy $E_{\min}\left(  Z|\widehat{\Lambda}\left(  \mathbf{x}\right)
\right)  $ of the classification system: for all random vectors $\mathbf{x}$
that are generated according to $p\left(  \mathbf{x}|\omega_{1}\right)  $ and
$p\left(  \mathbf{x}|\omega_{2}\right)  $, where $p\left(  \mathbf{x}%
|\omega_{1}\right)  $ and $p\left(  \mathbf{x}|\omega_{2}\right)  $ are
related to statistical distributions of random vectors $\mathbf{x}$ that have
constant or unchanging statistics.

The total eigenenergy of a binary classification system $p\left(
\widehat{\Lambda}\left(  \mathbf{x}\right)  |\omega_{1}\right)  -p\left(
\widehat{\Lambda}\left(  \mathbf{x}\right)  |\omega_{2}\right)
\overset{\omega_{1}}{\underset{\omega_{2}}{\gtrless}}0$ is found by adding up
contributions from characteristics of the classification system:

The eigenenergies $E_{\min}\left(  Z|p\left(  \widehat{\Lambda}\left(
\mathbf{x}\right)  |\omega_{1}\right)  \right)  $ and $E_{\min}\left(
Z|p\left(  \widehat{\Lambda}\left(  \mathbf{x}\right)  |\omega_{2}\right)
\right)  $ associated with the positions or locations of the class-conditional
likelihood ratios $p\left(  \widehat{\Lambda}\left(  \mathbf{x}\right)
|\omega_{1}\right)  $ and $p\left(  \widehat{\Lambda}\left(  \mathbf{x}%
\right)  |\omega_{2}\right)  $, where $E_{\min}\left(  Z|p\left(
\widehat{\Lambda}\left(  \mathbf{x}\right)  |\omega_{1}\right)  \right)  $ and
$E_{\min}\left(  Z|p\left(  \widehat{\Lambda}\left(  \mathbf{x}\right)
|\omega_{2}\right)  \right)  $ are related to eigenenergies associated with
positions and potential locations of extreme points that lie in either
overlapping regions or tails regions of statistical distributions related to
the class-conditional probability density functions $p\left(  \mathbf{x}%
|\omega_{1}\right)  $ and $p\left(  \mathbf{x}|\omega_{2}\right)  $.

Any given binary classification system that is determined by a likelihood
ratio test:%
\[
p\left(  \widehat{\Lambda}\left(  \mathbf{x}\right)  |\omega_{1}\right)
-p\left(  \widehat{\Lambda}\left(  \mathbf{x}\right)  |\omega_{2}\right)
\overset{\omega_{1}}{\underset{\omega_{2}}{\gtrless}}0\text{,}%
\]
where the probability density functions $p\left(  \mathbf{x}|\omega
_{1}\right)  $ and $p\left(  \mathbf{x}|\omega_{2}\right)  $ are related to
statistical distributions of random vectors $\mathbf{x}$ that have constant or
unchanging statistics, and the locus of a decision boundary:%
\[
D\left(  \mathbf{x}\right)  :p\left(  \widehat{\Lambda}\left(  \mathbf{x}%
\right)  |\omega_{1}\right)  -p\left(  \widehat{\Lambda}\left(  \mathbf{x}%
\right)  |\omega_{2}\right)  =0
\]
is governed by the locus of a likelihood ratio $p\left(  \widehat{\Lambda
}\left(  \mathbf{x}\right)  |\omega_{1}\right)  -p\left(  \widehat{\Lambda
}\left(  \mathbf{x}\right)  |\omega_{2}\right)  $ in statistical equilibrium:%
\[
p\left(  \widehat{\Lambda}\left(  \mathbf{x}\right)  |\omega_{1}\right)
\rightleftharpoons p\left(  \widehat{\Lambda}\left(  \mathbf{x}\right)
|\omega_{2}\right)  \text{,}%
\]
is a closed classification system.

Thus, the total eigenenergy $E_{\min}\left(  Z|\widehat{\Lambda}\left(
\mathbf{x}\right)  \right)  $ of any given binary classification system
$p\left(  \widehat{\Lambda}\left(  \mathbf{x}\right)  |\omega_{1}\right)
-p\left(  \widehat{\Lambda}\left(  \mathbf{x}\right)  |\omega_{2}\right)
\overset{\omega_{1}}{\underset{\omega_{2}}{\gtrless}}0$ is conserved and
remains relatively constant.

Therefore, the eigenenergy $E_{\min}\left(  Z|\widehat{\Lambda}\left(
\mathbf{x}\right)  \right)  $ of a binary classification system $p\left(
\widehat{\Lambda}\left(  \mathbf{x}\right)  |\omega_{1}\right)  -p\left(
\widehat{\Lambda}\left(  \mathbf{x}\right)  |\omega_{2}\right)
\overset{\omega_{1}}{\underset{\omega_{2}}{\gtrless}}0$ cannot be created or
destroyed, but only transferred from one classification system to another.

It follows that the corresponding expected risk $\mathfrak{R}_{\mathfrak{\min
}}\left(  Z|\widehat{\Lambda}\left(  \mathbf{x}\right)  \right)  $ of a binary
classification system $p\left(  \widehat{\Lambda}\left(  \mathbf{x}\right)
|\omega_{1}\right)  -p\left(  \widehat{\Lambda}\left(  \mathbf{x}\right)
|\omega_{2}\right)  \overset{\omega_{1}}{\underset{\omega_{2}}{\gtrless}}0$
cannot be created or destroyed, but only transferred from one classification
system to another.

I\ will now devise a system of data-driven, locus equations that generate
computer-implemented, optimal linear classification systems. My discoveries
are based on useful relations between geometric locus methods, geometric
methods in Hilbert spaces, statistics, and the binary classification theorem
that I\ have derived.

\section{Optimal Linear Classification Systems}

I will begin by outlining my design for computer-implemented optimal linear
classification systems. Such computer-implemented systems are scalable modules
for optimal statistical pattern recognition systems, all of which are capable
of performing a wide variety of statistical pattern recognition tasks, where
any given $M$-class statistical pattern recognition system exhibits optimal
generalization performance for an $M$-class feature space.

\subsection*{Problem Formulation}

\begin{flushleft}
The formulation of a system of data-driven, locus equations that generates
computer-implemented, optimal linear classification systems requires solving
three fundamental problems:
\end{flushleft}

\paragraph{Problem $\mathbf{1}$}

\textit{Define the geometric figures in a linear classification system, where
geometric figures involve points, vectors, lines, line segments, angles,
regions, planes, and hyperplanes.}

\paragraph{Problem $\mathbf{2}$}

\textit{Define the geometric and statistical properties exhibited by each of
the geometric figures.}

\paragraph{Problem $\mathbf{3}$}

\textit{Define the forms of the data-driven, locus equations that determine
the geometric figures.}

\subsection*{The Solution}

\begin{flushleft}
I\ have formulated a solution that answers all three problems. My solution is
based on three ideas:
\end{flushleft}

\paragraph{Idea $\mathbf{1}$}

Devise \emph{a dual locus of data points} that determines optimal linear
decision boundaries \emph{and} likelihood ratios that achieve the lowest
possible error rate.

\paragraph{Idea $\mathbf{2}$}

The dual locus of data points must \emph{determine} the \emph{coordinate
system} of the \emph{linear decision boundary}.

\paragraph{Idea $\mathbf{3}$}

The dual locus of data points \emph{must satisfy discrete versions of the
fundamental equations of binary classification for a classification system in
statistical equilibrium.}

\subsection*{Key Elements of the Solution}

\begin{flushleft}
The essential elements of the solution are outlined below.
\end{flushleft}

\paragraph{Locus of Principal Eigenaxis Components}

Returning to Eqs (\ref{Normal Eigenaxis Functional}) and
(\ref{Normal Form Normal Eigenaxis}), given that the vector components of a
principal eigenaxis specify all forms of linear curves and surfaces, and all
of the points on any given line, plane, or hyperplane explicitly and
exclusively reference the principal eigenaxis of the linear locus, it follows
that the principal eigenaxis of a linear decision boundary provides an
elegant, statistical eigen-coordinate system for a linear classification system.

Therefore, the dual locus of data points \emph{must} be a \emph{locus of
principal eigenaxis components}.

\paragraph{Critical Minimum Eigenenergy Constraint}

Given Eq. (\ref{Characteristic Eigenenergy}), it follows that the principal
eigenaxis of a linear decision boundary satisfies the linear decision boundary
in terms of its eigenenergy. Accordingly, the principal eigenaxis of a linear
decision boundary exhibits a characteristic eigenenergy that is unique for the
linear decision boundary. Thereby, the\ \emph{important generalizations} for a
linear decision boundary are \emph{determined by} the \emph{eigenenergy}
exhibited by its \emph{principal eigenaxis}.

Therefore, the locus of principal eigenaxis components \emph{must satisfy} a
critical minimum, i.e., a total allowed, \emph{eigenenergy constraint} such
that the locus of principal eigenaxis components satisfies a \emph{linear
decision boundary} in \emph{terms of} its critical minimum or total allowed
\emph{eigenenergies.} Thus, the locus of principal eigenaxis components must
satisfy \emph{discrete and data-driven versions} of the vector and equilibrium
equation in Eqs
(\ref{Vector Equation of Likelihood Ratio and Decision Boundary}) and
(\ref{Equilibrium Equation of Likelihood Ratio and Decision Boundary}), the
integral equation in Eq.
(\ref{Integral Equation of Likelihood Ratio and Decision Boundary}), the
fundamental integral equation of binary classification in Eq.
(\ref{Equalizer Rule}), and the corresponding integral equation in Eq.
(\ref{Balancing of Bayes' Risks and Counteracting Risks}) \emph{in terms of
its total allowed eigenenergies}.

\paragraph{Extreme Points}

In order for the locus of principal eigenaxis components to implement a
likelihood ratio, the locus of principal eigenaxis components \emph{must} be
\emph{formed by data points} that lie in either overlapping regions or tail
region of data distributions, thereby determining \emph{decision regions}
based on forces associated with \emph{risks and counter risks}:\emph{ }which
are related to positions and potential locations of data points that lie in
either overlapping regions or tail region of data distributions, where an
unknown portion of the data points are the \emph{source of decision error}.
Data points that lie in either overlapping regions or tail region of data
distributions will be called extreme points.

\paragraph{Parameter Vector of Class-conditional Densities}

Given that the locus of principal eigenaxis components must determine a
likelihood ratio, it follows that the locus of principal eigenaxis components
\emph{must also} be a \emph{parameter vector} that \emph{provides an estimate
of class-conditional density functions}. Given Eq.
(\emph{\ref{Equilibrium Equation of Likelihood Ratio and Decision Boundary}}),
it also follows that the parameter vector must be in statistical equilibrium.

\paragraph{Minimum Conditional Risk Constraint}

Given Eqs (\ref{Equalizer Rule}) and
(\ref{Balancing of Bayes' Risks and Counteracting Risks}), it follows that the
parameter vector must satisfy a discrete and data-driven version of the
fundamental integral equation of binary classification in Eq.
(\ref{Equalizer Rule}) and the corresponding integral equation in
(\ref{Balancing of Bayes' Risks and Counteracting Risks})\emph{,} where
extreme data points that lie in decision regions involve forces associated
with risks or counter risks, such that the parameter vector satisfies the
linear decision boundary in terms of a minimum risk.

Thus, the dual locus of extreme data points must jointly satisfy a discrete
and data-driven version of the fundamental integral equation of binary
classification in Eq. \emph{(\ref{Equalizer Rule})} in terms of forces
associated with risks and counter risks which are related to positions and
potential locations of extreme data points and corresponding total allowed
eigenenergies of principal eigenaxis components. Moreover, the forces
associated with risks and counter risks that are related to positions and
potential locations of extreme data points and the corresponding total allowed
eigenenergies of principal eigenaxis components must jointly satisfy a
discrete and data-driven version of Eq.
(\ref{Balancing of Bayes' Risks and Counteracting Risks}) so that $\left(
1\right)  $ the forces associated with risks and counter risks that are
related to positions and potential locations of extreme data points are
effectively balanced with each other, and $\left(  2\right)  $ the total
allowed eigenenergies of the principal eigenaxis components are effectively
balanced with each other.

I will now show that distributions of extreme points determine decision
regions for binary classification systems.

\subsection{Distributions of Extreme Points}

\begin{figure}[ptb]%
\centering
\fbox{\includegraphics[
height=2.5875in,
width=3.4411in
]%
{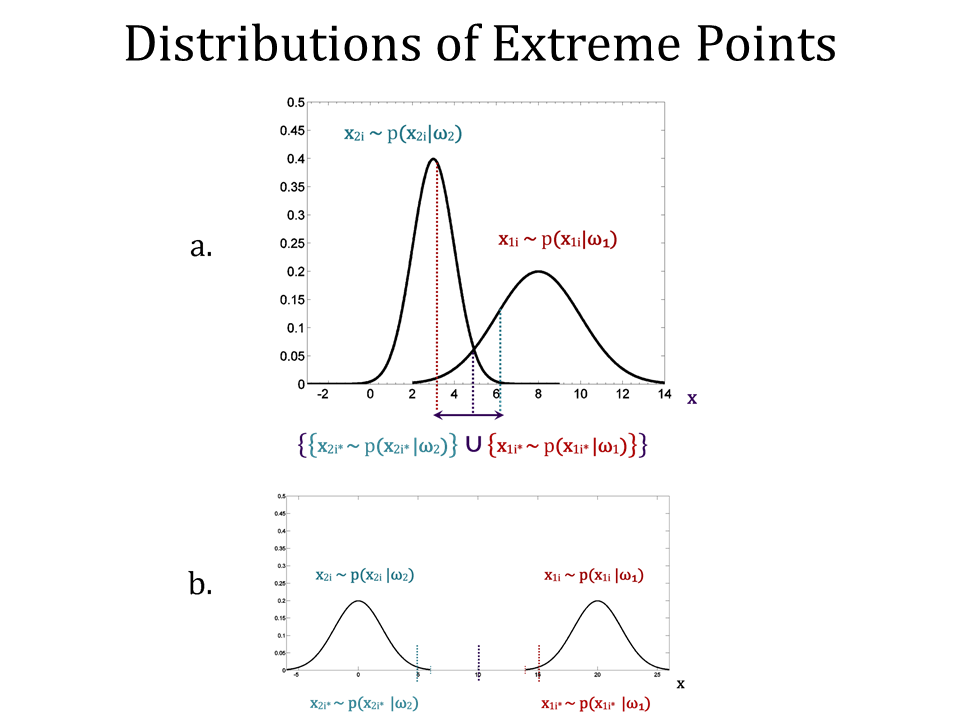}%
}\caption{Location properties of extreme data points are determined by
overlapping or non-overlapping intervals of probability density functions.
$\left(  a\right)  $ For overlapping data distributions, overlapping intervals
of likelihoods determine numbers and locations of extreme data points.
$\left(  b\right)  $ Relatively few extreme data points are located within the
tail regions of non-overlapping data distributions.}%
\label{Location Properties Extreme Data Points}%
\end{figure}

Take a collection of feature vectors for any two pattern classes, where the
data distributions are either overlapping or non-overlapping with each other.
Data points located in overlapping regions or tails regions between two data
distributions specify directions for which a given collection of data is most
variable or spread out. Call these data points "extreme points," where any
given extreme point is the endpoint of an extreme vector. Any given extreme
point is characterized by an expected value (a central location) and a
covariance (a spread). Figure $\ref{Location Properties Extreme Data Points}$a
depicts how overlapping intervals of probability density functions determine
locations of extreme points for two overlapping regions, and Fig.
$\ref{Location Properties Extreme Data Points}$b depicts how non-overlapping
intervals of probability density functions determine locations of extreme
points in two tail regions. Accordingly, distributions of extreme points
determine decision regions for binary classification systems, where the forces
associated with risks and counter risks are related to positions and potential
locations of extreme data points.%
\textbf{\ }

I\ will call a dual locus of principal eigenaxis components that determines an
estimate of class-conditional densities for extreme points a "linear
eigenlocus." I\ will refer to the parameter vector that provides an estimate
of class-conditional densities for extreme points as a "locus of likelihoods"
or a "parameter vector of likelihoods."

A linear eigenlocus, which is formed by a dual locus of principal eigenaxis
components and likelihoods, is a data-driven likelihood ratio and decision
boundary that determines computer-implemented, optimal linear statistical
classification systems that minimize the expected risk: for data drawn from
statistical distributions that have similar covariance matrices. I will call
the system of data-driven, mathematical laws that generates a linear
eigenlocus a "linear eigenlocus transform." I\ will introduce the primal
equation of a linear eigenlocus in the next section. I\ will begin the next
section by defining important geometric and statistical properties exhibited
by weighted extreme points on a linear eigenlocus. I\ will define these
properties in terms of geometric and statistical criterion.

\subsection{Linear Eigenlocus Transforms}

A\ high level description of linear eigenlocus transforms is outlined below.
The high level description specifies essential geometric and statistical
properties exhibited by the weighted extreme points on a linear eigenlocus.

\textbf{Linear eigenlocus transforms generate a locus of weighted extreme
points that is a dual locus of likelihoods and principal eigenaxis components,
where each weight specifies a class membership statistic and conditional
density for an extreme point, and each weight determines the magnitude and the
total allowed eigenenergy of an extreme vector.}

\begin{flushleft}
\textbf{Linear eigenlocus transforms choose each weight in a manner which
ensures that:}
\end{flushleft}

\paragraph{Criterion $\mathbf{1}$}

Each conditional density of an extreme point describes the central location
(expected value) and the spread (covariance) of the extreme point.

\paragraph{Criterion $\mathbf{2}$}

Distributions of the extreme points are distributed over the locus of
likelihoods in a symmetrically balanced and well-proportioned manner.

\paragraph{Criterion $\mathbf{3}$}

The total allowed eigenenergy possessed by each weighted extreme vector
specifies the probability of observing the extreme point within a localized region.

\paragraph{Criterion $\mathbf{4}$}

The total allowed eigenenergies of the weighted extreme vectors are
symmetrically balanced with each other about a center of total allowed eigenenergy.

\paragraph{Criterion $\mathbf{5}$}

The forces associated with risks and counter risks related to the weighted
extreme points are symmetrically balanced with each other about a center of
minimum risk.

\paragraph{Criterion $\mathbf{6}$}

The locus of principal eigenaxis components formed by weighted extreme vectors
partitions any given feature space into congruent decision regions which are
symmetrically partitioned by a linear decision boundary.

\paragraph{Criterion $\mathbf{7}$}

The locus of principal eigenaxis components is the focus of a linear decision boundary.

\paragraph{Criterion $\mathbf{8}$}

The locus of principal eigenaxis components formed by weighted extreme vectors
satisfies the linear decision boundary in terms of a critical minimum eigenenergy.

\paragraph{Criterion $\mathbf{9}$}

The locus of likelihoods formed by weighted extreme points satisfies the
linear decision boundary in terms of a minimum probability of decision error.

\paragraph{Criterion $\mathbf{10}$}

For data distributions that have dissimilar covariance matrices, the forces
associated with counter risks and risks, within each of the congruent decision
regions, are balanced with each other. For data distributions that have
similar covariance matrices, the forces associated with counter risks within
each of the congruent decision regions are equal to each other, and the forces
associated with risks within each of the congruent decision regions are equal
to each other.

\paragraph{Criterion $\mathbf{11}$}

For data distributions that have dissimilar covariance matrices, the
eigenenergies associated with counter risks and the eigenenergies associated
with risks, within each of the congruent decision regions, are balanced with
other. For data distributions that have similar covariance matrices, the
eigenenergies associated with counter risks within each of the congruent
decision regions are equal to each other, and the eigenenergies associated
with risks within each of the congruent decision regions are equal to each other.

I\ will devise a system of data-driven, locus equations that determines
likelihood ratios and decision boundaries which satisfy all of the above
criteria. Linear eigenlocus discriminant functions satisfy a fundamental
statistical property that I\ will call "symmetrical balance." The statistical
property of symmetrical balance is introduced next.

\subsection{Design of Balanced Fittings}

Learning machine architectures with $N$ free parameters have a learning
capacity to fit $N$ data points, where curves or surfaces can be made to pass
through every data point. However, highly flexible architectures with
indefinite parameter sets overfit training data
\citep
{Wahba1987,Breiman1991,Geman1992,Barron1998,Boser1992,Gershenfeld1999,Duda2001,Hastie2001,Haykin2009}%
, whereas architectures with too few parameters underfit training data
\citep{Guyon1992,Ivanciuc2007applications}%
.

\begin{figure}[ptb]%
\centering
\fbox{\includegraphics[
height=2.5875in,
width=3.4411in
]%
{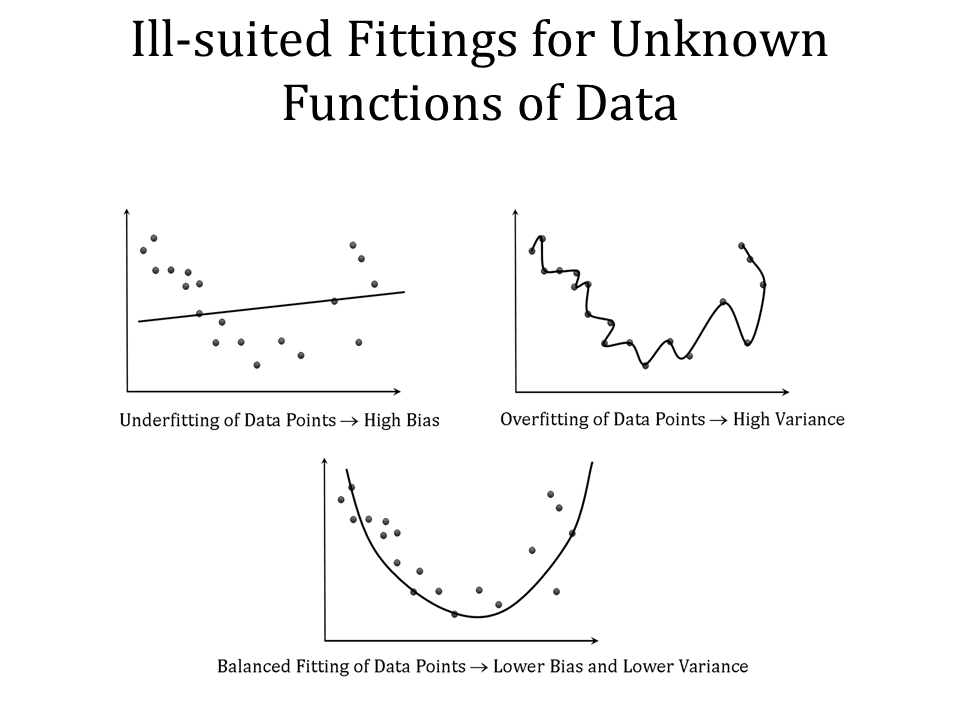}%
}\caption{ Illustration of the difficulties associated with fitting an unknown
function to a collection of random data points using classical interpolation
methods.}%
\label{Interpolation of Random Data Points}%
\end{figure}

I will show that fitting learning machine architectures to unknown
discriminant functions of data involves the design of \emph{balanced fittings}
for given sets of data points. But how do we define balanced fittings for
learning machines? If we think in terms of classical interpolation methods,
Fig. $\ref{Interpolation of Random Data Points}$ depicts a cartoon of an
underfitting, overfitting, and balanced fitting of a given set of data points;
Fig. $\ref{Interpolation of Random Data Points}$ also illustrates that
classical interpolation methods provide ill-suited fittings for unknown
functions of random data points.%

I\ will devise a system of data-driven, locus equations that determine
balanced fittings for learning machine architectures, where an unknown
function is a linear or quadratic discriminant function. The general idea is
outlined below.

\subsection{Statistical Property of Symmetrical Balance}

In this paper, I\ will design balanced fittings for learning machine
architectures, for any given set of data points, in terms of a fundamental
statistical property that I have named "symmetrical balance."

Generally speaking, symmetrical balance can be described as having an even
distribution of "weight" or a similar "load" on equal sides of a centrally
placed fulcrum. When a set of elements are arranged equally on either side of
a central axis, the result is bilateral symmetry
\citep{Jirousek1995}%
. Objects which exhibit bilateral symmetry look the same on both sides of a
central axis or a midline.

The physical property of symmetrical balance involves a physical system in
equilibrium, whereby opposing forces or influences of the system are balanced
with each other.

\subsubsection{Physical Property of Symmetrical Balance}

The physical property of symmetrical balance involves sets of elements which
are evenly or equally distributed over either side of an axis or a lever,
where a fulcrum is placed directly under the center of the axis or the lever.
Accordingly, symmetrical balance involves an axis or a lever in equilibrium,
where different elements are equal or in correct proportions, relative to the
center of an axis or a lever, such that the opposing forces and influences of
a system are balanced with each other.

\subsubsection{General Machinery of a Fulcrum and a Lever}

As a practical example, consider the general machinery of a fulcrum and a
lever, where a lever is any rigid object capable of turning about some fixed
point called a fulcrum. If a fulcrum is placed under directly under a lever's
center of gravity, the lever will remain balanced. Accordingly, the center of
gravity is the point at which the entire weight of a lever is considered to be
concentrated, so that if a fulcrum is placed at this point, the lever will
remain in equilibrium. If a lever is of uniform dimensions and density, then
the center of gravity is at the geometric center of the lever. Consider for
example, the playground device known as a seesaw or teeter-totter. The center
of gravity is at the geometric center of a teeter-totter, which is where the
fulcrum of a seesaw is located\
\citep{Asimov1966}%
.

I\ will use the idea of symmetrical balance, in terms of the general machinery
of a fulcrum and a lever, to devise a system of data-driven, locus equations
which determines the elegant, statical balancing feat that is outlined next.

\subsection{An Elegant Statistical Balancing Feat}

I\ will devise a system of data-driven, locus equations that determines a dual
locus of principal eigenaxis components and likelihoods, all of which satisfy
the statistical property of symmetrical balance described in the above set of
criteria. The dual locus provides an estimate of a principal eigenaxis that
has symmetrically balanced distributions of eigenenergies on equal sides of a
centrally placed fulcrum, which is located at its center of total allowed
eigenenergy. The dual locus also provides an estimate of a parameter vector of
likelihoods that has symmetrically balanced distributions of forces associated
with risks and counter risks on equal sides of a centrally placed fulcrum,
which is located at the center of risk. Thereby, a dual locus is in
\emph{statistical equilibrium}.

\subsubsection{Statistical Equilibrium}

I\ will show that the total allowed eigenenergies possessed by the principal
eigenaxis components on a dual locus are distributed over its axis in a
symmetrically balanced and well-proportioned manner, such that the total
allowed eigenenergies of a dual locus are symmetrically balanced with each
other about its center of total allowed eigenenergy, which is at the geometric
center of the dual locus.

\begin{figure}[ptb]%
\centering
\fbox{\includegraphics[
height=2.5875in,
width=3.4411in
]%
{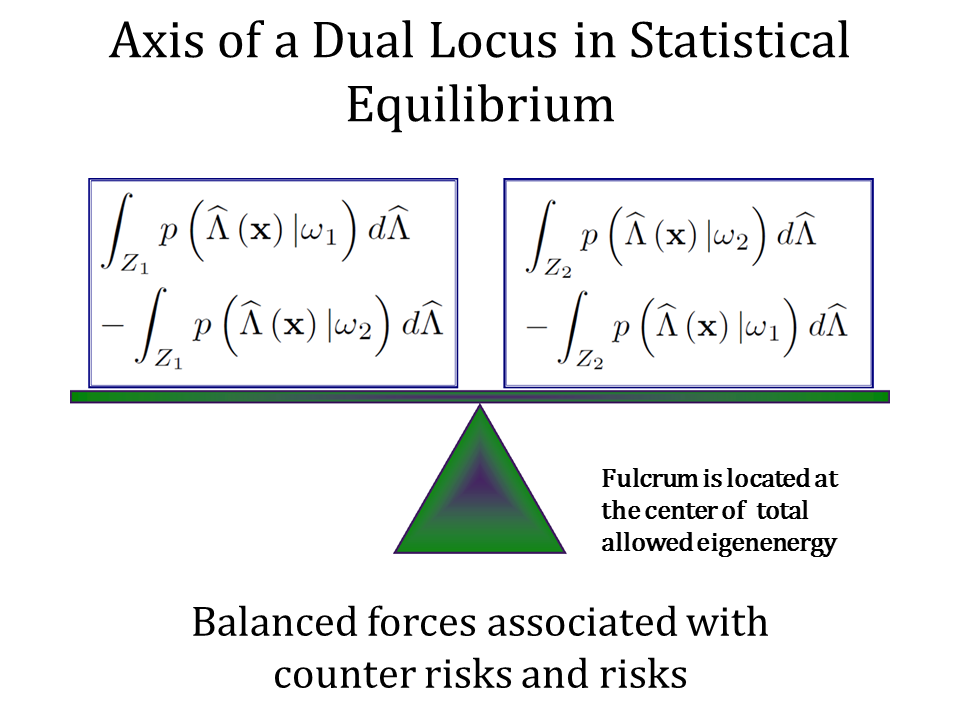}%
}\caption{Linear eigenlocus transforms generate linear discriminant functions
and linear decision boundaries which possess the statistical property of
symmetrical balance, whereby a dual locus determines decision regions $Z_{1}$
and $Z_{2}$ that have symmetrically balanced forces associated with risks and
counter risks: $\mathfrak{R}_{\mathfrak{\min}}\left(
Z|\protect\widehat{\Lambda}\left(  \mathbf{x}\right)  \right)  :\overline
{\mathfrak{R}}_{\mathfrak{\min}}\left(  Z_{1}|\omega_{1}\right)
-\mathfrak{R}_{\mathfrak{\min}}\left(  Z_{1}|\omega_{2}\right)
\rightleftharpoons\overline{\mathfrak{R}}_{\mathfrak{\min}}\left(
Z_{2}|\omega_{2}\right)  -\mathfrak{R}_{\mathfrak{\min}}\left(  Z_{2}%
|\omega_{1}\right)  $}%
\label{Dual Statistical Balancing Feat}%
\end{figure}

I will also demonstrate that the utility of the statistical balancing feat
involves \emph{balancing} all of the forces associated with the counter risk
$\overline{\mathfrak{R}}_{\mathfrak{\min}}\left(  Z_{1}|\omega_{1}\right)  $
and the risk $\mathfrak{R}_{\mathfrak{\min}}\left(  Z_{1}|\omega_{2}\right)  $
in the $Z_{1}$ decision region \emph{with} all of the forces associated with
the counter risk $\overline{\mathfrak{R}}_{\mathfrak{\min}}\left(
Z_{2}|\omega_{2}\right)  $ and the risk $\mathfrak{R}_{\mathfrak{\min}}\left(
Z_{2}|\omega_{1}\right)  $ in the $Z_{2}$ decision region:%
\[
\mathfrak{R}_{\mathfrak{\min}}\left(  Z|\widehat{\Lambda}\left(
\mathbf{x}\right)  \right)  :\overline{\mathfrak{R}}_{\mathfrak{\min}}\left(
Z_{1}|\omega_{1}\right)  -\mathfrak{R}_{\mathfrak{\min}}\left(  Z_{1}%
|\omega_{2}\right)  \rightleftharpoons\overline{\mathfrak{R}}_{\mathfrak{\min
}}\left(  Z_{2}|\omega_{2}\right)  -\mathfrak{R}_{\mathfrak{\min}}\left(
Z_{2}|\omega_{1}\right)  \text{,}%
\]
where the forces associated with risks and counter risks are related to
extreme point positions and potential locations, such that the eigenenergy and
the corresponding expected risk $\mathfrak{R}_{\mathfrak{\min}}\left(
Z|\widehat{\Lambda}\left(  \mathbf{x}\right)  \right)  $ of a binary
classification system are both minimized. Figure
$\ref{Dual Statistical Balancing Feat}$ illustrates that linear eigenlocus
transforms routinely accomplish an elegant, statistical balancing feat in
eigenspace which facilitates a surprising, statistical balancing feat in
decision space $Z$.%

Linear eigenlocus transforms are generated by solving the inequality
constrained optimization problem that is introduced next.

\subsection{Primal Problem of a Linear Eigenlocus}

Take any given collection of training data for a binary classification problem
of the form:%
\[
\left(  \mathbf{x}_{1},y_{1}\right)  ,\ldots,\left(  \mathbf{x}_{N}%
,y_{N}\right)  \in%
\mathbb{R}
^{d}\times Y,Y=\left\{  \pm1\right\}  \text{,}%
\]
where feature vectors $\mathbf{x}$ from class $\omega_{1}$ and class
$\omega_{2}$ are drawn from unknown, class-conditional probability density
functions $p\left(  \mathbf{x}|\omega_{1}\right)  $ and $p\left(
\mathbf{x}|\omega_{2}\right)  $ and are identically distributed.

A linear eigenlocus $\boldsymbol{\tau}$ is estimated by solving an inequality
constrained optimization problem:%
\begin{align}
\min\Psi\left(  \boldsymbol{\tau}\right)   &  =\left\Vert \boldsymbol{\tau
}\right\Vert ^{2}/2+C/2\sum\nolimits_{i=1}^{N}\xi_{i}^{2}\text{,}%
\label{Primal Normal Eigenlocus}\\
\text{s.t. }y_{i}\left(  \mathbf{x}_{i}^{T}\boldsymbol{\tau}+\tau_{0}\right)
&  \geq1-\xi_{i},\ \xi_{i}\geq0,\ i=1,...,N\text{,}\nonumber
\end{align}
where $\boldsymbol{\tau}$ is a $d\times1$ constrained, primal linear
eigenlocus which is a dual locus of likelihoods and principal eigenaxis
components $\widehat{\Lambda}_{\boldsymbol{\tau}}\left(  \mathbf{x}\right)  $,
$\left\Vert \boldsymbol{\tau}\right\Vert ^{2}$ is the total allowed
eigenenergy exhibited by $\boldsymbol{\tau}$, $\tau_{0}$ is a functional of
$\boldsymbol{\tau}$, $C$ and $\xi_{i}$ are regularization parameters, and
$y_{i}$ are class membership statistics: if $\mathbf{x}_{i}\in\omega_{1}$,
assign $y_{i}=1$; if $\mathbf{x}_{i}\in\omega_{2}$, assign $y_{i}=-1$.

Equation (\ref{Primal Normal Eigenlocus}) is the primal problem of a linear
eigenlocus, where the system of $N$ inequalities must be satisfied:%
\[
y_{i}\left(  \mathbf{x}_{i}^{T}\boldsymbol{\tau}+\tau_{0}\right)  \geq
1-\xi_{i},\ \xi_{i}\geq0,\ i=1,...,N\text{,}%
\]
such that a constrained, primal linear eigenlocus $\boldsymbol{\tau}$
satisfies a critical minimum eigenenergy constraint:%
\begin{equation}
\gamma\left(  \boldsymbol{\tau}\right)  =\left\Vert \boldsymbol{\tau
}\right\Vert _{\min_{c}}^{2}\text{,}
\label{Minimum Total Eigenenergy Primal Normal Eigenlocus}%
\end{equation}
where $\left\Vert \boldsymbol{\tau}\right\Vert _{\min_{c}}^{2}$ determines the
minimum risk $\mathfrak{R}_{\mathfrak{\min}}\left(  Z|\boldsymbol{\tau
}\right)  $ of a linear classification system.

Solving the inequality constrained optimization problem in Eq.
(\ref{Primal Normal Eigenlocus}) involves solving a dual optimization problem
that determines the fundamental unknowns of Eq.
(\ref{Primal Normal Eigenlocus}). Denote a Wolfe dual linear eigenlocus by
$\boldsymbol{\psi}$ and the Lagrangian dual problem of $\boldsymbol{\psi}$ by
$\max\Xi\left(  \boldsymbol{\psi}\right)  $. Let $\boldsymbol{\psi}$ be a
Wolfe dual of $\boldsymbol{\tau}$ such that proper and effective strong
duality relationships exist between the algebraic systems of $\min\Psi\left(
\boldsymbol{\tau}\right)  $ and $\max\Xi\left(  \boldsymbol{\psi}\right)  $.
Thereby, let $\boldsymbol{\psi}$ be related with $\boldsymbol{\tau}$ in a
symmetrical manner that specifies the locations of the principal eigenaxis
components on $\boldsymbol{\tau}$. The Wolfe dual linear eigenlocus
$\boldsymbol{\psi}$ is important for the following reasons.

\subsection{Why the Wolfe Dual Linear Eigenlocus Matters}

Duality relationships for Lagrange multiplier problems are based on the
premise that it is the Lagrange multipliers which are the fundamental unknowns
associated with a constrained problem. Dual methods solve an alternate
problem, termed the dual problem, whose unknowns are the Lagrange multipliers
of the first problem, termed the primal problem. Once the Lagrange multipliers
are known, the solution to a primal problem can be determined
\citep{Luenberger2003}%
.

\subsubsection{The Real Unknowns}

A constrained, primal linear eigenlocus is a dual locus of principal eigenaxis
components and likelihoods formed by weighted extreme points, where each
weight is specified by a class membership statistic and a scale factor. Each
scale factor specifies a conditional density for a weighted extreme point on a
locus of likelihoods, and each scale factor determines the magnitude and the
eigenenergy of a weighted extreme vector on a locus of principal eigenaxis
components. The main issue concerns how the scale factors are determined.

\subsubsection{The Fundamental Unknowns}

The fundamental unknowns are the scale factors of the principal eigenaxis
components on a Wolfe dual linear eigenlocus $\boldsymbol{\psi}$.

\subsection{Strong Dual Linear Eigenlocus Transforms}

For the problem of linear eigenlocus transforms, the Lagrange multipliers
method introduces a Wolfe dual linear eigenlocus $\boldsymbol{\psi}$ of
principal eigenaxis components, for which the Lagrange multipliers $\left\{
\psi_{i}\right\}  _{i=1}^{N}$ are the magnitudes or lengths of a set of Wolfe
dual principal eigenaxis components $\left\{  \psi_{i}%
\overrightarrow{\mathbf{e}}_{i}\right\}  _{i=1}^{N}$, where $\left\{
\overrightarrow{\mathbf{e}}_{i}\right\}  _{i=1}^{N}$ are non-orthogonal unit
vectors, and finds extrema for the restriction of a primal linear eigenlocus
$\boldsymbol{\tau}$ to a Wolfe dual eigenspace. Accordingly, the fundamental
unknowns associated with Eq. (\ref{Primal Normal Eigenlocus}) are the
magnitudes or lengths of the Wolfe dual principal eigenaxis components on
$\boldsymbol{\psi}$.

\subsubsection{Strong Duality}

Because Eq. (\ref{Primal Normal Eigenlocus}) is a convex programming problem,
the theorem for convex duality guarantees an equivalence and corresponding
symmetry between a constrained, primal linear eigenlocus $\boldsymbol{\tau}$
and its Wolfe dual $\boldsymbol{\psi}$
\citep{Nash1996,Luenberger2003}%
. Strong duality holds between the systems of locus equations denoted by
$\min\Psi\left(  \boldsymbol{\tau}\right)  $ and $\max\Xi\left(
\boldsymbol{\psi}\right)  $, so that the duality gap between the constrained
primal and the Wolfe dual linear eigenlocus solution is zero
\citep{Luenberger1969,Nash1996,Fletcher2000,Luenberger2003}%
.

The Lagrangian dual problem of a Wolfe dual linear eigenlocus will be derived
by means of the Lagrangian functional that is introduced next.

\subsection{The Lagrangian of the Linear Eigenlocus}

The inequality constrained optimization problem in Eq.
(\ref{Primal Normal Eigenlocus}) is solved by using Lagrange multipliers
$\psi_{i}\geq0$ and the Lagrangian functional:%
\begin{align}
L_{\Psi\left(  \boldsymbol{\tau}\right)  }\left(  \boldsymbol{\tau}%
\mathbf{,}\tau_{0},\mathbf{\xi},\boldsymbol{\psi}\right)   &  =\left\Vert
\boldsymbol{\tau}\right\Vert ^{2}/2\label{Lagrangian Normal Eigenlocus}\\
&  +C/2\sum\nolimits_{i=1}^{N}\xi_{i}^{2}\nonumber\\
&  -\sum\nolimits_{i=1}^{N}\psi_{i}\nonumber\\
&  \times\left\{  y_{i}\left(  \mathbf{x}_{i}^{T}\boldsymbol{\tau}+\tau
_{0}\right)  -1+\xi_{i}\right\} \nonumber
\end{align}
which is minimized with respect to the primal variables $\boldsymbol{\tau}%
$\textbf{ }and $\tau_{0}$ and is maximized with respect to the dual variables
$\psi_{i}$.

The Karush-Kuhn-Tucker (KKT) conditions on the Lagrangian functional
$L_{\Psi\left(  \boldsymbol{\tau}\right)  }$:%
\begin{equation}
\boldsymbol{\tau}-\sum\nolimits_{i=1}^{N}\psi_{i}y_{i}\mathbf{x}_{i}=0,\text{
\ }i=1,...N\text{,} \label{KKTE1}%
\end{equation}%
\begin{equation}
\sum\nolimits_{i=1}^{N}\psi_{i}y_{i}=0,\text{ \ }i=1,...,N\text{,}
\label{KKTE2}%
\end{equation}%
\begin{equation}
C\sum\nolimits_{i=1}^{N}\xi_{i}-\sum\nolimits_{i=1}^{N}\psi_{i}=0\text{,}
\label{KKTE3}%
\end{equation}%
\begin{equation}
\psi_{i}\geq0,\text{ \ }i=1,...,N\text{,} \label{KKTE4}%
\end{equation}%
\begin{equation}
\psi_{i}\left[  y_{i}\left(  \mathbf{x}_{i}^{T}\boldsymbol{\tau}+\tau
_{0}\right)  -1+\xi_{i}\right]  \geq0,\ i=1,...,N\text{,} \label{KKTE5}%
\end{equation}
which can found in
\citep{Cortes1995,Burges1998,Cristianini2000,Scholkopf2002}%
, determine a system of data-driven, locus equations which are jointly
satisfied by a constrained primal and a Wolfe dual linear eigenlocus. I will
define the manner in which the KKT conditions determine geometric and
statistical properties exhibited by weighted extreme points on a Wolfe dual
$\boldsymbol{\psi}$ and a constrained primal $\boldsymbol{\tau}$ linear
eigenlocus. Thereby, I\ will demonstrate the manner in which the KKT
conditions ensure that $\boldsymbol{\psi}$ and $\boldsymbol{\tau}$ jointly
satisfy discrete and data-driven versions of the fundamental equations of
binary classification for a classification system in statistical equilibrium.

The Lagrangian dual problem of a Wolfe dual linear eigenlocus is introduced next.

\subsection{Lagrangian Dual Problem of a Linear Eigenlocus}

The resulting expressions for a primal linear eigenlocus $\boldsymbol{\tau}$
in Eq. (\ref{KKTE1}) and a Wolfe dual linear eigenlocus $\boldsymbol{\psi}$ in
Eq. (\ref{KKTE2}) are substituted into the Lagrangian functional
$L_{\Psi\left(  \boldsymbol{\tau}\right)  }$ of Eq.
(\ref{Lagrangian Normal Eigenlocus}) and simplified. This produces the
Lagrangian dual problem of a Wolfe dual linear eigenlocus: a quadratic
programming problem%
\begin{equation}
\max\Xi\left(  \boldsymbol{\psi}\right)  =\sum\nolimits_{i=1}^{N}\psi_{i}%
-\sum\nolimits_{i,j=1}^{N}\psi_{i}\psi_{j}y_{i}y_{j}\frac{\left[
\mathbf{x}_{i}^{T}\mathbf{x}_{j}+\delta_{ij}/C\right]  }{2}
\label{Wolfe Dual Normal Eigenlocus}%
\end{equation}
which is subject to the algebraic constraints $\sum\nolimits_{i=1}^{N}%
y_{i}\psi_{i}=0$ and $\psi_{i}\geq0$, where $\delta_{ij}$ is the Kronecker
$\delta$ defined as unity for $i=j$ and $0$ otherwise.

Equation (\ref{Wolfe Dual Normal Eigenlocus}) can be written in vector
notation by letting $\mathbf{Q}\triangleq\epsilon\mathbf{I}%
+\widetilde{\mathbf{X}}\widetilde{\mathbf{X}}^{T}$ and $\widetilde{\mathbf{X}%
}\triangleq\mathbf{D}_{y}\mathbf{X}$, where $\mathbf{D}_{y}$ is an $N\times N$
diagonal matrix of class membership statistics (labels) $y_{i}$, and the
$N\times d$ data matrix is $\mathbf{X}$ $=%
\begin{pmatrix}
\mathbf{x}_{1}, & \mathbf{x}_{2}, & \ldots, & \mathbf{x}_{N}%
\end{pmatrix}
^{T}$. This produces the matrix version of the Lagrangian dual problem of a
primal linear eigenlocus within its Wolfe dual eigenspace:%
\begin{equation}
\max\Xi\left(  \boldsymbol{\psi}\right)  =\mathbf{1}^{T}\boldsymbol{\psi
}-\frac{\boldsymbol{\psi}^{T}\mathbf{Q}\boldsymbol{\psi}}{2}
\label{Vector Form Wolfe Dual}%
\end{equation}
which is subject to the constraints $\boldsymbol{\psi}^{T}\mathbf{y}=0$ and
$\psi_{i}\geq0$
\citep{Reeves2009}%
. Given the theorem for convex duality, it follows that a Wolfe dual linear
eigenlocus $\boldsymbol{\psi}$ is a dual locus of likelihoods and principal
eigenaxis components $\widehat{\Lambda}_{\boldsymbol{\psi}}\left(
\mathbf{x}\right)  $, where $\boldsymbol{\psi}$ exhibits a total allowed
eigenenergy $\left\Vert \boldsymbol{\psi}\right\Vert _{\min_{c}}^{2}$ that is
symmetrically related to the total allowed eigenenergy $\left\Vert
\boldsymbol{\tau}\right\Vert _{\min_{c}}^{2}$ of $\boldsymbol{\tau}$:
$\left\Vert \boldsymbol{\psi}\right\Vert _{\min_{c}}^{2}\simeq\left\Vert
\boldsymbol{\tau}\right\Vert _{\min_{c}}^{2}$.

\subsection{Loci of Constrained Quadratic Forms}

The representation of a constrained, primal linear eigenlocus
$\boldsymbol{\tau}$ within its Wolfe dual eigenspace involves the eigensystem
of the constrained quadratic form $\boldsymbol{\psi}^{T}\mathbf{Q}%
\boldsymbol{\psi}$ in Eq. (\ref{Vector Form Wolfe Dual}), where
$\boldsymbol{\psi}$ is the principal eigenvector of $\mathbf{Q}$, such that
$\boldsymbol{\psi}^{T}\mathbf{y}=0$ and $\psi_{i}\geq0$. I will demonstrate
how the eigensystem in Eq. (\ref{Vector Form Wolfe Dual}) determines the
manner in which the total allowed eigenenergies $\left\Vert \boldsymbol{\tau
}\right\Vert _{\min_{c}}^{2}$ exhibited by $\boldsymbol{\tau}$ are
symmetrically balanced with each other.

The standard quadratic form%
\[
\mathbf{x}^{T}\mathbf{Ax}=1
\]
is a locus equation that determines $d$-dimensional circles, ellipses,
hyperbolas, parabolas, lines, or points, where a symmetric matrix $\mathbf{A}
$ specifies algebraic constraints that are satisfied by the points
$\mathbf{x}$ on a given locus. The eigenvalues of $\mathbf{A}$ determine the
shape of a given locus and the principal eigenvector of $\mathbf{A}$ is the
major (principal) axis of a given locus
\citep{Hewson2009}%
. Alternatively, principal eigenvectors of sample correlation and covariance
matrices determine principal axes that describe the direction in which the
data is the most variable or spread out
\citep{Duda2001,Hastie2001,Jolliffe2002}%
.

I\ will demonstrate that Eqs (\ref{Wolfe Dual Normal Eigenlocus}) and
(\ref{Vector Form Wolfe Dual}) determine a dual linear eigenlocus
$\boldsymbol{\psi}$ which is in statistical equilibrium such that the total
allowed eigenenergies $\left\Vert \boldsymbol{\tau}\right\Vert _{\min_{c}}%
^{2}$ exhibited by $\boldsymbol{\tau}$ are symmetrically balanced with each
other about a center of total allowed eigenenergy. I will also demonstrate
that the utility of the statistical balancing feat involves \emph{balancing}
all of the forces associated with the counter risk $\overline{\mathfrak{R}%
}_{\mathfrak{\min}}\left(  Z_{1}|\omega_{1}\right)  $ and the risk
$\mathfrak{R}_{\mathfrak{\min}}\left(  Z_{1}|\omega_{2}\right)  $ in the
$Z_{1}$ decision region \emph{with} all of the forces associated with the
counter risk $\overline{\mathfrak{R}}_{\mathfrak{\min}}\left(  Z_{2}%
|\omega_{2}\right)  $ and the risk $\mathfrak{R}_{\mathfrak{\min}}\left(
Z_{2}|\omega_{1}\right)  $ in the $Z_{2}$ decision region, where the forces
associated with risks and counter risks are related to positions and potential
locations of extreme points, such that the eigenenergy $\left\Vert
\boldsymbol{\tau}\right\Vert _{\min_{c}}^{2}$ and the expected risk
$\mathfrak{R}_{\mathfrak{\min}}\left(  Z|\boldsymbol{\tau}\right)  $ of a
discrete, linear classification system are both minimized.

I\ will now use the KKT conditions in Eqs (\ref{KKTE1}) and (\ref{KKTE4}) to
derive the locus equation of a constrained, primal linear eigenlocus
$\boldsymbol{\tau}$.

\subsection{The Constrained Primal Linear Eigenlocus}

Using the KKT\ conditions in Eqs (\ref{KKTE1}) and (\ref{KKTE4}), it follows
that an estimate for $\boldsymbol{\tau}$ satisfies the following vector
expression:%
\begin{equation}
\boldsymbol{\tau}=\sum\nolimits_{i=1}^{N}y_{i}\psi_{i}\mathbf{x}_{i}\text{,}
\label{Normal Eigenlocus Estimate}%
\end{equation}
where the $y_{i}$ terms are training set labels, i.e., class membership
statistics, (if $\mathbf{x}_{i}$ is a member of class $\omega_{1}$, assign
$y_{i}=1$; otherwise, assign $y_{i}=-1$), $\psi_{i}$ is a scale factor for
$\mathbf{x}_{i}$, and the magnitude $\psi_{i}$ of each principal eigenaxis
component $\psi_{i}\overrightarrow{\mathbf{e}}_{i}$ on $\boldsymbol{\psi}$ is
greater than or equal to zero: $\psi_{i}\geq0$.

The KKT condition in Eq. (\ref{KKTE4}) requires that the length $\psi_{i}$ of
each principal eigenaxis component $\psi_{i}\overrightarrow{\mathbf{e}}_{i}$
on $\boldsymbol{\psi}$ either satisfy or exceed zero: $\psi_{i}\geq0$. Any
principal eigenaxis component $\psi_{i}\overrightarrow{\mathbf{e}}_{i}$ which
has zero length, where $\psi_{i}=0$, satisfies the origin $P_{\mathbf{0}}%
\begin{pmatrix}
0, & 0, & \cdots, & 0
\end{pmatrix}
$ and is not on the Wolfe dual linear eigenlocus $\boldsymbol{\psi}$. It
follows that the constrained, primal principal eigenaxis component $\psi
_{i}\mathbf{x}_{i}$ also has zero length, i.e., $\left\Vert \psi_{i}%
\mathbf{x}_{i}\right\Vert =0$, and is not on the constrained, primal linear
eigenlocus $\boldsymbol{\tau}$.

Data points $\mathbf{x}_{i}$ correlated with Wolfe dual principal eigenaxis
components $\psi_{i}\overrightarrow{\mathbf{e}}_{i}$ that have non-zero
magnitudes $\psi_{i}>0$ are termed extreme vectors. Accordingly, extreme
vectors are unscaled, primal principal eigenaxis components on
$\boldsymbol{\tau}$. Geometric and statistical properties of extreme vectors
are outlined below.

\subsubsection{Properties of Extreme Vectors}

Take a collection of training data drawn from any two statistical
distributions. An extreme point is defined to be a data point which exhibits a
high variability of geometric location, that is, possesses a large covariance,
such that it is located $(1)$ relatively far from its distribution mean, $(2)$
relatively close to the mean of the other distribution, and $(3)$ relatively
close to other extreme points. Therefore, an extreme point is located
somewhere within either an overlapping region or a tail region between the two
data distributions.

Given the geometric and statistical properties exhibited by the locus of an
extreme point, it follows that a set of extreme vectors determine principal
directions of large covariance for a given collection of training data. Thus,
extreme vectors are discrete principal components that specify directions for
which a given collection of training data is most variable or spread out.
Accordingly, the loci of a set of extreme vectors span a region of large
covariance between two distributions of training data. Decision regions and
risks and counter risks for overlapping and non-overlapping data distributions
are defined next.

\paragraph{Overlapping Data Distributions}

For overlapping data distributions, the loci of the extreme vectors from each
pattern class are distributed within bipartite, joint geometric regions of
large covariance, both of which span the region of data distribution overlap.
Therefore, decision regions that have significant risks are functions of
overlapping intervals of probability density functions that determine numbers
and locations of extreme points. Figure
$\ref{Location Properties Extreme Data Points}$a depicts how extreme points
from two pattern classes are located within bipartite, joint geometric regions
of large variance that are located between two overlapping data distributions.

\paragraph{Non-overlapping Data Distributions}

For non-overlapping data distributions, the loci of the extreme vectors are
distributed within bipartite, disjoint geometric regions of large covariance,
i.e., separate tail regions, that are located between the data distributions.
Because tail regions of distributions are determined by non-overlapping
intervals of low likelihood, relatively few extreme points are located within
tail regions. Thereby, relatively few extreme points are located between
non-overlapping data distributions. Thus, decision regions that have
\emph{negligible or no risk} are functions of non-overlapping intervals of
probability density functions that determine \emph{tail regions}. Figure
$\ref{Location Properties Extreme Data Points}$b illustrates how small numbers
of extreme points are located within the tail regions of non-overlapping,
Gaussian data distributions.

\subsection{Primal Linear Eigenlocus Components}

All of the principal eigenaxis components on a constrained, primal linear
eigenlocus $\boldsymbol{\tau}$ are labeled, scaled extreme points in $%
\mathbb{R}
^{d}$. Denote the labeled, scaled extreme vectors that belong to class
$\omega_{1}$ and $\omega_{2}$ by $\psi_{1_{i\ast}}\mathbf{x}_{1_{i_{\ast}}}$
and $-\psi_{2_{i\ast}}\mathbf{x}_{2_{i\ast}}$, with scale factors
$\psi_{1_{i\ast}}$ and $\psi_{2_{i\ast}}$, extreme vectors $\mathbf{x}%
_{1_{i\ast}}$ and $\mathbf{x}_{2_{i\ast}}$, and class membership statistics
$y_{i}=1$ and $y_{i}=-1$ respectively. Let there be $l_{1}$ labeled, scaled
extreme points $\left\{  \psi_{1_{i\ast}}\mathbf{x}_{1_{i\ast}}\right\}
_{i=1}^{l_{1}}$ and $l_{2}$ labeled, scaled extreme points $\left\{
-\psi_{2_{i\ast}}\mathbf{x}_{2_{i\ast}}\right\}  _{i=1}^{l_{2}}$.

\begin{figure}[ptb]%
\centering
\fbox{\includegraphics[
height=2.5875in,
width=3.4411in
]%
{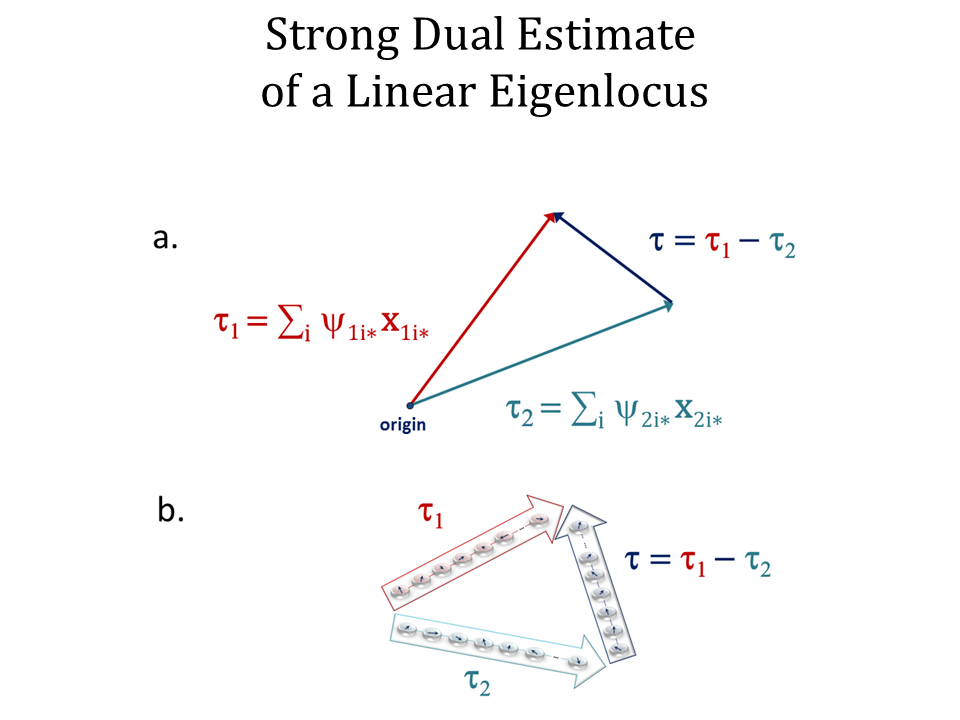}%
}\caption{$\left(  a\right)  $ A constrained, primal linear eigenlocus
$\boldsymbol{\tau}$ is determined by the vector difference $\boldsymbol{\tau
}_{1}-\boldsymbol{\tau}_{2}$ between a pair of constrained, primal linear
eigenlocus components $\boldsymbol{\tau}_{1}$ and $\boldsymbol{\tau}_{2}$.
$\left(  b\right)  $ The scaled extreme points on $\boldsymbol{\tau}_{1}$ and
$\boldsymbol{\tau}_{2}$ are endpoints of scaled extreme vectors that possess
unchanged directions and eigen-balanced lengths.}%
\label{Primal Linear Eigenlocus in Wolfe Dual Eigenspace}%
\end{figure}

Given Eq. (\ref{Normal Eigenlocus Estimate}) and the assumptions outlined
above, it follows that an estimate for a constrained, primal linear eigenlocus
$\boldsymbol{\tau}$ is based on the vector difference between a pair of
constrained, primal linear eigenlocus components:%
\begin{align}
\boldsymbol{\tau}  &  =\sum\nolimits_{i=1}^{l_{1}}\psi_{1_{i\ast}}%
\mathbf{x}_{1_{i\ast}}-\sum\nolimits_{i=1}^{l_{2}}\psi_{2_{i\ast}}%
\mathbf{x}_{2_{i\ast}}\label{Pair of Normal Eigenlocus Components}\\
&  =\boldsymbol{\tau}_{1}-\boldsymbol{\tau}_{2}\text{,}\nonumber
\end{align}
where the constrained, primal linear eigenlocus components $\sum
\nolimits_{i=1}^{l_{1}}\psi_{1_{i\ast}}\mathbf{x}_{1_{i\ast}}$ and
$\sum\nolimits_{i=1}^{l_{2}}\psi_{2_{i\ast}}\mathbf{x}_{2_{i\ast}}$ are
denoted by $\boldsymbol{\tau}_{1}$ and $\boldsymbol{\tau}_{2}$ respectively.
The sets of scaled extreme points $\left\{  \psi_{1_{1\ast}}\mathbf{x}%
_{1_{i\ast}}\right\}  _{i=1}^{l_{1}}$ and $\left\{  \psi_{2_{1\ast}}%
\mathbf{x}_{2_{i\ast}}\right\}  _{i=1}^{l_{2}}$ on $\boldsymbol{\tau}_{1}$ and
$\boldsymbol{\tau}_{2}$ determine the loci of $\boldsymbol{\tau}_{1}$ and
$\boldsymbol{\tau}_{2}$ and therefore determine the dual locus of
$\boldsymbol{\tau}=\boldsymbol{\tau}_{1}-\boldsymbol{\tau}_{2}$.
Figure\textbf{\ }$\ref{Primal Linear Eigenlocus in Wolfe Dual Eigenspace}$
depicts how the loci of $\boldsymbol{\tau}_{1}$ and $\boldsymbol{\tau}_{2}$
determine the dual locus of $\boldsymbol{\tau}$.%

I\ will now define how the regularization parameters $C$ and $\xi_{i}$ in Eqs
(\ref{Primal Normal Eigenlocus}) and (\ref{Vector Form Wolfe Dual}) affect the
dual locus of $\boldsymbol{\tau}$. Regularization components are essential
numerical ingredients in algorithms that involve inversions of data matrices
\citep
{Linz1979,Groetsch1984,Wahba1987,Groetsch1993,Hansen1998,Engl2000,Zhdanov2002,Linz2003}%
. Because linear eigenlocus transforms involve an inversion of the Gram matrix
$\mathbf{Q}$ in Eq. (\ref{Vector Form Wolfe Dual}), some type of
regularization is required for low rank Gram matrices
\citep{Reeves2009,Reeves2011}%
.

\subsection{Weak Dual Linear Eigenlocus Transforms}

It has been shown that the number and the locations of the principal eigenaxis
components on $\boldsymbol{\psi}$ and $\boldsymbol{\tau}$ are considerably
affected by the rank and eigenspectrum of $\mathbf{Q}$. In particular, low
rank Gram matrices $\mathbf{Q}$ generate "weak\emph{\ }dual" linear eigenlocus
transforms that produce irregular, linear partitions of decision spaces
\citep{,Reeves2015resolving}%
.

For example, given non-overlapping data distributions and low rank Gram
matrices, weak dual linear eigenlocus transforms produce asymmetric, linear
partitions that exhibit optimal generalization performance at the expense of
unnecessary principal eigenaxis components, where \emph{all} of the training
data is transformed into constrained, primal principal eigenaxis components.
For overlapping data distributions, incomplete eigenspectra of low rank Gram
matrices $\mathbf{Q}$ result in weak dual linear eigenlocus transforms which
determine ill-formed, linear decision boundaries that exhibit substandard
generalization performance
\citep{Reeves2009,Reeves2011,Reeves2015resolving}%
. All of these problems are solved by the regularization method that is
described next.

\subsubsection{Regularization of Linear Eigenlocus Transforms}

For any collection of $N$ training vectors of dimension $d$, where $d<N$, the
Gram matrix $\mathbf{Q}$ has low rank. It has been shown that the regularized
form of $\mathbf{Q}$, for which $\epsilon\ll1$ and $\mathbf{Q}\triangleq
\epsilon\mathbf{I}+\widetilde{\mathbf{X}}\widetilde{\mathbf{X}}^{T}$, ensures
that $\mathbf{Q}$ has full rank and a complete eigenvector set so that
$\mathbf{Q}$ has a complete eigenspectrum. The regularization constant $C$ is
related to the regularization parameter $\epsilon$\ by $\frac{1}{C}$
\citep{Reeves2011}%
.

For $N$ training vectors of dimension $d$, where $d<N$, all of the
regularization parameters $\left\{  \xi_{i}\right\}  _{i=1}^{N}$ in Eq.
(\ref{Primal Normal Eigenlocus}) and all of its derivatives are set equal to a
very small value: $\xi_{i}=\xi\ll1$. The regularization constant $C$ is set
equal to $\frac{1}{\xi}$: $C=\frac{1}{\xi}$.

For $N$ training vectors of dimension $d$, where $N<d$, all of the
regularization parameters $\left\{  \xi_{i}\right\}  _{i=1}^{N}$ in Eq.
(\ref{Primal Normal Eigenlocus}) and all of its derivatives are set equal to
zero: $\xi_{i}=\xi=0$. The regularization constant $C$ is set equal to
infinity: $C=\infty$.

In the next section, I will devise locus equations that determine the manner
in which a constrained, primal linear eigenlocus partitions any given feature
space into congruent decision regions.

\section{Equations of a Linear Discriminant Function}

For data distributions that have similar covariance matrices, a constrained,
primal linear eigenlocus is the primary basis of a linear discriminant
function that implements optimal likelihood ratio tests. The manner in which
the dual locus of $\boldsymbol{\tau}$ partitions a feature space is specified
by the KKT condition in Eq. (\ref{KKTE5}) and the KKT condition of
complementary slackness.

\subsection{KKT Condition of Complementary Slackness}

The KKT condition of complementary slackness requires that for all constraints
that are not active, where locus equations are \emph{ill-defined}:%
\[
y_{i}\left(  \mathbf{x}_{i}^{T}\boldsymbol{\tau}+\tau_{0}\right)  -1+\xi
_{i}>0
\]
because they are not satisfied as equalities, the corresponding magnitudes
$\psi_{i}$ of the Wolfe dual principal eigenaxis components $\psi
_{i}\overrightarrow{\mathbf{e}}_{i}$ must be zero: $\psi_{i}=0$. Accordingly,
if an inequality is "slack" (not strict), the other inequality cannot be
slack
\citep{Sundaram1996}%
.

Therefore, let there be $l$ active constraints, where $l=l_{1}+l_{2}$. Let
$\xi_{i}=\xi=0$ or $\xi_{i}=\xi\ll1$. The theorem of Karush, Kuhn, and Tucker
provides the guarantee that a Wolf dual linear eigenlocus $\boldsymbol{\psi}$
exists such that the following constraints are satisfied:%
\[
\left\{  \psi_{i\ast}>0\right\}  _{i=1}^{l}\text{,}%
\]
and the following locus equations are satisfied:%
\[
\psi_{i\ast}\left[  y_{i}\left(  \mathbf{x}_{i\ast}^{T}\boldsymbol{\tau}%
+\tau_{0}\right)  -1+\xi_{i}\right]  =0,\ i=1,...,l\text{,}%
\]
where $l$ Wolfe dual principal eigenaxis components $\psi_{i\ast
}\overrightarrow{\mathbf{e}}_{i}$ have non-zero magnitudes $\left\{
\psi_{i\ast}\overrightarrow{\mathbf{e}}_{i}|\psi_{i\ast}>0\right\}  _{i=1}%
^{l}$
\citep{Sundaram1996}%
.

The above condition is known as the \emph{condition of complementary
slackness}. So, in order for the constraint $\psi_{i\ast}>0$ to hold, the
following locus equation must be satisfied:%
\[
y_{i}\left(  \mathbf{x}_{i\ast}^{T}\boldsymbol{\tau}+\tau_{0}\right)
-1+\xi_{i}=0\text{.}%
\]

Accordingly, let there be $l_{1}$ locus equations:%
\[
\mathbf{x}_{1_{i\ast}}^{T}\boldsymbol{\tau}+\tau_{0}+\xi_{i}=1,\ i=1,...,l_{1}%
\text{,}%
\]
where $y_{i}=+1$, and let there be $l_{2}$ locus equations:%
\[
\mathbf{x}_{2_{i\ast}}^{T}\boldsymbol{\tau}+\tau_{0}-\xi_{i}%
=-1,\ i=1,...,l_{2}\text{,}%
\]
where $y_{i}=-1$.

It follows that the linear discriminant function%
\begin{equation}
D\left(  \mathbf{x}\right)  =\boldsymbol{\tau}^{T}\mathbf{x}+\tau_{0}
\label{Discriminant Function}%
\end{equation}
satisfies the set of constraints:%
\[
D_{0}\left(  \mathbf{x}\right)  =0\text{, }D_{+1}\left(  \mathbf{x}\right)
=+1\text{, and }D_{-1}\left(  \mathbf{x}\right)  =-1\text{,}%
\]
where $D_{0}\left(  \mathbf{x}\right)  $ denotes a linear decision boundary,
$D_{+1}\left(  \mathbf{x}\right)  $ denotes a linear decision border for the
$Z_{1}$ decision region, and $D_{-1}\left(  \mathbf{x}\right)  $ denotes a
linear decision border for the $Z_{2}$ decision region.

I will now show that the constraints on the linear discriminant function
$D\left(  \mathbf{x}\right)  =\boldsymbol{\tau}^{T}\mathbf{x}+\tau_{0}$
determine three equations of symmetrical, linear partitioning curves or
surfaces, where all of the points on all three linear loci reference the
constrained, primal linear eigenlocus $\boldsymbol{\tau}$. Returning to Eq.
(\ref{Normal Form Normal Eigenaxis}), recall that the equation of a linear
locus can be written as%
\[
\frac{\mathbf{x}^{T}\boldsymbol{\nu}}{\left\Vert \boldsymbol{\nu}\right\Vert
}=\left\Vert \boldsymbol{\nu}\right\Vert \text{,}%
\]
where the principal eigenaxis $\boldsymbol{\nu}/\left\Vert \boldsymbol{\nu
}\right\Vert $ has length $1$ and points in the direction of a principal
eigenvector $\boldsymbol{\nu}$, and $\left\Vert \boldsymbol{\nu}\right\Vert $
is the distance of a line, plane, or hyperplane to the origin. Any point
$\mathbf{x}$ that satisfies the above equation is on the linear locus of
points specified by $\boldsymbol{\nu}$, where all of the points $\mathbf{x}$
on the linear locus exclusively reference the principal eigenaxis
$\boldsymbol{\nu}$.

I will now use Eq. (\ref{Normal Form Normal Eigenaxis}) and the constraints on
the linear discriminant function in Eq. (\ref{Discriminant Function}) to
devise locus equations that determine the manner in which a constrained,
primal linear eigenlocus partitions any given feature space into congruent
decision regions.

\subsection{Linear Eigenlocus Partitions of Feature Spaces}

I\ will now derive the locus equation of a linear decision boundary
$D_{0}\left(  \mathbf{x}\right)  $.

\subsubsection{Equation of a Linear Decision Boundary $D_{0}\left(
\mathbf{x}\right)  $}

Given Eq. (\ref{Normal Form Normal Eigenaxis}) and the assumption that
$D\left(  \mathbf{x}\right)  =0$, it follows that the linear discriminant
function%
\[
D\left(  \mathbf{x}\right)  =\boldsymbol{\tau}^{T}\mathbf{x}+\tau_{0}%
\]
can be written as:%
\begin{equation}
\frac{\mathbf{x}^{T}\boldsymbol{\tau}}{\left\Vert \boldsymbol{\tau}\right\Vert
}=-\frac{\tau_{0}}{\left\Vert \boldsymbol{\tau}\right\Vert }\text{,}
\label{Decision Boundary}%
\end{equation}
where $\frac{\left\vert \tau_{0}\right\vert }{\left\Vert \boldsymbol{\tau
}\right\Vert }$ is the distance of a linear decision boundary $D_{0}\left(
\mathbf{x}\right)  $ to the origin.

Therefore, any point $\mathbf{x}$ that satisfies Eq. (\ref{Decision Boundary})
is on the linear decision boundary $D_{0}\left(  \mathbf{x}\right)  $, and all
of the points $\mathbf{x}$ on the linear decision boundary $D_{0}\left(
\mathbf{x}\right)  $ exclusively reference the constrained, primal linear
eigenlocus $\boldsymbol{\tau}$. Thereby, the constrained, linear discriminant
function $\boldsymbol{\tau}^{T}\mathbf{x}+\tau_{0}$ satisfies the boundary
value of a linear decision boundary $D_{0}\left(  \mathbf{x}\right)  $:
$\boldsymbol{\tau}^{T}\mathbf{x}+\tau_{0}=0$.

I will now derive the locus equation of the linear decision border
$D_{+1}\left(  \mathbf{x}\right)  $.

\subsubsection{Equation of the $D_{+1}\left(  \mathbf{x}\right)  $ Decision
Border}

Given Eq. (\ref{Normal Form Normal Eigenaxis}) and the assumption that
$D\left(  \mathbf{x}\right)  =1$, it follows that the linear discriminant
function can be written as:%
\begin{equation}
\frac{\mathbf{x}^{T}\boldsymbol{\tau}}{\left\Vert \boldsymbol{\tau}\right\Vert
}=-\frac{\tau_{0}}{\left\Vert \boldsymbol{\tau}\right\Vert }+\frac
{1}{\left\Vert \boldsymbol{\tau}\right\Vert }\text{,}
\label{Decision Border One}%
\end{equation}
where $\frac{\left\vert 1-\tau_{0}\right\vert }{\left\Vert \boldsymbol{\tau
}\right\Vert }$ is the distance of the linear decision border $D_{+1}\left(
\mathbf{x}\right)  $ to the origin.

Therefore, any point $\mathbf{x}$ that satisfies Eq.
(\ref{Decision Border One}) is on the linear decision border $D_{+1}\left(
\mathbf{x}\right)  $, and all of the points $\mathbf{x}$ on the linear
decision border $D_{+1}\left(  \mathbf{x}\right)  $ exclusively reference the
constrained, primal linear eigenlocus $\boldsymbol{\tau}$. Thereby, the
constrained, linear discriminant function $\boldsymbol{\tau}^{T}%
\mathbf{x}+\tau_{0}$ satisfies the boundary value of a linear decision border
$D_{+1}\left(  \mathbf{x}\right)  $: $\boldsymbol{\tau}^{T}\mathbf{x}+\tau
_{0}=1$.

I will now derive the locus equation of the linear decision border
$D_{-1}\left(  \mathbf{x}\right)  $.

\subsubsection{Equation of the $D_{-1}\left(  \mathbf{x}\right)  $ Decision
Border}

Given Eq. (\ref{Normal Form Normal Eigenaxis}) and the assumption that
$D\left(  \mathbf{x}\right)  =-1$, it follows that the linear discriminant
function can be written as:%

\begin{equation}
\frac{\mathbf{x}^{T}\boldsymbol{\tau}}{\left\Vert \boldsymbol{\tau}\right\Vert
}=-\frac{\tau_{0}}{\left\Vert \boldsymbol{\tau}\right\Vert }-\frac
{1}{\left\Vert \boldsymbol{\tau}\right\Vert }\text{,}
\label{Decision Border Two}%
\end{equation}
where $\frac{\left\vert -1-\tau_{0}\right\vert }{\left\Vert \boldsymbol{\tau
}\right\Vert }$ is the distance of the linear decision border $D_{-1}\left(
\mathbf{x}\right)  $ to the origin.

Therefore, any point $\mathbf{x}$ that satisfies Eq.
(\ref{Decision Border Two}) is on the linear decision border $D_{-1}\left(
\mathbf{x}\right)  $, and all of the points $\mathbf{x}$ on the linear
decision border $D_{-1}\left(  \mathbf{x}\right)  $ exclusively reference the
constrained, primal linear eigenlocus $\boldsymbol{\tau}$. Thereby, the
constrained, linear discriminant function $\boldsymbol{\tau}^{T}%
\mathbf{x}+\tau_{0}$ satisfies the boundary value of a linear decision border
$D_{-1}\left(  \mathbf{x}\right)  $: $\boldsymbol{\tau}^{T}\mathbf{x}+\tau
_{0}=-1$.

Given Eqs (\ref{Decision Boundary}) - (\ref{Decision Border Two}), it is
concluded that the constrained, linear discriminant function $D\left(
\mathbf{x}\right)  =\boldsymbol{\tau}^{T}\mathbf{x}+\tau_{0}$ determines
three, symmetrical, linear curves or surfaces, where all of the points on
$D_{0}\left(  \mathbf{x}\right)  $, $D_{+1}\left(  \mathbf{x}\right)  $, and
$D_{-1}\left(  \mathbf{x}\right)  $ exclusively reference the constrained,
primal linear eigenlocus $\boldsymbol{\tau}$.

Moreover, it is concluded that the constrained, linear discriminant function
$D\left(  \mathbf{x}\right)  =\boldsymbol{\tau}^{T}\mathbf{x}+\tau_{0}$
satisfies boundary values for a linear decision boundary $D_{0}\left(
\mathbf{x}\right)  $ and two linear decision borders $D_{+1}\left(
\mathbf{x}\right)  $ and $D_{-1}\left(  \mathbf{x}\right)  $.

I will now use the locus equations of the linear decision borders to derive an
expression for the distance between the decision borders.

\subsubsection{Distance Between the Linear Decision Borders}

Using Eqs (\ref{Decision Border One}) and (\ref{Decision Border Two}), it
follows that the distance between the linear decision borders $D_{+1}\left(
\mathbf{x}\right)  $ and $D_{-1}\left(  \mathbf{x}\right)  $:%
\begin{align}
D_{\left(  D_{+1}\left(  \mathbf{x}\right)  -D_{-1}\left(  \mathbf{x}\right)
\right)  }  &  =\left(  -\frac{\tau_{0}}{\left\Vert \boldsymbol{\tau
}\right\Vert }+\frac{1}{\left\Vert \boldsymbol{\tau}\right\Vert }\right)
\label{Distance Between Decision Borders}\\
&  -\left(  -\frac{\tau_{0}}{\left\Vert \boldsymbol{\tau}\right\Vert }%
-\frac{1}{\left\Vert \boldsymbol{\tau}\right\Vert }\right) \nonumber\\
&  =\frac{2}{\left\Vert \boldsymbol{\tau}\right\Vert }\nonumber
\end{align}
is equal to twice the inverted length of the constrained, primal linear
eigenlocus $\boldsymbol{\tau}$. Therefore, it is concluded that the span of
the constrained geometric region between the linear decision borders is
regulated by the statistic $2\left\Vert \boldsymbol{\tau}\right\Vert ^{-1}$.

I\ will now derive expressions for distances between the linear decision
borders and the linear decision boundary.

\subsubsection{Distances Between Decision Borders and Boundary}

Using Eqs (\ref{Decision Boundary}) and (\ref{Decision Border One}), it
follows that the distance between the linear decision border $D_{+1}\left(
\mathbf{x}\right)  $ and the linear decision boundary $D_{0}\left(
\mathbf{x}\right)  $ is $\frac{1}{\left\Vert \boldsymbol{\tau}\right\Vert }$:%
\begin{align}
D_{\left(  D_{+1}\left(  \mathbf{x}\right)  -D_{0}\left(  \mathbf{x}\right)
\right)  }  &  =\left(  -\frac{\tau_{0}}{\left\Vert \boldsymbol{\tau
}\right\Vert }+\frac{1}{\left\Vert \boldsymbol{\tau}\right\Vert }\right)
\label{Symmetrical Distance Between Border One and Boundary}\\
&  -\left(  -\frac{\tau_{0}}{\left\Vert \boldsymbol{\tau}\right\Vert }\right)
\nonumber\\
&  =\frac{1}{\left\Vert \boldsymbol{\tau}\right\Vert }\text{,}\nonumber
\end{align}
where the linear decision border $D_{+1}\left(  \mathbf{x}\right)  $ and the
linear decision boundary $D_{0}\left(  \mathbf{x}\right)  $ delineate a
constrained geometric region $R_{1}$ in $%
\mathbb{R}
^{d}$.

Using Eqs (\ref{Decision Boundary}) and (\ref{Decision Border Two}), it
follows that the distance between the linear decision boundary $D_{0}\left(
\mathbf{x}\right)  $ and the linear decision border $D_{-1}\left(
\mathbf{x}\right)  $ is also $\frac{1}{\left\Vert \boldsymbol{\tau}\right\Vert
}$:%
\begin{align}
D_{\left(  D_{0}\left(  \mathbf{x}\right)  -D_{-1}\left(  \mathbf{x}\right)
\right)  }  &  =\left(  -\frac{\tau_{0}}{\left\Vert \boldsymbol{\tau
}\right\Vert }\right)
\label{Symmetrical Distance Between Border Two and Boundary}\\
&  -\left(  -\frac{\tau_{0}}{\left\Vert \boldsymbol{\tau}\right\Vert }%
-\frac{1}{\left\Vert \boldsymbol{\tau}\right\Vert }\right) \nonumber\\
&  =\frac{1}{\left\Vert \boldsymbol{\tau}\right\Vert }\text{,}\nonumber
\end{align}
where the linear decision border $D_{-1}\left(  \mathbf{x}\right)  $ and the
linear decision boundary $D_{0}\left(  \mathbf{x}\right)  $ delineate a
constrained geometric region $R_{2}$ in $%
\mathbb{R}
^{d}$.

It follows that the constrained geometric region $R_{1}$ between the linear
decision border $D_{+1}\left(  \mathbf{x}\right)  $ and the linear decision
boundary $D_{0}\left(  \mathbf{x}\right)  $ is congruent to the constrained
geometric region $R_{2}$ between the linear decision boundary $D_{0}\left(
\mathbf{x}\right)  $ and the linear decision border $D_{-1}\left(
\mathbf{x}\right)  $, i.e., $R_{1}\cong R_{2}$.

The equivalent distance of $\frac{1}{\left\Vert \boldsymbol{\tau}\right\Vert
}$ between each linear decision border and the linear decision boundary
reveals that the bilateral symmetry exhibited by the linear decision borders
along the linear decision boundary is regulated by the inverted length
$\left\Vert \boldsymbol{\tau}\right\Vert ^{-1}$ of $\boldsymbol{\tau}$.
Therefore, it is concluded that the spans of the congruent geometric regions
$R_{1}\cong R_{2}$ delineated by the linear decision boundary of Eq.
(\ref{Decision Boundary}) and the linear decision borders of Eqs
(\ref{Decision Border One}) and (\ref{Decision Border Two}) are controlled by
the\ statistic $\left\Vert \boldsymbol{\tau}\right\Vert ^{-1}$.

\subsection{Eigenaxis of Symmetry}

It has been shown that a constrained, linear discriminant function $D\left(
\mathbf{x}\right)  =\boldsymbol{\tau}^{T}\mathbf{x}+\tau_{0}$ determines
three, symmetrical, linear partitioning curves or surfaces, where all of the
points on a linear decision boundary $D_{0}\left(  \mathbf{x}\right)  $ and
linear decision borders $D_{+1}\left(  \mathbf{x}\right)  $ and $D_{-1}\left(
\mathbf{x}\right)  $ exclusively reference a constrained, primal linear
eigenlocus $\boldsymbol{\tau}$. Using Eqs
(\ref{Distance Between Decision Borders}) -
(\ref{Symmetrical Distance Between Border Two and Boundary}), it follows that
$\boldsymbol{\tau}$ is an eigenaxis of symmetry which delineates congruent
decision regions $Z_{1}\cong Z_{2}$ that are symmetrically partitioned by a
linear decision boundary, where the span of both decision regions is regulated
by the\ inverted length\textit{\ }$\left\Vert \boldsymbol{\tau}\right\Vert
^{-1}$ of $\boldsymbol{\tau}$.

\subsubsection{New Notation and Terminology}

I\ will show that the \emph{constrained}, linear eigenlocus discriminant
function $D\left(  \mathbf{x}\right)  =\boldsymbol{\tau}^{T}\mathbf{x}%
+\tau_{0}$ determines a discrete, linear \emph{classification system}
$\boldsymbol{\tau}^{T}\mathbf{x}+\tau_{0}\overset{\omega_{1}}{\underset{\omega
_{2}}{\gtrless}}0$, where $\boldsymbol{\tau=\tau}_{1}-\boldsymbol{\tau}_{2}$
is the \emph{likelihood ratio} of the classification system. Define the
\emph{focus} of the linear classification system $\boldsymbol{\tau}%
^{T}\mathbf{x}+\tau_{0}\overset{\omega_{1}}{\underset{\omega_{2}}{\gtrless}}0$
to be an equilibrium \emph{point} that defines linear decision borders
$D_{+1}\left(  \mathbf{x}\right)  $ and $D_{-1}\left(  \mathbf{x}\right)  $
located at equal distances from a linear decision boundary $D_{0}\left(
\mathbf{x}\right)  $.

Therefore, let $\widetilde{\Lambda}_{\boldsymbol{\tau}}\left(  \mathbf{x}%
\right)  =\boldsymbol{\tau}^{T}\mathbf{x}+\tau_{0}$ denote a linear eigenlocus
discriminant function and let $\widehat{\Lambda}_{\boldsymbol{\tau}}\left(
\mathbf{x}\right)  =\boldsymbol{\tau}_{1}-\boldsymbol{\tau}_{2}$ denote the
likelihood ratio of the linear classification system $\boldsymbol{\tau}%
^{T}\mathbf{x}+\tau_{0}\overset{\omega_{1}}{\underset{\omega_{2}}{\gtrless}}0$
which is a likelihood ratio test $\widetilde{\Lambda}_{\boldsymbol{\tau}%
}\left(  \mathbf{x}\right)  \overset{\omega_{1}}{\underset{\omega
_{2}}{\gtrless}}0$. The likelihood ratio $\widehat{\Lambda}_{\boldsymbol{\tau
}}\left(  \mathbf{x}\right)  =\boldsymbol{\tau}_{1}-\boldsymbol{\tau}_{2}$ is
said to be the primary \emph{focus} of the linear classification system
$\boldsymbol{\tau}^{T}\mathbf{x}+\tau_{0}\overset{\omega_{1}}{\underset{\omega
_{2}}{\gtrless}}0$.

\begin{figure}[ptb]%
\centering
\fbox{\includegraphics[
height=2.5875in,
width=3.4411in
]%
{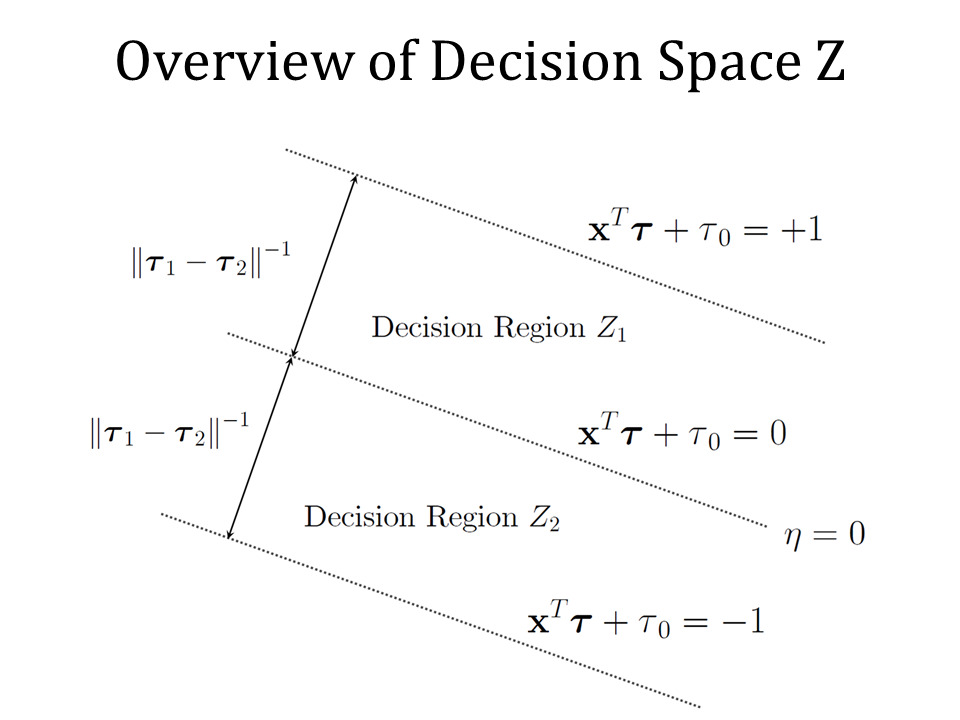}%
}\caption{Linear eigenlocus transforms generate a dual locus of principal
eigenaxis components and likelihoods $\boldsymbol{\tau}=\boldsymbol{\tau}%
_{1}-\boldsymbol{\tau}_{2}$: the basis of a linear classification system
$\protect\widetilde{\Lambda}_{\boldsymbol{\tau}}\left(  \mathbf{x}\right)
\protect\overset{\omega_{1}}{\protect\underset{\omega_{2}}{\gtrless}}0$ which
determines congruent decision regions $Z_{1}\protect\cong Z_{2}$ that will be
shown to have symmetrically balanced forces associated with counter risks and
risks: $\overline{\mathfrak{R}}_{\mathfrak{\min}}\left(  Z_{1}|\omega
_{1}\right)  -\mathfrak{R}_{\mathfrak{\min}}\left(  Z_{1}|\omega_{2}\right)
\rightleftharpoons\overline{\mathfrak{R}}_{\mathfrak{\min}}\left(
Z_{2}|\omega_{2}\right)  -\mathfrak{R}_{\mathfrak{\min}}\left(  Z_{2}%
|\omega_{1}\right)  $.}%
\label{Decision Space for Linear Eigenlocus Transforms}%
\end{figure}

Figure $\ref{Decision Space for Linear Eigenlocus Transforms}$ illustrates how
a constrained, linear discriminant function $\widetilde{\Lambda}%
_{\boldsymbol{\tau}}\left(  \mathbf{x}\right)  =\boldsymbol{\tau}%
^{T}\mathbf{x}+\tau_{0}$ determines three, symmetrical, linear partitioning
curves or surfaces which delineate congruent decision regions $Z_{1}\cong
Z_{2}$ that will be shown to have symmetrically balanced forces associated
with counter risks and risks: $\overline{\mathfrak{R}}_{\mathfrak{\min}%
}\left(  Z_{1}|\omega_{1}\right)  -\mathfrak{R}_{\mathfrak{\min}}\left(
Z_{1}|\omega_{2}\right)  \rightleftharpoons\overline{\mathfrak{R}%
}_{\mathfrak{\min}}\left(  Z_{2}|\omega_{2}\right)  -\mathfrak{R}%
_{\mathfrak{\min}}\left(  Z_{2}|\omega_{1}\right)  $.%

Let $Z$ denote the decision space determined by the decision regions $Z_{1}$
and $Z_{2}$, where $Z\subset%
\mathbb{R}
^{d}$, $Z=Z_{1}+Z_{2}$, $Z_{1}\cong Z_{2}$, and $Z_{1}$ and $Z_{2}$ are
contiguous. I will now demonstrate that the distance between the loci of the
constrained, primal linear eigenlocus components $\boldsymbol{\tau}_{1}$ and
$\boldsymbol{\tau}_{2}$ regulates the span of the decision space $Z$ between
the linear decision borders $D_{+1}\left(  \mathbf{x}\right)  $ and
$D_{-1}\left(  \mathbf{x}\right)  $.

\subsection{Regulation of the Decision Space $Z$}

Substitution of the expression for $\boldsymbol{\tau}$ in Eq.
(\ref{Pair of Normal Eigenlocus Components}) into Eq.
(\ref{Distance Between Decision Borders}) provides an expression for the span
of the decision space $Z$ between the linear decision borders $D_{+1}\left(
\mathbf{x}\right)  $ and $D_{-1}\left(  \mathbf{x}\right)  $:%
\begin{equation}
Z\propto\frac{2}{\left\Vert \boldsymbol{\tau}_{1}-\boldsymbol{\tau}%
_{2}\right\Vert }\text{,} \label{Width of Linear Decision Region}%
\end{equation}
where the constrained width of the decision space $Z$ is equal to twice the
inverted magnitude of the vector difference of $\boldsymbol{\tau}_{1}$ and
$\boldsymbol{\tau}_{2}$. Thus, the span of the decision space $Z$ between the
linear borders $D_{+1}\left(  \mathbf{x}\right)  $ and $D_{-1}\left(
\mathbf{x}\right)  $ is inversely proportional to the distance between the
loci of $\boldsymbol{\tau}_{1}$ and $\boldsymbol{\tau}_{2}$.

Therefore, the distance between the linear decision borders is regulated by
the magnitudes and the directions of the constrained, primal linear eigenlocus
components $\boldsymbol{\tau}_{1}$ and $\boldsymbol{\tau}_{2}$ on
$\boldsymbol{\tau}$. Using the algebraic and geometric relationships in Eq.
(\ref{Inner Product Statistic}) depicted in Fig.
$\ref{Second-order Distance Statisitcs}$a, it follows that the span of the
decision space $Z$ is regulated by the statistic $2\left(  \left\Vert
\boldsymbol{\tau}_{1}\right\Vert \left\Vert \boldsymbol{\tau}_{2}\right\Vert
\cos\theta_{\boldsymbol{\tau}_{1}\boldsymbol{\tau}_{2}}\right)  ^{-1}$, where
$\theta_{\boldsymbol{\tau}_{1}\boldsymbol{\tau}_{2}}$ is the angle between
$\boldsymbol{\tau}_{1}$ and $\boldsymbol{\tau}_{2}$.

\subsubsection{Regulation of the Decision Regions $Z_{1}$ and $Z_{2}$}

The distance between the loci of $\boldsymbol{\tau}_{1}$ and $\boldsymbol{\tau
}_{2}$ also regulates the span of the decision regions between the linear
decision boundary $D_{0}\left(  \mathbf{x}\right)  $ and the linear decision
borders $D_{+1}\left(  \mathbf{x}\right)  $ and $D_{-1}\left(  \mathbf{x}%
\right)  $. Substitution of the expression for $\boldsymbol{\tau}$ in Eq.
(\ref{Pair of Normal Eigenlocus Components}) into Eq.
(\ref{Symmetrical Distance Between Border One and Boundary}) provides an
expression for the span of the decision region $Z_{1}$ between the linear
decision border $D_{+1}\left(  \mathbf{x}\right)  $ and the linear decision
boundary $D_{0}\left(  \mathbf{x}\right)  $:%
\begin{align}
Z_{1}  &  \propto\left(  -\frac{\tau_{0}}{\left\Vert \boldsymbol{\tau}%
_{1}-\boldsymbol{\tau}_{2}\right\Vert }+\frac{1}{\left\Vert \boldsymbol{\tau
}_{1}-\boldsymbol{\tau}_{2}\right\Vert }\right)
\label{Large Covariance Region One}\\
&  -\left(  -\frac{\tau_{0}}{\left\Vert \boldsymbol{\tau}_{1}-\boldsymbol{\tau
}_{2}\right\Vert }\right) \nonumber\\
&  =\frac{1}{\left\Vert \boldsymbol{\tau}_{1}-\boldsymbol{\tau}_{2}\right\Vert
}\text{,}\nonumber
\end{align}
where the span of the decision region $Z_{1}$ satisfies the statistic
$\left\Vert \boldsymbol{\tau}_{1}-\boldsymbol{\tau}_{2}\right\Vert ^{-1}$.

The span of the decision region $Z_{2}$ between the linear decision boundary
$D_{0}\left(  \mathbf{x}\right)  $ and the linear decision border
$D_{-1}\left(  \mathbf{x}\right)  $:%
\begin{align}
Z_{2}  &  \propto\left(  -\frac{\tau_{0}}{\left\Vert \boldsymbol{\tau}%
_{1}-\boldsymbol{\tau}_{2}\right\Vert }\right)
\label{Large Covariance Region Two}\\
&  -\left(  -\frac{\tau_{0}}{\left\Vert \boldsymbol{\tau}_{1}-\boldsymbol{\tau
}_{2}\right\Vert }-\frac{1}{\left\Vert \boldsymbol{\tau}_{1}-\boldsymbol{\tau
}_{2}\right\Vert }\right) \nonumber\\
&  =\frac{1}{\left\Vert \boldsymbol{\tau}_{1}-\boldsymbol{\tau}_{2}\right\Vert
}\nonumber
\end{align}
also satisfies the statistic $\left\Vert \boldsymbol{\tau}_{1}%
-\boldsymbol{\tau}_{2}\right\Vert ^{-1}$.

Therefore, the span of the congruent decision regions $Z_{1}\cong Z_{2}$
between the linear decision boundary and the linear decision borders is
inversely proportional to the magnitude of the vector difference of
$\boldsymbol{\tau}_{1}$ and $\boldsymbol{\tau}_{2}$:%
\[
\frac{1}{\left\Vert \boldsymbol{\tau}_{1}-\boldsymbol{\tau}_{2}\right\Vert
}\text{,}%
\]
where $\left\Vert \boldsymbol{\tau}_{1}-\boldsymbol{\tau}_{2}\right\Vert $ is
the distance between the loci of $\boldsymbol{\tau}_{1}$ and $\boldsymbol{\tau
}_{2}$. Thereby, the widths of the decision regions $Z_{1}$ and $Z_{2}$ are
regulated by the statistic $\frac{1}{\left\Vert \boldsymbol{\tau}%
_{1}-\boldsymbol{\tau}_{2}\right\Vert }$.

\subsection{The Linear Eigenlocus Test}

I\ will now derive a statistic for the $\tau_{0}$ term in Eq.
(\ref{Discriminant Function}). I\ will use the statistic to derive a
likelihood statistic that is the basis of a linear eigenlocus decision rule.

\subsubsection{Estimate for the $\tau_{0}$ Term}

Using the KKT condition in Eq. (\ref{KKTE5}) and the KKT condition of
complementary slackness, it follows that the following set of locus equations
must be satisfied:%
\[
y_{i}\left(  \mathbf{x}_{i\ast}^{T}\boldsymbol{\tau}+\tau_{0}\right)
-1+\xi_{i}=0,\ i=1,...,l\text{,}%
\]
such that an estimate for $\tau_{0}$ satisfies the statistic:%
\begin{align}
\tau_{0}  &  =\sum\nolimits_{i=1}^{l}y_{i}\left(  1-\xi_{i}\right)
-\sum\nolimits_{i=1}^{l}\mathbf{x}_{i\ast}^{T}\boldsymbol{\tau}%
\label{Normal Eigenlocus Projection Factor}\\
&  =\sum\nolimits_{i=1}^{l}y_{i}\left(  1-\xi_{i}\right)  -\left(
\sum\nolimits_{i=1}^{l}\mathbf{x}_{i\ast}\right)  ^{T}\boldsymbol{\tau
}\text{.}\nonumber
\end{align}

I\ will now use the statistic for $\tau_{0}$ to derive a vector expression for
a linear eigenlocus test that is used to classify unknown pattern vectors. Let
$\widehat{\mathbf{x}}_{i\ast}\triangleq\sum\nolimits_{i=1}^{l}\mathbf{x}%
_{i\ast}$.

Substitution of the statistic for $\tau_{0}$ in Eq.
(\ref{Normal Eigenlocus Projection Factor}) into the expression for the
discriminant function in Eq. (\ref{Discriminant Function}) provides a linear
eigenlocus test $\widetilde{\Lambda}_{\boldsymbol{\tau}}\left(  \mathbf{x}%
\right)  \overset{\omega_{1}}{\underset{\omega_{2}}{\gtrless}}0$ for
classifying an unknown pattern vector $\mathbf{x}$:%
\begin{align}
\widetilde{\Lambda}_{\boldsymbol{\tau}}\left(  \mathbf{x}\right)   &  =\left(
\mathbf{x}-\widehat{\mathbf{x}}_{i\ast}\right)  ^{T}\boldsymbol{\tau
}\label{NormalEigenlocusTestStatistic}\\
&  \mathbf{+}\sum\nolimits_{i=1}^{l}y_{i}\left(  1-\xi_{i}\right)
\overset{\omega_{1}}{\underset{\omega_{2}}{\gtrless}}0\text{,}\nonumber
\end{align}
where the statistic $\widehat{\mathbf{x}}_{i\ast}$ is the locus of the
aggregate or cluster $\sum\nolimits_{i=1}^{l}\mathbf{x}_{i\ast}$ of a set of
$l$ extreme points, and the statistic $\sum\nolimits_{i=1}^{l}y_{i}\left(
1-\xi_{i}\right)  $ accounts for the class membership of the primal principal
eigenaxis components on $\boldsymbol{\tau}_{1}$ and $\boldsymbol{\tau}_{2}$.

\subsubsection{Locus of Aggregated Risk $\protect\widehat{\mathfrak{R}}$}

The cluster $\sum\nolimits_{i=1}^{l}\mathbf{x}_{i\ast}$ of a set of extreme
points represents the aggregated risk $\widehat{\mathfrak{R}}$ for a decision
space $Z$. Accordingly, the vector transform $\mathbf{x}-\widehat{\mathbf{x}%
}_{i\ast}$ accounts for the distance between the unknown vector $\mathbf{x}$
and the locus of aggregated risk $\widehat{\mathfrak{R}}$.

I will now derive a vector expression that provides geometric insight into how
the linear test $\widetilde{\Lambda}_{\boldsymbol{\tau}}\left(  \mathbf{x}%
\right)  \overset{\omega_{1}}{\underset{\omega_{2}}{\gtrless}}0$ in Eq.
(\ref{NormalEigenlocusTestStatistic}) makes a decision.

\subsection{Linear Decision Locus}

Denote a unit linear eigenlocus $\boldsymbol{\tau}\mathbf{/}\left\Vert
\boldsymbol{\tau}\right\Vert $ by $\widehat{\boldsymbol{\tau}}$. Letting
$\boldsymbol{\tau=\tau}\mathbf{/}\left\Vert \boldsymbol{\tau}\right\Vert $ in
Eq. (\ref{NormalEigenlocusTestStatistic}) provides an expression for a
decision locus%
\begin{align}
\widehat{D}\left(  \mathbf{x}\right)   &  =\left(  \mathbf{x}%
-\widehat{\mathbf{x}}_{i\ast}\right)  ^{T}\boldsymbol{\tau}\mathbf{/}%
\left\Vert \boldsymbol{\tau}\right\Vert
\label{Statistical Locus of Category Decision}\\
&  \mathbf{+}\frac{1}{\left\Vert \boldsymbol{\tau}\right\Vert }\sum
\nolimits_{i=1}^{l}y_{i}\left(  1-\xi_{i}\right) \nonumber
\end{align}
which is determined by the scalar projection of $\mathbf{x}%
-\widehat{\mathbf{x}}_{i\ast}$ onto $\widehat{\boldsymbol{\tau}}$.
Accordingly, the component of $\mathbf{x}-\widehat{\mathbf{x}}_{i\ast}$ along
$\widehat{\boldsymbol{\tau}}$ specifies a signed magnitude $\left\Vert
\mathbf{x}-\widehat{\mathbf{x}}_{i\ast}\right\Vert \cos\theta$ along the axis
of $\widehat{\boldsymbol{\tau}}$, where $\theta$ is the angle between the
transformed vector $\mathbf{x}-\widehat{\mathbf{x}}_{i\ast}$ and
$\widehat{\boldsymbol{\tau}}$.

It follows that the component $\operatorname{comp}%
_{\overrightarrow{\widehat{\boldsymbol{\tau}}}}\left(  \overrightarrow{\left(
\mathbf{x}-\widehat{\mathbf{x}}_{i\ast}\right)  }\right)  $ of the vector
transform $\mathbf{x}-\widehat{\mathbf{x}}_{i\ast}$ of an unknown pattern
vector $\mathbf{x}$ along the axis of a unit linear eigenlocus
$\widehat{\boldsymbol{\tau}}$%
\[
P_{\widehat{D}\left(  \mathbf{x}\right)  }=\operatorname{comp}%
_{\overrightarrow{\widehat{\boldsymbol{\tau}}}}\left(  \overrightarrow{\left(
\mathbf{x}-\widehat{\mathbf{x}}_{i\ast}\right)  }\right)  =\left\Vert
\mathbf{x}-\widehat{\mathbf{x}}_{i\ast}\right\Vert \cos\theta
\]
specifies a locus $P_{\widehat{D}\left(  \mathbf{x}\right)  }$ of a category
decision, where $P_{\widehat{D}\left(  \mathbf{x}\right)  }$ is at a distance
of $\left\Vert \mathbf{x}-\widehat{\mathbf{x}}_{i\ast}\right\Vert \cos\theta$
from the origin, along the axis of a linear eigenlocus $\boldsymbol{\tau}$.

\begin{figure}[ptb]%
\centering
\fbox{\includegraphics[
height=2.5875in,
width=3.4411in
]%
{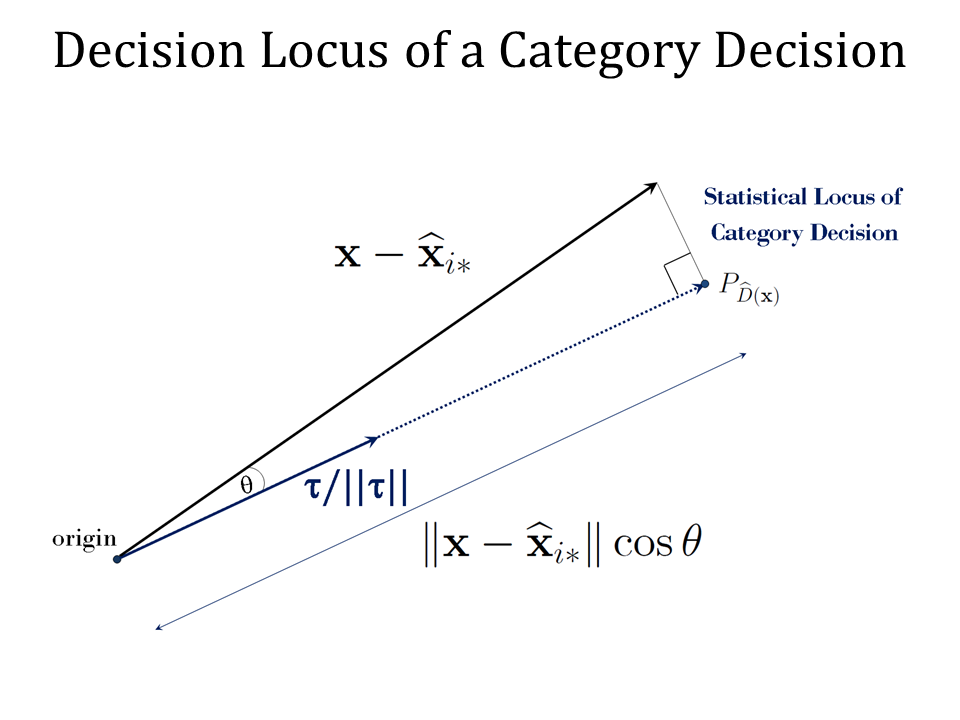}%
}\caption{Illustration of a statistical decision locus $P_{\protect\widehat{D}%
\left(  \mathbf{x}\right)  }$ for an unknown, transformed pattern vector
$\mathbf{x}-E[\mathbf{x}_{i\ast}]$ that is projected onto $\boldsymbol{\tau
}\mathbf{/}\left\Vert \boldsymbol{\tau}\right\Vert $.}%
\label{Statistical Decision Locus}%
\end{figure}

Figure $\ref{Statistical Decision Locus}$ depicts a decision locus generated
by the linear eigenlocus test $\widetilde{\Lambda}_{\boldsymbol{\tau}}\left(
\mathbf{x}\right)  \overset{\omega_{1}}{\underset{\omega_{2}}{\gtrless}}0$ in
Eq. (\ref{NormalEigenlocusTestStatistic}).%

The above expression for a decision locus $P_{\widehat{D}\left(
\mathbf{x}\right)  }$ provides geometric insight into how the linear
discriminant function $\widetilde{\Lambda}_{\boldsymbol{\tau}}\left(
\mathbf{x}\right)  =\boldsymbol{\tau}^{T}\mathbf{x}+\tau_{0}$ in Eq.
(\ref{Discriminant Function}) assigns an unknown pattern vector to a pattern
class. Using Eqs (\ref{Decision Boundary}), (\ref{Decision Border One}),
(\ref{Decision Border Two}), and (\ref{Statistical Locus of Category Decision}%
), it follows that the linear discriminant function in Eq.
(\ref{Discriminant Function}) generates a decision locus $P_{\widehat{D}%
\left(  \mathbf{x}\right)  }$ which lies in a geometric region that is either
$\left(  1\right)  $ inside or bordering one of the decision regions $Z_{1}$
and $Z_{2}$ depicted in Fig.
$\ref{Decision Space for Linear Eigenlocus Transforms}$, $\left(  2\right)  $
on the other side of the linear decision border $D_{+1}\left(  \mathbf{x}%
\right)  $, where $\boldsymbol{\tau}^{T}\mathbf{x}+\tau_{0}=+1$, or $\left(
3\right)  $ on the other side of the linear decision border $D_{-1}\left(
\mathbf{x}\right)  $, where $\boldsymbol{\tau}^{T}\mathbf{x}+\tau_{0}=-1$.

Using Eqs (\ref{NormalEigenlocusTestStatistic}) and
(\ref{Statistical Locus of Category Decision}), it follows that the linear
discriminant function%
\[
D\left(  \mathbf{x}\right)  =\operatorname{comp}%
_{\overrightarrow{\widehat{\boldsymbol{\tau}}}}\left(  \overrightarrow{\left(
\mathbf{x}-\widehat{\mathbf{x}}_{i\ast}\right)  }\right)  +\frac{1}{\left\Vert
\boldsymbol{\tau}\right\Vert }\sum\nolimits_{i=1}^{l}y_{i}\left(  1-\xi
_{i}\right)  \overset{\omega_{1}}{\underset{\omega_{2}}{\gtrless}}0\text{,}%
\]
where $\boldsymbol{\tau=\tau}\mathbf{/}\left\Vert \boldsymbol{\tau}\right\Vert
$, generates an output based on the decision locus $\operatorname{comp}%
_{\overrightarrow{\widehat{\boldsymbol{\tau}}}}\left(  \mathbf{x}%
-\overline{\mathbf{x}}_{i\ast}\right)  $ and the class membership statistic
$\frac{1}{\left\Vert \boldsymbol{\tau}\right\Vert }\sum\nolimits_{i=1}%
^{l}y_{i}\left(  1-\xi_{i}\right)  $.

\subsubsection{Linear Decision Threshold}

Returning to Eq. (\ref{General Form of Decision Function II}), recall that an
optimal decision function computes the likelihood ratio $\Lambda\left(
\mathbf{x}\right)  $ for a feature vector $\mathbf{x}$ and makes a decision by
comparing the ratio $\Lambda\left(  \mathbf{x}\right)  $ to the threshold
$\eta=0$. Given Eqs (\ref{Decision Boundary}) and
(\ref{Statistical Locus of Category Decision}), it follows that a linear
eigenlocus test $\widetilde{\Lambda}_{\boldsymbol{\tau}}\left(  \mathbf{x}%
\right)  \overset{\omega_{1}}{\underset{\omega_{2}}{\gtrless}}0$ makes a
decision by comparing the output%
\[
\operatorname{sign}\left(  \operatorname{comp}%
_{\overrightarrow{\widehat{\boldsymbol{\tau}}}}\left(  \overrightarrow{\left(
\mathbf{x}-\widehat{\mathbf{x}}_{i\ast}\right)  }\right)  +\frac{1}{\left\Vert
\boldsymbol{\tau}\right\Vert }\sum\nolimits_{i=1}^{l}y_{i}\left(  1-\xi
_{i}\right)  \right)  \text{,}%
\]
where $\operatorname{sign}\left(  x\right)  \equiv\frac{x}{\left\vert
x\right\vert }$ for $x\neq0$, to a threshold $\eta$ along the axis of
$\widehat{\boldsymbol{\tau}}$ in $%
\mathbb{R}
^{d}$, where $\eta=0$.

\subsection{Linear Eigenlocus Decision Rules}

Substitution of the expression for $\boldsymbol{\tau}$ in Eq.
(\ref{Pair of Normal Eigenlocus Components}) into Eq.
(\ref{NormalEigenlocusTestStatistic}) provides a linear eigenlocus test in
terms of the primal eigenlocus components $\boldsymbol{\tau}_{1}$ and
$\boldsymbol{\tau}_{2}$:%
\begin{align}
\widetilde{\Lambda}_{\boldsymbol{\tau}}\left(  \mathbf{x}\right)   &  =\left(
\mathbf{x}-\sum\nolimits_{i=1}^{l}\mathbf{x}_{i\ast}\right)  ^{T}%
\boldsymbol{\tau}_{1}\label{NormalEigenlocusTestStatistic2}\\
&  -\left(  \mathbf{x}-\sum\nolimits_{i=1}^{l}\mathbf{x}_{i\ast}\right)
^{T}\boldsymbol{\tau}_{2}\nonumber\\
&  \mathbf{+}\sum\nolimits_{i=1}^{l}y_{i}\left(  1-\xi_{i}\right)
\overset{\omega_{1}}{\underset{\omega_{2}}{\gtrless}}0\text{.}\nonumber
\end{align}

I will show that a constrained, primal linear eigenlocus $\boldsymbol{\tau}$
and its Wolfe dual $\boldsymbol{\psi}$ possess an essential statistical
property which enables linear eigenlocus discriminant functions
$\widetilde{\Lambda}_{\boldsymbol{\tau}}\left(  \mathbf{x}\right)
=\boldsymbol{\tau}^{T}\mathbf{x}+\tau_{0}$ to satisfy a discrete and
data=driven version of the fundamental integral equation of binary
classification:%
\begin{align*}
f\left(  \widetilde{\Lambda}_{\boldsymbol{\tau}}\left(  \mathbf{x}\right)
\right)   &  =\;\int\nolimits_{Z_{1}}p\left(  \mathbf{x}_{1_{i\ast}%
}|\boldsymbol{\tau}_{1}\right)  d\boldsymbol{\tau}_{1}+\int\nolimits_{Z_{2}%
}p\left(  \mathbf{x}_{1_{i\ast}}|\boldsymbol{\tau}_{1}\right)
d\boldsymbol{\tau}_{1}+\delta\left(  y\right)  \sum\nolimits_{i=1}^{l_{1}}%
\psi_{1_{i_{\ast}}}\\
&  =\int\nolimits_{Z_{1}}p\left(  \mathbf{x}_{2_{i\ast}}|\boldsymbol{\tau}%
_{2}\right)  d\boldsymbol{\tau}_{2}+\int\nolimits_{Z_{2}}p\left(
\mathbf{x}_{2_{i\ast}}|\boldsymbol{\tau}_{2}\right)  d\boldsymbol{\tau}%
_{2}-\delta\left(  y\right)  \sum\nolimits_{i=1}^{l_{2}}\psi_{2_{i_{\ast}}%
}\text{,}%
\end{align*}
over the decision space $Z=Z_{1}+Z_{2}$, where $\delta\left(  y\right)
\triangleq\sum\nolimits_{i=1}^{l}y_{i}\left(  1-\xi_{i}\right)  $, and all of
the forces associated with counter risks $\overline{\mathfrak{R}%
}_{\mathfrak{\min}}\left(  Z_{1}|\boldsymbol{\tau}_{1}\right)  $ and
$\overline{\mathfrak{R}}_{\mathfrak{\min}}\left(  Z_{2}|\boldsymbol{\tau}%
_{2}\right)  $ and risks $\mathfrak{R}_{\mathfrak{\min}}\left(  Z_{1}%
|\boldsymbol{\tau}_{2}\right)  $ and $\mathfrak{R}_{\mathfrak{\min}}\left(
Z_{2}|\boldsymbol{\tau}_{1}\right)  $ within the $Z_{1}$ and $Z_{2}$ decision
regions are symmetrically balanced with each other:%
\begin{align*}
f\left(  \widetilde{\Lambda}_{\boldsymbol{\tau}}\left(  \mathbf{x}\right)
\right)   &  :\;\int\nolimits_{Z_{1}}p\left(  \mathbf{x}_{1_{i\ast}%
}|\boldsymbol{\tau}_{1}\right)  d\boldsymbol{\tau}_{1}-\int\nolimits_{Z_{1}%
}p\left(  \mathbf{x}_{2_{i\ast}}|\boldsymbol{\tau}_{2}\right)
d\boldsymbol{\tau}_{2}+\delta\left(  y\right)  \sum\nolimits_{i=1}^{l_{1}}%
\psi_{1_{i_{\ast}}}\\
&  =\int\nolimits_{Z_{2}}p\left(  \mathbf{x}_{2_{i\ast}}|\boldsymbol{\tau}%
_{2}\right)  d\boldsymbol{\tau}_{2}-\int\nolimits_{Z_{2}}p\left(
\mathbf{x}_{1_{i\ast}}|\boldsymbol{\tau}_{1}\right)  d\boldsymbol{\tau}%
_{1}-\delta\left(  y\right)  \sum\nolimits_{i=1}^{l_{2}}\psi_{2_{i_{\ast}}}%
\end{align*}
by means of an integral equation:%
\begin{align*}
f\left(  \widetilde{\Lambda}_{\boldsymbol{\tau}}\left(  \mathbf{x}\right)
\right)   &  =\int\nolimits_{Z}p\left(  \mathbf{x}_{1_{i\ast}}%
|\boldsymbol{\tau}_{1}\right)  d\boldsymbol{\tau}_{1}=\left\Vert
\boldsymbol{\tau}_{1}\right\Vert _{\min_{c}}^{2}+C_{1}\\
&  =\int\nolimits_{Z}p\left(  \mathbf{x}_{2_{i\ast}}|\boldsymbol{\tau}%
_{2}\right)  d\boldsymbol{\tau}_{2}=\left\Vert \boldsymbol{\tau}%
_{2}\right\Vert _{\min_{c}}^{2}+C_{2}\text{,}%
\end{align*}
where $p\left(  \mathbf{x}_{2_{i\ast}}|\boldsymbol{\tau}_{2}\right)  $ and
$p\left(  \mathbf{x}_{1_{i\ast}}|\boldsymbol{\tau}_{1}\right)  $ are
class-conditional densities for respective extreme points $\mathbf{x}%
_{2_{i\ast}}$ and $\mathbf{x}_{1_{i\ast}}$, and $C_{1}$ and $C_{2}$ are
integration constants for $\boldsymbol{\tau}_{1}$ and $\boldsymbol{\tau}_{2}$
respectively. I will identify this essential statistical property after
I\ identify the fundamental properties possessed by a Wolfe dual linear
eigenlocus $\boldsymbol{\psi}$.

\section{The Wolfe Dual Eigenspace I}

Let there be $l$ principal eigenaxis components $\left\{  \psi_{i\ast
}\overrightarrow{\mathbf{e}}_{i}|\psi_{i\ast}>0\right\}  _{i=1}^{l}$ on a
constrained, primal linear eigenlocus within its Wolfe dual eigenspace:%
\[
\max\Xi\left(  \boldsymbol{\psi}\right)  =\mathbf{1}^{T}\boldsymbol{\psi
}-\frac{\boldsymbol{\psi}^{T}\mathbf{Q}\boldsymbol{\psi}}{2}\text{,}%
\]
where the Wolfe dual linear eigenlocus $\boldsymbol{\psi}$ satisfies the
constraints $\boldsymbol{\psi}^{T}\mathbf{y}=0$ and $\psi_{i\ast}>0$.
Quadratic forms $\boldsymbol{\psi}^{T}\mathbf{Q}\boldsymbol{\psi}$ determine
five classes of quadratic surfaces that include $N$-dimensional circles,
ellipses, hyperbolas, parabolas, and lines
\citep{Hewson2009}%
. I\ will now use Rayleigh's principle
\citep[see][]{Strang1986}
and the theorem for convex duality to define the geometric essence of
$\boldsymbol{\psi}$.

Rayleigh's principle guarantees that the quadratic%
\[
r\left(  \mathbf{Q},\mathbf{x}\right)  =\mathbf{x}^{T}\mathbf{Qx}%
\]
is maximized by the largest eigenvector $\mathbf{x}_{1}$, with its maximal
value equal to the largest eigenvalue $\lambda_{1}=\underset{0\neq
\mathbf{x\in}%
\mathbb{R}
^{N}}{\max}r\left(  \mathbf{Q},\mathbf{x}\right)  $. Raleigh's principle can
be used to find principal eigenvectors $\mathbf{x}_{1}$ which satisfy
additional constraints such as $a_{1}x_{1}+\cdots a_{N}x_{N}=c$, for which%
\[
\lambda_{1}=\underset{a_{1}x_{1}+\cdots a_{N}x_{N}=c}{\max}r\left(
\mathbf{Q},\mathbf{x}\right)  \text{.}%
\]

The theorem for convex duality guarantees an equivalence and corresponding
symmetry between a constrained, primal linear eigenlocus $\boldsymbol{\tau}$
and its Wolfe dual $\boldsymbol{\psi}$. Raleigh's principle and the theorem
for convex duality jointly indicate that Eq. (\ref{Vector Form Wolfe Dual})
provides an estimate of the largest eigenvector $\boldsymbol{\psi}$ of a Gram
matrix $\mathbf{Q}$ for which $\boldsymbol{\psi}$ satisfies the constraints
$\boldsymbol{\psi}^{T}\mathbf{y}=0$ and $\psi_{i}\geq0$, such that
$\boldsymbol{\psi}$ is a principal eigenaxis of three, symmetrical\textit{\ }%
hyperplane partitioning surfaces associated with the constrained quadratic
form $\boldsymbol{\psi}^{T}\mathbf{Q}\boldsymbol{\psi}$.

I will now show that maximization of the functional $\mathbf{1}^{T}%
\boldsymbol{\psi}-\boldsymbol{\psi}^{T}\mathbf{Q}\boldsymbol{\psi}\mathbf{/}2
$ requires that $\boldsymbol{\psi}$ satisfy an eigenenergy constraint which is
symmetrically related to the restriction of the primal linear eigenlocus
$\boldsymbol{\tau}$ to its Wolfe dual eigenspace.

\subsection{Eigenenergy Constraint on $\boldsymbol{\psi}$}

Equation (\ref{Minimum Total Eigenenergy Primal Normal Eigenlocus}) and the
theorem for convex duality jointly indicate that $\boldsymbol{\psi}$ satisfies
a critical minimum eigenenergy constraint that is symmetrically related to the
critical minimum eigenenergy constraint on $\boldsymbol{\tau}$:%
\[
\left\Vert \boldsymbol{\psi}\right\Vert _{\min_{c}}^{2}\cong\left\Vert
\boldsymbol{\tau}\right\Vert _{\min_{c}}^{2}\text{.}%
\]

Therefore, a Wolfe dual linear eigenlocus $\boldsymbol{\psi}$ satisfies a
critical minimum eigenenergy constraint%
\[
\max\boldsymbol{\psi}^{T}\mathbf{Q}\boldsymbol{\psi}=\lambda_{\max
\boldsymbol{\psi}}\left\Vert \boldsymbol{\psi}\right\Vert _{\min_{c}}^{2}%
\]
for which the functional $\mathbf{1}^{T}\boldsymbol{\psi}-\boldsymbol{\psi
}^{T}\mathbf{Q}\boldsymbol{\psi}\mathbf{/}2$ in Eq.
(\ref{Vector Form Wolfe Dual}) is maximized by the largest eigenvector
$\boldsymbol{\psi}$ of $\mathbf{Q}$, such that the constrained quadratic form
$\boldsymbol{\psi}^{T}\mathbf{Q}\boldsymbol{\psi}\mathbf{/}2$\textbf{,} where
$\boldsymbol{\psi}^{T}\mathbf{y}=0$ and $\psi_{i}\geq0$, reaches its smallest
possible value. This indicates that principal eigenaxis components on
$\boldsymbol{\psi}$ satisfy minimum length constraints. Principal eigenaxis
components on a Wolfe dual linear eigenlocus $\boldsymbol{\psi}$ also satisfy
an equilibrium constraint.

\subsection{Equilibrium Constraint on $\boldsymbol{\psi}$}

The KKT condition in Eq. (\ref{KKTE2}) requires that the magnitudes of the
Wolfe dual principal eigenaxis components on $\boldsymbol{\psi}$ satisfy the
equation:%
\[
\left(  y_{i}=1\right)  \sum\nolimits_{i=1}^{l_{1}}\psi_{1_{i\ast}}+\left(
y_{i}=-1\right)  \sum\nolimits_{i=1}^{l_{2}}\psi_{2_{i\ast}}=0
\]
so that
\begin{equation}
\sum\nolimits_{i=1}^{l_{1}}\psi_{1_{i\ast}}-\sum\nolimits_{i=1}^{l_{2}}%
\psi_{2_{i\ast}}=0\text{.} \label{Wolfe Dual Equilibrium Point}%
\end{equation}
It follows that the \emph{integrated lengths} of the Wolfe dual principal
eigenaxis components correlated with each pattern category must \emph{balance}
each other:%
\begin{equation}
\sum\nolimits_{i=1}^{l_{1}}\psi_{1_{i\ast}}\rightleftharpoons\sum
\nolimits_{i=1}^{l_{2}}\psi_{2_{i\ast}}\text{.}
\label{Equilibrium Constraint on Dual Eigen-components}%
\end{equation}

Accordingly, let $l_{1}+l_{2}=l$ and express a Wolfe dual linear eigenlocus
$\boldsymbol{\psi}$ in terms of $l$ non-orthogonal unit vectors $\left\{
\overrightarrow{\mathbf{e}}_{1\ast},\ldots,\overrightarrow{\mathbf{e}}_{l\ast
}\right\}  $:%
\begin{align}
\boldsymbol{\psi}  &  =\sum\nolimits_{i=1}^{l}\psi_{i\ast}%
\overrightarrow{\mathbf{e}}_{i\ast}\label{Wolfe Dual Vector Equation}\\
&  =\sum\nolimits_{i=1}^{l_{1}}\psi_{1i\ast}\overrightarrow{\mathbf{e}%
}_{1i\ast}+\sum\nolimits_{i=1}^{l_{2}}\psi_{2i\ast}\overrightarrow{\mathbf{e}%
}_{2i\ast}\nonumber\\
&  =\boldsymbol{\psi}_{1}+\boldsymbol{\psi}_{2}\text{,}\nonumber
\end{align}
where each scaled, non-orthogonal unit vector $\psi_{1i\ast}%
\overrightarrow{\mathbf{e}}_{1i\ast}$ or $\psi_{2i\ast}%
\overrightarrow{\mathbf{e}}_{2i\ast}$ is a \emph{displacement vector} that is
correlated with an $\mathbf{x}_{1_{i\ast}}$ or $\mathbf{x}_{2_{i\ast}}$
extreme vector respectively, $\boldsymbol{\psi}_{1}$ denotes the Wolfe dual
eigenlocus component $\sum\nolimits_{i=1}^{l_{1}}\psi_{1_{i\ast}%
}\overrightarrow{\mathbf{e}}_{1_{i\ast}}$, and $\boldsymbol{\psi}_{2}$ denotes
the Wolfe dual eigenlocus component $\sum\nolimits_{i=1}^{l_{2}}\psi
_{2_{i\ast}}\overrightarrow{\mathbf{e}}_{2_{i\ast}}$. I\ will demonstrate that
a Wolfe dual linear eigenlocus $\boldsymbol{\psi}$ is a displacement vector
that accounts for directions and magnitudes of $\mathbf{x}_{1_{i\ast}}$ and
$\mathbf{x}_{2_{i\ast}}$ extreme vectors.

Given Eq. (\ref{Equilibrium Constraint on Dual Eigen-components}) and data
distributions that have dissimilar covariance matrices, it follows that the
forces associated with counter risks and risks, within each of the congruent
decision regions, are balanced with each other. Given Eq.
(\ref{Equilibrium Constraint on Dual Eigen-components}) and data distributions
that have similar covariance matrices, it follows that the forces associated
with counter risks within each of the congruent decision regions are equal to
each other, and the forces associated with risks within each of the congruent
decision regions are equal to each other.

Given Eqs (\ref{Equilibrium Constraint on Dual Eigen-components}) and
(\ref{Wolfe Dual Vector Equation}), it follows that the axis of a Wolfe dual
linear eigenlocus $\boldsymbol{\psi}$ can be regarded as a lever that is
formed by \emph{sets of principal eigenaxis components which are evenly or
equally distributed over either side of the\emph{ axis }of }$\boldsymbol{\psi
}$\emph{, where a fulcrum is placed directly under the center of the axis of
}$\boldsymbol{\psi}$.

Thereby, the axis of $\boldsymbol{\psi}$ is in statistical equilibrium, where
all of the principal eigenaxis components on $\boldsymbol{\psi}$ are equal or
in correct proportions, relative to the center of $\boldsymbol{\psi}$, such
that the opposing forces associated with risks and counter risks of a linear
classification system are balanced with each other. Figure
$\ref{Linear Dual Locus in Statistical Equilibrium}$ illustrates the axis of
$\mathbf{\psi}$ in statistical equilibrium.\textbf{%
\begin{figure}[ptb]%
\centering
\fbox{\includegraphics[
height=2.5875in,
width=3.4411in
]%
{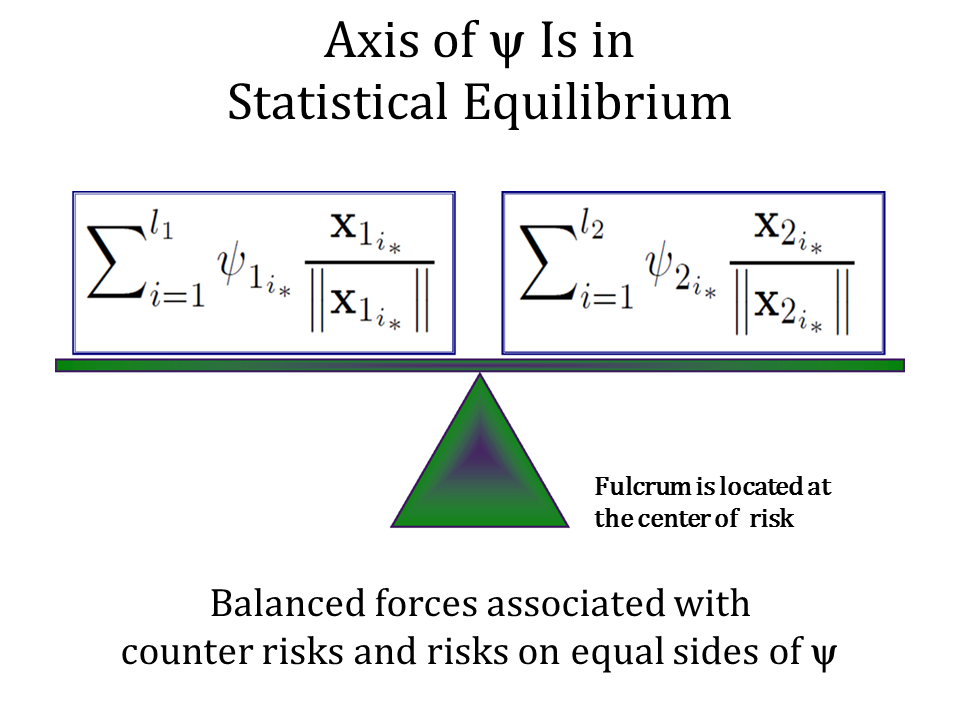}%
}\caption{All of the principal eigenaxis components on $\boldsymbol{\psi}$
have equal or correct proportions, relative to the center of $\boldsymbol{\psi
}$, so that opposing forces associated with risks and counter risks are
symmetrically balanced with each other.}%
\label{Linear Dual Locus in Statistical Equilibrium}%
\end{figure}
}

Using Eqs (\ref{Equilibrium Constraint on Dual Eigen-components}) and
(\ref{Wolfe Dual Vector Equation}), it follows that the length $\left\Vert
\boldsymbol{\psi}_{1}\right\Vert $ of the vector $\boldsymbol{\psi}_{1}$ is
balanced with the length $\left\Vert \boldsymbol{\psi}_{2}\right\Vert $ of the
vector $\boldsymbol{\psi}_{2}$:%
\begin{equation}
\left\Vert \boldsymbol{\psi}_{1}\right\Vert \rightleftharpoons\left\Vert
\boldsymbol{\psi}_{2}\right\Vert \text{,}
\label{Equilibrium Constraint on Dual Component Lengths}%
\end{equation}
and that the total allowed eigenenergies exhibited by $\boldsymbol{\psi}_{1}$
and $\boldsymbol{\psi}_{2}$ are balanced with each other:%
\begin{equation}
\left\Vert \boldsymbol{\psi}_{1}\right\Vert _{\min_{c}}^{2}\rightleftharpoons
\left\Vert \boldsymbol{\psi}_{2}\right\Vert _{\min_{c}}^{2}\text{.}
\label{Symmetrical Balance of Wolf Dual Eigenenergies}%
\end{equation}

Therefore, the equilibrium constraint on $\boldsymbol{\psi}$ in Eq.
(\ref{Equilibrium Constraint on Dual Eigen-components}) ensures that the total
allowed eigenenergies exhibited by the Wolfe dual principal eigenaxis
components on $\boldsymbol{\psi}_{1}$ and $\boldsymbol{\psi}_{2}$ are
symmetrically balanced with each other:%
\[
\left\Vert \sum\nolimits_{i=1}^{l_{1}}\psi_{1_{i\ast}}\right\Vert _{\min_{c}%
}^{2}\rightleftharpoons\left\Vert \sum\nolimits_{i=1}^{l_{2}}\psi_{2_{i\ast}%
}\right\Vert _{\min_{c}}^{2}%
\]
about the center of total allowed eigenenergy $\left\Vert \boldsymbol{\psi
}\right\Vert _{\min_{c}}^{2}$: which is located at the geometric center of
$\boldsymbol{\psi}$ because $\left\Vert \boldsymbol{\psi}_{1}\right\Vert
\equiv\left\Vert \boldsymbol{\psi}_{2}\right\Vert $. This indicates that the
total allowed eigenenergies of $\boldsymbol{\psi}$ are distributed over its
axis in a symmetrically balanced and well-proportioned manner.

\subsection{Symmetrical Balance Exhibited by the Axis of $\boldsymbol{\psi}$}

Given Eqs (\ref{Equilibrium Constraint on Dual Component Lengths}) and
(\ref{Symmetrical Balance of Wolf Dual Eigenenergies}), it follows that the
axis of a Wolfe dual linear eigenlocus $\boldsymbol{\psi}$ can be regarded as
a lever that has \emph{equal weight on equal sides of a centrally placed
fulcrum}.

\begin{figure}[ptb]%
\centering
\fbox{\includegraphics[
height=2.5875in,
width=3.4411in
]%
{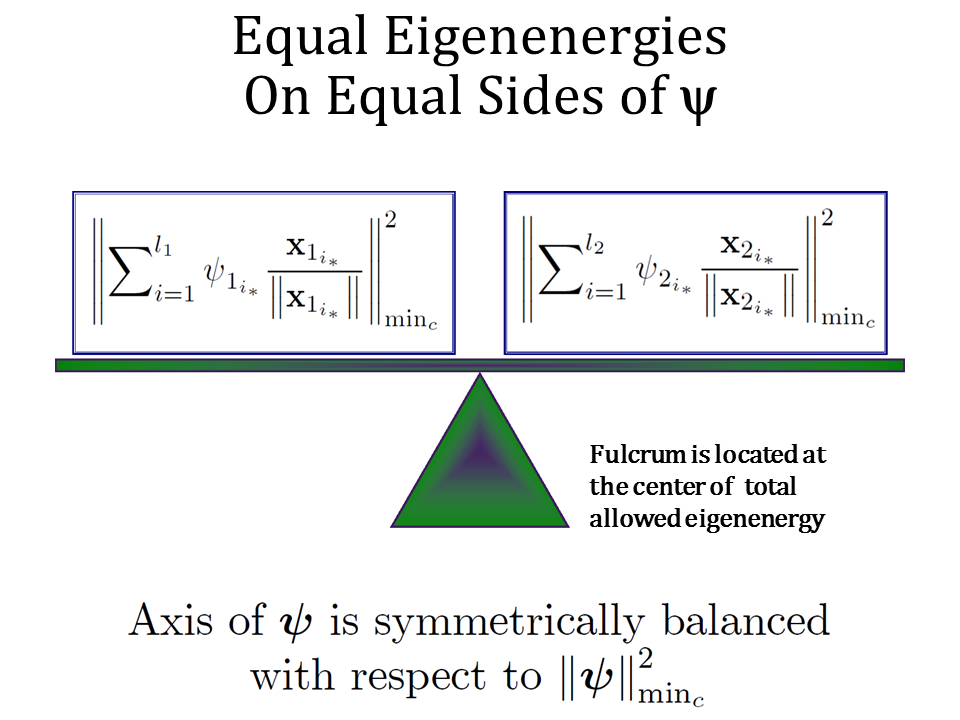}%
}\caption{The axis of $\boldsymbol{\psi}$ can be regarded as a lever that has
an equal distribution of eigenenergies on equal sides of a centrally placed
fulcrum which is located at the center of the total allowed eigenenergy
$\left\Vert \boldsymbol{\psi}\right\Vert _{\min_{c}}^{2}$ of $\boldsymbol{\psi
}$.}%
\label{Symmetrical Balance of Wolfe Dual Linear Eigenlocus}%
\end{figure}

Thereby, the axis of $\boldsymbol{\psi}$ is a lever that has an equal
distribution of eigenenergies on equal sides of a centrally placed fulcrum.
Later on, I\ will show that symmetrically balanced, joint distributions of
principal eigenaxis components on $\boldsymbol{\psi}$ and $\boldsymbol{\tau}$
are symmetrically distributed over the axes of the Wolfe dual principal
eigenaxis components on $\boldsymbol{\psi}$ and the unconstrained, correlated
primal principal eigenaxis components (the extreme vectors) on
$\boldsymbol{\tau}$. Figure\textbf{\ }%
$\ref{Symmetrical Balance of Wolfe Dual Linear Eigenlocus}$ depicts how the
axis of $\boldsymbol{\psi}$ can be regarded as a lever that has an equal
distribution of eigenenergies on equal sides of a centrally placed fulcrum
which is located at the geometric center, i.e., the critical minimum
eigenenergy $\left\Vert \boldsymbol{\psi}\right\Vert _{\min_{c}}^{2}$ of
$\boldsymbol{\psi}$.%

The eigenspectrum of $\mathbf{Q}$ plays a fundamental role in describing the
hyperplane surfaces associated with $\boldsymbol{\psi}$. I\ will now
demonstrate that the eigenspectrum of $\mathbf{Q}$ determines the shapes of
the quadratic surfaces which are specified by the constrained quadratic form
in Eq. (\ref{Vector Form Wolfe Dual}).

\subsection{Eigenspectrum Shaping of Quadratic Surfaces}

Take the standard equation of a quadratic form: $\mathbf{x}^{T}\mathbf{Qx}=1$.
Write $\mathbf{x}$ in terms of an orthogonal basis of unit eigenvectors
$\left\{  \mathbf{v}_{1},\ldots,\mathbf{v}_{N}\right\}  $ so that
$\mathbf{x=}\sum\nolimits_{i=1}^{N}x_{i}\mathbf{v}_{i}$. Substitution of this
expression into $\mathbf{x}^{T}\mathbf{Qx}$%
\[
\mathbf{x}^{T}\mathbf{Qx}=\left(  \sum\nolimits_{i=1}^{N}x_{i}\mathbf{v}%
_{i}\right)  ^{T}\mathbf{Q}\left(  \sum\nolimits_{j=1}^{N}x_{j}\mathbf{v}%
_{j}\right)
\]
produces a simple coordinate form equation of a quadratic surface%
\begin{equation}
\lambda_{1}x_{1}^{2}+\lambda_{2}x_{2}^{2}+\ldots+\lambda_{N}x_{N}^{2}=1
\label{Eigenvalue Coordinate Form Second Order Loci}%
\end{equation}
\emph{solely} in terms of the eigenvalues $\lambda_{N}\leq$ $\lambda
_{N-1}\ldots\leq\lambda_{1}$ of the matrix $\mathbf{Q}$
\citep{Hewson2009}%
.

Equation (\ref{Eigenvalue Coordinate Form Second Order Loci}) reveals that the
\emph{geometric shape} of a quadratic surface is completely determined by the
\emph{eigenvalues} of the matrix associated with a quadratic form. This
general property of quadratic forms will lead to far reaching consequences for
linear and quadratic eigenlocus transforms.

I will now show that the inner product statistics of a training data
collection essentially determine the geometric shapes of the quadratic
surfaces which are specified by the constrained quadratic form in Eq.
(\ref{Vector Form Wolfe Dual}).

Consider a Gram or kernel matrix $\mathbf{Q}$ associated with the constrained
quadratic form in Eq. (\ref{Vector Form Wolfe Dual}). Denote the elements of
the Gram or kernel matrix $\mathbf{Q}$ by $\varphi\left(  \mathbf{x}%
_{i},\mathbf{x}_{j}\right)  $, where $\varphi\left(  \mathbf{x}_{i}%
,\mathbf{x}_{j}\right)  $ denotes an inner product relationship between the
training vectors $\mathbf{x}_{i}$ and $\mathbf{x}_{j}$. The Cayley-Hamilton
theorem provides the result that the eigenvalues $\left\{  \lambda
_{i}\right\}  _{i=1}^{N}\in\Re$ of $\mathbf{Q}$ satisfy the characteristic
equation%
\[
\det\left(  \mathbf{Q-\lambda I}\right)  =0
\]
which is a polynomial of degree $N$. The roots $p\left(  \lambda\right)  =0$
of the characteristic polynomial $p\left(  \lambda\right)  $ of $\mathbf{Q}$:%
\[
\det\left(
\begin{bmatrix}
\varphi\left(  \mathbf{x}_{1},\mathbf{x}_{1}\right)  -\lambda_{1} & \cdots &
\varphi\left(  \mathbf{x}_{1},\mathbf{x}_{N}\right) \\
\varphi\left(  \mathbf{x}_{2},\mathbf{x}_{1}\right)  & \cdots & \varphi\left(
\mathbf{x}_{2},\mathbf{x}_{N}\right) \\
\vdots & \ddots & \vdots\\
\varphi\left(  \mathbf{x}_{N},\mathbf{x}_{1}\right)  & \cdots & \varphi\left(
\mathbf{x}_{N},\mathbf{x}_{N}\right)  -\lambda_{N}%
\end{bmatrix}
\right)  =0
\]
are also the eigenvalues $\lambda_{N}\leq$ $\lambda_{N-1}\leq\ldots\leq
\lambda_{1}$ of $\mathbf{Q}$
\citep{Lathi1998}%
.

Therefore, given that $(1)$ the roots of a characteristic polynomial $p\left(
\lambda\right)  $ vary continuously with its coefficients and that $(2)$ the
coefficients of $p\left(  \lambda\right)  $ can be expressed in terms of sums
of principal minors
\citep[see][]{Meyer2000}%
, it follows that the eigenvalues $\lambda_{i}$ of $\mathbf{Q}$ vary
continuously with the inner product elements $\varphi\left(  \mathbf{x}%
_{i},\mathbf{x}_{j}\right)  $ of $\mathbf{Q}$. Thereby, the eigenvalues
$\lambda_{N}\leq$ $\lambda_{N-1}\leq\ldots\leq\lambda_{1}$ of a Gram or kernel
matrix $\mathbf{Q}$ are essentially determined by its inner product elements
$\varphi\left(  \mathbf{x}_{i},\mathbf{x}_{j}\right)  $.

\subsection{Statistics for Linear Partitions}

Given Eq. (\ref{Eigenvalue Coordinate Form Second Order Loci}) and the
continuous functional relationship between the inner product elements and the
eigenvalues of a Gram or kernel matrix, it follows that the geometric shapes
of the three, symmetrical quadratic partitioning surfaces determined by Eqs
(\ref{Wolfe Dual Normal Eigenlocus}) and (\ref{Vector Form Wolfe Dual}) are an
inherent\textit{\ }function of inner product statistics $\varphi\left(
\mathbf{x}_{i},\mathbf{x}_{j}\right)  $ between vectors. Therefore, it is
concluded that the form of the inner product statistics contained within Gram
or kernel matrices essentially determines the shapes of the three, symmetrical
quadratic partitioning surfaces determined by Eqs
(\ref{Wolfe Dual Normal Eigenlocus}) and (\ref{Vector Form Wolfe Dual}).

I\ have conducted simulation studies which demonstrate that the eigenvalues of
a polynomial kernel matrix associated with the constrained quadratic form in
Eq. (\ref{Vector Form Wolfe Dual}), for which matrix elements $\varphi\left(
\mathbf{x}_{i},\mathbf{x}_{j}\right)  $ have the algebraic form of $\left(
\mathbf{x}_{i}^{T}\mathbf{x}_{j}+1\right)  ^{2}$, determine either\ $l-1$%
-dimensional circles, ellipses, hyperbolas, or parabolas. Such second-order
decision boundary estimates can be generated by solving an inequality
constrained optimization problem that is similar in nature to Eq.
(\ref{Primal Normal Eigenlocus}). I\ have also conducted simulation studies
which demonstrate that the eigenvalues of a Gram matrix associated with the
constrained quadratic form in Eq. (\ref{Vector Form Wolfe Dual}), for which
matrix elements $\varphi\left(  \mathbf{x}_{i},\mathbf{x}_{j}\right)  $ have
the algebraic form of $\mathbf{x}_{i}^{T}\mathbf{x}_{j}$, determine
$l-1$-dimensional hyperplane surfaces
\citep[see][]{ Reeves2015resolving}%
.

So, let the Gram matrix $\mathbf{Q}$ in Eq. (\ref{Vector Form Wolfe Dual})
contain inner product statistics $\varphi\left(  \mathbf{x}_{i},\mathbf{x}%
_{j}\right)  =\mathbf{x}_{i}^{T}\mathbf{x}_{j}$ for a separating hyperplane
$H_{0}\left(  \mathbf{x}\right)  $ and hyperplane decision borders
$H_{+1}\left(  \mathbf{x}\right)  $ and $H_{-1}\left(  \mathbf{x}\right)  $
that have bilateral symmetry along $H_{0}\left(  \mathbf{x}\right)  $. Later
on, I\ will show that linear eigenlocus transforms map the labeled $\pm1$,
inner product statistics $\mathbf{x}_{i}^{T}\mathbf{x}_{j}$ contained within
$\mathbf{Q}$%
\[
\mathbf{Q}\boldsymbol{\psi}=\lambda\mathbf{_{\max\boldsymbol{\psi}}%
}\boldsymbol{\psi}%
\]
into a Wolfe dual linear eigenlocus of principal eigenaxis components%
\[
\mathbf{Q\sum\nolimits_{i=1}^{l}}\psi_{i\ast}\overrightarrow{\mathbf{e}%
}_{i\ast}=\lambda\mathbf{_{\max\boldsymbol{\psi}}}\sum\nolimits_{i=1}^{l}%
\psi_{i\ast}\overrightarrow{\mathbf{e}}_{i\ast}\text{,}%
\]
formed by $l$ scaled $\psi_{i\ast}$, non-orthogonal unit vectors $\left\{
\overrightarrow{\mathbf{e}}_{1\ast},\ldots,\overrightarrow{\mathbf{e}}_{l\ast
}\right\}  $, where the locus of each Wolfe dual principal eigenaxis component
$\psi_{i\ast}\overrightarrow{\mathbf{e}}_{i\ast}\ $is determined by the
direction and well-proportioned magnitude of a correlated extreme vector
$\mathbf{x}_{i\ast}$.

By way of motivation, I\ will now identify the fundamental statistical
property possessed by $\boldsymbol{\tau}$ that enables a linear eigenlocus
discriminant function $\widetilde{\Lambda}_{\boldsymbol{\tau}}\left(
\mathbf{x}\right)  =\boldsymbol{\tau}^{T}\mathbf{x}+\tau_{0}$ to satisfy a
discrete and data-driven version of the fundamental integral equation of
binary classification in Eq. (\ref{Equalizer Rule}).

\section{Property of Symmetrical Balance I}

I\ have demonstrated that constrained, linear eigenlocus discriminant
functions $\widetilde{\Lambda}_{\boldsymbol{\tau}}\left(  \mathbf{x}\right)
=\boldsymbol{\tau}^{T}\mathbf{x}+\tau_{0}$ determine contiguous and congruent
$Z_{1}\cong Z_{2}$ decision regions $Z_{1}$ and $Z_{2}$ that are delineated by
linear decision borders $D_{+1}\left(  \mathbf{x}\right)  $ and $D_{-1}\left(
\mathbf{x}\right)  $ located at equal distances $\frac{2}{\left\Vert
\boldsymbol{\tau}\right\Vert }$ from a linear decision boundary $D_{0}\left(
\mathbf{x}\right)  $, where all of the points $\mathbf{x}$ on $D_{+1}\left(
\mathbf{x}\right)  $, $D_{-1}\left(  \mathbf{x}\right)  $, and $D_{0}\left(
\mathbf{x}\right)  $ reference a linear eigenlocus $\boldsymbol{\tau}$.

Therefore, I\ have shown that constrained, linear eigenlocus discriminant
functions $\widetilde{\Lambda}_{\boldsymbol{\tau}}\left(  \mathbf{x}\right)
=\boldsymbol{\tau}^{T}\mathbf{x}+\tau_{0}$ satisfy boundary values of linear
decision borders $D_{+1}\left(  \mathbf{x}\right)  $ and $D_{-1}\left(
\mathbf{x}\right)  $ and linear decision boundaries $D_{0}\left(
\mathbf{x}\right)  $, where the axis of $\boldsymbol{\tau}$ is an axis of
symmetry for $D_{+1}\left(  \mathbf{x}\right)  $, $D_{-1}\left(
\mathbf{x}\right)  $, and $D_{0}\left(  \mathbf{x}\right)  $.

The bilateral symmetry exhibited by linear decision borders $D_{+1}\left(
\mathbf{x}\right)  $ and $D_{-1}\left(  \mathbf{x}\right)  $ along a linear
decision boundary $D_{0}\left(  \mathbf{x}\right)  $ is also known as
symmetrical balance. Given that $\boldsymbol{\tau}$ is an axis of symmetry
which satisfies boundary values of linear decision borders $D_{+1}\left(
\mathbf{x}\right)  $ and $D_{-1}\left(  \mathbf{x}\right)  $ and linear
decision boundaries $D_{0}\left(  \mathbf{x}\right)  $, it follows that
$\boldsymbol{\tau}$ \emph{must posses} the statistical property of
\emph{symmetrical balance}. Recall that the physical property of symmetrical
balance involves an axis or lever in equilibrium where different elements are
equal or in correct proportions, relative to the center of an axis or a lever,
such that the opposing forces or influences of a system are balanced with each other.

\subsection{Symmetrical Balance Exhibited by the Axis of $\boldsymbol{\tau}$}

Returning to Eqs (\ref{Equilibrium Constraint on Dual Eigen-components}) and
(\ref{Wolfe Dual Vector Equation}), recall that the axis of $\boldsymbol{\psi
}$ can be regarded as a lever in statistical equilibrium where different
principal eigenaxis components are equal or in correct proportions, relative
to the center of $\boldsymbol{\psi}$, such that the opposing forces associated
with the risks and counter risks of a linear classification system are
balanced with each other. Thus, the axis of $\boldsymbol{\psi=\psi}%
_{1}+\boldsymbol{\psi}_{2}$ exhibits the statistical property of symmetrical
balance, where $\sum\nolimits_{i=1}^{l_{1}}\psi_{1_{i\ast}}%
\overrightarrow{\mathbf{e}}_{1_{i\ast}}\rightleftharpoons\sum\nolimits_{i=1}%
^{l_{2}}\psi_{2_{i\ast}}\overrightarrow{\mathbf{e}}_{2_{i\ast}}$.

Furthermore, given Eqs (\ref{Equilibrium Constraint on Dual Component Lengths}%
) and (\ref{Symmetrical Balance of Wolf Dual Eigenenergies}), the axis of
$\boldsymbol{\psi}$ can be regarded as a lever that has an equal distribution
of eigenenergies on equal sides of a centrally placed fulcrum which is located
at the center of total allowed eigenenergy $\left\Vert \boldsymbol{\psi
}\right\Vert _{\min_{c}}^{2}$ of $\boldsymbol{\psi}$. Accordingly, the total
allowed eigenenergies possessed by the principal eigenaxis components on
$\boldsymbol{\psi}$ are symmetrically balanced with each other about a center
of total allowed eigenenergy $\left\Vert \boldsymbol{\psi}\right\Vert
_{\min_{c}}^{2}$ which is located at the geometric center of $\boldsymbol{\psi
}$. Thus, the axis of $\boldsymbol{\psi=\psi}_{1}+\boldsymbol{\psi}_{2}$
exhibits the statistical property of symmetrical balance, where $\left\Vert
\boldsymbol{\psi}_{1}\right\Vert \rightleftharpoons\left\Vert \boldsymbol{\psi
}_{2}\right\Vert $ and $\left\Vert \sum\nolimits_{i=1}^{l_{1}}\psi_{1_{i\ast}%
}\overrightarrow{\mathbf{e}}_{1_{i\ast}}\right\Vert _{\min_{c}}^{2}%
\rightleftharpoons\left\Vert \sum\nolimits_{i=1}^{l_{2}}\psi_{2_{i\ast}%
}\overrightarrow{\mathbf{e}}_{2_{i\ast}}\right\Vert _{\min_{c}}^{2}$.

Returning to Eqs (\ref{Normal Eigenaxis Functional}) and
(\ref{Characteristic Eigenenergy}), recall that the locus of any given line,
plane, or hyperplane is determined by the eigenenergy $\left\Vert
\boldsymbol{\nu}\right\Vert ^{2}$ exhibited by the locus of its principal
eigenaxis $\boldsymbol{\nu}$, where any given principal eigenaxis
$\boldsymbol{\nu}$ and any given point $\mathbf{x}$ on a linear locus
satisfies the eigenenergy $\left\Vert \boldsymbol{\nu}\right\Vert ^{2}$ of
$\boldsymbol{\nu}$. Thereby, the inherent property of a linear locus and its
principal eigenaxis $\boldsymbol{\nu}$ is the eigenenergy $\left\Vert
\boldsymbol{\nu}\right\Vert ^{2}$ exhibited by $\boldsymbol{\nu}$.

\begin{figure}[ptb]%
\centering
\fbox{\includegraphics[
height=2.5875in,
width=3.4411in
]%
{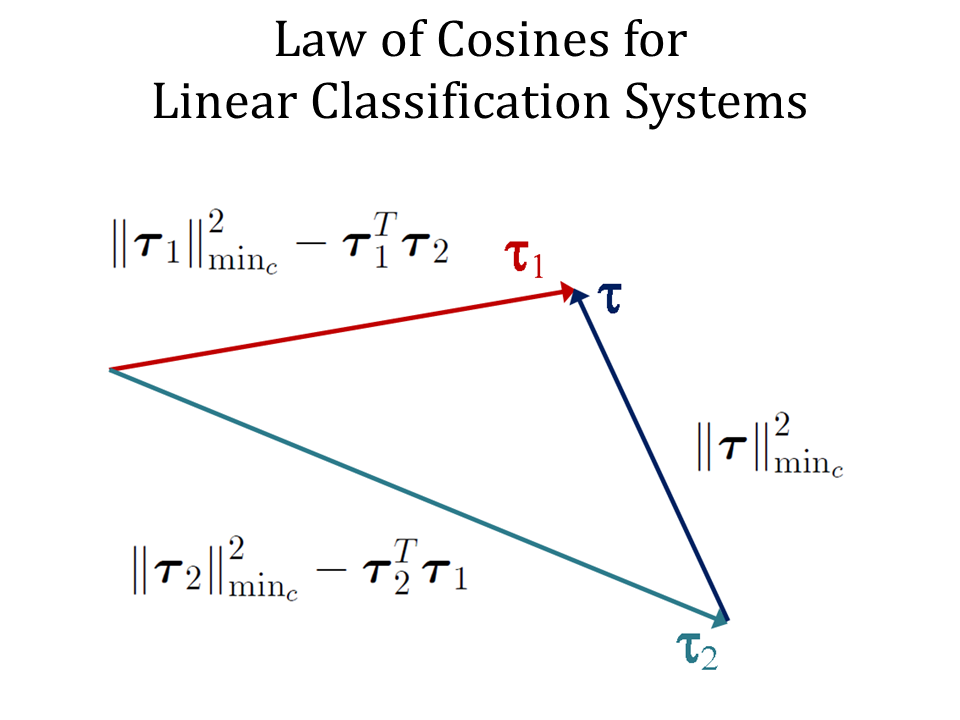}%
}\caption{The likelihood ratio $\protect\widehat{\Lambda}_{\boldsymbol{\tau}%
}\left(  \mathbf{x}\right)  =\boldsymbol{\tau}_{1}-\boldsymbol{\tau}_{2}$ of a
linear eigenlocus discriminant function $\protect\widetilde{\Lambda
}_{\boldsymbol{\tau}}\left(  \mathbf{x}\right)  =\boldsymbol{\tau}%
^{T}\mathbf{x}+\tau_{0}$ satisfies the law of cosines in a symmetrically
balanced manner.}%
\label{Law of Cosines for Binary Classification Systems}%
\end{figure}

Therefore, given Eqs (\ref{Normal Eigenaxis Functional}),
(\ref{Characteristic Eigenenergy}),
(\ref{Minimum Total Eigenenergy Primal Normal Eigenlocus}), and
(\ref{Pair of Normal Eigenlocus Components}), it follows that a constrained,
primal linear eigenlocus $\boldsymbol{\tau}$ satisfies the linear decision
boundary $D_{0}\left(  \mathbf{x}\right)  $ in Eq. (\ref{Decision Boundary})
and the linear decision borders $D_{+1}\left(  \mathbf{x}\right)  $ and
$D_{-1}\left(  \mathbf{x}\right)  $ in Eqs (\ref{Decision Border One}) and
(\ref{Decision Border Two}) in terms of its total allowed eigenenergies:%
\begin{align*}
\left\Vert \boldsymbol{\tau}\right\Vert _{\min_{c}}^{2}  &  =\left\Vert
\boldsymbol{\tau}_{1}-\boldsymbol{\tau}_{2}\right\Vert _{\min_{c}}^{2}\\
&  \cong\left[  \left\Vert \boldsymbol{\tau}_{1}\right\Vert _{\min_{c}}%
^{2}-\boldsymbol{\tau}_{1}^{T}\boldsymbol{\tau}_{2}\right]  +\left[
\left\Vert \boldsymbol{\tau}_{2}\right\Vert _{\min_{c}}^{2}-\boldsymbol{\tau
}_{2}^{T}\boldsymbol{\tau}_{1}\right]  \text{,}%
\end{align*}
where the functional $\left\Vert \boldsymbol{\tau}_{1}\right\Vert _{\min_{c}%
}^{2}-\boldsymbol{\tau}_{1}^{T}\boldsymbol{\tau}_{2}$ is associated with the
$D_{+1}\left(  \mathbf{x}\right)  $ linear decision border, the functional
$\left\Vert \boldsymbol{\tau}_{2}\right\Vert _{\min_{c}}^{2}-\boldsymbol{\tau
}_{2}^{T}\boldsymbol{\tau}_{1}$ is associated with the $D_{-1}\left(
\mathbf{x}\right)  $ linear decision border, and the functional $\left\Vert
\boldsymbol{\tau}\right\Vert _{\min_{c}}^{2}$ is associated with the linear
decision boundary $D_{0}\left(  \mathbf{x}\right)  $. Thus, the total allowed
eigenenergies of the principal eigenaxis components on a linear eigenlocus
$\boldsymbol{\tau=\tau}_{1}-\boldsymbol{\tau}_{2}$ must satisfy the law of
cosines in the symmetrically balanced manner depicted in Fig.
$\ref{Law of Cosines for Binary Classification Systems}$.%

Given that $\boldsymbol{\tau}$ must possess the statistical property of
symmetrical balance in terms of its principal eigenaxis components, it follows
that the axis of $\boldsymbol{\tau}$ is essentially a lever that is
symmetrically balanced with respect to the center of eigenenergy $\left\Vert
\boldsymbol{\tau}_{1}-\boldsymbol{\tau}_{2}\right\Vert _{\min_{c}}^{2}$ of
$\boldsymbol{\tau}$. Accordingly, the axis of $\boldsymbol{\tau}$ is said to
be in statistical equilibrium, where the constrained, primal principal
eigenaxis components on $\boldsymbol{\tau}$%
\begin{align*}
\boldsymbol{\tau}  &  =\boldsymbol{\tau}_{1}-\boldsymbol{\tau}_{2}\\
&  =\sum\nolimits_{i=1}^{l_{1}}\psi_{1_{i\ast}}\mathbf{x}_{1_{i\ast}}%
-\sum\nolimits_{i=1}^{l_{2}}\psi_{2_{i\ast}}\mathbf{x}_{2_{i\ast}}%
\end{align*}
are equal or in correct proportions, relative to the center of
$\boldsymbol{\tau}$, such that all of the forces associated with the risks
$\mathfrak{R}_{\mathfrak{\min}}\left(  Z_{1}|\boldsymbol{\tau}_{2}\right)  $
and $\mathfrak{R}_{\mathfrak{\min}}\left(  Z_{2}|\boldsymbol{\tau}_{1}\right)
$: for class $\omega_{2}$ and class $\omega_{1}$, and all of the forces
associated with the counter risks $\overline{\mathfrak{R}}_{\mathfrak{B}%
}\left(  Z_{1}|\boldsymbol{\tau}_{1}\right)  $ and $\overline{\mathfrak{R}%
}_{\mathfrak{B}}\left(  Z_{2}|\boldsymbol{\tau}_{2}\right)  $: for class
$\omega_{1}$ and class $\omega_{2}$, of a binary, linear classification system
$\mathbf{x}^{T}\boldsymbol{\tau}+\tau_{0}\overset{\omega_{1}}{\underset{\omega
_{2}}{\gtrless}}0$ are symmetrically balanced with each other.

\begin{figure}[ptb]%
\centering
\fbox{\includegraphics[
height=2.5875in,
width=3.4411in
]%
{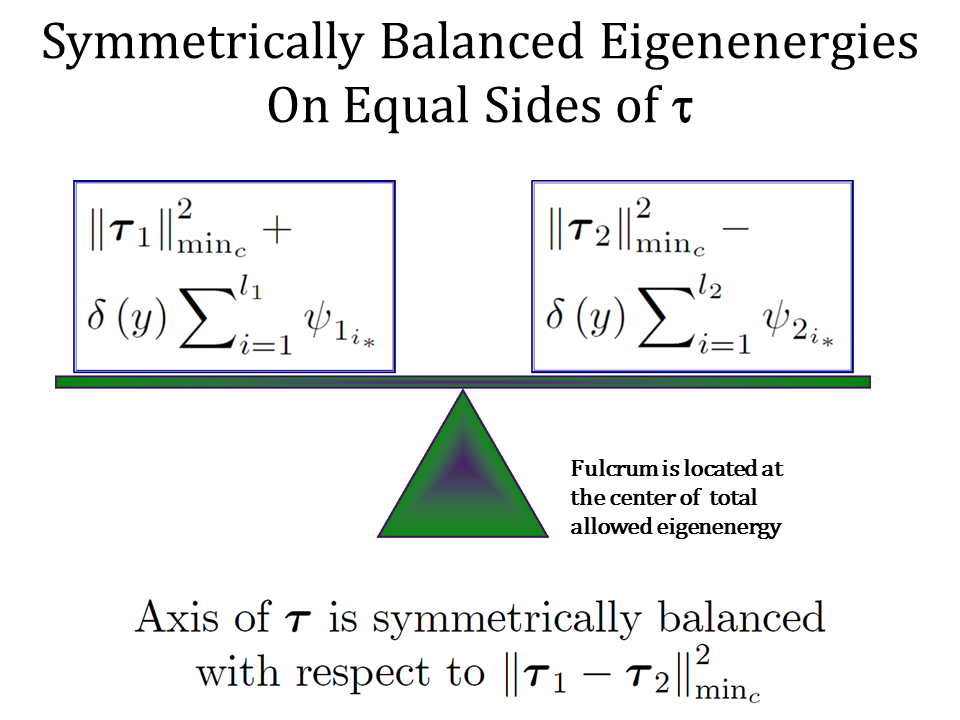}%
}\caption{A constrained, linear eigenlocus discriminant function
$\protect\widetilde{\Lambda}_{\boldsymbol{\tau}}\left(  \mathbf{x}\right)
=\boldsymbol{\tau}^{T}\mathbf{x}+\tau_{0}$ satisfies a fundamental integral
equation of binary classification for a classification system in statistical
equilibrium, where the expected risk $\mathfrak{R}_{\mathfrak{\min}}\left(
Z\mathbf{|}\boldsymbol{\tau}\right)  $ and the total allowed eigenenergy
$\left\Vert \boldsymbol{\tau}\right\Vert _{\min_{c}}^{2}$ of the
classification system are minimized, because the axis of $\boldsymbol{\tau}$
is essentially a lever that is symmetrically balanced with respect to the
center of eigenenergy $\left\Vert \boldsymbol{\tau}_{1}-\boldsymbol{\tau}%
_{2}\right\Vert _{\min_{c}}^{2}$ of $\boldsymbol{\tau}$.}%
\label{Symmetrical Balance of Constrained Primal Linear Eigenlocus}%
\end{figure}

I will prove that a constrained, linear eigenlocus discriminant function
$\widetilde{\Lambda}_{\boldsymbol{\tau}}\left(  \mathbf{x}\right)
=\mathbf{x}^{T}\boldsymbol{\tau}+\tau_{0}$ satisfies a discrete and
data-driven version of the fundamental integral equation of binary
classification for a classification system in statistical equilibrium in Eq.
(\ref{Equalizer Rule}) because the axis of $\boldsymbol{\tau}$ is essentially
a lever that is symmetrically balanced with respect to the center of
eigenenergy $\left\Vert \boldsymbol{\tau}_{1}-\boldsymbol{\tau}_{2}\right\Vert
_{\min_{c}}^{2}$ of $\boldsymbol{\tau}$ in the following manner:%
\[
\left\Vert \boldsymbol{\tau}_{1}\right\Vert _{\min_{c}}^{2}-\left\Vert
\boldsymbol{\tau}_{1}\right\Vert \left\Vert \boldsymbol{\tau}_{2}\right\Vert
\cos\theta_{\boldsymbol{\tau}_{1}\boldsymbol{\tau}_{2}}+\delta\left(
y\right)  \frac{1}{2}\sum\nolimits_{i=1}^{l}\psi_{_{i\ast}}\equiv\frac{1}%
{2}\left\Vert \boldsymbol{\tau}\right\Vert _{\min_{c}}^{2}%
\]
and%
\[
\left\Vert \boldsymbol{\tau}_{2}\right\Vert _{\min_{c}}^{2}-\left\Vert
\boldsymbol{\tau}_{2}\right\Vert \left\Vert \boldsymbol{\tau}_{1}\right\Vert
\cos\theta_{\boldsymbol{\tau}_{2}\boldsymbol{\tau}_{1}}-\delta\left(
y\right)  \frac{1}{2}\sum\nolimits_{i=1}^{l}\psi_{_{i\ast}}\equiv\frac{1}%
{2}\left\Vert \boldsymbol{\tau}\right\Vert _{\min_{c}}^{2}%
\]
where the equalizer statistic $\nabla_{eq}$%
\[
\nabla_{eq}\triangleq\delta\left(  y\right)  \frac{1}{2}\sum\nolimits_{i=1}%
^{l}\psi_{_{i\ast}}%
\]
for which $\delta\left(  y\right)  \triangleq\sum\nolimits_{i=1}^{l}%
y_{i}\left(  1-\xi_{i}\right)  $, equalizes the total allowed eigenenergies
$\left\Vert \boldsymbol{\tau}_{1}\right\Vert _{\min_{c}}^{2}$ and $\left\Vert
\boldsymbol{\tau}_{2}\right\Vert _{\min_{c}}^{2}$ exhibited by
$\boldsymbol{\tau}_{1}$ and $\boldsymbol{\tau}_{2}$:%
\begin{align*}
&  \left\Vert \boldsymbol{\tau}_{1}\right\Vert _{\min_{c}}^{2}+\delta\left(
y\right)  \frac{1}{2}\sum\nolimits_{i=1}^{l}\psi_{_{i\ast}}\\
&  \equiv\left\Vert \boldsymbol{\tau}_{2}\right\Vert _{\min_{c}}^{2}%
-\delta\left(  y\right)  \frac{1}{2}\sum\nolimits_{i=1}^{l}\psi_{_{i\ast}}%
\end{align*}
so that the dual locus of $\boldsymbol{\tau}_{1}-\boldsymbol{\tau}_{2}$ is in
statistical equilibrium. Figure
$\ref{Symmetrical Balance of Constrained Primal Linear Eigenlocus}$%
\textbf{\ }illustrates the property of symmetrical balance exhibited by the
dual locus of $\boldsymbol{\tau}$.%

I\ will obtain the above equations for a linear eigenlocus $\boldsymbol{\tau}$
in statistical equilibrium by devising a chain of arguments which demonstrate
that a constrained, linear eigenlocus discriminant function
$\widetilde{\Lambda}_{\boldsymbol{\tau}}\left(  \mathbf{x}\right)
=\mathbf{x}^{T}\boldsymbol{\tau}+\tau_{0}$ satisfies discrete and data-driven
versions of the fundamental equations of binary classification for a
classification system in statistical equilibrium in Eqs
(\ref{Vector Equation of Likelihood Ratio and Decision Boundary}) -
(\ref{Balancing of Bayes' Risks and Counteracting Risks}).

The general course of my argument is outlined next.

\section{General Course of Argument I}

In order to prove that a constrained, linear eigenlocus discriminant function
$\widetilde{\Lambda}_{\boldsymbol{\tau}}\left(  \mathbf{x}\right)
=\boldsymbol{\tau}^{T}\mathbf{x}+\tau_{0}$ satisfies $\left(  1\right)  $ the
vector equation%
\[
\boldsymbol{\tau}^{T}\mathbf{x}+\tau_{0}=0\text{,}%
\]
$\left(  2\right)  $ the statistical equilibrium equation:%
\[
p\left(  \widehat{\Lambda}_{\boldsymbol{\tau}}\left(  \mathbf{x}\right)
|\omega_{1}\right)  \rightleftharpoons p\left(  \widehat{\Lambda
}_{\boldsymbol{\tau}}\left(  \mathbf{x}\right)  |\omega_{2}\right)  \text{,}%
\]
$\left(  3\right)  $ the corresponding integral equation:%
\[
\int_{Z}p\left(  \widehat{\Lambda}_{\boldsymbol{\tau}}\left(  \mathbf{x}%
\right)  |\omega_{1}\right)  d\widehat{\Lambda}_{\boldsymbol{\tau}}=\int%
_{Z}p\left(  \widehat{\Lambda}_{\boldsymbol{\tau}}\left(  \mathbf{x}\right)
|\omega_{2}\right)  d\widehat{\Lambda}_{\boldsymbol{\tau}}\text{,}%
\]
$\left(  4\right)  $ a discrete and data-driven linear version of the
fundamental integral equation of binary classification for a classification
system in statistical equilibrium:%
\begin{align*}
f\left(  \widetilde{\Lambda}_{\boldsymbol{\tau}}\left(  \mathbf{x}\right)
\right)   &  =\int_{Z_{1}}p\left(  \widehat{\Lambda}_{\boldsymbol{\tau}%
}\left(  \mathbf{x}\right)  |\omega_{1}\right)  d\widehat{\Lambda
}_{\boldsymbol{\tau}}+\int_{Z_{2}}p\left(  \widehat{\Lambda}_{\boldsymbol{\tau
}}\left(  \mathbf{x}\right)  |\omega_{1}\right)  d\widehat{\Lambda
}_{\boldsymbol{\tau}}+\delta\left(  y\right)  \sum\nolimits_{i=1}^{l_{1}}%
\psi_{1_{i_{\ast}}}\\
&  =\int_{Z_{1}}p\left(  \widehat{\Lambda}_{\boldsymbol{\tau}}\left(
\mathbf{x}\right)  |\omega_{2}\right)  d\widehat{\Lambda}_{\boldsymbol{\tau}%
}+\int_{Z_{2}}p\left(  \widehat{\Lambda}_{\boldsymbol{\tau}}\left(
\mathbf{x}\right)  |\omega_{2}\right)  d\widehat{\Lambda}_{\boldsymbol{\tau}%
}-\delta\left(  y\right)  \sum\nolimits_{i=1}^{l_{2}}\psi_{2_{i_{\ast}}%
}\text{,}%
\end{align*}
and $\left(  5\right)  $ the corresponding integral equation:%
\begin{align*}
f\left(  \widetilde{\Lambda}_{\boldsymbol{\tau}}\left(  \mathbf{x}\right)
\right)   &  :\;\int_{Z_{1}}p\left(  \widehat{\Lambda}_{\boldsymbol{\tau}%
}\left(  \mathbf{x}\right)  |\omega_{1}\right)  d\widehat{\Lambda
}_{\boldsymbol{\tau}}-\int_{Z_{1}}p\left(  \widehat{\Lambda}_{\boldsymbol{\tau
}}\left(  \mathbf{x}\right)  |\omega_{2}\right)  d\widehat{\Lambda
}_{\boldsymbol{\tau}}+\delta\left(  y\right)  \sum\nolimits_{i=1}^{l_{1}}%
\psi_{1_{i_{\ast}}}\\
&  =\int_{Z_{2}}p\left(  \widehat{\Lambda}_{\boldsymbol{\tau}}\left(
\mathbf{x}\right)  |\omega_{2}\right)  d\widehat{\Lambda}_{\boldsymbol{\tau}%
}-\int_{Z_{2}}p\left(  \widehat{\Lambda}_{\boldsymbol{\tau}}\left(
\mathbf{x}\right)  |\omega_{1}\right)  d\widehat{\Lambda}_{\boldsymbol{\tau}%
}-\delta\left(  y\right)  \sum\nolimits_{i=1}^{l_{2}}\psi_{2_{i_{\ast}}%
}\text{,}%
\end{align*}
I will need to develop mathematical machinery for several systems of locus
equations. The fundamental equations of a binary classification system involve
mathematical machinery and systems of locus equations that determine the
following mathematical objects:

\begin{enumerate}
\item Total allowed eigenenergies of extreme points on a Wolfe dual and a
constrained primal linear eigenlocus.

\item Total allowed eigenenergies of Wolfe dual and constrained primal linear
eigenlocus components.

\item Total allowed eigenenergy of a Wolfe dual and a constrained primal
linear eigenlocus.

\item Class-conditional probability density functions for extreme points.

\item Conditional probability functions for extreme points.

\item Risks and counter risks of extreme points.

\item Conditional probability functions for the risks and the counter risks
related to positions and potential locations of extreme points.

\item Integral equations of class-conditional probability density functions.
\end{enumerate}

A high level overview of the development of the mathematical machinery and
systems of locus equations is outlined below.

I\ will develop class-conditional probability density functions and
conditional probability functions for extreme points in the following manner:

Using Eq. (\ref{Geometric Locus of Vector}), any given extreme point
$\mathbf{x}_{i\ast}$ is the endpoint on a locus (a position vector) of random
variables%
\[%
\begin{pmatrix}
\left\Vert \mathbf{x}_{i\ast}\right\Vert \cos\mathbb{\alpha}_{\mathbf{x}%
_{i\ast1}1}, & \left\Vert \mathbf{x}_{i\ast}\right\Vert \cos\mathbb{\alpha
}_{\mathbf{x}_{i\ast2}2}, & \cdots, & \left\Vert \mathbf{x}_{i\ast}\right\Vert
\cos\mathbb{\alpha}_{\mathbf{x}_{i\ast d}d}%
\end{pmatrix}
\text{,}%
\]
where each random variable $\left\Vert \mathbf{x}_{i\ast}\right\Vert
\cos\mathbb{\alpha}_{x\ast_{i}j}$ is characterized by an expected value
$E\left[  \left\Vert \mathbf{x}_{i\ast}\right\Vert \cos\mathbb{\alpha}%
_{x\ast_{i}j}\right]  $ and a variance $\operatorname{var}\left(  \left\Vert
\mathbf{x}_{i\ast}\right\Vert \cos\mathbb{\alpha}_{x_{i\ast}j}\right)  $.
Therefore, an extreme point $\mathbf{x}_{i\ast}$ is described by a central
location (an expected value) and a covariance (a spread). The relative
likelihood that an extreme point has a given location is described by a
conditional probability density function. The cumulative probability of given
locations for an extreme point, i.e., the probability of finding the extreme
point within a localized region, is described by a conditional probability
function
\citep{Ash1993,Flury1997}%
.

So, take the Wolfe dual linear eigenlocus in Eq.
(\ref{Wolfe Dual Vector Equation}):%
\begin{align*}
\boldsymbol{\psi}  &  =\sum\nolimits_{i=1}^{l_{1}}\psi_{1i\ast}%
\overrightarrow{\mathbf{e}}_{1i\ast}+\sum\nolimits_{i=1}^{l_{2}}\psi_{2i\ast
}\overrightarrow{\mathbf{e}}_{2i\ast}\\
&  =\boldsymbol{\psi}_{1}+\boldsymbol{\psi}_{2}\text{,}%
\end{align*}
where each scaled, non-orthogonal unit vector $\psi_{1i\ast}%
\overrightarrow{\mathbf{e}}_{1i\ast}$ or $\psi_{2i\ast}%
\overrightarrow{\mathbf{e}}_{2i\ast}$ is a displacement vector that is
correlated with an $\mathbf{x}_{1_{i\ast}}$ or $\mathbf{x}_{2_{i\ast}}$extreme
vector respectively. For a given set $\left\{  \left\{  \mathbf{x}_{1_{i\ast}%
}\right\}  _{i=1}^{l_{1}},\;\left\{  \mathbf{x}_{2_{i\ast}}\right\}
_{i=1}^{l_{2}}\right\}  $ of $\mathbf{x}_{1_{i_{\ast}}}$ and $\mathbf{x}%
_{2_{i_{\ast}}}$extreme points, I\ will show that each Wolfe dual principal
eigenaxis component $\psi_{i\ast}\overrightarrow{\mathbf{e}}_{i\ast}$ on
$\boldsymbol{\psi}$ specifies a class-conditional density $p\left(
\mathbf{x}_{i\ast}|\operatorname{comp}_{\overrightarrow{\mathbf{x}_{i\ast}}%
}\left(  \overrightarrow{\boldsymbol{\tau}}\right)  \right)  $ for a
correlated extreme point $\mathbf{x}_{i\ast}$, such that $\boldsymbol{\psi
}_{1}$ and $\boldsymbol{\psi}_{2}$ are class-conditional probability density
functions in Wolfe dual eigenspace, $\boldsymbol{\tau}_{1}$ is a parameter
vector for a class-conditional probability density $p\left(  \mathbf{x}%
_{1_{i\ast}}|\boldsymbol{\tau}_{1}\right)  $ for a given set $\left\{
\mathbf{x}_{1_{i\ast}}\right\}  _{i=1}^{l_{1}}$ of $\mathbf{x}_{1_{i_{\ast}}}$
extreme points:%
\[
\boldsymbol{\tau}_{1}=p\left(  \mathbf{x}_{1_{i\ast}}|\boldsymbol{\tau}%
_{1}\right)  \text{,}%
\]
and $\boldsymbol{\tau}_{2}$ is a parameter vector for a class-conditional
probability density $p\left(  \mathbf{x}_{2_{i\ast}}|\boldsymbol{\tau}%
_{2}\right)  $ for a given set $\left\{  \mathbf{x}_{2_{i\ast}}\right\}
_{i=1}^{l_{2}}$ of $\mathbf{x}_{2_{i_{\ast}}}$ extreme points:%
\[
\boldsymbol{\tau}_{2}=p\left(  \mathbf{x}_{2_{i\ast}}|\boldsymbol{\tau}%
_{2}\right)  \text{,}%
\]
where the area under each pointwise conditional density $p\left(
\mathbf{x}_{i\ast}|\operatorname{comp}_{\overrightarrow{\mathbf{x}_{i\ast}}%
}\left(  \overrightarrow{\boldsymbol{\tau}}\right)  \right)  $ is a
conditional probability that an extreme point $\mathbf{x}_{i\ast}$ will be
observed in a $Z_{1}$ or $Z_{2}$ decision region of a decision space $Z$.

In order to develop class-conditional probability densities for extreme
points, I will devise a system of data-driven, locus equations in Wolfe dual
eigenspace that provides tractable point and coordinate relationships between
the weighted (labeled and scaled) extreme points on $\boldsymbol{\tau}_{1}$
and $\boldsymbol{\tau}_{2}$ and the Wolfe dual principal eigenaxis components
on $\boldsymbol{\psi}_{1}$ and $\boldsymbol{\psi}_{2}$. I\ will use this
system of equations to develop equations for geometric and statistical
properties possessed by the Wolfe dual and the constrained, primal principal
eigenaxis components. Next, I\ will use these equations and identified
properties to define class-conditional probability densities for individual
extreme points and class-conditional probability densities $p\left(
\mathbf{x}_{1_{i\ast}}|\boldsymbol{\tau}_{1}\right)  $ and $p\left(
\mathbf{x}_{2_{i\ast}}|\boldsymbol{\tau}_{2}\right)  $ for labeled sets of
extreme points.

Thereby, I\ will demonstrate that the conditional probability function
$P\left(  \mathbf{x}_{1_{i\ast}}|\boldsymbol{\tau}_{1}\right)  $ for a given
set $\left\{  \mathbf{x}_{1_{i\ast}}\right\}  _{i=1}^{l_{1}}$ of
$\mathbf{x}_{1_{i_{\ast}}}$ extreme points is given by the area under the
class-conditional probability density function $p\left(  \mathbf{x}_{1_{i\ast
}}|\boldsymbol{\tau}_{1}\right)  $%
\begin{align*}
P\left(  \mathbf{x}_{1_{i\ast}}|\boldsymbol{\tau}_{1}\right)   &  =\int%
_{Z}\left(  \sum\nolimits_{i=1}^{l_{1}}\psi_{1_{i\ast}}\mathbf{x}_{1_{i\ast}%
}\right)  d\boldsymbol{\tau}_{1}=\int_{Z}p\left(  \mathbf{x}_{1_{i\ast}%
}|\boldsymbol{\tau}_{1}\right)  d\boldsymbol{\tau}_{1}\\
&  =\int_{Z}\boldsymbol{\tau}_{1}d\boldsymbol{\tau}_{1}=\left\Vert
\boldsymbol{\tau}_{1}\right\Vert ^{2}+C_{1}\text{,}%
\end{align*}
over the decision space $Z$, where $\left\Vert \boldsymbol{\tau}%
_{1}\right\Vert ^{2}$ is the total allowed eigenenergy exhibited by
$\boldsymbol{\tau}_{1}$ and $C_{1}$ is an integration constant.

Likewise, I\ will demonstrate that the conditional probability function
$P\left(  \mathbf{x}_{2_{i\ast}}|\boldsymbol{\tau}_{2}\right)  $ for a given
set $\left\{  \mathbf{x}_{2_{i\ast}}\right\}  _{i=1}^{l_{2}}$ of
$\mathbf{x}_{2_{i_{\ast}}}$ extreme points is given by the area under the
class-conditional probability density function $p\left(  \mathbf{x}_{2_{i\ast
}}|\boldsymbol{\tau}_{2}\right)  $%
\begin{align*}
P\left(  \mathbf{x}_{2_{i\ast}}|\boldsymbol{\tau}_{2}\right)   &  =\int%
_{Z}\left(  \sum\nolimits_{i=1}^{l_{2}}\psi_{2_{i\ast}}\mathbf{x}_{2_{i\ast}%
}\right)  d\boldsymbol{\tau}_{2}=\int_{Z}p\left(  \mathbf{x}_{2_{i\ast}%
}|\boldsymbol{\tau}_{2}\right)  d\boldsymbol{\tau}_{2}\\
&  =\int_{Z}\boldsymbol{\tau}_{2}d\boldsymbol{\tau}_{2}=\left\Vert
\boldsymbol{\tau}_{2}\right\Vert ^{2}+C_{2}\text{,}%
\end{align*}
over the decision space $Z$, where $\left\Vert \boldsymbol{\tau}%
_{2}\right\Vert ^{2}$ is the total allowed eigenenergy exhibited by
$\boldsymbol{\tau}_{2}$ and $C_{2}$ is an integration constant.

In order to define the $C_{1}$ and $C_{2}$ integration constants, I\ will need
to define the manner in which the total allowed eigenenergies possessed by the
scaled extreme vectors on $\boldsymbol{\tau}_{1}$ and $\boldsymbol{\tau}_{2}$
are symmetrically balanced with each other. I\ will use these results to
define the manner in which the area under the class-conditional probability
density functions $p\left(  \mathbf{x}_{1_{i\ast}}|\boldsymbol{\tau}%
_{1}\right)  $ and $p\left(  \mathbf{x}_{2_{i_{\ast}}}|\boldsymbol{\tau}%
_{2}\right)  $ and the corresponding conditional probability functions
$P\left(  \mathbf{x}_{1_{i\ast}}|\boldsymbol{\tau}_{1}\right)  $ and $P\left(
\mathbf{x}_{2_{i\ast}}|\boldsymbol{\tau}_{2}\right)  $ for class $\omega_{1}$
and class $\omega_{2}$ are symmetrically balanced with each other.

I\ will define the $C_{1}$ and $C_{2}$ integration constants in the following
manner: I will use the KKT condition in Eq. (\ref{KKTE5}) and the theorem of
Karush, Kuhn, and Tucker to devise a system of data-driven, locus equations
that determines the manner in which the total allowed eigenenergies of the
scaled extreme vectors on $\boldsymbol{\tau}_{1}$ and $\boldsymbol{\tau}_{2}$
are symmetrically balanced with each other. I\ will use these results along
with results obtained from the analysis of the Wolfe dual eigenspace to devise
a system of data-driven, locus equations that determines the manner in which
the class-conditional density functions $p\left(  \mathbf{x}_{1_{i\ast}%
}|\boldsymbol{\tau}_{1}\right)  $ and $p\left(  \mathbf{x}_{2_{i_{\ast}}%
}|\boldsymbol{\tau}_{2}\right)  $ satisfy an integral equation%
\[
f\left(  \widetilde{\Lambda}_{\boldsymbol{\tau}}\left(  \mathbf{x}\right)
\right)  :\int_{Z}p\left(  \mathbf{x}_{1_{i\ast}}|\boldsymbol{\tau}%
_{1}\right)  d\boldsymbol{\tau}_{1}+\nabla_{eq}=\int_{Z}p\left(
\mathbf{x}_{2_{i\ast}}|\boldsymbol{\tau}_{2}\right)  d\boldsymbol{\tau}%
_{2}-\nabla_{eq}\text{,}%
\]
over the decision space $Z$, where $\nabla_{eq}$ is a symmetric equalizer statistic.

Thereby, I\ will demonstrate that the statistical property of symmetrical
balance exhibited by the principal eigenaxis components on $\boldsymbol{\psi}$
and $\boldsymbol{\tau}$ ensures that the conditional probability functions
$P\left(  \mathbf{x}_{1_{i\ast}}|\boldsymbol{\tau}_{1}\right)  $ and $P\left(
\mathbf{x}_{2_{i\ast}}|\boldsymbol{\tau}_{2}\right)  $ for class $\omega_{1}$
and class $\omega_{2}$ are equal to each other, so that a linear eigenlocus
discriminant function $\widetilde{\Lambda}_{\boldsymbol{\tau}}\left(
\mathbf{x}\right)  =\mathbf{x}^{T}\boldsymbol{\tau}+\tau_{0}$ satisfies a
data-driven version of the integral equation in Eq.
(\ref{Integral Equation of Likelihood Ratio and Decision Boundary}).

I will use these results along with results obtained from the analysis of the
Wolfe dual eigenspace to prove that linear eigenlocus discriminant functions
$\widetilde{\Lambda}_{\boldsymbol{\tau}}\left(  \mathbf{x}\right)
=\mathbf{x}^{T}\boldsymbol{\tau}+\tau_{0}$ satisfy a data-driven version of
the fundamental integral equation of binary classification in Eq.
(\ref{Equalizer Rule}) along with the corresponding integral equation in Eq.
(\ref{Balancing of Bayes' Risks and Counteracting Risks}).

I\ will also devise an integral equation $f\left(  \widetilde{\Lambda
}_{\boldsymbol{\tau}}\left(  \mathbf{x}\right)  \right)  $ that illustrates
the manner in which the property of symmetrical balance exhibited by the
principal eigenaxis components on $\boldsymbol{\psi}$ and $\boldsymbol{\tau}$
enables linear eigenlocus discriminant functions $\widetilde{\Lambda
}_{\boldsymbol{\tau}}\left(  \mathbf{x}\right)  =\boldsymbol{\tau}%
^{T}\mathbf{x}+\tau_{0}$ to effectively balance the forces associated with the
expected risk $\mathfrak{R}_{\mathfrak{\min}}\left(  Z\mathbf{|}%
\boldsymbol{\tau}\right)  $ of a classification system $\boldsymbol{\tau}%
^{T}\mathbf{x}+\tau_{0}\overset{\omega_{1}}{\underset{\omega_{2}}{\gtrless}}%
0$: where all of the forces associated with the counter risk $\overline
{\mathfrak{R}}_{\mathfrak{\min}}\left(  Z_{1}|\boldsymbol{\tau}_{1}\right)  $
and the risk $\mathfrak{R}_{\mathfrak{\min}}\left(  Z_{1}|\boldsymbol{\tau
}_{2}\right)  $ within the $Z_{1}$ decision region are symmetrically balanced
with all of the forces associated with the counter risk $\overline
{\mathfrak{R}}_{\mathfrak{\min}}\left(  Z_{2}|\boldsymbol{\tau}_{2}\right)  $
and the risk $\mathfrak{R}_{\mathfrak{\min}}\left(  Z_{2}|\boldsymbol{\tau
}_{1}\right)  $ within the $Z_{2}$ decision region.

Thereby, I will devise integral equations that are satisfied by linear
eigenlocus discriminant functions $\widetilde{\Lambda}_{\boldsymbol{\tau}%
}\left(  \mathbf{x}\right)  =\mathbf{x}^{T}\boldsymbol{\tau}+\tau_{0}$, by
which the discriminant function $\widetilde{\Lambda}_{\boldsymbol{\tau}%
}\left(  \mathbf{x}\right)  =\mathbf{x}^{T}\boldsymbol{\tau}+\tau_{0}$ is the
solution to the fundamental integral equation of binary classification for a
linear classification system in statistical equilibrium:%
\begin{align*}
f\left(  \widetilde{\Lambda}_{\boldsymbol{\tau}}\left(  \mathbf{x}\right)
\right)   &  =\;\int\nolimits_{Z_{1}}p\left(  \mathbf{x}_{1_{i\ast}%
}|\boldsymbol{\tau}_{1}\right)  d\boldsymbol{\tau}_{1}+\int\nolimits_{Z_{2}%
}p\left(  \mathbf{x}_{1_{i\ast}}|\boldsymbol{\tau}_{1}\right)
d\boldsymbol{\tau}_{1}+\nabla_{eq}\\
&  =\int\nolimits_{Z_{1}}p\left(  \mathbf{x}_{2_{i\ast}}|\boldsymbol{\tau}%
_{2}\right)  d\boldsymbol{\tau}_{2}+\int\nolimits_{Z_{2}}p\left(
\mathbf{x}_{2_{i\ast}}|\boldsymbol{\tau}_{2}\right)  d\boldsymbol{\tau}%
_{2}-\nabla_{eq}\text{,}%
\end{align*}
where all of the forces associated with the counter risks and the risks for
class $\omega_{1}$ and class $\omega_{2}$ are symmetrically balanced with each
other%
\begin{align*}
f\left(  \widetilde{\Lambda}_{\boldsymbol{\tau}}\left(  \mathbf{x}\right)
\right)   &  :\;\int\nolimits_{Z_{1}}p\left(  \mathbf{x}_{1_{i\ast}%
}|\boldsymbol{\tau}_{1}\right)  d\boldsymbol{\tau}_{1}-\int\nolimits_{Z_{1}%
}p\left(  \mathbf{x}_{2_{i\ast}}|\boldsymbol{\tau}_{2}\right)
d\boldsymbol{\tau}_{2}+\nabla_{eq}\\
&  =\int\nolimits_{Z_{2}}p\left(  \mathbf{x}_{2_{i\ast}}|\boldsymbol{\tau}%
_{2}\right)  d\boldsymbol{\tau}_{2}-\int\nolimits_{Z_{2}}p\left(
\mathbf{x}_{1_{i\ast}}|\boldsymbol{\tau}_{1}\right)  d\boldsymbol{\tau}%
_{1}-\nabla_{eq}\text{,}%
\end{align*}
over the $Z_{1}$ and $Z_{2}$ decision regions.

Linear eigenlocus transforms involve symmetrically balanced, first and
second-order statistical moments of extreme data points. I\ will begin my
analysis by defining first and second-order statistical moments of data points.

\section{First and Second-Order Statistical Moments}

Consider again the Gram matrix $\mathbf{Q}$ associated with the constrained
quadratic form in Eq. (\ref{Vector Form Wolfe Dual})%
\begin{equation}
\mathbf{Q}=%
\begin{pmatrix}
\mathbf{x}_{1}^{T}\mathbf{x}_{1} & \mathbf{x}_{1}^{T}\mathbf{x}_{2} & \cdots &
-\mathbf{x}_{1}^{T}\mathbf{x}_{N}\\
\mathbf{x}_{2}^{T}\mathbf{x}_{1} & \mathbf{x}_{2}^{T}\mathbf{x}_{2} & \cdots &
-\mathbf{x}_{2}^{T}\mathbf{x}_{N}\\
\vdots & \vdots & \ddots & \vdots\\
-\mathbf{x}_{N}^{T}\mathbf{x}_{1} & -\mathbf{x}_{N}^{T}\mathbf{x}_{2} & \cdots
& \mathbf{x}_{N}^{T}\mathbf{x}_{N}%
\end{pmatrix}
\text{,} \label{Autocorrelation Matrix}%
\end{equation}
where $\mathbf{Q}\triangleq\widetilde{\mathbf{X}}\widetilde{\mathbf{X}}^{T}$,
$\widetilde{\mathbf{X}}\triangleq\mathbf{D}_{y}\mathbf{X}$, $\mathbf{D}_{y} $
is a $N\times N$ diagonal matrix of training labels $y_{i}$ and the $N\times
d$ data matrix is $\mathbf{X}$ $=%
\begin{pmatrix}
\mathbf{x}_{1}, & \mathbf{x}_{2}, & \ldots, & \mathbf{x}_{N}%
\end{pmatrix}
^{T}$. Without loss of generality (WLOG), assume that $N$ is an even number,
let the first $N/2$ vectors have the label $y_{i}=1$ and let the last $N/2$
vectors have the label $y_{i}=-1$. WLOG, the analysis that follows does not
take label information into account.

Using Eq. (\ref{Scalar Projection}), let the inner product statistic
$\mathbf{x}_{i}^{T}\mathbf{x}_{j}$ be interpreted as $\left\Vert
\mathbf{x}_{i}\right\Vert $ times the scalar projection $\left\Vert
\mathbf{x}_{j}\right\Vert \cos\theta_{\mathbf{x}_{i}\mathbf{x}_{j}}$ of
$\mathbf{x}_{j} $ onto $\mathbf{x}_{i}$. It follows that row $\mathbf{Q}%
\left(  i,:\right)  $ in Eq. (\ref{Autocorrelation Matrix}) contains uniformly
weighted $\left\Vert \mathbf{x}_{i}\right\Vert $ scalar projections
$\left\Vert \mathbf{x}_{j}\right\Vert \cos\theta_{\mathbf{x}_{i}\mathbf{x}%
_{j}}$ for each of the $N$ vectors $\left\{  \mathbf{x}_{j}\right\}
_{j=1}^{N} $ onto the vector $\mathbf{x}_{i}$:%
\begin{equation}
\widetilde{\mathbf{Q}}=%
\begin{pmatrix}
\left\Vert \mathbf{x}_{1}\right\Vert \left\Vert \mathbf{x}_{1}\right\Vert
\cos\theta_{\mathbf{x_{1}x}_{1}} & \cdots & -\left\Vert \mathbf{x}%
_{1}\right\Vert \left\Vert \mathbf{x}_{N}\right\Vert \cos\theta_{\mathbf{x}%
_{1}\mathbf{x}_{N}}\\
\left\Vert \mathbf{x}_{2}\right\Vert \left\Vert \mathbf{x}_{1}\right\Vert
\cos\theta_{\mathbf{x}_{2}\mathbf{x}_{1}} & \cdots & -\left\Vert
\mathbf{x}_{2}\right\Vert \left\Vert \mathbf{x}_{N}\right\Vert \cos
\theta_{\mathbf{x}_{2}\mathbf{x}_{N}}\\
\vdots & \ddots & \vdots\\
-\left\Vert \mathbf{x}_{N}\right\Vert \left\Vert \mathbf{x}_{1}\right\Vert
\cos\theta_{\mathbf{x}_{N}\mathbf{x}_{1}} & \cdots & \left\Vert \mathbf{x}%
_{N}\right\Vert \left\Vert \mathbf{x}_{N}\right\Vert \cos\theta_{\mathbf{x}%
_{N}\mathbf{x}_{N}}%
\end{pmatrix}
\text{,} \label{Inner Product Matrix}%
\end{equation}
where $0<\theta_{\mathbf{x}_{i}\mathbf{x}_{j}}\leq\frac{\pi}{2}$ or $\frac
{\pi}{2}<\theta_{\mathbf{x}_{i}\mathbf{x}_{j}}\leq\pi$. Alternatively, column
$\mathbf{Q}\left(  :,j\right)  $ in Eq. (\ref{Autocorrelation Matrix})
contains weighted $\left\Vert \mathbf{x}_{i}\right\Vert $ scalar
projections$\left\Vert \mathbf{x}_{j}\right\Vert \cos\theta_{\mathbf{x}%
_{i}\mathbf{x}_{j}}$ for the vector $\mathbf{x}_{j}$ onto each of the $N$
vectors $\left\{  \mathbf{x}_{i}\right\}  _{i=1}^{N}$.

Now consider the $i$th row $\widetilde{\mathbf{Q}}\left(  i,:\right)  $ of
$\widetilde{\mathbf{Q}}$ in Eq. (\ref{Inner Product Matrix}). Again, using Eq.
(\ref{Scalar Projection}), it follows that element $\widetilde{\mathbf{Q}%
}\left(  i,j\right)  $ of row $\widetilde{\mathbf{Q}}\left(  i,:\right)  $
specifies the length $\left\Vert \mathbf{x}_{i}\right\Vert $ of the vector
$\mathbf{x}_{i}$ multiplied by the scalar projection $\left\Vert
\mathbf{x}_{j}\right\Vert \cos\theta_{\mathbf{x}_{i}\mathbf{x}_{j}}$
of\ $\mathbf{x}_{j}$ onto $\mathbf{x}_{i}$:%
\[
\widetilde{\mathbf{Q}}\left(  i,j\right)  =\left\Vert \mathbf{x}%
_{i}\right\Vert \left[  \left\Vert \mathbf{x}_{j}\right\Vert \cos
\theta_{\mathbf{x}_{i}\mathbf{x}_{j}}\right]  \text{,}%
\]
where the signed magnitude of the vector projection of $\mathbf{x}_{j}$ along
the axis of $\mathbf{x}_{i}$%
\[
\operatorname{comp}_{\overrightarrow{\mathbf{x}}_{i}}\left(
\overrightarrow{\mathbf{x}}_{j}\right)  =\left\Vert \mathbf{x}_{j}\right\Vert
\cos\theta_{\mathbf{x}_{i}\mathbf{x}_{j}}=\mathbf{x}_{j}^{T}\left(
\frac{\mathbf{x}_{i}}{\left\Vert \mathbf{x}_{i}\right\Vert }\right)
\]
provides a measure $\widehat{\mathbf{x}}_{j}$ of the first degree components
of the vector $\mathbf{x}_{j}$%
\[
\mathbf{x}_{j}=\left(  x_{j_{1}},x_{j_{2}},\cdots,x_{j_{d}}\right)  ^{T}%
\]
along the axis of the vector $\mathbf{x}_{i}$%
\[
\mathbf{x}_{i}=\left(  x_{i_{1}},x_{i_{2}},\cdots,x_{i_{d}}\right)
^{T}\text{.}%
\]

Accordingly, the signed magnitude $\left\Vert \mathbf{x}_{j}\right\Vert
\cos\theta_{\mathbf{x}_{i}\mathbf{x}_{j}}$ provides an estimate
$\widehat{\mathbf{x}}_{i}$ for the amount of first degree components of the
vector $\mathbf{x}_{i}$ that are distributed over the axis of the vector
$\mathbf{x}_{j}$. This indicates that signed magnitudes $\left\Vert
\mathbf{x}_{j}\right\Vert \cos\theta_{\mathbf{x}_{i}\mathbf{x}_{j}}$ contained
with $\widetilde{\mathbf{Q}}$ account for how the first degree coordinates of
a data point $\mathbf{x}_{i}$ are distributed along the axes of a set of
vectors $\left\{  \mathbf{x}_{j}\right\}  _{j=1}^{N}$ within Euclidean space.

Using the above assumptions and notation, for any given row
$\widetilde{\mathbf{Q}}\left(  i,:\right)  $ of Eq.
(\ref{Inner Product Matrix}), it follows that the statistic denoted by
$E_{\widehat{\mathbf{x}}_{i}}\left[  \mathbf{x}_{i}|\left\{  \mathbf{x}%
_{j}\right\}  _{j=1}^{N}\right]  $%
\begin{align}
E_{\widehat{\mathbf{x}}_{i}}\left[  \mathbf{x}_{i}|\left\{  \mathbf{x}%
_{j}\right\}  _{j=1}^{N}\right]   &  =\left\Vert \mathbf{x}_{i}\right\Vert
{\displaystyle\sum\nolimits_{j}}
\operatorname{comp}_{\overrightarrow{\mathbf{x}}_{i}}\left(
\overrightarrow{\mathbf{x}}_{j}\right)
\label{Row Distribution First Order Vector Coordinates}\\
&  =\left\Vert \mathbf{x}_{i}\right\Vert
{\displaystyle\sum\nolimits_{j}}
\left\Vert \mathbf{x}_{j}\right\Vert \cos\theta_{\mathbf{x}_{i}\mathbf{x}_{j}%
}\nonumber
\end{align}
provides an estimate $E_{\widehat{\mathbf{x}}_{i}}\left[  \mathbf{x}%
_{i}|\left\{  \mathbf{x}_{j}\right\}  _{j=1}^{N}\right]  $ for the amount of
first degree components of a vector $\mathbf{x}_{i}$ that are distributed over
the axes of a set of vectors $\left\{  \mathbf{x}_{j}\right\}  _{j=1}^{N}$,
where labels have not been taken into account.

Thereby, Eq. (\ref{Row Distribution First Order Vector Coordinates}) describes
a distribution of first degree coordinates for a vector $\mathbf{x}_{i}$ in a
data collection.

Given that Eq. (\ref{Row Distribution First Order Vector Coordinates})
involves signed magnitudes of vector projections along the axis of a fixed
vector $\mathbf{x}_{i}$, the distribution of first degree vector coordinates
described by Eq. (\ref{Row Distribution First Order Vector Coordinates}) is
said to determine a \emph{first-order statistical moment about the locus of a
data point} $\mathbf{x}_{i}$. Because the statistic $E_{\widehat{\mathbf{x}%
}_{i}}\left[  \mathbf{x}_{i}|\left\{  \mathbf{x}_{j}\right\}  _{j=1}%
^{N}\right]  $ depends on the uniform direction of a fixed vector
$\mathbf{x}_{i}$, the statistic $E_{\widehat{\mathbf{x}}_{i}}\left[
\mathbf{x}_{i}|\left\{  \mathbf{x}_{j}\right\}  _{j=1}^{N}\right]  $ is said
to be unidirectional.

In the next section, I will devise pointwise covariance statistics that
provide unidirectional estimates of covariance along a fixed reference axis.

\subsection{Unidirectional Covariance Statistics}

Classical covariance statistics provide omnidirectional estimates of
covariance along $N$ axes of $N$ vectors. I\ will now argue that such
omnidirectional statistics provide non-coherent estimates of covariance.

\subsubsection{Omnidirectional Covariance Statistics}

Take the data matrix $\mathbf{X}$ $=%
\begin{pmatrix}
\mathbf{x}_{1}, & \mathbf{x}_{2}, & \ldots, & \mathbf{x}_{N}%
\end{pmatrix}
^{T}$ and consider the classical covariance statistic:%
\begin{align*}
\widehat{\operatorname{cov}}\left(  \mathbf{X}\right)   &  =\frac{1}{N}%
{\displaystyle\sum\nolimits_{i}}
\left(  \mathbf{x}_{i}-\overline{\mathbf{x}}\right)  ^{2}\text{,}\\
&  =\frac{1}{N}%
{\displaystyle\sum\nolimits_{i}}
\left(  \mathbf{x}_{i}-\left(  \frac{1}{N}%
{\displaystyle\sum\nolimits_{i}}
\mathbf{x}_{i}\right)  \right)  ^{2}%
\end{align*}
written in vector notation. The statistic $\widehat{\operatorname{cov}}\left(
\mathbf{X}\right)  $ measures the square of the Euclidean distance between a
common mean vector $\overline{\mathbf{x}}$ and each of the vectors
$\mathbf{x}_{i}$ in a collection of data $\left\{  \mathbf{x}_{i}\right\}
_{i=1}^{N}$
\citep{Ash1993,Flury1997}%
.

Because the statistic $\widehat{\operatorname{cov}}\left(  \mathbf{X}\right)
$ depends on $N$ directions of $N$ training vectors, the statistic
$\widehat{\operatorname{cov}}\left(  \mathbf{X}\right)  $ is said to be
omnidirectional and is considered to be a non-coherent estimate of covariance.
Accordingly, the statistic $\widehat{\operatorname{cov}}\left(  \mathbf{X}%
\right)  $ provides an omnidirectional, non-coherent estimate of the joint
variations of the $d\times N$ random variables of a collection of $N$ random
vectors $\left\{  \mathbf{x}_{i}\right\}  _{i=1}^{N}$ about the $d$ random
variables of the mean vector $\overline{\mathbf{x}}$.

I will now devise pointwise covariance statistics $\widehat{\operatorname{cov}%
}_{up}\left(  \mathbf{x}_{i}\right)  $ for individual vectors. Pointwise
covariance statistics are unidirectional statistics that provide coherent
estimates of covariance along a fixed reference axis. WLOG, label information
is not taken into consideration.

\subsubsection{Pointwise Covariance Statistics}

Take any row $\widetilde{\mathbf{Q}}\left(  i,:\right)  $ of the matrix
$\widetilde{\mathbf{Q}}$ in Eq. (\ref{Inner Product Matrix}) and consider the
inner product statistic $\left\Vert \mathbf{x}_{i}\right\Vert \left\Vert
\mathbf{x}_{j}\right\Vert \cos\theta_{\mathbf{x}_{i}\mathbf{x}_{j}}$ in
element $\widetilde{\mathbf{Q}}\left(  i,j\right)  $. Using Eqs
(\ref{Geometric Locus of Vector}) and (\ref{Inner Product Statistic}), it
follows that element $\widetilde{\mathbf{Q}}\left(  i,j\right)  $ in row
$\widetilde{\mathbf{Q}}\left(  i,:\right)  $ specifies the joint variations
$\operatorname{cov}\left(  \mathbf{x}_{i},\mathbf{x}_{j}\right)  $%
\[
\operatorname{cov}\left(  \mathbf{x}_{i},\mathbf{x}_{j}\right)  =\left\Vert
\mathbf{x}_{i}\right\Vert \left\Vert \mathbf{x}_{j}\right\Vert \cos
\theta_{\mathbf{x}_{i}\mathbf{x}_{j}}%
\]
between the components of the vector $\mathbf{x}_{i}$%
\[%
\begin{pmatrix}
\left\Vert \mathbf{x}_{i}\right\Vert \cos\mathbb{\alpha}_{\mathbf{x}_{i1}1}, &
\left\Vert \mathbf{x}_{i}\right\Vert \cos\mathbb{\alpha}_{\mathbf{x}_{i2}2}, &
\cdots, & \left\Vert \mathbf{x}_{i}\right\Vert \cos\mathbb{\alpha}%
_{\mathbf{x}_{id}d}%
\end{pmatrix}
\]
and the components of the vector $\mathbf{x}_{j}$%
\[%
\begin{pmatrix}
\left\Vert \mathbf{x}_{j}\right\Vert \cos\mathbb{\alpha}_{\mathbf{x}_{j1}1}, &
\left\Vert \mathbf{x}_{j}\right\Vert \cos\mathbb{\alpha}_{\mathbf{x}_{j2}2}, &
\cdots, & \left\Vert \mathbf{x}_{j}\right\Vert \cos\mathbb{\alpha}%
_{\mathbf{x}_{jd}d}%
\end{pmatrix}
\text{,}%
\]
where the $d$ components $\left\{  \left\Vert \mathbf{x}\right\Vert
\cos\mathbb{\alpha}_{x_{i}i}\right\}  _{i=1}^{d}$ of any given vector
$\mathbf{x}$ are random variables, each of which is characterized by an
expected value $E\left[  \left\Vert \mathbf{x}\right\Vert \cos\mathbb{\alpha
}_{x_{i}i}\right]  $ and a variance $\operatorname{var}\left(  \left\Vert
\mathbf{x}\right\Vert \cos\mathbb{\alpha}_{x_{i}i}\right)  $. It follows that
the $j$th element $\widetilde{\mathbf{Q}}\left(  i,j\right)  $ of row
$\widetilde{\mathbf{Q}}\left(  i,:\right)  $ specifies the joint variations of
the $d$ random variables of a vector $\mathbf{x}_{j}$ about the $d$ random
variables of the vector $\mathbf{x}_{i}$.

Thus, row $\widetilde{\mathbf{Q}}\left(  i,:\right)  $ specifies the joint
variations between the random variables of a fixed vector $\mathbf{x}_{i}$ and
the random variables of an entire collection of data.

Again, take any row $\widetilde{\mathbf{Q}}\left(  i,:\right)  $ of the matrix
$\widetilde{\mathbf{Q}}$ in Eq. (\ref{Inner Product Matrix}). Using the
definition of the scalar project statistic in Eq. (\ref{Scalar Projection}),
it follows that the statistic $\widehat{\operatorname{cov}}_{up}\left(
\mathbf{x}_{i}\right)  $:%
\begin{align}
\widehat{\operatorname{cov}}_{up}\left(  \mathbf{x}_{i}\right)   &  =%
{\displaystyle\sum\nolimits_{j=1}^{N}}
\left\Vert \mathbf{x}_{i}\right\Vert \left\Vert \mathbf{x}_{j}\right\Vert
\cos\theta_{\mathbf{x}_{i}\mathbf{x}_{j}}%
\label{Pointwise Covariance Statistic}\\
&  =%
{\displaystyle\sum\nolimits_{j=1}^{N}}
\mathbf{x}_{i}^{T}\mathbf{x}_{j}=\mathbf{x}_{i}^{T}\left(
{\displaystyle\sum\nolimits_{j=1}^{N}}
\mathbf{x}_{j}\right) \nonumber\\
&  =\left\Vert \mathbf{x}_{i}\right\Vert
{\displaystyle\sum\nolimits_{j=1}^{N}}
\left\Vert \mathbf{x}_{j}\right\Vert \cos\theta_{\mathbf{x}_{i}\mathbf{x}_{j}%
}\nonumber
\end{align}
provides a unidirectional estimate of the joint variations of the $d$ random
variables of each of the $N$ vectors of a data collection $\left\{
\mathbf{x}_{j}\right\}  _{j=1}^{N}$ and a unidirectional estimate of the joint
variations of the $d$ random variables of the common mean $%
{\displaystyle\sum\nolimits_{j=1}^{N}}
\mathbf{x}_{j}$ of the data, about the $d$ random variables of a fixed vector
$\mathbf{x}_{i}$, along the axis of the fixed vector $\mathbf{x}_{i}$.

Thereby, the statistic $\widehat{\operatorname{cov}}_{up}\left(
\mathbf{x}_{i}\right)  $ specifies the direction of a vector $\mathbf{x}_{i}$
and a signed magnitude along the axis of the vector $\mathbf{x}_{i}$.

The statistic $\widehat{\operatorname{cov}}_{up}\left(  \mathbf{x}_{i}\right)
$ in Eq. (\ref{Pointwise Covariance Statistic}) is defined to be a pointwise
covariance estimate for a data point $\mathbf{x}_{i}$, where the statistic
$\widehat{\operatorname{cov}}_{up}\left(  \mathbf{x}_{i}\right)  $ provides a
unidirectional estimate of the joint variations between the random variables
of each vector $\mathbf{x}_{j}$ in a data collection and the random variables
of a fixed vector $\mathbf{x}_{i}$ and a unidirectional estimate of the joint
variations between the random variables of the mean vector $%
{\displaystyle\sum\nolimits_{j=1}^{N}}
\mathbf{x}_{j}$ and the fixed vector $\mathbf{x}_{i} $. Given that the joint
variations estimated by the statistic $\widehat{\operatorname{cov}}%
_{up}\left(  \mathbf{x}_{i}\right)  $ are derived from second-order distance
statistics $\left\Vert \mathbf{x}_{i}-\mathbf{x}_{j}\right\Vert ^{2} $ which
involve signed magnitudes of vector projections along the common axis of a
fixed vector $\mathbf{x}_{i}$, a pointwise covariance estimate
$\widehat{\operatorname{cov}}_{up}\left(  \mathbf{x}_{i}\right)  $ is said to
determine a \emph{second-order statistical moment about the locus of a data
point} $\mathbf{x}_{i}$. Using Eq.
(\ref{Row Distribution First Order Vector Coordinates}), Eq.
(\ref{Pointwise Covariance Statistic}) also specifies a distribution of first
order coordinates for a given vector $\mathbf{x}_{i}$ which determines a
first-order statistical moment about the locus of the data point
$\mathbf{x}_{i}$.

I\ will now demonstrate that pointwise covariance statistics can be used to
discover extreme points.

\subsection{Discovery of Extreme Data Points}

The Gram matrix associated with the constrained quadratic form in Eq.
(\ref{Vector Form Wolfe Dual}) contains inner product statistics for\ two
labeled collections of data. Denote those data points that belong to class
$\omega_{1}$ by $\mathbf{x}_{1_{i}}$ and those that belong to class
$\omega_{2}$ by $\mathbf{x}_{2_{i}}$. Let $\overline{\mathbf{x}}_{1}$ and
$\overline{\mathbf{x}}_{2}$ denote the mean vectors of class $\omega_{1}$ and
class $\omega_{2}$. Let $i=1:n_{1}$ where the vector $\mathbf{x}_{1_{i}}$ has
the label $y_{i}=1$ and let $i=n_{1}+1:n_{1}+n_{2}$ where the vector
$\mathbf{x}_{2_{i}}$ has the label $y_{i}=-1$. Using label information, Eq.
(\ref{Pointwise Covariance Statistic}) can be rewritten as%
\[
\widehat{\operatorname{cov}}_{up}\left(  \mathbf{x}_{1_{i}}\right)
=\mathbf{x}_{1_{i}}^{T}\left(  \sum\nolimits_{j=1}^{n_{1}}\mathbf{x}_{1_{j}%
}-\sum\nolimits_{j=n_{1}+1}^{n_{1}+n_{2}}\mathbf{x}_{2_{j}}\right)
\]
and%
\[
\widehat{\operatorname{cov}}_{up}\left(  \mathbf{x}_{2_{i}}\right)
=\mathbf{x}_{2_{i}}^{T}\left(  \sum\nolimits_{j=n_{1}+1}^{n_{1}+n_{2}%
}\mathbf{x}_{2_{j}}-\sum\nolimits_{j=1}^{n_{1}}\mathbf{x}_{1_{j}}\right)
\text{.}%
\]

I will now show that extreme points possess large pointwise covariances
relative to the non-extreme points in each respective pattern class. Recall
that an extreme point is located relatively far from its distribution mean,
relatively close to the mean of the other distribution and relatively close to
other extreme points. Denote an extreme point by $\mathbf{x}_{1_{i\ast}}$ or
$\mathbf{x}_{2_{i\ast}}$ and a non-extreme point by $\mathbf{x}_{1_{i}}$ or
$\mathbf{x}_{2_{i}}$.

Take any extreme point $\mathbf{x}_{1_{i\ast}}$ and any non-extreme point
$\mathbf{x}_{1_{i}}$ that belong to class $\omega_{1}$ and consider the
pointwise covariance estimates for $\mathbf{x}_{1_{i\ast}}$:%
\[
\widehat{\operatorname{cov}}_{up}\left(  \mathbf{x}_{1_{i\ast}}\right)
=\mathbf{x}_{1_{i\ast}}^{T}\overline{\mathbf{x}}_{1}-\mathbf{x}_{1_{i\ast}%
}^{T}\overline{\mathbf{x}}_{2}%
\]
and for $\mathbf{x}_{1_{i}}$:%
\[
\widehat{\operatorname{cov}}_{up}\left(  \mathbf{x}_{1}\right)  =\mathbf{x}%
_{1_{i}}^{T}\overline{\mathbf{x}}_{1}-\mathbf{x}_{1_{i}}^{T}\overline
{\mathbf{x}}_{2}\text{.}%
\]

Because $\mathbf{x}_{1_{i\ast}}$ is an extreme point, it follows that
$\mathbf{x}_{1_{i\ast}}^{T}\overline{\mathbf{x}}_{1}>$ $\mathbf{x}_{1_{i}}%
^{T}\overline{\mathbf{x}}_{1}$ and that $\mathbf{x}_{1_{i\ast}}^{T}%
\overline{\mathbf{x}}_{2}<$ $\mathbf{x}_{1_{i}}^{T}\overline{\mathbf{x}}_{2}$.
Thus, $\widehat{\operatorname{cov}}_{up}\left(  \mathbf{x}_{1_{i\ast}}\right)
>\widehat{\operatorname{cov}}_{up}\left(  \mathbf{x}_{1_{i}}\right)  $.
Therefore, each extreme point $\mathbf{x}_{1_{i\ast}}$ exhibits a pointwise
covariance $\widehat{\operatorname{cov}}_{up}\left(  \mathbf{x}_{1_{i\ast}%
}\right)  $ that exceeds the pointwise covariance $\widehat{\operatorname{cov}%
}_{up}\left(  \mathbf{x}_{1}\right)  $ of all of the non-extreme points
$\mathbf{x}_{1}$ in class $\omega_{1}$.

Now take any extreme point $\mathbf{x}_{2_{i\ast}}$ and any non-extreme point
$\mathbf{x}_{2_{i}}$ that belong to class $\omega_{2}$ and consider the
pointwise covariance estimates for $\mathbf{x}_{2_{i\ast}}$:%
\[
\widehat{\operatorname{cov}}_{up}\left(  \mathbf{x}_{2_{i\ast}}\right)
=\mathbf{x}_{2_{i\ast}}^{T}\overline{\mathbf{x}}_{2}-\mathbf{x}_{2_{i\ast}%
}^{T}\overline{\mathbf{x}}_{1}%
\]
and for $\mathbf{x}_{2_{i}}$:%
\[
\widehat{\operatorname{cov}}_{up}\left(  \mathbf{x}_{2}\right)  =\mathbf{x}%
_{2_{i}}^{T}\overline{\mathbf{x}}_{2}-\mathbf{x}_{2_{i}}^{T}\overline
{\mathbf{x}}_{1}\text{.}%
\]

Because $\mathbf{x}_{2_{i\ast}}$ is an extreme point, it follows that
$\mathbf{x}_{2_{i\ast}}^{T}\overline{\mathbf{x}}_{2}>$ $\mathbf{x}_{2_{i}}%
^{T}\overline{\mathbf{x}}_{2}$ and that $\mathbf{x}_{2_{i\ast}}^{T}%
\overline{\mathbf{x}}_{1}<$ $\mathbf{x}_{2_{i}}^{T}\overline{\mathbf{x}}_{1}$.
Thus, $\widehat{\operatorname{cov}}_{up}\left(  \mathbf{x}_{2_{i\ast}}\right)
>\widehat{\operatorname{cov}}_{up}\left(  \mathbf{x}_{2_{i}}\right)  $.
Therefore, each extreme point $\mathbf{x}_{2_{i\ast}}$ exhibits a pointwise
covariance $\widehat{\operatorname{cov}}_{up}\left(  \mathbf{x}_{2_{i\ast}%
}\right)  $ that exceeds the pointwise covariance $\widehat{\operatorname{cov}%
}_{up}\left(  \mathbf{x}_{2}\right)  $ of all of the non-extreme points
$\mathbf{x}_{2}$ in class $\omega_{2}$.

Thereby, it is concluded that extreme points possess large pointwise
covariances relative to non-extreme points in their respective pattern class.
It is also concluded that the pointwise covariance
$\widehat{\operatorname{cov}}_{up}\left(  \mathbf{x}_{1_{i\ast}}\right)  $ or
$\widehat{\operatorname{cov}}_{up}\left(  \mathbf{x}_{2_{i\ast}}\right)  $
exhibited by any given extreme point $\mathbf{x}_{1_{i\ast}}$ or
$\mathbf{x}_{2_{i\ast}}$ may exceed pointwise covariances of other extreme
points in each respective pattern class.

Therefore, it will be assumed that each extreme point $\mathbf{x}_{1_{i_{\ast
}}}$ or $\mathbf{x}_{2_{i_{\ast}}}$ exhibits a critical first and second-order
statistical moment $\widehat{\operatorname{cov}}_{up}\left(  \mathbf{x}%
_{1_{i_{\ast}}}\right)  $ or $\widehat{\operatorname{cov}}_{up}\left(
\mathbf{x}_{2_{i\ast}}\right)  $ that exceeds some threshold $\varrho$, for
which each corresponding scale factor $\psi_{1i\ast}$ or $\psi_{2i\ast}$
exhibits a critical value that exceeds zero $\psi_{1i\ast}>0$ or $\psi
_{2i\ast}>0$. Accordingly, first and second-order statistical moments
$\widehat{\operatorname{cov}}_{up}\left(  \mathbf{x}_{1_{i}}\right)  $ or
$\widehat{\operatorname{cov}}_{up}\left(  \mathbf{x}_{2_{i}}\right)  $ about
the loci of non-extreme points $\mathbf{x}_{1_{i}}$ or $\mathbf{x}_{2_{i}}$ do
not exceed the threshold $\varrho$ and their corresponding scale factors
$\psi_{1i}$ or $\psi_{2i}$ are effectively zero: $\psi_{1i}=0$ or $\psi
_{2i}=0$.

I will now devise a system of equations for a principal eigen-decomposition of
the Gram matrix $\mathbf{Q}$ denoted in Eqs (\ref{Autocorrelation Matrix}) and
(\ref{Inner Product Matrix}) that describes tractable point and coordinate
relationships between the scaled extreme points on $\boldsymbol{\tau}_{1}$ and
$\boldsymbol{\tau}_{2}$ and the Wolfe dual principal eigenaxis components on
$\boldsymbol{\psi}_{1}$ and $\boldsymbol{\psi}_{2}$.

\section{Inside the Wolfe Dual Eigenspace I}

Take the Gram matrix $\mathbf{Q}$ associated with the quadratic form in Eq.
(\ref{Vector Form Wolfe Dual}). Let $\mathbf{q}_{j\text{ }}$denote the $j$th
column of $\mathbf{Q}$, which is an $N$-vector. Let $\lambda_{\max
_{\boldsymbol{\psi}}}$ and $\boldsymbol{\psi}$ denote the largest eigenvalue
and largest eigenvector of $\mathbf{Q}$ respectively. Using this notation
\citep[see][]{Trefethen1998}%
, the principal eigen-decomposition of $\mathbf{Q}$%
\[
\mathbf{Q}\boldsymbol{\psi}=\lambda\mathbf{_{\max_{\boldsymbol{\psi}}}%
}\boldsymbol{\psi}%
\]
can be rewritten as%
\[
\lambda_{\max_{\boldsymbol{\psi}}}\boldsymbol{\psi}=%
{\displaystyle\sum\nolimits_{j=1}^{N}}
\psi_{_{j}}\mathbf{q}_{j\text{ }}%
\]
so that the principal eigenaxis $\boldsymbol{\psi}$ of $\mathbf{Q}$ is
expressed as a linear combination of transformed vectors $\frac{\psi_{j}%
}{\lambda_{\max_{\boldsymbol{\psi}}}}\mathbf{q}_{j\text{ }}$:%
\begin{equation}
\left[
\begin{array}
[c]{c}%
\\
\boldsymbol{\psi}\\
\\
\end{array}
\right]  =\frac{\psi_{1}}{\lambda_{\max_{\boldsymbol{\psi}}}}\left[
\begin{array}
[c]{c}%
\\
\mathbf{q}_{1\text{ }}\\
\\
\end{array}
\right]  +\frac{\psi_{2}}{\lambda_{\max_{\boldsymbol{\psi}}}}\left[
\begin{array}
[c]{c}%
\\
\mathbf{q}_{2\text{ }}\\
\\
\end{array}
\right]  +\cdots+\frac{\psi_{N}}{\lambda_{\max_{\boldsymbol{\psi}}}}\left[
\begin{array}
[c]{c}%
\\
\mathbf{q}_{N\text{ }}\\
\\
\end{array}
\right]  \text{,} \label{Alternate Eigendecomposition Equation}%
\end{equation}
where the $i$th element of the vector $\mathbf{q}_{j\text{ }}$ specifies an
inner product statistic $\mathbf{x}_{i}^{T}\mathbf{x}_{j}$ between the vectors
$\mathbf{x}_{i}$ and $\mathbf{x}_{j}$.

Using Eqs (\ref{Autocorrelation Matrix}) and
(\ref{Alternate Eigendecomposition Equation}), a Wolfe dual linear eigenlocus
$\left(  \psi_{1},\cdots,\psi_{N}\right)  ^{T}$ can be written as:%
\begin{align}
\boldsymbol{\psi}  &  =\frac{\psi_{1}}{\lambda_{\max_{\boldsymbol{\psi}}}}%
\begin{pmatrix}
\mathbf{x}_{1}^{T}\mathbf{x}_{1}\\
\mathbf{x}_{2}^{T}\mathbf{x}_{1}\\
\vdots\\
-\mathbf{x}_{N}^{T}\mathbf{x}_{1}%
\end{pmatrix}
+\frac{\psi_{2}}{\lambda_{\max_{\boldsymbol{\psi}}}}%
\begin{pmatrix}
\mathbf{x}_{1}^{T}\mathbf{x}_{2}\\
\mathbf{x}_{2}^{T}\mathbf{x}_{2}\\
\vdots\\
-\mathbf{x}_{N}^{T}\mathbf{x}_{2}%
\end{pmatrix}
+\cdots\label{Dual Normal Eigenlocus Components}\\
\cdots &  +\frac{\psi_{N-1}}{\lambda_{\max_{\boldsymbol{\psi}}}}%
\begin{pmatrix}
-\mathbf{x}_{1}^{T}\mathbf{x}_{N-1}\\
-\mathbf{x}_{2}^{T}\mathbf{x}_{N-1}\\
\vdots\\
\mathbf{x}_{N}^{T}\mathbf{x}_{N-1}%
\end{pmatrix}
+\frac{\psi_{N}}{\lambda_{\max_{\boldsymbol{\psi}}}}%
\begin{pmatrix}
-\mathbf{x}_{1}^{T}\mathbf{x}_{N}\\
-\mathbf{x}_{2}^{T}\mathbf{x}_{N}\\
\vdots\\
\mathbf{x}_{N}^{T}\mathbf{x}_{N}%
\end{pmatrix}
\nonumber
\end{align}
which illustrates that the magnitude $\psi_{j}$ of the $j^{th}$ Wolfe dual
principal eigenaxis component $\psi_{j}\overrightarrow{\mathbf{e}}_{j}$ is
correlated with joint variations of labeled vectors $\mathbf{x}_{i}$ about the
vector $\mathbf{x}_{j}$.

Alternatively, using Eqs (\ref{Inner Product Matrix}) and
(\ref{Alternate Eigendecomposition Equation}), a Wolfe dual linear eigenlocus
$\boldsymbol{\psi}$ can be written as:%
\begin{align}
\boldsymbol{\psi}  &  =\frac{\psi_{1}}{\lambda_{\max_{\boldsymbol{\psi}}}%
}\left(
\begin{array}
[c]{c}%
\left\Vert \mathbf{x}_{1}\right\Vert \left\Vert \mathbf{x}_{1}\right\Vert
\cos\theta_{\mathbf{x}_{1}^{T}\mathbf{x}_{1}}\\
\left\Vert \mathbf{x}_{2}\right\Vert \left\Vert \mathbf{x}_{1}\right\Vert
\cos\theta_{\mathbf{x}_{2}^{T}\mathbf{x}_{1}}\\
\vdots\\
-\left\Vert \mathbf{x}_{N}\right\Vert \left\Vert \mathbf{x}_{1}\right\Vert
\cos\theta_{\mathbf{x}_{N}^{T}\mathbf{x}_{1}}%
\end{array}
\right)  +\cdots\label{Dual Normal Eigenlocus Component Projections}\\
&  \cdots+\frac{\psi_{N}}{\lambda_{\max_{\boldsymbol{\psi}}}}\left(
\begin{array}
[c]{c}%
-\left\Vert \mathbf{x}_{1}\right\Vert \left\Vert \mathbf{x}_{N}\right\Vert
\cos\theta_{\mathbf{x}_{1}^{T}\mathbf{x}_{N}}\\
-\left\Vert \mathbf{x}_{2}\right\Vert \left\Vert \mathbf{x}_{N}\right\Vert
\cos\theta_{\mathbf{x}_{2}^{T}\mathbf{x}_{N}}\\
\vdots\\
\left\Vert \mathbf{x}_{N}\right\Vert \left\Vert \mathbf{x}_{N}\right\Vert
\cos\theta_{\mathbf{x}_{N}^{T}\mathbf{x}_{N}}%
\end{array}
\right) \nonumber
\end{align}
which illustrates that the magnitude $\psi_{j}$ of the $j^{th}$ Wolfe dual
principal eigenaxis component $\psi_{j}\overrightarrow{\mathbf{e}}_{j}$ on
$\boldsymbol{\psi}$ is correlated with scalar projections $\left\Vert
\mathbf{x}_{j}\right\Vert \cos\theta_{\mathbf{x}_{i}\mathbf{x}_{j}}$ of the
vector $\mathbf{x}_{j}$ onto labeled vectors $\mathbf{x}_{i}$.

Equations (\ref{Dual Normal Eigenlocus Components}) and
(\ref{Dual Normal Eigenlocus Component Projections}) both indicate that the
magnitude $\psi_{j} $ of the $j^{th}$ Wolfe dual principal eigenaxis component
$\psi_{j}\overrightarrow{\mathbf{e}}_{j}$ on $\boldsymbol{\psi}$ is correlated
with a first and second-order statistical moment about the locus of the data
point $\mathbf{x}_{j}$.

\subsection{Assumptions}

It will be assumed that each extreme point $\mathbf{x}_{1_{i_{\ast}}}$ or
$\mathbf{x}_{2_{i_{\ast}}}$ exhibits a critical first and second-order
statistical moment $\widehat{\operatorname{cov}}_{up}\left(  \mathbf{x}%
_{1_{i_{\ast}}}\right)  $ or $\widehat{\operatorname{cov}}_{up}\left(
\mathbf{x}_{2_{i\ast}}\right)  $ that exceeds some threshold $\varrho$, for
which each corresponding scale factor $\psi_{1i\ast}$ or $\psi_{2i\ast}$
exhibits a critical value that exceeds zero: $\psi_{1i\ast}>0$ or
$\psi_{2i\ast}>0$. It will also be assumed that first and second-order
statistical moments $\widehat{\operatorname{cov}}_{up}\left(  \mathbf{x}%
_{1_{i}}\right)  $ or $\widehat{\operatorname{cov}}_{up}\left(  \mathbf{x}%
_{2_{i}}\right)  $ about the loci of non-extreme points $\mathbf{x}_{1_{i}}$
or $\mathbf{x}_{2_{i}}$ do not exceed the threshold $\varrho$ so that the
corresponding scale factors $\psi_{1i}$ or $\psi_{2i}$ are effectively zero:
$\psi_{1i}=0$ or $\psi_{2i}=0$.

Express a Wolfe dual linear eigenlocus $\boldsymbol{\psi}$ in terms of $l$
non-orthogonal unit vectors $\left\{  \overrightarrow{\mathbf{e}}_{1\ast
},\ldots,\overrightarrow{\mathbf{e}}_{l\ast}\right\}  $%
\begin{align}
\boldsymbol{\psi}  &  =\sum\nolimits_{i=1}^{l}\psi_{i\ast}%
\overrightarrow{\mathbf{e}}_{i\ast}%
\label{Non-orthogonal Eigenaxes of Dual Normal Eigenlocus}\\
&  =\sum\nolimits_{i=1}^{l_{1}}\psi_{1i\ast}\overrightarrow{\mathbf{e}%
}_{1i\ast}+\sum\nolimits_{i=1}^{l_{2}}\psi_{2i\ast}\overrightarrow{\mathbf{e}%
}_{2i\ast}\text{,}\nonumber
\end{align}
where each scaled, non-orthogonal unit vector denoted by $\psi_{1i\ast
}\overrightarrow{\mathbf{e}}_{1i\ast}$ or $\psi_{2i\ast}%
\overrightarrow{\mathbf{e}}_{2i\ast}$ is correlated with an extreme vector
$\mathbf{x}_{1_{i\ast}}$ or $\mathbf{x}_{2_{i\ast}}$ respectively.
Accordingly, each Wolfe dual principal eigenaxis component $\psi_{1i\ast
}\overrightarrow{\mathbf{e}}_{1i\ast}$ or $\psi_{2i\ast}%
\overrightarrow{\mathbf{e}}_{2i\ast}$ is a scaled, non-orthogonal unit vector
that contributes to the estimation of $\boldsymbol{\psi}$ and
$\boldsymbol{\tau}$.

WLOG, indices do not indicate locations of inner product expressions in Eq.
(\ref{Dual Normal Eigenlocus Component Projections}).

\subsubsection*{Extreme Point Notation}

Denote the extreme points that belong to class $\omega_{1}$ and $\omega_{2}$
by $\mathbf{x}_{1_{i_{\ast}}}$ and $\mathbf{x}_{2_{i\ast}}$ with labels
$y_{i}=1$ and $y_{i}=-1$ respectively. Let there be $l_{1}$ extreme points
$\mathbf{x}_{1_{i\ast}}$ from class $\omega_{1}$ and $l_{2}$ extreme points
$\mathbf{x}_{2_{i\ast}}$ from class $\omega_{2}$.

Let there be $l_{1}$ principal eigenaxis components $\psi_{1i\ast
}\overrightarrow{\mathbf{e}}_{1i\ast}$, where each scale factor $\psi_{1i\ast
}$ is correlated with an extreme vector $\mathbf{x}_{1_{i_{\ast}}}$. Let there
be $l_{2}$ principal eigenaxis components $\psi_{2i\ast}%
\overrightarrow{\mathbf{e}}_{2i\ast}$, where each scale factor $\psi_{2i\ast}$
is correlated with an extreme vector $\mathbf{x}_{2_{i_{\ast}}}$. Let
$l_{1}+l_{2}=l$.

Recall that the risk $\mathfrak{R}_{\mathfrak{\min}}\left(  Z|p\left(
\widehat{\Lambda}\left(  \mathbf{x}\right)  \right)  \right)  $:%
\[
\mathfrak{R}_{\mathfrak{\min}}\left(  Z|p\left(  \widehat{\Lambda}\left(
\mathbf{x}\right)  \right)  \right)  =\mathfrak{R}_{\mathfrak{\min}}\left(
Z|p\left(  \widehat{\Lambda}\left(  \mathbf{x}\right)  |\omega_{1}\right)
\right)  +\mathfrak{R}_{\mathfrak{\min}}\left(  Z|p\left(  \widehat{\Lambda
}\left(  \mathbf{x}\right)  |\omega_{2}\right)  \right)
\]
for a binary classification system involves \emph{opposing forces} that depend
on the likelihood ratio test $\widehat{\Lambda}\left(  \mathbf{x}\right)
=p\left(  \widehat{\Lambda}\left(  \mathbf{x}\right)  |\omega_{1}\right)
-p\left(  \widehat{\Lambda}\left(  \mathbf{x}\right)  |\omega_{2}\right)
\overset{\omega_{1}}{\underset{\omega_{2}}{\gtrless}}0$ and the corresponding
decision boundary $p\left(  \widehat{\Lambda}\left(  \mathbf{x}\right)
|\omega_{1}\right)  -p\left(  \widehat{\Lambda}\left(  \mathbf{x}\right)
|\omega_{2}\right)  =0$.

In particular, the forces associated with the counter risk $\overline
{\mathfrak{R}}_{\mathfrak{\min}}\left(  Z_{1}|p\left(  \widehat{\Lambda
}\left(  \mathbf{x}\right)  |\omega_{1}\right)  \right)  $ in the $Z_{1}$
decision region and the risk $\mathfrak{R}_{\mathfrak{\min}}\left(
Z_{2}|p\left(  \widehat{\Lambda}\left(  \mathbf{x}\right)  |\omega_{1}\right)
\right)  $ in the $Z_{2}$ decision region are forces associated with positions
and potential locations of pattern vectors $\mathbf{x}$ that are generated
according to $p\left(  \mathbf{x}|\omega_{1}\right)  $, and the forces
associated with the risk $\mathfrak{R}_{\mathfrak{\min}}\left(  Z_{1}|p\left(
\widehat{\Lambda}\left(  \mathbf{x}\right)  |\omega_{2}\right)  \right)  $ in
the $Z_{1}$ decision region and the counter risk $\overline{\mathfrak{R}%
}_{\mathfrak{\min}}\left(  Z_{2}|p\left(  \widehat{\Lambda}\left(
\mathbf{x}\right)  |\omega_{2}\right)  \right)  $ in th $Z_{2}$ decision
region are forces associated with positions and potential locations of pattern
vectors $\mathbf{x}$ that are generated according to $p\left(  \mathbf{x}%
|\omega_{2}\right)  $.

Linear eigenlocus transforms define the opposing forces of a classification
system in terms of forces associated with counter risks $\overline
{\mathfrak{R}}_{\mathfrak{\min}}\left(  Z_{1}|\psi_{1i\ast}\mathbf{x}%
_{1_{i_{\ast}}}\right)  $ and $\overline{\mathfrak{R}}_{\mathfrak{\min}%
}\left(  Z_{2}|\psi_{2i\ast}\mathbf{x}_{2_{i_{\ast}}}\right)  $ and risks
$\mathfrak{R}_{\mathfrak{\min}}\left(  Z_{1}|\psi_{2i\ast}\mathbf{x}%
_{2_{i_{\ast}}}\right)  $ and $\mathfrak{R}_{\mathfrak{\min}}\left(
Z_{2}|\psi_{1i\ast}\mathbf{x}_{1_{i_{\ast}}}\right)  $ related to scaled
extreme points $\psi_{1i\ast}\mathbf{x}_{1_{i_{\ast}}}$ and $\psi_{2i\ast
}\mathbf{x}_{2_{i_{\ast}}}$: which are forces associated with positions and
potential locations of extreme points $\mathbf{x}_{1_{i_{\ast}}}$ and
$\mathbf{x}_{2_{i_{\ast}}}$ in the $Z_{1}$ and $Z_{2}$ decision regions of a
decision space $Z$.

In particular, the forces associated with counter risks $\overline
{\mathfrak{R}}_{\mathfrak{\min}}\left(  Z_{1}|\psi_{1i\ast}\mathbf{x}%
_{1_{i_{\ast}}}\right)  $ and risks $\mathfrak{R}_{\mathfrak{\min}}\left(
Z_{2}|\psi_{1i\ast}\mathbf{x}_{1_{i_{\ast}}}\right)  $ for class $\omega_{1}$
are determined by magnitudes and directions of scaled extreme vectors
$\psi_{1i\ast}\mathbf{x}_{1_{i_{\ast}}}$ on $\boldsymbol{\tau}_{1}$, and the
forces associated with counter risks $\overline{\mathfrak{R}}_{\mathfrak{\min
}}\left(  Z_{2}|\psi_{2i\ast}\mathbf{x}_{2_{i_{\ast}}}\right)  $ and risks
$\mathfrak{R}_{\mathfrak{\min}}\left(  Z_{1}|\psi_{2i\ast}\mathbf{x}%
_{2_{i_{\ast}}}\right)  $ for class $\omega_{2}$ are determined by magnitudes
and directions of scaled extreme vectors $\psi_{2i\ast}\mathbf{x}_{2_{i_{\ast
}}}$ on $\boldsymbol{\tau}_{2}$.

I\ will show that a Wolfe dual linear eigenlocus $\mathbf{\psi}$ is a
displacement vector that accounts for the magnitudes and the directions of all
of the scaled extreme vectors on $\boldsymbol{\tau}_{1}-\boldsymbol{\tau}_{2}%
$. Linear eigenlocus transforms determine the opposing forces of a
classification system by means of symmetrically balanced, pointwise covariance statistics.

Symmetrically balanced, pointwise covariance statistics determine forces
associated with the counter risk $\overline{\mathfrak{R}}_{\mathfrak{\min}%
}\left(  Z_{1}|p\left(  \widehat{\Lambda}_{\boldsymbol{\tau}}\left(
\mathbf{x}\right)  |\omega_{1}\right)  \right)  $ for class $\omega_{1}$ and
the risk $\mathfrak{R}_{\mathfrak{\min}}\left(  Z_{1}|p\left(
\widehat{\Lambda}_{\boldsymbol{\tau}}\left(  \mathbf{x}\right)  |\omega
_{2}\right)  \right)  $ for class $\omega_{2}$ in the $Z_{1}$ decision region
that are balanced with forces associated with the counter risk $\overline
{\mathfrak{R}}_{\mathfrak{\min}}\left(  Z_{2}|p\left(  \widehat{\Lambda
}_{\boldsymbol{\tau}}\left(  \mathbf{x}\right)  |\omega_{2}\right)  \right)  $
for class $\omega_{2}$ and the risk $\mathfrak{R}_{\mathfrak{\min}}\left(
Z_{2}|p\left(  \widehat{\Lambda}_{\boldsymbol{\tau}}\left(  \mathbf{x}\right)
|\omega_{1}\right)  \right)  $ for class $\omega_{1}$ in the $Z_{2}$ decision
region:%
\begin{align*}
f\left(  \widetilde{\Lambda}_{\boldsymbol{\tau}}\left(  \mathbf{x}\right)
\right)   &  :\overline{\mathfrak{R}}_{\mathfrak{\min}}\left(  Z_{1}|p\left(
\widehat{\Lambda}_{\boldsymbol{\tau}}\left(  \mathbf{x}\right)  |\omega
_{1}\right)  \right)  -\mathfrak{R}_{\mathfrak{\min}}\left(  Z_{1}|p\left(
\widehat{\Lambda}_{\boldsymbol{\tau}}\left(  \mathbf{x}\right)  |\omega
_{2}\right)  \right) \\
&  \rightleftharpoons\overline{\mathfrak{R}}_{\mathfrak{\min}}\left(
Z_{2}|p\left(  \widehat{\Lambda}_{\boldsymbol{\tau}}\left(  \mathbf{x}\right)
|\omega_{2}\right)  \right)  -\mathfrak{R}_{\mathfrak{\min}}\left(
Z_{2}|p\left(  \widehat{\Lambda}_{\boldsymbol{\tau}}\left(  \mathbf{x}\right)
|\omega_{1}\right)  \right)  \text{.}%
\end{align*}

I\ will now define symmetrically balanced, pointwise covariance statistics.
The geometric nature of the statistics is outlined first.

\subsection{Symmetrically Balanced Covariance Statistics I}

Take two labeled sets of extreme vectors, where each extreme vector is
correlated with a scale factor that determines scaled, signed magnitudes,
i.e., scaled components of the scaled extreme vector, along the axes of the
extreme vectors in each pattern class, such that the integrated scale factors
from each pattern class balance each other.

Generally speaking, for any given set of extreme vectors, all of the scaled,
signed magnitudes along the axis of any given extreme vector from a given
pattern class, which are determined by vector projections of scaled extreme
vectors from the \emph{other} pattern class, \emph{are distributed in opposite
directions}.

Thereby, for two labeled sets of extreme vectors, where each extreme vector is
correlated with a scale factor and the integrated scale factors from each
pattern class balance each other, it follows that scaled, signed magnitudes
along the axis of any given extreme vector, which are determined by vector
projections of scaled extreme vectors from the \emph{other} pattern class,
\emph{are distributed on the opposite side of the origin}.

Accordingly, scaled, signed magnitudes along the axes of all of the extreme
vectors are distributed in a symmetrically balanced manner, where each scale
factor specifies a symmetrically balanced distribution for an extreme point
which ensures that the \emph{components of} an extreme vector are
\emph{distributed over} the axes of a given \emph{collection} of extreme
vectors in a symmetrically balanced \emph{and} well-proportioned manner.

I will show that symmetrically balanced covariance statistics are the basis of
linear eigenlocus transforms. For any given set of extreme points, I will
demonstrate that linear eigenlocus transforms find a set of scale factors in
Wolfe dual eigenspace, which are determined by the symmetrically balanced
covariance statistics in Eqs
(\ref{Eigen-balanced Pointwise Covariance Estimate Class One}) and
(\ref{Eigen-balanced Pointwise Covariance Estimate Class Two}), such that
congruent decision regions $Z_{1}\cong Z_{1}$ are determined by symmetrically
balanced forces associated with counter risks and risks:%
\begin{align*}
\mathfrak{R}_{\mathfrak{\min}}\left(  Z:Z_{1}\cong Z\right)   &
:\overline{\mathfrak{R}}_{\mathfrak{\min}}\left(  Z_{1}|\mathbf{\tau}%
_{1}\right)  -\mathfrak{R}_{\mathfrak{\min}}\left(  Z_{1}|\mathbf{\tau}%
_{2}\right) \\
&  \rightleftharpoons\overline{\mathfrak{R}}_{\mathfrak{\min}}\left(
Z_{2}|\mathbf{\tau}_{2}\right)  -\mathfrak{R}_{\mathfrak{\min}}\left(
Z_{2}|\mathbf{\tau}_{1}\right)  \text{,}%
\end{align*}
where forces associated with the counter risk $\overline{\mathfrak{R}%
}_{\mathfrak{\min}}\left(  Z_{1}|\mathbf{\tau}_{1}\right)  $ for class
$\omega_{1}$ and the risk $\mathfrak{R}_{\mathfrak{\min}}\left(
Z_{1}|\mathbf{\tau}_{2}\right)  $ for class $\omega_{2}$ in the $Z_{1}$
decision region are symmetrically balanced with forces associated with the
counter risk $\overline{\mathfrak{R}}_{\mathfrak{\min}}\left(  Z_{2}|p\left(
\widehat{\Lambda}\left(  \mathbf{x}\right)  |\omega_{2}\right)  \right)  $ for
class $\omega_{2}$ and the risk $\mathfrak{R}_{\mathfrak{\min}}\left(
Z_{2}|p\left(  \widehat{\Lambda}\left(  \mathbf{x}\right)  |\omega_{1}\right)
\right)  $ for class $\omega_{1}$ in the $Z_{2}$ decision region.

\begin{figure}[ptb]%
\centering
\fbox{\includegraphics[
height=2.5875in,
width=3.4411in
]%
{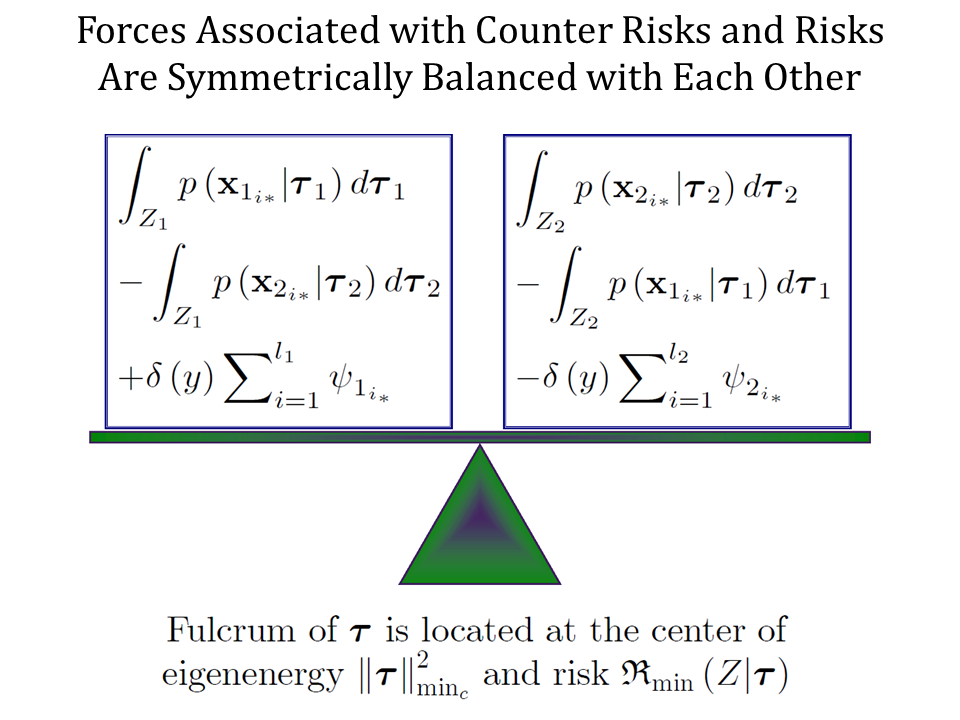}%
}\caption{Symmetrically balanced covariance statistics
$\protect\widehat{\operatorname{cov}}_{up_{\updownarrow}}\left(
\mathbf{x}_{1_{i_{\ast}}}\right)  $ and $\protect\widehat{\operatorname{cov}%
}_{up_{\updownarrow}}\left(  \mathbf{x}_{2_{i\ast}}\right)  $ for extreme
points $\mathbf{x}_{1_{i_{\ast}}}$ and $\mathbf{x}_{2_{i\ast}}$ are the basis
of linear eigenlocus transforms. Note: $\delta\left(  y\right)  \triangleq
\sum\nolimits_{i=1}^{l}y_{i}\left(  1-\xi_{i}\right)  $.}%
\label{Balancing Feat in Wolfe Dual Eigenspace}%
\end{figure}

Figure $\ref{Balancing Feat in Wolfe Dual Eigenspace}$ illustrates that
symmetrically balanced covariance statistics determine linear discriminant
functions that satisfy a fundamental integral equation of binary
classification for a linear classification system in statistical equilibrium,
where the expected risk $\mathfrak{R}_{\mathfrak{\min}}\left(  Z\mathbf{|}%
\boldsymbol{\tau}\right)  $ and the total allowed eigenenergy $\left\Vert
\boldsymbol{\tau}\right\Vert _{\min_{c}}^{2}$ of the linear classification
system are minimized.%

Using Eqs (\ref{Equilibrium Constraint on Dual Eigen-components}) and
(\ref{Pointwise Covariance Statistic}), along with the notation and
assumptions outlined above, it follows that summation over the $l$ components
of $\boldsymbol{\psi}$ in Eq.
(\ref{Dual Normal Eigenlocus Component Projections}) provides symmetrically
balanced, covariance statistics for the $\mathbf{x}_{1_{i_{\ast}}}$ extreme
vectors, where each extreme point $\mathbf{x}_{1_{i_{\ast}}}$ exhibits a
symmetrically balanced, first and second-order statistical moment
$\widehat{\operatorname{cov}}_{up_{\updownarrow}}\left(  \mathbf{x}%
_{1_{i_{\ast}}}\right)  $%
\begin{align}
\widehat{\operatorname{cov}}_{up_{\updownarrow}}\left(  \mathbf{x}%
_{1_{i_{\ast}}}\right)   &  =\left\Vert \mathbf{x}_{1_{i_{\ast}}}\right\Vert
\sum\nolimits_{j=1}^{l_{1}}\psi_{1_{j\ast}}\left\Vert \mathbf{x}_{1_{j\ast}%
}\right\Vert \cos\theta_{\mathbf{x}_{1_{i\ast}}\mathbf{x}_{1_{j\ast}}%
}\label{Eigen-balanced Pointwise Covariance Estimate Class One}\\
&  -\left\Vert \mathbf{x}_{1_{i_{\ast}}}\right\Vert \sum\nolimits_{j=1}%
^{l_{2}}\psi_{2_{j\ast}}\left\Vert \mathbf{x}_{2_{j\ast}}\right\Vert
\cos\theta_{\mathbf{x}_{1_{i\ast}}\mathbf{x}_{2_{j\ast}}}\nonumber
\end{align}
relative to $l$ symmetrically balanced, scaled, signed magnitudes determined
by vector projections of scaled extreme vectors in each respective pattern class.

Likewise, summation over the $l$ components of $\boldsymbol{\psi}$ in Eq.
(\ref{Dual Normal Eigenlocus Component Projections}) provides symmetrically
balanced, covariance statistics for the $\mathbf{x}_{2_{i_{\ast}}}$ extreme
vectors, where each extreme point $\mathbf{x}_{2_{i_{\ast}}}$ exhibits a a
symmetrically balanced, first and second-order statistical moment
$\widehat{\operatorname{cov}}_{up_{\updownarrow}}\left(  \mathbf{x}_{2_{i\ast
}}\right)  $%
\begin{align}
\widehat{\operatorname{cov}}_{up_{\updownarrow}}\left(  \mathbf{x}_{2_{i\ast}%
}\right)   &  =\left\Vert \mathbf{x}_{2_{i\ast}}\right\Vert \sum
\nolimits_{j=1}^{l_{2}}\psi_{2_{j\ast}}\left\Vert \mathbf{x}_{2_{j\ast}%
}\right\Vert \cos\theta_{\mathbf{x}_{2_{i\ast}}\mathbf{x}_{2_{j\ast}}%
}\label{Eigen-balanced Pointwise Covariance Estimate Class Two}\\
&  -\left\Vert \mathbf{x}_{2_{i_{\ast}}}\right\Vert \sum\nolimits_{j=1}%
^{l_{1}}\psi_{1_{j\ast}}\left\Vert \mathbf{x}_{1_{j\ast}}\right\Vert
\cos\theta_{\mathbf{x}_{1_{i\ast}}\mathbf{x}_{2_{j\ast}}}\nonumber
\end{align}
relative to $l$ symmetrically balanced, scaled, signed magnitudes determined
by vector projections of scaled extreme vectors in each respective pattern class.

\subsection{Common Geometrical and Statistical Properties}

I\ will now use Eqs
(\ref{Eigen-balanced Pointwise Covariance Estimate Class One}) and
(\ref{Eigen-balanced Pointwise Covariance Estimate Class Two}) to identify
symmetrical, geometric and statistical properties possessed by the principal
eigenaxis components on $\boldsymbol{\tau}$ and $\boldsymbol{\psi}$.

\subsubsection{Loci of the $\psi_{1i\ast}\protect\overrightarrow{\mathbf{e}%
}_{1i\ast}$ Components}

Let $i=1:l_{1}$, where each extreme vector $\mathbf{x}_{1_{i_{\ast}}}$ is
correlated with a Wolfe principal eigenaxis component $\psi_{1i\ast
}\overrightarrow{\mathbf{e}}_{1i\ast}$. Using Eqs
(\ref{Dual Normal Eigenlocus Component Projections}) and
(\ref{Non-orthogonal Eigenaxes of Dual Normal Eigenlocus}), it follows that
the locus of the $i^{th}$ principal eigenaxis component $\psi_{1i\ast
}\overrightarrow{\mathbf{e}}_{1i\ast}$ on $\boldsymbol{\psi}_{1}$ is a
function of the expression:%
\begin{align}
\psi_{1i\ast}  &  =\lambda_{\max_{\boldsymbol{\psi}}}^{-1}\left\Vert
\mathbf{x}_{1_{i_{\ast}}}\right\Vert \sum\nolimits_{j=1}^{l_{1}}\psi
_{1_{j\ast}}\left\Vert \mathbf{x}_{1_{j\ast}}\right\Vert \cos\theta
_{\mathbf{x}_{1_{i\ast}}\mathbf{x}_{1_{j\ast}}}%
\label{Dual Eigen-coordinate Locations Component One}\\
&  -\lambda_{\max_{\boldsymbol{\psi}}}^{-1}\left\Vert \mathbf{x}_{1_{i_{\ast}%
}}\right\Vert \sum\nolimits_{j=1}^{l_{2}}\psi_{2_{j\ast}}\left\Vert
\mathbf{x}_{2_{j\ast}}\right\Vert \cos\theta_{\mathbf{x}_{1_{i\ast}}%
\mathbf{x}_{2_{j\ast}}}\text{,}\nonumber
\end{align}
where $\psi_{1i\ast}$ provides a scale factor for the non-orthogonal unit
vector $\overrightarrow{\mathbf{e}}_{1i\ast}$. Geometric and statistical
explanations for the eigenlocus statistics%
\begin{equation}
\psi_{1_{j\ast}}\left\Vert \mathbf{x}_{1_{j\ast}}\right\Vert \cos
\theta_{\mathbf{x}_{1_{i\ast}}\mathbf{x}_{1_{j\ast}}}\text{ and }%
\psi_{2_{j\ast}}\left\Vert \mathbf{x}_{2_{j\ast}}\right\Vert \cos
\theta_{\mathbf{x}_{1_{i\ast}}\mathbf{x}_{2_{j\ast}}}
\label{Projection Statistics psi1}%
\end{equation}
in Eq. (\ref{Dual Eigen-coordinate Locations Component One}) are considered next.

\subsubsection{Geometric Nature of Eigenlocus Statistics}

The first geometric interpretation of the eigenlocus statistics in Eq.
(\ref{Projection Statistics psi1}) defines $\psi_{1_{j\ast}}$ and
$\psi_{2_{j\ast}}$ to be scale factors for the signed magnitudes of the vector
projections%
\[
\left\Vert \mathbf{x}_{1_{j\ast}}\right\Vert \cos\theta_{\mathbf{x}_{1_{i\ast
}}\mathbf{x}_{1_{j\ast}}}\text{ and }\left\Vert \mathbf{x}_{2_{j\ast}%
}\right\Vert \cos\theta_{\mathbf{x}_{1_{i\ast}}\mathbf{x}_{2_{j\ast}}}%
\]
of the scaled extreme vectors $\psi_{1_{j\ast}}\mathbf{x}_{1_{j\ast}}$ and
$\psi_{2_{j\ast}}\mathbf{x}_{2_{j\ast}}$ along the axis of the extreme vector
$\mathbf{x}_{1_{i\ast}}$, where $\cos\theta_{\mathbf{x}_{1_{i\ast}}%
\mathbf{x}_{1_{j\ast}}}$ and $\cos\theta_{\mathbf{x}_{1_{i\ast}}%
\mathbf{x}_{2_{j\ast}}}$ specify the respective angles between the axes of the
scaled extreme vectors $\psi_{1_{j\ast}}\mathbf{x}_{1_{j\ast}}$ and
$\psi_{2_{j\ast}}\mathbf{x}_{2_{j\ast}}$ and the axis of the extreme vector
$\mathbf{x}_{1_{i\ast}}$. Note that the signed magnitude $\psi_{2_{j\ast}%
}\left\Vert \mathbf{x}_{2_{j\ast}}\right\Vert \cos\theta_{\mathbf{x}%
_{1_{i\ast}}\mathbf{x}_{2_{j\ast}}}$ is distributed in the opposite direction,
so that the locus of $\psi_{2_{j\ast}}\left\Vert \mathbf{x}_{2_{j\ast}%
}\right\Vert \cos\theta_{\mathbf{x}_{1_{i\ast}}\mathbf{x}_{2_{j\ast}}}$ is on
the opposite side of the origin, along the axis of the extreme vector
$\mathbf{x}_{1_{i\ast}}$.

\begin{figure}[ptb]%
\centering
\fbox{\includegraphics[
height=2.5875in,
width=3.4411in
]%
{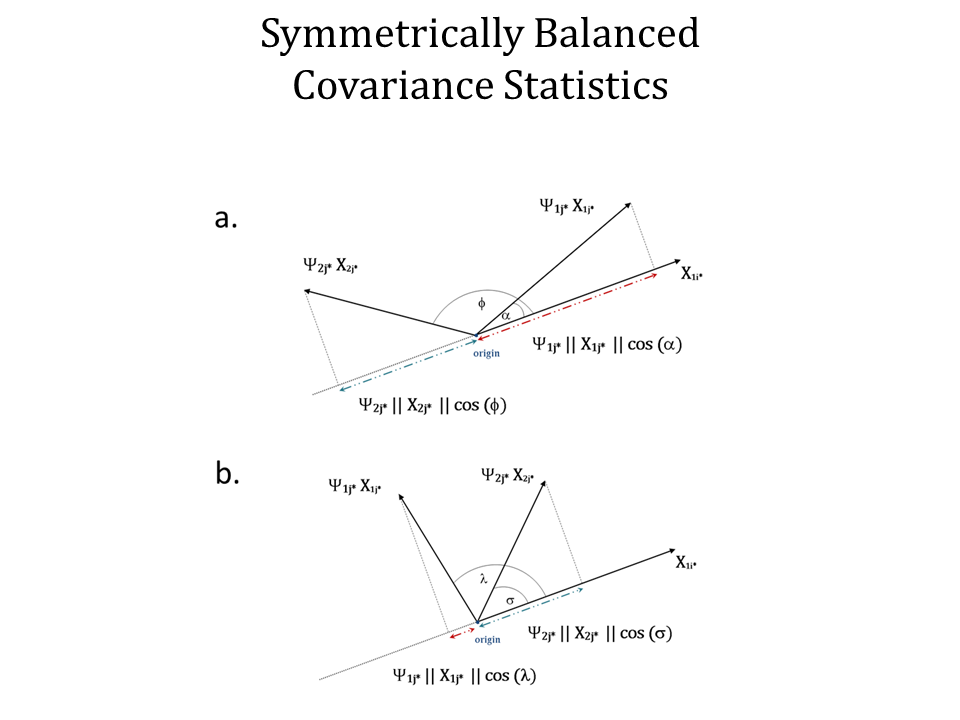}%
}\caption{Examples of positive and negative, scaled, signed magnitudes of
vector projections of scaled extreme vectors $\psi_{1_{j\ast}}\mathbf{x}%
_{1_{j\ast}}$ and $\psi_{2_{j\ast}}\mathbf{x}_{2_{j\ast}}$, along the axis of
an extreme vector $\mathbf{x}_{1_{i\ast}}$ which is correlated with a Wolfe
dual principal eigenaxis component $\psi_{1i\ast}%
\protect\overrightarrow{\mathbf{e}}_{1i\ast}$.}%
\label{Wolfe Dual Linear Eigenlocus Statistics}%
\end{figure}

Figure $\ref{Wolfe Dual Linear Eigenlocus Statistics}$ illustrates the
geometric and statistical nature of the eigenlocus statistics in Eq.
(\ref{Projection Statistics psi1}), where any given scaled, signed magnitude
$\psi_{1_{j\ast}}\left\Vert \mathbf{x}_{1_{j\ast}}\right\Vert \cos
\theta_{\mathbf{x}_{1_{i\ast}}\mathbf{x}_{1_{j\ast}}}$ or $\psi_{2_{j\ast}%
}\left\Vert \mathbf{x}_{2_{j\ast}}\right\Vert \cos\theta_{\mathbf{x}%
_{1_{i\ast}}\mathbf{x}_{2_{j\ast}}}$ may be positive or negative (see Figs
$\ref{Wolfe Dual Linear Eigenlocus Statistics}$a and
$\ref{Wolfe Dual Linear Eigenlocus Statistics}$b).%

\subsubsection{An Alternative Geometric Interpretation}

An alternative geometric explanation for the eigenlocus statistics in Eq.
(\ref{Projection Statistics psi1}) accounts for the representation of
$\boldsymbol{\tau}_{1}$ and $\boldsymbol{\tau}_{2}$ within the Wolfe dual
eigenspace. Consider the vector relationships%
\[
\psi_{1_{j\ast}}\left\Vert \mathbf{x}_{1_{j\ast}}\right\Vert =\left\Vert
\psi_{1_{j\ast}}\mathbf{x}_{1_{j\ast}}\right\Vert =\left\Vert \boldsymbol{\tau
}_{1}(j)\right\Vert
\]
and%
\[
\psi_{2_{j\ast}}\left\Vert \mathbf{x}_{2_{j\ast}}\right\Vert =\left\Vert
\psi_{2_{j\ast}}\mathbf{x}_{2_{j\ast}}\right\Vert =\left\Vert \boldsymbol{\tau
}_{2}(j)\right\Vert \text{,}%
\]
where $\boldsymbol{\tau}_{1}(j)$ and $\boldsymbol{\tau}_{2}(j)$ are the $j$th
constrained, primal principal eigenaxis components on $\boldsymbol{\tau}_{1}$
and $\boldsymbol{\tau}_{2}$.

It follows that the scaled $\psi_{1_{j\ast}}$, signed magnitude $\left\Vert
\mathbf{x}_{1_{j\ast}}\right\Vert \cos\theta_{\mathbf{x}_{1_{i\ast}}%
\mathbf{x}_{1_{j\ast}}}$ of the vector projection of the scaled extreme vector
$\psi_{1_{j\ast}}\mathbf{x}_{1_{j\ast}}$ along the axis of the extreme vector
$\mathbf{x}_{1_{i\ast}}$%
\[
\psi_{1_{j\ast}}\left\Vert \mathbf{x}_{1_{j\ast}}\right\Vert \cos
\theta_{\mathbf{x}_{1_{i\ast}}\mathbf{x}_{1_{j\ast}}}%
\]
determines the scaled $\cos\theta_{\mathbf{x}_{1_{i\ast}}\mathbf{x}_{1_{j\ast
}}}$ length of the $j$th constrained, primal principal eigenaxis component
$\boldsymbol{\tau}_{1}(j)$ on $\boldsymbol{\tau}_{1}$%
\[
\cos\theta_{\mathbf{x}_{1_{i\ast}}\mathbf{x}_{1_{j\ast}}}\left\Vert
\boldsymbol{\tau}_{1}(j)\right\Vert \text{,}%
\]
where $\psi_{1_{j\ast}}$ is the length of the $\psi_{1j\ast}%
\overrightarrow{\mathbf{e}}_{1j\ast}$ Wolfe dual principal eigenaxis component
and $\cos\theta_{\mathbf{x}_{1_{i\ast}}\mathbf{x}_{1_{j\ast}}}$ specifies the
angle between the extreme vectors $\mathbf{x}_{1_{i\ast}}$ and $\mathbf{x}%
_{1_{j\ast}}$.

Likewise, the scaled $\psi_{2_{j\ast}}$, signed magnitude $\left\Vert
\mathbf{x}_{2_{j\ast}}\right\Vert \cos\theta_{\mathbf{x}_{1_{i\ast}}%
\mathbf{x}_{2_{j\ast}}}$ of the vector projection of the scaled extreme vector
$\psi_{2_{j\ast}}\mathbf{x}_{2_{j\ast}}$ along the axis of the extreme vector
$\mathbf{x}_{1_{i\ast}}$%
\[
\psi_{2_{j\ast}}\left\Vert \mathbf{x}_{2_{j\ast}}\right\Vert \cos
\theta_{\mathbf{x}_{1_{i\ast}}\mathbf{x}_{2_{j\ast}}}%
\]
determines the scaled $\cos\theta_{\mathbf{x}_{1_{i\ast}}\mathbf{x}_{2_{j\ast
}}}$ length of the $j$th constrained, primal principal eigenaxis component
$\boldsymbol{\tau}_{2}(j)$ on $\boldsymbol{\tau}_{2}$%
\[
\cos\theta_{\mathbf{x}_{1_{i\ast}}\mathbf{x}_{2_{j\ast}}}\left\Vert
\boldsymbol{\tau}_{2}(j)\right\Vert \text{,}%
\]
where $\psi_{2_{j\ast}}$ is the length of the $\psi_{2j\ast}%
\overrightarrow{\mathbf{e}}_{2j\ast}$ Wolfe dual principal eigenaxis component
and $\cos\theta_{\mathbf{x}_{1_{i\ast}}\mathbf{x}_{2_{j\ast}}}$ specifies the
angle between the extreme vectors $\mathbf{x}_{1_{i\ast}}$ and $\mathbf{x}%
_{2_{j\ast}}$.

Therefore, each Wolfe dual principal eigenaxis component $\psi_{1i\ast
}\overrightarrow{\mathbf{e}}_{1i\ast}$ is a function of the constrained,
primal principal eigenaxis components on $\boldsymbol{\tau}_{1}$ and
$\boldsymbol{\tau}_{2}$:%
\begin{align}
\psi_{1i\ast}  &  =\lambda_{\max_{\boldsymbol{\psi}}}^{-1}\left\Vert
\mathbf{x}_{1_{i_{\ast}}}\right\Vert \sum\nolimits_{j=1}^{l_{1}}\cos
\theta_{\mathbf{x}_{1_{i\ast}}\mathbf{x}_{1_{j\ast}}}\left\Vert
\boldsymbol{\tau}_{1}(j)\right\Vert \label{Constrained Primal Eigenlocus psi1}%
\\
&  -\lambda_{\max_{\boldsymbol{\psi}}}^{-1}\left\Vert \mathbf{x}_{1_{i_{\ast}%
}}\right\Vert \sum\nolimits_{j=1}^{l_{2}}\cos\theta_{\mathbf{x}_{1_{i\ast}%
}\mathbf{x}_{2_{j\ast}}}\left\Vert \boldsymbol{\tau}_{2}(j)\right\Vert
\text{,}\nonumber
\end{align}
where the angle between each principal eigenaxis component $\boldsymbol{\tau
}_{1}(j)$ and $\boldsymbol{\tau}_{2}(j)$ and the extreme vector $\mathbf{x}%
_{1_{i\ast}}$ is fixed.

I will now define the significant geometric and statistical properties which
are jointly exhibited by Wolfe dual $\psi_{1i\ast}\overrightarrow{\mathbf{e}%
}_{1i\ast}$\textbf{\ }and constrained, primal $\psi_{1i\ast}\mathbf{x}%
_{1_{i\ast}}$ principal eigenaxis components that regulate the symmetric
partitioning of a feature space $Z$.

\subsection{Significant Geometric and Statistical Properties}

Using the definition of Eq.
(\ref{Eigen-balanced Pointwise Covariance Estimate Class One}), Eq.
(\ref{Dual Eigen-coordinate Locations Component One}) indicates that the locus
of the principal eigenaxis component $\psi_{1i\ast}\overrightarrow{\mathbf{e}%
}_{1i\ast}$ is determined by a symmetrically balanced, signed magnitude along
the axis of an extreme vector $\mathbf{x}_{1_{i\ast}}$, relative to
symmetrically balanced, scaled, signed magnitudes of extreme vector
projections in each respective pattern class.

\subsubsection{Symmetrically Balanced Signed Magnitudes}

Let $\operatorname{comp}_{\overrightarrow{\mathbf{x}_{1i\ast}}}\left(
\overrightarrow{\widetilde{\psi}_{1i\ast}\left\Vert \widetilde{\mathbf{x}%
}_{\ast}\right\Vert _{_{1i_{\ast}}}}\right)  $ denote the symmetrically
balanced, signed magnitude%
\begin{align}
\operatorname{comp}_{\overrightarrow{\mathbf{x}_{1i\ast}}}\left(
\overrightarrow{\widetilde{\psi}_{1i\ast}\left\Vert \widetilde{\mathbf{x}%
}_{\ast}\right\Vert _{_{1i_{\ast}}}}\right)   &  =\sum\nolimits_{j=1}^{l_{1}%
}\psi_{1_{j\ast}}\label{Unidirectional Scaling Term One1}\\
&  \times\left[  \left\Vert \mathbf{x}_{1_{j\ast}}\right\Vert \cos
\theta_{\mathbf{x}_{1_{i\ast}}\mathbf{x}_{1_{j\ast}}}\right] \nonumber\\
&  -\sum\nolimits_{j=1}^{l_{2}}\psi_{2_{j\ast}}\nonumber\\
&  \times\left[  \left\Vert \mathbf{x}_{2_{j\ast}}\right\Vert \cos
\theta_{\mathbf{x}_{1_{i\ast}}\mathbf{x}_{2_{j\ast}}}\right] \nonumber
\end{align}
along the axis of the extreme vector $\mathbf{x}_{1_{i\ast}}$ that is
correlated with the Wolfe dual principal eigenaxis component $\psi_{1i\ast
}\overrightarrow{\mathbf{e}}_{1i\ast}$.

\subsubsection{Symmetrically Balanced Distributions}

Using the definitions of Eqs (\ref{Pointwise Covariance Statistic}) and
(\ref{Eigen-balanced Pointwise Covariance Estimate Class One}), it follows
that Eq. (\ref{Unidirectional Scaling Term One1}) specifies a symmetrically
balanced distribution of scaled, first degree coordinates of extreme vectors
along the axis of $\mathbf{x}_{1_{i\ast}}$, where each scale factor
$\psi_{1_{j\ast}}$ or $\psi_{2_{j\ast}}$ specifies how an extreme vector
$\mathbf{x}_{1_{j\ast}}$ or $\mathbf{x}_{2_{j\ast}}$ is distributed along the
axis of $\mathbf{x}_{1_{i\ast}}$, and each scale factor $\psi_{1_{j\ast}}$ or
$\psi_{2_{j\ast}}$ specifies a symmetrically balanced distribution of scaled,
first degree coordinates of extreme vectors $\left\{  \psi_{_{j\ast}%
}\mathbf{x}_{j\ast}\right\}  _{j=1}^{l}$ along the axis of an extreme vector
$\mathbf{x}_{1_{j\ast}}$ or $\mathbf{x}_{2_{j\ast}}$.

Therefore, each scaled, signed magnitude%
\[
\psi_{1_{j\ast}}\left\Vert \mathbf{x}_{1_{j\ast}}\right\Vert \cos
\theta_{\mathbf{x}_{1_{i\ast}}\mathbf{x}_{1_{j\ast}}}\text{ \ or \ }%
\psi_{2_{j\ast}}\left\Vert \mathbf{x}_{2_{j\ast}}\right\Vert \cos
\theta_{\mathbf{x}_{1_{i\ast}}\mathbf{x}_{2_{j\ast}}}%
\]
provides an estimate for how the components of the extreme vector
$\mathbf{x}_{1_{i_{\ast}}}$ are symmetrically distributed over the axis of a
scaled extreme vector $\psi_{1_{j\ast}}\mathbf{x}_{1_{j\ast}}$ or
$\psi_{2_{j\ast}}\mathbf{x}_{2_{j\ast}}$, where each scale factor
$\psi_{1_{j\ast}}$ or $\psi_{2_{j\ast}}$ specifies a symmetrically balanced
distribution of scaled, first degree coordinates of extreme vectors $\left\{
\psi_{_{j\ast}}\mathbf{x}_{j\ast}\right\}  _{j=1}^{l}$ along the axis of an
extreme vector $\mathbf{x}_{1_{j\ast}}$ or $\mathbf{x}_{2_{j\ast}}$.

Again, using Eqs (\ref{Pointwise Covariance Statistic} and
(\ref{Eigen-balanced Pointwise Covariance Estimate Class One}), it follows
that Eq. (\ref{Unidirectional Scaling Term One1}) determines a symmetrically
balanced, first and second-order statistical moment about the locus of
$\mathbf{x}_{1_{i\ast}}$, where each scale factor $\psi_{1_{j\ast}}$ or
$\psi_{1_{j\ast}}$ specifies how the components of an extreme vector
$\mathbf{x}_{1_{j\ast}}$ or $\mathbf{x}_{2_{j\ast}}$ are distributed along the
axis of $\mathbf{x}_{1_{i\ast}}$, and each scale factor $\psi_{1_{j\ast}}$ or
$\psi_{1_{j\ast}}$ specifies a symmetrically balanced distribution for an
extreme vector $\mathbf{x}_{1_{j\ast}}$ or $\mathbf{x}_{2_{j\ast}}$.

\subsubsection{Distributions of Eigenaxis Components}

Using Eqs (\ref{Equilibrium Constraint on Dual Eigen-components}),
(\ref{Dual Eigen-coordinate Locations Component One}), and
(\ref{Unidirectional Scaling Term One1}), it follows that symmetrically
balanced, joint distributions of the principal eigenaxis components on
$\boldsymbol{\psi}$ and $\boldsymbol{\tau}$ are distributed over the axis of
the extreme vector $\mathbf{x}_{1_{i\ast}}$.

Again, using Eq. (\ref{Dual Eigen-coordinate Locations Component One}), it
follows that identical, symmetrically balanced, joint distributions of the
principal eigenaxis components on $\boldsymbol{\psi}$ and $\boldsymbol{\tau}$
are distributed over the axis of the Wolfe dual principal eigenaxis component
$\psi_{1i\ast}\overrightarrow{\mathbf{e}}_{1i\ast}$.

Thereby, symmetrically balanced, joint distributions of the principal
eigenaxis components on $\boldsymbol{\psi}$ and $\boldsymbol{\tau}$ are
identically and symmetrically distributed over the respective axes of each
Wolfe dual principal eigenaxis component $\psi_{1i\ast}%
\overrightarrow{\mathbf{e}}_{1i\ast}$ and each correlated extreme vector
$\mathbf{x}_{1_{i\ast}}$.

Alternatively, using Eq. (\ref{Constrained Primal Eigenlocus psi1}), the
symmetrically balanced, signed magnitude in Eq.
(\ref{Unidirectional Scaling Term One1}) depends upon the difference between
integrated cosine-scaled lengths of the constrained, primal principal
eigenaxis components on $\boldsymbol{\tau}_{1}$ and $\boldsymbol{\tau}_{2}$:%
\begin{align}
\operatorname{comp}_{\overrightarrow{\mathbf{x}_{1i\ast}}}\left(
\overrightarrow{\widetilde{\psi}_{1i\ast}\left\Vert \widetilde{\mathbf{x}%
}_{\ast}\right\Vert _{_{1_{i\ast}}}}\right)   &  =\sum\nolimits_{j=1}^{l_{1}%
}\cos\theta_{\mathbf{x}_{1_{i\ast}}\mathbf{x}_{1_{j\ast}}}\left\Vert
\boldsymbol{\tau}_{1}(j)\right\Vert \label{Unidirectional Scaling Term One2}\\
&  -\sum\nolimits_{j=1}^{l_{2}}\cos\theta_{\mathbf{x}_{1_{i\ast}}%
\mathbf{x}_{2_{j\ast}}}\left\Vert \boldsymbol{\tau}_{2}(j)\right\Vert
\nonumber
\end{align}
which also shows that symmetrically balanced, joint distributions of the
principal eigenaxis components on $\boldsymbol{\psi}$ and $\boldsymbol{\tau}$
are identically and symmetrically distributed along the respective axes of
each Wolfe dual principal eigenaxis component $\psi_{1i\ast}%
\overrightarrow{\mathbf{e}}_{1i\ast}$ and each correlated extreme vector
$\mathbf{x}_{1_{i\ast}}$.

Using Eqs (\ref{Dual Eigen-coordinate Locations Component One}) and
(\ref{Unidirectional Scaling Term One1}), it follows that the length
$\psi_{1i\ast}$ of each Wolfe dual principal eigenaxis component $\psi
_{1i\ast}\overrightarrow{\mathbf{e}}_{1i\ast}$ is determined by the weighted
length of a correlated extreme vector $\mathbf{x}_{1_{i\ast}}$%
\begin{equation}
\psi_{1i\ast}=\left[  \lambda_{\max_{\boldsymbol{\psi}}}^{-1}\times
\operatorname{comp}_{\overrightarrow{\mathbf{x}_{1i\ast}}}\left(
\overrightarrow{\widetilde{\psi}_{1i\ast}\left\Vert \widetilde{\mathbf{x}%
}_{\ast}\right\Vert _{_{1i_{\ast}}}}\right)  \right]  \left\Vert
\mathbf{x}_{1_{i\ast}}\right\Vert \text{,}
\label{Magnitude Dual Normal Eigenaxis Component Class One}%
\end{equation}
where the weighting factor is determined by an eigenvalue $\lambda
_{\max_{\boldsymbol{\psi}}}^{-1}$ scaling of a symmetrically balanced, signed
magnitude $\operatorname{comp}_{\overrightarrow{\mathbf{x}_{1i\ast}}}\left(
\overrightarrow{\widetilde{\psi}_{1i\ast}\left\Vert \widetilde{\mathbf{x}%
}_{\ast}\right\Vert _{_{1i_{\ast}}}}\right)  $ along the axis of
$\mathbf{x}_{1_{i\ast}}$.

\subsubsection{Symmetrically Balanced Lengths}

Given that $\psi_{1i\ast}>0$, $\lambda_{\max_{\boldsymbol{\psi}}}^{-1}>0$ and
$\left\Vert \mathbf{x}_{1_{i\ast}}\right\Vert >0$, it follows that the
symmetrically balanced, signed magnitude along the axis of each extreme vector
$\mathbf{x}_{1_{i\ast}}$ is a positive number%
\[
\operatorname{comp}_{\overrightarrow{\mathbf{x}_{1i\ast}}}\left(
\overrightarrow{\widetilde{\psi}_{1i\ast}\left\Vert \widetilde{\mathbf{x}%
}_{\ast}\right\Vert _{_{1i_{\ast}}}}\right)  >0
\]
which indicates that the weighting factor in Eq.
(\ref{Magnitude Dual Normal Eigenaxis Component Class One}) determines a
well-proportioned length%
\[
\lambda_{\max_{\boldsymbol{\psi}}}^{-1}\operatorname{comp}%
_{\overrightarrow{\mathbf{x}_{1i\ast}}}\left(  \overrightarrow{\widetilde{\psi
}_{1i\ast}\left\Vert \widetilde{\mathbf{x}}_{\ast}\right\Vert _{_{1i_{\ast}}}%
}\right)  \left\Vert \mathbf{x}_{1_{i\ast}}\right\Vert
\]
for an extreme vector $\mathbf{x}_{1_{i\ast}}$. Thereby, the length
$\psi_{1i\ast}$ of each Wolfe dual principal eigenaxis component $\psi
_{1i\ast}\overrightarrow{\mathbf{e}}_{1i\ast}$ is determined by a
well-proportioned length of a correlated extreme vector $\mathbf{x}_{1_{i\ast
}}$.

Returning to Eqs (\ref{Eigen-balanced Pointwise Covariance Estimate Class One}%
) and (\ref{Dual Eigen-coordinate Locations Component One}), it follows that
the length $\psi_{1i\ast}$ of each Wolfe dual principal eigenaxis component
$\psi_{1i\ast}\overrightarrow{\mathbf{e}}_{1i\ast}$ on $\boldsymbol{\psi}$%
\[
\psi_{1i\ast}=\lambda_{\max_{\boldsymbol{\psi}}}^{-1}\operatorname{comp}%
_{\overrightarrow{\mathbf{x}_{1i\ast}}}\left(  \overrightarrow{\widetilde{\psi
}_{1i\ast}\left\Vert \widetilde{\mathbf{x}}_{\ast}\right\Vert _{_{1i_{\ast}}}%
}\right)  \left\Vert \mathbf{x}_{1_{i\ast}}\right\Vert
\]
is shaped by a symmetrically balanced, first and second-order statistical
moment about the locus of a correlated extreme vector $\mathbf{x}_{1_{i\ast}}
$.

Now, take any given correlated pair $\left\{  \psi_{1i\ast}%
\overrightarrow{\mathbf{e}}_{1i\ast},\psi_{1_{i\ast}}\mathbf{x}_{1_{i\ast}%
}\right\}  $ of Wolfe dual and constrained, primal principal eigenaxis
components. I will now show that the direction of $\psi_{1i\ast}%
\overrightarrow{\mathbf{e}}_{1i\ast}$ is identical to the direction of
$\psi_{1_{i\ast}}\mathbf{x}_{1_{i\ast}}$.

\subsubsection{Directional Symmetries}

The vector direction of each Wolfe dual principal eigenaxis component
$\psi_{1i\ast}\overrightarrow{\mathbf{e}}_{1i\ast}$ is implicitly specified by
Eq. (\ref{Dual Eigen-coordinate Locations Component One}), where it has been
assumed that $\psi_{1i\ast}$ provides a scale factor for a non-orthogonal unit
vector $\overrightarrow{\mathbf{e}}_{1i\ast}$. Using the definitions of Eqs
(\ref{Pointwise Covariance Statistic}) and
(\ref{Eigen-balanced Pointwise Covariance Estimate Class One}), it follows
that the symmetrically balanced, pointwise covariance statistic in Eq.
(\ref{Dual Eigen-coordinate Locations Component One}) specifies the direction
of a correlated extreme vector $\mathbf{x}_{1_{i_{\ast}}}$ and a
well-proportioned magnitude along the axis of the extreme vector
$\mathbf{x}_{1_{i_{\ast}}}$.

Returning to Eqs (\ref{Unidirectional Scaling Term One1}),
(\ref{Unidirectional Scaling Term One2}), and
(\ref{Magnitude Dual Normal Eigenaxis Component Class One}), take any given
Wolfe dual principal eigenaxis component $\psi_{1i\ast}%
\overrightarrow{\mathbf{e}}_{1i\ast}$ that is correlated with an extreme
vector $\mathbf{x}_{1_{i_{\ast}}}$. Given that the magnitude $\psi_{1i\ast}$
of each Wolfe dual principal eigenaxis component $\psi_{1i\ast}%
\overrightarrow{\mathbf{e}}_{1i\ast}$ is determined by a well-proportioned
magnitude of a correlated extreme vector $\mathbf{x}_{1_{i_{\ast}}}$%
\[
\psi_{1i\ast}=\lambda_{\max_{\boldsymbol{\psi}}}^{-1}\operatorname{comp}%
_{\overrightarrow{\mathbf{x}_{1i\ast}}}\left(  \overrightarrow{\widetilde{\psi
}_{1i\ast}\left\Vert \widetilde{\mathbf{x}}_{\ast}\right\Vert _{_{1i_{\ast}}}%
}\right)  \left\Vert \mathbf{x}_{1_{i\ast}}\right\Vert \text{,}%
\]
it follows that each non-orthogonal unit vector $\overrightarrow{\mathbf{e}%
}_{1i\ast}$ has the same direction as an extreme vector $\mathbf{x}%
_{1_{i_{\ast}}}$%
\[
\overrightarrow{\mathbf{e}}_{1i\ast}\equiv\frac{\mathbf{x}_{1_{i_{\ast}}}%
}{\left\Vert \mathbf{x}_{1_{i_{\ast}}}\right\Vert }\text{.}%
\]

Thereby, the direction of each Wolfe dual principal eigenaxis component
$\psi_{1i\ast}\overrightarrow{\mathbf{e}}_{1i\ast}$ on $\boldsymbol{\psi}$ is
identical to the direction of a correlated, constrained primal principal
eigenaxis component $\psi_{1_{i\ast}}\mathbf{x}_{1_{i\ast}}$ on
$\boldsymbol{\tau}_{1}$ which is determined by the direction of a scaled
$\psi_{1i\ast}$ extreme vector $\mathbf{x}_{1_{i\ast}}$. Each Wolfe dual and
correlated, constrained primal principal eigenaxis component are said to
exhibit \emph{directional symmetry}.

Therefore, it is concluded that correlated principal eigenaxis components on
$\boldsymbol{\psi}_{1}$ and $\boldsymbol{\tau}_{1}$ exhibit directional symmetry.

\subsubsection{Directions of Large Covariance}

It is concluded that the uniform directions of the Wolfe dual and the
correlated, constrained primal principal eigenaxis components specify
directions of large covariance which contribute to a symmetric partitioning of
a minimal geometric region of constant width that spans a region of large
covariance between two data distributions. It is also concluded that each of
the correlated principal eigenaxis components on $\boldsymbol{\psi}_{1}$ and
$\boldsymbol{\tau}_{1}$ possess well-proportioned magnitudes for which the
constrained, linear discriminant function $D\left(  \mathbf{x}\right)
=\mathbf{x}^{T}\boldsymbol{\tau}+\tau_{0}$ delineates centrally located,
bipartite, congruent regions of large covariance between any two data distributions.

\subsubsection{Loci of the $\psi_{2i\ast}\protect\overrightarrow{\mathbf{e}%
}_{2i\ast}$ Components}

Let $i=1:l_{2}$, where each extreme vector $\mathbf{x}_{2_{i_{\ast}}}$ is
correlated with a Wolfe dual principal eigenaxis component $\psi_{2i\ast
}\overrightarrow{\mathbf{e}}_{2i\ast}$. Using Eqs
(\ref{Dual Normal Eigenlocus Component Projections}) and
(\ref{Non-orthogonal Eigenaxes of Dual Normal Eigenlocus}), it follows that
the locus of the $i^{th}$ Wolfe dual principal eigenaxis component
$\psi_{2i\ast}\overrightarrow{\mathbf{e}}_{2i\ast}$ on $\boldsymbol{\psi}_{2}$
is a function of the expression:%
\begin{align}
\psi_{2i\ast}  &  =\lambda_{\max_{\boldsymbol{\psi}}}^{-1}\left\Vert
\mathbf{x}_{2_{i\ast}}\right\Vert \sum\nolimits_{j=1}^{l_{2}}\psi_{2_{j\ast}%
}\left\Vert \mathbf{x}_{2_{j\ast}}\right\Vert \cos\theta_{\mathbf{x}%
_{2_{i\ast}}\mathbf{x}_{2_{j\ast}}}%
\label{Dual Eigen-coordinate Locations Component Two}\\
&  -\lambda_{\max_{\boldsymbol{\psi}}}^{-1}\left\Vert \mathbf{x}_{2_{i\ast}%
}\right\Vert \sum\nolimits_{j=1}^{l_{1}}\psi_{1_{j\ast}}\left\Vert
\mathbf{x}_{1_{j\ast}}\right\Vert \cos\theta_{\mathbf{x}_{2_{i\ast}}%
\mathbf{x}_{1_{j\ast}}}\text{,}\nonumber
\end{align}
where $\psi_{2i\ast}$ provides a scale factor for the non-orthogonal unit
vector $\overrightarrow{\mathbf{e}}_{2i\ast}$.

Results obtained from the previous analysis are readily generalized to the
Wolfe dual $\psi_{2i\ast}\overrightarrow{\mathbf{e}}_{2i\ast}$ and the
constrained, primal $\psi_{2_{i\ast}}\mathbf{x}_{2_{i\ast}}$ principal
eigenaxis components, so the analysis will not be replicated. However, the
counterpart to Eq. (\ref{Unidirectional Scaling Term One1}) is necessary for a
future argument. Let $i=l_{1}+1:l_{1}+l_{2}$, where each extreme vector
$\mathbf{x}_{2_{i_{\ast}}}$ is correlated with a Wolfe principal eigenaxis
component $\psi_{2i\ast}\overrightarrow{\mathbf{e}}_{2i\ast}$. Accordingly,
let $\operatorname{comp}_{\overrightarrow{\mathbf{x}_{2i\ast}}}\left(
\overrightarrow{\widetilde{\psi}_{2i\ast}\left\Vert \widetilde{\mathbf{x}%
}_{\ast}\right\Vert _{_{2i_{\ast}}}}\right)  $ denote the symmetrically
balanced, signed magnitude%
\begin{align}
\operatorname{comp}_{\overrightarrow{\mathbf{x}_{2i\ast}}}\left(
\overrightarrow{\widetilde{\psi}_{2i\ast}\left\Vert \widetilde{\mathbf{x}%
}_{\ast}\right\Vert _{_{2i_{\ast}}}}\right)   &  =\sum\nolimits_{j=1}^{l_{2}%
}\psi_{2_{j\ast}}\label{Unidirectional Scaling Term Two1}\\
&  \times\left[  \left\Vert \mathbf{x}_{2_{j\ast}}\right\Vert \cos
\theta_{\mathbf{x}_{2_{i\ast}}\mathbf{x}_{2_{j\ast}}}\right] \nonumber\\
&  -\sum\nolimits_{j=1}^{l_{1}}\psi_{1_{j\ast}}\nonumber\\
&  \times\left[  \left\Vert \mathbf{x}_{1_{j\ast}}\right\Vert \cos
\theta_{\mathbf{x}_{2_{i\ast}}\mathbf{x}_{1_{j\ast}}}\right] \nonumber
\end{align}
along the axis of the extreme vector $\mathbf{x}_{2_{i\ast}}$ that is
correlated with the Wolfe dual principal eigenaxis component $\psi_{2i\ast
}\overrightarrow{\mathbf{e}}_{2i\ast}$.

\subsubsection{Similar Properties Exhibited by $\boldsymbol{\psi}$ and
$\boldsymbol{\tau}$}

I\ will now identify similar, geometric and statistical properties which are
jointly exhibited by the Wolfe dual principal eigenaxis components on
$\boldsymbol{\psi}$ and the correlated, constrained, primal principal
eigenaxis components on $\boldsymbol{\tau}$. The properties are summarized below.

\paragraph{Directional Symmetry}

\begin{enumerate}
\item The direction of each Wolfe dual principal eigenaxis component
$\psi_{1i\ast}\overrightarrow{\mathbf{e}}_{1i\ast}$ on $\boldsymbol{\psi
}\mathbf{_{1}}$ is identical to the direction of a correlated, constrained,
primal principal eigenaxis component $\psi_{1_{i\ast}}\mathbf{x}_{1_{i\ast}}$
on$\mathbf{\ }\boldsymbol{\tau}\mathbf{_{1}}$.

\item The direction of each Wolfe dual principal eigenaxis component
$\psi_{2i\ast}\overrightarrow{\mathbf{e}}_{2i\ast}$ on $\boldsymbol{\psi
}\mathbf{_{2}}$ is identical to the direction of a correlated, constrained,
primal principal eigenaxis component $\psi_{2_{i\ast}}\mathbf{x}_{2_{i\ast}}$
on$\mathbf{\ }\boldsymbol{\tau}\mathbf{_{2}}$.
\end{enumerate}

\paragraph{Symmetrically Balanced Lengths}

\begin{enumerate}
\item The lengths of each Wolfe dual principal eigenaxis component
$\psi_{1i\ast}\overrightarrow{\mathbf{e}}_{1i\ast}$ on $\boldsymbol{\psi
}\mathbf{_{1}}$ and each correlated, constrained primal principal eigenaxis
component $\psi_{1_{i\ast}}\mathbf{x}_{1_{i\ast}}$ on$\mathbf{\ }%
\boldsymbol{\tau}\mathbf{_{1}}$ are shaped by identical, symmetrically
balanced, joint distributions of the principal eigenaxis components on
$\boldsymbol{\psi}$ and $\boldsymbol{\tau}$.

\item The lengths of each Wolfe dual principal eigenaxis component
$\psi_{2i\ast}\overrightarrow{\mathbf{e}}_{2i\ast}$ on $\boldsymbol{\psi
}\mathbf{_{2}}$ and each correlated, constrained primal principal eigenaxis
component $\psi_{2_{i\ast}}\mathbf{x}_{2_{i\ast}}$ on$\mathbf{\ }%
\boldsymbol{\tau}\mathbf{_{2}}$ are shaped by identical, symmetrically
balanced, joint distributions of the principal eigenaxis components on
$\boldsymbol{\psi}$ and $\boldsymbol{\tau}$.
\end{enumerate}

\paragraph{Symmetrically Balanced Pointwise Covariance Statistics}

\begin{enumerate}
\item The magnitude $\psi_{1i\ast}$ of each Wolfe dual principal eigenaxis
component $\psi_{1i\ast}\overrightarrow{\mathbf{e}}_{1i\ast}$ on
$\boldsymbol{\psi}\mathbf{_{1}}$%
\[
\psi_{1i\ast}=\lambda_{\max_{\boldsymbol{\psi}}}^{-1}\operatorname{comp}%
_{\overrightarrow{\mathbf{x}_{1i\ast}}}\left(  \overrightarrow{\widetilde{\psi
}_{1i\ast}\left\Vert \widetilde{\mathbf{x}}_{\ast}\right\Vert _{_{1i_{\ast}}}%
}\right)  \left\Vert \mathbf{x}_{1_{i\ast}}\right\Vert
\]
is determined by a symmetrically balanced, pointwise covariance estimate%
\begin{align*}
\widehat{\operatorname{cov}}_{up_{\updownarrow}}\left(  \mathbf{x}%
_{1_{i_{\ast}}}\right)   &  =\lambda_{\max_{\boldsymbol{\psi}}}^{-1}\left\Vert
\mathbf{x}_{1_{i_{\ast}}}\right\Vert \\
&  \times\sum\nolimits_{j=1}^{l_{1}}\psi_{1_{j\ast}}\left\Vert \mathbf{x}%
_{1_{j\ast}}\right\Vert \cos\theta_{\mathbf{x}_{1_{i\ast}}\mathbf{x}%
_{1_{j\ast}}}\\
&  -\lambda_{\max_{\boldsymbol{\psi}}}^{-1}\left\Vert \mathbf{x}_{1_{i_{\ast}%
}}\right\Vert \\
&  \times\sum\nolimits_{j=1}^{l_{2}}\psi_{2_{j\ast}}\left\Vert \mathbf{x}%
_{2_{j\ast}}\right\Vert \cos\theta_{\mathbf{x}_{1_{i\ast}}\mathbf{x}%
_{2_{j\ast}}}%
\end{align*}
for a correlated extreme vector\textbf{\ }$\mathbf{x}_{1_{i\ast}}$, such that
the locus of each constrained, primal principal eigenaxis component\textbf{\ }%
$\psi_{1_{i\ast}}\mathbf{x}_{1_{i\ast}}$\textbf{\ }on $\boldsymbol{\tau
}\mathbf{_{1}}$\textbf{\ }provides a maximum\textit{\ }covariance estimate in
a principal location $\mathbf{x}_{1_{i\ast}}$, in the form of a symmetrically
balanced, first and second-order statistical moment about the locus of an
extreme data point $\mathbf{x}_{1_{i\ast}}$.

\item The magnitude $\psi_{2i\ast}$ of each Wolfe dual principal eigenaxis
component $\psi_{2i\ast}\overrightarrow{\mathbf{e}}_{2i\ast}$ on
$\boldsymbol{\psi}\mathbf{_{2}}$%
\[
\psi_{2i\ast}=\lambda_{\max_{\boldsymbol{\psi}}}^{-1}\operatorname{comp}%
_{\overrightarrow{\mathbf{x}_{2i\ast}}}\left(  \overrightarrow{\widetilde{\psi
}_{2i\ast}\left\Vert \widetilde{\mathbf{x}}_{\ast}\right\Vert _{_{2i_{\ast}}}%
}\right)  \left\Vert \mathbf{x}_{2_{i\ast}}\right\Vert
\]
is determined by a symmetrically balanced, pointwise covariance estimate%
\begin{align*}
\widehat{\operatorname{cov}}_{up_{\updownarrow}}\left(  \mathbf{x}%
_{2_{i_{\ast}}}\right)   &  =\lambda_{\max_{\boldsymbol{\psi}}}^{-1}\left\Vert
\mathbf{x}_{2_{i\ast}}\right\Vert \\
&  \times\sum\nolimits_{j=1}^{l_{2}}\psi_{2_{j\ast}}\left\Vert \mathbf{x}%
_{2_{j\ast}}\right\Vert \cos\theta_{\mathbf{x}_{2_{i\ast}}\mathbf{x}%
_{2_{j\ast}}}\\
&  -\lambda_{\max_{\boldsymbol{\psi}}}^{-1}\left\Vert \mathbf{x}_{2_{i\ast}%
}\right\Vert \\
&  \times\sum\nolimits_{j=1}^{l_{1}}\psi_{1_{j\ast}}\left\Vert \mathbf{x}%
_{1_{j\ast}}\right\Vert \cos\theta_{\mathbf{x}_{2_{i\ast}}\mathbf{x}%
_{1_{j\ast}}}%
\end{align*}
for a correlated extreme vector\textbf{\ }$\mathbf{x}_{2_{i\ast}}$, such that
the locus of each constrained, primal principal eigenaxis component\textbf{\ }%
$\psi_{2_{i\ast}}\mathbf{x}_{2_{i\ast}}$\textbf{\ }on $\boldsymbol{\tau
}\mathbf{_{2}}$\textbf{\ }provides a maximum\textit{\ }covariance estimate in
a principal location $\mathbf{x}_{2_{i\ast}}$, in the form of a symmetrically
balanced, first and second-order statistical moment about the locus of an
extreme point $\mathbf{x}_{2_{i\ast}}$.
\end{enumerate}

\paragraph{Symmetrically Balanced Statistical Moments}

\begin{enumerate}
\item Each Wolfe dual principal eigenaxis component\textbf{\ }$\psi_{1i\ast
}\overrightarrow{\mathbf{e}}_{1i\ast}$ on $\boldsymbol{\psi}\mathbf{_{1}}%
$\textbf{\ }specifies a symmetrically balanced, first and second-order
statistical moment about the locus of a correlated extreme point
$\mathbf{x}_{1_{i\ast}}$ relative to the loci of all of the scaled extreme
points which determines the locus of a constrained, primal principal eigenaxis
component $\psi_{1_{i\ast}}\mathbf{x}_{1_{i\ast}}$ on $\boldsymbol{\tau}_{1}$.

\item Each Wolfe dual principal eigenaxis component\textbf{\ }$\psi_{2i\ast
}\overrightarrow{\mathbf{e}}_{1i\ast}$ on $\boldsymbol{\psi}\mathbf{_{2}}%
$\textbf{\ }specifies a symmetrically balanced, first and second-order
statistical moment about the locus of a correlated extreme point
$\mathbf{x}_{2_{i\ast}}$ relative to the loci of all of the scaled extreme
points which determines the locus of a constrained, primal principal eigenaxis
component $\psi_{2_{i\ast}}\mathbf{x}_{2_{i\ast}}$ on $\boldsymbol{\tau}_{2}$.
\end{enumerate}

\paragraph{Symmetrically Balanced Distributions of Extreme Points}

\begin{enumerate}
\item Any given maximum covariance estimate $\widehat{\operatorname{cov}%
}_{up_{\updownarrow}}\left(  \mathbf{x}_{1_{i_{\ast}}}\right)  $ describes how
the components of $l$ scaled extreme vectors $\left\{  \psi_{1_{j\ast}%
}\mathbf{x}_{1_{j_{\ast}}}\right\}  _{j=1}^{l_{1}}$ and $\left\{
\psi_{2_{j\ast}}\mathbf{x}_{2_{j_{\ast}}}\right\}  _{j=1}^{l_{2}}$ are
distributed along the axis of an extreme vector $\mathbf{x}_{1_{i\ast}}$,
where each scale factor $\psi_{1_{j\ast}}$ or $\psi_{2_{j\ast}}$ specifies a
symmetrically balanced distribution of $l$ scaled extreme vectors along the
axis of an extreme vector $\mathbf{x}_{1_{j\ast}}$ or $\mathbf{x}_{2_{j\ast}}%
$, such that a pointwise covariance estimate $\widehat{\operatorname{cov}%
}_{up_{\updownarrow}}\left(  \mathbf{x}_{1_{i_{\ast}}}\right)  $ provides an
estimate for how the components of the extreme vector $\mathbf{x}_{1_{i_{\ast
}}}$ are symmetrically distributed over the axes of the $l$ scaled extreme
vectors. Thus, $\widehat{\operatorname{cov}}_{up_{\updownarrow}}\left(
\mathbf{x}_{1_{i_{\ast}}}\right)  $ describes a distribution of first degree
coordinates for $\mathbf{x}_{1_{i_{\ast}}}$.

\item Any given maximum\textit{\ }covariance estimate
$\widehat{\operatorname{cov}}_{up_{\updownarrow}}\left(  \mathbf{x}%
_{2_{i_{\ast}}}\right)  $ describes how the components of $l$ scaled extreme
vectors $\left\{  \psi_{1_{j\ast}}\mathbf{x}_{1_{j_{\ast}}}\right\}
_{j=1}^{l_{1}}$ and $\left\{  \psi_{2_{j\ast}}\mathbf{x}_{2_{j_{\ast}}%
}\right\}  _{j=1}^{l_{2}}$ are distributed along the axis of an extreme vector
$\mathbf{x}_{2_{i\ast}}$, where each scale factor $\psi_{1_{j\ast}}$ or
$\psi_{2_{j\ast}}$ specifies a symmetrically balanced distribution of $l$
scaled extreme vectors along the axis of an extreme vector $\mathbf{x}%
_{1_{j\ast}}$ or $\mathbf{x}_{2_{j\ast}}$, such that a pointwise covariance
estimate $\widehat{\operatorname{cov}}_{up_{\updownarrow}}\left(
\mathbf{x}_{2_{i_{\ast}}}\right)  $ provides an estimate for how the
components of the extreme vector $\mathbf{x}_{2_{i_{\ast}}}$ are symmetrically
distributed over the axes of the $l$ scaled extreme vectors. Thus,
$\widehat{\operatorname{cov}}_{up_{\updownarrow}}\left(  \mathbf{x}%
_{2_{i_{\ast}}}\right)  $ describes a distribution of first degree coordinates
for $\mathbf{x}_{2_{i_{\ast}}}$.
\end{enumerate}

I\ will now define the equivalence between the total allowed eigenenergies
exhibited by $\boldsymbol{\psi}$ and $\boldsymbol{\tau}$.

\subsection{Equivalence Between Eigenenergies of $\boldsymbol{\psi}$ and
$\boldsymbol{\tau}$}

The inner product between the integrated Wolf dual principal eigenaxis
components on $\boldsymbol{\psi}$%
\begin{align*}
\left\Vert \boldsymbol{\psi}\right\Vert _{\min_{c}}^{2}  &  =\left(
\sum\nolimits_{i=1}^{l_{1}}\psi_{1i\ast}\frac{\mathbf{x}_{1_{i\ast}}^{T}%
}{\left\Vert \mathbf{x}_{1_{i\ast}}\right\Vert }+\sum\nolimits_{i=1}^{l_{2}%
}\psi_{2i\ast}\frac{\mathbf{x}_{2_{i\ast}}^{T}}{\left\Vert \mathbf{x}%
_{2_{i\ast}}\right\Vert }\right) \\
&  \times\left(  \sum\nolimits_{i=1}^{l_{1}}\psi_{1i\ast}\frac{\mathbf{x}%
_{1_{i\ast}}}{\left\Vert \mathbf{x}_{1_{i\ast}}\right\Vert }+\sum
\nolimits_{i=1}^{l_{2}}\psi_{2i\ast}\frac{\mathbf{x}_{2_{i\ast}}}{\left\Vert
\mathbf{x}_{2_{i\ast}}\right\Vert }\right)
\end{align*}
determines the total allowed eigenenergy $\left\Vert \boldsymbol{\psi
}\right\Vert _{\min_{c}}^{2}$ of $\boldsymbol{\psi}$: which is symmetrically
equivalent with the critical minimum eigenenergy $\left\Vert \boldsymbol{\tau
}\right\Vert _{\min_{c}}^{2}$ of $\boldsymbol{\tau}$ within its Wolfe dual
eigenspace%
\begin{align*}
\left\Vert \boldsymbol{\tau}\right\Vert _{\min_{c}}^{2}  &  =\left(
\sum\nolimits_{i=1}^{l_{1}}\psi_{1_{i\ast}}\mathbf{x}_{1_{i\ast}}^{T}%
-\sum\nolimits_{i=1}^{l_{2}}\psi_{2_{i\ast}}\mathbf{x}_{2_{i\ast}}^{T}\right)
\\
&  \times\left(  \sum\nolimits_{i=1}^{l_{1}}\psi_{1_{i\ast}}\mathbf{x}%
_{1_{i\ast}}-\sum\nolimits_{i=1}^{l_{2}}\psi_{2_{i\ast}}\mathbf{x}_{2_{i\ast}%
}\right)  \text{.}%
\end{align*}

I will now argue that the equivalence $\left\Vert \boldsymbol{\psi}\right\Vert
_{\min_{c}}^{2}\simeq\left\Vert \boldsymbol{\tau}\right\Vert _{\min_{c}}^{2}$
between the total allowed eigenenergies exhibited by $\boldsymbol{\psi}$ and
$\boldsymbol{\tau}$ involves symmetrically balanced, joint eigenenergy
distributions with respect to the principal eigenaxis components on
$\boldsymbol{\psi}$ and $\boldsymbol{\tau}$.

\paragraph{Symmetrical Equivalence of Eigenenergy Distributions}

Using Eqs (\ref{Equilibrium Constraint on Dual Eigen-components}),
(\ref{Dual Eigen-coordinate Locations Component One}),
(\ref{Unidirectional Scaling Term One1}), and
(\ref{Unidirectional Scaling Term Two1}), it follows that identical,
symmetrically balanced, joint distributions of principal eigenaxis components
on $\boldsymbol{\psi}$ and $\boldsymbol{\tau}$ are symmetrically distributed
over the respective axes of each Wolfe dual principal eigenaxis component on
$\boldsymbol{\psi}$ and each correlated and unconstrained primal principal
eigenaxis component (extreme vector) on $\boldsymbol{\tau}$. Therefore,
constrained primal and Wolfe dual principal eigenaxis components that are
correlated with each other are formed by equivalent, symmetrically balanced,
joint distributions of principal eigenaxis components on $\boldsymbol{\psi}$
and $\boldsymbol{\tau}$.

Thereby, symmetrically balanced, joint distributions of principal eigenaxis
components on $\boldsymbol{\psi}$ and $\boldsymbol{\tau}$ are symmetrically
distributed over the axes of all of the Wolf dual principal eigenaxis
components $\left\{  \psi_{1i\ast}\overrightarrow{\mathbf{e}}_{1i\ast
}\right\}  _{i=1}^{l_{1}}$ and $\left\{  \psi_{2i\ast}%
\overrightarrow{\mathbf{e}}_{2i\ast}\right\}  _{i=1}^{l_{2}}$ on
$\boldsymbol{\psi}_{1}$ and $\boldsymbol{\psi}_{2}$ and all of the
constrained, primal principal eigenaxis components $\left\{  \psi_{1_{i\ast}%
}\mathbf{x}_{1_{i\ast}}\right\}  _{i=1}^{l_{1}}$ and $\left\{  \psi_{2_{i\ast
}}\mathbf{x}_{2_{i\ast}}\right\}  _{i=1}^{l_{2}}$ on $\boldsymbol{\tau}_{1}$
and $\boldsymbol{\tau}_{2}$, where $\overrightarrow{\mathbf{e}}_{1i\ast}%
=\frac{\mathbf{x}_{1_{i\ast}}}{\left\Vert \mathbf{x}_{1_{i\ast}}\right\Vert }$
and $\overrightarrow{\mathbf{e}}_{2i\ast}=\frac{\mathbf{x}_{2_{i\ast}}%
}{\left\Vert \mathbf{x}_{2_{i\ast}}\right\Vert }$.

Therefore, the distribution of eigenenergies with respect to the Wolfe dual
principal eigenaxis components on $\boldsymbol{\psi}$ is symmetrically
equivalent to the distribution of eigenenergies with respect to the
constrained, primal principal eigenaxis components on $\boldsymbol{\tau}$,
such that the total allowed eigenenergies $\left\Vert \boldsymbol{\psi
}\right\Vert _{\min_{c}}^{2}$ and $\left\Vert \boldsymbol{\tau}\right\Vert
_{\min_{c}}^{2}$ exhibited by $\boldsymbol{\psi}$ and $\boldsymbol{\tau}$
satisfy symmetrically balanced, joint eigenenergy distributions with respect
to the principal eigenaxis components on\emph{\ }$\boldsymbol{\psi}$ and
$\boldsymbol{\tau}$. Thus, all of the constrained, primal principal eigenaxis
components on $\boldsymbol{\tau}_{1}-\boldsymbol{\tau}_{2}$ possess
eigenenergies that satisfy symmetrically balanced, joint eigenenergy
distributions with respect to the principal eigenaxis components on
$\boldsymbol{\psi}$ and $\boldsymbol{\tau}$.

Later on, I\ will show that the critical minimum eigenenergies exhibited by
the scaled extreme vectors determine conditional probabilities of
classification error for extreme points, where any given extreme point has a
risk or counter risk that is determined by a measure of central location and a
measure of spread, both of which are described by a conditional probability density.

In the next section, I\ will show that each Wolfe dual principal eigenaxis
component specifies a standardized, conditional probability density for an
extreme point. By way of motivation, an overview of Bayesian estimates for
parameter vectors $\mathbf{\theta}$ of unknown probability densities $p\left(
\mathbf{x}\right)  $ is presented next.

\subsection{Parameter Vectors $\mathbf{\theta}$ of Probability Densities}

Take any given class of feature vectors $\mathbf{x}$ that is described by an
unknown probability density function $p\left(  \mathbf{x}\right)  $. Let the
unknown density function $p\left(  \mathbf{x}\right)  $ have a known
parametric form in terms of a parameter vector $\mathbf{\theta}$, where the
unknowns are the components of $\mathbf{\theta}$. Any information about
$\mathbf{\theta}$ prior to observing a set of data samples is assumed to be
contained in a known prior density $p\left(  \mathbf{\theta}\right)  $, where
observation of a set of data samples produces a posterior density $p\left(
\mathbf{\theta}|D\right)  $ which is sharply peaked about the true value of
$\mathbf{\theta}$ if $\widehat{\mathbf{\theta}}\simeq\mathbf{\theta}$.

Given these assumptions, $p\left(  \mathbf{x}|D\right)  $ can be computed by
integrating the joint density $p\left(  \mathbf{x},\mathbf{\theta|}D\right)  $
over $\mathbf{\theta}$%
\[
p\left(  \mathbf{x}|D\right)  =\int p\left(  \mathbf{x},\mathbf{\theta
|}D\right)  d\mathbf{\theta}\text{.}%
\]
Write $p\left(  \mathbf{x},\mathbf{\theta|}D\right)  $ as $p\left(
\mathbf{x}|\mathbf{\theta}\right)  p\left(  \mathbf{\theta}|D\right)  $. If
$p\left(  \mathbf{\theta}|D\right)  $ peaks sharply about some value
$\widehat{\mathbf{\theta}}$, then the integral%
\[
p\left(  \mathbf{x}|D\right)  =\int p\left(  \mathbf{x}|\mathbf{\theta
}\right)  p\left(  \mathbf{\theta}|D\right)  d\mathbf{\theta}\simeq p\left(
\mathbf{x}|\widehat{\mathbf{\theta}}\right)
\]
produces an estimate $\widehat{\mathbf{\theta}}$ for the desired parameter
vector $\mathbf{\theta}$
\citep{Duda2001}%
. Accordingly, unknown probability density functions $p\left(  \mathbf{x}%
\right)  $ can be determined by Bayesian estimation of a parameter vector
$\mathbf{\theta}$ with a known prior density $p\left(  \mathbf{\theta}\right)
$.

\subsubsection{Learning an Unknown Vector $\mathbf{\theta}$ of Two Conditional
Densities}

Given the previous assumptions for Bayesian estimation of parameter vectors
$\mathbf{\theta}$, it is reasonable to assume that information about an
unknown probability density function $p\left(  \mathbf{x}\right)  $ is
distributed over the components of a parameter vector $\widehat{\mathbf{\theta
}}$. It has been demonstrated that symmetrically balanced, joint distributions
of principal eigenaxis components on $\boldsymbol{\psi}$ and $\boldsymbol{\tau
}$ are symmetrically distributed over the axes of all of the Wolf dual
principal eigenaxis components on $\boldsymbol{\psi}$ and all of the
constrained, primal principal eigenaxis components on $\boldsymbol{\tau}$. In
the next analysis, I\ will show that information for two unknown conditional
density functions $p\left(  \mathbf{x}_{1_{i\ast}}|\widehat{\mathbf{\theta}%
}_{1}\right)  $ and $p\left(  \mathbf{x}_{2_{i\ast}}|\widehat{\mathbf{\theta}%
}_{2}\right)  $ is distributed over the scaled extreme points on
$\boldsymbol{\tau}_{1}-\boldsymbol{\tau}_{2}$, where $\boldsymbol{\tau}$ is an
unknown parameter vector $\widehat{\mathbf{\theta}}$ that contains information
about the unknown conditional densities $\widehat{\mathbf{\theta}%
}=\widehat{\mathbf{\theta}}_{1}-\widehat{\mathbf{\theta}}_{2}$.

I will now define pointwise conditional densities which are determined by the
components of a constrained, primal linear eigenlocus $\boldsymbol{\tau}= $
$\boldsymbol{\tau}_{1}-\boldsymbol{\tau}_{2}$, where each conditional density
$p\left(  \mathbf{x}_{1_{i_{\ast}}}|\operatorname{comp}%
_{\overrightarrow{\mathbf{x}_{1i\ast}}}\left(
\overrightarrow{\boldsymbol{\tau}}\right)  \right)  $ or $p\left(
\mathbf{x}_{2_{i_{\ast}}}|\operatorname{comp}_{\overrightarrow{\mathbf{x}%
_{2i\ast}}}\left(  \overrightarrow{\boldsymbol{\tau}}\right)  \right)  $ for
an $\mathbf{x}_{1_{i\ast}}$ or $\mathbf{x}_{2_{i\ast}}$ extreme point is given
by components $\operatorname{comp}_{\overrightarrow{\mathbf{x}_{1i\ast}}%
}\left(  \overrightarrow{\boldsymbol{\tau}}\right)  $ or $\operatorname{comp}%
_{\overrightarrow{\mathbf{x}_{2i\ast}}}\left(
\overrightarrow{\boldsymbol{\tau}}\right)  $ of $\boldsymbol{\tau}$ along the
corresponding extreme vector $\mathbf{x}_{1_{i\ast}}$ or $\mathbf{x}%
_{2_{i\ast}}$.

\subsection{Pointwise Conditional Densities}

Consider again the equations for the loci of the $\psi_{1i\ast}%
\overrightarrow{\mathbf{e}}_{1i\ast}$ and $\psi_{2i\ast}%
\overrightarrow{\mathbf{e}}_{2i\ast}$ Wolfe dual principal eigenaxis
components in Eqs (\ref{Dual Eigen-coordinate Locations Component One}) and
(\ref{Dual Eigen-coordinate Locations Component Two}). It has been
demonstrated that any given Wolfe dual principal eigenaxis component
$\psi_{1i\ast}\overrightarrow{\mathbf{e}}_{1i\ast}$ correlated with an
$\mathbf{x}_{1_{i_{\ast}}}$ extreme point and any given Wolfe dual principal
eigenaxis component $\psi_{2i\ast}\overrightarrow{\mathbf{e}}_{2i\ast}$
correlated with an $\mathbf{x}_{2_{i\ast}}$ extreme point provides an estimate
for how the components of $l$ scaled extreme vectors $\left\{  \psi_{_{j\ast}%
}\mathbf{x}_{j\ast}\right\}  _{j=1}^{l}$ are symmetrically distributed along
the axis of a correlated extreme vector $\mathbf{x}_{1_{i_{\ast}}}$ or
$\mathbf{x}_{2_{i\ast}}$, where components of scaled extreme vectors
$\psi_{_{j\ast}}\mathbf{x}_{j\ast}$ are symmetrically distributed according to
class labels $\pm1$, signed magnitudes $\left\Vert \mathbf{x}_{j\ast
}\right\Vert \cos\theta_{\mathbf{x}_{1_{i\ast}}\mathbf{x}_{_{j\ast}}}$ or
$\left\Vert \mathbf{x}_{j\ast}\right\Vert \cos\theta_{\mathbf{x}_{2_{i\ast}%
}\mathbf{x}_{_{j\ast}}}$, and symmetrically balanced distributions of scaled
extreme vectors $\left\{  \psi_{_{j\ast}}\mathbf{x}_{j\ast}\right\}
_{j=1}^{l}$ specified by scale factors $\psi_{_{j\ast}}$.

Thereby, symmetrically balanced distributions of first degree coordinates of
all of the extreme points are symmetrically distributed along the axes of all
of the extreme vectors, where all of the scale factors satisfy the equivalence
relation $\sum\nolimits_{j=1}^{l_{1}}\psi_{1_{j\ast}}=\sum\nolimits_{j=1}%
^{l_{2}}\psi_{2_{j\ast}}$. Accordingly, principal eigenaxis components
$\psi_{1i\ast}\overrightarrow{\mathbf{e}}_{1i\ast}$ or $\psi_{2i\ast
}\overrightarrow{\mathbf{e}}_{2i\ast}$ describe distributions of first degree
coordinates for extreme points $\mathbf{x}_{1_{i_{\ast}}}$ or $\mathbf{x}%
_{2_{i_{\ast}}}$.

Therefore, for any given extreme vector $\mathbf{x}_{1_{i\ast}}$, the relative
likelihood that the extreme point $\mathbf{x}_{1_{i\ast}}$ has a given
location is specified by the locus of the Wolfe dual principal eigenaxis
component $\psi_{1i\ast}\frac{\mathbf{x}_{1_{i\ast}}}{\left\Vert
\mathbf{x}_{1_{i\ast}}\right\Vert }$:%
\begin{align*}
\psi_{1i\ast}\frac{\mathbf{x}_{1_{i\ast}}}{\left\Vert \mathbf{x}_{1_{i\ast}%
}\right\Vert }  &  =\lambda_{\max_{\boldsymbol{\psi}}}^{-1}\left\Vert
\mathbf{x}_{1_{i_{\ast}}}\right\Vert \sum\nolimits_{j=1}^{l_{1}}\psi
_{1_{j\ast}}\left\Vert \mathbf{x}_{1_{j\ast}}\right\Vert \cos\theta
_{\mathbf{x}_{1_{i\ast}}\mathbf{x}_{1_{j\ast}}}\\
&  -\lambda_{\max_{\boldsymbol{\psi}}}^{-1}\left\Vert \mathbf{x}_{1_{i_{\ast}%
}}\right\Vert \sum\nolimits_{j=1}^{l_{2}}\psi_{2_{j\ast}}\left\Vert
\mathbf{x}_{2_{j\ast}}\right\Vert \cos\theta_{\mathbf{x}_{1_{i\ast}}%
\mathbf{x}_{2_{j\ast}}}\text{,}%
\end{align*}
where $\psi_{1i\ast}\frac{\mathbf{x}_{1_{i\ast}}}{\left\Vert \mathbf{x}%
_{1_{i\ast}}\right\Vert }$ describes a conditional expectation (a measure of
central location) and a conditional covariance (a measure of spread) for the
extreme point $\mathbf{x}_{1_{i_{\ast}}}$. Thereby, it is concluded that the
principal eigenaxis component $\psi_{1i\ast}\frac{\mathbf{x}_{1_{i\ast}}%
}{\left\Vert \mathbf{x}_{1_{i\ast}}\right\Vert }$ specifies a\emph{\ }%
conditional density $p\left(  \mathbf{x}_{1_{i_{\ast}}}|\operatorname{comp}%
_{\overrightarrow{\mathbf{x}_{1i\ast}}}\left(
\overrightarrow{\boldsymbol{\tau}}\right)  \right)  $ for the extreme point
$\mathbf{x}_{1_{i_{\ast}}}$, where the scale factor $\psi_{1i\ast}$ is a
\emph{unit }measure or estimate of density and likelihood for the extreme
point $\mathbf{x}_{1_{i\ast}}$.

Likewise, for any given extreme vector $\mathbf{x}_{2_{i\ast}}$, the relative
likelihood that the extreme point $\mathbf{x}_{2_{i\ast}}$ has a given
location is specified by the locus of the Wolfe dual principal eigenaxis
component $\psi_{2i\ast}\frac{\mathbf{x}_{2_{i\ast}}}{\left\Vert
\mathbf{x}_{2_{i\ast}}\right\Vert }$:%
\begin{align*}
\psi_{2i\ast}\frac{\mathbf{x}_{2_{i\ast}}}{\left\Vert \mathbf{x}_{2_{i\ast}%
}\right\Vert }  &  =\lambda_{\max_{\boldsymbol{\psi}}}^{-1}\left\Vert
\mathbf{x}_{2_{i\ast}}\right\Vert \sum\nolimits_{j=1}^{l_{2}}\psi_{2_{j\ast}%
}\left\Vert \mathbf{x}_{2_{j\ast}}\right\Vert \cos\theta_{\mathbf{x}%
_{2_{i\ast}}\mathbf{x}_{2_{j\ast}}}\\
&  -\lambda_{\max_{\boldsymbol{\psi}}}^{-1}\left\Vert \mathbf{x}_{2_{i\ast}%
}\right\Vert \sum\nolimits_{j=1}^{l_{1}}\psi_{1_{j\ast}}\left\Vert
\mathbf{x}_{1_{j\ast}}\right\Vert \cos\theta_{\mathbf{x}_{2_{i\ast}}%
\mathbf{x}_{1_{j\ast}}}\text{,}%
\end{align*}
where $\psi_{2i\ast}\frac{\mathbf{x}_{2_{i\ast}}}{\left\Vert \mathbf{x}%
_{2_{i\ast}}\right\Vert }$ describes a conditional expectation (a measure of
central location) and a conditional covariance (a measure of spread) for the
extreme point $\mathbf{x}_{2_{i\ast}}$. Thereby, it is concluded that the
principal eigenaxis component $\psi_{2i\ast}\frac{\mathbf{x}_{2_{i\ast}}%
}{\left\Vert \mathbf{x}_{2_{i\ast}}\right\Vert }$ specifies a conditional
density $p\left(  \mathbf{x}_{2_{i_{\ast}}}|\operatorname{comp}%
_{\overrightarrow{\mathbf{x}_{2i\ast}}}\left(
\overrightarrow{\boldsymbol{\tau}}\right)  \right)  $ for the extreme point
$\mathbf{x}_{2_{i\ast}}$, where the scale factor $\psi_{2i\ast}$ is a
\emph{unit }measure or estimate of density and likelihood for the extreme
point $\mathbf{x}_{2_{i\ast}}$.

It has been shown that a Wolfe dual linear eigenlocus $\boldsymbol{\psi}$ is
formed by a locus of scaled, normalized extreme vectors%
\begin{align*}
\boldsymbol{\psi}  &  =\sum\nolimits_{i=1}^{l_{1}}\psi_{1i\ast}\frac
{\mathbf{x}_{1_{i\ast}}}{\left\Vert \mathbf{x}_{1_{i\ast}}\right\Vert }%
+\sum\nolimits_{i=1}^{l_{2}}\psi_{2i\ast}\frac{\mathbf{x}_{2_{i\ast}}%
}{\left\Vert \mathbf{x}_{2_{i\ast}}\right\Vert }\\
&  =\boldsymbol{\psi}_{1}+\boldsymbol{\psi}_{2}\text{,}%
\end{align*}
where $\boldsymbol{\psi}_{1}=\sum\nolimits_{i=1}^{l_{1}}\psi_{1i\ast}%
\frac{\mathbf{x}_{1_{i\ast}}}{\left\Vert \mathbf{x}_{1_{i\ast}}\right\Vert }$
and $\boldsymbol{\psi}_{2}=\sum\nolimits_{i=1}^{l_{2}}\psi_{2i\ast}%
\frac{\mathbf{x}_{2_{i\ast}}}{\left\Vert \mathbf{x}_{2_{i\ast}}\right\Vert }$,
and each $\psi_{1i\ast}$ or $\psi_{2i\ast}$ scale factor provides a unit
measure or estimate of density and likelihood for an $\mathbf{x}_{1_{i\ast}}$
or $\mathbf{x}_{2_{i\ast}}$ extreme point.

Given that each Wolfe dual principal eigenaxis component $\psi_{1i\ast}%
\frac{\mathbf{x}_{1_{i\ast}}}{\left\Vert \mathbf{x}_{1_{i\ast}}\right\Vert }$
on $\boldsymbol{\psi}_{1}$ specifies a conditional density $p\left(
\mathbf{x}_{1_{i_{\ast}}}|\operatorname{comp}_{\overrightarrow{\mathbf{x}%
_{1i\ast}}}\left(  \overrightarrow{\boldsymbol{\tau}}\right)  \right)  $ for a
correlated extreme point $\mathbf{x}_{1_{i_{\ast}}}$, it follows that
conditional densities for the $\mathbf{x}_{1_{i_{\ast}}}$ extreme points are
distributed over the principal eigenaxis components of $\boldsymbol{\psi}_{1}$%
\begin{align}
\boldsymbol{\psi}_{1}  &  =\sum\nolimits_{i=1}^{l_{1}}p\left(  \mathbf{x}%
_{1_{i_{\ast}}}|\operatorname{comp}_{\overrightarrow{\mathbf{x}_{1i\ast}}%
}\left(  \overrightarrow{\boldsymbol{\tau}}\right)  \right)  \frac
{\mathbf{x}_{1_{i\ast}}}{\left\Vert \mathbf{x}_{1_{i\ast}}\right\Vert
}\label{Wolfe Dual Conditional Density Extreme Points 1}\\
&  =\sum\nolimits_{i=1}^{l_{1}}\psi_{1_{i\ast}}\frac{\mathbf{x}_{1_{i\ast}}%
}{\left\Vert \mathbf{x}_{1_{i\ast}}\right\Vert }\text{,}\nonumber
\end{align}
where $\psi_{1_{i\ast}}\frac{\mathbf{x}_{1_{i\ast}}}{\left\Vert \mathbf{x}%
_{1_{i\ast}}\right\Vert }$ specifies a conditional density for $\mathbf{x}%
_{1_{i_{\ast}}}$, such that $\boldsymbol{\psi}_{1}$ is a parameter vector for
a class-conditional probability density $p\left(  \frac{\mathbf{x}_{1_{i\ast}%
}}{\left\Vert \mathbf{x}_{1_{i\ast}}\right\Vert }|\boldsymbol{\psi}%
_{1}\right)  $ for a given set $\left\{  \mathbf{x}_{1_{i\ast}}\right\}
_{i=1}^{l_{1}}$ of $\mathbf{x}_{1_{i_{\ast}}}$ extreme points:%
\[
\boldsymbol{\psi}_{1}=p\left(  \frac{\mathbf{x}_{1_{i\ast}}}{\left\Vert
\mathbf{x}_{1_{i\ast}}\right\Vert }|\boldsymbol{\psi}_{1}\right)  \text{.}%
\]

Given that each Wolfe dual principal eigenaxis component $\psi_{2i\ast}%
\frac{\mathbf{x}_{2_{i\ast}}}{\left\Vert \mathbf{x}_{2_{i\ast}}\right\Vert }$
on $\boldsymbol{\psi}_{2}$ specifies a conditional density $p\left(
\mathbf{x}_{2_{i_{\ast}}}|\operatorname{comp}_{\overrightarrow{\mathbf{x}%
_{2i\ast}}}\left(  \overrightarrow{\boldsymbol{\tau}}\right)  \right)  $ for a
correlated extreme point $\mathbf{x}_{2_{i_{\ast}}}$, it follows that
conditional densities for the $\mathbf{x}_{2_{i_{\ast}}}$ extreme points are
distributed over the principal eigenaxis components of $\boldsymbol{\psi}_{2}$%
\begin{align}
\boldsymbol{\psi}_{2}  &  =\sum\nolimits_{i=1}^{l_{2}}p\left(  \mathbf{x}%
_{2_{i_{\ast}}}|\operatorname{comp}_{\overrightarrow{\mathbf{x}_{2i\ast}}%
}\left(  \overrightarrow{\boldsymbol{\tau}}\right)  \right)  \frac
{\mathbf{x}_{2_{i\ast}}}{\left\Vert \mathbf{x}_{2_{i\ast}}\right\Vert
}\label{Wolfe Dual Conditional Density Extreme Points 2}\\
&  =\sum\nolimits_{i=1}^{l_{2}}\psi_{2_{i\ast}}\frac{\mathbf{x}_{2_{i\ast}}%
}{\left\Vert \mathbf{x}_{2_{i\ast}}\right\Vert }\text{,}\nonumber
\end{align}
where $\psi_{2_{i\ast}}\frac{\mathbf{x}_{2_{i\ast}}}{\left\Vert \mathbf{x}%
_{2_{i\ast}}\right\Vert }$ specifies a conditional density for $\mathbf{x}%
_{2_{i\ast}}$, such that $\boldsymbol{\psi}_{2}$ is a parameter vector for a
class-conditional probability density $p\left(  \frac{\mathbf{x}_{2_{i\ast}}%
}{\left\Vert \mathbf{x}_{2_{i\ast}}\right\Vert }|\boldsymbol{\psi}_{2}\right)
$ for a given set $\left\{  \mathbf{x}_{2_{i_{\ast}}}\right\}  _{i=1}^{l_{2}}$
of $\mathbf{x}_{2_{i_{\ast}}}$ extreme points:%
\[
\boldsymbol{\psi}_{2}=p\left(  \frac{\mathbf{x}_{2_{i\ast}}}{\left\Vert
\mathbf{x}_{2_{i\ast}}\right\Vert }|\boldsymbol{\psi}_{2}\right)  \text{.}%
\]

Therefore, it is concluded that $\boldsymbol{\psi}_{1}$ is a parameter vector
for the class-conditional probability density function $p\left(
\frac{\mathbf{x}_{1_{i\ast}}}{\left\Vert \mathbf{x}_{1_{i\ast}}\right\Vert
}|\boldsymbol{\psi}_{1}\right)  $ and $\boldsymbol{\psi}_{2}$ is a parameter
vector for the class-conditional probability density function $p\left(
\frac{\mathbf{x}_{2_{i\ast}}}{\left\Vert \mathbf{x}_{2_{i\ast}}\right\Vert
}|\boldsymbol{\psi}_{2}\right)  $.

Returning to Eq. (\ref{Equilibrium Constraint on Dual Eigen-components}), it
follows that the pointwise conditional densities $\psi_{1i\ast}\frac
{\mathbf{x}_{1_{i\ast}}}{\left\Vert \mathbf{x}_{1_{i\ast}}\right\Vert }$ and
$\psi_{2i\ast}\frac{\mathbf{x}_{2_{i\ast}}}{\left\Vert \mathbf{x}_{2_{i\ast}%
}\right\Vert }$ for all of the extreme points in class $\omega_{1}$ and class
$\omega_{2}$ are symmetrically balanced with each other:%
\[
\sum\nolimits_{i=1}^{l_{1}}\psi_{1i\ast}\frac{\mathbf{x}_{1_{i\ast}}%
}{\left\Vert \mathbf{x}_{1_{i\ast}}\right\Vert }\rightleftharpoons
\sum\nolimits_{i=1}^{l_{2}}\psi_{2i\ast}\frac{\mathbf{x}_{2_{i\ast}}%
}{\left\Vert \mathbf{x}_{2_{i\ast}}\right\Vert }%
\]
\qquad in the Wolfe dual eigenspace. Therefore, the class-conditional
probability density functions $\boldsymbol{\psi}_{1}$ and $\boldsymbol{\psi
}_{2}$ in the Wolfe dual eigenspace for class $\omega_{1}$ and class
$\omega_{2}$ are \emph{symmetrically balanced with each other}:%
\[
p\left(  \frac{\mathbf{x}_{1_{i\ast}}}{\left\Vert \mathbf{x}_{1_{i\ast}%
}\right\Vert }|\boldsymbol{\psi}_{1}\right)  \rightleftharpoons p\left(
\frac{\mathbf{x}_{2_{i\ast}}}{\left\Vert \mathbf{x}_{2_{i\ast}}\right\Vert
}|\boldsymbol{\psi}_{2}\right)  \text{.}%
\]

I\ will now devise expressions for the class-conditional probability density
functions in the decision space $Z$ for class $\omega_{1}$ and class
$\omega_{2}$.

\subsection{Class-conditional Probability Densities}

I\ will now show that a linear eigenlocus $\boldsymbol{\tau=\tau}%
_{1}-\boldsymbol{\tau}_{2}$ is a parameter vector for class-conditional
probability density functions $p\left(  \mathbf{x}_{1_{i\ast}}|\omega
_{1}\right)  $ and $p\left(  \mathbf{x}_{2_{i\ast}}|\omega_{2}\right)  $.

\subsubsection{Class-Conditional Density for Class $\omega_{1}$}

Given that each Wolfe dual principal eigenaxis component $\psi_{1i\ast
}\overrightarrow{\mathbf{e}}_{1i\ast}$ specifies a conditional density
$p\left(  \mathbf{x}_{1_{i_{\ast}}}|\operatorname{comp}%
_{\overrightarrow{\mathbf{x}_{1i\ast}}}\left(
\overrightarrow{\boldsymbol{\tau}}\right)  \right)  $ for a correlated extreme
point $\mathbf{x}_{1_{i_{\ast}}}$, it follows that conditional densities for
the $\mathbf{x}_{1_{i_{\ast}}}$ extreme points are distributed over the
principal eigenaxis components of $\boldsymbol{\tau}_{1}$%
\begin{align}
\boldsymbol{\tau}_{1}  &  =\sum\nolimits_{i=1}^{l_{1}}p\left(  \mathbf{x}%
_{1_{i_{\ast}}}|\operatorname{comp}_{\overrightarrow{\mathbf{x}_{1i\ast}}%
}\left(  \overrightarrow{\boldsymbol{\tau}}\right)  \right)  \mathbf{x}%
_{1_{i\ast}}\label{Conditional Density Extreme Points 1}\\
&  =\sum\nolimits_{i=1}^{l_{1}}\psi_{1_{i\ast}}\mathbf{x}_{1_{i\ast}}%
\text{,}\nonumber
\end{align}
where $\psi_{1_{i\ast}}\mathbf{x}_{1_{i\ast}}$ specifies a conditional density
for $\mathbf{x}_{1_{i_{\ast}}}$, such that $\boldsymbol{\tau}_{1}$ is a
parameter vector for a class-conditional probability density $p\left(
\mathbf{x}_{1_{i\ast}}|\boldsymbol{\tau}_{1}\right)  $ for a given set
$\left\{  \mathbf{x}_{1_{i\ast}}\right\}  _{i=1}^{l_{1}}$ of $\mathbf{x}%
_{1_{i_{\ast}}}$ extreme points:%
\[
\boldsymbol{\tau}_{1}=p\left(  \mathbf{x}_{1_{i\ast}}|\boldsymbol{\tau}%
_{1}\right)  \text{.}%
\]

\subsubsection{Class-Conditional Density for Class $\omega_{2}$}

Given that each Wolfe dual principal eigenaxis component $\psi_{2i\ast
}\overrightarrow{\mathbf{e}}_{2i\ast}$ specifies a conditional density
$p\left(  \mathbf{x}_{2_{i_{\ast}}}|\operatorname{comp}%
_{\overrightarrow{\mathbf{x}_{2i\ast}}}\left(
\overrightarrow{\boldsymbol{\tau}}\right)  \right)  $ for a correlated extreme
point $\mathbf{x}_{2_{i_{\ast}}}$, it follows that conditional densities for
the $\mathbf{x}_{2_{i_{\ast}}}$ extreme points are distributed over the
principal eigenaxis components of $\boldsymbol{\tau}_{2}$%
\begin{align}
\boldsymbol{\tau}_{2}  &  =\sum\nolimits_{i=1}^{l_{2}}p\left(  \mathbf{x}%
_{2_{i_{\ast}}}|\operatorname{comp}_{\overrightarrow{\mathbf{x}_{2i\ast}}%
}\left(  \overrightarrow{\boldsymbol{\tau}}\right)  \right)  \mathbf{x}%
_{2_{i\ast}}\label{Conditional Density Extreme Points 2}\\
&  =\sum\nolimits_{i=1}^{l_{2}}\psi_{2_{i\ast}}\mathbf{x}_{2_{i\ast}}%
\text{,}\nonumber
\end{align}
where $\psi_{2_{i\ast}}\mathbf{x}_{2_{i\ast}}$ specifies a conditional density
for $\mathbf{x}_{2_{i\ast}}$, such that $\boldsymbol{\tau}_{2}$ is a parameter
vector for a class-conditional probability density $p\left(  \mathbf{x}%
_{2_{i_{\ast}}}|\boldsymbol{\tau}_{2}\right)  $ for a given set $\left\{
\mathbf{x}_{2_{i_{\ast}}}\right\}  _{i=1}^{l_{2}}$ of $\mathbf{x}_{2_{i_{\ast
}}}$ extreme points:%
\[
\boldsymbol{\tau}_{2}=p\left(  \mathbf{x}_{2_{i_{\ast}}}|\boldsymbol{\tau}%
_{2}\right)  \text{.}%
\]
Therefore, it is concluded that $\boldsymbol{\tau}_{1}$ is a parameter vector
for the class-conditional probability density function $p\left(
\mathbf{x}_{1_{i\ast}}|\boldsymbol{\tau}_{1}\right)  $ and $\boldsymbol{\tau
}_{2}$ is a parameter vector for the class-conditional probability density
function $p\left(  \mathbf{x}_{2_{i_{\ast}}}|\boldsymbol{\tau}_{2}\right)  $.

I will now devise integrals for the conditional probability functions for
class $\omega_{1}$ and class $\omega_{2}$.

\subsection{Conditional Probability Functions}

I\ will now show that the conditional probability function $P\left(
\mathbf{x}_{1_{i\ast}}|\boldsymbol{\tau}_{1}\right)  $ for class $\omega_{1}$
is given by the area under the class-conditional probability density function
$p\left(  \mathbf{x}_{1_{i\ast}}|\boldsymbol{\tau}_{1}\right)  $ over the
decision space $Z$.

\subsubsection{Conditional Probability Function for Class $\omega_{1}$}

A linear eigenlocus $\boldsymbol{\tau}=\sum\nolimits_{i=1}^{l_{1}}%
\psi_{1_{i\ast}}\mathbf{x}_{1_{i\ast}}-\sum\nolimits_{i=1}^{l_{2}}%
\psi_{2_{i\ast}}\mathbf{x}_{2_{i\ast}}$ is the basis of a linear
classification system $\boldsymbol{\tau}^{T}\mathbf{x}+\tau_{0}\overset{\omega
_{1}}{\underset{\omega_{2}}{\gtrless}}0$ that partitions any given feature
space into congruent decision regions $Z_{1}\cong Z_{2}$, whereby, for any two
overlapping data distributions, an $\mathbf{x}_{1_{i_{\ast}}}$ or
$\mathbf{x}_{2_{i_{\ast}}}$ extreme point lies in either region $Z_{1}$ or
region $Z_{2}$, and for any two non-overlapping data distributions,
$\mathbf{x}_{1_{i_{\ast}}}$ extreme points lie in region $Z_{1}$ and
$\mathbf{x}_{2_{i_{\ast}}}$ extreme points lie in region $Z_{2}$.

Therefore, the area under each pointwise conditional density in Eq.
(\ref{Conditional Density Extreme Points 1})%
\[
\int_{Z_{1}}p\left(  \mathbf{x}_{1_{i_{\ast}}}|\operatorname{comp}%
_{\overrightarrow{\mathbf{x}_{1i\ast}}}\left(
\overrightarrow{\boldsymbol{\tau}}\right)  \right)  d\boldsymbol{\tau}%
_{1}\left(  \mathbf{x}_{1_{i_{\ast}}}\right)  \text{ or }\int_{Z_{2}}p\left(
\mathbf{x}_{1_{i_{\ast}}}|\operatorname{comp}_{\overrightarrow{\mathbf{x}%
_{1i\ast}}}\left(  \overrightarrow{\boldsymbol{\tau}}\right)  \right)
d\boldsymbol{\tau}_{1}\left(  \mathbf{x}_{1_{i_{\ast}}}\right)
\]
is a conditional probability that an $\mathbf{x}_{1_{i_{\ast}}}$ extreme point
will be observed in either region $Z_{1}$ or region $Z_{2}$.

Thus, the area $P\left(  \mathbf{x}_{1_{i\ast}}|\boldsymbol{\tau}_{1}\right)
$ under the class-conditional probability density function $p\left(
\mathbf{x}_{1_{i\ast}}|\boldsymbol{\tau}_{1}\right)  $ in Eq.
(\ref{Conditional Density Extreme Points 1})%
\begin{align*}
P\left(  \mathbf{x}_{1_{i\ast}}|\boldsymbol{\tau}_{1}\right)   &  =\int%
_{Z}\left(  \sum\nolimits_{i=1}^{l_{1}}p\left(  \mathbf{x}_{1_{i_{\ast}}%
}|\operatorname{comp}_{\overrightarrow{\mathbf{x}_{1i\ast}}}\left(
\overrightarrow{\boldsymbol{\tau}}\right)  \right)  \mathbf{x}_{1_{i\ast}%
}\right)  d\boldsymbol{\tau}_{1}\\
&  =\int_{Z}\left(  \sum\nolimits_{i=1}^{l_{1}}\psi_{1_{i\ast}}\mathbf{x}%
_{1_{i\ast}}\right)  d\boldsymbol{\tau}_{1}=\int_{Z}p\left(  \mathbf{x}%
_{1_{i\ast}}|\boldsymbol{\tau}_{1}\right)  d\boldsymbol{\tau}_{1}\\
&  =\int_{Z}\boldsymbol{\tau}_{1}d\boldsymbol{\tau}_{1}=\frac{1}{2}\left\Vert
\boldsymbol{\tau}_{1}\right\Vert ^{2}+C=\left\Vert \boldsymbol{\tau}%
_{1}\right\Vert ^{2}+C_{1}%
\end{align*}
specifies the conditional probability of observing a set $\left\{
\mathbf{x}_{1_{i\ast}}\right\}  _{i=1}^{l_{1}}$ of $\mathbf{x}_{1_{i_{\ast}}}$
extreme points within \emph{localized regions} of the decision space $Z$,
where conditional densities $\psi_{1_{i\ast}}\mathbf{x}_{1_{i\ast}}$ for
$\mathbf{x}_{1_{i_{\ast}}}$ extreme points that lie in the $Z_{2}$ decision
region \emph{contribute} to the cost or risk $\mathfrak{R}_{\mathfrak{\min}%
}\left(  Z_{2}|\psi_{1i\ast}\mathbf{x}_{1_{i_{\ast}}}\right)  $ of making a
decision error, and conditional densities $\psi_{1_{i\ast}}\mathbf{x}%
_{1_{i\ast}}$ for $\mathbf{x}_{1_{i_{\ast}}}$ extreme points that lie in the
$Z_{1}$ decision region \emph{counteract }the cost or risk $\overline
{\mathfrak{R}}_{\mathfrak{\min}}\left(  Z_{1}|\psi_{1i\ast}\mathbf{x}%
_{1_{i_{\ast}}}\right)  $ of making a decision error.

It follows that the area $P\left(  \mathbf{x}_{1_{i\ast}}|\boldsymbol{\tau
}_{1}\right)  $ under the class-conditional probability density function
$p\left(  \mathbf{x}_{1_{i\ast}}|\boldsymbol{\tau}_{1}\right)  $ is determined
by regions of risk $\mathfrak{R}_{\mathfrak{\min}}\left(  Z_{2}|\psi_{1i\ast
}\mathbf{x}_{1_{i_{\ast}}}\right)  $ and regions of counter risk
$\overline{\mathfrak{R}}_{\mathfrak{\min}}\left(  Z_{1}|\psi_{1i\ast
}\mathbf{x}_{1_{i_{\ast}}}\right)  $ for the $\mathbf{x}_{1_{i_{\ast}}}$
extreme points, where regions of risk $\mathfrak{R}_{\mathfrak{\min}}\left(
Z_{2}|\psi_{1i\ast}\mathbf{x}_{1_{i_{\ast}}}\right)  $ and regions of counter
risk $\overline{\mathfrak{R}}_{\mathfrak{\min}}\left(  Z_{1}|\psi_{1i\ast
}\mathbf{x}_{1_{i_{\ast}}}\right)  $ are localized regions in decision space
$Z$ that are determined by central locations (expected values) and spreads
(covariances) of $\mathbf{x}_{1_{i\ast}}$ extreme points.

Therefore, the conditional probability function $P\left(  \mathbf{x}%
_{1_{i\ast}}|\boldsymbol{\tau}_{1}\right)  $ for class $\omega_{1}$ is given
by the integral%
\begin{align}
P\left(  \mathbf{x}_{1_{i\ast}}|\boldsymbol{\tau}_{1}\right)   &  =\int%
_{Z}p\left(  \widehat{\Lambda}_{\boldsymbol{\tau}}\left(  \mathbf{x}\right)
|\omega_{1}\right)  d\widehat{\Lambda}=\int_{Z}p\left(  \mathbf{x}_{1_{i\ast}%
}|\boldsymbol{\tau}_{1}\right)  d\boldsymbol{\tau}_{1}%
\label{Conditional Probability Function for Class One}\\
&  =\int_{Z}\boldsymbol{\tau}_{1}d\boldsymbol{\tau}_{1}=\left\Vert
\boldsymbol{\tau}_{1}\right\Vert ^{2}+C_{1}\text{,}\nonumber
\end{align}
over the decision space $Z$, which has a solution in terms of the critical
minimum eigenenergy $\left\Vert \boldsymbol{\tau}_{1}\right\Vert _{\min_{c}%
}^{2}$ exhibited by $\boldsymbol{\tau}_{1}$ and an integration constant
$C_{1}$.

I\ will now demonstrate that the conditional probability function $P\left(
\mathbf{x}_{2_{i\ast}}|\boldsymbol{\tau}_{2}\right)  $ for class $\omega_{2}$
is given by the area under the class-conditional probability density function
$p\left(  \mathbf{x}_{2_{i_{\ast}}}|\boldsymbol{\tau}_{2}\right)  $ over the
decision space $Z.$

\subsubsection{Conditional Probability Function for Class $\omega_{2}$}

The area under each pointwise conditional density in Eq.
(\ref{Conditional Density Extreme Points 2})%
\[
\int_{Z_{1}}p\left(  \mathbf{x}_{2_{i_{\ast}}}|\operatorname{comp}%
_{\overrightarrow{\mathbf{x}_{2i\ast}}}\left(
\overrightarrow{\boldsymbol{\tau}}\right)  \right)  d\boldsymbol{\tau}%
_{2}\left(  \mathbf{x}_{2_{i\ast}}\right)  \text{ or }\int_{Z_{2}}p\left(
\mathbf{x}_{2_{i_{\ast}}}|\operatorname{comp}_{\overrightarrow{\mathbf{x}%
_{2i\ast}}}\left(  \overrightarrow{\boldsymbol{\tau}}\right)  \right)
d\boldsymbol{\tau}_{2}\left(  \mathbf{x}_{2_{i\ast}}\right)
\]
is a conditional probability that an $\mathbf{x}_{2_{i_{\ast}}}$ extreme point
will be observed in either region $Z_{1}$ or region $Z_{2}$.

Thus, the area $P\left(  \mathbf{x}_{2_{i\ast}}|\boldsymbol{\tau}_{2}\right)
$ under the class-conditional probability density function $p\left(
\mathbf{x}_{2_{i_{\ast}}}|\boldsymbol{\tau}_{2}\right)  $ in Eq.
(\ref{Conditional Density Extreme Points 2})%
\begin{align*}
P\left(  \mathbf{x}_{2_{i\ast}}|\boldsymbol{\tau}_{2}\right)   &  =\int%
_{Z}\left(  \sum\nolimits_{i=1}^{l_{2}}p\left(  \mathbf{x}_{2_{i_{\ast}}%
}|\operatorname{comp}_{\overrightarrow{\mathbf{x}_{2i\ast}}}\left(
\overrightarrow{\boldsymbol{\tau}}\right)  \right)  \mathbf{x}_{2_{i\ast}%
}\right)  d\boldsymbol{\tau}_{2}\\
&  =\int_{Z}\left(  \sum\nolimits_{i=1}^{l_{2}}\psi_{2_{i\ast}}\mathbf{x}%
_{2_{i\ast}}\right)  d\boldsymbol{\tau}_{2}=\int_{Z}p\left(  \mathbf{x}%
_{2_{i\ast}}|\boldsymbol{\tau}_{2}\right)  d\boldsymbol{\tau}_{2}\\
&  =\int_{Z}\boldsymbol{\tau}_{2}d\boldsymbol{\tau}_{2}=\frac{1}{2}\left\Vert
\boldsymbol{\tau}_{2}\right\Vert ^{2}+C=\left\Vert \boldsymbol{\tau}%
_{2}\right\Vert ^{2}+C_{2}%
\end{align*}
specifies the conditional probability of observing a set $\left\{
\mathbf{x}_{2_{i_{\ast}}}\right\}  _{i=1}^{l_{2}}$ of $\mathbf{x}_{2_{i_{\ast
}}}$ extreme points within localized regions of the decision space $Z$, where
conditional densities $\psi_{2i\ast}\mathbf{x}_{2_{i_{\ast}}}$ for
$\mathbf{x}_{2_{i_{\ast}}}$ extreme points that lie in the $Z_{1}$ decision
region \emph{contribute} to the cost or risk $\mathfrak{R}_{\mathfrak{\min}%
}\left(  Z_{1}|\psi_{2i\ast}\mathbf{x}_{2_{i_{\ast}}}\right)  $ of making a
decision error, and conditional densities $\psi_{2i\ast}\mathbf{x}%
_{2_{i_{\ast}}}$ for $\mathbf{x}_{2_{i_{\ast}}}$ extreme points that lie in
the $Z_{2}$ decision region \emph{counteract }the cost or risk $\overline
{\mathfrak{R}}_{\mathfrak{\min}}\left(  Z_{2}|\psi_{2i\ast}\mathbf{x}%
_{2_{i_{\ast}}}\right)  $ of making a decision error.

It follows that the area $P\left(  \mathbf{x}_{2_{i_{\ast}}}|\boldsymbol{\tau
}_{2}\right)  $ under the class-conditional probability density function
$p\left(  \mathbf{x}_{2_{i_{\ast}}}|\boldsymbol{\tau}_{2}\right)  $ is
determined by regions of risk $\mathfrak{R}_{\mathfrak{\min}}\left(
Z_{1}|\psi_{2i\ast}\mathbf{x}_{2_{i_{\ast}}}\right)  $ and regions of counter
risk $\overline{\mathfrak{R}}_{\mathfrak{\min}}\left(  Z_{2}|\psi_{2i\ast
}\mathbf{x}_{2_{i_{\ast}}}\right)  $ for the $\mathbf{x}_{2_{i_{\ast}}}$
extreme points, where regions of risk $\mathfrak{R}_{\mathfrak{\min}}\left(
Z_{1}|\psi_{2i\ast}\mathbf{x}_{2_{i_{\ast}}}\right)  $ and regions of counter
risk $\overline{\mathfrak{R}}_{\mathfrak{\min}}\left(  Z_{2}|\psi_{2i\ast
}\mathbf{x}_{2_{i_{\ast}}}\right)  $ are localized regions in decision space
$Z$ that are determined by central locations (expected values) and spreads
(covariances) of $\mathbf{x}_{2_{i\ast}}$ extreme points.

Therefore, the conditional probability function $P\left(  \mathbf{x}%
_{2_{i\ast}}|\boldsymbol{\tau}_{2}\right)  $ for class $\omega_{2}$ is given
by the integral%
\begin{align}
P\left(  \mathbf{x}_{2_{i\ast}}|\boldsymbol{\tau}_{2}\right)   &  =\int%
_{Z}p\left(  \widehat{\Lambda}_{\boldsymbol{\tau}}\left(  \mathbf{x}\right)
|\omega_{2}\right)  d\widehat{\Lambda}=\int_{Z}p\left(  \mathbf{x}_{2_{i\ast}%
}|\boldsymbol{\tau}_{2}\right)  d\boldsymbol{\tau}_{2}%
\label{Conditional Probability Function for  Class Two}\\
&  =\int_{Z}\boldsymbol{\tau}_{2}d\boldsymbol{\tau}_{2}=\left\Vert
\boldsymbol{\tau}_{2}\right\Vert ^{2}+C_{2}\text{,}\nonumber
\end{align}
over the decision space $Z$, which has a solution in terms of the critical
minimum eigenenergy $\left\Vert \boldsymbol{\tau}_{2}\right\Vert _{\min_{c}%
}^{2}$ exhibited by $\boldsymbol{\tau}_{2}$ and an integration constant
$C_{2}$.

In order to precisely define the manner in which constrained linear eigenlocus
discriminant functions $\widetilde{\Lambda}_{\boldsymbol{\tau}}\left(
\mathbf{x}\right)  =\mathbf{x}^{T}\boldsymbol{\tau}+\tau_{0}$ satisfy a
data-driven version of the fundamental integral equation of binary
classification for a classification system in statistical equilibrium, I need
to precisely define the manner in which the total allowed eigenenergies of the
principal eigenaxis components on $\boldsymbol{\tau}$ are symmetrically
balanced with each other. Furthermore, I need to identify the manner in which
the property of symmetrical balance exhibited by the principal eigenaxis
components on $\boldsymbol{\psi}$ \emph{and} $\boldsymbol{\tau}$ enables
linear eigenlocus classification systems $\mathbf{x}^{T}\boldsymbol{\tau}%
+\tau_{0}\overset{\omega_{1}}{\underset{\omega_{2}}{\gtrless}}0$ to
\emph{effectively balance all of the forces associated with }the risk
$\mathfrak{R}_{\mathfrak{\min}}\left(  Z_{1}|\boldsymbol{\tau}_{2}\right)  $
and the counter risk $\overline{\mathfrak{R}}_{\mathfrak{\min}}\left(
Z_{1}|\mathbf{\tau}_{1}\right)  $ in the $Z_{1}$ decision region \emph{with
all of the forces associated with} the risk $\mathfrak{R}_{\mathfrak{\min}%
}\left(  Z_{2}|\mathbf{\tau}_{1}\right)  $ and the counter risk $\overline
{\mathfrak{R}}_{\mathfrak{\min}}\left(  Z_{2}|\boldsymbol{\tau}_{2}\right)  $
in the $Z_{2}$ decision region.

Recall that the risk $\mathfrak{R}_{\mathfrak{\min}}\left(  Z|\widehat{\Lambda
}\left(  \mathbf{x}\right)  \right)  $ of a binary classification system%
\[
\mathfrak{R}_{\mathfrak{\min}}\left(  Z|\widehat{\Lambda}\left(
\mathbf{x}\right)  \right)  =\mathfrak{R}_{\mathfrak{\min}}\left(
Z_{2}|p\left(  \widehat{\Lambda}\left(  \mathbf{x}\right)  |\omega_{1}\right)
\right)  +\mathfrak{R}_{\mathfrak{\min}}\left(  Z_{1}|p\left(
\widehat{\Lambda}\left(  \mathbf{x}\right)  |\omega_{2}\right)  \right)
\]
involves opposing forces that depend on the likelihood ratio test
$\widehat{\Lambda}\left(  \mathbf{x}\right)  =p\left(  \widehat{\Lambda
}\left(  \mathbf{x}\right)  |\omega_{1}\right)  -p\left(  \widehat{\Lambda
}\left(  \mathbf{x}\right)  |\omega_{2}\right)  \overset{\omega_{1}%
}{\underset{\omega_{2}}{\gtrless}}0$ and the corresponding decision boundary
$p\left(  \widehat{\Lambda}\left(  \mathbf{x}\right)  |\omega_{1}\right)
-p\left(  \widehat{\Lambda}\left(  \mathbf{x}\right)  |\omega_{2}\right)  =0$.

It has been demonstrated that linear eigenlocus transforms define these
opposing forces in terms of symmetrically balanced, pointwise covariance
statistics:%
\begin{align*}
\psi_{1i\ast}\frac{\mathbf{x}_{1_{i\ast}}}{\left\Vert \mathbf{x}_{1_{i\ast}%
}\right\Vert }  &  =\lambda_{\max_{\boldsymbol{\psi}}}^{-1}\left\Vert
\mathbf{x}_{1_{i_{\ast}}}\right\Vert \sum\nolimits_{j=1}^{l_{1}}\psi
_{1_{j\ast}}\left\Vert \mathbf{x}_{1_{j\ast}}\right\Vert \cos\theta
_{\mathbf{x}_{1_{i\ast}}\mathbf{x}_{1_{j\ast}}}\\
&  -\lambda_{\max_{\boldsymbol{\psi}}}^{-1}\left\Vert \mathbf{x}_{1_{i_{\ast}%
}}\right\Vert \sum\nolimits_{j=1}^{l_{2}}\psi_{2_{j\ast}}\left\Vert
\mathbf{x}_{2_{j\ast}}\right\Vert \cos\theta_{\mathbf{x}_{1_{i\ast}}%
\mathbf{x}_{2_{j\ast}}}\text{,}%
\end{align*}
and%
\begin{align*}
\psi_{2i\ast}\frac{\mathbf{x}_{2_{i\ast}}}{\left\Vert \mathbf{x}_{2_{i\ast}%
}\right\Vert }  &  =\lambda_{\max_{\boldsymbol{\psi}}}^{-1}\left\Vert
\mathbf{x}_{2_{i\ast}}\right\Vert \sum\nolimits_{j=1}^{l_{2}}\psi_{2_{j\ast}%
}\left\Vert \mathbf{x}_{2_{j\ast}}\right\Vert \cos\theta_{\mathbf{x}%
_{2_{i\ast}}\mathbf{x}_{2_{j\ast}}}\\
&  -\lambda_{\max_{\boldsymbol{\psi}}}^{-1}\left\Vert \mathbf{x}_{2_{i\ast}%
}\right\Vert \sum\nolimits_{j=1}^{l_{1}}\psi_{1_{j\ast}}\left\Vert
\mathbf{x}_{1_{j\ast}}\right\Vert \cos\theta_{\mathbf{x}_{2_{i\ast}}%
\mathbf{x}_{1_{j\ast}}}\text{,}%
\end{align*}
such that any given conditional density%
\[
p\left(  \mathbf{x}_{1_{i_{\ast}}}|\operatorname{comp}%
_{\overrightarrow{\mathbf{x}_{1i\ast}}}\left(
\overrightarrow{\boldsymbol{\tau}}\right)  \right)  \text{ \ or \ }p\left(
\mathbf{x}_{2_{i_{\ast}}}|\operatorname{comp}_{\overrightarrow{\mathbf{x}%
_{2i\ast}}}\left(  \overrightarrow{\boldsymbol{\tau}}\right)  \right)
\]
for a respective extreme point $\mathbf{x}_{1_{i\ast}}$ or $\mathbf{x}%
_{2_{i\ast}}$ is defined in terms of forces related to counter risks and risks
associated with positions and potential locations of $\mathbf{x}_{1_{i\ast}}$
and $\mathbf{x}_{2_{i\ast}}$ extreme points within the $Z_{1}$ and $Z_{2}$
decision regions of a decision space $Z$.

Linear eigenlocus transforms routinely accomplish an elegant, statistical
balancing feat that involves finding the right mix of principal eigenaxis
components on $\boldsymbol{\psi}$ and $\boldsymbol{\tau}$. I\ will now show
that the scale factors $\left\{  \psi_{i\ast}\right\}  _{i=1}^{l}$ of the
Wolfe dual principal eigenaxis components $\left\{  \psi_{i\ast}%
\frac{\mathbf{x}_{i\ast}}{\left\Vert \mathbf{x}_{i\ast}\right\Vert }%
|\psi_{i\ast}>0\right\}  _{i=1}^{l}$ on $\boldsymbol{\psi}$ play a fundamental
role in this statistical balancing feat. I\ will develop an equation of
statistical equilibrium for the axis of $\boldsymbol{\tau}$ that is determined
by the equation of statistical equilibrium:%
\[
\sum\nolimits_{i=1}^{l_{1}}\psi_{1i\ast}\frac{\mathbf{x}_{1_{i\ast}}%
}{\left\Vert \mathbf{x}_{1_{i\ast}}\right\Vert }\rightleftharpoons
\sum\nolimits_{i=1}^{l_{2}}\psi_{2i\ast}\frac{\mathbf{x}_{2_{i\ast}}%
}{\left\Vert \mathbf{x}_{2_{i\ast}}\right\Vert }%
\]
for the axis of $\boldsymbol{\psi}$.

\subsection{Finding the Right Mix of Component Lengths}

It has been demonstrated that the directions of the constrained primal and the
Wolfe dual principal eigenaxis components are fixed, along with the angles
between all of the extreme vectors. I will now show that the lengths of the
Wolfe dual principal eigenaxis components on $\boldsymbol{\psi}$ must satisfy
critical magnitude constraints.

Using Eq. (\ref{Dual Eigen-coordinate Locations Component One}), it follows
that the integrated lengths $\sum\nolimits_{i=1}^{l_{1}}\psi_{1i\ast}$ of the
components $\psi_{1i\ast}\frac{\mathbf{x}_{1_{i\ast}}}{\left\Vert
\mathbf{x}_{1_{i\ast}}\right\Vert }$ on $\boldsymbol{\psi}_{1}$ must satisfy
the equation:%
\begin{align}
\sum\nolimits_{i=1}^{l_{1}}\psi_{1i\ast}  &  =\lambda_{\max_{\boldsymbol{\psi
}}}^{-1}\sum\nolimits_{i=1}^{l_{1}}\left\Vert \mathbf{x}_{1_{i\ast}%
}\right\Vert \label{integrated dual loci one1}\\
&  \times\sum\nolimits_{j=1}^{l_{1}}\psi_{1_{j\ast}}\left\Vert \mathbf{x}%
_{1_{j\ast}}\right\Vert \cos\theta_{\mathbf{x}_{1_{i\ast}}\mathbf{x}%
_{1_{j\ast}}}\nonumber\\
&  -\lambda_{\max_{\boldsymbol{\psi}}}^{-1}\sum\nolimits_{i=1}^{l_{1}%
}\left\Vert \mathbf{x}_{1_{i\ast}}\right\Vert \nonumber\\
&  \times\sum\nolimits_{j=1}^{l_{2}}\psi_{2_{j\ast}}\left\Vert \mathbf{x}%
_{2_{j\ast}}\right\Vert \cos\theta_{\mathbf{x}_{1_{i\ast}}\mathbf{x}%
_{2_{j\ast}}}\nonumber
\end{align}
which reduces to%
\[
\sum\nolimits_{i=1}^{l_{1}}\psi_{1i\ast}=\lambda_{\max_{\boldsymbol{\psi}}%
}^{-1}\sum\nolimits_{i=1}^{l_{1}}\mathbf{x}_{1_{i\ast}}^{T}\left(
\sum\nolimits_{j=1}^{l_{1}}\psi_{1_{j\ast}}\mathbf{x}_{1_{j\ast}}%
-\sum\nolimits_{j=1}^{l_{2}}\psi_{2_{j\ast}}\mathbf{x}_{2_{j\ast}}\right)
\text{.}%
\]
Using Eq. (\ref{Dual Eigen-coordinate Locations Component Two}), it follows
that the integrated lengths $\sum\nolimits_{i=1}^{l_{2}}\psi_{2i\ast}$ of the
components $\psi_{2i\ast}\frac{\mathbf{x}_{2_{i\ast}}}{\left\Vert
\mathbf{x}_{2_{i\ast}}\right\Vert }$ on $\boldsymbol{\psi}_{2}$ must satisfy
the equation:%
\begin{align}
\sum\nolimits_{i=1}^{l_{2}}\psi_{2i\ast}  &  =\lambda_{\max_{\boldsymbol{\psi
}}}^{-1}\sum\nolimits_{i=1}^{l_{2}}\left\Vert \mathbf{x}_{2_{i\ast}%
}\right\Vert \label{integrated dual loci two1}\\
&  \times\sum\nolimits_{j=1}^{l_{2}}\psi_{2_{j\ast}}\left\Vert \mathbf{x}%
_{2_{j\ast}}\right\Vert \cos\theta_{\mathbf{x}_{2_{i\ast}}\mathbf{x}%
_{2_{j\ast}}}\nonumber\\
&  -\lambda_{\max_{\boldsymbol{\psi}}}^{-1}\sum\nolimits_{i=1}^{l_{2}%
}\left\Vert \mathbf{x}_{2_{i\ast}}\right\Vert \nonumber\\
&  \times\sum\nolimits_{j=1}^{l_{1}}\psi_{1_{j\ast}}\left\Vert \mathbf{x}%
_{1_{j\ast}}\right\Vert \cos\theta_{\mathbf{x}_{2_{i\ast}}\mathbf{x}%
_{1_{j\ast}}}\nonumber
\end{align}
which reduces to%
\[
\sum\nolimits_{i=1}^{l_{2}}\psi_{2i\ast}=\lambda_{\max_{\boldsymbol{\psi}}%
}^{-1}\sum\nolimits_{i=1}^{l_{2}}\mathbf{x}_{2_{i\ast}}^{T}\left(
\sum\nolimits_{j=1}^{l_{2}}\psi_{2_{j\ast}}\mathbf{x}_{2_{j\ast}}%
-\sum\nolimits_{j=1}^{l_{1}}\psi_{1_{j\ast}}\mathbf{x}_{1_{j\ast}}\right)
\text{.}%
\]

I will now show that Eqs (\ref{integrated dual loci one1}) and
(\ref{integrated dual loci two1}) determine a balanced eigenlocus equation,
where RHS Eq. (\ref{integrated dual loci one1}) $=$ RHS Eq.
(\ref{integrated dual loci two1}).

\subsection{Balanced Linear Eigenlocus Equations}

Returning to Eq. (\ref{Equilibrium Constraint on Dual Eigen-components})%
\[
\sum\nolimits_{i=1}^{l_{1}}\psi_{1_{i\ast}}\rightleftharpoons\sum
\nolimits_{i=1}^{l_{2}}\psi_{2_{i\ast}}\text{,}%
\]
where the axis of $\boldsymbol{\psi}$ is in statistical equilibrium, it
follows that the RHS\ of Eq. (\ref{integrated dual loci one1}) must equal the
RHS\ of Eq. (\ref{integrated dual loci two1}):%
\begin{align}
&  \sum\nolimits_{i=1}^{l_{1}}\mathbf{x}_{1_{i\ast}}^{T}\left(  \sum
\nolimits_{j=1}^{l_{1}}\psi_{1_{j\ast}}\mathbf{x}_{1_{j\ast}}-\sum
\nolimits_{j=1}^{l_{2}}\psi_{2_{j\ast}}\mathbf{x}_{2_{j\ast}}\right)
\label{Balanced Eigenlocus Equation Linear}\\
&  =\sum\nolimits_{i=1}^{l_{2}}\mathbf{x}_{2_{i\ast}}^{T}\left(
\sum\nolimits_{j=1}^{l_{2}}\psi_{2_{j\ast}}\mathbf{x}_{2_{j\ast}}%
-\sum\nolimits_{j=1}^{l_{1}}\psi_{1_{j\ast}}\mathbf{x}_{1_{j\ast}}\right)
\text{.}\nonumber
\end{align}

Therefore, all of the $\mathbf{x}_{1_{i\ast}}$ and $\mathbf{x}_{2_{i\ast}}$
extreme points are distributed over the axes of $\boldsymbol{\tau}_{1}%
$\textbf{ }and $\boldsymbol{\tau}_{2}$ in the symmetrically balanced manner:%
\begin{equation}
\sum\nolimits_{i=1}^{l_{1}}\mathbf{x}_{1_{i\ast}}^{T}\left(  \boldsymbol{\tau
}_{1}\mathbf{-}\boldsymbol{\tau}_{2}\right)  =\sum\nolimits_{i=1}^{l_{2}%
}\mathbf{x}_{2_{i\ast}}^{T}\left(  \boldsymbol{\tau}_{2}\mathbf{-}%
\boldsymbol{\tau}_{1}\right)  \text{,} \label{Balanced Eigenlocus Equation}%
\end{equation}
where the components of the $\mathbf{x}_{1_{i\ast}}$ extreme vectors along the
axis of $\boldsymbol{\tau}_{2}$ oppose the components of the $\mathbf{x}%
_{1_{i\ast}}$ extreme vectors along the axis of $\boldsymbol{\tau}_{1}$, and
the components of the $\mathbf{x}_{2_{i\ast}}$ extreme vectors along the axis
of $\boldsymbol{\tau}_{1}$ oppose the components of the $\mathbf{x}_{2_{i\ast
}}$ extreme vectors along the axis of $\boldsymbol{\tau}_{2}$.

Let $\widehat{\mathbf{x}}_{i\ast}\triangleq\sum\nolimits_{i=1}^{l}%
\mathbf{x}_{i\ast}$, where $\sum\nolimits_{i=1}^{l}\mathbf{x}_{i\ast}%
=\sum\nolimits_{i=1}^{l_{1}}\mathbf{x}_{1_{i\ast}}+\sum\nolimits_{i=1}^{l_{2}%
}\mathbf{x}_{2_{i\ast}}$. Using Eq. (\ref{Balanced Eigenlocus Equation}), it
follows that the component of $\widehat{\mathbf{x}}_{i\ast}$ along
$\boldsymbol{\tau}_{1}$ is symmetrically balanced with the component of
$\widehat{\mathbf{x}}_{i\ast}$ along $\boldsymbol{\tau}_{2}$%
\[
\operatorname{comp}_{\overrightarrow{\boldsymbol{\tau}_{1}}}\left(
\overrightarrow{\widehat{\mathbf{x}}_{i\ast}}\right)  =\operatorname{comp}%
_{\overrightarrow{\boldsymbol{\tau}_{2}}}\left(
\overrightarrow{\widehat{\mathbf{x}}_{i\ast}}\right)
\]
so that the components $\operatorname{comp}_{\overrightarrow{\boldsymbol{\tau
}_{1}}}\left(  \overrightarrow{\widehat{\mathbf{x}}_{i\ast}}\right)  $ and
$\operatorname{comp}_{\overrightarrow{\boldsymbol{\tau}_{2}}}\left(
\overrightarrow{\widehat{\mathbf{x}}_{i\ast}}\right)  $ of clusters or
aggregates of the extreme vectors from both pattern classes have \emph{equal
forces associated with risks and counter risks} on opposite sides of the axis
of $\boldsymbol{\tau}$.

\subsubsection{Statistical Equilibrium of Risks and Counter Risks}

\begin{figure}[ptb]%
\centering
\fbox{\includegraphics[
height=2.5875in,
width=3.4411in
]%
{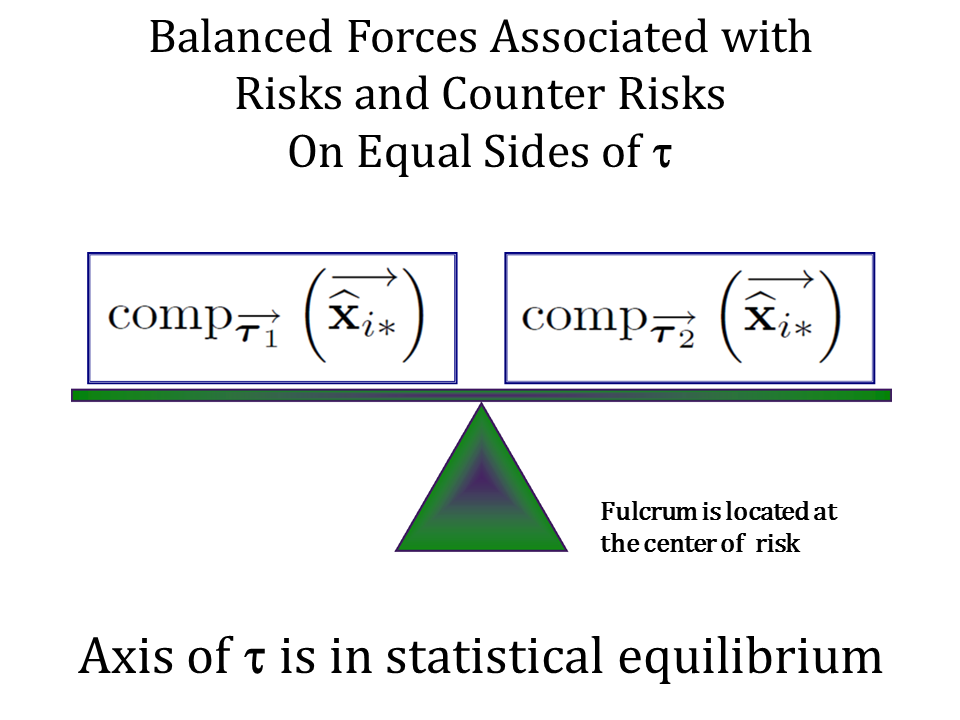}%
}\caption{The axis of $\boldsymbol{\tau}$ is in statistical equilibrium, where
two equal weights $\operatorname{comp}%
_{\protect\overrightarrow{\boldsymbol{\tau}_{1}}}\left(
\protect\overrightarrow{\protect\widehat{\mathbf{x}}_{i\ast}}\right)  $ and
$\operatorname{comp}_{\protect\overrightarrow{\boldsymbol{\tau}_{2}}}\left(
\protect\overrightarrow{\protect\widehat{\mathbf{x}}_{i\ast}}\right)  $ are
placed on opposite sides of the fulcrum of $\boldsymbol{\tau}$ which is
located at the center of total allowed eigenenergy $\left\Vert
\boldsymbol{\tau}_{1}-\boldsymbol{\tau}_{2}\right\Vert _{\min_{c}}^{2}$ of
$\boldsymbol{\tau}$.}%
\label{Statistical Equilibrium of Primal Linear Eigenlocus}%
\end{figure}

Given Eq. (\ref{Balanced Eigenlocus Equation}), it follows that the axis of
$\boldsymbol{\tau}$ is a lever of uniform density, where the center of
$\boldsymbol{\tau}$ is $\left\Vert \boldsymbol{\tau}\right\Vert _{\min_{c}%
}^{2}$, for which two equal weights $\operatorname{comp}%
_{\overrightarrow{\boldsymbol{\tau}_{1}}}\left(
\overrightarrow{\widehat{\mathbf{x}}_{i\ast}}\right)  $ and
$\operatorname{comp}_{\overrightarrow{\boldsymbol{\tau}_{2}}}\left(
\overrightarrow{\widehat{\mathbf{x}}_{i\ast}}\right)  $ are placed on opposite
sides of the fulcrum of $\boldsymbol{\tau}$, whereby the axis of
$\boldsymbol{\tau}$ is in \emph{statistical equilibrium}. Figure
$\ref{Statistical Equilibrium of Primal Linear Eigenlocus}$ illustrates the
axis of $\boldsymbol{\tau}$ in statistical equilibrium, where forces
associated with counter risks and risks of aggregates of extreme points are
balanced with each other.%

\subsection{Critical Magnitude Constraints}

Equation (\ref{Balanced Eigenlocus Equation}) indicates that the lengths%
\[
\left\{  \psi_{1_{i\ast}}|\psi_{1_{i\ast}}>0\right\}  _{i=1}^{l_{1}}\text{ and
}\left\{  \psi_{2_{i\ast}}|\psi_{2_{i\ast}}>0\right\}  _{i=1}^{l_{2}}%
\]
of the $l$ Wolfe dual principal eigenaxis components on $\boldsymbol{\psi}$
satisfy critical magnitude constraints, such that the Wolfe dual eigensystem
in Eq. (\ref{Dual Normal Eigenlocus Components}), which specifies highly
interconnected, balanced sets of inner product relationships amongst the Wolfe
dual and the constrained, primal principal eigenaxis components in Eqs
(\ref{integrated dual loci one1}) and (\ref{integrated dual loci two1}),
determines well-proportioned lengths $\psi_{1i\ast}$ or $\psi_{2i\ast}$ for
each Wolfe dual principal eigenaxis component $\psi_{1i\ast}\frac
{\mathbf{x}_{1_{i\ast}}}{\left\Vert \mathbf{x}_{1_{i\ast}}\right\Vert }$ or
$\psi_{2i\ast}\frac{\mathbf{x}_{2_{i\ast}}}{\left\Vert \mathbf{x}_{2_{i\ast}%
}\right\Vert }$ on $\boldsymbol{\psi}_{1}$ or $\boldsymbol{\psi}_{2}$, where
each scale factor $\psi_{1i\ast}$ or $\psi_{2i\ast}$ determines a
well-proportioned length for a correlated, constrained primal principal
eigenaxis component $\psi_{1_{i\ast}}\mathbf{x}_{1_{i\ast}}$ or $\psi
_{2_{i\ast}}\mathbf{x}_{2_{i\ast}}$ on $\boldsymbol{\tau}_{1}$ or
$\boldsymbol{\tau}_{2}$.

I will demonstrate that the axis of $\boldsymbol{\psi}$, which is constrained
to be in statistical equilibrium $\sum\nolimits_{i=1}^{l_{1}}\psi_{1i\ast
}\frac{\mathbf{x}_{1_{i\ast}}}{\left\Vert \mathbf{x}_{1_{i\ast}}\right\Vert
}\rightleftharpoons\sum\nolimits_{i=1}^{l_{2}}\psi_{2i\ast}\frac
{\mathbf{x}_{2_{i\ast}}}{\left\Vert \mathbf{x}_{2_{i\ast}}\right\Vert }$,
determines an equilibrium point $p\left(  \widehat{\Lambda}_{\boldsymbol{\psi
}}\left(  \mathbf{x}\right)  |\omega_{1}\right)  -p\left(  \widehat{\Lambda
}_{\boldsymbol{\psi}}\left(  \mathbf{x}\right)  |\omega_{2}\right)  =0$ of an
integral equation $f\left(  \widetilde{\Lambda}_{\boldsymbol{\tau}}\left(
\mathbf{x}\right)  \right)  $ such that a linear eigenlocus discriminant
function $\widetilde{\Lambda}_{\boldsymbol{\tau}}\left(  \mathbf{x}\right)
=\boldsymbol{\tau}^{T}\mathbf{x}+\tau_{0}$ is the solution to a fundamental
integral equation of binary classification for a linear classification system
$\boldsymbol{\tau}^{T}\mathbf{x}+\tau_{0}\overset{\omega_{1}}{\underset{\omega
_{2}}{\gtrless}}0$ in statistical equilibrium.

Let $\mathfrak{R}_{\mathfrak{\min}}\left(  Z|\mathbf{\tau}\right)  $ denote
the risk of a linear classification system $\boldsymbol{\tau}^{T}%
\mathbf{x}+\tau_{0}\overset{\omega_{1}}{\underset{\omega_{2}}{\gtrless}}0$
that is determined by a linear eigenlocus transform. Take any given set
$\left\{  \left\{  \mathbf{x}_{1_{i\ast}}\right\}  _{i=1}^{l_{1}},\;\left\{
\mathbf{x}_{2_{i\ast}}\right\}  _{i=1}^{l_{2}}\right\}  $ of extreme points
and take the set of scale factors $\left\{  \left\{  \psi_{1i\ast}\right\}
_{i=1}^{l_{1}},\left\{  \psi_{2i\ast}\right\}  _{i=1}^{l_{2}}\right\}  $ that
are determined by a linear eigenlocus transform.

Let $\overleftrightarrow{\mathfrak{R}}_{\mathfrak{\min}}\left(  Z|\psi
_{1_{j\ast}}\mathbf{x}_{1_{j\ast}}\right)  $ denote the force associated with
either the counter risk or the risk that is related to the locus of the scaled
extreme vector $\psi_{1_{j\ast}}\mathbf{x}_{1_{j\ast}}$ in the decision space
$Z$, where the force associated with the risk $\mathfrak{R}_{\mathfrak{\min}%
}\left(  Z|\mathbf{\tau}\right)  $ may be positive or negative. Let
$\overleftrightarrow{\mathfrak{R}}_{\mathfrak{\min}}\left(  Z|\psi_{2_{j\ast}%
}\mathbf{x}_{2_{j\ast}}\right)  $ denote the force associated with either the
counter risk or the risk that is related to the locus of the scaled extreme
vector $\psi_{2_{j\ast}}\mathbf{x}_{2_{j\ast}}$ in the decision space $Z$,
where the force associated with the risk $\mathfrak{R}_{\mathfrak{\min}%
}\left(  Z|\mathbf{\tau}\right)  $ may be positive or negative.

Take any given extreme point $\mathbf{x}_{1_{i\ast}}$ from class $\omega_{1}$.
Let $\overleftrightarrow{\mathfrak{R}}_{\mathfrak{\min}}\left(  Z|\psi
_{1_{j\ast}}\mathbf{x}_{1_{i\ast}}^{T}\mathbf{x}_{1_{j\ast}}\right)  $ denote
the force associated with either the counter risk or the risk that is related
to the locus of the component of the extreme vector $\mathbf{x}_{1_{i\ast}}$
along the scaled extreme vector $\psi_{1_{j\ast}}\mathbf{x}_{1_{j\ast}}$ in
the decision space $Z$, where the force associated with the risk
$\mathfrak{R}_{\mathfrak{\min}}\left(  Z|\mathbf{\tau}\right)  $ may be
positive or negative. Likewise, let $\overleftrightarrow{\mathfrak{R}%
}_{\mathfrak{\min}}\left(  Z|\psi_{2_{j\ast}}\mathbf{x}_{1_{i\ast}}%
^{T}\mathbf{x}_{2_{j\ast}}\right)  $ denote the force associated with either
the counter risk or the risk that is related to the locus of the component of
the extreme vector $\mathbf{x}_{1_{i\ast}}$ along the scaled extreme vector
$\psi_{2_{j\ast}}\mathbf{x}_{2_{j\ast}}$ in the decision space $Z$, where the
force associated with the risk $\mathfrak{R}_{\mathfrak{\min}}\left(
Z|\mathbf{\tau}\right)  $ may be positive or negative.

Take any given extreme point $\mathbf{x}_{2_{i_{\ast}}}$ from class
$\omega_{2}$. Let $\overleftrightarrow{\mathfrak{R}}_{\mathfrak{\min}}\left(
Z|\psi_{2_{j\ast}}\mathbf{x}_{2_{i\ast}}^{T}\mathbf{x}_{2_{j\ast}}\right)  $
denote the force associated with either the counter risk or the risk that is
related to the locus of the component of the extreme vector $\mathbf{x}%
_{2_{i\ast}}$ along the scaled extreme vector $\psi_{2_{j\ast}}\mathbf{x}%
_{2_{j\ast}}$ in the decision space $Z$, where the force associated with the
risk $\mathfrak{R}_{\mathfrak{\min}}\left(  Z|\mathbf{\tau}\right)  $ may be
positive or negative. Likewise, let $\overleftrightarrow{\mathfrak{R}%
}_{\mathfrak{\min}}\left(  Z|\psi_{1_{j\ast}}\mathbf{x}_{2_{i\ast}}%
^{T}\mathbf{x}_{1_{j\ast}}\right)  $ denote the force associated with either
the counter risk or the risk that is related to the locus of the component of
the extreme vector $\mathbf{x}_{2_{i\ast}}$ along the scaled extreme vector
$\psi_{1_{j\ast}}\mathbf{x}_{1_{j\ast}}$ in the decision space $Z$, where the
force associated with the risk $\mathfrak{R}_{\mathfrak{\min}}\left(
Z|\mathbf{\tau}\right)  $ may be positive or negative.

Returning to Eq. (\ref{Balanced Eigenlocus Equation Linear})%
\begin{align*}
&  \sum\nolimits_{i=1}^{l_{1}}\mathbf{x}_{1_{i\ast}}^{T}\left(  \sum
\nolimits_{j=1}^{l_{1}}\psi_{1_{j\ast}}\mathbf{x}_{1_{j\ast}}-\sum
\nolimits_{j=1}^{l_{2}}\psi_{2_{j\ast}}\mathbf{x}_{2_{j\ast}}\right) \\
&  =\sum\nolimits_{i=1}^{l_{2}}\mathbf{x}_{2_{i\ast}}^{T}\left(
\sum\nolimits_{j=1}^{l_{2}}\psi_{2_{j\ast}}\mathbf{x}_{2_{j\ast}}%
-\sum\nolimits_{j=1}^{l_{1}}\psi_{1_{j\ast}}\mathbf{x}_{1_{j\ast}}\right)
\text{,}%
\end{align*}
it follows that the collective forces associated with risks and counter risks,
which are related to the positions and the potential locations of all of the
extreme points, are balanced in the following manner:%
\begin{align*}
\mathfrak{R}_{\mathfrak{\min}}\left(  Z|\mathbf{\tau}\right)   &
:\sum\nolimits_{i=1}^{l_{1}}\left[  \sum\nolimits_{j=1}^{l_{1}}%
\overleftrightarrow{\mathfrak{R}}_{\mathfrak{\min}}\left(  Z|\psi_{1_{j\ast}%
}\mathbf{x}_{1_{i\ast}}^{T}\mathbf{x}_{1_{j\ast}}\right)  -\sum\nolimits_{j=1}%
^{l_{2}}\overleftrightarrow{\mathfrak{R}}_{\mathfrak{\min}}\left(
Z|\psi_{2_{j\ast}}\mathbf{x}_{1_{i\ast}}^{T}\mathbf{x}_{2_{j\ast}}\right)
\right] \\
&  =\sum\nolimits_{i=1}^{l_{2}}\left[  \sum\nolimits_{j=1}^{l_{2}%
}\overleftrightarrow{\mathfrak{R}}_{\mathfrak{\min}}\left(  Z|\psi_{2_{j\ast}%
}\mathbf{x}_{2_{i\ast}}^{T}\mathbf{x}_{2_{j\ast}}\right)  -\sum\nolimits_{j=1}%
^{l_{1}}\overleftrightarrow{\mathfrak{R}}_{\mathfrak{\min}}\left(
Z|\psi_{1_{j\ast}}\mathbf{x}_{2_{i\ast}}^{T}\mathbf{x}_{1_{j\ast}}\right)
\right]  \text{.}%
\end{align*}

So, take any given set $\left\{  \left\{  \mathbf{x}_{1_{i\ast}}\right\}
_{i=1}^{l_{1}},\;\left\{  \mathbf{x}_{2_{i\ast}}\right\}  _{i=1}^{l_{2}%
}\right\}  $ of extreme points and take the set of scale factors $\left\{
\left\{  \psi_{1i\ast}\right\}  _{i=1}^{l_{1}},\left\{  \psi_{2i\ast}\right\}
_{i=1}^{l_{2}}\right\}  $ that are determined by a linear eigenlocus transform.

I\ will show that linear eigenlocus transforms choose magnitudes or scale
factors for the Wolfe dual principal eigenaxis components:%
\[
\left\{  \psi_{1i\ast}\frac{\mathbf{x}_{1_{i\ast}}}{\left\Vert \mathbf{x}%
_{1_{i\ast}}\right\Vert }\right\}  _{i=1}^{l_{1}}\text{ and }\left\{
\psi_{2i\ast}\frac{\mathbf{x}_{2_{i\ast}}}{\left\Vert \mathbf{x}_{2_{i\ast}%
}\right\Vert }\right\}  _{i=1}^{l_{2}}%
\]
on $\mathbf{\psi}$, which is constrained to satisfy the equation of
statistical equilibrium:%
\[
\sum\nolimits_{i=1}^{l_{1}}\psi_{1i\ast}\frac{\mathbf{x}_{1_{i\ast}}%
}{\left\Vert \mathbf{x}_{1_{i\ast}}\right\Vert }\rightleftharpoons
\sum\nolimits_{i=1}^{l_{2}}\psi_{2i\ast}\frac{\mathbf{x}_{2_{i\ast}}%
}{\left\Vert \mathbf{x}_{2_{i\ast}}\right\Vert }\text{,}%
\]
such that the likelihood ratio $\widehat{\Lambda}_{\boldsymbol{\tau}}\left(
\mathbf{x}\right)  =\boldsymbol{\tau}_{1}-\boldsymbol{\tau}_{2}$ and the
classification system $\boldsymbol{\tau}^{T}\mathbf{x}+\tau_{0}\overset{\omega
_{1}}{\underset{\omega_{2}}{\gtrless}}0$ are in statistical equilibrium, and
the risk $\mathfrak{R}_{\mathfrak{\min}}\left(  Z|\boldsymbol{\tau}\right)  $
and the corresponding total allowed eigenenergy $\left\Vert \boldsymbol{\tau
}_{1}-\boldsymbol{\tau}_{2}\right\Vert _{\min_{c}}^{2}$ exhibited by the
classification system $\boldsymbol{\tau}^{T}\mathbf{x}+\tau_{0}\overset{\omega
_{1}}{\underset{\omega_{2}}{\gtrless}}0$ are minimized.

In the next section, I\ will explicitly define the manner in which
constrained, linear eigenlocus discriminant functions $\widetilde{\Lambda
}_{\boldsymbol{\tau}}\left(  \mathbf{x}\right)  =\boldsymbol{\tau}%
^{T}\mathbf{x}+\tau_{0}$ satisfy linear decision boundaries $D_{0}\left(
\mathbf{x}\right)  $ and linear decision borders $D_{+1}\left(  \mathbf{x}%
\right)  $ and $D_{-1}\left(  \mathbf{x}\right)  $. I will use these results
to show that the principal eigenaxis $\boldsymbol{\tau}$ of a linear
eigenlocus discriminant function $\widetilde{\Lambda}_{\boldsymbol{\tau}%
}\left(  \mathbf{x}\right)  =\boldsymbol{\tau}^{T}\mathbf{x}+\tau_{0}$ is a
lever that is symmetrically balanced with respect to the center of eigenenergy
$\left\Vert \boldsymbol{\tau}_{1}-\boldsymbol{\tau}_{2}\right\Vert _{\min_{c}%
}^{2}$ of $\boldsymbol{\tau}$, such that the total allowed eigenenergies
exhibited by the scaled extreme vectors on $\boldsymbol{\tau}_{1}%
-\boldsymbol{\tau}_{2}$ are symmetrically balanced about the fulcrum
$\left\Vert \boldsymbol{\tau}\right\Vert _{\min_{c}}^{2}$ of $\boldsymbol{\tau
}$. Thereby, I\ will show that the likelihood ratio $\widehat{\Lambda
}_{\boldsymbol{\tau}}\left(  \mathbf{x}\right)  =\boldsymbol{\tau}%
_{1}-\boldsymbol{\tau}_{2}$ and the classification system $\boldsymbol{\tau
}^{T}\mathbf{x}+\tau_{0}\overset{\omega_{1}}{\underset{\omega_{2}}{\gtrless}%
}0$ are in statistical equilibrium.

I\ will use all of these results to identify the manner in which the property
of symmetrical balance exhibited by the principal eigenaxis components on
$\boldsymbol{\psi}$ and $\boldsymbol{\tau}$ enables linear eigenlocus
classification systems $\boldsymbol{\tau}^{T}\mathbf{x}+\tau_{0}%
\overset{\omega_{1}}{\underset{\omega_{2}}{\gtrless}}0$ to effectively balance
all of the forces associated with the counter risk $\overline{\mathfrak{R}%
}_{\mathfrak{\min}}\left(  Z_{1}|\boldsymbol{\tau}_{1}\right)  $ and the risk
$\mathfrak{R}_{\mathfrak{\min}}\left(  Z_{1}|\boldsymbol{\tau}_{2}\right)  $
in the $Z_{1}$ decision region with all of the forces associated with the
counter risk $\overline{\mathfrak{R}}_{\mathfrak{\min}}\left(  Z_{2}%
|\boldsymbol{\tau}_{2}\right)  $ and the risk $\mathfrak{R}_{\mathfrak{\min}%
}\left(  Z_{2}|\boldsymbol{\tau}_{1}\right)  $ in the $Z_{2}$ decision region:%
\begin{align*}
f\left(  \widetilde{\Lambda}_{\boldsymbol{\tau}}\left(  \mathbf{x}\right)
\right)   &  :\int\nolimits_{Z_{1}}p\left(  \mathbf{x}_{1_{i\ast}%
}|\boldsymbol{\tau}_{1}\right)  d\boldsymbol{\tau}_{1}-\int\nolimits_{Z_{1}%
}p\left(  \mathbf{x}_{2_{i\ast}}|\boldsymbol{\tau}_{2}\right)
d\boldsymbol{\tau}_{2}+\delta\left(  y\right)  \boldsymbol{\psi}_{1}\\
&  =\int\nolimits_{Z_{2}}p\left(  \mathbf{x}_{2_{i\ast}}|\boldsymbol{\tau}%
_{2}\right)  d\boldsymbol{\tau}_{2}-\int\nolimits_{Z_{2}}p\left(
\mathbf{x}_{1_{i\ast}}|\boldsymbol{\tau}_{1}\right)  d\boldsymbol{\tau}%
_{1}-\delta\left(  y\right)  \boldsymbol{\psi}_{2}\text{,}%
\end{align*}
where $\delta\left(  y\right)  \triangleq\sum\nolimits_{i=1}^{l}y_{i}\left(
1-\xi_{i}\right)  $, and $Z_{1}$ and $Z_{2}$ are congruent decision regions
$Z_{1}\cong Z_{2}$, given the equilibrium point $\boldsymbol{\psi}%
_{1}-\boldsymbol{\psi}_{2}=0$ and the class-conditional probability density
functions $p\left(  \mathbf{x}_{1_{i\ast}}|\boldsymbol{\tau}_{1}\right)  $ and
$p\left(  \mathbf{x}_{2_{i_{\ast}}}|\boldsymbol{\tau}_{2}\right)  $, where the
areas under the probability density functions $p\left(  \mathbf{x}_{1_{i\ast}%
}|\boldsymbol{\tau}_{1}\right)  $ and $p\left(  \mathbf{x}_{2_{i_{\ast}}%
}|\boldsymbol{\tau}_{2}\right)  $ are symmetrically balanced with each other
over the $Z_{1}$ and $Z_{2}$ decision regions.

\section{Risk Minimization for Linear Classifiers}

In the next two sections, I will show that the conditional probability
function $P\left(  \mathbf{x}_{1_{i\ast}}|\boldsymbol{\tau}_{1}\right)  $ for
class $\omega_{1}$, which is given by the integral%
\[
P\left(  \mathbf{x}_{1_{i\ast}}|\boldsymbol{\tau}_{1}\right)  =\int%
_{Z}\boldsymbol{\tau}_{1}d\boldsymbol{\tau}_{1}=\left\Vert \boldsymbol{\tau
}_{1}\right\Vert _{\min_{c}}^{2}+C_{1}\text{,}%
\]
over the decision space $Z$, and the conditional probability function
$P\left(  \mathbf{x}_{2_{i\ast}}|\boldsymbol{\tau}_{2}\right)  $ for class
$\omega_{2}$, which is given by the integral%
\[
P\left(  \mathbf{x}_{2_{i\ast}}|\boldsymbol{\tau}_{2}\right)  =\int%
_{Z}\boldsymbol{\tau}_{2}d\boldsymbol{\tau}_{2}=\left\Vert \boldsymbol{\tau
}_{2}\right\Vert _{\min_{c}}^{2}+C_{2}\text{,}%
\]
over the decision space $Z$, satisfy an integral equation where the area under
the probability density function $p\left(  \mathbf{x}_{1_{i\ast}%
}|\boldsymbol{\tau}_{1}\right)  $ for class $\omega_{1}$ is
\emph{symmetrically balanced with} the area under the probability density
function $p\left(  \mathbf{x}_{2_{i_{\ast}}}|\boldsymbol{\tau}_{2}\right)  $
for class $\omega_{2}$%
\[
f\left(  \widetilde{\Lambda}_{\boldsymbol{\tau}}\left(  \mathbf{x}\right)
\right)  :\int_{Z}\boldsymbol{\tau}_{1}d\boldsymbol{\tau}_{1}+\nabla
_{eq}\equiv\int_{Z}\boldsymbol{\tau}_{2}d\boldsymbol{\tau}_{2}-\nabla
_{eq}\text{,}%
\]
where $\nabla_{eq}$ is an equalizer statistic, such that the likelihood ratio
$\widehat{\Lambda}_{\boldsymbol{\tau}}\left(  \mathbf{x}\right)
=\boldsymbol{\tau}_{1}-\boldsymbol{\tau}_{2}$ and the classification system
$\boldsymbol{\tau}^{T}\mathbf{x}+\tau_{0}\overset{\omega_{1}}{\underset{\omega
_{2}}{\gtrless}}0$ are in statistical equilibrium.

Accordingly, I\ will formulate a system of data-driven, locus equations that
determines the total allowed eigenenergies $\left\Vert \boldsymbol{\tau}%
_{1}\right\Vert _{\min_{c}}^{2}$ and $\left\Vert \boldsymbol{\tau}%
_{2}\right\Vert _{\min_{c}}^{2}$ exhibited by $\boldsymbol{\tau}_{1}$ and
$\boldsymbol{\tau}_{2}$, and I will derive values for the integration
constants $C_{1}$ and $C_{2}$. I will use these results to devise an equalizer
statistic $\nabla_{eq}$ for an integral equation that is satisfied by the
class-conditional probability density functions $p\left(  \mathbf{x}%
_{1_{i\ast}}|\boldsymbol{\tau}_{1}\right)  $ and $p\left(  \mathbf{x}%
_{2_{i_{\ast}}}|\boldsymbol{\tau}_{2}\right)  $.

I\ will now devise a system of data-driven, locus equations that determines
the manner in which the total allowed eigenenergies of the scaled extreme
points on $\boldsymbol{\tau}_{1}-\boldsymbol{\tau}_{2}$ are symmetrically
balanced about the fulcrum $\left\Vert \boldsymbol{\tau}\right\Vert _{\min
_{c}}^{2}$ of $\boldsymbol{\tau}$. Accordingly, I will devise three systems of
data-driven, locus equations that explicitly determine the total allowed
eigenenergy $\left\Vert \boldsymbol{\tau}_{1}\right\Vert _{\min_{c}}^{2}$
exhibited by $\boldsymbol{\tau}_{1}$, the total allowed eigenenergy
$\left\Vert \boldsymbol{\tau}_{2}\right\Vert _{\min_{c}}^{2}$ exhibited by
$\boldsymbol{\tau}_{2}$, and the total allowed eigenenergy $\left\Vert
\boldsymbol{\tau}\right\Vert _{\min_{c}}^{2}$ exhibited by $\boldsymbol{\tau}$.

\subsection{Critical Minimum Eigenenergy Constraints I}

Let there be $l$ labeled, scaled extreme points on a linear eigenlocus
$\boldsymbol{\tau}$. Given the theorem of Karush, Kuhn, and Tucker and the KKT
condition in Eq. (\ref{KKTE5}), it follows that a Wolf dual linear eigenlocus
$\boldsymbol{\psi}$ exists, for which%
\[
\left\{  \psi_{i\ast}>0\right\}  _{i=1}^{l}\text{,}%
\]
such that the $l$ constrained, primal principal eigenaxis components $\left\{
\psi_{i_{\ast}}\mathbf{x}_{i_{\ast}}\right\}  _{i=1}^{l}$ on $\boldsymbol{\tau
}$ satisfy a system of $l$ eigenlocus equations:%
\begin{equation}
\psi_{i_{\ast}}\left[  y_{i}\left(  \mathbf{x}_{i_{\ast}}^{T}\boldsymbol{\tau
}+\tau_{0}\right)  -1+\xi_{i}\right]  =0,\ i=1,...,l\text{.}
\label{Minimum Eigenenergy Functional System}%
\end{equation}

I\ will now use Eq. (\ref{Minimum Eigenenergy Functional System}) to define
critical minimum eigenenergy constraints on $\boldsymbol{\tau}_{1}$ and
$\boldsymbol{\tau}_{2}$. The analysis begins with the critical minimum
eigenenergy constraint on $\boldsymbol{\tau}_{1}$.

\subsubsection{Total Allowed Eigenenergy of $\boldsymbol{\tau}_{1}$}

Take any scaled extreme vector $\psi_{1_{i_{\ast}}}\mathbf{x}_{1_{i_{\ast}}}$
that belongs to class $\omega_{1}$. Using Eq.
(\ref{Minimum Eigenenergy Functional System}) and letting $y_{i}=+1$, it
follows that the constrained, primal principal eigenaxis component
$\psi_{1_{i_{\ast}}}\mathbf{x}_{1_{i_{\ast}}}$ on $\boldsymbol{\tau}_{1}$ is
specified by the equation:%
\[
\psi_{1_{i_{\ast}}}\mathbf{x}_{1_{i_{\ast}}}^{T}\boldsymbol{\tau}%
=\psi_{1_{i_{\ast}}}\left(  1-\xi_{i}-\tau_{0}\right)
\]
which is part of a system of $l_{1}$ eigenlocus equations. Therefore, each
constrained, primal principal eigenaxis component $\psi_{1_{i_{\ast}}%
}\mathbf{x}_{1_{i_{\ast}}}$ on $\boldsymbol{\tau}_{1}$ satisfies the above
locus equation.

Now take all of the $l_{1}$ scaled extreme vectors $\left\{  \psi_{1_{i_{\ast
}}}\mathbf{x}_{1_{i_{\ast}}}\right\}  _{i=1}^{l_{1}}$ that belong to class
$\omega_{1}$. Again, using Eq. (\ref{Minimum Eigenenergy Functional System})
and letting $y_{i}=+1$, it follows that the complete set $\left\{
\psi_{1_{i_{\ast}}}\mathbf{x}_{1_{i_{\ast}}}\right\}  _{i=1}^{l_{1}}$ of
$l_{1}$ constrained, primal principal eigenaxis components $\psi_{1_{i_{\ast}%
}}\mathbf{x}_{1_{i_{\ast}}}$ on $\boldsymbol{\tau}_{1}$ is determined by the
system of $l_{1}$ eigenlocus equations:%
\begin{equation}
\psi_{1_{i_{\ast}}}\mathbf{x}_{1_{i_{\ast}}}^{T}\boldsymbol{\tau}%
=\psi_{1_{i_{\ast}}}\left(  1-\xi_{i}-\tau_{0}\right)  ,\ i=1,...,l_{1}%
\text{.} \label{Minimum Eigenenergy Class One}%
\end{equation}

Using Eq. (\ref{Minimum Eigenenergy Class One}), it follows that the entire
set $\left\{  \psi_{1_{i_{\ast}}}\mathbf{x}_{1_{i_{\ast}}}\right\}
_{i=1}^{l_{1}}$ of $l_{1}\times d$ transformed, extreme vector coordinates
satisfies the system of $l_{1}$ eigenlocus equations:%
\[
\text{ }(1)\text{ \ }\psi_{1_{1_{\ast}}}\mathbf{x}_{1_{1_{\ast}}}%
^{T}\boldsymbol{\tau}=\psi_{1_{1_{\ast}}}\left(  1-\xi_{i}-\tau_{0}\right)
\text{,}%
\]%
\[
\text{ }(2)\text{ \ }\psi_{1_{2_{\ast}}}\mathbf{x}_{1_{2_{\ast}}}%
^{T}\boldsymbol{\tau}=\psi_{1_{2_{\ast}}}\left(  1-\xi_{i}-\tau_{0}\right)
\text{,}%
\]

\[
\vdots
\]%
\[
(l_{1})\text{\ \ }\psi_{1_{l_{\ast}}}\mathbf{x}_{1_{l_{\ast}}}^{T}%
\boldsymbol{\tau}=\psi_{1_{l_{\ast}}}\left(  1-\xi_{i}-\tau_{0}\right)
\text{,}%
\]
where each constrained, primal principal eigenaxis component $\psi
_{1_{i_{\ast}}}\mathbf{x}_{1_{i_{\ast}}}$ on $\boldsymbol{\tau}_{1}$ satisfies
the identity:%
\[
\psi_{1_{i\ast}}\mathbf{x}_{1_{i\ast}}^{T}\boldsymbol{\tau}\equiv
\psi_{1_{i_{\ast}}}\left(  1-\xi_{i}-\tau_{0}\right)  \text{.}%
\]

I\ will now derive an identity for the total allowed eigenenergy of
$\boldsymbol{\tau}_{1}$. Let $E_{\boldsymbol{\tau}_{1}}$ denote the functional
of the total allowed eigenenergy $\left\Vert \boldsymbol{\tau}_{1}\right\Vert
_{\min_{c}}^{2}$ of $\boldsymbol{\tau}_{1}$ and let $\boldsymbol{\tau=\tau
}_{1}-\boldsymbol{\tau}_{2}$. Summation over the above system of $l_{1}$
eigenlocus equations produces the following equation for the total allowed
eigenenergy $\left\Vert \boldsymbol{\tau}_{1}\right\Vert _{\min_{c}}^{2}$ of
$\boldsymbol{\tau}_{1}$:%
\[
\left(  \sum\nolimits_{i=1}^{l_{1}}\psi_{1_{i_{\ast}}}\mathbf{x}_{1_{i_{\ast}%
}}^{T}\right)  \left(  \boldsymbol{\tau}_{1}-\boldsymbol{\tau}_{2}\right)
\equiv\sum\nolimits_{i=1}^{l_{1}}\psi_{1_{i_{\ast}}}\left(  1-\xi_{i}-\tau
_{0}\right)
\]
which reduces to%
\[
\boldsymbol{\tau}_{1}^{T}\boldsymbol{\tau}_{1}-\boldsymbol{\tau}_{1}%
^{T}\boldsymbol{\tau}_{2}\equiv\sum\nolimits_{i=1}^{l_{1}}\psi_{1_{i_{\ast}}%
}\left(  1-\xi_{i}-\tau_{0}\right)
\]
so that the functional $E_{\boldsymbol{\tau}_{1}}$ satisfies the identity%
\[
\left\Vert \boldsymbol{\tau}_{1}\right\Vert _{\min_{c}}^{2}-\boldsymbol{\tau
}_{1}^{T}\boldsymbol{\tau}_{2}\equiv\sum\nolimits_{i=1}^{l_{1}}\psi
_{1_{i_{\ast}}}\left(  1-\xi_{i}-\tau_{0}\right)  \text{.}%
\]

Therefore, the total allowed eigenenergy $\left\Vert \boldsymbol{\tau}%
_{1}\right\Vert _{\min_{c}}^{2}$ exhibited by the constrained, primal
principal eigenlocus component $\boldsymbol{\tau}_{1}$ is determined by the
identity%
\begin{equation}
\left\Vert \boldsymbol{\tau}_{1}\right\Vert _{\min_{c}}^{2}-\left\Vert
\boldsymbol{\tau}_{1}\right\Vert \left\Vert \boldsymbol{\tau}_{2}\right\Vert
\cos\theta_{\boldsymbol{\tau}_{1}\boldsymbol{\tau}_{2}}\equiv\sum
\nolimits_{i=1}^{l_{1}}\psi_{1_{i_{\ast}}}\left(  1-\xi_{i}-\tau_{0}\right)
\text{,} \label{TAE Eigenlocus Component One}%
\end{equation}
where the functional $E_{\boldsymbol{\tau}_{1}}$ of the total allowed
eigenenergy $\left\Vert \boldsymbol{\tau}_{1}\right\Vert _{\min_{c}}^{2}$
exhibited by $\boldsymbol{\tau}_{1}$%
\[
E_{\boldsymbol{\tau}_{1}}=\left\Vert \boldsymbol{\tau}_{1}\right\Vert
_{\min_{c}}^{2}-\left\Vert \boldsymbol{\tau}_{1}\right\Vert \left\Vert
\boldsymbol{\tau}_{2}\right\Vert \cos\theta_{\boldsymbol{\tau}_{1}%
\boldsymbol{\tau}_{2}}%
\]
is equivalent to the functional $E_{\boldsymbol{\psi}_{1}}$%
\[
E_{\boldsymbol{\psi}_{1}}=\sum\nolimits_{i=1}^{l_{1}}\psi_{1_{i_{\ast}}%
}\left(  1-\xi_{i}-\tau_{0}\right)
\]
of the integrated magnitudes $\sum\nolimits_{i=1}^{l_{1}}\psi_{1_{i_{\ast}}}$
of the Wolfe dual principal eigenaxis components $\psi_{1_{i_{\ast}}}%
\frac{\mathbf{x}_{1_{i_{\ast}}}}{\left\Vert \mathbf{x}_{1_{i_{\ast}}%
}\right\Vert }$ and the $\tau_{0}$ statistic.

Returning to Eq. (\ref{Decision Border One}), it follows that the functionals
$E_{\boldsymbol{\tau}_{1}}$ and $E_{\boldsymbol{\psi}_{1}}$ specify the manner
in which linear eigenlocus discriminant functions $\widetilde{\Lambda
}_{\boldsymbol{\tau}}\left(  \mathbf{x}\right)  =\mathbf{x}^{T}%
\boldsymbol{\tau}+\tau_{0}$ satisfy the linear decision border $D_{+1}\left(
\mathbf{x}\right)  $: $\boldsymbol{\tau}^{T}\mathbf{x}+\tau_{0}=1$.

Given Eq. (\ref{TAE Eigenlocus Component One}), it is concluded that a linear
eigenlocus discriminant function $\widetilde{\Lambda}_{\boldsymbol{\tau}%
}\left(  \mathbf{x}\right)  =\mathbf{x}^{T}\boldsymbol{\tau}+\tau_{0}$
satisfies the linear decision border $D_{+1}\left(  \mathbf{x}\right)  $ in
terms of the total allowed eigenenergy $\left\Vert \boldsymbol{\tau}%
_{1}\right\Vert _{\min_{c}}^{2}$ exhibited by $\boldsymbol{\tau}_{1}$, where
the functional $\left\Vert \boldsymbol{\tau}_{1}\right\Vert _{\min_{c}}%
^{2}-\left\Vert \boldsymbol{\tau}_{1}\right\Vert \left\Vert \boldsymbol{\tau
}_{2}\right\Vert \cos\theta_{\boldsymbol{\tau}_{1}\boldsymbol{\tau}_{2}}$ is
constrained by the functional $\sum\nolimits_{i=1}^{l_{1}}\psi_{1_{i_{\ast}}%
}\left(  1-\xi_{i}-\tau_{0}\right)  $.

The critical minimum eigenenergy constraint on $\boldsymbol{\tau}_{2}$ is
examined next.

\subsubsection{Total Allowed Eigenenergy of $\boldsymbol{\tau}_{2}$}

Take any scaled extreme vector $\psi_{2_{i_{\ast}}}\mathbf{x}_{2_{i_{\ast}}}$
that belongs to class $\omega_{2}$. Using Eq.
(\ref{Minimum Eigenenergy Functional System}) and letting $y_{i}=-1$, it
follows that the constrained, primal principal eigenaxis component
$\psi_{2_{i_{\ast}}}\mathbf{x}_{2_{i_{\ast}}}$ on $\boldsymbol{\tau}_{2}$ is
specified by the equation:%
\[
-\psi_{2_{i_{\ast}}}\mathbf{x}_{2_{i_{\ast}}}^{T}\boldsymbol{\tau}%
=\psi_{2_{i_{\ast}}}\left(  1-\xi_{i}+\tau_{0}\right)
\]
which is part of a system of $l_{2}$ eigenlocus equations. Therefore, each
constrained, primal principal eigenaxis component $\psi_{2_{i_{\ast}}%
}\mathbf{x}_{2_{i_{\ast}}}$ on $\boldsymbol{\tau}_{2}$ satisfies the above
locus equation.

Now take all of the $l_{2}$ scaled, extreme vectors $\left\{  \psi
_{2_{i_{\ast}}}\mathbf{x}_{2_{i_{\ast}}}\right\}  _{i=1}^{l_{2}}$ that belong
to class $\omega_{2}$. Again, using Eq.
(\ref{Minimum Eigenenergy Functional System}) and letting $y_{i}=-1$, it
follows that the complete set $\left\{  \psi_{2_{i_{\ast}}}\mathbf{x}%
_{2_{i_{\ast}}}\right\}  _{i=1}^{l_{2}}$ of $l_{2}$ constrained, primal
principal eigenaxis components $\psi_{2_{i_{\ast}}}\mathbf{x}_{2_{i_{\ast}}}$
on $\boldsymbol{\tau}_{2}$ is determined by the system of $l_{2}$ eigenlocus
equations:%
\begin{equation}
-\psi_{2_{i_{\ast}}}\mathbf{x}_{2_{i_{\ast}}}^{T}\boldsymbol{\tau}%
=\psi_{2_{i_{\ast}}}\left(  1-\xi_{i}+\tau_{0}\right)  ,\ i=1,...,l_{2}%
\text{.} \label{Minimum Eigenenergy Class Two}%
\end{equation}

Using Eq. (\ref{Minimum Eigenenergy Class Two}), it follows that the entire
set $\left\{  \psi_{2_{i_{\ast}}}\mathbf{x}_{2_{i_{\ast}}}\right\}
_{i=1}^{l_{2}}$ of $l_{2}\times d$ transformed, extreme vector coordinates
satisfies the system of $l_{2}$ eigenlocus equations:%
\[
\text{ }(1)\text{ \ }-\psi_{2_{1_{\ast}}}\mathbf{x}_{2_{1_{\ast}}}%
^{T}\boldsymbol{\tau}=\psi_{2_{1_{\ast}}}\left(  1-\xi_{i}+\tau_{0}\right)
\text{,}%
\]%
\[
(2)\text{ \ }-\psi_{2_{2_{\ast}}}\mathbf{x}_{2_{2_{\ast}}}^{T}\boldsymbol{\tau
}=\psi_{2_{2_{\ast}}}\left(  1-\xi_{i}+\tau_{0}\right)  \text{,}%
\]%
\[
\vdots
\]%
\[
(l_{2})\ -\psi_{2_{l_{2}\ast}}\mathbf{x}_{2_{_{l_{2}\ast}}}^{T}%
\boldsymbol{\tau}=\psi_{2_{l_{2}\ast}}\left(  1-\xi_{i}+\tau_{0}\right)
\text{,}%
\]
where each constrained, primal principal eigenaxis component $\psi_{2_{\ast}%
}\mathbf{x}_{2_{_{\ast}}}$ on $\boldsymbol{\tau}_{2}$ satisfies the identity:%
\[
-\psi_{2_{i_{\ast}}}\mathbf{x}_{2_{i_{\ast}}}^{T}\boldsymbol{\tau}\equiv
\psi_{2_{i_{\ast}}}\left(  1-\xi_{i}+\tau_{0}\right)  \text{.}%
\]

I will now derive an identity for the total allowed eigenenergy of
$\boldsymbol{\tau}_{2}$. Let $E_{\boldsymbol{\tau}_{2}}$ denote the functional
of the total allowed eigenenergy $\left\Vert \boldsymbol{\tau}_{2}\right\Vert
_{\min_{c}}^{2}$ of $\boldsymbol{\tau}_{2}$ and let $\boldsymbol{\tau=\tau
}_{1}-\boldsymbol{\tau}_{2}$. Summation over the above system of $l_{2}$
eigenlocus equations produces the following equation for the total allowed
eigenenergy $\left\Vert \boldsymbol{\tau}_{2}\right\Vert _{\min_{c}}^{2}$ of
$\boldsymbol{\tau}_{2}$:%
\[
-\left(  \sum\nolimits_{i=1}^{l_{2}}\psi_{2_{i_{\ast}}}\mathbf{x}_{2_{i_{\ast
}}}^{T}\right)  \left(  \boldsymbol{\tau}_{1}-\boldsymbol{\tau}_{2}\right)
\equiv\sum\nolimits_{i=1}^{l_{2}}\psi_{2_{i_{\ast}}}\left(  1-\xi_{i}+\tau
_{0}\right)
\]
which reduces to%
\[
\boldsymbol{\tau}_{2}^{T}\boldsymbol{\tau}_{2}-\boldsymbol{\tau}_{2}%
^{T}\boldsymbol{\tau}_{1}\equiv\sum\nolimits_{i=1}^{l_{2}}\psi_{2_{i_{\ast}}%
}\left(  1-\xi_{i}+\tau_{0}\right)
\]
so that the functional $E_{\boldsymbol{\tau}_{2}}$ satisfies the identity%
\[
\left\Vert \boldsymbol{\tau}_{2}\right\Vert _{\min_{c}}^{2}-\boldsymbol{\tau
}_{2}^{T}\boldsymbol{\tau}_{1}\equiv\sum\nolimits_{i=1}^{l_{2}}\psi
_{2_{i_{\ast}}}\left(  1-\xi_{i}+\tau_{0}\right)  \text{.}%
\]

Therefore, the total allowed eigenenergy $\left\Vert \boldsymbol{\tau}%
_{2}\right\Vert _{\min_{c}}^{2}$ exhibited by the constrained, primal
eigenlocus component $\boldsymbol{\tau}_{2}$ is determined by the identity%
\begin{equation}
\left\Vert \boldsymbol{\tau}_{2}\right\Vert _{\min_{c}}^{2}-\left\Vert
\boldsymbol{\tau}_{2}\right\Vert \left\Vert \boldsymbol{\tau}_{1}\right\Vert
\cos\theta_{\boldsymbol{\tau}_{2}\boldsymbol{\tau}_{1}}\equiv\sum
\nolimits_{i=1}^{l_{2}}\psi_{2_{i_{\ast}}}\left(  1-\xi_{i}+\tau_{0}\right)
\text{,} \label{TAE Eigenlocus Component Two}%
\end{equation}
where the functional $E_{\boldsymbol{\tau}_{2}}$ of the total allowed
eigenenergy $\left\Vert \boldsymbol{\tau}_{2}\right\Vert _{\min_{c}}^{2}$
exhibited by $\boldsymbol{\tau}_{2}$%
\[
E_{\boldsymbol{\tau}_{2}}=\left\Vert \boldsymbol{\tau}_{2}\right\Vert
_{\min_{c}}^{2}-\left\Vert \boldsymbol{\tau}_{2}\right\Vert \left\Vert
\boldsymbol{\tau}_{1}\right\Vert \cos\theta_{\boldsymbol{\tau}_{2}%
\boldsymbol{\tau}_{1}}%
\]
is equivalent to the functional $E_{\boldsymbol{\psi}_{2}}$%
\[
E_{\boldsymbol{\psi}_{2}}=\sum\nolimits_{i=1}^{l_{2}}\psi_{2_{i_{\ast}}%
}\left(  1-\xi_{i}+\tau_{0}\right)
\]
of the integrated magnitudes $\sum\nolimits_{i=1}^{l_{2}}\psi_{2_{i_{\ast}}}$
of the Wolfe dual principal eigenaxis components $\psi_{2_{i_{\ast}}}%
\frac{\mathbf{x}_{2_{i_{\ast}}}}{\left\Vert \mathbf{x}_{2_{i_{\ast}}%
}\right\Vert }$ and the $\tau_{0}$ statistic.

Returning to Eq. (\ref{Decision Border Two}), it follows the functionals
$E_{\boldsymbol{\tau}_{2}}$ and $E_{\boldsymbol{\psi}_{2}}$ specify the manner
in which linear eigenlocus discriminant functions $\widetilde{\Lambda
}_{\boldsymbol{\tau}}\left(  \mathbf{x}\right)  =\mathbf{x}^{T}%
\boldsymbol{\tau}+\tau_{0}$ satisfy the linear decision border $D_{-1}\left(
\mathbf{x}\right)  $: $\boldsymbol{\tau}^{T}\mathbf{x}+\tau_{0}=-1$.

Given Eq. (\ref{TAE Eigenlocus Component Two}), it is concluded that a linear
eigenlocus discriminant function $\widetilde{\Lambda}_{\boldsymbol{\tau}%
}\left(  \mathbf{x}\right)  =\mathbf{x}^{T}\boldsymbol{\tau}+\tau_{0}$
satisfies the linear decision border $D_{-1}\left(  \mathbf{x}\right)  $ in
terms of the total allowed eigenenergy $\left\Vert \boldsymbol{\tau}%
_{2}\right\Vert _{\min_{c}}^{2}$ exhibited by $\boldsymbol{\tau}_{2}$, where
the functional $\left\Vert \boldsymbol{\tau}_{2}\right\Vert _{\min_{c}}%
^{2}-\left\Vert \boldsymbol{\tau}_{2}\right\Vert \left\Vert \boldsymbol{\tau
}_{1}\right\Vert \cos\theta_{\boldsymbol{\tau}_{2}\boldsymbol{\tau}_{1}}$ is
constrained by the functional $\sum\nolimits_{i=1}^{l_{2}}\psi_{2_{i_{\ast}}%
}\left(  1-\xi_{i}+\tau_{0}\right)  $.

The critical minimum eigenenergy constraint on $\boldsymbol{\tau}$ is examined next.

\subsubsection{Total Allowed Eigenenergy of $\boldsymbol{\tau}$}

I\ will now derive an identity for the total allowed eigenenergy of a
constrained, primal linear eigenlocus $\boldsymbol{\tau}$. Let
$E_{\boldsymbol{\tau}}$ denote the functional of the total allowed eigenenergy
$\left\Vert \boldsymbol{\tau}\right\Vert _{\min_{c}}^{2}$ of $\boldsymbol{\tau
}$.

Summation over the complete system of eigenlocus equations satisfied by
$\boldsymbol{\tau}_{1}$%
\[
\left(  \sum\nolimits_{i=1}^{l_{1}}\psi_{1_{i_{\ast}}}\mathbf{x}_{1_{i_{\ast}%
}}^{T}\right)  \boldsymbol{\tau}\equiv\sum\nolimits_{i=1}^{l_{1}}%
\psi_{1_{i_{\ast}}}\left(  1-\xi_{i}-\tau_{0}\right)
\]
and by $\boldsymbol{\tau}_{2}$%
\[
\left(  -\sum\nolimits_{i=1}^{l_{2}}\psi_{2_{i_{\ast}}}\mathbf{x}_{2_{i\ast}%
}^{T}\right)  \boldsymbol{\tau}\equiv\sum\nolimits_{i=1}^{l_{2}}%
\psi_{2_{i_{\ast}}}\left(  1-\xi_{i}+\tau_{0}\right)
\]
produces the following identity for the functional $E_{\boldsymbol{\tau}}$ of
the total allowed eigenenergy $\left\Vert \boldsymbol{\tau}\right\Vert
_{\min_{c}}^{2}$ of $\boldsymbol{\tau}$%
\begin{align*}
\left\Vert \boldsymbol{\tau}\right\Vert _{\min_{c}}^{2}  &  :\left(
\sum\nolimits_{i=1}^{l_{1}}\psi_{1_{i_{\ast}}}\mathbf{x}_{1_{i_{\ast}}}%
^{T}-\sum\nolimits_{i=1}^{l_{2}}\psi_{2_{i_{\ast}}}\mathbf{x}_{2_{i\ast}}%
^{T}\right)  \boldsymbol{\tau}\\
&  \equiv\sum\nolimits_{i=1}^{l_{1}}\psi_{1_{i_{\ast}}}\left(  1-\xi_{i}%
-\tau_{0}\right)  +\sum\nolimits_{i=1}^{l_{2}}\psi_{2_{i_{\ast}}}\left(
1-\xi_{i}+\tau_{0}\right)
\end{align*}
which reduces to%
\begin{align}
\left(  \boldsymbol{\tau}_{1}-\boldsymbol{\tau}_{2}\right)  ^{T}%
\boldsymbol{\tau}  &  \equiv\sum\nolimits_{i=1}^{l_{1}}\psi_{1i\ast}\left(
1-\xi_{i}-\tau_{0}\right) \label{Symmetrical Balance of TAE SDE}\\
&  +\sum\nolimits_{i=1}^{l_{2}}\psi_{2_{i_{\ast}}}\left(  1-\xi_{i}+\tau
_{0}\right) \nonumber\\
&  \equiv\sum\nolimits_{i=1}^{l}\psi_{i_{\ast}}\left(  1-\xi_{i}\right)
\text{,}\nonumber
\end{align}
where I\ have used the equilibrium constraint on $\boldsymbol{\psi}$ in Eq.
(\ref{Equilibrium Constraint on Dual Eigen-components}).

Thereby, the functional $E_{\boldsymbol{\tau}}$ of the total allowed
eigenenergy $\left\Vert \boldsymbol{\tau}\right\Vert _{\min_{c}}^{2}$
exhibited by $\boldsymbol{\tau}$%
\begin{align*}
E_{\boldsymbol{\tau}}  &  =\left(  \boldsymbol{\tau}_{1}-\boldsymbol{\tau}%
_{2}\right)  ^{T}\boldsymbol{\tau}\\
&  \mathbf{=}\left\Vert \boldsymbol{\tau}\right\Vert _{\min_{c}}^{2}%
\end{align*}
is equivalent to the functional $E_{\boldsymbol{\psi}}$%
\[
E_{\boldsymbol{\psi}}=\sum\nolimits_{i=1}^{l}\psi_{i_{\ast}}\left(  1-\xi
_{i}\right)
\]
solely in terms of the integrated magnitudes $\sum\nolimits_{i=1}^{l}%
\psi_{i_{\ast}}$ of the Wolfe dual principal eigenaxis components on
$\boldsymbol{\psi}$.

Thus, the total allowed eigenenergy $\left\Vert \boldsymbol{\tau}\right\Vert
_{\min_{c}}^{2}$ exhibited by a constrained, primal linear eigenlocus
$\boldsymbol{\tau}$ is specified by the integrated magnitudes $\psi_{i_{\ast}%
}$ of the Wolfe dual principal eigenaxis components $\psi_{i\ast}%
\frac{\mathbf{x}_{i\ast}}{\left\Vert \mathbf{x}_{i\ast}\right\Vert }$ on
$\boldsymbol{\psi}$%
\begin{align}
\left\Vert \boldsymbol{\tau}\right\Vert _{\min_{c}}^{2}  &  \equiv
\sum\nolimits_{i=1}^{l}\psi_{i_{\ast}}\left(  1-\xi_{i}\right) \label{TAE SDE}%
\\
&  \equiv\sum\nolimits_{i=1}^{l}\psi_{i_{\ast}}-\sum\nolimits_{i=1}^{l}\xi
_{i}\psi_{i_{\ast}}\text{,}\nonumber
\end{align}
where the regularization parameters $\xi_{i}=\xi\ll1$ are seen to determine
negligible constraints on $\left\Vert \boldsymbol{\tau}\right\Vert _{\min_{c}%
}^{2}$. Therefore, it is concluded that the total allowed eigenenergy
$\left\Vert \boldsymbol{\tau}\right\Vert _{\min_{c}}^{2}$ exhibited by a
constrained, primal linear eigenlocus $\boldsymbol{\tau}$ is determined by the
integrated magnitudes $\sum\nolimits_{i=1}^{l}\psi_{i_{\ast}}$ of the Wolfe
dual principal eigenaxis components $\psi_{i\ast}\frac{\mathbf{x}_{i\ast}%
}{\left\Vert \mathbf{x}_{i\ast}\right\Vert }$ on $\boldsymbol{\psi}$.

Returning to Eq. (\ref{Decision Boundary}), it follows that the equilibrium
constraint on $\boldsymbol{\psi}$ and the corresponding functionals
$E_{\boldsymbol{\tau}}$ and $E_{\boldsymbol{\psi}}$ specify the manner in
which linear eigenlocus discriminant functions $\widetilde{\Lambda
}_{\boldsymbol{\tau}}\left(  \mathbf{x}\right)  =\mathbf{x}^{T}%
\boldsymbol{\tau}+\tau_{0}$ satisfy linear decision boundaries $D_{0}\left(
\mathbf{x}\right)  $: $\boldsymbol{\tau}^{T}\mathbf{x}+\tau_{0}=0$.

Given Eq. (\ref{TAE SDE}), it is concluded that a linear eigenlocus
discriminant function $\widetilde{\Lambda}_{\boldsymbol{\tau}}\left(
\mathbf{x}\right)  =\mathbf{x}^{T}\boldsymbol{\tau}+\tau_{0}$ satisfies a
linear decision boundary $D_{0}\left(  \mathbf{x}\right)  $ in terms of its
total allowed eigenenergy $\left\Vert \boldsymbol{\tau}\right\Vert _{\min_{c}%
}^{2}$, where the functional $\left\Vert \boldsymbol{\tau}\right\Vert
_{\min_{c}}^{2}$ is constrained by the functional $\sum\nolimits_{i=1}^{l}%
\psi_{i_{\ast}}\left(  1-\xi_{i}\right)  $.

Using Eqs (\ref{TAE Eigenlocus Component One}),
(\ref{TAE Eigenlocus Component Two}), and
(\ref{Symmetrical Balance of TAE SDE}), it follows that the symmetrically
balanced constraints%
\[
E_{\boldsymbol{\psi}_{1}}=\sum\nolimits_{i=1}^{l_{1}}\psi_{1_{i_{\ast}}%
}\left(  1-\xi_{i}-\tau_{0}\right)  \text{ \ and \ }E_{\boldsymbol{\psi}_{2}%
}=\sum\nolimits_{i=1}^{l_{2}}\psi_{2_{i_{\ast}}}\left(  1-\xi_{i}+\tau
_{0}\right)
\]
satisfied by a linear eigenlocus discriminant function $\widetilde{\Lambda
}_{\boldsymbol{\tau}}\left(  \mathbf{x}\right)  =\mathbf{x}^{T}%
\boldsymbol{\tau}+\tau_{0}$ on the respective linear decision borders
$D_{+1}\left(  \mathbf{x}\right)  $ and $D_{-1}\left(  \mathbf{x}\right)  $,
and the corresponding constraint%
\[
E_{\boldsymbol{\psi}}=\sum\nolimits_{i=1}^{l_{1}}\psi_{1i\ast}\left(
1-\xi_{i}-\tau_{0}\right)  +\sum\nolimits_{i=1}^{l_{2}}\psi_{2_{i_{\ast}}%
}\left(  1-\xi_{i}+\tau_{0}\right)
\]
satisfied by a linear eigenlocus discriminant function $\widetilde{\Lambda
}_{\boldsymbol{\tau}}\left(  \mathbf{x}\right)  =\mathbf{x}^{T}%
\boldsymbol{\tau}+\tau_{0}$ on the linear decision boundary $D_{0}\left(
\mathbf{x}\right)  $, ensure that the total allowed eigenenergies $\left\Vert
\boldsymbol{\tau}_{1}-\boldsymbol{\tau}_{2}\right\Vert _{\min_{c}}^{2}$
exhibited by the scaled extreme points on $\boldsymbol{\tau}_{1}%
-\boldsymbol{\tau}_{2}$:%
\begin{align*}
\left\Vert \boldsymbol{\tau}\right\Vert _{\min_{c}}^{2}  &  =\left\Vert
\boldsymbol{\tau}_{1}\right\Vert _{\min_{c}}^{2}-\left\Vert \boldsymbol{\tau
}_{1}\right\Vert \left\Vert \boldsymbol{\tau}_{2}\right\Vert \cos
\theta_{\boldsymbol{\tau}_{1}\boldsymbol{\tau}_{2}}\\
&  +\left\Vert \boldsymbol{\tau}_{2}\right\Vert _{\min_{c}}^{2}-\left\Vert
\boldsymbol{\tau}_{2}\right\Vert \left\Vert \boldsymbol{\tau}_{1}\right\Vert
\cos\theta_{\boldsymbol{\tau}_{2}\boldsymbol{\tau}_{1}}%
\end{align*}
satisfy the law of cosines in the symmetrically balanced manner depicted in
Fig. $\ref{Law of Cosines for Binary Classification Systems}$.

Given the binary classification theorem, it follows that linear eigenlocus
likelihood ratios $\widehat{\Lambda}_{\boldsymbol{\tau}}\left(  \mathbf{x}%
\right)  =\boldsymbol{\tau}_{1}-\boldsymbol{\tau}_{2}$ and corresponding
decision boundaries $D_{0}\left(  \mathbf{x}\right)  $ satisfy an integral
equation $f\left(  \widetilde{\Lambda}_{\boldsymbol{\tau}}\left(
\mathbf{x}\right)  \right)  $ where the areas $\int\nolimits_{Z}p\left(
\mathbf{x}_{1_{i\ast}}|\boldsymbol{\tau}_{1}\right)  d\boldsymbol{\tau}_{1}$
and $\int\nolimits_{Z}p\left(  \mathbf{x}_{2_{i\ast}}|\boldsymbol{\tau}%
_{2}\right)  d\boldsymbol{\tau}_{2}$ under the class-conditional probability
density functions $p\left(  \mathbf{x}_{1_{i\ast}}|\boldsymbol{\tau}%
_{1}\right)  $ and $p\left(  \mathbf{x}_{2_{i_{\ast}}}|\boldsymbol{\tau}%
_{2}\right)  $ are balanced with each other. Furthermore, the eigenenergy
associated with the position or location of the likelihood ratio $p\left(
\widehat{\Lambda}_{\boldsymbol{\tau}}\left(  \mathbf{x}\right)  |\omega
_{2}\right)  $ given class $\omega_{2}$ must be balanced with the eigenenergy
associated with the position or location of the likelihood ratio $p\left(
\widehat{\Lambda}_{\boldsymbol{\tau}}\left(  \mathbf{x}\right)  |\omega
_{1}\right)  $ given class $\omega_{1}$. Therefore, linear eigenlocus
likelihood ratios $\widehat{\Lambda}_{\boldsymbol{\tau}}\left(  \mathbf{x}%
\right)  =\boldsymbol{\tau}_{1}-\boldsymbol{\tau}_{2}$ and corresponding
decision boundaries $D_{0}\left(  \mathbf{x}\right)  $ also satisfy an
integral equation where the total allowed eigenenergies $\left\Vert
\boldsymbol{\tau}_{1}-\boldsymbol{\tau}_{2}\right\Vert _{\min_{c}}^{2}$ of a
linear eigenlocus $\boldsymbol{\tau}=\boldsymbol{\tau}_{1}-\boldsymbol{\tau
}_{2}$ are balanced with each other.

I\ will show that the discrete, linear classification system $\boldsymbol{\tau
}^{T}\mathbf{x}+\tau_{0}\overset{\omega_{1}}{\underset{\omega_{2}}{\gtrless}%
}0$ seeks an equilibrium point $p\left(  \widehat{\Lambda}_{\boldsymbol{\psi}%
}\left(  \mathbf{x}\right)  |\omega_{1}\right)  -p\left(  \widehat{\Lambda
}_{\boldsymbol{\psi}}\left(  \mathbf{x}\right)  |\omega_{2}\right)  =0$ of an
integral equation $f\left(  \widetilde{\Lambda}_{\boldsymbol{\tau}}\left(
\mathbf{x}\right)  \right)  $, where the total allowed eigenenergies
$\left\Vert \boldsymbol{\tau}_{1}-\boldsymbol{\tau}_{2}\right\Vert _{\min_{c}%
}^{2}$ of the classification system are balanced with each other, such that
the eigenenergy and the risk of the classification system $\boldsymbol{\tau
}^{T}\mathbf{x}+\tau_{0}\overset{\omega_{1}}{\underset{\omega_{2}}{\gtrless}%
}0$ are minimized, and the classification system $\boldsymbol{\tau}%
^{T}\mathbf{x}+\tau_{0}\overset{\omega_{1}}{\underset{\omega_{2}}{\gtrless}}0$
is in statistical equilibrium.

In the next section, I will develop an integral equation $f\left(
\widetilde{\Lambda}_{\boldsymbol{\tau}}\left(  \mathbf{x}\right)  \right)  $
that is satisfied by linear eigenlocus discriminant functions
$\widetilde{\Lambda}_{\boldsymbol{\tau}}\left(  \mathbf{x}\right)
=\boldsymbol{\tau}^{T}\mathbf{x}+\tau_{0}$, where the total allowed
eigenenergies $\left\Vert \boldsymbol{\tau}_{1}-\boldsymbol{\tau}%
_{2}\right\Vert _{\min_{c}}^{2}$ exhibited by the principal eigenaxis
components on a linear eigenlocus $\boldsymbol{\tau}=\boldsymbol{\tau}%
_{1}-\boldsymbol{\tau}_{2}$ are symmetrically balanced with each other.
Thereby, I will show that the likelihood ratio $p\left(  \widehat{\Lambda
}_{\boldsymbol{\tau}}\left(  \mathbf{x}\right)  |\omega_{1}\right)  -p\left(
\widehat{\Lambda}_{\boldsymbol{\tau}}\left(  \mathbf{x}\right)  |\omega
_{2}\right)  $ is in statistical equilibrium, and that the areas under the
class-conditional probability density functions $p\left(  \mathbf{x}%
_{1_{i\ast}}|\boldsymbol{\tau}_{1}\right)  $ and $p\left(  \mathbf{x}%
_{2_{i_{\ast}}}|\boldsymbol{\tau}_{2}\right)  $, over the decision space
$Z=Z_{1}+Z_{2}$, are symmetrically balanced with each other. I will use these
results to show that the linear eigenlocus discriminant function
$\widetilde{\Lambda}_{\boldsymbol{\tau}}\left(  \mathbf{x}\right)
=\boldsymbol{\tau}^{T}\mathbf{x}+\tau_{0}$ is the solution to a fundamental
integral equation of binary classification for a linear classification system
in statistical equilibrium. The solution involves a surprising, statistical
balancing feat in decision space $Z$ which hinges on an elegant, statistical
balancing feat in eigenspace $\widetilde{Z}$.

\section{The Balancing Feat in Eigenspace I}

A linear eigenlocus%
\[
\boldsymbol{\tau}=\boldsymbol{\tau}_{1}-\boldsymbol{\tau}_{2}%
\]
which is formed by a locus of labeled ($+1$ or $-1$), scaled ($\psi_{1_{i\ast
}}$ or $\psi_{2_{i\ast}}$) extreme vectors ($\mathbf{x}_{1_{i\ast}}$ or
$\mathbf{x}_{2_{i\ast}}$)%
\[
\boldsymbol{\tau}=\sum\nolimits_{i=1}^{l_{1}}\psi_{1_{i\ast}}\mathbf{x}%
_{1_{i\ast}}-\sum\nolimits_{i=1}^{l_{2}}\psi_{2_{i\ast}}\mathbf{x}_{2_{i\ast}}%
\]
has a \emph{dual nature} that is \emph{twofold}:

Each $\psi_{1_{i\ast}}$ or $\psi_{2_{i\ast}}$ scale factor determines the
total allowed eigenenergy%
\[
\left\Vert \psi_{1_{i\ast}}\mathbf{x}_{1_{i\ast}}\right\Vert _{\min_{c}}%
^{2}\text{ \ or \ }\left\Vert \psi_{2_{i\ast}}\mathbf{x}_{2_{i\ast}%
}\right\Vert _{\min_{c}}^{2}%
\]
of a principal eigenaxis component $\psi_{1_{i\ast}}\mathbf{x}_{1_{i\ast}}$ or
$\psi_{2_{i\ast}}\mathbf{x}_{2_{i\ast}}$ on $\boldsymbol{\tau}_{1}%
-\boldsymbol{\tau}_{2}$ in decision space $Z$, and each $\psi_{1_{i\ast}}$ or
$\psi_{2_{i\ast}}$ scale factor determines the total allowed eigenenergy%
\[
\left\Vert \psi_{1_{i\ast}}\frac{\mathbf{x}_{1_{i\ast}}}{\left\Vert
\mathbf{x}_{1_{i\ast}}\right\Vert }\right\Vert _{\min_{c}}^{2}\text{ \ or
\ }\left\Vert \psi_{2_{i\ast}}\frac{\mathbf{x}_{2_{i\ast}}}{\left\Vert
\mathbf{x}_{2_{i\ast}}\right\Vert }\right\Vert _{\min_{c}}^{2}%
\]
of a principal eigenaxis component $\psi_{1_{i\ast}}\frac{\mathbf{x}%
_{1_{i\ast}}}{\left\Vert \mathbf{x}_{1_{i\ast}}\right\Vert }$ or
$\psi_{2_{i\ast}}\frac{\mathbf{x}_{2_{i\ast}}}{\left\Vert \mathbf{x}%
_{2_{i\ast}}\right\Vert }$ on $\boldsymbol{\psi}_{1}+\boldsymbol{\psi}_{2}$ in
Wolfe dual eigenspace $\widetilde{Z}$.

In addition, each $\psi_{1_{i\ast}}$ scale factor specifies dual conditional
densities for an $\mathbf{x}_{1_{i\ast}}$ extreme point:%
\[
p\left(  \mathbf{x}_{1_{i_{\ast}}}|\operatorname{comp}%
_{\overrightarrow{\mathbf{x}_{1i\ast}}}\left(
\overrightarrow{\boldsymbol{\tau}}\right)  \right)  \frac{\mathbf{x}%
_{1_{i\ast}}}{\left\Vert \mathbf{x}_{1_{i\ast}}\right\Vert }\text{ \ and
\ }p\left(  \mathbf{x}_{1_{i_{\ast}}}|\operatorname{comp}%
_{\overrightarrow{\mathbf{x}_{1i\ast}}}\left(
\overrightarrow{\boldsymbol{\tau}}\right)  \right)  \mathbf{x}_{1_{i\ast}%
}\text{,}%
\]
and each $\psi_{2_{i\ast}}$ scale factor specifies dual conditional densities
for an $\mathbf{x}_{2_{i\ast}}$ extreme point:%
\[
p\left(  \mathbf{x}_{2_{i_{\ast}}}|\operatorname{comp}%
_{\overrightarrow{\mathbf{x}_{2i\ast}}}\left(
\overrightarrow{\boldsymbol{\tau}}\right)  \right)  \frac{\mathbf{x}%
_{2_{i\ast}}}{\left\Vert \mathbf{x}_{2_{i\ast}}\right\Vert }\text{ \ and
\ }p\left(  \mathbf{x}_{2_{i_{\ast}}}|\operatorname{comp}%
_{\overrightarrow{\mathbf{x}_{2i\ast}}}\left(
\overrightarrow{\boldsymbol{\tau}}\right)  \right)  \mathbf{x}_{2_{i\ast}%
}\text{.}%
\]

Accordingly, a Wolfe dual linear eigenlocus $\boldsymbol{\psi}%
=\boldsymbol{\psi}_{1}+\boldsymbol{\psi}_{2}$ is a parameter vector of
likelihoods:%
\begin{align*}
\widehat{\Lambda}_{\boldsymbol{\psi}}\left(  \mathbf{x}\right)   &
=\sum\nolimits_{i=1}^{l_{1}}p\left(  \mathbf{x}_{1_{i_{\ast}}}%
|\operatorname{comp}_{\overrightarrow{\mathbf{x}_{1i\ast}}}\left(
\overrightarrow{\boldsymbol{\tau}}\right)  \right)  \frac{\mathbf{x}%
_{1_{i\ast}}}{\left\Vert \mathbf{x}_{1_{i\ast}}\right\Vert }\\
&  +\sum\nolimits_{i=1}^{l_{2}}p\left(  \mathbf{x}_{2_{i_{\ast}}%
}|\operatorname{comp}_{\overrightarrow{\mathbf{x}_{2i\ast}}}\left(
\overrightarrow{\boldsymbol{\tau}}\right)  \right)  \frac{\mathbf{x}%
_{2_{i\ast}}}{\left\Vert \mathbf{x}_{2_{i\ast}}\right\Vert }%
\end{align*}
\emph{and} a locus of principal eigenaxis components in Wolfe dual eigenspace
$\widetilde{Z}$, and a primal linear eigenlocus $\boldsymbol{\tau=\tau}%
_{1}-\boldsymbol{\tau}_{2}$ is a parameter vector of likelihoods:%
\begin{align*}
\widehat{\Lambda}_{\boldsymbol{\tau}}\left(  \mathbf{x}\right)   &
=\sum\nolimits_{i=1}^{l_{1}}p\left(  \mathbf{x}_{1_{i_{\ast}}}%
|\operatorname{comp}_{\overrightarrow{\mathbf{x}_{1i\ast}}}\left(
\overrightarrow{\boldsymbol{\tau}}\right)  \right)  \mathbf{x}_{1_{i\ast}}\\
&  -\sum\nolimits_{i=1}^{l_{2}}p\left(  \mathbf{x}_{2_{i_{\ast}}%
}|\operatorname{comp}_{\overrightarrow{\mathbf{x}_{2i\ast}}}\left(
\overrightarrow{\boldsymbol{\tau}}\right)  \right)  \mathbf{x}_{2_{i\ast}}%
\end{align*}
\emph{and} a locus of principal eigenaxis components in decision space $Z$,
that jointly determine the basis of a linear classification system
$\boldsymbol{\tau}^{T}\mathbf{x}+\tau_{0}\overset{\omega_{1}}{\underset{\omega
_{2}}{\gtrless}}0$.

Moreover, the Wolfe dual likelihood ratio%
\begin{align*}
\widehat{\Lambda}_{\boldsymbol{\psi}}\left(  \mathbf{x}\right)   &  =p\left(
\widehat{\Lambda}_{\boldsymbol{\psi}}\left(  \mathbf{x}\right)  |\omega
_{1}\right)  +p\left(  \widehat{\Lambda}_{\boldsymbol{\psi}}\left(
\mathbf{x}\right)  |\omega_{2}\right) \\
&  =\boldsymbol{\psi}_{1}+\boldsymbol{\psi}_{2}%
\end{align*}
is constrained to satisfy the equilibrium equation:%
\[
p\left(  \widehat{\Lambda}_{\boldsymbol{\psi}}\left(  \mathbf{x}\right)
|\omega_{1}\right)  =p\left(  \widehat{\Lambda}_{\boldsymbol{\psi}}\left(
\mathbf{x}\right)  |\omega_{2}\right)
\]
so that the Wolfe dual likelihood ratio $\widehat{\Lambda}_{\boldsymbol{\psi}%
}\left(  \mathbf{x}\right)  =\boldsymbol{\psi}_{1}+\boldsymbol{\psi}_{2}$ is
in statistical equilibrium:%
\[
\boldsymbol{\psi}_{1}=\boldsymbol{\psi}_{2}\text{.}%
\]

I will demonstrate that the dual nature of $\boldsymbol{\tau}$ enables a
linear eigenlocus discriminant function%
\[
\widetilde{\Lambda}_{\boldsymbol{\tau}}\left(  \mathbf{x}\right)
=\boldsymbol{\tau}^{T}\mathbf{x}+\tau_{0}%
\]
to be the solution to a fundamental integral equation of binary classification
for a classification system in statistical equilibrium:%
\begin{align*}
f\left(  \widetilde{\Lambda}_{\boldsymbol{\tau}}\left(  \mathbf{x}\right)
\right)   &  =\;\int\nolimits_{Z_{1}}p\left(  \mathbf{x}_{1_{i\ast}%
}|\boldsymbol{\tau}_{1}\right)  d\boldsymbol{\tau}_{1}+\int\nolimits_{Z_{2}%
}p\left(  \mathbf{x}_{1_{i\ast}}|\boldsymbol{\tau}_{1}\right)
d\boldsymbol{\tau}_{1}+\delta\left(  y\right)  \boldsymbol{\psi}_{1}\\
&  =\int\nolimits_{Z_{1}}p\left(  \mathbf{x}_{2_{i\ast}}|\boldsymbol{\tau}%
_{2}\right)  d\boldsymbol{\tau}_{2}+\int\nolimits_{Z_{2}}p\left(
\mathbf{x}_{2_{i\ast}}|\boldsymbol{\tau}_{2}\right)  d\boldsymbol{\tau}%
_{2}-\delta\left(  y\right)  \boldsymbol{\psi}_{2}\text{,}%
\end{align*}
where all of the forces associated with the counter risks and the risks for
class $\omega_{1}$ and class $\omega_{2}$ are symmetrically balanced with each
other%
\begin{align*}
f\left(  \widetilde{\Lambda}_{\boldsymbol{\tau}}\left(  \mathbf{x}\right)
\right)   &  :\int\nolimits_{Z_{1}}p\left(  \mathbf{x}_{1_{i\ast}%
}|\boldsymbol{\tau}_{1}\right)  d\boldsymbol{\tau}_{1}-\int\nolimits_{Z_{1}%
}p\left(  \mathbf{x}_{2_{i\ast}}|\boldsymbol{\tau}_{2}\right)
d\boldsymbol{\tau}_{2}+\delta\left(  y\right)  \boldsymbol{\psi}_{1}\\
&  =\int\nolimits_{Z_{2}}p\left(  \mathbf{x}_{2_{i\ast}}|\boldsymbol{\tau}%
_{2}\right)  d\boldsymbol{\tau}_{2}-\int\nolimits_{Z_{2}}p\left(
\mathbf{x}_{1_{i\ast}}|\boldsymbol{\tau}_{1}\right)  d\boldsymbol{\tau}%
_{1}-\delta\left(  y\right)  \boldsymbol{\psi}_{2}\text{,}%
\end{align*}
over the $Z_{1}$ and $Z_{2}$ decision regions, by means of an elegant,
statistical balancing feat in Wolfe dual eigenspace $\widetilde{Z}$, where the
functional $E_{\boldsymbol{\tau}_{1}}$ of $\left\Vert \boldsymbol{\tau}%
_{1}\right\Vert _{\min_{c}}^{2}$ in Eq. (\ref{TAE Eigenlocus Component One})
and the functional $E_{\boldsymbol{\tau}_{2}}$ of $\left\Vert \boldsymbol{\tau
}_{2}\right\Vert _{\min_{c}}^{2}$ in Eq. (\ref{TAE Eigenlocus Component Two})
are constrained to be equal to each other by means of a symmetric equalizer
statistic $\nabla_{eq}$: $\frac{\delta\left(  y\right)  }{2}\boldsymbol{\psi}%
$, where $\delta\left(  y\right)  \triangleq\sum\nolimits_{i=1}^{l}%
y_{i}\left(  1-\xi_{i}\right)  $.

I have shown that each of the constrained, primal principal eigenaxis
components $\psi_{1_{i\ast}}\mathbf{x}_{1_{i\ast}}$ or $\psi_{2_{i\ast}%
}\mathbf{x}_{2_{i\ast}}$ on $\boldsymbol{\tau}=\boldsymbol{\tau}%
_{1}-\boldsymbol{\tau}_{2}$ have such magnitudes and directions that a
constrained, linear eigenlocus discriminant function $\widetilde{\Lambda
}_{\boldsymbol{\tau}}\left(  \mathbf{x}\right)  =\boldsymbol{\tau}%
^{T}\mathbf{x}+\tau_{0}$ partitions any given feature space into congruent
decision regions $Z_{1}\cong Z_{2}$, which are symmetrically partitioned by a
linear decision boundary, by means of three, symmetrical linear loci, all of
which reference $\boldsymbol{\tau}$.

I\ will show that linear eigenlocus classification systems $\boldsymbol{\tau
}^{T}\mathbf{x}+\tau_{0}\overset{\omega_{1}}{\underset{\omega_{2}}{\gtrless}%
}0$ generate decision regions $Z_{1}$ and $Z_{2}$ for which the dual parameter
vectors of likelihoods $\widehat{\Lambda}_{\boldsymbol{\psi}}\left(
\mathbf{x}\right)  =\boldsymbol{\psi}_{1}+\boldsymbol{\psi}_{2}$ and
$\widehat{\Lambda}_{\boldsymbol{\tau}}\left(  \mathbf{x}\right)
=\boldsymbol{\tau}_{1}-\boldsymbol{\tau}_{2}$ are in statistical equilibrium.
Thereby, I\ will demonstrate that balancing the forces associated with the
risk $\mathfrak{R}_{\mathfrak{\min}}\left(  Z|\boldsymbol{\tau}\right)  $ of
the linear classification system $\boldsymbol{\tau}^{T}\mathbf{x}+\tau
_{0}\overset{\omega_{1}}{\underset{\omega_{2}}{\gtrless}}0$ hinges on
balancing the eigenenergies associated with the positions or locations of the
dual likelihood ratios $\widehat{\Lambda}_{\boldsymbol{\psi}}\left(
\mathbf{x}\right)  $ and $\widehat{\Lambda}_{\boldsymbol{\tau}}\left(
\mathbf{x}\right)  $:%
\begin{align*}
\widehat{\Lambda}_{\boldsymbol{\psi}}\left(  \mathbf{x}\right)   &
=\boldsymbol{\psi}_{1}+\boldsymbol{\psi}_{2}\\
&  =\sum\nolimits_{i=1}^{l_{1}}\psi_{1_{i\ast}}\frac{\mathbf{x}_{1_{i\ast}}%
}{\left\Vert \mathbf{x}_{1_{i\ast}}\right\Vert }+\sum\nolimits_{i=1}^{l_{2}%
}\psi_{2_{i\ast}}\frac{\mathbf{x}_{2_{i\ast}}}{\left\Vert \mathbf{x}%
_{2_{i\ast}}\right\Vert }%
\end{align*}
and%
\begin{align*}
\widehat{\Lambda}_{\boldsymbol{\tau}}\left(  \mathbf{x}\right)   &
=\boldsymbol{\tau}_{1}-\boldsymbol{\tau}_{2}\\
&  =\sum\nolimits_{i=1}^{l_{1}}\psi_{1_{i\ast}}\mathbf{x}_{1_{i\ast}}%
-\sum\nolimits_{i=1}^{l_{2}}\psi_{2_{i\ast}}\mathbf{x}_{2_{i\ast}}\text{.}%
\end{align*}

\subsection{Balancing the Eigenenergies of $\boldsymbol{\tau}_{1}%
-\boldsymbol{\tau}_{2}$}

I will now devise an equation that determines how the total allowed
eigenenergies $\left\Vert \boldsymbol{\tau}_{1}\right\Vert _{\min_{c}}^{2}$
and $\left\Vert \boldsymbol{\tau}_{2}\right\Vert _{\min_{c}}^{2}$ exhibited by
$\boldsymbol{\tau}_{1}$ and $\boldsymbol{\tau}_{2}$ are symmetrically balanced
with each other.

Using Eq. (\ref{TAE Eigenlocus Component One}) and the equilibrium constraint
on $\boldsymbol{\psi}$ in Eq.
(\ref{Equilibrium Constraint on Dual Eigen-components})%
\begin{align*}
\left\Vert \boldsymbol{\tau}_{1}\right\Vert _{\min_{c}}^{2}-\left\Vert
\boldsymbol{\tau}_{1}\right\Vert \left\Vert \boldsymbol{\tau}_{2}\right\Vert
\cos\theta_{\boldsymbol{\tau}_{1}\boldsymbol{\tau}_{2}}  &  \equiv
\sum\nolimits_{i=1}^{l_{1}}\psi_{1_{i_{\ast}}}\left(  1-\xi_{i}-\tau
_{0}\right) \\
&  \equiv\frac{1}{2}\sum\nolimits_{i=1}^{l}\psi_{_{i_{\ast}}}\left(  1-\xi
_{i}-\tau_{0}\right)  \text{,}%
\end{align*}
it follows that the functional $E_{\boldsymbol{\tau}_{1}}$ of the total
allowed eigenenergy $\left\Vert \boldsymbol{\tau}_{1}\right\Vert _{\min_{c}%
}^{2}$ of $\boldsymbol{\tau}_{1}$%
\[
E_{\boldsymbol{\tau}_{1}}=\left\Vert \boldsymbol{\tau}_{1}\right\Vert
_{\min_{c}}^{2}-\left\Vert \boldsymbol{\tau}_{1}\right\Vert \left\Vert
\boldsymbol{\tau}_{2}\right\Vert \cos\theta_{\boldsymbol{\tau}_{1}%
\boldsymbol{\tau}_{2}}%
\]
is equivalent to the functional $E_{\boldsymbol{\psi}_{1}}$:%
\begin{equation}
E_{\boldsymbol{\psi}_{1}}=\frac{1}{2}\sum\nolimits_{i=1}^{l}\psi_{_{i_{\ast}}%
}\left(  1-\xi_{i}\right)  -\tau_{0}\sum\nolimits_{i=1}^{l_{1}}\psi
_{1_{i_{\ast}}}\text{.} \label{TAE Constraint COMP1}%
\end{equation}

Using Eq. (\ref{TAE Eigenlocus Component Two}) and the equilibrium constraint
on $\boldsymbol{\psi}$ in Eq.
(\ref{Equilibrium Constraint on Dual Eigen-components})%
\begin{align*}
\left\Vert \boldsymbol{\tau}_{2}\right\Vert _{\min_{c}}^{2}-\left\Vert
\boldsymbol{\tau}_{2}\right\Vert \left\Vert \boldsymbol{\tau}_{1}\right\Vert
\cos\theta_{\boldsymbol{\tau}_{2}\boldsymbol{\tau}_{1}}  &  \equiv
\sum\nolimits_{i=1}^{l_{2}}\psi_{2_{i_{\ast}}}\left(  1-\xi_{i}+\tau
_{0}\right) \\
&  \equiv\frac{1}{2}\sum\nolimits_{i=1}^{l}\psi_{_{i_{\ast}}}\left(  1-\xi
_{i}+\tau_{0}\right)  \text{,}%
\end{align*}
it follows that the functional $E_{\boldsymbol{\tau}_{2}}$ of the total
allowed eigenenergy $\left\Vert \boldsymbol{\tau}_{2}\right\Vert _{\min_{c}%
}^{2}$ of $\boldsymbol{\tau}_{2}$%
\[
E_{\boldsymbol{\tau}_{2}}=\left\Vert \boldsymbol{\tau}_{2}\right\Vert
_{\min_{c}}^{2}-\left\Vert \boldsymbol{\tau}_{2}\right\Vert \left\Vert
\boldsymbol{\tau}_{1}\right\Vert \cos\theta_{\boldsymbol{\tau}_{2}%
\boldsymbol{\tau}_{1}}%
\]
is equivalent to the functional $E_{\boldsymbol{\psi}_{2}}$:%
\begin{equation}
E_{\boldsymbol{\psi}_{2}}=\frac{1}{2}\sum\nolimits_{i=1}^{l}\psi_{_{i_{\ast}}%
}\left(  1-\xi_{i}\right)  +\tau_{0}\sum\nolimits_{i=1}^{l_{2}}\psi
_{2_{i_{\ast}}}\text{.} \label{TAE Constraint COMP2}%
\end{equation}

Next, I will use the identity for $\left\Vert \boldsymbol{\tau}\right\Vert
_{\min_{c}}^{2}$ in Eq. (\ref{TAE SDE})%
\[
\left\Vert \boldsymbol{\tau}\right\Vert _{\min_{c}}^{2}\equiv\sum
\nolimits_{i=1}^{l}\psi_{i_{\ast}}\left(  1-\xi_{i}\right)
\]
to rewrite $E_{\boldsymbol{\psi}_{1}}$%
\begin{align*}
E_{\boldsymbol{\psi}_{1}}  &  =\frac{1}{2}\sum\nolimits_{i=1}^{l}%
\psi_{_{i_{\ast}}}\left(  1-\xi_{i}\right)  -\tau_{0}\sum\nolimits_{i=1}%
^{l_{1}}\psi_{1_{i_{\ast}}}\\
&  \equiv\frac{1}{2}\left\Vert \boldsymbol{\tau}\right\Vert _{\min_{c}}%
^{2}-\tau_{0}\sum\nolimits_{i=1}^{l_{1}}\psi_{1_{i_{\ast}}}%
\end{align*}
and $E_{\boldsymbol{\psi}_{2}}$%
\begin{align*}
E_{\boldsymbol{\psi}_{2}}  &  =\frac{1}{2}\sum\nolimits_{i=1}^{l}%
\psi_{_{i_{\ast}}}\left(  1-\xi_{i}\right)  +\tau_{0}\sum\nolimits_{i=1}%
^{l_{2}}\psi_{2_{i_{\ast}}}\\
&  \equiv\frac{1}{2}\left\Vert \boldsymbol{\tau}\right\Vert _{\min_{c}}%
^{2}+\tau_{0}\sum\nolimits_{i=1}^{l_{2}}\psi_{2_{i_{\ast}}}%
\end{align*}
in terms of $\frac{1}{2}\left\Vert \boldsymbol{\tau}\right\Vert _{\min_{c}%
}^{2}$ and a symmetric equalizer statistic.

Substituting the rewritten expressions for $E_{\boldsymbol{\psi}_{1}}$ and
$E_{\boldsymbol{\psi}_{2}}$ into Eqs (\ref{TAE Eigenlocus Component One}) and
(\ref{TAE Eigenlocus Component Two}) produces the equations%
\[
\left(  \left\Vert \boldsymbol{\tau}_{1}\right\Vert _{\min_{c}}^{2}-\left\Vert
\boldsymbol{\tau}_{1}\right\Vert \left\Vert \boldsymbol{\tau}_{2}\right\Vert
\cos\theta_{\boldsymbol{\tau}_{1}\boldsymbol{\tau}_{2}}\right)  +\tau_{0}%
\sum\nolimits_{i=1}^{l_{1}}\psi_{1_{i_{\ast}}}\equiv\frac{1}{2}\left\Vert
\boldsymbol{\tau}\right\Vert _{\min_{c}}^{2}%
\]
and
\[
\left(  \left\Vert \boldsymbol{\tau}_{2}\right\Vert _{\min_{c}}^{2}-\left\Vert
\boldsymbol{\tau}_{2}\right\Vert \left\Vert \boldsymbol{\tau}_{1}\right\Vert
\cos\theta_{\boldsymbol{\tau}_{2}\boldsymbol{\tau}_{1}}\right)  -\tau_{0}%
\sum\nolimits_{i=1}^{l_{2}}\psi_{2_{i_{\ast}}}\equiv\frac{1}{2}\left\Vert
\boldsymbol{\tau}\right\Vert _{\min_{c}}^{2}\text{,}%
\]
where the terms $\tau_{0}\sum\nolimits_{i=1}^{l_{1}}\psi_{1_{i_{\ast}}}$ and
$-\tau_{0}\sum\nolimits_{i=1}^{l_{2}}\psi_{2_{i_{\ast}}}$ specify a symmetric
equalizer statistic $\nabla_{eq}$ for integrals of class-conditional
probability density functions $p\left(  \mathbf{x}_{1_{i\ast}}%
|\boldsymbol{\tau}_{1}\right)  $ and $p\left(  \mathbf{x}_{2_{i_{\ast}}%
}|\boldsymbol{\tau}_{2}\right)  $ that determine conditional probability
functions $P\left(  \mathbf{x}_{1_{i\ast}}|\boldsymbol{\tau}_{1}\right)  $ and
$P\left(  \mathbf{x}_{2_{i\ast}}|\boldsymbol{\tau}_{2}\right)  $.

Therefore, let$\ \nabla_{eq}$ denote $\tau_{0}\sum\nolimits_{i=1}^{l_{1}}%
\psi_{1_{i_{\ast}}}$ and $\tau_{0}\sum\nolimits_{i=1}^{l_{2}}\psi_{2_{i_{\ast
}}}$, where%
\[
\nabla_{eq}\triangleq\frac{\tau_{0}}{2}\sum\nolimits_{i=1}^{l}\psi_{_{i\ast}%
}\text{.}%
\]

It follows that the dual, class-conditional parameter vectors of likelihoods
$\boldsymbol{\psi}_{1}$, $\boldsymbol{\psi}_{2}$, $\boldsymbol{\tau}_{1}$, and
$\boldsymbol{\tau}_{2}$ satisfy the eigenlocus equations%
\begin{equation}
\left\Vert \boldsymbol{\tau}_{1}\right\Vert _{\min_{c}}^{2}-\left\Vert
\boldsymbol{\tau}_{1}\right\Vert \left\Vert \boldsymbol{\tau}_{2}\right\Vert
\cos\theta_{\boldsymbol{\tau}_{1}\boldsymbol{\tau}_{2}}+\nabla_{eq}\equiv
\frac{1}{2}\left\Vert \boldsymbol{\tau}\right\Vert _{\min_{c}}^{2}
\label{Balancing Feat SDEC1}%
\end{equation}
and%
\begin{equation}
\left\Vert \boldsymbol{\tau}_{2}\right\Vert _{\min_{c}}^{2}-\left\Vert
\boldsymbol{\tau}_{2}\right\Vert \left\Vert \boldsymbol{\tau}_{1}\right\Vert
\cos\theta_{\boldsymbol{\tau}_{2}\boldsymbol{\tau}_{1}}-\nabla_{eq}\equiv
\frac{1}{2}\left\Vert \boldsymbol{\tau}\right\Vert _{\min_{c}}^{2}\text{,}
\label{Balancing Feat SDEC2}%
\end{equation}
where $\nabla_{eq}\triangleq\frac{\tau_{0}}{2}\sum\nolimits_{i=1}^{l}%
\psi_{_{i\ast}}$.

I will now examine the geometric and statistical properties of the equalizer
statistic $\nabla_{eq}$ in eigenspace.

\subsubsection{Properties of $\nabla_{eq}$ in Eigenspace}

Substituting the vector expression $\boldsymbol{\tau}=\sum\nolimits_{i=1}%
^{l_{1}}\psi_{1_{i_{\ast}}}\mathbf{x}_{1_{i_{\ast}}}-\sum\nolimits_{i=1}%
^{l_{2}}\psi_{2_{i_{\ast}}}\mathbf{x}_{2_{i_{\ast}}}$ for $\boldsymbol{\tau}$
in Eq. (\ref{Pair of Normal Eigenlocus Components}) into the expression for
$\tau_{0}$ in Eq. (\ref{Normal Eigenlocus Projection Factor}) produces the
statistic for $\tau_{0}$:%
\begin{align}
\tau_{0}  &  =-\sum\nolimits_{i=1}^{l}\mathbf{x}_{i\ast}^{T}\sum
\nolimits_{j=1}^{l_{1}}\psi_{1_{j_{\ast}}}\mathbf{x}_{1_{j_{\ast}}%
}\label{Eigenlocus Projection Factor Two}\\
&  +\sum\nolimits_{i=1}^{l}\mathbf{x}_{i\ast}^{T}\sum\nolimits_{j=1}^{l_{2}%
}\psi_{2_{j_{\ast}}}\mathbf{x}_{2_{j_{\ast}}}+\sum\nolimits_{i=1}^{l}%
y_{i}\left(  1-\xi_{i}\right)  \text{.}\nonumber
\end{align}

Substituting the statistic for $\tau_{0}$ in Eq.
(\ref{Eigenlocus Projection Factor Two}) into the expression for $\nabla_{eq}$
produces the statistic for $\nabla_{eq}$:%
\begin{align*}
\nabla_{eq}  &  =\frac{\tau_{0}}{2}\sum\nolimits_{i=1}^{l}\psi_{_{i\ast}}\\
&  =-\left(  \sum\nolimits_{i=1}^{l}\mathbf{x}_{i\ast}^{T}\boldsymbol{\tau
}_{1}\right)  \frac{1}{2}\sum\nolimits_{i=1}^{l}\psi_{_{i\ast}}\\
&  +\left(  \sum\nolimits_{i=1}^{l}\mathbf{x}_{i\ast}^{T}\boldsymbol{\tau}%
_{2}\right)  \frac{1}{2}\sum\nolimits_{i=1}^{l}\psi_{_{i\ast}}+\delta\left(
y\right)  \frac{1}{2}\sum\nolimits_{i=1}^{l}\psi_{_{i\ast}}\text{,}%
\end{align*}
where $\delta\left(  y\right)  \triangleq\sum\nolimits_{i=1}^{l}y_{i}\left(
1-\xi_{i}\right)  $. Let $\widehat{\mathbf{x}}_{i\ast}\triangleq
\sum\nolimits_{i=1}^{l}\mathbf{x}_{i\ast}$.

It follows that $\tau_{0}$ regulates a symmetrical balancing act for
components of $\widehat{\mathbf{x}}_{i\ast}$ along $\boldsymbol{\tau}_{1}$ and
$\boldsymbol{\tau}_{2}$, where the statistic $\nabla_{eq}$ is written as%
\[
+\nabla_{eq}=\left[  \operatorname{comp}_{\overrightarrow{\boldsymbol{\tau
}_{2}}}\left(  \overrightarrow{\widehat{\mathbf{x}}_{i\ast}}\right)
-\operatorname{comp}_{\overrightarrow{\boldsymbol{\tau}_{1}}}\left(
\overrightarrow{\widehat{\mathbf{x}}_{i\ast}}\right)  +\delta\left(  y\right)
\right]  \frac{1}{2}\sum\nolimits_{i=1}^{l}\psi_{_{i\ast}}%
\]
and%
\[
-\nabla_{eq}=\left[  \operatorname{comp}_{\overrightarrow{\boldsymbol{\tau
}_{1}}}\left(  \overrightarrow{\widehat{\mathbf{x}}_{i\ast}}\right)
-\operatorname{comp}_{\overrightarrow{\boldsymbol{\tau}_{2}}}\left(
\overrightarrow{\widehat{\mathbf{x}}_{i\ast}}\right)  -\delta\left(  y\right)
\right]  \frac{1}{2}\sum\nolimits_{i=1}^{l}\psi_{_{i\ast}}\text{.}%
\]

Returning to Eq. (\ref{Balanced Eigenlocus Equation}):%
\[
\widehat{\mathbf{x}}_{1i\ast}^{T}\left(  \boldsymbol{\tau}_{1}\mathbf{-}%
\boldsymbol{\tau}_{2}\right)  =\widehat{\mathbf{x}}_{2i\ast}^{T}\left(
\boldsymbol{\tau}_{2}\mathbf{-}\boldsymbol{\tau}_{1}\right)  \text{,}%
\]
given that the components of $\widehat{\mathbf{x}}_{1i\ast}$ and
$\widehat{\mathbf{x}}_{2i\ast}$ along $\boldsymbol{\tau}_{1}$ and
$\boldsymbol{\tau}_{2}$ are in statistical equilibrium:%
\[
\left[  \operatorname{comp}_{\overrightarrow{\boldsymbol{\tau}_{1}}}\left(
\overrightarrow{\widehat{\mathbf{x}}_{1i\ast}}\right)  -\operatorname{comp}%
_{\overrightarrow{\boldsymbol{\tau}_{2}}}\left(
\overrightarrow{\widehat{\mathbf{x}}_{1i\ast}}\right)  \right]
\rightleftharpoons\left[  \operatorname{comp}%
_{\overrightarrow{\boldsymbol{\tau}_{2}}}\left(
\overrightarrow{\widehat{\mathbf{x}}_{2i\ast}}\right)  -\operatorname{comp}%
_{\overrightarrow{\boldsymbol{\tau}_{1}}}\left(
\overrightarrow{\widehat{\mathbf{x}}_{2i\ast}}\right)  \right]  \text{,}%
\]
it follows that%
\[
+\nabla_{eq}=\delta\left(  y\right)  \frac{1}{2}\sum\nolimits_{i=1}^{l}%
\psi_{_{i\ast}}\equiv\delta\left(  y\right)  \boldsymbol{\psi}_{1}%
\]
and%
\[
-\nabla_{eq}=-\delta\left(  y\right)  \frac{1}{2}\sum\nolimits_{i=1}^{l}%
\psi_{_{i\ast}}\equiv-\delta\left(  y\right)  \boldsymbol{\psi}_{2}\text{.}%
\]

I\ will now demonstrate that the symmetric equalizer statistic $\nabla_{eq}$
ensures that the class-conditional probability density functions $p\left(
\mathbf{x}_{1_{i\ast}}|\boldsymbol{\tau}_{1}\right)  $ and $p\left(
\mathbf{x}_{2_{i_{\ast}}}|\boldsymbol{\tau}_{2}\right)  $ satisfy the integral
equation%
\begin{align*}
f\left(  \widetilde{\Lambda}_{\boldsymbol{\tau}}\left(  \mathbf{x}\right)
\right)  =  &  \int_{Z}p\left(  \mathbf{x}_{1_{i\ast}}|\boldsymbol{\tau}%
_{1}\right)  d\boldsymbol{\tau}_{1}+\delta\left(  y\right)  \frac{1}{2}%
\sum\nolimits_{i=1}^{l}\psi_{_{i\ast}}\\
&  =\int_{Z}p\left(  \mathbf{x}_{2_{i\ast}}|\boldsymbol{\tau}_{2}\right)
d\boldsymbol{\tau}_{2}-\delta\left(  y\right)  \sum\nolimits_{i=1}^{l}%
\psi_{_{i\ast}}\text{,}%
\end{align*}
over the decision space $Z$: $Z=Z_{1}+Z_{2}$ and $Z_{1}\cong Z_{2}$, whereby
the likelihood ratio $\widehat{\Lambda}_{\boldsymbol{\tau}}\left(
\mathbf{x}\right)  =p\left(  \widehat{\Lambda}_{\boldsymbol{\tau}}\left(
\mathbf{x}\right)  |\omega_{1}\right)  -p\left(  \widehat{\Lambda
}_{\boldsymbol{\tau}}\left(  \mathbf{x}\right)  |\omega_{2}\right)  $ of the
classification system $\boldsymbol{\tau}^{T}\left(  \mathbf{x}%
-\widehat{\mathbf{x}}_{i\ast}\right)  +\delta\left(  y\right)  \overset{\omega
_{1}}{\underset{\omega_{2}}{\gtrless}}0$ is in statistical equilibrium. In the
process, I will formulate an equation which ensures that $\left\Vert
\boldsymbol{\tau}_{1}\right\Vert _{\min_{c}}^{2}$ and $\left\Vert
\boldsymbol{\tau}_{2}\right\Vert _{\min_{c}}^{2}$ are symmetrically balanced
with each other.

\subsection{Linear Eigenlocus Integral Equation}

Let $\nabla_{eq}=\delta\left(  y\right)  \frac{1}{2}\sum\nolimits_{i=1}%
^{l}\psi_{_{i\ast}}$. Substituting the expression for $+\nabla_{eq}$ into Eq.
(\ref{Balancing Feat SDEC1}) produces an equation that is satisfied by the
conditional probabilities of locations for the set $\left\{  \mathbf{x}%
_{1_{i\ast}}\right\}  _{i=1}^{l_{1}}$ of $\mathbf{x}_{1_{i\ast}}$ extreme
points within the decision space $Z$:%
\begin{align*}
P\left(  \mathbf{x}_{1_{i\ast}}|\boldsymbol{\tau}_{1}\right)   &  =\left\Vert
\boldsymbol{\tau}_{1}\right\Vert _{\min_{c}}^{2}-\left\Vert \boldsymbol{\tau
}_{1}\right\Vert \left\Vert \boldsymbol{\tau}_{2}\right\Vert \cos
\theta_{\boldsymbol{\tau}_{1}\boldsymbol{\tau}_{2}}+\delta\left(  y\right)
\frac{1}{2}\sum\nolimits_{i=1}^{l}\psi_{_{i\ast}}\\
&  \equiv\frac{1}{2}\left\Vert \boldsymbol{\tau}\right\Vert _{\min_{c}}%
^{2}\text{,}%
\end{align*}
and substituting the expression for $-\nabla_{eq}$ into Eq.
(\ref{Balancing Feat SDEC2}) produces an equation that is satisfied by the
conditional probabilities of locations for the set $\left\{  \mathbf{x}%
_{2_{i\ast}}\right\}  _{i=1}^{l_{2}}$ of $\mathbf{x}_{2_{i\ast}}$ extreme
points within the decision space $Z$:%
\begin{align*}
P\left(  \mathbf{x}_{2_{i\ast}}|\boldsymbol{\tau}_{2}\right)   &  =\left\Vert
\boldsymbol{\tau}_{2}\right\Vert _{\min_{c}}^{2}-\left\Vert \boldsymbol{\tau
}_{2}\right\Vert \left\Vert \boldsymbol{\tau}_{1}\right\Vert \cos
\theta_{\boldsymbol{\tau}_{2}\boldsymbol{\tau}_{1}}-\delta\left(  y\right)
\frac{1}{2}\sum\nolimits_{i=1}^{l}\psi_{_{i\ast}}\\
&  \equiv\frac{1}{2}\left\Vert \boldsymbol{\tau}\right\Vert _{\min_{c}}%
^{2}\text{,}%
\end{align*}
where the equalizer statistic $\nabla_{eq}$%
\[
\nabla_{eq}=\delta\left(  y\right)  \frac{1}{2}\sum\nolimits_{i=1}^{l}%
\psi_{_{i\ast}}%
\]
\emph{equalizes} the conditional probabilities $P\left(  \mathbf{x}_{1_{i\ast
}}|\boldsymbol{\tau}_{1}\right)  $ and $P\left(  \mathbf{x}_{2_{i\ast}%
}|\boldsymbol{\tau}_{2}\right)  $ of observing the set $\left\{
\mathbf{x}_{1_{i\ast}}\right\}  _{i=1}^{l_{1}}$ of $\mathbf{x}_{1_{i\ast}}$
extreme points and the set $\left\{  \mathbf{x}_{2_{i\ast}}\right\}
_{i=1}^{l_{2}}$ of $\mathbf{x}_{2_{i\ast}}$ extreme points within the $Z_{1}$
and $Z_{2}$ decision regions of the decision space $Z$.

Therefore, the equalizer statistic $\delta\left(  y\right)  \frac{1}{2}%
\sum\nolimits_{i=1}^{l}\psi_{_{i\ast}}$ ensures that $\left\Vert
\boldsymbol{\tau}_{1}\right\Vert _{\min_{c}}^{2}$ and $\left\Vert
\boldsymbol{\tau}_{2}\right\Vert _{\min_{c}}^{2}$ are symmetrically balanced
with each other in the following manner:%
\[
\left\Vert \boldsymbol{\tau}_{1}\right\Vert _{\min_{c}}^{2}-\left\Vert
\boldsymbol{\tau}_{1}\right\Vert \left\Vert \boldsymbol{\tau}_{2}\right\Vert
\cos\theta_{\boldsymbol{\tau}_{1}\boldsymbol{\tau}_{2}}+\delta\left(
y\right)  \frac{1}{2}\sum\nolimits_{i=1}^{l}\psi_{_{i\ast}}\equiv\frac{1}%
{2}\left\Vert \boldsymbol{\tau}\right\Vert _{\min_{c}}^{2}%
\]
and%
\[
\left\Vert \boldsymbol{\tau}_{2}\right\Vert _{\min_{c}}^{2}-\left\Vert
\boldsymbol{\tau}_{2}\right\Vert \left\Vert \boldsymbol{\tau}_{1}\right\Vert
\cos\theta_{\boldsymbol{\tau}_{2}\boldsymbol{\tau}_{1}}-\delta\left(
y\right)  \frac{1}{2}\sum\nolimits_{i=1}^{l}\psi_{_{i\ast}}\equiv\frac{1}%
{2}\left\Vert \boldsymbol{\tau}\right\Vert _{\min_{c}}^{2}\text{.}%
\]

Thereby, the equalizer statistic%
\[
\delta\left(  y\right)  \frac{1}{2}\sum\nolimits_{i=1}^{l}\psi_{_{i\ast}}%
\]
\emph{equalizes} the total allowed eigenenergies $\left\Vert \boldsymbol{\tau
}_{1}\right\Vert _{\min_{c}}^{2}$ and $\left\Vert \boldsymbol{\tau}%
_{2}\right\Vert _{\min_{c}}^{2}$ exhibited by $\boldsymbol{\tau}_{1}$ and
$\boldsymbol{\tau}_{2}$ so that the total allowed eigenenergies $\left\Vert
\boldsymbol{\tau}_{1}-\boldsymbol{\tau}_{2}\right\Vert _{\min_{c}}^{2}$
exhibited by the scaled extreme points on $\boldsymbol{\tau}_{1}%
-\boldsymbol{\tau}_{2}$ are symmetrically balanced with each other about the
fulcrum of $\boldsymbol{\tau}$:%
\begin{equation}
\left\Vert \boldsymbol{\tau}_{1}\right\Vert _{\min_{c}}^{2}+\delta\left(
y\right)  \frac{1}{2}\sum\nolimits_{i=1}^{l}\psi_{_{i\ast}}\equiv\left\Vert
\boldsymbol{\tau}_{2}\right\Vert _{\min_{c}}^{2}-\delta\left(  y\right)
\frac{1}{2}\sum\nolimits_{i=1}^{l}\psi_{_{i\ast}}
\label{Symmetrical Balance of Total Allowed Eigenenergies}%
\end{equation}
which is located at the center of eigenenergy $\left\Vert \boldsymbol{\tau
}\right\Vert _{\min_{c}}^{2}$: the geometric center of $\boldsymbol{\tau}$.
Thus, the likelihood ratio%
\begin{align*}
\widehat{\Lambda}_{\boldsymbol{\tau}}\left(  \mathbf{x}\right)   &  =p\left(
\widehat{\Lambda}_{\boldsymbol{\tau}}\left(  \mathbf{x}\right)  |\omega
_{1}\right)  -p\left(  \widehat{\Lambda}_{\boldsymbol{\tau}}\left(
\mathbf{x}\right)  |\omega_{2}\right) \\
&  =\boldsymbol{\tau}_{1}-\boldsymbol{\tau}_{2}%
\end{align*}
of the classification system $\boldsymbol{\tau}^{T}\left(  \mathbf{x}%
-\widehat{\mathbf{x}}_{i\ast}\right)  +\delta\left(  y\right)  \overset{\omega
_{1}}{\underset{\omega_{2}}{\gtrless}}0$ is in statistical equilibrium.

It follows that the eigenenergy $\left\Vert \boldsymbol{\tau}_{1}\right\Vert
_{\min_{c}}^{2}$ associated with the position or location of the parameter
vector of likelihoods $p\left(  \widehat{\Lambda}_{\boldsymbol{\tau}}\left(
\mathbf{x}\right)  |\omega_{1}\right)  $ given class $\omega_{1}$ is
symmetrically balanced with the eigenenergy $\left\Vert \boldsymbol{\tau}%
_{2}\right\Vert _{\min_{c}}^{2}$ associated with the position or location of
the parameter vector of likelihoods $p\left(  \widehat{\Lambda}%
_{\boldsymbol{\tau}}\left(  \mathbf{x}\right)  |\omega_{2}\right)  $ given
class $\omega_{2}$.

Returning to Eq. (\ref{Conditional Probability Function for Class One})%
\[
P\left(  \mathbf{x}_{1_{i\ast}}|\boldsymbol{\tau}_{1}\right)  =\int%
_{Z}\boldsymbol{\tau}_{1}d\boldsymbol{\tau}_{1}=\left\Vert \boldsymbol{\tau
}_{1}\right\Vert _{\min_{c}}^{2}+C_{1}%
\]
and Eq. (\ref{Conditional Probability Function for Class Two})%
\[
P\left(  \mathbf{x}_{2_{i\ast}}|\boldsymbol{\tau}_{2}\right)  =\int%
_{Z}\boldsymbol{\tau}_{2}d\boldsymbol{\tau}_{2}=\left\Vert \boldsymbol{\tau
}_{2}\right\Vert _{\min_{c}}^{2}+C_{2}\text{,}%
\]
it follows that the value for the integration constant $C_{1}$ in Eq.
(\ref{Conditional Probability Function for Class One}) is%
\[
C_{1}=-\left\Vert \boldsymbol{\tau}_{1}\right\Vert \left\Vert \boldsymbol{\tau
}_{2}\right\Vert \cos\theta_{\boldsymbol{\tau}_{1}\boldsymbol{\tau}_{2}}%
\]
and the value for the integration constant $C_{2}$ in Eq.
(\ref{Conditional Probability Function for Class Two}) is%
\[
C_{2}=-\left\Vert \boldsymbol{\tau}_{2}\right\Vert \left\Vert \boldsymbol{\tau
}_{1}\right\Vert \cos\theta_{\boldsymbol{\tau}_{2}\boldsymbol{\tau}_{1}%
}\text{.}%
\]

Therefore, the area $P\left(  \mathbf{x}_{1_{i\ast}}|\boldsymbol{\tau}%
_{1}\right)  $ under the class-conditional probability density function
$p\left(  \mathbf{x}_{1_{i\ast}}|\boldsymbol{\tau}_{1}\right)  $ in Eq.
(\ref{Conditional Density Extreme Points 1}):%
\begin{align}
P\left(  \mathbf{x}_{1_{i\ast}}|\boldsymbol{\tau}_{1}\right)   &  =\int%
_{Z}p\left(  \mathbf{x}_{1_{i\ast}}|\boldsymbol{\tau}_{1}\right)
d\boldsymbol{\tau}_{1}+\delta\left(  y\right)  \frac{1}{2}\sum\nolimits_{i=1}%
^{l}\psi_{_{i\ast}}\label{Integral Equation Class One}\\
&  =\int_{Z}\boldsymbol{\tau}_{1}d\boldsymbol{\tau}_{1}+\delta\left(
y\right)  \frac{1}{2}\sum\nolimits_{i=1}^{l}\psi_{_{i\ast}}\nonumber\\
&  =\left\Vert \boldsymbol{\tau}_{1}\right\Vert _{\min_{c}}^{2}-\left\Vert
\boldsymbol{\tau}_{1}\right\Vert \left\Vert \boldsymbol{\tau}_{2}\right\Vert
\cos\theta_{\boldsymbol{\tau}_{1}\boldsymbol{\tau}_{2}}+\delta\left(
y\right)  \frac{1}{2}\sum\nolimits_{i=1}^{l}\psi_{_{i\ast}}\nonumber\\
&  =\left\Vert \boldsymbol{\tau}_{1}\right\Vert _{\min_{c}}^{2}-\left\Vert
\boldsymbol{\tau}_{1}\right\Vert \left\Vert \boldsymbol{\tau}_{2}\right\Vert
\cos\theta_{\boldsymbol{\tau}_{1}\boldsymbol{\tau}_{2}}+\delta\left(
y\right)  \sum\nolimits_{i=1}^{l_{1}}\psi_{1_{i_{\ast}}}\nonumber\\
&  \equiv\frac{1}{2}\left\Vert \boldsymbol{\tau}\right\Vert _{\min_{c}}%
^{2}\text{,}\nonumber
\end{align}
over the decision space $Z$, is \emph{symmetrically balanced} with the area
$P\left(  \mathbf{x}_{2_{i\ast}}|\boldsymbol{\tau}_{2}\right)  $ under the
class-conditional probability density function $p\left(  \mathbf{x}%
_{2_{i_{\ast}}}|\boldsymbol{\tau}_{2}\right)  $ in Eq.
(\ref{Conditional Density Extreme Points 2}):%
\begin{align}
P\left(  \mathbf{x}_{2_{i\ast}}|\boldsymbol{\tau}_{2}\right)   &  =\int%
_{Z}p\left(  \mathbf{x}_{2_{i\ast}}|\boldsymbol{\tau}_{2}\right)
d\boldsymbol{\tau}_{2}-\delta\left(  y\right)  \frac{1}{2}\sum\nolimits_{i=1}%
^{l}\psi_{_{i\ast}}\label{Integral Equation Class Two}\\
&  =\int_{Z}\boldsymbol{\tau}_{2}d\boldsymbol{\tau}_{2}-\delta\left(
y\right)  \frac{1}{2}\left\Vert \boldsymbol{\tau}\right\Vert _{\min_{c}}%
^{2}\nonumber\\
&  =\left\Vert \boldsymbol{\tau}_{2}\right\Vert _{\min_{c}}^{2}-\left\Vert
\boldsymbol{\tau}_{2}\right\Vert \left\Vert \boldsymbol{\tau}_{1}\right\Vert
\cos\theta_{\boldsymbol{\tau}_{2}\boldsymbol{\tau}_{1}}-\delta\left(
y\right)  \frac{1}{2}\sum\nolimits_{i=1}^{l}\psi_{_{i\ast}}\nonumber\\
&  =\left\Vert \boldsymbol{\tau}_{2}\right\Vert _{\min_{c}}^{2}-\left\Vert
\boldsymbol{\tau}_{2}\right\Vert \left\Vert \boldsymbol{\tau}_{1}\right\Vert
\cos\theta_{\boldsymbol{\tau}_{2}\boldsymbol{\tau}_{1}}-\delta\left(
y\right)  \sum\nolimits_{i=1}^{l_{2}}\psi_{2_{i_{\ast}}}\nonumber\\
&  \equiv\frac{1}{2}\left\Vert \boldsymbol{\tau}\right\Vert _{\min_{c}}%
^{2}\text{,}\nonumber
\end{align}
over the decision space $Z$, where the area $P\left(  \mathbf{x}_{1_{i\ast}%
}|\boldsymbol{\tau}_{1}\right)  $ under $p\left(  \mathbf{x}_{1_{i\ast}%
}|\boldsymbol{\tau}_{1}\right)  $ and the area $P\left(  \mathbf{x}_{2_{i\ast
}}|\boldsymbol{\tau}_{2}\right)  $ under $p\left(  \mathbf{x}_{2_{i_{\ast}}%
}|\boldsymbol{\tau}_{2}\right)  $ are constrained to be equal to $\frac{1}%
{2}\left\Vert \boldsymbol{\tau}\right\Vert _{\min_{c}}^{2}$:%
\[
P\left(  \mathbf{x}_{1_{i\ast}}|\boldsymbol{\tau}_{1}\right)  \equiv P\left(
\mathbf{x}_{2_{i\ast}}|\boldsymbol{\tau}_{2}\right)  \equiv\frac{1}%
{2}\left\Vert \boldsymbol{\tau}\right\Vert _{\min_{c}}^{2}\text{.}%
\]

It follows that the linear eigenlocus discriminant function
$\widetilde{\Lambda}_{\boldsymbol{\tau}}\left(  \mathbf{x}\right)
=\boldsymbol{\tau}^{T}\mathbf{x}+\tau_{0}$ is the solution to the integral
equation%
\begin{align}
f\left(  \widetilde{\Lambda}_{\boldsymbol{\tau}}\left(  \mathbf{x}\right)
\right)  =  &  \int_{Z_{1}}\boldsymbol{\tau}_{1}d\boldsymbol{\tau}_{1}%
+\int_{Z_{2}}\boldsymbol{\tau}_{1}d\boldsymbol{\tau}_{1}+\delta\left(
y\right)  \sum\nolimits_{i=1}^{l_{1}}\psi_{1_{i_{\ast}}}%
\label{Linear Eigenlocus Integral Equation}\\
&  =\int_{Z_{1}}\boldsymbol{\tau}_{2}d\boldsymbol{\tau}_{2}+\int_{Z_{2}%
}\boldsymbol{\tau}_{2}d\boldsymbol{\tau}_{2}-\delta\left(  y\right)
\sum\nolimits_{i=1}^{l_{2}}\psi_{2_{i_{\ast}}}\text{,}\nonumber
\end{align}
over the decision space $Z=Z_{1}+Z_{2}$, where the dual likelihood ratios
$\widehat{\Lambda}_{\boldsymbol{\psi}}\left(  \mathbf{x}\right)
=\boldsymbol{\psi}_{1}+\boldsymbol{\psi}_{2}$ and $\widehat{\Lambda
}_{\boldsymbol{\tau}}\left(  \mathbf{x}\right)  =\boldsymbol{\tau}%
_{1}-\boldsymbol{\tau}_{2}$ are in statistical equilibrium, all of the forces
associated with the counter risks $\overline{\mathfrak{R}}_{\mathfrak{\min}%
}\left(  Z_{1}|p\left(  \widehat{\Lambda}_{\boldsymbol{\tau}}\left(
\mathbf{x}\right)  |\omega_{1}\right)  \right)  $ and the risks $\mathfrak{R}%
_{\mathfrak{\min}}\left(  Z_{2}|p\left(  \widehat{\Lambda}_{\boldsymbol{\tau}%
}\left(  \mathbf{x}\right)  |\omega_{1}\right)  \right)  $ in the $Z_{1}$ and
$Z_{2}$ decision regions: which are related to positions and potential
locations of extreme points $\mathbf{x}_{1_{i_{\ast}}}$ that are generated
according to $p\left(  \mathbf{x}|\omega_{1}\right)  $, are equal to all of
the forces associated with the risks $\mathfrak{R}_{\mathfrak{\min}}\left(
Z_{1}|p\left(  \widehat{\Lambda}_{\boldsymbol{\tau}}\left(  \mathbf{x}\right)
|\omega_{2}\right)  \right)  $ and the counter risks $\overline{\mathfrak{R}%
}_{\mathfrak{\min}}\left(  Z_{2}|p\left(  \widehat{\Lambda}_{\boldsymbol{\tau
}}\left(  \mathbf{x}\right)  |\omega_{2}\right)  \right)  $ in the $Z_{1}$ and
$Z_{2}$ decision regions: which are related to positions and potential
locations of extreme points $\mathbf{x}_{2_{i_{\ast}}}$ that are generated
according to $p\left(  \mathbf{x}|\omega_{2}\right)  $, and the eigenenergy
associated with the position or location of the parameter vector of
likelihoods $p\left(  \widehat{\Lambda}_{\boldsymbol{\tau}}\left(
\mathbf{x}\right)  |\omega_{1}\right)  $ given class $\omega_{1}$ is balanced
with the eigenenergy associated with the position or location of the parameter
vector of likelihoods $p\left(  \widehat{\Lambda}_{\boldsymbol{\tau}}\left(
\mathbf{x}\right)  |\omega_{2}\right)  $ given class $\omega_{2}$.

So, let $p\left(  \frac{\mathbf{x}_{1_{i\ast}}}{\left\Vert \mathbf{x}%
_{1_{i\ast}}\right\Vert }|\boldsymbol{\psi}_{1}\right)  $ and $p\left(
\frac{\mathbf{x}_{2_{i\ast}}}{\left\Vert \mathbf{x}_{2_{i\ast}}\right\Vert
}|\boldsymbol{\psi}_{2}\right)  $ denote the Wolfe dual parameter vectors of
likelihoods $\boldsymbol{\psi}_{1}=\sum\nolimits_{i=1}^{l_{1}}\psi
_{1_{i_{\ast}}}\frac{\mathbf{x}_{1_{i\ast}}}{\left\Vert \mathbf{x}_{1_{i\ast}%
}\right\Vert }$ and $\boldsymbol{\psi}_{2}=\sum\nolimits_{i=1}^{l_{2}}%
\psi_{2_{i_{\ast}}}\frac{\mathbf{x}_{2_{i\ast}}}{\left\Vert \mathbf{x}%
_{2_{i\ast}}\right\Vert }$. It follows that the class-conditional probability
density functions $p\left(  \frac{\mathbf{x}_{1_{i\ast}}}{\left\Vert
\mathbf{x}_{1_{i\ast}}\right\Vert }|\boldsymbol{\psi}_{1}\right)  $ and
$p\left(  \frac{\mathbf{x}_{2_{i\ast}}}{\left\Vert \mathbf{x}_{2_{i\ast}%
}\right\Vert }|\boldsymbol{\psi}_{2}\right)  $ in the Wolfe dual eigenspace
$\widetilde{Z}$ and the class-conditional probability density functions
$p\left(  \mathbf{x}_{1_{i\ast}}|\boldsymbol{\tau}_{1}\right)  $ and $p\left(
\mathbf{x}_{2_{i_{\ast}}}|\boldsymbol{\tau}_{2}\right)  $ in the decision
space $Z$ satisfy the integral equation%
\begin{align*}
f\left(  \widetilde{\Lambda}_{\boldsymbol{\tau}}\left(  \mathbf{x}\right)
\right)  =  &  \int_{Z}p\left(  \mathbf{x}_{1_{i\ast}}|\boldsymbol{\tau}%
_{1}\right)  d\boldsymbol{\tau}_{1}+\delta\left(  y\right)  p\left(
\frac{\mathbf{x}_{1_{i\ast}}}{\left\Vert \mathbf{x}_{1_{i\ast}}\right\Vert
}|\boldsymbol{\psi}_{1}\right) \\
&  =\int_{Z}p\left(  \mathbf{x}_{2_{i\ast}}|\boldsymbol{\tau}_{2}\right)
d\boldsymbol{\tau}_{2}-\delta\left(  y\right)  p\left(  \frac{\mathbf{x}%
_{2_{i\ast}}}{\left\Vert \mathbf{x}_{2_{i\ast}}\right\Vert }|\boldsymbol{\psi
}_{2}\right)  \text{,}%
\end{align*}
over the decision space $Z$, where $Z=Z_{1}+Z_{2}$, $Z_{1}$ and $Z_{2}$ are
contiguous decision regions, $Z_{1}\cong Z_{2}$, and $Z\subset%
\mathbb{R}
^{d}$.

Thus, it is concluded that the linear eigenlocus discriminant function%
\[
\widetilde{\Lambda}_{\boldsymbol{\tau}}\left(  \mathbf{x}\right)  =\left(
\mathbf{x}-\widehat{\mathbf{x}}_{i\ast}\right)  ^{T}\left(  \boldsymbol{\tau
}_{1}-\boldsymbol{\tau}_{2}\right)  \mathbf{+}\sum\nolimits_{i=1}^{l}%
y_{i}\left(  1-\xi_{i}\right)
\]
is the solution to the integral equation:%
\begin{align}
f\left(  \widetilde{\Lambda}_{\boldsymbol{\tau}}\left(  \mathbf{x}\right)
\right)  =  &  \int_{Z_{1}}p\left(  \mathbf{x}_{1_{i\ast}}|\boldsymbol{\tau
}_{1}\right)  d\boldsymbol{\tau}_{1}+\int_{Z_{2}}p\left(  \mathbf{x}%
_{1_{i\ast}}|\boldsymbol{\tau}_{1}\right)  d\boldsymbol{\tau}_{1}%
\label{Linear Eigenlocus Integral Equation I}\\
&  +\delta\left(  y\right)  p\left(  \frac{\mathbf{x}_{1_{i\ast}}}{\left\Vert
\mathbf{x}_{1_{i\ast}}\right\Vert }|\boldsymbol{\psi}_{1}\right) \nonumber\\
&  =\int_{Z_{1}}p\left(  \mathbf{x}_{2_{i\ast}}|\boldsymbol{\tau}_{2}\right)
d\boldsymbol{\tau}_{2}+\int_{Z_{2}}p\left(  \mathbf{x}_{2_{i\ast}%
}|\boldsymbol{\tau}_{2}\right)  d\boldsymbol{\tau}_{2}\nonumber\\
&  -\delta\left(  y\right)  p\left(  \frac{\mathbf{x}_{2_{i\ast}}}{\left\Vert
\mathbf{x}_{2_{i\ast}}\right\Vert }|\boldsymbol{\psi}_{2}\right)
\text{,}\nonumber
\end{align}
over the decision space $Z=Z_{1}+Z_{2}$, where $\delta\left(  y\right)
\triangleq\sum\nolimits_{i=1}^{l}y_{i}\left(  1-\xi_{i}\right)  $, the
integral $\int_{Z_{1}}p\left(  \mathbf{x}_{1_{i\ast}}|\boldsymbol{\tau}%
_{1}\right)  d\boldsymbol{\tau}_{1}$ accounts for all of the forces associated
with counter risks $\overline{\mathfrak{R}}_{\mathfrak{\min}}\left(
Z_{1}|\psi_{1i\ast}\mathbf{x}_{1_{i_{\ast}}}\right)  $ that are related to
positions and potential locations of corresponding $\mathbf{x}_{1_{i_{\ast}}}$
extreme points that lie in the $Z_{1}$ decision region, the integral
$\int_{Z_{2}}p\left(  \mathbf{x}_{1_{i\ast}}|\boldsymbol{\tau}_{1}\right)
d\boldsymbol{\tau}_{1}$ accounts for all of the forces associated with risks
$\mathfrak{R}_{\mathfrak{\min}}\left(  Z_{2}|\psi_{1i\ast}\mathbf{x}%
_{1_{i_{\ast}}}\right)  $ that are related to positions and potential
locations of corresponding $\mathbf{x}_{1_{i_{\ast}}}$ extreme points that lie
in the $Z_{2}$ decision region, the integral $\int_{Z_{1}}p\left(
\mathbf{x}_{2_{i\ast}}|\boldsymbol{\tau}_{2}\right)  d\boldsymbol{\tau}_{2}$
accounts for all of the forces associated with risks $\mathfrak{R}%
_{\mathfrak{\min}}\left(  Z_{1}|\psi_{2i\ast}\mathbf{x}_{2_{i_{\ast}}}\right)
$ that are related to positions and potential locations of corresponding
$\mathbf{x}_{2_{i_{\ast}}}$ extreme points that lie in the $Z_{1}$ decision
region, and the integral $\int_{Z_{2}}p\left(  \mathbf{x}_{2_{i\ast}%
}|\boldsymbol{\tau}_{2}\right)  d\boldsymbol{\tau}_{2}$ accounts for all of
the forces associated with counter risks $\overline{\mathfrak{R}%
}_{\mathfrak{\min}}\left(  Z_{2}|\psi_{2i\ast}\mathbf{x}_{2_{i_{\ast}}%
}\right)  $ that are related to positions and potential locations of
corresponding $\mathbf{x}_{2_{i_{\ast}}}$ extreme points that lie in the
$Z_{2}$ decision region.

The equalizer statistics $+\delta\left(  y\right)  p\left(  \frac
{\mathbf{x}_{1_{i\ast}}}{\left\Vert \mathbf{x}_{1_{i\ast}}\right\Vert
}|\boldsymbol{\psi}_{1}\right)  $ and $-\delta\left(  y\right)  p\left(
\frac{\mathbf{x}_{2_{i\ast}}}{\left\Vert \mathbf{x}_{2_{i\ast}}\right\Vert
}|\boldsymbol{\psi}_{2}\right)  $ ensure that the collective forces associated
with the risks $\mathfrak{R}_{\mathfrak{\min}}\left(  Z|\boldsymbol{\tau}%
_{1}\right)  $ and $\mathfrak{R}_{\mathfrak{\min}}\left(  Z|\boldsymbol{\tau
}_{2}\right)  $ for class $\omega_{1}$ and class $\omega_{2}$, which are given
by the respective integrals $\int_{Z}p\left(  \mathbf{x}_{1_{i\ast}%
}|\boldsymbol{\tau}_{1}\right)  d\boldsymbol{\tau}_{1}$ and $\int_{Z}p\left(
\mathbf{x}_{2_{i\ast}}|\boldsymbol{\tau}_{2}\right)  d\boldsymbol{\tau}_{2}$,
are symmetrically balanced with each other.

Therefore, the classification system%
\[
\left(  \mathbf{x}-\widehat{\mathbf{x}}_{i\ast}\right)  ^{T}\left(
\boldsymbol{\tau}_{1}-\boldsymbol{\tau}_{2}\right)  +\delta\left(  y\right)
\overset{\omega_{1}}{\underset{\omega_{2}}{\gtrless}}0
\]
is in statistical equilibrium:%
\begin{align}
f\left(  \widetilde{\Lambda}_{\boldsymbol{\tau}}\left(  \mathbf{x}\right)
\right)  :  &  \int_{Z_{1}}p\left(  \mathbf{x}_{1_{i\ast}}|\boldsymbol{\tau
}_{1}\right)  d\boldsymbol{\tau}_{1}-\int_{Z_{1}}p\left(  \mathbf{x}%
_{2_{i\ast}}|\boldsymbol{\tau}_{2}\right)  d\boldsymbol{\tau}_{2}%
\label{Linear Eigenlocus Integral Equation II}\\
&  +\delta\left(  y\right)  p\left(  \frac{\mathbf{x}_{1_{i\ast}}}{\left\Vert
\mathbf{x}_{1_{i\ast}}\right\Vert }|\boldsymbol{\psi}_{1}\right) \nonumber\\
&  =\int_{Z_{2}}p\left(  \mathbf{x}_{2_{i\ast}}|\boldsymbol{\tau}_{2}\right)
d\boldsymbol{\tau}_{2}-\int_{Z_{2}}p\left(  \mathbf{x}_{1_{i\ast}%
}|\boldsymbol{\tau}_{1}\right)  d\boldsymbol{\tau}_{1}\nonumber\\
&  -\delta\left(  y\right)  p\left(  \frac{\mathbf{x}_{2_{i\ast}}}{\left\Vert
\mathbf{x}_{2_{i\ast}}\right\Vert }|\boldsymbol{\psi}_{2}\right)
\text{,}\nonumber
\end{align}
where all of the forces associated with the counter risk $\overline
{\mathfrak{R}}_{\mathfrak{\min}}\left(  Z_{1}|\mathbf{\tau}_{1}\right)  $ for
class $\omega_{1}$ and the risk $\mathfrak{R}_{\mathfrak{\min}}\left(
Z_{1}|\boldsymbol{\tau}_{2}\right)  $ for class $\omega_{2}$ in the $Z_{1}$
decision region are symmetrically balanced with all of the forces associated
with the counter risk $\overline{\mathfrak{R}}_{\mathfrak{\min}}\left(
Z_{2}|\boldsymbol{\tau}_{2}\right)  $ for class $\omega_{2}$ and the risk
$\mathfrak{R}_{\mathfrak{\min}}\left(  Z_{2}|\mathbf{\tau}_{1}\right)  $ for
class $\omega_{1}$ in the $Z_{2}$ decision region, such that the risk
$\mathfrak{R}_{\mathfrak{\min}}\left(  Z|\widehat{\Lambda}_{\boldsymbol{\tau}%
}\left(  \mathbf{x}\right)  \right)  $ of the classification system is
minimized, and the eigenenergies associated with the counter risk
$\overline{\mathfrak{R}}_{\mathfrak{\min}}\left(  Z_{1}|\mathbf{\tau}%
_{1}\right)  $ for class $\omega_{1}$ and the risk $\mathfrak{R}%
_{\mathfrak{\min}}\left(  Z_{1}|\boldsymbol{\tau}_{2}\right)  $ for class
$\omega_{2}$ in the $Z_{1}$ decision region are balanced with the
eigenenergies associated with the counter risk $\overline{\mathfrak{R}%
}_{\mathfrak{\min}}\left(  Z_{2}|p\left(  \widehat{\Lambda}_{\boldsymbol{\tau
}}\left(  \mathbf{x}\right)  |\omega_{2}\right)  \right)  $ for class
$\omega_{2}$ and the risk $\mathfrak{R}_{\mathfrak{\min}}\left(
Z_{2}|p\left(  \widehat{\Lambda}_{\boldsymbol{\tau}}\left(  \mathbf{x}\right)
|\omega_{1}\right)  \right)  $ for class $\omega_{1}$ in the $Z_{2}$ decision
region:%
\begin{align*}
f\left(  \widetilde{\Lambda}_{\boldsymbol{\tau}}\left(  \mathbf{x}\right)
\right)   &  :E_{\min}\left(  Z_{1}|p\left(  \widehat{\Lambda}%
_{\boldsymbol{\tau}}\left(  \mathbf{x}\right)  |\omega_{1}\right)  \right)
-E_{\min}\left(  Z_{1}|p\left(  \widehat{\Lambda}_{\boldsymbol{\tau}}\left(
\mathbf{x}\right)  |\omega_{2}\right)  \right) \\
&  =E_{\min}\left(  Z_{2}|p\left(  \widehat{\Lambda}_{\boldsymbol{\tau}%
}\left(  \mathbf{x}\right)  |\omega_{2}\right)  \right)  -E_{\min}\left(
Z_{2}|p\left(  \widehat{\Lambda}_{\boldsymbol{\tau}}\left(  \mathbf{x}\right)
|\omega_{1}\right)  \right)
\end{align*}
such that the eigenenergy $E_{\min}\left(  Z|\widehat{\Lambda}%
_{\boldsymbol{\tau}}\left(  \mathbf{x}\right)  \right)  $ of the
classification system is minimized.

Thus, the locus of principal eigenaxis components on $\boldsymbol{\tau
}=\boldsymbol{\tau}_{1}-\boldsymbol{\tau}_{2}$ satisfies the integral equation%
\begin{align*}
f\left(  \widetilde{\Lambda}_{\boldsymbol{\tau}}\left(  \mathbf{x}\right)
\right)  =  &  \int_{Z}\boldsymbol{\tau}_{1}d\boldsymbol{\tau}_{1}%
+\delta\left(  y\right)  \frac{1}{2}\sum\nolimits_{i=1}^{l}\psi_{_{i\ast}}\\
&  =\int_{Z}\boldsymbol{\tau}_{2}d\boldsymbol{\tau}_{2}-\delta\left(
y\right)  \frac{1}{2}\sum\nolimits_{i=1}^{l}\psi_{_{i\ast}}\\
&  \equiv\frac{1}{2}\left\Vert \boldsymbol{\tau}\right\Vert _{\min_{c}}%
^{2}\text{,}%
\end{align*}
over the decision space $Z=Z_{1}+Z_{2}$, where $\boldsymbol{\tau}_{1}$ and
$\boldsymbol{\tau}_{2}$ are components of a principal eigenaxis
$\boldsymbol{\tau}$, and $\left\Vert \boldsymbol{\tau}\right\Vert _{\min_{c}%
}^{2}$ is the total allowed eigenenergy exhibited by $\boldsymbol{\tau}$.

The above integral equation can be written as:%
\begin{align}
f\left(  \widetilde{\Lambda}_{\boldsymbol{\tau}}\left(  \mathbf{x}\right)
\right)  =  &  \int_{Z_{1}}\boldsymbol{\tau}_{1}d\boldsymbol{\tau}_{1}%
+\int_{Z_{2}}\boldsymbol{\tau}_{1}d\boldsymbol{\tau}_{1}+\delta\left(
y\right)  \frac{1}{2}\sum\nolimits_{i=1}^{l}\psi_{_{i\ast}}%
\label{Linear Eigenlocus Integral Equation III}\\
&  =\int_{Z_{1}}\boldsymbol{\tau}_{2}d\boldsymbol{\tau}_{2}+\int_{Z_{2}%
}\boldsymbol{\tau}_{2}d\boldsymbol{\tau}_{2}-\delta\left(  y\right)  \frac
{1}{2}\sum\nolimits_{i=1}^{l}\psi_{_{i\ast}}\text{,}\nonumber
\end{align}
where the integral $\int_{Z_{1}}\boldsymbol{\tau}_{1}d\boldsymbol{\tau}_{1}$
accounts for all of the eigenenergies $\left\Vert \psi_{1_{i_{\ast}}%
}\mathbf{x}_{1_{i_{\ast}}}\right\Vert _{\min_{c}}^{2}$ exhibited by all of the
$\mathbf{x}_{1_{i_{\ast}}}$ extreme points that lie in the $Z_{1}$ decision
region, the integral $\int_{Z_{2}}\boldsymbol{\tau}_{1}d\boldsymbol{\tau}_{1}$
accounts for all of the eigenenergies $\left\Vert \psi_{1_{i_{\ast}}%
}\mathbf{x}_{1_{i_{\ast}}}\right\Vert _{\min_{c}}^{2}$ exhibited by all of the
$\mathbf{x}_{1_{i_{\ast}}}$ extreme points that lie in the $Z_{2}$ decision
region, the integral $\int_{Z_{1}}\boldsymbol{\tau}_{2}d\boldsymbol{\tau}_{2}$
accounts for all of the eigenenergies $\left\Vert \psi_{2_{i_{\ast}}%
}\mathbf{x}_{2_{i_{\ast}}}\right\Vert _{\min_{c}}^{2}$ exhibited by all of the
$\mathbf{x}_{2_{i_{\ast}}}$ extreme points that lie in the $Z_{2}$ decision
region, and the integral $\int_{Z_{2}}\boldsymbol{\tau}_{2}d\boldsymbol{\tau
}_{2}$ accounts for all of the eigenenergies $\left\Vert \psi_{2_{i_{\ast}}%
}\mathbf{x}_{2_{i_{\ast}}}\right\Vert _{\min_{c}}^{2}$ exhibited by all of the
$\mathbf{x}_{2_{i_{\ast}}}$ extreme points that lie in the $Z_{1}$ decision
region. The equalizer statistics $+\delta\left(  y\right)  \frac{1}{2}%
\sum\nolimits_{i=1}^{l}\psi_{_{i\ast}}$ and $-\delta\left(  y\right)  \frac
{1}{2}\sum\nolimits_{i=1}^{l}\psi_{_{i\ast}}$ ensure that the integrals
$\int_{Z}\boldsymbol{\tau}_{1}d\boldsymbol{\tau}_{1}$ and $\int_{Z}%
\boldsymbol{\tau}_{2}d\boldsymbol{\tau}_{2}$ are symmetrically balanced with
each other.

Equation (\ref{Linear Eigenlocus Integral Equation III}) can be rewritten as%
\begin{align}
f\left(  \widetilde{\Lambda}_{\boldsymbol{\tau}}\left(  \mathbf{x}\right)
\right)  =  &  \int_{Z_{1}}\boldsymbol{\tau}_{1}d\boldsymbol{\tau}_{1}%
-\int_{Z_{1}}\boldsymbol{\tau}_{2}d\boldsymbol{\tau}_{2}+\delta\left(
y\right)  \frac{1}{2}\sum\nolimits_{i=1}^{l}\psi_{_{i\ast}}%
\label{Linear Eigenlocus Integral Equation IV}\\
&  =\int_{Z_{2}}\boldsymbol{\tau}_{2}d\boldsymbol{\tau}_{2}-\int_{Z_{2}%
}\boldsymbol{\tau}_{1}d\boldsymbol{\tau}_{1}-\delta\left(  y\right)  \frac
{1}{2}\sum\nolimits_{i=1}^{l}\psi_{_{i\ast}}\text{,}\nonumber
\end{align}
where all of the eigenenergies $\left\Vert \psi_{1_{i_{\ast}}}\mathbf{x}%
_{1_{i_{\ast}}}\right\Vert _{\min_{c}}^{2}$ and $\left\Vert \psi_{2_{i_{\ast}%
}}\mathbf{x}_{2_{i_{\ast}}}\right\Vert _{\min_{c}}^{2}$ associated with the
counter risk $\overline{\mathfrak{R}}_{\mathfrak{\min}}\left(  Z_{1}%
|\mathbf{\tau}_{1}\right)  $ and the risk $\mathfrak{R}_{\mathfrak{\min}%
}\left(  Z_{1}|\boldsymbol{\tau}_{2}\right)  $ in the $Z_{1}$ decision region
are \emph{symmetrically balanced} with all of the eigenenergies $\left\Vert
\psi_{1_{i_{\ast}}}\mathbf{x}_{1_{i_{\ast}}}\right\Vert _{\min_{c}}^{2}$ and
$\left\Vert \psi_{2_{i_{\ast}}}\mathbf{x}_{2_{i_{\ast}}}\right\Vert _{\min
_{c}}^{2}$ associated with the counter risk $\overline{\mathfrak{R}%
}_{\mathfrak{\min}}\left(  Z_{2}|\boldsymbol{\tau}_{2}\right)  $ and the risk
$\mathfrak{R}_{\mathfrak{\min}}\left(  Z_{2}|\mathbf{\tau}_{1}\right)  $ in
the $Z_{2}$ decision region.

Given Eqs (\ref{Linear Eigenlocus Integral Equation I}) -
(\ref{Linear Eigenlocus Integral Equation IV}), it is concluded that linear
eigenlocus discriminant functions $\widehat{\Lambda}_{\boldsymbol{\tau}%
}\left(  \mathbf{x}\right)  =\boldsymbol{\tau}^{T}\mathbf{x}+\tau_{0}$ satisfy
discrete and data-driven versions of the integral equation of binary
classification in Eq.
(\ref{Integral Equation of Likelihood Ratio and Decision Boundary}), the
fundamental integral equation of binary classification for a classification
system in statistical equilibrium in Eq. (\ref{Equalizer Rule}), and the
corresponding integral equation for a classification system in statistical
equilibrium in Eq. (\ref{Balancing of Bayes' Risks and Counteracting Risks}).

\subsection{Equilibrium Points of Integral Equations I}

Returning to the binary classification theorem, recall that the risk
$\mathfrak{R}_{\mathfrak{\min}}\left(  Z|\widehat{\Lambda}\left(
\mathbf{x}\right)  \right)  $ and the corresponding eigenenergy $E_{\min
}\left(  Z|\widehat{\Lambda}\left(  \mathbf{x}\right)  \right)  $ of a
classification system $p\left(  \widehat{\Lambda}\left(  \mathbf{x}\right)
|\omega_{1}\right)  -p\left(  \widehat{\Lambda}\left(  \mathbf{x}\right)
|\omega_{2}\right)  \overset{\omega_{1}}{\underset{\omega_{2}}{\gtrless}}0$
are governed by the equilibrium point%
\[
p\left(  \widehat{\Lambda}\left(  \mathbf{x}\right)  |\omega_{1}\right)
-p\left(  \widehat{\Lambda}\left(  \mathbf{x}\right)  |\omega_{2}\right)  =0
\]
of the integral equation%
\begin{align*}
f\left(  \widehat{\Lambda}\left(  \mathbf{x}\right)  \right)   &  =\int%
_{Z_{1}}p\left(  \widehat{\Lambda}\left(  \mathbf{x}\right)  |\omega
_{2}\right)  d\widehat{\Lambda}+\int_{Z_{2}}p\left(  \widehat{\Lambda}\left(
\mathbf{x}\right)  |\omega_{2}\right)  d\widehat{\Lambda}\\
&  =\int_{Z_{2}}p\left(  \widehat{\Lambda}\left(  \mathbf{x}\right)
|\omega_{1}\right)  d\widehat{\Lambda}+\int_{Z_{1}}p\left(  \widehat{\Lambda
}\left(  \mathbf{x}\right)  |\omega_{1}\right)  d\widehat{\Lambda}\text{,}%
\end{align*}
over the decision space $Z=Z_{1}+Z_{2}$, where the equilibrium point $p\left(
\widehat{\Lambda}\left(  \mathbf{x}\right)  |\omega_{1}\right)  -p\left(
\widehat{\Lambda}\left(  \mathbf{x}\right)  |\omega_{2}\right)  =0$ is the
focus of a decision boundary $D\left(  \mathbf{x}\right)  $.

Returning to Eq. (\ref{Wolfe Dual Equilibrium Point}):%
\[
\sum\nolimits_{i=1}^{l_{1}}\psi_{1i\ast}-\sum\nolimits_{i=1}^{l_{2}}%
\psi_{2i\ast}=0\text{,}%
\]
it follows that the Wolfe dual linear eigenlocus $\boldsymbol{\psi}$ of
likelihoods and principal eigenaxis components $\widehat{\Lambda
}_{\boldsymbol{\psi}}\left(  \mathbf{x}\right)  $%
\begin{align*}
\boldsymbol{\psi}  &  =\sum\nolimits_{i=1}^{l}\psi_{i\ast}\frac{\mathbf{x}%
_{i\ast}}{\left\Vert \mathbf{x}_{i\ast}\right\Vert }\\
&  =\sum\nolimits_{i=1}^{l_{1}}\psi_{1i\ast}\frac{\mathbf{x}_{1_{i\ast}}%
}{\left\Vert \mathbf{x}_{1_{i\ast}}\right\Vert }+\sum\nolimits_{i=1}^{l_{2}%
}\psi_{2i\ast}\frac{\mathbf{x}_{2_{i\ast}}}{\left\Vert \mathbf{x}_{2_{i\ast}%
}\right\Vert }\\
&  =\boldsymbol{\psi}_{1}+\boldsymbol{\psi}_{2}%
\end{align*}
is the \emph{equilibrium point} $p\left(  \widehat{\Lambda}_{\boldsymbol{\psi
}}\left(  \mathbf{x}\right)  |\omega_{1}\right)  -p\left(  \widehat{\Lambda
}_{\boldsymbol{\psi}}\left(  \mathbf{x}\right)  |\omega_{2}\right)  =0$:%
\[
\sum\nolimits_{i=1}^{l_{1}}\psi_{1i\ast}\frac{\mathbf{x}_{1_{i\ast}}%
}{\left\Vert \mathbf{x}_{1_{i\ast}}\right\Vert }-\sum\nolimits_{i=1}^{l_{2}%
}\psi_{2i\ast}\frac{\mathbf{x}_{2_{i\ast}}}{\left\Vert \mathbf{x}_{2_{i\ast}%
}\right\Vert }=0
\]
of the integral equation in Eq. (\ref{Linear Eigenlocus Integral Equation})
and all of its derivatives in Eqs (\ref{Linear Eigenlocus Integral Equation I}%
) - (\ref{Linear Eigenlocus Integral Equation IV}).

Therefore, it is concluded that the risk $\mathfrak{R}_{\mathfrak{\min}%
}\left(  Z|\widehat{\Lambda}_{\boldsymbol{\tau}}\left(  \mathbf{x}\right)
\right)  $ and the eigenenergy $E_{\min}\left(  Z|\widehat{\Lambda
}_{\boldsymbol{\tau}}\left(  \mathbf{x}\right)  \right)  $ of the
classification system $\boldsymbol{\tau}^{T}\left(  \mathbf{x}%
-\widehat{\mathbf{x}}_{i\ast}\right)  +\delta\left(  y\right)  \overset{\omega
_{1}}{\underset{\omega_{2}}{\gtrless}}0$ are governed by the equilibrium point
$p\left(  \widehat{\Lambda}_{\boldsymbol{\psi}}\left(  \mathbf{x}\right)
|\omega_{1}\right)  -p\left(  \widehat{\Lambda}_{\boldsymbol{\psi}}\left(
\mathbf{x}\right)  |\omega_{2}\right)  =0$:%
\[
\sum\nolimits_{i=1}^{l_{1}}\psi_{1i\ast}\frac{\mathbf{x}_{1_{i\ast}}%
}{\left\Vert \mathbf{x}_{1_{i\ast}}\right\Vert }-\sum\nolimits_{i=1}^{l_{2}%
}\psi_{2i\ast}\frac{\mathbf{x}_{2_{i\ast}}}{\left\Vert \mathbf{x}_{2_{i\ast}%
}\right\Vert }=0
\]
of the integral equation $f\left(  \widetilde{\Lambda}_{\boldsymbol{\tau}%
}\left(  \mathbf{x}\right)  \right)  $:%
\begin{align*}
f\left(  \widetilde{\Lambda}_{\boldsymbol{\tau}}\left(  \mathbf{x}\right)
\right)  =  &  \int_{Z_{1}}p\left(  \mathbf{x}_{1_{i\ast}}|\boldsymbol{\tau
}_{1}\right)  d\boldsymbol{\tau}_{1}+\int_{Z_{2}}p\left(  \mathbf{x}%
_{1_{i\ast}}|\boldsymbol{\tau}_{1}\right)  d\boldsymbol{\tau}_{1}\\
&  +\delta\left(  y\right)  p\left(  \frac{\mathbf{x}_{1_{i\ast}}}{\left\Vert
\mathbf{x}_{1_{i\ast}}\right\Vert }|\boldsymbol{\psi}_{1}\right) \\
&  =\int_{Z_{1}}p\left(  \mathbf{x}_{2_{i\ast}}|\boldsymbol{\tau}_{2}\right)
d\boldsymbol{\tau}_{2}+\int_{Z_{2}}p\left(  \mathbf{x}_{2_{i\ast}%
}|\boldsymbol{\tau}_{2}\right)  d\boldsymbol{\tau}_{2}\\
&  -\delta\left(  y\right)  p\left(  \frac{\mathbf{x}_{2_{i\ast}}}{\left\Vert
\mathbf{x}_{2_{i\ast}}\right\Vert }|\boldsymbol{\psi}_{2}\right)  \text{,}%
\end{align*}
over the decision space $Z=Z_{1}+Z_{2}$, where the equilibrium point
$\sum\nolimits_{i=1}^{l_{1}}\psi_{1i\ast}\frac{\mathbf{x}_{1_{i\ast}}%
}{\left\Vert \mathbf{x}_{1_{i\ast}}\right\Vert }-\sum\nolimits_{i=1}^{l_{2}%
}\psi_{2i\ast}\frac{\mathbf{x}_{2_{i\ast}}}{\left\Vert \mathbf{x}_{2_{i\ast}%
}\right\Vert }=0$ is the dual focus of a linear decision boundary $D\left(
\mathbf{x}\right)  $.

I will now develop an integral equation that explicitly accounts for the
primal focus $\widehat{\Lambda}_{\boldsymbol{\tau}}\left(  \mathbf{x}\right)
$ of a linear decision boundary $D\left(  \mathbf{x}\right)  $ and the
equilibrium point $\widehat{\Lambda}_{\boldsymbol{\psi}}\left(  \mathbf{x}%
\right)  $ of the integral equation $f\left(  \widetilde{\Lambda
}_{\boldsymbol{\tau}}\left(  \mathbf{x}\right)  \right)  $.

\section{The Balancing Feat in Dual Space I}

Let $\boldsymbol{\tau=\tau}_{1}-\boldsymbol{\tau}_{2}$ and substitute the
statistic for $\tau_{0}$ in Eq. (\ref{Eigenlocus Projection Factor Two})%
\begin{align*}
\tau_{0}  &  =-\sum\nolimits_{i=1}^{l}\mathbf{x}_{i\ast}^{T}\sum
\nolimits_{j=1}^{l_{1}}\psi_{1_{j_{\ast}}}\mathbf{x}_{1_{j_{\ast}}}\\
&  +\sum\nolimits_{i=1}^{l}\mathbf{x}_{i\ast}^{T}\sum\nolimits_{j=1}^{l_{2}%
}\psi_{2_{j_{\ast}}}\mathbf{x}_{2_{j_{\ast}}}+\sum\nolimits_{i=1}^{l}%
y_{i}\left(  1-\xi_{i}\right)
\end{align*}
into Eq. (\ref{Minimum Eigenenergy Class One})%
\[
\psi_{1_{i_{\ast}}}\mathbf{x}_{1_{i_{\ast}}}^{T}\boldsymbol{\tau}%
=\psi_{1_{i_{\ast}}}\left(  1-\xi_{i}-\tau_{0}\right)  ,\ i=1,...,l_{1}%
\text{,}%
\]
where each conditional density $\psi_{1_{i_{\ast}}}\frac{\mathbf{x}%
_{1_{i_{\ast}}}}{\left\Vert \mathbf{x}_{1_{i_{\ast}}}\right\Vert }$ of an
$\mathbf{x}_{1_{i_{\ast}}}$ extreme point satisfies the identity:%
\[
\psi_{1_{i_{\ast}}}\left(  1-\xi_{i}\right)  \equiv\psi_{1_{i_{\ast}}%
}\mathbf{x}_{1_{i_{\ast}}}^{T}\boldsymbol{\tau}+\psi_{1_{i_{\ast}}}\tau
_{0}\text{.}%
\]

Accordingly, the above identity can be rewritten in terms of an eigenlocus
equation that is satisfied by the conditional density $\psi_{1_{i_{\ast}}%
}\frac{\mathbf{x}_{1_{i_{\ast}}}}{\left\Vert \mathbf{x}_{1_{i_{\ast}}%
}\right\Vert }$ of an $\mathbf{x}_{1_{i_{\ast}}}$ extreme point:%
\begin{align}
\psi_{1_{i_{\ast}}}  &  =\psi_{1_{i\ast}}\mathbf{x}_{1_{i\ast}}^{T}\left(
\boldsymbol{\tau}_{1}-\boldsymbol{\tau}_{2}\right)
\label{Pointwise Conditional Density Constraint One}\\
&  +\psi_{1_{i_{\ast}}}\left\{  \sum\nolimits_{j=1}^{l}\mathbf{x}_{j\ast}%
^{T}\left(  \boldsymbol{\tau}_{2}-\boldsymbol{\tau}_{1}\right)  \right\}
\nonumber\\
&  +\xi_{i}\psi_{1_{i_{\ast}}}+\delta\left(  y\right)  \psi_{1_{i_{\ast}}%
}\text{,}\nonumber
\end{align}
where $\delta\left(  y\right)  \triangleq\sum\nolimits_{i=1}^{l}y_{i}\left(
1-\xi_{i}\right)  $, $\psi_{1_{i\ast}}\mathbf{x}_{1_{i\ast}}$ is a principal
eigenaxis component on $\boldsymbol{\tau}_{1}$, and the set of scaled extreme
vectors\emph{\ }$\psi_{1_{i_{\ast}}}\left(  \sum\nolimits_{j=1}^{l}%
\mathbf{x}_{j\ast}\right)  $ are symmetrically distributed over
$\boldsymbol{\tau}_{2}-\boldsymbol{\tau}_{1}$:%
\[
\psi_{1_{i_{\ast}}}\left(  \sum\nolimits_{j=1}^{l}\mathbf{x}_{j\ast}%
^{T}\right)  \boldsymbol{\tau}_{2}-\psi_{1_{i_{\ast}}}\left(  \sum
\nolimits_{j=1}^{l}\mathbf{x}_{j\ast}^{T}\right)  \boldsymbol{\tau}%
_{1}\text{.}%
\]

Again, let $\boldsymbol{\tau=\tau}_{1}-\boldsymbol{\tau}_{2}$. Substitute the
statistic for $\tau_{0}$ in Eq. (\ref{Eigenlocus Projection Factor Two})%
\begin{align*}
\tau_{0}  &  =-\sum\nolimits_{i=1}^{l}\mathbf{x}_{i\ast}^{T}\sum
\nolimits_{j=1}^{l_{1}}\psi_{1_{j_{\ast}}}\mathbf{x}_{1_{j_{\ast}}}\\
&  +\sum\nolimits_{i=1}^{l}\mathbf{x}_{i\ast}^{T}\sum\nolimits_{j=1}^{l_{2}%
}\psi_{2_{j_{\ast}}}\mathbf{x}_{2_{j_{\ast}}}+\sum\nolimits_{i=1}^{l}%
y_{i}\left(  1-\xi_{i}\right)
\end{align*}
into Eq. (\ref{Minimum Eigenenergy Class Two})%
\[
-\psi_{2_{i_{\ast}}}\mathbf{x}_{2_{i_{\ast}}}^{T}\boldsymbol{\tau}%
=\psi_{2_{i_{\ast}}}\left(  1-\xi_{i}+\tau_{0}\right)  ,\text{\ }%
i=1,...,l_{2}\text{,}%
\]
where each conditional density $\psi_{2_{i_{\ast}}}\frac{\mathbf{x}%
_{2_{i_{\ast}}}}{\left\Vert \mathbf{x}_{2_{i_{\ast}}}\right\Vert }$ of an
$\mathbf{x}_{2_{i_{\ast}}}$ extreme point satisfies the identity:%
\[
\psi_{2_{i_{\ast}}}\left(  1-\xi_{i}\right)  =-\psi_{2_{i_{\ast}}}%
\mathbf{x}_{2_{i_{\ast}}}^{T}\boldsymbol{\tau}-\psi_{2_{i_{\ast}}}\tau
_{0}\text{.}%
\]

Accordingly, the above identity can be rewritten in terms of an eigenlocus
equation that is satisfied by the conditional density $\psi_{2_{i_{\ast}}%
}\frac{\mathbf{x}_{2_{i_{\ast}}}}{\left\Vert \mathbf{x}_{2_{i_{\ast}}%
}\right\Vert }$ of an $\mathbf{x}_{2_{i_{\ast}}}$ extreme point:%
\begin{align}
\psi_{2_{i_{\ast}}}  &  =\psi_{2_{i_{\ast}}}\mathbf{x}_{2_{i_{\ast}}}%
^{T}\left(  \boldsymbol{\tau}_{2}-\boldsymbol{\tau}_{1}\right)
\label{Pointwise Conditional Density Constraint Two}\\
&  +\psi_{2_{i_{\ast}}}\left\{  \sum\nolimits_{j=1}^{l}\mathbf{x}_{j\ast}%
^{T}\left(  \boldsymbol{\tau}_{1}-\boldsymbol{\tau}_{2}\right)  \right\}
\nonumber\\
&  +\xi_{i}\psi_{2_{i_{\ast}}}-\delta\left(  y\right)  \psi_{2_{i_{\ast}}%
}\text{,}\nonumber
\end{align}
where $\delta\left(  y\right)  \triangleq\sum\nolimits_{i=1}^{l}y_{i}\left(
1-\xi_{i}\right)  $, $\psi_{2_{i_{\ast}}}\mathbf{x}_{2_{i_{\ast}}}$ is a
principal eigenaxis component on $\boldsymbol{\tau}_{2}$, and the set of
scaled extreme vectors $\psi_{2_{i_{\ast}}}\left(  \sum\nolimits_{j=1}%
^{l}\mathbf{x}_{j\ast}\right)  $ are symmetrically distributed over
$\boldsymbol{\tau}_{1}-\boldsymbol{\tau}_{2}$:%
\[
\psi_{2_{i_{\ast}}}\left(  \sum\nolimits_{j=1}^{l}\mathbf{x}_{j\ast}%
^{T}\right)  \boldsymbol{\tau}_{1}-\psi_{2_{i_{\ast}}}\left(  \sum
\nolimits_{j=1}^{l}\mathbf{x}_{j\ast}^{T}\right)  \boldsymbol{\tau}%
_{2}\text{.}%
\]

Using Eqs (\ref{Integral Equation Class One}) and
(\ref{Pointwise Conditional Density Constraint One}), it follows that the
conditional probability $P\left(  \mathbf{x}_{1_{i\ast}}|\boldsymbol{\tau}%
_{1}\right)  $ of observing the set $\left\{  \mathbf{x}_{1_{i\ast}}\right\}
_{i=1}^{l_{1}}$ of $\mathbf{x}_{1_{i\ast}}$ extreme points within localized
regions of the decision space $Z$ is determined by the eigenlocus equation:%
\begin{align}
P\left(  \mathbf{x}_{1_{i\ast}}|\boldsymbol{\tau}_{1}\right)   &
=\sum\nolimits_{i=1}^{l_{1}}\psi_{1_{i\ast}}\mathbf{x}_{1_{i\ast}}^{T}\left(
\boldsymbol{\tau}_{1}-\boldsymbol{\tau}_{2}\right) \label{Bayes' Risk One}\\
&  +\sum\nolimits_{i=1}^{l_{1}}\psi_{1_{i_{\ast}}}\left\{  \sum\nolimits_{j=1}%
^{l}\mathbf{x}_{j\ast}^{T}\left(  \boldsymbol{\tau}_{2}-\boldsymbol{\tau}%
_{1}\right)  \right\} \nonumber\\
&  +\delta\left(  y\right)  \sum\nolimits_{i=1}^{l_{1}}\psi_{1_{i_{\ast}}%
}+\sum\nolimits_{i=1}^{l_{1}}\xi_{i}\psi_{1_{i_{\ast}}}\nonumber\\
&  \equiv\sum\nolimits_{i=1}^{l_{1}}\psi_{1_{i_{\ast}}}\text{,}\nonumber
\end{align}
where $P\left(  \mathbf{x}_{1_{i\ast}}|\boldsymbol{\tau}_{1}\right)  $
evaluates to $\sum\nolimits_{i=1}^{l_{1}}\psi_{1_{i_{\ast}}}$, and scaled
extreme vectors are symmetrically distributed over $\boldsymbol{\tau}%
_{2}-\boldsymbol{\tau}_{1}$ in the following manner:%
\[
\left(  \sum\nolimits_{i=1}^{l_{1}}\psi_{1_{i_{\ast}}}\sum\nolimits_{j=1}%
^{l}\mathbf{x}_{j\ast}^{T}\right)  \boldsymbol{\tau}_{2}-\left(
\sum\nolimits_{i=1}^{l_{1}}\psi_{1_{i_{\ast}}}\sum\nolimits_{j=1}%
^{l}\mathbf{x}_{j\ast}^{T}\right)  \boldsymbol{\tau}_{1}\text{.}%
\]

Using Eqs (\ref{Integral Equation Class Two}) and
(\ref{Pointwise Conditional Density Constraint Two}), it follows that the
conditional probability $P\left(  \mathbf{x}_{2_{i\ast}}|\boldsymbol{\tau}%
_{2}\right)  $ of observing the set $\left\{  \mathbf{x}_{2_{i\ast}}\right\}
_{i=1}^{l_{2}}$ of $\mathbf{x}_{2_{i\ast}}$ extreme points within localized
regions of the decision space $Z$ is determined by the eigenlocus equation:%
\begin{align}
P\left(  \mathbf{x}_{2_{i\ast}}|\boldsymbol{\tau}_{2}\right)   &
=\sum\nolimits_{i=1}^{l_{2}}\psi_{2_{i_{\ast}}}\mathbf{x}_{2_{i_{\ast}}}%
^{T}\left(  \boldsymbol{\tau}_{2}-\boldsymbol{\tau}_{1}\right)
\label{Bayes' Risk Two}\\
&  -\sum\nolimits_{i=1}^{l_{2}}\psi_{2_{i_{\ast}}}\left\{  \sum\nolimits_{j=1}%
^{l}\mathbf{x}_{j\ast}^{T}\left(  \boldsymbol{\tau}_{1}-\boldsymbol{\tau}%
_{2}\right)  \right\} \nonumber\\
&  -\delta\left(  y\right)  \sum\nolimits_{i=1}^{l_{2}}\psi_{2_{i_{\ast}}%
}+\sum\nolimits_{i=1}^{l_{2}}\xi_{i}\psi_{2_{i_{\ast}}}\nonumber\\
&  \equiv\sum\nolimits_{i=1}^{l_{2}}\psi_{2_{i_{\ast}}}\text{,}\nonumber
\end{align}
where $P\left(  \mathbf{x}_{2_{i\ast}}|\boldsymbol{\tau}_{2}\right)  $
evaluates to $\sum\nolimits_{i=1}^{l_{2}}\psi_{2_{i_{\ast}}}$, and scaled
extreme vectors are symmetrically distributed over $\boldsymbol{\tau}%
_{1}-\boldsymbol{\tau}_{2}$ in the following manner:%
\[
\left(  \sum\nolimits_{i=1}^{l_{2}}\psi_{2_{i_{\ast}}}\sum\nolimits_{j=1}%
^{l}\mathbf{x}_{j\ast}^{T}\right)  \boldsymbol{\tau}_{1}-\left(
\sum\nolimits_{i=1}^{l_{2}}\psi_{2_{i_{\ast}}}\sum\nolimits_{j=1}%
^{l}\mathbf{x}_{j\ast}^{T}\right)  \boldsymbol{\tau}_{2}\text{.}%
\]

I will now use Eqs (\ref{Bayes' Risk One}) and (\ref{Bayes' Risk Two}) to
devise an equilibrium equation that determines the overall manner in which
linear eigenlocus discriminant functions $\widehat{\Lambda}_{\boldsymbol{\tau
}}\left(  \mathbf{x}\right)  =\boldsymbol{\tau}^{T}\mathbf{x}+\tau_{0}$
minimize the risk $\mathfrak{R}_{\mathfrak{\min}}\left(  Z|\boldsymbol{\tau
}\right)  $ and the total allowed eigenenergy $\left\Vert \boldsymbol{\tau
}\right\Vert _{\min_{c}}^{2}$ for a given decision space $Z$.

\subsection{Minimization of Risk $\mathfrak{R}_{\mathfrak{\min}}\left(
Z|\boldsymbol{\tau}\right)  $ and Eigenenergy$\left\Vert \boldsymbol{\tau
}\right\Vert _{\min_{c}}^{2}$}

Take the estimates in Eqs (\ref{Bayes' Risk One}) and (\ref{Bayes' Risk Two}):%
\[
P\left(  \mathbf{x}_{1_{i\ast}}|\boldsymbol{\tau}_{1}\right)  =\sum
\nolimits_{i=1}^{l_{1}}\psi_{1_{i_{\ast}}}%
\]
and%
\[
P\left(  \mathbf{x}_{2_{i\ast}}|\boldsymbol{\tau}_{2}\right)  =\sum
\nolimits_{i=1}^{l_{2}}\psi_{2_{i_{\ast}}}%
\]
for the conditional probabilities $P\left(  \mathbf{x}_{1_{i\ast}%
}|\boldsymbol{\tau}_{1}\right)  $ and $P\left(  \mathbf{x}_{2_{i\ast}%
}|\boldsymbol{\tau}_{2}\right)  $ of observing the $\mathbf{x}_{1_{i\ast}}$
and $\mathbf{x}_{2_{i\ast}}$ extreme points within localized regions of the
decision space $Z$.

Given that the Wolfe dual eigenlocus of principal eigenaxis components and
likelihoods:%
\begin{align*}
\widehat{\Lambda}_{\boldsymbol{\psi}}\left(  \mathbf{x}\right)   &
=\boldsymbol{\psi}_{1}+\boldsymbol{\psi}_{2}\\
&  =\sum\nolimits_{i=1}^{l_{1}}\psi_{1_{i_{\ast}}}\frac{\mathbf{x}_{1_{i\ast}%
}}{\left\Vert \mathbf{x}_{1_{i\ast}}\right\Vert }+\sum\nolimits_{i=1}^{l_{2}%
}\psi_{2_{i_{\ast}}}\frac{\mathbf{x}_{2_{i\ast}}}{\left\Vert \mathbf{x}%
_{2_{i\ast}}\right\Vert }%
\end{align*}
satisfies the equilibrium equation:%
\[
\sum\nolimits_{i=1}^{l_{1}}\psi_{1_{i_{\ast}}}\frac{\mathbf{x}_{1_{i\ast}}%
}{\left\Vert \mathbf{x}_{1_{i\ast}}\right\Vert }=\sum\nolimits_{i=1}^{l_{2}%
}\psi_{2_{i_{\ast}}}\frac{\mathbf{x}_{2_{i\ast}}}{\left\Vert \mathbf{x}%
_{2_{i\ast}}\right\Vert }\text{,}%
\]
it follows that the conditional probabilities of observing the $\mathbf{x}%
_{1_{i\ast}}$ and the $\mathbf{x}_{2_{i\ast}}$ extreme points within localized
regions of the decision space $Z$ are equal to each other:%
\[
P\left(  \mathbf{x}_{1_{i\ast}}|\boldsymbol{\tau}_{1}\right)  =P\left(
\mathbf{x}_{2_{i\ast}}|\boldsymbol{\tau}_{2}\right)  \text{.}%
\]

Accordingly, set the vector expressions in Eqs (\ref{Bayes' Risk One}) and
(\ref{Bayes' Risk Two}) equal to each other:%
\begin{align*}
&  \sum\nolimits_{i=1}^{l_{1}}\psi_{1_{i\ast}}\mathbf{x}_{1_{i\ast}}%
^{T}\left(  \boldsymbol{\tau}_{1}-\boldsymbol{\tau}_{2}\right)  +\sum
\nolimits_{i=1}^{l_{1}}\psi_{1_{i_{\ast}}}\left\{  \sum\nolimits_{j=1}%
^{l}\mathbf{x}_{j\ast}^{T}\left(  \boldsymbol{\tau}_{2}-\boldsymbol{\tau}%
_{1}\right)  \right\} \\
&  +\delta\left(  y\right)  \sum\nolimits_{i=1}^{l_{1}}\psi_{1_{i_{\ast}}%
}+\sum\nolimits_{i=1}^{l_{1}}\xi_{i}\psi_{1_{i_{\ast}}}\\
&  =\sum\nolimits_{i=1}^{l_{2}}\psi_{2_{i_{\ast}}}\mathbf{x}_{2_{i_{\ast}}%
}^{T}\left(  \boldsymbol{\tau}_{2}-\boldsymbol{\tau}_{1}\right)
-\sum\nolimits_{i=1}^{l_{2}}\psi_{2_{i_{\ast}}}\left\{  \sum\nolimits_{j=1}%
^{l}\mathbf{x}_{j\ast}^{T}\left(  \boldsymbol{\tau}_{1}-\boldsymbol{\tau}%
_{2}\right)  \right\} \\
&  -\delta\left(  y\right)  \sum\nolimits_{i=1}^{l_{2}}\psi_{2_{i_{\ast}}%
}+\sum\nolimits_{i=1}^{l_{2}}\xi_{i}\psi_{2_{i_{\ast}}}\text{.}%
\end{align*}

It follows that the equilibrium equation:%
\begin{align}
&  \left\Vert \boldsymbol{\tau}_{1}\right\Vert _{\min_{c}}^{2}-\left\Vert
\boldsymbol{\tau}_{1}\right\Vert \left\Vert \boldsymbol{\tau}_{2}\right\Vert
\cos\theta_{\boldsymbol{\tau}_{1}\boldsymbol{\tau}_{2}}+\delta\left(
y\right)  \sum\nolimits_{i=1}^{l_{1}}\psi_{1_{i_{\ast}}}%
\label{Balancing Bayes' Risk Linear}\\
&  +\sum\nolimits_{i=1}^{l_{1}}\psi_{1_{i_{\ast}}}\left\{  \sum\nolimits_{j=1}%
^{l}\mathbf{x}_{j\ast}^{T}\left(  \boldsymbol{\tau}_{2}-\boldsymbol{\tau}%
_{1}\right)  \right\}  +\sum\nolimits_{i=1}^{l_{1}}\xi_{i}\psi_{1_{i_{\ast}}%
}\nonumber\\
&  =\left\Vert \boldsymbol{\tau}_{2}\right\Vert _{\min_{c}}^{2}-\left\Vert
\boldsymbol{\tau}_{2}\right\Vert \left\Vert \boldsymbol{\tau}_{1}\right\Vert
\cos\theta_{\boldsymbol{\tau}_{2}\boldsymbol{\tau}_{1}}-\delta\left(
y\right)  \sum\nolimits_{i=1}^{l_{2}}\psi_{2_{i_{\ast}}}\nonumber\\
&  -\sum\nolimits_{i=1}^{l_{2}}\psi_{2_{i_{\ast}}}\left\{  \sum\nolimits_{j=1}%
^{l}\mathbf{x}_{j\ast}^{T}\left(  \boldsymbol{\tau}_{1}-\boldsymbol{\tau}%
_{2}\right)  \right\}  +\sum\nolimits_{i=1}^{l_{2}}\xi_{i}\psi_{2_{i_{\ast}}%
}\nonumber
\end{align}
is satisfied by the equilibrium point:%
\[
p\left(  \widehat{\Lambda}_{\boldsymbol{\psi}}\left(  \mathbf{x}\right)
|\omega_{1}\right)  -p\left(  \widehat{\Lambda}_{\boldsymbol{\psi}}\left(
\mathbf{x}\right)  |\omega_{2}\right)  =0
\]
and the primal likelihood ratio:%
\[
\widehat{\Lambda}_{\boldsymbol{\tau}}\left(  \mathbf{x}\right)  =p\left(
\widehat{\Lambda}_{\boldsymbol{\tau}}\left(  \mathbf{x}\right)  |\omega
_{1}\right)  -p\left(  \widehat{\Lambda}_{\boldsymbol{\tau}}\left(
\mathbf{x}\right)  |\omega_{2}\right)
\]
of the classification system $\boldsymbol{\tau}^{T}\mathbf{x}+\tau
_{0}\overset{\omega_{1}}{\underset{\omega_{2}}{\gtrless}}0$, where the
equilibrium equation in Eq. (\ref{Balancing Bayes' Risk Linear}) is
constrained by the equilibrium point in Eq.
(\ref{Wolfe Dual Equilibrium Point}).

I will now use Eq. (\ref{Balancing Bayes' Risk Linear}) to develop a
fundamental linear eigenlocus integral equation of binary classification for a
classification system in statistical equilibrium.

\subsection{Fundamental Balancing Feat in Dual Spaces I}

Returning to Eq. (\ref{Conditional Probability Function for Class One})%
\begin{align*}
P\left(  \mathbf{x}_{1_{i\ast}}|\boldsymbol{\tau}_{1}\right)   &  =\int%
_{Z}p\left(  \mathbf{x}_{1_{i\ast}}|\boldsymbol{\tau}_{1}\right)
d\boldsymbol{\tau}_{1}\\
&  =\int_{Z}\boldsymbol{\tau}_{1}d\boldsymbol{\tau}_{1}=\left\Vert
\boldsymbol{\tau}_{1}\right\Vert _{\min_{c}}^{2}+C_{1}%
\end{align*}
and Eq.(\ref{Conditional Probability Function for Class Two})%
\begin{align*}
P\left(  \mathbf{x}_{2_{i\ast}}|\boldsymbol{\tau}_{2}\right)   &  =\int%
_{Z}p\left(  \mathbf{x}_{2_{i\ast}}|\boldsymbol{\tau}_{2}\right)
d\boldsymbol{\tau}_{2}\\
&  =\int_{Z}\boldsymbol{\tau}_{2}d\boldsymbol{\tau}_{2}=\left\Vert
\boldsymbol{\tau}_{2}\right\Vert _{\min_{c}}^{2}+C_{2}\text{,}%
\end{align*}
and using Eq. (\ref{Balancing Bayes' Risk Linear})%
\begin{align*}
&  \left\Vert \boldsymbol{\tau}_{1}\right\Vert _{\min_{c}}^{2}-\left\Vert
\boldsymbol{\tau}_{1}\right\Vert \left\Vert \boldsymbol{\tau}_{2}\right\Vert
\cos\theta_{\boldsymbol{\tau}_{1}\boldsymbol{\tau}_{2}}+\delta\left(
y\right)  \sum\nolimits_{i=1}^{l_{1}}\psi_{1_{i_{\ast}}}\\
&  +\sum\nolimits_{i=1}^{l_{1}}\psi_{1_{i_{\ast}}}\left\{  \sum\nolimits_{j=1}%
^{l}\mathbf{x}_{j\ast}^{T}\left(  \boldsymbol{\tau}_{2}-\boldsymbol{\tau}%
_{1}\right)  \right\}  +\sum\nolimits_{i=1}^{l_{1}}\xi_{i}\psi_{1_{i_{\ast}}%
}\\
&  =\left\Vert \boldsymbol{\tau}_{2}\right\Vert _{\min_{c}}^{2}-\left\Vert
\boldsymbol{\tau}_{2}\right\Vert \left\Vert \boldsymbol{\tau}_{1}\right\Vert
\cos\theta_{\boldsymbol{\tau}_{2}\boldsymbol{\tau}_{1}}-\delta\left(
y\right)  \sum\nolimits_{i=1}^{l_{2}}\psi_{2_{i_{\ast}}}\\
&  -\sum\nolimits_{i=1}^{l_{2}}\psi_{2_{i_{\ast}}}\left\{  \sum\nolimits_{j=1}%
^{l}\mathbf{x}_{j\ast}^{T}\left(  \boldsymbol{\tau}_{1}-\boldsymbol{\tau}%
_{2}\right)  \right\}  +\sum\nolimits_{i=1}^{l_{2}}\xi_{i}\psi_{2_{i_{\ast}}}%
\end{align*}
where $P\left(  \mathbf{x}_{1_{i\ast}}|\boldsymbol{\tau}_{1}\right)  =P\left(
\mathbf{x}_{2_{i\ast}}|\boldsymbol{\tau}_{2}\right)  $, it follows that the
value for the integration constant $C_{1}$ in Eq.
(\ref{Conditional Probability Function for Class One}) is:%
\[
C_{1}=-\left\Vert \boldsymbol{\tau}_{1}\right\Vert \left\Vert \boldsymbol{\tau
}_{2}\right\Vert \cos\theta_{\boldsymbol{\tau}_{1}\boldsymbol{\tau}_{2}}%
+\sum\nolimits_{i=1}^{l_{1}}\xi_{i}\psi_{1_{i_{\ast}}}\text{,}%
\]
and that the value for the integration constant $C_{2}$ in
Eq.(\ref{Conditional Probability Function for Class Two}) is:%
\[
C_{2}=-\left\Vert \boldsymbol{\tau}_{2}\right\Vert \left\Vert \boldsymbol{\tau
}_{1}\right\Vert \cos\theta_{\boldsymbol{\tau}_{2}\boldsymbol{\tau}_{1}}%
+\sum\nolimits_{i=1}^{l_{2}}\xi_{i}\psi_{2_{i_{\ast}}}\text{.}%
\]

Substituting the values for $C_{1}$ and $C_{2}$ into Eqs
(\ref{Conditional Probability Function for Class One}) and
(\ref{Conditional Probability Function for Class Two}) produces the integral
equation%
\begin{align*}
&  f\left(  \widetilde{\Lambda}_{\boldsymbol{\tau}}\left(  \mathbf{x}\right)
\right)  =%
{\displaystyle\int\nolimits_{Z}}
\boldsymbol{\tau}_{1}d\boldsymbol{\tau}_{1}+\delta\left(  y\right)
\sum\nolimits_{i=1}^{l_{1}}\psi_{1_{i_{\ast}}}\\
&  +\sum\nolimits_{i=1}^{l_{1}}\psi_{1_{i_{\ast}}}\left\{  \sum\nolimits_{j=1}%
^{l}\mathbf{x}_{j\ast}^{T}\left(  \boldsymbol{\tau}_{2}-\boldsymbol{\tau}%
_{1}\right)  \right\} \\
&  =\int\nolimits_{Z}\boldsymbol{\tau}_{2}d\boldsymbol{\tau}_{2}-\delta\left(
y\right)  \sum\nolimits_{i=1}^{l_{2}}\psi_{2_{i_{\ast}}}\\
&  -\sum\nolimits_{i=1}^{l_{2}}\psi_{2_{i_{\ast}}}\left\{  \sum\nolimits_{j=1}%
^{l}\mathbf{x}_{j\ast}^{T}\left(  \boldsymbol{\tau}_{1}-\boldsymbol{\tau}%
_{2}\right)  \right\}  \text{,}%
\end{align*}
over the decision space $Z$, where the equalizer statistics%
\[
\nabla_{eq}\left(  p\left(  \widehat{\Lambda}_{\boldsymbol{\tau}}\left(
\mathbf{x}\right)  |\omega_{1}\right)  \right)  =\delta\left(  y\right)
\sum\nolimits_{i=1}^{l_{1}}\psi_{1_{i_{\ast}}}+\sum\nolimits_{i=1}^{l_{1}}%
\psi_{1_{i_{\ast}}}\left\{  \sum\nolimits_{j=1}^{l}\mathbf{x}_{j\ast}%
^{T}\left(  \boldsymbol{\tau}_{2}-\boldsymbol{\tau}_{1}\right)  \right\}
\]
and%
\[
\nabla_{eq}\left(  p\left(  \widehat{\Lambda}_{\boldsymbol{\tau}}\left(
\mathbf{x}\right)  |\omega_{2}\right)  \right)  =-\delta\left(  y\right)
\sum\nolimits_{i=1}^{l_{2}}\psi_{2_{i_{\ast}}}-\sum\nolimits_{i=1}^{l_{2}}%
\psi_{2_{i_{\ast}}}\left\{  \sum\nolimits_{j=1}^{l}\mathbf{x}_{j\ast}%
^{T}\left(  \boldsymbol{\tau}_{1}-\boldsymbol{\tau}_{2}\right)  \right\}
\]
ensure that the eigenenergy $\left\Vert \boldsymbol{\tau}_{1}\right\Vert
_{\min_{c}}^{2}$ associated with the position or location of the parameter
vector of likelihoods $p\left(  \widehat{\Lambda}_{\boldsymbol{\tau}}\left(
\mathbf{x}\right)  |\omega_{1}\right)  $ given class $\omega_{1}$ is
symmetrically balanced with the eigenenergy $\left\Vert \boldsymbol{\tau}%
_{2}\right\Vert _{\min_{c}}^{2}$ associated with the position or location of
the parameter vector of likelihoods $p\left(  \widehat{\Lambda}%
_{\boldsymbol{\tau}}\left(  \mathbf{x}\right)  |\omega_{2}\right)  $ given
class $\omega_{2}$.

The primal class-conditional probability density functions $p\left(
\mathbf{x}_{1_{i\ast}}|\boldsymbol{\tau}_{1}\right)  $ and $p\left(
\mathbf{x}_{2_{i\ast}}|\boldsymbol{\tau}_{2}\right)  $ and the Wolfe dual
class-conditional probability density functions $p\left(  \frac{\mathbf{x}%
_{1_{i\ast}}}{\left\Vert \mathbf{x}_{1_{i\ast}}\right\Vert }|\boldsymbol{\psi
}_{1}\right)  $ and $p\left(  \frac{\mathbf{x}_{2_{i\ast}}}{\left\Vert
\mathbf{x}_{2_{i\ast}}\right\Vert }|\boldsymbol{\psi}_{2}\right)  $ satisfy
the integral equation in the following manner:%
\begin{align}
&  f\left(  \widetilde{\Lambda}_{\boldsymbol{\tau}}\left(  \mathbf{x}\right)
\right)  =%
{\displaystyle\int\nolimits_{Z}}
p\left(  \mathbf{x}_{1_{i\ast}}|\boldsymbol{\tau}_{1}\right)
d\boldsymbol{\tau}_{1}+\delta\left(  y\right)  p\left(  \frac{\mathbf{x}%
_{1_{i\ast}}}{\left\Vert \mathbf{x}_{1_{i\ast}}\right\Vert }|\boldsymbol{\psi
}_{1}\right) \label{Linear Eigenlocus Integral Equation V}\\
&  +\widehat{\mathbf{x}}_{i\ast}\left[  p\left(  \mathbf{x}_{2_{i\ast}%
}|\boldsymbol{\tau}_{2}\right)  -p\left(  \mathbf{x}_{1_{i\ast}}%
|\boldsymbol{\tau}_{1}\right)  \right]  p\left(  \frac{\mathbf{x}_{1_{i\ast}}%
}{\left\Vert \mathbf{x}_{1_{i\ast}}\right\Vert }|\boldsymbol{\psi}_{1}\right)
\nonumber\\
&  =\int\nolimits_{Z}p\left(  \mathbf{x}_{2_{i\ast}}|\boldsymbol{\tau}%
_{2}\right)  d\boldsymbol{\tau}_{2}-\delta\left(  y\right)  p\left(
\frac{\mathbf{x}_{2_{i\ast}}}{\left\Vert \mathbf{x}_{2_{i\ast}}\right\Vert
}|\boldsymbol{\psi}_{2}\right) \nonumber\\
&  +\widehat{\mathbf{x}}_{i\ast}\left[  p\left(  \mathbf{x}_{2_{i\ast}%
}|\boldsymbol{\tau}_{2}\right)  -p\left(  \mathbf{x}_{1_{i\ast}}%
|\boldsymbol{\tau}_{1}\right)  \right]  p\left(  \frac{\mathbf{x}_{2_{i\ast}}%
}{\left\Vert \mathbf{x}_{2_{i\ast}}\right\Vert }|\boldsymbol{\psi}_{2}\right)
\text{,}\nonumber
\end{align}
over the $Z_{1}$ and $Z_{2}$ decision regions.

The equalizer statistics:%
\begin{align}
\nabla_{eq}\left(  p\left(  \widehat{\Lambda}_{\boldsymbol{\psi}}\left(
\mathbf{x}\right)  |\omega_{1}\right)  \right)   &  =\delta\left(  y\right)
p\left(  \frac{\mathbf{x}_{1_{i\ast}}}{\left\Vert \mathbf{x}_{1_{i\ast}%
}\right\Vert }|\boldsymbol{\psi}_{1}\right)
\label{Equalizer Statistic Class One}\\
&  +\widehat{\mathbf{x}}_{i\ast}p\left(  \mathbf{x}_{2_{i\ast}}%
|\boldsymbol{\tau}_{2}\right)  p\left(  \frac{\mathbf{x}_{1_{i\ast}}%
}{\left\Vert \mathbf{x}_{1_{i\ast}}\right\Vert }|\boldsymbol{\psi}_{1}\right)
\nonumber\\
&  -\widehat{\mathbf{x}}_{i\ast}p\left(  \mathbf{x}_{1_{i\ast}}%
|\boldsymbol{\tau}_{1}\right)  p\left(  \frac{\mathbf{x}_{1_{i\ast}}%
}{\left\Vert \mathbf{x}_{1_{i\ast}}\right\Vert }|\boldsymbol{\psi}_{1}\right)
\nonumber
\end{align}
and%
\begin{align}
\nabla_{eq}\left(  p\left(  \widehat{\Lambda}_{\boldsymbol{\psi}}\left(
\mathbf{x}\right)  |\omega_{2}\right)  \right)   &  =-\delta\left(  y\right)
p\left(  \frac{\mathbf{x}_{2_{i\ast}}}{\left\Vert \mathbf{x}_{2_{i\ast}%
}\right\Vert }|\boldsymbol{\psi}_{2}\right)
\label{Equalizer Statistic Class Two}\\
&  +\widehat{\mathbf{x}}_{i\ast}p\left(  \mathbf{x}_{2_{i\ast}}%
|\boldsymbol{\tau}_{2}\right)  p\left(  \frac{\mathbf{x}_{2_{i\ast}}%
}{\left\Vert \mathbf{x}_{2_{i\ast}}\right\Vert }|\boldsymbol{\psi}_{2}\right)
\nonumber\\
&  -\widehat{\mathbf{x}}_{i\ast}p\left(  \mathbf{x}_{1_{i\ast}}%
|\boldsymbol{\tau}_{1}\right)  p\left(  \frac{\mathbf{x}_{2_{i\ast}}%
}{\left\Vert \mathbf{x}_{2_{i\ast}}\right\Vert }|\boldsymbol{\psi}_{2}\right)
\text{,}\nonumber
\end{align}
where $p\left(  \frac{\mathbf{x}_{1_{i\ast}}}{\left\Vert \mathbf{x}_{1_{i\ast
}}\right\Vert }|\boldsymbol{\psi}_{1}\right)  $ and $p\left(  \frac
{\mathbf{x}_{2_{i\ast}}}{\left\Vert \mathbf{x}_{2_{i\ast}}\right\Vert
}|\boldsymbol{\psi}_{2}\right)  $ determine the equilibrium point%
\[
p\left(  \widehat{\Lambda}_{\boldsymbol{\psi}}\left(  \mathbf{x}\right)
|\omega_{1}\right)  -p\left(  \widehat{\Lambda}_{\boldsymbol{\psi}}\left(
\mathbf{x}\right)  |\omega_{2}\right)  =0
\]
of the integral equation $f\left(  \widetilde{\Lambda}_{\boldsymbol{\tau}%
}\left(  \mathbf{x}\right)  \right)  $ in Eq.
(\ref{Linear Eigenlocus Integral Equation V}), ensure that all of the forces
associated with counter risks $\overline{\mathfrak{R}}_{\mathfrak{\min}%
}\left(  Z_{1}|\psi_{1i\ast}\mathbf{x}_{1_{i_{\ast}}}\right)  $ and risks
$\mathfrak{R}_{\mathfrak{\min}}\left(  Z_{1}|\psi_{2i\ast}\mathbf{x}%
_{2_{i_{\ast}}}\right)  $ in the $Z_{1}$ decision region, which are related to
positions and potential locations of corresponding $\mathbf{x}_{1_{i_{\ast}}}$
and $\mathbf{x}_{2_{i_{\ast}}}$ extreme points in the $Z_{1}$ decision region,
are \emph{balanced with} all of the forces associated with counter risks
$\overline{\mathfrak{R}}_{\mathfrak{\min}}\left(  Z_{2}|\psi_{2i\ast
}\mathbf{x}_{2_{i_{\ast}}}\right)  $ and risks $\mathfrak{R}_{\mathfrak{\min}%
}\left(  Z_{2}|\psi_{1i\ast}\mathbf{x}_{1_{i_{\ast}}}\right)  $ in the $Z_{2}$
decision region, which are related to positions and potential locations of
corresponding $\mathbf{x}_{2_{i_{\ast}}}$ and $\mathbf{x}_{1_{i_{\ast}}}$
extreme points in the $Z_{2}$ decision region, such that the collective forces
associated with the integrals $\int_{Z}p\left(  \mathbf{x}_{1_{i\ast}%
}|\boldsymbol{\tau}_{1}\right)  d\boldsymbol{\tau}_{1}$ and $\int_{Z}p\left(
\mathbf{x}_{2_{i\ast}}|\boldsymbol{\tau}_{2}\right)  d\boldsymbol{\tau}_{2}$
are \emph{symmetrically balanced with} each other.

The above integral equation can be written as:%
\begin{align*}
f\left(  \widetilde{\Lambda}_{\boldsymbol{\tau}}\left(  \mathbf{x}\right)
\right)  =  &  \int_{Z_{1}}p\left(  \mathbf{x}_{1_{i\ast}}|\boldsymbol{\tau
}_{1}\right)  d\boldsymbol{\tau}_{1}+\int_{Z_{2}}p\left(  \mathbf{x}%
_{1_{i\ast}}|\boldsymbol{\tau}_{1}\right)  d\boldsymbol{\tau}_{1}+\nabla
_{eq}\left(  \widehat{\Lambda}_{\boldsymbol{\psi}_{1}}\left(  \mathbf{x}%
\right)  \right) \\
&  =\int_{Z_{1}}p\left(  \mathbf{x}_{2_{i\ast}}|\boldsymbol{\tau}_{2}\right)
d\boldsymbol{\tau}_{2}+\int_{Z_{2}}p\left(  \mathbf{x}_{2_{i\ast}%
}|\boldsymbol{\tau}_{2}\right)  d\boldsymbol{\tau}_{2}+\nabla_{eq}\left(
\widehat{\Lambda}_{\boldsymbol{\psi}_{2}}\left(  \mathbf{x}\right)  \right)
\text{,}%
\end{align*}
where the integral $\int_{Z_{1}}p\left(  \mathbf{x}_{1_{i\ast}}%
|\boldsymbol{\tau}_{1}\right)  d\boldsymbol{\tau}_{1}$ accounts for all of the
forces associated with counter risks $\overline{\mathfrak{R}}_{\mathfrak{\min
}}\left(  Z_{1}|\psi_{1i\ast}\mathbf{x}_{1_{i_{\ast}}}\right)  $ which are
related to positions and potential locations of corresponding $\mathbf{x}%
_{1_{i_{\ast}}}$ extreme points that lie in the $Z_{1}$ decision region, the
integral $\int_{Z_{2}}p\left(  \mathbf{x}_{1_{i\ast}}|\boldsymbol{\tau}%
_{1}\right)  d\boldsymbol{\tau}_{1}$ accounts for all of the forces associated
with risks $\mathfrak{R}_{\mathfrak{\min}}\left(  Z_{2}|\psi_{1i\ast
}\mathbf{x}_{1_{i_{\ast}}}\right)  $ which are related to positions and
potential locations of corresponding $\mathbf{x}_{1_{i_{\ast}}}$ extreme
points that lie in the $Z_{2}$ decision region, the integral $\int_{Z_{2}%
}p\left(  \mathbf{x}_{2_{i\ast}}|\boldsymbol{\tau}_{2}\right)
d\boldsymbol{\tau}_{2}$ accounts for all of the forces associated with counter
risks $\overline{\mathfrak{R}}_{\mathfrak{\min}}\left(  Z_{2}|\psi_{2i\ast
}\mathbf{x}_{2_{i_{\ast}}}\right)  $ which are related to positions and
potential locations of corresponding $\mathbf{x}_{2_{i_{\ast}}}$ extreme
points that lie in the $Z_{2}$ decision region, and the integral $\int_{Z_{1}%
}p\left(  \mathbf{x}_{2_{i\ast}}|\boldsymbol{\tau}_{2}\right)
d\boldsymbol{\tau}_{2}$ accounts for all of the forces associated with risks
$\mathfrak{R}_{\mathfrak{\min}}\left(  Z_{1}|\psi_{2i\ast}\mathbf{x}%
_{2_{i_{\ast}}}\right)  $ which are related to positions and potential
locations of corresponding $\mathbf{x}_{2_{i_{\ast}}}$ extreme points that lie
in the $Z_{1}$ decision region.

It follows that the classification system $\boldsymbol{\tau}^{T}%
\mathbf{x}+\tau_{0}\overset{\omega_{1}}{\underset{\omega_{2}}{\gtrless}}0$
seeks a point of statistical equilibrium $p\left(  \widehat{\Lambda
}_{\boldsymbol{\psi}}\left(  \mathbf{x}\right)  |\omega_{1}\right)  -p\left(
\widehat{\Lambda}_{\boldsymbol{\psi}}\left(  \mathbf{x}\right)  |\omega
_{2}\right)  =0$ where the opposing forces and influences of the
classification system are balanced with each other, such that the eigenenergy
and the risk of the classification system are minimized, and the
classification system is in statistical equilibrium.

Therefore, it is concluded that the linear eigenlocus discriminant function%
\[
\widetilde{\Lambda}_{\boldsymbol{\tau}}\left(  \mathbf{x}\right)  =\left(
\mathbf{x}-\sum\nolimits_{i=1}^{l}\mathbf{x}_{i\ast}\right)  ^{T}\left(
\boldsymbol{\tau}_{1}-\boldsymbol{\tau}_{2}\right)  +\sum\nolimits_{i=1}%
^{l}y_{i}\left(  1-\xi_{i}\right)
\]
is the solution to the integral equation:%
\begin{align*}
&  f\left(  \widetilde{\Lambda}_{\boldsymbol{\tau}}\left(  \mathbf{x}\right)
\right)  =\int_{Z_{1}}\boldsymbol{\tau}_{1}d\boldsymbol{\tau}_{1}+\int_{Z_{2}%
}\boldsymbol{\tau}_{1}d\boldsymbol{\tau}_{1}+\delta\left(  y\right)
\sum\nolimits_{i=1}^{l_{1}}\psi_{1_{i_{\ast}}}\\
&  +\sum\nolimits_{i=1}^{l_{1}}\psi_{1_{i_{\ast}}}\left\{  \sum\nolimits_{j=1}%
^{l}\mathbf{x}_{j\ast}^{T}\left(  \boldsymbol{\tau}_{2}-\boldsymbol{\tau}%
_{1}\right)  \right\} \\
&  =\int_{Z_{1}}\boldsymbol{\tau}_{2}d\boldsymbol{\tau}_{2}+\int_{Z_{2}%
}\boldsymbol{\tau}_{2}d\boldsymbol{\tau}_{2}-\delta\left(  y\right)
\sum\nolimits_{i=1}^{l_{2}}\psi_{2_{i_{\ast}}}\\
&  -\sum\nolimits_{i=1}^{l_{2}}\psi_{2_{i_{\ast}}}\left\{  \sum\nolimits_{j=1}%
^{l}\mathbf{x}_{j\ast}^{T}\left(  \boldsymbol{\tau}_{1}-\boldsymbol{\tau}%
_{2}\right)  \right\}  \text{,}%
\end{align*}
over the decision space $Z=Z_{1}+Z_{2}$, where $Z_{1}$ and $Z_{2}$ are
congruent decision regions $Z_{1}\cong Z_{2}$ that have respective counter
risks:%
\[
\overline{\mathfrak{R}}_{\mathfrak{\min}}\left(  Z_{1}|\boldsymbol{\tau}%
_{1}\right)  \text{ \ and \ }\overline{\mathfrak{R}}_{\mathfrak{\min}}\left(
Z_{2}|\boldsymbol{\tau}_{2}\right)
\]
and respective risks:%
\[
\mathfrak{R}_{\mathfrak{\min}}\left(  Z_{1}|\boldsymbol{\tau}_{2}\right)
\text{ \ and \ }\mathfrak{R}_{\mathfrak{\min}}\left(  Z_{2}|\boldsymbol{\tau
}_{1}\right)  \text{,}%
\]
where all of the forces associated with the risk $\mathfrak{R}_{\mathfrak{\min
}}\left(  Z_{1}|\boldsymbol{\tau}_{2}\right)  $ for class $\omega_{2}$ in the
$Z_{1}$ decision region and the counter risk $\overline{\mathfrak{R}%
}_{\mathfrak{\min}}\left(  Z_{2}|\boldsymbol{\tau}_{2}\right)  $ for class
$\omega_{2}$ in the $Z_{2}$ decision region are balanced with all of the
forces associated with the counter risk $\overline{\mathfrak{R}}%
_{\mathfrak{\min}}\left(  Z_{1}|\boldsymbol{\tau}_{1}\right)  $ for class
$\omega_{1}$ in the $Z_{1}$ decision region and the risk $\mathfrak{R}%
_{\mathfrak{\min}}\left(  Z_{2}|\boldsymbol{\tau}_{1}\right)  $ for class
$\omega_{1}$ in the $Z_{2}$ decision region:%
\begin{align*}
f\left(  \widehat{\Lambda}\left(  \mathbf{x}\right)  \right)   &
:\mathfrak{R}_{\mathfrak{\min}}\left(  Z_{1}|\boldsymbol{\tau}_{2}\right)
+\overline{\mathfrak{R}}_{\mathfrak{\min}}\left(  Z_{2}|\boldsymbol{\tau}%
_{2}\right) \\
&  \rightleftharpoons\overline{\mathfrak{R}}_{\mathfrak{\min}}\left(
Z_{1}|\boldsymbol{\tau}_{1}\right)  +\mathfrak{R}_{\mathfrak{\min}}\left(
Z_{2}|\boldsymbol{\tau}_{1}\right)
\end{align*}
such that the risk $\mathfrak{R}_{\mathfrak{\min}}\left(  Z|\widehat{\Lambda
}_{\boldsymbol{\tau}}\left(  \mathbf{x}\right)  \right)  $ and the eigenenergy
$E_{\min}\left(  Z|\widehat{\Lambda}_{\boldsymbol{\tau}}\left(  \mathbf{x}%
\right)  \right)  $ of the classification system $\boldsymbol{\tau}%
^{T}\mathbf{x}+\tau_{0}\overset{\omega_{1}}{\underset{\omega_{2}}{\gtrless}}0$
are governed by the equilibrium point $p\left(  \widehat{\Lambda
}_{\boldsymbol{\psi}}\left(  \mathbf{x}\right)  |\omega_{1}\right)  -p\left(
\widehat{\Lambda}_{\boldsymbol{\psi}}\left(  \mathbf{x}\right)  |\omega
_{2}\right)  =0$:%
\[
\sum\nolimits_{i=1}^{l_{1}}\psi_{1_{i\ast}}\frac{\mathbf{x}_{1_{i\ast}}%
}{\left\Vert \mathbf{x}_{1_{i\ast}}\right\Vert }-\sum\nolimits_{i=1}^{l_{2}%
}\psi_{2_{i\ast}}\frac{\mathbf{x}_{2_{i\ast}}}{\left\Vert \mathbf{x}%
_{2_{i\ast}}\right\Vert }=0
\]
of the integral equation $f\widehat{\Lambda}_{\boldsymbol{\tau}}\left(
\mathbf{x}\right)  $.

It follows that the classification system%
\[
\left(  \mathbf{x}-\sum\nolimits_{i=1}^{l}\mathbf{x}_{i\ast}\right)
^{T}\left(  \boldsymbol{\tau}_{1}-\boldsymbol{\tau}_{2}\right)  +\sum
\nolimits_{i=1}^{l}y_{i}\left(  1-\xi_{i}\right)  \overset{\omega
_{1}}{\underset{\omega_{2}}{\gtrless}}0
\]
is in statistical equilibrium:%
\begin{align*}
f\left(  \widetilde{\Lambda}_{\boldsymbol{\tau}}\left(  \mathbf{x}\right)
\right)   &  :\int_{Z_{1}}p\left(  \mathbf{x}_{1_{i\ast}}|\boldsymbol{\tau
}_{1}\right)  d\boldsymbol{\tau}_{1}-\int_{Z_{1}}p\left(  \mathbf{x}%
_{2_{i\ast}}|\boldsymbol{\tau}_{2}\right)  d\boldsymbol{\tau}_{2}+\nabla
_{eq}\left(  p\left(  \widehat{\Lambda}_{\boldsymbol{\psi}}\left(
\mathbf{x}\right)  |\omega_{1}\right)  \right) \\
&  =\int_{Z_{2}}p\left(  \mathbf{x}_{2_{i\ast}}|\boldsymbol{\tau}_{2}\right)
d\boldsymbol{\tau}_{2}-\int_{Z_{2}}p\left(  \mathbf{x}_{1_{i\ast}%
}|\boldsymbol{\tau}_{1}\right)  d\boldsymbol{\tau}_{1}+\nabla_{eq}\left(
p\left(  \widehat{\Lambda}_{\boldsymbol{\psi}}\left(  \mathbf{x}\right)
|\omega_{2}\right)  \right)  \text{,}%
\end{align*}
where all of the forces associated with the counter risk $\overline
{\mathfrak{R}}_{\mathfrak{\min}}\left(  Z_{1}|\boldsymbol{\tau}_{1}\right)  $
and the risk $\mathfrak{R}_{\mathfrak{\min}}\left(  Z_{1}|\boldsymbol{\tau
}_{2}\right)  $ in the $Z_{1}$ decision region are balanced with all of the
forces associated with the counter risk $\overline{\mathfrak{R}}%
_{\mathfrak{\min}}\left(  Z_{2}|\boldsymbol{\tau}_{2}\right)  $ and the risk
$\mathfrak{R}_{\mathfrak{\min}}\left(  Z_{2}|\boldsymbol{\tau}_{1}\right)  $
in the $Z_{2}$ decision region:%
\begin{align*}
f\left(  \widetilde{\Lambda}_{\boldsymbol{\tau}}\left(  \mathbf{x}\right)
\right)   &  :\overline{\mathfrak{R}}_{\mathfrak{\min}}\left(  Z_{1}%
|\boldsymbol{\tau}_{1}\right)  -\mathfrak{R}_{\mathfrak{\min}}\left(
Z_{1}|\boldsymbol{\tau}_{2}\right) \\
&  =\overline{\mathfrak{R}}_{\mathfrak{\min}}\left(  Z_{2}|\boldsymbol{\tau
}_{2}\right)  -\mathfrak{R}_{\mathfrak{\min}}\left(  Z_{2}|\boldsymbol{\tau
}_{1}\right)  \text{,}%
\end{align*}
and the eigenenergies associated with the counter risk $\overline
{\mathfrak{R}}_{\mathfrak{\min}}\left(  Z_{1}|\boldsymbol{\tau}_{1}\right)  $
and the risk $\mathfrak{R}_{\mathfrak{\min}}\left(  Z_{1}|\boldsymbol{\tau
}_{2}\right)  $ in the $Z_{1}$ decision region are balanced with the
eigenenergies associated with the counter risk $\overline{\mathfrak{R}%
}_{\mathfrak{\min}}\left(  Z_{2}|\boldsymbol{\tau}_{2}\right)  $ and the risk
$\mathfrak{R}_{\mathfrak{\min}}\left(  Z_{2}|\boldsymbol{\tau}_{1}\right)  $
in the $Z_{2}$ decision region:%
\begin{align*}
f\left(  \widetilde{\Lambda}_{\boldsymbol{\tau}}\left(  \mathbf{x}\right)
\right)   &  :E_{\min_{c}}\left(  Z_{1}|\boldsymbol{\tau}_{1}\right)
-E_{\min_{c}}\left(  Z_{1}|\boldsymbol{\tau}_{2}\right) \\
&  =E_{\min_{c}}\left(  Z_{2}|\boldsymbol{\tau}_{2}\right)  -E_{\min_{c}%
}\left(  Z_{2}|\boldsymbol{\tau}_{1}\right)  \text{.}%
\end{align*}

Therefore, it is concluded that the risk $\mathfrak{R}_{\mathfrak{\min}%
}\left(  Z|\widehat{\Lambda}_{\boldsymbol{\tau}}\left(  \mathbf{x}\right)
\right)  $ and the eigenenergy $E_{\min}\left(  Z|\widehat{\Lambda
}_{\boldsymbol{\tau}}\left(  \mathbf{x}\right)  \right)  $ of the
classification system $\boldsymbol{\tau}^{T}\mathbf{x}+\tau_{0}\overset{\omega
_{1}}{\underset{\omega_{2}}{\gtrless}}0$ are governed by the equilibrium point
$\widehat{\Lambda}_{\boldsymbol{\psi}}\left(  \mathbf{x}\right)  $:%
\[
p\left(  \widehat{\Lambda}_{\boldsymbol{\psi}}\left(  \mathbf{x}\right)
|\omega_{1}\right)  -p\left(  \widehat{\Lambda}_{\boldsymbol{\psi}}\left(
\mathbf{x}\right)  |\omega_{2}\right)  =0
\]
of the integral equation%
\begin{align*}
f\left(  \widetilde{\Lambda}_{\boldsymbol{\tau}}\left(  \mathbf{x}\right)
\right)  =  &  \int_{Z_{1}}p\left(  \mathbf{x}_{1_{i\ast}}|\boldsymbol{\tau
}_{1}\right)  d\boldsymbol{\tau}_{1}+\int_{Z_{2}}p\left(  \mathbf{x}%
_{1_{i\ast}}|\boldsymbol{\tau}_{1}\right)  d\boldsymbol{\tau}_{1}+\nabla
_{eq}\left(  p\left(  \widehat{\Lambda}_{\boldsymbol{\psi}}\left(
\mathbf{x}\right)  |\omega_{1}\right)  \right) \\
&  =\int_{Z_{1}}p\left(  \mathbf{x}_{2_{i\ast}}|\boldsymbol{\tau}_{2}\right)
d\boldsymbol{\tau}_{2}+\int_{Z_{2}}p\left(  \mathbf{x}_{2_{i\ast}%
}|\boldsymbol{\tau}_{2}\right)  d\boldsymbol{\tau}_{2}+\nabla_{eq}\left(
p\left(  \widehat{\Lambda}_{\boldsymbol{\psi}}\left(  \mathbf{x}\right)
|\omega_{2}\right)  \right)  \text{,}%
\end{align*}
where the opposing forces and influences of the classification system are
balanced with each other, such that the eigenenergy and the risk of the
classification system are minimized, and the classification system is in
statistical equilibrium.

\begin{figure}[ptb]%
\centering
\fbox{\includegraphics[
height=2.5875in,
width=3.4411in
]%
{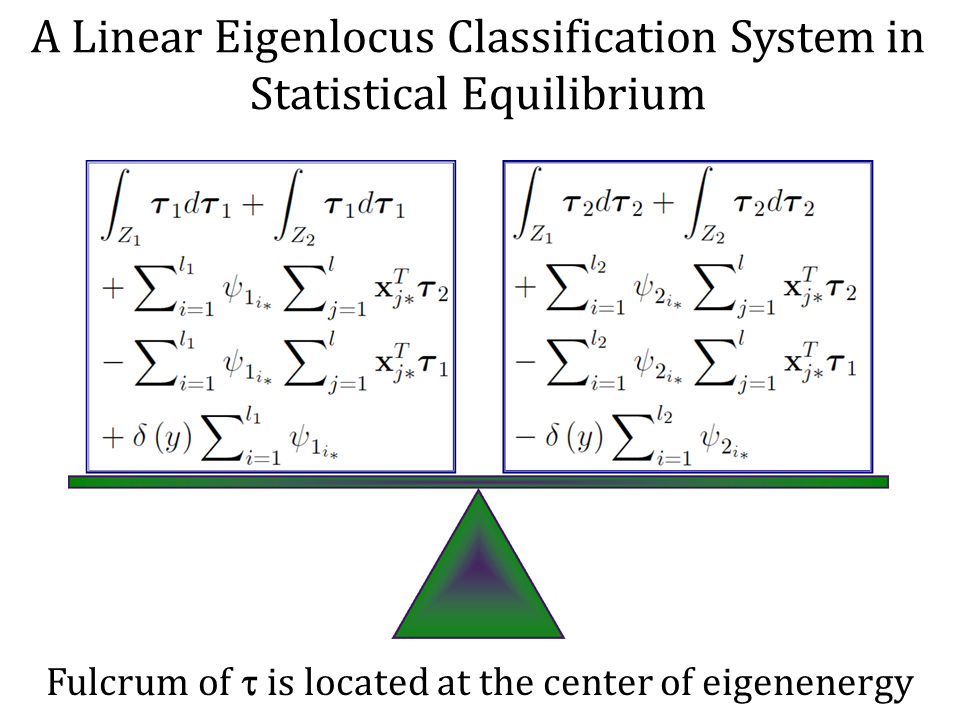}%
}\caption{Linear eigenlocus transforms generate linear classification systems
$\boldsymbol{\tau}^{T}\mathbf{x}+\tau_{0}\protect\overset{\omega
_{1}}{\protect\underset{\omega_{2}}{\gtrless}}0$ that satisfy a fundamental
integral equation of binary classification for a classification system in
statistical equilibrium.}%
\label{Symmetrical Balance of Bayes' Error Linear}%
\end{figure}

Figure $\ref{Symmetrical Balance of Bayes' Error Linear}$ illustrates the
manner in which linear eigenlocus discriminant functions $\widetilde{\Lambda
}_{\boldsymbol{\tau}}\left(  \mathbf{x}\right)  =\boldsymbol{\tau}%
^{T}\mathbf{x}+\tau_{0}$ minimize the total allowed eigenenergy $\left\Vert
\boldsymbol{\tau}\right\Vert _{\min_{c}}^{2}$ and the risk $\mathfrak{R}%
_{\mathfrak{\min}}\left(  Z|\mathbf{\tau}\right)  $ of linear classification
systems.%

\begin{figure}[ptb]%
\centering
\fbox{\includegraphics[
height=2.5875in,
width=3.4411in
]%
{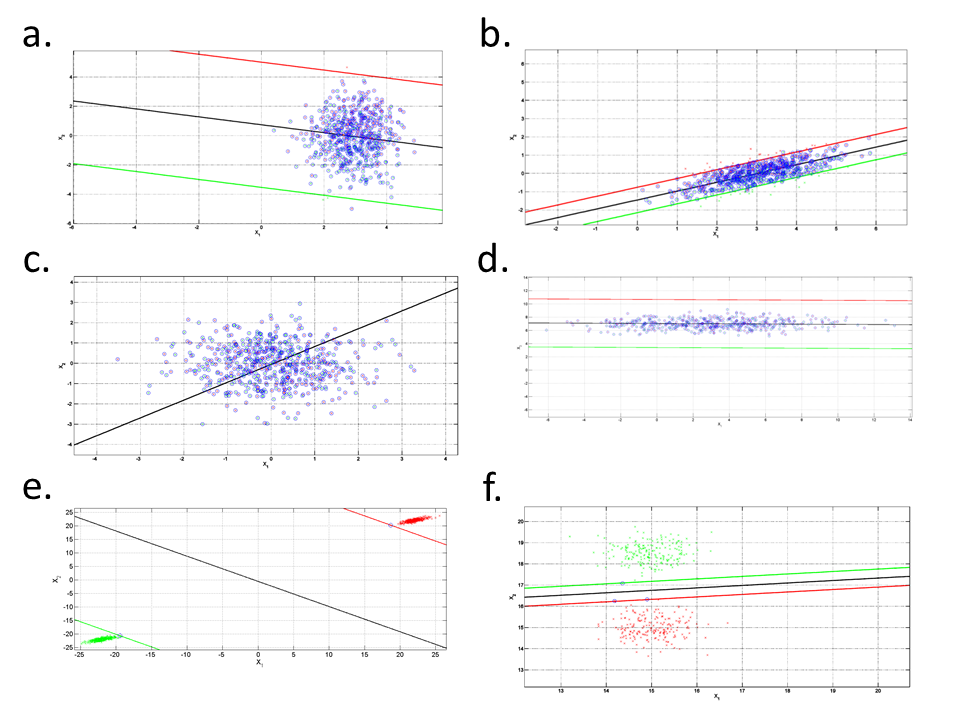}%
}\caption{Linear eigenlocus classification systems $\boldsymbol{\tau}%
^{T}\mathbf{x}+\tau_{0}\protect\overset{\omega_{1}}{\protect\underset{\omega
_{2}}{\gtrless}}0$ generate optimal linear decision boundaries for $\left(
1\right)  $ highly overlapping data distributions: see $\left(  a\right)  $
and $\left(  b\right)  $, $\left(  2\right)  $ completely overlapping data
distributions: see $\left(  c\right)  $ and $\left(  d\right)  $, and $\left(
3\right)  $ non-overlapping data distributions: see $\left(  e\right)  $ and
$\left(  f\right)  $.}%
\label{Bayes' Decision Boundaries Linear}%
\end{figure}

By way of illustration, Fig. $\ref{Bayes' Decision Boundaries Linear}$ shows
that linear eigenlocus transforms generate decision regions $Z_{1}$ and
$Z_{2}$ that minimize the risk $\mathfrak{R}_{\mathfrak{\min}}\left(
Z|\mathbf{\tau}\right)  $, where $\mathfrak{R}_{\mathfrak{\min}}\left(
Z_{1}|\boldsymbol{\tau}_{2}\right)  =\mathfrak{R}_{\mathfrak{\min}}\left(
Z_{2}|\boldsymbol{\tau}_{1}\right)  $, for highly overlapping data
distributions, completely overlapping data distributions and non-overlapping
data distributions. Accordingly, given data distributions that have similar
covariance matrices, linear eigenlocus classification systems
$\boldsymbol{\tau}^{T}\mathbf{x}+\tau_{0}\overset{\omega_{1}}{\underset{\omega
_{2}}{\gtrless}}0$ generate optimal linear decision boundaries for highly
overlapping data distributions (see Fig.
$\ref{Bayes' Decision Boundaries Linear}$a and Fig.
$\ref{Bayes' Decision Boundaries Linear}$b), completely overlapping data
distributions (see Fig. $\ref{Bayes' Decision Boundaries Linear}$c and Fig.
$\ref{Bayes' Decision Boundaries Linear}$d) and non-overlapping data
distributions (see Fig. $\ref{Bayes' Decision Boundaries Linear}$e and Fig.
$\ref{Bayes' Decision Boundaries Linear}$f), where unconstrained, primal
principal eigenaxis components (extreme points) are enclosed in blue circles.%

I\ am now in a position to formally state a \emph{discrete linear
classification theorem.}

\section*{Discrete Linear Classification Theorem}

Take a collection of $d$-component random vectors $\mathbf{x}$ that are
generated according to probability density functions $p\left(  \mathbf{x}%
|\omega_{1}\right)  $ and $p\left(  \mathbf{x}|\omega_{2}\right)  $ related to
statistical distributions of random vectors $\mathbf{x}$ that have constant or
unchanging statistics and similar covariance matrices, and let
$\widetilde{\Lambda}_{\boldsymbol{\tau}}\left(  \mathbf{x}\right)
=\boldsymbol{\tau}^{T}\mathbf{x}+\tau_{0}\overset{\omega_{1}}{\underset{\omega
_{2}}{\gtrless}}0$ denote the likelihood ratio test for a discrete, linear
classification system, where $\omega_{1}$ or $\omega_{2}$ is the true data
category, $\boldsymbol{\tau}$ is a locus of principal eigenaxis components and
likelihoods:%
\begin{align*}
\boldsymbol{\tau}  &  \triangleq\widehat{\Lambda}_{\boldsymbol{\tau}}\left(
\mathbf{x}\right)  =p\left(  \widehat{\Lambda}_{\boldsymbol{\tau}}\left(
\mathbf{x}\right)  |\omega_{1}\right)  -p\left(  \widehat{\Lambda
}_{\boldsymbol{\tau}}\left(  \mathbf{x}\right)  |\omega_{2}\right) \\
&  =\boldsymbol{\tau}_{1}-\boldsymbol{\tau}_{2}\\
&  =\sum\nolimits_{i=1}^{l_{1}}\psi_{1_{i_{\ast}}}\mathbf{x}_{1_{i_{\ast}}%
}-\sum\nolimits_{i=1}^{l_{2}}\psi_{2_{i_{\ast}}}\mathbf{x}_{2_{i_{\ast}}%
}\text{,}%
\end{align*}
where $\mathbf{x}_{1_{i_{\ast}}}\sim p\left(  \mathbf{x}|\omega_{1}\right)  $,
$\mathbf{x}_{2_{i_{\ast}}}\sim p\left(  \mathbf{x}|\omega_{2}\right)  $,
$\psi_{1_{i_{\ast}}}$ and $\psi_{2_{i_{\ast}}}$ are scale factors that provide
unit measures of likelihood for respective data points $\mathbf{x}%
_{1_{i_{\ast}}}$ and $\mathbf{x}_{2_{i_{\ast}}}$ which lie in either
overlapping regions or tails regions of data distributions related to
$p\left(  \mathbf{x}|\omega_{1}\right)  $ and $p\left(  \mathbf{x}|\omega
_{2}\right)  $, and $\tau_{0}$ is a functional of $\boldsymbol{\tau}$:%
\[
\tau_{0}=\sum\nolimits_{i=1}^{l}y_{i}\left(  1-\xi_{i}\right)  -\sum
\nolimits_{i=1}^{l}\mathbf{x}_{i\ast}^{T}\boldsymbol{\tau}\text{,}%
\]
where $\sum\nolimits_{i=1}^{l}\mathbf{x}_{i\ast}=\sum\nolimits_{i=1}^{l_{1}%
}\mathbf{x}_{1_{i_{\ast}}}+\sum\nolimits_{i=1}^{l_{2}}\mathbf{x}_{2_{i_{\ast}%
}}$ is a cluster of the data points $\mathbf{x}_{1_{i_{\ast}}}$ and
$\mathbf{x}_{2_{i_{\ast}}}$ used to form $\boldsymbol{\tau}$, $y_{i}$ are
class membership statistics: if $\mathbf{x}_{i\ast}\in\omega_{1}$, assign
$y_{i}=1$; if $\mathbf{x}_{i\ast}\in\omega_{2}$, assign $y_{i}=-1$, and
$\xi_{i}$ are regularization parameters: $\xi_{i}=\xi=0$ for full rank Gram
matrices or $\xi_{i}=\xi\ll1$ for low rank Gram matrices.

The linear discriminant function%
\[
\widetilde{\Lambda}_{\boldsymbol{\tau}}\left(  \mathbf{x}\right)
=\boldsymbol{\tau}^{T}\mathbf{x}+\tau_{0}%
\]
is the solution to the integral equation%
\begin{align*}
f\left(  \widetilde{\Lambda}_{\boldsymbol{\tau}}\left(  \mathbf{x}\right)
\right)  =  &  \int_{Z_{1}}\boldsymbol{\tau}_{1}d\boldsymbol{\tau}_{1}%
+\int_{Z_{2}}\boldsymbol{\tau}_{1}d\boldsymbol{\tau}_{1}+\delta\left(
y\right)  \sum\nolimits_{i=1}^{l_{1}}\psi_{1_{i_{\ast}}}\\
&  =\int_{Z_{1}}\boldsymbol{\tau}_{2}d\boldsymbol{\tau}_{2}+\int_{Z_{2}%
}\boldsymbol{\tau}_{2}d\boldsymbol{\tau}_{2}-\delta\left(  y\right)
\sum\nolimits_{i=1}^{l_{2}}\psi_{2_{i_{\ast}}}\text{,}%
\end{align*}
over the decision space $Z=Z_{1}+Z_{2}$, where $Z_{1}$ and $Z_{2}$ are
congruent decision regions: $Z_{1}\cong Z_{2}$ and $\delta\left(  y\right)
\triangleq\sum\nolimits_{i=1}^{l}y_{i}\left(  1-\xi_{i}\right)  $, such that
the expected risk $\mathfrak{R}_{\mathfrak{\min}}\left(  Z|\widehat{\Lambda
}_{\boldsymbol{\tau}}\left(  \mathbf{x}\right)  \right)  $ and the
corresponding eigenenergy $E_{\min}\left(  Z|\widehat{\Lambda}%
_{\boldsymbol{\tau}}\left(  \mathbf{x}\right)  \right)  $ of the
classification system $\boldsymbol{\tau}^{T}\mathbf{x}+\tau_{0}\overset{\omega
_{1}}{\underset{\omega_{2}}{\gtrless}}0$ are governed by the equilibrium point%
\[
\sum\nolimits_{i=1}^{l_{1}}\psi_{1i\ast}-\sum\nolimits_{i=1}^{l_{2}}%
\psi_{2i\ast}=0
\]
of the integral equation $f\left(  \widetilde{\Lambda}_{\boldsymbol{\tau}%
}\left(  \mathbf{x}\right)  \right)  $, where the equilibrium point is a dual
locus of principal eigenaxis components and likelihoods:%
\begin{align*}
\boldsymbol{\psi}  &  \triangleq\widehat{\Lambda}_{\boldsymbol{\psi}}\left(
\mathbf{x}\right)  =p\left(  \widehat{\Lambda}_{\boldsymbol{\psi}}\left(
\mathbf{x}\right)  |\omega_{1}\right)  +p\left(  \widehat{\Lambda
}_{\boldsymbol{\psi}}\left(  \mathbf{x}\right)  |\omega_{2}\right) \\
&  =\boldsymbol{\psi}_{1}+\boldsymbol{\psi}_{2}\\
&  =\sum\nolimits_{i=1}^{l_{1}}\psi_{1_{i_{\ast}}}\frac{\mathbf{x}%
_{1_{i_{\ast}}}}{\left\Vert \mathbf{x}_{1_{i_{\ast}}}\right\Vert }%
+\sum\nolimits_{i=1}^{l_{2}}\psi_{2_{i_{\ast}}}\frac{\mathbf{x}_{2_{i_{\ast}}%
}}{\left\Vert \mathbf{x}_{2_{i_{\ast}}}\right\Vert }%
\end{align*}
that is constrained to be in statistical equilibrium:%
\[
\sum\nolimits_{i=1}^{l_{1}}\psi_{1_{i_{\ast}}}\frac{\mathbf{x}_{1_{i_{\ast}}}%
}{\left\Vert \mathbf{x}_{1_{i_{\ast}}}\right\Vert }=\sum\nolimits_{i=1}%
^{l_{2}}\psi_{2_{i_{\ast}}}\frac{\mathbf{x}_{2_{i_{\ast}}}}{\left\Vert
\mathbf{x}_{2_{i_{\ast}}}\right\Vert }\text{.}%
\]

Therefore, the forces associated with the counter risk $\overline
{\mathfrak{R}}_{\mathfrak{\min}}\left(  Z_{1}|p\left(  \widehat{\Lambda
}_{\boldsymbol{\tau}}\left(  \mathbf{x}\right)  |\omega_{1}\right)  \right)  $
and the risk $\mathfrak{R}_{\mathfrak{\min}}\left(  Z_{2}|p\left(
\widehat{\Lambda}_{\boldsymbol{\tau}}\left(  \mathbf{x}\right)  |\omega
_{1}\right)  \right)  $ in the $Z_{1}$ and $Z_{2}$ decision regions: which are
related to positions and potential locations of data points $\mathbf{x}%
_{1_{i_{\ast}}}$ that are generated according to $p\left(  \mathbf{x}%
|\omega_{1}\right)  $, are balanced with the forces associated with the risk
$\mathfrak{R}_{\mathfrak{\min}}\left(  Z_{1}|p\left(  \widehat{\Lambda
}_{\boldsymbol{\tau}}\left(  \mathbf{x}\right)  |\omega_{2}\right)  \right)  $
and the counter risk $\overline{\mathfrak{R}}_{\mathfrak{\min}}\left(
Z_{2}|p\left(  \widehat{\Lambda}_{\boldsymbol{\tau}}\left(  \mathbf{x}\right)
|\omega_{2}\right)  \right)  $ in the $Z_{1}$ and $Z_{2}$ decision regions:
which are related to positions and potential locations of data points
$\mathbf{x}_{2_{i_{\ast}}}$ that are generated according to $p\left(
\mathbf{x}|\omega_{2}\right)  $.

Furthermore, the eigenenergy $E_{\min}\left(  Z|p\left(  \widehat{\Lambda
}_{\boldsymbol{\tau}}\left(  \mathbf{x}\right)  |\omega_{1}\right)  \right)  $
associated with the position or location of the parameter vector of
likelihoods $p\left(  \widehat{\Lambda}_{\boldsymbol{\tau}}\left(
\mathbf{x}\right)  |\omega_{1}\right)  $ given class $\omega_{1}$ is balanced
with the eigenenergy $E_{\min}\left(  Z|p\left(  \widehat{\Lambda
}_{\boldsymbol{\tau}}\left(  \mathbf{x}\right)  |\omega_{2}\right)  \right)  $
associated with the position or location of the parameter vector of
likelihoods $p\left(  \widehat{\Lambda}_{\boldsymbol{\tau}}\left(
\mathbf{x}\right)  |\omega_{2}\right)  $ given class $\omega_{2}$:%
\[
\left\Vert \boldsymbol{\tau}_{1}\right\Vert _{\min_{c}}^{2}+\delta\left(
y\right)  \sum\nolimits_{i=1}^{l_{1}}\psi_{1_{i_{\ast}}}\equiv\left\Vert
\boldsymbol{\tau}_{2}\right\Vert _{\min_{c}}^{2}-\delta\left(  y\right)
\sum\nolimits_{i=1}^{l_{2}}\psi_{2_{i_{\ast}}}\text{,}%
\]
where the total eigenenergy%
\begin{align*}
\left\Vert \boldsymbol{\tau}\right\Vert _{\min_{c}}^{2}  &  =\left\Vert
\boldsymbol{\tau}_{1}-\boldsymbol{\tau}_{2}\right\Vert _{\min_{c}}^{2}\\
&  =\left\Vert \boldsymbol{\tau}_{1}\right\Vert _{\min_{c}}^{2}-\left\Vert
\boldsymbol{\tau}_{1}\right\Vert \left\Vert \boldsymbol{\tau}_{2}\right\Vert
\cos\theta_{\boldsymbol{\tau}_{1}\boldsymbol{\tau}_{2}}\\
&  +\left\Vert \boldsymbol{\tau}_{2}\right\Vert _{\min_{c}}^{2}-\left\Vert
\boldsymbol{\tau}_{2}\right\Vert \left\Vert \boldsymbol{\tau}_{1}\right\Vert
\cos\theta_{\boldsymbol{\tau}_{2}\boldsymbol{\tau}_{1}}\\
&  =\sum\nolimits_{i=1}^{l_{1}}\psi_{1i\ast}\left(  1-\xi_{i}\right)
+\sum\nolimits_{i=1}^{l_{2}}\psi_{2i\ast}\left(  1-\xi_{i}\right)
\end{align*}
of the discrete, linear classification system $\boldsymbol{\tau}^{T}%
\mathbf{x}+\tau_{0}\overset{\omega_{1}}{\underset{\omega_{2}}{\gtrless}}0$ is
determined by the eigenenergies associated with the position or location of
the likelihood ratio $\boldsymbol{\tau=\tau}_{1}-\boldsymbol{\tau}_{2}$ and
the locus of a corresponding, linear decision boundary $\boldsymbol{\tau}%
^{T}\mathbf{x}+\tau_{0}=0$.

It follows that the discrete, linear classification system $\boldsymbol{\tau
}^{T}\mathbf{x}+\tau_{0}\overset{\omega_{1}}{\underset{\omega_{2}}{\gtrless}%
}0$ is in statistical equilibrium:%
\begin{align*}
f\left(  \widetilde{\Lambda}_{\boldsymbol{\tau}}\left(  \mathbf{x}\right)
\right)  :  &  \int_{Z_{1}}\boldsymbol{\tau}_{1}d\boldsymbol{\tau}_{1}%
-\int_{Z_{1}}\boldsymbol{\tau}_{2}d\boldsymbol{\tau}_{2}+\delta\left(
y\right)  \sum\nolimits_{i=1}^{l_{1}}\psi_{1_{i_{\ast}}}\\
&  =\int_{Z_{2}}\boldsymbol{\tau}_{2}d\boldsymbol{\tau}_{2}-\int_{Z_{2}%
}\boldsymbol{\tau}_{1}d\boldsymbol{\tau}_{1}-\delta\left(  y\right)
\sum\nolimits_{i=1}^{l_{2}}\psi_{2_{i_{\ast}}}\text{,}%
\end{align*}
where the forces associated with the counter risk $\overline{\mathfrak{R}%
}_{\mathfrak{\min}}\left(  Z_{1}|p\left(  \widehat{\Lambda}_{\boldsymbol{\tau
}}\left(  \mathbf{x}\right)  |\omega_{1}\right)  \right)  $ for class
$\omega_{1}$ and the risk $\mathfrak{R}_{\mathfrak{\min}}\left(
Z_{1}|p\left(  \widehat{\Lambda}_{\boldsymbol{\tau}}\left(  \mathbf{x}\right)
|\omega_{2}\right)  \right)  $ for class $\omega_{2}$ in the $Z_{1}$ decision
region are balanced with the forces associated with the counter risk
$\overline{\mathfrak{R}}_{\mathfrak{\min}}\left(  Z_{2}|p\left(
\widehat{\Lambda}_{\boldsymbol{\tau}}\left(  \mathbf{x}\right)  |\omega
_{2}\right)  \right)  $ for class $\omega_{2}$ and the risk $\mathfrak{R}%
_{\mathfrak{\min}}\left(  Z_{2}|p\left(  \widehat{\Lambda}_{\boldsymbol{\tau}%
}\left(  \mathbf{x}\right)  |\omega_{1}\right)  \right)  $ for class
$\omega_{1}$ in the $Z_{2}$ decision region:%
\begin{align*}
f\left(  \widetilde{\Lambda}_{\boldsymbol{\tau}}\left(  \mathbf{x}\right)
\right)   &  :\overline{\mathfrak{R}}_{\mathfrak{\min}}\left(  Z_{1}|p\left(
\widehat{\Lambda}_{\boldsymbol{\tau}}\left(  \mathbf{x}\right)  |\omega
_{1}\right)  \right)  -\mathfrak{R}_{\mathfrak{\min}}\left(  Z_{1}|p\left(
\widehat{\Lambda}_{\boldsymbol{\tau}}\left(  \mathbf{x}\right)  |\omega
_{2}\right)  \right) \\
&  =\overline{\mathfrak{R}}_{\mathfrak{\min}}\left(  Z_{2}|p\left(
\widehat{\Lambda}_{\boldsymbol{\tau}}\left(  \mathbf{x}\right)  |\omega
_{2}\right)  \right)  -\mathfrak{R}_{\mathfrak{\min}}\left(  Z_{2}|p\left(
\widehat{\Lambda}_{\boldsymbol{\tau}}\left(  \mathbf{x}\right)  |\omega
_{1}\right)  \right)
\end{align*}
such that the expected risk $\mathfrak{R}_{\mathfrak{\min}}\left(
Z|\widehat{\Lambda}_{\boldsymbol{\tau}}\left(  \mathbf{x}\right)  \right)  $
of the classification system is minimized, and the eigenenergies associated
with the counter risk $\overline{\mathfrak{R}}_{\mathfrak{\min}}\left(
Z_{1}|p\left(  \widehat{\Lambda}_{\boldsymbol{\tau}}\left(  \mathbf{x}\right)
|\omega_{1}\right)  \right)  $ for class $\omega_{1}$ and the risk
$\mathfrak{R}_{\mathfrak{\min}}\left(  Z_{1}|p\left(  \widehat{\Lambda
}_{\boldsymbol{\tau}}\left(  \mathbf{x}\right)  |\omega_{2}\right)  \right)  $
for class $\omega_{2}$ in the $Z_{1}$ decision region are balanced with the
eigenenergies associated with the counter risk $\overline{\mathfrak{R}%
}_{\mathfrak{\min}}\left(  Z_{2}|p\left(  \widehat{\Lambda}_{\boldsymbol{\tau
}}\left(  \mathbf{x}\right)  |\omega_{2}\right)  \right)  $ for class
$\omega_{2}$ and the risk $\mathfrak{R}_{\mathfrak{\min}}\left(
Z_{2}|p\left(  \widehat{\Lambda}_{\boldsymbol{\tau}}\left(  \mathbf{x}\right)
|\omega_{1}\right)  \right)  $ for class $\omega_{1}$ in the $Z_{2}$ decision
region:%
\begin{align*}
f\left(  \widetilde{\Lambda}_{\boldsymbol{\tau}}\left(  \mathbf{x}\right)
\right)   &  :E_{\min}\left(  Z_{1}|p\left(  \widehat{\Lambda}%
_{\boldsymbol{\tau}}\left(  \mathbf{x}\right)  |\omega_{1}\right)  \right)
-E_{\min}\left(  Z_{1}|p\left(  \widehat{\Lambda}_{\boldsymbol{\tau}}\left(
\mathbf{x}\right)  |\omega_{2}\right)  \right) \\
&  =E_{\min}\left(  Z_{2}|p\left(  \widehat{\Lambda}_{\boldsymbol{\tau}%
}\left(  \mathbf{x}\right)  |\omega_{2}\right)  \right)  -E_{\min}\left(
Z_{2}|p\left(  \widehat{\Lambda}_{\boldsymbol{\tau}}\left(  \mathbf{x}\right)
|\omega_{1}\right)  \right)
\end{align*}
such that the eigenenergy $E_{\min}\left(  Z|\widehat{\Lambda}%
_{\boldsymbol{\tau}}\left(  \mathbf{x}\right)  \right)  $ of the
classification system is minimized.

Thus, any given discrete, linear classification system $\boldsymbol{\tau}%
^{T}\mathbf{x}+\tau_{0}\overset{\omega_{1}}{\underset{\omega_{2}}{\gtrless}}0$
exhibits an error rate that is consistent with the risk $\mathfrak{R}%
_{\mathfrak{\min}}\left(  Z|\widehat{\Lambda}_{\boldsymbol{\tau}}\left(
\mathbf{x}\right)  \right)  $ and the corresponding eigenenergy $E_{\min
}\left(  Z|\widehat{\Lambda}_{\boldsymbol{\tau}}\left(  \mathbf{x}\right)
\right)  $ of the classification system: for all random vectors $\mathbf{x}$
that are generated according to $p\left(  \mathbf{x}|\omega_{1}\right)  $ and
$p\left(  \mathbf{x}|\omega_{2}\right)  $, where $p\left(  \mathbf{x}%
|\omega_{1}\right)  $ and $p\left(  \mathbf{x}|\omega_{2}\right)  $ are
related to statistical distributions of random vectors $\mathbf{x}$ that have
similar covariance matrices and constant or unchanging statistics.

Therefore, a discrete, linear classification system $\boldsymbol{\tau}%
^{T}\mathbf{x}+\tau_{0}\overset{\omega_{1}}{\underset{\omega_{2}}{\gtrless}}0$
seeks a point of statistical equilibrium where the opposing forces and
influences of the classification system are balanced with each other, such
that the eigenenergy and the expected risk of the classification system are
minimized, and the classification system is in statistical equilibrium.

I\ will now show that the eigenenergy of a discrete, linear classification
system is conserved and remains relatively constant, so that the eigenenergy
and the corresponding expected risk of a discrete, linear classification
system cannot be created or destroyed, but only transferred from one
classification system to another.

\section*{Law of Conservation of Eigenenergy:}

\subsection*{For Discrete Linear Classification Systems}

Take a collection of $N$ random vectors $\mathbf{x}$ of dimension $d$ that are
generated according to probability density functions $p\left(  \mathbf{x}%
|\omega_{1}\right)  $ and $p\left(  \mathbf{x}|\omega_{2}\right)  $ related to
statistical distributions of random vectors $\mathbf{x}$ that have constant or
unchanging statistics and similar covariance matrices, where the number of
random vectors $\mathbf{x}\sim p\left(  \mathbf{x}|\omega_{1}\right)  $ equals
the number of random vectors $\mathbf{x}\sim p\left(  \mathbf{x}|\omega
_{2}\right)  $, and let $\widetilde{\Lambda}_{\boldsymbol{\tau}}\left(
\mathbf{x}\right)  =\boldsymbol{\tau}^{T}\mathbf{x}+\tau_{0}\overset{\omega
_{1}}{\underset{\omega_{2}}{\gtrless}}0$ denote the likelihood ratio test for
a discrete, linear classification system, where $\omega_{1}$ or $\omega_{2}$
is the true data category, $\boldsymbol{\tau}$ is a locus of principal
eigenaxis components and likelihoods:%
\begin{align*}
\boldsymbol{\tau}  &  \triangleq\widehat{\Lambda}_{\boldsymbol{\tau}}\left(
\mathbf{x}\right)  =p\left(  \widehat{\Lambda}_{\boldsymbol{\tau}}\left(
\mathbf{x}\right)  |\omega_{1}\right)  -p\left(  \widehat{\Lambda
}_{\boldsymbol{\tau}}\left(  \mathbf{x}\right)  |\omega_{2}\right) \\
&  =\boldsymbol{\tau}_{1}-\boldsymbol{\tau}_{2}\\
&  =\sum\nolimits_{i=1}^{l_{1}}\psi_{1_{i_{\ast}}}\mathbf{x}_{1_{i_{\ast}}%
}-\sum\nolimits_{i=1}^{l_{2}}\psi_{2_{i_{\ast}}}\mathbf{x}_{2_{i_{\ast}}%
}\text{,}%
\end{align*}
where $\mathbf{x}_{1_{i_{\ast}}}\sim p\left(  \mathbf{x}|\omega_{1}\right)  $,
$\mathbf{x}_{2_{i_{\ast}}}\sim p\left(  \mathbf{x}|\omega_{2}\right)  $,
$\psi_{1_{i_{\ast}}}$ and $\psi_{2_{i_{\ast}}}$ are scale factors that provide
unit measures of likelihood for respective data points $\mathbf{x}%
_{1_{i_{\ast}}}$ and $\mathbf{x}_{2_{i_{\ast}}}$ which lie in either
overlapping regions or tails regions of data distributions related to
$p\left(  \mathbf{x}|\omega_{1}\right)  $ and $p\left(  \mathbf{x}|\omega
_{2}\right)  $, and $\tau_{0}$ is a functional of $\boldsymbol{\tau}$:%
\[
\tau_{0}=\sum\nolimits_{i=1}^{l}y_{i}\left(  1-\xi_{i}\right)  -\sum
\nolimits_{i=1}^{l}\mathbf{x}_{i\ast}^{T}\boldsymbol{\tau}\text{,}%
\]
where $\sum\nolimits_{i=1}^{l}\mathbf{x}_{i\ast}=\sum\nolimits_{i=1}^{l_{1}%
}\mathbf{x}_{1_{i_{\ast}}}+\sum\nolimits_{i=1}^{l_{2}}\mathbf{x}_{2_{i_{\ast}%
}}$ is a cluster of the data points $\mathbf{x}_{1_{i_{\ast}}}$ and
$\mathbf{x}_{2_{i_{\ast}}}$ used to form $\boldsymbol{\tau}$, $y_{i}$ are
class membership statistics: if $\mathbf{x}_{i\ast}\in\omega_{1}$, assign
$y_{i}=1$; if $\mathbf{x}_{i\ast}\in\omega_{2}$, assign $y_{i}=-1$, and
$\xi_{i}$ are regularization parameters: $\xi_{i}=\xi=0$ for full rank Gram
matrices or $\xi_{i}=\xi\ll1$ for low rank Gram matrices.

The expected risk $\mathfrak{R}_{\mathfrak{\min}}\left(  Z|\widehat{\Lambda
}_{\boldsymbol{\tau}}\left(  \mathbf{x}\right)  \right)  $ and the
corresponding eigenenergy $E_{\min}\left(  Z|\widehat{\Lambda}%
_{\boldsymbol{\tau}}\left(  \mathbf{x}\right)  \right)  $ of a discrete,
linear classification system $\boldsymbol{\tau}^{T}\mathbf{x}+\tau
_{0}\overset{\omega_{1}}{\underset{\omega_{2}}{\gtrless}}0$ are governed by
the equilibrium point%
\[
\sum\nolimits_{i=1}^{l_{1}}\psi_{1i\ast}-\sum\nolimits_{i=1}^{l_{2}}%
\psi_{2i\ast}=0
\]
of the integral equation%
\begin{align*}
f\left(  \widetilde{\Lambda}_{\boldsymbol{\tau}}\left(  \mathbf{x}\right)
\right)  =  &  \int_{Z_{1}}\boldsymbol{\tau}_{1}d\boldsymbol{\tau}_{1}%
+\int_{Z_{2}}\boldsymbol{\tau}_{1}d\boldsymbol{\tau}_{1}+\delta\left(
y\right)  \sum\nolimits_{i=1}^{l_{1}}\psi_{1_{i_{\ast}}}\\
&  =\int_{Z_{1}}\boldsymbol{\tau}_{2}d\boldsymbol{\tau}_{2}+\int_{Z_{2}%
}\boldsymbol{\tau}_{2}d\boldsymbol{\tau}_{2}-\delta\left(  y\right)
\sum\nolimits_{i=1}^{l_{2}}\psi_{2_{i_{\ast}}}\text{,}%
\end{align*}
over the decision space $Z=Z_{1}+Z_{2}$, where $Z_{1}$ and $Z_{2}$ are
congruent decision regions: $Z_{1}\cong Z_{2}$, $\delta\left(  y\right)
\triangleq\sum\nolimits_{i=1}^{l}y_{i}\left(  1-\xi_{i}\right)  $, and the
forces associated with the counter risk $\overline{\mathfrak{R}}%
_{\mathfrak{\min}}\left(  Z_{1}|p\left(  \widehat{\Lambda}_{\boldsymbol{\tau}%
}\left(  \mathbf{x}\right)  |\omega_{1}\right)  \right)  $ and the risk
$\mathfrak{R}_{\mathfrak{\min}}\left(  Z_{2}|p\left(  \widehat{\Lambda
}_{\boldsymbol{\tau}}\left(  \mathbf{x}\right)  |\omega_{1}\right)  \right)  $
in the $Z_{1}$ and $Z_{2}$ decision regions: which are related to positions
and potential locations of data points $\mathbf{x}_{1_{i_{\ast}}}$ that are
generated according to $p\left(  \mathbf{x}|\omega_{1}\right)  $, are balanced
with the forces associated with the risk $\mathfrak{R}_{\mathfrak{\min}%
}\left(  Z_{1}|p\left(  \widehat{\Lambda}_{\boldsymbol{\tau}}\left(
\mathbf{x}\right)  |\omega_{2}\right)  \right)  $ and the counter risk
$\overline{\mathfrak{R}}_{\mathfrak{\min}}\left(  Z_{2}|p\left(
\widehat{\Lambda}_{\boldsymbol{\tau}}\left(  \mathbf{x}\right)  |\omega
_{2}\right)  \right)  $ in the $Z_{1}$ and $Z_{2}$ decision regions: which are
related to positions and potential locations of data points $\mathbf{x}%
_{2_{i_{\ast}}}$ that are generated according to $p\left(  \mathbf{x}%
|\omega_{2}\right)  $.

Furthermore, the eigenenergy $E_{\min}\left(  Z|p\left(  \widehat{\Lambda
}_{\boldsymbol{\tau}}\left(  \mathbf{x}\right)  |\omega_{1}\right)  \right)  $
associated with the position or location of the parameter vector of
likelihoods $p\left(  \widehat{\Lambda}_{\boldsymbol{\tau}}\left(
\mathbf{x}\right)  |\omega_{1}\right)  $ given class $\omega_{1}$ is balanced
with the eigenenergy $E_{\min}\left(  Z|p\left(  \widehat{\Lambda
}_{\boldsymbol{\tau}}\left(  \mathbf{x}\right)  |\omega_{2}\right)  \right)  $
associated with the position or location of the parameter vector of
likelihoods $p\left(  \widehat{\Lambda}_{\boldsymbol{\tau}}\left(
\mathbf{x}\right)  |\omega_{2}\right)  $ given class $\omega_{2}$:%
\[
\left\Vert \boldsymbol{\tau}_{1}\right\Vert _{\min_{c}}^{2}+\delta\left(
y\right)  \sum\nolimits_{i=1}^{l_{1}}\psi_{1_{i_{\ast}}}\equiv\left\Vert
\boldsymbol{\tau}_{2}\right\Vert _{\min_{c}}^{2}-\delta\left(  y\right)
\sum\nolimits_{i=1}^{l_{2}}\psi_{2_{i_{\ast}}}\text{,}%
\]
where%
\begin{align*}
\left\Vert \boldsymbol{\tau}\right\Vert _{\min_{c}}^{2}  &  =\left\Vert
\boldsymbol{\tau}_{1}\right\Vert _{\min_{c}}^{2}-\left\Vert \boldsymbol{\tau
}_{1}\right\Vert \left\Vert \boldsymbol{\tau}_{2}\right\Vert \cos
\theta_{\boldsymbol{\tau}_{1}\boldsymbol{\tau}_{2}}\\
&  +\left\Vert \boldsymbol{\tau}_{2}\right\Vert _{\min_{c}}^{2}-\left\Vert
\boldsymbol{\tau}_{2}\right\Vert \left\Vert \boldsymbol{\tau}_{1}\right\Vert
\cos\theta_{\boldsymbol{\tau}_{2}\boldsymbol{\tau}_{1}}\\
&  =\sum\nolimits_{i=1}^{l_{1}}\psi_{1i\ast}\left(  1-\xi_{i}\right)
+\sum\nolimits_{i=1}^{l_{2}}\psi_{2i\ast}\left(  1-\xi_{i}\right)  \text{.}%
\end{align*}

The eigenenergy $\left\Vert \boldsymbol{\tau}\right\Vert _{\min_{c}}%
^{2}=\left\Vert \boldsymbol{\tau}_{1}-\boldsymbol{\tau}_{2}\right\Vert
_{\min_{c}}^{2}$ is the state of a discrete, linear classification system
$\boldsymbol{\tau}^{T}\mathbf{x}+\tau_{0}\overset{\omega_{1}}{\underset{\omega
_{2}}{\gtrless}}0$ that is associated with the position or location of a dual
likelihood ratio:%
\begin{align*}
\boldsymbol{\psi}  &  \triangleq\widehat{\Lambda}_{\boldsymbol{\psi}}\left(
\mathbf{x}\right)  =p\left(  \widehat{\Lambda}_{\boldsymbol{\psi}}\left(
\mathbf{x}\right)  |\omega_{1}\right)  +p\left(  \widehat{\Lambda
}_{\boldsymbol{\psi}}\left(  \mathbf{x}\right)  |\omega_{2}\right) \\
&  =\boldsymbol{\psi}_{1}+\boldsymbol{\psi}_{2}\\
&  =\sum\nolimits_{i=1}^{l_{1}}\psi_{1_{i_{\ast}}}\frac{\mathbf{x}%
_{1_{i_{\ast}}}}{\left\Vert \mathbf{x}_{1_{i_{\ast}}}\right\Vert }%
+\sum\nolimits_{i=1}^{l_{2}}\psi_{2_{i_{\ast}}}\frac{\mathbf{x}_{2_{i_{\ast}}%
}}{\left\Vert \mathbf{x}_{2_{i_{\ast}}}\right\Vert }\text{,}%
\end{align*}
which is constrained to be in statistical equilibrium:%
\[
\sum\nolimits_{i=1}^{l_{1}}\psi_{1_{i_{\ast}}}\frac{\mathbf{x}_{1_{i_{\ast}}}%
}{\left\Vert \mathbf{x}_{1_{i_{\ast}}}\right\Vert }=\sum\nolimits_{i=1}%
^{l_{2}}\psi_{2_{i_{\ast}}}\frac{\mathbf{x}_{2_{i_{\ast}}}}{\left\Vert
\mathbf{x}_{2_{i_{\ast}}}\right\Vert }\text{,}%
\]
and the locus of a corresponding linear decision boundary $\boldsymbol{\tau
}^{T}\mathbf{x}+\tau_{0}=0$.

Thus, any given discrete, linear classification system $\boldsymbol{\tau}%
^{T}\mathbf{x}+\tau_{0}\overset{\omega_{1}}{\underset{\omega_{2}}{\gtrless}}0$
exhibits an error rate that is consistent with the expected risk
$\mathfrak{R}_{\mathfrak{\min}}\left(  Z|\widehat{\Lambda}_{\boldsymbol{\tau}%
}\left(  \mathbf{x}\right)  \right)  $ and the corresponding eigenenergy
$E_{\min}\left(  Z|\widehat{\Lambda}_{\boldsymbol{\tau}}\left(  \mathbf{x}%
\right)  \right)  $ of the classification system: for all random vectors
$\mathbf{x}$ that are generated according to $p\left(  \mathbf{x}|\omega
_{1}\right)  $ and $p\left(  \mathbf{x}|\omega_{2}\right)  $, where $p\left(
\mathbf{x}|\omega_{1}\right)  $ and $p\left(  \mathbf{x}|\omega_{2}\right)  $
are related to statistical distributions of random vectors $\mathbf{x}$ that
have similar covariance matrices and constant or unchanging statistics.

The total eigenenergy of a discrete, linear classification system
$\boldsymbol{\tau}^{T}\mathbf{x}+\tau_{0}\overset{\omega_{1}}{\underset{\omega
_{2}}{\gtrless}}0$ is found by adding up contributions from characteristics of
the classification system:

The eigenenergies $E_{\min}\left(  Z|p\left(  \widehat{\Lambda}%
_{\boldsymbol{\tau}}\left(  \mathbf{x}\right)  |\omega_{1}\right)  \right)  $
and $E_{\min}\left(  Z|p\left(  \widehat{\Lambda}_{\boldsymbol{\tau}}\left(
\mathbf{x}\right)  |\omega_{2}\right)  \right)  $ associated with the
positions or locations of the parameter vectors of likelihoods $p\left(
\widehat{\Lambda}_{\boldsymbol{\tau}}\left(  \mathbf{x}\right)  |\omega
_{1}\right)  $ and $p\left(  \widehat{\Lambda}_{\boldsymbol{\tau}}\left(
\mathbf{x}\right)  |\omega_{2}\right)  $, where%
\[
E_{\min}\left(  Z|p\left(  \widehat{\Lambda}_{\boldsymbol{\tau}}\left(
\mathbf{x}\right)  |\omega_{1}\right)  \right)  =\left\Vert \boldsymbol{\tau
}_{1}\right\Vert _{\min_{c}}^{2}\text{ \ and \ }E_{\min}\left(  Z|p\left(
\widehat{\Lambda}_{\boldsymbol{\tau}}\left(  \mathbf{x}\right)  |\omega
_{2}\right)  \right)  =\left\Vert \boldsymbol{\tau}_{2}\right\Vert _{\min_{c}%
}^{2}%
\]
are related to eigenenergies associated with positions and potential locations
of extreme points that lie in either overlapping regions or tails regions of
statistical distributions related to the class-conditional probability density
functions $p\left(  \mathbf{x}|\omega_{1}\right)  $ and $p\left(
\mathbf{x}|\omega_{2}\right)  $, and the total eigenenergy $\left\Vert
\boldsymbol{\tau}\right\Vert _{\min_{c}}^{2}$ satisfies the vector equations
\begin{align*}
\left\Vert \boldsymbol{\tau}\right\Vert _{\min_{c}}^{2}  &  =\left\Vert
\boldsymbol{\tau}_{1}-\boldsymbol{\tau}_{2}\right\Vert _{\min_{c}}^{2}\\
&  =\left\Vert \boldsymbol{\tau}_{1}\right\Vert _{\min_{c}}^{2}-\left\Vert
\boldsymbol{\tau}_{1}\right\Vert \left\Vert \boldsymbol{\tau}_{2}\right\Vert
\cos\theta_{\boldsymbol{\tau}_{1}\boldsymbol{\tau}_{2}}\\
&  +\left\Vert \boldsymbol{\tau}_{2}\right\Vert _{\min_{c}}^{2}-\left\Vert
\boldsymbol{\tau}_{2}\right\Vert \left\Vert \boldsymbol{\tau}_{1}\right\Vert
\cos\theta_{\boldsymbol{\tau}_{2}\boldsymbol{\tau}_{1}}%
\end{align*}
and%
\[
\left\Vert \boldsymbol{\tau}\right\Vert _{\min_{c}}^{2}=\sum\nolimits_{i=1}%
^{l_{1}}\psi_{1i\ast}\left(  1-\xi_{i}\right)  +\sum\nolimits_{i=1}^{l_{2}%
}\psi_{2i\ast}\left(  1-\xi_{i}\right)  \text{.}%
\]

Any given discrete, linear classification system that is determined by a
likelihood ratio test:%
\[
\boldsymbol{\tau}^{T}\mathbf{x}+\tau_{0}\overset{\omega_{1}}{\underset{\omega
_{2}}{\gtrless}}0\text{,}%
\]
where the class-conditional probability density functions $p\left(
\mathbf{x}|\omega_{1}\right)  $ and $p\left(  \mathbf{x}|\omega_{2}\right)  $
are related to statistical distributions of random vectors $\mathbf{x}$ that
have constant or unchanging statistics and similar covariance matrices, and
the locus of a linear decision boundary:%
\[
D\left(  \mathbf{x}\right)  :\boldsymbol{\tau}^{T}\mathbf{x}+\tau_{0}=0
\]
is governed by the locus of a dual likelihood ratio $p\left(  \widehat{\Lambda
}_{\boldsymbol{\psi}}\left(  \mathbf{x}\right)  |\omega_{1}\right)  +p\left(
\widehat{\Lambda}_{\boldsymbol{\psi}}\left(  \mathbf{x}\right)  |\omega
_{2}\right)  $ in statistical equilibrium:%
\[
p\left(  \widehat{\Lambda}_{\boldsymbol{\psi}}\left(  \mathbf{x}\right)
|\omega_{1}\right)  \rightleftharpoons p\left(  \widehat{\Lambda
}_{\boldsymbol{\psi}}\left(  \mathbf{x}\right)  |\omega_{2}\right)  \text{,}%
\]
is a closed classification system.

Thus, the total eigenenergy $E_{\min}\left(  Z|\widehat{\Lambda}%
_{\boldsymbol{\tau}}\left(  \mathbf{x}\right)  \right)  $%
\begin{align*}
E_{\min}\left(  Z|\widehat{\Lambda}_{\boldsymbol{\tau}}\left(  \mathbf{x}%
\right)  \right)   &  \triangleq\left\Vert \boldsymbol{\tau}\right\Vert
_{\min_{c}}^{2}\\
&  =\left\Vert \boldsymbol{\tau}_{1}-\boldsymbol{\tau}_{2}\right\Vert
_{\min_{c}}^{2}\\
&  =\sum\nolimits_{i=1}^{l_{1}}\psi_{1i\ast}\left(  1-\xi_{i}\right)
+\sum\nolimits_{i=1}^{l_{2}}\psi_{2i\ast}\left(  1-\xi_{i}\right)
\end{align*}
of any given discrete, linear classification system $\boldsymbol{\tau}%
^{T}\mathbf{x}+\tau_{0}\overset{\omega_{1}}{\underset{\omega_{2}}{\gtrless}}0$
is conserved and remains relatively constant.

Therefore, the eigenenergy $E_{\min}\left(  Z|\widehat{\Lambda}%
_{\boldsymbol{\tau}}\left(  \mathbf{x}\right)  \right)  $ of a discrete,
linear classification system $\boldsymbol{\tau}^{T}\mathbf{x}+\tau
_{0}\overset{\omega_{1}}{\underset{\omega_{2}}{\gtrless}}0$ cannot be created
or destroyed, but only transferred from one classification system to another.

It follows that the corresponding expected risk $\mathfrak{R}_{\mathfrak{\min
}}\left(  Z|\widehat{\Lambda}_{\boldsymbol{\tau}}\left(  \mathbf{x}\right)
\right)  $ of a discrete, linear classification system $\boldsymbol{\tau}%
^{T}\mathbf{x}+\tau_{0}\overset{\omega_{1}}{\underset{\omega_{2}}{\gtrless}}0$
cannot be created or destroyed, but only transferred from one classification
system to another.

I\ will now identify the fundamental property which is common to each of the
scaled extreme points on any given likelihood ratio $\widehat{\Lambda
}_{\boldsymbol{\tau}}\left(  \mathbf{x}\right)  $ and linear decision boundary
$D_{0}\left(  \mathbf{x}\right)  $ that is determined by a linear eigenlocus
classification system $\boldsymbol{\tau}^{T}\mathbf{x}+\tau_{0}\overset{\omega
_{1}}{\underset{\omega_{2}}{\gtrless}}0$.

\subsubsection{Inherent Property of Eigen-scaled Extreme Points on
$\boldsymbol{\tau}_{1}-\boldsymbol{\tau}_{2}$}

Given that a linear eigenlocus $\boldsymbol{\tau}=\boldsymbol{\tau}%
_{1}-\boldsymbol{\tau}_{2}$ is a locus of likelihoods that determines a
likelihood ratio $\widehat{\Lambda}_{\boldsymbol{\tau}}\left(  \mathbf{x}%
\right)  $ and a locus of principal eigenaxis components that determines the
coordinate system of a linear decision boundary $D_{0}\left(  \mathbf{x}%
\right)  $, it follows that the total allowed eigenenergy%
\[
\left\Vert \psi_{1_{i_{\ast}}}\mathbf{x}_{1_{i_{\ast}}}\right\Vert _{\min_{c}%
}^{2}\text{ or \ }\left\Vert \psi_{2_{i_{\ast}}}\mathbf{x}_{2_{i_{\ast}}%
}\right\Vert _{\min_{c}}^{2}%
\]
exhibited by each scaled extreme vector%
\[
\psi_{1_{i_{\ast}}}\mathbf{x}_{1_{i_{\ast}}}\text{ or \ }\psi_{2_{i_{\ast}}%
}\mathbf{x}_{2_{i_{\ast}}}%
\]
on $\boldsymbol{\tau}_{1}-\boldsymbol{\tau}_{2}$ and the corresponding
class-conditional risk:%
\[
\int_{Z_{2}}p\left(  \mathbf{x}_{1_{i_{\ast}}}|\operatorname{comp}%
_{\overrightarrow{\mathbf{x}_{1i\ast}}}\left(
\overrightarrow{\boldsymbol{\tau}}\right)  \right)  d\boldsymbol{\tau}%
_{1}\left(  \mathbf{x}_{1_{i_{\ast}}}\right)  \text{ or }\int_{Z_{1}}p\left(
\mathbf{x}_{2_{i_{\ast}}}|\operatorname{comp}_{\overrightarrow{\mathbf{x}%
_{2i\ast}}}\left(  \overrightarrow{\boldsymbol{\tau}}\right)  \right)
d\boldsymbol{\tau}_{2}\left(  \mathbf{x}_{2_{i\ast}}\right)
\]
or class-conditional counter risk:%
\[
\int_{Z_{1}}p\left(  \mathbf{x}_{1_{i_{\ast}}}|\operatorname{comp}%
_{\overrightarrow{\mathbf{x}_{1i\ast}}}\left(
\overrightarrow{\boldsymbol{\tau}}\right)  \right)  d\boldsymbol{\tau}%
_{1}\left(  \mathbf{x}_{1_{i_{\ast}}}\right)  \text{ or }\int_{Z_{2}}p\left(
\mathbf{x}_{2_{i_{\ast}}}|\operatorname{comp}_{\overrightarrow{\mathbf{x}%
_{2i\ast}}}\left(  \overrightarrow{\boldsymbol{\tau}}\right)  \right)
d\boldsymbol{\tau}_{2}\left(  \mathbf{x}_{2_{i\ast}}\right)
\]
possessed by each extreme point $\mathbf{x}_{1_{i_{\ast}}}$ or $\mathbf{x}%
_{2_{i_{\ast}}}$, which are determined by $\left\Vert \psi_{1_{i_{\ast}}%
}\mathbf{x}_{1_{i_{\ast}}}\right\Vert _{\min_{c}}^{2}$ or $\left\Vert
\psi_{2_{i_{\ast}}}\mathbf{x}_{2_{i_{\ast}}}\right\Vert _{\min_{c}}^{2}$,
\emph{jointly satisfy} the fundamental linear eigenlocus integral equation of
binary classification in Eq. (\ref{Linear Eigenlocus Integral Equation V}).
Thereby, it is concluded that the \emph{fundamental property} possessed by
each of the scaled extreme points on a linear eigenlocus $\boldsymbol{\tau
}_{1}-\boldsymbol{\tau}_{2}$ is the \emph{total allowed eigenenergy} exhibited
by a corresponding, scaled extreme vector.

I will now devise an expression for a linear eigenlocus that is a locus of
discrete conditional probabilities.

\section{Linear Eigenlocus of Probabilities}

Write a linear eigenlocus $\boldsymbol{\tau}$ in terms of%
\begin{align*}
\boldsymbol{\tau}  &  =\lambda_{\max_{\psi}}^{-1}\sum\nolimits_{i=1}^{l_{1}%
}\frac{\mathbf{x}_{1_{i_{\ast}}}}{\left\Vert \mathbf{x}_{1_{i_{\ast}}%
}\right\Vert }\left\Vert \mathbf{x}_{1_{i_{\ast}}}\right\Vert ^{2}%
\widehat{\operatorname{cov}}_{sm_{\updownarrow}}\left(  \mathbf{x}%
_{1_{i_{\ast}}}\right) \\
&  -\lambda_{\max_{\psi}}^{-1}\sum\nolimits_{i=1}^{l_{2}}\frac{\mathbf{x}%
_{2_{i_{\ast}}}}{\left\Vert \mathbf{x}_{2_{i_{\ast}}}\right\Vert }\left\Vert
\mathbf{x}_{2_{i_{\ast}}}\right\Vert ^{2}\widehat{\operatorname{cov}%
}_{sm_{\updownarrow}}\left(  \mathbf{x}_{2_{i_{\ast}}}\right)  \text{,}%
\end{align*}
where $\widehat{\operatorname{cov}}_{sm_{\updownarrow}}\left(  \mathbf{x}%
_{1_{i_{\ast}}}\right)  $ and $\widehat{\operatorname{cov}}_{sm_{\updownarrow
}}\left(  \mathbf{x}_{2_{i_{\ast}}}\right)  $ denote the symmetrically
balanced, signed magnitudes in Eqs (\ref{Unidirectional Scaling Term One1})
and (\ref{Unidirectional Scaling Term Two1}): the terms $\frac{\left\Vert
\mathbf{x}_{1_{i_{\ast}}}\right\Vert }{\left\Vert \mathbf{x}_{1_{i_{\ast}}%
}\right\Vert }$ and $\frac{\left\Vert \mathbf{x}_{2_{i_{\ast}}}\right\Vert
}{\left\Vert \mathbf{x}_{2_{i_{\ast}}}\right\Vert }$ have been introduced and rearranged.

Next, rewrite $\boldsymbol{\tau}$ in terms of total allowed eigenenergies%
\begin{align}
\boldsymbol{\tau}  &  =\sum\nolimits_{i=1}^{l_{1}}\left\Vert \lambda
_{\max_{\psi}}^{-1}\left(  \widehat{\operatorname{cov}}_{sm_{\updownarrow}%
}\left(  \mathbf{x}_{1_{i_{\ast}}}\right)  \right)  ^{\frac{1}{2}}%
\mathbf{x}_{1_{i_{\ast}}}\right\Vert _{\min_{c}}^{2}\frac{\mathbf{x}%
_{1_{i_{\ast}}}}{\left\Vert \mathbf{x}_{1_{i_{\ast}}}\right\Vert
}\label{Probabilisitc Expression for Normal Eigenlocus}\\
&  -\sum\nolimits_{i=1}^{l_{2}}\left\Vert \lambda_{\max_{\psi}}^{-1}\left(
\widehat{\operatorname{cov}}_{sm_{\updownarrow}}\left(  \mathbf{x}%
_{2_{i_{\ast}}}\right)  \right)  ^{\frac{1}{2}}\mathbf{x}_{2_{i_{\ast}}%
}\right\Vert _{\min_{c}}^{2}\frac{\mathbf{x}_{2_{i_{\ast}}}}{\left\Vert
\mathbf{x}_{2_{i_{\ast}}}\right\Vert }\text{,}\nonumber
\end{align}
where the conditional probability $\mathcal{P}\left(  \mathbf{x}_{1_{i_{\ast}%
}}\mathbf{|}\tilde{Z}\left(  \mathbf{x}_{1_{i_{\ast}}}\right)  \right)  $ of
observing an $\mathbf{x}_{1_{i_{\ast}}}$ extreme point within a localized
region $\tilde{Z}\left(  \mathbf{x}_{1_{i_{\ast}}}\right)  $ of a decision
space $Z=Z_{1}+Z_{2}$ is given by the expression:%
\[
\mathcal{P}\left(  \mathbf{x}_{1_{i_{\ast}}}\mathbf{|}\tilde{Z}\left(
\mathbf{x}_{1_{i_{\ast}}}\right)  \right)  =\left\Vert \lambda_{\max_{\psi}%
}^{-1}\left(  \widehat{\operatorname{cov}}_{sm_{\updownarrow}}\left(
\mathbf{x}_{1_{i_{\ast}}}\right)  \right)  ^{\frac{1}{2}}\mathbf{x}%
_{1_{i_{\ast}}}\right\Vert _{\min_{c}}^{2}\text{,}%
\]
where $\tilde{Z}\left(  \mathbf{x}_{1_{i_{\ast}}}\right)  \subset Z_{1}$ or
$\tilde{Z}\left(  \mathbf{x}_{1_{i_{\ast}}}\right)  \subset Z_{2}$, and the
conditional probability $\mathcal{P}\left(  \mathbf{x}_{2_{i_{\ast}}}%
|\tilde{Z}\left(  \mathbf{x}_{2_{i_{\ast}}}\right)  \right)  $ of observing an
$\mathbf{x}_{2_{i_{\ast}}}$ extreme point within a localized region $\tilde
{Z}\left(  \mathbf{x}_{2_{i_{\ast}}}\right)  $ of a decision space $Z$ is
given by the expression:%
\[
\mathcal{P}\left(  \mathbf{x}_{2_{i_{\ast}}}|\tilde{Z}\left(  \mathbf{x}%
_{2_{i_{\ast}}}\right)  \right)  =\left\Vert \lambda_{\max_{\psi}}^{-1}\left(
\widehat{\operatorname{cov}}_{sm_{\updownarrow}}\left(  \mathbf{x}%
_{2_{i_{\ast}}}\right)  \right)  ^{\frac{1}{2}}\mathbf{x}_{2_{i_{\ast}}%
}\right\Vert _{\min_{c}}^{2}\text{,}%
\]
where $\tilde{Z}\left(  \mathbf{x}_{2_{i_{\ast}}}\right)  \subset Z_{1}$ or
$\tilde{Z}\left(  \mathbf{x}_{2_{i_{\ast}}}\right)  \subset Z_{2}$.

Now rewrite Eq. (\ref{Probabilisitc Expression for Normal Eigenlocus}) as a
locus of discrete conditional probabilities:%
\begin{align}
\boldsymbol{\tau}_{1}-\boldsymbol{\tau}_{2}  &  =\sum\nolimits_{i=1}^{l_{1}%
}\mathcal{P}\left(  \mathbf{x}_{1_{i_{\ast}}}|\tilde{Z}\left(  \mathbf{x}%
_{1_{i_{\ast}}}\right)  \right)  \frac{\mathbf{x}_{1_{i_{\ast}}}}{\left\Vert
\mathbf{x}_{1_{i_{\ast}}}\right\Vert }%
\label{SDNE Conditional Likelihood Ratio}\\
&  -\sum\nolimits_{i=1}^{l_{2}}\mathcal{P}\left(  \mathbf{x}_{2_{i_{\ast}}%
}\mathbf{|}\tilde{Z}\left(  \mathbf{x}_{2_{i_{\ast}}}\right)  \right)
\frac{\mathbf{x}_{2_{i_{\ast}}}}{\left\Vert \mathbf{x}_{2_{i_{\ast}}%
}\right\Vert }\nonumber
\end{align}
which provides discrete measures for conditional probabilities of
classification errors $\mathcal{P}_{\min_{e}}\left(  \mathbf{x}_{1_{i_{\ast}}%
}|Z_{2}\left(  \mathbf{x}_{1_{i_{\ast}}}\right)  \right)  $ and $\mathcal{P}%
_{\min_{e}}\left(  \mathbf{x}_{2_{i_{\ast}}}|Z_{1}\left(  \mathbf{x}%
_{2_{i_{\ast}}}\right)  \right)  $ for $\mathbf{x}_{1_{i_{\ast}}}$ extreme
points that lie in the $Z_{2}$ decision region and $\mathbf{x}_{2_{i_{\ast}}}$
extreme points that lie in the $Z_{1}$ decision region.

I\ will now use Eq. (\ref{SDNE Conditional Likelihood Ratio}) to devise a
probabilistic expression for a linear eigenlocus discriminant function.

\subsection{A Probabilistic Expression for $\boldsymbol{\tau}$}

Returning to Eq. (\ref{Statistical Locus of Category Decision}), consider the
estimate $\widehat{\Lambda}_{\boldsymbol{\tau}}\left(  \mathbf{x}\right)  $
that an unknown pattern vector $\mathbf{x}$ is located within some particular
region of $%
\mathbb{R}
^{d}$%
\begin{align*}
\widehat{\Lambda}_{\boldsymbol{\tau}}\left(  \mathbf{x}\right)   &  =\left(
\mathbf{x}-\widehat{\mathbf{x}}_{i\ast}\right)  ^{T}\boldsymbol{\tau
}\mathbf{/}\left\Vert \boldsymbol{\tau}\right\Vert \\
&  \mathbf{+}\frac{1}{\left\Vert \boldsymbol{\tau}\right\Vert }\sum
\nolimits_{i=1}^{l}y_{i}\left(  1-\xi_{i}\right)
\end{align*}
based on the value of the decision locus $\operatorname{comp}%
_{\overrightarrow{\widehat{\boldsymbol{\tau}}}}\left(  \overrightarrow{\left(
\mathbf{x}-\widehat{\mathbf{x}}_{i\ast}\right)  }\right)  $ and class
membership statistic $\frac{1}{\left\Vert \boldsymbol{\tau}\right\Vert }%
\sum\nolimits_{i=1}^{l}y_{i}\left(  1-\xi_{i}\right)  $, where
$\operatorname{comp}_{\overrightarrow{\widehat{\boldsymbol{\tau}}}}\left(
\overrightarrow{\left(  \mathbf{x}-\widehat{\mathbf{x}}_{i\ast}\right)
}\right)  $ denotes a signed magnitude $\left\Vert \mathbf{x}%
-\widehat{\mathbf{x}}_{i\ast}\right\Vert \cos\theta$ along the axis of
$\widehat{\boldsymbol{\tau}}$, $\theta$ is the angle between the vector
$\mathbf{x}-\widehat{\mathbf{x}}_{i\ast}$ and $\widehat{\boldsymbol{\tau}}$,
and $\widehat{\boldsymbol{\tau}}$ denotes the unit linear eigenlocus
$\boldsymbol{\tau}\mathbf{/}\left\Vert \boldsymbol{\tau}\right\Vert $.

I will now demonstrate that the signed magnitude expression $\left(
\mathbf{x}-\widehat{\mathbf{x}}_{i\ast}\right)  ^{T}\boldsymbol{\tau
}\mathbf{/}\left\Vert \boldsymbol{\tau}\right\Vert $ is a locus of discrete
conditional probabilities.

Substitute the expression for $\boldsymbol{\tau}_{1}-\boldsymbol{\tau}_{2}$ in
Eq. (\ref{SDNE Conditional Likelihood Ratio}) into the expression for the
linear eigenlocus test in Eq. (\ref{NormalEigenlocusTestStatistic}). Denote
the unit primal principal eigenaxis components $\frac{\mathbf{x}_{1_{i_{\ast}%
}}}{\left\Vert \mathbf{x}_{1_{i_{\ast}}}\right\Vert }$ and $\frac
{\mathbf{x}_{2_{i_{\ast}}}}{\left\Vert \mathbf{x}_{2_{i_{\ast}}}\right\Vert }$
by $\widehat{\mathbf{x}}_{1_{i_{\ast}}}$ and $\widehat{\mathbf{x}}%
_{2_{i_{\ast}}}$. It follows that the probability that the unknown pattern
vector $\mathbf{x}$ is located within a specific region of $%
\mathbb{R}
^{d}$ is provided by the expression%
\begin{align*}
\widehat{\Lambda}_{\boldsymbol{\tau}}\left(  \mathbf{x}\right)   &
=\sum\nolimits_{i=1}^{l_{1}}\left[  \left(  \mathbf{x}-\widehat{\mathbf{x}%
}_{i\ast}\right)  ^{T}\widehat{\mathbf{x}}_{1_{i_{\ast}}}\right]
\mathcal{P}\left(  \mathbf{x}_{1_{i_{\ast}}}|\tilde{Z}\left(  \mathbf{x}%
_{1_{i_{\ast}}}\right)  \right) \\
&  -\sum\nolimits_{i=1}^{l_{2}}\left[  \left(  \mathbf{x}-\widehat{\mathbf{x}%
}_{i\ast}\right)  ^{T}\widehat{\mathbf{x}}_{2_{i_{\ast}}}\right]
\mathcal{P}\left(  \mathbf{x}_{2_{i_{\ast}}}|\tilde{Z}\left(  \mathbf{x}%
_{2_{i_{\ast}}}\right)  \right) \\
&  \mathbf{+}\frac{1}{\left\Vert \boldsymbol{\tau}\right\Vert }\sum
\nolimits_{i=1}^{l}y_{i}\left(  1-\xi_{i}\right)  \text{,}%
\end{align*}
where $\mathcal{P}\left(  \mathbf{x}_{1_{i_{\ast}}}|\tilde{Z}\left(
\mathbf{x}_{1_{i_{\ast}}}\right)  \right)  $ and $\mathcal{P}\left(
\mathbf{x}_{2_{i_{\ast}}}|\tilde{Z}\left(  \mathbf{x}_{2_{i_{\ast}}}\right)
\right)  $ provides discrete measures for conditional probabilities of
classification errors%
\[
\mathcal{P}_{\min_{e}}\left(  \mathbf{x}_{1_{i_{\ast}}}|Z_{2}\left(
\mathbf{x}_{1_{i_{\ast}}}\right)  \right)  \text{ and }\mathcal{P}_{\min_{e}%
}\left(  \mathbf{x}_{2_{i_{\ast}}}|Z_{1}\left(  \mathbf{x}_{2_{i_{\ast}}%
}\right)  \right)
\]
for $\mathbf{x}_{1_{i_{\ast}}}$ extreme points that lie in the $Z_{2}$
decision region and $\mathbf{x}_{2_{i_{\ast}}}$ extreme points that lie in the
$Z_{1}$ decision region or conditional probabilities of counter risks%
\[
\mathcal{P}_{\min_{e}}\left(  \mathbf{x}_{1_{i_{\ast}}}|Z_{1}\left(
\mathbf{x}_{1_{i_{\ast}}}\right)  \right)  \text{ and }\mathcal{P}_{\min_{e}%
}\left(  \mathbf{x}_{2_{i_{\ast}}}|Z_{2}\left(  \mathbf{x}_{2_{i_{\ast}}%
}\right)  \right)
\]
for $\mathbf{x}_{1_{i_{\ast}}}$ extreme points that lie in the $Z_{1}$
decision region and $\mathbf{x}_{2_{i_{\ast}}}$ extreme points that lie in the
$Z_{2}$ decision region.

The above expression reduces to a locus of discrete conditional probabilities%
\begin{align}
\Lambda_{\boldsymbol{\tau}}\left(  \mathbf{x}\right)   &  =\sum\nolimits_{i=1}%
^{l_{1}}\operatorname{comp}_{\overrightarrow{\widehat{\mathbf{x}}_{1_{i_{\ast
}}}}}\left(  \overrightarrow{\left(  \mathbf{x}-\widehat{\mathbf{x}}_{i\ast
}\right)  }\right)  \mathcal{P}\left(  \mathbf{x}_{1_{i_{\ast}}}|\tilde
{Z}\left(  \mathbf{x}_{1_{i_{\ast}}}\right)  \right)
\label{Normal Eigenlocus Likelihood Ratio}\\
&  -\sum\nolimits_{i=1}^{l_{2}}\operatorname{comp}%
_{\overrightarrow{\widehat{\mathbf{x}}_{2_{i_{\ast}}}}}\left(
\overrightarrow{\left(  \mathbf{x}-\widehat{\mathbf{x}}_{i\ast}\right)
}\right)  \mathcal{P}\left(  \mathbf{x}_{2_{i_{\ast}}}|\tilde{Z}\left(
\mathbf{x}_{2_{i_{\ast}}}\right)  \right) \nonumber\\
&  \mathbf{+}\frac{1}{\left\Vert \boldsymbol{\tau}\right\Vert }\sum
\nolimits_{i=1}^{l}y_{i}\left(  1-\xi_{i}\right)  \text{,}\nonumber
\end{align}
so that the conditional probability $\mathcal{P}\left(  \mathbf{x|}\tilde
{Z}\left(  \mathbf{x}_{1_{i_{\ast}}}\right)  \right)  $ of finding the unknown
pattern vector $\mathbf{x}$ within the localized region $\tilde{Z}\left(
\mathbf{x}_{1_{i_{\ast}}}\right)  $ of the decision space $Z$ is determined by
the likelihood statistic:%
\begin{equation}
\mathcal{P}\left(  \mathbf{x|}\tilde{Z}\left(  \mathbf{x}_{1_{i_{\ast}}%
}\right)  \right)  =\operatorname{comp}_{\overrightarrow{\widehat{\mathbf{x}%
}_{1_{i_{\ast}}}}}\left(  \overrightarrow{\left(  \mathbf{x}%
-\widehat{\mathbf{x}}_{i\ast}\right)  }\right)  \mathcal{P}\left(
\mathbf{x}_{1_{i_{\ast}}}|\tilde{Z}\left(  \mathbf{x}_{1_{i_{\ast}}}\right)
\right)  \text{,} \label{Probability Estimate One}%
\end{equation}
and the conditional probability $\mathcal{P}\left(  \mathbf{x|}\tilde
{Z}\left(  \mathbf{x}_{2_{i_{\ast}}}\right)  \right)  $ of finding the unknown
pattern vector $\mathbf{x}$ within the localized region $\tilde{Z}\left(
\mathbf{x}_{2_{i_{\ast}}}\right)  $ of the decision space $Z$ is determined by
the likelihood statistic:%
\begin{equation}
\mathcal{P}\left(  \mathbf{x|}\tilde{Z}\left(  \mathbf{x}_{2_{i_{\ast}}%
}\right)  \right)  =\operatorname{comp}_{\overrightarrow{\widehat{\mathbf{x}%
}_{2_{i_{\ast}}}}}\left(  \overrightarrow{\left(  \mathbf{x}%
-\widehat{\mathbf{x}}_{i\ast}\right)  }\right)  \mathcal{P}\left(
\mathbf{x}_{2_{i_{\ast}}}|\tilde{Z}\left(  \mathbf{x}_{2_{i_{\ast}}}\right)
\right)  \text{.} \label{Probability Estimate Two}%
\end{equation}

Accordingly, the likelihood statistic $\mathcal{P}\left(  \mathbf{x|}\tilde
{Z}\left(  \mathbf{x}_{1_{i_{\ast}}}\right)  \right)  $ in Eq.
(\ref{Probability Estimate One}) is proportional, according to the signed
magnitude $\operatorname{comp}_{\overrightarrow{\widehat{\mathbf{x}%
}_{1_{i_{\ast}}}}}\left(  \overrightarrow{\left(  \mathbf{x}%
-\widehat{\mathbf{x}}_{i\ast}\right)  }\right)  $ along the axis of
$\widehat{\mathbf{x}}_{1_{i_{\ast}}}$, to the conditional probability
$\mathcal{P}\left(  \mathbf{x}_{1_{i_{\ast}}}|\tilde{Z}\left(  \mathbf{x}%
_{1_{i_{\ast}}}\right)  \right)  $ of finding the extreme point $\mathbf{x}%
_{1_{i_{\ast}}}$ within a localized region $\tilde{Z}\left(  \mathbf{x}%
_{1_{i_{\ast}}}\right)  $ of the decision space $Z$. Similarly, the likelihood
statistic $\mathcal{P}\left(  \mathbf{x|}\tilde{Z}\left(  \mathbf{x}%
_{2_{i_{\ast}}}\right)  \right)  $ in Eq. (\ref{Probability Estimate Two}) is
proportional, according to the signed magnitude $\operatorname{comp}%
_{\overrightarrow{\widehat{\mathbf{x}}_{2_{i_{\ast}}}}}\left(
\overrightarrow{\left(  \mathbf{x}-\widehat{\mathbf{x}}_{i\ast}\right)
}\right)  $ along the axis of $\widehat{\mathbf{x}}_{2_{i_{\ast}}}$, to the
conditional probability $P\left(  \mathbf{x}_{2_{i_{\ast}}}|\tilde{Z}\left(
\mathbf{x}_{2_{i_{\ast}}}\right)  \right)  $ of finding the extreme point
$\mathbf{x}_{2_{i_{\ast}}}$ within a localized region $\tilde{Z}\left(
\mathbf{x}_{2_{i_{\ast}}}\right)  $ of the decision space $Z$.

Thus, it is concluded that the signed magnitude expression $\left(
\mathbf{x}-\widehat{\mathbf{x}}_{i\ast}\right)  ^{T}\boldsymbol{\tau
}\mathbf{/}\left\Vert \boldsymbol{\tau}\right\Vert $ in Eq.
(\ref{Statistical Locus of Category Decision}) is a locus of discrete
conditional probabilities that satisfies the linear eigenlocus integral
equation of binary classification in Eq.
(\ref{Linear Eigenlocus Integral Equation V}).

I\ will now devise a system of data-driven, locus equations that generate
computer-implemented, optimal quadratic classification systems. The general
trend of my arguments is similar to the general trend of my arguments for
linear eigenlocus transforms. My discoveries are based on useful relations
between geometric locus methods, geometric methods in reproducing kernel
Hilbert spaces, statistics, and the binary classification theorem that I\ have derived.

\section{Optimal Quadratic Classification Systems}

I will begin by outlining my design for computer-implemented, optimal
quadratic classification systems. Such computer-implemented systems are
scalable modules for optimal statistical pattern recognition systems, all of
which are capable of performing a wide variety of statistical pattern
recognition tasks, where any given $M$-class statistical pattern recognition
system achieves the lowest possible error rate and has the best generalization
performance for its $M$-class feature space.

\subsection*{Problem Formulation}

\begin{flushleft}
The formulation of a system of data-driven, locus equations that generates
computer-implemented, optimal quadratic classification systems requires
solving three fundamental problems:
\end{flushleft}

\paragraph{Problem $\mathbf{1}$}

\textit{Define the geometric figures in a quadratic classification system,
where geometric figures involve points, vectors, line segments, angles,
regions, and quadratic curves or surfaces.}

\paragraph{Problem $\mathbf{2}$}

\textit{Define the geometric and statistical properties exhibited by each of
the geometric figures.}

\paragraph{Problem $\mathbf{3}$}

\textit{Define the forms of the data-driven, locus equations that determine
the geometric figures.}

\subsection*{The Solution}

\begin{flushleft}
I\ have formulated a solution that answers all three problems. My solution is
based on three ideas:
\end{flushleft}

\paragraph{Idea $\mathbf{1}$}

Devise \emph{a dual locus of data points} that determines quadratic decision
boundaries \emph{and} likelihood ratios that achieve the lowest possible error rate.

\paragraph{Idea $\mathbf{2}$}

The dual locus of data points must \emph{determine} the \emph{coordinate
system} of the \emph{quadratic decision boundary}.

\paragraph{Idea $\mathbf{3}$}

The dual locus of data points \emph{must satisfy discrete versions of the
fundamental equations of binary classification for a classification system in
statistical equilibrium.}

\subsection*{Key Elements of the Solution}

\begin{flushleft}
The essential elements of the solution are outlined below.
\end{flushleft}

\paragraph{Locus of Principal Eigenaxis Components}

Returning to Eqs (\ref{Vector Equation of a Conic}) and
(\ref{Vector Equation of Circles and Spheres}), given that the vector
components of a principal eigenaxis specify all forms of quadratic curves and
surfaces, and all of the points on any given quadratic curve or surface
explicitly and exclusively reference the principal eigenaxis of the quadratic
locus, it follows that the principal eigenaxis of a quadratic decision
boundary provides an elegant, statistical eigen-coordinate system for a
quadratic classification system.

Therefore, the dual locus of data points \emph{must} be a \emph{locus of
principal eigenaxis components}.

\paragraph{Critical Minimum Eigenenergy Constraint}

Given Eqs (\ref{Characteristic Eigenenergy of Quadratic}) and
(\ref{Characteristic Eigenenergy of Quadratic 2}), it follows that the
principal eigenaxis of a quadratic decision boundary satisfies the quadratic
decision boundary in terms of its eigenenergy. Accordingly, the principal
eigenaxis of a quadratic decision boundary exhibits a characteristic
eigenenergy that is unique for the quadratic decision boundary. Thereby,
the\ \emph{important generalizations} for a quadratic decision boundary are
\emph{determined by} the \emph{eigenenergy} exhibited by its \emph{principal
eigenaxis}.

Therefore, the locus of principal eigenaxis components \emph{must satisfy} a
critical minimum, i.e., a total allowed, \emph{eigenenergy constraint}, such
that the locus of principal eigenaxis components satisfies a \emph{quadratic
decision boundary} in \emph{terms of} its critical minimum or total allowed
\emph{eigenenergies. }Thus, the locus of principal eigenaxis components must
satisfy the vector and equilibrium equation in Eqs
(\ref{Vector Equation of Likelihood Ratio and Decision Boundary}) and
(\ref{Equilibrium Equation of Likelihood Ratio and Decision Boundary}), the
integral equation in Eq.
(\ref{Integral Equation of Likelihood Ratio and Decision Boundary}), the
fundamental integral equation of binary classification in Eq.
(\ref{Equalizer Rule}), and the corresponding integral equation in Eq.
(\ref{Balancing of Bayes' Risks and Counteracting Risks}) \emph{in terms of
its total allowed eigenenergies}.

\paragraph{Extreme Points}

In order for the locus of principal eigenaxis components to implement a
likelihood ratio, the locus of principal eigenaxis components \emph{must} be
\emph{formed by data points} that lie in either overlapping regions or tail
region of data distributions, thereby determining \emph{decision regions}
based on forces associated with \emph{risks and counter risks}:\emph{ }which
are related to positions and potential locations of data points that lie in
either overlapping regions or tail region of data distributions, where an
unknown portion of the data points are the \emph{source of decision errors}.
Data points that lie in either overlapping regions or tail region of data
distributions will be called extreme points.

\paragraph{Parameter Vector of Class-conditional Densities}

Given that the locus of principal eigenaxis components must determine a
likelihood ratio for binary classification, it follows that the locus of
principal eigenaxis components \emph{must also} be a \emph{parameter vector}
that \emph{provides an estimate of class-conditional density functions}. Given
Eq.
(\emph{\ref{Equilibrium Equation of Likelihood Ratio and Decision Boundary}}),
it also follows that the parameter vector must be in statistical equilibrium.

\paragraph{Locus of Reproducing Kernels}

Descriptions of quadratic decision boundaries involve first and second degree
point coordinates and first and second degree vector components.

Therefore, the dual locus of data points \emph{must} be a \emph{locus of
reproducing kernels}, where either second-order, polynomial reproducing
kernels $\left(  \mathbf{x}^{T}\mathbf{s}+1\right)  ^{2}$ or Gaussian
reproducing kernels $k_{\mathbf{s}}=\exp\left(  -\gamma\left\Vert
\mathbf{x}-\mathbf{s}\right\Vert ^{2}\right)  $, where $\gamma=0.01$, are
deemed sufficient.

\paragraph{Minimum Conditional Risk Constraint}

Given Eqs (\ref{Equalizer Rule}) and
(\ref{Balancing of Bayes' Risks and Counteracting Risks}), it follows that the
parameter vector must satisfy a discrete version of the fundamental integral
equation of binary classification in Eq. (\ref{Equalizer Rule}) and the
corresponding integral equation in
(\ref{Balancing of Bayes' Risks and Counteracting Risks})\emph{,} where
extreme data points that lie in decision regions involve forces associated
with risks or counter risks, such that the parameter vector satisfies the
quadratic decision boundary in terms of minimum risk.

Thus, the dual locus of extreme data points must jointly satisfy a discrete
version of the fundamental integral equation of binary classification in Eq.
\emph{(\ref{Equalizer Rule})} in terms of forces associated with risks and
counter risks which are related to positions and potential locations of
extreme data points and corresponding total allowed eigenenergies of principal
eigenaxis components. Moreover, the forces associated with risks and counter
risks related to positions and potential locations of extreme data points and
the corresponding total allowed eigenenergies of principal eigenaxis
components must jointly satisfy a discrete version of Eq.
(\ref{Balancing of Bayes' Risks and Counteracting Risks}) so that $\left(
1\right)  $ all of the forces associated with the risks and counter risks that
are related to positions and potential locations of extreme data points are
effectively balanced with each other, and $\left(  2\right)  $ the total
allowed eigenenergies of the principal eigenaxis components are effectively
balanced with each other.

I\ will call a dual locus of principal eigenaxis components formed by weighted
reproducing kernels of extreme points that determines an estimate of
class-conditional densities for extreme points a "quadratic eigenlocus."
I\ will refer to the parameter vector that provides an estimate of
class-conditional densities for extreme points as a "locus of likelihoods" or
a "parameter vector or likelihoods."

A quadratic eigenlocus, which is formed by a dual locus of principal eigenaxis
components and likelihoods, is a data-driven likelihood ratio and decision
boundary that determines computer-implemented, quadratic classification
systems that minimize the expected risk. In this paper, the dual locus is
based on second-order, polynomial reproducing kernels of extreme points, where
each reproducing kernel replaces a directed, straight line segment of an
extreme vector with a second-order, polynomial curve. All of the locus
equations are readily generalized for Gaussian reproducing kernels. I\ will
refer to second-order, polynomial reproducing kernels as reproducing kernels,
where any given reproducing kernel is an extreme vector. I will call the
system of data-driven, mathematical laws that generates a quadratic eigenlocus
a "quadratic eigenlocus transform." I\ will introduce the primal equation of a
quadratic eigenlocus in the next section. I\ will begin the next section by
defining important geometric and statistical properties exhibited by weighted
reproducing kernels of extreme points on a quadratic eigenlocus. I\ will
define these properties in terms of geometric and statistical criterion.

\subsection{Quadratic Eigenlocus Transforms}

A\ high level description of quadratic eigenlocus transforms is outlined
below. The high level description specifies essential geometric and
statistical properties exhibited by weighted reproducing kernels of extreme
points on a quadratic eigenlocus.

\textbf{Quadratic eigenlocus transforms generate a locus of weighted
reproducing kernels of extreme points that is a dual locus of likelihoods and
principal eigenaxis components, where each weight specifies a class membership
statistic and conditional density for an extreme point, and each weight
determines the magnitude and the total allowed eigenenergy of an extreme
vector.}

\begin{flushleft}
\textbf{Quadratic eigenlocus transforms choose each weight in a manner which
ensures that:}
\end{flushleft}

\paragraph{Criterion $\mathbf{1}$}

Each conditional density of an extreme point describes the central location
(expected value) and the spread (covariance) of the extreme point.

\paragraph{Criterion $\mathbf{2}$}

Distributions of the extreme points are distributed over the locus of
likelihoods in a symmetrically balanced and well-proportioned manner.

\paragraph{Criterion $\mathbf{3}$}

The total allowed eigenenergy possessed by each weighted extreme vector
specifies the probability of observing the extreme point within a localized region.

\paragraph{Criterion $\mathbf{4}$}

The total allowed eigenenergies of the weighted extreme vectors are
symmetrically balanced with each other about the center of total allowed eigenenergy.

\paragraph{Criterion $\mathbf{5}$}

The forces associated with risks and counter risks related to the weighted
extreme points are symmetrically balanced with each other about a center of
minimum risk.

\paragraph{Criterion $\mathbf{6}$}

The locus of principal eigenaxis components formed by weighted extreme vectors
partitions any given feature space into symmetrical decision regions which are
symmetrically partitioned by a quadratic decision boundary.

\paragraph{Criterion $\mathbf{7}$}

The locus of principal eigenaxis components is the focus of a quadratic
decision boundary.

\paragraph{Criterion $\mathbf{8}$}

The locus of principal eigenaxis components formed by weighted extreme vectors
satisfies the quadratic decision boundary in terms of a critical minimum eigenenergy.

\paragraph{Criterion $\mathbf{9}$}

The locus of likelihoods formed by weighted reproducing kernels of extreme
points satisfies the quadratic decision boundary in terms of a minimum
probability of decision error.

\paragraph{Criterion $\mathbf{10}$}

For data distributions that have dissimilar covariance matrices, the forces
associated with counter risks and risks, within each of the symmetrical
decision regions, are balanced with each other. For data distributions that
have similar covariance matrices, the forces associated with counter risks
within each of the symmetrical decision regions are equal to each other, and
the forces associated with risks within each of the symmetrical decision
regions are equal to each other.

\paragraph{Criterion $\mathbf{11}$}

For data distributions that have dissimilar covariance matrices, the
eigenenergies associated with counter risks and the eigenenergies associated
with risks, within each of the symmetrical decision regions, are balanced with
other. For data distributions that have similar covariance matrices, the
eigenenergies associated with counter risks within each of the symmetrical
decision regions are equal to each other, and the eigenenergies associated
with risks within each of the symmetrical decision regions are equal to each other.

I\ will devise a system of data-driven, locus equations that determines
likelihood ratios and decision boundaries which satisfy all of the above
criteria. Accordingly, quadratic eigenlocus transforms generate a dual locus
of principal eigenaxis components and likelihoods that exhibits the
statistical property of symmetrical balance which is illustrated in Fig.
($\ref{Dual Statistical Balancing Feat}$). Quadratic eigenlocus transforms are
generated by solving the inequality constrained optimization problem that is
introduced next.

\subsection{Primal Problem of a Quadratic Eigenlocus}

Take any given collection of training data for a binary classification problem
of the form:%
\[
\left(  \mathbf{x}_{1},y_{1}\right)  ,\ldots,\left(  \mathbf{x}_{N}%
,y_{N}\right)  \in%
\mathbb{R}
^{d}\times Y,Y=\left\{  \pm1\right\}  \text{,}%
\]
where feature vectors $\mathbf{x}$ from class $\omega_{1}$ and class
$\omega_{2}$ are drawn from unknown, class-conditional probability density
functions $p\left(  \mathbf{x}|\omega_{1}\right)  $ and $p\left(
\mathbf{x}|\omega_{2}\right)  $ and are identically distributed.

A quadratic eigenlocus $\boldsymbol{\kappa}$ is estimated by solving an
inequality constrained optimization problem:%
\begin{align}
\min\Psi\left(  \boldsymbol{\kappa}\right)   &  =\left\Vert \boldsymbol{\kappa
}\right\Vert ^{2}/2+C/2\sum\nolimits_{i=1}^{N}\xi_{i}^{2}\text{,}%
\label{Primal Normal Eigenlocus Q}\\
\text{s.t. }y_{i}\left(  \left(  \mathbf{x}^{T}\mathbf{x}_{i}+1\right)
^{2}\boldsymbol{\kappa}+\kappa_{0}\right)   &  \geq1-\xi_{i},\ \xi_{i}%
\geq0,\ i=1,...,N\text{,}\nonumber
\end{align}
where $\boldsymbol{\kappa}$ is a $d\times1$ constrained, primal quadratic
eigenlocus which is a dual locus of likelihoods and principal eigenaxis
components, $\left(  \mathbf{x}^{T}\mathbf{x}_{i}+1\right)  ^{2}$ is a
reproducing kernel for the point $\mathbf{x}_{i}$, $\left\Vert
\boldsymbol{\kappa}\right\Vert ^{2}$ is the total allowed eigenenergy
exhibited by $\boldsymbol{\kappa}$, $\kappa_{0}$ is a functional of
$\boldsymbol{\kappa}$, $C$ and $\xi_{i}$ are regularization parameters, and
$y_{i}$ are class membership statistics: if $\mathbf{x}_{i}\in\omega_{1}$,
assign $y_{i}=1$; if $\mathbf{x}_{i}\in\omega_{2}$, assign $y_{i}=-1$.

Equation (\ref{Primal Normal Eigenlocus Q}) is the primal problem of a
quadratic eigenlocus, where the system of $N$ inequalities must be satisfied:%
\[
y_{i}\left(  \left(  \mathbf{x}^{T}\mathbf{x}_{i}+1\right)  ^{2}%
\boldsymbol{\kappa}+\kappa_{0}\right)  \geq1-\xi_{i},\ \xi_{i}\geq
0,\ i=1,...,N\text{,}%
\]
such that a constrained, primal quadratic eigenlocus $\boldsymbol{\kappa}$
satisfies a critical minimum eigenenergy constraint:%
\begin{equation}
\gamma\left(  \boldsymbol{\kappa}\right)  =\left\Vert \boldsymbol{\kappa
}\right\Vert _{\min_{c}}^{2}\text{,}
\label{Minimum Total Eigenenergy Primal Normal Eigenlocus Q}%
\end{equation}
where $\left\Vert \boldsymbol{\kappa}\right\Vert _{\min_{c}}^{2}$ determines
the minimum risk $\mathfrak{R}_{\mathfrak{\min}}\left(  Z|\boldsymbol{\kappa
}\right)  $ of a quadratic classification system.

Solving the inequality constrained optimization problem in Eq.
(\ref{Primal Normal Eigenlocus Q}) involves solving a dual optimization
problem that determines the fundamental unknowns of Eq.
(\ref{Primal Normal Eigenlocus Q}). Denote a Wolfe dual quadratic eigenlocus
by $\boldsymbol{\psi}$ and the Lagrangian dual problem of $\boldsymbol{\psi}$
by $\max\Xi\left(  \boldsymbol{\psi}\right)  $. Let $\boldsymbol{\psi}$ be a
Wolfe dual of $\boldsymbol{\kappa}$ such that proper and effective strong
duality relationships exist between the algebraic systems of $\min\Psi\left(
\boldsymbol{\kappa}\right)  $ and $\max\Xi\left(  \boldsymbol{\psi}\right)  $.
Thereby, let $\boldsymbol{\psi}$ be related with $\boldsymbol{\kappa}$ in a
symmetrical manner that specifies the locations of the principal eigenaxis
components on $\boldsymbol{\kappa}$.

\subsubsection{The Real Unknowns}

A constrained, primal quadratic eigenlocus is a dual locus of principal
eigenaxis components and likelihoods formed by weighted reproducing kernels of
extreme points, where each weight is specified by a class membership statistic
and a scale factor. Each scale factor specifies a conditional density for a
weighted extreme point on a locus of likelihoods, and each scale factor
determines the magnitude and the eigenenergy of a weighted extreme vector on a
locus of principal eigenaxis components. The main issue concerns how the scale
factors are determined.

\subsubsection{The Fundamental Unknowns}

The fundamental unknowns are the scale factors of the principal eigenaxis
components on a Wolfe dual quadratic eigenlocus $\boldsymbol{\psi}$.

\subsection{Strong Dual Quadratic Eigenlocus Transforms}

For the problem of quadratic eigenlocus transforms, the Lagrange multipliers
method introduces a Wolfe dual quadratic eigenlocus $\boldsymbol{\psi}$ of
principal eigenaxis components, for which the Lagrange multipliers $\left\{
\psi_{i}\right\}  _{i=1}^{N}$ are the magnitudes or lengths of a set of Wolfe
dual principal eigenaxis components $\left\{  \psi_{i}%
\overrightarrow{\mathbf{e}}_{i}\right\}  _{i=1}^{N}$, where $\left\{
\overrightarrow{\mathbf{e}}_{i}\right\}  _{i=1}^{N}$ are non-orthogonal unit
vectors and finds extrema for the restriction of a primal quadratic eigenlocus
$\boldsymbol{\kappa}$ to a Wolfe dual eigenspace. Accordingly, the fundamental
unknowns associated with Eq. (\ref{Primal Normal Eigenlocus Q}) are the
magnitudes or lengths of the Wolfe dual principal eigenaxis components on
$\boldsymbol{\psi}$.

\subsubsection{Strong Duality}

Because Eq. (\ref{Primal Normal Eigenlocus Q}) is a convex programming
problem, the theorem for convex duality guarantees an equivalence and
corresponding symmetry between a constrained, primal quadratic eigenlocus
$\boldsymbol{\kappa}$ and its Wolfe dual $\boldsymbol{\psi}$
\citep{Nash1996,Luenberger2003}%
. Strong duality holds between the systems of locus equations denoted by
$\min\Psi\left(  \boldsymbol{\tau}\right)  $ and $\max\Xi\left(
\boldsymbol{\psi}\right)  $, so that the duality gap between the constrained
primal and the Wolfe dual quadratic eigenlocus solution is zero
\citep{Luenberger1969,Nash1996,Fletcher2000,Luenberger2003}%
.

The Lagrangian dual problem of a Wolfe dual quadratic eigenlocus will be
derived by means of the Lagrangian equation that is introduced next.

\subsection{The Lagrangian of the Quadratic Eigenlocus}

The inequality constrained optimization problem in Eq.
(\ref{Primal Normal Eigenlocus Q}) is solved by using Lagrange multipliers
$\psi_{i}\geq0$ and the Lagrangian:%
\begin{align}
L_{\Psi\left(  \boldsymbol{\kappa}\right)  }\left(  \boldsymbol{\kappa
}\mathbf{,}\kappa_{0},\mathbf{\xi},\boldsymbol{\psi}\right)   &  =\left\Vert
\boldsymbol{\kappa}\right\Vert ^{2}/2\label{Lagrangian Normal Eigenlocus Q}\\
&  +C/2\sum\nolimits_{i=1}^{N}\xi_{i}^{2}\nonumber\\
&  -\sum\nolimits_{i=1}^{N}\psi_{i}\nonumber\\
&  \times\left\{  y_{i}\left(  \left(  \mathbf{x}^{T}\mathbf{x}_{i}+1\right)
^{2}\boldsymbol{\kappa}+\kappa_{0}\right)  -1+\xi_{i}\right\} \nonumber
\end{align}
which is minimized with respect to the primal variables $\boldsymbol{\kappa}%
$\textbf{ }and $\kappa_{0}$ and is maximized with respect to the dual
variables $\psi_{i}$.

The Karush-Kuhn-Tucker (KKT) conditions on the Lagrangian $L_{\Psi\left(
\boldsymbol{\kappa}\right)  }$:%
\begin{equation}
\boldsymbol{\kappa}-\sum\nolimits_{i=1}^{N}\psi_{i}y_{i}\left(  \mathbf{x}%
^{T}\mathbf{x}_{i}+1\right)  ^{2}=0,\text{ \ }i=1,...N\text{,} \label{KKTE1 Q}%
\end{equation}%
\begin{equation}
\sum\nolimits_{i=1}^{N}\psi_{i}y_{i}=0,\text{ \ }i=1,...,N\text{,}
\label{KKTE2 Q}%
\end{equation}%
\begin{equation}
C\sum\nolimits_{i=1}^{N}\xi_{i}-\sum\nolimits_{i=1}^{N}\psi_{i}=0\text{,}
\label{KKTE3 Q}%
\end{equation}%
\begin{equation}
\psi_{i}\geq0,\text{ \ }i=1,...,N\text{,} \label{KKTE4 Q}%
\end{equation}%
\begin{equation}
\psi_{i}\left[  y_{i}\left(  \left(  \mathbf{x}^{T}\mathbf{x}_{i}+1\right)
^{2}\boldsymbol{\kappa}+\kappa_{0}\right)  -1+\xi_{i}\right]  \geq
0,\ i=1,...,N\text{,} \label{KKTE5 Q}%
\end{equation}
which can found in
\citep{Cortes1995,Burges1998,Cristianini2000,Scholkopf2002}%
, determine a system of data-driven, locus equations which are jointly
satisfied by a constrained primal and a Wolfe dual quadratic eigenlocus. I
will define the manner in which the KKT conditions determine geometric and
statistical properties exhibited by weighted reproducing kernels of extreme
points on a Wolfe dual $\boldsymbol{\psi}$ and a constrained primal
$\boldsymbol{\kappa}$ quadratic eigenlocus. Thereby, I\ will demonstrate the
manner in which the KKT conditions ensure that $\boldsymbol{\psi}$ and
$\boldsymbol{\kappa}$ jointly satisfy discrete and data-driven versions of the
fundamental equations of binary classification for a classification system in
statistical equilibrium.

The Lagrangian dual problem of a Wolfe dual quadratic eigenlocus is introduced next.

\subsection{Lagrangian Dual Problem of a Quadratic Eigenlocus}

The resulting expressions for a primal quadratic eigenlocus
$\boldsymbol{\kappa}$ in Eq. (\ref{KKTE1 Q}) and a Wolfe dual quadratic
eigenlocus $\boldsymbol{\psi}$ in Eq. (\ref{KKTE2 Q}) are substituted into the
Lagrangian functional $L_{\Psi\left(  \boldsymbol{\tau}\right)  }$ of Eq.
(\ref{Lagrangian Normal Eigenlocus Q}) and simplified. This produces the
Lagrangian dual problem of a Wolfe dual quadratic eigenlocus: a quadratic
programming problem%
\begin{equation}
\max\Xi\left(  \boldsymbol{\psi}\right)  =\sum\nolimits_{i=1}^{N}\psi_{i}%
-\sum\nolimits_{i,j=1}^{N}\psi_{i}\psi_{j}y_{i}y_{j}\frac{\left[  \left(
\mathbf{x}_{i}^{T}\mathbf{x}_{j}+1\right)  ^{2}+\delta_{ij}/C\right]  }{2}
\label{Wolfe Dual Normal Eigenlocus Q}%
\end{equation}
which is subject to the algebraic constraints $\sum\nolimits_{i=1}^{N}%
y_{i}\psi_{i}=0$ and $\psi_{i}\geq0$, where $\delta_{ij}$ is the Kronecker
$\delta$ defined as unity for $i=j$ and $0$ otherwise.

Equation (\ref{Wolfe Dual Normal Eigenlocus Q}) can be written in vector
notation by letting $\mathbf{Q}\triangleq\epsilon\mathbf{I}%
+\widetilde{\mathbf{X}}\widetilde{\mathbf{X}}^{T}$ and $\widetilde{\mathbf{X}%
}\triangleq\mathbf{D}_{y}\mathbf{X}$, where $\mathbf{D}_{y}$ is an $N\times N$
diagonal matrix of training labels $y_{i}$ and the $N\times d$ data matrix is%
\[
\mathbf{X}=%
\begin{pmatrix}
\left(  \mathbf{x}^{T}\mathbf{x}_{1}+1\right)  ^{2}, & \left(  \mathbf{x}%
^{T}\mathbf{x}_{2}+1\right)  ^{2}, & \ldots, & \left(  \mathbf{x}%
^{T}\mathbf{x}_{N}+1\right)  ^{2}%
\end{pmatrix}
^{T}\text{.}%
\]
This produces the matrix version of the Lagrangian dual problem of a primal
quadratic eigenlocus within its Wolfe dual eigenspace:%
\begin{equation}
\max\Xi\left(  \boldsymbol{\psi}\right)  =\mathbf{1}^{T}\boldsymbol{\psi
}-\frac{\boldsymbol{\psi}^{T}\mathbf{Q}\boldsymbol{\psi}}{2}
\label{Vector Form Wolfe Dual Q}%
\end{equation}
which is subject to the constraints $\boldsymbol{\psi}^{T}\mathbf{y}=0$ and
$\psi_{i}\geq0$
\citep{Reeves2009}%
. Given the theorem for convex duality, it follows that a Wolfe dual quadratic
eigenlocus $\boldsymbol{\psi}$ is a dual locus of likelihoods and principal
eigenaxis components $\widehat{\Lambda}_{\boldsymbol{\psi}}\left(
\mathbf{x}\right)  $, where $\boldsymbol{\psi}$ exhibits a total allowed
eigenenergy $\left\Vert \boldsymbol{\psi}\right\Vert _{\min_{c}}^{2}$ that is
symmetrically related to the total allowed eigenenergy $\left\Vert
\boldsymbol{\kappa}\right\Vert _{\min_{c}}^{2}$ of $\boldsymbol{\kappa}$:
$\left\Vert \boldsymbol{\psi}\right\Vert _{\min_{c}}^{2}\simeq\left\Vert
\boldsymbol{\kappa}\right\Vert _{\min_{c}}^{2}$.

\subsection{Loci of Constrained Quadratic Forms}

The representation of a constrained, primal quadratic eigenlocus
$\boldsymbol{\kappa}$ within its Wolfe dual eigenspace involves the
eigensystem of the constrained quadratic form $\boldsymbol{\psi}^{T}%
\mathbf{Q}\boldsymbol{\psi}$ in Eq. (\ref{Vector Form Wolfe Dual Q}), where
$\boldsymbol{\psi}$ is the principal eigenvector of $\mathbf{Q}$, such that
$\boldsymbol{\psi}^{T}\mathbf{y}=0$ and $\psi_{i}\geq0$. I\ will demonstrate
that Eq. (\ref{Vector Form Wolfe Dual Q}) determines a dual quadratic
eigenlocus $\boldsymbol{\psi}$ which is in statistical equilibrium such that
the total allowed eigenenergies $\left\Vert \boldsymbol{\kappa}\right\Vert
_{\min_{c}}^{2}$ exhibited by $\boldsymbol{\kappa}$ are symmetrically balanced
with each other about a center of total allowed eigenenergy. I will also
demonstrate that the utility of the statistical balancing feat involves
\emph{balancing} all of the forces associated with the counter risk
$\overline{\mathfrak{R}}_{\mathfrak{\min}}\left(  Z_{1}|\omega_{1}\right)  $
and the risk $\mathfrak{R}_{\mathfrak{\min}}\left(  Z_{1}|\omega_{2}\right)  $
in the $Z_{1}$ decision region \emph{with} all of the forces associated with
the counter risk $\overline{\mathfrak{R}}_{\mathfrak{\min}}\left(
Z_{2}|\omega_{2}\right)  $ and the risk $\mathfrak{R}_{\mathfrak{\min}}\left(
Z_{2}|\omega_{1}\right)  $ in the $Z_{2}$ decision region, where the forces
associated with risks and counter risks are related to positions and potential
locations of extreme points, such that the eigenenergy $\left\Vert
\boldsymbol{\kappa}\right\Vert _{\min_{c}}^{2}$ and the risk $\mathfrak{R}%
_{\mathfrak{\min}}\left(  Z|\boldsymbol{\kappa}\right)  $ of a discrete,
quadratic classification system are both minimized.

I\ will now use the KKT conditions in Eqs (\ref{KKTE1 Q}) and (\ref{KKTE4 Q})
to derive the locus equation of a constrained, primal quadratic eigenlocus
$\boldsymbol{\kappa}$.

\subsection{The Constrained Primal Quadratic Eigenlocus}

Using the KKT\ conditions in Eqs (\ref{KKTE1 Q}) and (\ref{KKTE4 Q}), it
follows that an estimate for $\boldsymbol{\kappa}$ satisfies the following
locus equation:%
\begin{equation}
\boldsymbol{\kappa}=\sum\nolimits_{i=1}^{N}y_{i}\psi_{i}\left(  \mathbf{x}%
^{T}\mathbf{x}_{i}+1\right)  ^{2}\text{,} \label{Normal Eigenlocus Estimate Q}%
\end{equation}
where the $y_{i}$ terms are class membership statistics (if $\mathbf{x}_{i}$
is a member of class $\omega_{1}$, assign $y_{i}=1$; otherwise, assign
$y_{i}=-1$) and the magnitude $\psi_{i}$ of each principal eigenaxis component
$\psi_{i}\overrightarrow{\mathbf{e}}_{i}$ on $\boldsymbol{\psi}$ is greater
than or equal to zero: $\psi_{i}\geq0$.

The KKT condition in Eq. (\ref{KKTE4 Q}) requires that the length $\psi_{i}$
of each principal eigenaxis component $\psi_{i}\overrightarrow{\mathbf{e}}%
_{i}$ on $\boldsymbol{\psi}$ either satisfy or exceed zero: $\psi_{i}\geq0$.
Any principal eigenaxis component $\psi_{i}\overrightarrow{\mathbf{e}}_{i}$
which has zero length ($\psi_{i}=0$) satisfies the origin $P_{\mathbf{0}}%
\begin{pmatrix}
0, & 0, & \cdots, & 0
\end{pmatrix}
$ and is not on the Wolfe dual quadratic eigenlocus $\boldsymbol{\psi}$. It
follows that the constrained, primal principal eigenaxis component $\psi
_{i}\left(  \mathbf{x}^{T}\mathbf{x}_{i}+1\right)  ^{2}$ also has zero length
($\psi_{i}\left(  \mathbf{x}_{i}^{2}+1\right)  =0$) and is not on the
constrained, primal quadratic eigenlocus $\boldsymbol{\kappa}$.

Reproducing kernels $\left(  \mathbf{x}^{T}\mathbf{x}_{i}+1\right)  ^{2}$ of
data points $\mathbf{x}_{i}$ correlated with Wolfe dual principal eigenaxis
components $\psi_{i}\overrightarrow{\mathbf{e}}_{i}$ that have non-zero
magnitudes $\psi_{i}>0$ are termed extreme vectors. Accordingly, extreme
vectors are unscaled, primal principal eigenaxis components on
$\boldsymbol{\kappa}$. Recall that a set of extreme vectors specify principal
directions of large covariance for a given collection of training data. Thus,
extreme vectors are discrete principal components that determine directions
for which a given collection of training data is most variable or spread out.
Therefore, the loci of a set of extreme vectors span a region of large
covariance between two distributions of training data. See Fig.
$\ref{Location Properties Extreme Data Points}$.

\subsection{Primal Quadratic Eigenlocus Components}

All of the principal eigenaxis components on a constrained, primal quadratic
eigenlocus $\boldsymbol{\kappa}$ are labeled, scaled reproducing kernels of
extreme points in $%
\mathbb{R}
^{d}$. Denote the labeled, scaled extreme vectors that belong to class
$\omega_{1}$ and $\omega_{2}$ by $\psi_{1_{i\ast}}\left(  \mathbf{x}%
^{T}\mathbf{x}_{1_{i\ast}}+1\right)  ^{2}$ and $-\psi_{2_{i\ast}}\left(
\mathbf{x}^{T}\mathbf{x}_{2_{i\ast}}+1\right)  ^{2}$, with scale factors
$\psi_{1_{i\ast}}$ and $\psi_{2_{i\ast}}$, extreme vectors $\left(
\mathbf{x}^{T}\mathbf{x}_{1_{i\ast}}+1\right)  ^{2}$ and $\left(
\mathbf{x}^{T}\mathbf{x}_{2_{i\ast}}+1\right)  ^{2}$, and labels $y_{i}=1$ and
$y_{i}=-1$ respectively. Let there be $l_{1}$ labeled, scaled reproducing
kernels $\left\{  \psi_{1_{i\ast}}\left(  \mathbf{x}^{T}\mathbf{x}_{1_{i\ast}%
}+1\right)  ^{2}\right\}  _{i=1}^{l_{1}}$ and $l_{2}$ labeled, scaled
reproducing kernels $\left\{  -\left(  \mathbf{x}^{T}\mathbf{x}_{2_{i\ast}%
}+1\right)  ^{2}\right\}  _{i=1}^{l_{2}}$.

Given Eq. (\ref{Normal Eigenlocus Estimate Q}) and the assumptions outlined
above, it follows that an estimate for a constrained, primal quadratic
eigenlocus $\boldsymbol{\kappa}$ is based on the vector difference between a
pair of constrained, primal quadratic eigenlocus components:%
\begin{align}
\boldsymbol{\kappa}  &  =\sum\nolimits_{i=1}^{l_{1}}\psi_{1_{i\ast}}\left(
\mathbf{x}^{T}\mathbf{x}_{1_{i\ast}}+1\right)  ^{2}-\sum\nolimits_{i=1}%
^{l_{2}}\psi_{2_{i\ast}}\left(  \mathbf{x}^{T}\mathbf{x}_{2_{i\ast}}+1\right)
^{2}\label{Pair of Normal Eigenlocus Components Q}\\
&  =\boldsymbol{\kappa}_{1}-\boldsymbol{\kappa}_{2}\text{,}\nonumber
\end{align}
where the constrained, primal quadratic eigenlocus components%
\[
\boldsymbol{\kappa}_{1}=\sum\nolimits_{i=1}^{l_{1}}\psi_{1_{i\ast}}\left(
\mathbf{x}^{T}\mathbf{x}_{1_{i\ast}}+1\right)  ^{2}\text{ and }%
\boldsymbol{\kappa}_{2}=\sum\nolimits_{i=1}^{l_{2}}\psi_{2_{i\ast}}\left(
\mathbf{x}^{T}\mathbf{x}_{2_{i\ast}}+1\right)  ^{2}%
\]
are denoted by $\boldsymbol{\kappa}_{1}$ and $\boldsymbol{\kappa}_{2}$
respectively. The scaled reproducing kernels on $\boldsymbol{\kappa}_{1}$ and
$\boldsymbol{\kappa}_{2}$ determine the loci of $\boldsymbol{\kappa}_{1}$ and
$\boldsymbol{\kappa}_{2}$ and therefore determine the dual locus of
$\boldsymbol{\kappa}=\boldsymbol{\kappa}_{1}-\boldsymbol{\kappa}_{2}$.

\begin{figure}[ptb]%
\centering
\fbox{\includegraphics[
height=2.5875in,
width=3.4411in
]%
{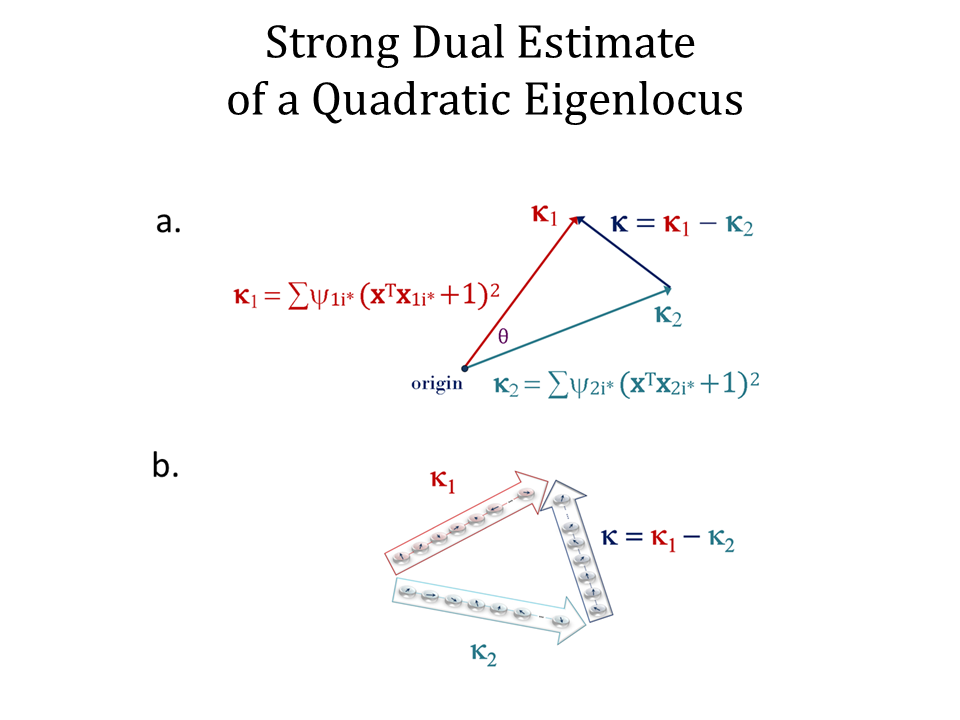}%
}\caption{$\left(  a\right)  $ A constrained, primal quadratic eigenlocus
$\boldsymbol{\kappa}$ is determined by the vector difference
$\boldsymbol{\kappa}_{1}-\boldsymbol{\kappa}_{2}$ between a pair of
constrained, primal quadratic eigenlocus components $\boldsymbol{\kappa}_{1}$
and $\boldsymbol{\kappa}_{2}$. $\left(  b\right)  $ The scaled extreme points
on $\boldsymbol{\kappa}_{1}$ and $\boldsymbol{\kappa}_{2}$ are endpoints of
scaled extreme vectors that possess unchanged directions and eigen-balanced
lengths.}%
\label{Primal Quadratic Eigenlocus in Wolfe Dual Eigenspace}%
\end{figure}

Figure\textbf{\ }$\ref{Primal Quadratic Eigenlocus in Wolfe Dual Eigenspace}$
depicts how the configurations of $\boldsymbol{\kappa}_{1}$ and
$\boldsymbol{\kappa}_{2}$ determine the configuration of $\boldsymbol{\kappa}%
$.%

I\ will now define values for the regularization parameters $C$ and $\xi_{i}$
in Eqs (\ref{Primal Normal Eigenlocus Q}) and (\ref{Vector Form Wolfe Dual Q}).

\subsection{Weak Dual Quadratic Eigenlocus Transforms}

The number and the locations of the principal eigenaxis components on
$\boldsymbol{\psi}$ and $\boldsymbol{\kappa}$ are considerably affected by the
rank and eigenspectrum of $\mathbf{Q}$. Low rank kernel matrices $\mathbf{Q}$
generate "weak\emph{\ }dual" quadratic eigenlocus transforms that produce
irregular, quadratic partitions of decision spaces.

Given non-overlapping data distributions and low rank kernel matrices
$\mathbf{Q}$, weak dual quadratic eigenlocus transforms produce asymmetric,
quadratic partitions that exhibit optimal generalization performance at the
expense of unnecessary principal eigenaxis components, where \emph{all} of the
training data is transformed into constrained, primal principal eigenaxis
components. For overlapping data distributions, incomplete eigenspectra of low
rank kernel matrices $\mathbf{Q}$ result in weak dual quadratic eigenlocus
transforms which determine ill-formed, quadratic decision boundaries that
exhibit substandard generalization performance. All of these problems are
solved by the regularization method that is described next.

\subsubsection{Regularization of Quadratic Eigenlocus Transforms}

For any collection of $N$ training vectors of dimension $d$, where $d<N$, the
kernel matrix $\mathbf{Q}$ has low rank. The results for low rank Gram
matrices in
\citet{Reeves2011}
are readily extended to kernel matrices. Accordingly, the regularized form of
$\mathbf{Q}$, for which $\epsilon\ll1$ and $\mathbf{Q}\triangleq
\epsilon\mathbf{I}+\widetilde{\mathbf{X}}\widetilde{\mathbf{X}}^{T}$, ensures
that $\mathbf{Q}$ has full rank and a complete eigenvector set so that
$\mathbf{Q}$ has a complete eigenspectrum. The regularization constant $C$ is
related to the regularization parameter $\epsilon$\ by $\frac{1}{C}$.

For $N$ training vectors of dimension $d$, where $d<N$, all of the
regularization parameters $\left\{  \xi_{i}\right\}  _{i=1}^{N}$ in Eq.
(\ref{Primal Normal Eigenlocus Q}) and all of its derivatives are set equal to
a very small value: $\xi_{i}=\xi\ll1$. The regularization constant $C$ is set
equal to $\frac{1}{\xi}$: $C=\frac{1}{\xi}$.

For $N$ training vectors of dimension $d$, where $N<d$, all of the
regularization parameters $\left\{  \xi_{i}\right\}  _{i=1}^{N}$ in Eq.
(\ref{Primal Normal Eigenlocus Q}) and all of its derivatives are set equal to
zero: $\xi_{i}=\xi=0$. The regularization constant $C$ is set equal to
infinity: $C=\infty$.

In the next section, I will devise locus equations that determine the manner
in which a constrained, primal quadratic eigenlocus partitions any given
feature space into symmetrical decision regions.

\section{Equations of a Quadratic Discriminant Function}

A constrained, primal quadratic eigenlocus is the primary basis of a quadratic
discriminant function that implements an optimal likelihood ratio test. The
manner in which the dual locus of $\boldsymbol{\kappa}$ partitions a feature
space is specified by the KKT condition in Eq. (\ref{KKTE5 Q}) and the KKT
condition of complementary slackness.

\subsection{KKT Condition of Complementary Slackness}

The KKT condition of complementary slackness requires that for all constraints
that are not active, where locus equations are \emph{ill-defined}:%
\[
y_{i}\left(  \left(  \mathbf{x}^{T}\mathbf{x}_{i}+1\right)  ^{2}%
\boldsymbol{\kappa}+\kappa_{0}\right)  -1+\xi_{i}>0
\]
because they are not satisfied as equalities, the corresponding magnitudes
$\psi_{i}$ of the Wolfe dual principal eigenaxis components $\psi
_{i}\overrightarrow{\mathbf{e}}_{i}$ must be zero: $\psi_{i}=0$
\citep{Sundaram1996}%
. Accordingly, if an inequality is "slack" (not strict), the other inequality
cannot be slack.

Therefore, let there be $l$ active constraints, where $l=l_{1}+l_{2}$. Let
$\xi_{i}=\xi=0$ or $\xi_{i}=\xi\ll1$. The theorem of Karush, Kuhn, and Tucker
provides the guarantee that a Wolf dual quadratic eigenlocus $\boldsymbol{\psi
}$ exists such that the following constraints are satisfied:%
\[
\left\{  \psi_{i\ast}>0\right\}  _{i=1}^{l}\text{,}%
\]
and the following locus equations are satisfied:%
\[
\psi_{i\ast}\left[  y_{i}\left(  \left(  \mathbf{x}^{T}\mathbf{x}_{i\ast
}+1\right)  ^{2}\boldsymbol{\kappa}+\kappa_{0}\right)  -1+\xi_{i}\right]
=0,\ i=1,...,l\text{,}%
\]
where $l$ Wolfe dual principal eigenaxis components $\psi_{i\ast
}\overrightarrow{\mathbf{e}}_{i}$ have non-zero magnitudes $\left\{
\psi_{i\ast}\overrightarrow{\mathbf{e}}_{i}|\psi_{i\ast}>0\right\}  _{i=1}^{l}
$
\citep{Sundaram1996}%
.

The above condition is known as the condition of complementary slackness. So,
in order for the constraint $\psi_{i\ast}>0$ to hold, the following locus
equation must be satisfied:%
\[
y_{i}\left(  \left(  \mathbf{x}^{T}\mathbf{x}_{i\ast}+1\right)  ^{2}%
\boldsymbol{\kappa}+\kappa_{0}\right)  -1+\xi_{i}=0\text{.}%
\]

Accordingly, let there be $l_{1}$ locus equations:%
\[
\left(  \mathbf{x}^{T}\mathbf{x}_{1i\ast}+1\right)  ^{2}\boldsymbol{\kappa
}+\kappa_{0}+\xi_{i}=1,\ i=1,...,l_{1}\text{,}%
\]
where $y_{i}=+1$, and $l_{2}$ locus equations:%
\[
\left(  \mathbf{x}^{T}\mathbf{x}_{2i\ast}+1\right)  ^{2}\boldsymbol{\kappa
}-\kappa_{0}-\xi_{i}=-1,\ i=1,...,l_{2}\text{,}%
\]
where $y_{i}=-1$.

It follows that the quadratic discriminant function%
\begin{equation}
D\left(  \mathbf{s}\right)  =\left(  \mathbf{x}^{T}\mathbf{s}+1\right)
^{2}\boldsymbol{\kappa}+\kappa_{0} \label{Discriminant Function Q}%
\end{equation}
satisfies the set of constraints:%
\[
D_{0}\left(  \mathbf{s}\right)  =0\text{, }D_{+1}\left(  \mathbf{s}\right)
=+1\text{, and }D_{-1}\left(  \mathbf{s}\right)  =-1\text{,}%
\]
where $D_{0}\left(  \mathbf{s}\right)  $ denotes a quadratic decision
boundary, $D_{+1}\left(  \mathbf{x}\right)  $ denotes a quadratic decision
border for the $Z_{1}$ decision region, and $D_{-1}\left(  \mathbf{x}\right)
$ denotes a quadratic decision border for the $Z_{2}$ decision region.

I will now show that the constraints on the quadratic discriminant function
$D\left(  \mathbf{s}\right)  $ determine three equations of symmetrical,
quadratic partitioning curves or surfaces, where all of the points on all
three quadratic loci reference the constrained, primal quadratic eigenlocus
$\boldsymbol{\kappa}$.

Returning to Eq. (\ref{Normal Form Second-order Locus}), recall that the
equation of a quadratic locus can be written as%
\[
\frac{2\mathbf{x}^{T}\boldsymbol{\nu}+\left(  e^{2}\cos^{2}\theta-1\right)
\left\Vert \mathbf{x}\right\Vert ^{2}}{\left\Vert \boldsymbol{\nu}\right\Vert
}=\left\Vert \boldsymbol{\nu}\right\Vert \text{,}%
\]
where $e$ is the eccentricity, $\theta$ is the angle between $\mathbf{x}$ and
$\boldsymbol{\nu}$, and the principal eigenaxis $\boldsymbol{\nu}/\left\Vert
\boldsymbol{\nu}\right\Vert $ has length $1$ and points in the direction of a
principal eigenvector $\boldsymbol{\nu}$. Any point $\mathbf{x}$ that
satisfies the above equation is on the quadratic locus of points specified by
$\boldsymbol{\nu}$, where all of the points $\mathbf{x}$ on the quadratic
locus exclusively reference the principal eigenaxis $\boldsymbol{\nu}$.

Returning to Eq. (\ref{Normal Form Circle and Sphere}), recall that the
equation of a spherically symmetric, quadratic locus can be written as%
\[
\frac{2\left(  \mathbf{x-r}\right)  ^{T}\boldsymbol{\nu}}{\left\Vert
\boldsymbol{\nu}\right\Vert }=\left\Vert \boldsymbol{\nu}\right\Vert \text{,}%
\]
where $\mathbf{r}$ is the radius of a spherically symmetric, quadratic locus,
and the principal eigenaxis $\boldsymbol{\nu}/\left\Vert \boldsymbol{\nu
}\right\Vert $ has length $1$ and points in the direction of a principal
eigenvector $\boldsymbol{\nu}$. Any point $\mathbf{x}$ that satisfies the
above equation is on the spherically symmetric, quadratic locus of points
specified by $\boldsymbol{\nu}$, where all of the points $\mathbf{x}$ on the
spherically symmetric, quadratic locus exclusively reference the principal
eigenaxis $\boldsymbol{\nu}$.

I\ will now use Eqs (\ref{Normal Form Second-order Locus}) and
(\ref{Normal Form Circle and Sphere}) along with the constraints on the
quadratic discriminant function in Eq. (\ref{Discriminant Function Q}) to
devise locus equations that determine the manner in which a constrained,
primal quadratic eigenlocus partitions any given feature space into
symmetrical decision regions.

\subsection{Quadratic Partitions of Feature Spaces}

I\ will now derive the locus equation of a quadratic decision boundary
$D_{0}\left(  \mathbf{s}\right)  $.

\subsubsection{Equation of a Quadratic Decision Boundary $D_{0}\left(
\mathbf{s}\right)  $}

Using Eqs (\ref{Normal Form Second-order Locus}) and
(\ref{Normal Form Circle and Sphere}), along with the assumption that
$D\left(  \mathbf{s}\right)  =0$, it follows that the quadratic discriminant
function%
\[
D\left(  \mathbf{s}\right)  =\left(  \mathbf{x}^{T}\mathbf{s}+1\right)
^{2}\boldsymbol{\kappa}+\kappa_{0}%
\]
can be rewritten as:%
\begin{equation}
\frac{\left(  \mathbf{x}^{T}\mathbf{s}+1\right)  ^{2}\boldsymbol{\kappa}%
}{\left\Vert \boldsymbol{\kappa}\right\Vert }=-\frac{\kappa_{0}}{\left\Vert
\boldsymbol{\kappa}\right\Vert }\text{.} \label{Decision Boundary Q}%
\end{equation}

Therefore, any point $\mathbf{s}$ that satisfies Eq.
(\ref{Decision Boundary Q}) is on the quadratic decision boundary
$D_{0}\left(  \mathbf{s}\right)  $, and all of the points $\mathbf{s}$ on the
quadratic decision boundary $D_{0}\left(  \mathbf{s}\right)  $ exclusively
reference the constrained, primal quadratic eigenlocus $\boldsymbol{\kappa}$.
Thereby, the constrained, quadratic discriminant function $\boldsymbol{\kappa
}^{T}k_{\mathbf{s}}+\kappa_{0}$ satisfies the boundary value of a quadratic
decision boundary $D_{0}\left(  \mathbf{s}\right)  $: $\boldsymbol{\kappa}%
^{T}k_{\mathbf{s}}+\kappa_{0}=0$.

I will now derive the locus equation of the $D_{+1}\left(  \mathbf{s}\right)
$ quadratic decision border.

\subsubsection{Equation of the $D_{+1}\left(  \mathbf{s}\right)  $ Decision
Border}

Using Eqs (\ref{Normal Form Second-order Locus}) and
(\ref{Normal Form Circle and Sphere}), along with the assumption that
$D\left(  \mathbf{s}\right)  =1$, it follows that the quadratic discriminant
function in Eq. (\ref{Discriminant Function Q}) can be rewritten as:%
\begin{equation}
\frac{\left(  \mathbf{x}^{T}\mathbf{s}+1\right)  ^{2}\boldsymbol{\kappa}%
}{\left\Vert \boldsymbol{\kappa}\right\Vert }=-\frac{\kappa_{0}}{\left\Vert
\boldsymbol{\kappa}\right\Vert }+\frac{1}{\left\Vert \boldsymbol{\kappa
}\right\Vert }\text{.} \label{Decision Border One Q}%
\end{equation}

Therefore, any point $\mathbf{s}$ that satisfies Eq.
(\ref{Decision Border One Q}) is on the quadratic decision border
$D_{+1}\left(  \mathbf{s}\right)  $, and all of the points $\mathbf{s}$ on the
quadratic decision border $D_{+1}\left(  \mathbf{s}\right)  $ exclusively
reference the constrained, primal quadratic eigenlocus $\boldsymbol{\kappa}$.
Thereby, the constrained, quadratic discriminant function $\boldsymbol{\kappa
}^{T}k_{\mathbf{s}}+\kappa_{0}$ satisfies the boundary value of a quadratic
decision border $D_{+1}\left(  \mathbf{s}\right)  $: $\boldsymbol{\kappa}%
^{T}k_{\mathbf{s}}+\kappa_{0}=1$.

I will now derive the locus equation of the $D_{-1}\left(  \mathbf{s}\right)
$ quadratic decision border.

\subsubsection{Equation of the $D_{-1}\left(  \mathbf{s}\right)  $ Decision
Border}

Using Eqs (\ref{Normal Form Second-order Locus}) and
(\ref{Normal Form Circle and Sphere}), along with the assumption that
$D\left(  \mathbf{s}\right)  =-1$, it follows that the quadratic discriminant
function in Eq. (\ref{Discriminant Function Q}) can be rewritten as:%

\begin{equation}
\frac{\left(  \mathbf{x}^{T}\mathbf{s}+1\right)  ^{2}\boldsymbol{\kappa}%
}{\left\Vert \boldsymbol{\kappa}\right\Vert }=-\frac{\kappa_{0}}{\left\Vert
\boldsymbol{\kappa}\right\Vert }-\frac{1}{\left\Vert \boldsymbol{\kappa
}\right\Vert }\text{.} \label{Decision Border Two Q}%
\end{equation}

Therefore, any point $\mathbf{s}$ that satisfies Eq.
(\ref{Decision Border Two Q}) is on the quadratic decision border
$D_{-1}\left(  \mathbf{s}\right)  $, and all of the points $\mathbf{s}$ on the
quadratic decision border $D_{-1}\left(  \mathbf{s}\right)  $ exclusively
reference the constrained, primal quadratic eigenlocus $\boldsymbol{\kappa}$.
Thereby, the constrained, quadratic discriminant function $\boldsymbol{\kappa
}^{T}k_{\mathbf{s}}+\kappa_{0}$ satisfies the boundary value of a quadratic
decision border $D_{-1}\left(  \mathbf{s}\right)  $: $\boldsymbol{\kappa}%
^{T}k_{\mathbf{s}}+\kappa_{0}=-1$.

Given Eqs (\ref{Decision Boundary Q}), (\ref{Decision Border One Q}), and
(\ref{Decision Border Two Q}), it is concluded that the constrained, quadratic
discriminant function $D\left(  \mathbf{s}\right)  =\boldsymbol{\kappa}%
^{T}k_{\mathbf{s}}+\kappa_{0}$ determines three, symmetrical, quadratic curves
or surfaces, where all of the points on $D_{0}\left(  \mathbf{s}\right)  $,
$D_{+1}\left(  \mathbf{s}\right)  $, and $D_{-1}\left(  \mathbf{s}\right)  $
exclusively reference the constrained, primal quadratic eigenlocus
$\boldsymbol{\kappa}$.

Moreover, it is concluded that the constrained, quadratic discriminant
function $\boldsymbol{\kappa}^{T}k_{\mathbf{s}}+\kappa_{0}$ satisfies boundary
values for a quadratic decision boundary $D_{0}\left(  \mathbf{s}\right)  $
and two quadratic decision borders $D_{+1}\left(  \mathbf{s}\right)  $ and
$D_{-1}\left(  \mathbf{s}\right)  $.

The quadratic decision borders $D_{+1}\left(  \mathbf{s}\right)  $ and
$D_{-1}\left(  \mathbf{s}\right)  $ in Eqs (\ref{Decision Border One Q}) and
(\ref{Decision Border Two Q}) satisfy the symmetrically balanced constraints%
\[
-\frac{\kappa_{0}}{\left\Vert \boldsymbol{\kappa}\right\Vert }+\frac
{1}{\left\Vert \boldsymbol{\kappa}\right\Vert }\text{ and }-\frac{\kappa_{0}%
}{\left\Vert \boldsymbol{\kappa}\right\Vert }-\frac{1}{\left\Vert
\boldsymbol{\kappa}\right\Vert }%
\]
with respect to the constraint satisfied by the quadratic decision boundary
$D_{0}\left(  \mathbf{s}\right)  $%
\[
-\frac{\kappa_{0}}{\left\Vert \boldsymbol{\kappa}\right\Vert }%
\]
so that a constrained, quadratic discriminant function%
\[
D\left(  \mathbf{s}\right)  =\left(  \mathbf{x}^{T}\mathbf{s}+1\right)
^{2}\boldsymbol{\kappa}+\kappa_{0}%
\]
delineates symmetrical decision regions $Z_{1}\simeq Z_{2}$ that are
symmetrically partitioned by the quadratic decision boundary in Eq.
(\ref{Decision Boundary Q}).

\subsection{Eigenaxis of Symmetry}

It has been shown that a constrained, quadratic discriminant function
$D\left(  \mathbf{s}\right)  =\left(  \mathbf{x}^{T}\mathbf{s}+1\right)
^{2}\boldsymbol{\kappa}+\kappa_{0}$ determines three, symmetrical quadratic
partitioning curves or surfaces, where all of the points on a quadratic
decision boundary $D_{0}\left(  \mathbf{s}\right)  $ and quadratic decision
borders $D_{+1}\left(  \mathbf{s}\right)  $ and $D_{-1}\left(  \mathbf{s}%
\right)  $ exclusively reference a constrained, primal quadratic eigenlocus
$\boldsymbol{\kappa}$. Thereby, $\boldsymbol{\kappa}$ is an eigenaxis of
symmetry which delineates symmetrical decision regions $Z_{1}\simeq Z_{2}$
that are symmetrically partitioned by a quadratic decision boundary, where the
span of both decision regions is regulated by the\ constraints in Eqs
(\ref{Decision Boundary Q}), (\ref{Decision Border One Q}), and
(\ref{Decision Border Two Q}).

\subsubsection{Illustrations of Eigenaxes of Symmetry}

\begin{figure}[ptb]%
\centering
\fbox{\includegraphics[
height=2.5875in,
width=3.813in
]%
{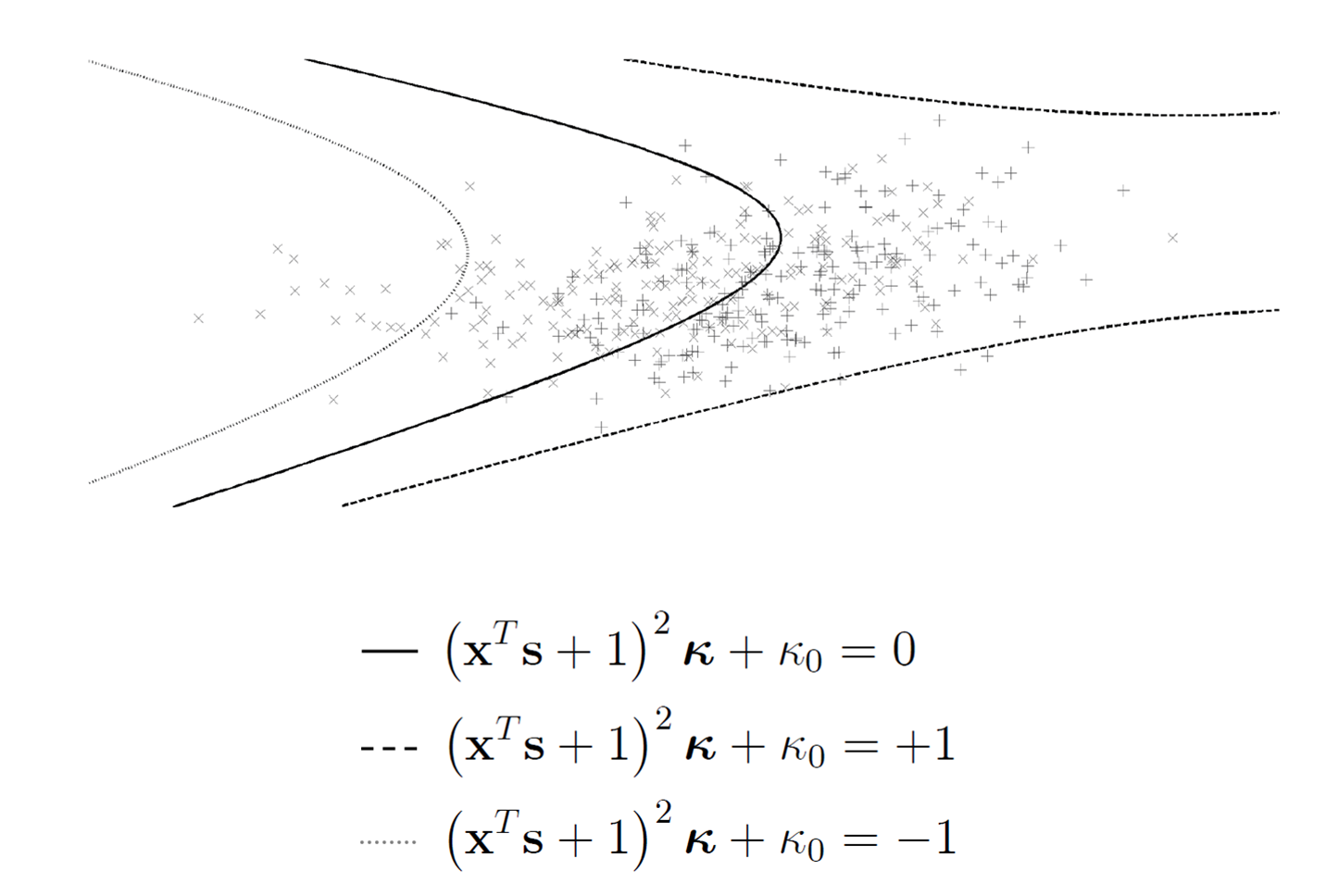}%
}\caption{Simulation example where a constrained, quadratic eigenlocus
discriminant function $D\left(  \mathbf{s}\right)  =\left(  \mathbf{x}%
^{T}\mathbf{s}+1\right)  ^{2}\boldsymbol{\kappa}+\kappa_{0}$ describes three,
symmetrical parabolic partitioning curves or surfaces, where all of the points
on a parabolic decision boundary $D_{0}\left(  \mathbf{s}\right)  $ and
parabolic decision borders $D_{+1}\left(  \mathbf{s}\right)  $ and
$D_{-1}\left(  \mathbf{s}\right)  $ exclusively reference a constrained,
primal quadratic eigenlocus $\boldsymbol{\kappa}$.}%
\label{Axis of Symmetry One}%
\end{figure}

\begin{figure}[ptb]%
\centering
\fbox{\includegraphics[
height=2.5867in,
width=3.8865in
]%
{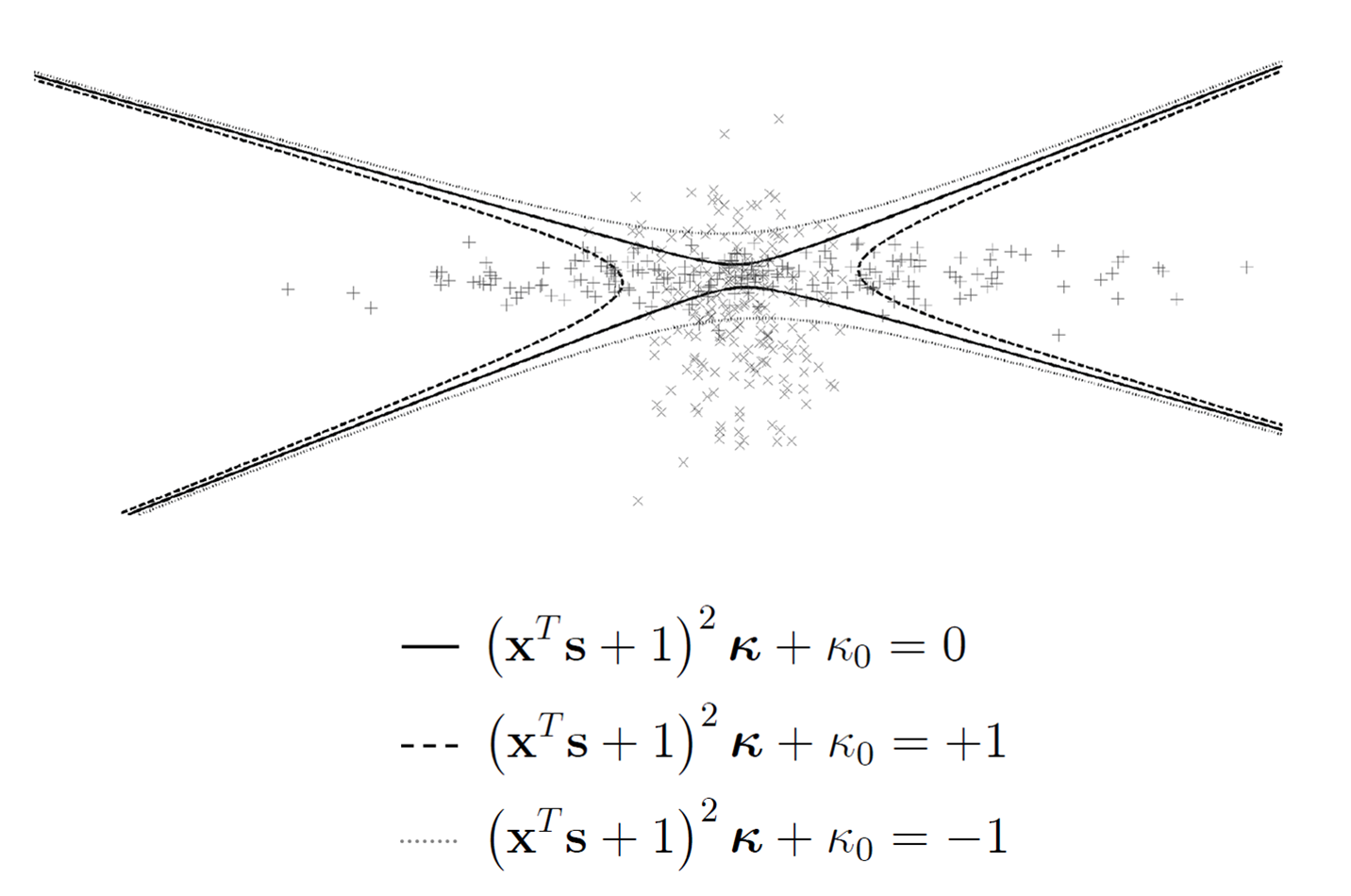}%
}\caption{Simulation example where a constrained, quadratic eigenlocus
discriminant function $D\left(  \mathbf{s}\right)  =\left(  \mathbf{x}%
^{T}\mathbf{s}+1\right)  ^{2}\boldsymbol{\kappa}+\kappa_{0}$ describes three,
symmetrical hyperbolic partitioning curves or surfaces, where all of the
points on a hyperbolic decision boundary $D_{0}\left(  \mathbf{s}\right)  $
and hyperbolic decision borders $D_{+1}\left(  \mathbf{s}\right)  $ and
$D_{-1}\left(  \mathbf{s}\right)  $ exclusively reference a constrained,
primal quadratic eigenlocus $\boldsymbol{\kappa}$.}%
\label{Axis of Symmetry Two}%
\end{figure}

\begin{figure}[ptb]%
\centering
\fbox{\includegraphics[
height=2.5875in,
width=4.1485in
]%
{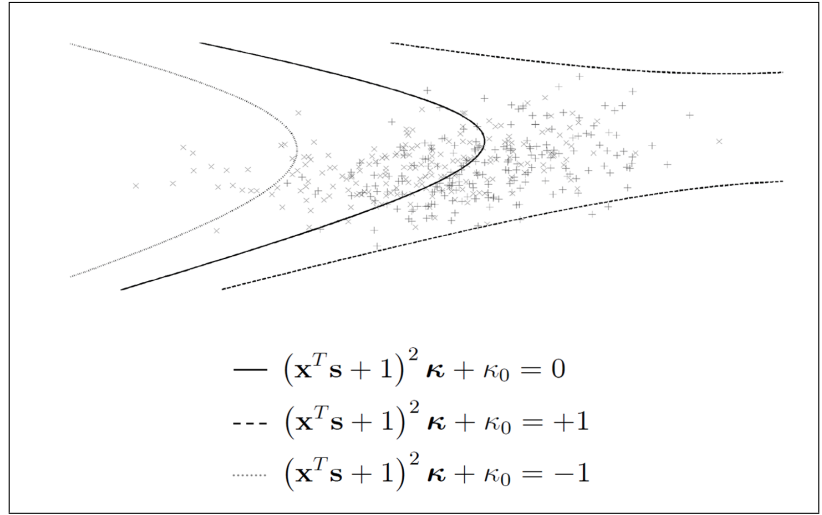}%
}\caption{Simulation example where a constrained, quadratic eigenlocus
discriminant function $D\left(  \mathbf{s}\right)  =\left(  \mathbf{x}%
^{T}\mathbf{s}+1\right)  ^{2}\boldsymbol{\kappa}+\kappa_{0}$ describes three,
symmetrical parabolic partitioning curves or surfaces, where all of the points
on a parabolic decision boundary $D_{0}\left(  \mathbf{s}\right)  $ and
parabolic decision borders $D_{+1}\left(  \mathbf{s}\right)  $ and
$D_{-1}\left(  \mathbf{s}\right)  $ exclusively reference a constrained,
primal quadratic eigenlocus $\boldsymbol{\kappa}$.}%
\label{Axis of Symmetry Three}%
\end{figure}

Figures $\ref{Axis of Symmetry One}$, $\ref{Axis of Symmetry Two}$, and
$\ref{Axis of Symmetry Three}$ show that $\boldsymbol{\kappa}$ is an eigenaxis
of symmetry which delineates symmetrical decision regions $Z_{1}\simeq Z_{2}$
that are symmetrically partitioned by a quadratic decision boundary. The
examples have been produced by simulation case studies for Gaussian data in MATLAB.

\paragraph{Eigenaxis of Symmetry for Parabolic Decision Boundary}

Figure $\ref{Axis of Symmetry One}$ illustrates a case where
$\boldsymbol{\kappa}$ is an eigenaxis of symmetry which delineates symmetrical
decision regions $Z_{1}\simeq Z_{2}$ determined by parabolic decision borders
that are symmetrically partitioned by a parabolic decision boundary.
Accordingly, a constrained, quadratic discriminant function $D\left(
\mathbf{s}\right)  =\left(  \mathbf{x}^{T}\mathbf{s}+1\right)  ^{2}%
\boldsymbol{\kappa}+\kappa_{0}$ satisfies the boundary value of a parabolic
decision boundary and the boundary values of two parabolic decision borders.%

\paragraph{Eigenaxis of Symmetry for Hyperbolic Decision Boundary}

Figure $\ref{Axis of Symmetry Two}$ illustrates a case where
$\boldsymbol{\kappa}$ is an eigenaxis of symmetry which delineates symmetrical
decision regions $Z_{1}\simeq Z_{2}$ determined by hyperbolic decision borders
that are symmetrically partitioned by a hyperbolic decision boundary.
Accordingly, a constrained, quadratic discriminant function $D\left(
\mathbf{s}\right)  =\left(  \mathbf{x}^{T}\mathbf{s}+1\right)  ^{2}%
\boldsymbol{\kappa}+\kappa_{0}$ satisfies the boundary value of a hyperbolic
decision boundary and the boundary values of two hyperbolic decision borders.%

\paragraph{Eigenaxis of Symmetry for Parabolic Decision Boundary}

Figure $\ref{Axis of Symmetry Three}$ illustrates another case where
$\boldsymbol{\kappa}$ is an eigenaxis of symmetry which delineates symmetrical
decision regions $Z_{1}\simeq Z_{2}$ determined by parabolic decision borders
that are symmetrically partitioned by a parabolic decision boundary.
Accordingly, a constrained, quadratic discriminant function $D\left(
\mathbf{s}\right)  =\left(  \mathbf{x}^{T}\mathbf{s}+1\right)  ^{2}%
\boldsymbol{\kappa}+\kappa_{0}$ satisfies the boundary value of a parabolic
decision boundary and the boundary values of two parabolic decision borders.%

\subsubsection{New Notation and Terminology}

I\ will show that the \emph{constrained}, quadratic eigenlocus discriminant
function $D\left(  \mathbf{s}\right)  =\left(  \mathbf{x}^{T}\mathbf{s}%
+1\right)  ^{2}\boldsymbol{\kappa}+\kappa_{0}$ determines a discrete,
quadratic \emph{classification system} $\left(  \mathbf{x}^{T}\mathbf{s}%
+1\right)  ^{2}\boldsymbol{\kappa}+\kappa_{0}\overset{\omega_{1}%
}{\underset{\omega_{2}}{\gtrless}}0$, where $\boldsymbol{\kappa=\kappa}%
_{1}-\boldsymbol{\kappa}_{2}$ is the \emph{likelihood ratio} of the
classification system. Define the \emph{primal focus} of the quadratic
classification system $\left(  \mathbf{x}^{T}\mathbf{s}+1\right)
^{2}\boldsymbol{\kappa}+\kappa_{0}\overset{\omega_{1}}{\underset{\omega
_{2}}{\gtrless}}0$ to be an \emph{equilibrium point} that defines quadratic
decision borders $D_{+1}\left(  \mathbf{s}\right)  $ and $D_{-1}\left(
\mathbf{s}\right)  $ located at symmetrically constrained distances from a
quadratic decision boundary $D_{0}\left(  \mathbf{s}\right)  $.

Therefore, let $\widetilde{\Lambda}_{\boldsymbol{\kappa}}\left(
\mathbf{s}\right)  =\left(  \mathbf{x}^{T}\mathbf{s}+1\right)  ^{2}%
\boldsymbol{\kappa}+\kappa_{0}$ denote a quadratic eigenlocus discriminant
function and let $\widehat{\Lambda}_{\boldsymbol{\kappa}}\left(
\mathbf{s}\right)  =\boldsymbol{\kappa}_{1}-\boldsymbol{\kappa}_{2}$ denote
the likelihood ratio of the quadratic classification system $\left(
\mathbf{x}^{T}\mathbf{s}+1\right)  ^{2}\boldsymbol{\kappa}+\kappa
_{0}\overset{\omega_{1}}{\underset{\omega_{2}}{\gtrless}}0$ which is a
likelihood ratio test $\widetilde{\Lambda}_{\boldsymbol{\kappa}}\left(
\mathbf{s}\right)  \overset{\omega_{1}}{\underset{\omega_{2}}{\gtrless}}0$.
The likelihood ratio $\widehat{\Lambda}_{\boldsymbol{\kappa}}\left(
\mathbf{s}\right)  =\boldsymbol{\kappa}_{1}-\boldsymbol{\kappa}_{2}$ is said
to be the primary \emph{focus} of the quadratic classification system $\left(
\mathbf{x}^{T}\mathbf{s}+1\right)  ^{2}\boldsymbol{\kappa}+\kappa
_{0}\overset{\omega_{1}}{\underset{\omega_{2}}{\gtrless}}0$.

\subsection{The Quadratic Eigenlocus Test}

I\ will now derive a statistic for the $\kappa_{0}$ term in Eq.
(\ref{Discriminant Function Q}). I\ will use the statistic to derive a
likelihood statistic that is the basis of a quadratic eigenlocus decision rule.

\subsubsection{Locus Equation for the $\kappa_{0}$ Term}

Using the KKT condition in Eq. (\ref{KKTE5 Q}) and the KKT condition of
complementary slackness, it follows that the following set of locus equations
must be satisfied:%
\[
y_{i}\left(  \left(  \mathbf{x}^{T}\mathbf{x}_{i\ast}+1\right)  ^{2}%
\boldsymbol{\kappa}+\kappa_{0}\right)  -1+\xi_{i}=0,\ i=1,...,l\text{,}%
\]
such that an estimate for $\kappa_{0}$ satisfies the locus equation:%
\begin{align}
\kappa_{0}  &  =\sum\nolimits_{i=1}^{l}y_{i}\left(  1-\xi_{i}\right)
-\sum\nolimits_{i=1}^{l}\left(  \left(  \mathbf{x}^{T}\mathbf{x}_{i\ast
}+1\right)  ^{2}\right)  \boldsymbol{\kappa}%
\label{Normal Eigenlocus Projection Factor Q}\\
&  =\sum\nolimits_{i=1}^{l}y_{i}\left(  1-\xi_{i}\right)  -\left(
\sum\nolimits_{i=1}^{l}\left(  \mathbf{x}^{T}\mathbf{x}_{i\ast}+1\right)
^{2}\right)  \boldsymbol{\kappa}\text{.}\nonumber
\end{align}

I\ will now use the statistic for $\kappa_{0}$ to derive a vector expression
for a quadratic eigenlocus test that is used to classify unknown pattern vectors.

Substitution of the statistic for $\kappa_{0}$ in Eq.
(\ref{Normal Eigenlocus Projection Factor Q}) into the expression for the
quadratic discriminant function $D\left(  \mathbf{s}\right)  $ in Eq.
(\ref{Discriminant Function Q}) provides a quadratic eigenlocus test
$\widetilde{\Lambda}_{\boldsymbol{\kappa}}\left(  \mathbf{s}\right)
\overset{H_{1}}{\underset{H_{2}}{\gtrless}}0$ for classifying an unknown
pattern vector $\mathbf{s}$:%
\begin{align}
\widetilde{\Lambda}_{\boldsymbol{\kappa}}\left(  \mathbf{s}\right)   &
=\left(  \left(  \mathbf{x}^{T}\mathbf{s}+1\right)  ^{2}\right)
\boldsymbol{\kappa}-\sum\nolimits_{i=1}^{l}\left(  \left(  \mathbf{x}%
^{T}\mathbf{x}_{i\ast}+1\right)  ^{2}\right)  \boldsymbol{\kappa
}\label{NormalEigenlocusTestStatistic Q}\\
&  \mathbf{+}\sum\nolimits_{i=1}^{l}y_{i}\left(  1-\xi_{i}\right)
\overset{\omega_{1}}{\underset{\omega_{2}}{\gtrless}}0\nonumber\\
&  =\left(  \left(  \mathbf{x}^{T}\mathbf{s}+1\right)  ^{2}-\sum
\nolimits_{i=1}^{l}\left(  \mathbf{x}^{T}\mathbf{x}_{i\ast}+1\right)
^{2}\right)  \boldsymbol{\kappa}\nonumber\\
&  \mathbf{+}\sum\nolimits_{i=1}^{l}y_{i}\left(  1-\xi_{i}\right)
\overset{\omega_{1}}{\underset{\omega_{2}}{\gtrless}}0\text{,}\nonumber
\end{align}
where the statistic $\sum\nolimits_{i=1}^{l}\left(  \mathbf{x}^{T}%
\mathbf{x}_{i\ast}+1\right)  ^{2}$ is the locus of an aggregate or cluster of
a set of $l$ extreme points, and the statistic $\sum\nolimits_{i=1}^{l}%
y_{i}\left(  1-\xi_{i}\right)  $ accounts for the class membership of the
primal principal eigenaxis components on $\boldsymbol{\kappa}_{1}$ and
$\boldsymbol{\kappa}_{2}$.

\subsubsection{Locus of Aggregated Risk $\protect\widehat{\mathfrak{R}}$}

The cluster $\sum\nolimits_{i=1}^{l}\left(  \mathbf{x}^{T}\mathbf{x}_{i\ast
}+1\right)  ^{2}$ of a set of extreme points represents the aggregated risk
$\widehat{\mathfrak{R}}$ for a decision space $Z$. Accordingly, the vector
transform%
\[
\left(  \mathbf{x}^{T}\mathbf{s}+1\right)  ^{2}-\sum\nolimits_{i=1}^{l}\left(
\mathbf{x}^{T}\mathbf{x}_{i\ast}+1\right)  ^{2}%
\]
accounts for the distance between the unknown vector $\mathbf{s}$ and the
locus of aggregated risk $\widehat{\mathfrak{R}}$. Let $\widehat{\mathbf{x}%
}_{i\ast}\triangleq\sum\nolimits_{i=1}^{l}\left(  \mathbf{x}^{T}%
\mathbf{x}_{i\ast}+1\right)  ^{2}$.

\subsubsection{Quadratic Decision Locus}

Denote a unit quadratic eigenlocus $\boldsymbol{\kappa}\mathbf{/}\left\Vert
\boldsymbol{\kappa}\right\Vert $ by $\widehat{\boldsymbol{\kappa}}$. Letting
$\boldsymbol{\kappa}=\boldsymbol{\kappa}\mathbf{/}\left\Vert
\boldsymbol{\kappa}\right\Vert $ in Eq. (\ref{NormalEigenlocusTestStatistic Q}%
) provides an expression for a decision locus%
\begin{align}
\widehat{D}\left(  \mathbf{s}\right)   &  =\left(  k_{\mathbf{s}%
}-k_{\widehat{\mathbf{x}}_{i\ast}}\right)  \boldsymbol{\kappa}\mathbf{/}%
\left\Vert \boldsymbol{\kappa}\right\Vert
\label{Statistical Locus of Category Decision Q}\\
&  \mathbf{+}\frac{1}{\left\Vert \boldsymbol{\kappa}\right\Vert }%
\sum\nolimits_{i=1}^{l}y_{i}\left(  1-\xi_{i}\right) \nonumber
\end{align}
which is determined by the scalar projection of $k_{\mathbf{s}}%
-k_{\widehat{\mathbf{x}}_{i\ast}}$ onto $\widehat{\boldsymbol{\kappa}}$.
Accordingly, the component of $k_{\mathbf{s}}-k_{\widehat{\mathbf{x}}_{i\ast}%
}$ along $\widehat{\boldsymbol{\kappa}}$ specifies a signed magnitude
$\left\Vert k_{\mathbf{s}}-k_{\widehat{\mathbf{x}}_{i\ast}}\right\Vert
\cos\theta$ along the axis of $\widehat{\boldsymbol{\kappa}}$, where $\theta$
is the angle between the transformed vector $k_{\mathbf{s}}%
-k_{\widehat{\mathbf{x}}_{i\ast}}$ and $\widehat{\boldsymbol{\kappa}}$.

\begin{figure}[ptb]%
\centering
\fbox{\includegraphics[
height=2.5875in,
width=3.4411in
]%
{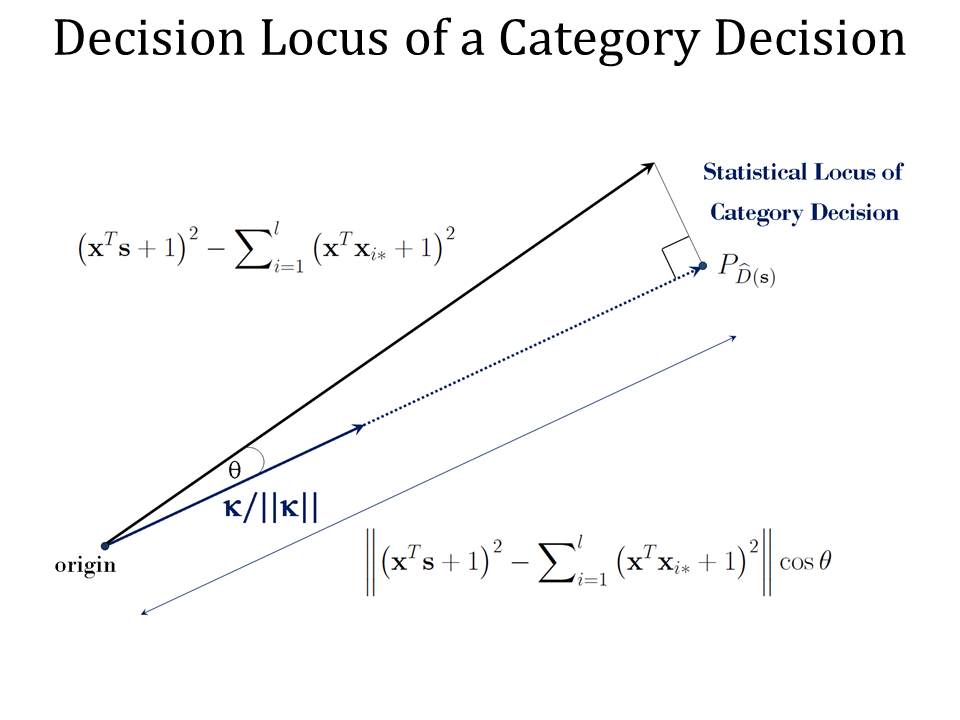}%
}\caption{A statistical decision locus $P_{\protect\widehat{D}\left(
\mathbf{s}\right)  }$ for an unknown, transformed pattern vector
$k_{\mathbf{s}}-k_{\overline{\mathbf{x}}_{i\ast}}$ that is projected onto
$\boldsymbol{\kappa}\mathbf{/}\left\Vert \boldsymbol{\kappa}\right\Vert $.}%
\label{Statistical Decision Locus Q}%
\end{figure}

It follows that the component $\operatorname{comp}%
_{\overrightarrow{\widehat{\boldsymbol{\kappa}}}}\left(
\overrightarrow{\left(  k_{\mathbf{s}}-k_{\widehat{\mathbf{x}}_{i\ast}%
}\right)  }\right)  $ of the vector transform $k_{\mathbf{s}}%
-k_{\widehat{\mathbf{x}}_{i\ast}}$ of an unknown pattern vector $k_{\mathbf{s}%
}$ along the axis of a unit quadratic eigenlocus $\widehat{\boldsymbol{\kappa
}}$%
\[
P_{\widehat{D}\left(  \mathbf{s}\right)  }=\operatorname{comp}%
_{\overrightarrow{\widehat{\boldsymbol{\kappa}}}}\left(
\overrightarrow{\left(  k_{\mathbf{s}}-k_{\widehat{\mathbf{x}}_{i\ast}%
}\right)  }\right)  =\left\Vert k_{\mathbf{s}}-k_{\widehat{\mathbf{x}}_{i\ast
}}\right\Vert \cos\theta
\]
specifies a locus $P_{\widehat{D}\left(  \mathbf{s}\right)  }$ of a category
decision, where $P_{\widehat{D}\left(  \mathbf{s}\right)  }$ is at a distance
of $\left\Vert k_{\mathbf{s}}-k_{\widehat{\mathbf{x}}_{i\ast}}\right\Vert
\cos\theta$ from the origin, along the axis of a quadratic eigenlocus
$\boldsymbol{\kappa}$. Accordingly, the quadratic discriminant function
$D\left(  \mathbf{s}\right)  $ in Eq. (\ref{Discriminant Function Q}) makes a
decision based on the value of the decision locus $\operatorname{comp}%
_{\overrightarrow{\widehat{\boldsymbol{\kappa}}}}\left(
\overrightarrow{\left(  k_{\mathbf{s}}-k_{\widehat{\mathbf{x}}_{i\ast}%
}\right)  }\right)  $ and the class membership statistic $\frac{1}{\left\Vert
\boldsymbol{\kappa}\right\Vert }\sum\nolimits_{i=1}^{l}y_{i}\left(  1-\xi
_{i}\right)  $. Figure $\ref{Statistical Decision Locus Q}$ depicts a decision
locus generated by the quadratic discriminant function $\widehat{D}\left(
\mathbf{s}\right)  $ in Eq. (\ref{Statistical Locus of Category Decision Q}).%

\subsubsection{Quadratic Decision Threshold}

Returning to Eq. (\ref{General Form of Decision Function II}), recall that an
optimal decision function computes the likelihood ratio $\Lambda\left(
\mathbf{x}\right)  $ for a feature vector $\mathbf{x}$ and makes a decision by
comparing the ratio $\Lambda\left(  \mathbf{x}\right)  $ to the threshold
$\eta=0$. Given Eqs (\ref{Decision Boundary Q}) and
(\ref{Statistical Locus of Category Decision Q}), it follows that a quadratic
eigenlocus test $\widetilde{\Lambda}_{\boldsymbol{\kappa}}\left(
\mathbf{s}\right)  \overset{H_{1}}{\underset{H_{2}}{\gtrless}}0$ makes a
decision by comparing the output%
\[
\operatorname{sign}\left(  \operatorname{comp}%
_{\overrightarrow{\widehat{\boldsymbol{\kappa}}}}\left(
\overrightarrow{\left(  k_{\mathbf{s}}-k_{\widehat{\mathbf{x}}_{i\ast}%
}\right)  }\right)  +\frac{1}{\left\Vert \boldsymbol{\kappa}\right\Vert }%
\sum\nolimits_{i=1}^{l}y_{i}\left(  1-\xi_{i}\right)  \right)  \text{,}%
\]
where $\operatorname{sign}\left(  x\right)  \equiv\frac{x}{\left\vert
x\right\vert }$ for $x\neq0$, to a threshold $\eta$ along the axis of
$\widehat{\boldsymbol{\kappa}}$ in $%
\mathbb{R}
^{d}$, where $\eta=0$.

\subsection{Quadratic Eigenlocus Decision Rules}

Substitution of the equation for $\boldsymbol{\kappa}$ in Eq.
(\ref{Pair of Normal Eigenlocus Components Q}) into Eq.
(\ref{NormalEigenlocusTestStatistic Q}) provides a quadratic eigenlocus test
in terms of the primal eigenlocus components $\boldsymbol{\kappa}_{1}$ and
$\boldsymbol{\kappa}_{2}$:%
\begin{align}
\widetilde{\Lambda}_{\boldsymbol{\kappa}}\left(  \mathbf{s}\right)   &
=\left(  \left(  \mathbf{x}^{T}\mathbf{s}+1\right)  ^{2}-\sum\nolimits_{i=1}%
^{l}\left(  \mathbf{x}^{T}\mathbf{x}_{i\ast}+1\right)  ^{2}\right)
\boldsymbol{\kappa}_{1}\label{NormalEigenlocusTestStatistic2 Q}\\
&  -\left(  \left(  \mathbf{x}^{T}\mathbf{s}+1\right)  ^{2}-\sum
\nolimits_{i=1}^{l}\left(  \mathbf{x}^{T}\mathbf{x}_{i\ast}+1\right)
^{2}\right)  \boldsymbol{\kappa}_{2}\nonumber\\
&  \mathbf{+}\sum\nolimits_{i=1}^{l}y_{i}\left(  1-\xi_{i}\right)
\overset{\omega_{1}}{\underset{\omega_{2}}{\gtrless}}0\text{.}\nonumber
\end{align}

I will show that a constrained, primal quadratic eigenlocus
$\boldsymbol{\kappa}$ and its Wolfe dual $\boldsymbol{\psi}$ possess an
essential statistical property which enables quadratic eigenlocus discriminant
functions $\widetilde{\Lambda}_{\boldsymbol{\kappa}}\left(  \mathbf{s}\right)
=\left(  \mathbf{x}^{T}\mathbf{s}+1\right)  ^{2}\boldsymbol{\kappa}+\kappa
_{0}$ to satisfy a discrete version of the fundamental integral equation of
binary classification:%
\begin{align*}
f\left(  \widetilde{\Lambda}_{\boldsymbol{\kappa}}\left(  \mathbf{s}\right)
\right)   &  =\;\int\nolimits_{Z_{1}}p\left(  k_{\mathbf{x}_{1_{i\ast}}%
}|\boldsymbol{\kappa}_{1}\right)  d\boldsymbol{\kappa}_{1}+\int%
\nolimits_{Z_{2}}p\left(  k_{\mathbf{x}_{1_{i\ast}}}|\boldsymbol{\kappa}%
_{1}\right)  d\boldsymbol{\kappa}_{1}+\delta\left(  y\right)  \sum
\nolimits_{i=1}^{l_{1}}\psi_{1_{i_{\ast}}}\\
&  =\int\nolimits_{Z_{1}}p\left(  k_{\mathbf{x}_{2_{i\ast}}}%
|\boldsymbol{\kappa}_{2}\right)  d\boldsymbol{\kappa}_{2}+\int\nolimits_{Z_{2}%
}p\left(  k_{\mathbf{x}_{2_{i\ast}}}|\boldsymbol{\kappa}_{2}\right)
d\boldsymbol{\kappa}_{2}-\delta\left(  y\right)  \sum\nolimits_{i=1}^{l_{2}%
}\psi_{2_{i_{\ast}}}\text{,}%
\end{align*}
over the decision space $Z=Z_{1}+Z_{2}$, where $\delta\left(  y\right)
\triangleq\sum\nolimits_{i=1}^{l}y_{i}\left(  1-\xi_{i}\right)  $, and all of
the forces associated with counter risks $\overline{\mathfrak{R}%
}_{\mathfrak{\min}}\left(  Z_{1}|\boldsymbol{\kappa}_{1}\right)  $ and
$\overline{\mathfrak{R}}_{\mathfrak{\min}}\left(  Z_{2}|\boldsymbol{\kappa
}_{2}\right)  $ and risks $\mathfrak{R}_{\mathfrak{\min}}\left(
Z_{1}|\boldsymbol{\kappa}_{2}\right)  $ and $\mathfrak{R}_{\mathfrak{\min}%
}\left(  Z_{2}|\boldsymbol{\kappa}_{1}\right)  $ within the $Z_{1}$ and
$Z_{2}$ decision regions are symmetrically balanced with each other:%
\begin{align*}
f\left(  \widetilde{\Lambda}_{\boldsymbol{\kappa}}\left(  \mathbf{s}\right)
\right)   &  :\;\int\nolimits_{Z_{1}}p\left(  k_{\mathbf{x}_{1_{i\ast}}%
}|\boldsymbol{\kappa}_{1}\right)  d\boldsymbol{\kappa}_{1}-\int%
\nolimits_{Z_{1}}p\left(  k_{\mathbf{x}_{2_{i\ast}}}|\boldsymbol{\kappa}%
_{2}\right)  d\boldsymbol{\kappa}_{2}+\delta\left(  y\right)  \sum
\nolimits_{i=1}^{l_{1}}\psi_{1_{i_{\ast}}}\\
&  =\int\nolimits_{Z_{2}}p\left(  k_{\mathbf{x}_{2_{i\ast}}}%
|\boldsymbol{\kappa}_{2}\right)  d\boldsymbol{\kappa}_{2}-\int\nolimits_{Z_{2}%
}p\left(  k_{\mathbf{x}_{1_{i\ast}}}|\boldsymbol{\kappa}_{1}\right)
d\boldsymbol{\kappa}_{1}-\delta\left(  y\right)  \sum\nolimits_{i=1}^{l_{2}%
}\psi_{2_{i_{\ast}}}%
\end{align*}
by means of an integral equation:%
\begin{align*}
f\left(  \widetilde{\Lambda}_{\boldsymbol{\kappa}}\left(  \mathbf{s}\right)
\right)   &  =\int\nolimits_{Z}p\left(  k_{\mathbf{x}_{1_{i\ast}}%
}|\boldsymbol{\kappa}_{1}\right)  d\boldsymbol{\kappa}_{1}=\left\Vert
\boldsymbol{\kappa}_{1}\right\Vert _{\min_{c}}^{2}+C_{1}\\
&  =\int\nolimits_{Z}p\left(  k_{\mathbf{x}_{2_{i\ast}}}|\boldsymbol{\kappa
}_{2}\right)  d\boldsymbol{\kappa}_{2}=\left\Vert \boldsymbol{\kappa}%
_{2}\right\Vert _{\min_{c}}^{2}+C_{2}\text{,}%
\end{align*}
where $p\left(  k_{\mathbf{x}_{2_{i\ast}}}|\boldsymbol{\kappa}_{2}\right)  $
and $p\left(  k_{\mathbf{x}_{1_{i\ast}}}|\boldsymbol{\kappa}_{1}\right)  $ are
class-conditional densities for respective extreme points $k_{\mathbf{x}%
_{2_{i\ast}}}$ and $k_{\mathbf{x}_{1_{i\ast}}}$, and $C_{1}$ and $C_{2}$ are
integration constants for $\boldsymbol{\kappa}_{1}$ and $\boldsymbol{\kappa
}_{2}$ respectively. I will define this property after I\ define the
fundamental properties possessed by a Wolfe dual quadratic eigenlocus
$\boldsymbol{\psi}$.

\section{The Wolfe Dual Eigenspace II}

Let there be $l$ principal eigenaxis components $\left\{  \psi_{i\ast
}\overrightarrow{\mathbf{e}}_{i}|\psi_{i\ast}>0\right\}  _{i=1}^{l}$ on a
constrained, primal quadratic eigenlocus within its Wolfe dual eigenspace:%
\[
\max\Xi\left(  \boldsymbol{\psi}\right)  =\mathbf{1}^{T}\boldsymbol{\psi
}-\frac{\boldsymbol{\psi}^{T}\mathbf{Q}\boldsymbol{\psi}}{2}\text{,}%
\]
where the Wolfe dual quadratic eigenlocus $\boldsymbol{\psi}$ satisfies the
constraints $\boldsymbol{\psi}^{T}\mathbf{y}=0$ and $\psi_{i\ast}>0$.

The theorem for convex duality guarantees an equivalence and corresponding
symmetry between a constrained, primal quadratic eigenlocus
$\boldsymbol{\kappa}$ and its Wolfe dual $\boldsymbol{\psi}$. Raleigh's
principle
\citep[see][]{Strang1986}
and the theorem for convex duality jointly indicate that Eq.
(\ref{Vector Form Wolfe Dual Q}) provides an estimate of the largest
eigenvector $\boldsymbol{\psi}$ of a kernel matrix $\mathbf{Q}$, where
$\boldsymbol{\psi}$ satisfies the constraints $\boldsymbol{\psi}^{T}%
\mathbf{y}=0$ and $\psi_{i}\geq0$, such that $\boldsymbol{\psi}$ is a
principal eigenaxis of three, symmetrical\textit{\ }quadratic partitioning
surfaces associated with the constrained quadratic form $\boldsymbol{\psi}%
^{T}\mathbf{Q}\boldsymbol{\psi}$.

I will now show that maximization of the functional $\mathbf{1}^{T}%
\boldsymbol{\psi}-\boldsymbol{\psi}^{T}\mathbf{Q}\boldsymbol{\psi}\mathbf{/}2$
requires that $\boldsymbol{\psi}$ satisfy an eigenenergy constraint which is
symmetrically related to the restriction of the primal quadratic eigenlocus
$\boldsymbol{\kappa}$ to its Wolfe dual eigenspace.

\subsection{Eigenenergy Constraint on $\boldsymbol{\psi}$}

Equation (\ref{Minimum Total Eigenenergy Primal Normal Eigenlocus Q}) and the
theorem for convex duality jointly indicate that $\boldsymbol{\psi}$ satisfies
an eigenenergy constraint that is symmetrically related to the eigenenergy
constraint on $\boldsymbol{\kappa}$ within its Wolfe dual eigenspace:%
\[
\left\Vert \boldsymbol{\psi}\right\Vert _{\min_{c}}^{2}\cong\left\Vert
\boldsymbol{\kappa}\right\Vert _{\min_{c}}^{2}\text{.}%
\]
Therefore, a Wolfe dual quadratic eigenlocus $\boldsymbol{\psi}$ satisfies an
eigenenergy constraint%
\[
\max\boldsymbol{\psi}^{T}\mathbf{Q}\boldsymbol{\psi}=\lambda_{\max
\boldsymbol{\psi}}\left\Vert \boldsymbol{\psi}\right\Vert _{\min_{c}}^{2}%
\]
for which the functional $\mathbf{1}^{T}\boldsymbol{\psi}-\boldsymbol{\psi
}^{T}\mathbf{Q}\boldsymbol{\psi}\mathbf{/}2$ in Eq.
(\ref{Vector Form Wolfe Dual Q}) is maximized by the largest eigenvector
$\boldsymbol{\psi}$ of $\mathbf{Q}$, such that the constrained quadratic form
$\boldsymbol{\psi}^{T}\mathbf{Q}\boldsymbol{\psi}\mathbf{/}2$\textbf{,} where
$\boldsymbol{\psi}^{T}\mathbf{y}=0$ and $\psi_{i}\geq0$, reaches its smallest
possible value. This indicates that principal eigenaxis components on a Wolfe
dual quadratic eigenlocus $\boldsymbol{\psi}$ satisfy minimum length
constraints. Principal eigenaxis components on a Wolfe dual quadratic
eigenlocus $\boldsymbol{\psi}$ also satisfy an equilibrium constraint.

\subsection{Equilibrium Constraint on $\boldsymbol{\psi}$}

The KKT condition in Eq. (\ref{KKTE2 Q}) requires that the magnitudes of the
Wolfe dual principal eigenaxis components on $\boldsymbol{\psi}$ satisfy the
equation:%
\[
\left(  y_{i}=1\right)  \sum\nolimits_{i=1}^{l_{1}}\psi_{1_{i\ast}}+\left(
y_{i}=-1\right)  \sum\nolimits_{i=1}^{l_{2}}\psi_{2_{i\ast}}=0
\]
so that%
\begin{equation}
\sum\nolimits_{i=1}^{l_{1}}\psi_{1_{i\ast}}-\sum\nolimits_{i=1}^{l_{2}}%
\psi_{2_{i\ast}}=0\text{.} \label{Wolfe Dual Equilibrium Point Q}%
\end{equation}
It follows that the integrated lengths of the Wolfe dual principal eigenaxis
components correlated with each pattern category must \emph{balance} each
other:%
\begin{equation}
\sum\nolimits_{i=1}^{l_{1}}\psi_{1_{i\ast}}\rightleftharpoons\sum
\nolimits_{i=1}^{l_{2}}\psi_{2_{i\ast}}\text{.}
\label{Equilibrium Constraint on Dual Eigen-components Q}%
\end{equation}

Accordingly, let $l_{1}+l_{2}=l$ and express a Wolfe dual quadratic eigenlocus
$\boldsymbol{\psi}$ in terms of $l$ non-orthogonal unit vectors $\left\{
\overrightarrow{\mathbf{e}}_{1\ast},\ldots,\overrightarrow{\mathbf{e}}_{l\ast
}\right\}  $:%
\begin{align}
\boldsymbol{\psi}  &  =\sum\nolimits_{i=1}^{l}\psi_{i\ast}%
\overrightarrow{\mathbf{e}}_{i\ast}\label{Wolfe Dual Vector Equation Q}\\
&  =\sum\nolimits_{i=1}^{l_{1}}\psi_{1i\ast}\overrightarrow{\mathbf{e}%
}_{1i\ast}+\sum\nolimits_{i=1}^{l_{2}}\psi_{2i\ast}\overrightarrow{\mathbf{e}%
}_{2i\ast}\nonumber\\
&  =\boldsymbol{\psi}_{1}+\boldsymbol{\psi}_{2}\text{,}\nonumber
\end{align}
where each scaled, non-orthogonal unit vector $\psi_{1i\ast}%
\overrightarrow{\mathbf{e}}_{1i\ast}$ or $\psi_{2i\ast}%
\overrightarrow{\mathbf{e}}_{2i\ast}$ is correlated with an extreme vector
$\left(  \mathbf{x}^{T}\mathbf{x}_{1_{i\ast}}+1\right)  ^{2}$ or $\left(
\mathbf{x}^{T}\mathbf{x}_{2_{i\ast}}+1\right)  ^{2}$ respectively,
$\boldsymbol{\psi}_{1}$ denotes the Wolfe dual eigenlocus component
$\sum\nolimits_{i=1}^{l_{1}}\psi_{1_{i\ast}}\overrightarrow{\mathbf{e}%
}_{1_{i\ast}}$, and $\boldsymbol{\psi}_{2}$ denotes the Wolfe dual eigenlocus
component $\sum\nolimits_{i=1}^{l_{2}}\psi_{2_{i\ast}}%
\overrightarrow{\mathbf{e}}_{2_{i\ast}}$.

Given Eq. (\ref{Equilibrium Constraint on Dual Eigen-components Q}) and data
distributions that have dissimilar covariance matrices, it follows that the
forces associated with counter risks and risks, within each of the symmetrical
decision regions, are balanced with each other. Given Eq.
(\ref{Equilibrium Constraint on Dual Eigen-components Q}) and data
distributions that have similar covariance matrices, it follows that the
forces associated with counter risks within each of the symmetrical decision
regions are equal to each other, and the forces associated with risks within
each of the symmetrical decision regions are equal to each other.

Given Eqs (\ref{Equilibrium Constraint on Dual Eigen-components Q}) and
(\ref{Wolfe Dual Vector Equation Q}), it follows that the axis of a Wolfe dual
quadratic eigenlocus $\boldsymbol{\psi}$ can be regarded as a lever that is
formed by \emph{sets of principal eigenaxis components which are evenly or
equally distributed over either side of the\emph{ axis }of }$\boldsymbol{\psi
}$\emph{, where a fulcrum is placed directly under the center of the axis of
}$\boldsymbol{\psi}$.

Thereby, the axis of $\boldsymbol{\psi}$ is in statistical equilibrium, where
all of the principal eigenaxis components on $\boldsymbol{\psi}$ are equal or
in correct proportions, relative to the center of $\boldsymbol{\psi}$, such
that the opposing forces associated with risks and counter risks of a
quadratic classification system are symmetrically balanced with each other.
Figure $\ref{Quadratic Dual Locus in Statistical Equilibrium}$\textbf{
}illustrates the axis of $\mathbf{\psi}$ in statistical equilibrium.\textbf{%
\begin{figure}[ptb]%
\centering
\fbox{\includegraphics[
height=2.5875in,
width=3.4411in
]%
{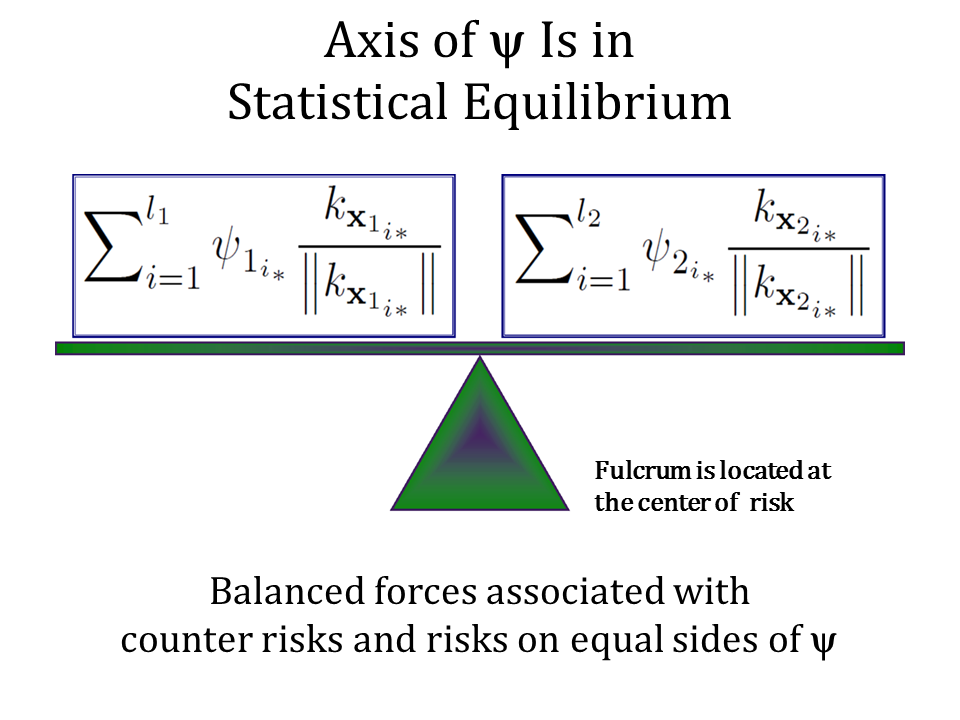}%
}\caption{All of the principal eigenaxis components on $\boldsymbol{\psi}$
have equal or correct proportions, relative to the center of $\boldsymbol{\psi
}$, so that opposing forces associated with risks and counter risks are
symmetrically balanced with each other.}%
\label{Quadratic Dual Locus in Statistical Equilibrium}%
\end{figure}
}

Using Eq. (\ref{Equilibrium Constraint on Dual Eigen-components Q}), it
follows that the length $\left\Vert \boldsymbol{\psi}_{1}\right\Vert $ of
$\boldsymbol{\psi}_{1}$ is balanced with the length $\left\Vert
\boldsymbol{\psi}_{2}\right\Vert $ of $\boldsymbol{\psi}_{2}$:%
\begin{equation}
\left\Vert \boldsymbol{\psi}_{1}\right\Vert \rightleftharpoons\left\Vert
\boldsymbol{\psi}_{2}\right\Vert
\label{Equilibrium Constraint on Dual Component Lengths Q}%
\end{equation}
and that the total allowed eigenenergies exhibited by $\boldsymbol{\psi}_{1}$
and $\boldsymbol{\psi}_{2}$ are balanced with each other:%
\begin{equation}
\left\Vert \boldsymbol{\psi}_{1}\right\Vert _{\min_{c}}^{2}\rightleftharpoons
\left\Vert \boldsymbol{\psi}_{2}\right\Vert _{\min_{c}}^{2}\text{.}
\label{Symmetrical Balance of Wolf Dual Eigenenergies Q}%
\end{equation}

Therefore, the equilibrium constraint on $\boldsymbol{\psi}$ in Eq.
(\ref{Equilibrium Constraint on Dual Eigen-components Q}) ensures that the
total allowed eigenenergies exhibited by the Wolfe dual principal eigenaxis
components on $\boldsymbol{\psi}_{1}$ and $\boldsymbol{\psi}_{2}$ are
symmetrically balanced with each other:%
\[
\left\Vert \sum\nolimits_{i=1}^{l_{1}}\psi_{1_{i\ast}}%
\overrightarrow{\mathbf{e}}_{1_{i\ast}}\right\Vert _{\min_{c}}^{2}%
\rightleftharpoons\left\Vert \sum\nolimits_{i=1}^{l_{2}}\psi_{2_{i\ast}%
}\overrightarrow{\mathbf{e}}_{2_{i\ast}}\right\Vert _{\min_{c}}^{2}%
\]
about the center of total allowed eigenenergy $\left\Vert \boldsymbol{\psi
}\right\Vert _{\min_{c}}^{2}$: which is located at the geometric center of
$\boldsymbol{\psi}$ because $\left\Vert \boldsymbol{\psi}_{1}\right\Vert
\equiv\left\Vert \boldsymbol{\psi}_{2}\right\Vert $. This indicates that the
total allowed eigenenergies of $\boldsymbol{\psi}$ are distributed over its
axis in a symmetrically balanced and well-proportioned manner.

\subsection{Symmetrical Balance Exhibited by the Axis of $\boldsymbol{\psi}$}

Given Eqs (\ref{Equilibrium Constraint on Dual Component Lengths Q}) and
(\ref{Symmetrical Balance of Wolf Dual Eigenenergies Q}), it follows that the
axis of a Wolfe dual quadratic eigenlocus $\boldsymbol{\psi}$ can be regarded
as a lever that has \emph{equal weight on equal sides of a centrally placed
fulcrum}.

Thereby, the axis of $\boldsymbol{\psi}$ is a lever that has an equal
distribution of eigenenergies on equal sides of a centrally placed fulcrum.
Later on, I\ will show that symmetrically balanced, joint distributions of
principal eigenaxis components on $\boldsymbol{\psi}$ and $\boldsymbol{\kappa
}$ are symmetrically distributed over the axes of the Wolfe dual principal
eigenaxis components on $\boldsymbol{\psi}$ and the unconstrained, correlated
primal principal eigenaxis components (extreme vectors) on $\boldsymbol{\kappa
}$.

\begin{figure}[ptb]%
\centering
\fbox{\includegraphics[
height=2.5875in,
width=3.4411in
]%
{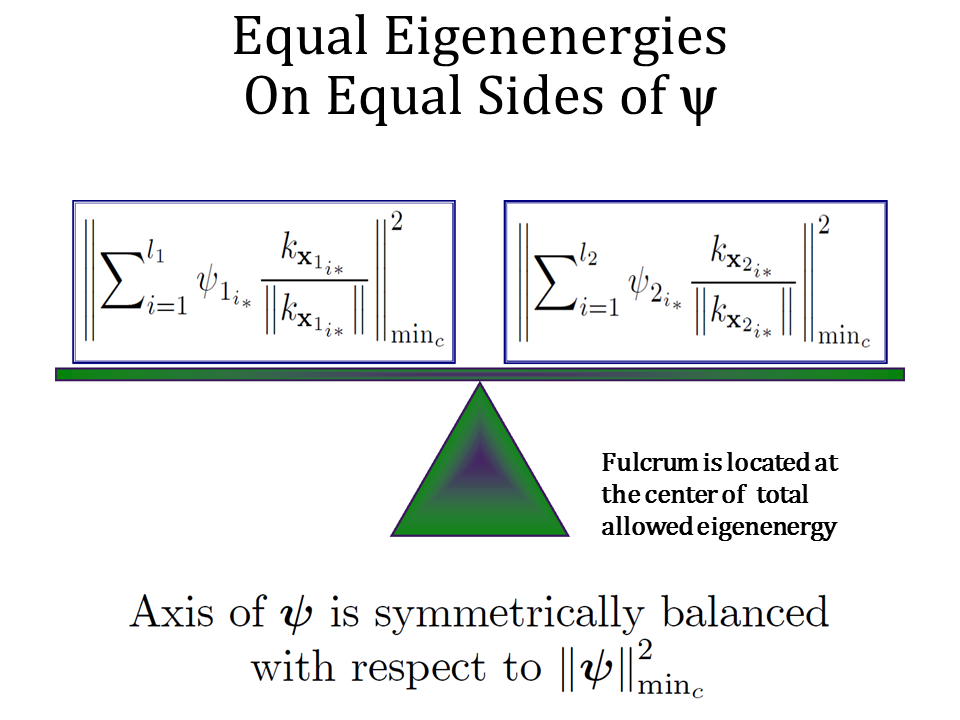}%
}\caption{The axis of $\boldsymbol{\psi}$ can be regarded as a lever that has
an equal distribution of eigenenergies on equal sides of a centrally placed
fulcrum which is located at the center of total allowed eigenenergy
$\left\Vert \boldsymbol{\psi}\right\Vert _{\min_{c}}^{2}$ of $\boldsymbol{\psi
}$.}%
\label{Symmetrical Balance of Wolfe Dual Quadratic Eigenlocus}%
\end{figure}

Figure $\ref{Symmetrical Balance of Wolfe Dual Quadratic Eigenlocus}$ depicts
how the axis of $\boldsymbol{\psi}$ can be regarded as a lever that has an
equal distribution of eigenenergies on equal sides of a centrally placed
fulcrum which is located at the geometric center, i.e., the critical minimum
eigenenergy $\left\Vert \boldsymbol{\psi}\right\Vert _{\min_{c}}^{2}$, of
$\boldsymbol{\psi}$.%

\subsection{Statistics for Quadratic Partitions}

Returning to Section $10.4$, recall that the eigenspectrum of a kernel matrix
$\mathbf{Q}$ determines the shapes of the quadratic surfaces which are
specified by the constrained quadratic form in Eq.
(\ref{Vector Form Wolfe Dual Q}), where the eigenvalues $\lambda_{N}\leq$
$\lambda_{N-1}\leq\ldots\leq\lambda_{1}$ of a kernel matrix $\mathbf{Q}$ are
essentially determined by its inner product elements $\varphi\left(
\mathbf{x}_{i},\mathbf{x}_{j}\right)  $.

Therefore, let the kernel matrix $\mathbf{Q}$ in Eq.
(\ref{Vector Form Wolfe Dual Q}) contain inner product statistics
$\varphi\left(  \mathbf{x}_{i},\mathbf{x}_{j}\right)  =\left(  \mathbf{x}%
_{i}^{T}\mathbf{x}_{j}+1\right)  ^{2}$ for a quadratic decision boundary
$Q_{0}\left(  \mathbf{x}\right)  $ and quadratic decision borders
$Q_{+1}\left(  \mathbf{x}\right)  $ and $Q_{-1}\left(  \mathbf{x}\right)  $
that delineate symmetrical decision regions $Z_{1}\simeq Z_{2}$. Later on,
I\ will show that quadratic eigenlocus transforms map the labeled $\pm1$,
inner product statistics $\left(  \mathbf{x}_{i}^{T}\mathbf{x}_{j}+1\right)
^{2}$ contained within $\mathbf{Q}$%
\[
\mathbf{Q}\boldsymbol{\psi}=\lambda\mathbf{_{\max\boldsymbol{\psi}}%
}\boldsymbol{\psi}%
\]
into a Wolfe dual quadratic eigenlocus of principal eigenaxis components%
\[
\mathbf{Q\sum\nolimits_{i=1}^{l}}\psi_{i\ast}\overrightarrow{\mathbf{e}%
}_{i\ast}=\lambda\mathbf{_{\max\boldsymbol{\psi}}}\sum\nolimits_{i=1}^{l}%
\psi_{i\ast}\overrightarrow{\mathbf{e}}_{i\ast}\text{,}%
\]
formed by $l$ scaled $\psi_{i\ast}$, non-orthogonal unit vectors $\left\{
\overrightarrow{\mathbf{e}}_{1\ast},\ldots,\overrightarrow{\mathbf{e}}_{l\ast
}\right\}  $, where the locus of each Wolfe dual principal eigenaxis component
$\psi_{i\ast}\overrightarrow{\mathbf{e}}_{i\ast}\ $is determined by the
direction and well-proportioned magnitude of a correlated extreme vector
$\left(  \mathbf{x}^{T}\mathbf{x}_{i\ast}+1\right)  ^{2}$.

By way of motivation, I\ will now define the fundamental property possessed by
a quadratic eigenlocus $\boldsymbol{\kappa}$ that enables a quadratic
eigenlocus discriminant function $\widetilde{\Lambda}_{\boldsymbol{\kappa}%
}\left(  \mathbf{s}\right)  =\left(  \mathbf{x}^{T}\mathbf{s}+1\right)
^{2}\boldsymbol{\kappa}+\kappa_{0}$ to satisfy the fundamental integral
equation of binary classification in Eq. (\ref{Equalizer Rule}).

\section{The Property of Symmetrical Balance II}

I\ have demonstrated that constrained, quadratic eigenlocus discriminant
functions $D\left(  \mathbf{s}\right)  =\left(  \mathbf{x}^{T}\mathbf{s}%
+1\right)  ^{2}\boldsymbol{\kappa}+\kappa_{0}$ determine symmetrical
$Z_{1}\simeq Z_{2}$ decision regions $Z_{1}$ and $Z_{2}$ that are delineated
by quadratic decision borders $D_{+1}\left(  \mathbf{s}\right)  $ and
$D_{-1}\left(  \mathbf{s}\right)  $, which satisfy symmetrically balanced
constraints relative to the constraint on a quadratic decision boundary
$D_{0}\left(  \mathbf{s}\right)  $, where all of the points $\mathbf{s}$ on
$D_{+1}\left(  \mathbf{s}\right)  $, $D_{-1}\left(  \mathbf{s}\right)  $, and
$D_{0}\left(  \mathbf{s}\right)  $ reference $\boldsymbol{\kappa}$.

Therefore, I\ have shown that constrained, quadratic eigenlocus discriminant
functions $D\left(  \mathbf{s}\right)  =\left(  \mathbf{x}^{T}\mathbf{s}%
+1\right)  ^{2}\boldsymbol{\kappa}+\kappa_{0}$ satisfy boundary values of
quadratic decision borders $D_{+1}\left(  \mathbf{s}\right)  $ and
$D_{-1}\left(  \mathbf{s}\right)  $ and quadratic decision boundaries
$D_{0}\left(  \mathbf{s}\right)  $, where the axis of $\boldsymbol{\kappa}$ is
an axis of symmetry for $D_{+1}\left(  \mathbf{s}\right)  $, $D_{-1}\left(
\mathbf{s}\right)  $, and $D_{0}\left(  \mathbf{s}\right)  $.

Given that $\boldsymbol{\kappa}$ is an axis of symmetry which satisfies
boundary values of quadratic decision borders $D_{+1}\left(  \mathbf{s}%
\right)  $ and $D_{-1}\left(  \mathbf{s}\right)  $ and quadratic decision
boundaries $D_{0}\left(  \mathbf{s}\right)  $, it follows that
$\boldsymbol{\kappa}$ \emph{must posses} the statistical property of
\emph{symmetrical balance}. Recall that the physical property of symmetrical
balance involves an axis or lever in equilibrium where different elements are
equal or in correct proportions, relative to the center of an axis or a lever,
such that the opposing forces or influences of a system are balanced with each other.

\subsection{Symmetrical Balance Exhibited by the Axis of $\boldsymbol{\kappa}%
$}

Returning to Eqs (\ref{Equilibrium Constraint on Dual Eigen-components Q}) and
(\ref{Wolfe Dual Vector Equation Q}), recall that the axis of
$\boldsymbol{\psi}$ can be regarded as a lever in statistical equilibrium
where different principal eigenaxis components are equal or in correct
proportions, relative to the center of $\boldsymbol{\psi}$, such that the
opposing forces associated with the risks and the counter risks of a quadratic
classification system are balanced with each other. Thus, the axis of
$\boldsymbol{\psi=\psi}_{1}+\boldsymbol{\psi}_{2}$ exhibits the statistical
property of symmetrical balance, where $\sum\nolimits_{i=1}^{l_{1}}%
\psi_{1_{i\ast}}\overrightarrow{\mathbf{e}}_{1_{i\ast}}\equiv\sum
\nolimits_{i=1}^{l_{2}}\psi_{2_{i\ast}}\overrightarrow{\mathbf{e}}_{2_{i\ast}%
}$.

Furthermore, given Eqs
(\ref{Equilibrium Constraint on Dual Component Lengths Q}) and
(\ref{Symmetrical Balance of Wolf Dual Eigenenergies Q}), the axis of
$\boldsymbol{\psi}$ can be regarded as a lever that has an equal distribution
of eigenenergies on equal sides of a centrally placed fulcrum which is located
at the center of total allowed eigenenergy $\left\Vert \boldsymbol{\psi
}\right\Vert _{\min_{c}}^{2}$ of $\boldsymbol{\psi}$. Accordingly, the total
allowed eigenenergies possessed by the principal eigenaxis components on
$\boldsymbol{\psi}$ are symmetrically balanced with each other about a center
of total allowed eigenenergy $\left\Vert \boldsymbol{\psi}\right\Vert
_{\min_{c}}^{2}$ which is located at the geometric center of $\boldsymbol{\psi
}$. Thus, the axis of $\boldsymbol{\psi=\psi}_{1}+\boldsymbol{\psi}_{2}$
exhibits the statistical property of symmetrical balance, where $\left\Vert
\boldsymbol{\psi}_{1}\right\Vert \equiv\left\Vert \boldsymbol{\psi}%
_{2}\right\Vert $ and $\left\Vert \sum\nolimits_{i=1}^{l_{1}}\psi_{1_{i\ast}%
}\overrightarrow{\mathbf{e}}_{1_{i\ast}}\right\Vert _{\min_{c}}^{2}%
\equiv\left\Vert \sum\nolimits_{i=1}^{l_{2}}\psi_{2_{i\ast}}%
\overrightarrow{\mathbf{e}}_{2_{i\ast}}\right\Vert _{\min_{c}}^{2}$.

Returning to Eqs (\ref{Characteristic Eigenenergy of Quadratic}) and
(\ref{Characteristic Eigenenergy of Quadratic 2}), recall that the locus of
any quadratic curve or surface is determined by the eigenenergy $\left\Vert
\boldsymbol{\nu}\right\Vert ^{2}$ exhibited by the locus of its principal
eigenaxis $\boldsymbol{\nu}$, where any given principal eigenaxis
$\boldsymbol{\nu}$ and any given point $\mathbf{x}$ on a quadratic locus
satisfies the eigenenergy $\left\Vert \boldsymbol{\nu}\right\Vert ^{2}$ of
$\boldsymbol{\nu}$. Accordingly, the inherent property of a quadratic locus
and its principal eigenaxis $\boldsymbol{\nu}$ is the eigenenergy $\left\Vert
\boldsymbol{\nu}\right\Vert ^{2}$ exhibited by $\boldsymbol{\nu}$.

\begin{figure}[ptb]%
\centering
\fbox{\includegraphics[
height=2.5875in,
width=2.6809in
]%
{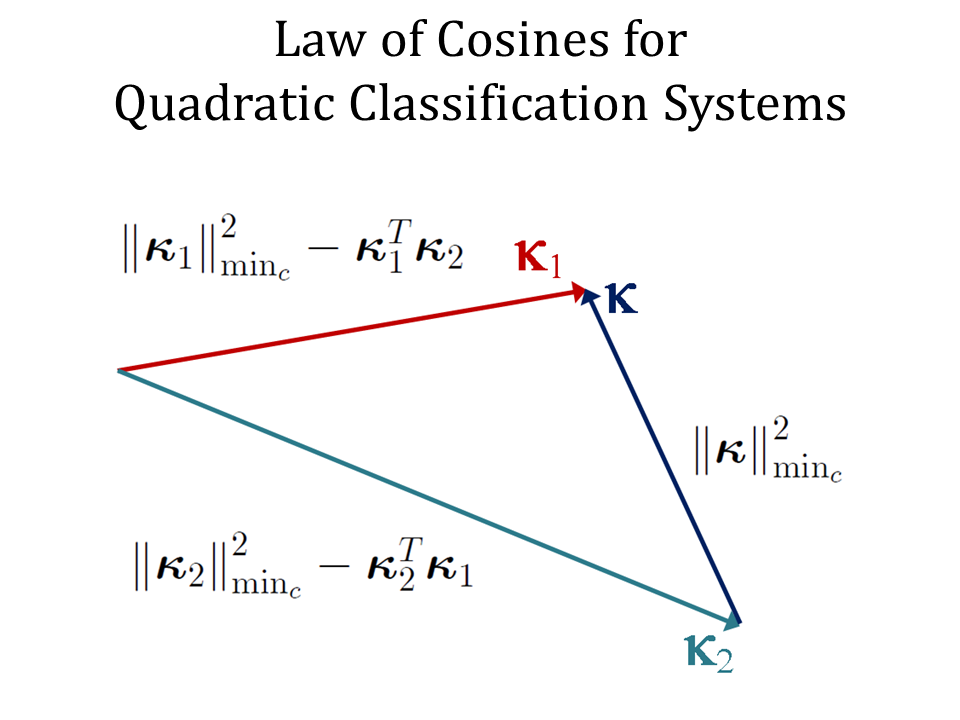}%
}\caption{The likelihood ratio $\protect\widehat{\Lambda}_{\boldsymbol{\kappa
}}\left(  \mathbf{s}\right)  =\boldsymbol{\kappa}_{1}-\boldsymbol{\kappa}_{2}$
of a quadratic eigenlocus discriminant function $\protect\widetilde{\Lambda
}_{\boldsymbol{\kappa}}\left(  \mathbf{s}\right)  =\left(  \mathbf{x}%
^{T}\mathbf{s}+1\right)  ^{2}\boldsymbol{\kappa}+\kappa_{0}$ satisfies the law
of cosines in a symmetrically balanced manner.}%
\label{Law of Cosines for Quadratic Classification Systems}%
\end{figure}

Therefore, Eqs (\ref{Characteristic Eigenenergy of Quadratic}),
(\ref{Characteristic Eigenenergy of Quadratic 2}),
(\ref{Minimum Total Eigenenergy Primal Normal Eigenlocus Q}), and
(\ref{Pair of Normal Eigenlocus Components Q}) jointly indicate that a
constrained, primal quadratic eigenlocus $\boldsymbol{\kappa}$ satisfies the
quadratic decision boundary $D_{0}\left(  \mathbf{s}\right)  $ in Eq
(\ref{Decision Boundary Q}) and the quadratic decision borders $D_{+1}\left(
\mathbf{s}\right)  $ and $D_{-1}\left(  \mathbf{s}\right)  $ in Eqs
(\ref{Decision Border One Q}) and (\ref{Decision Border Two Q}) in terms of
its total allowed eigenenergies%
\begin{align*}
\left\Vert \boldsymbol{\kappa}\right\Vert _{\min_{c}}^{2}  &  =\left\Vert
\boldsymbol{\kappa}_{1}-\boldsymbol{\kappa}_{2}\right\Vert _{\min_{c}}^{2}\\
&  \cong\left[  \left\Vert \boldsymbol{\kappa}_{1}\right\Vert _{\min_{c}}%
^{2}-\boldsymbol{\kappa}_{1}^{T}\boldsymbol{\kappa}_{2}\right]  +\left[
\left\Vert \boldsymbol{\kappa}_{2}\right\Vert _{\min_{c}}^{2}%
-\boldsymbol{\kappa}_{2}^{T}\boldsymbol{\kappa}_{1}\right]  \text{,}%
\end{align*}
where the functional $\left\Vert \boldsymbol{\kappa}_{1}\right\Vert _{\min
_{c}}^{2}-\boldsymbol{\kappa}_{1}^{T}\boldsymbol{\kappa}_{2}$ is associated
with the $D_{+1}\left(  \mathbf{s}\right)  $ quadratic decision border, the
functional $\left\Vert \boldsymbol{\kappa}_{2}\right\Vert _{\min_{c}}%
^{2}-\boldsymbol{\kappa}_{2}^{T}\boldsymbol{\kappa}_{1}$ is associated with
the $D_{-1}\left(  \mathbf{s}\right)  $ quadratic decision border, and the
functional $\left\Vert \boldsymbol{\kappa}\right\Vert _{\min_{c}}^{2}$ is
associated with the quadratic decision boundary $D_{0}\left(  \mathbf{s}%
\right)  $. Thus, the total allowed eigenenergies of the principal eigenaxis
components on a quadratic eigenlocus $\boldsymbol{\kappa=\kappa}%
_{1}-\boldsymbol{\kappa}_{2}$ must satisfy the law of cosines in the
symmetrically balanced manner depicted in Fig.
$\ref{Law of Cosines for Quadratic Classification Systems}$.%

Given that $\boldsymbol{\kappa}$ must possess the statistical property of
symmetrical balance in terms of its principal eigenaxis components, it follows
that the axis of $\boldsymbol{\kappa}$ is essentially a lever that is
symmetrically balanced with respect to the center of eigenenergy $\left\Vert
\boldsymbol{\kappa}_{1}-\boldsymbol{\kappa}_{2}\right\Vert _{\min_{c}}^{2}$ of
$\boldsymbol{\kappa}$. Accordingly, the axis of $\boldsymbol{\kappa}$ is said
to be in statistical equilibrium, where the constrained, primal principal
eigenaxis components on $\boldsymbol{\kappa}$%
\begin{align*}
\boldsymbol{\kappa}  &  =\boldsymbol{\kappa}_{1}-\boldsymbol{\kappa}_{2}\\
&  =\sum\nolimits_{i=1}^{l_{1}}\psi_{1_{i\ast}}\left(  \mathbf{x}%
^{T}\mathbf{x}_{1_{i\ast}}+1\right)  ^{2}-\sum\nolimits_{i=1}^{l_{2}}%
\psi_{2_{i\ast}}\left(  \mathbf{x}^{T}\mathbf{x}_{2_{i\ast}}+1\right)  ^{2}%
\end{align*}
are equal or in correct proportions, relative to the center of
$\boldsymbol{\kappa}$, such that all of the forces associated with the risks
$\mathfrak{R}_{\mathfrak{\min}}\left(  Z_{1}|\boldsymbol{\kappa}_{2}\right)  $
and $\mathfrak{R}_{\mathfrak{\min}}\left(  Z_{2}|\boldsymbol{\kappa}%
_{1}\right)  $ and all of the forces associated with the counter risks
$\overline{\mathfrak{R}}_{\mathfrak{\min}}\left(  Z_{1}|\boldsymbol{\kappa
}_{1}\right)  $ and $\overline{\mathfrak{R}}_{\mathfrak{\min}}\left(
Z_{2}|\boldsymbol{\kappa}_{2}\right)  $ of a binary, quadratic classification
system $\left(  \mathbf{x}^{T}\mathbf{s}+1\right)  ^{2}\boldsymbol{\kappa
}+\kappa_{0}\overset{\omega_{1}}{\underset{\omega_{2}}{\gtrless}}0$ are
symmetrically balanced with each other.

\begin{figure}[ptb]%
\centering
\fbox{\includegraphics[
height=2.5875in,
width=3.4411in
]%
{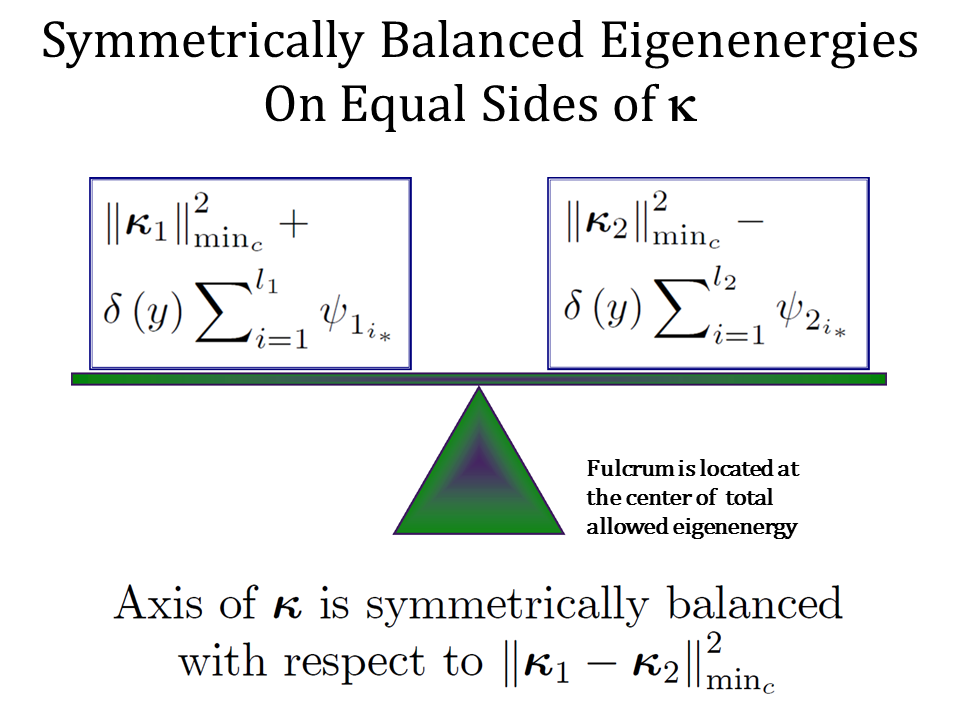}%
}\caption{A constrained, quadratic eigenlocus discriminant function
$\protect\widetilde{\Lambda}_{\boldsymbol{\kappa}}\left(  \mathbf{s}\right)
=\left(  \mathbf{x}^{T}\mathbf{s}+1\right)  ^{2}\boldsymbol{\kappa}+\kappa
_{0}$ satisfies a fundamental integral equation of binary classification for a
quadratic classification system in statistical equilibrium, where the expected
risk $\mathfrak{R}_{\mathfrak{\min}}\left(  Z\mathbf{|}\boldsymbol{\kappa
}\right)  $ and the total allowed eigenenergy $\left\Vert \boldsymbol{\kappa
}\right\Vert _{\min_{c}}^{2}$ of the system are minimized, because the axis of
$\boldsymbol{\kappa}$ is essentially a lever that is symmetrically balanced
with respect to the center of eigenenergy $\left\Vert \boldsymbol{\kappa}%
_{1}-\boldsymbol{\kappa}_{2}\right\Vert _{\min_{c}}^{2}$ of
$\boldsymbol{\kappa}$.}%
\label{Symmetrical Balance of Constrained Primal Quadratic Eigenlocus}%
\end{figure}

I will prove that a constrained, quadratic eigenlocus discriminant function
$\widetilde{\Lambda}_{\boldsymbol{\kappa}}\left(  \mathbf{s}\right)  =\left(
\mathbf{x}^{T}\mathbf{s}+1\right)  ^{2}\boldsymbol{\kappa}+\kappa_{0}$
satisfies a discrete and data-driven version of the fundamental integral
equation of binary classification for a classification system in statistical
equilibrium in Eq. (\ref{Equalizer Rule}) because the axis of $\mathbf{\kappa
}$ is essentially a lever that is symmetrically balanced with respect to the
center of eigenenergy $\left\Vert \mathbf{\kappa}_{1}-\mathbf{\kappa}%
_{2}\right\Vert _{\min_{c}}^{2}$ of $\mathbf{\kappa}$ in the following manner:%
\[
\left\Vert \mathbf{\kappa}_{1}\right\Vert _{\min_{c}}^{2}-\left\Vert
\mathbf{\kappa}_{1}\right\Vert \left\Vert \mathbf{\kappa}_{2}\right\Vert
\cos\theta_{\mathbf{\kappa}_{1}\mathbf{\kappa}_{2}}+\delta\left(  y\right)
\frac{1}{2}\sum\nolimits_{i=1}^{l}\psi_{_{i\ast}}\equiv\frac{1}{2}\left\Vert
\mathbf{\kappa}\right\Vert _{\min_{c}}^{2}%
\]
and%
\[
\left\Vert \mathbf{\kappa}_{2}\right\Vert _{\min_{c}}^{2}-\left\Vert
\mathbf{\kappa}_{2}\right\Vert \left\Vert \mathbf{\kappa}_{1}\right\Vert
\cos\theta_{\mathbf{\kappa}_{2}\mathbf{\kappa}_{1}}-\delta\left(  y\right)
\frac{1}{2}\sum\nolimits_{i=1}^{l}\psi_{_{i\ast}}\equiv\frac{1}{2}\left\Vert
\mathbf{\kappa}\right\Vert _{\min_{c}}^{2}%
\]
where the equalizer statistic $\nabla_{eq}$%
\[
\nabla_{eq}\triangleq\delta\left(  y\right)  \frac{1}{2}\sum\nolimits_{i=1}%
^{l}\psi_{_{i\ast}}%
\]
for which $\delta\left(  y\right)  \triangleq\sum\nolimits_{i=1}^{l}%
y_{i}\left(  1-\xi_{i}\right)  $, equalizes the total allowed eigenenergies
$\left\Vert \mathbf{\kappa}_{1}\right\Vert _{\min_{c}}^{2}$ and $\left\Vert
\mathbf{\kappa}_{2}\right\Vert _{\min_{c}}^{2}$ exhibited by $\mathbf{\kappa
}_{1}$ and $\mathbf{\kappa}_{2}$:%
\[
\left\Vert \mathbf{\kappa}_{1}\right\Vert _{\min_{c}}^{2}+\delta\left(
y\right)  \frac{1}{2}\sum\nolimits_{i=1}^{l}\psi_{_{i\ast}}\equiv\left\Vert
\mathbf{\kappa}_{2}\right\Vert _{\min_{c}}^{2}-\delta\left(  y\right)
\frac{1}{2}\sum\nolimits_{i=1}^{l}\psi_{_{i\ast}}%
\]
so that the dual locus of $\mathbf{\kappa}_{1}-\mathbf{\kappa}_{2}$ is in
statistical equilibrium. Figure
$\ref{Symmetrical Balance of Constrained Primal Quadratic Eigenlocus}$
illustrates the property of symmetrical balance exhibited by the dual locus of
$\boldsymbol{\kappa}$.%

I\ will obtain the above equations for a quadratic eigenlocus
$\boldsymbol{\kappa}$ in statistical equilibrium by devising a chain of
arguments which demonstrate that a constrained quadratic eigenlocus
discriminant function $\widetilde{\Lambda}_{\boldsymbol{\kappa}}\left(
\mathbf{s}\right)  =\left(  \mathbf{x}^{T}\mathbf{s}+1\right)  ^{2}%
\boldsymbol{\kappa}+\kappa_{0}$ satisfies discrete and data-driven versions of
the fundamental equations of binary classification for a classification system
in statistical equilibrium in Eqs
(\ref{Vector Equation of Likelihood Ratio and Decision Boundary}) -
(\ref{Balancing of Bayes' Risks and Counteracting Risks}).

The general course of my argument, which is outlined below, follows the
general course of my argument for linear eigenlocus transforms.

\section{General Course of Argument II}

In order to prove that a constrained quadratic eigenlocus discriminant
function $\widetilde{\Lambda}_{\boldsymbol{\kappa}}\left(  \mathbf{s}\right)
=\left(  \mathbf{x}^{T}\mathbf{s}+1\right)  ^{2}\boldsymbol{\kappa}+\kappa
_{0}$ satisfies $\left(  1\right)  $ the vector equation%
\[
\left(  \mathbf{x}^{T}\mathbf{s}+1\right)  ^{2}\boldsymbol{\kappa}+\kappa
_{0}=0\text{,}%
\]
$\left(  2\right)  $ the statistical equilibrium equation:%
\[
p\left(  \widehat{\Lambda}_{\boldsymbol{\kappa}}\left(  \mathbf{s}\right)
|\omega_{1}\right)  \rightleftharpoons p\left(  \widehat{\Lambda
}_{\boldsymbol{\kappa}}\left(  \mathbf{s}\right)  |\omega_{2}\right)  \text{,}%
\]
$\left(  3\right)  $ the corresponding integral equation:%
\[
\int_{Z}p\left(  \widehat{\Lambda}_{\boldsymbol{\kappa}}\left(  \mathbf{s}%
\right)  |\omega_{1}\right)  d\widehat{\Lambda}_{\boldsymbol{\kappa}}=\int%
_{Z}p\left(  \widehat{\Lambda}_{\boldsymbol{\kappa}}\left(  \mathbf{s}\right)
|\omega_{2}\right)  d\widehat{\Lambda}_{\boldsymbol{\kappa}}\text{,}%
\]
$\left(  4\right)  $ a discrete, quadratic version of the fundamental integral
equation of binary classification for a classification system in statistical
equilibrium:%
\begin{align*}
f\left(  \widetilde{\Lambda}_{\boldsymbol{\kappa}}\left(  \mathbf{s}\right)
\right)   &  =\int_{Z_{1}}p\left(  \widehat{\Lambda}_{\boldsymbol{\kappa}%
}\left(  \mathbf{s}\right)  |\omega_{1}\right)  d\widehat{\Lambda
}_{\boldsymbol{\kappa}}+\int_{Z_{2}}p\left(  \widehat{\Lambda}%
_{\boldsymbol{\kappa}}\left(  \mathbf{s}\right)  |\omega_{1}\right)
d\widehat{\Lambda}_{\boldsymbol{\kappa}}+\delta\left(  y\right)
\sum\nolimits_{i=1}^{l_{1}}\psi_{1_{i_{\ast}}}\\
&  =\int_{Z_{1}}p\left(  \widehat{\Lambda}_{\boldsymbol{\kappa}}\left(
\mathbf{s}\right)  |\omega_{2}\right)  d\widehat{\Lambda}_{\boldsymbol{\kappa
}}+\int_{Z_{2}}p\left(  \widehat{\Lambda}_{\boldsymbol{\kappa}}\left(
\mathbf{s}\right)  |\omega_{2}\right)  d\widehat{\Lambda}_{\boldsymbol{\kappa
}}-\delta\left(  y\right)  \sum\nolimits_{i=1}^{l_{2}}\psi_{2_{i_{\ast}}%
}\text{,}%
\end{align*}
and $\left(  5\right)  $ the corresponding integral equation:%
\begin{align*}
f\left(  \widetilde{\Lambda}_{\boldsymbol{\kappa}}\left(  \mathbf{s}\right)
\right)   &  :\;\int_{Z_{1}}p\left(  \widehat{\Lambda}_{\boldsymbol{\kappa}%
}\left(  \mathbf{s}\right)  |\omega_{1}\right)  d\widehat{\Lambda
}_{\boldsymbol{\kappa}}-\int_{Z_{1}}p\left(  \widehat{\Lambda}%
_{\boldsymbol{\kappa}}\left(  \mathbf{s}\right)  |\omega_{2}\right)
\widehat{\Lambda}_{\boldsymbol{\kappa}}+\delta\left(  y\right)  \sum
\nolimits_{i=1}^{l_{1}}\psi_{1_{i_{\ast}}}\\
&  =\int_{Z_{2}}p\left(  \widehat{\Lambda}_{\boldsymbol{\kappa}}\left(
\mathbf{s}\right)  |\omega_{2}\right)  d\widehat{\Lambda}_{\boldsymbol{\kappa
}}-\int_{Z_{2}}p\left(  \widehat{\Lambda}_{\boldsymbol{\kappa}}\left(
\mathbf{s}\right)  |\omega_{1}\right)  d\widehat{\Lambda}_{\boldsymbol{\kappa
}}-\delta\left(  y\right)  \sum\nolimits_{i=1}^{l_{2}}\psi_{2_{i_{\ast}}%
}\text{,}%
\end{align*}
I will need to develop mathematical machinery for several systems of locus
equations. The fundamental equations of a binary classification system involve
mathematical machinery and systems of locus equations that determine the
following mathematical objects:

\begin{enumerate}
\item Total allowed eigenenergies of extreme points on a Wolfe dual and a
constrained primal quadratic eigenlocus.

\item Total allowed eigenenergies of Wolfe dual and constrained quadratic
linear eigenlocus components.

\item Total allowed eigenenergy of a Wolfe dual and a constrained primal
quadratic eigenlocus.

\item Class-conditional probability density functions for extreme points.

\item Conditional probability functions for extreme points.

\item Risks and counter risks of extreme points.

\item Conditional probability functions for the risks and the counter risks
related to positions and potential locations of extreme points.

\item Integral equations of class-conditional probability density functions.
\end{enumerate}

A high level overview of the development of the mathematical machinery and
systems of locus equations is outlined below.

I\ will develop class-conditional probability density functions and
conditional probability functions for extreme points in the following manner:

Let $k_{\mathbf{x}_{i\ast}}$ denote an extreme point, where $k_{\mathbf{x}%
_{i\ast}}$ is a reproducing kernel of an extreme point $\mathbf{x}_{i\ast}$.
Using Eq. (\ref{Geometric Locus of Second-order Reproducing Kernel}), any
given extreme point $k_{\mathbf{x}_{i\ast}}$ is the endpoint on a locus of
\emph{random} \emph{variables}%
\[%
\begin{pmatrix}
\left(  \left\Vert \mathbf{x}_{i\ast}\right\Vert ^{2}+1\right)  \cos
\mathbb{\alpha}_{k\left(  \mathbf{x}_{i\ast1}\right)  1}, & \left(  \left\Vert
\mathbf{x}_{i\ast}\right\Vert ^{2}+1\right)  \cos\mathbb{\alpha}_{k\left(
\mathbf{x}_{i\ast2}\right)  2}, & \cdots, & \left(  \left\Vert \mathbf{x}%
_{i\ast}\right\Vert ^{2}+1\right)  \cos\mathbb{\alpha}_{k\left(
\mathbf{x}_{i\ast d}\right)  d}%
\end{pmatrix}
\text{,}%
\]
where each random variable is characterized by an expected value%
\[
E\left[  \left(  \left\Vert \mathbf{x}_{i\ast}\right\Vert ^{2}+1\right)
\cos\mathbb{\alpha}_{k\left(  \mathbf{x}_{i\ast}\right)  j}\right]
\]
and a variance%
\[
\operatorname{var}\left(  \left(  \left\Vert \mathbf{x}_{i\ast}\right\Vert
^{2}+1\right)  \cos\mathbb{\alpha}_{k\left(  \mathbf{x}_{i\ast}\right)
j}\right)  \text{.}%
\]
\qquad

Therefore, an extreme point $k_{\mathbf{x}_{i\ast}}$ is described by a central
location (an expected value) and a covariance (a spread). The relative
likelihood that an extreme point has a given location is described by a
conditional probability density function. The cumulative probability of given
locations for an extreme point, i.e., the probability of finding the extreme
point within a localized region, is described by a conditional probability
function
\citep{Ash1993,Flury1997}%
.

So, take the Wolfe dual quadratic eigenlocus in Eq.
(\ref{Wolfe Dual Vector Equation}):%
\begin{align*}
\boldsymbol{\psi}  &  =\sum\nolimits_{i=1}^{l_{1}}\psi_{1i\ast}%
\overrightarrow{\mathbf{e}}_{1i\ast}+\sum\nolimits_{i=1}^{l_{2}}\psi_{2i\ast
}\overrightarrow{\mathbf{e}}_{2i\ast}\\
&  =\boldsymbol{\psi}_{1}+\boldsymbol{\psi}_{2}\text{,}%
\end{align*}
where each scaled, non-orthogonal unit vector $\psi_{1i\ast}%
\overrightarrow{\mathbf{e}}_{1i\ast}$ or $\psi_{2i\ast}%
\overrightarrow{\mathbf{e}}_{2i\ast}$ is a displacement vector that is
correlated with an extreme vector $\left(  \mathbf{x}^{T}\mathbf{x}_{1_{i\ast
}}+1\right)  ^{2}$ or $\left(  \mathbf{x}^{T}\mathbf{x}_{2_{i\ast}}+1\right)
^{2}$ respectively.

For a given set%
\[
\left\{  \left\{  \left(  \mathbf{x}^{T}\mathbf{x}_{1_{i\ast}}+1\right)
^{2}\right\}  _{i=1}^{l_{1}},\;\left\{  \left(  \mathbf{x}^{T}\mathbf{x}%
_{2_{i\ast}}+1\right)  ^{2}\right\}  _{i=1}^{l_{2}}\right\}
\]
of $k_{\mathbf{x}_{1i\ast}}$ and $k_{\mathbf{x}_{2i\ast}}$ reproducing
kernels, I\ will show that each Wolfe dual principal eigenaxis component
$\psi_{i\ast}\overrightarrow{\mathbf{e}}_{i\ast}$ on $\boldsymbol{\psi}$
specifies a class-conditional density $p\left(  k_{\mathbf{x}_{i\ast}%
}|\operatorname{comp}_{\overrightarrow{k_{\mathbf{x}_{i\ast}}}}\left(
\overrightarrow{\boldsymbol{\kappa}}\right)  \right)  $ for a correlated
reproducing kernel $k_{\mathbf{x}_{i\ast}}$ of an extreme point $\mathbf{x}%
_{i\ast}$, such that $\boldsymbol{\psi}_{1}$ and $\boldsymbol{\psi}_{2}$ are
class-conditional probability density functions in Wolfe dual eigenspace,
$\boldsymbol{\kappa}_{1}$ is a parameter vector for a class-conditional
probability density $p\left(  k_{\mathbf{x}_{1i\ast}}|\boldsymbol{\kappa}%
_{1}\right)  $ for a given set $\left\{  \left(  \mathbf{x}^{T}\mathbf{x}%
_{1_{i\ast}}+1\right)  ^{2}\right\}  _{i=1}^{l_{1}}$ of $k_{\mathbf{x}%
_{1i\ast}}$ reproducing kernels:%
\[
\boldsymbol{\kappa}_{1}=p\left(  \left(  \mathbf{x}^{T}\mathbf{x}_{1_{i\ast}%
}+1\right)  ^{2}|\boldsymbol{\kappa}_{1}\right)  \text{,}%
\]
and $\boldsymbol{\kappa}_{2}$ is a parameter vector for a class-conditional
probability density $p\left(  k_{\mathbf{x}_{2i\ast}}|\boldsymbol{\kappa}%
_{2}\right)  $ for a given set $\left\{  \left(  \mathbf{x}^{T}\mathbf{x}%
_{2_{i\ast}}+1\right)  ^{2}\right\}  _{i=1}^{l_{2}}$ of $k_{\mathbf{x}%
_{2i\ast}}$ reproducing kernels:%
\[
\boldsymbol{\kappa}_{2}=p\left(  \left(  \mathbf{x}^{T}\mathbf{x}_{2_{i\ast}%
}+1\right)  ^{2}|\boldsymbol{\kappa}_{2}\right)  \text{,}%
\]
where the area under each pointwise conditional density $p\left(
k_{\mathbf{x}_{i\ast}}|\operatorname{comp}_{\overrightarrow{k_{\mathbf{x}%
_{i\ast}}}}\left(  \overrightarrow{\boldsymbol{\kappa}}\right)  \right)  $ is
a conditional probability that an extreme point $k_{\mathbf{x}_{i\ast}}$ will
be observed in a $Z_{1}$ or $Z_{2}$ decision region of a decision space $Z$.

In order to develop class-conditional probability densities for extreme
points, I will devise a system of data-driven, locus equations in Wolfe dual
eigenspace that provides tractable point and coordinate relationships between
the weighted (labeled and scaled) extreme points on $\boldsymbol{\kappa}_{1}$
and $\boldsymbol{\kappa}_{2}$ and the Wolfe dual principal eigenaxis
components on $\boldsymbol{\psi}_{1}$ and $\boldsymbol{\psi}_{2}$. I\ will use
this system of equations to develop equations for geometric and statistical
properties possessed by the Wolfe dual and the constrained, primal principal
eigenaxis components. Next, I\ will use these equations and identified
properties to define class-conditional probability densities for individual
extreme points and class-conditional probability densities $p\left(  \left(
\mathbf{x}^{T}\mathbf{x}_{1_{i\ast}}+1\right)  ^{2}|\boldsymbol{\kappa}%
_{1}\right)  $ and $p\left(  \left(  \mathbf{x}^{T}\mathbf{x}_{2_{i\ast}%
}+1\right)  ^{2}|\boldsymbol{\kappa}_{2}\right)  $ for labeled sets of extreme points.

Thereby, I will demonstrate that the conditional probability function
$P\left(  k_{\mathbf{x}_{1i\ast}}|\boldsymbol{\kappa}_{1}\right)  $ for a
given set $\left\{  \left(  \mathbf{x}^{T}\mathbf{x}_{1_{i\ast}}+1\right)
^{2}\right\}  _{i=1}^{l_{1}}$ of $k_{\mathbf{x}_{1i\ast}}$ reproducing kernels
is given by the area under the class-conditional probability density function
$p\left(  k_{\mathbf{x}_{1i\ast}}|\boldsymbol{\kappa}_{1}\right)  $:%
\begin{align*}
P\left(  k_{\mathbf{x}_{1i\ast}}|\boldsymbol{\kappa}_{1}\right)   &  =\int%
_{Z}\left(  \sum\nolimits_{i=1}^{l_{1}}\psi_{1_{i\ast}}\left(  \mathbf{x}%
^{T}\mathbf{x}_{1_{i\ast}}+1\right)  ^{2}\right)  d\boldsymbol{\kappa}_{1}\\
&  =\int_{Z}p\left(  k_{\mathbf{x}_{1i\ast}}|\boldsymbol{\kappa}_{1}\right)
d\boldsymbol{\kappa}_{1}\\
&  =\int_{Z}\boldsymbol{\kappa}_{1}d\boldsymbol{\kappa}_{1}=\left\Vert
\boldsymbol{\kappa}_{1}\right\Vert ^{2}+C_{1}\text{,}%
\end{align*}
over the decision space $Z$, where $\left\Vert \boldsymbol{\kappa}%
_{1}\right\Vert ^{2}$ is the total allowed eigenenergy exhibited by
$\boldsymbol{\kappa}_{1}$ and $C_{1}$ is an integration constant.

Likewise, I will demonstrate that the conditional probability function
$P\left(  k_{\mathbf{x}_{2i\ast}}|\boldsymbol{\kappa}_{2}\right)  $ for a
given set $\left\{  \left(  \mathbf{x}^{T}\mathbf{x}_{2_{i\ast}}+1\right)
^{2}\right\}  _{i=l_{1}+1}^{l_{2}}$ of $k_{\mathbf{x}_{2i\ast}}$ reproducing
kernels is given by the area under the class-conditional probability density
function $p\left(  k_{\mathbf{x}_{2i\ast}}|\boldsymbol{\kappa}_{2}\right)  $:%
\begin{align*}
P\left(  k_{\mathbf{x}_{2i\ast}}|\boldsymbol{\kappa}_{2}\right)   &  =\int%
_{Z}\left(  \sum\nolimits_{i=1}^{l_{2}}\psi_{2_{i\ast}}\left(  \mathbf{x}%
^{T}\mathbf{x}_{2_{i\ast}}+1\right)  ^{2}\right)  d\boldsymbol{\kappa}_{2}\\
&  =\int_{Z}p\left(  k_{\mathbf{x}_{2i\ast}}|\boldsymbol{\kappa}_{2}\right)
d\boldsymbol{\kappa}_{2}\\
&  =\int_{Z}\boldsymbol{\kappa}_{2}d\boldsymbol{\kappa}_{2}=\left\Vert
\boldsymbol{\kappa}_{2}\right\Vert ^{2}+C_{2}\text{,}%
\end{align*}
over the decision space $Z$, where $\left\Vert \boldsymbol{\kappa}%
_{2}\right\Vert ^{2}$ is the total allowed eigenenergy exhibited by
$\boldsymbol{\kappa}_{2}$ and $C_{2}$ is an integration constant.

In order to define the $C_{1}$ and $C_{2}$ integration constants, I\ will need
to define the manner in which the total allowed eigenenergies possessed by the
scaled extreme vectors on $\boldsymbol{\kappa}_{1}$ and $\boldsymbol{\kappa
}_{2}$ are symmetrically balanced with each other. I\ will use these results
to define the manner in which the area under the class-conditional probability
density functions $p\left(  \left(  \mathbf{x}^{T}\mathbf{x}_{1_{i\ast}%
}+1\right)  ^{2}|\boldsymbol{\kappa}_{1}\right)  $ and $p\left(  \left(
\mathbf{x}^{T}\mathbf{x}_{2_{i\ast}}+1\right)  ^{2}|\boldsymbol{\kappa}%
_{2}\right)  $ and the corresponding conditional probability functions
$P\left(  k_{\mathbf{x}_{1i\ast}}|\boldsymbol{\kappa}_{1}\right)  $ and
$P\left(  k_{\mathbf{x}_{2i\ast}}|\boldsymbol{\kappa}_{2}\right)  $ for class
$\omega_{1}$ and class $\omega_{2}$ are symmetrically balanced with each other.

I\ will define the $C_{1}$ and $C_{2}$ integration constants in the following
manner: I will use the KKT condition in Eq. (\ref{KKTE5 Q}) and the theorem of
Karush, Kuhn, and Tucker to devise a system of data-driven, locus equations
that determines the manner in which the total allowed eigenenergies of the
scaled extreme vectors on $\boldsymbol{\kappa}_{1}$ and $\boldsymbol{\kappa
}_{2}$ are symmetrically balanced with each other. I\ will use these results
along with results obtained from the analysis of the Wolfe dual eigenspace to
devise a system of data-driven, locus equations that determines the manner in
which the class-conditional density functions $p\left(  \left(  \mathbf{x}%
^{T}\mathbf{x}_{1_{i\ast}}+1\right)  ^{2}|\boldsymbol{\kappa}_{1}\right)  $
and $p\left(  \left(  \mathbf{x}^{T}\mathbf{x}_{2_{i\ast}}+1\right)
^{2}|\boldsymbol{\kappa}_{2}\right)  $ satisfy an integral equation%
\[
f\left(  \widetilde{\Lambda}_{\boldsymbol{\kappa}}\left(  \mathbf{s}\right)
\right)  :\int_{Z}p\left(  k_{\mathbf{x}_{1i\ast}}|\boldsymbol{\kappa}%
_{1}\right)  d\boldsymbol{\kappa}_{1}+\nabla_{eq}=\int_{Z}p\left(
k_{\mathbf{x}_{2i\ast}}|\boldsymbol{\kappa}_{2}\right)  d\boldsymbol{\kappa
}_{2}-\nabla_{eq}\text{,}%
\]
over the decision space $Z$, where $\nabla_{eq}$ is an equalizer statistic.

Thereby, I\ will demonstrate that the statistical property of symmetrical
balance exhibited by the principal eigenaxis components on $\boldsymbol{\psi}$
and $\boldsymbol{\kappa}$ ensures that the conditional probability functions
$P\left(  k_{\mathbf{x}_{1i\ast}}|\boldsymbol{\kappa}_{1}\right)  $ and
$P\left(  k_{\mathbf{x}_{2i\ast}}|\boldsymbol{\kappa}_{2}\right)  $ for class
$\omega_{1}$ and class $\omega_{2}$ are equal to each other, so that a
quadratic eigenlocus discriminant function $\widetilde{\Lambda}%
_{\boldsymbol{\kappa}}\left(  \mathbf{s}\right)  =\left(  \mathbf{x}%
^{T}\mathbf{s}+1\right)  ^{2}\boldsymbol{\kappa}+\kappa_{0}$ satisfies the
integral equation in Eq.
(\ref{Integral Equation of Likelihood Ratio and Decision Boundary}).

I will use these results along with results obtained from the analysis of the
Wolfe dual eigenspace to prove that quadratic eigenlocus discriminant
functions $\widetilde{\Lambda}_{\boldsymbol{\kappa}}\left(  \mathbf{s}\right)
=\left(  \mathbf{x}^{T}\mathbf{s}+1\right)  ^{2}\boldsymbol{\kappa}+\kappa
_{0}$ satisfy the fundamental integral equation of binary classification in
Eq. (\ref{Equalizer Rule}) along with the corresponding integral equation in
Eq. (\ref{Balancing of Bayes' Risks and Counteracting Risks}).

I\ will also devise an integral equation $f\left(  \widetilde{\Lambda
}_{\boldsymbol{\kappa}}\left(  \mathbf{s}\right)  \right)  $ that illustrates
the manner in which the property of symmetrical balance exhibited by the
principal eigenaxis components on $\boldsymbol{\psi}$ and $\boldsymbol{\kappa
}$ enables quadratic eigenlocus discriminant functions $\widetilde{\Lambda
}_{\boldsymbol{\kappa}}\left(  \mathbf{s}\right)  =\left(  \mathbf{x}%
^{T}\mathbf{s}+1\right)  ^{2}\boldsymbol{\kappa}+\kappa_{0}$ to effectively
balance all of the forces associated with the risk $\mathfrak{R}%
_{\mathfrak{B}}\left(  Z\mathbf{|}\boldsymbol{\kappa}\right)  $ of a quadratic
classification system $\left(  \mathbf{x}^{T}\mathbf{s}+1\right)
^{2}\boldsymbol{\kappa}+\kappa_{0}\overset{\omega_{1}}{\underset{\omega
_{2}}{\gtrless}}0$: where all of the forces associated with the counter risk
$\overline{\mathfrak{R}}_{\mathfrak{\min}}\left(  Z_{1}|\boldsymbol{\kappa
}_{1}\right)  $ and the risk $\mathfrak{R}_{\mathfrak{\min}}\left(
Z_{1}|\boldsymbol{\kappa}_{2}\right)  $ within the $Z_{1}$ decision region are
symmetrically balanced with all of the forces associated with the counter risk
$\overline{\mathfrak{R}}_{\mathfrak{\min}}\left(  Z_{2}|\boldsymbol{\kappa
}_{2}\right)  $ and the risk $\mathfrak{R}_{\mathfrak{\min}}\left(
Z_{2}|\boldsymbol{\kappa}_{1}\right)  $ within the $Z_{2}$ decision region.

Thereby, I will devise integral equations that are satisfied by quadratic
eigenlocus discriminant functions $\widetilde{\Lambda}_{\boldsymbol{\kappa}%
}\left(  \mathbf{s}\right)  =\left(  \mathbf{x}^{T}\mathbf{s}+1\right)
^{2}\boldsymbol{\kappa}+\kappa_{0}$, by which the discriminant function
$\widetilde{\Lambda}_{\boldsymbol{\kappa}}\left(  \mathbf{s}\right)  =\left(
\mathbf{x}^{T}\mathbf{s}+1\right)  ^{2}\boldsymbol{\kappa}+\kappa_{0}$ is the
solution to a discrete version of the fundamental integral equation of binary
classification for a classification system in statistical equilibrium:%
\begin{align*}
f\left(  \widetilde{\Lambda}_{\boldsymbol{\kappa}}\left(  \mathbf{s}\right)
\right)   &  =\;\int\nolimits_{Z_{1}}p\left(  k_{\mathbf{x}_{1i\ast}%
}|\boldsymbol{\kappa}_{1}\right)  d\boldsymbol{\kappa}_{1}+\int%
\nolimits_{Z_{2}}p\left(  k_{\mathbf{x}_{1i\ast}}|\boldsymbol{\kappa}%
_{1}\right)  d\boldsymbol{\kappa}_{1}+\nabla_{eq}\\
&  =\int\nolimits_{Z_{1}}p\left(  k_{\mathbf{x}_{2i\ast}}|\boldsymbol{\kappa
}_{2}\right)  d\boldsymbol{\kappa}_{2}+\int\nolimits_{Z_{2}}p\left(
k_{\mathbf{x}_{2i\ast}}|\boldsymbol{\kappa}_{2}\right)  d\boldsymbol{\kappa
}_{2}-\nabla_{eq}\text{,}%
\end{align*}
where all of the forces associated with counter risks and risks for class
$\omega_{1}$ and class $\omega_{2}$ are symmetrically balanced with each other%
\begin{align*}
f\left(  \widetilde{\Lambda}_{\boldsymbol{\kappa}}\left(  \mathbf{s}\right)
\right)   &  :\;\int\nolimits_{Z_{1}}p\left(  k_{\mathbf{x}_{1i\ast}%
}|\boldsymbol{\kappa}_{1}\right)  d\boldsymbol{\kappa}_{1}-\int%
\nolimits_{Z_{1}}p\left(  k_{\mathbf{x}_{2i\ast}}|\boldsymbol{\kappa}%
_{2}\right)  d\boldsymbol{\kappa}_{2}+\nabla_{eq}\\
&  =\int\nolimits_{Z_{2}}p\left(  k_{\mathbf{x}_{2i\ast}}|\boldsymbol{\kappa
}_{2}\right)  d\boldsymbol{\kappa}_{2}-\int\nolimits_{Z_{2}}p\left(
k_{\mathbf{x}_{1i\ast}}|\boldsymbol{\kappa}_{1}\right)  d\boldsymbol{\kappa
}_{1}-\nabla_{eq}\text{,}%
\end{align*}
over the $Z_{1}$ and $Z_{2}$ decision regions.

Quadratic eigenlocus transforms involve symmetrically balanced, first and
second-order statistical moments of reproducing kernels of extreme data
points. I\ will begin my analysis by defining first and second-order
statistical moments of reproducing kernels of data points.

\section{Statistical Moments II}

Consider again the matrix $\mathbf{Q}$ associated with the constrained
quadratic form in Eq. (\ref{Vector Form Wolfe Dual Q}):%
\begin{equation}
\mathbf{Q}=%
\begin{pmatrix}
\left(  \mathbf{x}_{1}^{T}\mathbf{x}_{1}+1\right)  ^{2} & \left(
\mathbf{x}_{1}^{T}\mathbf{x}_{2}+1\right)  ^{2} & \cdots & -\left(
\mathbf{x}_{1}^{T}\mathbf{x}_{N}+1\right)  ^{2}\\
\left(  \mathbf{x}_{2}^{T}\mathbf{x}_{1}+1\right)  ^{2} & \left(
\mathbf{x}_{2}^{T}\mathbf{x}_{2}+1\right)  ^{2} & \cdots & -\left(
\mathbf{x}_{2}^{T}\mathbf{x}_{N}+1\right)  ^{2}\\
\vdots & \vdots & \ddots & \vdots\\
-\left(  \mathbf{x}_{N}^{T}\mathbf{x}_{1}+1\right)  ^{2} & -\left(
\mathbf{x}_{N}^{T}\mathbf{x}_{2}+1\right)  ^{2} & \cdots & \left(
\mathbf{x}_{N}^{T}\mathbf{x}_{N}+1\right)  ^{2}%
\end{pmatrix}
\text{,} \label{Autocorrelation Matrix Q}%
\end{equation}
where $\mathbf{Q}\triangleq\widetilde{\mathbf{X}}\widetilde{\mathbf{X}}^{T}$,
$\widetilde{\mathbf{X}}\triangleq\mathbf{D}_{y}\mathbf{X}$, $\mathbf{D}_{y}$
is a $N\times N$ diagonal matrix of training labels $y_{i}$ and the $N\times
d$ reproducing kernel matrix is%
\[
\mathbf{X}=%
\begin{pmatrix}
\left(  \mathbf{x}^{T}\mathbf{x}_{1}+1\right)  ^{2}, & \left(  \mathbf{x}%
^{T}\mathbf{x}_{2}+1\right)  ^{2}, & \ldots, & \left(  \mathbf{x}%
^{T}\mathbf{x}_{N}+1\right)  ^{2}%
\end{pmatrix}
^{T}\text{.}%
\]

Without loss of generality (WLOG), assume that $N$ is an even number, the
first $N/2$ vectors have the training label $y_{i}=1$ and the last $N/2$
vectors have the training label $y_{i}=-1$. WLOG, the analysis that follows
does not take training label information into account.

Recall that reproducing kernels $k_{\mathbf{x}_{i}}=\left(  \mathbf{x}%
^{T}\mathbf{x}_{i}+1\right)  ^{2}$ replace straight lines of vectors with
second-order polynomial curves. Using Eq.
(\ref{Scalar Projection Reproducing Kernels}), let the inner product statistic
$\left(  \mathbf{x}_{i}^{T}\mathbf{x}_{j}+1\right)  ^{2}$ be interpreted as
$\left\Vert k_{\mathbf{x}_{i}}\right\Vert $ times the scalar projection
$\left\Vert k_{\mathbf{x}_{j}}\right\Vert \cos\theta_{k_{\mathbf{x_{i}}%
}k_{\mathbf{x}_{j}}}$ of $k_{\mathbf{x}_{j}}$ onto $k_{\mathbf{x}_{i}}$%
\begin{align*}
\left(  \mathbf{x}_{i}^{T}\mathbf{x}_{j}+1\right)  ^{2}  &  =\left(
\mathbf{x}^{T}\mathbf{x}_{i}+1\right)  ^{2}\left(  \mathbf{x}^{T}%
\mathbf{x}_{j}+1\right)  ^{2}\\
&  =\left\Vert k_{\mathbf{x}_{i}}\right\Vert \times\left[  \left\Vert
k_{\mathbf{x}_{j}}\right\Vert \cos\theta_{k_{\mathbf{x_{i}}}k_{\mathbf{x}_{j}%
}}\right]  \text{.}%
\end{align*}

It follows that row $\mathbf{Q}\left(  i,:\right)  $ in Eq.
(\ref{Autocorrelation Matrix Q}) contains uniformly weighted $\left\Vert
k_{\mathbf{x}_{i}}\right\Vert $ scalar projections $\left\Vert k_{\mathbf{x}%
_{j}}\right\Vert \cos\theta_{k_{\mathbf{x_{i}}}k_{\mathbf{x}_{j}}}$ for each
of the $N$ vectors $\left\{  \left(  \mathbf{x}^{T}\mathbf{x}_{j}+1\right)
^{2}\right\}  _{j=1}^{N}$ onto the vector $\left(  \mathbf{x}^{T}%
\mathbf{x}_{i}+1\right)  ^{2}$:%
\begin{equation}
\widetilde{\mathbf{Q}}=%
\begin{pmatrix}
\left\Vert f\left(  \mathbf{x}_{1}\right)  \right\Vert \left\Vert f\left(
\mathbf{x}_{1}\right)  \right\Vert \cos\theta_{k_{\mathbf{x_{1}}}%
k_{\mathbf{x}_{1}}} & \cdots & -\left\Vert f\left(  \mathbf{x}_{1}\right)
\right\Vert \left\Vert f\left(  \mathbf{x}_{N}\right)  \right\Vert \cos
\theta_{k_{\mathbf{x}_{1}}k_{\mathbf{x}_{N}}}\\
\left\Vert f\left(  \mathbf{x}_{2}\right)  \right\Vert \left\Vert f\left(
\mathbf{x}_{1}\right)  \right\Vert \cos\theta_{k_{\mathbf{x}_{2}}%
k_{\mathbf{x}_{1}}} & \cdots & -\left\Vert f\left(  \mathbf{x}_{2}\right)
\right\Vert \left\Vert f\left(  \mathbf{x}_{N}\right)  \right\Vert \cos
\theta_{k_{\mathbf{x}_{2}}k_{\mathbf{x}_{N}}}\\
\vdots & \ddots & \vdots\\
-\left\Vert f\left(  \mathbf{x}_{N}\right)  \right\Vert \left\Vert f\left(
\mathbf{x}_{1}\right)  \right\Vert \cos\theta_{k_{\mathbf{x}_{N}}%
k_{\mathbf{x}_{1}}} & \cdots & \left\Vert f\left(  \mathbf{x}_{N}\right)
\right\Vert \left\Vert f\left(  \mathbf{x}_{N}\right)  \right\Vert \cos
\theta_{k_{\mathbf{x}_{N}}k_{\mathbf{x}_{N}}}%
\end{pmatrix}
\text{,} \label{Inner Product Matrix Q}%
\end{equation}
where $f\left(  \mathbf{x}_{i}\right)  =$ $k_{\mathbf{x}_{i}}$ and
$0<\theta_{\mathbf{x}_{i}\mathbf{x}_{j}}\leq\frac{\pi}{2}$ or $\frac{\pi}%
{2}<\theta_{\mathbf{x}_{i}\mathbf{x}_{j}}\leq\pi$. Alternatively, column
$\mathbf{Q}\left(  :,j\right)  $ in Eq. (\ref{Autocorrelation Matrix Q})
contains weighted $\left\Vert k_{\mathbf{x}_{i}}\right\Vert $ scalar
projections $\left\Vert k_{\mathbf{x}_{j}}\right\Vert \cos\theta
_{k_{\mathbf{x_{i}}}k_{\mathbf{x}_{j}}}$ for the vector $\left(
\mathbf{x}^{T}\mathbf{x}_{j}+1\right)  ^{2}$ onto each of the $N$ vectors
$\left\{  \left(  \mathbf{x}^{T}\mathbf{x}_{i}+1\right)  ^{2}\right\}
_{i=1}^{N}$.

Now consider the $i$th row $\widetilde{\mathbf{Q}}\left(  i,:\right)  $ of
$\widetilde{\mathbf{Q}}$ in Eq. (\ref{Inner Product Matrix Q}). Again, using
Eq. (\ref{Scalar Projection Reproducing Kernels}), it follows that element
$\widetilde{\mathbf{Q}}\left(  i,j\right)  $ of row $\widetilde{\mathbf{Q}%
}\left(  i,:\right)  $ specifies the length $\left\Vert \left(  \mathbf{x}%
^{T}\mathbf{x}_{i}+1\right)  ^{2}\right\Vert $ of the vector $\left(
\mathbf{x}^{T}\mathbf{x}_{i}+1\right)  ^{2}$ multiplied by the scalar
projection $\left\Vert \left(  \mathbf{x}^{T}\mathbf{x}_{j}+1\right)
^{2}\right\Vert \cos\theta_{k_{\mathbf{x_{i}}}k_{\mathbf{x}_{j}}}$
of\ $\left(  \mathbf{x}^{T}\mathbf{x}_{j}+1\right)  ^{2}$ onto $\left(
\mathbf{x}^{T}\mathbf{x}_{i}+1\right)  ^{2}$:%
\[
\widetilde{\mathbf{Q}}\left(  i,j\right)  =\left\Vert \left(  \mathbf{x}%
^{T}\mathbf{x}_{i}+1\right)  ^{2}\right\Vert \left[  \left\Vert \left(
\mathbf{x}^{T}\mathbf{x}_{j}+1\right)  ^{2}\right\Vert \cos\theta
_{k_{\mathbf{x}_{i}}k_{\mathbf{x}_{j}}}\right]  \text{,}%
\]
where the signed magnitude of the vector projection of $\left(  \mathbf{x}%
^{T}\mathbf{x}_{j}+1\right)  ^{2}$ along the axis of $\left(  \mathbf{x}%
^{T}\mathbf{x}_{i}+1\right)  ^{2}$%
\begin{align*}
\operatorname{comp}_{\overrightarrow{k_{\mathbf{x}_{i}}}}\left(
\overrightarrow{k_{\mathbf{x}_{j}}}\right)   &  =\left\Vert \left(
\mathbf{x}^{T}\mathbf{x}_{j}+1\right)  ^{2}\right\Vert \cos\theta
_{k_{\mathbf{x}_{i}}k_{\mathbf{x}_{j}}}\text{,}\\
&  =\left(  \mathbf{x}^{T}\mathbf{x}_{j}+1\right)  ^{2}\left(  \frac{\left(
\mathbf{x}^{T}\mathbf{x}_{i}+1\right)  ^{2}}{\left\Vert \left(  \mathbf{x}%
^{T}\mathbf{x}_{i}+1\right)  ^{2}\right\Vert }\right)  \text{,}%
\end{align*}
provides a measure of the first and second degree components of the vector
$k_{\mathbf{x}_{j}}$%
\[
k_{\mathbf{x}_{j}}=%
\begin{pmatrix}
x_{j1}^{2}+2x_{j1}+1, & x_{j2}^{2}+2x_{j2}+1, & \cdots, & x_{jd}^{2}+2x_{jd}+1
\end{pmatrix}
^{T}%
\]
along the axis of the vector $k_{\mathbf{x}_{i}}$%
\[
k_{\mathbf{x}_{i}}=%
\begin{pmatrix}
x_{i1}^{2}+2x_{i1}+1, & x_{i2}^{2}+2x_{i2}+1, & \cdots, & x_{id}^{2}+2x_{id}+1
\end{pmatrix}
^{T}\text{.}%
\]

Accordingly, the signed magnitude $\left\Vert k_{\mathbf{x}_{j}}\right\Vert
\cos\theta_{k_{\mathbf{x}_{i}}k_{\mathbf{x}_{j}}}$ provides an estimate for
the amount of first degree and second degree components of the vector
$k_{\mathbf{x}_{i}}$ that are distributed over the axis of the vector
$k_{\mathbf{x}_{j}}$. This indicates that signed magnitudes $\left\Vert
k_{\mathbf{x}_{j}}\right\Vert \cos\theta_{k_{\mathbf{x}_{i}}k_{\mathbf{x}_{j}%
}}$ contained with $\widetilde{\mathbf{Q}}$ account for how the first degree
and second coordinates of a data point $k_{\mathbf{x}_{i}}$ are distributed
along the axes of a set of vectors $\left\{  k_{\mathbf{x}_{j}}\right\}
_{j=1}^{N}$ within Euclidean space.

Using the above assumptions and notation, for any given row
$\widetilde{\mathbf{Q}}\left(  i,:\right)  $ of Eq.
(\ref{Inner Product Matrix Q}), it follows that the statistic denoted by
$E_{k_{\mathbf{x}_{i}}}\left[  k_{\mathbf{x}_{i}}|\left\{  k_{\mathbf{x}_{j}%
}\right\}  _{j=1}^{N}\right]  $%
\begin{align}
E_{k_{\mathbf{x}_{i}}}\left[  k_{\mathbf{x}_{i}}|\left\{  k_{\mathbf{x}_{j}%
}\right\}  _{j=1}^{N}\right]   &  =\left\Vert k_{\mathbf{x}_{i}}\right\Vert
{\displaystyle\sum\nolimits_{j}}
\operatorname{comp}_{\overrightarrow{k_{\mathbf{x}_{i}}}}\left(
\overrightarrow{k_{\mathbf{x}_{j}}}\right)
\label{Row Distribution First Order Vector Coordinates Q}\\
&  =\left\Vert k_{\mathbf{x}_{i}}\right\Vert
{\displaystyle\sum\nolimits_{j}}
\left\Vert k_{\mathbf{x}_{j}}\right\Vert \cos\theta_{k_{\mathbf{x}_{i}%
}k_{\mathbf{x}_{j}}}\nonumber
\end{align}
provides an estimate $E_{k_{\mathbf{x}_{i}}}\left[  k_{\mathbf{x}_{i}%
}|\left\{  k_{\mathbf{x}_{j}}\right\}  _{j=1}^{N}\right]  $ for the amount of
first and second degree components of a vector $k_{\mathbf{x}_{i}}$ that are
distributed over the axes of a set of vectors $\left\{  k_{\mathbf{x}_{j}%
}\right\}  _{j=1}^{N}$, where labels have not been taken into account.

Thereby, Eq. (\ref{Row Distribution First Order Vector Coordinates Q})
describes a distribution of first and second degree coordinates for a vector
$k_{\mathbf{x}_{i}}$ in a data collection.

Given that Eq. (\ref{Row Distribution First Order Vector Coordinates Q})
involves signed magnitudes of vector projections along the axis of a fixed
vector $k_{\mathbf{x}_{i}}$, the distribution of first and second degree
vector coordinates described by Eq.
(\ref{Row Distribution First Order Vector Coordinates Q}) is said to determine
a \emph{first-order statistical moment about the locus of a reproducing
kernel} $k_{\mathbf{x}_{i}}$. Because the statistic $E_{k_{\mathbf{x}_{i}}%
}\left[  k_{\mathbf{x}_{i}}|\left\{  k_{\mathbf{x}_{j}}\right\}  _{j=1}%
^{N}\right]  $ depends on the uniform direction of a fixed vector
$k_{\mathbf{x}_{i}}$, the statistic $E_{k_{\mathbf{x}_{i}}}\left[
k_{\mathbf{x}_{i}}|\left\{  k_{\mathbf{x}_{j}}\right\}  _{j=1}^{N}\right]  $
is said to be unidirectional. In the next section, I will define pointwise
covariance statistics that provide unidirectional estimates of covariance
along a fixed reference axis.

\subsection{Unidirectional Covariance Statistics}

Recall that classical covariance statistics provide omnidirectional estimates
of covariance along $N$ axes of $N$ vectors. (See Section $13.1$). I will now
devise pointwise covariance statistics $\widehat{\operatorname{cov}}%
_{up}\left(  k_{\mathbf{x}_{i}}\right)  $ for individual vectors
$k_{\mathbf{x}_{i}}$. Pointwise covariance statistics for reproducing kernels
are unidirectional statistics that provide coherent estimates of covariance
along a fixed reference axis. WLOG, label information is not taken into consideration.

\subsubsection{Pointwise Covariance Statistics for Reproducing Kernels}

Take any row $\widetilde{\mathbf{Q}}\left(  i,:\right)  $ of the reproducing
kernel matrix $\widetilde{\mathbf{Q}}$ in Eq. (\ref{Inner Product Matrix Q})
and consider the inner product statistic $\left\Vert k_{\mathbf{x}_{i}%
}\right\Vert \left\Vert k_{\mathbf{x}_{j}}\right\Vert \cos\theta
_{k_{\mathbf{x}_{i}}k_{\mathbf{x}_{j}}}$ in element $\widetilde{\mathbf{Q}%
}\left(  i,j\right)  $. Using Eqs
(\ref{Geometric Locus of Second-order Reproducing Kernel}) and
(\ref{Inner Product Statistic Reproducing Kernels}), it follows that element
$\widetilde{\mathbf{Q}}\left(  i,j\right)  $ in row $\widetilde{\mathbf{Q}%
}\left(  i,:\right)  $ specifies the joint variations $\operatorname{cov}%
\left(  k_{\mathbf{x}_{i}},k_{\mathbf{x}_{j}}\right)  $%
\[
\operatorname{cov}\left(  k_{\mathbf{x}_{i}},k_{\mathbf{x}_{j}}\right)
=\left\Vert k_{\mathbf{x}_{i}}\right\Vert \left\Vert k_{\mathbf{x}_{j}%
}\right\Vert \cos\theta_{k_{\mathbf{x}_{i}}k_{\mathbf{x}_{j}}}%
\]
between the components of the vector $\left(  \mathbf{x}^{T}\mathbf{x}%
_{i}+1\right)  ^{2}$%
\[%
\begin{pmatrix}
\left(  \left\Vert \mathbf{x}_{i}\right\Vert ^{2}+1\right)  \cos
\mathbb{\alpha}_{k\left(  \mathbf{x}_{i1}\right)  1}, & \left(  \left\Vert
\mathbf{x}_{i}\right\Vert ^{2}+1\right)  \cos\mathbb{\alpha}_{k\left(
\mathbf{x}_{i2}\right)  2}, & \cdots, & \left(  \left\Vert \mathbf{x}%
_{i}\right\Vert ^{2}+1\right)  \cos\mathbb{\alpha}_{k\left(  \mathbf{x}%
_{id}\right)  d}%
\end{pmatrix}
\]
and the components of the vector $\left(  \mathbf{x}^{T}\mathbf{x}%
_{j}+1\right)  ^{2}$%
\[%
\begin{pmatrix}
\left(  \left\Vert \mathbf{x}_{j}\right\Vert ^{2}+1\right)  \cos
\mathbb{\alpha}_{k\left(  \mathbf{x}_{j1}\right)  1}, & \left(  \left\Vert
\mathbf{x}_{j}\right\Vert ^{2}+1\right)  \cos\mathbb{\alpha}_{k\left(
\mathbf{x}_{j2}\right)  2}, & \cdots, & \left(  \left\Vert \mathbf{x}%
_{j}\right\Vert ^{2}+1\right)  \cos\mathbb{\alpha}_{k\left(  \mathbf{x}%
_{jd}\right)  d}%
\end{pmatrix}
\text{,}%
\]
where the $d$ components $\left\{  \left(  \left\Vert \mathbf{s}\right\Vert
^{2}+1\right)  \cos\mathbb{\alpha}_{k\left(  s_{i}\right)  i}\right\}
_{i=1}^{d} $ of any given vector $\left(  \mathbf{x}^{T}\mathbf{s}+1\right)
^{2}$ are random variables, each of which is characterized by an expected
value and variance%
\[
E\left[  \left(  \left\Vert \mathbf{s}\right\Vert ^{2}+1\right)
\cos\mathbb{\alpha}_{k\left(  s_{i}\right)  i}\right]  \text{ and
}\operatorname{var}\left(  \left(  \left\Vert \mathbf{s}\right\Vert
^{2}+1\right)  \cos\mathbb{\alpha}_{k\left(  s_{i}\right)  i}\right)  \text{.}%
\]
It follows that the $j$th element $\widetilde{\mathbf{Q}}\left(  i,j\right)  $
of row $\widetilde{\mathbf{Q}}\left(  i,:\right)  $ specifies the joint
variations of the $d$ random variables of a vector $\left(  \mathbf{x}%
^{T}\mathbf{x}_{j}+1\right)  ^{2}$ about the $d$ random variables of the
vector $\left(  \mathbf{x}^{T}\mathbf{x}_{i}+1\right)  ^{2}$.

Thus, row $\widetilde{\mathbf{Q}}\left(  i,:\right)  $ specifies the joint
variations between the random variables of a fixed vector $\left(
\mathbf{x}^{T}\mathbf{x}_{i}+1\right)  ^{2}$ and the random variables of an
entire collection of vectors $\left(  \mathbf{x}^{T}\mathbf{x}_{j}+1\right)
^{2}$ of data.

Again, take any row $\widetilde{\mathbf{Q}}\left(  i,:\right)  $ of the matrix
$\widetilde{\mathbf{Q}}$ in Eq. (\ref{Inner Product Matrix Q}). Using Eq.
(\ref{Scalar Projection Reproducing Kernels}), it follows that the statistic
$\widehat{\operatorname{cov}}_{up}\left(  k_{\mathbf{x}_{i}}\right)  $:%
\begin{align}
\widehat{\operatorname{cov}}_{up}\left(  \left(  \mathbf{x}^{T}\mathbf{x}%
_{i}+1\right)  ^{2}\right)   &  =%
{\displaystyle\sum\nolimits_{j=1}^{N}}
\left\Vert \left(  \mathbf{x}^{T}\mathbf{x}_{i}+1\right)  ^{2}\right\Vert
\left\Vert \left(  \mathbf{x}^{T}\mathbf{x}_{j}+1\right)  ^{2}\right\Vert
\cos\theta_{k_{\mathbf{x_{i}}}k_{\mathbf{x}_{j}}}%
\label{Pointwise Covariance Statistic Q}\\
&  =%
{\displaystyle\sum\nolimits_{j=1}^{N}}
\left(  \mathbf{x}^{T}\mathbf{x}_{i}+1\right)  ^{2}\left(  \mathbf{x}%
^{T}\mathbf{x}_{j}+1\right)  ^{2}\nonumber\\
&  =\left(  \mathbf{x}^{T}\mathbf{x}_{i}+1\right)  ^{2}%
{\displaystyle\sum\nolimits_{j=1}^{N}}
\left(  \mathbf{x}^{T}\mathbf{x}_{j}+1\right)  ^{2}\nonumber\\
&  =\left\Vert \left(  \mathbf{x}^{T}\mathbf{x}_{i}+1\right)  ^{2}\right\Vert
{\displaystyle\sum\nolimits_{j=1}^{N}}
\left\Vert \left(  \mathbf{x}^{T}\mathbf{x}_{j}+1\right)  ^{2}\right\Vert
\cos\theta_{k_{\mathbf{x_{i}}}k_{\mathbf{x}_{j}}}\nonumber
\end{align}
provides a unidirectional estimate of the joint variations of the $d$ random
variables of each of the $N$ vectors of a training data collection $\left\{
k_{\mathbf{x}_{j}}\right\}  _{j=1}^{N}$ and a unidirectional estimate of the
joint variations of the $d$ random variables of the common mean $%
{\displaystyle\sum\nolimits_{j=1}^{N}}
k_{\mathbf{x}_{j}}$ of the $N$ vectors, about the $d$ random variables of a
fixed vector $k_{\mathbf{x}_{i}}$, along the axis of the fixed vector
$k_{\mathbf{x}_{i}}=\left(  \mathbf{x}^{T}\mathbf{x}_{i}+1\right)  ^{2}$.

Thereby, the statistic $\widehat{\operatorname{cov}}_{up}\left(
k_{\mathbf{x}_{i}}\right)  $ specifies the direction of the vector
$k_{\mathbf{x}_{i}}=\left(  \mathbf{x}^{T}\mathbf{x}_{i}+1\right)  ^{2}$ and a
signed magnitude along the axis of the vector $k_{\mathbf{x}_{i}}$.

The statistic $\widehat{\operatorname{cov}}_{up}\left(  \left(  \mathbf{x}%
^{T}\mathbf{x}_{i}+1\right)  ^{2}\right)  $ in Eq.
(\ref{Pointwise Covariance Statistic Q}) is defined to be a pointwise
covariance estimate for a reproducing kernel $\left(  \mathbf{x}^{T}%
\mathbf{x}_{i}+1\right)  ^{2}$, where the statistic
$\widehat{\operatorname{cov}}_{up}\left(  \left(  \mathbf{x}^{T}\mathbf{x}%
_{i}+1\right)  ^{2}\right)  $ provides a unidirectional estimate of the joint
variations between the random variables of each training vector $k_{\mathbf{x}%
_{j}}$ in a training data collection and the random variables of a fixed
vector $k_{\mathbf{x}_{i}}$ and a unidirectional estimate of the joint
variations between the random variables of the mean vector $%
{\displaystyle\sum\nolimits_{j=1}^{N}}
k_{\mathbf{x}_{j}}$ and the fixed vector $k_{\mathbf{x}_{i}}$. The statistic
$\widehat{\operatorname{cov}}_{up}\left(  \left(  \mathbf{x}^{T}\mathbf{x}%
_{i}+1\right)  ^{2}\right)  $ also accounts for first and second degree vector components.

Given that the joint variations estimated by the statistic
$\widehat{\operatorname{cov}}_{up}\left(  k_{\mathbf{x}_{i}}\right)  $ are
derived from second-order distance statistics $\left\Vert k_{\mathbf{x}_{i}%
}-k_{\mathbf{x}_{j}}\right\Vert ^{2}$ which involve signed magnitudes of
vector projections along the common axis of a fixed vector $k_{\mathbf{x}_{i}%
}$, a pointwise covariance estimate $\widehat{\operatorname{cov}}_{up}\left(
k_{\mathbf{x}_{i}}\right)  $ is said to determine a \emph{second-order
statistical moment about the locus of a reproducing kernel} $k_{\mathbf{x}%
_{i}}$. Using Eq. (\ref{Row Distribution First Order Vector Coordinates Q}),
Eq. (\ref{Pointwise Covariance Statistic Q}) also specifies a distribution of
first and second degree coordinates for a given reproducing kernel
$k_{\mathbf{x}_{i}}$ which determines a first-order statistical moment about
the locus of the data point $k_{\mathbf{x}_{i}}$.

I\ will now demonstrate that pointwise covariance statistics can be used to
discover reproducing kernels of extreme points.

\subsection{Discovery of Extreme Reproducing Kernels}

The kernel matrix associated with the constrained quadratic form in Eq.
(\ref{Vector Form Wolfe Dual Q}) contains inner product statistics for\ two
labeled collections of data. Denote those data points that belong to class
$\omega_{1}$ by $k_{\mathbf{x}_{1_{i}}}$ and those that belong to class
$\omega_{2}$ by $k_{\mathbf{x}_{2_{i}}}$. Let $\overline{k}_{\mathbf{x}_{1}}$
and $\overline{k}_{\mathbf{x}_{2}}$ denote the mean vectors of class
$\omega_{1}$ and class $\omega_{2}$. Let $i=1:n_{1}$ where the vector
$k_{\mathbf{x}_{1_{i}}}$ has the label $y_{i}=1$ and let $i=n_{1}%
+1:n_{1}+n_{2}$ where the vector $k_{\mathbf{x}_{2_{i}}}$ has the label
$y_{i}=-1$. Using label information, Eq.
(\ref{Pointwise Covariance Statistic Q}) can be rewritten as%
\[
\widehat{\operatorname{cov}}_{up}\left(  k_{\mathbf{x}_{1_{i}}}\right)
=k_{\mathbf{x}_{1_{i}}}\left(  \sum\nolimits_{j=1}^{n_{1}}k_{\mathbf{x}%
_{1_{j}}}-\sum\nolimits_{j=n_{1}+1}^{n_{1}+n_{2}}k_{\mathbf{x}_{2_{j}}%
}\right)
\]
and%
\[
\widehat{\operatorname{cov}}_{up}\left(  k_{\mathbf{x}_{2_{i}}}\right)
=k_{\mathbf{x}_{2_{i}}}\left(  \sum\nolimits_{j=n_{1}+1}^{n_{1}+n_{2}%
}k_{\mathbf{x}_{2_{j}}}-\sum\nolimits_{j=1}^{n_{1}}k_{\mathbf{x}_{1_{j}}%
}\right)  \text{.}%
\]

I will now show that extreme reproducing kernels possess large pointwise
covariances relative to the non-extreme reproducing kernels in each respective
pattern class. Recall that an extreme point is located relatively far from its
distribution mean, relatively close to the mean of the other distribution and
relatively close to other extreme points. Denote an extreme reproducing kernel
by $k_{\mathbf{x}_{1_{i\ast}}}$ or $k_{\mathbf{x}_{2_{i\ast}}}$ and a
non-extreme reproducing kernel by $k_{\mathbf{x}_{1_{i}}}$ or $k_{\mathbf{x}%
_{2_{i}}}$.

Take any extreme reproducing kernel $k_{\mathbf{x}_{1_{i\ast}}}$ and any
non-extreme reproducing kernel $k_{\mathbf{x}_{1_{i}}}$ that belong to class
$\omega_{1}$ and consider the pointwise covariance estimates for
$k_{\mathbf{x}_{1_{i\ast}}}$:%
\[
\widehat{\operatorname{cov}}_{up}\left(  k_{\mathbf{x}_{1_{i\ast}}}\right)
=k\left(  \mathbf{x}_{1_{i\ast}},\overline{\mathbf{x}}_{1}\right)  -k\left(
\mathbf{x}_{1_{i\ast}},\overline{\mathbf{x}}_{2}\right)
\]
and for $k_{\mathbf{x}_{1_{i}}}$:%
\[
\widehat{\operatorname{cov}}_{up}\left(  k_{\mathbf{x}_{1_{i}}}\right)
=k\left(  \mathbf{x}_{1_{i}},\overline{\mathbf{x}}_{1}\right)  -k\left(
\mathbf{x}_{1_{i}},\overline{\mathbf{x}}_{2}\right)  \text{.}%
\]

Because $k_{\mathbf{x}_{1_{i\ast}}}$ is an extreme reproducing kernel, it
follows that $k\left(  \mathbf{x}_{1_{i\ast}},\overline{\mathbf{x}}%
_{1}\right)  >$ $k\left(  \mathbf{x}_{1_{i}},\overline{\mathbf{x}}_{1}\right)
$ and that $k\left(  \mathbf{x}_{1_{i\ast}},\overline{\mathbf{x}}_{2}\right)
<$ $k\left(  \mathbf{x}_{1_{i}},\overline{\mathbf{x}}_{2}\right)  $. Thus,
$\widehat{\operatorname{cov}}_{up}\left(  k_{\mathbf{x}_{1_{i\ast}}}\right)
>\widehat{\operatorname{cov}}_{up}\left(  k_{\mathbf{x}_{1_{i}}}\right)  $.
Therefore, each extreme reproducing kernel $k_{\mathbf{x}_{1_{i\ast}}}$
exhibits a pointwise covariance $\widehat{\operatorname{cov}}_{up}\left(
k_{\mathbf{x}_{1_{i\ast}}}\right)  $ that exceeds the pointwise covariance
$\widehat{\operatorname{cov}}_{up}\left(  k_{\mathbf{x}_{1_{i}}}\right)  $ of
all of the non-extreme reproducing kernels $k_{\mathbf{x}_{1_{i}}}$ in class
$\omega_{1}$.

Now take any extreme reproducing kernel $k_{\mathbf{x}_{2_{i\ast}}}$ and any
non-extreme reproducing kernel $k_{\mathbf{x}_{2_{i}}}$ that belong to class
$\omega_{2}$ and consider the pointwise covariance estimates for
$k_{\mathbf{x}_{2_{i\ast}}}$:%
\[
\widehat{\operatorname{cov}}_{up}\left(  k_{\mathbf{x}_{2_{i\ast}}}\right)
=k\left(  \mathbf{x}_{2_{i\ast}},\overline{\mathbf{x}}_{2}\right)  -k\left(
\mathbf{x}_{2_{i\ast}},\overline{\mathbf{x}}_{1}\right)
\]
and for $k_{\mathbf{x}_{2_{i}}}$:%
\[
\widehat{\operatorname{cov}}_{up}\left(  k_{\mathbf{x}_{2_{i}}}\right)
=k\left(  \mathbf{x}_{2_{i}},\overline{\mathbf{x}}_{2}\right)  -k\left(
\mathbf{x}_{2_{i}},\overline{\mathbf{x}}_{1}\right)  \text{.}%
\]

Because $k_{\mathbf{x}_{2_{i\ast}}}$ is an extreme reproducing kernel, it
follows that $k\left(  \mathbf{x}_{2_{i\ast}},\overline{\mathbf{x}}%
_{2}\right)  >$ $k\left(  \mathbf{x}_{2_{i}},\overline{\mathbf{x}}_{2}\right)
$ and that $k\left(  \mathbf{x}_{2_{i\ast}},\overline{\mathbf{x}}_{1}\right)
<k\left(  \mathbf{x}_{2_{i}},\overline{\mathbf{x}}_{1}\right)  $. Thus,
$\widehat{\operatorname{cov}}_{up}\left(  k_{\mathbf{x}_{2_{i\ast}}}\right)
>\widehat{\operatorname{cov}}_{up}\left(  k_{\mathbf{x}_{2_{i}}}\right)  $.
Therefore, each extreme reproducing kernel $k_{\mathbf{x}_{2_{i\ast}}}$
exhibits a pointwise covariance $\widehat{\operatorname{cov}}_{up}\left(
k_{\mathbf{x}_{2_{i\ast}}}\right)  $ that exceeds the pointwise covariance
$\widehat{\operatorname{cov}}_{up}\left(  k_{\mathbf{x}_{2_{i}}}\right)  $ of
all of the non-extreme reproducing kernels $k_{\mathbf{x}_{2_{i}}}$ in class
$\omega_{2}$.

Thereby, it is concluded that extreme reproducing kernels possess large
pointwise covariances relative to non-extreme reproducing kernels in their
respective pattern class. It is also concluded that the pointwise covariance
$\widehat{\operatorname{cov}}_{up}\left(  k_{\mathbf{x}_{1_{i\ast}}}\right)  $
or $\widehat{\operatorname{cov}}_{up}\left(  k_{\mathbf{x}_{2_{i\ast}}%
}\right)  $ exhibited by any given extreme reproducing kernel $k_{\mathbf{x}%
_{1_{i\ast}}} $ or $k_{\mathbf{x}_{2_{i\ast}}}$ may exceed pointwise
covariances of other extreme reproducing kernels in each respective pattern class.

Therefore, it will be assumed that each extreme reproducing kernel
$k_{\mathbf{x}_{1_{i\ast}}}$ or $k_{\mathbf{x}_{2_{i\ast}}}$ exhibits a
critical first and second-order statistical moment
$\widehat{\operatorname{cov}}_{up}\left(  k_{\mathbf{x}_{1_{i\ast}}}\right)  $
or $\widehat{\operatorname{cov}}_{up}\left(  k_{\mathbf{x}_{2_{i\ast}}%
}\right)  $ that exceeds some threshold $\varrho$, for which each
corresponding scale factor $\psi_{1i\ast}$ or $\psi_{2i\ast}$ exhibits a
critical value that exceeds zero: $\psi_{1i\ast}>0$ or $\psi_{2i\ast}>0$.
Accordingly, first and second-order statistical moments
$\widehat{\operatorname{cov}}_{up}\left(  k_{\mathbf{x}_{1_{i}}}\right)  $ or
$\widehat{\operatorname{cov}}_{up}\left(  k_{\mathbf{x}_{2_{i}}}\right)  $
about the loci of non-extreme reproducing kernels $k_{\mathbf{x}_{1_{i}}}$ or
$k_{\mathbf{x}_{2_{i}}}$ do not exceed the threshold $\varrho$ and their
corresponding scale factors $\psi_{1i}$ or $\psi_{2i}$ are effectively zero:
$\psi_{1i}=0$ or $\psi_{2i}=0$.

I will now devise a system of equations for a principal eigen-decomposition of
the kernel matrix $\mathbf{Q}$ denoted in Eqs (\ref{Autocorrelation Matrix Q})
and (\ref{Inner Product Matrix Q}) that describes tractable point and
coordinate relationships between the scaled reproducing kernels on
$\boldsymbol{\kappa}_{1}$ and $\boldsymbol{\kappa}_{2}$ and the Wolfe dual
principal eigenaxis components on $\boldsymbol{\psi}_{1}$ and
$\boldsymbol{\psi}_{2}$.

\section{Inside the Wolfe Dual Eigenspace II}

Take the kernel matrix $\mathbf{Q}$ associated with the quadratic form in Eq
(\ref{Vector Form Wolfe Dual Q}). Let $\mathbf{q}_{j\text{ }}$denote the $j$th
column of $\mathbf{Q}$, which is an $N$-vector. Let $\lambda_{\max
_{\boldsymbol{\psi}}}$ and $\boldsymbol{\psi}$ denote the largest eigenvalue
and largest eigenvector of $\mathbf{Q}$ respectively. Using this notation
\citep[see][]{Trefethen1998}%
, the principal eigen-decomposition of $\mathbf{Q}$%
\[
\mathbf{Q}\boldsymbol{\psi}=\lambda\mathbf{_{\max_{\boldsymbol{\psi}}}%
}\boldsymbol{\psi}%
\]
can be rewritten as%
\[
\lambda_{\max_{\boldsymbol{\psi}}}\boldsymbol{\psi}=%
{\displaystyle\sum\nolimits_{j=1}^{N}}
\psi_{_{j}}\mathbf{q}_{j\text{ }}%
\]
so that the Wolfe dual principal eigenaxis $\boldsymbol{\psi}$ of $\mathbf{Q}$
is expressed as a linear combination of transformed vectors $\frac{\psi_{j}%
}{\lambda_{\max_{\boldsymbol{\psi}}}}\mathbf{q}_{j\text{ }}$:%
\begin{equation}
\left[
\begin{array}
[c]{c}%
\\
\boldsymbol{\psi}\\
\\
\end{array}
\right]  =\frac{\psi_{1}}{\lambda_{\max_{\boldsymbol{\psi}}}}\left[
\begin{array}
[c]{c}%
\\
\mathbf{q}_{1\text{ }}\\
\\
\end{array}
\right]  +\frac{\psi_{2}}{\lambda_{\max_{\boldsymbol{\psi}}}}\left[
\begin{array}
[c]{c}%
\\
\mathbf{q}_{2\text{ }}\\
\\
\end{array}
\right]  +\cdots+\frac{\psi_{N}}{\lambda_{\max_{\boldsymbol{\psi}}}}\left[
\begin{array}
[c]{c}%
\\
\mathbf{q}_{N\text{ }}\\
\\
\end{array}
\right]  \text{,} \label{Alternate Eigendecomposition Equation Q}%
\end{equation}
where the $i$th element of the vector $\mathbf{q}_{j\text{ }}$ specifies an
inner product statistic $K\left(  \mathbf{x}_{i},\mathbf{x}_{j}\right)  $
between the reproducing kernels $k_{\mathbf{x}_{i}}$ and $k_{\mathbf{x}_{j}}$.

Using Eqs (\ref{Autocorrelation Matrix Q}) and
(\ref{Alternate Eigendecomposition Equation Q}), a Wolfe dual quadratic
eigenlocus $\left(  \psi_{1},\cdots,\psi_{N}\right)  ^{T}$ can be written as:%
\begin{align}
\boldsymbol{\psi}  &  =\frac{\psi_{1}}{\lambda_{\max_{\boldsymbol{\psi}}}}%
\begin{pmatrix}
\left(  \mathbf{x}_{1}^{T}\mathbf{x}_{1}+1\right)  ^{2}\\
\left(  \mathbf{x}_{2}^{T}\mathbf{x}_{1}+1\right)  ^{2}\\
\vdots\\
-\left(  \mathbf{x}_{N}^{T}\mathbf{x}_{1}+1\right)  ^{2}%
\end{pmatrix}
+\frac{\psi_{2}}{\lambda_{\max_{\boldsymbol{\psi}}}}%
\begin{pmatrix}
\left(  \mathbf{x}_{1}^{T}\mathbf{x}_{2}+1\right)  ^{2}\\
\left(  \mathbf{x}_{2}^{T}\mathbf{x}_{2}+1\right)  ^{2}\\
\vdots\\
-\left(  \mathbf{x}_{N}^{T}\mathbf{x}_{2}+1\right)  ^{2}%
\end{pmatrix}
+\cdots\label{Dual Normal Eigenlocus Components Q}\\
\cdots &  +\frac{\psi_{N-1}}{\lambda_{\max_{\boldsymbol{\psi}}}}%
\begin{pmatrix}
-\left(  \mathbf{x}_{1}^{T}\mathbf{x}_{N-1}+1\right)  ^{2}\\
-\left(  \mathbf{x}_{2}^{T}\mathbf{x}_{N-1}+1\right)  ^{2}\\
\vdots\\
\left(  \mathbf{x}_{N}^{T}\mathbf{x}_{N-1}+1\right)  ^{2}%
\end{pmatrix}
+\frac{\psi_{N}}{\lambda_{\max_{\boldsymbol{\psi}}}}%
\begin{pmatrix}
-\left(  \mathbf{x}_{1}^{T}\mathbf{x}_{N}+1\right)  ^{2}\\
-\left(  \mathbf{x}_{2}^{T}\mathbf{x}_{N}+1\right)  ^{2}\\
\vdots\\
\left(  \mathbf{x}_{N}^{T}\mathbf{x}_{N}+1\right)  ^{2}%
\end{pmatrix}
\nonumber
\end{align}
which illustrates that the magnitude $\psi_{j}$ of the $j^{th}$ Wolfe dual
principal eigenaxis component $\psi_{j}\overrightarrow{\mathbf{e}}_{j}$ is
correlated with joint variations of labeled reproducing kernels about the
reproducing kernel $k_{\mathbf{x}_{j}}$.

Alternatively, using Eqs (\ref{Inner Product Matrix Q}) and
(\ref{Alternate Eigendecomposition Equation Q}), a Wolfe dual quadratic
eigenlocus $\boldsymbol{\psi}$ can be written as:%
\begin{align}
\boldsymbol{\psi}  &  =\frac{\psi_{1}}{\lambda_{\max_{\boldsymbol{\psi}}}%
}\left(
\begin{array}
[c]{c}%
\left\Vert \left(  \mathbf{x}^{T}\mathbf{x}_{1}+1\right)  ^{2}\right\Vert
\left\Vert \left(  \mathbf{x}^{T}\mathbf{x}_{1}+1\right)  ^{2}\right\Vert
\cos\theta_{k_{\mathbf{x_{1}}}k_{\mathbf{x}_{1}}}\\
\left\Vert \left(  \mathbf{x}^{T}\mathbf{x}_{2}+1\right)  ^{2}\right\Vert
\left\Vert \left(  \mathbf{x}^{T}\mathbf{x}_{1}+1\right)  ^{2}\right\Vert
\cos\theta_{k_{\mathbf{x}_{2}}k_{\mathbf{x}_{1}}}\\
\vdots\\
-\left\Vert \left(  \mathbf{x}^{T}\mathbf{x}_{N}+1\right)  ^{2}\right\Vert
\left\Vert \left(  \mathbf{x}^{T}\mathbf{x}_{1}+1\right)  ^{2}\right\Vert
\cos\theta_{k_{\mathbf{x}_{N}}k_{\mathbf{x}_{1}}}%
\end{array}
\right)  +\cdots\label{Dual Normal Eigenlocus Component Projections Q}\\
&  \cdots+\frac{\psi_{N}}{\lambda_{\max_{\boldsymbol{\psi}}}}\left(
\begin{array}
[c]{c}%
-\left\Vert \left(  \mathbf{x}^{T}\mathbf{x}_{1}+1\right)  ^{2}\right\Vert
\left\Vert \left(  \mathbf{x}^{T}\mathbf{x}_{N}+1\right)  ^{2}\right\Vert
\cos\theta_{k_{\mathbf{x}_{1}}k_{\mathbf{x}_{N}}}\\
-\left\Vert \left(  \mathbf{x}^{T}\mathbf{x}_{2}+1\right)  ^{2}\right\Vert
\left\Vert \left(  \mathbf{x}^{T}\mathbf{x}_{N}+1\right)  ^{2}\right\Vert
\cos\theta_{k_{\mathbf{x}_{2}}k_{\mathbf{x}_{N}}}\\
\vdots\\
\left\Vert \left(  \mathbf{x}^{T}\mathbf{x}_{N}+1\right)  ^{2}\right\Vert
\left\Vert \left(  \mathbf{x}^{T}\mathbf{x}_{N}+1\right)  ^{2}\right\Vert
\cos\theta_{k_{\mathbf{x}_{N}}k_{\mathbf{x}_{N}}}%
\end{array}
\right) \nonumber
\end{align}
which illustrates that the magnitude $\psi_{j}$ of the $j^{th}$ Wolfe dual
principal eigenaxis component $\psi_{j}\overrightarrow{\mathbf{e}}_{j}$ on
$\boldsymbol{\psi}$ is correlated with scalar projections $\left\Vert
k_{\mathbf{x}_{j}}\right\Vert \cos\theta_{k_{\mathbf{x}_{i}}k_{\mathbf{x}_{j}%
}}$ of the vector $k_{\mathbf{x}_{j}}$ onto labeled vectors $k_{\mathbf{x}%
_{i}}$.

Equations (\ref{Dual Normal Eigenlocus Components Q}) and
(\ref{Dual Normal Eigenlocus Component Projections Q}) both indicate that the
magnitude $\psi_{j}$ of the $j^{th}$ Wolfe dual principal eigenaxis component
$\psi_{j}\overrightarrow{\mathbf{e}}_{j}$ on $\boldsymbol{\psi}$ is correlated
with a first and second-order statistical moment about the locus of the
reproducing kernel $k_{\mathbf{x}_{j}}$.

\subsection{Assumptions}

It will be assumed that each extreme reproducing kernel $\left(
\mathbf{x}^{T}\mathbf{x}_{1_{i\ast}}+1\right)  ^{2}$ or $\left(
\mathbf{x}^{T}\mathbf{x}_{2_{i\ast}}+1\right)  ^{2}$ exhibits a critical first
and second-order statistical moment $\widehat{\operatorname{cov}}_{up}\left(
k_{\mathbf{x}_{1_{i_{\ast}}}}\right)  $ or $\widehat{\operatorname{cov}}%
_{up}\left(  k_{\mathbf{x}_{2_{i\ast}}}\right)  $ that exceeds some threshold
$\varrho$, for which each corresponding scale factor $\psi_{1i\ast}$ or
$\psi_{2i\ast}$ exhibits a critical value that exceeds zero: $\psi_{1i\ast}>0$
or $\psi_{2i\ast}>0$. It will also be assumed that first and second-order
statistical moments $\widehat{\operatorname{cov}}_{up}\left(  k_{\mathbf{x}%
_{1_{i}}}\right)  $ or $\widehat{\operatorname{cov}}_{up}\left(
k_{\mathbf{x}_{2_{i}}}\right)  $ about the loci of non-extreme reproducing
kernels $\left(  \mathbf{x}^{T}\mathbf{x}_{1_{i}}+1\right)  ^{2}$ or $\left(
\mathbf{x}^{T}\mathbf{x}_{2_{i}}+1\right)  ^{2}$ do not exceed the threshold
$\varrho$ and that the corresponding scale factors $\psi_{1i}$ or $\psi_{2i}$
are effectively zero: $\psi_{1i}=0$ or $\psi_{2i}=0$.

Express a Wolfe dual quadratic eigenlocus $\boldsymbol{\psi}$ in terms of $l$
non-orthogonal unit vectors $\left\{  \overrightarrow{\mathbf{e}}_{1\ast
},\ldots,\overrightarrow{\mathbf{e}}_{l\ast}\right\}  $%
\begin{align}
\boldsymbol{\psi}  &  =\sum\nolimits_{i=1}^{l}\psi_{i\ast}%
\overrightarrow{\mathbf{e}}_{i\ast}%
\label{Non-orthogonal Eigenaxes of Dual Normal Eigenlocus Q}\\
&  =\sum\nolimits_{i=1}^{l_{1}}\psi_{1i\ast}\overrightarrow{\mathbf{e}%
}_{1i\ast}+\sum\nolimits_{i=1}^{l_{2}}\psi_{2i\ast}\overrightarrow{\mathbf{e}%
}_{2i\ast}\text{,}\nonumber
\end{align}
where each scaled, non-orthogonal unit vector denoted by $\psi_{1i\ast
}\overrightarrow{\mathbf{e}}_{1i\ast}$ or $\psi_{2i\ast}%
\overrightarrow{\mathbf{e}}_{2i\ast}$ is correlated with an extreme vector
$\left(  \mathbf{x}^{T}\mathbf{x}_{1_{i\ast}}+1\right)  ^{2}$ or $\left(
\mathbf{x}^{T}\mathbf{x}_{2_{i\ast}}+1\right)  ^{2}$ respectively.
Accordingly, each Wolfe dual principal eigenaxis component $\psi_{1i\ast
}\overrightarrow{\mathbf{e}}_{1i\ast}$ or $\psi_{2i\ast}%
\overrightarrow{\mathbf{e}}_{2i\ast}$ is a scaled, non-orthogonal unit vector
that contributes to the estimation of $\boldsymbol{\psi}$ and
$\boldsymbol{\kappa}$.

WLOG, indices do not indicate locations of inner product expressions in Eq.
(\ref{Dual Normal Eigenlocus Component Projections Q}).

\subsubsection*{Notation}

Denote the extreme reproducing kernels $k_{\mathbf{x}_{1_{i\ast}}}$ or
$k_{\mathbf{x}_{2_{i\ast}}}$ that belong to class $\omega_{1}$ and $\omega
_{2}$ by $\left(  \mathbf{x}^{T}\mathbf{x}_{1_{i\ast}}+1\right)  ^{2}$ or
$\left(  \mathbf{x}^{T}\mathbf{x}_{2_{i\ast}}+1\right)  ^{2}$ with labels
$y_{i}=1$ and $y_{i}=-1$ respectively. Let there be $l_{1}$ extreme
reproducing kernels from class $\omega_{1}$ and $l_{2}$ extreme reproducing
kernels from class $\omega_{2}$.

Let there be $l_{1}$ principal eigenaxis components $\psi_{1i\ast
}\overrightarrow{\mathbf{e}}_{1i\ast}$, where each scale factor $\psi_{1i\ast
}$ is correlated with an extreme vector $\left(  \mathbf{x}^{T}\mathbf{x}%
_{1_{i\ast}}+1\right)  ^{2}$. Let there be $l_{2}$ principal eigenaxis
components $\psi_{2i\ast}\overrightarrow{\mathbf{e}}_{2i\ast}$, where each
scale factor $\psi_{2i\ast}$ is correlated with an extreme vector $\left(
\mathbf{x}^{T}\mathbf{x}_{2_{i\ast}}+1\right)  ^{2}$. Let $l_{1}+l_{2}=l$.

Recall that the risk $\mathfrak{R}_{\mathfrak{\min}}\left(  Z|p\left(
\widehat{\Lambda}\left(  \mathbf{x}\right)  \right)  \right)  $:%
\[
\mathfrak{R}_{\mathfrak{\min}}\left(  Z|p\left(  \widehat{\Lambda}\left(
\mathbf{x}\right)  \right)  \right)  =\mathfrak{R}_{\mathfrak{\min}}\left(
Z|p\left(  \widehat{\Lambda}\left(  \mathbf{x}\right)  |\omega_{1}\right)
\right)  +\mathfrak{R}_{\mathfrak{\min}}\left(  Z|p\left(  \widehat{\Lambda
}\left(  \mathbf{x}\right)  |\omega_{2}\right)  \right)
\]
for a binary classification system involves \emph{opposing forces} that depend
on the likelihood ratio test $\widehat{\Lambda}\left(  \mathbf{x}\right)
=p\left(  \widehat{\Lambda}\left(  \mathbf{x}\right)  |\omega_{1}\right)
-p\left(  \widehat{\Lambda}\left(  \mathbf{x}\right)  |\omega_{2}\right)
\overset{\omega_{1}}{\underset{\omega_{2}}{\gtrless}}0$ and the corresponding
decision boundary $p\left(  \widehat{\Lambda}\left(  \mathbf{x}\right)
|\omega_{1}\right)  -p\left(  \widehat{\Lambda}\left(  \mathbf{x}\right)
|\omega_{2}\right)  =0$.

In particular, the forces associated with the counter risk $\overline
{\mathfrak{R}}_{\mathfrak{\min}}\left(  Z_{1}|p\left(  \widehat{\Lambda
}\left(  \mathbf{x}\right)  |\omega_{1}\right)  \right)  $ in the $Z_{1}$
decision region and the risk $\mathfrak{R}_{\mathfrak{\min}}\left(
Z_{2}|p\left(  \widehat{\Lambda}\left(  \mathbf{x}\right)  |\omega_{1}\right)
\right)  $ in the $Z_{2}$ decision region are forces associated with positions
and potential locations of pattern vectors $\mathbf{x}$ that are generated
according to $p\left(  \mathbf{x}|\omega_{1}\right)  $, and the forces
associated with risk $\mathfrak{R}_{\mathfrak{\min}}\left(  Z_{1}|p\left(
\widehat{\Lambda}\left(  \mathbf{x}\right)  |\omega_{2}\right)  \right)  $ in
the $Z_{1}$ decision region and the counter risk $\overline{\mathfrak{R}%
}_{\mathfrak{\min}}\left(  Z_{2}|p\left(  \widehat{\Lambda}\left(
\mathbf{x}\right)  |\omega_{2}\right)  \right)  $ in th $Z_{2}$ decision
region are forces associated with positions and potential locations of pattern
vectors $\mathbf{x}$ that are generated according to $p\left(  \mathbf{x}%
|\omega_{2}\right)  $.

Quadratic eigenlocus transforms define the opposing forces of a classification
system in terms of forces associated with counter risks $\overline
{\mathfrak{R}}_{\mathfrak{\min}}\left(  Z_{1}|\psi_{1i\ast}k_{\mathbf{x}%
_{1_{i\ast}}}\right)  $ and $\overline{\mathfrak{R}}_{\mathfrak{\min}}\left(
Z_{2}|\psi_{2i\ast}k_{\mathbf{x}_{2_{i\ast}}}\right)  $ and risks
$\mathfrak{R}_{\mathfrak{\min}}\left(  Z_{1}|\psi_{2i\ast}k_{\mathbf{x}%
_{2_{i\ast}}}\right)  $ and $\mathfrak{R}_{\mathfrak{\min}}\left(  Z_{2}%
|\psi_{1i\ast}k_{\mathbf{x}_{1_{i\ast}}}\right)  $ related to scaled
reproducing kernels of extreme points $\psi_{1i\ast}k_{\mathbf{x}_{1_{i\ast}}%
}$ and $\psi_{2i\ast}k_{\mathbf{x}_{2_{i\ast}}}$: which are forces associated
with positions and potential locations of reproducing kernels of extreme
points $k_{\mathbf{x}_{1_{i\ast}}}$ and $k_{\mathbf{x}_{2_{i\ast}}}$ in the
$Z_{1}$ and $Z_{2}$ decision regions of a decision space $Z$.

In particular, the forces associated with the counter risks $\overline
{\mathfrak{R}}_{\mathfrak{\min}}\left(  Z_{1}|\psi_{1i\ast}k_{\mathbf{x}%
_{1_{i\ast}}}\right)  $ and the risks $\mathfrak{R}_{\mathfrak{\min}}\left(
Z_{2}|\psi_{1i\ast}k_{\mathbf{x}_{1_{i\ast}}}\right)  $ for class $\omega_{1}$
are determined by magnitudes and directions of scaled reproducing kernels
$\psi_{1i\ast}k_{\mathbf{x}_{1_{i\ast}}}$ on $\boldsymbol{\kappa}_{1}$, and
the forces associated with the counter risks $\overline{\mathfrak{R}%
}_{\mathfrak{\min}}\left(  Z_{2}|\psi_{2i\ast}k_{\mathbf{x}_{2_{i\ast}}%
}\right)  $ and the risks $\mathfrak{R}_{\mathfrak{\min}}\left(  Z_{1}%
|\psi_{2i\ast}k_{\mathbf{x}_{2_{i\ast}}}\right)  $ for class $\omega_{2}$ are
determined by magnitudes and directions of scaled reproducing kernels
$\psi_{2i\ast}k_{\mathbf{x}_{2_{i\ast}}}$ on $\boldsymbol{\kappa}_{2}$.

I\ will show that a Wolfe dual quadratic eigenlocus $\mathbf{\psi}$ is a
displacement vector that accounts for the magnitudes and the directions of all
of the scaled extreme vectors on $\boldsymbol{\kappa}_{1}-\boldsymbol{\kappa
}_{2}$. Quadratic eigenlocus transforms determine the opposing forces of a
classification system by means of symmetrically balanced, pointwise covariance statistics.

Symmetrically balanced, pointwise covariance statistics determine forces
associated with the counter risk $\overline{\mathfrak{R}}_{\mathfrak{\min}%
}\left(  Z_{1}|p\left(  \widehat{\Lambda}_{\boldsymbol{\kappa}}\left(
\mathbf{s}\right)  |\omega_{1}\right)  \right)  $ for class $\omega_{1}$ and
the risk $\mathfrak{R}_{\mathfrak{\min}}\left(  Z_{1}|p\left(
\widehat{\Lambda}_{\boldsymbol{\kappa}}\left(  \mathbf{s}\right)  |\omega
_{2}\right)  \right)  $ for class $\omega_{2}$ in the $Z_{1}$ decision region
that are balanced with forces associated with the counter risk $\overline
{\mathfrak{R}}_{\mathfrak{\min}}\left(  Z_{2}|p\left(  \widehat{\Lambda
}_{\boldsymbol{\kappa}}\left(  \mathbf{s}\right)  |\omega_{2}\right)  \right)
$ for class $\omega_{2}$ and the risk $\mathfrak{R}_{\mathfrak{\min}}\left(
Z_{2}|p\left(  \widehat{\Lambda}_{\boldsymbol{\kappa}}\left(  \mathbf{s}%
\right)  |\omega_{1}\right)  \right)  $ for class $\omega_{1}$ in the $Z_{2}$
decision region:%
\begin{align*}
f\left(  \widetilde{\Lambda}_{\boldsymbol{\tau}}\left(  \mathbf{x}\right)
\right)   &  :\overline{\mathfrak{R}}_{\mathfrak{\min}}\left(  Z_{1}|p\left(
\widehat{\Lambda}_{\boldsymbol{\kappa}}\left(  \mathbf{s}\right)  |\omega
_{1}\right)  \right)  -\mathfrak{R}_{\mathfrak{\min}}\left(  Z_{1}|p\left(
\widehat{\Lambda}_{\boldsymbol{\kappa}}\left(  \mathbf{s}\right)  |\omega
_{2}\right)  \right) \\
&  \rightleftharpoons\overline{\mathfrak{R}}_{\mathfrak{\min}}\left(
Z_{2}|p\left(  \widehat{\Lambda}_{\boldsymbol{\kappa}}\left(  \mathbf{s}%
\right)  |\omega_{2}\right)  \right)  -\mathfrak{R}_{\mathfrak{\min}}\left(
Z_{2}|p\left(  \widehat{\Lambda}_{\boldsymbol{\kappa}}\left(  \mathbf{s}%
\right)  |\omega_{1}\right)  \right)  \text{.}%
\end{align*}

I\ will now define symmetrically balanced, pointwise covariance statistics.
The geometric nature of the statistics is outlined first.

\subsection{Symmetrically Balanced Covariance Statistics II}

Take two labeled sets of extreme vectors, where each extreme vector is
correlated with a scale factor that determines scaled, signed magnitudes,
i.e., scaled components of the scaled extreme vector, along the axes of the
extreme vectors in each pattern class, such that the integrated scale factors
from each pattern class balance each other.

Generally speaking, for any given set of extreme vectors, all of the scaled,
signed magnitudes along the axis of any given extreme vector from a given
pattern class, which are determined by vector projections of scaled extreme
vectors from the \emph{other} pattern class, \emph{are distributed in opposite
directions}.

Thereby, for two labeled sets of extreme vectors, where each extreme vector is
correlated with a scale factor and the integrated scale factors from each
pattern class balance each other, it follows that scaled, signed magnitudes
along the axis of any given extreme vector, which are determined by vector
projections of scaled extreme vectors from the \emph{other} pattern class,
\emph{are distributed on the opposite side of the origin}.

Accordingly, scaled, signed magnitudes along the axes of all of the extreme
vectors are distributed in a symmetrically balanced manner, where each scale
factor specifies a symmetrically balanced distribution for an extreme point
which ensures that the \emph{components of} an extreme vector are
\emph{distributed over} the axes of a given \emph{collection} of extreme
vectors in a symmetrically balanced \emph{and} well-proportioned manner.

\begin{figure}[ptb]%
\centering
\fbox{\includegraphics[
height=2.5875in,
width=3.4411in
]%
{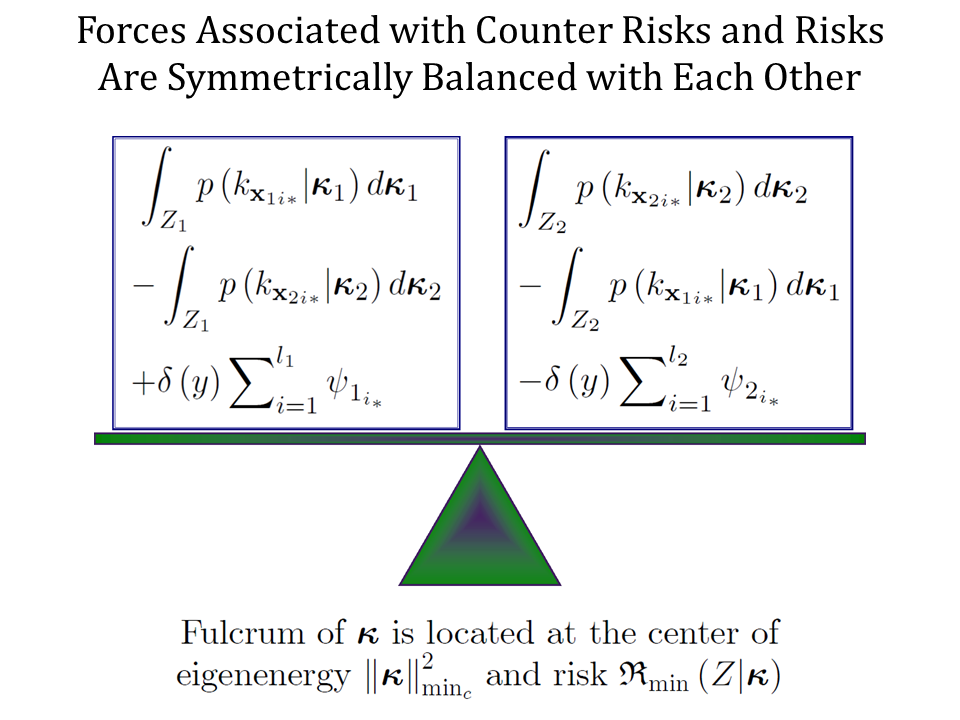}%
}\caption{Symmetrically balanced covariance statistics
$\protect\widehat{\operatorname{cov}}_{up_{\updownarrow}}\left(
k_{\mathbf{x}_{1_{i\ast}}}\right)  $ and $\protect\widehat{\operatorname{cov}%
}_{up_{\updownarrow}}\left(  k_{\mathbf{x}_{2_{i\ast}}}\right)  $ for extreme
points $\mathbf{x}_{1_{i_{\ast}}}$ and $\mathbf{x}_{2_{i\ast}}$ are the basis
of quadratic eigenlocus transforms. Note: $\delta\left(  y\right)
\triangleq\sum\nolimits_{i=1}^{l}y_{i}\left(  1-\xi_{i}\right)  $.}%
\label{Balancing Feat in Wolfe Dual Eigenspace Q}%
\end{figure}

I will show that symmetrically balanced covariance statistics are the basis of
quadratic eigenlocus transforms. For any given set of extreme points, I will
demonstrate that quadratic eigenlocus transforms find a set of scale factors
in Wolfe dual eigenspace, which are determined by the symmetrically balanced
covariance statistics in Eqs
(\ref{Eigen-balanced Pointwise Covariance Estimate Class One Q}) and
(\ref{Eigen-balanced Pointwise Covariance Estimate Class Two Q}), such that
symmetrical decision regions $Z_{1}\simeq Z_{1}$ are determined by
symmetrically balanced forces associated with counter risks and risks:%
\begin{align*}
\mathfrak{R}_{\mathfrak{\min}}\left(  Z:Z_{1}\simeq Z\right)   &
:\overline{\mathfrak{R}}_{\mathfrak{\min}}\left(  Z_{1}|\mathbf{\kappa}%
_{1}\right)  -\mathfrak{R}_{\mathfrak{\min}}\left(  Z_{1}|\mathbf{\kappa}%
_{2}\right) \\
&  \rightleftharpoons\overline{\mathfrak{R}}_{\mathfrak{\min}}\left(
Z_{2}|\mathbf{\kappa}_{2}\right)  -\mathfrak{R}_{\mathfrak{\min}}\left(
Z_{2}|\mathbf{\kappa}_{1}\right)  \text{.}%
\end{align*}
Figure $\ref{Balancing Feat in Wolfe Dual Eigenspace Q}$ illustrates that
symmetrically balanced covariance statistics determine quadratic discriminant
functions that satisfy a fundamental integral equation of binary
classification for a classification system in statistical equilibrium, where
the expected risk $\mathfrak{R}_{\mathfrak{\min}}\left(  Z\mathbf{|}%
\boldsymbol{\kappa}\right)  $ and the total allowed eigenenergy $\left\Vert
\boldsymbol{\kappa}\right\Vert _{\min_{c}}^{2}$ of the classification system
are minimized.%

Using Eqs (\ref{Equilibrium Constraint on Dual Eigen-components Q}) and
(\ref{Pointwise Covariance Statistic Q}), along with the notation and
assumptions outlined above, it follows that summation over the $l$ components
of $\boldsymbol{\psi}$ in Eq.
(\ref{Dual Normal Eigenlocus Component Projections Q}) provides symmetrically
balanced covariance statistics for the $\left(  \mathbf{x}^{T}\mathbf{x}%
_{1_{i_{\ast}}}+1\right)  ^{2}$ extreme vectors, where each extreme point
$k_{\mathbf{x}_{1_{i\ast}}}$ exhibits a symmetrically balanced, first and
second-order statistical moment $\widehat{\operatorname{cov}}%
_{up_{\updownarrow}}\left(  k_{\mathbf{x}_{1_{i\ast}}}\right)  $%
\begin{align}
\widehat{\operatorname{cov}}_{up_{\updownarrow}}\left(  \left(  \mathbf{x}%
^{T}\mathbf{x}_{_{1i\ast}}+1\right)  ^{2}\right)   &  =\left\Vert
k_{\mathbf{x}_{1_{i\ast}}}\right\Vert \sum\nolimits_{j=1}^{l_{1}}%
\psi_{1_{j\ast}}\left\Vert k_{\mathbf{x}_{1_{j\ast}}}\right\Vert \cos
\theta_{k_{\mathbf{x}_{1_{i\ast}}}k_{\mathbf{x}_{1_{j\ast}}}}%
\label{Eigen-balanced Pointwise Covariance Estimate Class One Q}\\
&  -\left\Vert k_{\mathbf{x}_{1_{i\ast}}}\right\Vert \sum\nolimits_{j=1}%
^{l_{2}}\psi_{2_{j\ast}}\left\Vert k_{\mathbf{x}_{2_{j\ast}}}\right\Vert
\cos\theta_{k_{\mathbf{x}_{1_{i\ast}}}k_{\mathbf{x}_{2_{j\ast}}}}\nonumber
\end{align}
relative to $l$ symmetrically balanced, scaled, signed magnitudes determined
by vector projections of scaled extreme vectors in each respective pattern class.

Likewise, summation over the $l$ components of $\boldsymbol{\psi}$ in Eq.
(\ref{Dual Normal Eigenlocus Component Projections Q}) provides symmetrically
balanced covariance statistics for the $\left(  \mathbf{x}^{T}\mathbf{x}%
_{2_{i_{\ast}}}+1\right)  ^{2}$ extreme vectors, where each extreme point
$k_{\mathbf{x}_{2_{i\ast}}}$ exhibits a a symmetrically balanced, first and
second-order statistical moment $\widehat{\operatorname{cov}}%
_{up_{\updownarrow}}\left(  k_{\mathbf{x}_{2_{i\ast}}}\right)  $%
\begin{align}
\widehat{\operatorname{cov}}_{up_{\updownarrow}}\left(  \left(  \mathbf{x}%
^{T}\mathbf{x}_{2_{i_{\ast}}}+1\right)  ^{2}\right)   &  =\left\Vert
k_{\mathbf{x}_{2_{i\ast}}}\right\Vert \sum\nolimits_{j=1}^{l_{2}}%
\psi_{2_{j\ast}}\left\Vert k_{\mathbf{x}_{2_{j\ast}}}\right\Vert \cos
\theta_{k_{\mathbf{x}_{2_{i\ast}}}k_{\mathbf{x}_{2_{j\ast}}}}%
\label{Eigen-balanced Pointwise Covariance Estimate Class Two Q}\\
&  -\left\Vert k_{\mathbf{x}_{2_{i\ast}}}\right\Vert \sum\nolimits_{j=1}%
^{l_{1}}\psi_{1_{j\ast}}\left\Vert k_{\mathbf{x}_{1_{j\ast}}}\right\Vert
\cos\theta_{k_{\mathbf{x}_{1_{i\ast}}}k_{\mathbf{x}_{2_{j\ast}}}}\nonumber
\end{align}
relative to $l$ symmetrically balanced, scaled, signed magnitudes determined
by vector projections of scaled extreme vectors in each respective pattern class.

\subsection{Common Geometrical and Statistical Properties}

I\ will now use Eqs
(\ref{Eigen-balanced Pointwise Covariance Estimate Class One Q}) and
(\ref{Eigen-balanced Pointwise Covariance Estimate Class Two Q}) to identify
symmetrical, geometric and statistical properties possessed by principal
eigenaxis components on $\boldsymbol{\kappa}$ and $\boldsymbol{\psi}$.

\subsection{Loci of the $\psi_{1i\ast}\protect\overrightarrow{\mathbf{e}%
}_{1i\ast}$ Components}

Let $i=1:l_{1}$, where each extreme vector $\left(  \mathbf{x}^{T}%
\mathbf{x}_{1_{i_{\ast}}}+1\right)  ^{2}$ is correlated with a Wolfe principal
eigenaxis component $\psi_{1i\ast}\overrightarrow{\mathbf{e}}_{1i\ast}$. Using
Eqs (\ref{Dual Normal Eigenlocus Component Projections Q}) and
(\ref{Non-orthogonal Eigenaxes of Dual Normal Eigenlocus Q}), it follows that
the locus of the $i^{th}$ principal eigenaxis component $\psi_{1i\ast
}\overrightarrow{\mathbf{e}}_{1i\ast}$ on $\boldsymbol{\psi}$ is a function of
the expression:%
\begin{align}
\psi_{1i\ast}  &  =\lambda_{\max_{\boldsymbol{\psi}}}^{-1}\left\Vert
k_{\mathbf{x}_{1_{i\ast}}}\right\Vert \sum\nolimits_{j=1}^{l_{1}}%
\psi_{1_{j\ast}}\left\Vert k_{\mathbf{x}_{1_{j\ast}}}\right\Vert \cos
\theta_{k_{\mathbf{x}_{1_{i\ast}}}k_{\mathbf{x}_{1_{j\ast}}}}%
\label{Dual Eigen-coordinate Locations Component One Q}\\
&  -\lambda_{\max_{\boldsymbol{\psi}}}^{-1}-\left\Vert k_{\mathbf{x}%
_{1_{i\ast}}}\right\Vert \sum\nolimits_{j=1}^{l_{2}}\psi_{2_{j\ast}}\left\Vert
k_{\mathbf{x}_{2_{j\ast}}}\right\Vert \cos\theta_{k_{\mathbf{x}_{1_{i\ast}}%
}k_{\mathbf{x}_{2_{j\ast}}}}\text{,}\nonumber
\end{align}
where $\psi_{1i\ast}$ provides a scale factor for the non-orthogonal unit
vector $\overrightarrow{\mathbf{e}}_{1i\ast}$. Geometric and statistical
explanations for the eigenlocus statistics%
\begin{equation}
\psi_{1_{j\ast}}\left\Vert k_{\mathbf{x}_{1_{j\ast}}}\right\Vert \cos
\theta_{k_{\mathbf{x}_{1_{i\ast}}}k_{\mathbf{x}_{1_{j\ast}}}}\text{ and }%
\psi_{2_{j\ast}}\left\Vert k_{\mathbf{x}_{2_{j\ast}}}\right\Vert \cos
\theta_{k_{\mathbf{x}_{1_{i\ast}}}k_{\mathbf{x}_{2_{j\ast}}}}
\label{Projection Statistics psi1 Q}%
\end{equation}
in Eq. (\ref{Dual Eigen-coordinate Locations Component One Q}) are considered next.

\subsubsection{Geometric Nature of Eigenlocus Statistics}

The first geometric interpretation of the eigenlocus statistics in Eq.
(\ref{Projection Statistics psi1 Q}) defines $\psi_{1_{j\ast}}$ and
$\psi_{2_{j\ast}}$ to be scale factors for the signed magnitudes of the vector
projections%
\[
\left\Vert \left(  \mathbf{x}^{T}\mathbf{x}_{1_{j_{\ast}}}+1\right)
^{2}\right\Vert \cos\theta_{k_{\mathbf{x}_{1_{i\ast}}}k_{\mathbf{x}_{1_{j\ast
}}}}\text{ and }\left\Vert \left(  \mathbf{x}^{T}\mathbf{x}_{2_{j_{\ast}}%
}+1\right)  ^{2}\right\Vert \cos\theta_{k_{\mathbf{x}_{1_{i\ast}}%
}k_{\mathbf{x}_{2_{j\ast}}}}%
\]
of the scaled extreme vectors $\psi_{1_{j\ast}}k_{\mathbf{x}_{1_{j\ast}}}$ and
$\psi_{2_{j\ast}}k_{\mathbf{x}_{2_{j\ast}}}$ along the axis of the extreme
vector $k_{\mathbf{x}_{1_{i\ast}}}$, where $\cos\theta_{k_{\mathbf{x}%
_{1_{i\ast}}}k_{\mathbf{x}_{1_{j\ast}}}}$ and $\cos\theta_{k_{\mathbf{x}%
_{1_{i\ast}}}k_{\mathbf{x}_{2_{j\ast}}}}$ specify the respective angles
between the axes of the scaled, extreme vectors $\psi_{1_{j\ast}}%
k_{\mathbf{x}_{1_{j\ast}}}$ and $\psi_{2_{j\ast}}k_{\mathbf{x}_{2_{j\ast}}}$
and the axis of the extreme vector $k_{\mathbf{x}_{1_{i\ast}}}$. Note that the
signed magnitude $-\left\Vert k_{\mathbf{x}_{2_{j\ast}}}\right\Vert \cos
\theta_{k_{\mathbf{x}_{1_{i\ast}}}k_{\mathbf{x}_{2_{j\ast}}}}$ is distributed
in the opposite direction, so that the locus of the signed magnitude is on the
opposite side of the origin, along the axis of the extreme vector
$k_{\mathbf{x}_{1_{i\ast}}}$.

\begin{figure}[ptb]%
\centering
\fbox{\includegraphics[
height=2.5875in,
width=3.4411in
]%
{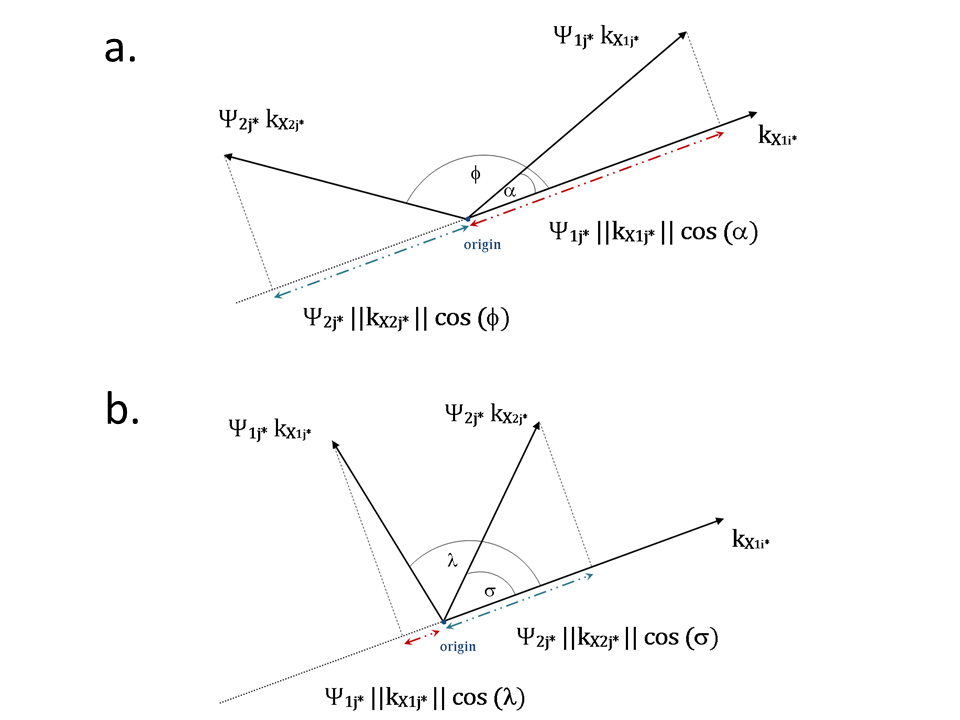}%
}\caption{Examples of positive and negative, eigen-scaled, signed magnitudes
of vector projections of eigen-scaled extreme vectors $\psi_{1_{j\ast}%
}k_{\mathbf{x}_{1_{j\ast}}}$ and $\psi_{2_{j\ast}}k_{\mathbf{x}_{2_{j\ast}}}$,
along the axis of an extreme vector $k_{\mathbf{x}_{1_{i\ast}}}$ which is
correlated with a Wolfe dual principal eigenaxis component $\psi_{1i\ast
}\protect\overrightarrow{\mathbf{e}}_{1i\ast}$.}%
\label{Wolfe Dual Quadratic Eigenlocus Statistics}%
\end{figure}

Figure $\ref{Wolfe Dual Quadratic Eigenlocus Statistics}$ illustrates the
geometric and statistical nature of the eigenlocus statistics in Eq.
(\ref{Projection Statistics psi1 Q}), where any given scaled, signed magnitude
$\psi_{1_{j\ast}}\left\Vert k_{\mathbf{x}_{1_{j\ast}}}\right\Vert \cos
\theta_{k_{\mathbf{x}_{1_{i\ast}}}k_{\mathbf{x}_{1_{j\ast}}}}$ or
$\psi_{2_{j\ast}}\left\Vert k_{\mathbf{x}_{2_{j\ast}}}\right\Vert \cos
\theta_{k_{\mathbf{x}_{1_{i\ast}}}k_{\mathbf{x}_{2_{j\ast}}}}$ may be positive
or negative (see Figs $\ref{Wolfe Dual Quadratic Eigenlocus Statistics}$a and
$\ref{Wolfe Dual Quadratic Eigenlocus Statistics}$b).%

\subsubsection{An Alternative Geometric Interpretation}

An alternative geometric explanation for the eigenlocus statistics in Eq.
(\ref{Projection Statistics psi1 Q}) accounts for the representation of the
$\boldsymbol{\kappa}_{1}$ and $\boldsymbol{\kappa}_{2}$ primal principal
eigenlocus components within the Wolfe dual eigenspace. Consider the
relationships%
\[
\psi_{1_{j\ast}}\left\Vert \left(  \mathbf{x}^{T}\mathbf{x}_{1_{j_{\ast}}%
}+1\right)  ^{2}\right\Vert =\left\Vert \psi_{1_{j\ast}}\left(  \mathbf{x}%
^{T}\mathbf{x}_{1_{j_{\ast}}}+1\right)  ^{2}\right\Vert =\left\Vert
\boldsymbol{\kappa}_{1}(j)\right\Vert
\]
and%
\[
\psi_{2_{j\ast}}\left\Vert \left(  \mathbf{x}^{T}\mathbf{x}_{2_{j_{\ast}}%
}+1\right)  ^{2}\right\Vert =\left\Vert \psi_{2_{j\ast}}\left(  \mathbf{x}%
^{T}\mathbf{x}_{2_{j_{\ast}}}+1\right)  ^{2}\right\Vert =\left\Vert
\boldsymbol{\kappa}_{2}(j)\right\Vert \text{,}%
\]
where $\boldsymbol{\kappa}_{1}(j)$ and $\boldsymbol{\kappa}_{2}(j)$ are the
$j$th constrained, primal principal eigenaxis components on
$\boldsymbol{\kappa}_{1}$ and $\boldsymbol{\kappa}_{2}$. Given the above
relationships, it follows that the scaled $\psi_{1_{j\ast}}$, signed magnitude
$\left\Vert \left(  \mathbf{x}^{T}\mathbf{x}_{1_{j_{\ast}}}+1\right)
^{2}\right\Vert \cos\theta_{k_{\mathbf{x}_{1_{i\ast}}}k_{\mathbf{x}_{1_{j\ast
}}}}$ of the vector projection of the scaled, extreme vector $\psi_{1_{j\ast}%
}k_{\mathbf{x}_{1_{j\ast}}}$ along the axis of the extreme vector
$k_{\mathbf{x}_{1_{i\ast}}}$%
\[
\psi_{1_{j\ast}}\left\Vert \left(  \mathbf{x}^{T}\mathbf{x}_{1_{j_{\ast}}%
}+1\right)  ^{2}\right\Vert \cos\theta_{k_{\mathbf{x}_{1_{i\ast}}%
}k_{\mathbf{x}_{1_{j\ast}}}}%
\]
determines the scaled $\cos\theta_{k_{\mathbf{x}_{1_{i\ast}}}k_{\mathbf{x}%
_{1_{j\ast}}}}$ length of the $j$th constrained, primal principal eigenaxis
component $\boldsymbol{\kappa}_{1}(j)$ on $\boldsymbol{\kappa}_{1}$%
\[
\cos\theta_{k_{\mathbf{x}_{1_{i\ast}}}k_{\mathbf{x}_{1_{j\ast}}}}\left\Vert
\boldsymbol{\kappa}_{1}(j)\right\Vert \text{,}%
\]
where $\psi_{1_{j\ast}}$ is the length of the $\psi_{1j\ast}%
\overrightarrow{\mathbf{e}}_{1j\ast}$ Wolfe dual principal eigenaxis component
and $\cos\theta_{k_{\mathbf{x}_{1_{i\ast}}}k_{\mathbf{x}_{1_{j\ast}}}}$
specifies the angle between the extreme vectors $k_{\mathbf{x}_{1_{i\ast}}}$
and $k_{\mathbf{x}_{1_{j\ast}}}$.

Likewise, the scaled $\psi_{2_{j\ast}}$, signed magnitude $\left\Vert \left(
\mathbf{x}^{T}\mathbf{x}_{2_{j_{\ast}}}+1\right)  ^{2}\right\Vert \cos
\theta_{k_{\mathbf{x}_{1_{i\ast}}}k_{\mathbf{x}_{2_{j\ast}}}}$ of the vector
projection of the scaled extreme vector $\psi_{2_{j\ast}}k_{\mathbf{x}%
_{2_{j\ast}}}$ along the axis of the extreme vector $k_{\mathbf{x}_{1_{i\ast}%
}}$%
\[
\psi_{2_{j\ast}}\left\Vert \left(  \mathbf{x}^{T}\mathbf{x}_{2_{j_{\ast}}%
}+1\right)  ^{2}\right\Vert \cos\theta_{k_{\mathbf{x}_{1_{i\ast}}%
}k_{\mathbf{x}_{2_{j\ast}}}}%
\]
determines the scaled $\cos\theta_{k_{\mathbf{x}_{1_{i\ast}}}k_{\mathbf{x}%
_{2_{j\ast}}}}$ length of the $j$th constrained, primal principal eigenaxis
component $\boldsymbol{\kappa}_{2}(j)$ on $\boldsymbol{\kappa}_{2}$%
\[
\cos\theta_{k_{\mathbf{x}_{1_{i\ast}}}k_{\mathbf{x}_{2_{j\ast}}}}\left\Vert
\boldsymbol{\kappa}_{2}(j)\right\Vert \text{,}%
\]
where $\psi_{2_{j\ast}}$ is the length of the $\psi_{2j\ast}%
\overrightarrow{\mathbf{e}}_{2j\ast}$ Wolfe dual principal eigenaxis component
and $\cos\theta_{k_{\mathbf{x}_{1_{i\ast}}}k_{\mathbf{x}_{2_{j\ast}}}}$
specifies the angle between the extreme vectors $k_{\mathbf{x}_{1_{i\ast}}}$
and $k_{\mathbf{x}_{2_{j\ast}}}$.

Therefore, the locus of each Wolfe dual principal eigenaxis component
$\psi_{1i\ast}\overrightarrow{\mathbf{e}}_{1i\ast}$ is a function of the
constrained, primal principal eigenaxis components on $\boldsymbol{\kappa}%
_{1}$ and $\boldsymbol{\kappa}_{2}$:%
\begin{align}
\psi_{1i\ast}  &  =\lambda_{\max_{\boldsymbol{\psi}}}^{-1}\left\Vert
k_{\mathbf{x}_{1_{i\ast}}}\right\Vert \sum\nolimits_{j=1}^{l_{1}}\cos
\theta_{k_{\mathbf{x}_{1_{i\ast}}}k_{\mathbf{x}_{1_{j\ast}}}}\left\Vert
\boldsymbol{\kappa}_{1}(j)\right\Vert
\label{Constrained Primal Eigenlocus psi1 Q}\\
&  -\lambda_{\max_{\boldsymbol{\psi}}}^{-1}\left\Vert k_{\mathbf{x}_{1_{i\ast
}}}\right\Vert \sum\nolimits_{j=1}^{l_{2}}\cos\theta_{k_{\mathbf{x}_{1_{i\ast
}}}k_{\mathbf{x}_{2_{j\ast}}}}\left\Vert \boldsymbol{\kappa}_{2}(j)\right\Vert
\text{,}\nonumber
\end{align}
where the angle between each principal eigenaxis component $\boldsymbol{\kappa
}_{1}(j)$ and $\boldsymbol{\kappa}_{2}(j)$ and the extreme vector
$k_{\mathbf{x}_{1_{i\ast}}}$ is fixed.

I will now define the significant geometric and statistical properties which
are jointly exhibited by Wolfe dual $\psi_{1i\ast}\overrightarrow{\mathbf{e}%
}_{1i\ast}$\textbf{\ }and constrained, primal $\psi_{1i\ast}k_{\mathbf{x}%
_{1_{i\ast}}}$ principal eigenaxis components that regulate the symmetric
partitioning of a feature space $Z$.

\subsection{Significant Geometric and Statistical Properties}

Using the definition of Eq.
(\ref{Eigen-balanced Pointwise Covariance Estimate Class One Q}), Eq.
(\ref{Dual Eigen-coordinate Locations Component One Q}) indicates that the
locus of the principal eigenaxis component $\psi_{1i\ast}%
\overrightarrow{\mathbf{e}}_{1i\ast}$ is determined by a symmetrically
balanced, signed magnitude along the axis of an extreme vector $k_{\mathbf{x}%
_{1_{i\ast}}}$, relative to symmetrically balanced, scaled, signed magnitudes
of extreme vector projections in each respective pattern class.

\subsubsection{Symmetrically Balanced Signed Magnitudes}

Let $\operatorname{comp}_{\overrightarrow{k_{\mathbf{x}_{1_{i\ast}}}}}\left(
\overrightarrow{\widetilde{\psi}_{1i\ast}\left\Vert k_{\widetilde{\mathbf{x}%
}\ast}\right\Vert _{_{1i_{\ast}}}}\right)  $ denote the symmetrically
balanced, signed magnitude%
\begin{align}
\operatorname{comp}_{\overrightarrow{k_{\mathbf{x}_{1_{i\ast}}}}}\left(
\overrightarrow{\widetilde{\psi}_{1i\ast}\left\Vert k_{\widetilde{\mathbf{x}%
}\ast}\right\Vert _{_{1i_{\ast}}}}\right)   &  =\sum\nolimits_{j=1}^{l_{1}%
}\psi_{1_{j\ast}}\label{Unidirectional Scaling Term One1 Q}\\
&  \times\left[  \left\Vert \left(  \mathbf{x}^{T}\mathbf{x}_{1_{j_{\ast}}%
}+1\right)  ^{2}\right\Vert \cos\theta_{k_{\mathbf{x}_{1_{i\ast}}%
}k_{\mathbf{x}_{1_{j\ast}}}}\right] \nonumber\\
&  -\sum\nolimits_{j=1}^{l_{2}}\psi_{2_{j\ast}}\nonumber\\
&  \times\left[  \left\Vert \left(  \mathbf{x}^{T}\mathbf{x}_{2_{j_{\ast}}%
}+1\right)  ^{2}\right\Vert \cos\theta_{k_{\mathbf{x}_{1_{i\ast}}%
}k_{\mathbf{x}_{2_{j\ast}}}}\right] \nonumber
\end{align}
along the axis of the extreme vector $k_{\mathbf{x}_{1_{i\ast}}}$ that is
correlated with the Wolfe dual principal eigenaxis component $\psi_{1i\ast
}\overrightarrow{\mathbf{e}}_{1i\ast}$, where%
\[
\left\Vert \left(  \mathbf{x}^{T}\mathbf{x}_{1_{j_{\ast}}}+1\right)
^{2}\right\Vert =\left\Vert \mathbf{x}_{1_{j\ast}}\right\Vert ^{2}+1
\]
and%
\[
\left\Vert \left(  \mathbf{x}^{T}\mathbf{x}_{2_{j_{\ast}}}+1\right)
^{2}\right\Vert =\left\Vert \mathbf{x}_{2_{j\ast}}\right\Vert ^{2}+1\text{.}%
\]

\subsubsection{Symmetrically Balanced Distributions}

Using the definitions of Eqs (\ref{Pointwise Covariance Statistic Q}) and
(\ref{Eigen-balanced Pointwise Covariance Estimate Class One Q}), it follows
that Eq. (\ref{Unidirectional Scaling Term One1 Q}) determines a symmetrically
balanced distribution of scaled, first and second degree coordinates of
extreme vectors along the axis of $k_{\mathbf{x}_{1_{i\ast}}}$, where each
scale factor $\psi_{1_{j\ast}}$ or $\psi_{2_{j\ast}}$ specifies how an extreme
vector $k_{\mathbf{x}_{1_{j\ast}}}$ or $k_{\mathbf{x}_{2_{j\ast}}}$ is
distributed along the axis of $k_{\mathbf{x}_{1_{i\ast}}}$, and each scale
factor $\psi_{1_{j\ast}}$ or $\psi_{2_{j\ast}}$ specifies a symmetrically
balanced distribution of scaled, first and second degree coordinates of
extreme vectors $\left\{  \psi_{_{j\ast}}k_{\mathbf{x}_{j\ast}}\right\}
_{j=1}^{l}$ along the axis of an extreme vector $k_{\mathbf{x}_{1_{j\ast}}}$
or $k_{\mathbf{x}_{2_{j\ast}}}$.

Therefore, each scaled, signed magnitude%
\[
\psi_{1_{j\ast}}\left\Vert k_{\mathbf{x}_{1_{j\ast}}}\right\Vert \cos
\theta_{k_{\mathbf{x}_{1_{i\ast}}}k_{\mathbf{x}_{1_{j\ast}}}}\text{ \ or
\ }\left\Vert k_{\mathbf{x}_{2_{j\ast}}}\right\Vert \cos\theta_{k_{\mathbf{x}%
_{1_{i\ast}}}k_{\mathbf{x}_{2_{j\ast}}}}%
\]
provides an estimate for how the components of the extreme vector
$k_{\mathbf{x}_{1_{i_{\ast}}}}$ are symmetrically distributed over the axis of
a scaled extreme vector $\psi_{1_{j\ast}}k_{\mathbf{x}_{1_{j\ast}}}$ or
$\psi_{2_{j\ast}}k_{\mathbf{x}_{2_{j\ast}}}$, where each scale factor
$\psi_{1_{j\ast}}$ or $\psi_{2_{j\ast}}$ specifies a symmetrically balanced
distribution of scaled, first and second degree coordinates of extreme vectors
$\left\{  \psi_{_{j\ast}}k_{\mathbf{x}_{j\ast}}\right\}  _{j=1}^{l}$ along the
axis of an extreme vector $k_{\mathbf{x}_{1_{j\ast}}}$ or $k_{\mathbf{x}%
_{2_{j\ast}}}$.

Again, using Eqs (\ref{Pointwise Covariance Statistic Q} and
(\ref{Eigen-balanced Pointwise Covariance Estimate Class One Q}), it follows
that Eq. (\ref{Unidirectional Scaling Term One1 Q}) determines a symmetrically
balanced, first and second-order statistical moment about the locus of
$k_{\mathbf{x}_{1_{i\ast}}}$, where each scale factor $\psi_{1_{j\ast}}$ or
$\psi_{1_{j\ast}}$ specifies how the components of an extreme vector
$k_{\mathbf{x}_{1_{j\ast}}}$ or $k_{\mathbf{x}_{2_{j\ast}}}$ are distributed
along the axis of $k_{\mathbf{x}_{1_{i\ast}}}$, and each scale factor
$\psi_{1_{j\ast}}$ or $\psi_{1_{j\ast}}$ specifies a symmetrically balanced
distribution for an extreme vector $k_{\mathbf{x}_{1_{j\ast}}}$ or
$k_{\mathbf{x}_{2_{j\ast}}}$.

\subsubsection{Distributions of Eigenaxis Components}

Using Eqs (\ref{Equilibrium Constraint on Dual Eigen-components Q}),
(\ref{Dual Eigen-coordinate Locations Component One Q}), and
(\ref{Unidirectional Scaling Term One1 Q}), it follows that symmetrically
balanced, joint distributions of the principal eigenaxis components on
$\boldsymbol{\psi}$ and $\boldsymbol{\kappa}$ are distributed over the axis of
the extreme vector $k_{\mathbf{x}_{1_{i\ast}}}$.

Again, using Eq. (\ref{Dual Eigen-coordinate Locations Component One Q}), it
follows that identical, symmetrically balanced, joint distributions of the
principal eigenaxis components on $\boldsymbol{\psi}$ and $\boldsymbol{\kappa
}$ are distributed over the axis of the Wolfe dual principal eigenaxis
component $\psi_{1i\ast}\overrightarrow{\mathbf{e}}_{1i\ast}$.

Thereby, symmetrically balanced, joint distributions of the principal
eigenaxis components on $\boldsymbol{\psi}$ and $\boldsymbol{\kappa}$ are
identically and symmetrically distributed over the respective axes of each
Wolfe dual principal eigenaxis component $\psi_{1i\ast}%
\overrightarrow{\mathbf{e}}_{1i\ast}$ and each correlated extreme vector
$k_{\mathbf{x}_{1_{i\ast}}}$.

Alternatively, using Eq. (\ref{Constrained Primal Eigenlocus psi1 Q}), the
symmetrically balanced, signed magnitude in Eq.
(\ref{Unidirectional Scaling Term One1 Q}) depends upon the difference between
integrated, cosine-scaled lengths of the constrained, primal principal
eigenaxis components on $\boldsymbol{\kappa}_{1}$ and $\boldsymbol{\kappa}%
_{2}$:%
\begin{align}
\operatorname{comp}_{\overrightarrow{k_{\mathbf{x}_{1_{i\ast}}}}}\left(
\overrightarrow{\widetilde{\psi}_{1i\ast}\left\Vert k_{\widetilde{\mathbf{x}%
}_{\ast}}\right\Vert _{_{1_{i\ast}}}}\right)   &  =\sum\nolimits_{j=1}^{l_{1}%
}\cos\theta_{k_{\mathbf{x}_{1_{i\ast}}}k_{\mathbf{x}_{1_{j\ast}}}}\left\Vert
\boldsymbol{\kappa}_{1}(j)\right\Vert
\label{Unidirectional Scaling Term One2 Q}\\
&  -\sum\nolimits_{j=1}^{l_{2}}\cos\theta_{k_{\mathbf{x}_{1_{i\ast}}%
}k_{\mathbf{x}_{2_{j\ast}}}}\left\Vert \boldsymbol{\kappa}_{2}(j)\right\Vert
\nonumber
\end{align}
which also shows that symmetrically balanced, joint distributions of the
eigenaxis components on $\boldsymbol{\psi}$ and $\boldsymbol{\kappa}$ are
identically distributed along the respective axes of each Wolfe dual principal
eigenaxis component $\psi_{1i\ast}\overrightarrow{\mathbf{e}}_{1i\ast}$ and
each correlated extreme vector $k_{\mathbf{x}_{1_{i\ast}}}$.

Using Eqs (\ref{Dual Eigen-coordinate Locations Component One Q}) and
(\ref{Unidirectional Scaling Term One1 Q}), it follows that the length
$\psi_{1i\ast}$ of each Wolfe dual principal eigenaxis component $\psi
_{1i\ast}\overrightarrow{\mathbf{e}}_{1i\ast}$ is determined by the weighted
length of a correlated, extreme vector $k_{\mathbf{x}_{1_{i\ast}}}$:%
\begin{equation}
\psi_{1i\ast}=\left[  \lambda_{\max_{\boldsymbol{\psi}}}^{-1}\times
\operatorname{comp}_{\overrightarrow{k_{\mathbf{x}_{1_{i\ast}}}}}\left(
\overrightarrow{\widetilde{\psi}_{1i\ast}\left\Vert k_{\widetilde{\mathbf{x}%
}_{\ast}}\right\Vert _{_{1_{i\ast}}}}\right)  \right]  \left\Vert
k_{\mathbf{x}_{1_{i\ast}}}\right\Vert \text{,}
\label{Magnitude Dual Normal Eigenaxis Component Class One Q}%
\end{equation}
where the weighting factor specifies an eigenvalue $\lambda_{\max
_{\boldsymbol{\psi}}}^{-1}$ scaling of a symmetrically balanced, signed
magnitude $\operatorname{comp}_{\overrightarrow{k_{\mathbf{x}_{1_{i\ast}}}}%
}\left(  \overrightarrow{\widetilde{\psi}_{1i\ast}\left\Vert
k_{\widetilde{\mathbf{x}}_{\ast}}\right\Vert _{_{1_{i\ast}}}}\right)  $ along
the axis of $k_{\mathbf{x}_{1_{i\ast}}}$.

\subsubsection{Symmetrically Balanced Lengths}

Given that $\psi_{1i\ast}>0$, $\lambda_{\max_{\boldsymbol{\psi}}}^{-1}>0$ and
$\left\Vert k_{\mathbf{x}_{1_{i\ast}}}\right\Vert >0$, it follows that the
symmetrically balanced, signed magnitude along the axis of each extreme vector
$k_{\mathbf{x}_{1_{i\ast}}}$ is a positive number%
\[
\operatorname{comp}_{\overrightarrow{k_{\mathbf{x}_{1_{i\ast}}}}}\left(
\overrightarrow{\widetilde{\psi}_{1i\ast}\left\Vert k_{\widetilde{\mathbf{x}%
}_{\ast}}\right\Vert _{_{1_{i\ast}}}}\right)  >0
\]
which indicates that the weighting factor in Eq.
(\ref{Magnitude Dual Normal Eigenaxis Component Class One Q}) determines a
well-proportioned length%
\[
\lambda_{\max_{\boldsymbol{\psi}}}^{-1}\operatorname{comp}%
_{\overrightarrow{k_{\mathbf{x}_{1_{i\ast}}}}}\left(
\overrightarrow{\widetilde{\psi}_{1i\ast}\left\Vert k_{\widetilde{\mathbf{x}%
}_{\ast}}\right\Vert _{_{1_{i\ast}}}}\right)  \left\Vert k_{\mathbf{x}%
_{1_{i\ast}}}\right\Vert
\]
for an extreme vector $k_{\mathbf{x}_{1_{i\ast}}}$. Thereby, the length
$\psi_{1i\ast}$ of each Wolfe dual principal eigenaxis component $\psi
_{1i\ast}\overrightarrow{\mathbf{e}}_{1i\ast}$ is determined by a
well-proportioned length of a correlated extreme vector $k_{\mathbf{x}%
_{1_{i\ast}}}$.

Returning to Eqs
(\ref{Eigen-balanced Pointwise Covariance Estimate Class One Q}) and
(\ref{Dual Eigen-coordinate Locations Component One Q}), it follows that the
length $\psi_{1i\ast}$ of each Wolfe dual principal eigenaxis component
$\psi_{1i\ast}\overrightarrow{\mathbf{e}}_{1i\ast}$ on $\boldsymbol{\psi}$%
\[
\psi_{1i\ast}=\lambda_{\max_{\boldsymbol{\psi}}}^{-1}\operatorname{comp}%
_{\overrightarrow{k_{\mathbf{x}_{1_{i\ast}}}}}\left(
\overrightarrow{\widetilde{\psi}_{1i\ast}\left\Vert k_{\widetilde{\mathbf{x}%
}_{\ast}}\right\Vert _{_{1_{i\ast}}}}\right)  \left\Vert k_{\mathbf{x}%
_{1_{i\ast}}}\right\Vert
\]
is shaped by a symmetrically balanced, first and second-order statistical
moment about the locus of a correlated extreme vector $k_{\mathbf{x}%
_{1_{i\ast}}}$.

Now, take any given correlated pair $\left\{  \psi_{1i\ast}%
\overrightarrow{\mathbf{e}}_{1i\ast},\psi_{1_{i\ast}}k_{\mathbf{x}_{1_{i\ast}%
}}\right\}  $ of Wolfe dual and constrained, primal principal eigenaxis
components. I will now show that the direction of $\psi_{1i\ast}%
\overrightarrow{\mathbf{e}}_{1i\ast}$ is identical to the direction of
$\psi_{1_{i\ast}}k_{\mathbf{x}_{1_{i\ast}}}$.

\subsubsection{Directional Symmetries}

The vector direction of each Wolfe dual principal eigenaxis component
$\psi_{1i\ast}\overrightarrow{\mathbf{e}}_{1i\ast}$ is implicitly specified by
Eq. (\ref{Dual Eigen-coordinate Locations Component One Q}), where it has been
assumed that $\psi_{1i\ast}$ provides a scale factor for a non-orthogonal unit
vector $\overrightarrow{\mathbf{e}}_{1i\ast}$. Using the definitions of Eqs
(\ref{Pointwise Covariance Statistic Q}) and
(\ref{Eigen-balanced Pointwise Covariance Estimate Class One Q}), it follows
that the symmetrically balanced, pointwise covariance statistic in Eq.
(\ref{Dual Eigen-coordinate Locations Component One Q}) specifies the
direction of a correlated extreme vector $k_{\mathbf{x}_{1_{i\ast}}}$ and a
well-proportioned magnitude along the axis of the extreme vector
$k_{\mathbf{x}_{1_{i\ast}}}$.

Returning to Eqs (\ref{Unidirectional Scaling Term One1 Q}),
(\ref{Unidirectional Scaling Term One2 Q}), and
(\ref{Magnitude Dual Normal Eigenaxis Component Class One Q}), take any given
Wolfe dual principal eigenaxis component $\psi_{1i\ast}%
\overrightarrow{\mathbf{e}}_{1i\ast}$ that is correlated with an extreme
vector $k_{\mathbf{x}_{1_{i\ast}}}$. Given that the magnitude $\psi_{1i\ast}$
of each Wolfe dual principal eigenaxis component $\psi_{1i\ast}%
\overrightarrow{\mathbf{e}}_{1i\ast}$ is determined by the well-proportioned
magnitude of a correlated extreme vector $k_{\mathbf{x}_{1_{i\ast}}}$%
\[
\psi_{1i\ast}=\lambda_{\max_{\boldsymbol{\psi}}}^{-1}\operatorname{comp}%
_{\overrightarrow{k_{\mathbf{x}_{1_{i\ast}}}}}\left(
\overrightarrow{\widetilde{\psi}_{1i\ast}\left\Vert k_{\widetilde{\mathbf{x}%
}_{\ast}}\right\Vert _{_{1_{i\ast}}}}\right)  \left\Vert k_{\mathbf{x}%
_{1_{i\ast}}}\right\Vert \text{,}%
\]
it follows that each non-orthogonal unit vector $\overrightarrow{\mathbf{e}%
}_{1i\ast}$ has the same direction as an extreme vector $k_{\mathbf{x}%
_{1_{i\ast}}}$%
\[
\overrightarrow{\mathbf{e}}_{1i\ast}\equiv\frac{k_{\mathbf{x}_{1_{i\ast}}}%
}{\left\Vert k_{\mathbf{x}_{1_{i\ast}}}\right\Vert }\text{.}%
\]
Thereby, the direction of each Wolfe dual principal eigenaxis component
$\psi_{1i\ast}\overrightarrow{\mathbf{e}}_{1i\ast}$ on $\boldsymbol{\psi}$ is
identical to the direction of a correlated, constrained primal principal
eigenaxis component $\psi_{1_{i\ast}}k_{\mathbf{x}_{1_{i\ast}}}$ on
$\boldsymbol{\kappa}_{1}$, which is determined by the direction of an extreme
vector $\mathbf{x}_{1_{i\ast}}$. Each Wolfe dual and correlated, constrained
primal principal eigenaxis component are said to exhibit directional symmetry.
Therefore, it is concluded that correlated principal eigenaxis components on
$\boldsymbol{\psi}_{1}$ and $\boldsymbol{\kappa}_{1}$ exhibit directional symmetry.

\subsubsection{Directions of Large Covariance}

It is concluded that the uniform directions of the Wolfe dual and the
correlated, constrained primal principal eigenaxis components determine
directions of large covariance which contribute to a symmetric partitioning of
a geometric region that spans a region of large covariance between two data
distributions. It is also concluded that each of the correlated principal
eigenaxis components on $\boldsymbol{\psi}_{1}$ and $\boldsymbol{\kappa}_{1}$
possess well-proportioned magnitudes for which the constrained, quadratic
eigenlocus discriminant function $\widetilde{\Lambda}_{\boldsymbol{\kappa}%
}\left(  \mathbf{s}\right)  =\left(  \mathbf{x}^{T}\mathbf{s}+1\right)
^{2}\boldsymbol{\kappa}+\kappa_{0}$ delineates symmetric regions of large
covariance between any two data distributions.

\subsection{Loci of the $\psi_{2i\ast}\protect\overrightarrow{\mathbf{e}%
}_{2i\ast}$ Components}

Let $i=1:l_{2}$, where each extreme vector $\left(  \mathbf{x}^{T}%
\mathbf{x}_{2_{i_{\ast}}}+1\right)  ^{2}$ is correlated with a Wolfe dual
principal eigenaxis component $\psi_{2i\ast}\overrightarrow{\mathbf{e}%
}_{2i\ast}$. Using Eqs (\ref{Dual Normal Eigenlocus Component Projections Q})
and (\ref{Non-orthogonal Eigenaxes of Dual Normal Eigenlocus Q}), it follows
that the locus of the $i^{th}$ Wolfe dual principal eigenaxis component
$\psi_{2i\ast}\overrightarrow{\mathbf{e}}_{2i\ast}$ on $\boldsymbol{\psi}$ is
a function of the expression:%
\begin{align}
\psi_{2i\ast}  &  =\lambda_{\max_{\boldsymbol{\psi}}}^{-1}\left\Vert
k_{\mathbf{x}_{2_{i\ast}}}\right\Vert \sum\nolimits_{j=1}^{l_{2}}%
\psi_{2_{j\ast}}\left\Vert k_{\mathbf{x}_{2_{j\ast}}}\right\Vert \cos
\theta_{k_{\mathbf{x}_{2_{i\ast}}}k_{\mathbf{x}_{2_{j\ast}}}}%
\label{Dual Eigen-coordinate Locations Component Two Q}\\
&  -\lambda_{\max_{\boldsymbol{\psi}}}^{-1}\left\Vert k_{\mathbf{x}_{2_{i\ast
}}}\right\Vert \sum\nolimits_{j=1}^{l_{1}}\psi_{1_{j\ast}}\left\Vert
k_{\mathbf{x}_{1_{j\ast}}}\right\Vert \cos\theta_{k_{\mathbf{x}_{1_{i\ast}}%
}k_{\mathbf{x}_{2_{j\ast}}}}\text{,}\nonumber
\end{align}
where $\psi_{2i\ast}$ provides a scale factor for the non-orthogonal unit
vector $\overrightarrow{\mathbf{e}}_{2i\ast}$.

Results obtained from the previous analysis are readily generalized to the
Wolfe dual $\psi_{2i\ast}\overrightarrow{\mathbf{e}}_{2i\ast}$ and the
constrained, primal $\psi_{2_{i\ast}}k_{\mathbf{x}_{2_{i\ast}}}$ principal
eigenaxis components, so the analysis will not be replicated. However, the
counterpart to Eq. (\ref{Unidirectional Scaling Term One1 Q}) is necessary for
a future argument. Let $i=1:l_{2}$, where each extreme vector $k_{\mathbf{x}%
_{2_{i_{\ast}}}}$ is correlated with a Wolfe principal eigenaxis component
$\psi_{2i\ast}\overrightarrow{\mathbf{e}}_{2i\ast}$. Accordingly, let
$\operatorname{comp}_{\overrightarrow{k_{\mathbf{x}_{2_{i\ast}}}}}\left(
\overrightarrow{\widetilde{\psi}_{2i\ast}\left\Vert k_{\widetilde{\mathbf{x}%
}_{\ast}}\right\Vert _{_{2_{i\ast}}}}\right)  $ denote the symmetrically
balanced, signed magnitude%
\begin{align}
\operatorname{comp}_{\overrightarrow{k_{\mathbf{x}_{2_{i\ast}}}}}\left(
\overrightarrow{\widetilde{\psi}_{2i\ast}\left\Vert k_{\widetilde{\mathbf{x}%
}_{\ast}}\right\Vert _{_{2_{i\ast}}}}\right)   &  =\sum\nolimits_{j=1}^{l_{2}%
}\psi_{2_{j\ast}}\label{Unidirectional Scaling Term Two1 Q}\\
&  \times\left[  \left\Vert \left(  \mathbf{x}^{T}\mathbf{x}_{2_{j_{\ast}}%
}+1\right)  ^{2}\right\Vert \cos\theta_{k_{\mathbf{x}_{2_{i\ast}}%
}k_{\mathbf{x}_{2_{j\ast}}}}\right] \nonumber\\
&  -\sum\nolimits_{j=1}^{l_{1}}\psi_{1_{j\ast}}\nonumber\\
&  \times\left[  \left\Vert \left(  \mathbf{x}^{T}\mathbf{x}_{1_{j_{\ast}}%
}+1\right)  ^{2}\right\Vert \cos\theta_{k_{\mathbf{x}_{2_{i\ast}}%
}k_{\mathbf{x}_{1_{j\ast}}}}\right] \nonumber
\end{align}
along the axis of the extreme vector $k_{\mathbf{x}_{2_{i\ast}}}$ that is
correlated with the Wolfe dual principal eigenaxis component $\psi_{2i\ast
}\overrightarrow{\mathbf{e}}_{2i\ast}$, where%
\[
\left\Vert \left(  \mathbf{x}^{T}\mathbf{x}_{1_{j_{\ast}}}+1\right)
^{2}\right\Vert =\left\Vert \mathbf{x}_{1_{j\ast}}\right\Vert ^{2}+1
\]
and%
\[
\left\Vert \left(  \mathbf{x}^{T}\mathbf{x}_{2_{j_{\ast}}}+1\right)
^{2}\right\Vert =\left\Vert \mathbf{x}_{2_{j\ast}}\right\Vert ^{2}+1\text{.}%
\]

\subsection{Similar Properties Exhibited by $\boldsymbol{\psi}$ and
$\boldsymbol{\kappa}$}

I\ will now identify similar geometric and statistical properties which are
jointly exhibited by the Wolfe dual principal eigenaxis components on
$\boldsymbol{\psi}$ and the correlated, constrained, primal principal
eigenaxis components on $\boldsymbol{\kappa}$. The properties are summarized below.

\paragraph{Directional Symmetry}

\begin{enumerate}
\item The direction of each Wolfe dual principal eigenaxis component
$\psi_{1i\ast}\overrightarrow{\mathbf{e}}_{1i\ast}$ on $\boldsymbol{\psi
}\mathbf{_{1}}$ is identical to the direction of a correlated, constrained
primal principal eigenaxis component $\psi_{1_{i\ast}}k_{\mathbf{x}_{1_{i\ast
}}}$ on$\mathbf{\ }\boldsymbol{\kappa}\mathbf{_{1}}$.

\item The direction of each Wolfe dual principal eigenaxis component
$\psi_{2i\ast}\overrightarrow{\mathbf{e}}_{2i\ast}$ on $\boldsymbol{\psi
}\mathbf{_{2}}$ is identical to the direction of a correlated, constrained
primal principal eigenaxis component $\psi_{2_{i\ast}}k_{\mathbf{x}_{2_{i\ast
}}}$ on$\mathbf{\ }\boldsymbol{\kappa}\mathbf{_{2}}$.
\end{enumerate}

\paragraph{Symmetrically Balanced Lengths}

\begin{enumerate}
\item The lengths of each Wolfe dual principal eigenaxis component
$\psi_{1i\ast}\overrightarrow{\mathbf{e}}_{1i\ast}$ on $\boldsymbol{\psi
}\mathbf{_{1}}$ and each correlated, constrained primal principal eigenaxis
component $\psi_{1_{i\ast}}k_{\mathbf{x}_{1_{i\ast}}}$ on$\mathbf{\ }%
\boldsymbol{\kappa}\mathbf{_{1}}$ are shaped by identical, symmetrically
balanced, joint distributions of the principal eigenaxis components on
$\boldsymbol{\psi}$ and $\boldsymbol{\kappa}$.

\item The lengths of each Wolfe dual principal eigenaxis component
$\psi_{2i\ast}\overrightarrow{\mathbf{e}}_{2i\ast}$ on $\boldsymbol{\psi
}\mathbf{_{2}}$ and each correlated, constrained primal principal eigenaxis
component $\psi_{2_{i\ast}}k_{\mathbf{x}_{2_{i\ast}}}$ on$\mathbf{\ }%
\boldsymbol{\kappa}\mathbf{_{2}}$ are shaped by identical, symmetrically
balanced, joint distributions of the principal eigenaxis components on
$\boldsymbol{\psi}$ and $\boldsymbol{\kappa}$.
\end{enumerate}

\paragraph{Symmetrically Balanced Pointwise Covariance Statistics}

\begin{enumerate}
\item The magnitude $\psi_{1i\ast}$ of each Wolfe dual principal eigenaxis
component $\psi_{1i\ast}\overrightarrow{\mathbf{e}}_{1i\ast}$ on
$\boldsymbol{\psi}\mathbf{_{1}}$%
\[
\psi_{1i\ast}=\lambda_{\max_{\boldsymbol{\psi}}}^{-1}\operatorname{comp}%
_{\overrightarrow{k_{\mathbf{x}_{1_{i\ast}}}}}\left(
\overrightarrow{\widetilde{\psi}_{1i\ast}\left\Vert k_{\widetilde{\mathbf{x}%
}_{\ast}}\right\Vert _{_{1_{i\ast}}}}\right)  \left\Vert k_{\mathbf{x}%
_{1_{i\ast}}}\right\Vert
\]
is determined by a symmetrically balanced, pointwise covariance estimate%
\begin{align*}
\widehat{\operatorname{cov}}_{up_{\updownarrow}}\left(  k_{\mathbf{x}%
_{1_{i_{\ast}}}}\right)   &  =\lambda_{\max_{\boldsymbol{\psi}}}%
^{-1}\left\Vert k_{\mathbf{x}_{1_{i_{\ast}}}}\right\Vert \\
&  \times\sum\nolimits_{j=1}^{l_{1}}\psi_{1_{j\ast}}\left\Vert k_{\mathbf{x}%
_{1_{j\ast}}}\right\Vert \cos\theta_{k_{\mathbf{x}_{1_{i\ast}}}k_{\mathbf{x}%
_{1_{j\ast}}}}\\
&  -\lambda_{\max_{\boldsymbol{\psi}}}^{-1}\left\Vert k_{\mathbf{x}%
_{1_{i_{\ast}}}}\right\Vert \\
&  \times\sum\nolimits_{j=1}^{l_{2}}\psi_{2_{j\ast}}\left\Vert k_{\mathbf{x}%
_{2_{j\ast}}}\right\Vert \cos\theta_{k_{\mathbf{x}_{1_{i\ast}}}k_{\mathbf{x}%
_{2_{j\ast}}}}%
\end{align*}
for a correlated extreme vector\textbf{\ }$k_{\mathbf{x}_{1_{i\ast}}}$, such
that the locus of each constrained, primal principal eigenaxis
component\textbf{\ }$\psi_{1_{i\ast}}k_{\mathbf{x}_{1_{i\ast}}}$\textbf{\ }on
$\boldsymbol{\kappa}\mathbf{_{1}}$\textbf{\ }provides a maximum\textit{\ }%
covariance estimate in a principal location $k_{\mathbf{x}_{1_{i_{\ast}}}}$,
in the form of a symmetrically balanced, first and second-order statistical
moment about the locus of an extreme point $k_{\mathbf{x}_{1_{i_{\ast}}}}$.

\item The magnitude $\psi_{2i\ast}$ of each Wolfe dual principal eigenaxis
component $\psi_{2i\ast}\overrightarrow{\mathbf{e}}_{2i\ast}$ on
$\boldsymbol{\psi}\mathbf{_{2}}$%
\[
\psi_{2i\ast}=\lambda_{\max_{\boldsymbol{\psi}}}^{-1}\operatorname{comp}%
_{\overrightarrow{k_{\mathbf{x}_{2i\ast}}}}\left(
\overrightarrow{\widetilde{\psi}_{2i\ast}\left\Vert k_{\widetilde{\mathbf{x}%
}_{\ast}}\right\Vert _{_{2i_{\ast}}}}\right)  \left\Vert k_{\mathbf{x}%
_{2_{i\ast}}}\right\Vert
\]
is determined by a symmetrically balanced, pointwise covariance estimate%
\begin{align*}
\widehat{\operatorname{cov}}_{up_{\updownarrow}}\left(  k_{\mathbf{x}%
_{2_{i_{\ast}}}}\right)   &  =\lambda_{\max_{\boldsymbol{\psi}}}%
^{-1}\left\Vert k_{\mathbf{x}_{2_{i\ast}}}\right\Vert \\
&  \times\sum\nolimits_{j=1}^{l_{2}}\psi_{2_{j\ast}}\left\Vert k_{\mathbf{x}%
_{2_{j\ast}}}\right\Vert \cos\theta_{k_{\mathbf{x}_{2_{i\ast}}}k_{\mathbf{x}%
_{2_{j\ast}}}}\\
&  -\lambda_{\max_{\boldsymbol{\psi}}}^{-1}\left\Vert k_{\mathbf{x}_{2_{i\ast
}}}\right\Vert \\
&  \times\sum\nolimits_{j=1}^{l_{1}}\psi_{1_{j\ast}}\left\Vert k_{\mathbf{x}%
_{1_{j\ast}}}\right\Vert \cos\theta_{k_{\mathbf{x}_{2_{i\ast}}}k_{\mathbf{x}%
_{1_{j\ast}}}}%
\end{align*}
for a correlated extreme vector\textbf{\ }$k_{\mathbf{x}_{2_{i\ast}}}$, such
that the locus of each constrained, primal principal eigenaxis
component\textbf{\ }$\psi_{2_{i\ast}}k_{\mathbf{x}_{2_{i\ast}}}$\textbf{\ }on
$\boldsymbol{\kappa}\mathbf{_{2}}$\textbf{\ }provides a maximum\textit{\ }%
covariance estimate in a principal location $k_{\mathbf{x}_{2_{i_{\ast}}}}$,
in the form of a symmetrically balanced, first and second-order statistical
moment about the locus of an extreme point $k_{\mathbf{x}_{2_{i_{\ast}}}}$.
\end{enumerate}

\paragraph{Symmetrically Balanced Statistical Moments}

\begin{enumerate}
\item Each Wolfe dual principal eigenaxis component\textbf{\ }$\psi_{1i\ast
}\overrightarrow{\mathbf{e}}_{1i\ast}$ on $\boldsymbol{\psi}\mathbf{_{1}}%
$\textbf{\ }specifies a symmetrically balanced, first and second-order
statistical moment about the locus of a correlated extreme point
$k_{\mathbf{x}_{1_{i_{\ast}}}}$, relative to the loci of all of the scaled
extreme points which determines the locus of a constrained, primal principal
eigenaxis component $\psi_{1_{i\ast}}k_{\mathbf{x}_{1_{i\ast}}}$ on
$\boldsymbol{\kappa}_{1}$.

\item Each Wolfe dual principal eigenaxis component\textbf{\ }$\psi_{2i\ast
}\overrightarrow{\mathbf{e}}_{1i\ast}$ on $\boldsymbol{\psi}\mathbf{_{2}}%
$\textbf{\ }specifies a symmetrically balanced, first and second-order
statistical moment about the locus of a correlated extreme point
$k_{\mathbf{x}_{2_{i_{\ast}}}}$, relative to the loci of all of the scaled
extreme points which determines the locus of a constrained, primal principal
eigenaxis component $\psi_{2_{i\ast}}k_{\mathbf{x}_{2_{i\ast}}}$ on
$\boldsymbol{\kappa}_{2}$.
\end{enumerate}

\paragraph{Symmetrically Balanced Distributions of Extreme Points}

\begin{enumerate}
\item Any given maximum covariance estimate $\widehat{\operatorname{cov}%
}_{up_{\updownarrow}}\left(  k_{\mathbf{x}_{1_{i_{\ast}}}}\right)  $ describes
how the components of $l$ scaled extreme vectors $\left\{  \psi_{1_{j\ast}%
}k_{\mathbf{x}_{1_{j_{\ast}}}}\right\}  _{j=1}^{l_{1}}$ and $\left\{
\psi_{2_{j\ast}}k_{\mathbf{x}_{2_{j_{\ast}}}}\right\}  _{j=1}^{l_{2}}$ are
distributed along the axis of an extreme vector $k_{\mathbf{x}_{1_{i\ast}}}$,
where each scale factor $\psi_{1_{j\ast}}$ or $\psi_{2_{j\ast}}$ specifies a
symmetrically balanced distribution of $l$ scaled extreme vectors along the
axis of an extreme vector $k_{\mathbf{x}_{1_{j_{\ast}}}}$ or $k_{\mathbf{x}%
_{k_{j_{\ast}}}}$, such that a pointwise covariance estimate
$\widehat{\operatorname{cov}}_{up_{\updownarrow}}\left(  k_{\mathbf{x}%
_{1_{i_{\ast}}}}\right)  $ provides an estimate for how the components of an
extreme vector $k_{\mathbf{x}_{1_{i\ast}}}$ are symmetrically distributed over
the axes of the $l$ scaled extreme vectors. Thus, $\widehat{\operatorname{cov}%
}_{up_{\updownarrow}}\left(  k_{\mathbf{x}_{1_{i_{\ast}}}}\right)  $ describes
a distribution of first and second degree coordinates for $k_{\mathbf{x}%
_{1_{i_{\ast}}}}$.

\item Any given maximum\textit{\ }covariance estimate
$\widehat{\operatorname{cov}}_{up_{\updownarrow}}\left(  k_{\mathbf{x}%
_{2_{i_{\ast}}}}\right)  $ describes how the components of $l$ scaled extreme
vectors $\left\{  \psi_{1_{j\ast}}k_{\mathbf{x}_{1_{j_{\ast}}}}\right\}
_{j=1}^{l_{1}}$ and $\left\{  \psi_{2_{j\ast}}k_{\mathbf{x}_{2_{j_{\ast}}}%
}\right\}  _{j=1}^{l_{2}}$ are distributed along the axis of an extreme vector
$k_{\mathbf{x}_{2_{i\ast}}}$, where each scale factor $\psi_{1_{j\ast}}$ or
$\psi_{2_{j\ast}}$ specifies a symmetrically balanced distribution of $l$
scaled extreme vectors along the axis of an extreme vector $k_{\mathbf{x}%
_{1_{j_{\ast}}}}$ or $k_{\mathbf{x}_{2_{j_{\ast}}}}$, such that a pointwise
covariance estimate $\widehat{\operatorname{cov}}_{up_{\updownarrow}}\left(
k_{\mathbf{x}_{2_{i_{\ast}}}}\right)  $ provides an estimate for how the
components of an extreme vector $k_{\mathbf{x}_{2_{i_{\ast}}}}$ are
symmetrically distributed over the axes of the $l$ scaled extreme vectors.
Thus, $\widehat{\operatorname{cov}}_{up_{\updownarrow}}\left(  k_{\mathbf{x}%
_{2_{i_{\ast}}}}\right)  $ describes a distribution of first and second degree
coordinates for $k_{\mathbf{x}_{2_{i_{\ast}}}}$.
\end{enumerate}

I\ will now define the equivalence between the total allowed eigenenergies
exhibited by $\boldsymbol{\psi}$ and $\boldsymbol{\kappa}$.

\subsection{Equivalence Between Eigenenergies of $\boldsymbol{\psi}$ and
$\boldsymbol{\kappa}$}

The inner product between the integrated Wolf dual principal eigenaxis
components on $\boldsymbol{\psi}$%
\begin{align*}
\left\Vert \boldsymbol{\psi}\right\Vert _{\min_{c}}^{2}  &  =\left(
\sum\nolimits_{i=1}^{l_{1}}\psi_{1i\ast}\frac{k_{\mathbf{x}_{1_{i_{\ast}}}}%
}{\left\Vert k_{\mathbf{x}_{1_{i_{\ast}}}}\right\Vert }+\sum\nolimits_{i=1}%
^{l_{2}}\psi_{2i\ast}\frac{k_{\mathbf{x}_{2_{i_{\ast}}}}}{\left\Vert
k_{\mathbf{x}_{2_{i_{\ast}}}}\right\Vert }\right) \\
&  \times\left(  \sum\nolimits_{i=1}^{l_{1}}\psi_{1i\ast}\frac{k_{\mathbf{x}%
_{1_{i_{\ast}}}}}{\left\Vert k_{\mathbf{x}_{1_{i_{\ast}}}}\right\Vert }%
+\sum\nolimits_{i=1}^{l_{2}}\psi_{2i\ast}\frac{k_{\mathbf{x}_{2_{i_{\ast}}}}%
}{\left\Vert k_{\mathbf{x}_{2_{i_{\ast}}}}\right\Vert }\right)
\end{align*}
determines the total allowed eigenenergy $\left\Vert \boldsymbol{\psi
}\right\Vert _{\min_{c}}^{2}$ of $\boldsymbol{\psi}$ which is symmetrically
equivalent with the critical minimum eigenenergy $\left\Vert
\boldsymbol{\kappa}\right\Vert _{\min_{c}}^{2}$ of $\boldsymbol{\kappa}$
within its Wolfe dual eigenspace%
\begin{align*}
\left\Vert \boldsymbol{\kappa}\right\Vert _{\min_{c}}^{2}  &  =\left(
\sum\nolimits_{i=1}^{l_{1}}\psi_{1_{i\ast}}\left(  \mathbf{x}^{T}%
\mathbf{x}_{1_{i_{\ast}}}+1\right)  ^{2}-\sum\nolimits_{i=1}^{l_{2}}%
\psi_{2_{i\ast}}\left(  \mathbf{x}^{T}\mathbf{x}_{2_{i_{\ast}}}+1\right)
^{2}\right) \\
&  \times\left(  \sum\nolimits_{i=1}^{l_{1}}\psi_{1_{i\ast}}\left(
\mathbf{x}^{T}\mathbf{x}_{1_{i_{\ast}}}+1\right)  ^{2}-\sum\nolimits_{i=1}%
^{l_{2}}\psi_{2_{i\ast}}\left(  \mathbf{x}^{T}\mathbf{x}_{2_{i_{\ast}}%
}+1\right)  ^{2}\right)  \text{.}%
\end{align*}
I will now argue that the equivalence $\left\Vert \boldsymbol{\psi}\right\Vert
_{\min_{c}}^{2}\simeq\left\Vert \boldsymbol{\kappa}\right\Vert _{\min_{c}}%
^{2}$ between the total allowed eigenenergies exhibited by $\boldsymbol{\psi}$
and $\boldsymbol{\kappa}$ involves symmetrically balanced, joint eigenenergy
distributions with respect to the principal eigenaxis components on
$\boldsymbol{\psi}$ and $\boldsymbol{\kappa}$.

\paragraph{Symmetrical Equivalence of Eigenenergy Distributions}

Using Eqs (\ref{Equilibrium Constraint on Dual Eigen-components Q}),
(\ref{Dual Eigen-coordinate Locations Component One Q}),
(\ref{Unidirectional Scaling Term One1 Q}), and
(\ref{Unidirectional Scaling Term Two1 Q}), it follows that identical,
symmetrically balanced, joint distributions of principal eigenaxis components
on $\boldsymbol{\psi}$ and $\boldsymbol{\kappa}$ are symmetrically distributed
over the respective axes of each Wolfe dual principal eigenaxis component on
$\boldsymbol{\psi}$ and each correlated and unconstrained primal principal
eigenaxis component (extreme vector) on $\boldsymbol{\kappa}$. Therefore,
constrained primal and Wolfe dual principal eigenaxis components that are
correlated with each other are formed by equivalent, symmetrically balanced,
joint distributions of principal eigenaxis components on $\boldsymbol{\psi}$
and $\boldsymbol{\kappa}$.

Thereby, symmetrically balanced, joint distributions of principal eigenaxis
components on $\boldsymbol{\psi}$ and $\boldsymbol{\kappa}$ are symmetrically
distributed over the axes of all of the Wolf dual principal eigenaxis
components $\left\{  \psi_{1i\ast}\overrightarrow{\mathbf{e}}_{1i\ast
}\right\}  _{i=1}^{l_{1}}$ and $\left\{  \psi_{2i\ast}%
\overrightarrow{\mathbf{e}}_{2i\ast}\right\}  _{i=1}^{l_{2}}$ on
$\boldsymbol{\psi}_{1}$ and $\boldsymbol{\psi}_{2}$ and all of the
constrained, primal principal eigenaxis components $\left\{  \psi_{1_{i\ast}%
}k_{\mathbf{x}_{1_{i\ast}}}\right\}  _{i=1}^{l_{1}}$ and $\left\{
\psi_{2_{i\ast}}k_{\mathbf{x}_{2_{i\ast}}}\right\}  _{i=1}^{l_{2}}$ on
$\boldsymbol{\kappa}_{1}$ and $\boldsymbol{\kappa}_{2}$, where
$\overrightarrow{\mathbf{e}}_{1i\ast}=\frac{k_{\mathbf{x}_{1_{i_{\ast}}}}%
}{\left\Vert k_{\mathbf{x}_{1_{i_{\ast}}}}\right\Vert }$ and
$\overrightarrow{\mathbf{e}}_{2i\ast}=\frac{k_{\mathbf{x}_{2_{i_{\ast}}}}%
}{\left\Vert k_{\mathbf{x}_{2_{i_{\ast}}}}\right\Vert }$.

Therefore, the distribution of eigenenergies with respect to the Wolfe dual
principal eigenaxis components on $\boldsymbol{\psi}$ is symmetrically
equivalent to the distribution of eigenenergies with respect to the
constrained, primal principal eigenaxis components on $\boldsymbol{\kappa}$,
such that the total allowed eigenenergies $\left\Vert \boldsymbol{\psi
}\right\Vert _{\min_{c}}^{2}$ and $\left\Vert \boldsymbol{\kappa}\right\Vert
_{\min_{c}}^{2}$ exhibited by $\boldsymbol{\psi}$ and $\boldsymbol{\kappa}$
satisfy symmetrically balanced, joint eigenenergy distributions with respect
to the principal eigenaxis components on\emph{\ }$\boldsymbol{\psi}$ and
$\boldsymbol{\kappa}$. Thus, all of the constrained, primal principal
eigenaxis components on $\boldsymbol{\kappa}_{1}-\boldsymbol{\kappa}_{2}$
possess eigenenergies that satisfy symmetrically balanced, joint eigenenergy
distributions with respect to the principal eigenaxis components on
$\boldsymbol{\psi}$ and $\boldsymbol{\kappa}$.

Later on, I\ will show that the critical minimum eigenenergies exhibited by
the scaled extreme vectors determine conditional probabilities of
classification error for extreme points (which are reproducing kernels), where
any given extreme point has a risk or a counter risk that is determined by a
measure of central location and a measure of spread, both of which are
described by a conditional probability density.

In the next section, I\ will show that each Wolfe dual principal eigenaxis
component specifies a conditional probability density for an extreme point
$k_{\mathbf{x}_{1_{i\ast}}}$ or $k_{\mathbf{x}_{2_{i\ast}}}$. Recall that it
is reasonable to assume that information about an unknown probability density
function $p\left(  \mathbf{x}\right)  $ is distributed over the components of
a parameter vector $\widehat{\mathbf{\theta}}$
\citep{Duda2001}%
. It has been demonstrated that symmetrically balanced, joint distributions of
principal eigenaxis components on $\boldsymbol{\psi}$ and $\boldsymbol{\kappa
}$ are symmetrically distributed over the axes of all of the Wolf dual
principal eigenaxis components on $\boldsymbol{\psi}$ and all of the
constrained, primal principal eigenaxis components on $\boldsymbol{\kappa}$.
In the next analysis, I\ will show that information for two unknown
conditional density functions $p\left(  k_{\mathbf{x}_{1_{i\ast}}%
}|\widehat{\mathbf{\theta}}_{1}\right)  $ and $p\left(  k_{\mathbf{x}%
_{2_{i\ast}}}|\widehat{\mathbf{\theta}}_{2}\right)  $ is distributed over the
scaled reproducing kernels of extreme points on $\boldsymbol{\kappa}%
_{1}-\boldsymbol{\kappa}_{2}$, where $\boldsymbol{\kappa}$ is an unknown
parameter vector $\widehat{\mathbf{\theta}}$ that contains information about
the unknown conditional densities $\widehat{\mathbf{\theta}}%
=\widehat{\mathbf{\theta}}_{1}-\widehat{\mathbf{\theta}}_{2}$.

I will now define pointwise conditional densities which are determined by the
components of a constrained, primal quadratic eigenlocus $\boldsymbol{\kappa
}=$ $\boldsymbol{\kappa}_{1}-\boldsymbol{\kappa}_{2}$, where each conditional
density $p\left(  k_{\mathbf{x}_{1i\ast}}|\operatorname{comp}%
_{\overrightarrow{k_{\mathbf{x}_{1i\ast}}}}\left(
\overrightarrow{\boldsymbol{\kappa}}\right)  \right)  $ or $p\left(
k_{\mathbf{x}_{2i\ast}}|\operatorname{comp}_{\overrightarrow{k_{\mathbf{x}%
_{2i\ast}}}}\left(  \overrightarrow{\boldsymbol{\kappa}}\right)  \right)  $
for an $k_{\mathbf{x}_{1_{i\ast}}}$ or $k_{\mathbf{x}_{2_{i\ast}}}$ extreme
point is given by components $\operatorname{comp}%
_{\overrightarrow{k_{\mathbf{x}_{1i\ast}}}}\left(
\overrightarrow{\boldsymbol{\kappa}}\right)  $ or $\operatorname{comp}%
_{\overrightarrow{k_{\mathbf{x}_{2i\ast}}}}\left(
\overrightarrow{\boldsymbol{\kappa}}\right)  $ of $\boldsymbol{\kappa} $ along
the corresponding extreme vector $k_{\mathbf{x}_{1_{i\ast}}}$ or
$k_{\mathbf{x}_{2_{i\ast}}}$.

\subsection{Pointwise Conditional Densities}

Consider again the equations for the loci of the $\psi_{1i\ast}%
\overrightarrow{\mathbf{e}}_{1i\ast}$ and $\psi_{2i\ast}%
\overrightarrow{\mathbf{e}}_{2i\ast}$ Wolfe dual principal eigenaxis
components in Eqs (\ref{Dual Eigen-coordinate Locations Component One Q}) and
(\ref{Dual Eigen-coordinate Locations Component Two Q}). It has been
demonstrated that any given Wolfe dual principal eigenaxis component
$\psi_{1i\ast}\overrightarrow{\mathbf{e}}_{1i\ast}$ correlated with a
reproducing kernel $k_{\mathbf{x}_{1_{i\ast}}}$ of an $\mathbf{x}_{1_{i_{\ast
}}}$ extreme point and any given Wolfe dual principal eigenaxis component
$\psi_{2i\ast}\overrightarrow{\mathbf{e}}_{2i\ast}$ correlated with a
reproducing kernel $k_{\mathbf{x}_{2_{i\ast}}}$ of an $\mathbf{x}_{2_{i_{\ast
}}}$ extreme point provides an estimate for how the components of $l$ scaled
extreme vectors $\left\{  \psi_{_{j\ast}}k_{\mathbf{x}_{j\ast}}\right\}
_{j=1}^{l}$ are symmetrically distributed along the axis of a correlated
extreme vector $k_{\mathbf{x}_{1_{i_{\ast}}}}$ or $k_{\mathbf{x}_{2_{i\ast}}}%
$, where components of scaled extreme vectors $\psi_{_{j\ast}}k_{\mathbf{x}%
_{j\ast}}$ are symmetrically distributed according to class labels $\pm1$,
signed magnitudes $\left\Vert k_{\mathbf{x}_{j\ast}}\right\Vert \cos
\theta_{k_{\mathbf{x}_{1_{i\ast}}}k_{\mathbf{x}_{_{j\ast}}}}$ or $\left\Vert
k_{\mathbf{x}_{j\ast}}\right\Vert \cos\theta_{k_{\mathbf{x}_{2_{i\ast}}%
}k_{\mathbf{x}_{_{j\ast}}}}$ and symmetrically balanced distributions of
scaled extreme vectors $\left\{  \psi_{_{j\ast}}k_{\mathbf{x}_{j\ast}%
}\right\}  _{j=1}^{l}$ specified by scale factors $\psi_{_{j\ast}}$.

Thereby, symmetrically balanced distributions of first and second degree
coordinates of all of the extreme points are symmetrically distributed along
the axes of all of the extreme vectors, where all of the scale factors satisfy
the equivalence relation $\sum\nolimits_{j=1}^{l_{1}}\psi_{1_{j\ast}}%
=\sum\nolimits_{j=1}^{l_{2}}\psi_{2_{j\ast}}$. Accordingly, principal
eigenaxis components $\psi_{1i\ast}\overrightarrow{\mathbf{e}}_{1i\ast}$ or
$\psi_{2i\ast}\overrightarrow{\mathbf{e}}_{2i\ast}$ describe distributions of
first and second degree coordinates for extreme points $k_{\mathbf{x}%
_{1_{i_{\ast}}}}$ or $k_{\mathbf{x}_{2_{i_{\ast}}}}$, where any given extreme
point is the endpoint of a directed line segment estimate.

Therefore, for any given extreme vector $k_{\mathbf{x}_{1_{i\ast}}}$, the
relative likelihood that the extreme point $k_{\mathbf{x}_{1_{i\ast}}}$ has a
given location is specified by the locus of the Wolfe dual principal eigenaxis
component $\psi_{1i\ast}\overrightarrow{\mathbf{e}}_{1i\ast}$:%
\begin{align*}
\psi_{1i\ast}\overrightarrow{\mathbf{e}}_{1i\ast}  &  =\lambda_{\max
_{\boldsymbol{\psi}}}^{-1}\left\Vert k_{\mathbf{x}_{1_{i_{\ast}}}}\right\Vert
\sum\nolimits_{j=1}^{l_{1}}\psi_{1_{j\ast}}\left\Vert k_{\mathbf{x}_{1_{j\ast
}}}\right\Vert \cos\theta_{k_{\mathbf{x}_{1_{i\ast}}}k_{\mathbf{x}_{1_{j\ast}%
}}}\\
&  -\lambda_{\max_{\boldsymbol{\psi}}}^{-1}\left\Vert k_{\mathbf{x}%
_{1_{i_{\ast}}}}\right\Vert \sum\nolimits_{j=1}^{l_{2}}\psi_{2_{j\ast}%
}\left\Vert k_{\mathbf{x}_{2_{j\ast}}}\right\Vert \cos\theta_{k_{\mathbf{x}%
_{1_{i\ast}}}k_{\mathbf{x}_{2_{j\ast}}}}\text{,}%
\end{align*}
where $\psi_{1i\ast}\overrightarrow{\mathbf{e}}_{1i\ast}$ describes a
conditional expectation (a measure of central location) and a conditional
covariance (a measure of spread) for the extreme point $k_{\mathbf{x}%
_{1_{i_{\ast}}}}$. Thereby, it is concluded that the principal eigenaxis
component $\psi_{1i\ast}\overrightarrow{\mathbf{e}}_{1i\ast}$ specifies
a\emph{\ }conditional density $p\left(  k_{\mathbf{x}_{1i\ast}}%
|\operatorname{comp}_{\overrightarrow{k_{\mathbf{x}_{1i\ast}}}}\left(
\overrightarrow{\boldsymbol{\kappa}}\right)  \right)  $ for the extreme point
$k_{\mathbf{x}_{1_{i_{\ast}}}}$, where the scale factor $\psi_{1i\ast}$ is a
\emph{unit }measure or estimate of density and likelihood for the extreme
point $k_{\mathbf{x}_{1_{i_{\ast}}}}$.

Likewise, for any given extreme vector $k_{\mathbf{x}_{2_{i\ast}}}$, the
relative likelihood that the extreme point $k_{\mathbf{x}_{2_{i\ast}}}$ has a
given location is specified by the locus of the Wolfe dual principal eigenaxis
component $\psi_{2i\ast}\overrightarrow{\mathbf{e}}_{2i\ast}$:%
\begin{align*}
\psi_{2i\ast}\overrightarrow{\mathbf{e}}_{2i\ast}  &  =\lambda_{\max
_{\boldsymbol{\psi}}}^{-1}\left\Vert k_{\mathbf{x}_{2_{i\ast}}}\right\Vert
\sum\nolimits_{j=1}^{l_{2}}\psi_{2_{j\ast}}\left\Vert k_{\mathbf{x}_{2_{j\ast
}}}\right\Vert \cos\theta_{k_{\mathbf{x}_{2_{i\ast}}}k_{\mathbf{x}_{2_{j\ast}%
}}}\\
&  -\lambda_{\max_{\boldsymbol{\psi}}}^{-1}\left\Vert k_{\mathbf{x}_{2_{i\ast
}}}\right\Vert \sum\nolimits_{j=1}^{l_{1}}\psi_{1_{j\ast}}\left\Vert
k_{\mathbf{x}_{1_{j\ast}}}\right\Vert \cos\theta_{k_{\mathbf{x}_{2_{i\ast}}%
}k_{\mathbf{x}_{1_{j\ast}}}}\text{,}%
\end{align*}
where $\psi_{2i\ast}\overrightarrow{\mathbf{e}}_{2i\ast}$ describes a
conditional expectation (a measure of central location) and a conditional
covariance (a measure of spread) for the extreme point $k_{\mathbf{x}%
_{2_{i\ast}}}$. Thereby, it is concluded that the principal eigenaxis
component $\psi_{2i\ast}\overrightarrow{\mathbf{e}}_{2i\ast}$ specifies
a\emph{\ }conditional density $p\left(  k_{\mathbf{x}_{2i\ast}}%
|\operatorname{comp}_{\overrightarrow{k_{\mathbf{x}_{2i\ast}}}}\left(
\overrightarrow{\boldsymbol{\kappa}}\right)  \right)  $ for the extreme point
$k_{\mathbf{x}_{2_{i\ast}}}$, where the scale factor $\psi_{2i\ast}$ is a
\emph{unit }measure or estimate of density and likelihood for the extreme
point $k_{\mathbf{x}_{2_{i\ast}}}$.

It has been shown that a Wolfe dual quadratic eigenlocus $\boldsymbol{\psi}$
is formed by a locus of scaled, normalized extreme vectors%
\begin{align*}
\boldsymbol{\psi}  &  =\sum\nolimits_{i=1}^{l_{1}}\psi_{1i\ast}\frac
{k_{\mathbf{x}_{1_{i_{\ast}}}}}{\left\Vert k_{\mathbf{x}_{1_{i_{\ast}}}%
}\right\Vert }+\sum\nolimits_{i=1}^{l_{2}}\psi_{2i\ast}\frac{k_{\mathbf{x}%
_{2_{i\ast}}}}{\left\Vert k_{\mathbf{x}_{2_{i\ast}}}\right\Vert }\\
&  =\boldsymbol{\psi}_{1}+\boldsymbol{\psi}_{2}\text{,}%
\end{align*}
where $\boldsymbol{\psi}_{1}=\sum\nolimits_{i=1}^{l_{1}}\psi_{1i\ast}%
\frac{k_{\mathbf{x}_{1_{i_{\ast}}}}}{\left\Vert k_{\mathbf{x}_{1_{i_{\ast}}}%
}\right\Vert }$ and $\boldsymbol{\psi}_{2}=\sum\nolimits_{i=1}^{l_{2}}%
\psi_{2i\ast}\frac{k_{\mathbf{x}_{2_{i\ast}}}}{\left\Vert k_{\mathbf{x}%
_{2_{i\ast}}}\right\Vert }$, and each $\psi_{1i\ast}$ or $\psi_{2i\ast}$ scale
factor provides a unit measure or estimate of density and likelihood for an
$k_{\mathbf{x}_{1_{i_{\ast}}}}$ or $k_{\mathbf{x}_{2_{i\ast}}}$ extreme point.

Given that each Wolfe dual principal eigenaxis component $\psi_{1i\ast}%
\frac{k_{\mathbf{x}_{1_{i_{\ast}}}}}{\left\Vert k_{\mathbf{x}_{1_{i_{\ast}}}%
}\right\Vert }$ on $\boldsymbol{\psi}_{1}$ specifies a conditional density
$p\left(  k_{\mathbf{x}_{1i\ast}}|\operatorname{comp}%
_{\overrightarrow{k_{\mathbf{x}_{1i\ast}}}}\left(
\overrightarrow{\boldsymbol{\kappa}}\right)  \right)  $ for a correlated
extreme point $k_{\mathbf{x}_{1_{i_{\ast}}}}$, it follows that conditional
densities for the $k_{\mathbf{x}_{1_{i_{\ast}}}}$ extreme points are
distributed over the principal eigenaxis components of $\boldsymbol{\psi}_{1}$%
\begin{align}
\boldsymbol{\psi}_{1}  &  =\sum\nolimits_{i=1}^{l_{1}}p\left(  k_{\mathbf{x}%
_{1i\ast}}|\operatorname{comp}_{\overrightarrow{k_{\mathbf{x}_{1i\ast}}}%
}\left(  \overrightarrow{\boldsymbol{\kappa}}\right)  \right)  \frac
{k_{\mathbf{x}_{1_{i_{\ast}}}}}{\left\Vert k_{\mathbf{x}_{1_{i_{\ast}}}%
}\right\Vert }\label{Wolfe Dual Conditional Density Extreme Points 1 Q}\\
&  =\sum\nolimits_{i=1}^{l_{1}}\psi_{1i\ast}\frac{k_{\mathbf{x}_{1_{i_{\ast}}%
}}}{\left\Vert k_{\mathbf{x}_{1_{i_{\ast}}}}\right\Vert }\text{,}\nonumber
\end{align}
where $\psi_{1i\ast}\frac{k_{\mathbf{x}_{1_{i_{\ast}}}}}{\left\Vert
k_{\mathbf{x}_{1_{i_{\ast}}}}\right\Vert }$ specifies a conditional density
for $k_{\mathbf{x}_{1i\ast}}$, such that $\boldsymbol{\psi}_{1}$ is a
parameter vector for a class-conditional probability density $p\left(
\frac{k_{\mathbf{x}_{1_{i_{\ast}}}}}{\left\Vert k_{\mathbf{x}_{1_{i_{\ast}}}%
}\right\Vert }|\boldsymbol{\psi}_{1}\right)  $ for a given set $\left\{
k_{\mathbf{x}_{1_{i_{\ast}}}}\right\}  _{i=1}^{l_{1}}$ of $k_{\mathbf{x}%
_{1_{i_{\ast}}}}$ extreme points:%
\[
\boldsymbol{\psi}_{1}=p\left(  \frac{k_{\mathbf{x}_{1_{i_{\ast}}}}}{\left\Vert
k_{\mathbf{x}_{1_{i_{\ast}}}}\right\Vert }|\boldsymbol{\psi}_{1}\right)
\text{.}%
\]

Given that each Wolfe dual principal eigenaxis component $\psi_{2i\ast}%
\frac{k_{\mathbf{x}_{2_{i\ast}}}}{\left\Vert k_{\mathbf{x}_{2_{i\ast}}%
}\right\Vert }$ on $\boldsymbol{\psi}_{2}$ specifies a conditional density
$p\left(  k_{\mathbf{x}_{2i\ast}}|\operatorname{comp}%
_{\overrightarrow{k_{\mathbf{x}_{2i\ast}}}}\left(
\overrightarrow{\boldsymbol{\kappa}}\right)  \right)  $ for a correlated
extreme point $k_{\mathbf{x}_{2_{i\ast}}}$, it follows that conditional
densities for the $k_{\mathbf{x}_{2_{i\ast}}}$ extreme points are distributed
over the principal eigenaxis components of $\boldsymbol{\psi}_{2}$%
\begin{align}
\boldsymbol{\psi}_{2}  &  =\sum\nolimits_{i=1}^{l_{2}}p\left(  k_{\mathbf{x}%
_{2i\ast}}|\operatorname{comp}_{\overrightarrow{k_{\mathbf{x}_{2i\ast}}}%
}\left(  \overrightarrow{\boldsymbol{\kappa}}\right)  \right)  \frac
{k_{\mathbf{x}_{2_{i\ast}}}}{\left\Vert k_{\mathbf{x}_{2_{i\ast}}}\right\Vert
}\label{Wolfe Dual Conditional Density Extreme Points 2 Q}\\
&  =\sum\nolimits_{i=1}^{l_{2}}\psi_{2_{i\ast}}\frac{k_{\mathbf{x}_{2_{i\ast}%
}}}{\left\Vert k_{\mathbf{x}_{2_{i\ast}}}\right\Vert }\text{,}\nonumber
\end{align}
where $\psi_{2_{i\ast}}\frac{k_{\mathbf{x}_{2_{i\ast}}}}{\left\Vert
k_{\mathbf{x}_{2_{i\ast}}}\right\Vert }$ specifies a conditional density for
$k_{\mathbf{x}_{2_{i\ast}}}$, such that $\boldsymbol{\psi}_{2}$ is a parameter
vector for a class-conditional probability density $p\left(  \frac
{k_{\mathbf{x}_{2_{i\ast}}}}{\left\Vert k_{\mathbf{x}_{2_{i\ast}}}\right\Vert
}|\boldsymbol{\psi}_{2}\right)  $ for a given set $\left\{  k_{\mathbf{x}%
_{2_{i\ast}}}\right\}  _{i=1}^{l_{2}}$ of $k_{\mathbf{x}_{2_{i\ast}}}$ extreme
points:%
\[
\boldsymbol{\psi}_{2}=p\left(  \frac{k_{\mathbf{x}_{2_{i\ast}}}}{\left\Vert
k_{\mathbf{x}_{2_{i\ast}}}\right\Vert }|\boldsymbol{\psi}_{2}\right)  \text{.}%
\]

Therefore, it is concluded that $\boldsymbol{\psi}_{1}$ is a parameter vector
for the class-conditional probability density function $p\left(
\frac{k_{\mathbf{x}_{1_{i_{\ast}}}}}{\left\Vert k_{\mathbf{x}_{1_{i_{\ast}}}%
}\right\Vert }|\boldsymbol{\psi}_{1}\right)  $ and $\boldsymbol{\psi}_{2}$ is
a parameter vector for the class-conditional probability density function
$p\left(  \frac{k_{\mathbf{x}_{2_{i\ast}}}}{\left\Vert k_{\mathbf{x}%
_{2_{i\ast}}}\right\Vert }|\boldsymbol{\psi}_{2}\right)  $.

Returning to Eq. (\ref{Equilibrium Constraint on Dual Eigen-components Q}), it
follows that the pointwise conditional densities $\psi_{1i\ast}\frac
{k_{\mathbf{x}_{1_{i_{\ast}}}}}{\left\Vert k_{\mathbf{x}_{1_{i_{\ast}}}%
}\right\Vert }$ and $\psi_{2i\ast}\frac{k_{\mathbf{x}_{2_{i_{\ast}}}}%
}{\left\Vert k_{\mathbf{x}_{2_{i_{\ast}}}}\right\Vert }$ for all of the
extreme points in class $\omega_{1}$ and class $\omega_{2}$ are symmetrically
balanced with each other:%
\[
\sum\nolimits_{i=1}^{l_{1}}\psi_{1i\ast}\frac{k_{\mathbf{x}_{1_{i_{\ast}}}}%
}{\left\Vert k_{\mathbf{x}_{1_{i_{\ast}}}}\right\Vert }\equiv\sum
\nolimits_{i=1}^{l_{2}}\psi_{2i\ast}\frac{k_{\mathbf{x}_{2_{i_{\ast}}}}%
}{\left\Vert k_{\mathbf{x}_{2_{i_{\ast}}}}\right\Vert }%
\]
\qquad in the Wolfe dual eigenspace. Therefore, the class-conditional
probability density functions $\boldsymbol{\psi}_{1}$ and $\boldsymbol{\psi
}_{2}$ in the Wolfe dual eigenspace for class $\omega_{1}$ and class
$\omega_{2}$ are \emph{symmetrically balanced with each other}:%
\[
p\left(  \frac{k_{\mathbf{x}_{1_{i_{\ast}}}}}{\left\Vert k_{\mathbf{x}%
_{1_{i_{\ast}}}}\right\Vert }|\boldsymbol{\psi}_{1}\right)  =p\left(
\frac{k_{\mathbf{x}_{2_{i_{\ast}}}}}{\left\Vert k_{\mathbf{x}_{2_{i_{\ast}}}%
}\right\Vert }|\boldsymbol{\psi}_{2}\right)  \text{.}%
\]

I\ will now devise expressions for the class-conditional probability density
functions in the decision space $Z$ for class $\omega_{1}$ and class
$\omega_{2}$.

\subsection{Class-conditional Probability Densities}

I\ will now show that a quadratic eigenlocus $\boldsymbol{\kappa=\kappa}%
_{1}-\boldsymbol{\kappa}_{2}$ is a parameter vector for class-conditional
probability density functions $p\left(  k_{\mathbf{x}_{1_{i\ast}}}|\omega
_{1}\right)  $ and $p\left(  k_{\mathbf{x}_{2_{i\ast}}}|\omega_{2}\right)  $.

\subsubsection{Class-Conditional Density for Class $\omega_{1}$}

Given that each Wolfe dual principal eigenaxis component $\psi_{1i\ast
}\overrightarrow{\mathbf{e}}_{1i\ast}$ specifies a conditional density
$p\left(  k_{\mathbf{x}_{1i\ast}}|\operatorname{comp}%
_{\overrightarrow{k_{\mathbf{x}_{1i\ast}}}}\left(
\overrightarrow{\boldsymbol{\kappa}}\right)  \right)  $ for a correlated
extreme point $k_{\mathbf{x}_{1i\ast}}$, it follows that conditional densities
for the $k_{\mathbf{x}_{1i\ast}}$ extreme points are distributed over the
principal eigenaxis components of $\boldsymbol{\kappa}_{1}$%
\begin{align}
\boldsymbol{\kappa}_{1}  &  =\sum\nolimits_{i=1}^{l_{1}}p\left(
k_{\mathbf{x}_{1i\ast}}|\operatorname{comp}_{\overrightarrow{k_{\mathbf{x}%
_{1i\ast}}}}\left(  \overrightarrow{\boldsymbol{\kappa}}\right)  \right)
k_{\mathbf{x}_{1_{i\ast}}}\label{Conditional Density Extreme Points 1 Q}\\
&  =\sum\nolimits_{i=1}^{l_{1}}\psi_{1_{i\ast}}k_{\mathbf{x}_{1_{i\ast}}%
}\text{,}\nonumber
\end{align}
where $\psi_{1_{i\ast}}k_{\mathbf{x}_{1_{i\ast}}}$ specifies a conditional
density for $k_{\mathbf{x}_{1_{i_{\ast}}}}$, such that $\boldsymbol{\kappa
}_{1}$ is a parameter vector for a class-conditional probability density
$p\left(  k_{\mathbf{x}_{1i\ast}}|\boldsymbol{\kappa}_{1}\right)  $ for a
given set $\left\{  k_{\mathbf{x}_{1_{i\ast}}}\right\}  _{i=1}^{l_{1}}$ of
$k_{\mathbf{x}_{1_{i_{\ast}}}}$ extreme points:%
\[
\boldsymbol{\kappa}_{1}=p\left(  k_{\mathbf{x}_{1i\ast}}|\boldsymbol{\kappa
}_{1}\right)  \text{.}%
\]

\subsubsection{Class-Conditional Density for Class $\omega_{2}$}

Given that each Wolfe dual principal eigenaxis component $\psi_{2i\ast
}\overrightarrow{\mathbf{e}}_{2i\ast}$ specifies a conditional density
$p\left(  k_{\mathbf{x}_{2i\ast}}|\operatorname{comp}%
_{\overrightarrow{k_{\mathbf{x}_{2i\ast}}}}\left(
\overrightarrow{\boldsymbol{\kappa}}\right)  \right)  $ for a correlated
extreme point $k_{\mathbf{x}_{2i\ast}}$, it follows that conditional densities
for the $k_{\mathbf{x}_{2i\ast}}$ extreme points are distributed over the
principal eigenaxis components of $\boldsymbol{\kappa}_{2}$%
\begin{align}
\boldsymbol{\kappa}_{2}  &  =\sum\nolimits_{i=1}^{l_{2}}p\left(
k_{\mathbf{x}_{2i\ast}}|\operatorname{comp}_{\overrightarrow{k_{\mathbf{x}%
_{2i\ast}}}}\left(  \overrightarrow{\boldsymbol{\kappa}}\right)  \right)
k_{\mathbf{x}_{2_{i\ast}}}\label{Conditional Density Extreme Points 2 Q}\\
&  =\sum\nolimits_{i=1}^{l_{2}}\psi_{2_{i\ast}}k_{\mathbf{x}_{2_{i\ast}}%
}\text{,}\nonumber
\end{align}
where $\psi_{2_{i\ast}}k_{\mathbf{x}_{2_{i\ast}}}$ specifies a conditional
density for $k_{\mathbf{x}_{2i\ast}}$, such that $\boldsymbol{\kappa}_{2}$ is
a parameter vector for a class-conditional probability density $p\left(
k_{\mathbf{x}_{2_{i_{\ast}}}}|\boldsymbol{\kappa}_{2}\right)  $ for a given
set $\left\{  k_{\mathbf{x}_{2i\ast}}\right\}  _{i=1}^{l_{2}}$ of
$k_{\mathbf{x}_{2i\ast}}$ extreme points:%
\[
\boldsymbol{\kappa}_{2}=p\left(  k_{\mathbf{x}_{2_{i_{\ast}}}}%
|\boldsymbol{\kappa}_{2}\right)  \text{.}%
\]

Therefore, it is concluded that $\boldsymbol{\kappa}_{1}$ is a parameter
vector for the class-conditional probability density function $p\left(
k_{\mathbf{x}_{1i\ast}}|\boldsymbol{\kappa}_{1}\right)  $ and
$\boldsymbol{\kappa}_{2}$ is a parameter vector for the class-conditional
probability density function $p\left(  k_{\mathbf{x}_{2_{i_{\ast}}}%
}|\boldsymbol{\kappa}_{2}\right)  $.

I will now devise integrals for the conditional probability functions for
class $\omega_{1}$ and class $\omega_{2}$.

\subsection{Conditional Probability Functions}

I\ will now show that the conditional probability function $P\left(
k_{\mathbf{x}_{1_{i\ast}}}|\boldsymbol{\kappa}_{1}\right)  $ for class
$\omega_{1}$ is given by the area under the class-conditional probability
density function $p\left(  k_{\mathbf{x}_{1i\ast}}|\boldsymbol{\kappa}%
_{1}\right)  $ over the decision space $Z$.

\subsubsection{Conditional Probability Function for Class $\omega_{1}$}

A quadratic eigenlocus $\boldsymbol{\kappa}=\sum\nolimits_{i=1}^{l_{1}}%
\psi_{1_{i\ast}}k_{\mathbf{x}_{1_{i\ast}}}-\sum\nolimits_{i=1}^{l_{2}}%
\psi_{2_{i\ast}}k_{\mathbf{x}_{2_{i\ast}}}$ is the basis of a quadratic
classification system $\boldsymbol{\kappa}^{T}k_{\mathbf{s}}+\kappa
_{0}\overset{\omega_{1}}{\underset{\omega_{2}}{\gtrless}}0$ that partitions
any given feature space into symmetrical decision regions $Z_{1}\simeq Z_{2}$,
whereby, for any two overlapping data distributions, an $k_{\mathbf{x}%
_{1i\ast}}$ or $k_{\mathbf{x}_{2i\ast}}$ extreme point lies in either region
$Z_{1}$ or region $Z_{2}$, and for any two non-overlapping data distributions,
$k_{\mathbf{x}_{1i\ast}}$ extreme points lie in region $Z_{1}$ and
$k_{\mathbf{x}_{2i\ast}}$ extreme points lie in region $Z_{2}$.

Therefore, the area under each pointwise conditional density in Eq.
(\ref{Conditional Density Extreme Points 1 Q})%
\[
\int_{Z_{1}}p\left(  k_{\mathbf{x}_{1i\ast}}|\operatorname{comp}%
_{\overrightarrow{k_{\mathbf{x}_{1i\ast}}}}\left(
\overrightarrow{\boldsymbol{\kappa}}\right)  \right)  d\boldsymbol{\kappa}%
_{1}\left(  k_{\mathbf{x}_{1i\ast}}\right)  \text{ or }\int_{Z_{2}}p\left(
k_{\mathbf{x}_{1i\ast}}|\operatorname{comp}_{\overrightarrow{k_{\mathbf{x}%
_{1i\ast}}}}\left(  \overrightarrow{\boldsymbol{\kappa}}\right)  \right)
d\boldsymbol{\kappa}_{1}\left(  k_{\mathbf{x}_{1i\ast}}\right)
\]
is a conditional probability that an $k_{\mathbf{x}_{1i\ast}}$ extreme point
will be observed in either region $Z_{1}$ or region $Z_{2}$.

Thus, the area $P\left(  k_{\mathbf{x}_{1_{i\ast}}}|\boldsymbol{\kappa}%
_{1}\right)  $ under the class-conditional density function $p\left(
k_{\mathbf{x}_{1i\ast}}|\boldsymbol{\kappa}_{1}\right)  $ in Eq.
(\ref{Conditional Density Extreme Points 1 Q})%
\begin{align*}
P\left(  k_{\mathbf{x}_{1_{i\ast}}}|\boldsymbol{\kappa}_{1}\right)   &
=\int_{Z}\left(  \sum\nolimits_{i=1}^{l_{1}}p\left(  k_{\mathbf{x}_{1i\ast}%
}|\operatorname{comp}_{\overrightarrow{k_{\mathbf{x}_{1i\ast}}}}\left(
\overrightarrow{\boldsymbol{\kappa}}\right)  \right)  k_{\mathbf{x}_{1_{i\ast
}}}\right)  d\boldsymbol{\kappa}_{1}\\
&  =\int_{Z}\left(  \sum\nolimits_{i=1}^{l_{1}}\psi_{1_{i\ast}}k_{\mathbf{x}%
_{1_{i\ast}}}\right)  d\boldsymbol{\kappa}_{1}=\int_{Z}p\left(  k_{\mathbf{x}%
_{1i\ast}}|\boldsymbol{\kappa}_{1}\right)  d\boldsymbol{\kappa}_{1}\\
&  =\int_{Z}\boldsymbol{\kappa}_{1}d\boldsymbol{\kappa}_{1}=\frac{1}%
{2}\left\Vert \boldsymbol{\kappa}_{1}\right\Vert ^{2}+C=\left\Vert
\boldsymbol{\kappa}_{1}\right\Vert ^{2}+C_{1}%
\end{align*}
specifies the conditional probability of observing a set $\left\{
k_{\mathbf{x}_{1i\ast}}\right\}  _{i=1}^{l_{1}}$ of $k_{\mathbf{x}_{1i\ast}}$
extreme points within \emph{localized regions} of the decision space $Z$,
where conditional densities $\psi_{1_{i\ast}}k_{\mathbf{x}_{1i\ast}}$ for
$k_{\mathbf{x}_{1i\ast}}$ extreme points that lie in the $Z_{2}$ decision
region \emph{contribute} to the cost or risk $\mathfrak{R}_{\mathfrak{\min}%
}\left(  Z_{2}|\psi_{1_{i\ast}}k_{\mathbf{x}_{1i\ast}}\right)  $ of making a
decision error, and conditional densities $\psi_{1_{i\ast}}k_{\mathbf{x}%
_{1i\ast}}$ for $k_{\mathbf{x}_{1i\ast}}$ extreme points that lie in the
$Z_{1}$ decision region \emph{counteract }the cost or risk $\overline
{\mathfrak{R}}_{\mathfrak{\min}}\left(  Z_{1}|\psi_{1_{i\ast}}k_{\mathbf{x}%
_{1i\ast}}\right)  $ of making a decision error.

It follows that the area $P\left(  k_{\mathbf{x}_{1_{i\ast}}}%
|\boldsymbol{\kappa}_{1}\right)  $ under the class-conditional probability
density function $p\left(  k_{\mathbf{x}_{1i\ast}}|\boldsymbol{\kappa}%
_{1}\right)  $ is determined by regions of risk $\mathfrak{R}_{\mathfrak{\min
}}\left(  Z_{2}|\psi_{1_{i\ast}}k_{\mathbf{x}_{1i\ast}}\right)  $ and regions
of counter risk $\overline{\mathfrak{R}}_{\mathfrak{\min}}\left(  Z_{1}%
|\psi_{1_{i\ast}}k_{\mathbf{x}_{1i\ast}}\right)  $ for the $k_{\mathbf{x}%
_{1_{i\ast}}}$ extreme points, where regions of risk $\mathfrak{R}%
_{\mathfrak{\min}}\left(  Z_{2}|\psi_{1_{i\ast}}k_{\mathbf{x}_{1i\ast}%
}\right)  $ and regions of counter risk $\overline{\mathfrak{R}}%
_{\mathfrak{\min}}\left(  Z_{1}|\psi_{1_{i\ast}}k_{\mathbf{x}_{1i\ast}%
}\right)  $ are localized regions in decision space $Z$ that are determined by
central locations (expected values) and spreads (covariances) of
$k_{\mathbf{x}_{1_{i\ast}}}$ extreme points.

Therefore, the conditional probability function $P\left(  k_{\mathbf{x}%
_{1_{i\ast}}}|\boldsymbol{\kappa}_{1}\right)  $ for class $\omega_{1}$ is
given by the integral%
\begin{align}
P\left(  k_{\mathbf{x}_{1_{i\ast}}}|\boldsymbol{\kappa}_{1}\right)   &
=\int_{Z}p\left(  \widehat{\Lambda}_{\boldsymbol{\kappa}}\left(
\mathbf{s}\right)  |\omega_{1}\right)  d\widehat{\Lambda}_{\boldsymbol{\kappa
}}=\int_{Z}p\left(  k_{\mathbf{x}_{1i\ast}}|\boldsymbol{\kappa}_{1}\right)
d\boldsymbol{\kappa}_{1}%
\label{Conditional Probability Function for Class One Q}\\
&  =\int_{Z}\boldsymbol{\kappa}_{1}d\boldsymbol{\kappa}_{1}=\left\Vert
\boldsymbol{\kappa}_{1}\right\Vert ^{2}+C_{1}\text{,}\nonumber
\end{align}
over the decision space $Z$, which has a solution in terms of the critical
minimum eigenenergy $\left\Vert \boldsymbol{\kappa}_{1}\right\Vert _{\min_{c}%
}^{2}$ exhibited by $\boldsymbol{\kappa}_{1}$ and an integration constant
$C_{1}$.

I\ will now demonstrate that the conditional probability function $P\left(
k_{\mathbf{x}_{2_{i\ast}}}|\boldsymbol{\kappa}_{2}\right)  $ for class
$\omega_{2}$ is given by the area under the class-conditional probability
density function $p\left(  k_{\mathbf{x}_{2i\ast}}|\boldsymbol{\kappa}%
_{2}\right)  $ over the decision space $Z.$

\subsubsection{Conditional Probability Function for Class $\omega_{2}$}

The area under each pointwise conditional density in Eq.
(\ref{Conditional Density Extreme Points 2 Q})%
\[
\int_{Z_{1}}p\left(  k_{\mathbf{x}_{2i\ast}}|\operatorname{comp}%
_{\overrightarrow{k_{\mathbf{x}_{2i\ast}}}}\left(
\overrightarrow{\boldsymbol{\kappa}}\right)  \right)  d\boldsymbol{\kappa}%
_{2}\left(  k_{\mathbf{x}_{2_{i\ast}}}\right)  \text{ or }\int_{Z_{2}}p\left(
k_{\mathbf{x}_{2i\ast}}|\operatorname{comp}_{\overrightarrow{k_{\mathbf{x}%
_{2i\ast}}}}\left(  \overrightarrow{\boldsymbol{\kappa}}\right)  \right)
d\boldsymbol{\kappa}_{2}\left(  k_{\mathbf{x}_{2_{i\ast}}}\right)
\]
is a conditional probability that an $k_{\mathbf{x}_{2_{i\ast}}}$ extreme
point will be observed in either region $Z_{1}$ or region $Z_{2}$.

Thus, the area $P\left(  k_{\mathbf{x}_{2_{i\ast}}}|\boldsymbol{\kappa}%
_{2}\right)  $ under the class-conditional density $p\left(  k_{\mathbf{x}%
_{2i\ast}}|\boldsymbol{\kappa}_{2}\right)  $ in Eq.
(\ref{Conditional Density Extreme Points 2 Q})%
\begin{align*}
P\left(  k_{\mathbf{x}_{2_{i\ast}}}|\boldsymbol{\kappa}_{2}\right)   &
=\int_{Z}\left(  \sum\nolimits_{i=1}^{l_{2}}p\left(  k_{\mathbf{x}_{2i\ast}%
}|\operatorname{comp}_{\overrightarrow{k_{\mathbf{x}_{2i\ast}}}}\left(
\overrightarrow{\boldsymbol{\kappa}}\right)  \right)  k_{\mathbf{x}_{2_{i\ast
}}}\right)  d\boldsymbol{\kappa}_{2}\\
&  =\int_{Z}\left(  \sum\nolimits_{i=1}^{l_{2}}\psi_{2_{i\ast}}k_{\mathbf{x}%
_{2_{i\ast}}}\right)  d\boldsymbol{\kappa}_{2}=\int_{Z}p\left(  k_{\mathbf{x}%
_{2i\ast}}|\boldsymbol{\kappa}_{2}\right)  d\boldsymbol{\kappa}_{2}\\
&  =\int_{Z}\boldsymbol{\kappa}_{2}d\boldsymbol{\kappa}_{2}=\frac{1}%
{2}\left\Vert \boldsymbol{\kappa}_{2}\right\Vert ^{2}+C=\left\Vert
\boldsymbol{\kappa}_{2}\right\Vert ^{2}+C_{2}%
\end{align*}
specifies the conditional probability of observing a set $\left\{
k_{\mathbf{x}_{2_{i\ast}}}\right\}  _{i=1}^{l_{2}}$ of $k_{\mathbf{x}%
_{2_{i\ast}}}$ extreme points within localized regions of the decision space
$Z$, where conditional densities $\psi_{2i\ast}k_{\mathbf{x}_{2_{i\ast}}}$ for
$k_{\mathbf{x}_{2_{i\ast}}}$ extreme points that lie in the $Z_{1}$ decision
region \emph{contribute} to the cost or risk $\mathfrak{R}_{\mathfrak{\min}%
}\left(  Z_{1}|\psi_{2i\ast}k_{\mathbf{x}_{2_{i\ast}}}\right)  $ of making a
decision error, and conditional densities $\psi_{2i\ast}k_{\mathbf{x}%
_{2_{i\ast}}}$ for $k_{\mathbf{x}_{2_{i\ast}}}$ extreme points that lie in the
$Z_{2}$ decision region \emph{counteract }the cost or risk $\overline
{\mathfrak{R}}_{\mathfrak{\min}}\left(  Z_{2}|\psi_{2i\ast}k_{\mathbf{x}%
_{2_{i\ast}}}\right)  $ of making a decision error.

It follows that the area $P\left(  k_{\mathbf{x}_{2_{i\ast}}}%
|\boldsymbol{\kappa}_{2}\right)  $ under the class-conditional probability
density function $p\left(  k_{\mathbf{x}_{2i\ast}}|\boldsymbol{\kappa}%
_{2}\right)  $ is determined by regions of risk $\mathfrak{R}_{\mathfrak{\min
}}\left(  Z_{1}|\psi_{2i\ast}k_{\mathbf{x}_{2_{i\ast}}}\right)  $ and regions
of counter risk $\overline{\mathfrak{R}}_{\mathfrak{\min}}\left(  Z_{2}%
|\psi_{2i\ast}k_{\mathbf{x}_{2_{i\ast}}}\right)  $ for the $k_{\mathbf{x}%
_{2_{i\ast}}}$ extreme points, where regions of risk $\mathfrak{R}%
_{\mathfrak{\min}}\left(  Z_{1}|\psi_{2i\ast}k_{\mathbf{x}_{2_{i\ast}}%
}\right)  $ and regions of counter risk $\overline{\mathfrak{R}}%
_{\mathfrak{\min}}\left(  Z_{2}|\psi_{2i\ast}k_{\mathbf{x}_{2_{i\ast}}%
}\right)  $ are localized regions in decision space $Z$ that are determined by
central locations (expected values) and spreads (covariances) of
$k_{\mathbf{x}_{2_{i\ast}}}$ extreme points.

Therefore, the conditional probability function $P\left(  k_{\mathbf{x}%
_{2_{i\ast}}}|\boldsymbol{\kappa}_{2}\right)  $ for class $\omega_{2}$ is
given by the integral%
\begin{align}
P\left(  k_{\mathbf{x}_{2_{i\ast}}}|\boldsymbol{\kappa}_{2}\right)   &
=\int_{Z}p\left(  \widehat{\Lambda}_{\boldsymbol{\kappa}}\left(
\mathbf{s}\right)  |\omega_{2}\right)  d\widehat{\Lambda}_{\boldsymbol{\kappa
}}=\int_{Z}p\left(  k_{\mathbf{x}_{2i\ast}}|\boldsymbol{\kappa}_{2}\right)
d\boldsymbol{\kappa}_{2}%
\label{Conditional Probability Function for Class Two Q}\\
&  =\int_{Z}\boldsymbol{\kappa}_{2}d\boldsymbol{\kappa}_{2}=\left\Vert
\boldsymbol{\kappa}_{2}\right\Vert ^{2}+C_{2}\text{,}\nonumber
\end{align}
over the decision space $Z$, which has a solution in terms of the critical
minimum eigenenergy $\left\Vert \boldsymbol{\kappa}_{2}\right\Vert ^{2}$
exhibited by $\boldsymbol{\kappa}_{2}$ and an integration constant $C_{2}$.

In order to precisely define the manner in which quadratic eigenlocus
discriminant functions $\widetilde{\Lambda}_{\boldsymbol{\kappa}}\left(
\mathbf{s}\right)  =\left(  \mathbf{x}^{T}\mathbf{s}+1\right)  ^{2}%
\boldsymbol{\kappa}+\kappa_{0}$ satisfy the fundamental integral equation of
binary classification for a classification system in statistical equilibrium,
I need to precisely define the manner in which the total allowed eigenenergies
of the principal eigenaxis components on $\boldsymbol{\kappa}$ are
symmetrically balanced with each other. Furthermore, I need to identify the
manner in which the property of symmetrical balance exhibited by the principal
eigenaxis components on $\boldsymbol{\psi}$ \emph{and} $\boldsymbol{\kappa}$
enables quadratic eigenlocus classification systems $\left(  \mathbf{x}%
^{T}\mathbf{s}+1\right)  ^{2}\boldsymbol{\kappa}+\kappa_{0}\overset{\omega
_{1}}{\underset{\omega_{2}}{\gtrless}}0$ to \emph{effectively balance} all of
the forces associated with the risk $\mathfrak{R}_{\mathfrak{\min}}\left(
Z_{1}|\boldsymbol{\kappa}_{2}\right)  $ and the counter risk $\overline
{\mathfrak{R}}_{\mathfrak{\min}}\left(  Z_{1}|\boldsymbol{\kappa}_{1}\right)
$ in the $Z_{1}$ decision region with all of the forces associated with the
risk $\mathfrak{R}_{\mathfrak{\min}}\left(  Z_{2}|\boldsymbol{\kappa}%
_{1}\right)  $ and the counter risk $\overline{\mathfrak{R}}_{\mathfrak{\min}%
}\left(  Z_{2}|\boldsymbol{\kappa}_{2}\right)  $ in the $Z_{2}$ decision region.

Recall that the expected risk $\mathfrak{R}_{\mathfrak{\min}}\left(
Z|\widehat{\Lambda}\left(  \mathbf{x}\right)  \right)  $ of a binary
classification system%
\[
\mathfrak{R}_{\mathfrak{\min}}\left(  Z|\widehat{\Lambda}\left(
\mathbf{x}\right)  \right)  =\mathfrak{R}_{\mathfrak{\min}}\left(
Z_{2}|p\left(  \widehat{\Lambda}\left(  \mathbf{x}\right)  |\omega_{1}\right)
\right)  +\mathfrak{R}_{\mathfrak{\min}}\left(  Z_{1}|p\left(
\widehat{\Lambda}\left(  \mathbf{x}\right)  |\omega_{2}\right)  \right)
\]
involves opposing forces that depend on the likelihood ratio test
$\widehat{\Lambda}\left(  \mathbf{x}\right)  =p\left(  \widehat{\Lambda
}\left(  \mathbf{x}\right)  |\omega_{1}\right)  -p\left(  \widehat{\Lambda
}\left(  \mathbf{x}\right)  |\omega_{2}\right)  \overset{\omega_{1}%
}{\underset{\omega_{2}}{\gtrless}}0$ and the corresponding decision boundary
$p\left(  \widehat{\Lambda}\left(  \mathbf{x}\right)  |\omega_{1}\right)
-p\left(  \widehat{\Lambda}\left(  \mathbf{x}\right)  |\omega_{2}\right)  =0$.

It has been demonstrated that quadratic eigenlocus transforms define these
opposing forces in terms of symmetrically balanced, pointwise covariance
statistics:%
\begin{align*}
\psi_{1i\ast}\frac{k_{\mathbf{x}_{1_{i_{\ast}}}}}{\left\Vert k_{\mathbf{x}%
_{1_{i_{\ast}}}}\right\Vert }  &  =\lambda_{\max_{\boldsymbol{\psi}}}%
^{-1}\left\Vert k_{\mathbf{x}_{1_{i\ast}}}\right\Vert \sum\nolimits_{j=1}%
^{l_{1}}\psi_{1_{j\ast}}\left\Vert k_{\mathbf{x}_{1_{j\ast}}}\right\Vert
\cos\theta_{k_{\mathbf{x}_{1_{i\ast}}}k_{\mathbf{x}_{1_{j\ast}}}}\\
&  -\lambda_{\max_{\boldsymbol{\psi}}}^{-1}-\left\Vert k_{\mathbf{x}%
_{1_{i\ast}}}\right\Vert \sum\nolimits_{j=1}^{l_{2}}\psi_{2_{j\ast}}\left\Vert
k_{\mathbf{x}_{2_{j\ast}}}\right\Vert \cos\theta_{k_{\mathbf{x}_{1_{i\ast}}%
}k_{\mathbf{x}_{2_{j\ast}}}}\text{,}%
\end{align*}
and%
\begin{align*}
\psi_{2i\ast}\frac{k_{\mathbf{x}_{2_{i_{\ast}}}}}{\left\Vert k_{\mathbf{x}%
_{2_{i_{\ast}}}}\right\Vert }  &  =\lambda_{\max_{\boldsymbol{\psi}}}%
^{-1}\left\Vert k_{\mathbf{x}_{2_{i\ast}}}\right\Vert \sum\nolimits_{j=1}%
^{l_{2}}\psi_{2_{j\ast}}\left\Vert k_{\mathbf{x}_{2_{j\ast}}}\right\Vert
\cos\theta_{k_{\mathbf{x}_{2_{i\ast}}}k_{\mathbf{x}_{2_{j\ast}}}}\\
&  -\lambda_{\max_{\boldsymbol{\psi}}}^{-1}\left\Vert k_{\mathbf{x}_{2_{i\ast
}}}\right\Vert \sum\nolimits_{j=1}^{l_{1}}\psi_{1_{j\ast}}\left\Vert
k_{\mathbf{x}_{1_{j\ast}}}\right\Vert \cos\theta_{k_{\mathbf{x}_{1_{i\ast}}%
}k_{\mathbf{x}_{2_{j\ast}}}}\text{,}%
\end{align*}
such that any given conditional density%
\[
p\left(  k_{\mathbf{x}_{1i\ast}}|\operatorname{comp}%
_{\overrightarrow{k_{\mathbf{x}_{1i\ast}}}}\left(
\overrightarrow{\boldsymbol{\kappa}}\right)  \right)  \text{ \ or \ }p\left(
k_{\mathbf{x}_{2i\ast}}|\operatorname{comp}_{\overrightarrow{k_{\mathbf{x}%
_{2i\ast}}}}\left(  \overrightarrow{\boldsymbol{\kappa}}\right)  \right)
\]
for a respective extreme point $k_{\mathbf{x}_{1_{i_{\ast}}}}$ or
$k_{\mathbf{x}_{2_{i\ast}}}$ is defined in terms of related counter risks and
risks associated with positions and potential locations of $k_{\mathbf{x}%
_{1_{i_{\ast}}}}$ and $k_{\mathbf{x}_{2_{i\ast}}}$ extreme points within the
$Z_{1}$ and $Z_{2}$ decision regions of a decision space $Z$.

Quadratic eigenlocus transforms routinely accomplish an elegant, statistical
balancing feat that involves finding the right mix of principal eigenaxis
components on $\boldsymbol{\psi}$ and $\boldsymbol{\kappa}$. I\ will now show
that the scale factors $\left\{  \psi_{i\ast}\right\}  _{i=1}^{l}$ of the
Wolfe dual principal eigenaxis components $\left\{  \psi_{i\ast}%
\frac{k_{\mathbf{x}_{i\ast}}}{\left\Vert k_{\mathbf{x}_{i\ast}}\right\Vert
}|\psi_{i\ast}>0\right\}  _{i=1}^{l}$ on $\boldsymbol{\psi}$ play a
fundamental role in this statistical balancing feat. I\ will develop an
equation of statistical equilibrium for the axis of $\boldsymbol{\kappa}$ that
is determined by the equation of statistical equilibrium:%
\[
\sum\nolimits_{i=1}^{l_{1}}\psi_{1i\ast}\frac{k_{\mathbf{x}_{1_{i_{\ast}}}}%
}{\left\Vert k_{\mathbf{x}_{1_{i_{\ast}}}}\right\Vert }=\sum\nolimits_{i=1}%
^{l_{2}}\psi_{2i\ast}\frac{k_{\mathbf{x}_{2_{i_{\ast}}}}}{\left\Vert
k_{\mathbf{x}_{2_{i_{\ast}}}}\right\Vert }%
\]
for the axis of $\boldsymbol{\psi}$.

\subsection{Finding the Right Mix of Component Lengths}

It has been demonstrated that the directions of the constrained primal and the
Wolfe dual principal eigenaxis components are fixed, along with the angles
between all of the extreme vectors. I will now argue that the lengths of the
Wolfe dual principal eigenaxis components on $\boldsymbol{\psi}$ must satisfy
critical magnitude constraints.

Using Eq. (\ref{Dual Eigen-coordinate Locations Component One Q}), it follows
that the integrated lengths $\sum\nolimits_{i=1}^{l_{1}}\psi_{1i\ast}$ of the
$\psi_{1i\ast}\overrightarrow{\mathbf{e}}_{1i\ast}$ components on
$\boldsymbol{\psi}_{1}$ must satisfy the equation:%
\begin{align}
\sum\nolimits_{i=1}^{l_{1}}\psi_{1i\ast}  &  =\lambda_{\max_{\boldsymbol{\psi
}}}^{-1}\sum\nolimits_{i=1}^{l_{1}}\left\Vert k_{\mathbf{x}_{1_{i\ast}}%
}\right\Vert \label{integrated dual loci one1 Q}\\
&  \times\sum\nolimits_{j=1}^{l_{1}}\psi_{1_{j\ast}}\left\Vert k_{\mathbf{x}%
_{1_{j\ast}}}\right\Vert \cos\theta_{k_{\mathbf{x}_{1_{i\ast}}}k_{\mathbf{x}%
_{1_{j\ast}}}}\nonumber\\
&  -\lambda_{\max_{\boldsymbol{\psi}}}^{-1}\sum\nolimits_{i=1}^{l_{1}%
}\left\Vert k_{\mathbf{x}_{1_{i\ast}}}\right\Vert \nonumber\\
&  \times\sum\nolimits_{j=1}^{l_{2}}\psi_{2_{j\ast}}\left\Vert k_{\mathbf{x}%
_{2_{j\ast}}}\right\Vert \cos\theta_{k_{\mathbf{x}_{1_{i\ast}}}k_{\mathbf{x}%
_{2_{j\ast}}}}\nonumber
\end{align}
which reduces to%
\[
\sum\nolimits_{i=1}^{l_{1}}\psi_{1i\ast}=\lambda_{\max_{\boldsymbol{\psi}}%
}^{-1}\sum\nolimits_{i=1}^{l_{1}}k_{\mathbf{x}_{1_{i\ast}}}^{T}\left(
\sum\nolimits_{j=1}^{l_{1}}\psi_{1_{j\ast}}k_{\mathbf{x}_{1_{j\ast}}}%
-\sum\nolimits_{j=1}^{l_{2}}\psi_{2_{j\ast}}k_{\mathbf{x}_{2_{j\ast}}}\right)
\text{.}%
\]
Using Eq. (\ref{Dual Eigen-coordinate Locations Component Two Q}), it follows
that the integrated lengths $\sum\nolimits_{i=1}^{l_{2}}\psi_{2i\ast}$ of the
$\psi_{2i\ast}\overrightarrow{\mathbf{e}}_{2i\ast}$ components on
$\boldsymbol{\psi}_{2}$ must satisfy the equation:%
\begin{align}
\sum\nolimits_{i=1}^{l_{2}}\psi_{2i\ast}  &  =\lambda_{\max_{\boldsymbol{\psi
}}}^{-1}\sum\nolimits_{i=1}^{l_{2}}\left\Vert k_{\mathbf{x}_{2_{i\ast}}%
}\right\Vert \label{integrated dual loci two1 Q}\\
&  \times\sum\nolimits_{j=1}^{l_{2}}\psi_{2_{j\ast}}\left\Vert k_{\mathbf{x}%
_{2_{j\ast}}}\right\Vert \cos\theta_{k_{\mathbf{x}_{2_{i\ast}}}k_{\mathbf{x}%
_{2_{j\ast}}}}\nonumber\\
&  -\lambda_{\max_{\boldsymbol{\psi}}}^{-1}\sum\nolimits_{i=1}^{l_{2}%
}\left\Vert k_{\mathbf{x}_{2_{i\ast}}}\right\Vert \nonumber\\
&  \times\sum\nolimits_{j=1}^{l_{1}}\psi_{1_{j\ast}}\left\Vert k_{\mathbf{x}%
_{1_{j\ast}}}\right\Vert \cos\theta_{k_{\mathbf{x}_{2_{i\ast}}}k_{\mathbf{x}%
_{1_{j\ast}}}}\nonumber
\end{align}
which reduces to%
\[
\sum\nolimits_{i=1}^{l_{2}}\psi_{2i\ast}=\lambda_{\max_{\boldsymbol{\psi}}%
}^{-1}\sum\nolimits_{i=1}^{l_{2}}k_{\mathbf{x}_{2_{i\ast}}}^{T}\left(
\sum\nolimits_{j=1}^{l_{2}}\psi_{2_{j\ast}}k_{\mathbf{x}_{2_{j\ast}}}%
-\sum\nolimits_{j=1}^{l_{1}}\psi_{1_{j\ast}}k_{\mathbf{x}_{1_{j\ast}}}\right)
\text{.}%
\]

I will now show that Eqs (\ref{integrated dual loci one1 Q}) and
(\ref{integrated dual loci two1 Q}) determine a balanced eigenlocus equation,
where RHS Eq. (\ref{integrated dual loci one1 Q}) $=$ RHS Eq.
(\ref{integrated dual loci two1 Q}).

\subsection{Balanced Quadratic Eigenlocus Equations}

Returning to Eq. (\ref{Equilibrium Constraint on Dual Eigen-components Q})%
\[
\sum\nolimits_{i=1}^{l_{1}}\psi_{1_{i\ast}}=\sum\nolimits_{i=1}^{l_{2}}%
\psi_{2_{i\ast}}\text{,}%
\]
where the axis of $\boldsymbol{\psi}$ is in statistical equilibrium, it
follows that the RHS\ of Eq. (\ref{integrated dual loci one1 Q}) must equal
the RHS\ of Eq. (\ref{integrated dual loci two1 Q}):%
\begin{align}
&  \sum\nolimits_{i=1}^{l_{1}}k_{\mathbf{x}_{1_{i\ast}}}^{T}\left(
\sum\nolimits_{j=1}^{l_{1}}\psi_{1_{j\ast}}k_{\mathbf{x}_{1_{j\ast}}}%
-\sum\nolimits_{j=1}^{l_{2}}\psi_{2_{j\ast}}k_{\mathbf{x}_{2_{j\ast}}}\right)
\label{Balanced Eigenlocus Equation Quadratic}\\
&  =\sum\nolimits_{i=1}^{l_{2}}k_{\mathbf{x}_{2_{i\ast}}}^{T}\left(
\sum\nolimits_{j=1}^{l_{2}}\psi_{2_{j\ast}}k_{\mathbf{x}_{2_{j\ast}}}%
-\sum\nolimits_{j=1}^{l_{1}}\psi_{1_{j\ast}}k_{\mathbf{x}_{1_{j\ast}}}\right)
\text{.}\nonumber
\end{align}

Therefore, all of the $k_{\mathbf{x}_{1_{i\ast}}}$ and $k_{\mathbf{x}%
_{2_{i\ast}}}$ extreme points are distributed over the axes of
$\boldsymbol{\kappa}_{1}$\textbf{ }and $\boldsymbol{\kappa}_{2}$ in the
symmetrically balanced manner:%
\begin{equation}
\sum\nolimits_{i=1}^{l_{1}}k_{\mathbf{x}_{1_{i\ast}}}^{T}\left(
\boldsymbol{\kappa}_{1}\mathbf{-}\boldsymbol{\kappa}_{2}\right)
=\sum\nolimits_{i=1}^{l_{2}}k_{\mathbf{x}_{2_{i\ast}}}^{T}\left(
\boldsymbol{\kappa}_{2}\mathbf{-}\boldsymbol{\kappa}_{1}\right)  \text{,}
\label{Balanced Eigenlocus Equation Q}%
\end{equation}
where the components of the $k_{\mathbf{x}_{1_{i\ast}}}$ extreme vectors along
the axis of $\boldsymbol{\kappa}_{2}$ oppose the components of the
$k_{\mathbf{x}_{1_{i\ast}}}$ extreme vectors along the axis of
$\boldsymbol{\kappa}_{1}$, and the components of the $k_{\mathbf{x}_{2_{i\ast
}}}$ extreme vectors along the axis of $\boldsymbol{\kappa}_{1}$ oppose the
components of the $k_{\mathbf{x}_{2_{i\ast}}}$ extreme vectors along the axis
of $\boldsymbol{\kappa}_{2}$.

Rewrite Eq. (\ref{Balanced Eigenlocus Equation Q}) as:%
\[
\sum\nolimits_{i=1}^{l_{1}}k_{\mathbf{x}_{1_{i\ast}}}\boldsymbol{\kappa}%
_{1}+\sum\nolimits_{i=1}^{l_{2}}k_{\mathbf{x}_{2_{i\ast}}}\boldsymbol{\kappa
}_{1}=\sum\nolimits_{i=1}^{l_{1}}k_{\mathbf{x}_{1_{i\ast}}}\boldsymbol{\kappa
}_{2}+\sum\nolimits_{i=1}^{l_{2}}k_{\mathbf{x}_{2_{i\ast}}}\boldsymbol{\kappa
}_{2}\text{,}%
\]
where components of the $k_{\mathbf{x}_{1_{i\ast}}}$ and $k_{\mathbf{x}%
_{2_{i\ast}}}$ extreme vectors along the axes of $\boldsymbol{\kappa}_{1}$ and
$\boldsymbol{\kappa}_{2}$ have forces associated with risks and counter risks
that are determined by expected values and spreads of $k_{\mathbf{x}%
_{1_{i\ast}}}$ and $k_{\mathbf{x}_{2_{i\ast}}}$ extreme points located in the
$Z_{1}$ and $Z_{2}$ or decision regions. It follows that, for any given
collection of extreme points drawn from any given statistical distribution,
the aggregate forces associated with the risks and the counter risks on the
axis of $\boldsymbol{\kappa}_{1}$ are balanced with the aggregate forces
associated with the risks and the counter risks on the axis of
$\boldsymbol{\kappa}_{2}$.

Let $\widehat{k}_{\mathbf{x}_{i\ast}}\triangleq\sum\nolimits_{i=1}%
^{l}k_{\mathbf{x}_{i\ast}}$, where $\sum\nolimits_{i=1}^{l}k_{\mathbf{x}%
_{i\ast}}=\sum\nolimits_{i=1}^{l_{1}}k_{\mathbf{x}_{1_{i\ast}}}+\sum
\nolimits_{i=1}^{l_{2}}k_{\mathbf{x}_{2_{i\ast}}}$. Using Eq.
(\ref{Balanced Eigenlocus Equation Q}), it follows that the component of
$\widehat{k}_{\mathbf{x}_{i\ast}}$ along $\boldsymbol{\kappa}_{1}$ is
symmetrically balanced with the component of $\widehat{k}_{\mathbf{x}_{i\ast}%
}$ along $\boldsymbol{\kappa}_{2}$%
\[
\operatorname{comp}_{\overrightarrow{\boldsymbol{\kappa}_{1}}}\left(
\overrightarrow{\widehat{k}_{\mathbf{x}_{i\ast}}}\right)  \rightleftharpoons
\operatorname{comp}_{\overrightarrow{\boldsymbol{\kappa}_{2}}}\left(
\overrightarrow{\widehat{k}_{\mathbf{x}_{i\ast}}}\right)
\]
so that the components $\operatorname{comp}%
_{\overrightarrow{\boldsymbol{\kappa}_{1}}}\left(  \overrightarrow{\widehat{k}%
_{\mathbf{x}_{i\ast}}}\right)  $ and $\operatorname{comp}%
_{\overrightarrow{\boldsymbol{\kappa}_{2}}}\left(  \overrightarrow{\widehat{k}%
_{\mathbf{x}_{i\ast}}}\right)  $ of clusters or aggregates of the extreme
vectors from both pattern classes have \emph{equal forces associated with
risks and counter risks} on opposite sides of the axis of $\boldsymbol{\kappa
}$.

\subsubsection{Statistical Equilibrium of Risks and Counter Risks}

\begin{figure}[ptb]%
\centering
\fbox{\includegraphics[
height=2.5875in,
width=3.4411in
]%
{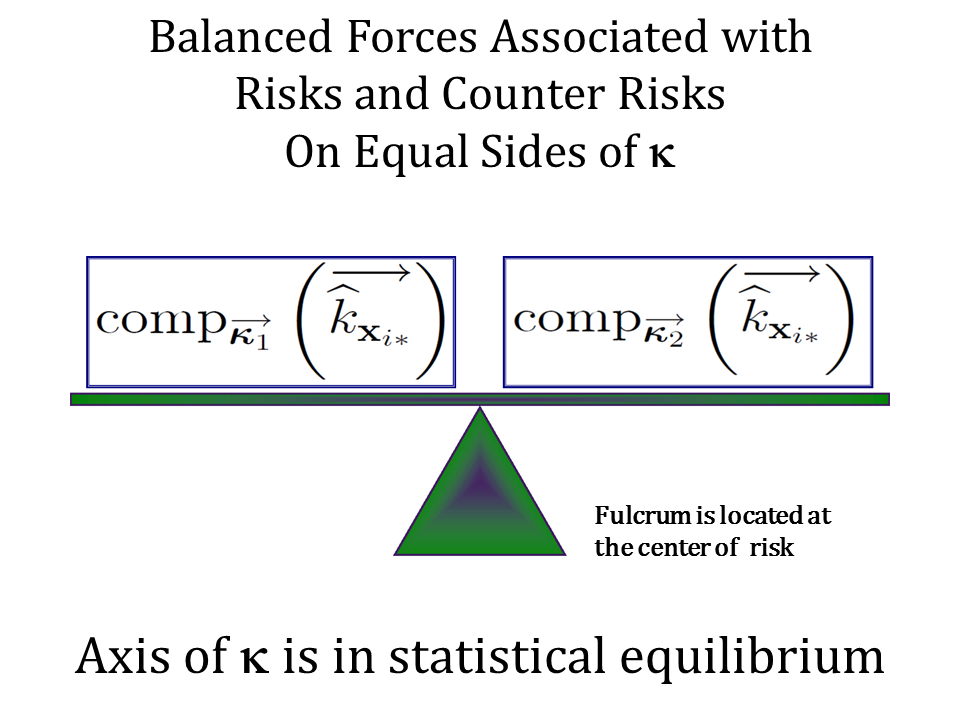}%
}\caption{The axis of $\boldsymbol{\kappa}$ is in statistical equilibrium,
where two equal weights $\operatorname{comp}%
_{\protect\overrightarrow{\boldsymbol{\kappa}_{1}}}\left(
\protect\overrightarrow{\protect\widehat{k}_{\mathbf{x}_{i\ast}}}\right)  $
and $\operatorname{comp}_{\protect\overrightarrow{\boldsymbol{\kappa}_{2}}%
}\left(  \protect\overrightarrow{\protect\widehat{k}_{\mathbf{x}_{i\ast}}%
}\right)  $ are placed on opposite sides of the fulcrum of $\boldsymbol{\kappa
}$ which is located at the center of total allowed eigenenergy $\left\Vert
\boldsymbol{\kappa}_{1}-\boldsymbol{\kappa}_{2}\right\Vert _{\min_{c}}^{2}$ of
$\boldsymbol{\kappa}$ .}%
\label{Statistical Equilibrium of Primal Quadratic Eigenlocus}%
\end{figure}

Given Eq. (\ref{Balanced Eigenlocus Equation Q}), it follows that the axis of
$\boldsymbol{\kappa}$ is a lever of uniform density, where the center of
$\boldsymbol{\kappa}$ is $\left\Vert \boldsymbol{\kappa}\right\Vert _{\min
_{c}}^{2}$, for which two equal weights $\operatorname{comp}%
_{\overrightarrow{\boldsymbol{\kappa}_{1}}}\left(  \overrightarrow{\widehat{k}%
_{\mathbf{x}_{i\ast}}}\right)  $ and $\operatorname{comp}%
_{\overrightarrow{\boldsymbol{\kappa}_{2}}}\left(  \overrightarrow{\widehat{k}%
_{\mathbf{x}_{i\ast}}}\right)  $ are placed on opposite sides of the fulcrum
of $\boldsymbol{\kappa}$, whereby the axis of $\boldsymbol{\kappa}$ is in
\emph{statistical equilibrium}. Figure
$\ref{Statistical Equilibrium of Primal Quadratic Eigenlocus}$ illustrates the
axis of $\boldsymbol{\kappa}$ in statistical equilibrium, where all of the
forces associated with the counter risks and the risks of aggregates of
extreme points are symmetrically balanced with each other.%

\subsubsection{Critical Magnitude Constraints}

Equation (\ref{Balanced Eigenlocus Equation Q}) indicates that the lengths%
\[
\left\{  \psi_{1_{i\ast}}|\psi_{1_{i\ast}}>0\right\}  _{i=1}^{l_{1}}\text{ and
}\left\{  \psi_{2_{i\ast}}|\psi_{2_{i\ast}}>0\right\}  _{i=1}^{l_{2}}%
\]
of the $l$ Wolfe dual principal eigenaxis components on $\boldsymbol{\psi}$
satisfy critical magnitude constraints, such that the Wolfe dual eigensystem
in Eq. (\ref{Dual Normal Eigenlocus Components Q}), which specifies highly
interconnected, balanced sets of inner product relationships amongst the Wolfe
dual and the constrained, primal principal eigenaxis components in Eqs
(\ref{integrated dual loci one1 Q}) and (\ref{integrated dual loci two1 Q}),
determines well-proportioned lengths $\psi_{1i\ast}$ or $\psi_{2i\ast}$ for
each Wolfe dual principal eigenaxis component $\psi_{1i\ast}\frac
{k_{\mathbf{x}_{1_{i_{\ast}}}}}{\left\Vert k_{\mathbf{x}_{1_{i_{\ast}}}%
}\right\Vert }$ or $\psi_{2i\ast}\frac{k_{\mathbf{x}_{2_{i_{\ast}}}}%
}{\left\Vert k_{\mathbf{x}_{2_{i_{\ast}}}}\right\Vert }$ on $\boldsymbol{\psi
}_{1}$ or $\boldsymbol{\psi}_{2}$, where each scale factor $\psi_{1i\ast}$ or
$\psi_{2i\ast}$ determines a well-proportioned length for a correlated,
constrained primal principal eigenaxis component $\psi_{1_{i\ast}%
}k_{\mathbf{x}_{1_{i\ast}}}$ or $\psi_{2_{i\ast}}k_{\mathbf{x}_{2_{i\ast}}}$
on $\boldsymbol{\kappa}_{1}$ or $\boldsymbol{\kappa}_{2}$.

I will demonstrate that the axis of $\boldsymbol{\psi}$, which is constrained
to be in statistical equilibrium $\sum\nolimits_{i=1}^{l_{1}}\psi_{1i\ast
}\frac{k_{\mathbf{x}_{1_{i_{\ast}}}}}{\left\Vert k_{\mathbf{x}_{1_{i_{\ast}}}%
}\right\Vert }=\sum\nolimits_{i=1}^{l_{2}}\psi_{2i\ast}\frac{k_{\mathbf{x}%
_{2_{i_{\ast}}}}}{\left\Vert k_{\mathbf{x}_{2_{i_{\ast}}}}\right\Vert }$,
determines an equilibrium point $p\left(  \widehat{\Lambda}_{\boldsymbol{\psi
}}\left(  \mathbf{s}\right)  |\omega_{1}\right)  -p\left(  \widehat{\Lambda
}_{\boldsymbol{\psi}}\left(  \mathbf{s}\right)  |\omega_{2}\right)  =0$ of an
integral equation $f\left(  \widetilde{\Lambda}_{\boldsymbol{\kappa}}\left(
\mathbf{s}\right)  \right)  $ such that a quadratic eigenlocus discriminant
function $\widetilde{\Lambda}_{\boldsymbol{\kappa}}\left(  \mathbf{s}\right)
=\left(  \mathbf{x}^{T}\mathbf{s}+1\right)  ^{2}\boldsymbol{\kappa}+\kappa
_{0}$ is the solution to a fundamental integral equation of binary
classification for a classification system $\left(  \mathbf{x}^{T}%
\mathbf{s}+1\right)  ^{2}\boldsymbol{\kappa}+\kappa_{0}\overset{\omega
_{1}}{\underset{\omega_{2}}{\gtrless}}0$ in statistical equilibrium.

Let $\mathfrak{R}_{\mathfrak{\min}}\left(  Z|\mathbf{\kappa}\right)  $ denote
the expected risk of a quadratic classification system $\left(  \mathbf{x}%
^{T}\mathbf{s}+1\right)  ^{2}\boldsymbol{\kappa}+\kappa_{0}\overset{\omega
_{1}}{\underset{\omega_{2}}{\gtrless}}0$ that is determined by a quadratic
eigenlocus transform. Take any given set $\left\{  \left\{  k_{\mathbf{x}%
_{1_{i_{\ast}}}}\right\}  _{i=1}^{l_{1}},\;\left\{  k_{\mathbf{x}_{2_{i_{\ast
}}}}\right\}  _{i=1}^{l_{2}}\right\}  $ of extreme points and take the set of
scale factors $\left\{  \left\{  \psi_{1i\ast}\right\}  _{i=1}^{l_{1}%
},\left\{  \psi_{2i\ast}\right\}  _{i=1}^{l_{2}}\right\}  $ that are
determined by a quadratic eigenlocus transform.

Let $\overleftrightarrow{\mathfrak{R}}_{\mathfrak{\min}}\left(  Z|\psi
_{1_{j\ast}}k_{\mathbf{x}_{1j\ast}}\right)  $ denote the force associated with
either the counter risk or the risk that is related to the locus of the scaled
extreme vector $\psi_{1_{j\ast}}k_{\mathbf{x}_{1j\ast}}$ in the decision space
$Z$, where the force associated with the expected risk $\mathfrak{R}%
_{\mathfrak{\min}}\left(  Z|\mathbf{\kappa}\right)  $ may be positive or
negative. Let $\overleftrightarrow{\mathfrak{R}}_{\mathfrak{\min}}\left(
Z|\psi_{2_{j\ast}}k_{\mathbf{x}_{2_{j\ast}}}\right)  $ denote the force
associated with either the counter risk or the risk that is related to the
locus of the scaled extreme vector $\psi_{2_{j\ast}}k_{\mathbf{x}_{2_{j\ast}}%
}$ in the decision space $Z$, where the force associated with the expected
risk $\mathfrak{R}_{\mathfrak{\min}}\left(  Z|\mathbf{\kappa}\right)  $ may be
positive or negative.

Take any given extreme point $k_{\mathbf{x}_{1_{i_{\ast}}}}$ from class
$\omega_{1}$. Let $\overleftrightarrow{\mathfrak{R}}_{\mathfrak{\min}}\left(
Z|\psi_{1_{j\ast}}k_{\mathbf{x}_{1_{i_{\ast}}}}^{T}k_{\mathbf{x}_{1j\ast}%
}\right)  $ denote the force associated with either the counter risk or the
risk that is related to the locus of the component of the extreme vector
$k_{\mathbf{x}_{1_{i_{\ast}}}}$ along the scaled extreme vector $\psi
_{1_{j\ast}}k_{\mathbf{x}_{1j\ast}}$ in the decision space $Z$, where the
force associated with the expected risk $\mathfrak{R}_{\mathfrak{\min}}\left(
Z|\mathbf{\kappa}\right)  $ may be positive or negative. Likewise, let
$\overleftrightarrow{\mathfrak{R}}_{\mathfrak{\min}}\left(  Z|\psi_{2_{j\ast}%
}k_{\mathbf{x}_{1_{i_{\ast}}}}^{T}k_{\mathbf{x}_{2_{j\ast}}}\right)  $ denote
the force associated with either the counter risk or the risk that is related
to the locus of the component of the extreme vector $k_{\mathbf{x}%
_{1_{i_{\ast}}}}$ along the scaled extreme vector $\psi_{2_{j\ast}%
}k_{\mathbf{x}_{2_{j\ast}}}$ in the decision space $Z$, where the force
associated with the expected risk $\mathfrak{R}_{\mathfrak{\min}}\left(
Z|\mathbf{\kappa}\right)  $ may be positive or negative.

Take any given extreme point $k_{\mathbf{x}_{2_{i_{\ast}}}}$ from class
$\omega_{2}$. Let $\overleftrightarrow{\mathfrak{R}}_{\mathfrak{\min}}\left(
Z|\psi_{2_{j\ast}}k_{\mathbf{x}_{2_{i_{\ast}}}}^{T}k_{\mathbf{x}_{2_{j\ast}}%
}\right)  $ denote the force associated with either the counter risk or the
risk that is related to the locus of the component of the extreme vector
$k_{\mathbf{x}_{2_{i_{\ast}}}}$ along the scaled extreme vector $\psi
_{2_{j\ast}}k_{\mathbf{x}_{2_{j\ast}}}$ in the decision space $Z$, where the
force associated with the expected risk $\mathfrak{R}_{\mathfrak{\min}}\left(
Z|\mathbf{\kappa}\right)  $ may be positive or negative. Likewise, let
$\overleftrightarrow{\mathfrak{R}}_{\mathfrak{\min}}\left(  Z|\psi_{1_{j\ast}%
}k_{\mathbf{x}_{2_{i_{\ast}}}}^{T}k_{\mathbf{x}_{1j\ast}}\right)  $ denote the
force associated with either the counter risk or the risk that is related to
the locus of the component of the extreme vector $k_{\mathbf{x}_{2_{i_{\ast}}%
}}$ along the scaled extreme vector $\psi_{1_{j\ast}}k_{\mathbf{x}_{1j\ast}}$
in the decision space $Z$, where the force associated with the expected risk
$\mathfrak{R}_{\mathfrak{\min}}\left(  Z|\mathbf{\kappa}\right)  $ may be
positive or negative.

Returning to Eq. (\ref{Balanced Eigenlocus Equation Quadratic})%
\begin{align*}
&  \sum\nolimits_{i=1}^{l_{1}}k_{\mathbf{x}_{1_{i\ast}}}^{T}\left(
\sum\nolimits_{j=1}^{l_{1}}\psi_{1_{j\ast}}k_{\mathbf{x}_{1_{j\ast}}}%
-\sum\nolimits_{j=1}^{l_{2}}\psi_{2_{j\ast}}k_{\mathbf{x}_{2_{j\ast}}}\right)
\\
&  =\sum\nolimits_{i=1}^{l_{2}}k_{\mathbf{x}_{2_{i\ast}}}^{T}\left(
\sum\nolimits_{j=1}^{l_{2}}\psi_{2_{j\ast}}k_{\mathbf{x}_{2_{j\ast}}}%
-\sum\nolimits_{j=1}^{l_{1}}\psi_{1_{j\ast}}k_{\mathbf{x}_{1_{j\ast}}}\right)
\text{.}%
\end{align*}
it follows that the collective forces associated with the risks and the
counter risks that are related to the positions and the potential locations of
all of the extreme points are balanced in the following manner:%
\begin{align*}
\mathfrak{R}_{\mathfrak{\min}}\left(  Z|\mathbf{\kappa}\right)   &
:\sum\nolimits_{i=1}^{l_{1}}\left[  \sum\nolimits_{j=1}^{l_{1}}%
\overleftrightarrow{\mathfrak{R}}_{\mathfrak{\min}}\left(  Z|\psi_{1_{j\ast}%
}k_{\mathbf{x}_{1_{i_{\ast}}}}^{T}k_{\mathbf{x}_{1j\ast}}\right)
-\sum\nolimits_{j=1}^{l_{2}}\overleftrightarrow{\mathfrak{R}}_{\mathfrak{\min
}}\left(  Z|\psi_{2_{j\ast}}k_{\mathbf{x}_{1_{i_{\ast}}}}^{T}k_{\mathbf{x}%
_{2_{j\ast}}}\right)  \right] \\
&  =\sum\nolimits_{i=1}^{l_{2}}\left[  \sum\nolimits_{j=1}^{l_{2}%
}\overleftrightarrow{\mathfrak{R}}_{\mathfrak{\min}}\left(  Z|\psi_{2_{j\ast}%
}k_{\mathbf{x}_{2_{i_{\ast}}}}^{T}k_{\mathbf{x}_{2_{j\ast}}}\right)
-\sum\nolimits_{j=1}^{l_{1}}\overleftrightarrow{\mathfrak{R}}_{\mathfrak{\min
}}\left(  Z|\psi_{1_{j\ast}}k_{\mathbf{x}_{2_{i_{\ast}}}}^{T}k_{\mathbf{x}%
_{1j\ast}}\right)  \right]  \text{.}%
\end{align*}

So, take any given set $\left\{  \left\{  k_{\mathbf{x}_{1_{i_{\ast}}}%
}\right\}  _{i=1}^{l_{1}},\;\left\{  k_{\mathbf{x}_{2_{i_{\ast}}}}\right\}
_{i=1}^{l_{2}}\right\}  $ of extreme points and take the set of scale factors
$\left\{  \left\{  \psi_{1i\ast}\right\}  _{i=1}^{l_{1}},\left\{  \psi
_{2i\ast}\right\}  _{i=1}^{l_{2}}\right\}  $ that are determined by a
quadratic eigenlocus transform.

I\ will show that quadratic eigenlocus transforms choose magnitudes or scale
factors for the Wolfe dual principal eigenaxis components:%
\[
\left\{  \psi_{1i\ast}\frac{k_{\mathbf{x}_{1_{i_{\ast}}}}}{\left\Vert
k_{\mathbf{x}_{1_{i_{\ast}}}}\right\Vert }\right\}  _{i=1}^{l_{1}}\text{ and
}\left\{  \psi_{2i\ast}\frac{k_{\mathbf{x}_{2_{i_{\ast}}}}}{\left\Vert
k_{\mathbf{x}_{2_{i_{\ast}}}}\right\Vert }\right\}  _{i=1}^{l_{2}}%
\]
on $\mathbf{\psi}$, which is constrained to satisfy the equation of
statistical equilibrium:%
\[
\sum\nolimits_{i=1}^{l_{1}}\psi_{1i\ast}\frac{k_{\mathbf{x}_{1_{i_{\ast}}}}%
}{\left\Vert k_{\mathbf{x}_{1_{i_{\ast}}}}\right\Vert }=\sum\nolimits_{i=1}%
^{l_{2}}\psi_{2i\ast}\frac{k_{\mathbf{x}_{2_{i_{\ast}}}}}{\left\Vert
k_{\mathbf{x}_{2_{i_{\ast}}}}\right\Vert }\text{,}%
\]
such that the likelihood ratio $\widehat{\Lambda}_{\boldsymbol{\kappa}}\left(
\mathbf{s}\right)  =\boldsymbol{\kappa}_{1}-\boldsymbol{\kappa}_{2}$ and the
classification system $\left(  \mathbf{x}^{T}\mathbf{s}+1\right)
^{2}\boldsymbol{\kappa}+\kappa_{0}\overset{\omega_{1}}{\underset{\omega
_{2}}{\gtrless}}0$ are in statistical equilibrium, and the expected risk
$\mathfrak{R}_{\mathfrak{\min}}\left(  Z|\boldsymbol{\kappa}\right)  $ and the
corresponding total allowed eigenenergies $\left\Vert \boldsymbol{\kappa}%
_{1}-\boldsymbol{\kappa}_{2}\right\Vert _{\min_{c}}^{2}$ exhibited by the
classification system $\left(  \mathbf{x}^{T}\mathbf{s}+1\right)
^{2}\boldsymbol{\kappa}+\kappa_{0}\overset{\omega_{1}}{\underset{\omega
_{2}}{\gtrless}}0$ are minimized.

In the next section, I\ will explicitly define the manner in which
constrained, quadratic eigenlocus discriminant functions $\widetilde{\Lambda
}_{\boldsymbol{\kappa}}\left(  \mathbf{s}\right)  =\left(  \mathbf{x}%
^{T}\mathbf{s}+1\right)  ^{2}\boldsymbol{\kappa}+\kappa_{0}$ satisfy quadratic
decision boundaries $D_{0}\left(  \mathbf{s}\right)  $ and quadratic decision
borders $D_{+1}\left(  \mathbf{s}\right)  $ and $D_{-1}\left(  \mathbf{s}%
\right)  $. I will use these results to show that the principal eigenaxis
$\boldsymbol{\kappa}$ of a quadratic eigenlocus discriminant functions
$\widetilde{\Lambda}_{\boldsymbol{\kappa}}\left(  \mathbf{s}\right)  =\left(
\mathbf{x}^{T}\mathbf{s}+1\right)  ^{2}\boldsymbol{\kappa}+\kappa_{0}$ is a
lever that is symmetrically balanced with respect to the center of eigenenergy
$\left\Vert \boldsymbol{\kappa}_{1}-\boldsymbol{\kappa}_{2}\right\Vert
_{\min_{c}}^{2}$ of $\boldsymbol{\kappa}$, such that the total allowed
eigenenergies of the scaled extreme vectors on $\boldsymbol{\kappa}%
_{1}-\boldsymbol{\kappa}_{2}$ are symmetrically balanced about the fulcrum
$\left\Vert \boldsymbol{\kappa}_{1}-\boldsymbol{\kappa}_{2}\right\Vert
_{\min_{c}}^{2}$ of $\boldsymbol{\kappa}$. Thereby, I\ will show that the
likelihood ratio $\widehat{\Lambda}_{\boldsymbol{\kappa}}\left(
\mathbf{s}\right)  =\boldsymbol{\kappa}_{1}-\boldsymbol{\kappa}_{2}$ and the
classification system $\left(  \mathbf{x}^{T}\mathbf{s}+1\right)
^{2}\boldsymbol{\kappa}+\kappa_{0}\overset{\omega_{1}}{\underset{\omega
_{2}}{\gtrless}}0$ are in statistical equilibrium.

I\ will use all of these results to identify the manner in which the property
of symmetrical balance exhibited by the principal eigenaxis components on
$\boldsymbol{\psi}$ and $\boldsymbol{\kappa}$ enables quadratic eigenlocus
classification systems $\left(  \mathbf{x}^{T}\mathbf{s}+1\right)
^{2}\boldsymbol{\kappa}+\kappa_{0}\overset{\omega_{1}}{\underset{\omega
_{2}}{\gtrless}}0$ to effectively balance all of the forces associated with
the counter risk $\overline{\mathfrak{R}}_{\mathfrak{\min}}\left(
Z_{1}|\boldsymbol{\kappa}_{1}\right)  $ and the risk $\mathfrak{R}%
_{\mathfrak{\min}}\left(  Z_{1}|\boldsymbol{\kappa}_{2}\right)  $ in the
$Z_{1}$ decision region with all of the forces associated with the counter
risk $\overline{\mathfrak{R}}_{\mathfrak{\min}}\left(  Z_{2}%
|\boldsymbol{\kappa}_{2}\right)  $ and the risk $\mathfrak{R}_{\mathfrak{\min
}}\left(  Z_{2}|\boldsymbol{\kappa}_{1}\right)  $ in the $Z_{2}$ decision
region:%
\begin{align*}
f\left(  \widetilde{\Lambda}_{\boldsymbol{\kappa}}\left(  \mathbf{s}\right)
\right)   &  :\int\nolimits_{Z_{1}}p\left(  k_{\mathbf{x}_{1i\ast}%
}|\boldsymbol{\kappa}_{1}\right)  d\boldsymbol{\kappa}_{1}-\int%
\nolimits_{Z_{1}}p\left(  k_{\mathbf{x}_{2i\ast}}|\boldsymbol{\kappa}%
_{2}\right)  d\boldsymbol{\kappa}_{2}+\delta\left(  y\right)  \boldsymbol{\psi
}_{1}\\
&  =\int\nolimits_{Z_{2}}p\left(  k_{\mathbf{x}_{2i\ast}}|\boldsymbol{\kappa
}_{2}\right)  d\boldsymbol{\kappa}_{2}-\int\nolimits_{Z_{2}}p\left(
k_{\mathbf{x}_{1i\ast}}|\boldsymbol{\kappa}_{1}\right)  d\boldsymbol{\kappa
}_{1}-\delta\left(  y\right)  \boldsymbol{\psi}_{2}\text{,}%
\end{align*}
where $\delta\left(  y\right)  \triangleq\sum\nolimits_{i=1}^{l}y_{i}\left(
1-\xi_{i}\right)  $, and $Z_{1}$ and $Z_{2}$ are symmetrical decision regions
$Z_{1}\simeq Z_{2}$, given the equilibrium point $\boldsymbol{\psi}%
_{1}-\boldsymbol{\psi}_{2}=0$ and the class-conditional probability density
functions $p\left(  k_{\mathbf{x}_{1i\ast}}|\boldsymbol{\kappa}_{1}\right)  $
and $p\left(  k_{\mathbf{x}_{2i\ast}}|\boldsymbol{\kappa}_{2}\right)  $, where
the areas under the probability density functions $p\left(  k_{\mathbf{x}%
_{1i\ast}}|\boldsymbol{\kappa}_{1}\right)  $ and $p\left(  k_{\mathbf{x}%
_{2i\ast}}|\boldsymbol{\kappa}_{2}\right)  $ are symmetrically balanced with
each other over the $Z_{1}$ and $Z_{2}$ decision regions.

\section{Risk Minimization for Quadratic Classifiers}

In the next two sections, I will show that the conditional probability
function $P\left(  k_{\mathbf{x}_{1_{i\ast}}}|\boldsymbol{\kappa}_{1}\right)
$ for class $\omega_{1}$, which is given by the integral%
\[
P\left(  k_{\mathbf{x}_{1_{i\ast}}}|\boldsymbol{\kappa}_{1}\right)  =\int%
_{Z}p\left(  k_{\mathbf{x}_{1i\ast}}|\boldsymbol{\kappa}_{1}\right)
d\boldsymbol{\kappa}_{1}=\left\Vert \boldsymbol{\kappa}_{1}\right\Vert
_{\min_{c}}^{2}+C_{1}\text{,}%
\]
over the decision space $Z$, and the conditional probability function
$P\left(  k_{\mathbf{x}_{2_{i\ast}}}|\boldsymbol{\kappa}_{2}\right)  $ for
class $\omega_{2}$, which is given by the integral%
\[
P\left(  k_{\mathbf{x}_{2_{i\ast}}}|\boldsymbol{\kappa}_{2}\right)  =\int%
_{Z}p\left(  k_{\mathbf{x}_{2i\ast}}|\boldsymbol{\kappa}_{2}\right)
d\boldsymbol{\kappa}_{2}=\left\Vert \boldsymbol{\kappa}_{2}\right\Vert
_{\min_{c}}^{2}+C_{2}\text{,}%
\]
over the decision space $Z$, satisfy an integral equation where the area under
the probability density function $p\left(  k_{\mathbf{x}_{1i\ast}%
}|\boldsymbol{\kappa}_{1}\right)  $ for class $\omega_{1}$ is
\emph{symmetrically balanced with} the area under the probability density
function $p\left(  k_{\mathbf{x}_{2i\ast}}|\boldsymbol{\kappa}_{2}\right)  $
for class $\omega_{2}$%
\[
f\left(  \widetilde{\Lambda}_{\boldsymbol{\kappa}}\left(  \mathbf{s}\right)
\right)  :\int_{Z}\boldsymbol{\kappa}_{1}d\boldsymbol{\kappa}_{1}+\nabla
_{eq}\equiv\int_{Z}\boldsymbol{\kappa}_{2}d\boldsymbol{\kappa}_{2}-\nabla
_{eq}\text{,}%
\]
where $\nabla_{eq}$ is an equalizer statistic, such that the likelihood ratio
$\widehat{\Lambda}_{\boldsymbol{\kappa}}\left(  \mathbf{s}\right)
=\boldsymbol{\kappa}_{1}-\boldsymbol{\kappa}_{2}$ and the classification
system $\left(  \mathbf{x}^{T}\mathbf{s}+1\right)  ^{2}\boldsymbol{\kappa
}+\kappa_{0}\overset{\omega_{1}}{\underset{\omega_{2}}{\gtrless}}0$ are in
statistical equilibrium.

Accordingly, I\ will formulate a system of data-driven, locus equations that
determines the total allowed eigenenergies $\left\Vert \boldsymbol{\kappa}%
_{1}\right\Vert _{\min_{c}}^{2}$ and $\left\Vert \boldsymbol{\kappa}%
_{2}\right\Vert _{\min_{c}}^{2}$ exhibited by $\boldsymbol{\kappa}_{1}$ and
$\boldsymbol{\kappa}_{2}$, and I will derive values for the integration
constants $C_{1}$ and $C_{2}$. I will use these results to devise an equalizer
statistic $\nabla_{eq}$ for an integral equation that is satisfied by the
class-conditional probability density functions $p\left(  k_{\mathbf{x}%
_{1i\ast}}|\boldsymbol{\kappa}_{1}\right)  $ and $p\left(  k_{\mathbf{x}%
_{2i\ast}}|\boldsymbol{\kappa}_{2}\right)  $.

I\ will now devise a system of data-driven, locus equations that determines
the manner in which the total allowed eigenenergies of the scaled extreme
points on $\boldsymbol{\kappa}_{1}-\boldsymbol{\kappa}_{2}$ are symmetrically
balanced about the fulcrum $\left\Vert \boldsymbol{\kappa}\right\Vert
_{\min_{c}}^{2}$ of $\boldsymbol{\kappa}$. Accordingly, I will devise three
systems of data-driven, locus equations that explicitly determine the total
allowed eigenenergy $\left\Vert \boldsymbol{\kappa}_{1}\right\Vert _{\min_{c}%
}^{2}$ exhibited by $\boldsymbol{\kappa}_{1}$, the total allowed eigenenergy
$\left\Vert \boldsymbol{\kappa}_{2}\right\Vert _{\min_{c}}^{2}$ exhibited by
$\boldsymbol{\kappa}_{2}$, and the total allowed eigenenergy $\left\Vert
\boldsymbol{\kappa}\right\Vert _{\min_{c}}^{2}$ exhibited by
$\boldsymbol{\kappa}$.

\subsection{Critical Minimum Eigenenergy Constraints II}

Let there be $l$ labeled, scaled reproducing kernels of extreme points on a
quadratic eigenlocus $\boldsymbol{\kappa}$. Given the theorem of Karush, Kuhn,
and Tucker and the KKT condition in Eq. (\ref{KKTE5 Q}), it follows that a
Wolf dual quadratic eigenlocus $\boldsymbol{\psi}$ exists, for which%
\[
\left\{  \psi_{i\ast}>0\right\}  _{i=1}^{l}\text{,}%
\]
such that the $l$ constrained, primal principal eigenaxis components $\left\{
\psi_{i_{\ast}}k_{\mathbf{x}_{i_{\ast}}}\right\}  _{i=1}^{l}$ on
$\boldsymbol{\kappa}$ satisfy a system of $l$ eigenlocus equations:%
\begin{equation}
\psi_{i\ast}\left[  y_{i}\left(  \left(  \mathbf{x}^{T}\mathbf{x}_{i\ast
}+1\right)  ^{2}\boldsymbol{\kappa}+\kappa_{0}\right)  -1+\xi_{i}\right]
=0,\ i=1,...,l\text{.} \label{Minimum Eigenenergy Functional System Q}%
\end{equation}

I\ will now use Eq. (\ref{Minimum Eigenenergy Functional System Q}) to define
critical minimum eigenenergy constraints on $\boldsymbol{\kappa}_{1}$ and
$\boldsymbol{\kappa}_{2}$. The analysis begins with the critical minimum
eigenenergy constraint on $\boldsymbol{\kappa}_{1}$.

\subsubsection{Total Allowed Eigenenergy of $\boldsymbol{\kappa}_{1}$}

Take any scaled extreme vector $\psi_{1_{i_{\ast}}}k_{\mathbf{x}_{1_{i_{\ast}%
}}}$ that belongs to class $\omega_{1}$. Using Eq.
(\ref{Minimum Eigenenergy Functional System Q}) and letting $y_{i}=+1$, it
follows that the constrained, primal principal eigenaxis component
$\psi_{1_{i_{\ast}}}k_{\mathbf{x}_{1_{i_{\ast}}}}$ on $\boldsymbol{\kappa}%
_{1}$ is specified by the equation:%
\[
\psi_{1_{i_{\ast}}}k_{\mathbf{x}_{1_{i_{\ast}}}}\boldsymbol{\kappa}%
=\psi_{1_{i_{\ast}}}\left(  1-\xi_{i}-\kappa_{0}\right)
\]
which is part of a system of $l_{1}$ eigenlocus equations. Therefore, each
constrained, primal principal eigenaxis component $\psi_{1_{i_{\ast}}%
}k_{\mathbf{x}_{1_{i_{\ast}}}}$ on $\boldsymbol{\kappa}_{1}$ satisfies the
above locus equation.

Now take all of the $l_{1}$ scaled extreme vectors $\left\{  \psi_{1_{i_{\ast
}}}k_{\mathbf{x}_{1_{i_{\ast}}}}\right\}  _{i=1}^{l_{1}}$ that belong to class
$\omega_{1}$. Again, using Eq. (\ref{Minimum Eigenenergy Functional System Q})
and letting $y_{i}=+1$, it follows that the complete set $\left\{
\psi_{1_{i_{\ast}}}k_{\mathbf{x}_{1_{i_{\ast}}}}\right\}  _{i=1}^{l_{1}}$ of
$l_{1}$ constrained, primal principal eigenaxis components $\psi_{1_{i_{\ast}%
}}k_{\mathbf{x}_{1_{i_{\ast}}}}$ on $\boldsymbol{\kappa}_{1}$ is determined by
the system of $l_{1}$ equations:%
\begin{equation}
\psi_{1_{i_{\ast}}}k_{\mathbf{x}_{1_{i_{\ast}}}}\boldsymbol{\kappa}%
=\psi_{1_{i_{\ast}}}\left(  1-\xi_{i}-\kappa_{0}\right)  ,\ i=1,...,l_{1}%
\text{.} \label{Minimum Eigenenergy Class One Q}%
\end{equation}

Using Eq. (\ref{Minimum Eigenenergy Class One Q}), it follows that the entire
set $\left\{  \psi_{1_{i_{\ast}}}k_{\mathbf{x}_{1_{i_{\ast}}}}\right\}
_{i=1}^{l_{1}}$ of $l_{1}\times d$ transformed, extreme vector coordinates
satisfies the system of $l_{1}$ eigenlocus equations:%
\[
\text{ }(1)\text{ \ }\psi_{1_{1_{\ast}}}k_{\mathbf{x}_{1_{1_{\ast}}}%
}\boldsymbol{\kappa}=\psi_{1_{1_{\ast}}}\left(  1-\xi_{i}-\kappa_{0}\right)
\text{,}%
\]%
\[
\text{ }(2)\text{ \ }\psi_{1_{2_{\ast}}}k_{\mathbf{x}_{1_{2_{\ast}}}%
}\boldsymbol{\kappa}=\psi_{1_{2_{\ast}}}\left(  1-\xi_{i}-\kappa_{0}\right)
\text{,}%
\]

\[
\vdots
\]%
\[
(l_{1})\text{\ \ }\psi_{1_{l_{\ast}}}k_{\mathbf{x}_{1_{1_{\ast}}}%
}\boldsymbol{\kappa}=\psi_{1_{l_{\ast}}}\left(  1-\xi_{i}-\kappa_{0}\right)
\text{,}%
\]
where each constrained, primal principal eigenaxis component $\psi
_{1_{i_{\ast}}}k_{\mathbf{x}_{1_{i_{\ast}}}}$ on $\boldsymbol{\kappa}_{1}$
satisfies the identity:%
\[
\psi_{1_{i_{\ast}}}k_{\mathbf{x}_{1_{i_{\ast}}}}\boldsymbol{\kappa}%
=\psi_{1_{i_{\ast}}}\left(  1-\xi_{i}-\kappa_{0}\right)  \text{.}%
\]

I\ will now formulate an identity for the total allowed eigenenergy of
$\boldsymbol{\kappa}_{1}$. Let $E_{\boldsymbol{\kappa}_{1}}$ denote the
functional of the total allowed eigenenergy $\left\Vert \boldsymbol{\kappa
}_{1}\right\Vert _{\min_{c}}^{2}$ of $\boldsymbol{\kappa}_{1}$ and let
$\boldsymbol{\kappa}=\boldsymbol{\kappa}_{1}-\boldsymbol{\kappa}_{2}$.
Summation over the above system of $l_{1}$ eigenlocus equations produces the
following equation for the total allowed eigenenergy $\left\Vert
\boldsymbol{\kappa}_{1}\right\Vert _{\min_{c}}^{2}$ of $\boldsymbol{\kappa
}_{1}$:%
\[
\left(  \sum\nolimits_{i=1}^{l_{1}}\psi_{1_{i_{\ast}}}k_{\mathbf{x}%
_{1_{i_{\ast}}}}\right)  \left(  \boldsymbol{\kappa}_{1}-\boldsymbol{\kappa
}_{2}\right)  \equiv\sum\nolimits_{i=1}^{l_{1}}\psi_{1_{i_{\ast}}}\left(
1-\xi_{i}-\kappa_{0}\right)
\]
which reduces to%
\[
\boldsymbol{\kappa}_{1}^{T}\boldsymbol{\kappa}_{1}-\boldsymbol{\kappa}_{1}%
^{T}\boldsymbol{\kappa}_{2}\equiv\sum\nolimits_{i=1}^{l_{1}}\psi_{1_{i_{\ast}%
}}\left(  1-\xi_{i}-\kappa_{0}\right)
\]
so that the functional $E_{\boldsymbol{\kappa}_{1}}$ satisfies the identity%
\[
\left\Vert \boldsymbol{\kappa}_{1}\right\Vert _{\min_{c}}^{2}%
-\boldsymbol{\kappa}_{1}^{T}\boldsymbol{\kappa}_{2}\equiv\sum\nolimits_{i=1}%
^{l_{1}}\psi_{1_{i_{\ast}}}\left(  1-\xi_{i}-\kappa_{0}\right)  \text{.}%
\]

Therefore, the total allowed eigenenergy $\left\Vert \boldsymbol{\kappa}%
_{1}\right\Vert _{\min_{c}}^{2}$ exhibited by the constrained, primal
principal eigenlocus component $\boldsymbol{\kappa}_{1}$ is determined by the
identity%
\begin{equation}
\left\Vert \boldsymbol{\kappa}_{1}\right\Vert _{\min_{c}}^{2}-\left\Vert
\boldsymbol{\kappa}_{1}\right\Vert \left\Vert \boldsymbol{\kappa}%
_{2}\right\Vert \cos\theta_{\boldsymbol{\kappa}_{1}\boldsymbol{\kappa}_{2}%
}\equiv\sum\nolimits_{i=1}^{l_{1}}\psi_{1_{i_{\ast}}}\left(  1-\xi_{i}%
-\kappa_{0}\right)  \text{,} \label{TAE Eigenlocus Component One Q}%
\end{equation}
where the functional $E_{\boldsymbol{\kappa}_{1}}$ of the total allowed
eigenenergy $\left\Vert \boldsymbol{\kappa}_{1}\right\Vert _{\min_{c}}^{2}$
exhibited by $\boldsymbol{\kappa}_{1}$%
\[
E_{\boldsymbol{\kappa}_{1}}=\left\Vert \boldsymbol{\kappa}_{1}\right\Vert
_{\min_{c}}^{2}-\left\Vert \boldsymbol{\kappa}_{1}\right\Vert \left\Vert
\boldsymbol{\kappa}_{2}\right\Vert \cos\theta_{\boldsymbol{\kappa}%
_{1}\boldsymbol{\kappa}_{2}}%
\]
is equivalent to the functional $E_{\boldsymbol{\psi}_{1}}$%
\[
E_{\boldsymbol{\psi}_{1}}=\sum\nolimits_{i=1}^{l_{1}}\psi_{1_{i_{\ast}}%
}\left(  1-\xi_{i}-\kappa_{0}\right)
\]
of the integrated magnitudes $\sum\nolimits_{i=1}^{l_{1}}\psi_{1_{i_{\ast}}}$
of the Wolfe dual principal eigenaxis components $\psi_{1_{i_{\ast}}}%
\frac{k_{\mathbf{x}_{2_{i_{\ast}}}}}{\left\Vert k_{\mathbf{x}_{1_{i_{\ast}}}%
}\right\Vert }$ and the $\kappa_{0}$ statistic.

Returning to Eq. (\ref{Decision Border One Q}), it follows that the
functionals $E_{\boldsymbol{\kappa}_{1}}$ and $E_{\boldsymbol{\psi}_{1}}$
specify the manner in which quadratic eigenlocus discriminant functions
$\widetilde{\Lambda}_{\boldsymbol{\kappa}}\left(  \mathbf{s}\right)  =\left(
\mathbf{x}^{T}\mathbf{s}+1\right)  ^{2}\boldsymbol{\kappa}+\kappa_{0}$ satisfy
the quadratic decision border $D_{+1}\left(  \mathbf{s}\right)  $: $\left(
\mathbf{x}^{T}\mathbf{s}+1\right)  ^{2}\boldsymbol{\kappa}+\kappa_{0}=1$.

Given Eq. (\ref{TAE Eigenlocus Component One Q}), it is concluded that a
quadratic eigenlocus discriminant function $\widetilde{\Lambda}%
_{\boldsymbol{\kappa}}\left(  \mathbf{s}\right)  =\left(  \mathbf{x}%
^{T}\mathbf{s}+1\right)  ^{2}\boldsymbol{\kappa}+\kappa_{0}$ satisfies the
quadratic decision border $D_{+1}\left(  \mathbf{s}\right)  $ in terms of the
total allowed eigenenergy $\left\Vert \boldsymbol{\kappa}_{1}\right\Vert
_{\min_{c}}^{2}$ exhibited by $\boldsymbol{\kappa}_{1}$, where the functional
$\left\Vert \boldsymbol{\kappa}_{1}\right\Vert _{\min_{c}}^{2}-\left\Vert
\boldsymbol{\kappa}_{1}\right\Vert \left\Vert \boldsymbol{\kappa}%
_{2}\right\Vert \cos\theta_{\boldsymbol{\kappa}_{1}\boldsymbol{\kappa}_{2}}$
is constrained by the functional $\sum\nolimits_{i=1}^{l_{1}}\psi_{1_{i_{\ast
}}}\left(  1-\xi_{i}-\kappa_{0}\right)  $.

The critical minimum eigenenergy constraint on $\boldsymbol{\kappa}_{2}$ is
examined next.

\subsubsection{Total Allowed Eigenenergy of $\boldsymbol{\kappa}_{2}$}

Take any scaled extreme vector $\psi_{2_{i_{\ast}}}k_{\mathbf{x}_{2_{i_{\ast}%
}}}$ that belongs to class $\omega_{2}$. Using Eq.
(\ref{Minimum Eigenenergy Functional System Q}) and letting $y_{i}=-1$, it
follows that the constrained, primal principal eigenaxis component
$\psi_{2_{i_{\ast}}}k_{\mathbf{x}_{2_{i_{\ast}}}}$ on $\boldsymbol{\kappa}%
_{2}$ is specified by the equation:%
\[
-\psi_{2_{i_{\ast}}}k_{\mathbf{x}_{2_{i_{\ast}}}}\boldsymbol{\kappa}%
=\psi_{2_{i_{\ast}}}\left(  1-\xi_{i}+\kappa_{0}\right)
\]
which is part of a system of $l_{2}$ eigenlocus equations. Therefore, each
constrained, primal principal eigenaxis component $\psi_{2_{i_{\ast}}%
}k_{\mathbf{x}_{2_{i_{\ast}}}}$ on $\boldsymbol{\kappa}_{2}$ satisfies the
above locus equation.

Now take all of the $l_{2}$ scaled extreme vectors $\left\{  \psi_{2_{i_{\ast
}}}k_{\mathbf{x}_{2_{i_{\ast}}}}\right\}  _{i=1}^{l_{2}}$ that belong to class
$\omega_{2}$. Again, using Eq. (\ref{Minimum Eigenenergy Functional System Q})
and letting $y_{i}=-1$, it follows that the complete set $\left\{
\psi_{2_{i_{\ast}}}k_{\mathbf{x}_{2_{i_{\ast}}}}\right\}  _{i=1}^{l_{2}}$ of
$l_{2}$ constrained, primal principal eigenaxis components $\psi_{2_{i_{\ast}%
}}k_{\mathbf{x}_{2_{i_{\ast}}}}$ on $\boldsymbol{\kappa}_{2}$ is determined by
the system of $l_{2}$ eigenlocus equations:%
\begin{equation}
-\psi_{2_{i_{\ast}}}k_{\mathbf{x}_{2_{i_{\ast}}}}\boldsymbol{\kappa}%
=\psi_{2_{i_{\ast}}}\left(  1-\xi_{i}+\kappa_{0}\right)  ,\text{\ }%
i=1,...,l_{2}\text{.} \label{Minimum Eigenenergy Class Two Q}%
\end{equation}

Using the above equation, it follows that the entire set $\left\{
\psi_{2_{i_{\ast}}}k_{\mathbf{x}_{2_{i_{\ast}}}}\right\}  _{i=1}^{l_{2}}$ of
$l_{2}\times d$ transformed, extreme vector coordinates satisfies the system
of $l_{2}$ eigenlocus equations:%
\[
\text{ }(1)\text{ \ }-\psi_{2_{i_{\ast}}}k_{\mathbf{x}_{2_{1+l_{1\ast}}}%
}\boldsymbol{\kappa}=\psi_{2_{1+l_{1\ast}}}\left(  1-\xi_{i}+\kappa
_{0}\right)  \text{,}%
\]%
\[
(2)\text{ \ }-\psi_{2_{i_{\ast}}}k_{\mathbf{x}_{2_{2_{\ast}}}}%
\boldsymbol{\kappa}=\psi_{2_{2_{\ast}}}\left(  1-\xi_{i}+\kappa_{0}\right)
\text{,}%
\]%
\[
\vdots
\]%
\[
(l_{2})\ -\psi_{2_{i_{\ast}}}k_{\mathbf{x}_{2_{l_{2\ast}}}}\boldsymbol{\kappa
}=\psi_{2_{l_{2}\ast}}\left(  1-\xi_{i}+\kappa_{0}\right)  \text{,}%
\]
where each constrained, primal principal eigenaxis component $\psi
_{2_{i_{\ast}}}k_{\mathbf{x}_{2_{i_{\ast}}}}$ on $\boldsymbol{\kappa}_{2}$
satisfies the identity:%
\[
-\psi_{2_{i_{\ast}}}k_{\mathbf{x}_{2_{i_{\ast}}}}\boldsymbol{\kappa}%
=\psi_{2_{i_{\ast}}}\left(  1-\xi_{i}+\kappa_{0}\right)  \text{.}%
\]

I will now formulate an identity for the total allowed eigenenergy of
$\boldsymbol{\kappa}_{2}$. Let $E_{\boldsymbol{\kappa}_{2}}$ denote the
functional of the total allowed eigenenergy $\left\Vert \boldsymbol{\kappa
}_{2}\right\Vert _{\min_{c}}^{2}$ of $\boldsymbol{\kappa}_{2}$ and let
$\boldsymbol{\kappa}=\boldsymbol{\kappa}_{1}-\boldsymbol{\kappa}_{2}$.
Summation over the above system of $l_{2}$ eigenlocus equations produces the
following equation for the total allowed eigenenergy $\left\Vert
\boldsymbol{\kappa}_{2}\right\Vert _{\min_{c}}^{2}$ of $\boldsymbol{\kappa
}_{2}$:%
\[
-\left(  \sum\nolimits_{i=1}^{l_{2}}\psi_{2_{i_{\ast}}}k_{\mathbf{x}%
_{2_{i_{\ast}}}}\right)  \left(  \boldsymbol{\kappa}_{1}-\boldsymbol{\kappa
}_{2}\right)  \equiv\sum\nolimits_{i=1}^{l_{2}}\psi_{2_{i_{\ast}}}\left(
1-\xi_{i}+\kappa_{0}\right)
\]
which reduces to%
\[
\boldsymbol{\kappa}_{2}^{T}\boldsymbol{\kappa}_{2}-\boldsymbol{\kappa}_{2}%
^{T}\boldsymbol{\kappa}_{1}\equiv\sum\nolimits_{i=1}^{l_{2}}\psi_{2_{i_{\ast}%
}}\left(  1-\xi_{i}+\kappa_{0}\right)
\]
so that the functional $E_{\boldsymbol{\kappa}_{2}}$ satisfies the identity%
\[
\left\Vert \boldsymbol{\kappa}_{2}\right\Vert _{\min_{c}}^{2}%
-\boldsymbol{\kappa}_{2}^{T}\boldsymbol{\kappa}_{1}\equiv\sum\nolimits_{i=1}%
^{l_{2}}\psi_{2_{i_{\ast}}}\left(  1-\xi_{i}+\kappa_{0}\right)  \text{.}%
\]

Therefore, the total allowed eigenenergy $\left\Vert \boldsymbol{\kappa}%
_{2}\right\Vert _{\min_{c}}^{2}$ exhibited by the constrained, primal
eigenlocus component $\boldsymbol{\kappa}_{2}$ is determined by the identity%
\begin{equation}
\left\Vert \boldsymbol{\kappa}_{2}\right\Vert _{\min_{c}}^{2}-\left\Vert
\boldsymbol{\kappa}_{2}\right\Vert \left\Vert \boldsymbol{\kappa}%
_{1}\right\Vert \cos\theta_{\boldsymbol{\kappa}_{2}\boldsymbol{\kappa}_{1}%
}\equiv\sum\nolimits_{i=1}^{l_{2}}\psi_{2_{i_{\ast}}}\left(  1-\xi_{i}%
+\kappa_{0}\right)  \text{,} \label{TAE Eigenlocus Component Two Q}%
\end{equation}
where the functional $E_{\boldsymbol{\kappa}_{2}}$ of the total allowed
eigenenergy $\left\Vert \boldsymbol{\kappa}_{2}\right\Vert _{\min_{c}}^{2}$
exhibited by $\boldsymbol{\kappa}_{2}$%
\[
E_{\boldsymbol{\kappa}_{2}}=\left\Vert \boldsymbol{\kappa}_{2}\right\Vert
_{\min_{c}}^{2}-\left\Vert \boldsymbol{\kappa}_{2}\right\Vert \left\Vert
\boldsymbol{\kappa}_{1}\right\Vert \cos\theta_{\boldsymbol{\kappa}%
_{2}\boldsymbol{\kappa}_{1}}%
\]
is equivalent to the functional $E_{\boldsymbol{\psi}_{2}}$%
\[
E_{\boldsymbol{\psi}_{2}}=\sum\nolimits_{i=1}^{l_{2}}\psi_{2_{i_{\ast}}%
}\left(  1-\xi_{i}+\kappa_{0}\right)
\]
of the integrated magnitudes $\sum\nolimits_{i=1}^{l_{2}}\psi_{2_{i_{\ast}}}$
of the Wolfe dual principal eigenaxis components $\psi_{2_{i_{\ast}}}%
\frac{k_{\mathbf{x}_{2_{i_{\ast}}}}}{\left\Vert k_{\mathbf{x}_{2_{i_{\ast}}}%
}\right\Vert }$ and the $\kappa_{0}$ statistic.

Returning to Eq. (\ref{Decision Border Two Q}), it follows the functionals
$E_{\boldsymbol{\kappa}_{2}}$ and $E_{\boldsymbol{\psi}_{2}}$ specify the
manner in which quadratic eigenlocus discriminant functions
$\widetilde{\Lambda}_{\boldsymbol{\kappa}}\left(  \mathbf{s}\right)  =\left(
\mathbf{x}^{T}\mathbf{s}+1\right)  ^{2}\boldsymbol{\kappa}+\kappa_{0}$ satisfy
the quadratic decision border $D_{-1}\left(  \mathbf{s}\right)  $: $\left(
\mathbf{x}^{T}\mathbf{s}+1\right)  ^{2}\boldsymbol{\kappa}+\kappa_{0}=-1$.

Given Eq. (\ref{TAE Eigenlocus Component Two Q}), it is concluded that a
quadratic eigenlocus discriminant function $\widetilde{\Lambda}%
_{\boldsymbol{\kappa}}\left(  \mathbf{s}\right)  =\left(  \mathbf{x}%
^{T}\mathbf{s}+1\right)  ^{2}\boldsymbol{\kappa}+\kappa_{0}$ satisfies the
quadratic decision border $D_{-1}\left(  \mathbf{s}\right)  $ in terms of the
total allowed eigenenergy $\left\Vert \boldsymbol{\kappa}_{2}\right\Vert
_{\min_{c}}^{2}$ exhibited by $\boldsymbol{\kappa}_{2}$, where the functional
$\left\Vert \boldsymbol{\kappa}_{2}\right\Vert _{\min_{c}}^{2}-\left\Vert
\boldsymbol{\kappa}_{2}\right\Vert \left\Vert \boldsymbol{\kappa}%
_{1}\right\Vert \cos\theta_{\boldsymbol{\kappa}_{2}\boldsymbol{\kappa}_{1}}$
is constrained by the functional $\sum\nolimits_{i=1}^{l_{2}}\psi_{2_{i_{\ast
}}}\left(  1-\xi_{i}+\kappa_{0}\right)  $.

The critical minimum eigenenergy constraint on $\boldsymbol{\kappa}$ is
examined next.

\subsubsection{Total Allowed Eigenenergy of $\boldsymbol{\kappa}$}

I\ will now formulate an identity for the total allowed eigenenergy of a
constrained, primal quadratic eigenlocus $\boldsymbol{\kappa}$. Let
$E_{\boldsymbol{\kappa}}$ denote the functional satisfied by the total allowed
eigenenergy $\left\Vert \boldsymbol{\kappa}\right\Vert _{\min_{c}}^{2}$ of
$\boldsymbol{\kappa}$.

Summation over the complete system of eigenlocus equations satisfied by
$\boldsymbol{\kappa}_{1}$%
\[
\left(  \sum\nolimits_{i=1}^{l_{1}}\psi_{1_{i_{\ast}}}k_{\mathbf{x}%
_{1_{i_{\ast}}}}\right)  \boldsymbol{\kappa}\equiv\sum\nolimits_{i=1}^{l_{1}%
}\psi_{1_{i_{\ast}}}\left(  1-\xi_{i}-\kappa_{0}\right)
\]
and by $\boldsymbol{\kappa}_{2}$%
\[
\left(  -\sum\nolimits_{i=1}^{l_{2}}\psi_{2_{i_{\ast}}}k_{\mathbf{x}%
_{2_{i\ast}}}\right)  \boldsymbol{\kappa}\equiv\sum\nolimits_{i=1}^{l_{2}}%
\psi_{2_{i_{\ast}}}\left(  1-\xi_{i}+\kappa_{0}\right)
\]
produces the following identity for the functional $E_{\boldsymbol{\kappa}}$
satisfied by the total allowed eigenenergy $\left\Vert \boldsymbol{\kappa
}\right\Vert _{\min_{c}}^{2}$ of $\boldsymbol{\kappa}$%
\begin{align*}
&  \left(  \sum\nolimits_{i=1}^{l_{1}}\psi_{1_{i_{\ast}}}k_{\mathbf{x}%
_{1_{i_{\ast}}}}-\sum\nolimits_{i=1}^{l_{2}}\psi_{2_{i_{\ast}}}k_{\mathbf{x}%
_{2_{i\ast}}}\right)  \boldsymbol{\kappa}\\
&  \equiv\sum\nolimits_{i=1}^{l_{1}}\psi_{1_{i_{\ast}}}\left(  1-\xi
_{i}-\kappa_{0}\right)  +\sum\nolimits_{i=1}^{l_{2}}\psi_{2_{i_{\ast}}}\left(
1-\xi_{i}+\kappa_{0}\right)
\end{align*}
which reduces to%
\begin{align}
\left(  \boldsymbol{\kappa}_{1}-\boldsymbol{\kappa}_{2}\right)  ^{T}%
\boldsymbol{\kappa}  &  \equiv\sum\nolimits_{i=1}^{l_{1}}\psi_{1i\ast}\left(
1-\xi_{i}-\kappa_{0}\right) \label{Symmetrical Balance of TAE SDE Q}\\
&  +\sum\nolimits_{i=1}^{l_{2}}\psi_{2_{i_{\ast}}}\left(  1-\xi_{i}+\kappa
_{0}\right) \nonumber\\
&  \equiv\sum\nolimits_{i=1}^{l}\psi_{i_{\ast}}\left(  1-\xi_{i}\right)
\text{,}\nonumber
\end{align}
where I\ have used the equilibrium constraint on $\boldsymbol{\psi}$ in Eq.
(\ref{Equilibrium Constraint on Dual Eigen-components Q}).

Thereby, the functional $E_{\boldsymbol{\kappa}}$ of the total allowed
eigenenergy $\left\Vert \boldsymbol{\kappa}\right\Vert _{\min_{c}}^{2}$
exhibited by $\boldsymbol{\kappa}$%
\begin{align*}
E_{\boldsymbol{\kappa}}  &  =\left(  \boldsymbol{\kappa}_{1}%
-\boldsymbol{\kappa}_{2}\right)  ^{T}\boldsymbol{\kappa}\\
&  \mathbf{=}\left\Vert \boldsymbol{\kappa}\right\Vert _{\min_{c}}^{2}%
\end{align*}
is equivalent to the functional $E_{\boldsymbol{\psi}}$%
\[
E_{\boldsymbol{\psi}}=\sum\nolimits_{i=1}^{l}\psi_{i_{\ast}}\left(  1-\xi
_{i}\right)
\]
solely in terms of the integrated magnitudes $\sum\nolimits_{i=1}^{l}%
\psi_{i_{\ast}}$ of the Wolfe dual principal eigenaxis components on
$\boldsymbol{\psi}$.

Thus, the total allowed eigenenergy $\left\Vert \boldsymbol{\kappa}\right\Vert
_{\min_{c}}^{2}$ exhibited by a constrained, primal quadratic eigenlocus
$\boldsymbol{\kappa}$ is specified by the integrated magnitudes $\psi
_{i_{\ast}}$ of the Wolfe dual principal eigenaxis components $\psi_{i\ast
}\frac{k_{\mathbf{x}i\ast}}{\left\Vert k_{\mathbf{x}_{i\ast}}\right\Vert }$ on
$\boldsymbol{\psi}$%
\begin{align}
\left\Vert \boldsymbol{\kappa}\right\Vert _{\min_{c}}^{2}  &  \equiv
\sum\nolimits_{i=1}^{l}\psi_{i_{\ast}}\left(  1-\xi_{i}\right)
\label{TAE SDE Q}\\
&  \equiv\sum\nolimits_{i=1}^{l}\psi_{i_{\ast}}-\sum\nolimits_{i=1}^{l}%
\psi_{i_{\ast}}\xi_{i}\text{,}\nonumber
\end{align}
where the regularization parameters $\xi_{i}=\xi\ll1$ are seen to determine
negligible constraints on $\left\Vert \boldsymbol{\kappa}\right\Vert
_{\min_{c}}^{2}$. Therefore, it is concluded that the total allowed
eigenenergy $\left\Vert \boldsymbol{\kappa}\right\Vert _{\min_{c}}^{2}$
exhibited by a constrained, primal quadratic eigenlocus $\boldsymbol{\kappa}$
is determined by the integrated magnitudes $\sum\nolimits_{i=1}^{l}%
\psi_{i_{\ast}}$ of the Wolfe dual principal eigenaxis components $\psi
_{i\ast}\frac{k_{\mathbf{x}i\ast}}{\left\Vert k_{\mathbf{x}_{i\ast}%
}\right\Vert }$ on $\boldsymbol{\psi}$.

Returning to Eq. (\ref{Decision Boundary Q}), it follows that the equilibrium
constraint on $\boldsymbol{\psi}$ and the corresponding functionals
$E_{\boldsymbol{\kappa}}$ and $E_{\boldsymbol{\psi}}$ specify the manner in
which quadratic eigenlocus discriminant functions $\widetilde{\Lambda
}_{\boldsymbol{\kappa}}\left(  \mathbf{s}\right)  =\left(  \mathbf{x}%
^{T}\mathbf{s}+1\right)  ^{2}\boldsymbol{\kappa}+\kappa_{0}$ satisfy quadratic
decision boundaries $D_{0}\left(  \mathbf{s}\right)  $: $\left(
\mathbf{x}^{T}\mathbf{s}+1\right)  ^{2}\boldsymbol{\kappa}+\kappa_{0}=0$.

Given Eq. (\ref{TAE SDE Q}), it is concluded that a quadratic eigenlocus
discriminant function $\widetilde{\Lambda}_{\boldsymbol{\kappa}}\left(
\mathbf{s}\right)  =\left(  \mathbf{x}^{T}\mathbf{s}+1\right)  ^{2}%
\boldsymbol{\kappa}+\kappa_{0}$ satisfies a quadratic decision boundary
$D_{0}\left(  \mathbf{s}\right)  $ in terms of its total allowed eigenenergy
$\left\Vert \boldsymbol{\kappa}\right\Vert _{\min_{c}}^{2}$, where the
functional $\left\Vert \boldsymbol{\kappa}\right\Vert _{\min_{c}}^{2}$ is
constrained by the functional $\sum\nolimits_{i=1}^{l}\psi_{i_{\ast}}\left(
1-\xi_{i}\right)  $.

Using Eqs (\ref{TAE Eigenlocus Component One Q}),
(\ref{TAE Eigenlocus Component Two Q}), and
(\ref{Symmetrical Balance of TAE SDE Q}), it follows that the symmetrically
balanced constraints%
\[
E_{\boldsymbol{\psi}_{1}}=\sum\nolimits_{i=1}^{l_{1}}\psi_{1_{i_{\ast}}%
}\left(  1-\xi_{i}-\kappa_{0}\right)  \text{ \ and \ }E_{\boldsymbol{\psi}%
_{2}}=\sum\nolimits_{i=1}^{l_{2}}\psi_{2_{i_{\ast}}}\left(  1-\xi_{i}%
+\kappa_{0}\right)
\]
satisfied by a quadratic eigenlocus discriminant function $\widetilde{\Lambda
}_{\boldsymbol{\kappa}}\left(  \mathbf{s}\right)  =\left(  \mathbf{x}%
^{T}\mathbf{s}+1\right)  ^{2}\boldsymbol{\kappa}+\kappa_{0}$ on the respective
quadratic decision borders $D_{+1}\left(  \mathbf{s}\right)  $ and
$D_{-1}\left(  \mathbf{s}\right)  $, and the corresponding constraint%
\[
E_{\boldsymbol{\psi}}=\sum\nolimits_{i=1}^{l_{1}}\psi_{1i\ast}\left(
1-\xi_{i}-\kappa_{0}\right)  +\sum\nolimits_{i=1}^{l_{2}}\psi_{2_{i_{\ast}}%
}\left(  1-\xi_{i}+\kappa_{0}\right)
\]
satisfied by a quadratic eigenlocus discriminant function $\widetilde{\Lambda
}_{\boldsymbol{\kappa}}\left(  \mathbf{s}\right)  =\left(  \mathbf{x}%
^{T}\mathbf{s}+1\right)  ^{2}\boldsymbol{\kappa}+\kappa_{0}$ on the quadratic
decision boundary $D_{0}\left(  \mathbf{s}\right)  $, ensure that the total
allowed eigenenergies $\left\Vert \boldsymbol{\kappa}_{1}-\boldsymbol{\kappa
}_{2}\right\Vert _{\min_{c}}^{2}$ exhibited by the scaled extreme points on
$\boldsymbol{\kappa}_{1}-\boldsymbol{\kappa}_{2}$:%
\begin{align*}
\left\Vert \boldsymbol{\kappa}\right\Vert _{\min_{c}}^{2}  &  =\left\Vert
\boldsymbol{\kappa}_{1}\right\Vert _{\min_{c}}^{2}-\left\Vert
\boldsymbol{\kappa}_{1}\right\Vert \left\Vert \boldsymbol{\kappa}%
_{2}\right\Vert \cos\theta_{\boldsymbol{\kappa}_{1}\boldsymbol{\kappa}_{2}}\\
&  +\left\Vert \boldsymbol{\kappa}_{2}\right\Vert _{\min_{c}}^{2}-\left\Vert
\boldsymbol{\kappa}_{2}\right\Vert \left\Vert \boldsymbol{\kappa}%
_{1}\right\Vert \cos\theta_{\boldsymbol{\kappa}_{2}\boldsymbol{\kappa}_{1}}%
\end{align*}
satisfy the law of cosines in the symmetrically balanced manner depicted in
Fig. $\ref{Law of Cosines for Quadratic Classification Systems}$.

Given the binary classification theorem, it follows that quadratic eigenlocus
likelihood ratios $\widehat{\Lambda}_{\boldsymbol{\kappa}}\left(
\mathbf{s}\right)  =\boldsymbol{\kappa}_{1}-\boldsymbol{\kappa}_{2}$ and
corresponding decision boundaries $D_{0}\left(  \mathbf{s}\right)  $ satisfy
an integral equation $f\left(  \widetilde{\Lambda}_{\boldsymbol{\kappa}%
}\left(  \mathbf{s}\right)  \right)  $ where the areas $\int\nolimits_{Z}%
p\left(  k_{\mathbf{x}_{1i\ast}}|\boldsymbol{\kappa}_{1}\right)
d\boldsymbol{\kappa}_{1}$ and $\int\nolimits_{Z}p\left(  k_{\mathbf{x}%
_{2i\ast}}|\boldsymbol{\kappa}_{2}\right)  d\boldsymbol{\kappa}_{2}$ under the
class-conditional probability density functions $p\left(  k_{\mathbf{x}%
_{1i\ast}}|\boldsymbol{\kappa}_{1}\right)  $ and $p\left(  k_{\mathbf{x}%
_{2i\ast}}|\boldsymbol{\kappa}_{2}\right)  $ are balanced with each other.
Furthermore, the eigenenergy associated with the position or location of the
likelihood ratio $p\left(  \widehat{\Lambda}_{\boldsymbol{\kappa}}\left(
\mathbf{s}\right)  |\omega_{2}\right)  $ given class $\omega_{2}$ must be
balanced with the eigenenergy associated with the position or location of the
likelihood ratio $p\left(  \widehat{\Lambda}_{\boldsymbol{\kappa}}\left(
\mathbf{s}\right)  |\omega_{1}\right)  $ given class $\omega_{1}$. Therefore,
quadratic eigenlocus likelihood ratios $\widehat{\Lambda}_{\boldsymbol{\kappa
}}\left(  \mathbf{s}\right)  =\boldsymbol{\kappa}_{1}-\boldsymbol{\kappa}_{2}$
and corresponding decision boundaries $D_{0}\left(  \mathbf{s}\right)  $ also
satisfy an integral equation where the total allowed eigenenergies $\left\Vert
\boldsymbol{\kappa}_{1}-\boldsymbol{\kappa}_{2}\right\Vert _{\min_{c}}^{2}$ of
a quadratic eigenlocus $\boldsymbol{\kappa}=\boldsymbol{\kappa}_{1}%
-\boldsymbol{\kappa}_{2}$ are balanced with each other.

I\ will show that the discrete, quadratic classification system $\left(
\mathbf{x}^{T}\mathbf{s}+1\right)  ^{2}\boldsymbol{\kappa}+\kappa
_{0}\overset{\omega_{1}}{\underset{\omega_{2}}{\gtrless}}0$ seeks an
equilibrium point $p\left(  \widehat{\Lambda}_{\boldsymbol{\psi}}\left(
\mathbf{s}\right)  |\omega_{1}\right)  -p\left(  \widehat{\Lambda
}_{\boldsymbol{\psi}}\left(  \mathbf{s}\right)  |\omega_{2}\right)  =0$ of an
integral equation $f\left(  \widetilde{\Lambda}_{\boldsymbol{\kappa}}\left(
\mathbf{s}\right)  \right)  $, where the total allowed eigenenergies
$\left\Vert \boldsymbol{\kappa}_{1}-\boldsymbol{\kappa}_{2}\right\Vert
_{\min_{c}}^{2}$ of the classification system are balanced with each other,
such that the eigenenergy and the expected risk of the classification system
$\left(  \mathbf{x}^{T}\mathbf{s}+1\right)  ^{2}\boldsymbol{\kappa}+\kappa
_{0}\overset{\omega_{1}}{\underset{\omega_{2}}{\gtrless}}0$ are minimized, and
the classification system $\left(  \mathbf{x}^{T}\mathbf{s}+1\right)
^{2}\boldsymbol{\kappa}+\kappa_{0}\overset{\omega_{1}}{\underset{\omega
_{2}}{\gtrless}}0$ is in statistical equilibrium.

In the next section, I will develop an integral equation $f\left(
\widetilde{\Lambda}_{\boldsymbol{\kappa}}\left(  \mathbf{s}\right)  \right)  $
that is satisfied by quadratic eigenlocus discriminant functions
$\widetilde{\Lambda}_{\boldsymbol{\kappa}}\left(  \mathbf{s}\right)  =\left(
\mathbf{x}^{T}\mathbf{s}+1\right)  ^{2}\boldsymbol{\kappa}+\kappa_{0}$, where
the total allowed eigenenergies $\left\Vert \boldsymbol{\kappa}_{1}%
-\boldsymbol{\kappa}_{2}\right\Vert _{\min_{c}}^{2}$ exhibited by the
principal eigenaxis components on a quadratic eigenlocus $\boldsymbol{\kappa
}=\boldsymbol{\kappa}_{1}-\boldsymbol{\kappa}_{2}$ are symmetrically balanced
with each other. Thereby, I will show that the likelihood ratio $p\left(
\widehat{\Lambda}_{\boldsymbol{\kappa}}\left(  \mathbf{s}\right)  |\omega
_{1}\right)  -p\left(  \widehat{\Lambda}_{\boldsymbol{\kappa}}\left(
\mathbf{s}\right)  |\omega_{2}\right)  $ is in statistical equilibrium, and
that the areas under the class-conditional density functions $p\left(
k_{\mathbf{x}_{1i\ast}}|\boldsymbol{\kappa}_{1}\right)  $ and $p\left(
k_{\mathbf{x}_{2i\ast}}|\boldsymbol{\kappa}_{2}\right)  $, over the decision
space $Z=Z_{1}+Z_{2}$, are symmetrically balanced with each other. I will use
these results to show that the discriminant function $\widetilde{\Lambda
}_{\boldsymbol{\kappa}}\left(  \mathbf{s}\right)  =\left(  \mathbf{x}%
^{T}\mathbf{s}+1\right)  ^{2}\boldsymbol{\kappa}+\kappa_{0}$ is the solution
to a fundamental integral equation of binary classification for a
classification system in statistical equilibrium. The solution involves a
surprising, statistical balancing feat in decision space $Z$ which hinges on
an elegant, statistical balancing feat in eigenspace $\widetilde{Z}$.

\section{The Balancing Feat in Eigenspace II}

A quadratic eigenlocus%
\[
\boldsymbol{\kappa}=\boldsymbol{\kappa}_{1}-\boldsymbol{\kappa}_{2}%
\]
which is formed by a locus of labeled ($+1$ or $-1$), scaled ($\psi_{1_{i\ast
}}$ or $\psi_{2_{i\ast}}$) extreme vectors ($k_{\mathbf{x}_{1_{i\ast}}}$ or
$k_{\mathbf{x}_{2_{i\ast}}}$)%
\[
\boldsymbol{\kappa}=\sum\nolimits_{i=1}^{l_{1}}\psi_{1_{i\ast}}k_{\mathbf{x}%
_{1_{i\ast}}}-\sum\nolimits_{i=1}^{l_{2}}\psi_{2_{i\ast}}k_{\mathbf{x}%
_{2_{i\ast}}}%
\]
has a \emph{dual nature} that is \emph{twofold}:

Each $\psi_{1_{i\ast}}$ or $\psi_{2_{i\ast}}$ scale factor determines the
total allowed eigenenergy%
\[
\left\Vert \psi_{1_{i\ast}}k_{\mathbf{x}_{1_{i\ast}}}\right\Vert _{\min_{c}%
}^{2}\text{ \ or \ }\left\Vert \psi_{2_{i\ast}}k_{\mathbf{x}_{2_{i\ast}}%
}\right\Vert _{\min_{c}}^{2}%
\]
of a principal eigenaxis component $\psi_{1_{i\ast}}k_{\mathbf{x}_{1_{i\ast}}%
}$ or $\psi_{2_{i\ast}}k_{\mathbf{x}_{2_{i\ast}}}$ on $\boldsymbol{\kappa}%
_{1}-\boldsymbol{\kappa}_{2}$ in decision space $Z$, and each $\psi_{1_{i\ast
}}$ or $\psi_{2_{i\ast}}$ scale factor determines the total allowed
eigenenergy%
\[
\left\Vert \psi_{1_{i\ast}}\frac{\mathbf{x}_{1_{i\ast}}}{\left\Vert
\mathbf{x}_{1_{i\ast}}\right\Vert }\right\Vert _{\min_{c}}^{2}\text{ \ or
\ }\left\Vert \psi_{2_{i\ast}}\frac{\mathbf{x}_{2_{i\ast}}}{\left\Vert
\mathbf{x}_{2_{i\ast}}\right\Vert }\right\Vert _{\min_{c}}^{2}%
\]
of a principal eigenaxis component $\psi_{1_{i\ast}}\frac{k_{\mathbf{x}%
_{1_{i\ast}}}}{\left\Vert k_{\mathbf{x}_{1_{i\ast}}}\right\Vert }$ or
$\psi_{2_{i\ast}}\frac{k_{\mathbf{x}_{2_{i\ast}}}}{\left\Vert k_{\mathbf{x}%
_{2_{i\ast}}}\right\Vert }$ on $\boldsymbol{\psi}_{1}+\boldsymbol{\psi}_{2}$
in Wolfe dual eigenspace $\widetilde{Z}$.

In addition, each $\psi_{1_{i\ast}}$ scale factor specifies dual conditional
densities for an $k_{\mathbf{x}_{1_{i\ast}}}$ extreme point:%
\[
p\left(  k_{\mathbf{x}_{1i\ast}}|\operatorname{comp}%
_{\overrightarrow{k_{\mathbf{x}_{1i\ast}}}}\left(
\overrightarrow{\boldsymbol{\kappa}}\right)  \right)  \frac{k_{\mathbf{x}%
_{1_{i\ast}}}}{\left\Vert k_{\mathbf{x}_{1_{i\ast}}}\right\Vert }\text{ \ and
\ }p\left(  k_{\mathbf{x}_{1i\ast}}|\operatorname{comp}%
_{\overrightarrow{k_{\mathbf{x}_{1i\ast}}}}\left(
\overrightarrow{\boldsymbol{\kappa}}\right)  \right)  k_{\mathbf{x}_{1_{i\ast
}}}\text{,}%
\]
and each $\psi_{2_{i\ast}}$ scale factor specifies dual conditional densities
for an $k_{\mathbf{x}_{2_{i\ast}}}$ extreme point:%
\[
p\left(  k_{\mathbf{x}_{2i\ast}}|\operatorname{comp}%
_{\overrightarrow{k_{\mathbf{x}_{2i\ast}}}}\left(
\overrightarrow{\boldsymbol{\kappa}}\right)  \right)  \frac{k_{k_{\mathbf{x}%
_{2i\ast}}}}{\left\Vert k_{k_{\mathbf{x}_{2i\ast}}}\right\Vert }\text{ \ and
\ }p\left(  k_{\mathbf{x}_{2i\ast}}|\operatorname{comp}%
_{\overrightarrow{k_{\mathbf{x}_{2i\ast}}}}\left(
\overrightarrow{\boldsymbol{\kappa}}\right)  \right)  k_{k_{\mathbf{x}%
_{2i\ast}}}\text{.}%
\]

Accordingly, a Wolfe dual quadratic eigenlocus $\boldsymbol{\psi
}=\boldsymbol{\psi}_{1}+\boldsymbol{\psi}_{2}$ is a parameter vector of
likelihoods:%
\begin{align*}
\widehat{\Lambda}_{\boldsymbol{\psi}}\left(  \mathbf{s}\right)   &
=\sum\nolimits_{i=1}^{l_{1}}p\left(  k_{\mathbf{x}_{1i\ast}}%
|\operatorname{comp}_{\overrightarrow{k_{\mathbf{x}_{1i\ast}}}}\left(
\overrightarrow{\boldsymbol{\kappa}}\right)  \right)  \frac{k_{\mathbf{x}%
_{1_{i\ast}}}}{\left\Vert k_{\mathbf{x}_{1_{i\ast}}}\right\Vert }\\
&  +\sum\nolimits_{i=1}^{l_{2}}p\left(  k_{\mathbf{x}_{2i\ast}}%
|\operatorname{comp}_{\overrightarrow{k_{\mathbf{x}_{2i\ast}}}}\left(
\overrightarrow{\boldsymbol{\kappa}}\right)  \right)  \frac{k_{k_{\mathbf{x}%
_{2i\ast}}}}{\left\Vert k_{k_{\mathbf{x}_{2i\ast}}}\right\Vert }%
\end{align*}
\emph{and} a locus of principal eigenaxis components in Wolfe dual eigenspace
$\widetilde{Z}$, and a primal quadratic eigenlocus $\boldsymbol{\kappa
}=\boldsymbol{\kappa}_{1}-\boldsymbol{\kappa}_{2}$ is a parameter vector of
likelihoods:%
\begin{align*}
\widehat{\Lambda}_{\boldsymbol{\kappa}}\left(  \mathbf{s}\right)   &
=\sum\nolimits_{i=1}^{l_{1}}p\left(  k_{\mathbf{x}_{1i\ast}}%
|\operatorname{comp}_{\overrightarrow{k_{\mathbf{x}_{1i\ast}}}}\left(
\overrightarrow{\boldsymbol{\kappa}}\right)  \right)  k_{\mathbf{x}_{1_{i\ast
}}}\\
&  -\sum\nolimits_{i=1}^{l_{2}}p\left(  k_{\mathbf{x}_{2i\ast}}%
|\operatorname{comp}_{\overrightarrow{k_{\mathbf{x}_{2i\ast}}}}\left(
\overrightarrow{\boldsymbol{\kappa}}\right)  \right)  k_{\mathbf{x}_{2_{i\ast
}}}%
\end{align*}
\emph{and} a locus of principal eigenaxis components in decision space $Z$,
that jointly determine the basis of a quadratic classification system $\left(
\mathbf{x}^{T}\mathbf{s}+1\right)  ^{2}\boldsymbol{\kappa}+\kappa
_{0}\overset{\omega_{1}}{\underset{\omega_{2}}{\gtrless}}0$.

Moreover, the Wolfe dual likelihood ratio%
\begin{align*}
\widehat{\Lambda}_{\boldsymbol{\psi}}\left(  \mathbf{s}\right)   &  =p\left(
\widehat{\Lambda}_{\boldsymbol{\psi}}\left(  \mathbf{x}\right)  |\omega
_{1}\right)  +p\left(  \widehat{\Lambda}_{\boldsymbol{\psi}}\left(
\mathbf{x}\right)  |\omega_{2}\right) \\
&  =\boldsymbol{\psi}_{1}+\boldsymbol{\psi}_{2}%
\end{align*}
is constrained to satisfy the equilibrium equation:%
\[
p\left(  \widehat{\Lambda}_{\boldsymbol{\psi}}\left(  \mathbf{x}\right)
|\omega_{1}\right)  =p\left(  \widehat{\Lambda}_{\boldsymbol{\psi}}\left(
\mathbf{x}\right)  |\omega_{2}\right)
\]
so that the Wolfe dual likelihood ratio $\widehat{\Lambda}_{\boldsymbol{\psi}%
}\left(  \mathbf{x}\right)  =\boldsymbol{\psi}_{1}+\boldsymbol{\psi}_{2}$ is
in statistical equilibrium:%
\[
\boldsymbol{\psi}_{1}=\boldsymbol{\psi}_{2}\text{.}%
\]

I will demonstrate that the dual nature of $\boldsymbol{\kappa}$ enables a
quadratic eigenlocus discriminant function%
\[
\widetilde{\Lambda}_{\boldsymbol{\kappa}}\left(  \mathbf{s}\right)  =\left(
\mathbf{x}^{T}\mathbf{s}+1\right)  ^{2}\boldsymbol{\kappa}+\kappa_{0}%
\]
to be the solution to a fundamental integral equation of binary classification
for a classification system in statistical equilibrium:%
\begin{align*}
f\left(  \widetilde{\Lambda}_{\boldsymbol{\kappa}}\left(  \mathbf{s}\right)
\right)   &  =\int\nolimits_{Z_{1}}p\left(  k_{\mathbf{x}_{1i\ast}%
}|\boldsymbol{\kappa}_{1}\right)  d\boldsymbol{\kappa}_{1}+\int%
\nolimits_{Z_{2}}p\left(  k_{\mathbf{x}_{1i\ast}}|\boldsymbol{\kappa}%
_{1}\right)  d\boldsymbol{\kappa}_{1}+\delta\left(  y\right)  \boldsymbol{\psi
}_{1}\\
&  =\int\nolimits_{Z_{1}}p\left(  k_{\mathbf{x}_{2_{i_{\ast}}}}%
|\boldsymbol{\kappa}_{2}\right)  d\boldsymbol{\kappa}_{2}+\int\nolimits_{Z_{2}%
}p\left(  k_{\mathbf{x}_{2_{i_{\ast}}}}|\boldsymbol{\kappa}_{2}\right)
d\boldsymbol{\kappa}_{2}-\delta\left(  y\right)  \boldsymbol{\psi}_{2}\text{,}%
\end{align*}
where all of the forces associated with the counter risks and the risks for
class $\omega_{1}$ and class $\omega_{2}$ are symmetrically balanced with each
other%
\begin{align*}
f\left(  \widetilde{\Lambda}_{\boldsymbol{\kappa}}\left(  \mathbf{s}\right)
\right)   &  :\int\nolimits_{Z_{1}}p\left(  k_{\mathbf{x}_{1i\ast}%
}|\boldsymbol{\kappa}_{1}\right)  d\boldsymbol{\kappa}_{1}-\int%
\nolimits_{Z_{1}}p\left(  k_{\mathbf{x}_{2_{i_{\ast}}}}|\boldsymbol{\kappa
}_{2}\right)  d\boldsymbol{\kappa}_{2}+\delta\left(  y\right)
\boldsymbol{\psi}_{1}\\
&  =\int\nolimits_{Z_{2}}p\left(  k_{\mathbf{x}_{2_{i_{\ast}}}}%
|\boldsymbol{\kappa}_{2}\right)  d\boldsymbol{\kappa}_{2}-\int\nolimits_{Z_{2}%
}p\left(  k_{\mathbf{x}_{1i\ast}}|\boldsymbol{\kappa}_{1}\right)
d\boldsymbol{\kappa}_{1}-\delta\left(  y\right)  \boldsymbol{\psi}_{2}\text{,}%
\end{align*}
over the $Z_{1}$ and $Z_{2}$ decision regions, by means of an elegant,
statistical balancing feat in Wolfe dual eigenspace $\widetilde{Z}$, where the
functional $E_{\boldsymbol{\kappa}_{1}}$ of $\left\Vert \boldsymbol{\kappa
}_{1}\right\Vert _{\min_{c}}^{2}$ in Eq. (\ref{TAE Eigenlocus Component One Q}%
) and the functional $E_{\boldsymbol{\kappa}_{2}}$ of $\left\Vert
\boldsymbol{\kappa}_{2}\right\Vert _{\min_{c}}^{2}$ in Eq.
(\ref{TAE Eigenlocus Component Two Q}) are constrained to be equal to each
other by means of a symmetric equalizer statistic $\nabla_{eq}$: $\frac
{\delta\left(  y\right)  }{2}\boldsymbol{\psi}$, where $\delta\left(
y\right)  \triangleq\sum\nolimits_{i=1}^{l}y_{i}\left(  1-\xi_{i}\right)  $.

I have shown that each of the constrained, primal principal eigenaxis
components $\psi_{1_{i\ast}}k_{\mathbf{x}_{1_{i\ast}}}$ or $\psi_{2_{i\ast}%
}k_{\mathbf{x}_{2_{i\ast}}}$ on $\boldsymbol{\kappa}=\boldsymbol{\kappa}%
_{1}-\boldsymbol{\kappa}_{2}$ have such magnitudes and directions that a
constrained, quadratic eigenlocus discriminant function $\widetilde{\Lambda
}_{\boldsymbol{\kappa}}\left(  \mathbf{s}\right)  =\left(  \mathbf{x}%
^{T}\mathbf{s}+1\right)  ^{2}\boldsymbol{\kappa}+\kappa_{0}$ partitions any
given feature space into symmetrical decision regions $Z_{1}\simeq Z_{2}$,
which are symmetrically partitioned by a quadratic decision boundary, by means
of three, symmetrical quadratic loci, all of which reference
$\boldsymbol{\kappa}$.

I\ will show that quadratic eigenlocus classification systems $\left(
\mathbf{x}^{T}\mathbf{s}+1\right)  ^{2}\boldsymbol{\kappa}+\kappa
_{0}\overset{\omega_{1}}{\underset{\omega_{2}}{\gtrless}}0$ generate decision
regions $Z_{1}$ and $Z_{2}$ for which the dual parameter vectors of
likelihoods $\widehat{\Lambda}_{\boldsymbol{\psi}}\left(  \mathbf{s}\right)
=\boldsymbol{\psi}_{1}+\boldsymbol{\psi}_{2}$ and $\widehat{\Lambda
}_{\boldsymbol{\kappa}}\left(  \mathbf{s}\right)  =\boldsymbol{\kappa}%
_{1}-\boldsymbol{\kappa}_{2}$ are in statistical equilibrium. Thereby, I\ will
demonstrate that balancing the forces associated with the expected risk
$\mathfrak{R}_{\mathfrak{\min}}\left(  Z|\boldsymbol{\kappa}\right)  $ of the
quadratic classification system $\left(  \mathbf{x}^{T}\mathbf{s}+1\right)
^{2}\boldsymbol{\kappa}+\kappa_{0}\overset{\omega_{1}}{\underset{\omega
_{2}}{\gtrless}}0$ hinges on balancing the eigenenergies associated with the
positions or locations of the dual likelihood ratios $\widehat{\Lambda
}_{\boldsymbol{\psi}}\left(  \mathbf{s}\right)  $ and $\widehat{\Lambda
}_{\boldsymbol{\kappa}}\left(  \mathbf{s}\right)  $:%
\begin{align*}
\widehat{\Lambda}_{\boldsymbol{\psi}}\left(  \mathbf{s}\right)   &
=\boldsymbol{\psi}_{1}+\boldsymbol{\psi}_{2}\\
&  =\sum\nolimits_{i=1}^{l_{1}}\psi_{1_{i\ast}}\frac{k_{\mathbf{x}_{1i\ast}}%
}{\left\Vert k_{\mathbf{x}_{1i\ast}}\right\Vert }+\sum\nolimits_{i=1}^{l_{2}%
}\psi_{2_{i\ast}}\frac{k_{\mathbf{x}_{2i\ast}}}{\left\Vert k_{\mathbf{x}%
_{2i\ast}}\right\Vert }%
\end{align*}
and%
\begin{align*}
\widehat{\Lambda}_{\boldsymbol{\kappa}}\left(  \mathbf{s}\right)   &
=\boldsymbol{\kappa}_{1}-\boldsymbol{\kappa}_{2}\\
&  =\sum\nolimits_{i=1}^{l_{1}}\psi_{1_{i\ast}}k_{\mathbf{x}_{1i\ast}}%
-\sum\nolimits_{i=1}^{l_{2}}\psi_{2_{i\ast}}k_{\mathbf{x}_{2i\ast}}\text{.}%
\end{align*}

\subsection{Balancing the Eigenenergies of $\boldsymbol{\kappa}_{1}%
-\boldsymbol{\kappa}_{2}$}

I will now devise an equation that determines how the total allowed
eigenenergies $\left\Vert \boldsymbol{\kappa}_{1}\right\Vert _{\min_{c}}^{2}$
and $\left\Vert \boldsymbol{\kappa}_{2}\right\Vert _{\min_{c}}^{2}$ exhibited
by $\boldsymbol{\kappa}_{1}$ and $\boldsymbol{\kappa}_{2}$ are symmetrically
balanced with each other.

Using Eq. (\ref{TAE Eigenlocus Component One Q}) and the equilibrium
constraint on $\boldsymbol{\psi}$ in Eq.
(\ref{Equilibrium Constraint on Dual Eigen-components Q})%
\begin{align*}
\left\Vert \boldsymbol{\kappa}_{1}\right\Vert _{\min_{c}}^{2}-\left\Vert
\boldsymbol{\kappa}_{1}\right\Vert \left\Vert \boldsymbol{\kappa}%
_{2}\right\Vert \cos\theta_{\boldsymbol{\kappa}_{1}\boldsymbol{\kappa}_{2}}
&  \equiv\sum\nolimits_{i=1}^{l_{1}}\psi_{1_{i_{\ast}}}\left(  1-\xi
_{i}-\kappa_{0}\right) \\
&  \equiv\frac{1}{2}\sum\nolimits_{i=1}^{l}\psi_{_{i_{\ast}}}\left(  1-\xi
_{i}-\kappa_{0}\right)  \text{,}%
\end{align*}
it follows that the functional $E_{\boldsymbol{\kappa}_{1}}$ of the total
allowed eigenenergy $\left\Vert \boldsymbol{\kappa}_{1}\right\Vert _{\min_{c}%
}^{2}$ of $\boldsymbol{\kappa}_{1}$%
\[
E_{\boldsymbol{\kappa}_{1}}=\left\Vert \boldsymbol{\kappa}_{1}\right\Vert
_{\min_{c}}^{2}-\left\Vert \boldsymbol{\kappa}_{1}\right\Vert \left\Vert
\boldsymbol{\kappa}_{2}\right\Vert \cos\theta_{\boldsymbol{\kappa}%
_{1}\boldsymbol{\kappa}_{2}}%
\]
is equivalent to the functional $E_{\boldsymbol{\psi}_{1}}$:%
\begin{equation}
E_{\boldsymbol{\psi}_{1}}=\frac{1}{2}\sum\nolimits_{i=1}^{l}\psi_{_{i_{\ast}}%
}\left(  1-\xi_{i}\right)  -\kappa_{0}\sum\nolimits_{i=1}^{l_{1}}%
\psi_{1_{i_{\ast}}}\text{.} \label{TAE Constraint COMP1 Q}%
\end{equation}

Using Eq. (\ref{TAE Eigenlocus Component Two Q}) and the equilibrium
constraint on $\boldsymbol{\psi}$ in Eq.
(\ref{Equilibrium Constraint on Dual Eigen-components Q})%
\begin{align*}
\left\Vert \boldsymbol{\kappa}_{2}\right\Vert _{\min_{c}}^{2}-\left\Vert
\boldsymbol{\kappa}_{2}\right\Vert \left\Vert \boldsymbol{\kappa}%
_{1}\right\Vert \cos\theta_{\boldsymbol{\kappa}_{2}\boldsymbol{\kappa}_{1}}
&  \equiv\sum\nolimits_{i=1}^{l_{2}}\psi_{2_{i_{\ast}}}\left(  1-\xi
_{i}+\kappa_{0}\right) \\
&  \equiv\frac{1}{2}\sum\nolimits_{i=1}^{l}\psi_{_{i_{\ast}}}\left(  1-\xi
_{i}+\kappa_{0}\right)  \text{,}%
\end{align*}
it follows that the functional $E_{\boldsymbol{\kappa}_{2}}$ of the total
allowed eigenenergy $\left\Vert \boldsymbol{\kappa}_{2}\right\Vert _{\min_{c}%
}^{2}$ of $\boldsymbol{\kappa}_{2}$%
\[
E_{\boldsymbol{\kappa}_{2}}=\left\Vert \boldsymbol{\kappa}_{2}\right\Vert
_{\min_{c}}^{2}-\left\Vert \boldsymbol{\kappa}_{2}\right\Vert \left\Vert
\boldsymbol{\kappa}_{1}\right\Vert \cos\theta_{\boldsymbol{\kappa}%
_{2}\boldsymbol{\kappa}_{1}}%
\]
is equivalent to the functional $E_{\boldsymbol{\psi}_{2}}$:%
\begin{equation}
E_{\boldsymbol{\psi}_{2}}=\frac{1}{2}\sum\nolimits_{i=1}^{l}\psi_{_{i_{\ast}}%
}\left(  1-\xi_{i}\right)  +\kappa_{0}\sum\nolimits_{i=1}^{l_{2}}%
\psi_{2_{i_{\ast}}}\text{.} \label{TAE Constraint COMP2 Q}%
\end{equation}

Next, I\ will use the identity for $\left\Vert \boldsymbol{\kappa}\right\Vert
_{\min_{c}}^{2}$ in Eq. (\ref{TAE SDE Q})%
\[
\left\Vert \boldsymbol{\kappa}\right\Vert _{\min_{c}}^{2}\equiv\sum
\nolimits_{i=1}^{l}\psi_{i_{\ast}}\left(  1-\xi_{i}\right)
\]
to rewrite $E_{\boldsymbol{\psi}_{1}}$%
\begin{align*}
E_{\boldsymbol{\psi}_{1}}  &  =\frac{1}{2}\sum\nolimits_{i=1}^{l}%
\psi_{_{i_{\ast}}}\left(  1-\xi_{i}\right)  -\kappa_{0}\sum\nolimits_{i=1}%
^{l_{1}}\psi_{1_{i_{\ast}}}\\
&  \equiv\frac{1}{2}\left\Vert \boldsymbol{\kappa}\right\Vert _{\min_{c}}%
^{2}-\kappa_{0}\sum\nolimits_{i=1}^{l_{1}}\psi_{1_{i_{\ast}}}%
\end{align*}
and $E_{\boldsymbol{\psi}_{2}}$%
\begin{align*}
E_{\boldsymbol{\psi}_{2}}  &  =\frac{1}{2}\sum\nolimits_{i=1}^{l}%
\psi_{_{i_{\ast}}}\left(  1-\xi_{i}\right)  +\kappa_{0}\sum\nolimits_{i=1}%
^{l_{2}}\psi_{2_{i_{\ast}}}\\
&  \equiv\frac{1}{2}\left\Vert \boldsymbol{\kappa}\right\Vert _{\min_{c}}%
^{2}+\kappa_{0}\sum\nolimits_{i=1}^{l_{2}}\psi_{2_{i_{\ast}}}%
\end{align*}
in terms of $\frac{1}{2}\left\Vert \boldsymbol{\kappa}\right\Vert _{\min_{c}%
}^{2}$ and a symmetric equalizer statistic.

Substituting the rewritten expressions for $E_{\boldsymbol{\psi}_{1}}$ and
$E_{\boldsymbol{\psi}_{2}}$ into Eqs (\ref{TAE Eigenlocus Component One Q})
and (\ref{TAE Eigenlocus Component Two Q}) produces the equations%
\[
\left(  \left\Vert \boldsymbol{\kappa}_{1}\right\Vert _{\min_{c}}%
^{2}-\left\Vert \boldsymbol{\kappa}_{1}\right\Vert \left\Vert
\boldsymbol{\kappa}_{2}\right\Vert \cos\theta_{\boldsymbol{\kappa}%
_{1}\boldsymbol{\kappa}_{2}}\right)  +\kappa_{0}\sum\nolimits_{i=1}^{l_{1}%
}\psi_{1_{i_{\ast}}}\equiv\frac{1}{2}\left\Vert \boldsymbol{\kappa}\right\Vert
_{\min_{c}}^{2}%
\]
and%
\[
\left(  \left\Vert \boldsymbol{\kappa}_{2}\right\Vert _{\min_{c}}%
^{2}-\left\Vert \boldsymbol{\kappa}_{2}\right\Vert \left\Vert
\boldsymbol{\kappa}_{1}\right\Vert \cos\theta_{\boldsymbol{\kappa}%
_{2}\boldsymbol{\kappa}_{1}}\right)  -\kappa_{0}\sum\nolimits_{i=1}^{l_{2}%
}\psi_{2_{i_{\ast}}}\equiv\frac{1}{2}\left\Vert \boldsymbol{\kappa}\right\Vert
_{\min_{c}}^{2}\text{,}%
\]
where the terms $\kappa_{0}\sum\nolimits_{i=1}^{l_{1}}\psi_{1_{i_{\ast}}}$ and
$-\kappa_{0}\sum\nolimits_{i=1}^{l_{2}}\psi_{2_{i_{\ast}}}$ specify a
symmetric equalizer statistic $\nabla_{eq}$ for integrals of class-conditional
probability density functions $p\left(  k_{\mathbf{x}_{1i\ast}}%
|\boldsymbol{\kappa}_{1}\right)  $ and $p\left(  k_{\mathbf{x}_{2_{i_{\ast}}}%
}|\boldsymbol{\kappa}_{2}\right)  $ that determine conditional probability
functions $P\left(  k_{\mathbf{x}_{1_{i\ast}}}|\boldsymbol{\kappa}_{1}\right)
$ and $P\left(  k_{\mathbf{x}_{2_{i\ast}}}|\boldsymbol{\kappa}_{2}\right)  $.

Therefore, let $\ \nabla_{eq}$ denote $\kappa_{0}\sum\nolimits_{i=1}^{l_{1}%
}\psi_{1_{i_{\ast}}}$ and $\kappa_{0}\sum\nolimits_{i=1}^{l_{2}}%
\psi_{2_{i_{\ast}}}$, where%
\[
\nabla_{eq}\triangleq\frac{\kappa_{0}}{2}\sum\nolimits_{i=1}^{l}\psi_{_{i\ast
}}\text{.}%
\]

It follows that the dual, class-conditional parameter vectors of likelihoods
$\boldsymbol{\psi}_{1}$, $\boldsymbol{\psi}_{2}$, $\boldsymbol{\kappa}_{1}$,
and $\boldsymbol{\kappa}_{2}$ satisfy the eigenlocus equations%
\begin{equation}
\left\Vert \boldsymbol{\kappa}_{1}\right\Vert _{\min_{c}}^{2}-\left\Vert
\boldsymbol{\kappa}_{1}\right\Vert \left\Vert \boldsymbol{\kappa}%
_{2}\right\Vert \cos\theta_{\boldsymbol{\kappa}_{1}\boldsymbol{\kappa}_{2}%
}+\nabla_{eq}\equiv\frac{1}{2}\left\Vert \boldsymbol{\kappa}\right\Vert
_{\min_{c}}^{2} \label{Balancing Feat SDEC1 Q}%
\end{equation}
and%
\begin{equation}
\left\Vert \boldsymbol{\kappa}_{2}\right\Vert _{\min_{c}}^{2}-\left\Vert
\boldsymbol{\kappa}_{2}\right\Vert \left\Vert \boldsymbol{\kappa}%
_{1}\right\Vert \cos\theta_{\boldsymbol{\kappa}_{2}\boldsymbol{\kappa}_{1}%
}-\nabla_{eq}\equiv\frac{1}{2}\left\Vert \boldsymbol{\kappa}\right\Vert
_{\min_{c}}^{2}\text{,} \label{Balancing Feat SDEC2 Q}%
\end{equation}
where $\nabla_{eq}\triangleq\frac{\kappa_{0}}{2}\sum\nolimits_{i=1}^{l}%
\psi_{_{i\ast}}$.

I will now examine the statistical and geometric properties of the equalizer
statistic $\nabla_{eq}$ in eigenspace.

\subsubsection{Properties of $\nabla_{eq}$ in Eigenspace}

Substituting the vector expression $\boldsymbol{\kappa}=\sum\nolimits_{i=1}%
^{l_{1}}\psi_{1_{i_{\ast}}}k_{\mathbf{x}_{1_{i_{\ast}}}}-\sum\nolimits_{i=1}%
^{l_{2}}\psi_{2_{i_{\ast}}}k_{\mathbf{x}_{2_{i_{\ast}}}}$ for
$\boldsymbol{\kappa}$ in Eq. (\ref{Pair of Normal Eigenlocus Components Q})
into the expression for $\kappa_{0}$ in Eq.
(\ref{Normal Eigenlocus Projection Factor Q}) produces the statistic for
$\kappa_{0}$:%
\begin{align}
\kappa_{0}  &  =-\sum\nolimits_{i=1}^{l}k_{\mathbf{x}_{i\ast}}\sum
\nolimits_{j=1}^{l_{1}}\psi_{1_{j_{\ast}}}k_{\mathbf{x}_{1_{j_{\ast}}}%
}\label{Eigenlocus Projection Factor Two Q}\\
&  +\sum\nolimits_{i=1}^{l}k_{\mathbf{x}_{i\ast}}\sum\nolimits_{j=1}^{l_{2}%
}\psi_{2_{j_{\ast}}}k_{\mathbf{x}_{2_{j_{\ast}}}}+\sum\nolimits_{i=1}^{l}%
y_{i}\left(  1-\xi_{i}\right)  \text{.}\nonumber
\end{align}

Substituting the statistic for $\kappa_{0}$ in Eq.
(\ref{Eigenlocus Projection Factor Two Q}) into the expression for
$\nabla_{eq}$ produces the statistic for $\nabla_{eq}$:%
\begin{align*}
\nabla_{eq}  &  =\frac{\kappa_{0}}{2}\sum\nolimits_{i=1}^{l}\psi_{_{i\ast}}\\
&  =-\left(  \sum\nolimits_{i=1}^{l}k_{\mathbf{x}_{i\ast}}\boldsymbol{\kappa
}_{1}\right)  \frac{1}{2}\sum\nolimits_{i=1}^{l}\psi_{_{i\ast}}\\
&  +\left(  \sum\nolimits_{i=1}^{l}k_{\mathbf{x}_{i\ast}}\boldsymbol{\kappa
}_{2}\right)  \frac{1}{2}\sum\nolimits_{i=1}^{l}\psi_{_{i\ast}}+\delta\left(
y\right)  \frac{1}{2}\sum\nolimits_{i=1}^{l}\psi_{_{i\ast}}\text{,}%
\end{align*}
where $\delta\left(  y\right)  \triangleq\sum\nolimits_{i=1}^{l}y_{i}\left(
1-\xi_{i}\right)  $. Let $k_{\widehat{\mathbf{x}}_{i\ast}}\triangleq
\sum\nolimits_{i=1}^{l}k_{\mathbf{x}_{i\ast}}$.

It follows that $\kappa_{0}$ regulates a symmetrical balancing act for
components of $k_{\widehat{\mathbf{x}}_{i\ast}}$ along $\boldsymbol{\kappa
}_{1}$ and $\boldsymbol{\kappa}_{2}$, where the statistic $\nabla_{eq}$ is
written as%
\[
+\nabla_{eq}=\left[  \operatorname{comp}_{\overrightarrow{\boldsymbol{\kappa
}_{2}}}\left(  \overrightarrow{k_{\widehat{\mathbf{x}}_{i\ast}}}\right)
-\operatorname{comp}_{\overrightarrow{\boldsymbol{\kappa}_{1}}}\left(
\overrightarrow{k_{\widehat{\mathbf{x}}_{i\ast}}}\right)  +\delta\left(
y\right)  \right]  \frac{1}{2}\sum\nolimits_{i=1}^{l}\psi_{_{i\ast}}%
\]
and%
\[
-\nabla_{eq}=\left[  \operatorname{comp}_{\overrightarrow{\boldsymbol{\kappa
}_{1}}}\left(  \overrightarrow{k_{\widehat{\mathbf{x}}_{i\ast}}}\right)
-\operatorname{comp}_{\overrightarrow{\boldsymbol{\kappa}_{2}}}\left(
\overrightarrow{k_{\widehat{\mathbf{x}}_{i\ast}}}\right)  -\delta\left(
y\right)  \right]  \frac{1}{2}\sum\nolimits_{i=1}^{l}\psi_{_{i\ast}}\text{.}%
\]

Returning to Eq. (\ref{Balanced Eigenlocus Equation Q}):%
\[
\sum\nolimits_{i=1}^{l_{1}}k_{\mathbf{x}_{1_{i\ast}}}\left(
\boldsymbol{\kappa}_{1}\mathbf{-}\boldsymbol{\kappa}_{2}\right)
=\sum\nolimits_{i=1}^{l_{2}}k_{\mathbf{x}_{2_{i\ast}}}\left(
\boldsymbol{\kappa}_{2}\mathbf{-}\boldsymbol{\kappa}_{1}\right)  \text{,}%
\]
given that components of $k_{\widehat{\mathbf{x}}_{1i\ast}}$ and
$k_{\widehat{\mathbf{x}}_{2i\ast}}$ along $\boldsymbol{\kappa}_{1}$ and
$\boldsymbol{\kappa}_{2}$ satisfy the state of statistical equilibrium%
\[
\left[  \operatorname{comp}_{\overrightarrow{\boldsymbol{\kappa}_{1}}}\left(
\overrightarrow{k_{\widehat{\mathbf{x}}_{1i\ast}}}\right)
-\operatorname{comp}_{\overrightarrow{\boldsymbol{\kappa}_{2}}}\left(
\overrightarrow{k_{\widehat{\mathbf{x}}_{1i\ast}}}\right)  \right]
\equiv\left[  \operatorname{comp}_{\overrightarrow{\boldsymbol{\kappa}_{2}}%
}\left(  \overrightarrow{k_{\widehat{\mathbf{x}}_{2i\ast}}}\right)
-\operatorname{comp}_{\overrightarrow{\boldsymbol{\kappa}_{1}}}\left(
\overrightarrow{k_{\widehat{\mathbf{x}}_{2i\ast}}}\right)  \right]  \text{,}%
\]
where $k_{\widehat{\mathbf{x}}_{1i\ast}}\triangleq\sum\nolimits_{i=1}^{l_{1}%
}k_{\mathbf{x}_{1_{i\ast}}}$ and $k_{\widehat{\mathbf{x}}_{2i\ast}}%
\triangleq\sum\nolimits_{i=1}^{l_{2}}k_{\mathbf{x}_{2_{i\ast}}}$, it follows
that%
\[
+\nabla_{eq}=\delta\left(  y\right)  \frac{1}{2}\sum\nolimits_{i=1}^{l}%
\psi_{_{i\ast}}\equiv\delta\left(  y\right)  \boldsymbol{\psi}_{1}%
\]
and%
\[
-\nabla_{eq}\equiv-\delta\left(  y\right)  \frac{1}{2}\sum\nolimits_{i=1}%
^{l}\psi_{_{i\ast}}\equiv-\delta\left(  y\right)  \boldsymbol{\psi}%
_{2}\text{.}%
\]

I\ will now demonstrate that the equalizer statistic $\nabla_{eq}$ ensures
that the class-conditional probability density functions $p\left(
k_{\mathbf{x}_{1i\ast}}|\boldsymbol{\kappa}_{1}\right)  $ and $p\left(
k_{\mathbf{x}_{2_{i_{\ast}}}}|\boldsymbol{\kappa}_{2}\right)  $ satisfy the
integral equation%
\begin{align*}
f\left(  \widetilde{\Lambda}_{\boldsymbol{\kappa}}\left(  \mathbf{s}\right)
\right)  =  &  \int_{Z}p\left(  k_{\mathbf{x}_{1_{i\ast}}}|\boldsymbol{\kappa
}_{1}\right)  d\boldsymbol{\kappa}_{1}+\delta\left(  y\right)  \frac{1}{2}%
\sum\nolimits_{i=1}^{l}\psi_{_{i\ast}}\\
&  =\int_{Z}p\left(  k_{\mathbf{x}_{2_{i\ast}}}|\boldsymbol{\kappa}%
_{2}\right)  d\boldsymbol{\kappa}_{2}-\delta\left(  y\right)  \frac{1}{2}%
\sum\nolimits_{i=1}^{l}\psi_{_{i\ast}}\text{,}%
\end{align*}
over the decision space $Z$: $Z=Z_{1}+Z_{2}$ and $Z_{1}\cong Z_{2}$, whereby
the likelihood ratio $\widehat{\Lambda}_{\boldsymbol{\kappa}}\left(
\mathbf{s}\right)  =p\left(  \widehat{\Lambda}_{\boldsymbol{\kappa}}\left(
\mathbf{s}\right)  |\omega_{1}\right)  -p\left(  \widehat{\Lambda
}_{\boldsymbol{\kappa}}\left(  \mathbf{s}\right)  |\omega_{2}\right)  $ of the
classification system $\left(  \mathbf{x}^{T}\mathbf{s}+1\right)
^{2}\boldsymbol{\kappa}+\kappa_{0}\overset{\omega_{1}}{\underset{\omega
_{2}}{\gtrless}}0$ is in statistical equilibrium. In the process, I will
formulate an equation which ensures that $\left\Vert \boldsymbol{\kappa}%
_{1}\right\Vert _{\min_{c}}^{2}$ and $\left\Vert \boldsymbol{\kappa}%
_{2}\right\Vert _{\min_{c}}^{2}$ are symmetrically balanced with each other.

\subsection{Quadratic Eigenlocus Integral Equation}

Let $\nabla_{eq}=\delta\left(  y\right)  \frac{1}{2}\sum\nolimits_{i=1}%
^{l}\psi_{_{i\ast}}$. Substituting the expression for $+\nabla_{eq}$ into Eq.
(\ref{Balancing Feat SDEC1 Q}) produces an equation that is satisfied by the
conditional probabilities of locations for the set $\left\{  k_{\mathbf{x}%
_{1_{i\ast}}}\right\}  _{i=1}^{l_{1}}$ of $k_{\mathbf{x}_{1_{i\ast}}}$ extreme
points within the decision space $Z$:%
\begin{align*}
P\left(  k_{\mathbf{x}_{1_{i\ast}}}|\boldsymbol{\kappa}_{1}\right)   &
=\left\Vert \boldsymbol{\kappa}_{1}\right\Vert _{\min_{c}}^{2}-\left\Vert
\boldsymbol{\kappa}_{1}\right\Vert \left\Vert \boldsymbol{\kappa}%
_{2}\right\Vert \cos\theta_{\boldsymbol{\kappa}_{1}\boldsymbol{\kappa}_{2}%
}+\delta\left(  y\right)  \frac{1}{2}\sum\nolimits_{i=1}^{l}\psi_{_{i\ast}}\\
&  \equiv\frac{1}{2}\left\Vert \boldsymbol{\kappa}\right\Vert _{\min_{c}}%
^{2}\text{,}%
\end{align*}
and substituting the expression for $-\nabla_{eq}$ into Eq.
(\ref{Balancing Feat SDEC2 Q}) produces an equation that is satisfied by the
conditional probabilities of locations for the set $\left\{  k_{\mathbf{x}%
_{2_{i\ast}}}\right\}  _{i=1}^{l_{2}}$ of $k_{\mathbf{x}_{2_{i\ast}}}$ extreme
points within the decision space $Z$:%
\begin{align*}
P\left(  k_{\mathbf{x}_{2_{i\ast}}}|\boldsymbol{\kappa}_{2}\right)   &
=\left\Vert \boldsymbol{\kappa}_{2}\right\Vert _{\min_{c}}^{2}-\left\Vert
\boldsymbol{\kappa}_{2}\right\Vert \left\Vert \boldsymbol{\kappa}%
_{1}\right\Vert \cos\theta_{\boldsymbol{\kappa}_{2}\boldsymbol{\kappa}_{1}%
}-\delta\left(  y\right)  \frac{1}{2}\sum\nolimits_{i=1}^{l}\psi_{_{i\ast}}\\
&  \equiv\frac{1}{2}\left\Vert \boldsymbol{\kappa}\right\Vert _{\min_{c}}%
^{2}\text{,}%
\end{align*}
where the equalizer statistic $\nabla_{eq}$%
\[
\nabla_{eq}=\delta\left(  y\right)  \frac{1}{2}\sum\nolimits_{i=1}^{l}%
\psi_{_{i\ast}}%
\]
\emph{equalizes} the conditional probabilities $P\left(  k_{\mathbf{x}%
_{1_{i\ast}}}|\boldsymbol{\kappa}_{1}\right)  $ and $P\left(  k_{\mathbf{x}%
_{2_{i\ast}}}|\boldsymbol{\kappa}_{2}\right)  $ of observing the set $\left\{
k_{\mathbf{x}_{1_{i\ast}}}\right\}  _{i=1}^{l_{1}}$ of $k_{\mathbf{x}%
_{1_{i\ast}}}$ extreme points and the set $\left\{  k_{\mathbf{x}_{2_{i\ast}}%
}\right\}  _{i=1}^{l_{2}}$ of $k_{\mathbf{x}_{2_{i\ast}}}$ extreme points
within the $Z_{1}$ and $Z_{2}$ decision regions of the decision space $Z$.

Therefore, the equalizer statistic $\delta\left(  y\right)  \frac{1}{2}%
\sum\nolimits_{i=1}^{l}\psi_{_{i\ast}}$ ensures that $\left\Vert
\boldsymbol{\kappa}_{1}\right\Vert _{\min_{c}}^{2}$ and $\left\Vert
\boldsymbol{\kappa}_{2}\right\Vert _{\min_{c}}^{2}$ are symmetrically balanced
with each other in the following manner:%
\[
\left\Vert \boldsymbol{\kappa}_{1}\right\Vert _{\min_{c}}^{2}-\left\Vert
\boldsymbol{\kappa}_{1}\right\Vert \left\Vert \boldsymbol{\kappa}%
_{2}\right\Vert \cos\theta_{\boldsymbol{\kappa}_{1}\boldsymbol{\kappa}_{2}%
}+\delta\left(  y\right)  \frac{1}{2}\sum\nolimits_{i=1}^{l}\psi_{_{i\ast}%
}\equiv\frac{1}{2}\left\Vert \boldsymbol{\kappa}\right\Vert _{\min_{c}}^{2}%
\]
and%
\[
\left\Vert \boldsymbol{\kappa}_{2}\right\Vert _{\min_{c}}^{2}-\left\Vert
\boldsymbol{\kappa}_{2}\right\Vert \left\Vert \boldsymbol{\kappa}%
_{1}\right\Vert \cos\theta_{\boldsymbol{\kappa}_{2}\boldsymbol{\kappa}_{1}%
}-\delta\left(  y\right)  \frac{1}{2}\sum\nolimits_{i=1}^{l}\psi_{_{i\ast}%
}\equiv\frac{1}{2}\left\Vert \boldsymbol{\kappa}\right\Vert _{\min_{c}}%
^{2}\text{.}%
\]

Thereby, the equalizer statistic%
\[
\delta\left(  y\right)  \frac{1}{2}\sum\nolimits_{i=1}^{l}\psi_{_{i\ast}}%
\]
\emph{equalizes} the total allowed eigenenergies $\left\Vert
\boldsymbol{\kappa}_{1}\right\Vert _{\min_{c}}^{2}$ and $\left\Vert
\boldsymbol{\kappa}_{2}\right\Vert _{\min_{c}}^{2}$ exhibited by
$\boldsymbol{\kappa}_{1}$ and $\boldsymbol{\kappa}_{2}$ so that the total
allowed eigenenergies $\left\Vert \boldsymbol{\kappa}_{1}-\boldsymbol{\kappa
}_{2}\right\Vert _{\min_{c}}^{2}$ exhibited by the scaled extreme points on
$\boldsymbol{\kappa}_{1}-\boldsymbol{\kappa}_{2}$ are symmetrically balanced
with each other about the fulcrum of $\boldsymbol{\kappa}$:%
\begin{equation}
\left\Vert \boldsymbol{\kappa}_{1}\right\Vert _{\min_{c}}^{2}+\delta\left(
y\right)  \frac{1}{2}\sum\nolimits_{i=1}^{l}\psi_{_{i\ast}}\equiv\left\Vert
\boldsymbol{\kappa}_{2}\right\Vert _{\min_{c}}^{2}-\delta\left(  y\right)
\frac{1}{2}\sum\nolimits_{i=1}^{l}\psi_{_{i\ast}}
\label{Symmetrical Balance of Total Allowed Eigenenergies Q}%
\end{equation}
which is located at the center of eigenenergy $\left\Vert \boldsymbol{\kappa
}\right\Vert _{\min_{c}}^{2}$: the geometric center of $\boldsymbol{\kappa}$.
Thus, the likelihood ratio%
\begin{align*}
\widehat{\Lambda}_{\boldsymbol{\kappa}}\left(  \mathbf{s}\right)   &
=p\left(  \widehat{\Lambda}_{\boldsymbol{\kappa}}\left(  \mathbf{s}\right)
|\omega_{1}\right)  -p\left(  \widehat{\Lambda}_{\boldsymbol{\kappa}}\left(
\mathbf{s}\right)  |\omega_{2}\right) \\
&  =\boldsymbol{\kappa}_{1}-\boldsymbol{\kappa}_{2}%
\end{align*}
of the classification system $\left(  \mathbf{x}^{T}\mathbf{s}+1\right)
^{2}\boldsymbol{\kappa}+\kappa_{0}\overset{\omega_{1}}{\underset{\omega
_{2}}{\gtrless}}0$ is in statistical equilibrium.

It follows that the eigenenergy $\left\Vert \boldsymbol{\kappa}_{1}\right\Vert
_{\min_{c}}^{2}$ associated with the position or location of the parameter
vector of likelihoods $p\left(  \widehat{\Lambda}_{\boldsymbol{\kappa}}\left(
\mathbf{s}\right)  |\omega_{1}\right)  $ given class $\omega_{1}$ is
symmetrically balanced with the eigenenergy $\left\Vert \boldsymbol{\kappa
}_{2}\right\Vert _{\min_{c}}^{2}$ associated with the position or location of
the parameter vector of likelihoods $p\left(  \widehat{\Lambda}%
_{\boldsymbol{\kappa}}\left(  \mathbf{s}\right)  |\omega_{2}\right)  $ given
class $\omega_{2}$.

Returning to Eq. (\ref{Conditional Probability Function for Class One Q})%
\[
P\left(  k_{\mathbf{x}_{1_{i\ast}}}|\boldsymbol{\kappa}_{1}\right)  =\int%
_{Z}\boldsymbol{\kappa}_{1}d\boldsymbol{\kappa}_{1}=\left\Vert
\boldsymbol{\kappa}_{1}\right\Vert _{\min_{c}}^{2}+C_{1}%
\]
and Eq. (\ref{Conditional Probability Function for Class Two Q})%
\[
P\left(  k_{\mathbf{x}_{2_{i\ast}}}|\boldsymbol{\kappa}_{2}\right)  =\int%
_{Z}\boldsymbol{\kappa}_{2}d\boldsymbol{\kappa}_{2}=\left\Vert
\boldsymbol{\kappa}_{2}\right\Vert _{\min_{c}}^{2}+C_{2}\text{,}%
\]
it follows that the value for the integration constant $C_{1}$ in Eq.
(\ref{Conditional Probability Function for Class One Q}) is%
\[
C_{1}=-\left\Vert \boldsymbol{\kappa}_{1}\right\Vert \left\Vert
\boldsymbol{\kappa}_{2}\right\Vert \cos\theta_{\boldsymbol{\kappa}%
_{1}\boldsymbol{\kappa}_{2}}%
\]
and the value for the integration constant $C_{2}$ in Eq.
(\ref{Conditional Probability Function for Class Two Q}) is%
\[
C_{2}=-\left\Vert \boldsymbol{\kappa}_{2}\right\Vert \left\Vert
\boldsymbol{\kappa}_{1}\right\Vert \cos\theta_{\boldsymbol{\kappa}%
_{2}\boldsymbol{\kappa}_{1}}\text{.}%
\]

Therefore, the area $P\left(  k_{\mathbf{x}_{1_{i\ast}}}|\boldsymbol{\kappa
}_{1}\right)  $ under the class-conditional density function $p\left(
k_{\mathbf{x}_{1_{i\ast}}}|\boldsymbol{\kappa}_{1}\right)  $ in Eq.
(\ref{Conditional Probability Function for Class One Q}):%
\begin{align}
P\left(  k_{\mathbf{x}_{1_{i\ast}}}|\boldsymbol{\kappa}_{1}\right)   &
=\int_{Z}p\left(  k_{\mathbf{x}_{1_{i\ast}}}|\boldsymbol{\kappa}_{1}\right)
d\boldsymbol{\kappa}_{1}+\delta\left(  y\right)  \frac{1}{2}\sum
\nolimits_{i=1}^{l}\psi_{_{i\ast}}\label{Integral Equation Class One Q}\\
&  =\int_{Z}\boldsymbol{\kappa}_{1}d\boldsymbol{\kappa}_{1}+\delta\left(
y\right)  \frac{1}{2}\sum\nolimits_{i=1}^{l}\psi_{_{i\ast}}\nonumber\\
&  =\left\Vert \boldsymbol{\kappa}_{1}\right\Vert _{\min_{c}}^{2}-\left\Vert
\boldsymbol{\kappa}_{1}\right\Vert \left\Vert \boldsymbol{\kappa}%
_{2}\right\Vert \cos\theta_{\boldsymbol{\kappa}_{1}\boldsymbol{\kappa}_{2}%
}+\delta\left(  y\right)  \frac{1}{2}\sum\nolimits_{i=1}^{l}\psi_{_{i\ast}%
}\nonumber\\
&  =\left\Vert \boldsymbol{\kappa}_{1}\right\Vert _{\min_{c}}^{2}-\left\Vert
\boldsymbol{\kappa}_{1}\right\Vert \left\Vert \boldsymbol{\kappa}%
_{2}\right\Vert \cos\theta_{\boldsymbol{\kappa}_{1}\boldsymbol{\kappa}_{2}%
}+\delta\left(  y\right)  \sum\nolimits_{i=1}^{l_{1}}\psi_{1_{i_{\ast}}%
}\nonumber\\
&  \equiv\frac{1}{2}\left\Vert \boldsymbol{\kappa}\right\Vert _{\min_{c}}%
^{2}\text{,}\nonumber
\end{align}
over the decision space $Z$, is \emph{symmetrically balanced} with the area
$P\left(  k_{\mathbf{x}_{2_{i\ast}}}|\boldsymbol{\kappa}_{2}\right)  $ under
the class-conditional density function $p\left(  k_{\mathbf{x}_{2_{i_{\ast}}}%
}|\boldsymbol{\kappa}_{2}\right)  $ in Eq.
(\ref{Conditional Probability Function for Class Two Q}):%
\begin{align}
P\left(  k_{\mathbf{x}_{2_{i\ast}}}|\boldsymbol{\kappa}_{2}\right)   &
=\int_{Z}p\left(  k_{\mathbf{x}_{2_{i\ast}}}|\boldsymbol{\kappa}_{2}\right)
d\boldsymbol{\kappa}_{2}-\delta\left(  y\right)  \frac{1}{2}\sum
\nolimits_{i=1}^{l}\psi_{_{i\ast}}\label{Integral Equation Class Two Q}\\
&  =\int_{Z}\boldsymbol{\kappa}_{2}d\boldsymbol{\kappa}_{2}-\delta\left(
y\right)  \frac{1}{2}\sum\nolimits_{i=1}^{l}\psi_{_{i\ast}}\nonumber\\
&  =\left\Vert \boldsymbol{\kappa}_{2}\right\Vert _{\min_{c}}^{2}-\left\Vert
\boldsymbol{\kappa}_{2}\right\Vert \left\Vert \boldsymbol{\kappa}%
_{1}\right\Vert \cos\theta_{\boldsymbol{\kappa}_{2}\boldsymbol{\kappa}_{1}%
}-\delta\left(  y\right)  \frac{1}{2}\sum\nolimits_{i=1}^{l}\psi_{_{i\ast}%
}\nonumber\\
&  =\left\Vert \boldsymbol{\kappa}_{2}\right\Vert _{\min_{c}}^{2}-\left\Vert
\boldsymbol{\kappa}_{2}\right\Vert \left\Vert \boldsymbol{\kappa}%
_{1}\right\Vert \cos\theta_{\boldsymbol{\kappa}_{2}\boldsymbol{\kappa}_{1}%
}-\delta\left(  y\right)  \sum\nolimits_{i=1}^{l_{2}}\psi_{2_{i_{\ast}}%
}\nonumber\\
&  \equiv\frac{1}{2}\left\Vert \boldsymbol{\kappa}\right\Vert _{\min_{c}}%
^{2}\text{,}\nonumber
\end{align}
over the decision space $Z$, where the area $P\left(  k_{\mathbf{x}_{1_{i\ast
}}}|\boldsymbol{\kappa}_{1}\right)  $ under $p\left(  k_{\mathbf{x}_{1_{i\ast
}}}|\boldsymbol{\kappa}_{1}\right)  $ and the area $P\left(  k_{\mathbf{x}%
_{2_{i\ast}}}|\boldsymbol{\kappa}_{2}\right)  $ under $p\left(  k_{\mathbf{x}%
_{2_{i_{\ast}}}}|\boldsymbol{\kappa}_{2}\right)  $ are constrained to be equal
to $\frac{1}{2}\left\Vert \boldsymbol{\kappa}\right\Vert _{\min_{c}}^{2}$:%
\[
P\left(  k_{\mathbf{x}_{1_{i\ast}}}|\boldsymbol{\kappa}_{1}\right)  =P\left(
k_{\mathbf{x}_{2_{i\ast}}}|\boldsymbol{\kappa}_{2}\right)  \equiv\frac{1}%
{2}\left\Vert \boldsymbol{\kappa}\right\Vert _{\min_{c}}^{2}\text{.}%
\]

It follows that the quadratic eigenlocus discriminant function
$\widetilde{\Lambda}_{\boldsymbol{\kappa}}\left(  \mathbf{s}\right)  =\left(
\mathbf{x}^{T}\mathbf{s}+1\right)  ^{2}\boldsymbol{\kappa}+\kappa_{0}$ is the
solution to the integral equation%
\begin{align}
f\left(  \widetilde{\Lambda}_{\boldsymbol{\kappa}}\left(  \mathbf{s}\right)
\right)  =  &  \int_{Z_{1}}\boldsymbol{\kappa}_{1}d\boldsymbol{\kappa}%
_{1}+\int_{Z_{2}}\boldsymbol{\kappa}_{1}d\boldsymbol{\kappa}_{1}+\delta\left(
y\right)  \sum\nolimits_{i=1}^{l_{1}}\psi_{1_{i_{\ast}}}%
\label{Quadratic Eigenlocus Integral Equation}\\
&  =\int_{Z_{1}}\boldsymbol{\kappa}_{2}d\boldsymbol{\kappa}_{2}+\int_{Z_{2}%
}\boldsymbol{\kappa}_{2}d\boldsymbol{\kappa}_{2}-\delta\left(  y\right)
\sum\nolimits_{i=1}^{l_{2}}\psi_{2_{i_{\ast}}}\text{,}\nonumber
\end{align}
over the decision space $Z=Z_{1}+Z_{2}$, where the dual likelihood ratios
$\widehat{\Lambda}_{\boldsymbol{\psi}}\left(  \mathbf{s}\right)
=\boldsymbol{\psi}_{1}+\boldsymbol{\psi}_{2}$ and $\widehat{\Lambda
}_{\boldsymbol{\kappa}}\left(  \mathbf{s}\right)  =\boldsymbol{\kappa}%
_{1}-\boldsymbol{\kappa}_{2}$ are in statistical equilibrium, all of the
forces associated with the counter risks $\overline{\mathfrak{R}%
}_{\mathfrak{\min}}\left(  Z_{1}|\boldsymbol{\kappa}_{1}\right)  $ and the
risks $\mathfrak{R}_{\mathfrak{\min}}\left(  Z_{2}|\boldsymbol{\kappa}%
_{1}\right)  $ in the $Z_{1}$ and $Z_{2}$ decision regions: which are related
to positions and potential locations of reproducing kernels $k_{\mathbf{x}%
_{1_{i\ast}}}$ of extreme points $\mathbf{x}_{1_{i_{\ast}}}$ that are
generated according to $p\left(  \mathbf{x}|\omega_{1}\right)  $, are equal to
all of the forces associated with the risks $\mathfrak{R}_{\mathfrak{\min}%
}\left(  Z_{1}|\boldsymbol{\kappa}_{2}\right)  $ and the counter risks
$\overline{\mathfrak{R}}_{\mathfrak{\min}}\left(  Z_{2}|\boldsymbol{\kappa
}_{2}\right)  $ in the $Z_{1}$ and $Z_{2}$ decision regions: which are related
to positions and potential locations of reproducing kernels $k_{\mathbf{x}%
_{2_{i\ast}}}$ of extreme points $\mathbf{x}_{2_{i_{\ast}}}$ that are
generated according to $p\left(  \mathbf{x}|\omega_{2}\right)  $, and the
eigenenergy associated with the position or location of the likelihood ratio
$p\left(  \widehat{\Lambda}_{\boldsymbol{\kappa}}\left(  \mathbf{s}\right)
|\omega_{1}\right)  $ given class $\omega_{1}$ is equal to the eigenenergy
associated with the position or location of the likelihood ratio $p\left(
\widehat{\Lambda}_{\boldsymbol{\kappa}}\left(  \mathbf{s}\right)  |\omega
_{2}\right)  $ given class $\omega_{2}$.

So, let $p\left(  \frac{k_{\mathbf{x}_{1i\ast}}}{\left\Vert k_{\mathbf{x}%
_{1i\ast}}\right\Vert }|\boldsymbol{\psi}_{1}\right)  $ and $p\left(
\frac{k_{\mathbf{x}_{2i\ast}}}{\left\Vert k_{\mathbf{x}_{2i\ast}}\right\Vert
}|\boldsymbol{\psi}_{2}\right)  $ denote the Wolfe dual parameter vectors of
likelihoods $\boldsymbol{\psi}_{1}=\sum\nolimits_{i=1}^{l_{1}}\psi
_{1_{i_{\ast}}}\frac{k_{\mathbf{x}_{1i\ast}}}{\left\Vert k_{\mathbf{x}%
_{1i\ast}}\right\Vert }$ and $\boldsymbol{\psi}_{2}=\sum\nolimits_{i=1}%
^{l_{2}}\psi_{2_{i_{\ast}}}\frac{k_{\mathbf{x}_{2i\ast}}}{\left\Vert
k_{\mathbf{x}_{2i\ast}}\right\Vert }$. It follows that the class-conditional
probability density functions $p\left(  \frac{k_{\mathbf{x}_{1i\ast}}%
}{\left\Vert k_{\mathbf{x}_{1i\ast}}\right\Vert }|\boldsymbol{\psi}%
_{1}\right)  $ and $p\left(  \frac{k_{\mathbf{x}_{2i\ast}}}{\left\Vert
k_{\mathbf{x}_{2i\ast}}\right\Vert }|\boldsymbol{\psi}_{2}\right)  $ in the
Wolfe dual eigenspace $\widetilde{Z}$ and the class-conditional probability
density functions $p\left(  k_{\mathbf{x}_{1i\ast}}|\boldsymbol{\kappa}%
_{1}\right)  $ and $p\left(  k_{\mathbf{x}_{2i\ast}}|\boldsymbol{\kappa}%
_{2}\right)  $ in the decision space $Z$ satisfy the integral equation%
\begin{align*}
f\left(  \widetilde{\Lambda}_{\boldsymbol{\kappa}}\left(  \mathbf{s}\right)
\right)  =  &  \int_{Z}p\left(  k_{\mathbf{x}_{1i\ast}}|\boldsymbol{\kappa
}_{1}\right)  d\boldsymbol{\kappa}_{1}+\delta\left(  y\right)  p\left(
\frac{k_{\mathbf{x}_{1i\ast}}}{\left\Vert k_{\mathbf{x}_{1i\ast}}\right\Vert
}|\boldsymbol{\psi}_{1}\right) \\
&  =\int_{Z}p\left(  k_{\mathbf{x}_{2i\ast}}|\boldsymbol{\kappa}_{2}\right)
d\boldsymbol{\kappa}_{2}-\delta\left(  y\right)  p\left(  \frac{k_{\mathbf{x}%
_{2i\ast}}}{\left\Vert k_{\mathbf{x}_{2i\ast}}\right\Vert }|\boldsymbol{\psi
}_{2}\right)  \text{,}%
\end{align*}
over the decision space $Z$, where $Z=Z_{1}+Z_{2}$, $Z_{1}\simeq Z_{2}$, and
$Z\subset%
\mathbb{R}
^{d}$.

Thus, it is concluded that the quadratic eigenlocus discriminant function%
\[
\widetilde{\Lambda}_{\boldsymbol{\kappa}}\left(  \mathbf{s}\right)  =\left(
k_{\mathbf{x}}-k_{\widehat{\mathbf{x}}_{i\ast}}\right)  \left(
\boldsymbol{\kappa}_{1}-\boldsymbol{\kappa}_{2}\right)  \mathbf{+}%
\sum\nolimits_{i=1}^{l}y_{i}\left(  1-\xi_{i}\right)
\]
is the solution to the integral equation:%
\begin{align}
f\left(  \widetilde{\Lambda}_{\boldsymbol{\kappa}}\left(  \mathbf{s}\right)
\right)  =  &  \int_{Z_{1}}p\left(  k_{\mathbf{x}_{1i\ast}}|\boldsymbol{\kappa
}_{1}\right)  d\boldsymbol{\kappa}_{1}+\int_{Z_{2}}p\left(  k_{\mathbf{x}%
_{1i\ast}}|\boldsymbol{\kappa}_{1}\right)  d\boldsymbol{\kappa}_{1}%
\label{Quadratic Eigenlocus Integral Equation I}\\
&  +\delta\left(  y\right)  p\left(  \frac{k_{\mathbf{x}_{1i\ast}}}{\left\Vert
k_{\mathbf{x}_{1i\ast}}\right\Vert }|\boldsymbol{\psi}_{1}\right) \nonumber\\
&  =\int_{Z_{1}}p\left(  k_{\mathbf{x}_{2i\ast}}|\boldsymbol{\kappa}%
_{2}\right)  d\boldsymbol{\kappa}_{2}+\int_{Z_{2}}p\left(  k_{\mathbf{x}%
_{2i\ast}}|\boldsymbol{\kappa}_{2}\right)  d\boldsymbol{\kappa}_{2}\nonumber\\
&  -\delta\left(  y\right)  p\left(  \frac{k_{\mathbf{x}_{2i\ast}}}{\left\Vert
k_{\mathbf{x}_{2i\ast}}\right\Vert }|\boldsymbol{\psi}_{2}\right)
\text{,}\nonumber
\end{align}
over the decision space $Z=Z_{1}+Z_{2}$, where $\delta\left(  y\right)
\triangleq\sum\nolimits_{i=1}^{l}y_{i}\left(  1-\xi_{i}\right)  $, the
integral $\int_{Z_{1}}p\left(  k_{\mathbf{x}_{1i\ast}}|\boldsymbol{\kappa}%
_{1}\right)  d\boldsymbol{\kappa}_{1}$ accounts for all of the forces
associated with the counter risks $\overline{\mathfrak{R}}_{\mathfrak{\min}%
}\left(  Z_{1}|\psi_{1i\ast}k_{\mathbf{x}_{1i\ast}}\right)  $ which are
related to positions and potential locations of corresponding $k_{\mathbf{x}%
_{1i\ast}}$ extreme points that lie in the $Z_{1}$ decision region, the
integral $\int_{Z_{2}}p\left(  k_{\mathbf{x}_{1i\ast}}|\boldsymbol{\kappa}%
_{1}\right)  d\boldsymbol{\kappa}_{1}$ accounts for all of the forces
associated with the risks $\mathfrak{R}_{\mathfrak{\min}}\left(  Z_{2}%
|\psi_{1i\ast}k_{\mathbf{x}_{1i\ast}}\right)  $ which are related to positions
and potential locations of corresponding $k_{\mathbf{x}_{1i\ast}}$ extreme
points that lie in the $Z_{2}$ decision region, the integral $\int_{Z_{1}%
}p\left(  k_{\mathbf{x}_{2i\ast}}|\boldsymbol{\kappa}_{2}\right)
d\boldsymbol{\kappa}_{2}$ accounts for all of the forces associated with the
risks $\mathfrak{R}_{\mathfrak{\min}}\left(  Z_{1}|\psi_{2i\ast}%
k_{\mathbf{x}_{2i\ast}}\right)  $ which are related to positions and potential
locations of corresponding $k_{\mathbf{x}_{2i\ast}}$ extreme points that lie
in the $Z_{1}$ decision region, and the integral $\int_{Z_{2}}p\left(
k_{\mathbf{x}_{2i\ast}}|\boldsymbol{\kappa}_{2}\right)  d\boldsymbol{\kappa
}_{2}$ accounts for all of the forces associated with the counter risks
$\overline{\mathfrak{R}}_{\mathfrak{\min}}\left(  Z_{2}|\psi_{2i\ast
}k_{\mathbf{x}_{2i\ast}}\right)  $ which are related to positions and
potential locations of corresponding $k_{\mathbf{x}_{2i\ast}}$ extreme points
that lie in the $Z_{2}$ decision region. The equalizer statistics
$+\delta\left(  y\right)  p\left(  \frac{k_{\mathbf{x}_{1i\ast}}}{\left\Vert
k_{\mathbf{x}_{1i\ast}}\right\Vert }|\boldsymbol{\psi}_{1}\right)  $ and
$-\delta\left(  y\right)  p\left(  \frac{k_{\mathbf{x}_{2i\ast}}}{\left\Vert
k_{\mathbf{x}_{2i\ast}}\right\Vert }|\boldsymbol{\psi}_{2}\right)  $ ensure
that the collective forces associated with the expected risks $\mathfrak{R}%
_{\mathfrak{\min}}\left(  Z|\boldsymbol{\kappa}_{1}\right)  $ and
$\mathfrak{R}_{\mathfrak{\min}}\left(  Z|\boldsymbol{\kappa}_{2}\right)  $ for
class $\omega_{1}$ and class $\omega_{2}$, which are given by the respective
integrals $\int_{Z}p\left(  k_{\mathbf{x}_{1i\ast}}|\boldsymbol{\kappa}%
_{1}\right)  d\boldsymbol{\kappa}_{1}$ and $\int_{Z}p\left(  k_{\mathbf{x}%
_{2i\ast}}|\boldsymbol{\kappa}_{2}\right)  d\boldsymbol{\kappa}_{2}$, are
symmetrically balanced with each other.

Therefore, the classification system%
\[
\left(  k_{\mathbf{x}}-k_{\widehat{\mathbf{x}}_{i\ast}}\right)  \left(
\boldsymbol{\kappa}_{1}-\boldsymbol{\kappa}_{2}\right)  +\delta\left(
y\right)  \overset{\omega_{1}}{\underset{\omega_{2}}{\gtrless}}0
\]
is in statistical equilibrium:%
\begin{align}
f\left(  \widetilde{\Lambda}_{\boldsymbol{\kappa}}\left(  \mathbf{s}\right)
\right)  :  &  \int_{Z_{1}}p\left(  k_{\mathbf{x}_{1i\ast}}|\boldsymbol{\kappa
}_{1}\right)  d\boldsymbol{\kappa}_{1}-\int_{Z_{1}}p\left(  k_{\mathbf{x}%
_{2i\ast}}|\boldsymbol{\kappa}_{2}\right)  d\boldsymbol{\kappa}_{2}%
\label{Quadratic Eigenlocus Integral Equation II}\\
&  +\delta\left(  y\right)  p\left(  \frac{k_{\mathbf{x}_{1i\ast}}}{\left\Vert
k_{\mathbf{x}_{1i\ast}}\right\Vert }|\boldsymbol{\psi}_{1}\right) \nonumber\\
&  =\int_{Z_{2}}p\left(  k_{\mathbf{x}_{2i\ast}}|\boldsymbol{\kappa}%
_{2}\right)  d\boldsymbol{\kappa}_{2}-\int_{Z_{2}}p\left(  k_{\mathbf{x}%
_{1i\ast}}|\boldsymbol{\kappa}_{1}\right)  d\boldsymbol{\kappa}_{1}\nonumber\\
&  -\delta\left(  y\right)  p\left(  \frac{k_{\mathbf{x}_{2i\ast}}}{\left\Vert
k_{\mathbf{x}_{2i\ast}}\right\Vert }|\boldsymbol{\psi}_{2}\right)
\text{,}\nonumber
\end{align}
where all of the forces associated with the counter risk $\overline
{\mathfrak{R}}_{\mathfrak{\min}}\left(  Z_{1}|\boldsymbol{\kappa}_{1}\right)
$ and the risk $\mathfrak{R}_{\mathfrak{\min}}\left(  Z_{1}|\boldsymbol{\kappa
}_{2}\right)  $ in the $Z_{1}$ decision region are symmetrically balanced with
all of the forces associated with the counter risk $\overline{\mathfrak{R}%
}_{\mathfrak{\min}}\left(  Z_{2}|\boldsymbol{\kappa}_{2}\right)  $ and the
risk $\mathfrak{R}_{\mathfrak{\min}}\left(  Z_{2}|\boldsymbol{\kappa}%
_{1}\right)  $ in the $Z_{2}$ decision region:%
\begin{align*}
f\left(  \widetilde{\Lambda}_{\boldsymbol{\kappa}}\left(  \mathbf{s}\right)
\right)   &  :\overline{\mathfrak{R}}_{\mathfrak{\min}}\left(  Z_{1}|p\left(
\widehat{\Lambda}_{\boldsymbol{\kappa}}\left(  \mathbf{s}\right)  |\omega
_{1}\right)  \right)  -\mathfrak{R}_{\mathfrak{\min}}\left(  Z_{1}|p\left(
\widehat{\Lambda}_{\boldsymbol{\kappa}}\left(  \mathbf{s}\right)  |\omega
_{2}\right)  \right) \\
&  =\overline{\mathfrak{R}}_{\mathfrak{\min}}\left(  Z_{2}|p\left(
\widehat{\Lambda}_{\boldsymbol{\kappa}}\left(  \mathbf{s}\right)  |\omega
_{2}\right)  \right)  -\mathfrak{R}_{\mathfrak{\min}}\left(  Z_{2}|p\left(
\widehat{\Lambda}_{\boldsymbol{\kappa}}\left(  \mathbf{s}\right)  |\omega
_{1}\right)  \right)
\end{align*}
such that the expected risk $\mathfrak{R}_{\mathfrak{\min}}\left(
Z|\widehat{\Lambda}_{\boldsymbol{\kappa}}\left(  \mathbf{s}\right)  \right)  $
of the classification system is minimized, and the eigenenergies associated
with the counter risk $\overline{\mathfrak{R}}_{\mathfrak{\min}}\left(
Z_{1}|p\left(  \widehat{\Lambda}_{\boldsymbol{\kappa}}\left(  \mathbf{s}%
\right)  |\omega_{1}\right)  \right)  $ and the risk $\mathfrak{R}%
_{\mathfrak{\min}}\left(  Z_{1}|p\left(  \widehat{\Lambda}_{\boldsymbol{\kappa
}}\left(  \mathbf{s}\right)  |\omega_{2}\right)  \right)  $ in the $Z_{1}$
decision region are balanced with the eigenenergies associated with the
counter risk $\overline{\mathfrak{R}}_{\mathfrak{\min}}\left(  Z_{2}|p\left(
\widehat{\Lambda}_{\boldsymbol{\kappa}}\left(  \mathbf{s}\right)  |\omega
_{2}\right)  \right)  $ and the risk $\mathfrak{R}_{\mathfrak{\min}}\left(
Z_{2}|p\left(  \widehat{\Lambda}_{\boldsymbol{\kappa}}\left(  \mathbf{s}%
\right)  |\omega_{1}\right)  \right)  $ in the $Z_{2}$ decision region:%
\begin{align*}
f\left(  \widetilde{\Lambda}_{\boldsymbol{\kappa}}\left(  \mathbf{s}\right)
\right)   &  :E_{\min}\left(  Z_{1}|p\left(  \widehat{\Lambda}%
_{\boldsymbol{\kappa}}\left(  \mathbf{s}\right)  |\omega_{1}\right)  \right)
-E_{\min}\left(  Z_{1}|p\left(  \widehat{\Lambda}_{\boldsymbol{\kappa}}\left(
\mathbf{s}\right)  |\omega_{2}\right)  \right) \\
&  =E_{\min}\left(  Z_{2}|p\left(  \widehat{\Lambda}_{\boldsymbol{\kappa}%
}\left(  \mathbf{s}\right)  |\omega_{2}\right)  \right)  -E_{\min}\left(
Z_{2}|p\left(  \widehat{\Lambda}_{\boldsymbol{\kappa}}\left(  \mathbf{s}%
\right)  |\omega_{1}\right)  \right)
\end{align*}
such that the eigenenergy $E_{\min}\left(  Z|\widehat{\Lambda}%
_{\boldsymbol{\kappa}}\left(  \mathbf{s}\right)  \right)  $ of the
classification system is minimized.

Thus, the locus of principal eigenaxis components on $\boldsymbol{\kappa
}=\boldsymbol{\kappa}_{1}-\boldsymbol{\kappa}_{2}$ satisfies the integral
equation%
\begin{align*}
f\left(  \widetilde{\Lambda}_{\boldsymbol{\kappa}}\left(  \mathbf{s}\right)
\right)  =  &  \int_{Z}\boldsymbol{\kappa}_{1}d\boldsymbol{\kappa}_{1}%
+\delta\left(  y\right)  \frac{1}{2}\sum\nolimits_{i=1}^{l}\psi_{_{i\ast}}\\
&  =\int_{Z}\boldsymbol{\kappa}_{2}d\boldsymbol{\kappa}_{2}-\delta\left(
y\right)  \frac{1}{2}\sum\nolimits_{i=1}^{l}\psi_{_{i\ast}}\\
&  \equiv\frac{1}{2}\left\Vert \boldsymbol{\kappa}\right\Vert _{\min_{c}}%
^{2}\text{,}%
\end{align*}
over the decision space $Z=Z_{1}+Z_{2}$, where $\boldsymbol{\kappa}_{1}$ and
$\boldsymbol{\kappa}_{2}$ are components of a principal eigenaxis
$\boldsymbol{\kappa}$, and $\left\Vert \boldsymbol{\kappa}\right\Vert
_{\min_{c}}^{2}$ is the total allowed eigenenergy exhibited by
$\boldsymbol{\kappa}$.

The above integral equation can be written as:%
\begin{align}
f\left(  \widetilde{\Lambda}_{\boldsymbol{\kappa}}\left(  \mathbf{s}\right)
\right)  =  &  \int_{Z_{1}}\boldsymbol{\kappa}_{1}d\boldsymbol{\kappa}%
_{1}+\int_{Z_{2}}\boldsymbol{\kappa}_{1}d\boldsymbol{\kappa}_{1}+\delta\left(
y\right)  \frac{1}{2}\sum\nolimits_{i=1}^{l}\psi_{_{i\ast}}%
\label{Quadratic Eigenlocus Integral Equation III}\\
&  =\int_{Z_{1}}\boldsymbol{\kappa}_{2}d\boldsymbol{\kappa}_{2}+\int_{Z_{2}%
}\boldsymbol{\kappa}_{2}d\boldsymbol{\kappa}_{2}-\delta\left(  y\right)
\frac{1}{2}\sum\nolimits_{i=1}^{l}\psi_{_{i\ast}}\text{,}\nonumber
\end{align}
where the integral $\int_{Z_{1}}\boldsymbol{\kappa}_{1}d\boldsymbol{\kappa
}_{1}$ accounts for all of the eigenenergies $\left\Vert \psi_{1_{i_{\ast}}%
}k_{\mathbf{x}_{1i\ast}}\right\Vert _{\min_{c}}^{2}$ exhibited by all of the
$k_{\mathbf{x}_{1i\ast}}$ extreme points that lie in the $Z_{1}$ decision
region, the integral $\int_{Z_{2}}\boldsymbol{\kappa}_{1}d\boldsymbol{\kappa
}_{1}$ accounts for all of the eigenenergies $\left\Vert \psi_{1_{i_{\ast}}%
}k_{\mathbf{x}_{1i\ast}}\right\Vert _{\min_{c}}^{2}$ exhibited by all of the
$k_{\mathbf{x}_{1i\ast}}$ extreme points that lie in the $Z_{2}$ decision
region, the integral $\int_{Z_{1}}\boldsymbol{\kappa}_{2}d\boldsymbol{\kappa
}_{2}$ accounts for all of the eigenenergies $\left\Vert \psi_{2_{i_{\ast}}%
}k_{\mathbf{x}_{2i\ast}}\right\Vert _{\min_{c}}^{2}$ exhibited by all of the
$k_{\mathbf{x}_{2i\ast}}$ extreme points that lie in the $Z_{2}$ decision
region, and the integral $\int_{Z_{2}}\boldsymbol{\kappa}_{2}%
d\boldsymbol{\kappa}_{2}$ accounts for all of the eigenenergies $\left\Vert
\psi_{2_{i_{\ast}}}k_{\mathbf{x}_{2i\ast}}\right\Vert _{\min_{c}}^{2}$
exhibited by all of the $k_{\mathbf{x}_{2i\ast}}$ extreme points that lie in
the $Z_{1}$ decision region. The equalizer statistics $+\delta\left(
y\right)  \frac{1}{2}\sum\nolimits_{i=1}^{l}\psi_{_{i\ast}}$ and
$-\delta\left(  y\right)  \frac{1}{2}\sum\nolimits_{i=1}^{l}\psi_{_{i\ast}}$
ensure that the integrals $\int_{Z}\boldsymbol{\kappa}_{1}d\boldsymbol{\kappa
}_{1}$ and $\int_{Z}\boldsymbol{\kappa}_{2}d\boldsymbol{\kappa}_{2}$ are
symmetrically balanced with each other.

Equation (\ref{Quadratic Eigenlocus Integral Equation III}) can be rewritten
as%
\begin{align}
f\left(  \widetilde{\Lambda}_{\boldsymbol{\kappa}}\left(  \mathbf{s}\right)
\right)  =  &  \int_{Z_{1}}\boldsymbol{\kappa}_{1}d\boldsymbol{\kappa}%
_{1}-\int_{Z_{1}}\boldsymbol{\kappa}_{2}d\boldsymbol{\kappa}_{2}+\delta\left(
y\right)  \frac{1}{2}\sum\nolimits_{i=1}^{l}\psi_{_{i\ast}}%
\label{Quadratic Eigenlocus Integral Equation IV}\\
&  =\int_{Z_{2}}\boldsymbol{\kappa}_{2}d\boldsymbol{\kappa}_{2}-\int_{Z_{2}%
}\boldsymbol{\kappa}_{1}d\boldsymbol{\kappa}_{1}-\delta\left(  y\right)
\frac{1}{2}\sum\nolimits_{i=1}^{l}\psi_{_{i\ast}}\text{,}\nonumber
\end{align}
where all of the eigenenergies $\left\Vert \psi_{1_{i_{\ast}}}k_{\mathbf{x}%
_{1i\ast}}\right\Vert _{\min_{c}}^{2}$ and $\left\Vert \psi_{2_{i_{\ast}}%
}k_{\mathbf{x}_{2i\ast}}\right\Vert _{\min_{c}}^{2}$ associated with the
counter risk $\overline{\mathfrak{R}}_{\mathfrak{\min}}\left(  Z_{1}%
|\mathbf{\kappa}_{1}\right)  $ and the risk $\mathfrak{R}_{\mathfrak{\min}%
}\left(  Z_{1}|\boldsymbol{\kappa}_{2}\right)  $ in the $Z_{1}$ decision
region are \emph{symmetrically balanced} with all of the eigenenergies
$\left\Vert \psi_{1_{i_{\ast}}}k_{\mathbf{x}_{1i\ast}}\right\Vert _{\min_{c}%
}^{2}$ and $\left\Vert \psi_{2_{i_{\ast}}}k_{\mathbf{x}_{2i\ast}}\right\Vert
_{\min_{c}}^{2}$ associated with the counter risk $\overline{\mathfrak{R}%
}_{\mathfrak{\min}}\left(  Z_{2}|\boldsymbol{\kappa}_{2}\right)  $ and the
risk $\mathfrak{R}_{\mathfrak{\min}}\left(  Z_{2}|\mathbf{\kappa}_{1}\right)
$ in the $Z_{2}$ decision region.

Given Eqs (\ref{Quadratic Eigenlocus Integral Equation I}) -
(\ref{Quadratic Eigenlocus Integral Equation IV}), it is concluded that
quadratic eigenlocus discriminant functions $\widetilde{\Lambda}%
_{\boldsymbol{\kappa}}\left(  \mathbf{s}\right)  =\left(  \mathbf{x}%
^{T}\mathbf{s}+1\right)  ^{2}\boldsymbol{\kappa}+\kappa_{0}$ satisfy discrete
and data-driven versions of the integral equation of binary classification in
Eq. (\ref{Integral Equation of Likelihood Ratio and Decision Boundary}), the
fundamental integral equation of binary classification for a classification
system in statistical equilibrium in Eq. (\ref{Equalizer Rule}), and the
corresponding integral equation for a classification system in statistical
equilibrium in Eq. (\ref{Balancing of Bayes' Risks and Counteracting Risks}).

\subsection{Equilibrium Points of Integral Equations II}

Returning to the binary classification theorem, recall that the expected risk
$\mathfrak{R}_{\mathfrak{\min}}\left(  Z|\widehat{\Lambda}\left(
\mathbf{x}\right)  \right)  $ and the corresponding eigenenergy $E_{\min
}\left(  Z|\widehat{\Lambda}\left(  \mathbf{x}\right)  \right)  $ of a
classification system $p\left(  \widehat{\Lambda}\left(  \mathbf{x}\right)
|\omega_{1}\right)  -p\left(  \widehat{\Lambda}\left(  \mathbf{x}\right)
|\omega_{2}\right)  \overset{\omega_{1}}{\underset{\omega_{2}}{\gtrless}}0$
are governed by the equilibrium point%
\[
p\left(  \widehat{\Lambda}\left(  \mathbf{x}\right)  |\omega_{1}\right)
-p\left(  \widehat{\Lambda}\left(  \mathbf{x}\right)  |\omega_{2}\right)  =0
\]
of the integral equation%
\begin{align*}
f\left(  \widehat{\Lambda}\left(  \mathbf{x}\right)  \right)   &  =\int%
_{Z_{1}}p\left(  \widehat{\Lambda}\left(  \mathbf{x}\right)  |\omega
_{1}\right)  d\widehat{\Lambda}+\int_{Z_{2}}p\left(  \widehat{\Lambda}\left(
\mathbf{x}\right)  |\omega_{1}\right)  d\widehat{\Lambda}\\
&  =\int_{Z_{1}}p\left(  \widehat{\Lambda}\left(  \mathbf{x}\right)
|\omega_{2}\right)  d\widehat{\Lambda}+\int_{Z_{2}}p\left(  \widehat{\Lambda
}\left(  \mathbf{x}\right)  |\omega_{2}\right)  d\widehat{\Lambda}\text{,}%
\end{align*}
over the decision space $Z=Z_{1}+Z_{2}$, where the equilibrium point $p\left(
\widehat{\Lambda}\left(  \mathbf{x}\right)  |\omega_{1}\right)  -p\left(
\widehat{\Lambda}\left(  \mathbf{x}\right)  |\omega_{2}\right)  =0$ is the
focus of a decision boundary $D\left(  \mathbf{x}\right)  $.

Returning to Eq. (\ref{Wolfe Dual Equilibrium Point Q}):%
\[
\sum\nolimits_{i=1}^{l_{1}}\psi_{1i\ast}-\sum\nolimits_{i=1}^{l_{2}}%
\psi_{2i\ast}=0\text{,}%
\]
it follows that the Wolfe dual quadratic eigenlocus $\boldsymbol{\psi}$ of
likelihoods and principal eigenaxis components $\widehat{\Lambda
}_{\boldsymbol{\psi}}\left(  \mathbf{s}\right)  $%
\begin{align*}
\boldsymbol{\psi}  &  =\sum\nolimits_{i=1}^{l}\psi_{i\ast}\frac{k_{\mathbf{x}%
_{i\ast}}}{\left\Vert k_{\mathbf{x}_{i\ast}}\right\Vert }\\
&  =\sum\nolimits_{i=1}^{l_{1}}\psi_{1i\ast}\frac{k_{\mathbf{x}_{1i\ast}}%
}{\left\Vert k_{\mathbf{x}_{1i\ast}}\right\Vert }+\sum\nolimits_{i=1}^{l_{2}%
}\psi_{2i\ast}\frac{k_{\mathbf{x}_{2i\ast}}}{\left\Vert k_{\mathbf{x}_{2i\ast
}}\right\Vert }\\
&  =\boldsymbol{\psi}_{1}+\boldsymbol{\psi}_{2}%
\end{align*}
is the \emph{equilibrium point} $p\left(  \widehat{\Lambda}_{\boldsymbol{\psi
}}\left(  \mathbf{s}\right)  |\omega_{1}\right)  -p\left(  \widehat{\Lambda
}_{\boldsymbol{\psi}}\left(  \mathbf{s}\right)  |\omega_{2}\right)  =0$:%
\[
\sum\nolimits_{i=1}^{l_{1}}\psi_{1i\ast}\frac{k_{\mathbf{x}_{1i\ast}}%
}{\left\Vert k_{\mathbf{x}_{1i\ast}}\right\Vert }-\sum\nolimits_{i=1}^{l_{2}%
}\psi_{2i\ast}\frac{k_{\mathbf{x}_{2i\ast}}}{\left\Vert k_{\mathbf{x}_{2i\ast
}}\right\Vert }=0
\]
of the integral equation in Eq. (\ref{Quadratic Eigenlocus Integral Equation})
and all of its derivatives in Eqs
(\ref{Quadratic Eigenlocus Integral Equation I}) -
(\ref{Quadratic Eigenlocus Integral Equation IV}).

Therefore, it is concluded that the expected risk $\mathfrak{R}%
_{\mathfrak{\min}}\left(  Z|\widehat{\Lambda}_{\boldsymbol{\kappa}}\left(
\mathbf{s}\right)  \right)  $ and the eigenenergy $E_{\min}\left(
Z|\widehat{\Lambda}_{\boldsymbol{\kappa}}\left(  \mathbf{s}\right)  \right)  $
of the classification system $\left(  \mathbf{x}^{T}\mathbf{s}+1\right)
^{2}\boldsymbol{\kappa}+\kappa_{0}\overset{\omega_{1}}{\underset{\omega
_{2}}{\gtrless}}0$ are governed by the equilibrium point $p\left(
\widehat{\Lambda}_{\boldsymbol{\psi}}\left(  \mathbf{s}\right)  |\omega
_{1}\right)  -p\left(  \widehat{\Lambda}_{\boldsymbol{\psi}}\left(
\mathbf{s}\right)  |\omega_{2}\right)  =0$:%
\[
\sum\nolimits_{i=1}^{l_{1}}\psi_{1i\ast}\frac{k_{\mathbf{x}_{1i\ast}}%
}{\left\Vert k_{\mathbf{x}_{1i\ast}}\right\Vert }-\sum\nolimits_{i=1}^{l_{2}%
}\psi_{2i\ast}\frac{k_{\mathbf{x}_{2i\ast}}}{\left\Vert k_{\mathbf{x}_{2i\ast
}}\right\Vert }=0
\]
of the integral equation $f\left(  \widetilde{\Lambda}_{\boldsymbol{\kappa}%
}\left(  \mathbf{s}\right)  \right)  $:%
\begin{align*}
f\left(  \widetilde{\Lambda}_{\boldsymbol{\kappa}}\left(  \mathbf{s}\right)
\right)  =  &  \int_{Z_{1}}p\left(  k_{\mathbf{x}_{1i\ast}}|\boldsymbol{\kappa
}_{1}\right)  d\boldsymbol{\kappa}_{1}+\int_{Z_{2}}p\left(  k_{\mathbf{x}%
_{1i\ast}}|\boldsymbol{\kappa}_{1}\right)  d\boldsymbol{\kappa}_{1}\\
&  +\delta\left(  y\right)  p\left(  \frac{k_{\mathbf{x}_{1i\ast}}}{\left\Vert
k_{\mathbf{x}_{1i\ast}}\right\Vert }|\boldsymbol{\psi}_{1}\right) \\
&  =\int_{Z_{1}}p\left(  k_{\mathbf{x}_{2i\ast}}|\boldsymbol{\kappa}%
_{2}\right)  d\boldsymbol{\kappa}_{2}+\int_{Z_{2}}p\left(  k_{\mathbf{x}%
_{2i\ast}}|\boldsymbol{\kappa}_{2}\right)  d\boldsymbol{\kappa}_{2}\\
&  -\delta\left(  y\right)  p\left(  \frac{k_{\mathbf{x}_{2i\ast}}}{\left\Vert
k_{\mathbf{x}_{2i\ast}}\right\Vert }|\boldsymbol{\psi}_{2}\right)  \text{,}%
\end{align*}
over the decision space $Z=Z_{1}+Z_{2}$, where the equilibrium point
$\sum\nolimits_{i=1}^{l_{1}}\psi_{1i\ast}\frac{k_{\mathbf{x}_{1i\ast}}%
}{\left\Vert k_{\mathbf{x}_{1i\ast}}\right\Vert }-\sum\nolimits_{i=1}^{l_{2}%
}\psi_{2i\ast}\frac{k_{\mathbf{x}_{2i\ast}}}{\left\Vert k_{\mathbf{x}_{2i\ast
}}\right\Vert }=0$ is the dual focus of a quadratic decision boundary
$D\left(  \mathbf{s}\right)  $.

I will now develop an integral equation that explicitly accounts for the
primal focus $\widehat{\Lambda}_{\boldsymbol{\kappa}}\left(  \mathbf{s}%
\right)  $ of a quadratic decision boundary $D\left(  \mathbf{s}\right)  $ and
the equilibrium point $\widehat{\Lambda}_{\boldsymbol{\psi}}\left(
\mathbf{s}\right)  $ of the integral equation $f\left(  \widetilde{\Lambda
}_{\boldsymbol{\kappa}}\left(  \mathbf{s}\right)  \right)  $.

\section{The Balancing Feat in Dual Space II}

Let $\boldsymbol{\kappa}=\boldsymbol{\kappa}_{1}-\boldsymbol{\kappa}_{2}$ and
substitute the statistic for $\kappa_{0}$ in Eq.
(\ref{Eigenlocus Projection Factor Two Q})%
\begin{align*}
\kappa_{0}  &  =-\sum\nolimits_{i=1}^{l}k_{\mathbf{x}_{i\ast}}\sum
\nolimits_{j=1}^{l_{1}}\psi_{1_{j_{\ast}}}k_{\mathbf{x}_{1_{j_{\ast}}}}\\
&  +\sum\nolimits_{i=1}^{l}k_{\mathbf{x}_{i\ast}}\sum\nolimits_{j=1}^{l_{2}%
}\psi_{2_{j_{\ast}}}k_{\mathbf{x}_{2_{j_{\ast}}}}+\sum\nolimits_{i=1}^{l}%
y_{i}\left(  1-\xi_{i}\right)
\end{align*}
into Eq. (\ref{Minimum Eigenenergy Class One Q})%
\[
\psi_{1_{i_{\ast}}}k_{\mathbf{x}_{1_{i_{\ast}}}}\boldsymbol{\kappa}%
=\psi_{1_{i_{\ast}}}\left(  1-\xi_{i}-\kappa_{0}\right)  ,\ i=1,...,l_{1}%
\text{,}%
\]
where each conditional density $\psi_{1_{i_{\ast}}}\frac{k_{\mathbf{x}%
_{1i\ast}}}{\left\Vert k_{\mathbf{x}_{1i\ast}}\right\Vert }$ of an
$k_{\mathbf{x}_{1i\ast}}$ extreme point satisfies the identity:%
\[
\psi_{1_{i_{\ast}}}\left(  1-\xi_{i}\right)  \equiv\psi_{1_{i_{\ast}}%
}k_{\mathbf{x}_{1_{i_{\ast}}}}\boldsymbol{\kappa}+\psi_{1_{i_{\ast}}}%
\kappa_{0}\text{.}%
\]

Accordingly, the above identity can be rewritten in terms of an eigenlocus
equation that is satisfied by the conditional density $\psi_{1_{i_{\ast}}%
}\frac{k_{\mathbf{x}_{1i\ast}}}{\left\Vert k_{\mathbf{x}_{1i\ast}}\right\Vert
}$ of an $k_{\mathbf{x}_{1i\ast}}$ extreme point:%
\begin{align}
\psi_{1_{i_{\ast}}}  &  =\psi_{1_{i_{\ast}}}k_{\mathbf{x}_{1_{i_{\ast}}}%
}\left(  \boldsymbol{\kappa}_{1}-\boldsymbol{\kappa}_{2}\right)
\label{Pointwise Conditional Density Constraint One Q}\\
&  +\psi_{1_{i_{\ast}}}\left\{  \sum\nolimits_{j=1}^{l}k_{\mathbf{x}_{j\ast}%
}\left(  \boldsymbol{\kappa}_{2}-\boldsymbol{\kappa}_{1}\right)  \right\}
\nonumber\\
&  +\xi_{i}\psi_{1_{i_{\ast}}}+\delta\left(  y\right)  \psi_{1_{i_{\ast}}%
}\text{,}\nonumber
\end{align}
where $\delta\left(  y\right)  \triangleq\sum\nolimits_{i=1}^{l}y_{i}\left(
1-\xi_{i}\right)  $, $\psi_{1_{i_{\ast}}}k_{\mathbf{x}_{1_{i_{\ast}}}}$ is a
principal eigenaxis component on $\boldsymbol{\kappa}_{1}$, and the set of
scaled extreme vectors\emph{\ }$\psi_{1_{i_{\ast}}}\left(  \sum\nolimits_{j=1}%
^{l}k_{\mathbf{x}_{j\ast}}\right)  $ are symmetrically distributed over
$\boldsymbol{\kappa}_{2}-\boldsymbol{\kappa}_{1}$:%
\[
\psi_{1_{i_{\ast}}}\left(  \sum\nolimits_{j=1}^{l}k_{\mathbf{x}_{j\ast}%
}\right)  \boldsymbol{\kappa}_{2}-\psi_{1_{i_{\ast}}}\left(  \sum
\nolimits_{j=1}^{l}k_{\mathbf{x}_{j\ast}}\right)  \boldsymbol{\kappa}%
_{1}\text{.}%
\]

Again, let $\boldsymbol{\kappa}=\boldsymbol{\kappa}_{1}-\boldsymbol{\kappa
}_{2}$. Substitute the statistic for $\kappa_{0}$ in Eq.
(\ref{Eigenlocus Projection Factor Two Q})%
\begin{align*}
\kappa_{0}  &  =-\sum\nolimits_{i=1}^{l}k_{\mathbf{x}_{i\ast}}\sum
\nolimits_{j=1}^{l_{1}}\psi_{1_{j_{\ast}}}k_{\mathbf{x}_{1_{j_{\ast}}}}\\
&  +\sum\nolimits_{i=1}^{l}k_{\mathbf{x}_{i\ast}}\sum\nolimits_{j=1}^{l_{2}%
}\psi_{2_{j_{\ast}}}k_{\mathbf{x}_{2_{j_{\ast}}}}+\sum\nolimits_{i=1}^{l}%
y_{i}\left(  1-\xi_{i}\right)
\end{align*}
into Eq. (\ref{Minimum Eigenenergy Class Two Q})%
\[
-\psi_{2_{i_{\ast}}}k_{\mathbf{x}_{2_{i_{\ast}}}}\boldsymbol{\kappa}%
=\psi_{2_{i_{\ast}}}\left(  1-\xi_{i}+\kappa_{0}\right)  ,\ i=1,...,l_{2}%
\text{,}%
\]
where each conditional density $\psi_{2_{i_{\ast}}}\frac{k_{\mathbf{x}%
_{2i\ast}}}{\left\Vert k_{\mathbf{x}_{2i\ast}}\right\Vert }$ of an
$k_{\mathbf{x}_{2i\ast}}$ extreme point satisfies the identity:%
\[
\psi_{2_{i_{\ast}}}\left(  1-\xi_{i}\right)  =-\psi_{2_{i_{\ast}}%
}k_{\mathbf{x}_{2_{i_{\ast}}}}\boldsymbol{\kappa}-\psi_{2_{i_{\ast}}}%
\kappa_{0}\text{.}%
\]

Accordingly, the above identity can be rewritten in terms of an eigenlocus
equation that is satisfied by the conditional density $\psi_{2_{i_{\ast}}%
}\frac{k_{\mathbf{x}_{2i\ast}}}{\left\Vert k_{\mathbf{x}_{2i\ast}}\right\Vert
}$ of an $k_{\mathbf{x}_{2i\ast}}$ extreme point:%
\begin{align}
\psi_{2_{i_{\ast}}}  &  =\psi_{2_{i_{\ast}}}k_{\mathbf{x}_{2_{i_{\ast}}}%
}\left(  \boldsymbol{\kappa}_{2}-\boldsymbol{\kappa}_{1}\right)
\label{Pointwise Conditional Density Constraint Two Q}\\
&  +\psi_{2_{i_{\ast}}}\left\{  \sum\nolimits_{j=1}^{l}k_{\mathbf{x}_{j\ast}%
}\left(  \boldsymbol{\kappa}_{1}-\boldsymbol{\kappa}_{2}\right)  \right\}
\nonumber\\
&  +\xi_{i}\psi_{2_{i_{\ast}}}-\delta\left(  y\right)  \psi_{2_{i_{\ast}}%
}\text{,}\nonumber
\end{align}
where $\delta\left(  y\right)  \triangleq\sum\nolimits_{i=1}^{l}y_{i}\left(
1-\xi_{i}\right)  $, $\psi_{2_{i_{\ast}}}\mathbf{x}_{2_{i_{\ast}}}$ is a
principal eigenaxis component on $\boldsymbol{\kappa}_{2}$, and the set of
scaled extreme vectors $\psi_{2_{i_{\ast}}}\left(  \sum\nolimits_{j=1}%
^{l}k_{\mathbf{x}_{j\ast}}\right)  $ are symmetrically distributed over
$\boldsymbol{\kappa}_{1}-\boldsymbol{\kappa}_{2}$:%
\[
\psi_{2_{i_{\ast}}}\left(  \sum\nolimits_{j=1}^{l}k_{\mathbf{x}_{j\ast}%
}\right)  \boldsymbol{\kappa}_{1}-\psi_{2_{i_{\ast}}}\left(  \sum
\nolimits_{j=1}^{l}k_{\mathbf{x}_{j\ast}}\right)  \boldsymbol{\kappa}%
_{2}\text{.}%
\]
Using Eqs (\ref{Integral Equation Class One Q}) and
(\ref{Pointwise Conditional Density Constraint One Q}), it follows that the
conditional probability $P\left(  k_{\mathbf{x}_{1_{i\ast}}}%
|\boldsymbol{\kappa}_{1}\right)  $ of observing the set $\left\{
k_{\mathbf{x}_{1_{i_{\ast}}}}\right\}  _{i=1}^{l_{1}}$ of $k_{\mathbf{x}%
_{1_{i_{\ast}}}}$ extreme points within localized regions of the decision
space $Z$ is determined by the eigenlocus equation:%
\begin{align}
P\left(  k_{\mathbf{x}_{1_{i\ast}}}|\boldsymbol{\kappa}_{1}\right)   &
=\sum\nolimits_{i=1}^{l_{1}}\psi_{1_{i_{\ast}}}k_{\mathbf{x}_{1_{i_{\ast}}}%
}\left(  \boldsymbol{\kappa}_{1}-\boldsymbol{\kappa}_{2}\right)
\label{Bayes' Risk One Q}\\
&  +\sum\nolimits_{i=1}^{l_{1}}\psi_{1_{i_{\ast}}}\left\{  \sum\nolimits_{j=1}%
^{l}k_{\mathbf{x}_{j\ast}}\left(  \boldsymbol{\kappa}_{2}-\boldsymbol{\kappa
}_{1}\right)  \right\} \nonumber\\
&  +\delta\left(  y\right)  \sum\nolimits_{i=1}^{l_{1}}\psi_{1_{i_{\ast}}%
}+\sum\nolimits_{i=1}^{l_{1}}\xi_{i}\psi_{1_{i_{\ast}}}\nonumber\\
&  \equiv\sum\nolimits_{i=1}^{l_{1}}\psi_{1_{i_{\ast}}}\text{,}\nonumber
\end{align}
where $P\left(  k_{\mathbf{x}_{1_{i\ast}}}|\boldsymbol{\kappa}_{1}\right)  $
evaluates to $\sum\nolimits_{i=1}^{l_{1}}\psi_{1_{i_{\ast}}}$, and scaled
extreme vectors are symmetrically distributed over $\boldsymbol{\kappa}%
_{2}-\boldsymbol{\kappa}_{1}$ in the following manner:%
\[
\left(  \sum\nolimits_{i=1}^{l_{1}}\psi_{1_{i_{\ast}}}\sum\nolimits_{j=1}%
^{l}k_{\mathbf{x}_{j\ast}}\right)  \boldsymbol{\kappa}_{2}-\left(
\sum\nolimits_{i=1}^{l_{1}}\psi_{1_{i_{\ast}}}\sum\nolimits_{j=1}%
^{l}k_{\mathbf{x}_{j\ast}}\right)  \boldsymbol{\kappa}_{1}\text{.}%
\]

Using Eqs (\ref{Integral Equation Class Two Q}) and
(\ref{Pointwise Conditional Density Constraint Two Q}), it follows that the
conditional probability $P\left(  k_{\mathbf{x}_{2_{i\ast}}}%
|\boldsymbol{\kappa}_{2}\right)  $ of observing the set $\left\{
k_{\mathbf{x}_{2_{i\ast}}}\right\}  _{i=1}^{l_{2}}$ of $k_{\mathbf{x}%
_{2_{i\ast}}}$ extreme points within localized regions of the decision space
$Z$ is determined by the eigenlocus equation:%
\begin{align}
P\left(  k_{\mathbf{x}_{2_{i\ast}}}|\boldsymbol{\kappa}_{2}\right)   &
=\sum\nolimits_{i=1}^{l_{2}}\psi_{2_{i_{\ast}}}k_{\mathbf{x}_{2_{i_{\ast}}}%
}\left(  \boldsymbol{\kappa}_{2}-\boldsymbol{\kappa}_{1}\right)
\label{Bayes' Risk Two Q}\\
&  -\sum\nolimits_{i=1}^{l_{2}}\psi_{2_{i_{\ast}}}\left\{  \sum\nolimits_{j=1}%
^{l}k_{\mathbf{x}_{j\ast}}\left(  \boldsymbol{\kappa}_{1}-\boldsymbol{\kappa
}_{2}\right)  \right\} \nonumber\\
&  -\delta\left(  y\right)  \sum\nolimits_{i=1}^{l_{2}}\psi_{2_{i_{\ast}}%
}+\sum\nolimits_{i=1}^{l_{2}}\xi_{i}\psi_{2_{i_{\ast}}}\nonumber\\
&  \equiv\sum\nolimits_{i=1}^{l_{2}}\psi_{2_{i_{\ast}}}\text{,}\nonumber
\end{align}
where $P\left(  k_{\mathbf{x}_{2_{i\ast}}}|\boldsymbol{\kappa}_{2}\right)  $
evaluates to $\sum\nolimits_{i=1}^{l_{2}}\psi_{2_{i_{\ast}}}$, and scaled
extreme vectors are symmetrically distributed over $\boldsymbol{\kappa}%
_{1}-\boldsymbol{\kappa}_{2}$ in the following manner:%
\[
\left(  \sum\nolimits_{i=1}^{l_{2}}\psi_{2_{i_{\ast}}}\sum\nolimits_{j=1}%
^{l}k_{\mathbf{x}_{j\ast}}\right)  \boldsymbol{\kappa}_{1}-\left(
\sum\nolimits_{i=1}^{l_{2}}\psi_{2_{i_{\ast}}}\sum\nolimits_{j=1}%
^{l}k_{\mathbf{x}_{j\ast}}\right)  \boldsymbol{\kappa}_{2}\text{.}%
\]

I will now use Eqs (\ref{Bayes' Risk One Q}) and (\ref{Bayes' Risk Two Q}) to
devise an equilibrium equation that determines the overall manner in which
quadratic eigenlocus discriminant functions $\widetilde{\Lambda}%
_{\boldsymbol{\kappa}}\left(  \mathbf{s}\right)  =\left(  \mathbf{x}%
^{T}\mathbf{s}+1\right)  ^{2}\boldsymbol{\kappa}+\kappa_{0}$ minimize the
expected risk $\mathfrak{R}_{\mathfrak{\min}}\left(  Z|\boldsymbol{\kappa
}\right)  $ and the total allowed eigenenergy $\left\Vert \boldsymbol{\kappa
}\right\Vert _{\min_{c}}^{2}$ for a given decision space $Z$.

\subsection{Minimization of Risk $\mathfrak{R}_{\mathfrak{\min}}\left(
Z|\boldsymbol{\kappa}\right)  $ and Eigenenergy$\left\Vert \boldsymbol{\kappa
}\right\Vert _{\min_{c}}^{2}$}

Take the estimates in Eqs (\ref{Bayes' Risk One Q}) and
(\ref{Bayes' Risk Two Q}):%
\[
P\left(  k_{\mathbf{x}_{1_{i\ast}}}|\boldsymbol{\kappa}_{1}\right)
=\sum\nolimits_{i=1}^{l_{1}}\psi_{1_{i_{\ast}}}%
\]
and%
\[
P\left(  k_{\mathbf{x}_{2_{i\ast}}}|\boldsymbol{\kappa}_{2}\right)
=\sum\nolimits_{i=1}^{l_{2}}\psi_{2_{i_{\ast}}}%
\]
for the conditional probabilities $P\left(  k_{\mathbf{x}_{1_{i\ast}}%
}|\boldsymbol{\kappa}_{1}\right)  $ and $P\left(  k_{\mathbf{x}_{2_{i\ast}}%
}|\boldsymbol{\kappa}_{2}\right)  $ of observing the $k_{\mathbf{x}_{1_{i\ast
}}}$ and $k_{\mathbf{x}_{2_{i\ast}}}$ extreme points within localized regions
of the decision space $Z$.

Given that the Wolfe dual eigenlocus of principal eigenaxis components and
likelihoods:%
\begin{align*}
\widehat{\Lambda}_{\boldsymbol{\psi}}\left(  \mathbf{s}\right)   &
=\boldsymbol{\psi}_{1}+\boldsymbol{\psi}_{2}\\
&  =\sum\nolimits_{i=1}^{l_{1}}\psi_{1i\ast}\frac{k_{\mathbf{x}_{1i\ast}}%
}{\left\Vert k_{\mathbf{x}_{1i\ast}}\right\Vert }+\sum\nolimits_{i=1}^{l_{2}%
}\psi_{2i\ast}\frac{k_{\mathbf{x}_{2i\ast}}}{\left\Vert k_{\mathbf{x}_{2i\ast
}}\right\Vert }%
\end{align*}
satisfies the equilibrium equation:%
\[
\sum\nolimits_{i=1}^{l_{1}}\psi_{1i\ast}\frac{k_{\mathbf{x}_{1i\ast}}%
}{\left\Vert k_{\mathbf{x}_{1i\ast}}\right\Vert }=\sum\nolimits_{i=1}^{l_{2}%
}\psi_{2i\ast}\frac{k_{\mathbf{x}_{2i\ast}}}{\left\Vert k_{\mathbf{x}_{2i\ast
}}\right\Vert }\text{,}%
\]
it follows that the conditional probabilities of observing the $k_{\mathbf{x}%
_{1_{i\ast}}}$ and the $k_{\mathbf{x}_{2_{i\ast}}}$ extreme points within
localized regions of the decision space $Z$ are equal to each other:%
\[
P\left(  k_{\mathbf{x}_{1_{i\ast}}}|\boldsymbol{\kappa}_{1}\right)  =P\left(
k_{\mathbf{x}_{2_{i\ast}}}|\boldsymbol{\kappa}_{2}\right)  \text{.}%
\]

Accordingly, set the vector expressions in Eqs (\ref{Bayes' Risk One Q}) and
(\ref{Bayes' Risk Two Q}) equal to each other:%
\begin{align*}
&  \sum\nolimits_{i=1}^{l_{1}}\psi_{1_{i_{\ast}}}k_{\mathbf{x}_{1_{i_{\ast}}}%
}\left(  \boldsymbol{\kappa}_{1}-\boldsymbol{\kappa}_{2}\right)
+\sum\nolimits_{i=1}^{l_{1}}\psi_{1_{i_{\ast}}}\left\{  \sum\nolimits_{j=1}%
^{l}k_{\mathbf{x}_{j\ast}}\left(  \boldsymbol{\kappa}_{2}-\boldsymbol{\kappa
}_{1}\right)  \right\} \\
&  +\delta\left(  y\right)  \sum\nolimits_{i=1}^{l_{1}}\psi_{1_{i_{\ast}}%
}+\sum\nolimits_{i=1}^{l_{1}}\xi_{i}\psi_{1_{i_{\ast}}}\\
&  =\sum\nolimits_{i=1}^{l_{2}}\psi_{2_{i_{\ast}}}k_{\mathbf{x}_{2_{i_{\ast}}%
}}\left(  \boldsymbol{\kappa}_{2}-\boldsymbol{\kappa}_{1}\right)
-\sum\nolimits_{i=1}^{l_{2}}\psi_{2_{i_{\ast}}}\left\{  \sum\nolimits_{j=1}%
^{l}k_{\mathbf{x}_{j\ast}}\left(  \boldsymbol{\kappa}_{1}-\boldsymbol{\kappa
}_{2}\right)  \right\} \\
&  -\delta\left(  y\right)  \sum\nolimits_{i=1}^{l_{2}}\psi_{2_{i_{\ast}}%
}+\sum\nolimits_{i=1}^{l_{2}}\xi_{i}\psi_{2_{i_{\ast}}}\text{.}%
\end{align*}

It follows that the equilibrium equation:%
\begin{align}
&  \left\Vert \boldsymbol{\kappa}_{1}\right\Vert _{\min_{c}}^{2}-\left\Vert
\boldsymbol{\kappa}_{1}\right\Vert \left\Vert \boldsymbol{\kappa}%
_{2}\right\Vert \cos\theta_{\boldsymbol{\kappa}_{1}\boldsymbol{\kappa}_{2}%
}+\delta\left(  y\right)  \sum\nolimits_{i=1}^{l_{1}}\psi_{1_{i_{\ast}}%
}\label{Balancing Bayes' Risk Quadratic}\\
&  +\sum\nolimits_{i=1}^{l_{1}}\psi_{1_{i_{\ast}}}\left\{  \sum\nolimits_{j=1}%
^{l}k_{\mathbf{x}_{j\ast}}\left(  \boldsymbol{\kappa}_{2}-\boldsymbol{\kappa
}_{1}\right)  \right\}  +\sum\nolimits_{i=1}^{l_{1}}\xi_{i}\psi_{1_{i_{\ast}}%
}\nonumber\\
&  =\left\Vert \boldsymbol{\kappa}_{2}\right\Vert _{\min_{c}}^{2}-\left\Vert
\boldsymbol{\kappa}_{2}\right\Vert \left\Vert \boldsymbol{\kappa}%
_{1}\right\Vert \cos\theta_{\boldsymbol{\kappa}_{2}\boldsymbol{\kappa}_{1}%
}-\delta\left(  y\right)  \sum\nolimits_{i=1}^{l_{2}}\psi_{2_{i_{\ast}}%
}\nonumber\\
&  -\sum\nolimits_{i=1}^{l_{2}}\psi_{2_{i_{\ast}}}\left\{  \sum\nolimits_{j=1}%
^{l}k_{\mathbf{x}_{j\ast}}\left(  \boldsymbol{\kappa}_{1}-\boldsymbol{\kappa
}_{2}\right)  \right\}  +\sum\nolimits_{i=1}^{l_{2}}\xi_{i}\psi_{2_{i_{\ast}}%
}\nonumber
\end{align}
is satisfied by the equilibrium point:%
\[
p\left(  \widehat{\Lambda}_{\boldsymbol{\psi}}\left(  \mathbf{s}\right)
|\omega_{1}\right)  -p\left(  \widehat{\Lambda}_{\boldsymbol{\psi}}\left(
\mathbf{s}\right)  |\omega_{2}\right)  =0
\]
and the likelihood ratio:%
\[
\widehat{\Lambda}_{\boldsymbol{\kappa}}\left(  \mathbf{s}\right)  =p\left(
\widehat{\Lambda}_{\boldsymbol{\kappa}}\left(  \mathbf{s}\right)  |\omega
_{1}\right)  -p\left(  \widehat{\Lambda}_{\boldsymbol{\kappa}}\left(
\mathbf{s}\right)  |\omega_{2}\right)
\]
of the classification system $\left(  \mathbf{x}^{T}\mathbf{s}+1\right)
^{2}\boldsymbol{\kappa}+\kappa_{0}\overset{\omega_{1}}{\underset{\omega
_{2}}{\gtrless}}0$, where the equilibrium equation in Eq.
(\ref{Balancing Bayes' Risk Quadratic}) is constrained by the equilibrium
point in Eq. (\ref{Wolfe Dual Equilibrium Point Q}).

I will now use Eq. (\ref{Balancing Bayes' Risk Quadratic}) to develop a
fundamental quadratic eigenlocus integral equation of binary classification
for a classification system in statistical equilibrium.

\subsection{Fundamental Balancing Feat in Dual Spaces II}

Returning to Eq. (\ref{Conditional Probability Function for Class One Q})%
\begin{align*}
P\left(  k_{\mathbf{x}_{1_{i\ast}}}|\boldsymbol{\kappa}_{1}\right)   &
=\int_{Z}p\left(  k_{\mathbf{x}_{1_{i\ast}}}|\boldsymbol{\kappa}_{1}\right)
d\boldsymbol{\kappa}_{1}\\
&  =\int_{Z}\boldsymbol{\kappa}_{1}d\boldsymbol{\kappa}_{1}=\left\Vert
\boldsymbol{\kappa}_{1}\right\Vert _{\min_{c}}^{2}+C_{1}%
\end{align*}
and Eq.(\ref{Conditional Probability Function for Class Two Q})%
\begin{align*}
P\left(  k_{\mathbf{x}_{2_{i\ast}}}|\boldsymbol{\kappa}_{2}\right)   &
=\int_{Z}p\left(  k_{\mathbf{x}_{2_{i\ast}}}|\boldsymbol{\kappa}_{2}\right)
d\boldsymbol{\kappa}_{2}\\
&  =\int_{Z}\boldsymbol{\kappa}_{2}d\boldsymbol{\kappa}_{2}=\left\Vert
\boldsymbol{\kappa}_{2}\right\Vert _{\min_{c}}^{2}+C_{2}\text{,}%
\end{align*}
and using Eq. (\ref{Balancing Bayes' Risk Quadratic})%
\begin{align*}
&  \left\Vert \boldsymbol{\kappa}_{1}\right\Vert _{\min_{c}}^{2}-\left\Vert
\boldsymbol{\kappa}_{1}\right\Vert \left\Vert \boldsymbol{\kappa}%
_{2}\right\Vert \cos\theta_{\boldsymbol{\kappa}_{1}\boldsymbol{\kappa}_{2}%
}+\delta\left(  y\right)  \sum\nolimits_{i=1}^{l_{1}}\psi_{1_{i_{\ast}}}\\
&  +\sum\nolimits_{i=1}^{l_{1}}\psi_{1_{i_{\ast}}}\left\{  \sum\nolimits_{j=1}%
^{l}k_{\mathbf{x}_{j\ast}}\left(  \boldsymbol{\kappa}_{2}-\boldsymbol{\kappa
}_{1}\right)  \right\}  +\sum\nolimits_{i=1}^{l_{1}}\xi_{i}\psi_{1_{i_{\ast}}%
}\\
&  =\left\Vert \boldsymbol{\kappa}_{2}\right\Vert _{\min_{c}}^{2}-\left\Vert
\boldsymbol{\kappa}_{2}\right\Vert \left\Vert \boldsymbol{\kappa}%
_{1}\right\Vert \cos\theta_{\boldsymbol{\kappa}_{2}\boldsymbol{\kappa}_{1}%
}-\delta\left(  y\right)  \sum\nolimits_{i=1}^{l_{2}}\psi_{2_{i_{\ast}}}\\
&  -\sum\nolimits_{i=1}^{l_{2}}\psi_{2_{i_{\ast}}}\left\{  \sum\nolimits_{j=1}%
^{l}k_{\mathbf{x}_{j\ast}}\left(  \boldsymbol{\kappa}_{1}-\boldsymbol{\kappa
}_{2}\right)  \right\}  +\sum\nolimits_{i=1}^{l_{2}}\xi_{i}\psi_{2_{i_{\ast}}}%
\end{align*}
where $P\left(  k_{\mathbf{x}_{1_{i\ast}}}|\boldsymbol{\kappa}_{1}\right)
=P\left(  k_{\mathbf{x}_{2_{i\ast}}}|\boldsymbol{\kappa}_{2}\right)  $, it
follows that the value for the integration constant $C_{1}$ in Eq.
(\ref{Conditional Probability Function for Class One Q}) is:%
\[
C_{1}=-\left\Vert \boldsymbol{\kappa}_{1}\right\Vert \left\Vert
\boldsymbol{\kappa}_{2}\right\Vert \cos\theta_{\boldsymbol{\kappa}%
_{1}\boldsymbol{\kappa}_{2}}+\sum\nolimits_{i=1}^{l_{1}}\xi_{i}\psi
_{1_{i_{\ast}}}\text{,}%
\]
and that the value for the integration constant $C_{2}$ in
Eq.(\ref{Conditional Probability Function for Class Two Q}) is:%
\[
C_{2}=-\left\Vert \boldsymbol{\kappa}_{2}\right\Vert \left\Vert
\boldsymbol{\kappa}_{1}\right\Vert \cos\theta_{\boldsymbol{\kappa}%
_{2}\boldsymbol{\kappa}_{1}}+\sum\nolimits_{i=1}^{l_{2}}\xi_{i}\psi
_{2_{i_{\ast}}}\text{.}%
\]

Substituting the values for $C_{1}$ and $C_{2}$ into Eqs
(\ref{Conditional Probability Function for Class One Q}) and
(\ref{Conditional Probability Function for Class Two Q}) produces the integral
equation%
\begin{align*}
&  f\left(  \widetilde{\Lambda}_{\boldsymbol{\kappa}}\left(  \mathbf{s}%
\right)  \right)  =\int_{Z}\boldsymbol{\kappa}_{1}d\boldsymbol{\kappa}%
_{1}+\delta\left(  y\right)  \sum\nolimits_{i=1}^{l_{1}}\psi_{1_{i_{\ast}}}\\
&  +\sum\nolimits_{i=1}^{l_{1}}\psi_{1_{i_{\ast}}}\left\{  \sum\nolimits_{j=1}%
^{l}k_{\mathbf{x}_{j\ast}}\left(  \boldsymbol{\kappa}_{2}-\boldsymbol{\kappa
}_{1}\right)  \right\} \\
&  =\int_{Z}\boldsymbol{\kappa}_{2}d\boldsymbol{\kappa}_{2}-\delta\left(
y\right)  \sum\nolimits_{i=1}^{l_{2}}\psi_{2_{i_{\ast}}}\\
&  -\sum\nolimits_{i=1}^{l_{2}}\psi_{2_{i_{\ast}}}\left\{  \sum\nolimits_{j=1}%
^{l}k_{\mathbf{x}_{j\ast}}\left(  \boldsymbol{\kappa}_{1}-\boldsymbol{\kappa
}_{2}\right)  \right\}  \text{,}%
\end{align*}
over the decision space $Z$, where the equalizer statistics%
\[
\nabla_{eq}\left(  p\left(  \widehat{\Lambda}_{\boldsymbol{\kappa}}\left(
\mathbf{s}\right)  |\omega_{1}\right)  \right)  =\delta\left(  y\right)
\sum\nolimits_{i=1}^{l_{1}}\psi_{1_{i_{\ast}}}+\sum\nolimits_{i=1}^{l_{1}}%
\psi_{1_{i_{\ast}}}\left\{  \sum\nolimits_{j=1}^{l}k_{\mathbf{x}_{j\ast}%
}\left(  \boldsymbol{\kappa}_{2}-\boldsymbol{\kappa}_{1}\right)  \right\}
\]
and%
\[
\nabla_{eq}\left(  p\left(  \widehat{\Lambda}_{\boldsymbol{\kappa}}\left(
\mathbf{s}\right)  |\omega_{2}\right)  \right)  =-\delta\left(  y\right)
\sum\nolimits_{i=1}^{l_{2}}\psi_{2_{i_{\ast}}}-\sum\nolimits_{i=1}^{l_{2}}%
\psi_{2_{i_{\ast}}}\left\{  \sum\nolimits_{j=1}^{l}k_{\mathbf{x}_{j\ast}%
}\left(  \boldsymbol{\kappa}_{1}-\boldsymbol{\kappa}_{2}\right)  \right\}
\]
ensure that the eigenenergy $\left\Vert \boldsymbol{\kappa}_{1}\right\Vert
_{\min_{c}}^{2}$ associated with the position or location of the likelihood
ratio $p\left(  \widehat{\Lambda}_{\boldsymbol{\kappa}}\left(  \mathbf{s}%
\right)  |\omega_{1}\right)  $ given class $\omega_{1}$ is symmetrically
balanced with the eigenenergy $\left\Vert \boldsymbol{\kappa}_{2}\right\Vert
_{\min_{c}}^{2}$ associated with the position or location of the likelihood
ratio $p\left(  \widehat{\Lambda}_{\boldsymbol{\kappa}}\left(  \mathbf{s}%
\right)  |\omega_{2}\right)  $ given class $\omega_{2}$.

The primal class-conditional probability density functions $p\left(
k_{\mathbf{x}_{1_{i\ast}}}|\boldsymbol{\kappa}_{1}\right)  $ and $p\left(
k_{\mathbf{x}_{2_{i\ast}}}|\boldsymbol{\kappa}_{2}\right)  $ and the Wolfe
dual class-conditional probability density functions $p\left(  \frac
{k_{\mathbf{x}_{1i\ast}}}{\left\Vert k_{\mathbf{x}_{1i\ast}}\right\Vert
}|\boldsymbol{\psi}_{1}\right)  $ and $p\left(  \frac{k_{\mathbf{x}_{2i\ast}}%
}{\left\Vert k_{\mathbf{x}_{2i\ast}}\right\Vert }|\boldsymbol{\psi}%
_{2}\right)  $ satisfy the integral equation in the following manner:%
\begin{align}
&  f\left(  \widetilde{\Lambda}_{\boldsymbol{\kappa}}\left(  \mathbf{s}%
\right)  \right)  =%
{\displaystyle\int\nolimits_{Z}}
p\left(  k_{\mathbf{x}_{1_{i\ast}}}|\boldsymbol{\kappa}_{1}\right)
d\boldsymbol{\kappa}_{1}+\delta\left(  y\right)  p\left(  \frac{k_{\mathbf{x}%
_{1i\ast}}}{\left\Vert k_{\mathbf{x}_{1i\ast}}\right\Vert }|\boldsymbol{\psi
}_{1}\right) \label{Quadratic Eigenlocus Integral Equation V}\\
&  +k_{\widehat{\mathbf{x}}_{i\ast}}\left[  p\left(  k_{\mathbf{x}_{2_{i\ast}%
}}|\boldsymbol{\kappa}_{2}\right)  -p\left(  k_{\mathbf{x}_{1_{i\ast}}%
}|\boldsymbol{\kappa}_{1}\right)  \right]  p\left(  \frac{k_{\mathbf{x}%
_{1i\ast}}}{\left\Vert k_{\mathbf{x}_{1i\ast}}\right\Vert }|\boldsymbol{\psi
}_{1}\right) \nonumber\\
&  =\int\nolimits_{Z}p\left(  k_{\mathbf{x}_{2_{i\ast}}}|\boldsymbol{\kappa
}_{2}\right)  d\boldsymbol{\kappa}_{2}-\delta\left(  y\right)  p\left(
\frac{k_{\mathbf{x}_{2i\ast}}}{\left\Vert k_{\mathbf{x}_{2i\ast}}\right\Vert
}|\boldsymbol{\psi}_{2}\right) \nonumber\\
&  +k_{\widehat{\mathbf{x}}_{i\ast}}\left[  p\left(  k_{\mathbf{x}_{2_{i\ast}%
}}|\boldsymbol{\kappa}_{2}\right)  -p\left(  k_{\mathbf{x}_{1_{i\ast}}%
}|\boldsymbol{\kappa}_{1}\right)  \right]  p\left(  \frac{k_{\mathbf{x}%
_{2i\ast}}}{\left\Vert k_{\mathbf{x}_{2i\ast}}\right\Vert }|\boldsymbol{\psi
}_{2}\right)  \text{,}\nonumber
\end{align}
over the $Z_{1}$ and $Z_{2}$ decision regions, where $k_{\widehat{\mathbf{x}%
}_{i\ast}}\triangleq\sum\nolimits_{i=1}^{l}k_{\mathbf{x}_{i\ast}}$.

The equalizer statistics:%
\begin{align}
\nabla_{eq}\left(  p\left(  \widehat{\Lambda}_{\boldsymbol{\psi}}\left(
\mathbf{s}\right)  |\omega_{1}\right)  \right)   &  =\delta\left(  y\right)
p\left(  \frac{k_{\mathbf{x}_{1i\ast}}}{\left\Vert k_{\mathbf{x}_{1i\ast}%
}\right\Vert }|\boldsymbol{\psi}_{1}\right)
\label{Equalizer Statistic Class One Q}\\
&  +k_{\widehat{\mathbf{x}}_{i\ast}}p\left(  k_{\mathbf{x}_{2_{i\ast}}%
}|\boldsymbol{\kappa}_{2}\right)  p\left(  \frac{k_{\mathbf{x}_{1i\ast}}%
}{\left\Vert k_{\mathbf{x}_{1i\ast}}\right\Vert }|\boldsymbol{\psi}_{1}\right)
\nonumber\\
&  -k_{\widehat{\mathbf{x}}_{i\ast}}p\left(  k_{\mathbf{x}_{1_{i\ast}}%
}|\boldsymbol{\kappa}_{1}\right)  p\left(  \frac{k_{\mathbf{x}_{1i\ast}}%
}{\left\Vert k_{\mathbf{x}_{1i\ast}}\right\Vert }|\boldsymbol{\psi}_{1}\right)
\nonumber
\end{align}
and%
\begin{align}
\nabla_{eq}\left(  p\left(  \widehat{\Lambda}_{\boldsymbol{\psi}}\left(
\mathbf{s}\right)  |\omega_{2}\right)  \right)   &  =-\delta\left(  y\right)
p\left(  \frac{k_{\mathbf{x}_{2i\ast}}}{\left\Vert k_{\mathbf{x}_{2i\ast}%
}\right\Vert }|\boldsymbol{\psi}_{2}\right)
\label{Equalizer Statistic Class Two Q}\\
&  +k_{\widehat{\mathbf{x}}_{i\ast}}p\left(  k_{\mathbf{x}_{2_{i\ast}}%
}|\boldsymbol{\kappa}_{2}\right)  p\left(  \frac{k_{\mathbf{x}_{2i\ast}}%
}{\left\Vert k_{\mathbf{x}_{2i\ast}}\right\Vert }|\boldsymbol{\psi}_{2}\right)
\nonumber\\
&  -k_{\widehat{\mathbf{x}}_{i\ast}}p\left(  k_{\mathbf{x}_{1_{i\ast}}%
}|\boldsymbol{\kappa}_{1}\right)  p\left(  \frac{k_{\mathbf{x}_{2i\ast}}%
}{\left\Vert k_{\mathbf{x}_{2i\ast}}\right\Vert }|\boldsymbol{\psi}%
_{2}\right)  \text{,}\nonumber
\end{align}
where $p\left(  \frac{k_{\mathbf{x}_{1i\ast}}}{\left\Vert k_{\mathbf{x}%
_{1i\ast}}\right\Vert }|\boldsymbol{\psi}_{1}\right)  $ and $p\left(
\frac{k_{\mathbf{x}_{2i\ast}}}{\left\Vert k_{\mathbf{x}_{2i\ast}}\right\Vert
}|\boldsymbol{\psi}_{2}\right)  $ determine the equilibrium point%
\[
p\left(  \widehat{\Lambda}_{\boldsymbol{\psi}}\left(  \mathbf{s}\right)
|\omega_{1}\right)  -p\left(  \widehat{\Lambda}_{\boldsymbol{\psi}}\left(
\mathbf{s}\right)  |\omega_{2}\right)  =0
\]
of the integral equation $f\left(  \widetilde{\Lambda}_{\boldsymbol{\kappa}%
}\left(  \mathbf{s}\right)  \right)  $ in Eq.
(\ref{Quadratic Eigenlocus Integral Equation V}), ensure that all of the
forces associated with the counter risks $\overline{\mathfrak{R}%
}_{\mathfrak{\min}}\left(  Z_{1}|\psi_{1i\ast}k_{\mathbf{x}_{1i\ast}}\right)
$ and the risks $\mathfrak{R}_{\mathfrak{\min}}\left(  Z_{1}|\psi_{2i\ast
}k_{\mathbf{x}_{2i\ast}}\right)  $ in the $Z_{1}$ decision region, which are
related to positions and potential locations of corresponding $k_{\mathbf{x}%
_{1i\ast}}$ and $k_{\mathbf{x}_{2i\ast}}$ extreme points in the $Z_{1}$
decision region, are \emph{symmetrically balanced with} all of the forces
associated with the counter risks $\overline{\mathfrak{R}}_{\mathfrak{\min}%
}\left(  Z_{2}|\psi_{2i\ast}k_{\mathbf{x}_{2i\ast}}\right)  $ and the risks
$\mathfrak{R}_{\mathfrak{\min}}\left(  Z_{2}|\psi_{1i\ast}k_{\mathbf{x}%
_{1i\ast}}\right)  $ in the $Z_{2}$ decision region, which are related to
positions and potential locations of corresponding $k_{\mathbf{x}_{2i\ast}}$
and $k_{\mathbf{x}_{1i\ast}}$ extreme points in the $Z_{2}$ decision region,
such that the collective forces associated with the integrals $p\left(
k_{\mathbf{x}_{1_{i\ast}}}|\boldsymbol{\kappa}_{1}\right)  d\boldsymbol{\kappa
}_{1}$ and $\int_{Z}p\left(  k_{\mathbf{x}_{2_{i\ast}}}|\boldsymbol{\kappa
}_{2}\right)  d\boldsymbol{\kappa}_{2}$ are \emph{symmetrically balanced with}
each other.

The above integral equation can be written as:%
\begin{align*}
f\left(  \widetilde{\Lambda}_{\boldsymbol{\kappa}}\left(  \mathbf{s}\right)
\right)  =  &  \int_{Z_{1}}p\left(  k_{\mathbf{x}_{1_{i\ast}}}%
|\boldsymbol{\kappa}_{1}\right)  d\boldsymbol{\kappa}_{1}+\int_{Z_{2}}p\left(
k_{\mathbf{x}_{1_{i\ast}}}|\boldsymbol{\kappa}_{1}\right)  d\boldsymbol{\kappa
}_{1}+\nabla_{eq}\left(  \widehat{\Lambda}_{\boldsymbol{\psi}_{1}}\left(
\mathbf{s}\right)  \right) \\
&  =\int_{Z_{1}}p\left(  k_{\mathbf{x}_{2_{i\ast}}}|\boldsymbol{\kappa}%
_{2}\right)  d\boldsymbol{\kappa}_{2}+\int_{Z_{2}}p\left(  k_{\mathbf{x}%
_{2_{i\ast}}}|\boldsymbol{\kappa}_{2}\right)  d\boldsymbol{\kappa}_{2}%
+\nabla_{eq}\left(  \widehat{\Lambda}_{\boldsymbol{\psi}_{2}}\left(
\mathbf{s}\right)  \right)  \text{,}%
\end{align*}
where the integral $\int_{Z_{1}}p\left(  k_{\mathbf{x}_{1_{i\ast}}%
}|\boldsymbol{\kappa}_{1}\right)  d\boldsymbol{\kappa}_{1}$ accounts for all
of the forces associated with the counter risks $\overline{\mathfrak{R}%
}_{\mathfrak{\min}}\left(  Z_{1}|\psi_{1i\ast}k_{\mathbf{x}_{1_{i\ast}}%
}\right)  $ which are related to positions and potential locations of
corresponding $k_{\mathbf{x}_{1_{i\ast}}}$ extreme points that lie in the
$Z_{1}$ decision region, the integral $\int_{Z_{2}}p\left(  k_{\mathbf{x}%
_{1_{i\ast}}}|\boldsymbol{\kappa}_{1}\right)  d\boldsymbol{\kappa}_{1}$
accounts for all of the forces associated with the risks $\mathfrak{R}%
_{\mathfrak{\min}}\left(  Z_{2}|\psi_{1i\ast}k_{\mathbf{x}_{1_{i\ast}}%
}\right)  $ which are related to positions and potential locations of
corresponding $k_{\mathbf{x}_{1_{i\ast}}}$ extreme points that lie in the
$Z_{2}$ decision region, the integral $\int_{Z_{2}}p\left(  k_{\mathbf{x}%
_{2_{i\ast}}}|\boldsymbol{\kappa}_{2}\right)  d\boldsymbol{\kappa}_{2}$
accounts for all of the forces associated with the counter risks
$\overline{\mathfrak{R}}_{\mathfrak{\min}}\left(  Z_{2}|\psi_{2i\ast
}k_{\mathbf{x}_{2_{i\ast}}}\right)  $ which are related to positions and
potential locations of corresponding $k_{\mathbf{x}_{2_{i\ast}}}$ extreme
points that lie in the $Z_{2}$ decision region, and the integral $\int_{Z_{1}%
}p\left(  k_{\mathbf{x}_{2_{i\ast}}}|\boldsymbol{\kappa}_{2}\right)
d\boldsymbol{\kappa}_{2}$ accounts for all of the forces associated with the
risks $\mathfrak{R}_{\mathfrak{\min}}\left(  Z_{1}|\psi_{2i\ast}%
k_{\mathbf{x}_{2_{i\ast}}}\right)  $ which are related to positions and
potential locations of corresponding $k_{\mathbf{x}_{2_{i\ast}}}$ extreme
points that lie in the $Z_{1}$ decision region.

It follows that the classification system $\left(  \mathbf{x}^{T}%
\mathbf{s}+1\right)  ^{2}\boldsymbol{\kappa}+\kappa_{0}$ $\overset{\omega
_{1}}{\underset{\omega_{2}}{\gtrless}}0$ seeks a point of statistical
equilibrium $p\left(  \widehat{\Lambda}_{\boldsymbol{\psi}}\left(
\mathbf{s}\right)  |\omega_{1}\right)  -p\left(  \widehat{\Lambda
}_{\boldsymbol{\psi}}\left(  \mathbf{s}\right)  |\omega_{2}\right)  =0$ where
the opposing forces and influences of the classification system are balanced
with each other, such that the eigenenergy and the expected risk of the
classification system are minimized, and the classification system is in
statistical equilibrium.

Therefore, it is concluded that the quadratic eigenlocus discriminant function%
\[
\widetilde{\Lambda}_{\boldsymbol{\kappa}}\left(  \mathbf{s}\right)  =\left(
k_{\mathbf{x}}-k_{\widehat{\mathbf{x}}_{i\ast}}\right)  \left(
\boldsymbol{\kappa}_{1}-\boldsymbol{\kappa}_{2}\right)  \mathbf{+}%
\sum\nolimits_{i=1}^{l}y_{i}\left(  1-\xi_{i}\right)
\]
is the solution to the integral equation:%
\begin{align*}
&  f\left(  \widetilde{\Lambda}_{\boldsymbol{\kappa}}\left(  \mathbf{s}%
\right)  \right)  =\int_{Z_{1}}\boldsymbol{\kappa}_{1}d\boldsymbol{\kappa}%
_{1}+\int_{Z_{2}}\boldsymbol{\kappa}_{1}d\boldsymbol{\kappa}_{1}+\delta\left(
y\right)  \sum\nolimits_{i=1}^{l_{1}}\psi_{1_{i_{\ast}}}\\
&  +\sum\nolimits_{i=1}^{l_{1}}\psi_{1_{i_{\ast}}}\left\{  \sum\nolimits_{j=1}%
^{l}k_{\mathbf{x}_{j\ast}}\left(  \boldsymbol{\kappa}_{2}-\boldsymbol{\kappa
}_{1}\right)  \right\} \\
&  =\int_{Z_{1}}\boldsymbol{\kappa}_{2}d\boldsymbol{\kappa}_{2}+\int_{Z_{2}%
}\boldsymbol{\kappa}_{2}d\boldsymbol{\kappa}_{2}-\delta\left(  y\right)
\sum\nolimits_{i=1}^{l_{2}}\psi_{2_{i_{\ast}}}\\
&  -\sum\nolimits_{i=1}^{l_{2}}\psi_{2_{i_{\ast}}}\left\{  \sum\nolimits_{j=1}%
^{l}k_{\mathbf{x}_{j\ast}}\left(  \boldsymbol{\kappa}_{1}-\boldsymbol{\kappa
}_{2}\right)  \right\}  \text{,}%
\end{align*}
over the decision space $Z=Z_{1}+Z_{2}$, where $Z_{1}$ and $Z_{2}$ are
symmetrical decision regions $Z_{1}\simeq Z_{2}$ that have respective counter
risks:%
\[
\overline{\mathfrak{R}}_{\mathfrak{\min}}\left(  Z_{1}|\boldsymbol{\kappa}%
_{1}\right)  \text{ \ and \ }\overline{\mathfrak{R}}_{\mathfrak{\min}}\left(
Z_{2}|\boldsymbol{\kappa}_{2}\right)
\]
and respective risks:%
\[
\mathfrak{R}_{\mathfrak{\min}}\left(  Z_{1}|\boldsymbol{\kappa}_{2}\right)
\text{ \ and \ }\mathfrak{R}_{\mathfrak{\min}}\left(  Z_{2}|\boldsymbol{\kappa
}_{1}\right)  \text{,}%
\]
where all of the forces associated with the risk $\mathfrak{R}_{\mathfrak{\min
}}\left(  Z_{1}|\boldsymbol{\kappa}_{2}\right)  $ for class $\omega_{2}$ in
the $Z_{1}$ decision region and the counter risk $\overline{\mathfrak{R}%
}_{\mathfrak{\min}}\left(  Z_{2}|\boldsymbol{\kappa}_{2}\right)  $ for class
$\omega_{2}$ in the $Z_{2}$ decision region are balanced with all of the
forces associated with the counter risk $\overline{\mathfrak{R}}%
_{\mathfrak{\min}}\left(  Z_{1}|\boldsymbol{\kappa}_{1}\right)  $ for class
$\omega_{1}$ in the $Z_{1}$ decision region and the risk $\mathfrak{R}%
_{\mathfrak{\min}}\left(  Z_{2}|\boldsymbol{\kappa}_{1}\right)  $ for class
$\omega_{1}$ in the $Z_{2}$ decision region:%
\begin{align*}
f\left(  \widehat{\Lambda}\left(  \mathbf{x}\right)  \right)   &
:\mathfrak{R}_{\mathfrak{\min}}\left(  Z_{1}|\boldsymbol{\kappa}_{2}\right)
+\overline{\mathfrak{R}}_{\mathfrak{\min}}\left(  Z_{2}|\boldsymbol{\kappa
}_{2}\right) \\
&  =\overline{\mathfrak{R}}_{\mathfrak{\min}}\left(  Z_{1}|\boldsymbol{\kappa
}_{1}\right)  +\mathfrak{R}_{\mathfrak{\min}}\left(  Z_{2}|\boldsymbol{\kappa
}_{1}\right)
\end{align*}
such that the expected risk $\mathfrak{R}_{\mathfrak{\min}}\left(
Z|\widehat{\Lambda}_{\boldsymbol{\kappa}}\left(  \mathbf{s}\right)  \right)  $
and the eigenenergy $E_{\min}\left(  Z|\widehat{\Lambda}_{\boldsymbol{\kappa}%
}\left(  \mathbf{s}\right)  \right)  $ of the classification system $\left(
\mathbf{x}^{T}\mathbf{s}+1\right)  ^{2}\boldsymbol{\kappa}+\kappa
_{0}\overset{\omega_{1}}{\underset{\omega_{2}}{\gtrless}}0$ are governed by
the equilibrium point $p\left(  \widehat{\Lambda}_{\boldsymbol{\psi}}\left(
\mathbf{s}\right)  |\omega_{1}\right)  -p\left(  \widehat{\Lambda
}_{\boldsymbol{\psi}}\left(  \mathbf{s}\right)  |\omega_{2}\right)  =0$:%
\[
\sum\nolimits_{i=1}^{l_{1}}\psi_{1i\ast}\frac{k_{\mathbf{x}_{1i\ast}}%
}{\left\Vert k_{\mathbf{x}_{1i\ast}}\right\Vert }-\sum\nolimits_{i=1}^{l_{2}%
}\psi_{2i\ast}\frac{k_{\mathbf{x}_{2i\ast}}}{\left\Vert k_{\mathbf{x}_{2i\ast
}}\right\Vert }=0
\]
of the integral equation $f\left(  \widetilde{\Lambda}_{\boldsymbol{\kappa}%
}\left(  \mathbf{s}\right)  \right)  $.

It follows that the classification system%
\[
\left(  k_{\mathbf{x}}-k_{\widehat{\mathbf{x}}_{i\ast}}\right)  \left(
\boldsymbol{\kappa}_{1}-\boldsymbol{\kappa}_{2}\right)  \mathbf{+}%
\sum\nolimits_{i=1}^{l}y_{i}\left(  1-\xi_{i}\right)  \overset{\omega
_{1}}{\underset{\omega_{2}}{\gtrless}}0
\]
is in statistical equilibrium:%
\begin{align*}
f\left(  \widetilde{\Lambda}_{\boldsymbol{\kappa}}\left(  \mathbf{s}\right)
\right)   &  :\int_{Z_{1}}p\left(  k_{\mathbf{x}_{1_{i\ast}}}%
|\boldsymbol{\kappa}_{1}\right)  d\boldsymbol{\kappa}_{1}-\int_{Z_{1}}p\left(
k_{\mathbf{x}_{2_{i\ast}}}|\boldsymbol{\kappa}_{2}\right)  d\boldsymbol{\kappa
}_{2}+\nabla_{eq}\left(  \widehat{\Lambda}_{\boldsymbol{\psi}_{1}}\left(
\mathbf{s}\right)  \right) \\
&  =\int_{Z_{2}}p\left(  k_{\mathbf{x}_{2_{i\ast}}}|\boldsymbol{\kappa}%
_{2}\right)  d\boldsymbol{\kappa}_{2}-\int_{Z_{2}}p\left(  k_{\mathbf{x}%
_{1_{i\ast}}}|\boldsymbol{\kappa}_{1}\right)  d\boldsymbol{\kappa}_{1}%
+\nabla_{eq}\left(  \widehat{\Lambda}_{\boldsymbol{\psi}_{2}}\left(
\mathbf{s}\right)  \right)  \text{,}%
\end{align*}
where all of the forces associated with the counter risk $\overline
{\mathfrak{R}}_{\mathfrak{\min}}\left(  Z_{1}|\boldsymbol{\kappa}_{1}\right)
$ and the risk $\mathfrak{R}_{\mathfrak{\min}}\left(  Z_{1}|\boldsymbol{\kappa
}_{2}\right)  $ in the $Z_{1}$ decision region are balanced with all of the
forces associated with the counter risk $\overline{\mathfrak{R}}%
_{\mathfrak{\min}}\left(  Z_{2}|\boldsymbol{\kappa}_{2}\right)  $ and the risk
$\mathfrak{R}_{\mathfrak{\min}}\left(  Z_{2}|\boldsymbol{\kappa}_{1}\right)  $
in the $Z_{2}$ decision region:%
\begin{align*}
f\left(  \widetilde{\Lambda}_{\boldsymbol{\kappa}}\left(  \mathbf{s}\right)
\right)   &  :\overline{\mathfrak{R}}_{\mathfrak{\min}}\left(  Z_{1}%
|\boldsymbol{\kappa}_{1}\right)  -\mathfrak{R}_{\mathfrak{\min}}\left(
Z_{1}|\boldsymbol{\kappa}_{2}\right) \\
&  =\overline{\mathfrak{R}}_{\mathfrak{\min}}\left(  Z_{2}|\boldsymbol{\kappa
}_{2}\right)  -\mathfrak{R}_{\mathfrak{\min}}\left(  Z_{2}|\boldsymbol{\kappa
}_{1}\right)  \text{,}%
\end{align*}
and the eigenenergies associated with the counter risk $\overline
{\mathfrak{R}}_{\mathfrak{\min}}\left(  Z_{1}|\boldsymbol{\kappa}_{1}\right)
$ and the risk $\mathfrak{R}_{\mathfrak{\min}}\left(  Z_{1}|\boldsymbol{\kappa
}_{2}\right)  $ in the $Z_{1}$ decision region are balanced with the
eigenenergies associated with the counter risk $\overline{\mathfrak{R}%
}_{\mathfrak{\min}}\left(  Z_{2}|\boldsymbol{\kappa}_{2}\right)  $ and the
risk $\mathfrak{R}_{\mathfrak{\min}}\left(  Z_{2}|\boldsymbol{\kappa}%
_{1}\right)  $ in the $Z_{2}$ decision region:%
\begin{align*}
f\left(  \widetilde{\Lambda}_{\boldsymbol{\kappa}}\left(  \mathbf{s}\right)
\right)   &  :E_{\min_{c}}\left(  Z_{1}|\boldsymbol{\kappa}_{1}\right)
-E_{\min_{c}}\left(  Z_{1}|\boldsymbol{\kappa}_{2}\right) \\
&  =E_{\min_{c}}\left(  Z_{2}|\boldsymbol{\kappa}_{2}\right)  -E_{\min_{c}%
}\left(  Z_{2}|\boldsymbol{\kappa}_{1}\right)  \text{.}%
\end{align*}

Therefore, it is concluded that the expected risk $\mathfrak{R}%
_{\mathfrak{\min}}\left(  Z|\widehat{\Lambda}_{\boldsymbol{\kappa}}\left(
\mathbf{s}\right)  \right)  $ and the eigenenergy $E_{\min}\left(
Z|\widehat{\Lambda}_{\boldsymbol{\kappa}}\left(  \mathbf{s}\right)  \right)  $
of the classification system $\left(  \mathbf{x}^{T}\mathbf{s}+1\right)
^{2}\boldsymbol{\kappa}+\kappa_{0}\overset{\omega_{1}}{\underset{\omega
_{2}}{\gtrless}}0$ are governed by the equilibrium point $\widehat{\Lambda
}_{\boldsymbol{\psi}}\left(  \mathbf{s}\right)  $:%
\[
p\left(  \widehat{\Lambda}_{\boldsymbol{\psi}}\left(  \mathbf{s}\right)
|\omega_{1}\right)  -p\left(  \widehat{\Lambda}_{\boldsymbol{\psi}}\left(
\mathbf{s}\right)  |\omega_{2}\right)  =0
\]
of the integral equation%
\begin{align*}
f\left(  \widetilde{\Lambda}_{\boldsymbol{\kappa}}\left(  \mathbf{s}\right)
\right)  =  &  \int_{Z_{1}}p\left(  k_{\mathbf{x}_{1_{i\ast}}}%
|\boldsymbol{\kappa}_{1}\right)  d\boldsymbol{\kappa}_{1}+\int_{Z_{2}}p\left(
k_{\mathbf{x}_{1_{i\ast}}}|\boldsymbol{\kappa}_{1}\right)  d\boldsymbol{\kappa
}_{1}\\
&  +\nabla_{eq}\left(  p\left(  \widehat{\Lambda}_{\boldsymbol{\psi}}\left(
\mathbf{s}\right)  |\omega_{1}\right)  \right) \\
&  =\int_{Z_{2}}p\left(  k_{\mathbf{x}_{2_{i\ast}}}|\boldsymbol{\kappa}%
_{2}\right)  d\boldsymbol{\kappa}_{2}+\int_{Z_{1}}p\left(  k_{\mathbf{x}%
_{2_{i\ast}}}|\boldsymbol{\kappa}_{2}\right)  d\boldsymbol{\kappa}_{2}\\
&  +\nabla_{eq}\left(  p\left(  \widehat{\Lambda}_{\boldsymbol{\psi}}\left(
\mathbf{s}\right)  |\omega_{2}\right)  \right)  \text{,}%
\end{align*}
where the opposing forces and influences of the classification system are
balanced with each other, such that the eigenenergy and the expected risk of
the classification system are minimized, and the classification system is in
statistical equilibrium.

\begin{figure}[ptb]%
\centering
\fbox{\includegraphics[
height=2.5875in,
width=3.6063in
]%
{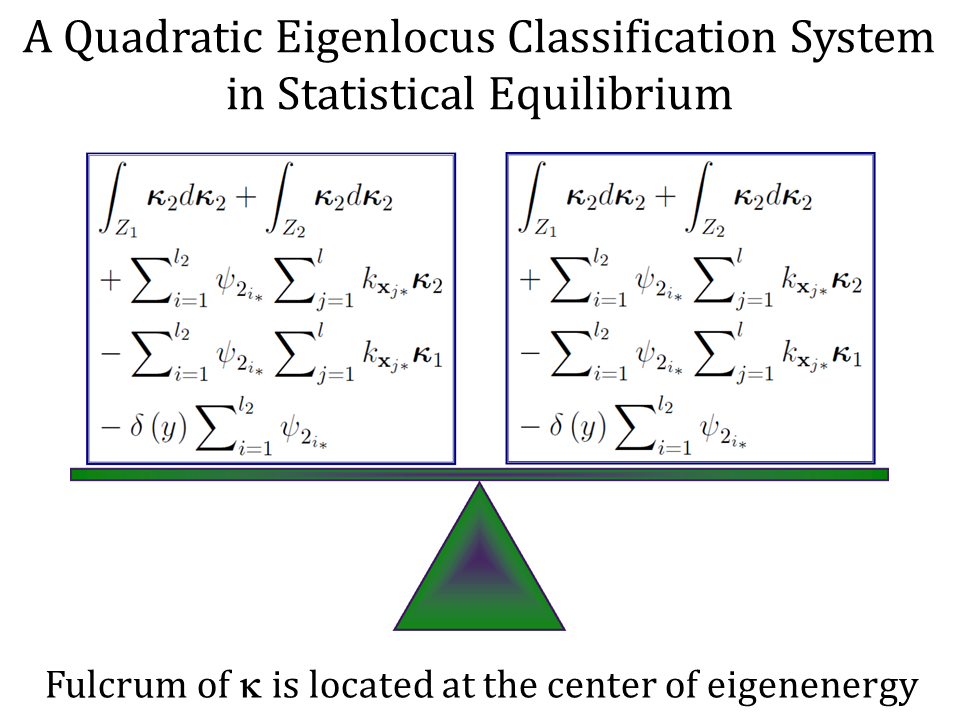}%
}\caption{Quadratic eigenlocus transforms generate quadratic classification
systems $\left(  \mathbf{x}^{T}\mathbf{s}+1\right)  ^{2}\boldsymbol{\kappa
}+\kappa_{0}\protect\overset{\omega_{1}}{\protect\underset{\omega
_{2}}{\gtrless}}0$ that satisfy a fundamental integral equation of binary
classification for a classification system in statistical equilibrium.}%
\label{Symmetrical Balance of Bayes' Error Quadratic}%
\end{figure}

Figure $\ref{Symmetrical Balance of Bayes' Error Quadratic}$ illustrates the
manner in which quadratic eigenlocus discriminant functions
$\widetilde{\Lambda}_{\boldsymbol{\kappa}}\left(  \mathbf{s}\right)  =\left(
\mathbf{x}^{T}\mathbf{s}+1\right)  ^{2}\boldsymbol{\kappa}+\kappa_{0}$
minimize the total allowed eigenenergy $\left\Vert \boldsymbol{\kappa
}\right\Vert _{\min_{c}}^{2}$ and the expected risk $\mathfrak{R}%
_{\mathfrak{\min}}\left(  Z|\mathbf{\kappa}\right)  $ of quadratic
classification systems.%

\begin{figure}[ptb]%
\centering
\fbox{\includegraphics[
height=2.5875in,
width=3.4411in
]%
{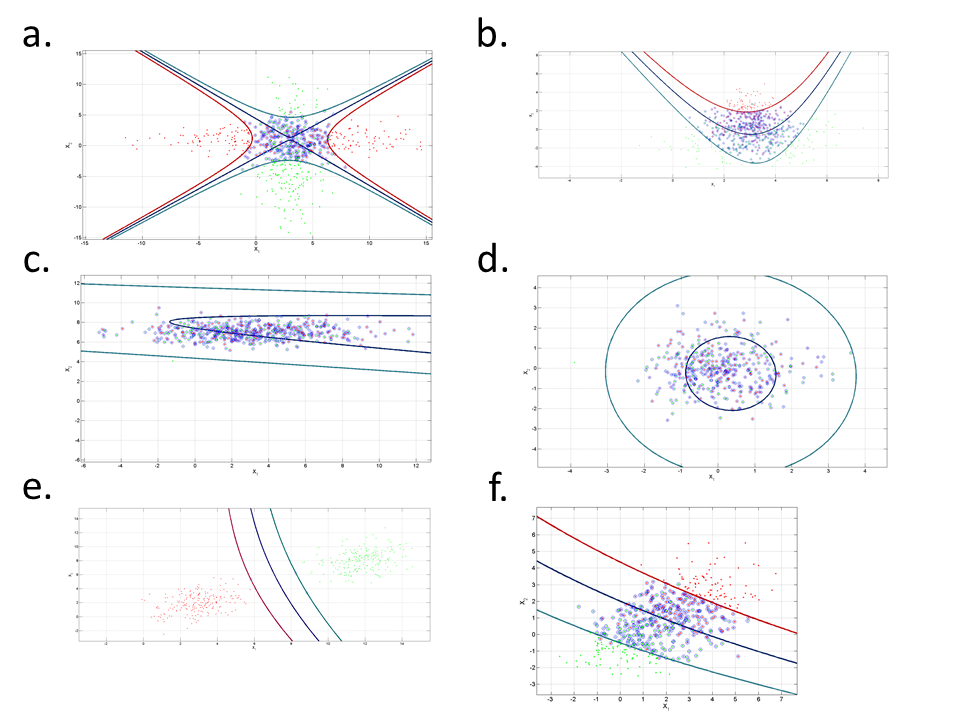}%
}\caption{Quadratic eigenlocus classification systems $\left(  \mathbf{x}%
^{T}\mathbf{s}+1\right)  ^{2}\boldsymbol{\kappa}+\kappa_{0}%
\protect\overset{\omega_{1}}{\protect\underset{\omega_{2}}{\gtrless}}0$
generate optimal quadratic decision boundaries for $\left(  1\right)  $
overlapping data distributions that have dissimilar covariance matrices: see
$\left(  a\right)  $ and $\left(  b\right)  $, $\left(  2\right)  $ completely
overlapping data distributions: see $\left(  c\right)  $ and $\left(
d\right)  $, and $\left(  3\right)  $ linear decision boundaries for data
distributions that have similar covariance matrices: see $\left(  e\right)  $
and $\left(  f\right)  $.}%
\label{Bayes' Decision Boundaries Quadratic}%
\end{figure}

By way of illustration, Fig. $\ref{Bayes' Decision Boundaries Quadratic}$
shows that quadratic eigenlocus transforms generate decision regions $Z_{1}$
and $Z_{2}$ that minimize the expected risk $\mathfrak{R}_{\mathfrak{\min}%
}\left(  Z|\boldsymbol{\kappa}\right)  $ for overlapping data distributions
that have dissimilar covariance matrices, completely overlapping data
distributions, and non-overlapping and overlapping data distributions that
have similar covariance matrices.

Accordingly, quadratic eigenlocus classification systems $\left(
\mathbf{x}^{T}\mathbf{s}+1\right)  ^{2}\boldsymbol{\kappa}+\kappa
_{0}\overset{\omega_{1}}{\underset{\omega_{2}}{\gtrless}}0$ generate optimal
quadratic decision boundaries for overlapping data distributions that have
dissimilar covariance matrices (see Fig.
$\ref{Bayes' Decision Boundaries Quadratic}$a and Fig.
$\ref{Bayes' Decision Boundaries Quadratic}$b) and completely overlapping data
distributions (see Fig. $\ref{Bayes' Decision Boundaries Quadratic}$c and Fig.
$\ref{Bayes' Decision Boundaries Quadratic}$d), where unconstrained, primal
principal eigenaxis components (extreme points) are enclosed in blue circles.
Moreover, quadratic eigenlocus discriminant functions provide estimates of
linear decision boundaries for non-overlapping data distributions that have
similar covariance matrices (see Fig.
$\ref{Bayes' Decision Boundaries Quadratic}$e) and overlapping data
distributions that have similar covariance matrices (see Fig.
$\ref{Bayes' Decision Boundaries Quadratic}$f).%

I\ will now show that quadratic eigenlocus transforms generate linear decision
boundaries that are approximated by second-order curves.

\subsection{Approximations of Linear Decision Boundaries}

Consider the decision rule in Eq. (\ref{General Gaussian Equalizer Rule}) for
two classes of Gaussian data that have similar covariance matrices
($\mathbf{\Sigma}_{1}=$ $\mathbf{\Sigma}_{2}=\mathbf{\Sigma}$), where the
likelihood ratio test $\widehat{\Lambda}\left(  \mathbf{x}\right)
\overset{H_{1}}{\underset{H_{2}}{\gtrless}}0$ generates linear decision boundaries.

So, take any given quadratic eigenlocus discriminant function
$\widetilde{\Lambda}_{\boldsymbol{\kappa}}\left(  \mathbf{s}\right)  =\left(
\mathbf{x}^{T}\mathbf{s}+1\right)  ^{2}\boldsymbol{\kappa}+\kappa$ that is
determined by a quadratic eigenlocus transform for any two sets of Gaussian
data that have similar covariance matrices: $\mathbf{\Sigma}_{1}=$
$\mathbf{\Sigma}_{2}=\mathbf{\Sigma}$.

Recall that, for any given linear or quadratic eigenlocus transform, the inner
product statistics contained within a Gram or kernel matrix determine the
shapes of three, symmetrical quadratic partitioning surfaces. Recall also that
topological properties exhibited by loci of vectors are unchanged if directed
line segments of vectors are replaced by sinuous curves. Given that
second-order, polynomial reproducing kernels approximate vectors with
continuous, second-order curves, it follows that the inner product elements of
the kernel matrix determine eigenvalues which determine three, symmetrical
hyperplane partitioning surfaces: for the two given sets of Gaussian data that
have similar covariance matrices: $\mathbf{\Sigma}_{1}=$ $\mathbf{\Sigma}%
_{2}=\mathbf{\Sigma}$.

Therefore, the class-conditional probability densities $p\left(
k_{\mathbf{x}_{1_{i\ast}}}|\boldsymbol{\kappa}_{1}\right)  $ and $p\left(
k_{\mathbf{x}_{2_{i\ast}}}|\boldsymbol{\kappa}_{2}\right)  $ specified by the
parameter vector of likelihoods $\boldsymbol{\kappa}=\boldsymbol{\kappa}%
_{1}-\boldsymbol{\kappa}_{2}$%
\begin{align*}
\widehat{\Lambda}_{\boldsymbol{\kappa}}\left(  \mathbf{s}\right)   &
=\sum\nolimits_{i=1}^{l_{1}}p\left(  k_{\mathbf{x}_{1i\ast}}%
|\operatorname{comp}_{\overrightarrow{k_{\mathbf{x}_{1i\ast}}}}\left(
\overrightarrow{\boldsymbol{\kappa}}\right)  \right)  k_{\mathbf{x}_{1_{i\ast
}}}\\
&  -\sum\nolimits_{i=1}^{l_{2}}p\left(  k_{\mathbf{x}_{2i\ast}}%
|\operatorname{comp}_{\overrightarrow{k_{\mathbf{x}_{2i\ast}}}}\left(
\overrightarrow{\boldsymbol{\kappa}}\right)  \right)  k_{\mathbf{x}_{2_{i\ast
}}}%
\end{align*}
describe similar covariance matrices $\mathbf{\Sigma}_{1}$ and $\mathbf{\Sigma
}_{2}$ for class $\omega_{1}$ and class $\omega_{2}$%
\[
\mathbf{\Sigma}_{1}\approx\mathbf{\Sigma}_{2}\text{,}%
\]
where each pointwise conditional density%
\[
p\left(  k_{\mathbf{x}_{1i\ast}}|\operatorname{comp}%
_{\overrightarrow{k_{\mathbf{x}_{1i\ast}}}}\left(
\overrightarrow{\boldsymbol{\kappa}}\right)  \right)  k_{\mathbf{x}_{1_{i\ast
}}}\text{ or }p\left(  k_{\mathbf{x}_{2i\ast}}|\operatorname{comp}%
_{\overrightarrow{k_{\mathbf{x}_{2i\ast}}}}\left(
\overrightarrow{\boldsymbol{\kappa}}\right)  \right)  k_{\mathbf{x}_{2_{i\ast
}}}%
\]
describes a distribution of first and second degree coordinates for an extreme
point $\mathbf{x}_{1_{i_{\ast}}}$ or $\mathbf{x}_{2_{i_{\ast}}}$. It follows
that quadratic eigenlocus transforms generate linear decision boundaries that
are approximated by second-order curves.

Therefore, it is concluded that linear decision boundaries are generated by
quadratic eigenlocus classification systems $\left(  \mathbf{x}^{T}%
\mathbf{s}+1\right)  ^{2}\boldsymbol{\kappa}+\kappa_{0}\overset{\omega
_{1}}{\underset{\omega_{2}}{\gtrless}}0$. It is also concluded that linear
decision boundaries are generated by quadratic eigenlocus classification
systems $\widetilde{\Lambda}_{\boldsymbol{\kappa}}\left(  \mathbf{s}\right)
\overset{H_{1}}{\underset{H_{2}}{\gtrless}}0$ based on Gaussian reproducing
kernels $\exp\left(  -\gamma\left\Vert \mathbf{x}-\mathbf{s}\right\Vert
^{2}\right)  $, where the hyperparameter $\gamma=1/100$ and the quadratic
eigenlocus decision rule is:%
\begin{equation}
\widetilde{\Lambda}_{\boldsymbol{\kappa}}\left(  \mathbf{s}\right)
=\exp\left(  -0.01\left\Vert \mathbf{x}-\mathbf{s}\right\Vert ^{2}\right)
\boldsymbol{\kappa}+\kappa_{0}\text{.}
\label{Quadratic Equalizer Rule Based on Gaussian RK}%
\end{equation}

I\ am now in a position to formally state a \emph{discrete, quadratic
classification theorem}.

\section*{Discrete Quadratic Classification Theorem}

Take a collection of $d$-component random vectors $\mathbf{x}$ that are
generated according to probability density functions $p\left(  \mathbf{x}%
|\omega_{1}\right)  $ and $p\left(  \mathbf{x}|\omega_{2}\right)  $ related to
statistical distributions of random vectors $\mathbf{x}$ that have constant or
unchanging statistics, and let $\widetilde{\Lambda}_{\boldsymbol{\kappa}%
}\left(  \mathbf{s}\right)  =\boldsymbol{\kappa}^{T}k_{\mathbf{s}}+\kappa
_{0}\overset{\omega_{1}}{\underset{\omega_{2}}{\gtrless}}0$ denote the
likelihood ratio test for a discrete, quadratic classification system, where
$\omega_{1}$ or $\omega_{2}$ is the true data category, $\boldsymbol{\kappa}$
is a locus of principal eigenaxis components and likelihoods:%
\begin{align*}
\boldsymbol{\kappa}  &  \triangleq\widehat{\Lambda}_{\boldsymbol{\kappa}%
}\left(  \mathbf{s}\right)  =p\left(  \widehat{\Lambda}_{\boldsymbol{\kappa}%
}\left(  \mathbf{s}\right)  |\omega_{1}\right)  -p\left(  \widehat{\Lambda
}_{\boldsymbol{\kappa}}\left(  \mathbf{s}\right)  |\omega_{2}\right) \\
&  =\boldsymbol{\kappa}_{1}-\boldsymbol{\kappa}_{2}\\
&  =\sum\nolimits_{i=1}^{l_{1}}\psi_{1_{i_{\ast}}}k_{\mathbf{x}_{1_{i\ast}}%
}-\sum\nolimits_{i=1}^{l_{2}}\psi_{2_{i_{\ast}}}k_{\mathbf{x}_{2_{i\ast}}%
}\text{,}%
\end{align*}
where $k_{\mathbf{x}_{1_{i\ast}}}$ and $k_{\mathbf{x}_{2_{i\ast}}}$ are
reproducing kernels for respective data points $\mathbf{x}_{1_{i_{\ast}}}$ and
$\mathbf{x}_{2_{i_{\ast}}}$: the reproducing kernel $K(\mathbf{x,s}%
)=k_{\mathbf{s}}(\mathbf{x})$ is either $k_{\mathbf{s}}(\mathbf{x}%
)\triangleq\left(  \mathbf{x}^{T}\mathbf{s}+1\right)  ^{2}$ or $k_{\mathbf{s}%
}(\mathbf{x})\triangleq\exp\left(  -0.01\left\Vert \mathbf{x}-\mathbf{s}%
\right\Vert ^{2}\right)  $, $\mathbf{x}_{1_{i_{\ast}}}\sim p\left(
\mathbf{x}|\omega_{1}\right)  $, $\mathbf{x}_{2_{i_{\ast}}}\sim p\left(
\mathbf{x}|\omega_{2}\right)  $, $\psi_{1_{i_{\ast}}}$ and $\psi_{2_{i_{\ast}%
}}$ are scale factors that provide unit measures of likelihood for respective
data points $\mathbf{x}_{1_{i_{\ast}}}$ and $\mathbf{x}_{2_{i_{\ast}}}$ which
lie in either overlapping regions or tails regions of data distributions
related to $p\left(  \mathbf{x}|\omega_{1}\right)  $ and $p\left(
\mathbf{x}|\omega_{2}\right)  $, and $\kappa_{0}$ is a functional of
$\boldsymbol{\kappa}$:%
\[
\kappa_{0}=\sum\nolimits_{i=1}^{l}y_{i}\left(  1-\xi_{i}\right)
-\sum\nolimits_{i=1}^{l}k_{\mathbf{x}_{i\ast}}\boldsymbol{\kappa}\text{,}%
\]
where $\sum\nolimits_{i=1}^{l}k_{\mathbf{x}_{i\ast}}=\sum\nolimits_{i=1}%
^{l_{1}}k_{\mathbf{x}_{1_{i\ast}}}+\sum\nolimits_{i=1}^{l_{2}}k_{\mathbf{x}%
_{2_{i\ast}}}$ is a cluster of reproducing kernels of the data points
$\mathbf{x}_{1_{i_{\ast}}}$ and $\mathbf{x}_{2_{i_{\ast}}}$ used to form
$\boldsymbol{\kappa}$, $y_{i}$ are class membership statistics: if
$\mathbf{x}_{i\ast}\in\omega_{1}$, assign $y_{i}=1$; if $\mathbf{x}_{i\ast}%
\in\omega_{2}$, assign $y_{i}=-1$, and $\xi_{i}$ are regularization
parameters: $\xi_{i}=\xi=0$ for full rank kernel matrices or $\xi_{i}=\xi\ll1$
for low rank kernel matrices.

The quadratic discriminant function%
\[
\widetilde{\Lambda}_{\boldsymbol{\kappa}}\left(  \mathbf{s}\right)
=\boldsymbol{\kappa}^{T}k_{\mathbf{s}}+\kappa_{0}%
\]
is the solution to the integral equation%
\begin{align*}
f\left(  \widetilde{\Lambda}_{\boldsymbol{\kappa}}\left(  \mathbf{s}\right)
\right)  =  &  \int_{Z_{1}}\boldsymbol{\kappa}_{1}d\boldsymbol{\kappa}%
_{1}+\int_{Z_{2}}\boldsymbol{\kappa}_{1}d\boldsymbol{\kappa}_{1}+\delta\left(
y\right)  \sum\nolimits_{i=1}^{l_{1}}\psi_{1_{i_{\ast}}}\\
&  =\int_{Z_{1}}\boldsymbol{\kappa}_{2}d\boldsymbol{\kappa}_{2}+\int_{Z_{2}%
}\boldsymbol{\kappa}_{2}d\boldsymbol{\kappa}_{2}-\delta\left(  y\right)
\sum\nolimits_{i=1}^{l_{2}}\psi_{2_{i_{\ast}}}\text{,}%
\end{align*}
over the decision space $Z=Z_{1}+Z_{2}$, where $Z_{1}$ and $Z_{2}$ are
symmetrical decision regions: $Z_{1}\simeq Z_{2}$ and $\delta\left(  y\right)
\triangleq\sum\nolimits_{i=1}^{l}y_{i}\left(  1-\xi_{i}\right)  $, such that
the expected risk $\mathfrak{R}_{\mathfrak{\min}}\left(  Z|\widehat{\Lambda
}_{\boldsymbol{\kappa}}\left(  \mathbf{s}\right)  \right)  $ and the
corresponding eigenenergy $E_{\min}\left(  Z|\widehat{\Lambda}%
_{\boldsymbol{\kappa}}\left(  \mathbf{s}\right)  \right)  $ of the
classification system $\boldsymbol{\kappa}^{T}k_{\mathbf{s}}+\kappa
_{0}\overset{\omega_{1}}{\underset{\omega_{2}}{\gtrless}}0$ are governed by
the equilibrium point%
\[
\sum\nolimits_{i=1}^{l_{1}}\psi_{1i\ast}-\sum\nolimits_{i=1}^{l_{2}}%
\psi_{2i\ast}=0
\]
of the integral equation $f\left(  \widetilde{\Lambda}_{\boldsymbol{\kappa}%
}\left(  \mathbf{s}\right)  \right)  $, where the equilibrium point is a dual
locus of principal eigenaxis components and likelihoods:%
\begin{align*}
\boldsymbol{\psi}  &  \triangleq\widehat{\Lambda}_{\boldsymbol{\psi}}\left(
\mathbf{s}\right)  =p\left(  \widehat{\Lambda}_{\boldsymbol{\psi}}\left(
\mathbf{s}\right)  |\omega_{1}\right)  +p\left(  \widehat{\Lambda
}_{\boldsymbol{\psi}}\left(  \mathbf{s}\right)  |\omega_{2}\right) \\
&  =\boldsymbol{\psi}_{1}+\boldsymbol{\psi}_{2}\\
&  =\sum\nolimits_{i=1}^{l_{1}}\psi_{1_{i_{\ast}}}\frac{k_{\mathbf{x}%
_{1_{i\ast}}}}{\left\Vert k_{\mathbf{x}_{1_{i\ast}}}\right\Vert }%
+\sum\nolimits_{i=1}^{l_{2}}\psi_{2_{i_{\ast}}}\frac{k_{\mathbf{x}_{2_{i\ast}%
}}}{\left\Vert k_{\mathbf{x}_{2_{i\ast}}}\right\Vert }%
\end{align*}
that is constrained to be in statistical equilibrium:%
\[
\sum\nolimits_{i=1}^{l_{1}}\psi_{1_{i_{\ast}}}\frac{k_{\mathbf{x}_{1_{i\ast}}%
}}{\left\Vert k_{\mathbf{x}_{1_{i\ast}}}\right\Vert }=\sum\nolimits_{i=1}%
^{l_{2}}\psi_{2_{i_{\ast}}}\frac{k_{\mathbf{x}_{2_{i\ast}}}}{\left\Vert
k_{\mathbf{x}_{2_{i\ast}}}\right\Vert }\text{.}%
\]

Therefore, the forces associated with the counter risk $\overline
{\mathfrak{R}}_{\mathfrak{\min}}\left(  Z_{1}|p\left(  \widehat{\Lambda
}_{\boldsymbol{\kappa}}\left(  \mathbf{s}\right)  |\omega_{1}\right)  \right)
$ and the risk $\mathfrak{R}_{\mathfrak{\min}}\left(  Z_{2}|p\left(
\widehat{\Lambda}_{\boldsymbol{\kappa}}\left(  \mathbf{s}\right)  |\omega
_{1}\right)  \right)  $ in the $Z_{1}$ and $Z_{2}$ decision regions: which are
related to positions and potential locations of reproducing kernels
$k_{\mathbf{x}_{1_{i\ast}}}$ of data points $\mathbf{x}_{1_{i_{\ast}}}$ that
are generated according to $p\left(  \mathbf{x}|\omega_{1}\right)  $, are
balanced with the forces associated with the risk $\mathfrak{R}%
_{\mathfrak{\min}}\left(  Z_{1}|p\left(  \widehat{\Lambda}_{\boldsymbol{\kappa
}}\left(  \mathbf{s}\right)  |\omega_{2}\right)  \right)  $ and the counter
risk $\overline{\mathfrak{R}}_{\mathfrak{\min}}\left(  Z_{2}|p\left(
\widehat{\Lambda}_{\boldsymbol{\kappa}}\left(  \mathbf{s}\right)  |\omega
_{2}\right)  \right)  $ in the $Z_{1}$ and $Z_{2}$ decision regions: which are
related to positions and potential locations of reproducing kernels
$k_{\mathbf{x}_{2_{i\ast}}}$ of data points $\mathbf{x}_{2_{i_{\ast}}}$ that
are generated according to $p\left(  \mathbf{x}|\omega_{2}\right)  $.

Furthermore, the eigenenergy $E_{\min}\left(  Z|p\left(  \widehat{\Lambda
}_{\boldsymbol{\kappa}}\left(  \mathbf{s}\right)  |\omega_{1}\right)  \right)
$ associated with the position or location of the likelihood ratio $p\left(
\widehat{\Lambda}_{\boldsymbol{\kappa}}\left(  \mathbf{s}\right)  |\omega
_{1}\right)  $ given class $\omega_{1}$ is balanced with the eigenenergy
$E_{\min}\left(  Z|p\left(  \widehat{\Lambda}_{\boldsymbol{\kappa}}\left(
\mathbf{s}\right)  |\omega_{2}\right)  \right)  $ associated with the position
or location of the likelihood ratio $p\left(  \widehat{\Lambda}%
_{\boldsymbol{\kappa}}\left(  \mathbf{s}\right)  |\omega_{2}\right)  $ given
class $\omega_{2}$:%
\[
\left\Vert \boldsymbol{\kappa}_{1}\right\Vert _{\min_{c}}^{2}+\delta\left(
y\right)  \sum\nolimits_{i=1}^{l_{1}}\psi_{1_{i_{\ast}}}\equiv\left\Vert
\boldsymbol{\kappa}_{2}\right\Vert _{\min_{c}}^{2}-\delta\left(  y\right)
\sum\nolimits_{i=1}^{l_{2}}\psi_{2_{i_{\ast}}}\text{,}%
\]
where the total eigenenergy%
\begin{align*}
\left\Vert \boldsymbol{\kappa}\right\Vert _{\min_{c}}^{2}  &  =\left\Vert
\boldsymbol{\kappa}_{1}-\boldsymbol{\kappa}_{2}\right\Vert _{\min_{c}}^{2}\\
&  =\left\Vert \boldsymbol{\kappa}_{1}\right\Vert _{\min_{c}}^{2}-\left\Vert
\boldsymbol{\kappa}_{1}\right\Vert \left\Vert \boldsymbol{\kappa}%
_{2}\right\Vert \cos\theta_{\boldsymbol{\kappa}_{1}\boldsymbol{\kappa}_{2}}\\
&  +\left\Vert \boldsymbol{\kappa}_{2}\right\Vert _{\min_{c}}^{2}-\left\Vert
\boldsymbol{\kappa}_{2}\right\Vert \left\Vert \boldsymbol{\kappa}%
_{1}\right\Vert \cos\theta_{\boldsymbol{\kappa}_{2}\boldsymbol{\kappa}_{1}}\\
&  =\sum\nolimits_{i=1}^{l_{1}}\psi_{1i\ast}\left(  1-\xi_{i}\right)
+\sum\nolimits_{i=1}^{l_{2}}\psi_{2i\ast}\left(  1-\xi_{i}\right)
\end{align*}
of the discrete, quadratic classification system $\boldsymbol{\kappa}%
^{T}k_{\mathbf{s}}+\kappa_{0}\overset{\omega_{1}}{\underset{\omega
_{2}}{\gtrless}}0$ is determined by the eigenenergies associated with the
position or location of the likelihood ratio $\boldsymbol{\kappa=\kappa}%
_{1}-\boldsymbol{\kappa}_{2}$ and the locus of a corresponding, quadratic
decision boundary $\boldsymbol{\kappa}^{T}k_{\mathbf{s}}+\kappa_{0}=0$.

It follows that the discrete, quadratic classification system
$\boldsymbol{\kappa}^{T}k_{\mathbf{s}}+\kappa_{0}\overset{\omega
_{1}}{\underset{\omega_{2}}{\gtrless}}0$ is in statistical equilibrium:%
\begin{align*}
f\left(  \widetilde{\Lambda}_{\boldsymbol{\kappa}}\left(  \mathbf{s}\right)
\right)  :  &  \int_{Z_{1}}\boldsymbol{\kappa}_{1}d\boldsymbol{\kappa}%
_{1}-\int_{Z_{1}}\boldsymbol{\kappa}_{2}d\boldsymbol{\kappa}_{2}+\delta\left(
y\right)  \sum\nolimits_{i=1}^{l_{1}}\psi_{1_{i_{\ast}}}\\
&  =\int_{Z_{2}}\boldsymbol{\kappa}_{2}d\boldsymbol{\kappa}_{2}-\int_{Z_{2}%
}\boldsymbol{\kappa}_{1}d\boldsymbol{\kappa}_{1}-\delta\left(  y\right)
\sum\nolimits_{i=1}^{l_{2}}\psi_{2_{i_{\ast}}}\text{,}%
\end{align*}
where the forces associated with the counter risk $\overline{\mathfrak{R}%
}_{\mathfrak{\min}}\left(  Z_{1}|p\left(  \widehat{\Lambda}%
_{\boldsymbol{\kappa}}\left(  \mathbf{s}\right)  |\omega_{1}\right)  \right)
$ and the risk $\mathfrak{R}_{\mathfrak{\min}}\left(  Z_{1}|p\left(
\widehat{\Lambda}_{\boldsymbol{\kappa}}\left(  \mathbf{s}\right)  |\omega
_{2}\right)  \right)  $ in the $Z_{1}$ decision region are balanced with the
forces associated with the counter risk $\overline{\mathfrak{R}}%
_{\mathfrak{\min}}\left(  Z_{2}|p\left(  \widehat{\Lambda}_{\boldsymbol{\kappa
}}\left(  \mathbf{s}\right)  |\omega_{2}\right)  \right)  $ and the risk
$\mathfrak{R}_{\mathfrak{\min}}\left(  Z_{2}|p\left(  \widehat{\Lambda
}_{\boldsymbol{\kappa}}\left(  \mathbf{s}\right)  |\omega_{1}\right)  \right)
$ in the $Z_{2}$ decision region:%
\begin{align*}
f\left(  \widetilde{\Lambda}_{\boldsymbol{\kappa}}\left(  \mathbf{s}\right)
\right)   &  :\overline{\mathfrak{R}}_{\mathfrak{\min}}\left(  Z_{1}|p\left(
\widehat{\Lambda}_{\boldsymbol{\kappa}}\left(  \mathbf{s}\right)  |\omega
_{1}\right)  \right)  -\mathfrak{R}_{\mathfrak{\min}}\left(  Z_{1}|p\left(
\widehat{\Lambda}_{\boldsymbol{\kappa}}\left(  \mathbf{s}\right)  |\omega
_{2}\right)  \right) \\
&  =\overline{\mathfrak{R}}_{\mathfrak{\min}}\left(  Z_{2}|p\left(
\widehat{\Lambda}_{\boldsymbol{\kappa}}\left(  \mathbf{s}\right)  |\omega
_{2}\right)  \right)  -\mathfrak{R}_{\mathfrak{\min}}\left(  Z_{2}|p\left(
\widehat{\Lambda}_{\boldsymbol{\kappa}}\left(  \mathbf{s}\right)  |\omega
_{1}\right)  \right)
\end{align*}
such that the expected risk $\mathfrak{R}_{\mathfrak{\min}}\left(
Z|\widehat{\Lambda}_{\boldsymbol{\kappa}}\left(  \mathbf{s}\right)  \right)  $
of the classification system is minimized, and the eigenenergies associated
with the counter risk $\overline{\mathfrak{R}}_{\mathfrak{\min}}\left(
Z_{1}|p\left(  \widehat{\Lambda}_{\boldsymbol{\kappa}}\left(  \mathbf{s}%
\right)  |\omega_{1}\right)  \right)  $ and the risk $\mathfrak{R}%
_{\mathfrak{\min}}\left(  Z_{1}|p\left(  \widehat{\Lambda}_{\boldsymbol{\kappa
}}\left(  \mathbf{s}\right)  |\omega_{2}\right)  \right)  $ in the $Z_{1}$
decision region are balanced with the eigenenergies associated with the
counter risk $\overline{\mathfrak{R}}_{\mathfrak{\min}}\left(  Z_{2}|p\left(
\widehat{\Lambda}_{\boldsymbol{\kappa}}\left(  \mathbf{s}\right)  |\omega
_{2}\right)  \right)  $ and the risk $\mathfrak{R}_{\mathfrak{\min}}\left(
Z_{2}|p\left(  \widehat{\Lambda}_{\boldsymbol{\kappa}}\left(  \mathbf{s}%
\right)  |\omega_{1}\right)  \right)  $ in the $Z_{2}$ decision region:%
\begin{align*}
f\left(  \widetilde{\Lambda}_{\boldsymbol{\kappa}}\left(  \mathbf{s}\right)
\right)   &  :E_{\min}\left(  Z_{1}|p\left(  \widehat{\Lambda}%
_{\boldsymbol{\kappa}}\left(  \mathbf{s}\right)  |\omega_{1}\right)  \right)
-E_{\min}\left(  Z_{1}|p\left(  \widehat{\Lambda}_{\boldsymbol{\kappa}}\left(
\mathbf{s}\right)  |\omega_{2}\right)  \right) \\
&  =E_{\min}\left(  Z_{2}|p\left(  \widehat{\Lambda}_{\boldsymbol{\kappa}%
}\left(  \mathbf{s}\right)  |\omega_{2}\right)  \right)  -E_{\min}\left(
Z_{2}|p\left(  \widehat{\Lambda}_{\boldsymbol{\kappa}}\left(  \mathbf{s}%
\right)  |\omega_{1}\right)  \right)
\end{align*}
such that the eigenenergy $E_{\min}\left(  Z|\widehat{\Lambda}%
_{\boldsymbol{\kappa}}\left(  \mathbf{s}\right)  \right)  $ of the
classification system is minimized.

Thus, any given discrete, quadratic classification system $\boldsymbol{\kappa
}^{T}k_{\mathbf{s}}+\kappa_{0}\overset{\omega_{1}}{\underset{\omega
_{2}}{\gtrless}}0$ exhibits an error rate that is consistent with the expected
risk $\mathfrak{R}_{\mathfrak{\min}}\left(  Z|\widehat{\Lambda}%
_{\boldsymbol{\kappa}}\left(  \mathbf{s}\right)  \right)  $ and the
corresponding eigenenergy $E_{\min}\left(  Z|\widehat{\Lambda}%
_{\boldsymbol{\kappa}}\left(  \mathbf{s}\right)  \right)  $ of the
classification system: for all random vectors $\mathbf{x}$ that are generated
according to $p\left(  \mathbf{x}|\omega_{1}\right)  $ and $p\left(
\mathbf{x}|\omega_{2}\right)  $, where $p\left(  \mathbf{x}|\omega_{1}\right)
$ and $p\left(  \mathbf{x}|\omega_{2}\right)  $ are related to statistical
distributions of random vectors $\mathbf{x}$ that have constant or unchanging statistics.

Therefore, a discrete, quadratic classification system $\boldsymbol{\kappa
}^{T}k_{\mathbf{s}}+\kappa_{0}\overset{\omega_{1}}{\underset{\omega
_{2}}{\gtrless}}0$ seeks a point of statistical equilibrium where the opposing
forces and influences of the classification system are balanced with each
other, such that the eigenenergy and the expected risk of the classification
system are minimized, and the classification system is in statistical equilibrium.

I\ will now show that the eigenenergy of a discrete, quadratic classification
system is conserved and remains relatively constant, so that the eigenenergy
and the expected risk of a discrete, quadratic classification system cannot be
created or destroyed, but only transferred from one classification system to another.

\section*{Law of Conservation of Eigenenergy:}

\subsection*{For Discrete Quadratic Classification Systems}

Take a collection of $N$ random vectors $\mathbf{x}$ of dimension $d$ that are
generated according to probability density functions $p\left(  \mathbf{x}%
|\omega_{1}\right)  $ and $p\left(  \mathbf{x}|\omega_{2}\right)  $ related to
statistical distributions of random vectors $\mathbf{x}$ that have constant or
unchanging statistics, where the number of random vectors $\mathbf{x}\sim
p\left(  \mathbf{x}|\omega_{1}\right)  $ equals the number of random vectors
$\mathbf{x}\sim p\left(  \mathbf{x}|\omega_{2}\right)  $, and let
$\widetilde{\Lambda}_{\boldsymbol{\kappa}}\left(  \mathbf{s}\right)
=\boldsymbol{\kappa}^{T}k_{\mathbf{s}}+\kappa_{0}\overset{\omega
_{1}}{\underset{\omega_{2}}{\gtrless}}0$ denote the likelihood ratio test for
a discrete, quadratic classification system, where $\omega_{1}$ or $\omega
_{2}$ is the true data category, $\boldsymbol{\kappa}$ is a locus of principal
eigenaxis components and likelihoods:%
\begin{align*}
\boldsymbol{\kappa}  &  \triangleq\widehat{\Lambda}_{\boldsymbol{\kappa}%
}\left(  \mathbf{s}\right)  =p\left(  \widehat{\Lambda}_{\boldsymbol{\kappa}%
}\left(  \mathbf{s}\right)  |\omega_{1}\right)  -p\left(  \widehat{\Lambda
}_{\boldsymbol{\kappa}}\left(  \mathbf{s}\right)  |\omega_{2}\right) \\
&  =\boldsymbol{\kappa}_{1}-\boldsymbol{\kappa}_{2}\\
&  =\sum\nolimits_{i=1}^{l_{1}}\psi_{1_{i_{\ast}}}k_{\mathbf{x}_{1_{i\ast}}%
}-\sum\nolimits_{i=1}^{l_{2}}\psi_{2_{i_{\ast}}}k_{\mathbf{x}_{2_{i\ast}}%
}\text{,}%
\end{align*}
where $k_{\mathbf{x}_{1_{i\ast}}}$ and $k_{\mathbf{x}_{2_{i\ast}}}$ are
reproducing kernels for respective data points $\mathbf{x}_{1_{i_{\ast}}}$ and
$\mathbf{x}_{2_{i_{\ast}}}$: the reproducing kernel $K(\mathbf{x,s}%
)=k_{\mathbf{s}}(\mathbf{x})$ is either $k_{\mathbf{s}}(\mathbf{x}%
)\triangleq\left(  \mathbf{x}^{T}\mathbf{s}+1\right)  ^{2}$ or $k_{\mathbf{s}%
}(\mathbf{x})\triangleq\exp\left(  -0.01\left\Vert \mathbf{x}-\mathbf{s}%
\right\Vert ^{2}\right)  $, $\mathbf{x}_{1_{i_{\ast}}}\sim p\left(
\mathbf{x}|\omega_{1}\right)  $, $\mathbf{x}_{2_{i_{\ast}}}\sim p\left(
\mathbf{x}|\omega_{2}\right)  $, $\psi_{1_{i_{\ast}}}$ and $\psi_{2_{i_{\ast}%
}}$ are scale factors that provide unit measures of likelihood for respective
data points $\mathbf{x}_{1_{i_{\ast}}}$ and $\mathbf{x}_{2_{i_{\ast}}}$ which
lie in either overlapping regions or tails regions of data distributions
related to $p\left(  \mathbf{x}|\omega_{1}\right)  $ and $p\left(
\mathbf{x}|\omega_{2}\right)  $, and $\kappa_{0}$ is a functional of
$\boldsymbol{\kappa}$:%
\[
\kappa_{0}=\sum\nolimits_{i=1}^{l}y_{i}\left(  1-\xi_{i}\right)
-\sum\nolimits_{i=1}^{l}k_{\mathbf{x}_{i\ast}}\boldsymbol{\kappa}\text{,}%
\]
where $\sum\nolimits_{i=1}^{l}k_{\mathbf{x}_{i\ast}}=\sum\nolimits_{i=1}%
^{l_{1}}k_{\mathbf{x}_{1_{i\ast}}}+\sum\nolimits_{i=1}^{l_{2}}k_{\mathbf{x}%
_{2_{i\ast}}}$ is a cluster of reproducing kernels of the data points
$\mathbf{x}_{1_{i_{\ast}}}$ and $\mathbf{x}_{2_{i_{\ast}}}$ used to form
$\boldsymbol{\kappa}$, $y_{i}$ are class membership statistics: if
$\mathbf{x}_{i\ast}\in\omega_{1}$, assign $y_{i}=1$; if $\mathbf{x}_{i\ast}%
\in\omega_{2}$, assign $y_{i}=-1$, and $\xi_{i}$ are regularization
parameters: $\xi_{i}=\xi=0$ for full rank kernel matrices or $\xi_{i}=\xi\ll1$
for low rank kernel matrices.

The expected risk $\mathfrak{R}_{\mathfrak{\min}}\left(  Z|\widehat{\Lambda
}_{\boldsymbol{\kappa}}\left(  \mathbf{s}\right)  \right)  $ and the
corresponding eigenenergy $E_{\min}\left(  Z|\widehat{\Lambda}%
_{\boldsymbol{\kappa}}\left(  \mathbf{s}\right)  \right)  $ of a discrete,
quadratic classification system $\boldsymbol{\kappa}^{T}k_{\mathbf{s}}%
+\kappa_{0}\overset{\omega_{1}}{\underset{\omega_{2}}{\gtrless}}0$ are
governed by the equilibrium point%
\[
\sum\nolimits_{i=1}^{l_{1}}\psi_{1i\ast}-\sum\nolimits_{i=1}^{l_{2}}%
\psi_{2i\ast}=0
\]
of the integral equation%
\begin{align*}
f\left(  \widetilde{\Lambda}_{\boldsymbol{\kappa}}\left(  \mathbf{s}\right)
\right)  =  &  \int_{Z_{1}}\boldsymbol{\kappa}_{1}d\boldsymbol{\kappa}%
_{1}+\int_{Z_{2}}\boldsymbol{\kappa}_{1}d\boldsymbol{\kappa}_{1}+\delta\left(
y\right)  \sum\nolimits_{i=1}^{l_{1}}\psi_{1_{i_{\ast}}}\\
&  =\int_{Z_{1}}\boldsymbol{\kappa}_{2}d\boldsymbol{\kappa}_{2}+\int_{Z_{2}%
}\boldsymbol{\kappa}_{2}d\boldsymbol{\kappa}_{2}-\delta\left(  y\right)
\sum\nolimits_{i=1}^{l_{2}}\psi_{2_{i_{\ast}}}\text{,}%
\end{align*}
over the decision space $Z=Z_{1}+Z_{2}$, where $Z_{1}$ and $Z_{2}$ are
symmetrical decision regions: $Z_{1}\simeq Z_{2}$, $\delta\left(  y\right)
\triangleq\sum\nolimits_{i=1}^{l}y_{i}\left(  1-\xi_{i}\right)  $, and the
forces associated with the counter risk $\overline{\mathfrak{R}}%
_{\mathfrak{\min}}\left(  Z_{1}|p\left(  \widehat{\Lambda}_{\boldsymbol{\kappa
}}\left(  \mathbf{s}\right)  |\omega_{1}\right)  \right)  $ and the risk
$\mathfrak{R}_{\mathfrak{\min}}\left(  Z_{2}|p\left(  \widehat{\Lambda
}_{\boldsymbol{\kappa}}\left(  \mathbf{s}\right)  |\omega_{1}\right)  \right)
$ in the $Z_{1}$ and $Z_{2}$ decision regions: which are related to positions
and potential locations of reproducing kernels $k_{\mathbf{x}_{1_{i\ast}}}$ of
data points $\mathbf{x}_{1_{i_{\ast}}}$ that are generated according to
$p\left(  \mathbf{x}|\omega_{1}\right)  $, are balanced with the forces
associated with the risk $\mathfrak{R}_{\mathfrak{\min}}\left(  Z_{1}|p\left(
\widehat{\Lambda}_{\boldsymbol{\kappa}}\left(  \mathbf{s}\right)  |\omega
_{2}\right)  \right)  $ and the counter risk $\overline{\mathfrak{R}%
}_{\mathfrak{\min}}\left(  Z_{2}|p\left(  \widehat{\Lambda}%
_{\boldsymbol{\kappa}}\left(  \mathbf{s}\right)  |\omega_{2}\right)  \right)
$ in the $Z_{1}$ and $Z_{2}$ decision regions: which are related to positions
and potential locations of reproducing kernels $k_{\mathbf{x}_{2_{i\ast}}}$ of
data points $\mathbf{x}_{2_{i_{\ast}}}$ that are generated according to
$p\left(  \mathbf{x}|\omega_{2}\right)  $.

Furthermore, the eigenenergy $E_{\min}\left(  Z|p\left(  \widehat{\Lambda
}_{\boldsymbol{\kappa}}\left(  \mathbf{s}\right)  |\omega_{1}\right)  \right)
$ associated with the position or location of the likelihood ratio $p\left(
\widehat{\Lambda}_{\boldsymbol{\kappa}}\left(  \mathbf{s}\right)  |\omega
_{1}\right)  $ given class $\omega_{1}$ is balanced with the eigenenergy
$E_{\min}\left(  Z|p\left(  \widehat{\Lambda}_{\boldsymbol{\kappa}}\left(
\mathbf{s}\right)  |\omega_{2}\right)  \right)  $ associated with the position
or location of the likelihood ratio $p\left(  \widehat{\Lambda}%
_{\boldsymbol{\kappa}}\left(  \mathbf{s}\right)  |\omega_{2}\right)  $ given
class $\omega_{2}$:%
\[
\left\Vert \boldsymbol{\kappa}_{1}\right\Vert _{\min_{c}}^{2}+\delta\left(
y\right)  \sum\nolimits_{i=1}^{l_{1}}\psi_{1_{i_{\ast}}}\equiv\left\Vert
\boldsymbol{\kappa}_{2}\right\Vert _{\min_{c}}^{2}-\delta\left(  y\right)
\sum\nolimits_{i=1}^{l_{2}}\psi_{2_{i_{\ast}}}\text{,}%
\]
where%
\begin{align*}
\left\Vert \boldsymbol{\kappa}\right\Vert _{\min_{c}}^{2}  &  =\left\Vert
\boldsymbol{\kappa}_{1}\right\Vert _{\min_{c}}^{2}-\left\Vert
\boldsymbol{\kappa}_{1}\right\Vert \left\Vert \boldsymbol{\kappa}%
_{2}\right\Vert \cos\theta_{\boldsymbol{\kappa}_{1}\boldsymbol{\kappa}_{2}}\\
&  +\left\Vert \boldsymbol{\kappa}_{2}\right\Vert _{\min_{c}}^{2}-\left\Vert
\boldsymbol{\kappa}_{2}\right\Vert \left\Vert \boldsymbol{\kappa}%
_{1}\right\Vert \cos\theta_{\boldsymbol{\kappa}_{2}\boldsymbol{\kappa}_{1}}\\
&  =\sum\nolimits_{i=1}^{l_{1}}\psi_{1i\ast}\left(  1-\xi_{i}\right)
+\sum\nolimits_{i=1}^{l_{2}}\psi_{2i\ast}\left(  1-\xi_{i}\right)  \text{.}%
\end{align*}

The eigenenergy $\left\Vert \boldsymbol{\kappa}\right\Vert _{\min_{c}}%
^{2}=\left\Vert \boldsymbol{\kappa}_{1}-\boldsymbol{\kappa}_{2}\right\Vert
_{\min_{c}}^{2}$ is the state of a discrete, quadratic classification system
$\boldsymbol{\kappa}^{T}k_{\mathbf{s}}+\kappa_{0}\overset{\omega
_{1}}{\underset{\omega_{2}}{\gtrless}}0$ that is associated with the position
or location of a dual likelihood ratio:%
\begin{align*}
\boldsymbol{\psi}  &  \triangleq\widehat{\Lambda}_{\boldsymbol{\psi}}\left(
\mathbf{x}\right)  =p\left(  \widehat{\Lambda}_{\boldsymbol{\psi}}\left(
\mathbf{x}\right)  |\omega_{1}\right)  +p\left(  \widehat{\Lambda
}_{\boldsymbol{\psi}}\left(  \mathbf{x}\right)  |\omega_{2}\right) \\
&  =\boldsymbol{\psi}_{1}+\boldsymbol{\psi}_{2}\\
&  =\sum\nolimits_{i=1}^{l_{1}}\psi_{1_{i_{\ast}}}\frac{k_{\mathbf{x}%
_{1_{i\ast}}}}{\left\Vert k_{\mathbf{x}_{1_{i\ast}}}\right\Vert }%
+\sum\nolimits_{i=1}^{l_{2}}\psi_{2_{i_{\ast}}}\frac{k_{\mathbf{x}_{2_{i\ast}%
}}}{\left\Vert k_{\mathbf{x}_{2_{i\ast}}}\right\Vert }\text{,}%
\end{align*}
which is constrained to be in statistical equilibrium:%
\[
\sum\nolimits_{i=1}^{l_{1}}\psi_{1_{i_{\ast}}}\frac{k_{\mathbf{x}_{1_{i\ast}}%
}}{\left\Vert k_{\mathbf{x}_{1_{i\ast}}}\right\Vert }=\sum\nolimits_{i=1}%
^{l_{2}}\psi_{2_{i_{\ast}}}\frac{k_{\mathbf{x}_{2_{i\ast}}}}{\left\Vert
k_{\mathbf{x}_{2_{i\ast}}}\right\Vert }\text{,}%
\]
and the locus of a corresponding, quadratic decision boundary
$\boldsymbol{\kappa}^{T}k_{\mathbf{s}}+\kappa_{0}=0$.

Thus, any given discrete, quadratic classification system $\boldsymbol{\kappa
}^{T}k_{\mathbf{s}}+\kappa_{0}\overset{\omega_{1}}{\underset{\omega
_{2}}{\gtrless}}0$ exhibits an error rate that is consistent with the expected
risk $\mathfrak{R}_{\mathfrak{\min}}\left(  Z|\widehat{\Lambda}%
_{\boldsymbol{\kappa}}\left(  \mathbf{s}\right)  \right)  $ and the
corresponding eigenenergy $E_{\min}\left(  Z|\widehat{\Lambda}%
_{\boldsymbol{\kappa}}\left(  \mathbf{s}\right)  \right)  $ of the
classification system: for all random vectors $\mathbf{x}$ that are generated
according to $p\left(  \mathbf{x}|\omega_{1}\right)  $ and $p\left(
\mathbf{x}|\omega_{2}\right)  $, where $p\left(  \mathbf{x}|\omega_{1}\right)
$ and $p\left(  \mathbf{x}|\omega_{2}\right)  $ are related to statistical
distributions of random vectors $\mathbf{x}$ have constant or unchanging statistics.

The total eigenenergy of a discrete, quadratic classification system
$\boldsymbol{\kappa}^{T}k_{\mathbf{s}}+\kappa_{0}\overset{\omega
_{1}}{\underset{\omega_{2}}{\gtrless}}0$ is found by adding up contributions
from characteristics of the classification system:

The eigenenergies $E_{\min}\left(  Z|p\left(  \widehat{\Lambda}%
_{\boldsymbol{\kappa}}\left(  \mathbf{s}\right)  |\omega_{1}\right)  \right)
$ and $E_{\min}\left(  Z|p\left(  \widehat{\Lambda}_{\boldsymbol{\kappa}%
}\left(  \mathbf{s}\right)  |\omega_{2}\right)  \right)  $ associated with the
positions or locations of the class-conditional likelihood ratios $p\left(
\widehat{\Lambda}_{\boldsymbol{\kappa}}\left(  \mathbf{s}\right)  |\omega
_{1}\right)  $ and $p\left(  \widehat{\Lambda}_{\boldsymbol{\kappa}}\left(
\mathbf{s}\right)  |\omega_{2}\right)  $, where%
\[
E_{\min}\left(  Z|p\left(  \widehat{\Lambda}_{\boldsymbol{\kappa}}\left(
\mathbf{s}\right)  |\omega_{1}\right)  \right)  =\left\Vert \boldsymbol{\kappa
}_{1}\right\Vert _{\min_{c}}^{2}\text{ \ and \ }E_{\min}\left(  Z|p\left(
\widehat{\Lambda}_{\boldsymbol{\kappa}}\left(  \mathbf{s}\right)  |\omega
_{2}\right)  \right)  =\left\Vert \boldsymbol{\kappa}_{2}\right\Vert
_{\min_{c}}^{2}%
\]
are related to eigenenergies associated with positions and potential locations
of extreme points that lie in either overlapping regions or tails regions of
statistical distributions related to the class-conditional probability density
functions $p\left(  \mathbf{x}|\omega_{1}\right)  $ and $p\left(
\mathbf{x}|\omega_{2}\right)  $, and the total eigenenergy $\left\Vert
\boldsymbol{\kappa}\right\Vert _{\min_{c}}^{2}$ satisfies the vector
equations
\begin{align*}
\left\Vert \boldsymbol{\kappa}\right\Vert _{\min_{c}}^{2}  &  =\left\Vert
\boldsymbol{\kappa}_{1}-\boldsymbol{\kappa}_{2}\right\Vert _{\min_{c}}^{2}\\
&  =\left\Vert \boldsymbol{\kappa}_{1}\right\Vert _{\min_{c}}^{2}-\left\Vert
\boldsymbol{\kappa}_{1}\right\Vert \left\Vert \boldsymbol{\kappa}%
_{2}\right\Vert \cos\theta_{\boldsymbol{\kappa}_{1}\boldsymbol{\kappa}_{2}}\\
&  +\left\Vert \boldsymbol{\kappa}_{2}\right\Vert _{\min_{c}}^{2}-\left\Vert
\boldsymbol{\kappa}_{2}\right\Vert \left\Vert \boldsymbol{\kappa}%
_{1}\right\Vert \cos\theta_{\boldsymbol{\kappa}_{2}\boldsymbol{\kappa}_{1}}%
\end{align*}
and%
\[
\left\Vert \boldsymbol{\kappa}\right\Vert _{\min_{c}}^{2}=\sum\nolimits_{i=1}%
^{l_{1}}\psi_{1i\ast}\left(  1-\xi_{i}\right)  +\sum\nolimits_{i=1}^{l_{2}%
}\psi_{2i\ast}\left(  1-\xi_{i}\right)  \text{.}%
\]

Any given discrete, quadratic classification system that is determined by a
likelihood ratio test:%
\[
\boldsymbol{\kappa}^{T}k_{\mathbf{s}}+\kappa_{0}\overset{\omega_{1}%
}{\underset{\omega_{2}}{\gtrless}}0\text{,}%
\]
where the class-conditional probability density functions $p\left(
\mathbf{x}|\omega_{1}\right)  $ and $p\left(  \mathbf{x}|\omega_{2}\right)  $
are related to statistical distributions of random vectors $\mathbf{x}$ that
have constant or unchanging statistics, and the locus of a quadratic decision
boundary:%
\[
D\left(  \mathbf{x}\right)  :\boldsymbol{\kappa}^{T}k_{\mathbf{s}}+\kappa
_{0}=0
\]
is governed by the locus of a dual likelihood ratio $p\left(  \widehat{\Lambda
}_{\boldsymbol{\psi}}\left(  \mathbf{x}\right)  |\omega_{1}\right)  +p\left(
\widehat{\Lambda}_{\boldsymbol{\psi}}\left(  \mathbf{x}\right)  |\omega
_{2}\right)  $ in statistical equilibrium:%
\[
p\left(  \widehat{\Lambda}_{\boldsymbol{\psi}}\left(  \mathbf{x}\right)
|\omega_{1}\right)  \rightleftharpoons p\left(  \widehat{\Lambda
}_{\boldsymbol{\psi}}\left(  \mathbf{x}\right)  |\omega_{2}\right)  \text{,}%
\]
is a closed classification system.

Thus, the total eigenenergy $E_{\min}\left(  Z|\widehat{\Lambda}%
_{\boldsymbol{\kappa}}\left(  \mathbf{s}\right)  \right)  $%
\begin{align*}
E_{\min}\left(  Z|\widehat{\Lambda}_{\boldsymbol{\kappa}}\left(
\mathbf{s}\right)  \right)   &  \triangleq\left\Vert \boldsymbol{\kappa
}\right\Vert _{\min_{c}}^{2}\\
&  =\left\Vert \boldsymbol{\kappa}_{1}-\boldsymbol{\kappa}_{2}\right\Vert
_{\min_{c}}^{2}\\
&  =\sum\nolimits_{i=1}^{l_{1}}\psi_{1i\ast}\left(  1-\xi_{i}\right)
+\sum\nolimits_{i=1}^{l_{2}}\psi_{2i\ast}\left(  1-\xi_{i}\right)
\end{align*}
of any given discrete, quadratic classification system $\boldsymbol{\kappa
}^{T}k_{\mathbf{s}}+\kappa_{0}\overset{\omega_{1}}{\underset{\omega
_{2}}{\gtrless}}0$ is conserved and remains relatively constant.

Therefore, the eigenenergy $E_{\min}\left(  Z|\widehat{\Lambda}%
_{\boldsymbol{\kappa}}\left(  \mathbf{s}\right)  \right)  $ of a discrete,
quadratic classification system $\boldsymbol{\kappa}^{T}k_{\mathbf{s}}%
+\kappa_{0}\overset{\omega_{1}}{\underset{\omega_{2}}{\gtrless}}0$ cannot be
created or destroyed, but only transferred from one classification system to another.

It follows that the corresponding expected risk $\mathfrak{R}_{\mathfrak{\min
}}\left(  Z|\widehat{\Lambda}_{\boldsymbol{\kappa}}\left(  \mathbf{s}\right)
\right)  $ of a discrete, quadratic classification system $\boldsymbol{\kappa
}^{T}k_{\mathbf{s}}+\kappa_{0}\overset{\omega_{1}}{\underset{\omega
_{2}}{\gtrless}}0$ cannot be created or destroyed, but only transferred from
one classification system to another.

I\ will now identify the fundamental property which is common to each of the
scaled extreme points on any given likelihood ratio $\widehat{\Lambda
}_{\boldsymbol{\kappa}}\left(  \mathbf{s}\right)  $ and quadratic decision
boundary $D_{0}\left(  \mathbf{s}\right)  $ that is determined by a quadratic
eigenlocus decision rule $\widetilde{\Lambda}_{\boldsymbol{\kappa}}\left(
\mathbf{s}\right)  \overset{H_{1}}{\underset{H_{2}}{\gtrless}}0$.

\subsubsection{Inherent Property of Eigen-scaled Extreme Points on
$\boldsymbol{\kappa}_{1}-\boldsymbol{\kappa}_{2}$}

Given that a quadratic eigenlocus $\boldsymbol{\kappa}=\boldsymbol{\kappa}%
_{1}-\boldsymbol{\kappa}_{2}$ is a locus of likelihoods that determines a
likelihood ratio $\widetilde{\Lambda}_{\boldsymbol{\kappa}}\left(
\mathbf{s}\right)  $ and a locus of principal eigenaxis components that
determines the coordinate system of a quadratic decision boundary
$D_{0}\left(  \mathbf{s}\right)  $, it follows that the total allowed
eigenenergy%
\[
\left\Vert \psi_{1_{i_{\ast}}}k_{\mathbf{x}_{1_{i_{\ast}}}}\right\Vert
_{\min_{c}}^{2}\text{ or \ }\left\Vert \psi_{2_{i_{\ast}}}k_{\mathbf{x}%
_{2_{i\ast}}}\right\Vert _{\min_{c}}^{2}%
\]
exhibited by each scaled extreme vector%
\[
\psi_{1_{i_{\ast}}}k_{\mathbf{x}_{1_{i_{\ast}}}}\text{ or \ }\psi_{2_{i_{\ast
}}}k_{\mathbf{x}_{2_{i\ast}}}%
\]
on $\boldsymbol{\kappa}_{1}-\boldsymbol{\kappa}_{2}$ and the corresponding
class-conditional risk:%
\[
\int_{Z_{2}}p\left(  k_{\mathbf{x}_{1i\ast}}|\operatorname{comp}%
_{\overrightarrow{k_{\mathbf{x}_{1i\ast}}}}\left(
\overrightarrow{\boldsymbol{\kappa}}\right)  \right)  d\boldsymbol{\kappa}%
_{1}\left(  k_{\mathbf{x}_{1i\ast}}\right)  \text{ or \ }\int_{Z_{1}}p\left(
k_{\mathbf{x}_{2i\ast}}|\operatorname{comp}_{\overrightarrow{k_{\mathbf{x}%
_{2i\ast}}}}\left(  \overrightarrow{\boldsymbol{\kappa}}\right)  \right)
d\boldsymbol{\kappa}_{2}\left(  k_{\mathbf{x}_{2_{i\ast}}}\right)
\]
or class-conditional counter risk:%
\[
\int_{Z_{1}}p\left(  k_{\mathbf{x}_{1i\ast}}|\operatorname{comp}%
_{\overrightarrow{k_{\mathbf{x}_{1i\ast}}}}\left(
\overrightarrow{\boldsymbol{\kappa}}\right)  \right)  d\boldsymbol{\kappa}%
_{1}\left(  k_{\mathbf{x}_{1i\ast}}\right)  \text{ or \ }\int_{Z_{2}}p\left(
k_{\mathbf{x}_{2i\ast}}|\operatorname{comp}_{\overrightarrow{k_{\mathbf{x}%
_{2i\ast}}}}\left(  \overrightarrow{\boldsymbol{\kappa}}\right)  \right)
d\boldsymbol{\kappa}_{2}\left(  k_{\mathbf{x}_{2_{i\ast}}}\right)
\]
possessed by each extreme point $k_{\mathbf{x}_{1_{i_{\ast}}}}$ or
$k_{\mathbf{x}_{2_{i\ast}}}$, which are determined by $\left\Vert
\psi_{1_{i_{\ast}}}k_{\mathbf{x}_{1_{i_{\ast}}}}\right\Vert _{\min_{c}}^{2}$
or\ $\left\Vert \psi_{2_{i_{\ast}}}k_{\mathbf{x}_{2_{i\ast}}}\right\Vert
_{\min_{c}}^{2}$, \emph{jointly satisfy} the fundamental quadratic eigenlocus
integral equation of binary classification in Eq.
(\ref{Quadratic Eigenlocus Integral Equation V}). Thereby, it is concluded
that the \emph{fundamental property} possessed by each of the scaled extreme
points on a quadratic eigenlocus $\boldsymbol{\kappa}_{1}-\boldsymbol{\kappa
}_{2}$ is the \emph{total allowed eigenenergy} exhibited by a corresponding,
scaled extreme vector.

I will now devise an expression for a quadratic eigenlocus that is a locus of
discrete conditional probabilities.

\section{Quadratic Eigenlocus of Probabilities}

Write a quadratic eigenlocus $\boldsymbol{\kappa}$ in terms of%
\begin{align*}
\boldsymbol{\kappa}  &  =\lambda_{\max_{\psi}}^{-1}\sum\nolimits_{i=1}^{l_{1}%
}\frac{k_{\mathbf{x}_{1_{i_{\ast}}}}}{\left\Vert k_{\mathbf{x}_{1_{i_{\ast}}}%
}\right\Vert }\left\Vert k_{\mathbf{x}_{1_{i_{\ast}}}}\right\Vert
^{2}\widehat{\operatorname{cov}}_{sm_{\updownarrow}}\left(  k_{\mathbf{x}%
_{1_{i_{\ast}}}}\right) \\
&  -\lambda_{\max_{\psi}}^{-1}\sum\nolimits_{i=1}^{l_{2}}\frac{k_{\mathbf{x}%
_{2_{i_{\ast}}}}}{\left\Vert k_{\mathbf{x}_{2_{i_{\ast}}}}\right\Vert
}\left\Vert k_{\mathbf{x}_{2_{i_{\ast}}}}\right\Vert ^{2}%
\widehat{\operatorname{cov}}_{sm_{\updownarrow}}\left(  k_{\mathbf{x}%
_{2_{i_{\ast}}}}\right)  \text{,}%
\end{align*}
where $\widehat{\operatorname{cov}}_{sm_{\updownarrow}}\left(  k_{\mathbf{x}%
_{1_{i_{\ast}}}}\right)  $ and $\widehat{\operatorname{cov}}_{sm_{\updownarrow
}}\left(  k_{\mathbf{x}_{2_{i_{\ast}}}}\right)  $ denote the symmetrically
balanced, signed magnitudes in Eqs (\ref{Unidirectional Scaling Term One1 Q})
and (\ref{Unidirectional Scaling Term Two1 Q}): the terms $\frac
{k_{\mathbf{x}_{1_{i_{\ast}}}}}{\left\Vert k_{\mathbf{x}_{1_{i_{\ast}}}%
}\right\Vert }$ and $\frac{k_{\mathbf{x}_{2_{i_{\ast}}}}}{\left\Vert
k_{\mathbf{x}_{2_{i_{\ast}}}}\right\Vert }$ have been introduced and rearranged.

Next, rewrite $\boldsymbol{\kappa}$ in terms of total allowed eigenenergies%
\begin{align}
\boldsymbol{\kappa}  &  =\sum\nolimits_{i=1}^{l_{1}}\left\Vert \lambda
_{\max_{\psi}}^{-1}\left(  \widehat{\operatorname{cov}}_{sm_{\updownarrow}%
}\left(  k_{\mathbf{x}_{1_{i_{\ast}}}}\right)  \right)  ^{\frac{1}{2}%
}k_{\mathbf{x}_{1_{i_{\ast}}}}\right\Vert _{\min_{c}}^{2}\frac{k_{\mathbf{x}%
_{1_{i_{\ast}}}}}{\left\Vert k_{\mathbf{x}_{1_{i_{\ast}}}}\right\Vert
}\label{Probabilisitc Expression for Normal Eigenlocus Q}\\
&  -\sum\nolimits_{i=1}^{l_{2}}\left\Vert \lambda_{\max_{\psi}}^{-1}\left(
\widehat{\operatorname{cov}}_{sm_{\updownarrow}}\left(  k_{\mathbf{x}%
_{2_{i_{\ast}}}}\right)  \right)  ^{\frac{1}{2}}k_{\mathbf{x}_{2_{i_{\ast}}}%
}\right\Vert _{\min_{c}}^{2}\frac{k_{\mathbf{x}_{2_{i_{\ast}}}}}{\left\Vert
k_{\mathbf{x}_{2_{i_{\ast}}}}\right\Vert }\text{,}\nonumber
\end{align}
where the conditional probability $\mathcal{P}\left(  k_{\mathbf{x}%
_{1_{i_{\ast}}}}|\tilde{Z}\left(  k_{\mathbf{x}_{1_{i_{\ast}}}}\right)
\right)  $ of observing an $k_{\mathbf{x}_{1_{i_{\ast}}}}$ extreme point
within a localized region $\tilde{Z}\left(  k_{\mathbf{x}_{1_{i_{\ast}}}%
}\right)  $ of a decision space $Z=Z_{1}+Z_{2}$ is given by the expression:%
\[
\mathcal{P}\left(  k_{\mathbf{x}_{1_{i_{\ast}}}}|\tilde{Z}\left(
k_{\mathbf{x}_{1_{i_{\ast}}}}\right)  \right)  =\left\Vert \lambda_{\max
_{\psi}}^{-1}\left(  \widehat{\operatorname{cov}}_{sm_{\updownarrow}}\left(
k_{\mathbf{x}_{1_{i_{\ast}}}}\right)  \right)  ^{\frac{1}{2}}k_{\mathbf{x}%
_{1_{i_{\ast}}}}\right\Vert _{\min_{c}}^{2}\text{,}%
\]
where $\tilde{Z}\left(  k_{\mathbf{x}_{1_{i_{\ast}}}}\right)  \subset Z_{1}$
or $\tilde{Z}\left(  k_{\mathbf{x}_{1_{i_{\ast}}}}\right)  \subset Z_{2}$, and
the conditional probability $\mathcal{P}\left(  k_{\mathbf{x}_{2_{i_{\ast}}}%
}|\tilde{Z}\left(  k_{\mathbf{x}_{2_{i_{\ast}}}}\right)  \right)
\mathcal{\ }$of observing an $k_{\mathbf{x}_{2_{i_{\ast}}}}$ extreme point
within a localized region $\tilde{Z}\left(  k_{\mathbf{x}_{2_{i_{\ast}}}%
}\right)  $ of a decision space $Z$ is given by the expression:%
\[
\mathcal{P}\left(  k_{\mathbf{x}_{2_{i_{\ast}}}}|\tilde{Z}\left(
k_{\mathbf{x}_{2_{i_{\ast}}}}\right)  \right)  =\left\Vert \lambda_{\max
_{\psi}}^{-1}\left(  \widehat{\operatorname{cov}}_{sm_{\updownarrow}}\left(
k_{\mathbf{x}_{2_{i_{\ast}}}}\right)  \right)  ^{\frac{1}{2}}k_{\mathbf{x}%
_{2_{i_{\ast}}}}\right\Vert _{\min_{c}}^{2}\text{,}%
\]
where $\tilde{Z}\left(  k_{\mathbf{x}_{2_{i_{\ast}}}}\right)  \subset Z_{1}$
or $\tilde{Z}\left(  k_{\mathbf{x}_{2_{i_{\ast}}}}\right)  \subset Z_{2}$.

Now rewrite Eq. (\ref{Probabilisitc Expression for Normal Eigenlocus Q}) as a
locus of discrete conditional probabilities:%
\begin{align}
\boldsymbol{\kappa}_{1}-\boldsymbol{\kappa}_{2}  &  =\sum\nolimits_{i=1}%
^{l_{1}}\mathcal{P}\left(  k_{\mathbf{x}_{1_{i_{\ast}}}}|\tilde{Z}\left(
k_{\mathbf{x}_{1_{i_{\ast}}}}\right)  \right)  \frac{k_{\mathbf{x}%
_{1_{i_{\ast}}}}}{\left\Vert k_{\mathbf{x}_{1_{i_{\ast}}}}\right\Vert
}\label{SDNE Conditional Likelihood Ratio Q}\\
&  -\sum\nolimits_{i=1}^{l_{2}}\mathcal{P}\left(  k_{\mathbf{x}_{2_{i_{\ast}}%
}}|\tilde{Z}\left(  k_{\mathbf{x}_{2_{i_{\ast}}}}\right)  \right)
\frac{k_{\mathbf{x}_{2_{i_{\ast}}}}}{\left\Vert k_{\mathbf{x}_{2_{i_{\ast}}}%
}\right\Vert }\nonumber
\end{align}
which provides discrete measures for conditional probabilities of
classification errors%
\[
\mathcal{P}_{\min_{e}}\left(  k_{\mathbf{x}_{1_{i_{\ast}}}}|Z_{2}\left(
k_{\mathbf{x}_{1_{i_{\ast}}}}\right)  \right)  \text{ and }\mathcal{P}%
_{\min_{e}}\left(  k_{\mathbf{x}_{2_{i_{\ast}}}}|Z_{1}\left(  k_{\mathbf{x}%
_{2_{i_{\ast}}}}\right)  \right)
\]
for $k_{\mathbf{x}_{1_{i_{\ast}}}}$ extreme points that lie in the $Z_{2}$
decision region and $k_{\mathbf{x}_{2_{i_{\ast}}}}$ extreme points that lie in
the $Z_{1}$ decision region.

I\ will now use Eq. (\ref{SDNE Conditional Likelihood Ratio Q}) to devise a
probabilistic expression for a quadratic eigenlocus discriminant function.

\subsection{A Probabilistic Expression for $\boldsymbol{\kappa}$}

Returning to Eq. (\ref{Statistical Locus of Category Decision Q}), consider
the estimate $\widetilde{\Lambda}_{\boldsymbol{\kappa}}\left(  \mathbf{s}%
\right)  $ that an unknown pattern vector $\mathbf{x}$ is located within some
particular region of $%
\mathbb{R}
^{d}$%
\begin{align*}
\widetilde{\Lambda}_{\boldsymbol{\kappa}}\left(  \mathbf{s}\right)   &
=\left(  k_{\mathbf{x}}-k_{\widehat{\mathbf{x}}_{i\ast}}\right)
\boldsymbol{\kappa}\mathbf{/}\left\Vert \boldsymbol{\kappa}\right\Vert \\
&  \mathbf{+}\frac{1}{\left\Vert \boldsymbol{\kappa}\right\Vert }%
\sum\nolimits_{i=1}^{l}y_{i}\left(  1-\xi_{i}\right)
\end{align*}
based on the value of the decision locus $\operatorname{comp}%
_{\overrightarrow{\widehat{\boldsymbol{\kappa}}}}\left(
\overrightarrow{\left(  k_{\mathbf{x}}-k_{\widehat{\mathbf{x}}_{i\ast}%
}\right)  }\right)  $ and class membership statistic $\frac{1}{\left\Vert
\boldsymbol{\kappa}\right\Vert }\sum\nolimits_{i=1}^{l}y_{i}\left(  1-\xi
_{i}\right)  $, where $\operatorname{comp}%
_{\overrightarrow{\widehat{\boldsymbol{\kappa}}}}\left(
\overrightarrow{\left(  k_{\mathbf{x}}-k_{\widehat{\mathbf{x}}_{i\ast}%
}\right)  }\right)  $ denotes a signed magnitude $\left\Vert k_{\mathbf{x}%
}-k_{\widehat{\mathbf{x}}_{i\ast}}\right\Vert \cos\theta$ along the axis of
$\widehat{\boldsymbol{\kappa}}$, $\theta$ is the angle between the vector
$\left(  k_{\mathbf{x}}-k_{\widehat{\mathbf{x}}_{i\ast}}\right)  $ and
$\widehat{\boldsymbol{\kappa}}$, and $\widehat{\boldsymbol{\kappa}}$ denotes
the unit quadratic eigenlocus $\boldsymbol{\kappa}\mathbf{/}\left\Vert
\boldsymbol{\kappa}\right\Vert $.

I will now demonstrate that the signed magnitude expression $\left(
k_{\mathbf{x}}-k_{\widehat{\mathbf{x}}_{i\ast}}\right)  \boldsymbol{\kappa
}\mathbf{/}\left\Vert \boldsymbol{\kappa}\right\Vert $ is a locus of discrete
conditional probabilities.

Substitute the expression for $\boldsymbol{\kappa}_{1}-\boldsymbol{\kappa}%
_{2}$ in Eq. (\ref{SDNE Conditional Likelihood Ratio Q}) into the expression
for the quadratic eigenlocus test in Eq.
(\ref{NormalEigenlocusTestStatistic Q}). Denote the unit primal principal
eigenaxis components $\frac{k_{\mathbf{x}_{1_{i_{\ast}}}}}{\left\Vert
k_{\mathbf{x}_{1_{i_{\ast}}}}\right\Vert }$ and $\frac{k_{\mathbf{x}%
_{2_{i_{\ast}}}}}{\left\Vert k_{\mathbf{x}_{2_{i_{\ast}}}}\right\Vert }$ by
$\widehat{k}_{\mathbf{x}_{1_{i_{\ast}}}}$ and $\widehat{k}_{\mathbf{x}%
_{2_{i_{\ast}}}}$. It follows that the probability that the unknown pattern
vector $\mathbf{x}$ is located within a specific region of $%
\mathbb{R}
^{d}$ is provided by the expression:%
\begin{align*}
\widetilde{\Lambda}_{\boldsymbol{\kappa}}\left(  \mathbf{s}\right)   &
=\sum\nolimits_{i=1}^{l_{1}}\left[  \left(  k_{\mathbf{x}}%
-k_{\widehat{\mathbf{x}}_{i\ast}}\right)  \widehat{k}_{\mathbf{x}_{1_{i_{\ast
}}}}\right]  \mathcal{P}\left(  k_{\mathbf{x}_{1_{i_{\ast}}}}|\tilde{Z}\left(
k_{\mathbf{x}_{1_{i_{\ast}}}}\right)  \right) \\
&  -\sum\nolimits_{i=1}^{l_{2}}\left[  \left(  k_{\mathbf{x}}%
-k_{\widehat{\mathbf{x}}_{i\ast}}\right)  \widehat{k}_{\mathbf{x}_{2_{i_{\ast
}}}}\right]  \mathcal{P}\left(  k_{\mathbf{x}_{2_{i_{\ast}}}}|\tilde{Z}\left(
k_{\mathbf{x}_{2_{i_{\ast}}}}\right)  \right) \\
&  \mathbf{+}\frac{1}{\left\Vert \boldsymbol{\kappa}\right\Vert }%
\sum\nolimits_{i=1}^{l}y_{i}\left(  1-\xi_{i}\right)  \text{,}%
\end{align*}
where $\mathcal{P}\left(  k_{\mathbf{x}_{1_{i_{\ast}}}}|\tilde{Z}\left(
k_{\mathbf{x}_{1_{i_{\ast}}}}\right)  \right)  $ and $\mathcal{P}\left(
k_{\mathbf{x}_{2_{i_{\ast}}}}|\tilde{Z}\left(  k_{\mathbf{x}_{2_{i_{\ast}}}%
}\right)  \right)  $ provides discrete measures for conditional probabilities
of classification errors $\mathcal{P}_{\min_{e}}\left(  k_{\mathbf{x}%
_{1_{i_{\ast}}}}|Z_{2}\left(  k_{\mathbf{x}_{1_{i_{\ast}}}}\right)  \right)  $
and $\mathcal{P}_{\min_{e}}\left(  k_{\mathbf{x}_{2_{i_{\ast}}}}|Z_{1}\left(
k_{\mathbf{x}_{2_{i_{\ast}}}}\right)  \right)  $ for $k_{\mathbf{x}%
_{1_{i_{\ast}}}}$ extreme points that lie in the $Z_{2}$ decision region and
$k_{\mathbf{x}_{2_{i_{\ast}}}}$ extreme points that lie in the $Z_{1}$
decision region.

The above expression reduces to a locus of discrete conditional probabilities%
\begin{align}
\widetilde{\Lambda}_{\boldsymbol{\kappa}}\left(  \mathbf{s}\right)   &
=\sum\nolimits_{i=1}^{l_{1}}\operatorname{comp}_{\overrightarrow{k_{\mathbf{x}%
_{1_{i_{\ast}}}}}}\left(  \overrightarrow{\left(  k_{\mathbf{x}}%
-k_{\widehat{\mathbf{x}}_{i\ast}}\right)  }\right)  \mathcal{P}\left(
k_{\mathbf{x}_{1_{i_{\ast}}}}|\tilde{Z}\left(  k_{\mathbf{x}_{1_{i_{\ast}}}%
}\right)  \right) \label{Normal Eigenlocus Likelihood Ratio Q}\\
&  -\sum\nolimits_{i=1}^{l_{2}}\operatorname{comp}%
_{\overrightarrow{k_{\mathbf{x}_{2_{i_{\ast}}}}}}\left(
\overrightarrow{\left(  k_{\mathbf{x}}-k_{\widehat{\mathbf{x}}_{i\ast}%
}\right)  }\right)  \mathcal{P}\left(  k_{\mathbf{x}_{2_{i_{\ast}}}}|\tilde
{Z}\left(  k_{\mathbf{x}_{2_{i_{\ast}}}}\right)  \right) \nonumber\\
&  \mathbf{+}\frac{1}{\left\Vert \boldsymbol{\kappa}\right\Vert }%
\sum\nolimits_{i=1}^{l}y_{i}\left(  1-\xi_{i}\right) \nonumber
\end{align}
so that the conditional probability $\mathcal{P}\left(  \mathbf{x}|\tilde
{Z}\left(  k_{\mathbf{x}_{1_{i_{\ast}}}}\right)  \right)  $ of finding the
unknown pattern vector $\mathbf{x}$ within the localized region $\tilde
{Z}\left(  k_{\mathbf{x}_{1_{i_{\ast}}}}\right)  $ of the decision space $Z$
is determined by the likelihood statistic:%
\begin{equation}
\mathcal{P}\left(  \mathbf{x}|\tilde{Z}\left(  k_{\mathbf{x}_{1_{i_{\ast}}}%
}\right)  \right)  =\operatorname{comp}_{\overrightarrow{k_{\mathbf{x}%
_{1_{i_{\ast}}}}}}\left(  \overrightarrow{\left(  k_{\mathbf{x}}%
-k_{\widehat{\mathbf{x}}_{i\ast}}\right)  }\right)  \mathcal{P}\left(
k_{\mathbf{x}_{1_{i_{\ast}}}}|\tilde{Z}\left(  k_{\mathbf{x}_{1_{i_{\ast}}}%
}\right)  \right)  \text{,} \label{Probability Estimate One Q}%
\end{equation}
and the conditional probability $\mathcal{P}\left(  \mathbf{x}|\tilde
{Z}\left(  k_{\mathbf{x}_{2_{i_{\ast}}}}\right)  \right)  $ of finding the
unknown pattern vector $\mathbf{x}$ within the localized region $\tilde
{Z}\left(  k_{\mathbf{x}_{2_{i_{\ast}}}}\right)  $ of the decision space $Z$
is determined by the likelihood statistic:%
\begin{equation}
\mathcal{P}\left(  \mathbf{x}|\tilde{Z}\left(  k_{\mathbf{x}_{2_{i_{\ast}}}%
}\right)  \right)  =\operatorname{comp}_{\overrightarrow{k_{\mathbf{x}%
_{2_{i_{\ast}}}}}}\left(  \overrightarrow{\left(  k_{\mathbf{x}}%
-k_{\widehat{\mathbf{x}}_{i\ast}}\right)  }\right)  \mathcal{P}\left(
k_{\mathbf{x}_{2_{i_{\ast}}}}|\tilde{Z}\left(  k_{\mathbf{x}_{2_{i_{\ast}}}%
}\right)  \right)  \text{.} \label{Probability Estimate Two Q}%
\end{equation}

Therefore, the likelihood statistic $\mathcal{P}\left(  \mathbf{x}|\tilde
{Z}\left(  k_{\mathbf{x}_{1_{i_{\ast}}}}\right)  \right)  $ in Eq.
(\ref{Probability Estimate One Q}) is proportional, according to the signed
magnitude $\operatorname{comp}_{\overrightarrow{k_{\mathbf{x}_{1_{i_{\ast}}}}%
}}\left(  \overrightarrow{\left(  k_{\mathbf{x}}-k_{\widehat{\mathbf{x}%
}_{i\ast}}\right)  }\right)  $ along the axis of $k_{\mathbf{x}_{1_{i_{\ast}}%
}}$, to the conditional probability $\mathcal{P}\left(  k_{\mathbf{x}%
_{1_{i_{\ast}}}}|\tilde{Z}\left(  k_{\mathbf{x}_{1_{i_{\ast}}}}\right)
\right)  $ of finding the extreme point $k_{\mathbf{x}_{1_{i_{\ast}}}}$ within
a localized region $\tilde{Z}\left(  k_{\mathbf{x}_{1_{i_{\ast}}}}\right)  $
of the decision space $Z$. Similarly, the likelihood statistic $\mathcal{P}%
\left(  \mathbf{x}|\tilde{Z}\left(  k_{\mathbf{x}_{2_{i_{\ast}}}}\right)
\right)  $ in Eq. (\ref{Probability Estimate Two Q}) is proportional,
according to the signed magnitude $\operatorname{comp}%
_{\overrightarrow{k_{\mathbf{x}_{2_{i_{\ast}}}}}}\left(
\overrightarrow{\left(  k_{\mathbf{x}}-k_{\widehat{\mathbf{x}}_{i\ast}%
}\right)  }\right)  $ along the axis of $k_{\mathbf{x}_{2_{i_{\ast}}}}$, to
the conditional probability $\mathcal{P}\left(  k_{\mathbf{x}_{2_{i_{\ast}}}%
}|\tilde{Z}\left(  k_{\mathbf{x}_{2_{i_{\ast}}}}\right)  \right)  $ of finding
the extreme point $k_{\mathbf{x}_{2_{i_{\ast}}}}$ within a localized region
$\tilde{Z}\left(  k_{\mathbf{x}_{2_{i_{\ast}}}}\right)  $ of the decision
space $Z$.

Thus, it is concluded that the signed magnitude expression $\left(
k_{\mathbf{x}}-k_{\widehat{\mathbf{x}}_{i\ast}}\right)  \boldsymbol{\kappa
}\mathbf{/}\left\Vert \boldsymbol{\kappa}\right\Vert $ in Eq.
(\ref{Normal Eigenlocus Likelihood Ratio Q}) is a locus of discrete
conditional probabilities that satisfies the quadratic eigenlocus integral
equation of binary classification in Eq.
(\ref{Quadratic Eigenlocus Integral Equation V}).

\section{Design of Optimal Decision Systems}

I will now consider the design and development of optimal, statistical pattern
recognition or classification systems using either linear eigenlocus or
quadratic eigenlocus discriminant and decision functions. I\ will show that
both classes of discriminant functions are scalable modules for optimal,
statistical classification systems where the eigenenergy and the expected risk
of the classification system are minimized. Accordingly, I\ will show that
both classes of discriminant functions are scalable, individual components of
optimal ensemble systems, where any given ensemble of linear or quadratic
binary classifiers exhibits optimal generalization performance for an
$M$-class feature space.

I\ will also show that both classes of decision functions provide a practical
statistical gauge for measuring data distribution overlap and the decision
error rate for two given sets of feature or pattern vectors. The statistical
gauge can also be used to identify homogeneous data distributions.

I will begin by demonstrating that linear and quadratic eigenlocus
discriminant functions are characteristic functions for a given class
$\mathcal{\omega}_{i}$ of data. Given a subset $A$ of a larger set, the
characteristic function $\chi_{A}$, sometimes also called the indicator
function, is the function defined to be identically one on $A$, and is zero elsewhere.

\subsection{Characteristic Functions}

Let $\widehat{\Lambda}_{\mathfrak{B}_{ij}}\left(  \mathbf{x}\right)  $ denote
a linear or quadratic discriminant function for two given pattern classes
$\mathcal{\omega}_{i}$ and $\omega_{j}$, where the feature vectors in class
$\mathcal{\omega}_{i}$ have the training label $+1$, and the feature vectors
in class $\omega_{j}$ have the training label $-1$. I will now demonstrate
that $\widehat{\Lambda}_{\mathfrak{B}_{ij}}\left(  \mathbf{x}\right)  $ is a
characteristic function or indicator function for feature vectors $\mathbf{x}$
that belong to class $\omega_{i}$: $\mathbf{x\in\,}\omega_{i}$.

Let $A$ be the event that an unseen feature vector $\mathbf{x\in\,}\omega_{i}$
lies in the decision region $Z_{1}$ so that $\operatorname{sign}\left(
\widehat{\Lambda}_{\mathfrak{B}_{ij}}\left(  \mathbf{x}\right)  \right)  =1$.
Then the probability $P\left(  \left(  A\right)  \right)  $ of $A$ can be
written as an expectation as follows: Define the characteristic function%
\[
\chi_{A}=\left\{
\begin{array}
[c]{cc}%
1\text{,} & \text{if event }A\text{ occurs}\\
0\text{,} & \text{otherwise.}%
\end{array}
\right.
\]
Therefore, $\chi_{A}$ is a random variable and%
\begin{align*}
E\left[  \chi_{A}\right]   &  =%
{\displaystyle\sum\nolimits_{r=o}^{1}}
rP\left(  \chi_{A}=r\right) \\
&  =P\left(  A\right)  \text{.}%
\end{align*}
Thus,%
\[
P\left(  A\right)  =E\left[  \chi_{A}\right]  \text{.}%
\]
\qquad\qquad\qquad\qquad

Thereby, the discriminant function $\widehat{\Lambda}_{\mathfrak{B}_{ij}%
}\left(  \mathbf{x}\right)  $ is an indicator\emph{\ }function $\chi
_{\omega_{i}}$ for feature vectors $\mathbf{x}$ that belong to class
$\omega_{i}$, where $\chi_{\omega_{i}}$ denotes the event that an unseen
feature vector $\mathbf{x\in\,}\omega_{i}$ lies in the decision region $Z_{1}$
so that $\operatorname{sign}\left(  \widehat{\Lambda}_{\mathfrak{B}_{ij}%
}\left(  \mathbf{x}\right)  \right)  =1$:%
\begin{align*}
E\left[  \chi_{\omega_{i}}\right]   &  =%
{\displaystyle\sum\nolimits_{r=o}^{1}}
rP\left(  \chi_{\omega_{i}}=r\right) \\
&  =P\left(  \operatorname{sign}\left(  \widehat{\Lambda}_{\mathfrak{B}_{ij}%
}\left(  \mathbf{x}\right)  \right)  =1\right)  \text{.}%
\end{align*}

It follows that, for any given $M$-class feature space $\left\{  \omega
_{i}\right\}  _{i=1}^{M}$, an ensemble of $M-1$ discriminant functions $%
{\displaystyle\sum\nolimits_{j=1}^{M-1}}
\widehat{\Lambda}_{\mathfrak{B}_{ij}}\left(  \mathbf{x}\right)  $, for which
the discriminant function $\widehat{\Lambda}_{\mathfrak{B}_{ij}}\left(
\mathbf{x}\right)  $ is an indicator function $\chi_{\omega_{i}}$ for class
$\omega_{i}$, provides $M-1$ characteristic functions $\chi_{\omega_{i}}$ for
feature vectors $\mathbf{x}$ that belong to class $\omega_{i}$:%
\begin{align*}
E\left[  \chi_{\omega_{i}}\right]   &  =%
{\displaystyle\sum\nolimits_{j=1}^{M-1}}
{\displaystyle\sum\nolimits_{r=o}^{1}}
rP\left(  \operatorname{sign}\left(  \widehat{\Lambda}_{\mathfrak{B}_{ij}%
}\left(  \mathbf{x}\right)  \right)  =1\right) \\
&  =%
{\displaystyle\sum\nolimits_{j=1}^{M-1}}
P\left(  \operatorname{sign}\left(  \widehat{\Lambda}_{\mathfrak{B}_{ij}%
}\left(  \mathbf{x}\right)  \right)  =1\right) \\
&  =%
{\displaystyle\sum\nolimits_{j=1}^{M-1}}
\operatorname{sign}\left(  \widehat{\Lambda}_{\mathfrak{B}_{ij}}\left(
\mathbf{x}\right)  \right)  =1\text{,}%
\end{align*}
where the probability $E\left[  \chi_{\omega_{i}}\right]  $ that an unseen
feature vector $\mathbf{x}$ belongs to class $\omega_{i}$ satisfies a
fundamental integral equation of binary classification for a classification
system in statistical equilibrium:%
\begin{align*}
\mathfrak{R}_{\mathfrak{B}}\left(  Z|\widehat{\Lambda}_{\mathfrak{B}_{ij}%
}\left(  \mathbf{x}\right)  \right)   &  =\int_{Z_{1}}p\left(
\widehat{\Lambda}_{\mathfrak{B}_{ij}}\left(  \mathbf{x}\right)  |\omega
_{2}\right)  d\widehat{\Lambda}_{\mathfrak{B}_{ij}}+\int_{Z_{2}}p\left(
\widehat{\Lambda}_{\mathfrak{B}_{ij}}\left(  \mathbf{x}\right)  |\omega
_{2}\right)  d\widehat{\Lambda}_{\mathfrak{B}_{ij}}+\nabla_{eq}\\
&  =\int_{Z_{2}}p\left(  \widehat{\Lambda}_{\mathfrak{B}_{ij}}\left(
\mathbf{x}\right)  |\omega_{1}\right)  d\widehat{\Lambda}_{\mathfrak{B}_{ij}%
}+\int_{Z_{1}}p\left(  \widehat{\Lambda}_{\mathfrak{B}_{ij}}\left(
\mathbf{x}\right)  |\omega_{1}\right)  d\widehat{\Lambda}_{\mathfrak{B}_{ij}%
}-\nabla_{eq}\text{,}%
\end{align*}
over the decision regions $Z_{1}$ and $Z_{2}$, where $\nabla_{eq}$ is an
equalizer statistic, for each discriminant function $\widehat{\Lambda
}_{\mathfrak{B}_{ij}}\left(  \mathbf{x}\right)  $ in the ensemble $%
{\displaystyle\sum\nolimits_{j=1}^{M-1}}
\widehat{\Lambda}_{\mathfrak{B}_{ij}}\left(  \mathbf{x}\right)  $.

Thus, $E\left[  \chi_{\omega_{i}}\right]  $ is determined by an ensemble of
$M-1$ discriminant functions $%
{\displaystyle\sum\nolimits_{j=1}^{M-1}}
\widehat{\Lambda}_{\mathfrak{B}_{ij}}\left(  \mathbf{x}\right)  $, where the
output of each characteristic function $\chi_{\omega_{i}}$ is determined by
the expected risk $\mathfrak{R}_{\mathfrak{\min}}\left(  \omega_{i}%
|Z_{2}\right)  $ for class $\omega_{i}$, given the discriminant function
$\widehat{\Lambda}_{\mathfrak{B}_{ij}}\left(  \mathbf{x}\right)  $ and the
decision region $Z_{2}$.

I will now show that the class of linear eigenlocus decision rules are
scalable modules for optimal linear classification systems.

\subsection{Linear Eigenlocus Decision Rules}

\begin{figure}[ptb]%
\centering
\fbox{\includegraphics[
height=2.5875in,
width=3.4411in
]%
{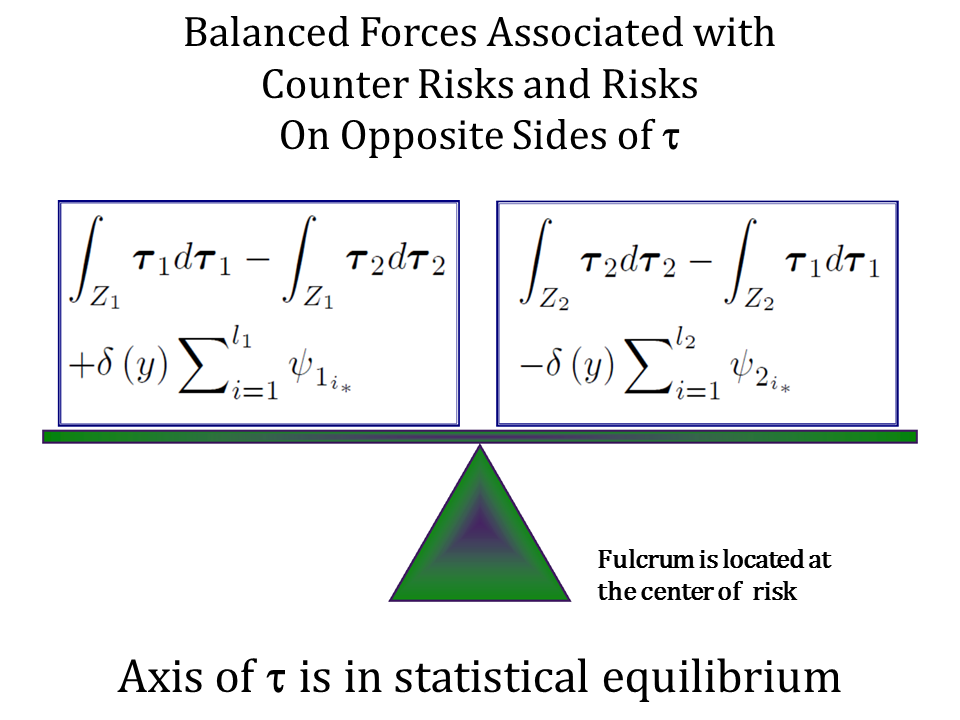}%
}\caption{Linear eigenlocus classification systems $\boldsymbol{\tau}%
^{T}\mathbf{x}+\tau_{0}\protect\overset{\omega_{1}}{\protect\underset{\omega
_{2}}{\gtrless}}0$ are in statistical equilibrium because the axis of a linear
eigenlocus $\boldsymbol{\tau}$ is in statistical equilibrium.}%
\label{Symetrically Balanced Eigenaxis of Linear Eigenlocus}%
\end{figure}

I\ have devised a system of data-driven, locus equations which determines
unknown, linear discriminant functions%
\[
\widetilde{\Lambda}_{\boldsymbol{\tau}}\left(  \mathbf{x}\right)
=\boldsymbol{\tau}^{T}\mathbf{x}+\tau_{0}%
\]
that are the basis of likelihood ratio tests%
\[
\widetilde{\Lambda}_{\boldsymbol{\tau}}\left(  \mathbf{x}\right)
=\boldsymbol{\tau}^{T}\mathbf{x}+\tau_{0}\overset{\omega_{1}}{\underset{\omega
_{2}}{\gtrless}}0
\]
which generate linear decision boundaries that satisfy a fundamental integral
equation of binary classification for a classification system in statistical
equilibrium (see Fig.
$\ref{Symetrically Balanced Eigenaxis of Linear Eigenlocus}$), whereby
two-class feature spaces are divided into congruent decision regions such that
for all data distributions, the forces associated with counter risks and
risks, within each of the congruent decision regions, are balanced with each
other, and for data distributions that have similar covariance matrices, the
forces associated with counter risks within each of the congruent decision
regions are equal to each other, and the forces associated with risks within
each of the congruent decision regions are equal to each other.%

I have demonstrated that a linear eigenlocus discriminant function
$\widetilde{\Lambda}_{\boldsymbol{\tau}}\left(  \mathbf{x}\right)
=\boldsymbol{\tau}^{T}\mathbf{x}+\tau_{0}$ is the solution to the integral
equation:%
\begin{align*}
f\left(  \widetilde{\Lambda}_{\boldsymbol{\tau}}\left(  \mathbf{x}\right)
\right)  =  &  \int_{Z_{1}}\boldsymbol{\tau}_{1}d\boldsymbol{\tau}_{1}%
+\int_{Z_{2}}\boldsymbol{\tau}_{1}d\boldsymbol{\tau}_{1}+\delta\left(
y\right)  \sum\nolimits_{i=1}^{l_{1}}\psi_{1_{i_{\ast}}}\\
&  =\int_{Z_{1}}\boldsymbol{\tau}_{2}d\boldsymbol{\tau}_{2}+\int_{Z_{2}%
}\boldsymbol{\tau}_{2}d\boldsymbol{\tau}_{2}-\delta\left(  y\right)
\sum\nolimits_{i=1}^{l_{2}}\psi_{2_{i_{\ast}}}\text{,}%
\end{align*}
over the decision space $Z=Z_{1}+Z_{2}$, where $\delta\left(  y\right)
\triangleq\sum\nolimits_{i=1}^{l}y_{i}\left(  1-\xi_{i}\right)  $, such that
the expected risk $\mathfrak{R}_{\mathfrak{\min}}\left(  Z|\widehat{\Lambda
}_{\boldsymbol{\tau}}\left(  \mathbf{x}\right)  \right)  $ and the eigenenergy
$E_{\min}\left(  Z|\widehat{\Lambda}_{\boldsymbol{\tau}}\left(  \mathbf{x}%
\right)  \right)  $ of the classification system $\boldsymbol{\tau}%
^{T}\mathbf{x}+\tau_{0}\overset{\omega_{1}}{\underset{\omega_{2}}{\gtrless}}0$
are minimized for data drawn from statistical distributions that have similar
covariance matrices.

Thereby, a linear eigenlocus classification system $\boldsymbol{\tau}%
^{T}\mathbf{x}+\tau_{0}\overset{\omega_{1}}{\underset{\omega_{2}}{\gtrless}}0$
generates the locus of a linear decision boundary for any two classes of
feature vectors, including completely overlapping data distributions. For data
distributions that have constant or unchanging statistics and similar
covariance matrices, linear eigenlocus classification systems
$\boldsymbol{\tau}^{T}\mathbf{x}+\tau_{0}\overset{\omega_{1}}{\underset{\omega
_{2}}{\gtrless}}0$ exhibit optimal generalization performance, where the
generalization error is the lowest possible decision error.

I\ will now argue that linear eigenlocus decision rules $\widetilde{\Lambda
}_{\boldsymbol{\tau}}\left(  \mathbf{x}\right)  \overset{\omega_{1}%
}{\underset{\omega_{2}}{\gtrless}}0$ are scalable, individual components of
superior or optimal ensemble systems, where any given ensemble of linear
eigenlocus decision rules exhibits superior or optimal generalization
performance for its $M$-class feature space.

\subsection{Ensemble Systems of Eigenlocus Decision Rules I}

Because linear eigenlocus decision rules involve linear combinations of
extreme points and scaled extreme points:%
\begin{align*}
\widetilde{\Lambda}_{\boldsymbol{\tau}}\left(  \mathbf{x}\right)   &  =\left(
\mathbf{x}-\sum\nolimits_{i=1}^{l}\mathbf{x}_{i\ast}\right)  ^{T}\left[
\sum\nolimits_{i=1}^{l_{1}}\psi_{1_{i\ast}}\mathbf{x}_{1_{i\ast}}%
-\sum\nolimits_{i=1}^{l_{2}}\psi_{2_{i\ast}}\mathbf{x}_{2_{i\ast}}\right] \\
&  \mathbf{+}\sum\nolimits_{i=1}^{l}y_{i}\left(  1-\xi_{i}\right)
\overset{\omega_{1}}{\underset{\omega_{2}}{\gtrless}}0\text{,}%
\end{align*}
it follows that linear combinations of linear eigenlocus discriminant
functions can be used to build optimal pattern recognition systems, where the
\emph{overall system complexity is scale-invariant} for the feature space
dimension and the number of pattern classes. Thus, linear eigenlocus decision
rules $\widetilde{\Lambda}_{\boldsymbol{\tau}}\left(  \mathbf{x}\right)
\overset{\omega_{1}}{\underset{\omega_{2}}{\gtrless}}0$ are scalable modules
for optimal linear classification systems. Denote an optimal pattern
recognition system by $\mathfrak{P}_{\boldsymbol{o}}\left(  \mathbf{x}\right)
$. I\ will now outline an architecture for an ensemble system of linear
eigenlocus decision rules.

Given that a linear eigenlocus discriminant function $\widetilde{\Lambda
}_{\boldsymbol{\tau}}\left(  \mathbf{x}\right)  =\boldsymbol{\tau}%
^{T}\mathbf{x}+\tau_{0}$ is an indicator\emph{\ }function $\chi_{\omega_{i}}$
for any given class of feature vectors $\mathcal{\omega}_{i}$ that have the
training label $+1$, it follows that the decision function
$\operatorname{sign}\left(  \widetilde{\Lambda}_{\boldsymbol{\tau}}\left(
\mathbf{x}\right)  \right)  $%
\[
\operatorname{sign}\left(  \widetilde{\Lambda}_{\boldsymbol{\tau}}\left(
\mathbf{x}\right)  \right)  =\operatorname{sign}\left(  \boldsymbol{\tau}%
^{T}\mathbf{x}+\tau_{0}\right)  \text{,}%
\]
where $\operatorname{sign}\left(  x\right)  \equiv\frac{x}{\left\vert
x\right\vert }$ for $x\neq0$, provides a natural means for discriminating
between multiple classes of data, where decisions can be made that are based
on the \emph{largest probabilistic output} of decision banks $\mathcal{DB}%
_{\mathcal{\omega}_{i}}\left(  \mathbf{x}\right)  $ formed by linear
combinations of linear eigenlocus decision functions $\operatorname{sign}%
\left(  \widetilde{\Lambda}_{\boldsymbol{\tau}}\left(  \mathbf{x}\right)
\right)  $:%
\[
\mathcal{DB}_{\omega_{i}}\left(  \mathbf{x}\right)  =%
{\textstyle\sum\nolimits_{j=1}^{M-1}}
\operatorname{sign}\left(  \widetilde{\Lambda}_{\boldsymbol{\tau}}\left(
\mathbf{x}\right)  \right)  \text{,}%
\]
where the decision bank $\mathcal{DB}_{\omega_{i}}\left(  \mathbf{x}\right)  $
for a pattern class $\omega_{i}$ is an ensemble%
\[%
{\textstyle\sum\nolimits_{j=1}^{M-1}}
\operatorname{sign}\left(  \widetilde{\Lambda}_{\boldsymbol{\tau}}\left(
\mathbf{x}\right)  \right)
\]
of $M-1$ decision functions $\left\{  \operatorname{sign}\left(
\widetilde{\Lambda}_{\boldsymbol{\tau}}\left(  \mathbf{x}\right)  \right)
\right\}  _{j=1}^{M-1}$ for which the pattern vectors in the given class
$\mathcal{\omega}_{i}$ have the training label $+1$, and the pattern vectors
in all of the other pattern classes have the training label $-1$, where the
probabilistic output of any given decision bank $\mathcal{DB}_{\omega_{i}%
}\left(  \mathbf{x}\right)  $ is given by a set of $M-1$ characteristic
functions:%
\begin{align*}
E\left[  \chi_{\omega_{i}}\right]   &  =%
{\displaystyle\sum\nolimits_{j=1}^{M-1}}
P\left(  \operatorname{sign}\left(  \widehat{\Lambda}_{\mathfrak{B}_{ij}%
}\left(  \mathbf{x}\right)  \right)  =1\right) \\
&  =%
{\displaystyle\sum\nolimits_{j=1}^{M-1}}
\operatorname{sign}\left(  \widehat{\Lambda}_{\mathfrak{B}_{ij}}\left(
\mathbf{x}\right)  \right)  =1\text{.}%
\end{align*}

\begin{figure}[ptb]%
\centering
\fbox{\includegraphics[
height=2.5875in,
width=3.4411in
]%
{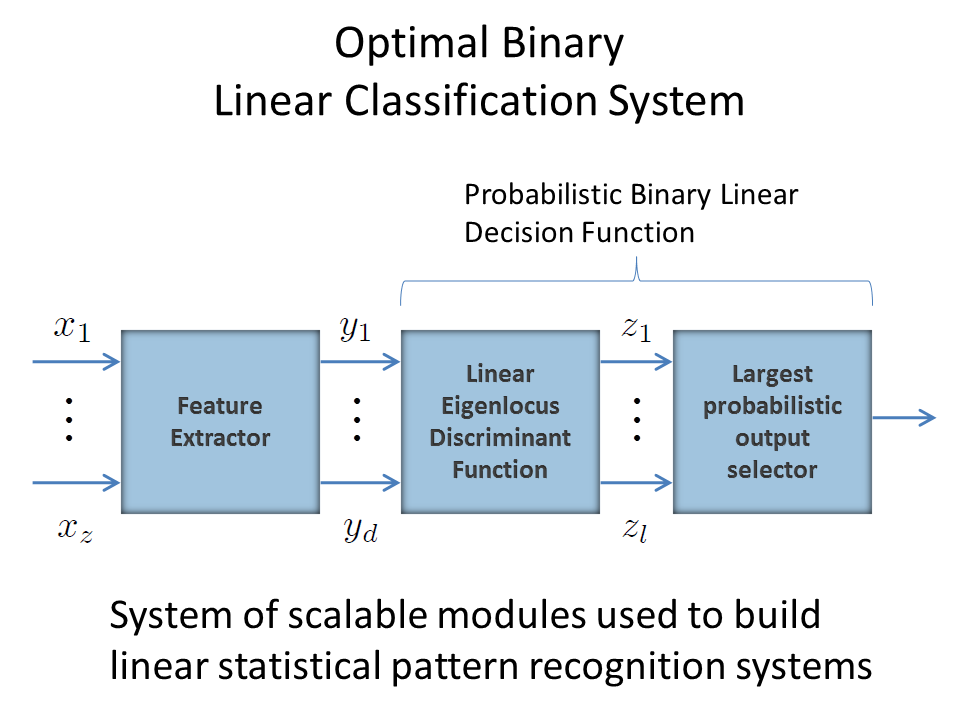}%
}\caption{Illustration of a system of scalable modules used to build optimal
statistical pattern recognition machines. The system includes a feature
extractor, a linear eigenlocus discriminant function
$\protect\widetilde{\Lambda}_{\boldsymbol{\tau}}\left(  \mathbf{x}\right)
=\boldsymbol{\tau}^{T}\mathbf{x}+\tau_{0}$ and a decision function
$\operatorname{sign}\left(  \boldsymbol{\tau}^{T}\mathbf{x}+\tau_{0}\right)
$.}%
\label{System of Scalable Modules Linear Eigenlocus}%
\end{figure}

Decision banks $\mathcal{DB}_{\omega_{i}}\left(  \mathbf{x}\right)  $ are
formed by the system of scalable modules depicted in Fig.
$\ref{System of Scalable Modules Linear Eigenlocus}$, where linear
combinations of optimal binary linear classification systems can be used to
build an optimal statistical pattern recognition system $\mathfrak{P}%
_{\boldsymbol{o}}\left(  \mathbf{x}\right)  $ which distinguishes between the
objects in $M$ different pattern classes: $\left\{  \mathcal{\omega}%
_{i}\right\}  _{i=1}^{M}$. Objects in pattern classes can involve any type of
distinguishing features that have been extracted from collections of: $\left(
1\right)  $ networks formed by interconnected systems of people and/or things:
material or biological, $\left(  2\right)  $ documents, $\left(  3\right)  $
images, or $\left(  4\right)  $ waveforms, signals, or sequences, including
stationary random processes.%

I will devise an optimal, linear statistical pattern recognition system
$\mathfrak{P}_{\boldsymbol{o}}\left(  \mathbf{x}\right)  $ formed by $M$
decision banks $\left\{  \mathcal{DB}_{\omega_{i}}\left(  \mathbf{x}\right)
\right\}  _{i=1}^{M}$ of linear eigenlocus decision functions
$\operatorname{sign}\left(  \widetilde{\Lambda}_{\boldsymbol{\tau}}\left(
\mathbf{x}\right)  \right)  $:%
\begin{equation}
\mathfrak{P}_{\boldsymbol{o}}\left(  \mathbf{x}\right)  =\left\{
\mathcal{DB}_{\omega_{i}}\left(
{\textstyle\sum\nolimits_{j=1}^{M-1}}
\operatorname{sign}\left(  \widetilde{\Lambda}_{\boldsymbol{\tau}}\left(
\mathbf{x}\right)  \right)  \right)  \right\}  _{i=1}^{M}\text{,}
\label{Ensemble of Linear Eigenlocus Decision Functions}%
\end{equation}
where%
\[
\operatorname{sign}\left(  \widetilde{\Lambda}_{\boldsymbol{\tau}}\left(
\mathbf{x}\right)  \right)  =\operatorname{sign}\left(  \boldsymbol{\tau}%
^{T}\mathbf{x}+\tau_{0}\right)  \text{,}%
\]
that provides a set of $M\times(M-1)$ decision statistics%
\[
\left\{  \operatorname{sign}\left(  \widetilde{\Lambda}_{\boldsymbol{\tau}%
}\left(  \mathbf{x}\right)  \right)  \right\}  _{j=1}^{_{M\times(M-1)}}%
\]
for $M$ pattern classes $\left\{  \mathcal{\omega}_{i}\right\}  _{i=1}^{M}$,
where each decision statistic is a characteristic function $\chi_{\omega_{i}%
}\mapsto P\left(  \operatorname{sign}\left(  \widehat{\Lambda}_{\mathfrak{B}%
_{ij}}\left(  \mathbf{x}\right)  \right)  =1\right)  $ that is determined by
an optimal likelihood ratio test for a two-class feature space:%
\[
\widetilde{\Lambda}_{\boldsymbol{\tau}}\left(  \mathbf{x}\right)
\overset{\omega_{1}}{\underset{\omega_{2}}{\gtrless}}0\text{,}%
\]
such that the maximum value selector of the pattern recognition system
$\mathfrak{P}_{\boldsymbol{o}}\left(  \mathbf{x}\right)  $ chooses the pattern
class $\mathcal{\omega}_{i}$ for which a decision bank $\mathcal{DB}%
_{\omega_{i}}\left(  \mathbf{x}\right)  $ has the maximum probabilistic
output:%
\[
D_{\mathfrak{B}}\left(  \mathbf{x}\right)  \underset{i\in1,\cdots
,M}{=ArgMax}\left(  \mathcal{DB}_{\omega_{i}}\left(  \mathbf{x}\right)
\right)  \text{,}%
\]
where the probabilistic output of each decision bank $\mathcal{DB}_{\omega
_{i}}\left(  \mathbf{x}\right)  $ is given by a set of $M-1$ characteristic
functions:%
\begin{align*}
E\left[  \chi_{\omega_{i}}\right]   &  =%
{\displaystyle\sum\nolimits_{j=1}^{M-1}}
P\left(  \operatorname{sign}\left(  \widehat{\Lambda}_{\mathfrak{B}_{ij}%
}\left(  \mathbf{x}\right)  \right)  =1\right) \\
&  =%
{\displaystyle\sum\nolimits_{j=1}^{M-1}}
\operatorname{sign}\left(  \widehat{\Lambda}_{\mathfrak{B}_{ij}}\left(
\mathbf{x}\right)  \right)  =1\text{.}%
\end{align*}

For data distributions that have similar covariance matrices, I\ will now show
that the ensemble system of linear eigenlocus decision functions in Eq.
(\ref{Ensemble of Linear Eigenlocus Decision Functions}) generates a set of
linear decision boundaries and decision statistics that minimize the
probability of misclassification or decision error for an $M$-class feature space.

\subsection{Expected Risk for Eigenlocus Ensemble Systems I}

Take any given $M$-class feature space, where all $M$ data distributions have
similar covariance matrices. Now take any given ensemble system $\mathfrak{P}%
_{\boldsymbol{o}}\left(  \mathbf{x}\right)  $ formed by $M$ decision banks
$\left\{  \mathcal{DB}_{\omega_{i}}\left(  \mathbf{x}\right)  \right\}
_{i=1}^{M}$ of linear eigenlocus decision functions $\operatorname{sign}%
\left(  \widetilde{\Lambda}_{\boldsymbol{\tau}}\left(  \mathbf{x}\right)
\right)  $:%
\[
\mathfrak{P}_{\boldsymbol{o}}\left(  \mathbf{x}\right)  =\left\{
\mathcal{DB}_{\omega_{i}}\left(
{\textstyle\sum\nolimits_{j=1}^{M-1}}
\operatorname{sign}\left(  \widetilde{\Lambda}_{\boldsymbol{\tau}}\left(
\mathbf{x}\right)  \right)  \right)  \right\}  _{i=1}^{M}\text{,}%
\]
where each likelihood ratio test $\widetilde{\Lambda}_{\boldsymbol{\tau}_{j}%
}\left(  \mathbf{x}\right)  \overset{\omega_{1}}{\underset{\omega
_{2}}{\gtrless}}0$ in an ensemble $%
{\textstyle\sum\nolimits_{j=1}^{M-1}}
\operatorname{sign}\left(  \widetilde{\Lambda}_{\boldsymbol{\tau}_{j}}\left(
\mathbf{x}\right)  \right)  $ minimizes the total probability of error and
achieves the minimum possible error rate for two given pattern classes.

Next, take the decision bank $\mathcal{DB}_{\omega_{i}}\left(  \mathbf{x}%
\right)  =%
{\textstyle\sum\nolimits_{j=1}^{M-1}}
\operatorname{sign}\left(  \widetilde{\Lambda}_{\boldsymbol{\tau}_{j}}\left(
\mathbf{x}\right)  \right)  $ for any given pattern class $\omega_{i}$, and
let $\mathfrak{R}_{\mathfrak{\min}_{ij}}\left(  Z_{2}|\widetilde{\Lambda
}_{\boldsymbol{\tau}_{j}}\right)  $ denote the expected risk for class
$\omega_{i}$ for any given likelihood ratio test $\widetilde{\Lambda
}_{\boldsymbol{\tau}_{j}}\left(  \mathbf{x}\right)  \overset{\omega
_{1}}{\underset{\omega_{2}}{\gtrless}}0$:%
\[
\mathfrak{R}_{\mathfrak{\min}ij}\left(  Z_{2}|\widetilde{\Lambda
}_{\boldsymbol{\tau}_{j}}\right)  =\int_{Z_{2}}p\left(  \widetilde{\Lambda
}_{\boldsymbol{\tau}_{j}}\left(  \mathbf{x}\right)  |\omega_{i}\right)
d\widetilde{\Lambda}_{\boldsymbol{\tau}_{j}}%
\]
which determines the conditional probability of classification error for class
$\omega_{i}$ in a two-class decision space, where $Z=Z_{1}+Z_{2}$. It follows
that each classification system $\widetilde{\Lambda}_{\boldsymbol{\tau}_{j}%
}\left(  \mathbf{x}\right)  \overset{\omega_{1}}{\underset{\omega
_{2}}{\gtrless}}0$ for class $\omega_{i}$ minimizes the expected risk
$\mathfrak{R}_{\mathfrak{\min}ij}\left(  Z_{2}|\widetilde{\Lambda
}_{\boldsymbol{\tau}_{j}}\right)  $ in a two-class decision space, where
$Z=Z_{1}+Z_{2}$.

So, let $\mathfrak{R}_{\mathfrak{\min}}\left(  \omega_{i}|\overline{Z}%
_{i}\right)  $ denote the expected risk for class $\omega_{i}$ for any given
ensemble of linear eigenlocus decision rules:%
\[
\mathcal{DB}_{\omega_{i}}\left(  \mathbf{x}\right)  =%
{\displaystyle\sum\nolimits_{j=1}^{M-1}}
\left[  \widetilde{\Lambda}_{\boldsymbol{\tau}_{j}}\left(  \mathbf{x}\right)
\overset{\omega_{1}}{\underset{\omega_{2}}{\gtrless}}0\right]
\]
which determines the conditional probability of classification error for class
$\omega_{i}$ in $M-1$ two-class decision spaces $\left\{  Z_{j}\right\}
_{i=1}^{M-1}$, where the decision space $\overline{Z}_{i}$ for the ensemble of
linear classifiers is the union of the decision spaces $Z_{1},Z_{2}%
,\ldots,Z_{M-1}$:%
\[
\overline{Z}_{i}=%
{\displaystyle\sum\nolimits_{j=1}^{M-1}}
Z_{j}=\cup_{j=1}^{M-1}\;Z_{j}\text{.}%
\]

Therefore, given an ensemble of linear eigenlocus decision rules
$\mathcal{DB}_{\omega_{i}}\left(  \mathbf{x}\right)  $, it follows that the
expected risk $\mathfrak{R}_{\mathfrak{\min}}\left(  \omega_{i}|\overline
{Z}_{i}\right)  $ for class $\omega_{i}$ in an $M$-class feature space is
determined by%
\begin{align*}
\mathfrak{R}_{\mathfrak{\min}}\left(  \omega_{i}|\overline{Z}_{i}\right)   &
=%
{\displaystyle\sum\nolimits_{j=1}^{M-1}}
\mathfrak{R}_{\mathfrak{\min}_{ij}}\left(  Z_{2}|\widetilde{\Lambda
}_{\boldsymbol{\tau}_{j}}\left(  \mathbf{x}\right)  \right) \\
&  =%
{\displaystyle\sum\nolimits_{j=1}^{M-1}}
\int_{Z_{2}}p\left(  \widetilde{\Lambda}_{\boldsymbol{\tau}_{j}}\left(
\mathbf{x}\right)  |\omega_{i}\right)  d\widetilde{\Lambda}_{\boldsymbol{\tau
}_{j}}\text{,}%
\end{align*}
where minimization of the expected risk $\mathfrak{R}_{\mathfrak{\min}}\left(
\omega_{i}|\overline{Z}_{i}\right)  $ is equivalent to minimizing the total
probability of decision error for class $\omega_{i}$ in the $M$-class decision
space $\overline{Z}_{i}$.

Finally, let $\widehat{Z}=\cup_{i=1}^{M}\;\overline{Z}_{i}$ denote the
decision space that is determined by the ensemble system%
\[
\mathfrak{P}_{\boldsymbol{o}}\left(  \mathbf{x}\right)  =\left\{
\mathcal{DB}_{\omega_{i}}\left(
{\textstyle\sum\nolimits_{j=1}^{M-1}}
\operatorname{sign}\left(  \widetilde{\Lambda}_{\boldsymbol{\tau}_{j}}\left(
\mathbf{x}\right)  \right)  \right)  \right\}  _{i=1}^{M}%
\]
of $M$ decision banks $\left\{  \mathcal{DB}_{\omega_{i}}\left(
\mathbf{x}\right)  \right\}  _{i=1}^{M}$. It follows that the expected risk
$\mathfrak{R}_{\mathfrak{\min}}\left(  \widehat{Z}\right)  $ for all $M$
pattern classes $\left\{  \omega_{i}\right\}  _{i=1}^{M}$ is:%
\begin{align*}
\mathfrak{R}_{\min}\left(  \widehat{Z}\right)   &  =%
{\displaystyle\sum\nolimits_{i=1}^{M}}
\mathfrak{R}_{\mathfrak{\min}}\left(  \omega_{i}|\overline{Z}_{i}\right)  =%
{\displaystyle\sum\nolimits_{i=1}^{M}}
{\displaystyle\sum\nolimits_{j=1}^{M-1}}
\mathfrak{R}_{\mathfrak{\min}ij}\left(  Z_{2}|\widetilde{\Lambda
}_{\boldsymbol{\tau}_{j}}\right) \\
&  =%
{\displaystyle\sum\nolimits_{i=1}^{M}}
{\displaystyle\sum\nolimits_{j=1}^{M-1}}
\int_{Z_{2}}p\left(  \widetilde{\Lambda}_{\boldsymbol{\tau}_{j}}\left(
\mathbf{x}\right)  |\omega_{i}\right)  d\widetilde{\Lambda}_{\boldsymbol{\tau
}_{j}}\text{,}%
\end{align*}
where the expected risk $\mathfrak{R}_{\mathfrak{\min}}\left(  \omega
_{i}|\overline{Z}_{i}\right)  $ for each pattern class $\omega_{i}$ is
determined by an ensemble of linear classifiers $\mathcal{DB}_{\omega_{i}%
}\left(  \mathbf{x}\right)  $ that minimize the total probability of error for
class $\omega_{i}$ in the $M$-class decision space $\widehat{Z}$.

Thus, it is concluded that the ensemble system $\mathfrak{P}_{\boldsymbol{o}%
}\left(  \mathbf{x}\right)  $ in Eq.
(\ref{Ensemble of Linear Eigenlocus Decision Functions}) generates a set of
linear decision boundaries and decision statistics that minimize the
probability of decision error for an $M$-class feature space, where all $M$
data distributions have similar covariance matrices.

\subsection{Quadratic Eigenlocus Decision Rules}

\begin{figure}[ptb]%
\centering
\fbox{\includegraphics[
height=2.5875in,
width=3.4411in
]%
{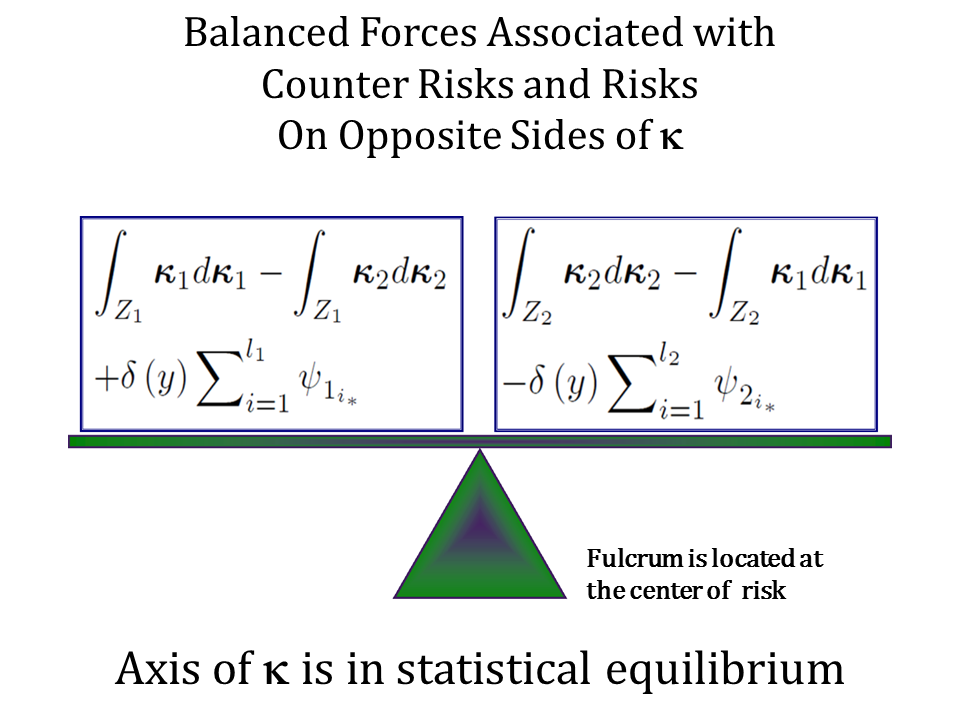}%
}\caption{Quadratic eigenlocus classification systems $\boldsymbol{\kappa
}k_{\mathbf{s}}+\kappa_{0}\protect\overset{\omega_{1}%
}{\protect\underset{\omega_{2}}{\gtrless}}0$ are in statistical equilibrium
because the axis of a quadratic eigenlocus $\boldsymbol{\kappa}$ is in
statistical equilibrium.}%
\label{Symetrically Balanced Eigenaxis of Quadratic Eigenlocus}%
\end{figure}

I\ have devised a system of data-driven, locus equations which determines
unknown, quadratic discriminant functions%
\[
\widetilde{\Lambda}_{\boldsymbol{\kappa}}\left(  \mathbf{s}\right)  =\left(
\mathbf{x}^{T}\mathbf{s}+1\right)  ^{2}\boldsymbol{\kappa}+\kappa_{0}\text{,}%
\]
based on second-order, polynomial reproducing kernels $\left(  \mathbf{x}%
^{T}\mathbf{s}+1\right)  ^{2}$, that are the basis of likelihood ratio tests%
\[
\widetilde{\Lambda}_{\boldsymbol{\kappa}}\left(  \mathbf{s}\right)  =\left(
\mathbf{x}^{T}\mathbf{s}+1\right)  ^{2}\boldsymbol{\kappa}+\kappa
_{0}\overset{\omega_{1}}{\underset{\omega_{2}}{\gtrless}}0
\]
which generate quadratic decision boundaries that satisfy a fundamental
integral equation of binary classification for a classification system in
statistical equilibrium (see Fig.
$\ref{Symetrically Balanced Eigenaxis of Quadratic Eigenlocus}$), whereby
two-class feature spaces are divided into symmetrical decision regions such
that for data distributions that have dissimilar covariance matrices, the
forces associated with the counter risks and the risks, within each of the
symmetrical decision regions, are balanced with each other, and for data
distributions that have similar covariance matrices, the forces associated
with the counter risks within each of the symmetrical decision regions are
equal to each other, and the forces associated with the risks within each of
the symmetrical decision regions are equal to each other. The system of
data-driven, locus equations is readily extended for Gaussian reproducing
kernels $\exp\left(  -0.01\left\Vert \mathbf{x}-\mathbf{s}\right\Vert
^{2}\right)  $, where unknown, quadratic discriminant functions are defined
by:%
\[
\widetilde{\Lambda}_{\boldsymbol{\kappa}}\left(  \mathbf{s}\right)
=\exp\left(  -0.01\left\Vert \mathbf{x}-\mathbf{s}\right\Vert ^{2}\right)
\boldsymbol{\kappa}+\kappa_{0}\text{.}%
\]%

I have demonstrated that a quadratic eigenlocus discriminant function
$\widetilde{\Lambda}_{\boldsymbol{\kappa}}\left(  \mathbf{s}\right)
=\boldsymbol{\kappa}^{T}k_{\mathbf{s}}+\kappa_{0}$, where the reproducing
kernel $k_{\mathbf{s}}$ for the data point $\mathbf{s}$ is either%
\[
k_{\mathbf{s}}=\left(  \mathbf{x}^{T}\mathbf{s}+1\right)  ^{2}\text{ \ or
\ }k_{\mathbf{s}}=\exp\left(  -0.01\left\Vert \mathbf{x}-\mathbf{s}\right\Vert
^{2}\right)  \text{,}%
\]
is the solution to the integral equation:%
\begin{align*}
f\left(  \widetilde{\Lambda}_{\boldsymbol{\kappa}}\left(  \mathbf{s}\right)
\right)  =  &  \int_{Z_{1}}\boldsymbol{\kappa}_{1}d\boldsymbol{\kappa}%
_{1}+\int_{Z_{2}}\boldsymbol{\kappa}_{1}d\boldsymbol{\kappa}_{1}+\delta\left(
y\right)  \sum\nolimits_{i=1}^{l_{1}}\psi_{1_{i_{\ast}}}\\
&  =\int_{Z_{1}}\boldsymbol{\kappa}_{2}d\boldsymbol{\kappa}_{2}+\int_{Z_{2}%
}\boldsymbol{\kappa}_{2}d\boldsymbol{\kappa}_{2}-\delta\left(  y\right)
\sum\nolimits_{i=1}^{l_{2}}\psi_{2_{i_{\ast}}}\text{,}%
\end{align*}
over the decision space $Z=Z_{1}+Z_{2}$, where $\delta\left(  y\right)
\triangleq\sum\nolimits_{i=1}^{l}y_{i}\left(  1-\xi_{i}\right)  $, such that
the expected risk $\mathfrak{R}_{\mathfrak{\min}}\left(  Z|\widehat{\Lambda
}_{\boldsymbol{\kappa}}\left(  \mathbf{s}\right)  \right)  $ and the
eigenenergy $E_{\min}\left(  Z|\widehat{\Lambda}_{\boldsymbol{\kappa}}\left(
\mathbf{s}\right)  \right)  $ of the classification system $\boldsymbol{\kappa
}^{T}k_{\mathbf{s}}+\kappa_{0}\overset{\omega_{1}}{\underset{\omega
_{2}}{\gtrless}}0$ are minimized.

Thereby, a quadratic eigenlocus classification system $\boldsymbol{\kappa}%
^{T}k_{\mathbf{s}}+\kappa_{0}\overset{\omega_{1}}{\underset{\omega
_{2}}{\gtrless}}0$ generates the locus of a quadratic decision boundary for
any two classes of feature vectors, including completely overlapping data
distributions. For any two classes of feature vectors $\mathbf{x}$ that have
common covariance matrices, quadratic eigenlocus classification systems
$\boldsymbol{\kappa}^{T}k_{\mathbf{s}}+\kappa_{0}\overset{\omega
_{1}}{\underset{\omega_{2}}{\gtrless}}0$ generate the locus of a quadratic
decision boundary that provides a robust estimate of a linear decision boundary.

Thus, for any given data distributions that have constant or unchanging
statistics and similar or dissimilar covariance matrices, quadratic eigenlocus
classification systems $\boldsymbol{\kappa}^{T}k_{\mathbf{s}}+\kappa
_{0}\overset{\omega_{1}}{\underset{\omega_{2}}{\gtrless}}0$ exhibit optimal
generalization performance, where the generalization error is the lowest
possible decision error. Therefore, quadratic eigenlocus classification
systems $\boldsymbol{\kappa}^{T}k_{\mathbf{s}}+\kappa_{0}\overset{\omega
_{1}}{\underset{\omega_{2}}{\gtrless}}0$ \emph{automatically} generate the
\emph{best decision boundary} for a wide variety of statistical pattern
recognition tasks. Accordingly, any given quadratic eigenlocus classification
system $\kappa^{T}k_{\mathbf{s}}+\kappa_{0}\overset{\omega_{1}%
}{\underset{\omega_{2}}{\gtrless}}0$ achieves the lowest error rate that can
be achieved by a discriminant function and the best generalization error that
can be achieved by a learning machine.

I\ will now argue that quadratic eigenlocus decision rules $\widetilde{\Lambda
}_{\boldsymbol{\kappa}}\left(  \mathbf{s}\right)  \overset{\omega
_{1}}{\underset{\omega_{2}}{\gtrless}}0$ are scalable, individual components
of optimal ensemble systems, where any given ensemble of quadratic eigenlocus
decision rules exhibits optimal generalization performance for its $M$-class
feature space. The argument is developed for the class of quadratic eigenlocus
decision rules defined by:%
\[
\widetilde{\Lambda}_{\boldsymbol{\kappa}}\left(  \mathbf{s}\right)  =\left(
\mathbf{x}^{T}\mathbf{s}+1\right)  ^{2}\boldsymbol{\kappa}+\kappa_{0}%
\]
and is applicable to the class of quadratic eigenlocus decision rules defined
by:%
\[
\widetilde{\Lambda}_{\boldsymbol{\kappa}}\left(  \mathbf{s}\right)
=\exp\left(  -0.01\left\Vert \mathbf{x}-\mathbf{s}\right\Vert ^{2}\right)
\boldsymbol{\kappa}+\kappa_{0}\text{.}%
\]

\subsection{Ensemble Systems of Eigenlocus Decision Rules II}

Because quadratic eigenlocus decision rules involve linear combinations of
extreme vectors, scaled reproducing kernels of extreme points, class
membership statistics, and regularization parameters%
\begin{align*}
\widetilde{\Lambda}_{\boldsymbol{\kappa}}\left(  \mathbf{s}\right)   &
=\left(  \left(  \mathbf{x}^{T}\mathbf{s}+1\right)  ^{2}-\sum\nolimits_{i=1}%
^{l}\left(  \mathbf{x}^{T}\mathbf{x}_{i\ast}+1\right)  ^{2}\right)
\boldsymbol{\kappa}_{1}\\
&  -\left(  \left(  \mathbf{x}^{T}\mathbf{s}+1\right)  ^{2}-\sum
\nolimits_{i=1}^{l}\left(  \mathbf{x}^{T}\mathbf{x}_{i\ast}+1\right)
^{2}\right)  \boldsymbol{\kappa}_{2}\\
&  \mathbf{+}\sum\nolimits_{i=1}^{l}y_{i}\left(  1-\xi_{i}\right)
\overset{\omega_{1}}{\underset{\omega_{2}}{\gtrless}}0\text{,}%
\end{align*}
where%
\[
\boldsymbol{\kappa}_{1}=\sum\nolimits_{i=1}^{l_{1}}\psi_{1_{i\ast}}\left(
\mathbf{x}^{T}\mathbf{x}_{1_{i\ast}}+1\right)  ^{2}%
\]
and%
\[
\boldsymbol{\kappa}_{2}=\sum\nolimits_{i=1}^{l_{2}}\psi_{2_{i\ast}}\left(
\mathbf{x}^{T}\mathbf{x}_{2_{i\ast}}+1\right)  ^{2}\text{,}%
\]
it follows that linear combinations of quadratic eigenlocus discriminant
functions can be used to build optimal statistical pattern recognition systems
$P_{\mathfrak{B}}\left(  \mathbf{s}\right)  $, where the \emph{overall system
complexity is scale-invariant} for the feature space dimension and the number
of pattern classes. Thus, quadratic eigenlocus decision rules
$\widetilde{\Lambda}_{\boldsymbol{\kappa}}\left(  \mathbf{s}\right)  $ are
scalable modules for optimal quadratic classification systems. I\ will now
outline an architecture for optimal ensemble systems of quadratic eigenlocus
decision rules.

Given that a quadratic eigenlocus discriminant function $\widetilde{\Lambda
}_{\boldsymbol{\kappa}}\left(  \mathbf{s}\right)  =\left(  \mathbf{x}%
^{T}\mathbf{s}+1\right)  ^{2}\boldsymbol{\kappa}+\kappa_{0}$ is an
indicator\emph{\ }function $\chi_{\omega_{i}}$ for any given class of feature
vectors $\mathcal{\omega}_{i}$ that have the training label $+1$, it follows
that the decision function $\operatorname{sign}\left(  \widetilde{\Lambda
}_{\boldsymbol{\kappa}}\left(  \mathbf{s}\right)  \right)  $%
\[
\operatorname{sign}\left(  \widetilde{\Lambda}_{\boldsymbol{\kappa}}\left(
\mathbf{s}\right)  \right)  =\operatorname{sign}\left(  \left(  \mathbf{x}%
^{T}\mathbf{s}+1\right)  ^{2}\boldsymbol{\kappa}+\kappa_{0}\right)  \text{,}%
\]
where $\operatorname{sign}\left(  x\right)  \equiv\frac{x}{\left\vert
x\right\vert }$ for $x\neq0$, provides a natural means for discriminating
between multiple classes of data, where decisions can be made that are based
on the largest probabilistic output of decision banks $\mathcal{DB}%
_{\omega_{i}}\left(  \mathbf{s}\right)  $ formed by linear combinations of
quadratic eigenlocus decision functions $\operatorname{sign}\left(
\widetilde{\Lambda}_{\boldsymbol{\kappa}}\left(  \mathbf{s}\right)  \right)
$:%
\[
\mathcal{DB}_{\omega_{i}}\left(  \mathbf{s}\right)  =%
{\textstyle\sum\nolimits_{j=1}^{M-1}}
\operatorname{sign}\left(  \widetilde{\Lambda}_{\boldsymbol{\kappa}_{j}%
}\left(  \mathbf{s}\right)  \right)  \text{,}%
\]
where the decision bank $\mathcal{DB}_{\omega_{i}}\left(  \mathbf{s}\right)  $
for a pattern class $\mathcal{\omega}_{i}$ is an ensemble%
\[%
{\textstyle\sum\nolimits_{j=1}^{M-1}}
\operatorname{sign}\left(  \widetilde{\Lambda}_{\boldsymbol{\kappa}_{j}%
}\left(  \mathbf{s}\right)  \right)
\]
of $M-1$ decision functions $\left\{  \operatorname{sign}\left(
\widetilde{\Lambda}_{\boldsymbol{\kappa}_{j}}\left(  \mathbf{s}\right)
\right)  \right\}  _{j=1}^{M-1}$ for which the pattern vectors in the given
class $\mathcal{\omega}_{i}$ have the training label $+1$, and the pattern
vectors in all of the other pattern classes have the training label $-1$.
Accordingly, the probabilistic output of any given decision bank
$\mathcal{DB}_{\omega_{i}}\left(  \mathbf{s}\right)  $ is given by a set of
$M-1$ characteristic functions:%
\begin{align*}
E\left[  \chi_{\omega_{i}}\right]   &  =%
{\displaystyle\sum\nolimits_{j=1}^{M-1}}
P\left(  \operatorname{sign}\left(  \widehat{\Lambda}_{\mathfrak{B}_{ij}%
}\left(  \widetilde{\mathbf{x}}\right)  \right)  =1\right) \\
&  =%
{\displaystyle\sum\nolimits_{j=1}^{M-1}}
\operatorname{sign}\left(  \widehat{\Lambda}_{\mathfrak{B}_{ij}}\left(
\widetilde{\mathbf{x}}\right)  \right)  =1\text{.}%
\end{align*}

\begin{figure}[ptb]%
\centering
\fbox{\includegraphics[
height=2.5875in,
width=3.4402in
]%
{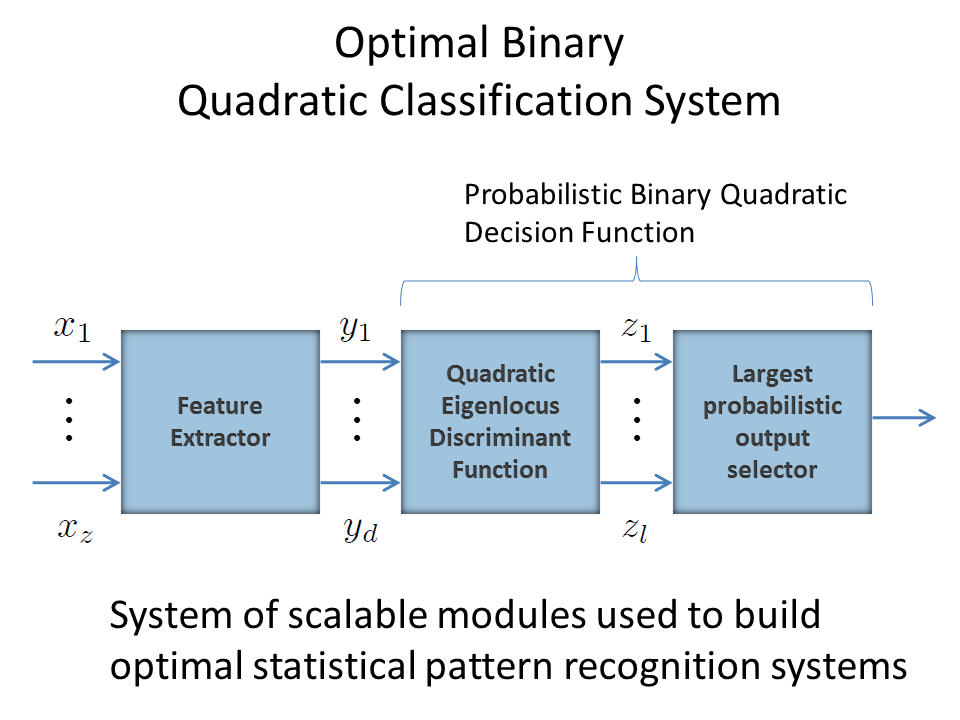}%
}\caption{Illustration of a system of scalable modules used to build optimal
statistical pattern recognition systems. The system includes a feature
extractor, a quadratic eigenlocus discriminant function
$\protect\widetilde{\Lambda}_{\boldsymbol{\kappa}}\left(  \mathbf{s}\right)
=\boldsymbol{\kappa}^{T}k_{\mathbf{s}}+\kappa_{0}$ and a decision function
$\operatorname{sign}\left(  \boldsymbol{\kappa}^{T}k_{\mathbf{s}}+\kappa
_{0}\right)  $.}%
\label{System of Scalable Modules Quadratic Eigenlocus}%
\end{figure}

Decision banks $\mathcal{DB}_{\omega_{i}}\left(  \mathbf{s}\right)  $ are
formed by the system of scalable modules depicted in Fig.
$\ref{System of Scalable Modules Quadratic Eigenlocus}$, where linear
combinations of optimal binary quadratic classification systems can be used to
build an optimal statistical pattern recognition system $\mathfrak{P}%
_{\boldsymbol{o}}\left(  \mathbf{x}\right)  $ which distinguishes between the
objects in $M$ different pattern classes $\left\{  \mathcal{\omega}%
_{i}\right\}  _{i=1}^{M}$: Objects in pattern classes can involve any type of
distinguishing features that have been extracted from collections of: $\left(
1\right)  $ networks formed by interconnected systems of people and/or things:
material or biological, $\left(  2\right)  $ documents, $\left(  3\right)  $
images, or $\left(  4\right)  $ waveforms, signals or sequences, including
stationary random processes.%

I will devise an optimal, statistical pattern recognition system
$P_{\mathfrak{B}}\left(  \mathbf{s}\right)  $ formed by $M$ decision banks
$\left\{  \mathcal{DB}_{\omega_{i}}\left(  \mathbf{s}\right)  \right\}
_{i=1}^{M}$ of quadratic eigenlocus decision functions $\operatorname{sign}%
\left(  \widetilde{\Lambda}_{\boldsymbol{\kappa}_{j}}\left(  \mathbf{s}%
\right)  \right)  $:%
\begin{equation}
\mathfrak{P}_{\boldsymbol{o}}\left(  \mathbf{x}\right)  =\left\{
\mathcal{DB}_{\omega_{i}}\left(
{\textstyle\sum\nolimits_{j=1}^{M-1}}
\operatorname{sign}\left(  \widetilde{\Lambda}_{\boldsymbol{\kappa}_{j}%
}\left(  \mathbf{s}\right)  \right)  \right)  \right\}  _{i=1}^{M}\text{,}
\label{Ensemble of Quadratic Eigenlocus Decision Functions}%
\end{equation}
where%
\[
\operatorname{sign}\left(  \widetilde{\Lambda}_{\boldsymbol{\kappa}_{j}%
}\left(  \mathbf{s}\right)  \right)  =\operatorname{sign}\left(  \left(
\mathbf{x}^{T}\mathbf{s}+1\right)  ^{2}\boldsymbol{\kappa}+\kappa_{0}\right)
\]
or%
\[
\operatorname{sign}\left(  \widetilde{\Lambda}_{\boldsymbol{\kappa}_{j}%
}\left(  \mathbf{s}\right)  \right)  =\operatorname{sign}\left(  \exp\left(
-0.01\left\Vert \mathbf{x}-\mathbf{s}\right\Vert ^{2}\right)
\boldsymbol{\kappa}+\kappa_{0}\right)  \text{,}%
\]
that provides a set of $M\times(M-1)$ decision statistics%
\[
\left\{  \operatorname{sign}\left(  \widetilde{\Lambda}_{\boldsymbol{\kappa
}_{j}}\left(  \mathbf{s}\right)  \right)  \right\}  _{j=1}^{_{M\times(M-1)}%
}\text{,}%
\]
for $M$ pattern classes $\left\{  \mathcal{\omega}_{i}\right\}  _{i=1}^{M}$,
where each decision statistic is a characteristic function $\chi_{\omega_{i}%
}\mapsto P\left(  \operatorname{sign}\left(  \widehat{\Lambda}_{\mathfrak{B}%
_{ij}}\left(  \mathbf{x}\right)  \right)  =1\right)  $ that is determined by
an optimal likelihood ratio test for a two-class feature space:%
\[
\widetilde{\Lambda}_{\boldsymbol{\kappa}_{j}}\left(  \mathbf{s}\right)
\overset{\omega_{1}}{\underset{\omega_{2}}{\gtrless}}0\text{,}%
\]
such that the maximum value selector of the pattern recognition system
$P_{\mathfrak{B}}\left(  \mathbf{s}\right)  $ chooses the pattern class
$\mathcal{\omega}_{i}$ for which a decision bank $\mathcal{DB}_{\omega_{i}%
}\left(  \mathbf{s}\right)  $ has the maximum probabilistic output:%
\[
D_{\mathfrak{B}}\left(  \mathbf{s}\right)  \underset{i\in1,\cdots
,M}{=ArgMax}\left(  \mathcal{DB}_{\omega_{i}}\left(  \mathbf{s}\right)
\right)  \text{,}%
\]
where the probabilistic output of each decision bank $\mathcal{DB}_{\omega
_{i}}\left(  \mathbf{s}\right)  $ is given by a set of $M-1$ characteristic
functions:%
\begin{align*}
E\left[  \chi_{\omega_{i}}\right]   &  =%
{\displaystyle\sum\nolimits_{j=1}^{M-1}}
P\left(  \operatorname{sign}\left(  \widehat{\Lambda}_{\mathfrak{B}_{ij}%
}\left(  \mathbf{s}\right)  \right)  =1\right) \\
&  =%
{\displaystyle\sum\nolimits_{j=1}^{M-1}}
\operatorname{sign}\left(  \widehat{\Lambda}_{\mathfrak{B}_{ij}}\left(
\mathbf{s}\right)  \right)  =1\text{.}%
\end{align*}

For data distributions that have unchanging mean and covariance functions,
I\ will now show that the ensemble of quadratic eigenlocus decision functions
in Eq. (\ref{Ensemble of Quadratic Eigenlocus Decision Functions}) generates a
set of quadratic decision boundaries and decision statistics that minimize the
probability of classification error for $M$ given pattern classes.

\subsection{Expected Risk for Eigenlocus Ensemble Systems II}

Take any given $M$-class feature space, where all $M$ data distributions have
unchanging mean and covariance functions. Now take any given ensemble system
$\mathfrak{P}_{\boldsymbol{o}}\left(  \mathbf{s}\right)  $ formed by $M$
decision banks $\left\{  \mathcal{DB}_{\omega_{i}}\left(  \mathbf{s}\right)
\right\}  _{i=1}^{M}$ of quadratic eigenlocus decision functions
$\operatorname{sign}\left(  \widetilde{\Lambda}_{\boldsymbol{\kappa}_{j}%
}\left(  \mathbf{s}\right)  \right)  $:%
\[
P_{\mathfrak{B}}\left(  \mathbf{s}\right)  =\left\{  \mathcal{DB}_{\omega_{i}%
}\left(
{\textstyle\sum\nolimits_{j=1}^{M-1}}
\operatorname{sign}\left(  \widetilde{\Lambda}_{\boldsymbol{\kappa}_{j}%
}\left(  \mathbf{s}\right)  \right)  \right)  \right\}  _{i=1}^{M}\text{,}%
\]
where each likelihood ratio test $\widetilde{\Lambda}_{\boldsymbol{\kappa}%
_{j}}\left(  \mathbf{s}\right)  \overset{\omega_{1}}{\underset{\omega
_{2}}{\gtrless}}0$ in an ensemble $%
{\textstyle\sum\nolimits_{j=1}^{M-1}}
\operatorname{sign}\left(  \widetilde{\Lambda}_{\boldsymbol{\kappa}_{j}%
}\left(  \mathbf{s}\right)  \right)  $ minimizes the total probability of
error and achieves the lowest possible error rate for two given pattern classes.

Next, take the decision bank $\mathcal{DB}_{\omega_{i}}\left(  \mathbf{x}%
\right)  =%
{\textstyle\sum\nolimits_{j=1}^{M-1}}
\operatorname{sign}\left(  \widetilde{\Lambda}_{\boldsymbol{\kappa}_{j}%
}\left(  \mathbf{s}\right)  \right)  $ for any given any given pattern class
$\omega_{i}$, and let $\mathfrak{R}_{\mathfrak{\min}ij}\left(  Z_{2}%
|\widetilde{\Lambda}_{\boldsymbol{\kappa}_{j}}\right)  $ denote the expected
risk for class $\omega_{i}$ for any given likelihood ratio test
$\widetilde{\Lambda}_{\boldsymbol{\kappa}_{j}}\left(  \mathbf{s}\right)
\overset{\omega_{1}}{\underset{\omega_{2}}{\gtrless}}0$:%
\[
\mathfrak{R}_{\mathfrak{\min}ij}\left(  Z_{2}|\widetilde{\Lambda
}_{\boldsymbol{\kappa}_{j}}\right)  =\int_{Z_{2}}p\left(  \widetilde{\Lambda
}_{\boldsymbol{\kappa}_{j}}\left(  \mathbf{s}\right)  |\omega_{i}\right)
d\widetilde{\Lambda}_{\boldsymbol{\kappa}_{j}}%
\]
which determines the conditional probability of classification error for class
$\omega_{i}$ in a two-class decision space, where $Z=Z_{1}+Z_{2}$. It follows
that each classification system $\widetilde{\Lambda}_{\boldsymbol{\kappa}_{j}%
}\left(  \mathbf{s}\right)  \overset{\omega_{1}}{\underset{\omega
_{2}}{\gtrless}}0$ for class $\omega_{i}$ minimizes the expected risk
$\mathfrak{R}_{\mathfrak{\min}_{ij}}\left(  Z_{2}|\widetilde{\Lambda
}_{\boldsymbol{\kappa}_{j}}\right)  $ in a two-class decision space, where
$Z=Z_{1}+Z_{2}$.

So, let $\mathfrak{R}_{\mathfrak{\min}}\left(  \omega_{i}|\overline{Z}%
_{i}\right)  $ denote the expected risk for class $\omega_{i}$ for any given
ensemble of quadratic eigenlocus decision rules:%
\[
\mathcal{DB}_{\omega_{i}}\left(  \mathbf{s}\right)  =%
{\displaystyle\sum\nolimits_{j=1}^{M-1}}
\left[  \widetilde{\Lambda}_{\boldsymbol{\kappa}_{j}}\left(  \mathbf{s}%
\right)  \overset{\omega_{1}}{\underset{\omega_{2}}{\gtrless}}0\right]
\]
which determines the conditional probability of classification error for class
$\omega_{i}$ in $M-1$ two-class decision spaces $\left\{  Z_{j}\right\}
_{i=1}^{M-1}$, where the decision space $\overline{Z}_{i}$ for the ensemble of
quadratic classifiers is the union of the decision spaces $Z_{1},Z_{2}%
,\ldots,Z_{M-1}$:%
\[
\overline{Z}_{i}=%
{\displaystyle\sum\nolimits_{j=1}^{M-1}}
Z_{j}=\cup_{j=1}^{M-1}\;Z_{j}\text{.}%
\]

Therefore, given an ensemble of quadratic eigenlocus decision rules
$\mathcal{DB}_{\omega_{i}}\left(  \mathbf{s}\right)  $, it follows that the
expected risk $\mathfrak{R}_{\mathfrak{\min}}\left(  \omega_{i}|\overline
{Z}_{i}\right)  $ for class $\omega_{i}$ in an $M$-class feature space is%
\begin{align*}
\mathfrak{R}_{\mathfrak{\min}}\left(  \omega_{i}|\overline{Z}_{i}\right)   &
=%
{\displaystyle\sum\nolimits_{j=1}^{M-1}}
\mathfrak{R}_{\mathfrak{\min}_{ij}}\left(  Z_{2}|\widetilde{\Lambda
}_{\boldsymbol{\kappa}_{j}}\right) \\
&  =%
{\displaystyle\sum\nolimits_{j=1}^{M-1}}
\int_{Z_{2}}p\left(  \widetilde{\Lambda}_{\boldsymbol{\kappa}_{j}}\left(
\mathbf{s}\right)  |\omega_{i}\right)  d\widetilde{\Lambda}%
_{\boldsymbol{\kappa}_{j}}\text{,}%
\end{align*}
where minimization of the expected risk $\mathfrak{R}_{\mathfrak{\min}}\left(
\omega_{i}|\overline{Z}_{i}\right)  $ is equivalent to minimizing the total
probability of decision error for class $\omega_{i}$ in the $M$-class decision
space $\overline{Z}_{i}$.

Finally, let $\widehat{Z}=\cup_{i=1}^{M}\;\overline{Z}_{i}$ denote the
decision space that is determined by the ensemble system%
\[
\mathfrak{P}_{\boldsymbol{o}}\left(  \mathbf{s}\right)  =\left\{
\mathcal{DB}_{\omega_{i}}\left(
{\textstyle\sum\nolimits_{j=1}^{M-1}}
\operatorname{sign}\left(  \widetilde{\Lambda}_{\boldsymbol{\kappa}_{j}%
}\left(  \mathbf{s}\right)  \right)  \right)  \right\}  _{i=1}^{M}%
\]
of $M$ decision banks $\left\{  \mathcal{DB}_{\omega_{i}}\left(
\mathbf{s}\right)  \right\}  _{i=1}^{M}$. It follows that the expected risk
for all $M$ pattern classes $\left\{  \omega_{i}\right\}  _{i=1}^{M}$ is:%
\begin{align*}%
{\displaystyle\sum\nolimits_{i=1}^{M}}
\mathfrak{R}_{\mathfrak{\min}}\left(  \omega_{i}|\overline{Z}_{i}\right)   &
=%
{\displaystyle\sum\nolimits_{i=1}^{M}}
{\displaystyle\sum\nolimits_{j=1}^{M-1}}
\mathfrak{R}_{\mathfrak{\min}_{ij}}\left(  Z_{2}|\widetilde{\Lambda
}_{\boldsymbol{\kappa}_{j}}\right) \\
&  =%
{\displaystyle\sum\nolimits_{i=1}^{M}}
{\displaystyle\sum\nolimits_{j=1}^{M-1}}
\int_{Z_{2}}p\left(  \widetilde{\Lambda}_{\boldsymbol{\kappa}_{j}}\left(
\mathbf{s}\right)  |\omega_{i}\right)  d\widetilde{\Lambda}%
_{\boldsymbol{\kappa}_{j}}\text{,}%
\end{align*}
where the expected risk $\mathfrak{R}_{\mathfrak{\min}}\left(  \omega
_{i}|\overline{Z}_{i}\right)  $ for each pattern class $\omega_{i}$ is
determined by an ensemble of quadratic classifiers $\mathcal{DB}_{\omega_{i}%
}\left(  \mathbf{s}\right)  $ that minimize the total probability of
misclassification or decision error for class $\omega_{i}$ in the $M$-class
decision space $\widehat{Z}$.

Thus, it is concluded that the ensemble system in Eq.
(\ref{Ensemble of Quadratic Eigenlocus Decision Functions}) generates a set of
quadratic decision boundaries and decision statistics that minimize the
probability of classification error for an $M$-class feature space, where all
$M$ data distributions have unchanging mean and covariance functions.

\subsection{Design of Decision Banks}

WLOG, let $\widetilde{\Lambda}\left(  \mathbf{x}\right)  $ denote a linear or
a quadratic eigenlocus discriminant function. The design of optimal,
statistical pattern recognition systems $\mathfrak{P}_{\boldsymbol{o}}\left(
\mathbf{x}\right)  $ involves designing $M$ decision banks, where each
decision bank contains an ensemble of $M-1$ decision functions
$\operatorname{sign}\left(  \widetilde{\Lambda}\left(  \mathbf{x}\right)
\right)  $, and each decision function is determined by a feature extractor
and a linear or a quadratic eigenlocus discriminant function
$\widetilde{\Lambda}\left(  \mathbf{x}\right)  $. A\ feature extractor
generates $d$-dimensional feature vectors from collections of networks,
documents, images, \emph{or} signals for \emph{all} of the $M$ pattern
classes. Alternatively, feature vectors from different data sources can be
fused by means of inductive matrix completion methods.

\paragraph{Fusion of Feature Vectors}

Inductive matrix completion methods enable the fusion of feature vectors that
have been extracted from collections of networks, documents, images, and
signals, where the dimension of the feature vectors for different collections
may differ
\citep[see][]{Jain2013}%
. Fusion of feature vectors enables the production of optimal, statistical
pattern recognition systems based on complex feature spaces.

I\ will now outline the process for producing optimal, statistical pattern
recognition systems $\mathfrak{P}_{\boldsymbol{o}}\left(  \mathbf{x}\right)  $.

Suppose that $M$ sets of $d$-dimensional feature vectors have been extracted
from either fused or non-fused collections of networks, documents, images, or
signals for $M$ pattern classes. Optimal, statistical pattern recognition
systems $\mathfrak{P}_{\boldsymbol{o}}\left(  \mathbf{x}\right)  $ are
produced in the following manner.

\subsubsection{Production of Decision Banks}

Let there be $M$ pattern classes $\left\{  \mathcal{\omega}_{i}\right\}
_{i=1}^{M}$ of $d$-dimensional feature vectors. Produce a decision bank%
\[
\mathcal{DB}_{\omega_{i}}\left(
{\textstyle\sum\nolimits_{j=1}^{M-1}}
\operatorname{sign}\left(  \widetilde{\Lambda}\left(  \mathbf{x}\right)
\right)  \right)
\]
for each pattern class $\mathcal{\omega}_{i}$ that consists of a bank or
ensemble $%
{\textstyle\sum\nolimits_{j=1}^{M-1}}
\operatorname{sign}\left(  \widetilde{\Lambda}\left(  \mathbf{x}\right)
\right)  $ of $M-1$ decision functions $\operatorname{sign}\left(
\widetilde{\Lambda}\left(  \mathbf{x}\right)  \right)  $. Accordingly, build
$M-1$ linear or quadratic eigenlocus discriminant functions
$\widetilde{\Lambda}\left(  \mathbf{x}\right)  $, where the feature vectors in
the given class $\mathcal{\omega}_{i}$ have the training label $+1$ and the
feature vectors in all of the other pattern classes have the training label
$-1$.

It has been demonstrated that the decision bank $\mathcal{DB}_{\omega_{i}%
}\left(  \mathbf{x}\right)  $ for each pattern class $\mathcal{\omega}_{i}$%
\[
\mathcal{DB}_{\omega_{i}}\left(  \mathbf{x}\right)  =%
{\textstyle\sum\nolimits_{j=1}^{M-1}}
\operatorname{sign}\left(  \widetilde{\Lambda}\left(  \mathbf{x}\right)
\right)
\]
generates a set of $M-1$ decision statistics $\left\{  \operatorname{sign}%
\left(  \widetilde{\Lambda}\left(  \mathbf{x}\right)  \right)  \right\}
_{j}^{M-1}$, where each decision statistic $\operatorname{sign}\left(
\widetilde{\Lambda}\left(  \mathbf{x}\right)  \right)  $ is a characteristic
function $\chi_{\omega_{i}}\mapsto P\left(  \operatorname{sign}\left(
\widehat{\Lambda}_{\mathfrak{B}_{ij}}\left(  \mathbf{x}\right)  \right)
=1\right)  $ that is determined by a likelihood ratio test%
\[
\widetilde{\Lambda}\left(  \mathbf{x}\right)  \overset{\omega_{1}%
}{\underset{\omega_{2}}{\gtrless}}0
\]
for a two-class feature space.

Accordingly, an optimal, statistical pattern recognition system $\mathfrak{P}%
_{\boldsymbol{o}}\left(  \mathbf{x}\right)  $%
\[
\mathfrak{P}_{\boldsymbol{o}}\left(  \mathbf{x}\right)  =\left\{
\mathcal{DB}_{\omega_{i}}\left(
{\textstyle\sum\nolimits_{j=1}^{M-1}}
\operatorname{sign}\left(  \widetilde{\Lambda}\left(  \mathbf{x}\right)
\right)  \right)  \right\}  _{i=1}^{M}\text{,}%
\]
contains $M$ decision banks $\left\{  \mathcal{DB}_{\omega_{i}}\left(
\mathbf{x}\right)  \right\}  _{i=1}^{M}$, i.e., $M$ ensembles $%
{\textstyle\sum\nolimits_{j=1}^{M-1}}
\operatorname{sign}\left(  \widetilde{\Lambda}\left(  \mathbf{x}\right)
\right)  $ of optimal decision functions $\operatorname{sign}\left(
\widetilde{\Lambda}\left(  \mathbf{x}\right)  \right)  $, all of which provide
a set of $M\times(M-1)$ decision statistics $\left\{  \operatorname{sign}%
\left(  \widetilde{\Lambda}\left(  \mathbf{x}\right)  \right)  \right\}
_{j=1}^{_{M\times(M-1)}}$ that minimize the probability of decision error for
an $M$-class feature space, such that the maximum value selector of the
pattern recognition system $\mathfrak{P}_{\boldsymbol{o}}\left(
\mathbf{x}\right)  $ chooses the pattern class $\mathcal{\omega}_{i}$ for
which a decision bank $\mathcal{DB}_{\omega_{i}}\left(  \mathbf{x}\right)  $
has the maximum probabilistic output:%
\[
D_{\mathfrak{B}}\left(  \mathbf{x}\right)  \underset{i\in1,\cdots
,M}{=ArgMax}\left(  \mathcal{DB}_{\omega_{i}}\left(  \mathbf{x}\right)
\right)  \text{,}%
\]
where the probabilistic output of each decision bank $\mathcal{DB}_{\omega
_{i}}\left(  \mathbf{x}\right)  $ is determined by a set of $M-1$
characteristic functions:%
\begin{align*}
E\left[  \chi_{\omega_{i}}\right]   &  =%
{\displaystyle\sum\nolimits_{j=1}^{M-1}}
P\left(  \operatorname{sign}\left(  \widehat{\Lambda}_{\mathfrak{B}_{ij}%
}\left(  \mathbf{x}\right)  \right)  =1\right) \\
&  =%
{\displaystyle\sum\nolimits_{j=1}^{M-1}}
\operatorname{sign}\left(  \widehat{\Lambda}_{\mathfrak{B}_{ij}}\left(
\mathbf{x}\right)  \right)  =1\text{.}%
\end{align*}

For data distributions that have unchanging mean and covariance functions,
optimal, statistical pattern recognition systems $\mathfrak{P}_{\boldsymbol{o}%
}\left(  \mathbf{x}\right)  $ that are formed by ensembles of quadratic
eigenlocus decision functions generate a set of quadratic decision boundaries
and decision statistics that minimize the probability of decision error.
Moreover, for data distributions that have common covariance functions,
ensembles of quadratic eigenlocus decision functions generate a set of linear
decision boundary estimates and decision statistics that minimize the
probability of decision error. It follows that ensembles of quadratic
eigenlocus decision functions generate a set of decision boundaries and
decision statistics that minimize the probability of decision error: for any
given sets of pattern or feature vectors which have been extracted from
digital signals, images, documents, or networks and are generated according to
probability density functions related to statistical distributions of random
vectors that have constant or unchanging statistics.

For data distributions that have similar covariance matrices, optimal,
statistical pattern recognition systems $\mathfrak{P}_{\boldsymbol{o}}\left(
\mathbf{x}\right)  $ that are formed by ensembles of linear eigenlocus
decision functions generate a set of linear decision boundaries and decision
statistics that minimize the probability of decision error.

Because an optimal, statistical pattern recognition system $\mathfrak{P}%
_{\boldsymbol{o}}\left(  \mathbf{x}\right)  $ is a linear combination of
linear or quadratic eigenlocus discriminant functions $\widetilde{\Lambda
}\left(  \mathbf{x}\right)  $, it follows that the overall network complexity
is scale-invariant for the feature space dimension and the number of pattern
classes. Moreover, optimal, statistical pattern recognition systems
$\mathfrak{P}_{\boldsymbol{o}}\left(  \mathbf{x}\right)  $ that achieve
minimum error rates are optimal ensembles of binary classifiers, where each
binary classifier in an ensemble based system $\mathfrak{P}_{\boldsymbol{o}%
}\left(  \mathbf{x}\right)  $ achieves the lowest possible error rate and
exhibits optimal generalization performance for its two-class feature space.

\subsection{Optimal Ensembles of Binary Classifiers}

Let $\mathfrak{P}_{\boldsymbol{o}}\left(  \mathbf{s}\right)  $ denote an
optimal, statistical classification system that is formed by ensembles of
quadratic eigenlocus discriminant functions. Optimal, statistical
classification systems $\mathfrak{P}_{\boldsymbol{o}}\left(  \mathbf{s}%
\right)  $ are ensemble based systems that \emph{outperform }other ensembles
of classifiers. I\ have demonstrated that optimal, statistical decision
systems $\mathfrak{P}_{\boldsymbol{o}}\left(  \mathbf{s}\right)  $ can be
formed by \emph{optimal ensembles} of individual binary classifiers, where
each binary classifier is a high-performance learning machine that achieves
the lowest possible error rate and exhibits optimal generalization performance
for its two-class feature space. In particular, optimal, statistical decision
systems $\mathfrak{P}_{\boldsymbol{o}}\left(  \mathbf{s}\right)  $ formed by
ensembles of quadratic eigenlocus decision functions $\operatorname{sign}%
\left(  \left(  \mathbf{x}^{T}\mathbf{s}+1\right)  ^{2}\boldsymbol{\kappa
}+\kappa_{0}\right)  $ or $\operatorname{sign}\left(  \exp\left(
-0.01\left\Vert \mathbf{x}-\mathbf{s}\right\Vert ^{2}\right)
\boldsymbol{\kappa}+\kappa_{0}\right)  $ achieve minimum error rates and
exhibit optimal generalization performance for $d$-component random vectors
$\mathbf{x}$ that are generated according to probability density functions
$p\left(  \mathbf{x}|\omega_{1}\right)  $ and $p\left(  \mathbf{x}|\omega
_{2}\right)  $ related to statistical distributions of random vectors
$\mathbf{x}$ that have constant or unchanging statistics and similar or
dissimilar covariance matrices.

\begin{figure}[ptb]%
\centering
\fbox{\includegraphics[
height=2.5875in,
width=3.4411in
]%
{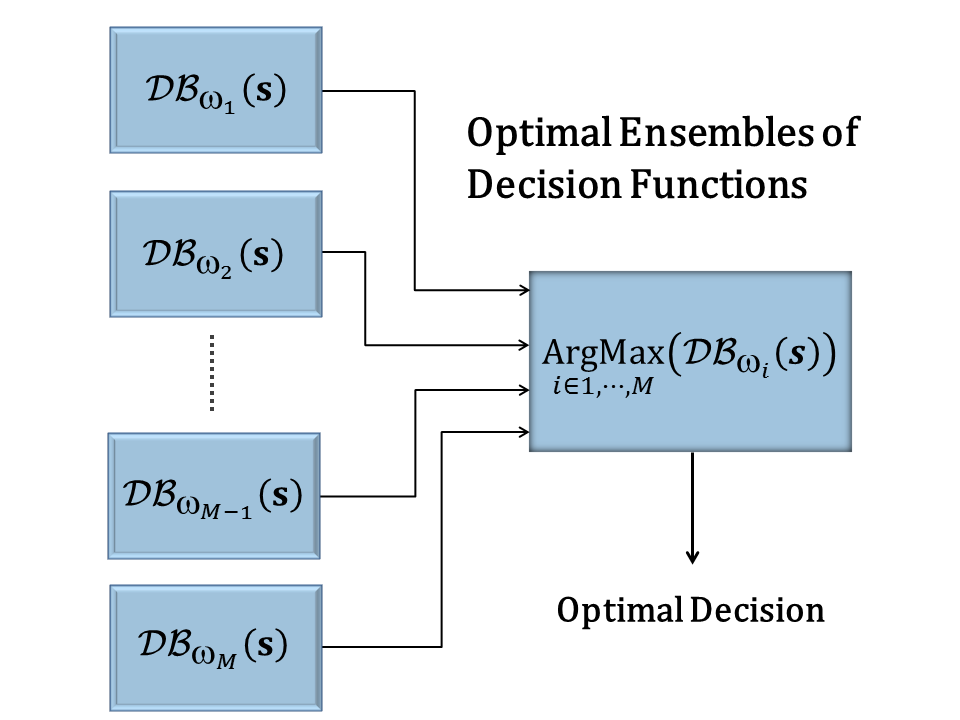}%
}\caption{Optimal, statistical decision systems $P_{\mathfrak{B}}\left(
\mathbf{s}\right)  $ are formed by $M$ decision banks\ or ensembles of optimal
likelihood ratio tests, where each decision bank $\mathcal{DB}_{\omega_{i}%
}\left(  \mathbf{s}\right)  $ consists of an ensemble of $M-1$ optimal
decision functions: ${\textstyle\protect\sum\nolimits_{j=1}^{M-1}%
}\operatorname{sign}\left(  \protect\widetilde{\Lambda}_{j}\left(
\mathbf{s}\right)  \right)  $.}%
\label{Ensembles of Bayes' Tests}%
\end{figure}

Figure $\ref{Ensembles of Bayes' Tests}$ illustrates ensemble based systems of
optimal likelihood ratio tests that generate optimal decision rules.%

I will now outline a method for fusing feature vectors from different data
sources that enables the production of optimal, statistical pattern
recognition systems based on complex feature spaces.

\subsection{Fusion of Decision Banks}

Feature vectors that have been extracted from collections of networks,
documents, images, or signals can be fused with each other by designing
decision banks for data obtained from different sources and combining the
outputs of the decision banks. I will outline the method for two different
data sources. The method is readily extended to $L$ sources of data.

Suppose that $M$ sets of $d$-dimensional and $n$-dimensional feature vectors
have been extracted from two different collections of networks, documents,
images, or signals for $M$ pattern classes. Optimal, statistical pattern
recognition systems $\mathfrak{P}_{\boldsymbol{o}}\left(  \mathbf{s}\right)  $
are produced in the following manner.

\subsubsection{Production of Fused Decision Banks}

Let there be $M$ pattern classes: $\left\{  \mathcal{\omega}_{i}\right\}
_{i=1}^{M}$. Let $\mathcal{DB}_{\omega_{i1}}$ and $\mathcal{DB}_{\omega_{i2}}$
denote decision banks for $d$-dimensional and $n$-dimensional feature vectors
respectively, where feature vectors in class $\mathcal{\omega}_{i}$ have the
training label $+1$ and feature vectors in all of the other pattern classes
have the training label $-1$. Produce the decision banks%
\[
\mathcal{DB}_{\omega_{i1}}\left(
{\textstyle\sum\nolimits_{j=1}^{M-1}}
\operatorname{sign}\left(  \widetilde{\Lambda}_{j}\left(  \mathbf{x}\right)
\right)  \right)
\]
and%
\[
\mathcal{DB}_{\omega_{i2}}\left(
{\textstyle\sum\nolimits_{j=1}^{M-1}}
\operatorname{sign}\left(  \widetilde{\Lambda}_{j}\left(  \mathbf{x}\right)
\right)  \right)
\]
for each pattern class $\mathcal{\omega}_{i}$, where $\mathcal{DB}%
_{\omega_{i1}}$ and $\mathcal{DB}_{\omega_{i2}}$ consist of a bank or ensemble
$%
{\textstyle\sum\nolimits_{j=1}^{M-1}}
\operatorname{sign}\left(  \widetilde{\Lambda}_{j}\left(  \mathbf{x}\right)
\right)  $ of $M-1$ quadratic eigenlocus decision functions
$\operatorname{sign}\left(  \widetilde{\Lambda}_{j}\left(  \mathbf{x}\right)
\right)  $. Accordingly, for each decision bank, build $M-1$ quadratic
eigenlocus discriminant functions $\widetilde{\Lambda}\left(  \mathbf{x}%
\right)  $, where the pattern vectors in the given class $\mathcal{\omega}%
_{i}$ have the training label $+1$ and the pattern vectors in all of the other
pattern classes have the training label $-1$.

So, for each pattern class $\mathcal{\omega}_{i}$, the decision banks
$\mathcal{DB}_{\omega_{i1}}$ and $\mathcal{DB}_{\omega_{i2}}$%
\[
\mathcal{DB}_{\omega_{i1}}\left(  \mathbf{x}\right)  =%
{\textstyle\sum\nolimits_{j=1}^{M-1}}
\operatorname{sign}\left(  \widetilde{\Lambda}_{j}\left(  \mathbf{x}\right)
\right)
\]
and%
\[
\mathcal{DB}_{\omega_{i2}}\left(  \mathbf{x}\right)  =%
{\textstyle\sum\nolimits_{j=1}^{M-1}}
\operatorname{sign}\left(  \widetilde{\Lambda}_{j}\left(  \mathbf{x}\right)
\right)
\]
generate two sets of $M-1$ decision statistics%
\[
\mathcal{DB}_{\omega_{i1}}\left(  \mathbf{x}\right)  =\left\{
\operatorname{sign}\left(  \widetilde{\Lambda}_{j}\left(  \mathbf{x}\right)
\right)  \right\}  _{j=1}^{M-1}%
\]
and%
\[
\mathcal{DB}_{\omega_{i2}}\left(  \mathbf{x}\right)  =\left\{
\operatorname{sign}\left(  \widetilde{\Lambda}_{j}\left(  \mathbf{x}\right)
\right)  \right\}  _{j=1}^{M-1}\text{,}%
\]
where each decision statistic $\operatorname{sign}\left(  \widetilde{\Lambda
}_{j}\left(  \mathbf{x}\right)  \right)  $ is a characteristic function
$\chi_{\omega_{i}}\mapsto P\left(  \operatorname{sign}\left(  \widehat{\Lambda
}_{\mathfrak{B}_{ij}}\left(  \mathbf{x}\right)  \right)  =1\right)  $ that is
determined by an optimal likelihood ratio test for a two-class feature space:%
\[
\widetilde{\Lambda}_{j}\left(  \mathbf{x}\right)  \overset{\omega
_{1}}{\underset{\omega_{2}}{\gtrless}}0\text{,}%
\]
such that the maximum value selector of the statistical pattern recognition
system $\mathfrak{P}_{\boldsymbol{o}}\left(  \mathbf{s}\right)  $ chooses the
pattern class $\mathcal{\omega}_{i}$ for which the fused decision banks $%
{\displaystyle\sum\nolimits_{j=1}^{2}}
\mathcal{DB}_{\omega_{ij}}\left(  \mathbf{x}\right)  $ have the maximum
probabilistic output:%
\[
D_{\mathfrak{B}}\left(  \mathbf{s}\right)  \underset{i\in1,\cdots
,M}{=ArgMax}\left(
{\displaystyle\sum\nolimits_{j=1}^{2}}
\mathcal{DB}_{\omega_{ij}}\left(  \mathbf{s}\right)  \right)  \text{.}%
\]

Accordingly, an optimal, statistical pattern recognition system $\mathfrak{P}%
_{\boldsymbol{o}}\left(  \mathbf{s}\right)  $ formed by the fused decision
banks $\mathcal{DB}_{\omega_{i1}}$ and $\mathcal{DB}_{\omega_{i2}}$:%
\[
\mathfrak{P}_{\boldsymbol{o}}\left(  \mathbf{s}\right)  =\left\{
\begin{array}
[c]{c}%
\mathcal{DB}_{\omega_{i1}}\left(
{\textstyle\sum\nolimits_{j=1}^{M-1}}
\operatorname{sign}\left(  \widehat{\Lambda}_{\mathfrak{BE}_{j}}\left(
\mathbf{s}\right)  \right)  \right) \\
+\mathcal{DB}_{\omega_{i2}}\left(
{\textstyle\sum\nolimits_{j=1}^{M-1}}
\operatorname{sign}\left(  \widehat{\Lambda}_{\mathfrak{BE}_{j}}\left(
\mathbf{s}\right)  \right)  \right)
\end{array}
\right\}  _{i=1}^{M}%
\]
contains $2\times M$ decision banks i.e., $2\times M$ ensembles of optimal,
statistical decision functions, all of which provide a set of $2\times
M\times(M-1)$ decision statistics%
\[
\left\{  \operatorname{sign}\left(  \widetilde{\Lambda}_{j}\left(
\mathbf{s}\right)  \right)  \right\}  _{j=1}^{2\times M\times(M-1)}%
\]
that minimize the probability of decision error for two sources of data in an
$M$-class feature space, such that the maximum value selector of the
statistical pattern recognition system $\mathfrak{P}_{\boldsymbol{o}}\left(
\mathbf{s}\right)  $%
\[
\mathfrak{P}_{\boldsymbol{o}}\left(  \mathbf{s}\right)  =\left\{
{\displaystyle\sum\limits_{j=1}^{2}}
\mathcal{DB}_{\omega_{ij}}\left(
{\textstyle\sum\nolimits_{k=1}^{M-1}}
\operatorname{sign}\left(  \widetilde{\Lambda}_{k}\left(  \mathbf{s}\right)
\right)  \right)  \right\}  _{i=1}^{M}%
\]
chooses the pattern class $\mathcal{\omega}_{i}$ for which two fused decision
banks $%
{\displaystyle\sum\nolimits_{j=1}^{2}}
\mathcal{DB}_{\omega_{ij}}\left(  \mathbf{s}\right)  $ have the maximum
probabilistic output:%
\[
D_{\mathfrak{B}}\left(  \mathbf{s}\right)  \underset{i\in1,\cdots
,M}{=ArgMax}\left(
{\displaystyle\sum\nolimits_{j=1}^{2}}
\mathcal{DB}_{\omega_{ij}}\left(  \mathbf{s}\right)  \right)  \text{.}%
\]

Thus, feature vectors from two different data sources can be fused by forming
fused ensembles of quadratic eigenlocus decision functions. Feature vectors
from two different data sources can also be fused by forming fused ensembles
of linear eigenlocus decision functions.

\begin{figure}[ptb]%
\centering
\fbox{\includegraphics[
height=2.5875in,
width=3.4411in
]%
{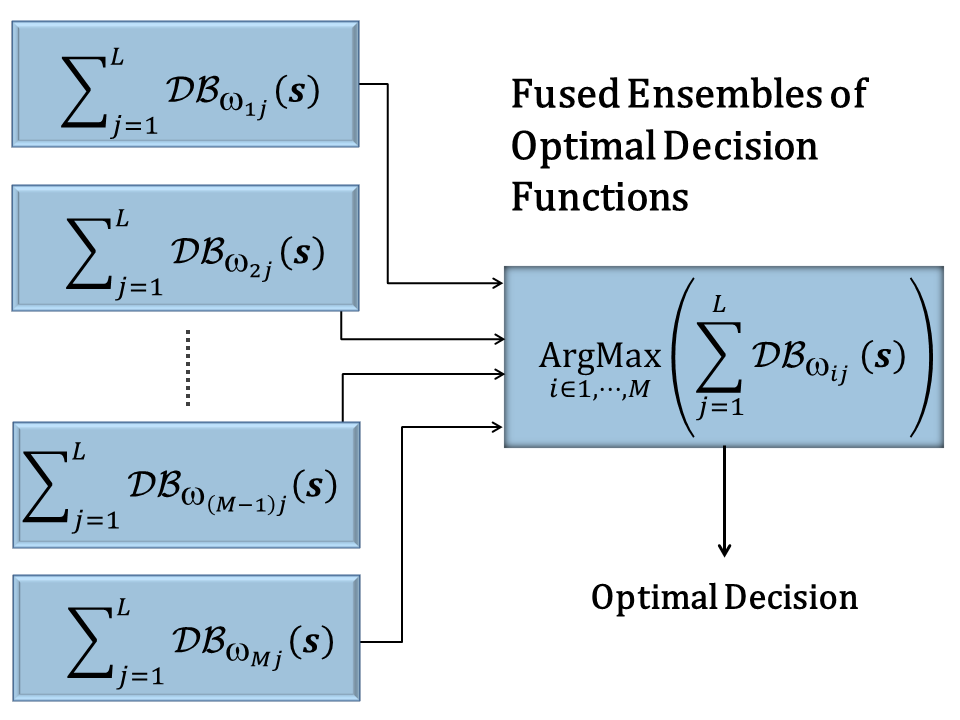}%
}\caption{Feature vectors from different data sources can be fused by forming
fused ensembles of linear or quadratic eigenlocus decision functions.}%
\label{Fused Ensembles of Bayes' Tests}%
\end{figure}

The ensemble method is readily extended to $L$ different data sources. Given
that fusion of decision banks based on different data sources involves linear
combinations of decision banks, it follows that optimal, statistical pattern
recognition systems $\mathfrak{P}_{\boldsymbol{o}}\left(  \mathbf{s}\right)  $
can be designed for data sets that have been collected from $L$ different data
sources:%
\[
\mathfrak{P}_{\boldsymbol{o}}\left(  \mathbf{s}\right)  =\left\{
{\displaystyle\sum\limits_{j=1}^{L}}
\mathcal{DB}_{\omega_{ij}}\left(
{\textstyle\sum\nolimits_{k=1}^{M-1}}
\operatorname{sign}\left(  \widetilde{\Lambda}_{k}\left(  \mathbf{s}\right)
\right)  \right)  \right\}  _{i=1}^{M}%
\]
such that the maximum value selector of the optimal, statistical pattern
recognition system $\mathfrak{P}_{\boldsymbol{o}}\left(  \mathbf{s}\right)  $
chooses the pattern class $\mathcal{\omega}_{i}$ for which $L$ fused decision
banks $%
{\displaystyle\sum\nolimits_{j=1}^{L}}
\mathcal{DB}_{\omega_{ij}}\left(  \mathbf{s}\right)  $ have the maximum
probabilistic output:%
\[
D_{\mathfrak{B}}\underset{i\in1,\cdots,M}{=ArgMax}\left(
{\displaystyle\sum\nolimits_{j=1}^{L}}
\mathcal{DB}_{\omega_{ij}}\left(  \mathbf{s}\right)  \right)  \text{.}%
\]
Figure $\ref{Fused Ensembles of Bayes' Tests}$ illustrates how feature vectors
from $L$ different data sources can be fused with each other by forming fused
ensembles of linear or quadratic eigenlocus decision rules.%

\subsection{Perfect Generalization Performance}

All classes of feature vectors drawn from non-overlapping statistical or data
distributions exhibit perfect discrimination capacity, where the expected risk
and the decision error is zero. Therefore, given $M$ pattern classes $\left\{
\mathcal{\omega}_{i}\right\}  _{i=1}^{M}$, if a feature extractor can be
developed for which negligible or no overlap exists between all $M$ data
distributions, then the scale-invariance of the optimal, statistical pattern
recognition system $\mathfrak{P}_{\boldsymbol{o}}\left(  \mathbf{s}\right)  $
described above ensures low estimation variance and \emph{perfect}
generalization performance. Moreover, $\mathfrak{P}_{\boldsymbol{o}}\left(
\mathbf{s}\right)  $ can be replaced by an optimal, \emph{linear}, statistical
pattern recognition system $\mathfrak{P}_{\boldsymbol{o}}\left(
\mathbf{x}\right)  $: formed by ensembles of linear eigenlocus decision functions.

The probability of error is the key parameter of all statistical pattern
recognition systems
\citep{Fukunaga1990,Jain2000}%
. The amount of overlap between data distributions determines the
classification error rate which is the lowest error rate that can be achieved
by any statistical classifier. In general, the classification error rate is
\emph{difficult to evaluate}
\citep{Fukunaga1990}%
.

\subsection{Design of Feature Vectors}%

\citet{Geman1992}
suggested that some important biases or generalizations can be achieved
through proper data representations. For the problem of learning discriminant
functions and decision boundaries, an important form of proper data
representations involves the identification and exploitation of distinguishing
features that are simple to extract, invariant to irrelevant transformations,
insensitive to noise, and useful for discriminating between objects in
different categories
\citep{Duda2001}%
.

\subsubsection{Sufficient Class Separability}

Useful sets of distinguishing features for discrimination tasks must exhibit
sufficient class separability, i.e., a negligible overlap exists between all
data distributions. In general, the design of distinguishing feature vectors
which exhibit sufficient class separability is the most fundamental and
difficult problem in the overall design of a statistical pattern recognition
system
\citep{Fukunaga1990,Jain2000,Duda2001}%
.

\subsection{Measures of Feature Vector Effectiveness}

The criteria to evaluate the effectiveness of feature vectors \emph{must be} a
measure of the \emph{overlap or class separability among data distributions}
and \emph{not} a \emph{measure of fit} such as the mean-square error of a
statistical model
\citep{Fukunaga1990}%
. For example, the Bhattacharyya distance provides a convenient measure of
class separability for two pattern classes. The measure provides an upper
bound of the Bayes' error if training data are drawn from Gaussian
distributions. However, the upper bound may be significantly higher than the
Bayes' error. Furthermore, the Bhattacharyya distance is difficult to evaluate
because the trace and the determinant of matrices are combined in the
criterion
\citep{Fukunaga1990}%
.

\subsubsection{Predicting Generalization Performance}

Because linear or quadratic eigenlocus classification systems optimize
trade-offs between \emph{decision errors}: based on trade-offs between counter
risks and risks, they can be used to \emph{predict} how well they will
\emph{generalize to new patterns}. Thereby, linear or quadratic eigenlocus
classification systems provide a robust measure of class separability and the
expected error rate for two given sets of feature vectors whose mean and
covariance functions remain constant over time. In addition, because quadratic
eigenlocus classification systems optimize trade-offs between counter risks
and risks for \emph{any two data distributions}, quadratic eigenlocus
classification systems provide \emph{accurate and precise} measures of data
distribution overlap and the expected error rate for any two given sets of
feature vectors.

\subsection{A Practical Statistical Multimeter}

Linear eigenlocus or quadratic eigenlocus decision functions provide a
practical statistical gauge for measuring data distribution overlap and the
expected error rate for two given sets of feature or pattern vectors.
Moreover, it has been demonstrated that quadratic eigenlocus classification
systems generate estimates of linear decision boundaries. WLOG, let
$\widetilde{\Lambda}\left(  \mathbf{x}\right)  \overset{\omega_{1}%
}{\underset{\omega_{2}}{\gtrless}}0$ denote a linear or quadratic eigenlocus
classification system: $\widetilde{\Lambda}\left(  \mathbf{x}\right)  $
denotes a discriminant function.

Recall that linear and quadratic eigenlocus discriminant functions are
likelihood ratio tests $\widetilde{\Lambda}\left(  \mathbf{x}\right)
\overset{\omega_{1}}{\underset{\omega_{2}}{\gtrless}}0$, where any given
likelihood ratio test $\widetilde{\Lambda}\left(  \mathbf{x}\right)
\overset{\omega_{1}}{\underset{\omega_{2}}{\gtrless}}0$ is based on trade-offs
between counter risks and risks (decision errors) for two given pattern
classes. Thereby, linear or quadratic eigenlocus decision functions
$\operatorname{sign}\left(  \widetilde{\Lambda}\left(  \mathbf{x}\right)
\right)  $ can be used to measure data distribution overlap and the expected
error rate for any two given sets of feature vectors whose mean and covariance
functions are constant over time.

To measure the expected error rate and data distribution overlap, build a
linear or quadratic eigenlocus classification system $\widetilde{\Lambda
}\left(  \mathbf{x}\right)  \overset{\omega_{1}}{\underset{\omega
_{2}}{\gtrless}}0$ using feature vectors which have been extracted from any
given collections of networks, documents, images, or signals for two pattern
classes. While equal numbers of training examples are not absolutely
necessary, the number of training examples from each of the pattern classes
should be reasonably balanced with each other.

Apply the decision function $\operatorname{sign}\left(  \widetilde{\Lambda
}\left(  \mathbf{x}\right)  \right)  $ to a collection of feature vectors
which have not been used to build the classification system
$\widetilde{\Lambda}\left(  \mathbf{x}\right)  \overset{\omega_{1}%
}{\underset{\omega_{2}}{\gtrless}}0$. In general, if $\boldsymbol{\tau}$ or
$\boldsymbol{\kappa}$ is based on a small number of principal eigenaxis
components, e.g., $2-10$, the data distributions are not overlapping with each
other and the expected error rate is negligible or zero. However, a large
number of principal eigenaxis components does not necessarily indicate a large
expected risk. Also, because quadratic eigenlocus classification systems
optimize trade-offs between counter risks and risks (decision errors) for
\emph{any two data distributions}, quadratic eigenlocus decision functions
$\operatorname{sign}\left(  \left(  \mathbf{x}^{T}\mathbf{s}+1\right)
^{2}\boldsymbol{\kappa}+\kappa_{0}\right)  $ or $\operatorname{sign}\left(
\exp\left(  -0.01\left\Vert \mathbf{x}-\mathbf{s}\right\Vert ^{2}\right)
\boldsymbol{\kappa}+\kappa_{0}\right)  $ provide \emph{accurate and precise}
measures of data distribution overlap and the expected error rate for any two
data distributions. Alternatively, linear eigenlocus decision functions
$\operatorname{sign}\left(  \boldsymbol{\tau}^{T}\mathbf{x}+\tau_{0}\right)  $
provide accurate and precise measures of data distribution overlap and the
expected error rate for any two data distributions that have similar
covariance matrices.

If data collection is cost prohibitive, use $k$-fold cross validation, where a
collection of feature vectors is split randomly into $k$ partitions. Build a
linear or quadratic eigenlocus classification system $\widetilde{\Lambda
}\left(  \mathbf{x}\right)  \overset{\omega_{1}}{\underset{\omega
_{2}}{\gtrless}}0$ with a data set consisting of $k-1$ of the original $k$
parts and use the remaining portion for testing. Repeat this process $k$
times. The expected error rate and data distribution overlap is the average
over the $k$ test runs.

\paragraph{Caveats:}

Because linear eigenlocus classification systems optimize trade-offs between
counter risks and risks (decision errors) for any two data distributions that
have similar covariance matrices, a\ statistical multimeter based on a linear
eigenlocus decision function \emph{may not} provide accurate and precise
measures of distribution overlap and the expected error rate for data
distributions that have different covariance matrices. Alternatively, because
quadratic eigenlocus classification systems optimize trade-offs between
counter risks and risks (decision errors) for any two data distributions, a
statistical multimeter based on a quadratic eigenlocus decision function
provides accurate and precise measures of data distribution overlap and the
expected error rate for any two data distributions.

\begin{figure}[ptb]%
\centering
\fbox{\includegraphics[
height=2.5875in,
width=3.4411in
]%
{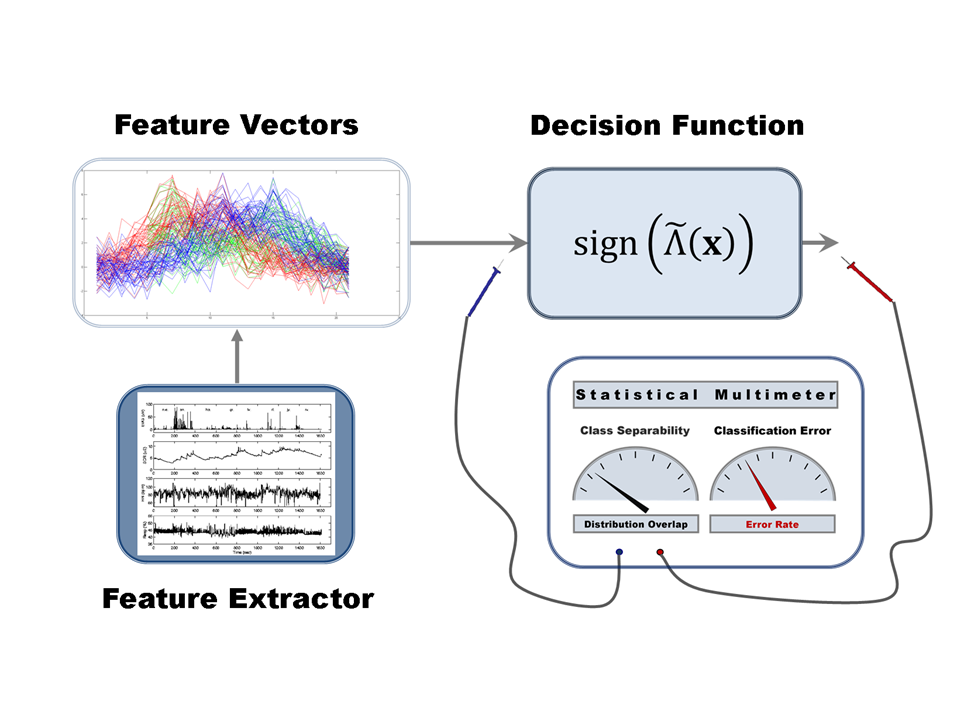}%
}\caption{Linear or quadratic eigenlocus decision functions
$\operatorname{sign}\left(  \protect\widetilde{\Lambda}\left(  \mathbf{x}%
\right)  \right)  $ provide a practical statistical multimeter for measuring
data distribution overlap and the expected error rate for two given sets of
feature vectors.}%
\label{Statistical Multimeter}%
\end{figure}

Figure $\ref{Statistical Multimeter}$ illustrates how linear or quadratic
eigenlocus decision functions $\operatorname{sign}\left(  \widetilde{\Lambda
}\left(  \mathbf{x}\right)  \right)  $ provide a practical statistical
multimeter for measuring data distribution overlap and the expected error rate
for two given sets of feature vectors.

\subsection{A Practical Statistical Gauge}

Linear or quadratic eigenlocus decision functions $\operatorname{sign}\left(
\widetilde{\Lambda}\left(  \mathbf{x}\right)  \right)  $ can also be used to
identify homogeneous data distributions. Build a linear or quadratic
eigenlocus classification system $\widetilde{\Lambda}\left(  \mathbf{x}%
\right)  \overset{\omega_{1}}{\underset{\omega_{2}}{\gtrless}}0$ using samples
drawn from two distributions. Apply the decision function $\operatorname{sign}%
\left(  \widetilde{\Lambda}\left(  \mathbf{x}\right)  \right)  $ to samples
which have not been used to build the classification system
$\widetilde{\Lambda}\left(  \mathbf{x}\right)  \overset{\omega_{1}%
}{\underset{\omega_{2}}{\gtrless}}0$.

Given homogeneous data distributions, essentially all of the training data are
transformed into constrained, primal principal eigenaxis components, such that
the error rate of the classification system $\widetilde{\Lambda}\left(
\mathbf{x}\right)  \overset{\omega_{1}}{\underset{\omega_{2}}{\gtrless}}0$ is
$\approx50\%$.

If data collection is cost prohibitive, use $k$-fold cross validation, where a
collection of feature vectors is split randomly into $k$ partitions. Build a
linear or quadratic eigenlocus classification system $\widetilde{\Lambda
}\left(  \mathbf{x}\right)  \overset{\omega_{1}}{\underset{\omega
_{2}}{\gtrless}}0$ with a data set consisting of $k-1$ of the original $k$
parts and use the remaining portion for testing. Repeat this process $k$
times. The expected error rate and data distribution overlap is the average
over the $k$ test runs.

\subsubsection{The Two Sample Problem}

Alternatively, linear or quadratic eigenlocus decision functions
$\operatorname{sign}\left(  \widetilde{\Lambda}\left(  \mathbf{x}\right)
\right)  $ can be used to determine if two samples are from different
distributions. Build a linear or quadratic eigenlocus classification system
$\widetilde{\Lambda}\left(  \mathbf{x}\right)  \overset{\omega_{1}%
}{\underset{\omega_{2}}{\gtrless}}0$ using samples drawn from two
distributions. Apply the decision function $\operatorname{sign}\left(
\widetilde{\Lambda}\left(  \mathbf{x}\right)  \right)  $ to samples which have
not been used to build the classification system. Given different data
distributions, the error rate of the classification system $\widetilde{\Lambda
}\left(  \mathbf{x}\right)  \overset{\omega_{1}}{\underset{\omega
_{2}}{\gtrless}}0$ is less than $50\%$.

\paragraph{Caveats:}

Because linear eigenlocus classification systems optimize trade-offs between
counter risks and risks (decision errors) for any two data distributions that
have similar covariance matrices, a\ statistical gauge based on a linear
eigenlocus decision function may \emph{not} provide accurate and precise
measures of error rates for samples drawn from distributions that have
different covariance matrices. Indeed, given two data distributions that have
different covariance matrices, the error rate of a linear eigenlocus
classification system $\widetilde{\Lambda}\left(  \mathbf{x}\right)
\overset{\omega_{1}}{\underset{\omega_{2}}{\gtrless}}0$ may be $50\%$.
Alternatively, because quadratic eigenlocus classification systems optimize
trade-offs between counter risks and risks (decision errors) for any two data
distributions, a statistical gauge based on a quadratic eigenlocus decision
function provides accurate and precise measures of error rates for samples
drawn from any two distributions.

\subsubsection{Primary Statistical Gauge}

Because quadratic eigenlocus decision functions $\operatorname{sign}\left(
\widetilde{\Lambda}\left(  \mathbf{x}\right)  \right)  $ provide accurate and
precise measures of data distribution overlap and the expected error rate for
any two data distributions in addition to accurate and precise statistical
indicators of homogenous and different data distributions, quadratic
eigenlocus decision functions $\operatorname{sign}\left(  \widetilde{\Lambda
}\left(  \mathbf{x}\right)  \right)  $ provide a primary statistical gauge.
Therefore, quadratic eigenlocus decision functions $\operatorname{sign}\left(
\widetilde{\Lambda}\left(  \mathbf{x}\right)  \right)  $ should always be used
to verify results that have been obtained using linear eigenlocus decision
functions $\operatorname{sign}\left(  \widetilde{\Lambda}\left(
\mathbf{x}\right)  \right)  $.

I have covered a lot of ground in this working paper. The main ideas of the
paper are summarized in my final remarks.

\section{Final Remarks}

I have devised data-driven, mathematical laws that generate optimal,
statistical classification systems which achieve minimum error rates for data
distributions with unchanging statistics. The data-driven, mathematical laws
involve finding a solution of locus equations, subject to fundamental
statistical laws for a classification system in statistical equilibrium.

I\ have also introduced new ways of thinking about learning machines: in terms
of fundamental statistical laws that model-based architectures of learning
machines are subject to. I have devised equations of statistical equilibrium
along with equations of minimization of eigenenergy and expected risk that
learning machines architectures are subject to. In addition, I\ have
introduced new ways of thinking about linear, polynomial, and Gaussian kernel
support vector machines (SVMs): I have defined effective hyperparameters for
polynomial and Gaussian kernel SVMs, and I\ have defined regularization
methods for linear, polynomial, and Gaussian kernel SVMs; I have also resolved
the geometric locus dilemma for all three classes of SVMs.

I\ have devised a system of fundamental equations of binary classification for
a classification system in statistical equilibrium that must be satisfied by
likelihood ratios and decision boundaries of binary classification systems. I
have demonstrated that classification systems seek a point of statistical
equilibrium where the opposing forces and influences of a classification
system are balanced with each other, and the eigenenergy and the expected risk
of a classification system are minimized. I\ have also demonstrated that the
total eigenenergy of any given binary classification system is conserved and
remains relatively constant, so that the eigenenergy and the corresponding
expected risk of a binary classification system cannot be created or
destroyed, but only transferred from one classification system to another. In
addition, I have demonstrated that the eigenenergy and the corresponding
expected risk of discrete, linear and quadratic classification systems cannot
be created or destroyed, but only transferred from one classification system
to another.

I\ have used these results to rigorously define three classes of learning
machines that are scalable modules for optimal, statistical classification or
pattern recognition systems, where each class of learning machines exhibits
optimal generalization performance for a category of statistical
distributions. One class of learning machines achieves minimum error rates for
data sets drawn from statistical distributions that have unchanging statistics
and similar covariance matrices. The other two classes of learning machines
achieve minimum error rates for any given data sets drawn from statistical
distributions that have either similar or dissimilar covariance matrices and
unchanging statistics.

All three classes of learning machines are solutions to fundamental integral
equations of likelihood ratios and corresponding decision boundaries, so that
each class of learning machines finds a point of statistical equilibrium where
the opposing forces and influences of a statistical classification system are
balanced with each other, and the eigenenergy and the corresponding expected
risk of the learning machine are minimized. The generalization error of each
class of learning machines is determined by the minimum probability of
classification error: which is the lowest error rate that can be achieved by a
discriminant function and the best generalization error that can be achieved
by a learning machine. I\ have also defined optimal ensemble systems for each
class of learning machines so that the generalization error of any given
ensemble system is determined by the minimum probability of classification error.

In this paper, I\ have formulated the problem of learning unknown, linear and
quadratic discriminant functions from data as a locus problem, thereby
formulating geometric locus methods within a statistical framework.
Accordingly, I\ have devised general eigen-coordinate systems for all forms of
linear and quadratic loci, and I\ have identified similar geometric properties
exhibited by the points on all forms of linear and quadratic loci. I have
devised fundamental, data-driven, locus equations of binary classification for
linear and quadratic classification systems in statistical equilibrium, where
the opposing forces and influences of a system are balanced with each other,
and the eigenenergy and the corresponding expected risk of a classification
system are minimized. Accordingly, I have devised three systems of
data-driven, vector-based locus equations that generate optimal discriminant
functions and decision boundaries. The three systems of locus equations
involve solving variants of the inequality constrained optimization problem
for linear, polynomial, and Gaussian kernel SVMs. All three classes of
learning machines are capable of performing a wide variety of statistical
pattern recognition tasks, where any given learning machine exhibits optimal
generalization performance for a two-class feature space. For each class of
learning machines, I have demonstrated that any given learning machine is a
scalable, individual component of an optimal ensemble system, where any given
ensemble system of learning machines exhibits optimal generalization
performance for an $M$-class feature space.

I have named the system of data-driven, mathematical laws that generates
optimal, linear classification systems a linear eigenlocus transform. Linear
eigenlocus transforms generate linear discriminant functions that are scalable
modules for optimal, binary and multiclass, linear classification systems.
Because any given statistical pattern recognition system is a linear
combination of linear eigenlocus discriminant functions, the overall network
complexity is scale-invariant for the feature space dimension and the number
of pattern classes.

I have named the system of data-driven, mathematical laws that generates
optimal, quadratic classification systems a quadratic eigenlocus transform.
I\ have demonstrated that quadratic eigenlocus classification systems
automatically generate the best decision boundary for a given set of data
distributions, so that any given quadratic eigenlocus classification system
achieves the lowest possible error rate that can be achieved by a discriminant
function and the best generalization error that can be achieved by a learning
machine. Quadratic eigenlocus transforms generate quadratic discriminant
functions that are scalable modules for optimal, binary and multiclass, linear
and quadratic classification systems. Because any given statistical pattern
recognition system is a linear combination of quadratic eigenlocus
discriminant functions, the overall network complexity is scale-invariant for
the feature space dimension and the number of pattern classes.

Because the generalization error of quadratic eigenlocus classification
systems is determined by the minimum probability of classification error for
any given data distributions that have unchanging statistics, optimal,
statistical classification systems can be designed for applications where
"black box" classification methods are too risky. For example, homeland
security, medical, and military decisions can be based on data representations
that have been designed to minimize expected error rates.

I\ have defined innovative and practical ways to use linear and quadratic
eigenlocus decision functions as statistical multimeters and statistical
gauges. Linear and quadratic eigenlocus decision functions provide a robust
measure of class separability and expected error rates for two given sets of
pattern or feature vectors whose mean and covariance functions remain constant
over time. Quadratic eigenlocus decision functions provide the most reliable
statistical multimeter: based on accurate and precise measures of data
distribution overlap and expected error rates for any two data distributions
that have unchanging statistics.

Linear and quadratic eigenlocus decision functions can also be used to
identify homogeneous data distributions. Alternatively, linear and quadratic
eigenlocus decision functions can be used to determine if two samples are from
different distributions. Again, quadratic eigenlocus decision functions
provide the most reliable statistical gauge: based on accurate and precise
measures of data distribution overlap and expected error rates for any two
data distributions that have unchanging statistics.

Finally, I have defined novel applications using linear or quadratic
eigenlocus transforms, where feature vectors from different data sources are
fused by forming fused ensembles of linear or quadratic eigenlocus decision
functions. Because quadratic eigenlocus transforms generate optimal
statistical classification systems that achieve minimum error rates for any
given data distributions that have unchanging statistics, quadratic eigenlocus
transforms are a gold standard for fusing feature vectors from different data sources.

\subsection{Acknowledgments}

My master's thesis
\citep{Reeves1995}
was the primary impetus for this work. I am indebted to Oscar Gonzalez who was
my master's thesis advisor. The counsel of Oscar Gonzalez has sustained the
trailblazer within me. The discoveries that I\ have presented in this paper
were also motivated by my Ph.D. dissertation
\citep{Reeves2009}%
. I am grateful to Garry Jacyna for illuminating conversations regarding my
dissertation research. The guidance of Garry Jacyna enabled me to navigate the
Ph.D. pipeline.

\bibliographystyle{abbrvnat}
\bibliography{locusedb}

\end{document}